# Principles and Practice of Deep Representation Learning

## or A Mathematical Theory of Memory

**Sam Buchanan***, UC Berkeley & TTIC

**Druv Pai***, UC Berkeley

**Peng Wang**, University of Macau & University of Michigan

**Yi Ma**, University of Hong Kong & UC Berkeley

This manuscript is prepared for a course on the principles of deep learning and intelligence in general, to be taught by the authors at the respective institutions.

March 1st, 2026

* These authors contributed equally.

# Preface

*"All roads lead to Rome."*

**Objective.** This book reveals and studies a common and fundamental problem behind almost all modern practices of machine (artificial) intelligence. That is, *how to effectively and efficiently learn a low-dimensional distribution of data in a high-dimensional space and then transform the distribution to a compact and structured representation?* For any intelligent system, natural or man-made, such a representation can be generally regarded as a *memory* or (empirical) *knowledge* learned from data sensed from the external world. In recent years, people often informally refer to it as a "world model."

**Intended Audience.** This textbook aims to provide a systematic introduction to the mathematical and computational principles for learning (deep) representations of such data distributions, as a computable form of memory, for *senior undergraduate students and beginning graduate students*. The main prerequisites for this book are undergraduate linear algebra, probability/statistics, and optimization. Some familiarity with basic concepts from signal processing (sparse representation and compressed sensing in particular), information theory, and feedback control would enhance your appreciation.

**Motivation.** The main motivation for writing this book is that there have been tremendous developments in the past several years, by the authors and many colleagues, that aim to establish a principled and rigorous approach to understand deep neural networks and, more generally, intelligence itself. The deductive methodology advocated by this new approach is in direct contrast, and highly complementary, to the dominant methodology behind current practices of artificial intelligence, which is largely inductive and trial-and-error. The lack of understanding about such powerful AI models and systems has led to increasing hype and fears in society. We believe that a serious attempt to establish a principled approach to understand intelligence is more needed than ever. An overarching goal of this book is to provide solid theoretical and experimental evidence showing that it is now possible to study intelligence as a scientific and mathematical subject. As we will argue that intelligence is the fundamental capability to develop new memory (or knowledge) or correct existing one. Hence, one may view this book as a first attempt to develop *a Mathematical Theory of Intelligence*, at the level of learning empirical knowledge as memory, as the subtitle of the book suggests.

At the technical level, the theoretical framework presented in this book helps reconcile a long-standing gap between the classical approach to modeling data structures that are mainly based on analytical geometric, algebraic, and probabilistic models (e.g., subspaces, Gaussians, and equations) and the "modern" approach that is based on empirically designed non-parametric data-driven models (e.g., deep networks). As it turns out, a unification of the two seemingly separate methodologies becomes possible and even natural if one realizes that they all try to learn and represent *low-dimensional* structures in the data distribution of interest. They are merely different ways to pursue, represent, and exploit the low-dimensional structures. From this perspective, even many seemingly unrelated computational techniques, developed independently in separate fields at different times, can now be better understood under a common theoretical and computational framework and probably can be studied together from now on. As we will see in this book, these techniques include but are not limited to:

1. Lossy compressive encoding-decoding developed in classic information theory and coding theory versus modern representation learning;
2. Denoising and dimensionality reduction in classical signal processing versus diffusion and denoising models for modern generative methods;
3. Bayesian inference via maximum a posteriori or conditioned sampling versus constrained optimization via continuation techniques[1].

**Main Content.** We believe that the unified conceptual and computational framework presented in this book will be of great value to readers who truly want to clarify mysteries and misunderstandings about deep neural networks and (artificial) intelligence. Furthermore, the framework is meant to provide readers with guiding principles for developing significantly better and truly intelligent systems in the future. More specifically, besides an informal introduction (chapter), the main technical content of the book will be organized as six closely related topics (chapters):

1. We will start with the classical and most basic models of Principal Component Analysis (PCA), Independent Component Analysis (ICA), and Dictionary Learning (DL), which assume that the low-dimensional distributions of interest have linear and independent structures. From these simple idealistic models that are well studied and understood in signal processing and compressed sensing, we will introduce the most basic and important ideas[2] for how to learn and represent low-dimensional distributions.

[1] such as the augmented Lagrangian method

[2] such as denoising and dimensionality reduction

2. To generalize these classical models and their solutions to low-dimensional distributions that may not be easily represented any simple analytic models, we introduce a universal computational principle for learning such distributions: *compression*. As we will see, data compression provides a unifying view of all seemingly different classic and modern approaches to distribution or representation learning, including entropy minimization, denoising via score-matching, lossy compression with rate distortion, and discriminative representation learning via maximizing information gain.

3. Within this unifying framework, modern Deep Neural Networks (DNNs), such as ResNet, CNN, and Transformer, can all be mathematically interpreted as (unrolled) optimization algorithms that iteratively achieve better compression and better representations by reducing coding length/rate or gaining information. Not only does this framework help explain empirically designed deep network architectures thus far, it also leads to new architecture designs that can be significantly simpler and more efficient.

4. Furthermore, to ensure that the learned representation for a data distribution is correct and consistent, the *auto-encoding* architectures that consist of both encoding and decoding become necessary. In order for a learning system to be fully automatic and continuous, we will introduce a powerful *closed-loop transcription framework* that enables an auto-encoding system to self-correct and thus self-improve via a minimax game between the encoder and decoder.

5. To connect theory to practice, we will then study how the learned data distribution and representation can be utilized as a powerful prior to conduct Bayesian inference or constrained optimization that manifests as almost all types of tasks and settings that are popular in the practice of modern artificial intelligence, including conditional estimation, completion, and generation of real-world high-dimensional data such as images and texts.

6. Last but not least, we will demonstrate step-by-step how to effectively and efficiently learn deep representations of low-dimensional data distributions with large-scale real-world datasets, including visual data, body motions, and text data, and use them in many practical applications such as image classification, image completion, image segmentation, image generation, motion estimation, and similar tasks for the text data.

**Summary.** To summarize, the technical content presented in this book establishes strong conceptual and technical connections between the classical analytical approach and the modern data-driven approach, between simple parametric models and deep non-parametric models, between diverse inductive practices and a unified deductive framework from first principles. We will reveal that many seemingly unrelated or even competing approaches, though developed in separate fields with different terminologies and at different times in history, yet all strive to achieve a common objective:

*pursuing and exploiting intrinsic low-dimensional structures of data distributions embedded in high-dimensional spaces.*

To this end, the book will take us through a complete journey from theoretical formulation to mathematical deduction, then to computational implementation and practical applications.

# Preface to Version 2.0

As an open-source book, the first version, entitled "*Learning Deep Representations of Data Distributions*", released on August 18, 2025 has been frequently and constantly updated in the past six months. Meanwhile, based on our experience from using the book for a new course taught at the University of Hong Kong in Fall 2025 and feedback that we have received from colleagues, teaching assistants and students, we have identified numerous points throughout the book that merit further revision and expansion.

Hence, we have decided to make some substantial changes and upgrades of the content and organization of the book with a new *Version 2.0* and changed its title to:

*Principles and Practice of Deep Representation Learning*
*or A Mathematical Theory of Memory*

The new version also allows us to explicitly reveal strong conceptual and technical connections among materials across different chapters and sections within the book, so that the overall pedagogical value of the book, we believe, has been improved significantly over the first version.

In our opinion, the new version gives a much more unified, complete, and stream-lined presentation of this subject. The new version also incorporates newly developed theoretical insights as well as a growing number of practical applications for real-world data that have not been properly documented and systematically explained elsewhere. The Version 2.0 serves as a timely remedy to this situation.

## Major Changes in Version 2.0

- We have split Chapter 3 in the first version into to two chapters, now Chapter 3 and 4. The new Chapter 3 focuses on the denoising process for learning low-dimensional distributions. We have added a new section that characterizes conditions under which the process leads to generalization or memorization. The new Chapter 4 focuses on representation learning based on a lossy coding approach and the principle of maximizing information gain. Since the old version only illustrated how to apply the principle to learn image representations in the supervised setting, the new version has added a case study with the unsupervised setting. It also provides theoretical justification for popular unsupervised learning methods such as contrastive learning and DINO in particular.

- In the updated Chapter 5 on designing deep representations via unrolled optimization, we have added a principled derivation of a causal version for the white-box transformer architecture CRATE, which is important for processing sequential data such as texts as we often see in applications.

- In the updated Chapter 6 on consistent representation learning, we added a new section that further elucidates the relationships between distribution learning and representation learning. This provides theoretical clarification and justification for popular practical autoencoding methods such as variational autoencoding and representation autoencoders.

- In the updated Chapter 7, we have added a section that provides a principled explanation of representation learning with paired data and conditioned generation through the perspective of mutual information. It provides theoretical justification for popular practical methods such as CLIP and cross attention.

- The expanded application Chapter 8 now features many more detailed implementations of distribution learning and representation learning for real-world data, including natural 2D images, natural 3D objects, human body motions, and natural languages, under almost all popular and practical settings, supervised, weakly supervised, unsupervised, and conditioned.

- The final Chapter 9 about open directions has also been significantly rewritten. We attempt to give a clearer taxonomy of different levels of intelligence so that we can clarify what we have done and understood about intelligence (with this book) and what we have not. We believe that open problems associated with more advanced forms of intelligence can be well posed so that they can be studied qualitatively and quantitatively via scientific and mathematical means, instead of remaining as a mysterious subject or merely a philosophical topic.

## Contributors for Version 2.0

Besides the authors, many students and colleagues have joined this project and contributed valuable content to different parts of the book during the preparation of *Version 2.0*. Below is an incomplete list of people and their specific contributions, in alphabetical order:

- Tianzhe Chu: Experiments in Section 8.4, and AI tooling for the book.

- Prof. Shenghua Gao: Sections 8.8 and 8.9.

- Bingbing Huang: Sections 8.8 and 8.9.

- Kerui Min: Chinese translation.

- Prof. Qing Qu: Section 3.3.
- Shengbang Peter Tong: Sections 6.3 and 8.6.2.
- Chengyu Wang: Sections 8.8 and 8.9.
- Ziyang Robin Wu: Section 8.3 and website development.
- Chun-Hsiao Daniel Yeh: Sections 8.6 and 8.7.
- Brent Yi: Section 8.10.
- Jingfeng Yang: Section 8.3.
- Dr. Yaodong Yu: Chapter 5 is based on his PhD thesis.
- Dr. Zibo Zhao: Section 8.8.

We also thank Dr. Kevin Murphy and Dr. Bill Mark, for extensive technical feedback on the manuscript; Jan Cavel, for contributing an unofficial Romanian translation; and Stephen Butterfill and Jeroen Van Goey, for contributing corrections and fixes to the manuscript.

# Declaration of Open Source

"*I often compare open source to science. Science took this whole notion of developing ideas in the open and improving on other peoples' ideas. It made science what it is today and made the incredible advances that we have had possible. And I compare that to witchcraft and alchemy, where openness was something you didn't do.*"

– Linus Torvalds

This book is meant to be an "open-source" project which is to be made publicly available at the website:

https://ma-lab-berkeley.github.io/deep-representation-learning-book/

for all people who are interested in the scientific, mathematical, and computational principles of intelligence. The book will be updated continuously and periodically by all who are willing to contribute to this project, including its translated versions, teaching materials, and various AI tools developed that support studying, teaching, and researching this topic.

Through this book, we hope to initiate an "open knowledge program" which advocates that all scientific knowledge should be open sourced, for its organization, pedagogy, dissemination, and advancement.

# Acknowledgment

**Funding Support.** This book is primarily based on research results that have been developed since 2018. Thanks to the generous startup funds from UC Berkeley (2018) and the University of Hong Kong (2023), Yi Ma was able to embark and focus on this new exciting research direction in the past eight years or so. Through these years, for research in this direction, Yi Ma and his research team at Berkeley have been supported by the following research grants:

- The multi-university *THEORINET* project for the Foundations of Deep Learning, jointly funded by the Simons Foundation and the National Science Foundation (DMS grant #2031899);
- The *Closed-Loop Data Transcription via Minimaxing Rate Reduction* project funded by the Office of Naval Research (grant N00014-22-1-2102);
- The *Principled Approaches to Deep Learning for Low-dimensional Structures* project funded by the National Science Foundation (CISE grant #2402951).

This book would have not been possible without the financial support for these research projects. The authors have drawn tremendous inspiration from research results by colleagues and students who have been involved in these projects.

# Notation

In this book we employ notation from a variety of mathematical disciplines, summarized below. The reader is encouraged to consult this list as a reference as they read the various chapters of the book. We have tried to keep their usage as coherent as possible across chapters.

As a first example, for a natural number $n$ the set $\{1, 2, \ldots, n\}$ is denoted $[n]$.

## Linear Algebra Notation

- Scalars (either deterministic or random) are non-bold, e.g., $x$, $X$.
- Vectors (either deterministic or random) are lower-case bold, e.g., $\boldsymbol{x}$, $\boldsymbol{\pi}$. Entries of vectors are denoted $x_i$ or $\pi_i$, alternatively $(\boldsymbol{x})_i$. Generic terms (which can be scalars, vectors, matrices, or higher-order tensors) are also lower-case bold.
- Matrices (either deterministic or random) are capital bold, e.g., $\boldsymbol{X}, \boldsymbol{\Pi}$. Entries of matrices are denoted $X_{ij}$ or $\Pi_{ij}$, alternatively $(\boldsymbol{X})_{ij}$. Columns of matrices are lower-case bold vectors, e.g., $\boldsymbol{x}_i$ is the $i^{\text{th}}$ column of $\boldsymbol{X}$, unless otherwise defined.
- The transpose of a matrix is $\top$, e.g., $\boldsymbol{A}^\top$. The adjoint is $^*$, e.g., $\boldsymbol{A}^*$. The pseudo-inverse is $\dagger$, e.g., $\boldsymbol{A}^\dagger$.
- The rank of a matrix is $\operatorname{rank}(\boldsymbol{A})$, the trace is $\operatorname{tr}(\boldsymbol{A})$, the determinant is $\det(\boldsymbol{A})$, and the log-determinant is $\operatorname{logdet}(\boldsymbol{A})$.
- The element-wise multiplication of matrices is $\boldsymbol{A} \odot \boldsymbol{B}$. Element-wise squaring is $\boldsymbol{A}^{\odot 2}$, etc. Similarly, the Kronecker product of matrices is $\boldsymbol{A} \otimes \boldsymbol{B}$. The iterated Kronecker product is $\boldsymbol{A}^{\otimes 2}$, etc.
- For a function $f \colon \mathbb{R} \to \mathbb{R}$, its element-wise application to a matrix $\boldsymbol{X}$ is $f[\boldsymbol{X}]$, i.e., with square brackets. For a function $f \colon \mathbb{R}^d \to \mathbb{R}^k$ and a matrix $\boldsymbol{X} \in \mathbb{R}^{d \times n}$, we may broadcast $f$ to apply to each column without special notation, i.e., $f(\boldsymbol{X}) = [f(\boldsymbol{x}_1), \ldots, f(\boldsymbol{x}_n)] \in \mathbb{R}^{k \times n}$.

- The (Euclidean) unit sphere is $\mathbb{S}^{n-1} \subseteq \mathbb{R}^n$. The (Euclidean) unit ball is $\mathbb{B}^n \subseteq \mathbb{R}^n$.
- The set of orthogonal matrices are $\mathsf{O}(m, n) \subseteq \mathbb{R}^{m \times n}$ or $\mathsf{O}(n) = \mathsf{O}(n, n)$.
- Symmetric matrices are $\mathsf{Sym}(n)$, symmetric PSD matrices are $\mathsf{PSD}(n)$, symmetric PD matrices are $\mathsf{PD}(n)$.
- The eigenvalues of symmetric matrix $\boldsymbol{S} \in \mathsf{Sym}(n)$ are all real by the spectral theorem; they are denoted $\lambda_i(\boldsymbol{S})$ and ordered such that $\lambda_1(\boldsymbol{S}) \geq \cdots \geq \lambda_n(\boldsymbol{S})$.
- The singular values of $\boldsymbol{A} \in \mathbb{R}^{m \times n}$ with $\operatorname{rank}(\boldsymbol{A}) = r$ are denoted $\sigma_i(\boldsymbol{A})$ and ordered such that $\sigma_1(\boldsymbol{A}) \geq \cdots \geq \sigma_r(\boldsymbol{A}) > 0$. By convention we set $\sigma_{r+1}(\boldsymbol{A}) = \sigma_{r+2}(\boldsymbol{A}) = \cdots = 0$.
- The projection onto a set $\mathcal{K}$ is denoted $\operatorname{proj}_{\mathcal{K}}$.
- The $\ell^p$ norms of vectors are denoted by $\|\boldsymbol{x}\|_p$, $p \in [0, \infty]$.
- The Euclidean operator norm on matrices is $\|\boldsymbol{A}\| = \sigma_1(\boldsymbol{A})$.
- The Frobenius norm on matrices is $\|\boldsymbol{A}\|_F = \operatorname{tr}(\boldsymbol{A}^\top \boldsymbol{A})$.
- The Euclidean inner product of vectors is $\langle \boldsymbol{x}, \boldsymbol{y} \rangle = \boldsymbol{y}^\top \boldsymbol{x}$.
- The Frobenius inner product on matrices is $\langle \boldsymbol{X}, \boldsymbol{Y} \rangle = \operatorname{tr}(\boldsymbol{Y}^\top \boldsymbol{X})$.
- The all-ones vector/matrix is denoted $\mathbf{1}$ (the shape should be obvious from context). Similarly, the all-zeros vector/matrix is denoted $\mathbf{0}$. The identity matrix is denoted $\boldsymbol{I}$.

# Probability Notation

- The base probability measure is $\mathbb{P}$, i.e., the probability of an event $A$ occurring is $\mathbb{P}[A]$. We may specify the distribution of a random variable using a subscript, i.e., $\mathbb{P}_{\boldsymbol{x} \sim \mu}[\boldsymbol{x} \in S]$ describes the probabilities associated to a random variable $\boldsymbol{x}$ with distribution (measure) $\mu$. If two random variables $\boldsymbol{x}$ and $\boldsymbol{x}'$ have the same distribution, we write $\boldsymbol{x} \overset{\mathrm{d}}{=} \boldsymbol{x}'$.
- The expected value operator is $\mathbb{E}$, with the same subscript caveat.
- The covariance operator is Cov, with the same subscript caveat.
- The correlation operator is Corr, with the same subscript caveat.
- In certain probabilistic modeling contexts where a density is assumed (especially in Chapters 3 and 7), we also use the notation $p_{\boldsymbol{x}}$ for the density of a random variable $\boldsymbol{x}$. This extends to conditional densities, e.g. $p_{\boldsymbol{x} \mid \boldsymbol{y}}$ for the density of $\boldsymbol{x}$ conditioned on $\boldsymbol{y}$.

- We generally use Greek letters to denote realizations of random variables (in contrast to the random variables themselves). For example, $p_{\boldsymbol{x}}(\boldsymbol{\xi})$, $p_{\boldsymbol{x}|\boldsymbol{y}}(\boldsymbol{\xi} \mid \boldsymbol{\nu})$, $\mathbb{E}[\boldsymbol{x} \mid \boldsymbol{y} = \boldsymbol{\nu}]$, etc.

- The set of probability distributions on a finite set $\mathcal{X}$ is denoted $\Delta(\mathcal{X})$. (For $\mathcal{X} = [n]$, this is the probability simplex which can be embedded into $\mathbb{R}^n$).

- The Gaussian distribution with mean $\boldsymbol{\mu}$ and covariance $\boldsymbol{\Sigma}$ is denoted $\mathcal{N}(\boldsymbol{\mu}, \boldsymbol{\Sigma})$.

- The uniform distribution over a compact set $\mathcal{X}$ is denoted $\mathcal{U}(\mathcal{X})$.

# Machine Learning Notation

- Encoders and denoisers are usually denoted by $f$ and usually have parameters $\theta \in \Theta$. We usually write the features $\boldsymbol{z} = f_\theta(\boldsymbol{x})$.

- Decoders are usually denoted by $g$ and usually have parameters $\eta$. We usually write the auto-encoding $\hat{\boldsymbol{x}} = g_\eta(\boldsymbol{z})$ corresponding to $\boldsymbol{x}$.

- When $f$ and $g$ are implemented as neural networks, they will have the same number of layers without loss of generality; we write them as $f = f^L \circ f^{L-1} \circ \cdots \circ f^2 \circ f^1 \circ f^{\mathrm{pre}}$ and $g = g^{\mathrm{post}} \circ g^L \circ g^{L-1} \circ \cdots \circ g^2 \circ g^1$.

- We write the features at the input to layer $\ell$ as $\boldsymbol{z}^\ell$ such that $\boldsymbol{z}^{\ell+1} = f^\ell(\boldsymbol{z}^\ell)$ with $\boldsymbol{z}^1 = f^{\mathrm{pre}}(\boldsymbol{x})$ and $\boldsymbol{z} = \boldsymbol{z}^{L+1}$. Similarly the autoencoding features are $\hat{\boldsymbol{x}}^\ell$ with $\hat{\boldsymbol{x}}^{\ell+1} = g^\ell(\hat{\boldsymbol{x}}^\ell)$ with $\hat{\boldsymbol{x}}^1 = \boldsymbol{z}$ and $\hat{\boldsymbol{x}} = g^{\mathrm{post}}(\hat{\boldsymbol{x}}^{L+1})$.

- For denoising, optimization, and processes which have a continuous time index, the time is almost always a subscript, e.g., $\boldsymbol{x}_t$. Discrete sequences can have the index be a superscript or subscript.

# Modeling and Optimization Notation

- As in the Probability and Machine Learning Notation sections above, when we have a model, say for the realizations of a data distribution supported on $\mathbb{R}^D$, we always denote it with "plain" accented variables, say $\boldsymbol{x}$.

- In cases where we assume our data $\boldsymbol{x}$ is generated by a model from a restricted class of *parametric models*, say with a parameter $\boldsymbol{\mu}$, we also denote the *true parameter* with plain accenting.[3]

[3] This could arise either due to us making an assumption about our data, or due to our data being synthetically generated. This convention allows us to adopt a concise notation across both 'probabilistic' and 'analytical' modeling setups. Readers more familiar with the latter kinds of frameworks may be used to using notation such as $\boldsymbol{\mu}_\sharp$ or $\boldsymbol{\mu}_o$ for this case; contrast with our accented decision variable notation.

- *Decision variables* (or learnable parameters) of a model fit to observed data are denoted with “tilde” accenting, e.g. $\tilde{\boldsymbol{x}}$, in cases where there is an underlying model $\boldsymbol{x}$, or with plain notation in cases where no confusion is possible (e.g., if an empirically-fit model for *nonparametric* data $\boldsymbol{x}$ involves “mean” parameters $\boldsymbol{\mu}$).
- *Optimal solutions* to optimization problems are denoted with “star” accenting, either as a superscript or a subscript (say $\boldsymbol{x}_\star$ or $\boldsymbol{x}^\star$).
- *Estimators* for data that correspond to a specific statistical model, especially for the mean of that statistical model, are denoted with “bar” accenting, say $\bar{\boldsymbol{x}}$ (especially for minimum mean-squared error denoising in Chapter 3).
- *Approximations* associated to computational procedures for modeling data are denoted with “hat” accenting, say $\hat{\boldsymbol{x}}$ for the output of an autoencoder or a generative model that approximates data $\boldsymbol{x}$.[4]

[4] We generally distinguish these cases from optimal solutions $\boldsymbol{x}_\star$, although in some special cases they are equal (e.g., in several examples in Chapter 2). Think of the distinction between the parameter $\boldsymbol{\mu}$ of a model, which might be fit with optimization, and the results of sampling with that learned parameter.

# Contents

# Chapter 1

# An Informal Introduction to Intelligence

"*Just as the constant increase of entropy is the basic law of the universe, so it is the basic law of life to be ever more highly structured and to struggle against entropy.*"

– Václav Havel

## 1.1 Intelligence, Cybernetics, and Artificial Intelligence

The world we inhabit is neither fully random nor completely unpredictable.[1] Instead, it follows certain orders, patterns, and laws that render it largely predictable.[2] The very emergence and persistence of life depend on this predictability. Only by learning and memorizing what is predictable in the environment can life survive and thrive, since sound decisions and actions hinge on reliable predictions. Because the world offers seemingly unlimited predictable phenomena, intelligent beings—animals and humans—have evolved ever more acute senses: vision, hearing, touch, taste, and smell. These senses harvest high-throughput sensory data to perceive environmental regularities. Hence, a fundamental task for all intelligent beings is to

*learn and memorize predictable information from massive amounts of sensed data.*

[1] If the world were fully random, an intelligent being would have no need to learn or memorize anything.

[2] Some deterministic, some probabilistic.

Before we can understand how this is accomplished, we must address three questions:

- How can predictable information be modeled and represented mathematically?
- How can such information be computationally learned effectively and efficiently from data?
- How should this information be best organized and structured as memory or (empirical) knowledge, in our brain or a machine, to support future prediction and inference?

This book aims to provide some answers to these questions. These answers will help us better understand intelligence, especially the computational principles and mechanisms that enable it. Evidence suggests that all forms of intelligence—from low-level intelligence seen in early primitive life to the highest form of intelligence, the practice of modern science—share a common set of principles and mechanisms. We elaborate below.

**Emergence and evolution of intelligence.** A necessary condition for the emergence of life on Earth about four billion years ago is that the environment is largely predictable. Life has developed mechanisms that allow it to learn what is predictable about the environment, encode this information, and use it for survival. Generally speaking, we call this ability to learn knowledge of the world *intelligence*. To a large extent, the evolution of life is the mechanisms of intelligence at work [Ben23], compared to the evolution of the universe being the laws of physics at work. In the early stages of life, intelligence is mainly developed through two types of learning mechanisms: *phylogenetic* and *ontogenetic* [Wie61].

*Phylogenetic intelligence* refers to learning through the evolution of species. Species inherit and survive mainly based on memory and knowledge encoded in the DNA or genes of their parents. To a large extent, we may call DNA nature's pre-trained large models because they play a very similar role. The main characteristic of phylogenetic intelligence is that individuals have limited learning capacity. Learning is carried out through a "trial-and-error" mechanism based on random mutation of genes, and species evolve based on natural selection—survival of the fittest—as shown in Figure 1.1. This can be viewed as nature's implementation of what is now known as "reinforcement learning," [SB18] or potentially "neural architecture search" [ZL17]. However, such a "trial-and-error" process can be extremely slow, costly, and unpredictable. From the emergence of the first life forms, about 4.4–3.8 billion years ago [BBH+15], life has relied on this form of evolution.[3]

*Ontogenetic intelligence* refers to the learning mechanisms that allow an individual to learn through its own senses, memories, and predictions within its specific environment, and to improve and adapt its behaviors. Ontogenetic learning became possible after the emergence of the nervous system about 550–600 million years ago (in worm-like organisms) [WVM+25], shown in Figure 1.2 middle. With a sensory and nervous system, an individual can continuously form and improve its own memory about the world, in addition to what is inherited from DNA or genes. This capability significantly enhanced individual survival and contributed to the Cambrian explosion of life forms about 530 million years ago [Par04]. Compared to phylogenetic learning, ontogenetic learning is more efficient and predictable, and can be realized within an individual's resource limits.

[3] Astute readers may notice an uncanny similarity between how early life evolves and how large language models (LLMs) evolve today.

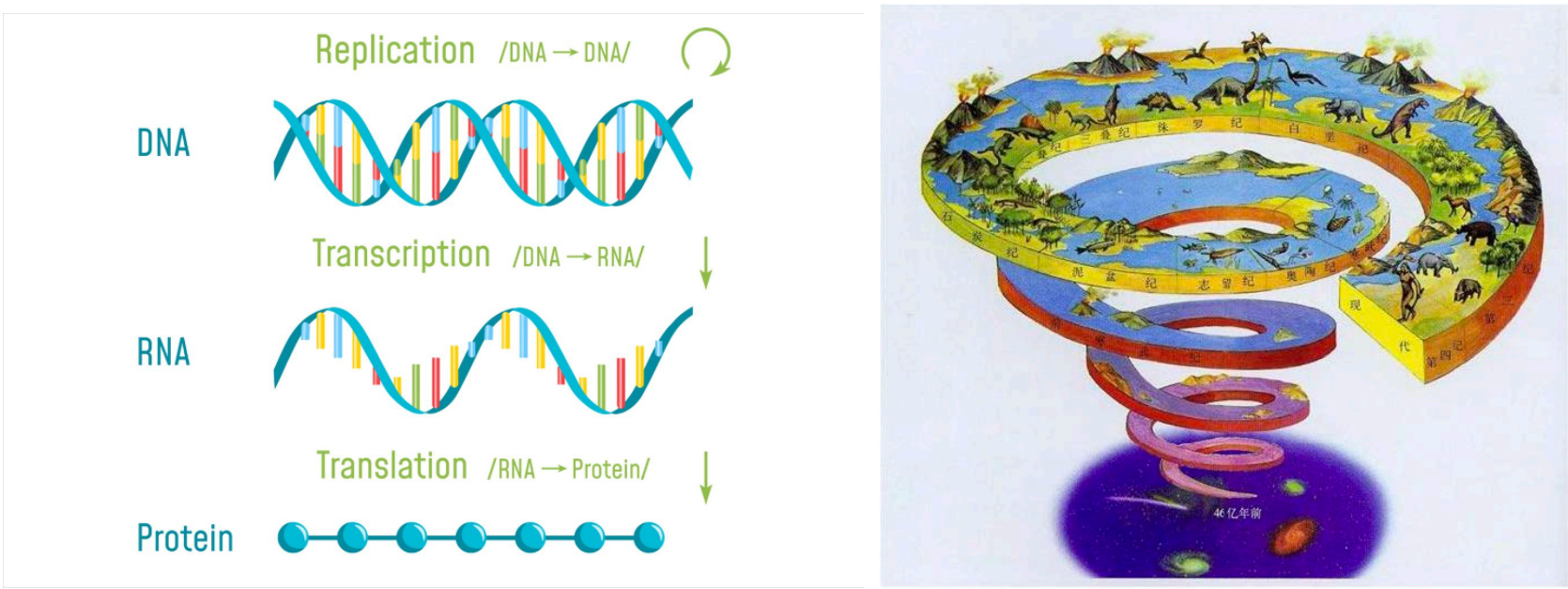

Figure 1.1: Evolution of phylogenetic intelligence: Memory or knowledge of the external world is encoded and passed on via DNA (left), and decoded from DNA to RNA and to proteins. In the early stage of life evolution (right), intelligence develops memory or knowledge at the species level via (random) gene mutation and natural selection—"may the fittest survive"—which can be viewed as a primitive form of reinforcement learning.

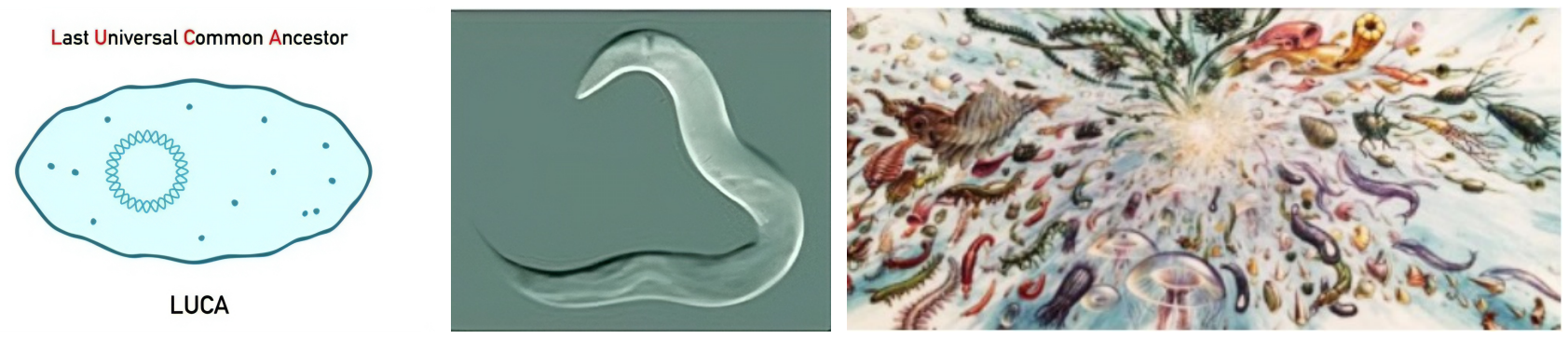

Figure 1.2: Evolution of life, from the ancestor of all life today (LUCA—last universal common ancestor), a single-cell-like organism that lived 3.5–4.3 billion years ago [MÁM+24], to the emergence of the first nervous system in worm-like species (middle), about 550 million years ago [WVM+25], to the explosion of life forms in the Cambrian period (right), about 530 million years ago.

Both types of learning rely on feedback from the external environment—penalties (death) or rewards (food)—applied to a species' or an individual's actions.[4] This insight inspired Norbert Wiener to conclude in his Cybernetics program [Wie48] that all intelligent beings, whether species or individuals, rely on closed-loop feedback mechanisms to learn and improve their memory and knowledge about the world. Furthermore, from plants to fish, birds, and mammals, more advanced species increasingly rely on ontogenetic learning: they remain with and learn from their parents for longer periods after birth, because individuals of the same species must survive in very diverse environments. In other words, as species become increasingly more "intelligent," they rely less and less on their inherited DNAs ("pre-trained foundation models") and more and more on their ability to learn themselves after birth.

**Evolution of human intelligence.** Since the emergence of *Homo sapiens* about 2.5 million years ago [Har15], a new, higher form of intelligence has emerged that evolves more efficiently and economically. Human societies developed languages—first spoken, later written—as shown in Figure 1.3. Language enables individuals to communicate and share useful information, allowing a human community to behave as a single intelligent organism that learns faster and retains more knowledge than any individual. Written texts thus play a role analogous to DNA and genes, enabling societies to accumulate and disseminate knowledge across lands and generations. We may refer to this type of intelligence as *societal intelligence*, distinguishing it from the phylogenetic intelligence of species and the ontogenetic intelligence of individuals. This knowledge accumulation underpins all (ancient) civilizations.

About two to three thousand years ago, human intelligence took another major quantum leap, enabling philosophers and mathematicians to develop knowledge that goes far beyond organizing empirical observations. The development of abstract concepts and symbols, such as numbers, time, space, logic, and geometry, gave rise to an entirely new and rigorous language of *mathematics*. In addition, the development of the ability to generate hypotheses and verify their correctness through logical deduction or experimentation laid the foundation for modern *science*. For the first time, humans could proactively and systematically discover and develop new knowledge with unprecedented accuracy and rigor.

[4] Gene mutation of the species or actions made by the individual.

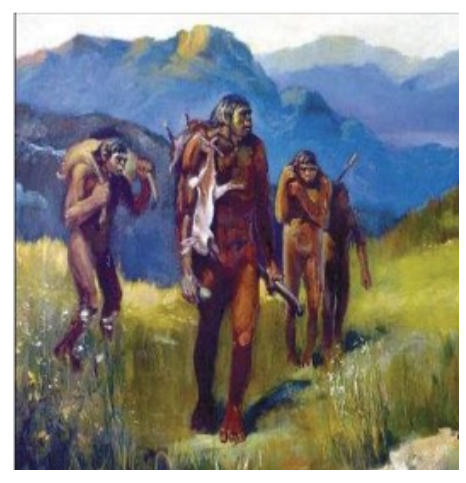

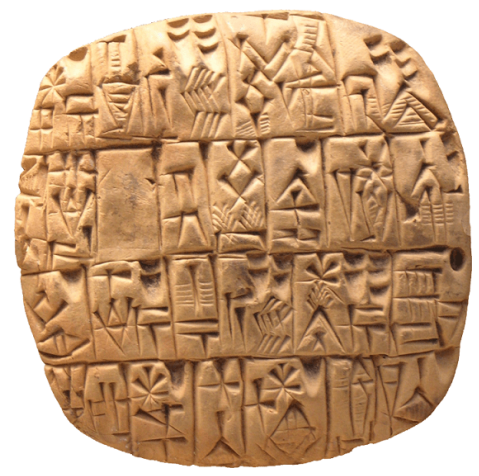

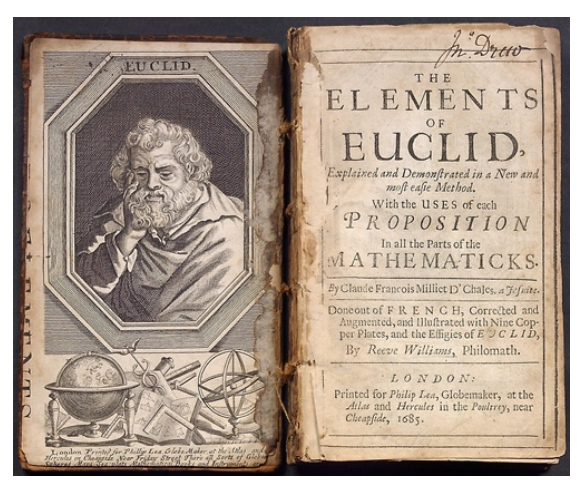

Figure 1.3: The development of verbal communication and spoken languages (70,000–30,000 years ago), written languages (about 3000 BC) [Sch14], and abstract mathematics (around 500–300 BC) [Hea+56] mark three key milestones in the evolution of human intelligence.

This form of intelligence has significantly improved the efficiency of acquiring knowledge about the unknown. We will call this advanced form of intelligence "scientific intelligence" due to its necessity for deductive and scientific discovery.

Hence, from what we can learn from nature, whenever we use the word "intelligence," we must be specific about which level or form we mean:

$$\textbf{phylogenetic} \implies \textbf{ontogenetic} \implies \textbf{societal} \implies \textbf{scientific intelligence}. \tag{1.1.1}$$

Clear characterization and distinction are necessary because we want to study intelligence as a scientific and mathematical subject. Although all forms may share the common objective of learning and improving memory and knowledge about the world, the computational mechanisms and physical implementations could differ. We believe the reader will better understand and appreciate their commonality and difference after studying this book. Therefore, we leave deeper scientific or philosophical discussions on the relationship among these different forms or levels of intelligence to the end of the book Chapter 9. For now, we only need to notice that, in terms of the type of knowledge learned, the first three stages of intelligence share more in common: they all learn knowledge (or memory) in a largely *empirical* fashion and then use it in a mostly *inductive* way. This book will mainly study and reveal how this type of knowledge can be learned from data sensed from an external world. At this point, the mechanisms behind how advanced mathematical and scentific knowledge is created remainly largely unclear although we believe it must be built on top of the ability to develop empirical knowledge. We leave this as an open problem for discussion in the final Chapter 9.

**Origin of machine intelligence—cybernetics.** In the 1940s, spurred by the war effort, scientists inspired by natural intelligence sought to emulate animal intelligence with machines, giving rise to the "Cybernetics" movement championed by Norbert Wiener [Kli11]. Wiener studied zoology at Harvard as an undergraduate before becoming a mathematician and control theorist. He devoted his life to understanding and building autonomous systems that could reproduce animal-like intelligence. Today, the Cybernetics program is often narrowly interpreted as being mainly about feedback control systems, the area in which Wiener made his most significant technical contributions. Yet the program was far broader and deeper: it aimed to understand intelligence as a whole—at least at the animal level—and influenced the work of an entire generation of renowned scientists, including Warren McCulloch, Walter Pitts, Claude Shannon, John von Neumann, and Alan Turing.

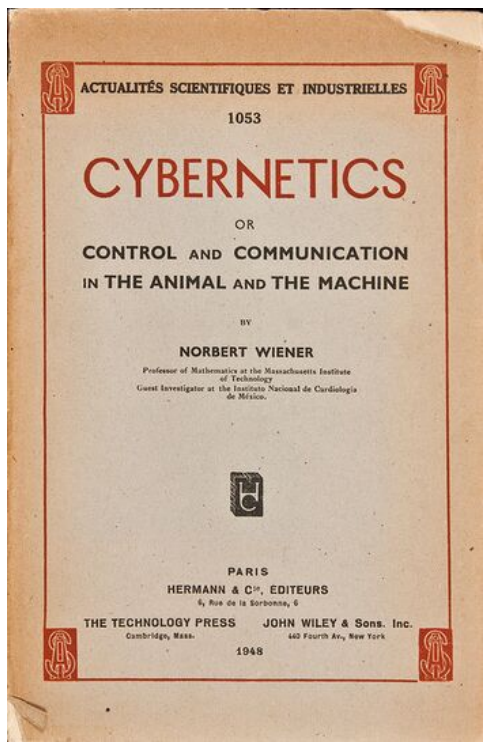

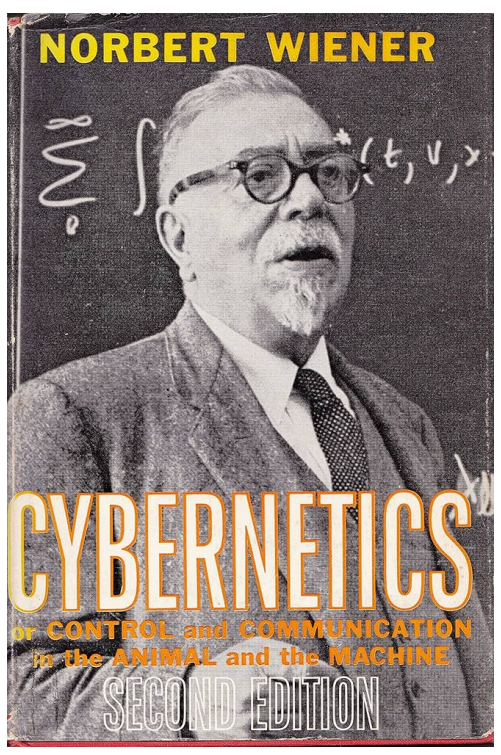

Figure 1.4: Norbert Wiener's book "Cybernetics" (1948) [Wie48] (left) and its second edition (1961) [Wie61] (right).

Wiener was arguably the first to study intelligence *as a system*, rather than focusing on isolated components or aspects of intelligence. His comprehensive views appeared in the celebrated 1948 book *Cybernetics: or Control and Communication in the Animal and the Machine* [Wie48]. In that book and its second edition published in 1961 [Wie61] (see Figure 1.4), he attempted to identify several necessary characteristics and mechanisms of intelligent systems, including (but not limited to):

- How to *measure and store* information (in the brain) and how to communicate with others.[5] This insight led Claude Shannon to formulate *information theory* in 1948 [Sha48].

- How to *correct errors* in prediction and estimation based on existing information. Wiener himself helped formalize the theory of (closed-loop) feedback control in the 1940s.

- How to learn to *make better decisions* when interacting with a non-cooperative or even adversarial environment. John von Neumann formalized this as *game theory* in 1944 [NMR44].

In 1943, inspired by Wiener's Cybernetics program, cognitive scientist Warren McCulloch and logician Walter Pitts jointly formalized the first computational model of a neuron [MP43], called an *artificial neuron* and illustrated later in Figure 1.15. Building on this model, Frank Rosenblatt constructed the Mark I Perceptron in the 1950s—a physical machine containing hundreds of such artificial neurons [Ros57]. The Perceptron was the first physically realized artificial neural network; see Figure 1.17. Notably, John von Neumann's universal computer architecture, proposed in 1945, was also designed to facilitate the goal of building *computing machines* that could physically realize the mechanisms suggested by the Cybernetics program [Neu58].

[5]Wiener was the first to point out that "information" is neither matter nor energy, but an independent quantity worthy of study.

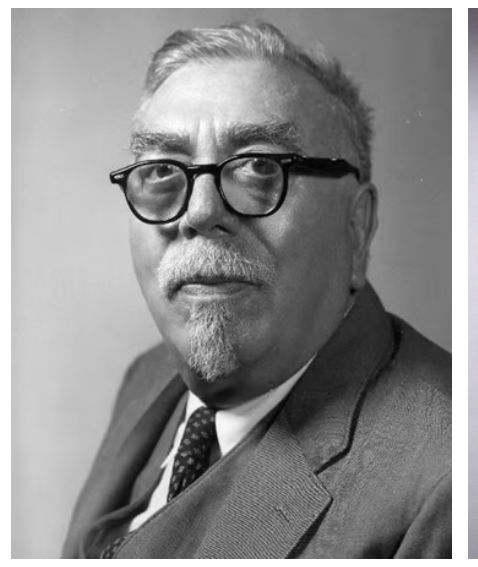 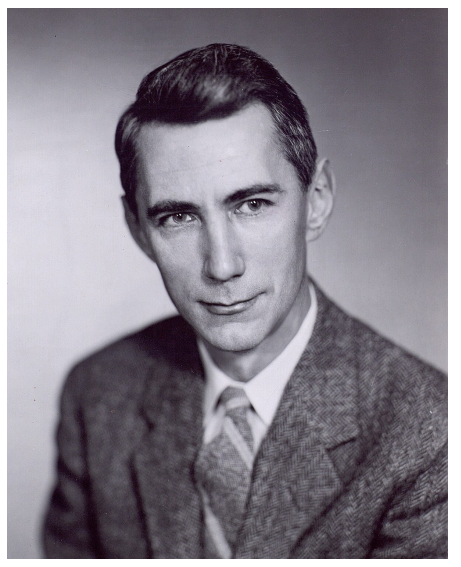 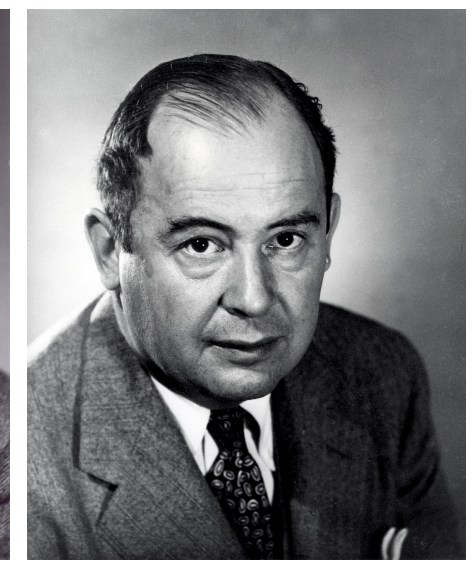 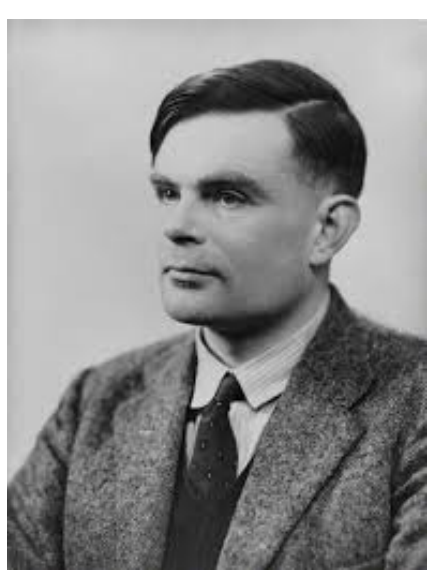

Figure 1.5: Pioneers of theoretical and computational foundations for intelligence: Norbert Wiener (cybernetics and control theory), Claude Shannon (information theory), John von Neumann (game theory), and Alan Turing (computing theory).

Astute readers will notice that the 1940s were truly magical: many fundamental ideas were invented and influential theories formalized, including the mathematical model of neurons, artificial neural networks, information theory, control theory, game theory, and computing machines. Figure 1.5 portrays some of the pioneers. Each of these contributions has grown to become the foundation of a scientific or engineering field and continues to have tremendous impact on our lives. All were inspired by the goal of developing machines that emulate intelligence in nature. Historical records show that Wiener's Cybernetics movement influenced nearly all of these pioneers and works. To a large extent, Wiener's program can be viewed as the true predecessor of today's "embodied intelligence" program; in fact, Wiener himself articulated a vision for such a program with remarkable clarity and concreteness [Wie61].

Although Wiener identified many key characteristics and mechanisms of (embodied) intelligence, he offered no clear recipe for integrating them into a complete autonomous intelligent system. From today's perspective, some of his views were incomplete or inaccurate. In particular, in the last chapter of the second edition of *Cybernetics* [Wie61], he stressed the need to *deal with nonlinearity* if machines are to emulate typical learning mechanisms in nature, yet he provided no concrete, effective solution for this difficult issue. In fairness, even the theory of linear systems was in its infancy at the time,[6] and nonlinear

[6] Wiener's famous achievement, the Wiener filter [Wie42; Wie49], which we will introduce later, was only for linear systems.

systems seemed far less approachable.

Nevertheless, we cannot help but marvel at Wiener's prescience about the importance of nonlinearity. As this book will show, the answer came only recently: nonlinearity can be handled effectively through progressive linearization and transformation realized by deep networks and representations (see Chapter 5). Moreover, we will demonstrate in this book how all the mechanisms listed above can be naturally integrated into a complete system which exhibits certain characteristics of an autonomous intelligent system (see Chapter 6).

**Origin of artificial intelligence.** The subtitle of Wiener's *Cybernetics—Control and Communication in the Animal and the Machine*—reveals that 1940s research aimed primarily at emulating animal-level intelligence. As noted earlier, the agendas of that era were dominated by Wiener's Cybernetics movement.

Alan Turing was among the first to recognize this limitation. In his celebrated 1950 paper "*Computing Machinery and Intelligence*" [Tur50], Turing formally posed the question of "can machines think?" In other words, whether machines could imitate human-level intelligence, to the point of machine intelligence being indistinguishable from human intelligence—now known as *the Turing test*.

Around 1955, a group of ambitious young scientists sought to break away from the then-dominant Cybernetics program and establish their own legacy. They accepted Turing's challenge of imitating human intelligence and proposed a workshop at Dartmouth College to be held in the summer of 1956. Their proposal stated [MMR+06]:

> "*The study is to proceed on the basis of the conjecture that every aspect of learning or any other feature of intelligence can in principle be so precisely described that a machine can be made to simulate it. An attempt will be made to find how to make machines use language, form abstractions and concepts, solve kinds of problems now reserved for humans, and improve themselves.*"

They aimed to formalize and study the higher-level intelligence that distinguishes humans from animals. Their agenda included abstraction, symbolic methods, natural language, and deductive reasoning (causal inference, logical deduction, etc.), i.e., scientific intelligence. The organizer of the workshop, John McCarthy—then a young assistant professor of mathematics at Dartmouth College—coined the now-famous term "Artificial Intelligence" (AI) to describe machines that exhibit scientific intelligence in order to formally describe the problems and goals of the workshop.

**Renaissance of "artificial intelligence" or "cybernetics"?** Over the past decade, machine intelligence has undergone explosive development, driven largely by deep artificial neural networks, sparked by the 2012 work of Geoffrey Hinton and his students [KSH12]. This period is hailed as the "Renaissance of AI." Yet, in terms of the tasks actually tackled (recognition, generation, prediction) and the techniques developed (reinforcement learning, imitation learning, encoding, decoding, denoising, and compression), we are largely emulating mechanisms common to the intelligence of early life and animals. Even in this regard, as we will clarify in this book, current "AI" models and systems have not fully or correctly implemented all necessary mechanisms for intelligence at the animal level which were known to the 1940s Cybernetics movement.

Strictly speaking, the recent advances of machine intelligence in the past decade do not align well with the 1956 Dartmouth AI program. Instead, what has been achieved is closer to the objectives of Wiener's classic 1940s Cybernetics program. It is probably more appropriate to call the current era the "Renaissance of Cybernetics." The recent rise of so-called "Embodied AI" for autonomous robots and agents aligns even more closely with the goals of the Cybernetics program. Only after we fully understand, from scientific and mathematical perspectives, what we have truly accomplished can we determine what problems remains open and which directions lead to the true nature of intelligence.[7] That is one of the main purposes of this book.

## 1.2 What to Learn?

### 1.2.1 Predictability

Data that carry useful information manifest in many forms. In their most natural form, they can be modeled as sequences that are predictable and computable. The notion of a predictable and computable sequence was central to the theory of information [Sha48] and computing and largely led to the invention of computers [Tur36]. The role of predictable sequences for (inductive) inference was studied by Ray Solomonoff, Andrey Kolmogorov, and others in the 1960s [Kol98] as a generalization of Claude Shannon's classic information theory [Sha48]. To help readers understand the concept of predictable sequences, we begin with some concrete examples.

**Scalar case.** The simplest predictable discrete sequence is arguably the sequence of natural numbers:

$$S = 1, 2, 3, 4, 5, 6, \ldots, n, n+1, \ldots \tag{1.2.1}$$

in which the next number $x_{n+1}$ is defined as the previous number $x_n$ plus 1:

$$x_{n+1} = x_n + 1. \tag{1.2.2}$$

One may generalize the notion of predictability to any sequence $\{x_n\}_{n=1}^{\infty}$ with $x_n \in \mathbb{R}$ if the next number $x_{n+1}$ can always be computed from its predecessor $x_n$:

$$x_{n+1} = f(x_n), \quad x_n \in \mathbb{R},\ n = 1, 2, 3, \ldots \tag{1.2.3}$$

[7] So we will resume our discussion at the end of this book in the last Chapter 9.

where $f(\cdot)$ is a *computable* (scalar) function.[8] Alan Turing's seminal work in 1936 [Tur36] gives a rigorous definition of computability. In practice, we often further assume that $f$ is efficiently computable and has nice properties such as continuity and differentiability. The necessity of these properties will become clear later once we understand more refined notions of computability and their roles in machine learning and intelligence.

**Multivariable case.** The next value may also depend on two predecessors. For example, the famous *Fibonacci sequence*

$$S = 1, 1, 2, 3, 5, 8, 13, 21, 34, 55, \ldots \tag{1.2.4}$$

satisfies

$$x_{n+2} = x_{n+1} + x_n, \quad x_n \in \mathbb{R},\ n = 1, 2, 3, \ldots \tag{1.2.5}$$

More generally, we may write

$$x_{n+2} = f(x_{n+1}, x_n), \quad x_n \in \mathbb{R},\ n = 1, 2, 3, \ldots \tag{1.2.6}$$

for any computable function $f$ that takes two inputs. Extending further, the next value may depend on the preceding $d$ values:

$$x_{n+d} = f(x_{n+d-1}, \ldots, x_n), \quad x_n \in \mathbb{R},\ n = 1, 2, 3, \ldots \tag{1.2.7}$$

The integer $d$ is called the *degree* of the recursion. The above expression (1.2.7) is called an *autoregression*, and the resulting sequence is *autoregressive*. When $f$ is linear, we say it is a *linear autoregression*.

**Vector case.** To simplify notation, we define a vector $\boldsymbol{x} \in \mathbb{R}^d$ that collects $d$ consecutive values in the sequence:

$$\boldsymbol{x}_n \doteq [x_{n+d-1}, \ldots, x_n]^\top, \quad \boldsymbol{x}_n \in \mathbb{R}^d,\ n = 1, 2, 3, \ldots \tag{1.2.8}$$

With this notation, the recursive relation (1.2.7) becomes

$$\boldsymbol{x}_{n+1} = g(\boldsymbol{x}_n) \in \mathbb{R}^d, \quad n = 1, 2, 3, \ldots \tag{1.2.9}$$

where $g(\cdot)$ is uniquely determined by the function $f$ in (1.2.7) and maps a $d$-dimensional vector to a $d$-dimensional vector. In different contexts, such a vector is sometimes called a "state" or a "token." Note that (1.2.7) defines a mapping $\mathbb{R}^d \to \mathbb{R}$, whereas here we have $g\colon \mathbb{R}^d \to \mathbb{R}^d$.

[8] Here we emphasize that the function $f(\cdot)$ itself is computable, meaning it can be implemented as a program on a computer.

**Controlled prediction.** We may also define a predictable sequence that depends on another predictable sequence as input:

$$\boldsymbol{x}_{n+1} = f(\boldsymbol{x}_n, \boldsymbol{u}_n) \in \mathbb{R}^d, \quad n = 1, 2, 3, \ldots, \tag{1.2.10}$$

where $\{\boldsymbol{u}_n\}$ with $\boldsymbol{u}_n \in \mathbb{R}^k$ is a (computable) predictable sequence. In other words, the next vector $\boldsymbol{x}_{n+1} \in \mathbb{R}^d$ depends on both $\boldsymbol{x}_n \in \mathbb{R}^d$ and $\boldsymbol{u}_n \in \mathbb{R}^k$. In control theory, the sequence $\{\boldsymbol{u}_n\}$ is often referred to as the "control input" and $\boldsymbol{x}_n$ as the "state" or "output" of the system (1.2.10). A classic example is a linear dynamical system:

$$\boldsymbol{x}_{n+1} = \boldsymbol{A}\boldsymbol{x}_n + \boldsymbol{B}\boldsymbol{u}_n, \quad \boldsymbol{A} \in \mathbb{R}^{d\times d}, \boldsymbol{B} \in \mathbb{R}^{d\times k}, \tag{1.2.11}$$

which is widely studied in control theory [CD91].

Often the control input is given by a computable function of the state $\boldsymbol{x}_n$ itself:

$$\boldsymbol{u}_n = h(\boldsymbol{x}_n), \quad n = 1, 2, 3, \ldots \tag{1.2.12}$$

As a result, the sequence $\{\boldsymbol{x}_n\}$ is given by composing the two computable functions $f$ and $h$:

$$\boldsymbol{x}_{n+1} = f(\boldsymbol{x}_n, h(\boldsymbol{x}_n)), \quad n = 1, 2, 3, \ldots \tag{1.2.13}$$

In this way, the sequence $\{\boldsymbol{x}_n\}$ again becomes an autoregressive predictable sequence. When the input $\boldsymbol{u}_n$ depends on the output $\boldsymbol{x}_n$, we say the resulting sequence is produced by a "closed-loop" system (1.2.13). As the closed-loop system no longer depends on any external input, we say such a system has become *autonomous*. It can be viewed as a special case of autoregression. For instance, if we choose $\boldsymbol{u}_n = \boldsymbol{F}\boldsymbol{x}_n$ in the above linear system (1.2.11), the closed-loop system becomes

$$\boldsymbol{x}_{n+1} = \boldsymbol{A}\boldsymbol{x}_n + \boldsymbol{B}\boldsymbol{u}_n = \boldsymbol{A}\boldsymbol{x}_n + \boldsymbol{B}\boldsymbol{F}\boldsymbol{x}_n = (\boldsymbol{A} + \boldsymbol{B}\boldsymbol{F})\boldsymbol{x}_n, \tag{1.2.14}$$

which is a linear autoregression.

**Continuous processes.** Predictable sequences have natural continuous counterparts, which we call predictable processes. The simplest such process is time itself, $x(t) = t$.

More generally, a process $\boldsymbol{x}(t)$ is *predictable* if, at every time $t$, its value at $t + \delta t$ is determined by its value at $t$, where $\delta t$ is an infinitesimal increment. Typically, $\boldsymbol{x}(t)$ is continuous and smooth, so the change $\delta\boldsymbol{x}(t) = \boldsymbol{x}(t+\delta t) - \boldsymbol{x}(t)$ is infinitesimally small. Such processes are typically described by (multivariate) differential equations

$$\dot{\boldsymbol{x}}(t) = f(\boldsymbol{x}(t)), \quad \boldsymbol{x} \in \mathbb{R}^d. \tag{1.2.15}$$

In systems theory [CD91; Sas99], the equation (1.2.15) is known as a state-space model. A controlled process is given by

$$\dot{\boldsymbol{x}}(t) = f(\boldsymbol{x}(t), \boldsymbol{u}(t)), \quad \boldsymbol{x} \in \mathbb{R}^d, \boldsymbol{u} \in \mathbb{R}^k, \tag{1.2.16}$$

where $\boldsymbol{u}(t)$ is a computable input process.

*Example* 1.1. Newton's second law predicts the trajectory $\boldsymbol{x}(t) \in \mathbb{R}^3$ of a moving object under a force $\boldsymbol{F}(t) \in \mathbb{R}^3$:

$$m\ddot{\boldsymbol{x}}(t) = \boldsymbol{F}(t). \tag{1.2.17}$$

When there is no force, i.e., $\boldsymbol{F}(t) \equiv \mathbf{0}$, this reduces to Newton's first law: the object moves at constant velocity $\boldsymbol{v} \in \mathbb{R}^3$:

$$\ddot{\boldsymbol{x}}(t) = \mathbf{0} \iff \text{there exists } \boldsymbol{v} \in \mathbb{R}^3 \text{ such that } \dot{\boldsymbol{x}}(t) = \boldsymbol{v}. \tag{1.2.18}$$

■

### 1.2.2 Low Dimensionality

**Learning to predict.** Now suppose you have observed or have been given many sequence segments:

$$\{\boldsymbol{x}_1, \boldsymbol{x}_2, \dots, \boldsymbol{x}_i, \dots, \boldsymbol{x}_N\} \tag{1.2.19}$$

drawn from a predictable sequence $\{x_n\}_{n=1}^{\infty}$. Without loss of generality, assume each segment has length $D \gg d$, so

$$\boldsymbol{x}_i = [x_{j(i)}, x_{j(i)+1}, \dots, x_{j(i)+D-1}]^\top \in \mathbb{R}^D \tag{1.2.20}$$

for some $j \in \mathbb{N}$. You are then given a new segment $\boldsymbol{x}_t$ and asked to predict its future values.

The difficulty is that the generating function $f$ and its order $d$ are unknown:

$$x_{n+d} = f(x_{n+d-1}, \dots, x_n). \tag{1.2.21}$$

The goal is therefore to learn $f$ and $d$ from the sample segments $\boldsymbol{x}_1, \boldsymbol{x}_2, \dots, \boldsymbol{x}_N$. The central task of learning to predict is:

*Given many sampled segments of a predictable sequence, how can we effectively and efficiently identify the function $f$?*

**Predictability and low-dimensionality.** To identify the predictive function $f$, we may notice a common characteristic of segments of any predictable sequence given by (1.2.21). If we take a long segment, say of length $D \gg d$, and view it as a vector

$$\boldsymbol{x}_i = [x_i, x_{i+1}, \dots, x_{i+D-1}]^\top \in \mathbb{R}^D, \tag{1.2.22}$$

then the set of all such vectors $\{\boldsymbol{x}_i\}$ is far from random and cannot occupy the entire space $\mathbb{R}^D$. Instead, it has at most $d$ degrees of freedom—given the first $d$ entries of any $\boldsymbol{x}_i$, the remaining entries are uniquely determined. In other words, all $\{\boldsymbol{x}_i\}$ lie on a $d$-dimensional surface. In mathematics, such a surface is called a submanifold, denoted $\mathcal{M} \subset \mathbb{R}^D$.

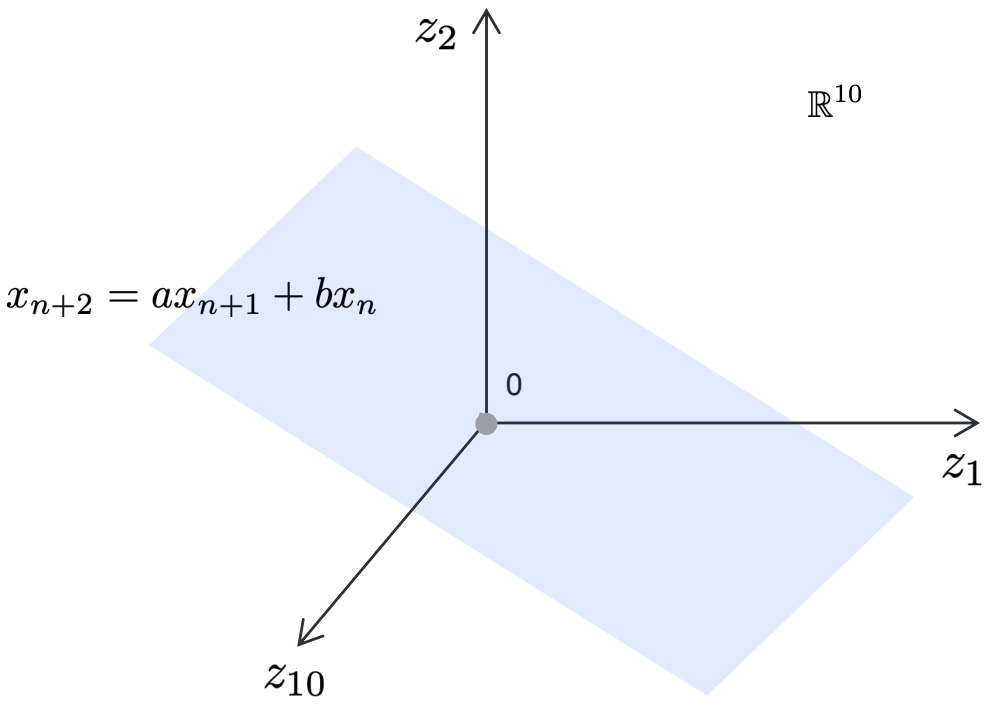

Figure 1.6: A two-dimensional subspace in a ten-dimensional ambient space.

In practice, if we choose the segment length $D$ large enough, all segments sampled from the same predicting function lie on a surface with intrinsic dimension $d$, significantly lower than that of the ambient space $D$. For example, if the sequence is given by the linear autoregression

$$x_{n+2} = ax_{n+1} + bx_n, \tag{1.2.23}$$

for some constants $a, b \in \mathbb{R}$, and we sample segments of length $D = 10$, then all samples lie on a two-dimensional plane in $\mathbb{R}^{10}$, as illustrated in Figure 1.6. Identifying this two-dimensional subspace fully determines the constants $a$ and $b$ in (1.2.23).

More generally, when the predicting function $f$ is linear, as in the systems given in (1.2.11) and (1.2.14), the long segments always lie on a low-dimensional linear subspace. Identifying the predicting function is then largely equivalent to identifying this subspace, a problem known as principal component analysis. We will discuss such classic models and methods in Chapter 2.

This observation extends to general predictable sequences: if we can identify the low-dimensional surface on which the segment samples lie, we can identify the predictive function $f$.[9] We cannot overemphasize the importance of this property: *All samples of long segments of a predictable sequence lie on a low-dimensional submanifold.* To a large extent, modern science and mathematics are precisely trying to identify and describe such low-dimensional structure of observed phenomena. This explains why the most important and famous scientific discoveries about our physical world are often described in terms of equations, as Figure 1.7 illustrated, because equations restrict the otherwise independent observed quantities onto lower-dimensional manifolds.

According the String Theory that unifies almost all physical laws we know about the world, the entire universe evolves in a space of 10 dimension, starting

[9] Under mild conditions, there is a one-to-one mapping between the low-dimensional surface and the function $f$. This fact has been exploited in problems such as system identification, which we will discuss later.

**17 Equations That Changed the World**
by Ian Stewart

| | | | |
|---|---|---|---|
| 1. | **Pythagoras's Theorem** | $a^2 + b^2 = c^2$ | Pythagoras,530 BC |
| 2. | **Logarithms** | $\log xy = \log x + \log y$ | John Napier, 1610 |
| 3. | **Calculus** | $\frac{\mathrm{d}f}{\mathrm{d}t} = \lim_{h \to 0} \frac{f(t+h) - f(t)}{h}$ | Newton, 1668 |
| 4. | **Law of Gravity** | $F = G\frac{m_1 m_2}{r^2}$ | Newton, 1687 |
| 5. | **The Square Root of Minus One** | $i^2 = -1$ | Euler, 1750 |
| 6. | **Euler's Formula for Polyhedra** | $V - E + F = 2$ | Euler, 1751 |
| 7. | **Normal Distribution** | $\Phi(x) = \frac{1}{\sqrt{2\pi\rho}} e^{\frac{(x-\mu)^2}{2\rho^2}}$ | C.F. Gauss, 1810 |
| 8. | **Wave Equation** | $\frac{\partial^2 u}{\partial t^2} = c^2 \frac{\partial^2 u}{\partial x^2}$ | J. d'Almbert, 1746 |
| 9. | **Fourier Transform** | $f(\omega) = \int_{\infty}^{\infty} f(x) e^{-2\pi i x \omega} \mathrm{d}\, x$ | J. Fourier, 1822 |
| 10. | **Navier-Stokes Equation** | $\rho \left( \frac{\partial \mathbf{v}}{\partial t} + \mathbf{v} \cdot \nabla \mathbf{v} \right) = -\nabla p + \nabla \cdot \mathbf{T} + \mathbf{f}$ | C. Navier, G. Stokes, 1845 |
| 11. | **Maxwell's Equations** | $\nabla \cdot \mathbf{E} = \frac{\rho}{\varepsilon_0}$ $\quad \nabla \cdot \mathbf{H} = 0$ <br> $\nabla \times \mathbf{E} = -\frac{1}{c}\frac{\partial \mathbf{H}}{\partial t}$ $\quad \nabla \times \mathbf{H} = \frac{1}{c}\frac{\partial E}{\partial t}$ | J.C. Maxwell, 1865 |
| 12. | **Second Law of Thermodynamics** | $\mathrm{d}S \geq 0$ | L. Boltzmann, 1874 |
| 13. | **Relativity** | $E = mc^2$ | Einstein, 1905 |
| 14. | **Schrodinger's Equation** | $i\hbar \frac{\partial}{\partial t} \Psi = H \Psi$ | E. Schrodinger, 1927 |
| 15. | **Information Theory** | $H = -\sum p(x) \log p(x)$ | C. Shannon, 1949 |
| 16. | **Chaos Theory** | $x_{t+1} = k x_t (1 - x_t)$ | Robert May, 1975 |
| 17. | **Black-Scholes Equation** | $\frac{1}{2}\sigma^2 S^2 \frac{\partial^2 V}{\partial S^2} + rS\frac{\partial V}{\partial S} + \frac{\partial V}{\partial t} - rV = 0$ | F. Black, M. Scholes, 1990 |

Figure 1.7: Famous 17 equations that changed the world. Most of these equations describe physical laws of the universe. For intelligence, the equation 15 of Entropy will play a very special role, as we will explain why in this book, especially in Chapter 3.

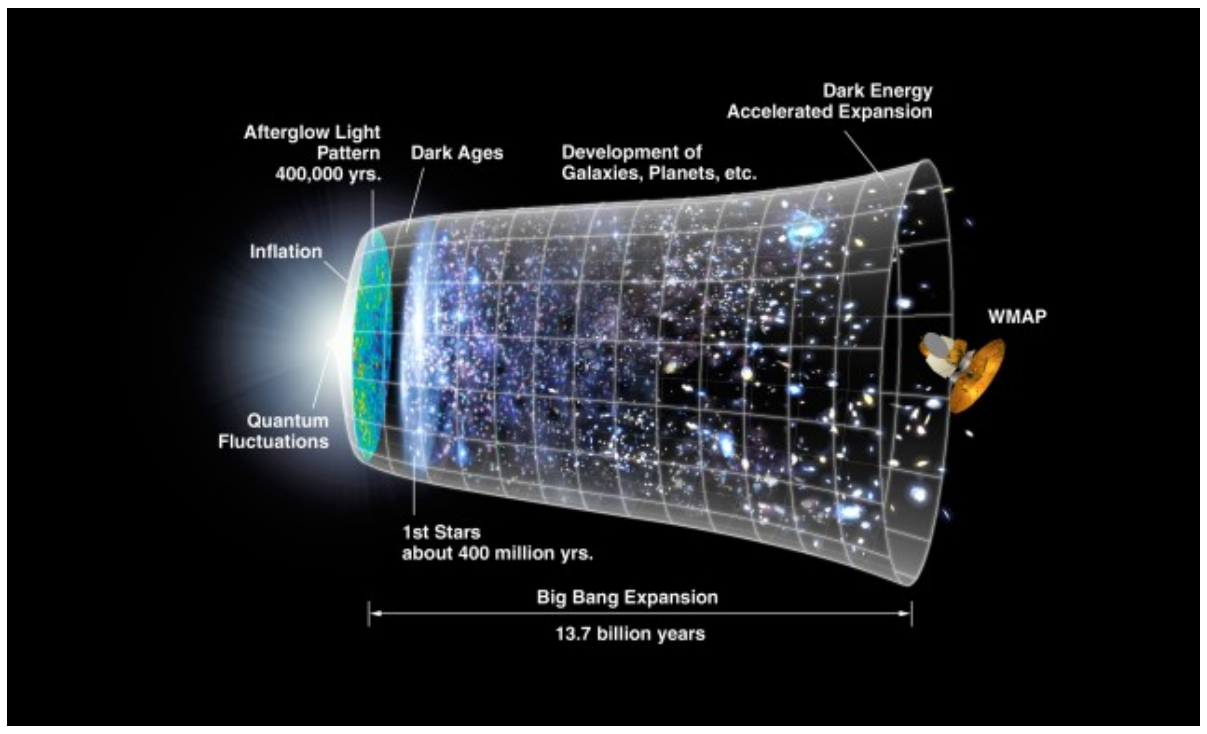

Figure 1.8: The evolution of the Universe that is, based on the best physics theory so far, believed to be in a space of 10 dimensions.

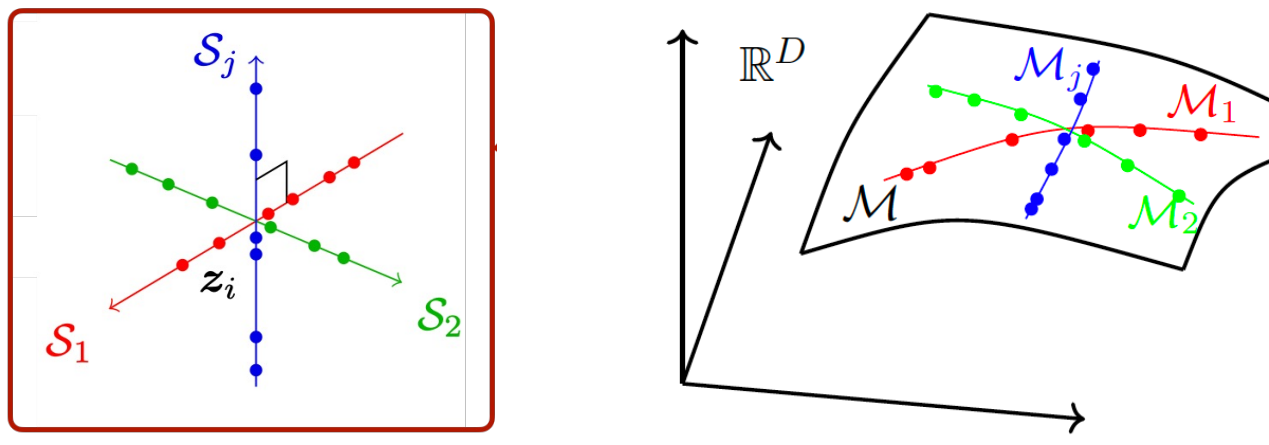

Figure 1.9: Data distributed on a mixture of (orthogonal) subspaces $\{\mathcal{S}_j\}$ (left) or submanifolds $\{\mathcal{M}_j\}$ (right).

from the Big Bang, see Figure 1.8. As we will see in this book, all modern learning methods also exploit the low-dimensionality property in data, either implicitly or explicitly.

*Remark* 1.1 (Entropy for Physics versus that for Intelligence). In Figure 1.7, we see that the only physical law that is *not* described in the form of an equation is the Second Law of Thermodynamics which states that the entropy of any closed physical system, like the whole universe, always *increases*. We will see in this book that the goal of intelligence (and science) is to identify structures in observed phenomena that can *decrease* the entropy in our prediction. This will be studied rather extensively in Chapter 3 and Appendix B.

In real-world scenarios, observed data often come from multiple predictable sequences. For example, a video sequence may contain several moving objects. In such cases, the data lie on a mixture of low-dimensional linear subspaces or nonlinear submanifolds, as illustrated in Figure 1.9.

**Properties of low-dimensionality.** Temporal correlation in predictable sequences is not the only reason data are low-dimensional. For example, the space of all images is vast, yet most of it consists of structureless random images, as shown in Figure 1.10 (left). Natural images and videos, however, are highly redundant because of strong spatial and temporal correlations among pixel values. This redundancy allows us to recognize easily whether an image is noisy or clean, as shown in Figure 1.10 (middle and right). Consequently, the distribution of natural images has a very low intrinsic dimension relative to the total number of pixels.

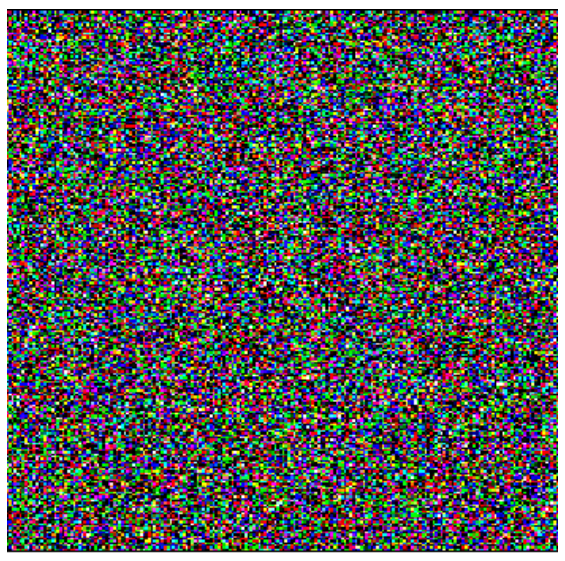 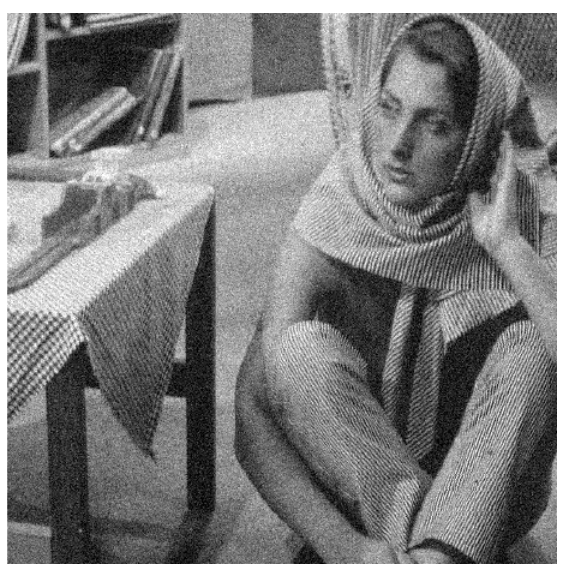 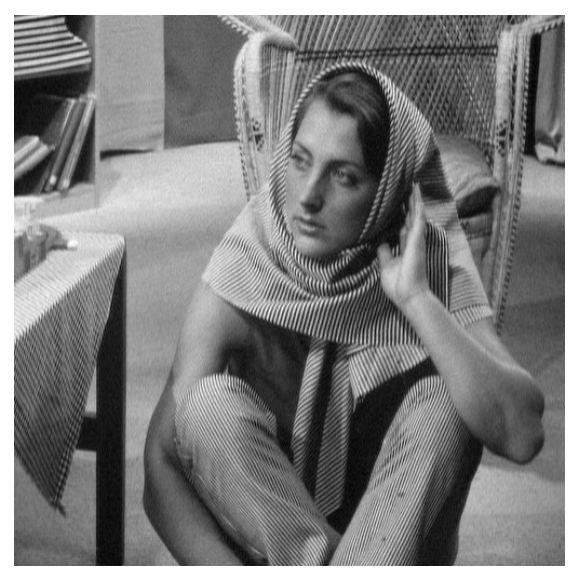

Figure 1.10: An image of random noise (left) versus a noisy image (middle) and the original clean image (right).

Because learning low-dimensional structures is both important and ubiquitous, the book *High-Dimensional Data Analysis with Low-Dimensional Models: Principles, Computation, and Applications* [WM22] begins with the statement: "*The problem of identifying the low-dimensional structure of signals or data in high-dimensional spaces is one of the most fundamental problems that, through a long history, interweaves many engineering and mathematical fields such as system theory, signal processing, pattern recognition, machine learning, and statistics.*"

By constraining the observed data point $\boldsymbol{x}$ to lie on a low-dimensional surface or submanifold, we make its entries highly dependent on or correlated to one another and, in a sense, "predictable" from the values of other entries. For example, if we know the data are constrained to a $d$-dimensional surface in $\mathbb{R}^D$, we can perform several useful tasks beyond prediction:

- **completion**: given more than $d$ entries of a typical sample $\boldsymbol{x}$, the remaining entries can usually be uniquely determined;[10]
- **denoising**: if the entries of a sample $\boldsymbol{x}$ are perturbed by small noise, the noise can be effectively removed by projecting $\boldsymbol{x}$ back onto the surface;
- **error correction**: if a small number of unknown entries of $\boldsymbol{x}$ are arbitrarily corrupted, they can be efficiently corrected.

Figure 1.11 illustrates these properties using a low-dimensional linear structure—a one-dimensional line in a two-dimensional plane.

Under mild conditions, these properties generalize to many other low-dimensional structures in high-dimensional spaces [WM22]. As we will see, these useful

[10]Prediction becomes a special case of this property.

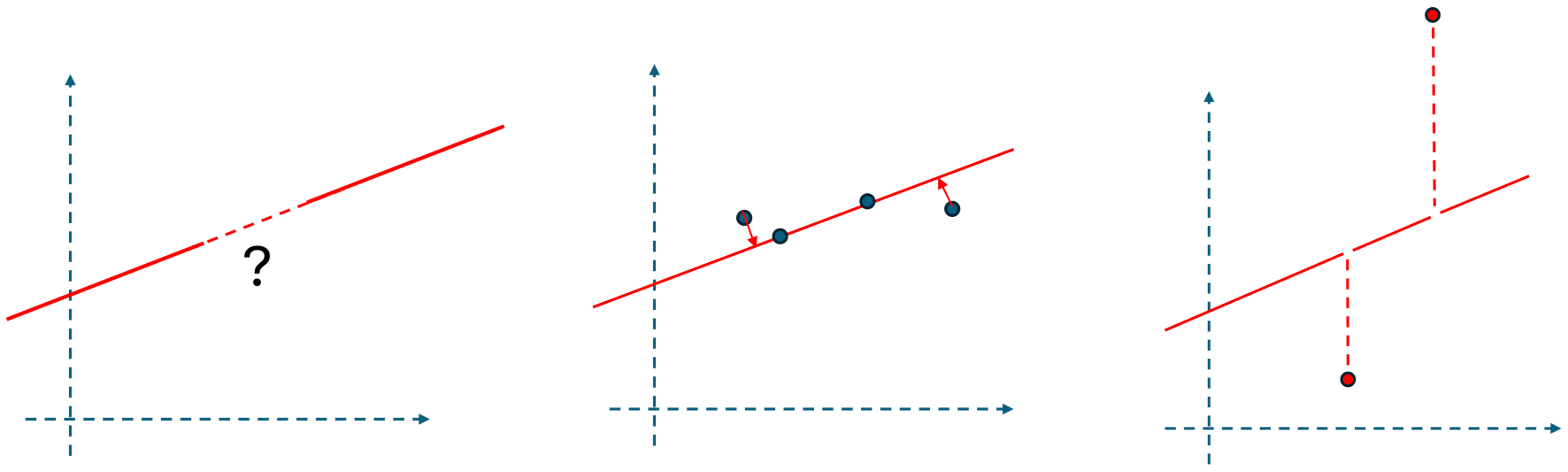

Figure 1.11: Illustration of properties of a low-dimensional (linear) structure: it enables completion (left), denoising (middle), and error correction (right).

properties—completion and denoising, for example—inspire effective methods for learning such structures.

For simplicity, we have so far used the deterministic case to introduce the notions of predictability and low-dimensionality, where data lie precisely on geometric structures such as subspaces or surfaces. In practice, however, data always contain some uncertainty or randomness. In this case, we may assume the data follow a probability distribution with density $p(\boldsymbol{x})$. A distribution is considered "low-dimensional" if its density concentrates around a low-dimensional geometric structure—a subspace, a surface, or a mixture thereof—as shown in Figure 1.9. This structure is also called the "support" of the distribution. Once learned, such a density $p(\boldsymbol{x})$ serves as a powerful prior for estimating $\boldsymbol{x}$ from partial, noisy, or corrupted observations:

$$\boldsymbol{y} = f(\boldsymbol{x}) + \boldsymbol{n}, \tag{1.2.24}$$

by computing the conditional estimate $\hat{\boldsymbol{x}}(\boldsymbol{y}) = \mathbb{E}(\boldsymbol{x} \mid \boldsymbol{y})$ or by sampling the conditional distribution $\hat{\boldsymbol{x}}(\boldsymbol{y}) \sim p(\boldsymbol{x} \mid \boldsymbol{y})$.[11]

Many prior works have studied the properties of data sets whose distributions are supported on a low-dimensional surface or submanifold, e.g., [ACM12; LMR17]. We here would like to single out its ultimate importance and claims it as the single assumption on which this book bases its deductive approach to understanding deep networks and intelligence in general:

> **The main assumption:** *Any intelligent system or learning method should (and can) rely on the predictability of the world; hence, the distribution of observed high-dimensional data samples has low-dimensional support.*

The remaining question is how to learn these low-dimensional structures correctly and computationally efficiently from high-dimensional data. As we will see, parametric models studied in classical analytical approaches and non-parametric models such as deep networks, popular in modern practice, are simply different means to the same end.

[11] Modern generative AI technologies, such as conditioned image generation, rely heavily on this principle, as we will discuss in Chapter 7.

## 1.3 How to Learn?

### 1.3.1 Analytical Model-Based Approaches

Note that even if a predictive function is tractable to compute, it does not imply that it is tractable or scalable to learn this function from a finite number of sampled segments. One classical approach to ensure tractability is to make explicit assumptions about the family of low-dimensional structures we are dealing with. Historically, due to limited computation and data, simple and idealistic analytical models were the first to be studied, as they often offer efficient closed-form or numerical solutions. In addition, they provide insights into more general problems and already yield useful solutions to important, though limited, cases. In the old days, when computational resources were scarce, analytical models that permitted efficient closed-form or numerical solutions were the only cases that could be implemented. *Linear structures* became the first class of models to be thoroughly studied.

For example, arguably the simplest case is to assume the data are distributed around a single low-dimensional subspace in a high-dimensional space. Somewhat equivalently, one may assume the data are distributed according to an almost degenerate low-dimensional Gaussian. Identifying such a subspace or Gaussian from a finite number of (noisy) samples is the classical problem of principal component analysis (PCA), and effective algorithms have been developed for this class of models [Jol02]. One can make the family of models increasingly more complex and expressive. For instance, one may assume the data are distributed around a mixture of low-dimensional components (subspaces or low-dimensional Gaussians), as in independent component analysis (ICA) [BJC85], dictionary learning (DL), generalized principal component analysis (GPCA) [VMS05], or the more general class of sparse low-dimensional models that have been studied extensively in recent years in fields such as compressive sensing [WM22].

Across all these analytical model families, the central problem is to identify the most compact model within each family that best fits the given data. Below, we give a brief account of these classical analytical models but leave a more systematic study to Chapter 2. In theory, these analytical models have provided tremendous insights into the geometric and statistical properties of low-dimensional structures. They often yield closed-form solutions or efficient and scalable algorithms, which are very useful for data whose distributions can be well approximated by such models. More importantly, for more general problems, they provide a sense of how easy or difficult the problem of identifying low-dimensional structures can be and what the basic ideas are to approach such a problem.

### Linear Dynamical Systems

**Wiener filter.** As discussed in Section 1.2.1, a central task of intelligence is to learn what is predictable in sequences of observations. The simplest class of predictable sequences—or signals—are those generated by a *linear time-invariant* (LTI) process:

$$x[n] = h[n] * z[n] + \epsilon[n], \tag{1.3.1}$$

where $*$ is the convolution operation, $z$ is the input and $h$ is the impulse response function.[12] Here $\epsilon[n]$ denotes additive observation noise. Given the input process $\{z[n]\}$ and observations of the output process $\{x[n]\}$, the goal is to find the optimal $h[n]$ such that $\hat{x}[n] = h[n] * z[n]$ predicts $x[n]$ optimally. Prediction quality (i.e., goodness) is measured by the mean squared error (MSE):

$$\min_h \mathbb{E}[\epsilon[n]^2] = \mathbb{E}[\|x[n] - h[n] * z[n]\|_2^2]. \tag{1.3.2}$$

The optimal solution $h[n]$ is called a (denoising) filter. Norbert Wiener—who also initiated the Cybernetics movement—studied this problem in the 1940s and derived an elegant closed-form solution known as the *Wiener filter* [Wie42; Wie49]. This result, also known as least-variance estimation and filtering, became one of the cornerstones of signal processing. Interested readers may refer to [MKS+04, Appendix B] for a detailed derivation of this type of estimator.

**Kalman filter.** The idea of denoising or filtering a dynamical system was later extended by Rudolph Kalman in the 1960s to a linear time-invariant system described by a finite-dimensional state-space model:

$$\boldsymbol{z}[n] = \boldsymbol{A}\boldsymbol{z}[n-1] + \boldsymbol{B}\boldsymbol{u}[n] + \boldsymbol{\epsilon}[n]. \tag{1.3.3}$$

The problem is to estimate the system state $\boldsymbol{z}[n]$ from noisy observations of the form

$$\boldsymbol{x}[n] = \boldsymbol{C}\boldsymbol{z}[n] + \boldsymbol{w}[n], \tag{1.3.4}$$

where $\boldsymbol{w}$ is (white) noise. The optimal causal[13] state estimator that minimizes the minimum-MSE (MMSE) prediction error

$$\min \mathbb{E}[\|\boldsymbol{x}[n] - \boldsymbol{C}\boldsymbol{z}[n]\|_2^2] \tag{1.3.5}$$

is given in closed form by the so-called *Kalman filter* [Kal60]. This result is a cornerstone of modern control theory because it enables estimation of a dynamical system's state from noisy observations. One can then introduce linear state feedback, for example $\boldsymbol{u}[n] = \boldsymbol{F}\hat{\boldsymbol{z}}[n]$, and render the closed-loop system fully autonomous, as shown in Equation (1.2.13). Interested readers may refer to [MKS+04, Appendix B] for a detailed derivation of the Kalman filter.

[12] Typically, $h$ is assumed to have certain desirable properties, such as finite length or a band-limited spectrum.

[13] This means the estimator can use only observations up to the current time step $n$. The Kalman filter is always causal, whereas the Wiener filter need not be.

**Identification of linear dynamical systems.** To derive the Kalman filter, the system parameters $\boldsymbol{A}, \boldsymbol{B}, \boldsymbol{C}$ are assumed to be known. If they are not given in advance, the problem becomes more challenging and is known as *system identification*: how to *learn* the parameters $\boldsymbol{A}, \boldsymbol{B}, \boldsymbol{C}$ from many samples of the input sequence $\{\boldsymbol{u}[n]\}$ and observation sequence $\{\boldsymbol{x}[n]\}$. This is a classic problem in systems theory. If the system is linear, the input and output sequences $\{\boldsymbol{u}[n], \boldsymbol{x}[n]\}$ jointly lie on a certain low-dimensional subspace[14]. Hence, the identification problem is essentially equivalent to identifying this low-dimensional subspace [LV09; LV10; VM96].

Note that the above problems have two things in common: first, the noise-free sequences or signals are assumed to be generated by an explicit family of parametric models; second, these models are essentially linear. Conceptually, let $\boldsymbol{x}_o$ be a random variable whose "true" distribution is supported on a low-dimensional linear subspace, say $S$. To a large extent, the Wiener filter and Kalman filter both try to estimate such an $\boldsymbol{x}_o$ from its noisy observations:

$$\boldsymbol{x} = \boldsymbol{x}_o + \boldsymbol{\epsilon}, \quad \boldsymbol{x}_o \sim S, \tag{1.3.6}$$

where $\boldsymbol{\epsilon}$ is typically a random Gaussian noise (or process). Hence, their solutions rely on identifying a low-dimensional linear subspace that best fits the observed noisy data. By projecting the data onto this subspace, one obtains the optimal denoising operations, all in closed form.

### Linear and Mixed Linear Models

**Principal component analysis.** From the above problems in classical signal processing and system identification, we see that the main task behind all these problems is to learn a *single* low-dimensional linear subspace from noisy observations. Mathematically, we may model such a structure as

$$\boldsymbol{x} = \boldsymbol{u}_1 z_1 + \boldsymbol{u}_2 z_2 + \cdots + \boldsymbol{u}_d z_d + \boldsymbol{\epsilon} = \boldsymbol{U}\boldsymbol{z} + \boldsymbol{\epsilon}, \quad \boldsymbol{U} \in \mathbb{R}^{D \times d}, \tag{1.3.7}$$

where $\boldsymbol{\epsilon} \in \mathbb{R}^D$ is small random noise. Figure 1.12 illustrates such a distribution with two principal components. The problem is to find the subspace basis $\boldsymbol{U}$ from many samples of $\boldsymbol{x}$. A typical approach is to minimize the projection error onto the subspace:

$$\min_{\boldsymbol{U}} \mathbb{E}[\|\boldsymbol{x} - \boldsymbol{U}\boldsymbol{U}^\top \boldsymbol{z}\|_2^2]. \tag{1.3.8}$$

This is essentially a denoising task: once the basis $\boldsymbol{U}$ is correctly found, we can denoise the noisy sample $\boldsymbol{x}$ by projecting it onto the low-dimensional subspace spanned by $\boldsymbol{U}$:

$$\boldsymbol{x} \to \hat{\boldsymbol{x}} = \boldsymbol{U}\boldsymbol{U}^\top \boldsymbol{x}. \tag{1.3.9}$$

If the noise is small and the low-dimensional subspace $\boldsymbol{U}$ is correctly learned, we expect $\boldsymbol{x} \approx \hat{\boldsymbol{x}}$. Thus, PCA is a special case of a so-called "auto-encoding"

[14] which has the same dimension as the order of the state-space model (1.3.3).

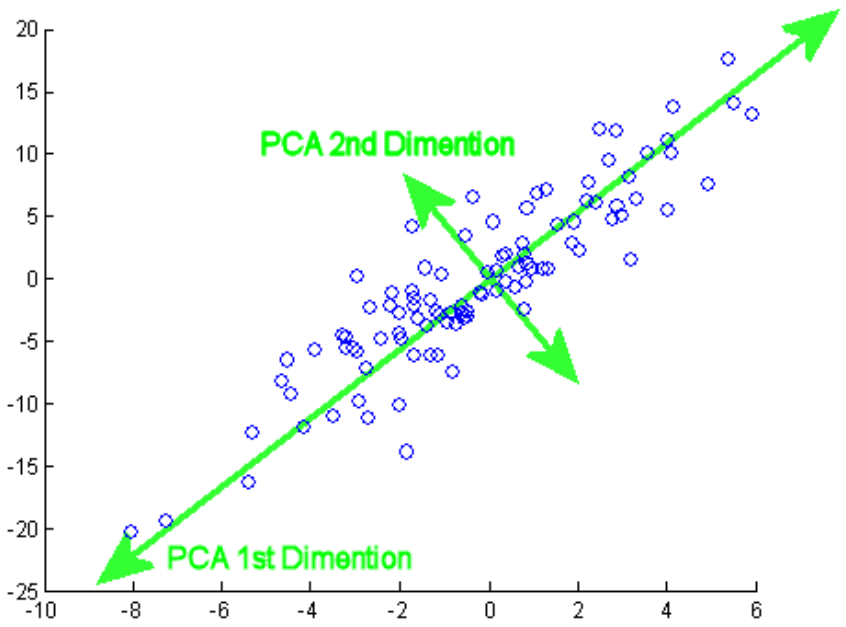

Figure 1.12: A distribution with two principal components.

scheme, which encodes the data by projecting it into a lower-dimensional space, and decodes a lower-dimensional code by it into the original space:

$$\boldsymbol{x} \xrightarrow{\boldsymbol{U}^\top} \boldsymbol{z} \xrightarrow{\boldsymbol{U}} \hat{\boldsymbol{x}}. \tag{1.3.10}$$

Because of the simple data structure, the encoder $\mathcal{E}$ and decoder $\mathcal{D}$ both become simple linear operators (projecting and lifting).

This classic problem in statistics is known as principal component analysis (PCA). It was first studied by Pearson in 1901 [Pea01] and later independently by Hotelling in 1933 [Hot33]. The topic is systematically summarized in the classic book [Jol02; Jol86]. One may also explicitly assume the data $\boldsymbol{x}$ is distributed according to a single low-dimensional Gaussian:

$$\boldsymbol{x} \sim \mathcal{N}(\mathbf{0}, \boldsymbol{U}\boldsymbol{U}^\top + \sigma \boldsymbol{I}), \quad \boldsymbol{U} \in \mathbb{R}^{D \times d}, \tag{1.3.11}$$

which is equivalent to assuming that $\boldsymbol{z}$ in the PCA model (1.3.7) is a standard normal distribution. This problem is known as probabilistic PCA [TB99] and has the same computational solution as PCA.

In this book, we revisit PCA in Chapter 2 from the perspective of learning a low-dimensional distribution. Our goal is to use this simple and idealistic model to convey some of the most fundamental ideas for learning a compact representation of a low-dimensional distribution, including the important notions of compression via denoising and auto-encoding for a consistent representation.

**Independent component analysis.** Independent component analysis (ICA) was originally proposed by [BJC85] as a classic model for *learning a good representation*. In fact, it was originally proposed as a simple mathematical model for our memory. The ICA model takes a deceptively similar form to the above PCA model (1.3.7) by assuming that the observed random variable $\boldsymbol{x}$ is a linear

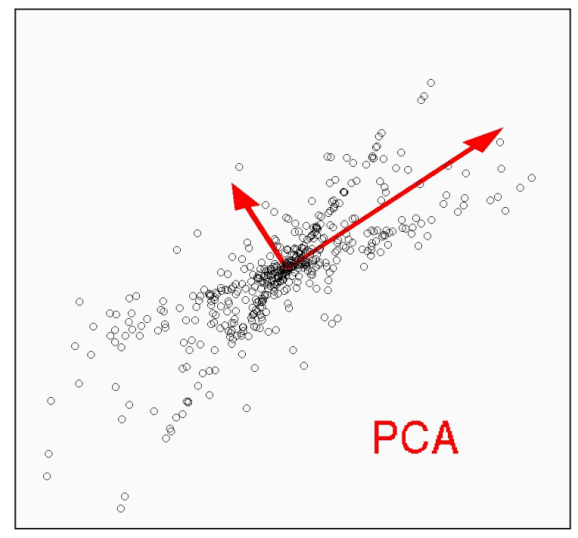

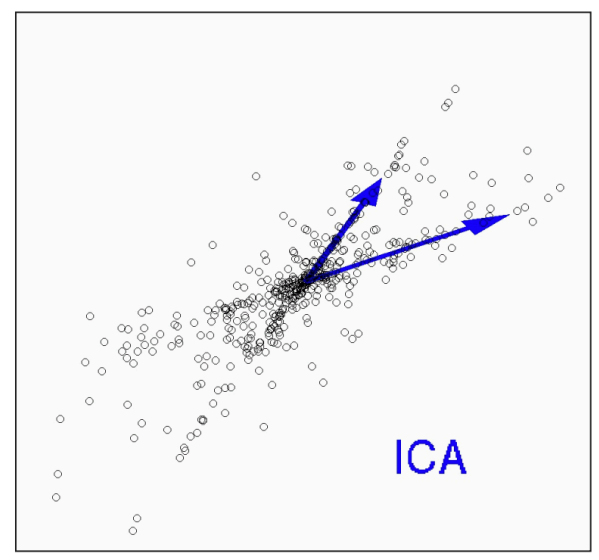

Figure 1.13: PCA (left) versus ICA (right). Note that PCA finds the *orthogonal* vectors that approximately span the data, while ICA finds (not necessarily orthogonal) vectors that *independently* combine to approximately form the data, as well as the coefficients of this combination.

superposition of multiple independent components $z_i$:[15]

$$\boldsymbol{x} = \boldsymbol{u}_1 z_1 + \boldsymbol{u}_2 z_2 + \cdots + \boldsymbol{u}_d z_d + \boldsymbol{\epsilon} = \boldsymbol{U}\boldsymbol{z} + \boldsymbol{\epsilon}. \tag{1.3.12}$$

However, here the components $z_i$ are assumed to be independent *non-Gaussian* variables. For example, a popular choice is

$$z_i = \sigma_i \cdot w_i, \quad \sigma_i \sim \mathrm{Ber}(p), \tag{1.3.13}$$

where $\sigma_i$ is a Bernoulli random variable and $w_i$ could be a constant value or another random variable, say Gaussian.[16] The ICA problem aims to identify both $\boldsymbol{U}$ and $\boldsymbol{z}$ from observed samples of $\boldsymbol{x}$. Figure 1.13 illustrates the difference between ICA and PCA.

Although the (decoding) mapping from $\boldsymbol{z}$ to $\boldsymbol{x}$ seems linear and straightforward once $\boldsymbol{U}$ and $\boldsymbol{z}$ are learned, the (encoding) mapping from $\boldsymbol{x}$ to $\boldsymbol{z}$ can be complicated and may not be represented by a simple linear mapping. Hence, ICA generally gives an auto-encoding of the form:

$$\boldsymbol{x} \xrightarrow{\mathcal{E}} \boldsymbol{z} \xrightarrow{\boldsymbol{U}} \hat{\boldsymbol{x}}. \tag{1.3.14}$$

Thus, unlike PCA, ICA is somewhat more difficult to analyze and solve. In the 1990s, researchers such as Erkki Oja and Aapo Hyvärinen [HO97; HO00b] made significant theoretical and algorithmic contributions to ICA. In Chapter 2, we will study and provide a solution to ICA from which the encoding mapping $\mathcal{E}$ will become clear. In fact, Aapo Hyvärinen proposed the general score matching method in 2005 [Hyv05], in part to learn distributions such as a mixture of Gaussian like in ICA. We will discuss more about this later and again in Chapter 3 in great details.

[15] In PCA the "components" are the learned vectors, whereas in ICA they are scalars. This is just a difference in names and convention.

[16] Even if $w$ is Gaussian, $\sigma w$ is no longer a Gaussian variable!

**Sparse structures and compressive sensing.** If $p$ in (1.3.13) is very small, the probability that any component is non-zero is small. In this case we say $\boldsymbol{x}$ is sparsely generated and concentrates on a union of linear subspaces whose dimension is $k = p \cdot d$. We may therefore extend the ICA model to a more general family of low-dimensional structures known as sparse models.

A $k$-sparse model is the set of all *$k$-sparse vectors*:

$$\mathcal{Z} = \{\boldsymbol{z} \in \mathbb{R}^n \mid \|\boldsymbol{z}\|_0 \leq k\}, \tag{1.3.15}$$

where $\|\cdot\|_0$ counts the number of non-zero entries. Thus $\mathcal{Z}$ is the union of all $k$-dimensional subspaces aligned with the coordinate axes, as illustrated in Figure 1.9 (left). A central problem in signal processing and statistics is to recover a sparse vector $\boldsymbol{z}$ from its linear observations

$$\boldsymbol{x} = \boldsymbol{A}\boldsymbol{z} + \boldsymbol{\epsilon}, \quad \boldsymbol{A} \in \mathbb{R}^{m \times n}, \tag{1.3.16}$$

where $\boldsymbol{A}$ is given, typically $m < n$, and $\boldsymbol{\epsilon}$ is small noise. This seemingly benign problem is NP-hard to solve and even hard to approximate (see [WM22] for details).

Despite a rich history dating back to the eighteenth century [Bos50], no provably efficient algorithm existed for this class of problems, although many heuristic algorithms were proposed between the 1960s and 1990s. Some were effective in practice but lacked rigorous justification. A major breakthrough came in the early 2000s when mathematicians including David Donoho, Emmanuel Candés, and Terence Tao [CT05a; CT05b; Don05] established a rigorous theoretical framework that characterizes precise conditions under which the sparse recovery problem can be solved effectively and efficiently via convex $\ell^1$ minimization:

$$\min \|\boldsymbol{z}\|_1 \quad \text{subject to} \quad \|\boldsymbol{x} - \boldsymbol{A}\boldsymbol{z}\|_2 \leq \epsilon, \tag{1.3.17}$$

where $\|\cdot\|_1$ is the sparsity-promoting $\ell^1$ norm of a vector and $\epsilon$ is a small constant. Any solution yields a sparse (auto) encoding

$$\boldsymbol{x} \xrightarrow{\mathcal{E}} \boldsymbol{z} \xrightarrow{\boldsymbol{A}} \hat{\boldsymbol{x}}. \tag{1.3.18}$$

We will describe such an algorithm (and thus mapping) in Chapter 2, revealing fundamental connections between sparse coding and deep learning.[17]

Conditions for $\ell^1$ minimization to succeed are surprisingly general: the minimum number of measurements $m$ required for a successful recovery is only proportional to the intrinsic dimension $k$. This is the *compressive sensing* phenomenon [Can06]. It extends to broad families of low-dimensional structures, including low-rank matrices. These results fundamentally changed our understanding of recovering low-dimensional structures in high-dimensional spaces. David Donoho, among others, celebrated this reversal as the "blessing of dimensionality" [D D00], in contrast to the usual pessimistic belief in the "curse

[17] Similarities between sparse-coding algorithms and deep networks were noted as early as 2010 [GL10].

of dimensionality.” The complete theory and body of results is presented in [WM22].

The computational significance of this framework cannot be overstated. It transformed problems previously believed intractable into ones that are not only tractable but scalably solvable using extremely efficient algorithms:

$$\text{intractable} \implies \text{tractable} \implies \text{scalable}. \tag{1.3.19}$$

The algorithms come with rigorous theoretical guarantees of correctness given precise requirements in data and computation. The deductive nature of this approach contrasts sharply with the empirical, inductive practice of deep neural networks. Yet we now know that both approaches share a common goal—pursuing low-dimensional structures in high-dimensional spaces.

**Dictionary learning.** Conceptually, an even harder problem than the sparse-coding problem (1.3.16) arises when the observation matrix $\boldsymbol{A}$ is unknown and must itself be learned from a set of (possibly noisy) observations $\boldsymbol{X} = [\boldsymbol{x}_1, \boldsymbol{x}_2, \ldots, \boldsymbol{x}_n]$:

$$\boldsymbol{X} = \boldsymbol{A}\boldsymbol{Z} + \boldsymbol{E}, \quad \boldsymbol{A} \in \mathbb{R}^{m \times n}. \tag{1.3.20}$$

Here we are given only $\boldsymbol{X}$, not the corresponding $\boldsymbol{Z} = [\boldsymbol{z}_1, \boldsymbol{z}_2, \ldots, \boldsymbol{z}_n]$ or the noise term $\boldsymbol{E} = [\boldsymbol{\epsilon}_1, \boldsymbol{\epsilon}_2, \ldots, \boldsymbol{\epsilon}_n]$, except that the $\boldsymbol{z}_i$ are known to be sparse. This is the *dictionary learning* problem, which generalizes the ICA problem (1.3.12) discussed earlier. In other words, given that the distribution of the data $\boldsymbol{X}$ is the image of a standard sparse distribution $\boldsymbol{Z}$ under a linear transform $\boldsymbol{A}$, we wish to learn both $\boldsymbol{A}$ and its “inverse” mapping $\mathcal{E}$ so as to obtain an autoencoder:

$$\boldsymbol{X} \xrightarrow{\ \mathcal{E}\ } \boldsymbol{Z} \xrightarrow{\ \boldsymbol{A}\ } \hat{\boldsymbol{X}}. \tag{1.3.21}$$

PCA, ICA, and dictionary learning all assume that the data distribution is supported on or near low-dimensional linear or piecewise-linear structures. Each method requires learning a (global or local) basis for these structures from noisy samples. In Chapter 2 we study how to identify such low-dimensional structures through these classical models. In particular, we will see that all of these low-dimensional (piecewise) linear models can be learned efficiently by the same family of fast algorithms known as *power iteration* [ZMZ+20]. Although these linear or piecewise-linear models are somewhat idealized for most real-world data, understanding them is an essential first step toward understanding more general low-dimensional distributions.

### General Distributions

The distributions of real-world data such as images, videos, and audio are too complex to be modeled by the above, somewhat idealistic, linear models or Gaussian processes. We normally do not know *a priori* from which family of parametric models they are generated. Historically, many attempts have been made to develop analytical models for these data. In particular, Fields Medalist

David Mumford spent considerable effort in the 1990s trying to understand and model the statistics of natural images [Mum96]. He and his students, including Song-Chun Zhu, drew inspiration and techniques from statistical physics and proposed many statistical and stochastic models for the distribution of natural images, such as random fields or stochastic processes [HM99; LPM03; MG99; ZM97a; ZWM97; ZM97b]. However, these analytical models achieved only limited success in producing samples that closely resemble natural images. Clearly, for real-world data like images, we need to develop more powerful and unifying methods to pursue their more general low-dimensional structures and obtain efficient representations, e.g., for encoding and decoding purposes [ACM12; LMR17].

In practice, for a general distribution of real-world data, we typically only have many samples from the distribution—the so-called *empirical distribution*. In such cases, we normally cannot expect to have a clear analytical form for their low-dimensional structures, nor for the resulting denoising operators.[18] We therefore need to develop a more general solution to these empirical distributions, not necessarily in closed form but at least efficiently computable. If we do this correctly, solutions to the aforementioned linear models should become their special cases.

**Denoising.** In the 1950s, statisticians became interested in the problem of denoising data drawn from an arbitrary distribution. Let $\boldsymbol{x}_o$ be a random variable with probability density function $p_o(\cdot)$. Suppose we observe a noisy version of $\boldsymbol{x}_o$:

$$\boldsymbol{x} = \boldsymbol{x}_o + \sigma \boldsymbol{g}, \tag{1.3.22}$$

where $\boldsymbol{g} \sim \mathcal{N}(\mathbf{0}, \boldsymbol{I})$ is standard Gaussian noise and $\sigma$ is the noise level of the observation. Let $p(\cdot)$ be the probability density function of $\boldsymbol{x}$.[19] Remarkably, the posterior expectation of $\boldsymbol{x}_o$ given $\boldsymbol{x}$ can be calculated by an elegant formula, known as Tweedie's formula [Rob56]:[20]

$$\hat{\boldsymbol{x}}_o = \mathbb{E}[\boldsymbol{x}_o \mid \boldsymbol{x}] = \boldsymbol{x} + \sigma^2 \nabla \log p(\boldsymbol{x}). \tag{1.3.23}$$

As can be seen from the formula, the function $\nabla \log p(\boldsymbol{x})$ plays a very special role in denoising the observation $\boldsymbol{x}$ here. The noise $\boldsymbol{g}$ can be explicitly estimated as

$$\hat{\boldsymbol{g}} = \frac{\boldsymbol{x} - \hat{\boldsymbol{x}}_o}{\sigma} = -\sigma \nabla \log p(\boldsymbol{x}), \tag{1.3.24}$$

for which we only need to know the distribution $p(\cdot)$ of $\boldsymbol{x}$, not the ground truth $p_o(\cdot)$ for $\boldsymbol{x}_o$. An important implication of this result is that if we add Gaussian noise to any distribution, the denoising process can be done easily if we can somehow obtain the function $\nabla \log p(\boldsymbol{x})$.

[18] This is unlike the cases of PCA, Wiener filter, and Kalman filter.

[19] That is, $p(\boldsymbol{x}) = \int_{-\infty}^{\infty} \varphi_\sigma(\boldsymbol{x} - \boldsymbol{z}) p_o(\boldsymbol{z}) \mathrm{d}\boldsymbol{z}$, where $\varphi_\sigma$ is the density function of the Gaussian distribution $\mathcal{N}(\mathbf{0}, \sigma^2 \boldsymbol{I})$.

[20] Herbert Robbins gave the credit for this formula to Maurice Kenneth Tweedie from their personal correspondence.

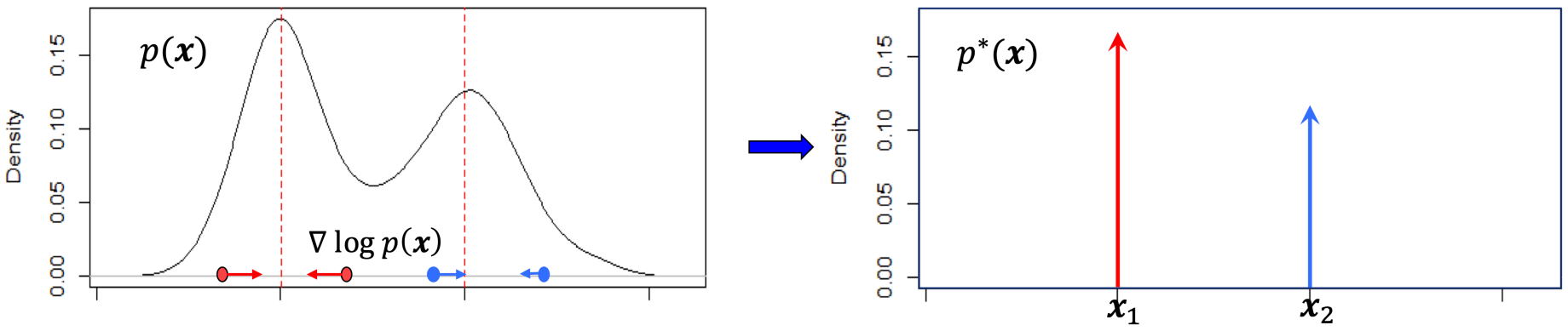

Figure 1.14: Geometric interpretation of a score function $\nabla \log p(\boldsymbol{x})$ for a distribution with density $p(\boldsymbol{x})$ on the left. The operation generated by the score function pushes the distribution towards areas of higher density. The goal is that, by a certain measure of compactness (e.g. entropy or coding length), the resulting distribution is more "compressed". Eventually, the distribution converges to one that has a lower-dimensional support, as $p^*(\boldsymbol{x})$ shown on the right.

Because this is such an important and useful result, it has been rediscovered and used in many different contexts and areas. For example, after Tweedie's formula [Rob56], it was rediscovered a few years later by [Miy61] where it was termed "empirical Bayesian denoising." In the early 2000s, the function $\nabla \log p(\boldsymbol{x})$ was rediscovered again in the context of learning a general distribution and was named the "score function" by Aapo Hyvärinen [Hyv05]. Its connection to (empirical Bayesian) denoising was soon recognized by [Vin11]. Generalizations to other measurement distributions (beyond Gaussian noise) have been made by Eero Simoncelli's group [RS11]. Today, the most direct application of Tweedie's formula and denoising is image generation via iterative denoising [HJA20; KS21].

**Entropy minimization.** The score has an intuitive information-theoretic and geometric interpretation. In information theory, $-\log p(\boldsymbol{x})$ corresponds to the number of bits needed to encode $\boldsymbol{x}$[21]. The gradient $\nabla \log p(\boldsymbol{x})$ points toward higher-probability-density regions, as shown in Figure 1.14 (left). Moving in this direction reduces the number of bits required to encode $\boldsymbol{x}$. Thus, the operator $\nabla \log p(\boldsymbol{x})$ pushes the distribution to "shrink" toward high-density regions. Formally, one can show that the (differential) entropy

$$H(\boldsymbol{x}) = -\int p(\boldsymbol{w}) \log p(\boldsymbol{w}) \mathrm{d}\boldsymbol{w} \tag{1.3.25}$$

decreases under this operation (see Chapter 3 and Chapter B). With an optimal codebook, the resulting distribution achieves a lower coding rate and is thus more compressed. Repeating this denoising process indefinitely produces a distribution whose probability mass is concentrated on a support of lower dimension. For instance, the distribution $p(\boldsymbol{x})$ in Figure 1.14 (left) converges to

[21]at least for discrete variables, as detailed in Chapter 3.

$p^*(\boldsymbol{x})$ (right):[22][23]

$$H(\boldsymbol{x}) = -\int p(\boldsymbol{w}) \log p(\boldsymbol{w}) \mathrm{d}\boldsymbol{w} \quad \xrightarrow{\text{decreasing}} \quad H^*(\boldsymbol{x}) = -\int p^*(\boldsymbol{w}) \log p^*(\boldsymbol{w}) \mathrm{d}\boldsymbol{w}. \tag{1.3.26}$$

As the distribution converges to $p^*(\boldsymbol{x})$, its differential entropy approaches negative infinity due to a technical difference between continuous and discrete entropy definitions. Chapter 3 resolves this using a unified *rate-distortion* measure.

Later in this chapter and in Chapters 3 and 4, we explore how this simple denoising-compression framework unifies powerful methods for learning low-dimensional distributions in high-dimensional spaces, including natural image distributions.

## 1.3.2 Empirical Data-Driven Approaches

In practice, it is difficult to model important real-world data—such as images, sounds, and text—with the idealized linear, piecewise-linear, or other analytical models discussed in the previous section. Historically, many non-parametric models and methods have therefore been proposed to model or capture the empirical distribution of the data and then use such "models" for inference. These models, especially deep neural networks, often drew inspiration from the biological nervous system, because the brain of an animal or human processes such data with remarkable efficiency and effectiveness. Compared to the analytical models introduced in the previous section, these data-driven models are the least understood despite the fact that they have very much dominated the practice of AI technology in the past decade or so. Hence, one of the main goals of this book is to provide a principled and unified framework (very much through Chapters 5 to 8) to understand both these "modern" empirical models and the classic analytical models and reveal fundamental connections between them.

### Classic Artificial Neural Networks

**Artificial neuron.** Inspired by the nervous system in the brain, the first mathematical model of an artificial neuron[24] was proposed by Warren McCulloch[25] and Walter Pitts in 1943 [MP43]. It describes the relationship between the input $x_i$ and output $o_j$ as:

$$o_j = \varphi\Big(\sum_i w_{ji} x_i\Big), \tag{1.3.27}$$

[22] Strictly speaking, $p^*(\boldsymbol{x})$ is a generalized function: $p^*(\boldsymbol{x}) = p^*(\boldsymbol{x}_1)\delta(\boldsymbol{x} - \boldsymbol{x}_1) + p^*(\boldsymbol{x}_2)\delta(\boldsymbol{x} - \boldsymbol{x}_2)$ with $p^*(\boldsymbol{x}_1) + p^*(\boldsymbol{x}_2) = 1$.

[23] Notice that in this section we discuss the process of iteratively denoising and compressing a high-entropy noise distribution until it converges to the low-entropy data distribution. As we will see in Chapter 3, this problem is dual to the problem of learning an optimal encoding for the data distribution, which is also a useful application [RAG+24].

[24] known as the Linear Threshold Unit, or perceptron.

[25] a professor of psychiatry at the University of Chicago at the time

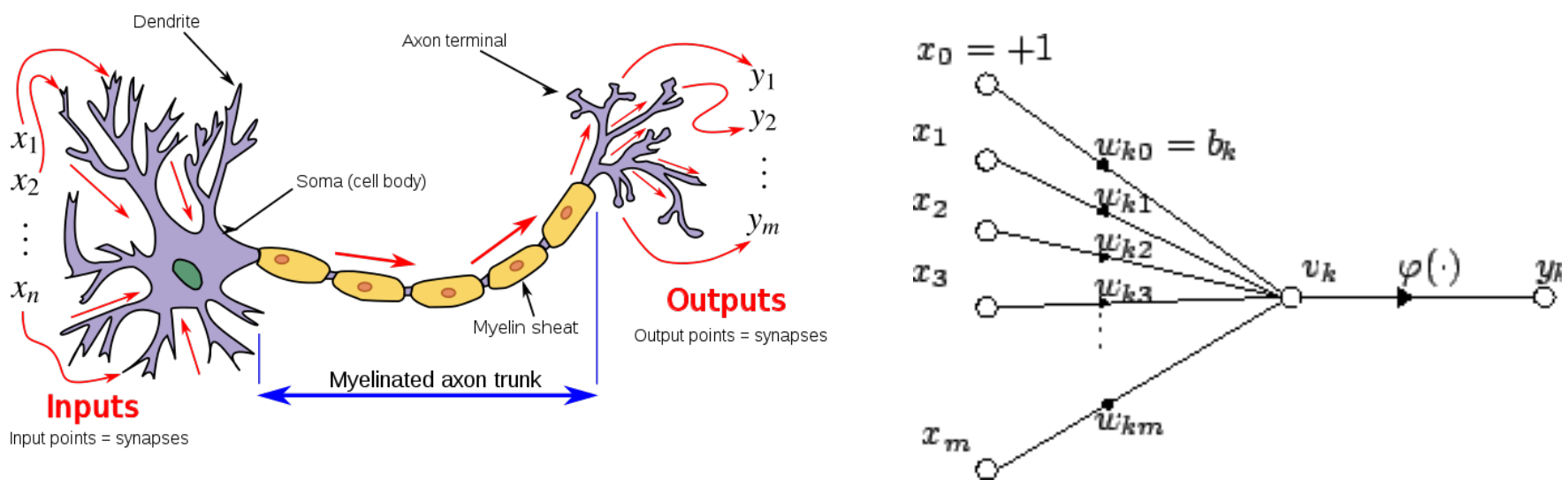

Figure 1.15: The first mathematical model of an artificial neuron (right) that emulates how a neuron (left) processes signals.

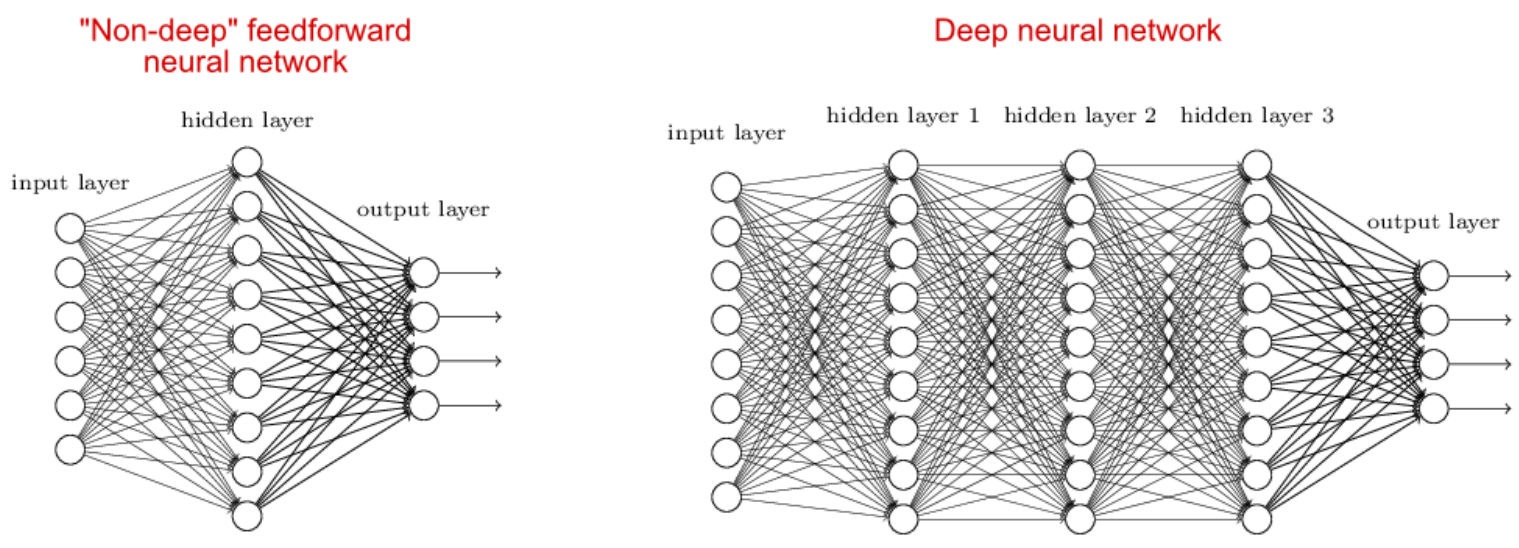

Figure 1.16: A network with one hidden layer (left) versus a deep network (right).

where $\varphi(\cdot)$ is some nonlinear activation, typically modeled by a threshold function. This model is illustrated in Figure 1.15. As we can see, this form already shares the main characteristics of a basic unit in modern deep neural networks. The model is derived from observations of how a single neuron works in our nervous system. However, researchers did not know exactly what functions a collection of such neurons could realize and perform. On a more technical level, they were also unsure which nonlinear activation function $\varphi(\cdot)$ should be used. Hence, historically many variants have been proposed.[26]

**Artificial neural network.** In the 1950s, Frank Rosenblatt was the first to build a machine with a *network* of such artificial neurons, shown in Figure 1.17. The machine, called Mark I Perceptron, consists of an input layer, an output layer, and a single hidden layer of 512 artificial neurons, as shown in Figure 1.17 left, which is similar to what is illustrated in Figure 1.16 (left). It was designed to classify optical images of letters. However, the capacity of a single-layer network is limited and can only learn linearly separable patterns. In the 1969 book *Perceptrons: An Introduction to Computational Geometry* by Marvin Minsky and Seymour Papert [MP69], it was shown that the single-layer architecture of Mark I Perceptron cannot learn an XOR function. This result significantly dampened interest in artificial neural networks, even though it was later proven that a multi-layer network can learn an XOR function [RHW86a]. In fact, a sufficiently large multi-layer network, as shown in Figure 1.16 (right), consisting of such simple neurons can simulate any finite-state machine, even the universal Turing machine.[27] Nevertheless, the study of artificial neural networks subsequently entered its first winter in the 1970s.

[26] Step function, hard or soft thresholding, rectified linear unit (ReLU), sigmoid, etc. [DSC22].

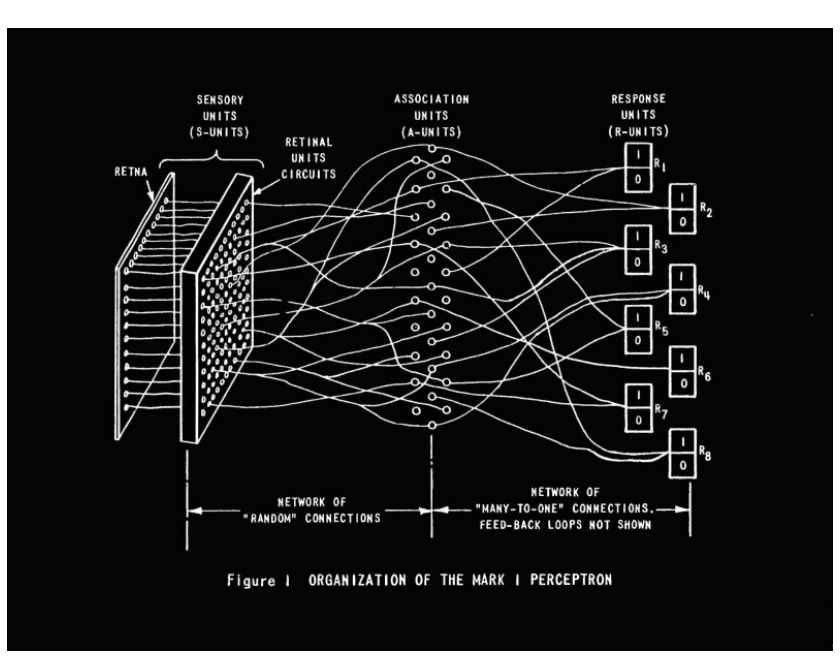

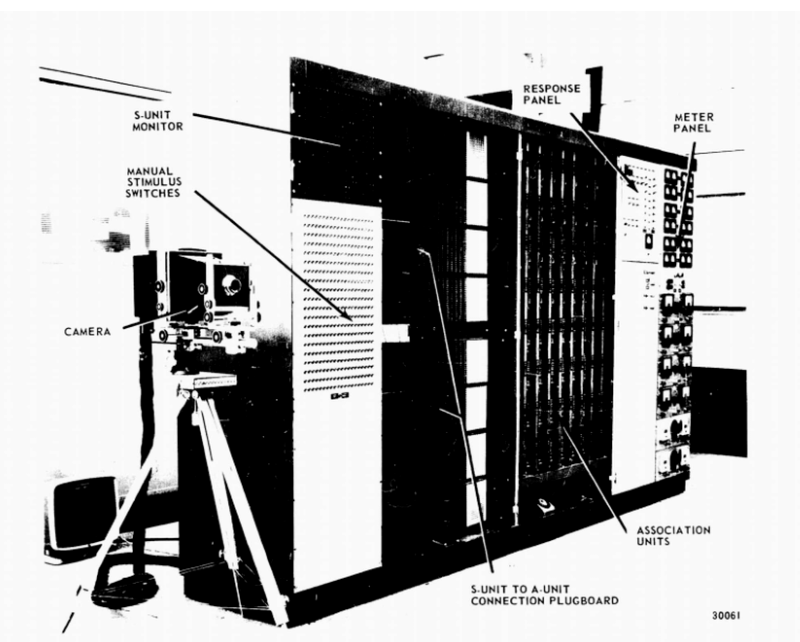

Figure 1.17: The Mark I Perceptron machine developed by Frank Rosenblatt in the late 1950s.

**Convolutional neural networks.** Early experiments with artificial neural networks such as the Mark I Perceptron in the 1950s and 1960s were somewhat disappointing. They suggested that simply connecting neurons in a general fashion, as in multi-layer perceptrons (MLPs), might not suffice. To build effective and efficient networks, it is extremely helpful to understand the collective purpose or function neurons in the network must achieve so that they can be organized and learned in a specialized way. Thus, at this juncture, once again the study of machine intelligence turned to the animal nervous system for inspiration.

It is known that most of our brain is dedicated to processing visual information [Pla99]. In the 1950s and 1960s, David Hubel and Torsten Wiesel systematically studied the visual cortices of cats. They discovered that the visual cortex contains different types of cells—simple cells and complex cells—which are sensitive to visual stimuli of different orientations and locations [HW59]. Hubel and Wiesel won the 1981 Nobel Prize in Physiology or Medicine for this groundbreaking discovery.

On the artificial neural network side, Hubel and Wiesel's work inspired Kunihiko Fukushima to design the "neocognitron" network in 1980, which consists of artificial neurons that emulate biological neurons in the visual cortices [Fuk80]. This is known as the first *convolutional neural network* (CNN), and its architec-ture is illustrated in Figure 1.18. Unlike the perceptron, the neocognitron had more than one hidden layer and could be viewed as a deep network, as shown in Figure 1.16 (right).

[27] Do not confuse what neural networks are capable of doing in principle with whether it is tractable or easy to learn a neural network that realizes certain desired functions.

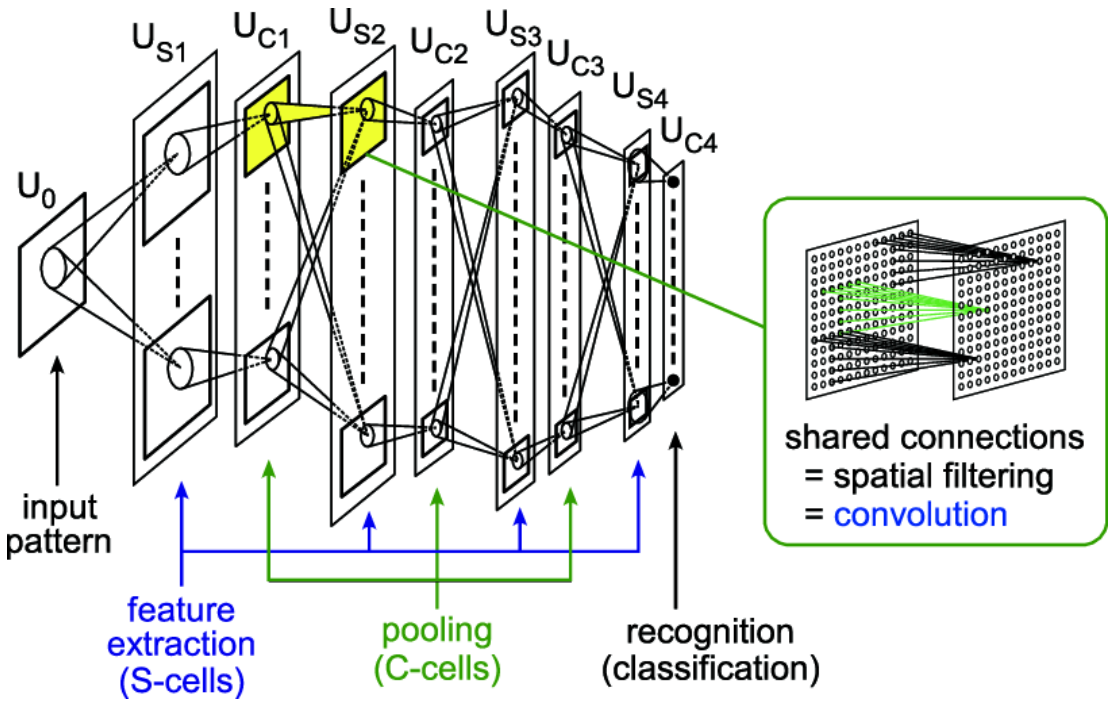

Figure 1.18: Origin of convolutional neural networks: the Neocognitron by Kunihiko Fukushima in 1980. Notice that the interleaving layers of convolutions and pooling emulate the functions of simple cells and complex cells discovered in the visual cortices of cats.

Also inspired by how neurons work in the cat's visual cortex, Fukushima was the first to introduce the *rectified linear unit* (ReLU):

$$\varphi(x) = \max\{0, x\} = \begin{cases} x, & \text{if } x > 0, \\ 0, & \text{if } x \leq 0, \end{cases} \tag{1.3.28}$$

as the activation function $\varphi(\cdot)$ in 1969 [Fuk69]. However, it was not until recent years that ReLU became widely used in modern deep (convolutional) neural networks. This book will explain why ReLU is a good choice once we discuss the main operations deep networks implement: compression.

CNN-type networks continued to evolve in the 1980s, and many different variants were introduced and studied. However, despite the remarkable capacities of deep networks and the improved architectures inspired by neuroscience, it remained extremely difficult to train such deep networks for real tasks such as image classification. Getting a network to work depended on many unexplainable heuristics and tricks, which limited the appeal and applicability of neural networks. A major breakthrough came around 1989 when Yann LeCun successfully used *back propagation* (BP) to learn a deep convolutional neural network for recognizing handwritten digits [LBD+89], later known as LeNet (see Figure 1.19). After several years of persistent development, his perseverance paid off: LeNet's performance eventually became good enough for practical use in the late 1990s [LBB+98a]. It was used by the US Post Office for recognizing handwritten digits (for zip codes). LeNet was considered the "prototype" network for all modern deep neural networks, such as AlexNet [KSH12] and ResNet [HZR+16b], which we will discuss later. For this work, Yann LeCun

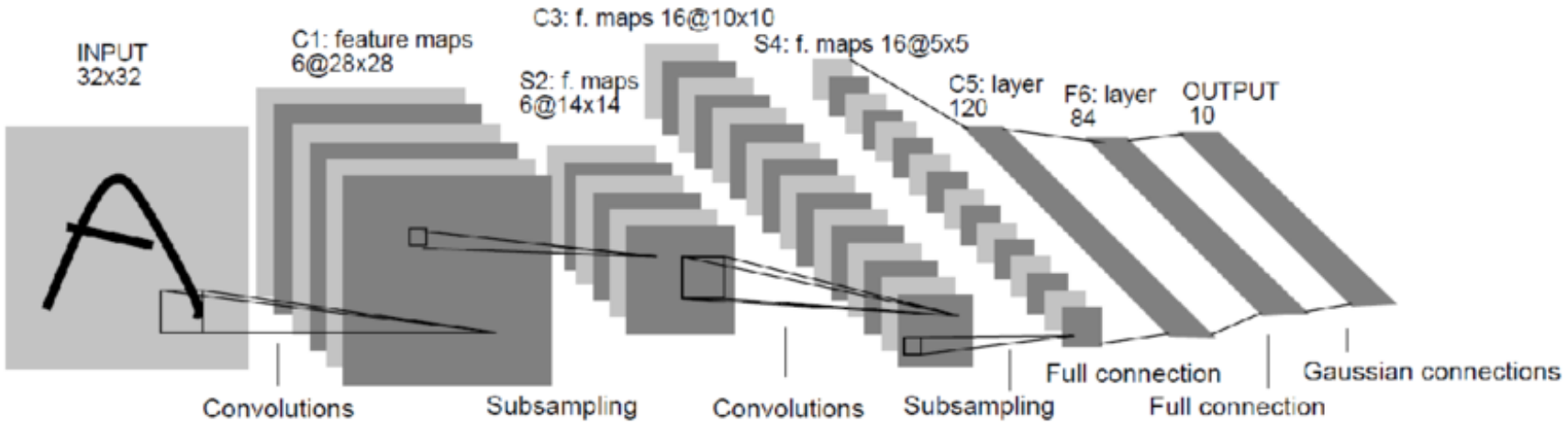

Figure 1.19: The LeNet-5 convolutional neural network designed by Yann LeCun in 1989.

was awarded the 2018 Turing Award.[28]

**Backpropagation.** Throughout history, the fate of deep neural networks has been tied to how easily and efficiently they can be trained. Backpropagation (BP) was introduced for this purpose. A multilayer perceptron can be expressed as a composition of linear mappings and nonlinear activations:

$$h(\boldsymbol{W}_1, \dots, \boldsymbol{W}_L) = f^L(\boldsymbol{W}_L f^{L-1}(\boldsymbol{W}_{L-1} \cdots f^2(\boldsymbol{W}_2 f^1(\boldsymbol{W}_1 \boldsymbol{x})))). \tag{1.3.29}$$

To train the network weights $\{\boldsymbol{W}_l\}_{l=1}^L$ via gradient descent, we must evaluate the gradient $\partial h / \partial \boldsymbol{W}_l$. The chain rule in calculus shows that gradients can be computed efficiently for such functions—a technique later termed backpropagation or BP; see Chapter A for details. BP was already known in optimal control and dynamic programming during the 1960s and 1970s, appearing in Paul Werbos's 1974 PhD thesis [Wer74; Wer94]. In 1986, David Rumelhart et al. were the first to apply it to train a multilayer perceptron (MLP) [RHW86a]. Since then, BP has become the dominant technique for training deep networks, as it is scalable and can be efficiently implemented on parallel and distributed computing platforms. However, nature likely does not use BP,[29] as the mechanism is too expensive for physical implementation.[30] This leaves ample room for future improvement.

Despite the aforementioned algorithmic advances, training deep networks remained finicky and computationally expensive in the 1980s and 1990s. By the late 1990s, support vector machines (SVMs) [CV95] had gained popularity as a superior alternative for classification tasks.[31] SVMs offered two advantages:

[28] Together with two other pioneers of deep networks, Yoshua Bengio and Geoffrey Hinton.

[29] As we have discussed earlier, nature almost ubiquitously learns to correct errors via closed-loop feedback.

[30] End-to-end BP is computationally intractable for neural structures as complex as the brain, especially if the updates need to happen in real time. Instead, localized updates to small sections of the neural circuitry are much more biologically feasible. There is a relatively small amount of work on transplanting such "local learning" rules to training deep networks, circumventing BP [BS16; LTP25; MSS+22]. We believe this is an exciting opportunity to improve the scalability of training deep networks.

[31] Similar ideas for classification problems trace back to Thomas Cover's 1964 PhD dissertation, which was condensed and published in a paper in 1964 [Cov64].

a rigorous statistical learning framework known as the Vapnik–Chervonenkis (VC) theory and efficient convex optimization algorithms [BV04]. The rise of SVMs ushered in a second winter for neural networks in the early 2000s.

**Compressive auto-encoding.** In the late 1980s and 1990s, artificial neural networks were already being used to learn low-dimensional representations of high-dimensional data such as images. It had been shown that neural networks could learn PCA directly from data [BH89; Oja82], rather than using the classic methods discussed in Section 1.3.1. It was also argued during this period that, due to their ability to model nonlinear transformations, neural networks could learn low-dimensional representations for data with nonlinear distributions. Similar to the linear PCA case, one can simultaneously learn a nonlinear dimension-reduction encoder $f$ and a decoder $g$, each modeled by a deep neural network [Kra91; RHW86a]:

$$\boldsymbol{X} \xrightarrow{\ f\ } \boldsymbol{Z} \xrightarrow{\ g\ } \hat{\boldsymbol{X}}. \tag{1.3.30}$$

By enforcing consistency between the decoded data $\hat{\boldsymbol{X}}$ and the original $\boldsymbol{X}$—for example, by minimizing a MSE-type reconstruction error[32]:

$$\min_{f,g} \|\boldsymbol{X} - \hat{\boldsymbol{X}}\|_2^2, \qquad \text{where} \quad \hat{\boldsymbol{X}} = g(f(\boldsymbol{X})), \tag{1.3.31}$$

an autoencoder can be learned directly from the data $\boldsymbol{X}$.

But how can we guarantee that such an auto-encoding indeed captures the true low-dimensional structures in $\boldsymbol{X}$ rather than yielding a trivial redundant representation? For instance, one could simply choose $f$ and $g$ to be identity maps and set $\boldsymbol{Z} = \boldsymbol{X}$. To ensure the auto-encoding is worthwhile, the resulting representation should be compressive according to some computable measure of complexity. In 1993, Geoffrey Hinton and colleagues proposed using coding length as such a measure, transforming the auto-encoding objective into finding a representation that minimizes coding length [HZ93]. This work established a fundamental connection between the principle of minimum description length [Ris78] and free (Helmholtz) energy minimization. Later work from Hinton's group [HS06] empirically demonstrated that such auto-encoding can learn meaningful low-dimensional representations for real-world images. Pierre Baldi provided a comprehensive survey of autoencoders in 2011 [Bal11], just before deep networks gained widespread popularity. We will discuss measures of complexity and auto-encoding further in Section 1.4, and present a systematic study of compressive auto-encoding in Chapter 6 from a more unified perspective.

### Modern Deep Neural Networks

For nearly 30 years—from the 1980s to the 2010s—neural networks were not taken seriously by the mainstream machine learning community. Early deep networks such as LeNet showed promising performance on small-scale classification problems like digit recognition, yet their design was largely empirical, the available datasets were tiny, and back-propagation was computationally prohibitive for the hardware of the era. These factors led to waning interest and stagnant progress, with only a handful of researchers persisting.

[32]MSE-type errors are known to be problematic for imagery data with complex nonlinear structures. As we will soon discuss, much recent work in generative methods, including GANs, has focused on finding better surrogate distance functions between the original data $\boldsymbol{X}$ and the regenerated $\hat{\boldsymbol{X}}$.

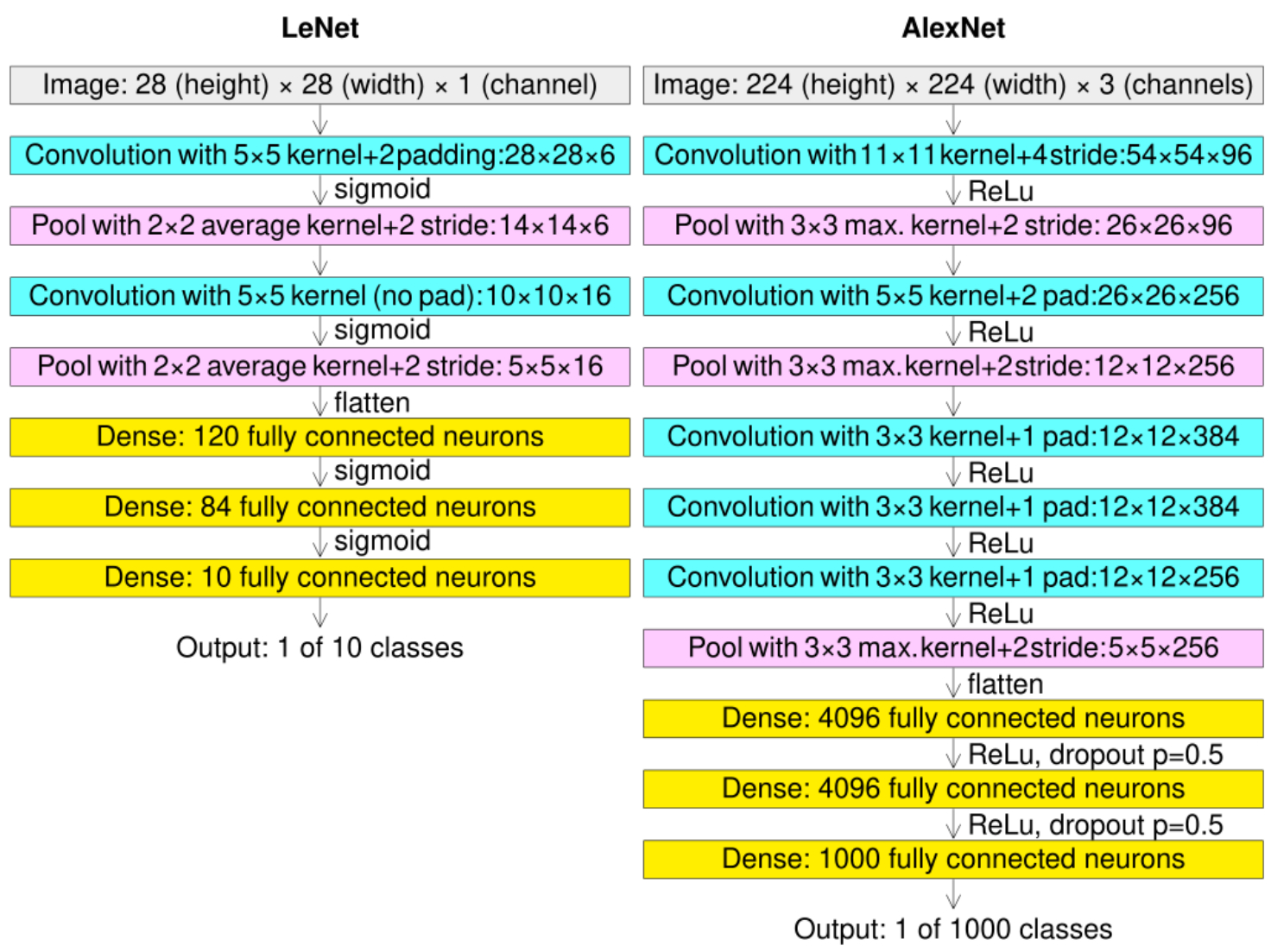

Figure 1.20: Architecture of LeNet [LBD+89] versus AlexNet [KSH12].

**Classification and recognition.** The tremendous potential of deep networks could be unleashed only once sufficient data and computational power became available. By the 2010s, large datasets such as ImageNet had emerged, and GPUs had become powerful enough to make back-propagation affordable even for networks far larger than LeNet. In 2012, the deep convolutional network AlexNet — named for Alex Krizhevsky, one of the authors [KSH12] — attracted widespread attention by surpassing existing classification methods on ImageNet by a significant margin.[33] Figure 1.20 compares AlexNet with LeNet. AlexNet retains many characteristics of LeNet—it is simply larger and replaces LeNet's sigmoid activations with ReLUs. This work contributed to Geoffrey Hinton's 2018 Turing Award.

This early success inspired the machine intelligence community to explore new neural network architectures, variations, and improvements. Empirically, it was discovered that larger and deeper networks yield better performance on tasks such as image classification. Notable architectures that emerged include VGG [SZ15], GoogLeNet [SLJ+14], ResNet [HZR+16a], and, more recently,

[33] Deep networks had already achieved state-of-the-art results on speech-recognition tasks, but these successes received far less attention than the breakthrough on image classification.

Transformers [VSP+17b]. Despite rapid empirically-driven performance improvements, theoretical explanations for these architectures—and the relationships among them, if any —remained scarce. One goal of this book is to uncover the common objective these networks optimize and to explain their shared characteristics, such as multiple layers of linear operators interleaved with nonlinear activations (see Chapter 5).

**Reinforcement learning.** Early deep network successes were mainly in supervised classification tasks such as speech and image recognition. Later, deep networks were adopted to learn decision-making or control policies for game playing. In this context, deep networks model the optimal decision/control policy (i.e., a probability distribution over actions to take in order to maximize the expected reward) or the associated optimal value function (an estimate of the expected reward from the given state), as shown in Figure 1.21. Network parameters are incrementally optimized[34] based on rewards returned from the success or failure of playing the game with the current policy. This learning paradigm is called *reinforcement learning* [SB18]; it originated in control-systems practice of the late 1960s [MM70; WF65] and traces back through a long and rich history to Richard Bellman's *dynamic programming* [Bel57] and Marvin Minsky's *trial-and-error learning* [Min54] in the 1950s.

From an implementation standpoint, the marriage of deep networks and reinforcement learning proved powerful: deep networks can approximate control policies and value functions for real-world environments that are difficult to model analytically. This culminated in DeepMind's AlphaGo system, which stunned the world in 2016 by defeating top Go player Lee Sedol and, in 2017, world champion Jie Ke.[35]

AlphaGo's success surprised the computing community, which had long regarded the game's state space as too vast for any efficient solution in terms of computation and sample size. The only plausible explanation is that the optimal value and policy function of Go possess significant favorable structure: qualitatively speaking, their intrinsic dimensions are low enough so that they can be well approximated by neural networks learnable from a manageable number of samples.

**Generation and prediction.** From a certain perspective, the early practice of deep networks in the 2010s was focused on extracting relevant information from the data $\boldsymbol{X}$ and encoding it into a task-specific representation $\boldsymbol{Z}$ (say $\boldsymbol{Z}$

[34] Typically via back-propagation (BP).

[35] In 1996, IBM's Deep Blue made history by defeating Russian grandmaster Garry Kasparov in chess using traditional machine learning techniques such as tree search and pruning, methods that have proven less scalable and unsuccessful for more complex games like Go.

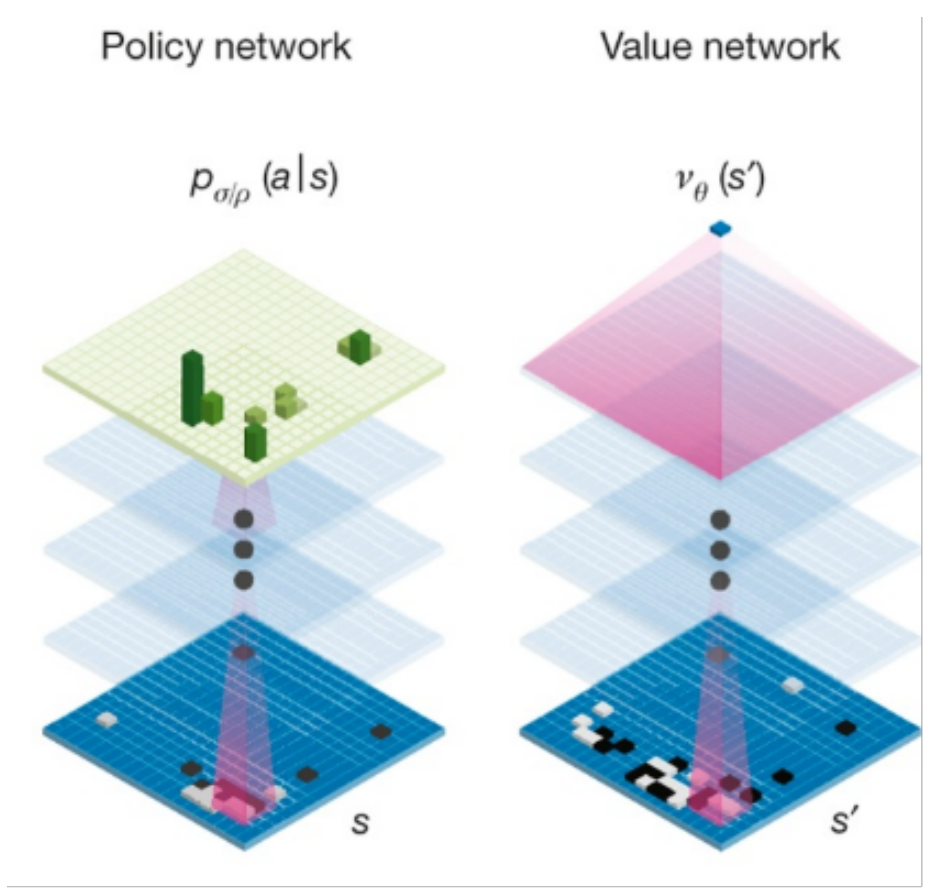

Figure 1.21: AlphaGo: using deep neural networks to model the optimal policy or value function for the game Go.

denotes class labels in classification tasks[36]):

$$\boldsymbol{X} \xrightarrow{\;f\;} \boldsymbol{Z}. \tag{1.3.32}$$

Here the learned mapping $f$ needs not preserve most distributional information about $\boldsymbol{X}$; it suffices to retain only the sufficient statistics for the task. For example, a sample $\boldsymbol{x} \in \boldsymbol{X}$ might be an image of an apple, mapped by $f$ to the class label $\boldsymbol{z} =$ "apple".

In many modern settings—such as the so-called large general-purpose ("foundation") models—we may need to also decode $\boldsymbol{Z}$ to recover the corresponding $\boldsymbol{X}$ to a prescribed precision:

$$\boldsymbol{Z} \xrightarrow{\;g\;} \hat{\boldsymbol{X}}. \tag{1.3.33}$$

Because $\boldsymbol{X}$ typically represents data observed from the external world, a good decoder allows us to simulate or predict what happens in the world. In a "text-to-image" or "text-to-video" task, for instance, $\boldsymbol{z}$ is the text describing the desired image $\boldsymbol{x}$, and the decoder should generate an $\hat{\boldsymbol{x}}$ whose content matches $\boldsymbol{x}$. Given an object class $\boldsymbol{z} =$ "apple", the decoder $g$ should produce an image $\hat{\boldsymbol{x}}$ that looks like an apple, though not necessarily identical to the original $\boldsymbol{x}$.

**Generation via discriminative approaches.** For the generated images $\hat{\boldsymbol{X}}$ to resemble true natural images $\boldsymbol{X}$, we must be able to evaluate and minimize some distance:

$$\min \operatorname{dist}(\boldsymbol{X}, \hat{\boldsymbol{X}}). \tag{1.3.34}$$

[36] This is a highly atypical notation for labels — the usual notation is $\boldsymbol{y}$ — but it is useful for our purposes to consider labels as highly compressed and sparse representations of the data. See Example 1.2 for more details.

As it turns out, most theoretically motivated distances are extremely difficult—if not impossible—to compute and optimize for distributions in high-dimensional space with low intrinsic dimension.[37]

In 2007, Zhuowen Tu, disappointed by early analytical attempts to model and generate natural images, decided to try a drastically different approach. In a paper published at CVPR 2007 [Tu07], he first suggested learning a generative model for images via a discriminative approach. The idea is simple: if evaluating the distance $\mathrm{dist}(\boldsymbol{X}, \hat{\boldsymbol{X}})$ proves difficult, one can instead learn a discriminator $d$ to separate $\hat{\boldsymbol{X}}$ from $\boldsymbol{X}$:

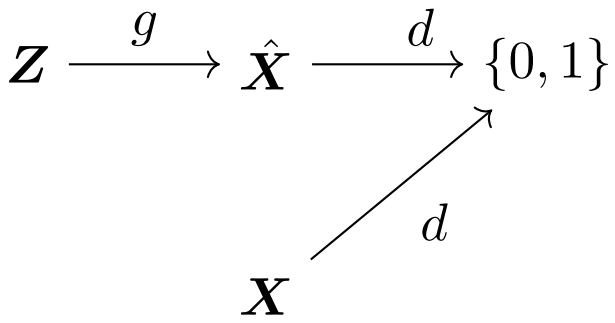

where labels 0 and 1 indicate whether an image is generated or real. Intuitively, the harder it becomes to separate $\hat{\boldsymbol{X}}$ and $\boldsymbol{X}$, the closer they likely are.

Tu's work [Tu07] first demonstrated the feasibility of learning a generative model from a discriminative approach. However, the work employed traditional methods for image generation and distribution classification (such as boosting), which proved slow and difficult to implement. After 2012, deep neural networks gained popularity for image classification. In 2014, Ian Goodfellow and colleagues again proposed generating natural images with a discriminative approach [GPM+14a]. They suggested modeling both the generator $g$ and discriminator $d$ with deep neural networks. Moreover, they proposed learning $g$ and $d$ via a minimax game:

$$\min_{g} \max_{d} \ell(\boldsymbol{X}, \hat{\boldsymbol{X}}), \tag{1.3.35}$$

where $\ell(\cdot)$ is some natural loss function associated with the classification task. In this formulation, the discriminator $d$ maximizes its success in separating $\boldsymbol{X}$ and $\hat{\boldsymbol{X}}$, while the generator $g$ does the opposite. This approach is named *generative adversarial networks* (GANs). It was shown that once trained on a large dataset, GANs can indeed generate photo-realistic images. Partially due to this work's influence, Yoshua Bengio received the 2018 Turing Award.

The discriminative approach appears to cleverly bypass a fundamental difficulty in distribution learning. However, rigorously speaking, this approach does not fully resolve the fundamental difficulty. It is shown in [GPM+14a] that with a properly chosen loss, the minimax formulation becomes mathematically equivalent to minimizing the Jensen-Shannon distance (see [CT91]) between $\boldsymbol{X}$ and $\hat{\boldsymbol{X}}$. This remains a hard problem for two low-dimensional distributions in high-dimensional space. Consequently, GANs typically rely on many heuristics and engineering tricks and often suffer from instability issues such as mode collapsing.[38]

[37] This remains true even when a parametric family of distributions for $\boldsymbol{X}$ is specified. The distance often becomes ill-conditioned or ill-defined for distributions with low-dimensional supports. Worse still, the chosen family might poorly approximate the true distribution of interest.

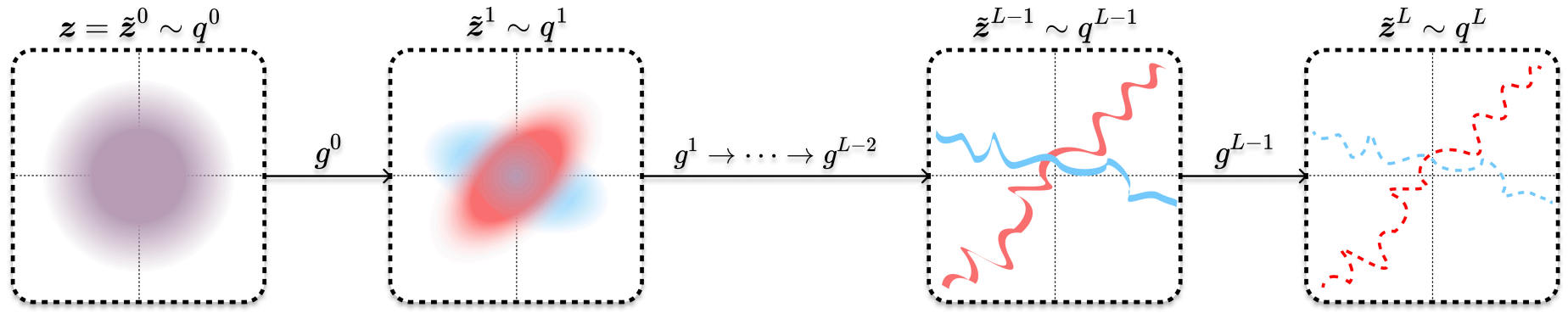

Figure 1.22: Illustration of an iterative denoising and compressing process that, starting from a Gaussian distribution $q^0 = \mathcal{N}(\mathbf{0}, \boldsymbol{I})$, converges to an arbitrary low-dimensional data distribution $q^L = p$.

**Generation via denoising and diffusion.** In 2015, shortly after GANs were introduced and gained popularity, Surya Ganguli and his students proposed that an iterative denoising process modeled by a deep network could be used to learn a general distribution, such as that of natural images [SWM+15]. Their method was inspired by properties of special Gaussian and binomial processes studied by William Feller in 1949 [Fel49].[39]

Soon, denoising operators based on the score function [Hyv05], briefly introduced in Section 1.3.1, were shown to be more general and unified the denoising and diffusion processes and algorithms [HJA20; SE19; SSK+21]. Figure 1.22 illustrates the process that transforms a generic Gaussian distribution $q^0 = \mathcal{N}(\mathbf{0}, \boldsymbol{I})$ to an (arbitrary) empirical distribution $p(\boldsymbol{x})$ by performing a sequence of iterative denoising (or compressing) operations:

$$\boldsymbol{z}^0 \sim \mathcal{N}(\mathbf{0}, \boldsymbol{I}) \xrightarrow{\ g^0\ } \boldsymbol{z}^1 \xrightarrow{\ g^1\ } \cdots \xrightarrow{\ g^{L-1}\ } \boldsymbol{z}^L \sim p(\boldsymbol{x}). \tag{1.3.36}$$

By now, denoising (and diffusion) has replaced GANs and become the mainstream method for learning distributions of images and videos, leading to popular commercial image generation engines such as Midjourney and Stability.ai. In Chapter 3 we will systematically introduce and study the denoising and diffusion method for learning a general low-dimensional distribution.

## 1.4 A Unifying Approach

So far, we have given a brief account of the main objective and history of machine intelligence, along with many important ideas and approaches associated with it. In recent years, following the empirical success of deep neural networks, tremendous efforts have been made to develop theoretical frameworks that help us understand empirically designed deep networks—whether specific seemingly necessary components (e.g., dropout [SHK+14], normalization [BKH16; IS15], attention [VSP+17b]) or their overall behaviors (e.g., double descent [BHM+19], neural collapse [PHD20]).

[38] Nevertheless, such a minimax formulation provides a practical approximation of the distance. It simplifies implementation and avoids certain caveats in directly computing the distance.

[39] Again, in the magical era of the 1940s!

Motivated in part by this trend, this book pursues several important and challenging goals:

- Develop a theoretical framework that yields rigorous mathematical interpretations of deep neural networks.
- Ensure correctness of the learned data distribution and consistency with the learned representation.
- Demonstrate that the framework leads to performant architectures and guides further practical improvements.

In recent years, mounting evidence suggests these goals can indeed be achieved by leveraging the theory and solutions of the classical analytical low-dimensional models discussed earlier (treated more thoroughly in Chapter 2) and by integrating fundamental ideas from related fields—namely information/coding theory, control/game theory, and optimization. This book aims to provide a systematic introduction to this new approach.

### 1.4.1 Learning Parsimonious Representations

One necessary condition for any learning task to be possible is that the sequences of interest must be *computable*, at least in the sense of Alan Turing [Tur36]. That is, a sequence can be computed via a program on a typical computer.[40] In addition to being computable, we require computation to be *tractable*.[41] That is, the computational cost (space and time) for learning and computing the sequence should not grow exponentially in length. Furthermore, as we see in nature (and in the modern practice of machine intelligence), for most practical tasks an intelligent system needs to learn what is predictable from massive data in a very high-dimensional space, such as from vision, sound, and touch. Hence, for intelligence we do not need to consider all computable and tractable sequences or structures; we should focus only on predictable sequences and structures that admit *scalable* realizations of their learning and computing algorithms:

$$\textbf{computable} \implies \textbf{tractable} \implies \textbf{scalable}. \tag{1.4.1}$$

This is because whatever algorithms intelligent beings use to learn useful information must be *scalable*. More specifically, the computational complexity of the algorithms should scale gracefully—typically linear or even sublinear—in the size and dimension of the data. On the technical level, this requires that the operations the algorithms rely on to learn can only utilize oracle information that can be efficiently computed from the data. More specifically, when the dimension is high and the scale is large, the only oracle one can afford to compute is either the first-order geometric information about the data[42] or the second-order statistical information[43]. The main goal of this book is to develop a theoretical and computational framework within which we can systematically develop efficient and effective solutions or algorithms with such scalable oracles and operations to *learn* low-dimensional structures from the sampled data and subsequently the predictive function.

[40] There are indeed well-defined sequences that are not computable.

[41] We do not need to consider predicting things whose computational complexity is intractable—say, grows exponentially in the length or dimension of the sequence.

**Pursuing low-dimensionality via compression.** From the examples of sequences we gave in Section 1.2.1, it is clear that some sequences are easy to model and compute, while others are more difficult. The computational cost of a sequence depends on the complexity of the predicting function $f$. The higher the degree of regression $d$, the more costly it is to compute. The function $f$ could be a simple linear function, or it could be a nonlinear function that is arbitrarily difficult to specify and compute.

It is reasonable to believe that if a sequence is harder—by whatever measure we choose—to specify and compute, then it will also be more difficult to learn from its sampled segments. Nevertheless, for any given predictable sequence, there are in fact many, often infinitely many, ways to specify it. For example, for the simple sequence $x_{n+1} = ax_n$, we could also define the same sequence with $x_{n+1} = ax_n + bx_{n-1} - bx_{n-1}$. Hence it would be very useful to develop an objective and rigorous notion of "complexity" for any given computable sequence.

Andrey Kolmogorov, a Russian mathematician, was one of the first to define complexity for any computable sequence.[44] He proposed that among all programs computing the same sequence, the length of the shortest program measures its complexity. This aligns with Occam's Razor—*choose the simplest theory explaining the same observation.* Formally, let $p$ be a program generating sequence $\boldsymbol{x}$ on universal computer $\mathcal{U}$. The Kolmogorov complexity of $\boldsymbol{x}$ is:

$$K(\boldsymbol{x}) = \min_{p\,:\,\mathcal{U}(p)=\boldsymbol{x}} L(p). \tag{1.4.2}$$

Thus, complexity measures how parsimoniously we can specify or compute the sequence. This definition is conceptually important and historically inspired profound studies in computational complexity and theoretical computer science.

The shortest program length represents the ultimate compression of the sequence, quantifying our gain from learning its generative mechanism. However,

[42] such as approximating a nonlinear structure locally with linear subspaces and computing the gradient of an objective function.

[43] such as covariance or correlation of the data or their features. As we will see, popular deep architectures such as Transformers rely on this type of information.

[44] Many contributed to this notion of sequence complexity, most notably Ray Solomonoff and Greg Chaitin. All three developed algorithmic information theory independently—Solomonoff in 1960, Kolmogorov in 1965 [Kol98], and Chaitin around 1966 [Cha66].

Kolmogorov complexity is generally *uncomputable* [CT91] and intractable to approximate. Consequently, it has little practical use—it cannot predict learning difficulty or assess how well we have learned.

**Computable measure of parsimony.** For practical purposes, we need an efficiently computable measure of complexity for sequences generated by the same predicting function.[45] Note that part of the reason Kolmogorov complexity is not computable is that its definition is non-constructive.

To introduce a computable measure of complexity, we may take a more constructive approach, as advocated by Claude Shannon through the framework of information theory [CT91; Sha48].[46] By assuming the sequence $\boldsymbol{x}$ is drawn from a probabilistic distribution $p(\boldsymbol{x})$, the entropy of the distribution:[47]

$$h(\boldsymbol{x}) \doteq -\int p(\boldsymbol{x}) \log p(\boldsymbol{x}) \mathrm{d}\boldsymbol{x} \tag{1.4.3}$$

provides a natural measure of its complexity. This measure also has a natural interpretation as the average number of binary bits needed to encode such a sequence, as we will see in Chapter 3.

To illustrate the main ideas of this view, let us take a large number of long sequence segments:

$$\{\boldsymbol{x}_1, \boldsymbol{x}_2, \ldots, \boldsymbol{x}_i, \ldots, \boldsymbol{x}_N\} \subset \mathbb{R}^D, \tag{1.4.4}$$

generated by a predicting function $f$. Without loss of generality, we assume all sequences have the same length $D$, so each sequence can be viewed as a vector in $\mathbb{R}^D$. We introduce a coding scheme (with a codebook), denoted $\mathcal{E}$, that maps every segment $\boldsymbol{x}_i$ to a unique stream of binary bits $\mathcal{E}(\boldsymbol{x}_i)$, and a corresponding decoding scheme $\mathcal{D}$ that maps binary bit strings back to sequence segments (say subject to error $\epsilon \geq 0$ as such an encoding scheme can be lossy). The simplest scheme fills the space spanned by all segments with $\epsilon$-balls, as shown in Figure 1.23. We then number the balls, and each sequence is encoded as the binary index of its closest ball, while each binary index is decoded back to the center of the corresponding ball. Thus, each segment can be recovered up to precision $\epsilon$ from its bit stream.

Then, for an encoding $\mathcal{E}$, the complexity of the predicting function $f$ can be evaluated as the average coding length $L(\mathcal{E}(\boldsymbol{x}))$ of all sequences $\boldsymbol{x}$ that it generates, known as the coding rate:[48]

$$R(\boldsymbol{x} \mid \mathcal{E}) = \mathbb{E}[L(\mathcal{E}(\boldsymbol{x}))] \approx \frac{1}{N} \sum_{i=1}^{N} L(\mathcal{E}(\boldsymbol{x}_i)). \tag{1.4.5}$$

[45] In practice, we typically care about learning the predicting function $f$, rather than any particular sequence generated by $f$.

[46] This framework has successfully guided engineering practice in the communication industry for over 80 years.

[47] Here we consider differential entropy as we assume the sequence consists of continuous variables. For discrete variables, we would use $H(\boldsymbol{x}) = -\sum_i p(\boldsymbol{x}_i) \log p(\boldsymbol{x}_i)$.

[48] One may make this more precise by taking $R(\boldsymbol{x} \mid \mathcal{E})$ to be the expected coding length for all segments $\boldsymbol{x}$ of length $D$.

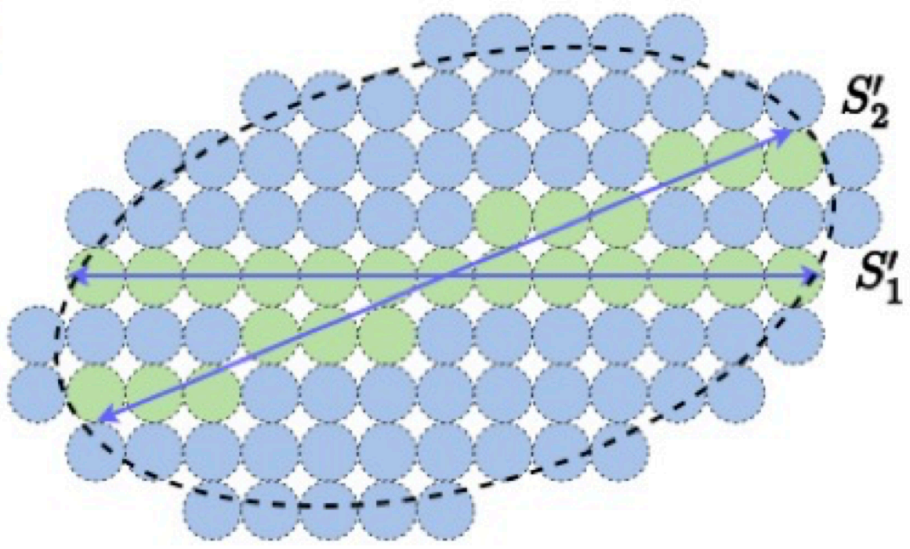

Figure 1.23: Comparison of two coding schemes. Imagine the true data distribution lies along the two arrowed lines $S_1'$ and $S_2'$. Samples from these lines can be encoded with a codebook of all blue balls, or with a codebook of only the green balls. The second scheme yields a much lower coding length/rate subject to the same precision.

The coding rate measure will change if we use a different coding scheme (or codebook). In practice, the better we know the low-dimensional structure around which the segments are distributed, the more efficient a codebook we can design, as illustrated in Figure 1.23. Astute readers may recognize that conceptually the denoising process illustrated in Figure 1.22 closely resembles going from the coding scheme with the blue balls to that with the green ones.

Given two coding schemes $\mathcal{E}_1$ and $\mathcal{E}_2$ for the segments, if the difference in the coding rates is positive:

$$R(\boldsymbol{x} \mid \mathcal{E}_1) - R(\boldsymbol{x} \mid \mathcal{E}_2) > 0, \tag{1.4.6}$$

we may say the coding scheme $\mathcal{E}_2$ is better. This difference can be viewed as a measure of how much more information $\mathcal{E}_2$ has over $\mathcal{E}_1$ about the distribution of the data. To a large extent, the goal of learning the data distribution is equivalent to finding the most efficient encoding and decoding scheme that minimizes the coding rate subject to a desired precision:

$$\min_{\mathcal{E},\mathcal{D}} R(\boldsymbol{x} \mid \mathcal{E}) \quad \text{subject to} \quad \operatorname{dist}(\boldsymbol{x}, \mathcal{D}(\mathcal{E}(\boldsymbol{x}))) \leq \epsilon. \tag{1.4.7}$$

As we will see in Chapter 4, the achievable minimal rate is closely related to the notion of entropy $H(\boldsymbol{x})$ (1.4.3).

*Remark* 1.2. The perspective of measuring data complexity with explicit encoding schemes has motivated several learning objectives proposed to revise Kolmogorov complexity for better computability [WD99], including the minimum message length (MML) proposed in 1968 [WB68] and the minimum description length (MDL) in 1978 [HY01; Ris78]. These objectives normally count the coding length for the coding scheme $\mathcal{E}$ itself (including its codebook) in addition to the data $\boldsymbol{x}$ of interest: $L(\mathcal{E}(\boldsymbol{x})) + L(\mathcal{E})$. However, if the goal is to learn a finite-sized codebook and apply it to a large number of sequences, the amortized

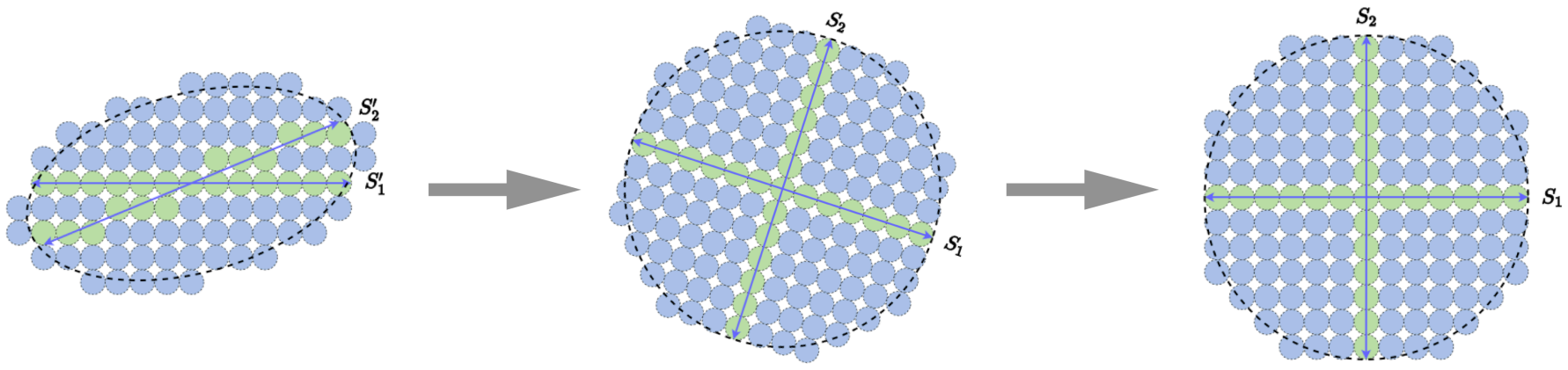

Figure 1.24: Transforming the identified low-dimensional data distribution, here on two linear subspaces, into a more informative and structured representation, where the two subspaces become more incoherent.

cost of the codebook can be ignored since

$$\frac{1}{N}\left(L(\mathcal{E})+\sum_{i=1}^{N} L(\mathcal{E}(\boldsymbol{x}_i))\right) \approx \frac{1}{N}\sum_{i=1}^{N} L(\mathcal{E}(\boldsymbol{x}_i)) \tag{1.4.8}$$

as $N$ becomes large.

Again, one may view the resulting optimal coding scheme as the one that achieves the best compression of the observed data. In general, compared to the Kolmogorov complexity, the coding length given by any encoding scheme will always be larger:

$$K(\boldsymbol{x}) < L(\mathcal{E}(\boldsymbol{x})). \tag{1.4.9}$$

Therefore, minimizing the coding rate/length essentially minimizes an upper bound of the otherwise uncomputable Kolmogorov complexity.

### 1.4.2 Learning Informative Representations

If the goal were simply to compress the given data for its own sake, the optimal codes approaching Kolmogorov complexity would in theory become nearly random or structureless [Cha66].[49] However, our true purpose in learning the data distribution of sequences $\boldsymbol{x}$ is to use it repeatedly with ease in future predictions. Hence, while compression allows us to identify the low-dimensional distribution in the data, we would like to encode this distribution in a *structured and organized* way so that the resulting representation is informative and efficient to use[50], as a "memory". Figure 1.24 illustrates intuitively why such a transformation is desirable.

As we will show in Chapter 4, these desired structures in the final representation can be precisely promoted by choosing a natural measure of *information gain* based on the coding rates of the chosen coding schemes.[51] Throughout this book, such an explicit and constructive coding approach provides a powerful computational framework for learning good representations of low-dimensional structures in real-world data. In many cases of practical importance, the coding length function can be efficiently computed or accurately approximated. In some benign cases, we can even obtain closed-form formulae—for example, subspace and Gaussian models.

[49] Any codes with structure can be further compressed.

[50] For example, to sample the distribution under different conditions.

[51] The information gain is a pervasive idea in algorithmic information theory and has been incorporated in several approaches to this topic, most recently $\mathcal{V}$-information [XZS+20]. This particular instantiation, similarly to many other measures of complexity we discussed, suffers

Moreover, this computational framework leads to a principled approach that naturally reveals the role deep networks play in this learning process. As we will derive systematically in Chapter 5, the layers of a deep network perform operations that incrementally optimize the objective function of interest. From this perspective, the role of deep networks can be precisely interpreted as emulating a certain iterative optimization algorithm—say, gradient descent—to optimize the objective of information gain. Layers of the resulting deep architectures can be endowed with precise statistical and geometric interpretations, namely performing incremental compressive encoding and decoding operations. Consequently, the derived deep networks become transparent "white boxes" that are mathematically fully explainable.

### 1.4.3 Learning Consistent Representations

To summarize our discussions so far, let us denote the data as

$$\boldsymbol{X} = \{\boldsymbol{x}_1, \boldsymbol{x}_2, \ldots, \boldsymbol{x}_i, \ldots, \boldsymbol{x}_N\} \subset \mathbb{R}^D, \tag{1.4.10}$$

and let $\boldsymbol{Z} = \mathcal{E}(\boldsymbol{X})$ be the codes of $\boldsymbol{X}$ via some encoder $\mathcal{E}$:

$$\boldsymbol{X} \xrightarrow{\mathcal{E}} \boldsymbol{Z}. \tag{1.4.11}$$

In the machine learning context, $\boldsymbol{Z}$ is often called the "features" of $\boldsymbol{X}$. Note that without knowing the underlying distribution of $\boldsymbol{X}$, we do not know which encoder $\mathcal{E}$ should be used to retain the most useful information about the distribution of $\boldsymbol{X}$. In practice, people often start by trying a compact encoding scheme that serves a specific task well. In particular, they try to learn an encoder that optimizes a certain (empirical) measure of parsimony for the learned representation:

$$\min \rho(\boldsymbol{Z}). \tag{1.4.12}$$

*Example* 1.2. For example, image classification is such a case: we assign all images in the same class to a single code and images in different classes to different codes, say "one-hot" vectors:

$$\boldsymbol{x} \mapsto \boldsymbol{z} \in \{[1, 0, 0, \ldots, 0, 0],\ [0, 1, 0 \ldots, 0, 0],\ \ldots,\ [0, 0, 0, \ldots, 0, 1].\} \tag{1.4.13}$$

Now, a classifier $f(\cdot)$ can be modeled as a function that predicts the probability of a given $\boldsymbol{x}$ belonging to each of the $k$ classes: $\hat{\boldsymbol{z}} = f(\boldsymbol{x}) \in \mathbb{R}^K$. Then the from being increasingly poorly-conditioned as the data support becomes more compact and low-dimensional, and so is not suitable for our needs.

"goodness" of a classifier can be measured by the so-called *cross entropy*:[52]

$$\ell(\hat{\boldsymbol{z}}, \boldsymbol{z}) = \sum_{k=1}^{K} -z_k \log \hat{z}_k, \tag{1.4.14}$$

where $z_k$ indicates the $k$-th entry of the vector $\boldsymbol{z}$. As the early practice of deep networks indicated [KSH12], if enough data are given, such an encoding scheme can often be represented by a deep network and learned in an end-to-end fashion by optimizing the cross-entropy. ■

The cross-entropy loss $\ell(\hat{\boldsymbol{z}}, \boldsymbol{z})$ can be viewed as a special measure of parsimony $\rho(\boldsymbol{z})$ associated with a particular family of encoding schemes that are suitable for classification. However, such an encoding is obviously *very lossy*. The learned $\boldsymbol{z}$ does not contain any other information about $\boldsymbol{x}$ except for its class type. For example, by assigning an image with (a code representing) the class label "apple", we no longer know which specific type of apple is in the original image from the label alone.

Of course, the other extreme is to require the coding scheme to be *lossless*. That is, there is a one-to-one mapping between $\boldsymbol{x}$ and its code $\boldsymbol{z}$. However, as we will see in Chapter 4, lossless coding (or compression) is impractical unless $\boldsymbol{x}$ is discrete. For a continuous random variable, we may only consider lossy coding schemes so that the coding length for the data can be finite. That is, we only encode the data up to a certain prescribed precision. As we will elaborate more in Chapter 4, lossy coding is not merely a practical choice; it plays a fundamental role in making learning of the underlying continuous distribution possible from finite samples of the distribution.

For many learning purposes, we want the feature $\boldsymbol{z}$, although *lossy*, to retain more information about $\boldsymbol{x}$ than just its class type. In this book, we will introduce a more general measure of parsimony based on coding length/rate associated with a more general family of coding schemes—coding with a mixture of subspaces or Gaussians. This family has the capability to closely approximate arbitrary real-world distributions up to a certain precision. As we will see in Chapter 4 and Chapter 5, such a measure will not only preserve most information about the distribution of $\boldsymbol{X}$ but will also promote certain desirable geometric and statistical structures for the learned representation $\boldsymbol{Z}$.

**Bidirectional auto-encoding for consistency.** In a broader learning context, the main goal of a compressive coding scheme $\mathcal{E}$ is to identify the low-dimensional structures in the data $\boldsymbol{X}$ so that they can be used to predict things in the original data space. This requires that the learned encoding scheme $\mathcal{E}$ admits an efficient decoding scheme, denoted $\mathcal{D}$, which maps the feature or

[52] The cross entropy can be viewed as a distance measure between the ground truth distribution of $\boldsymbol{z}$ and that of the prediction $\hat{\boldsymbol{z}}$. It can also be viewed as the expected coding length of $\boldsymbol{z}$ if we use the optimal code book for $\hat{\boldsymbol{z}}$ to encode $\boldsymbol{z}$. The cross entropy reaches its minimum when $\boldsymbol{z}$ and $\hat{\boldsymbol{z}}$ have the same distribution.

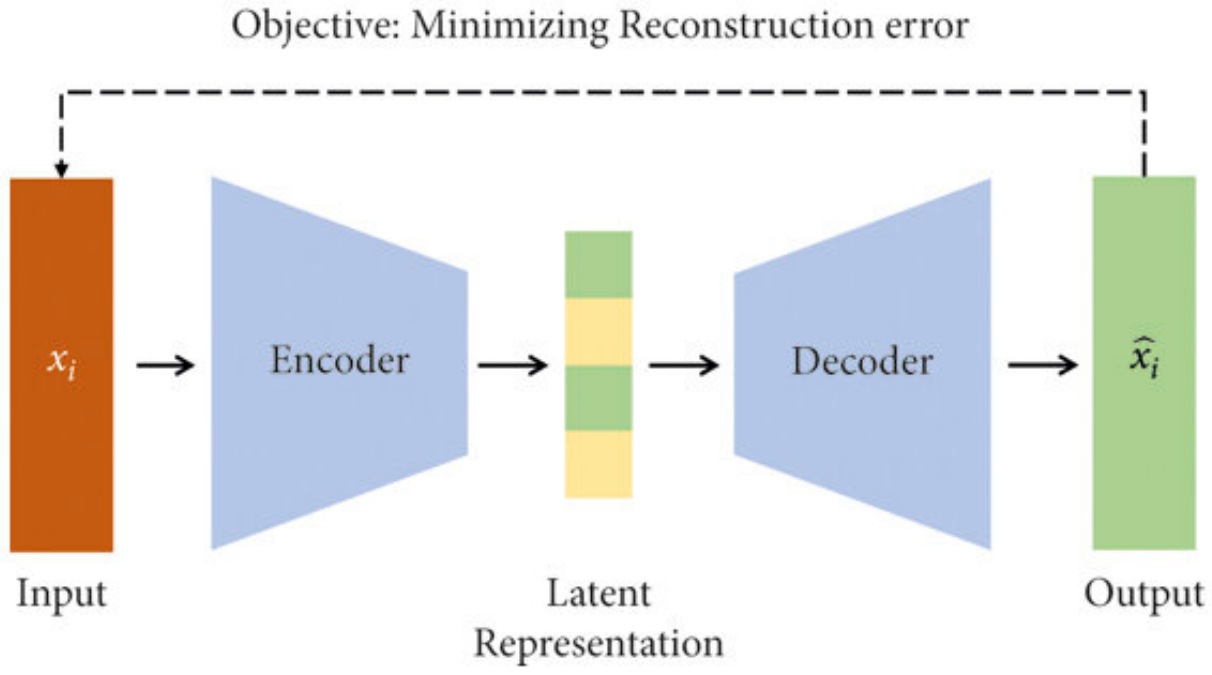

Figure 1.25: Illustration of the architecture of an autoencoder.

encoded data $\boldsymbol{Z}$ back to the data space:

$$\boldsymbol{X} \xrightarrow{\mathcal{E}} \boldsymbol{Z} \xrightarrow{\mathcal{D}} \hat{\boldsymbol{X}}. \tag{1.4.15}$$

We call such an encoding–decoding pair $(\mathcal{E}, \mathcal{D})$ an *auto-encoding*. Figure 1.25 illustrates this process.

Ideally, the decoding is approximately an "inverse" of the encoding, so that the data (distribution) $\hat{\boldsymbol{X}}$ decoded from $\boldsymbol{Z}$ is similar to the original data (distribution) $\boldsymbol{X}$ to some extent.[53] If this holds, we can recover or predict from $\boldsymbol{Z}$ what occurs in the original data space. In this case, we say the pair $(\mathcal{E}, \mathcal{D})$ provides a *consistent* auto-encoding. For most practical purposes, we not only need such a decoding to exist, but also need it to be efficiently realizable and physically implementable. For example, if $\boldsymbol{x}$ is real-valued, quantization is required for any decoding scheme to be realizable on a finite-state machine, as we will explain in Chapter 4. Thus, in general, realizable encoding and decoding schemes are necessarily lossy. A central question is how to learn a compact (lossy) representation $\boldsymbol{Z}$ that can still predict $\boldsymbol{X}$ well.

Generally speaking, both the encoder and decoder can be modeled and realized by deep networks and learned by solving an optimization problem of the following form:

$$\min_{f,g} [\ell(\boldsymbol{X}, \hat{\boldsymbol{X}}) + \rho(\boldsymbol{Z})], \qquad \text{where} \quad \boldsymbol{Z} = f(\boldsymbol{X}), \quad \hat{\boldsymbol{X}} = g(\boldsymbol{Z}) \tag{1.4.16}$$

where $\ell(\cdot, \cdot)$ is a distance function that promotes similarity between $\boldsymbol{X}$ and $\hat{\boldsymbol{X}}$[54] and $\rho(\boldsymbol{Z})$ is a measure that promotes parsimony and information richness of $\boldsymbol{Z}$. The classic principal component analysis (PCA) [Jol02] is a typical example of a consistent auto-encoding, which we will study in great detail in Chapter 2, as a precursor to more general low-dimensional structures. In Chapter 6, we will study how to learn consistent auto-encoding for general (say nonlinear) low-dimensional distributions.

[53] We will make precise what we mean by "similar" later.
[54] Either sample-wise or distribution-wise, depending on the choice of $\ell$.

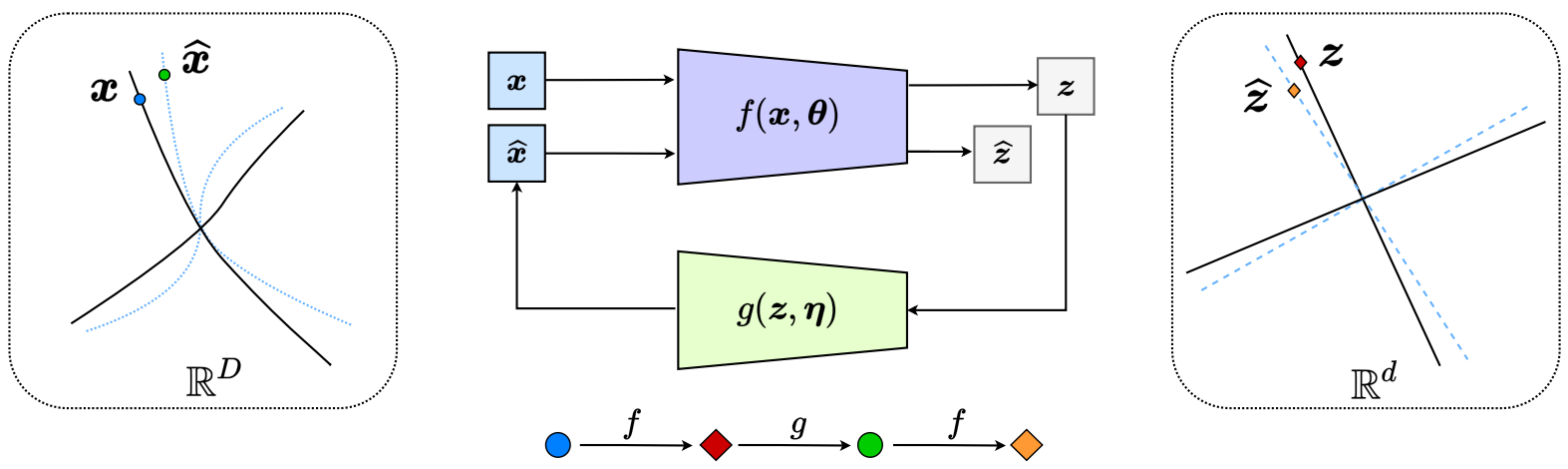

Figure 1.26: Illustration of a closed-loop transcription. Here we use a mapping $f$ to represent the encoder $\mathcal{E}$ and $g$ to represent the decoder $\mathcal{D}$.

### 1.4.4 Learning Self-Consistent Representations

In the auto-encoding objective above, one must evaluate how close the decoded data $\hat{\boldsymbol{X}}$ is to the original $\boldsymbol{X}$. This typically requires external supervision or knowledge of an appropriate similarity measure. Computing similarity between $\hat{\boldsymbol{X}}$ and $\boldsymbol{X}$ can be prohibitively expensive, or even impossible.[55] In nature, animals learn autonomously without comparing their estimate $\hat{\boldsymbol{X}}$ with the ground truth $\boldsymbol{X}$ in the data space; they typically lack that option.[56]

How, then, can a system learn without external supervision or comparison? How can it know that $\hat{\boldsymbol{X}}$ is consistent with $\boldsymbol{X}$ without direct comparison? This leads to the idea of "closing the loop." Under mild conditions (made precise in Chapter 6), ensuring consistency between $\boldsymbol{X}$ and $\hat{\boldsymbol{X}}$ reduces to encoding $\hat{\boldsymbol{X}}$ as $\hat{\boldsymbol{Z}}$ and checking consistency between $\boldsymbol{Z}$ and $\hat{\boldsymbol{Z}}$. We call this notion *self-consistency*, illustrated by:

$$\boldsymbol{X} \xrightarrow{\mathcal{E}} \boldsymbol{Z} \xrightarrow{\mathcal{D}} \hat{\boldsymbol{X}} \xrightarrow{\mathcal{E}} \hat{\boldsymbol{Z}}, \tag{1.4.17}$$

We call this process *closed-loop transcription*,[57] depicted in Figure 1.26.

It is arguably true that any autonomous intelligent being only needs to learn a self-consistent representation $\boldsymbol{Z}$ of the observed data $\boldsymbol{X}$, because checking consistency in the original data space—often meaning in the external world—is either too expensive or even physically infeasible. The closed-loop formulation allows one to learn an optimal encoding $f(\cdot, \boldsymbol{\theta})$ and decoding $g(\cdot, \boldsymbol{\eta})$ via a min-max game that depends only on the internal (learned) feature $\boldsymbol{Z}$:

$$\max_{\boldsymbol{\theta}} \min_{\boldsymbol{\eta}} [\ell(\boldsymbol{Z}, \hat{\boldsymbol{Z}}) + \rho(\boldsymbol{Z})], \qquad \text{where} \quad \boldsymbol{Z} = f(\boldsymbol{X}), \quad \hat{\boldsymbol{X}} = g(\boldsymbol{Z}), \quad \hat{\boldsymbol{Z}} = f(\hat{\boldsymbol{X}}) \tag{1.4.18}$$

where $\ell(\boldsymbol{Z}, \hat{\boldsymbol{Z}})$ is a loss function based on coding rates of the features $\boldsymbol{Z}$ and $\hat{\boldsymbol{Z}}$, which, as we will see, can be much easier to compute. Here again, $\rho(\boldsymbol{Z})$ is some measure that promotes parsimony and information richness of $\boldsymbol{Z}$. Somewhat surprisingly, as we will see in Chapter 6, under rather mild conditions such as $\boldsymbol{X}$ being sufficiently low-dimensional, self-consistency between $(\boldsymbol{Z}, \hat{\boldsymbol{Z}})$ implies consistency in $(\boldsymbol{X}, \hat{\boldsymbol{X}})$! In addition, we will also see that a closed-loop system will allow us to learn the distribution in a *continuous and incremental* manner,[58] without suffering from problems such as catastrophic forgetting associated with open-loop models.

[55] For instance, minimizing a distributional distance between the two random variables is statistically and computationally difficult.

[56] In animals and humans, the eyes process visual data long before it arrives at the brain; the brain does not have a mechanism to process raw visual data and must rely on the eyes' features to do any learning.

[57] Inspired by transcription between DNA and RNA or other proteins.

## 1.5 Bridging Theory and Practice for Machine Intelligence

So far, we have introduced several related frameworks for learning a compact and structured representation $\boldsymbol{Z}$ for a given data distribution $\boldsymbol{X}$:

- the open-ended encoding (1.4.11);
- the bi-directional auto-encoding (1.4.15);
- the closed-loop transcription (1.4.17).

In this book, we will systematically study all three frameworks, one after another:

$$\textbf{open-ended} \implies \textbf{bi-directional} \implies \textbf{closed-loop}, \tag{1.5.1}$$

in Chapter 5, Section 6.1, and Section 6.2, respectively.

As we will see in this book, these frameworks, including many discussed earlier in the chapter, can now be viewed as attempts to address *partially* the problem of learning a *complete and computable* representation of the distribution of the data $\boldsymbol{X}$:

$$\boldsymbol{X} \longleftrightarrow \boldsymbol{Z} \longleftrightarrow \boldsymbol{G}, \tag{1.5.2}$$

which includes two sets of mappings:

1. an autoencoding between the data $\boldsymbol{X}$ and its informative feature representation $\boldsymbol{Z}$,
2. a diffusion and denoising process between the learned representation $\boldsymbol{Z}$ and a simple distribution $\boldsymbol{G}$, say Gaussian.

The later process allows us to access and sample the learned structured low-dimensional features $\boldsymbol{Z}$ for tasks such as regenerating the original data $\hat{\boldsymbol{X}}$.[59] The overall geometric picture associated with these two (transforming) processes is illustrated in Figure 1.27.

[58] That is, to learn with one class at a time or even one sample at a time.

[59] Notice that even if the learned features lie on a single low-dimensional linear subspace, the most efficient ways to identify the subspace is essentially through a denoising process: the power iteration that we will elaborate in Chapter 2.

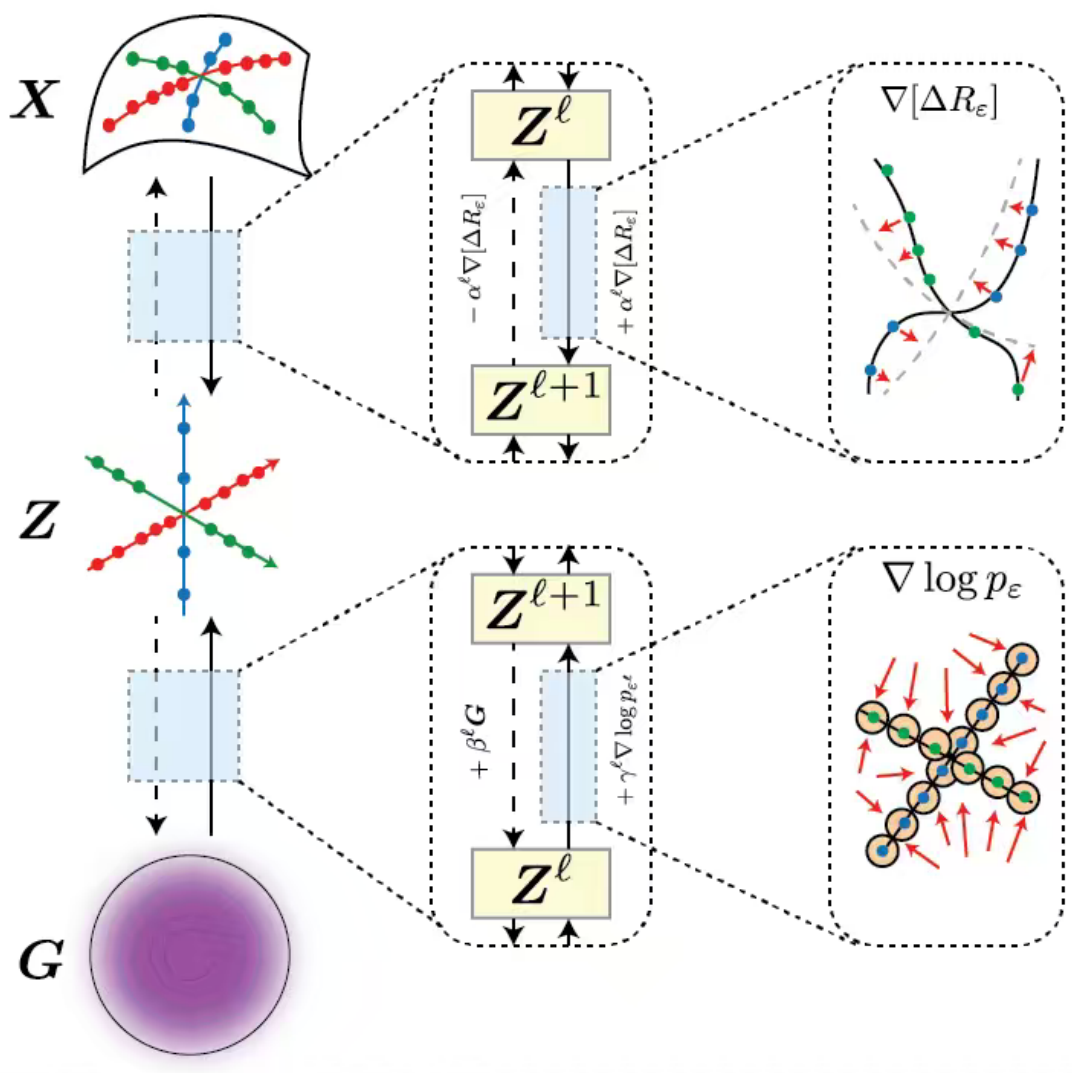

Figure 1.27: Illustration of a learned informative and structured feature representation $\boldsymbol{Z}$ for the data $\boldsymbol{X}$ that can be efficiently accessed through a denoising process from $\boldsymbol{G}$.

As we will see in Chapter 5, the role of modern deep networks can be precisely interpreted as to realize the operators[60] needed to conduct the above transformations among distributions. Popular architectures of deep neural networks (including CNNs, ResNet or Transformers) that were developed empirically will be mathematically interpreted, justified and improved through this approach.

In recent years, many theoretical frameworks have been proposed to help understand deep networks. However, most have failed to provide scalable solutions that match the performance of empirical methods on real-world data and tasks. Many theories offer little practical guidance for further empirical improvement. Chapter 7 and Chapter 8 will demonstrate how the framework presented in this book can bridge the gap between theory and practice. Chapter 7 will show how to use the learned distribution and its representation to conduct (Bayesian) inference for practical tasks involving (conditional) generation, estimation, and prediction. Chapter 8 will provide compelling experimental evidence that networks and systems designed from first principles can achieve competitive—and potentially superior—performance on tasks such as visual representation learning, image classification, image completion, segmentation, and text generation.

**Back to Intelligence.** As mentioned at the beginning, a fundamental task of any intelligent being is to learn predictable information from sensed data. We now understand the computational nature of this task and recognize that it is

[60]via unrolled optimization, as we will see

a never-ending process for two reasons:

- Memory or knowledge learned from data, say via encoding and decoding schemes, is unlikely to be correct or optimal. Intelligence must be able to improve when predictions of new observations contain errors.
- Observed data do not yet cover all predictable information. Intelligence must recognize when current memory or knowledge is inadequate and acquire new information whenever it becomes available.

Thus, intelligence is *not* about collecting all informative data in advance and training a model to memorize their (low-dimensional) distribution. Instead, it requires computational mechanisms that can continuously improve current knowledge of the distribution and acquire new information when needed. A fundamental characteristic of any intelligent being or system—whether an animal, a human, a civilization, or the entire scientific community—is *the ability to continuously improve or gain new information or knowledge on its own* through a closed-loop feedback about how good its current memory or knowledge is at explaining or predicting its observations. Hence, to some extent, we can symbolically express the relationship between intelligence and knowledge[61] as:

$$\begin{aligned} \text{Intelligence}(t) &= \frac{\mathrm{d}}{\mathrm{d}t}\text{Knowledge}(t), \\ \text{Knowledge}(t) &= \int_0^t \text{Intelligence}(s)\mathrm{d}s. \end{aligned}$$

A closed-loop feedback framework is a universal mechanism for any system capable of self-improvement and self-learning—reinforcement being a primitive form, as in natural selection—or gaming, with scientific inquiry representing its most advanced form through hypothesizing or conjecturing and falsification or verification. All intelligent beings and systems in nature employ such closed-loop mechanisms for learning at every level and scale. This ubiquity inspired early attempts to model intelligence with autonomous machines or semi-autonomous computers, particularly the Cybernetics movement pioneered and advocated by Norbert Wiener in the 1940s.

We hope this book helps readers better understand the scientific objectives, mathematical principles, and computational mechanisms underlying intelligence, at least the part for forming useful memory or empirical knowledge. We believe that this work not only bridges the theory and practice, but also connects the current to the future: It clarifies what we have done so far and at the same time reveals what we have not. Hence it lays the groundwork for future study of possibly new mechanisms behind higher-level intelligence that is unique to human—true "Artificial Intelligence" meant by the 1956 Dartmouth program. Such mechanisms can create abstract concepts and conduct deductive inference that, we believe, go beyond compression for memory from empirical data and conduct inductive inference that are studied and explained in this book and are being exploited in current "AI" technology. We will discuss open problems towards understanding and potentially realizing more advanced intelligence in the final Chapter 9.

[61] Here we mean "knowledge" in the broadest possible sense, including both empirical memory and scientific knowledge.

# Chapter 2

# Learning Linear and Independent Structures

> "*The art of doing mathematics consists in finding that special case which contains all the germs of generality.*"
>
> – David Hilbert

*Real data has low-dimensional structure.* To see why this is true, let us consider the unassuming case of static on a TV when the satellite isn't working. At each frame (approximately every $\frac{1}{30}$ seconds), the RGB static on a screen of size $H \times W$ is, roughly, sampled independently from a uniform distribution on $[0, 1]^{3 \times H \times W}$. In theory, the static *could* resolve to a natural image on any given frame, but even if you spend a thousand years looking at the TV screen, it will not. This discrepancy is explained by the fact that the set of $H \times W$ natural images takes up a vanishingly small fraction of the hypercube $[0, 1]^{3 \times H \times W}$. In particular, it is extremely low-dimensional, compared to the ambient space dimension. Similar phenomena occur for all other types of natural data, such as text, audio, and video. Thus, when we design systems and methods to process natural data and learn its structure or distribution, this is a central property of natural data which we need to take into account.

Therefore, our central task is to learn a distribution that has low intrinsic dimension in a high-dimensional space. In the remainder of this chapter, we discuss several *classical* methods to perform this task for several somewhat *idealistic* models for the distribution, namely models that are geometrically linear or statistically independent. While these models and methods are important and useful in their own right, we discuss them here as they motivate, inspire, and serve as a predecessor or analogue to more modern methods for more general distributions that involve deep (representation) learning.

Our main approach (and general problem formulation) can be summarized as:

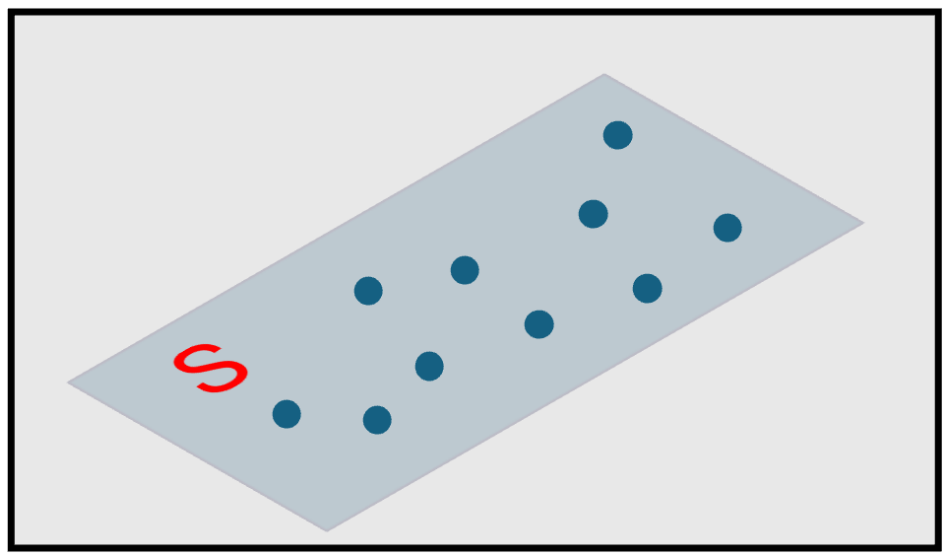

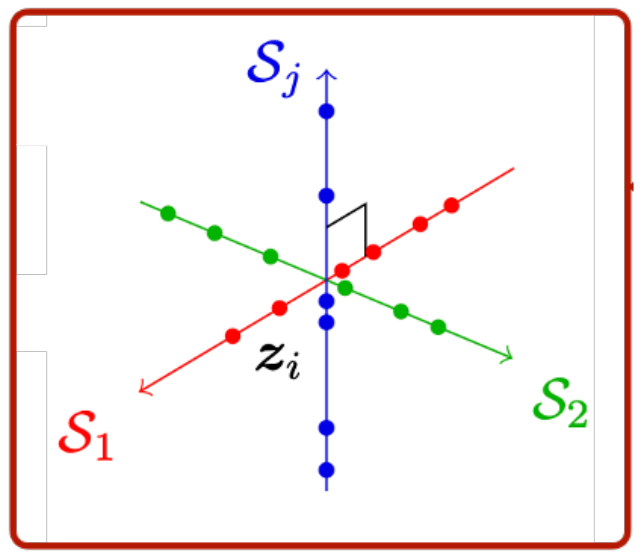

Figure 2.1: Data on a single low-dimensional subspace (left), say $\mathcal{S} = \operatorname{col}(\boldsymbol{U})$, or a mixture of low-dimensional subspaces (right), say $\mathcal{S}_j = \operatorname{col}(\boldsymbol{U}_j)$.

> **Problem:** *Given one or several (noisy or incomplete) observations of a ground truth sample from the data distribution, obtain an estimate of this sample.*

This approach underpins several classical methods for data processing, which we discuss in this chapter.

- Section 2.1 — Principal Components Analysis (PCA): Given noisy samples from a distribution supported on *one low-dimensional subspace*, obtain an estimate of the true sample that lies on this subspace.
- Section 2.2 — Complete Dictionary Learning and Independent Components Analysis (ICA): Given noisy samples from a distribution supported on *a union (not the span) of a few low-dimensional subspaces*, obtain an estimate of the true samples.
- Section 2.3 — Sparse Coding and Dictionary Learning: Given noisy samples from *a distribution supported on combinations of a few incoherent vectors*, such as the coordinate axes, obtain an estimate of the true sample, which also has this property.

In this chapter, as described above, we make simplifying modeling assumptions that essentially assume (the support of) the data distribution has geometrically (nearly, piece-wise) linear structures and statistically independent components, as illustrated in Figure 2.1. In Chapter 1, we have referred to such models of the data as "analytical models". These modeling assumptions allow us to derive efficient algorithms with provable efficiency guarantees, both in terms of data and computational complexity, for processing data at scale. However, they are imperfect models for often-complex real-world data distributions, and so their underlying assumptions only approximately hold. This means that the guarantees afforded by detailed analysis of these algorithms also only approximately hold in the case of real data.

Nonetheless, the techniques discussed in this chapter are useful in their own right, and beyond that, serve as the "special case with the germs of generality", so to speak, in that they present a guiding motivation and intuition for the more general paradigms within (deep) learning of more general distributions that we later address. As we will soon reveal in later chapters, in the deep learning era, modern "data-driven" or "non-parametric" approaches to learn (low-dimensional) data distributions essentially adopt the same concepts and ideas to learn.

**A Note on Notation.** We encourage the reader to consult the Notation chapter, particularly the "Modeling and Optimization Notation" section, as they begin to go through this chapter. We use the same notational conventions for modeling data and optimizing the parameters of these models throughout the book, and the simple, explicit models we study in this chapter help make the 'semantic' content of these conventions clear.

## 2.1 A Low-Dimensional Subspace

### 2.1.1 Principal Components Analysis (PCA)

Perhaps the simplest modeling assumption possible for low-dimensional structure is the so-called *low-rank* assumption. Letting $D$ be the dimension of our data space, we assume that our data belong to a low-dimensional subspace of dimension $d \ll D$, possibly plus some small disturbances.[1] This ends up being a nearly valid assumption for some surprisingly complex data, such as images of handwritten digits and face data [RD03] as shown in Figure 2.2, yet as we will see, it will lend itself extremely well to comprehensive analysis.

**Problem Formulation.** To write this in mathematical notation, we represent a subspace $\mathcal{S} \subseteq \mathbb{R}^D$ of dimension $d$ by an orthonormal matrix $\boldsymbol{U} \in \mathsf{O}(D, d) \subseteq \mathbb{R}^{D \times d}$ such that the columns of $\boldsymbol{U}$ span $\mathcal{S}$. Then, we say that our data $\{\boldsymbol{x}_i\}_{i=1}^N \subseteq \mathbb{R}^D$ have (approximate) low-rank structure if there exists an orthonormal matrix $\boldsymbol{U} \in \mathsf{O}(D, d)$, vectors $\{\boldsymbol{z}_i\}_{i=1}^N \subseteq \mathbb{R}^d$, and *small* vectors $\{\boldsymbol{\varepsilon}_i\}_{i=1}^N \subseteq \mathbb{R}^D$ such that

$$\boldsymbol{x}_i = \boldsymbol{U}\boldsymbol{z}_i + \boldsymbol{\varepsilon}_i, \quad \forall i \in [N]. \tag{2.1.1}$$

Here $\boldsymbol{\varepsilon}_i$ are disturbances that prevent the data from being perfectly low rank; their presence in our model allows us to quantify the degree to which our analysis remains relevant in the presence of deviations from our model. The true supporting subspace is $\mathcal{S} \doteq \mathrm{col}(\boldsymbol{U})$, the span of the columns of $\boldsymbol{U}$. To process all we can from these data, we need to recover $\mathcal{S}$; to do this it is sufficient to recover $\boldsymbol{U}$, also called the *principal components*. Fortunately, this is a computationally tractable task named **Principal Component Analysis**, and we discuss now how to solve it.

[1] Plus, potentially, a non-trivial offset from $\mathbf{0}$; in practice we account for this offset by subtracting the empirical mean over data samples from each sample, and in the following Chapter we pay no more heed to it.

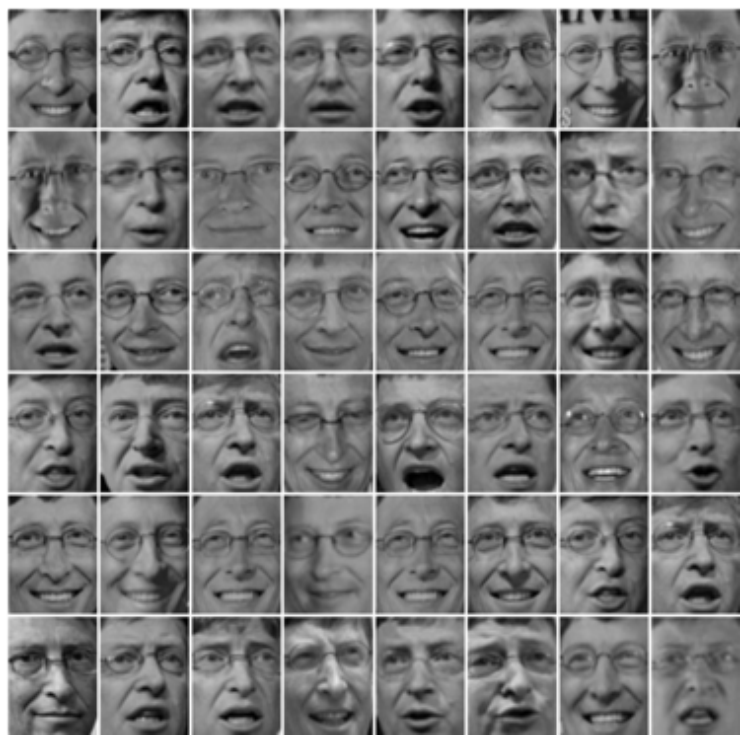

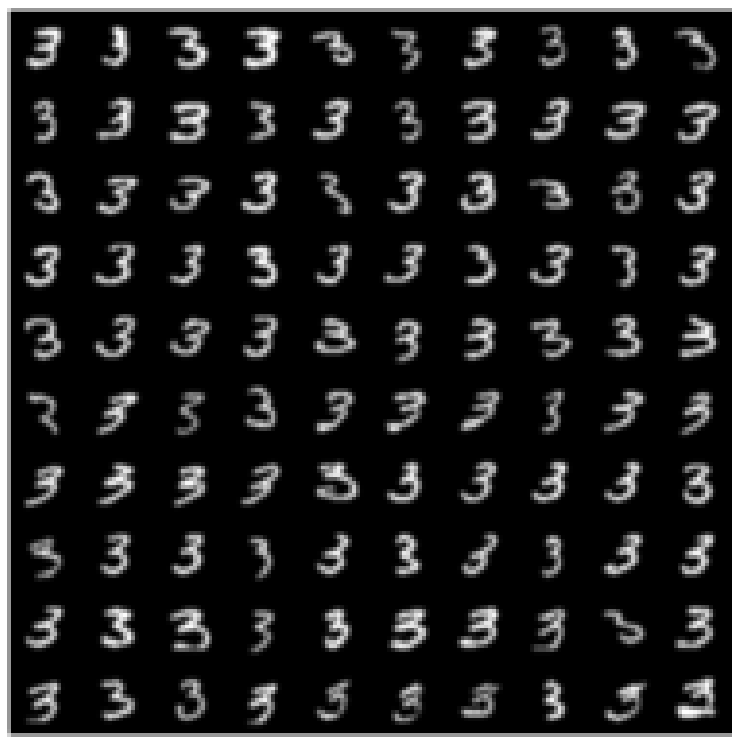

Figure 2.2: Images of aligned human faces and handwritten digits. Despite the seemingly large variety in their appearances, each set of these data spans (approximately) a very low-dimensional (nearly) linear subspace. This can be made technically precise in the notation of (2.1.1): for a relatively small subspace dimension $d$ (say, $\approx 20$, compared to the dimension $D = 784$ of MNIST digits), the error vectors $\boldsymbol{\varepsilon}_i$ are small for every $i \in [N]$ (where $N = 6 \cdot 10^4$ for MNIST is the dataset size). We can *learn* the optimal such subspace from data.

Given data $\{\boldsymbol{x}_i\}_{i=1}^N$ distributed as in (2.1.1), we aim to recover the model $\boldsymbol{U}$. A natural approach is to find the subspace $\boldsymbol{U}^\star$ and latent vectors $\{\boldsymbol{z}_i^\star\}_{i=1}^N$ which yield the best approximation $\boldsymbol{x}_i \approx \boldsymbol{U}^\star \boldsymbol{z}_i^\star$. Namely, we aim to solve the problem

$$\min_{\tilde{\boldsymbol{U}}, \{\tilde{\boldsymbol{z}}_i\}_{i=1}^N} \frac{1}{N} \sum_{i=1}^N \|\boldsymbol{x}_i - \tilde{\boldsymbol{U}} \tilde{\boldsymbol{z}}_i\|_2^2, \tag{2.1.2}$$

where $\tilde{\boldsymbol{U}}$ is constrained to be an orthonormal matrix, as above. We will omit this constraint in similar statements below for the sake of concision.

**Subspace Encoding-Decoding via Denoising.** To simplify this problem, for a fixed $\tilde{\boldsymbol{U}}$, we have (proof as exercise):

$$\min_{\{\tilde{\boldsymbol{z}}_i\}_{i=1}^N} \frac{1}{N} \sum_{i=1}^N \|\boldsymbol{x}_i - \tilde{\boldsymbol{U}} \tilde{\boldsymbol{z}}_i\|_2^2 = \frac{1}{N} \sum_{i=1}^N \min_{\tilde{\boldsymbol{z}}_i} \|\boldsymbol{x}_i - \tilde{\boldsymbol{U}} \tilde{\boldsymbol{z}}_i\|_2^2 \tag{2.1.3}$$

$$= \frac{1}{N} \sum_{i=1}^N \|\boldsymbol{x}_i - \tilde{\boldsymbol{U}} \tilde{\boldsymbol{U}}^\top \boldsymbol{x}_i\|_2^2. \tag{2.1.4}$$

That is, the optimal solution $(\boldsymbol{U}^\star, \{\boldsymbol{z}_i^\star\}_{i=1}^N)$ to the above optimization problem has $\boldsymbol{z}_i^\star = (\boldsymbol{U}^\star)^\top \boldsymbol{x}_i$.

Now, we can write the original optimization problem in $\tilde{\boldsymbol{U}}$ and $\{\tilde{\boldsymbol{z}}_i\}_{i=1}^N$ as an optimization problem just over $\tilde{\boldsymbol{U}}$, i.e., to obtain the basis $\boldsymbol{U}^\star$ and compact codes $\{\boldsymbol{z}_i^\star\}_{i=1}^N$ it suffices to solve either of the two following equivalent problems:

$$\min_{\tilde{\boldsymbol{U}}, \{\tilde{\boldsymbol{z}}_i\}_{i=1}^N} \frac{1}{N}\sum_{i=1}^N \|\boldsymbol{x}_i - \tilde{\boldsymbol{U}}\tilde{\boldsymbol{z}}_i\|_2^2 = \min_{\tilde{\boldsymbol{U}}} \frac{1}{N}\sum_{i=1}^N \|\boldsymbol{x}_i - \tilde{\boldsymbol{U}}\tilde{\boldsymbol{U}}^\top\boldsymbol{x}_i\|_2^2. \tag{2.1.5}$$

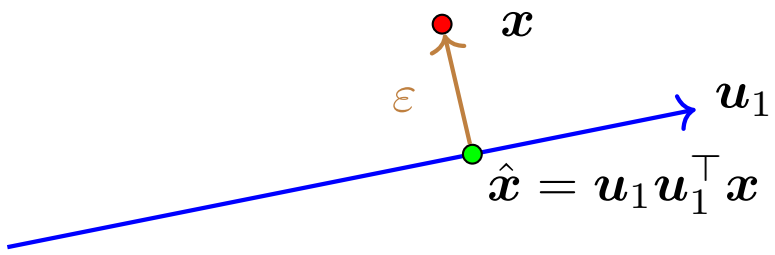

Figure 2.3: **Geometry of PCA.** A data point $\boldsymbol{x}$ (red) is projected onto the one-dimensional learned subspace spanned by the unit basis vector $\boldsymbol{u}_1$ (blue arrow). The projection $\boldsymbol{U}\boldsymbol{U}^\top\boldsymbol{x} = \boldsymbol{u}_1\boldsymbol{u}_1^\top\boldsymbol{x}$ (green) is the denoised version of $\boldsymbol{x}$ using the low-dimensional structure given by $\boldsymbol{u}_1$, and $\boldsymbol{\varepsilon}$ (brown arrow) represents the projection residual or noise.

Note that the problem on the right-hand side of (2.1.5) is a *denoising* problem: given noisy observations $\boldsymbol{x}_i$ of low-rank data, we aim to find the *noise-free* copy of $\boldsymbol{x}_i$, which under the model (2.1.1) is $\boldsymbol{U}\boldsymbol{z}_i$. That is, the denoised input $\hat{\boldsymbol{x}}_i = \boldsymbol{U}\boldsymbol{U}^\top\boldsymbol{x}_i$. Notice that this is the point on the subspace that is closest to $\boldsymbol{x}_i$, as visualized in Figure 2.3. Here by solving the equivalent problem of finding the best subspace, parameterized by the learned basis $\boldsymbol{U}^\star$, we learn an approximation to the *denoiser*, i.e., the projection matrix $\boldsymbol{U}^\star(\boldsymbol{U}^\star)^\top \approx \boldsymbol{U}\boldsymbol{U}^\top$ that projects the noisy data point $\boldsymbol{x}_i$ onto the subspace $\mathcal{S}$.

Putting the above process together, we essentially obtain a simple encoding-decoding scheme that maps a data point $\boldsymbol{x}$ in $\mathbb{R}^D$ to a lower-dimensional (latent) space $\mathbb{R}^d$ and then back to $\mathbb{R}^D$:

$$\boldsymbol{x} \xrightarrow{\mathcal{E}=(\boldsymbol{U}^\star)^\top} \boldsymbol{z} \xrightarrow{\mathcal{D}=\boldsymbol{U}^\star} \hat{\boldsymbol{x}}. \tag{2.1.6}$$

Here, $\boldsymbol{z} \in \mathbb{R}^d$ can be viewed as the low-dimensional compact code (or a latent representation) of a data point $\boldsymbol{x} \in \mathbb{R}^D$ and the learned subspace basis $\boldsymbol{U}^\star$ as the associated codebook whose columns are the (learned) optimal code words. The process achieves the function of denoising $\boldsymbol{x}$ by projecting it onto the subspace spanned by $\boldsymbol{U}^\star$.

**Computing the Subspace Basis.** For now, we continue our calculation. Let $\boldsymbol{X} = \begin{bmatrix}\boldsymbol{x}_1, \ldots, \boldsymbol{x}_N\end{bmatrix} \in \mathbb{R}^{D\times N}$ be the matrix whose columns are the observations $\boldsymbol{x}_i$. We have (proof as exercise)

$$\begin{aligned}\arg\min_{\tilde{\boldsymbol{U}}} \frac{1}{N}\sum_{i=1}^N \|\boldsymbol{x}_i - \tilde{\boldsymbol{U}}\tilde{\boldsymbol{U}}^\top\boldsymbol{x}_i\|_2^2 &= \arg\max_{\tilde{\boldsymbol{U}}} \frac{1}{N}\sum_{i=1}^N \|\tilde{\boldsymbol{U}}^\top\boldsymbol{x}_i\|_2^2 && (2.1.7)\\ &= \arg\max_{\tilde{\boldsymbol{U}}} \operatorname{tr}\left\{\tilde{\boldsymbol{U}}^\top\left(\frac{\boldsymbol{X}\boldsymbol{X}^\top}{N}\right)\tilde{\boldsymbol{U}}\right\}. && (2.1.8)\end{aligned}$$

Thus, to compute the principal components, we find the orthogonal matrix $\tilde{\boldsymbol{U}}$ which maximizes the term $\operatorname{tr}(\tilde{\boldsymbol{U}}^\top(\boldsymbol{X}\boldsymbol{X}^\top/N)\tilde{\boldsymbol{U}})$. We can prove via induction that this matrix $\boldsymbol{U}^\star$ has columns which are the *top $d$ unit eigenvectors* of $\boldsymbol{X}\boldsymbol{X}^\top/N$. We leave the whole proof to the reader in Exercise 2.1, but we handle the base case of the induction here. Suppose that $d = 1$. Then we only have a single unit vector $\boldsymbol{u}$ to recover, so the above problem reduces to

$$\max_{\tilde{\boldsymbol{u}} \colon \|\tilde{\boldsymbol{u}}\|_2 = 1} \tilde{\boldsymbol{u}}^\top(\boldsymbol{X}\boldsymbol{X}^\top/N)\tilde{\boldsymbol{u}}. \tag{2.1.9}$$

This is the so-called *Rayleigh quotient* of $\boldsymbol{X}\boldsymbol{X}^\top/N$. By invoking the spectral theorem we diagonalize $\boldsymbol{X}\boldsymbol{X}^\top/N = \boldsymbol{Q}\boldsymbol{\Lambda}\boldsymbol{Q}^\top$, where $\boldsymbol{Q}$ is orthogonal and $\boldsymbol{\Lambda}$ is diagonal with non-negative entries. Hence

$$\tilde{\boldsymbol{u}}^\top(\boldsymbol{X}\boldsymbol{X}^\top/N)\tilde{\boldsymbol{u}} = \tilde{\boldsymbol{u}}^\top\boldsymbol{Q}\boldsymbol{\Lambda}\boldsymbol{Q}^\top\tilde{\boldsymbol{u}} = (\boldsymbol{Q}^\top\tilde{\boldsymbol{u}})^\top\boldsymbol{\Lambda}(\boldsymbol{Q}^\top\tilde{\boldsymbol{u}}). \tag{2.1.10}$$

Since $\boldsymbol{Q}$ is an invertible orthogonal transformation, $\boldsymbol{Q}^\top\tilde{\boldsymbol{u}}$ is a unit vector, and optimizing over $\tilde{\boldsymbol{u}}$ is equivalent to optimizing over $\tilde{\boldsymbol{w}} \doteq \boldsymbol{Q}^\top\tilde{\boldsymbol{u}}$. Hence, we can write

$$\tilde{\boldsymbol{u}}^\top(\boldsymbol{X}\boldsymbol{X}^\top/N)\tilde{\boldsymbol{u}} = \tilde{\boldsymbol{w}}^\top\boldsymbol{\Lambda}\tilde{\boldsymbol{w}}, \tag{2.1.11}$$

whose optimal solutions $\boldsymbol{w}^\star$ among unit vectors are one-hot vectors whose only nonzero (hence unit) entry is in one of the indices corresponding to the largest eigenvalue of $\boldsymbol{X}\boldsymbol{X}^\top/N$. This means that $\boldsymbol{u}^\star = \boldsymbol{Q}\boldsymbol{w}^\star$, the optimal solution to the original problem, corresponds to a unit eigenvector of $\boldsymbol{X}\boldsymbol{X}^\top/N$ (i.e., column of $\boldsymbol{Q}$) which corresponds to the largest eigenvalue. Suitably generalizing this to the case $d > 1$, and summarizing all the previous discussion, we have the following informal Theorem.

**Theorem 2.1.** *Suppose that our dataset $\{\boldsymbol{x}_i\}_{i=1}^N \subseteq \mathbb{R}^D$ can be written as*

$$\boldsymbol{x}_i = \boldsymbol{U}\boldsymbol{z}_i + \boldsymbol{\varepsilon}_i, \qquad \forall i \in [N], \tag{2.1.12}$$

*where $\boldsymbol{U} \in \mathsf{O}(D, d)$ captures the low-rank structure, $\{\boldsymbol{z}_i\}_{i=1}^N \subseteq \mathbb{R}^d$ are the compact codes of the data, and $\{\boldsymbol{\varepsilon}_i\}_{i=1}^N \subseteq \mathbb{R}^D$ are small vectors indicating the deviation of our data from the low-rank model. Then the principal components $\boldsymbol{U}^\star \in \mathsf{O}(D, d)$ of our dataset are given by the top $d$ eigenvectors of $\boldsymbol{X}\boldsymbol{X}^\top/N$, where $\boldsymbol{X} = [\boldsymbol{x}_1, \ldots, \boldsymbol{x}_N] \in \mathbb{R}^{D\times N}$, and approximately correspond to the optimal linear denoiser: $\boldsymbol{U}^\star(\boldsymbol{U}^\star)^\top \approx \boldsymbol{U}\boldsymbol{U}^\top$.*

We do not give explicit rates of approximation here as they can become rather technical. In the special case that $\boldsymbol{\varepsilon}_i = \boldsymbol{0}$ for all $i$, the learned $\boldsymbol{U}^\star$ spans the support of the samples $\{\boldsymbol{x}_i\}_{i=1}^N$. If in addition the $\boldsymbol{z}_i$ are sufficiently diverse (say, spanning all of $\mathbb{R}^d$) then we would have perfect recovery: $\boldsymbol{U}^\star(\boldsymbol{U}^\star)^\top = \boldsymbol{U}\boldsymbol{U}^\top$.

*Remark* 2.1. In some data analysis tasks, the data matrix $\boldsymbol{X}$ is formatted such that each data point is a *row* rather than a *column* as is presented here. In this case the principal components are the top $d$ eigenvectors of $\boldsymbol{X}^\top\boldsymbol{X}/N$.

*Remark* 2.2 (Basis Selection via Denoising Eigenvalues)*.* In many cases, either our data will not truly be distributed according to a subspace-plus-noise model, or we will not know the true underlying dimension $d$ of the subspace. In this case, we have to choose $d$; this problem is called *model selection*. In the restricted case of PCA, one way to perform model selection is to compute $\boldsymbol{X}\boldsymbol{X}^\top/N$ and look for instances where adjacent eigenvalues sharply decrease; this is one indicator that the index of the larger eigenvalue is the "true dimension $d$", and the rest of the eigenvalues of $\boldsymbol{X}\boldsymbol{X}^\top/N$ are contributed by the noise or disturbances $\boldsymbol{\varepsilon}_i$. Model selection is a difficult problem and, nowadays in the era of deep learning where it is called "hyperparameter optimization", is usually done via brute force or Bayesian optimization, often using the loss on a holdout ("validation") dataset as an objective.

*Remark* 2.3 (Denoising Samples)*.* The expression on the right-hand side of (2.1.5), that is,

$$\min_{\tilde{\boldsymbol{U}}} \frac{1}{N}\sum_{i=1}^{N}\|\boldsymbol{x}_i - \tilde{\boldsymbol{U}}\tilde{\boldsymbol{U}}^\top \boldsymbol{x}_i\|_2^2, \tag{2.1.13}$$

is what's known as a *denoising problem*, thus named because it is an optimization problem whose solution *removes the noise from the samples so it fits on the subspace*. Denoising—learning a map which removes noise from noisy samples so that it fits on the data structure (such as in (2.1.1), but maybe more complicated)—is a common method for learning distributions that will be discussed in the sequel and throughout the manuscript. Note that we have already discussed this notion, but it bears repeating due to its central importance in later chapters.

*Remark* 2.4 (Neural Network Interpretation)*.* If we do a PCA, we approximately recover the support of the distribution encoded by the parameter $\boldsymbol{U}^\star$. The learned denoising map then takes the form $\boldsymbol{U}^\star(\boldsymbol{U}^\star)^\top$. On top of being a denoiser, this can be viewed as a *simple two-layer weight-tied linear neural network*: the first layer multiplies by $(\boldsymbol{U}^\star)^\top$, and the second layer multiplies by $\boldsymbol{U}^\star$, namely

$$\operatorname{denoise}(\boldsymbol{x}) = \underbrace{\boldsymbol{U}^\star \circ \underbrace{\mathrm{id} \circ \underbrace{(\boldsymbol{U}^\star)^\top \boldsymbol{x}}_{\text{first ``layer''}}}_{\text{post-activation of first ``layer''}}}_{\text{output of ``NN''}} \tag{2.1.14}$$

Contrasting this to a standard two-layer neural network, we see a structural similarity:

$$\operatorname{NN}(\boldsymbol{x}) = \underbrace{\boldsymbol{W}^\star \circ \underbrace{\mathrm{ReLU} \circ \underbrace{(\boldsymbol{U}^\star)^\top \boldsymbol{x}}_{\text{first layer}}}_{\text{post-activation of first layer}}}_{\text{output of NN}} \tag{2.1.15}$$

In particular, PCA can be interpreted as *learning a simple two-layer denoising*

*autoencoder*,[2] one of the simplest examples of a non-trivial neural network. In this framework, the *learned representations* would just be $(\boldsymbol{U}^\star)^\top \boldsymbol{x} \approx \boldsymbol{z}$. In this way, PCA serves as a model problem for (deep) representation learning, which we will draw upon further in the monograph. Notice that in this analogy, the representations reflect, or are projections of, the input data towards a learned low-dimensional structure. This property will be particularly relevant in the future.

### 2.1.2 Pursuing Low-Rank Structure via Power Iteration

There is a computationally efficient way to estimate the top eigenvectors of $\boldsymbol{X}\boldsymbol{X}^\top/N$ or any symmetric positive semidefinite matrix $\boldsymbol{M}$, called *power iteration*. This method is the building block of several algorithmic approaches to high-dimensional data analysis that we discuss later in the chapter, so we discuss it here.

Let $\boldsymbol{M}$ be a symmetric positive semidefinite matrix. There exists an orthonormal basis for $\mathbb{R}^D$ consisting of eigenvectors $(\boldsymbol{w}_i)_{i=1}^D$ of $\boldsymbol{M}$, with corresponding eigenvalues $\lambda_1 \geq \cdots \geq \lambda_D \geq 0$. By definition, any eigenvector $\boldsymbol{w}_i$ satisfies $\lambda_i \boldsymbol{w}_i = \boldsymbol{M}\boldsymbol{w}_i$. Therefore, for any $\lambda_i > 0$, $\boldsymbol{w}_i$ is a "fixed point" to the following equation:

$$\boldsymbol{w} = \frac{\boldsymbol{M}\boldsymbol{w}}{\|\boldsymbol{M}\boldsymbol{w}\|_2}. \tag{2.1.16}$$

**Theorem 2.2** (Power Iteration)**.** *Assume that $\lambda_1 > \lambda_i$ for all $i > 1$. If we compute the fixed point of (2.1.16) using the following iteration:*

$$\boldsymbol{v}_0 \sim \mathcal{N}(\boldsymbol{0}, \boldsymbol{1}), \qquad \boldsymbol{v}_{t+1} \leftarrow \frac{\boldsymbol{M}\boldsymbol{v}_t}{\|\boldsymbol{M}\boldsymbol{v}_t\|_2}, \tag{2.1.17}$$

*then, in the limit, $\boldsymbol{v}_t$ will converge to a top unit-norm eigenvector of $\boldsymbol{M}$.*

*Proof.* First, note that for all $t$, we have

$$\boldsymbol{v}_t = \frac{\boldsymbol{M}\boldsymbol{v}_{t-1}}{\|\boldsymbol{M}\boldsymbol{v}_{t-1}\|_2} = \frac{\boldsymbol{M}^2\boldsymbol{v}_{t-2}}{\|\boldsymbol{M}\boldsymbol{v}_{t-1}\|_2\|\boldsymbol{M}\boldsymbol{v}_{t-2}\|_2} = \cdots = \frac{\boldsymbol{M}^t\boldsymbol{v}_0}{\prod_{s=1}^t \|\boldsymbol{M}\boldsymbol{v}_s\|_2}. \tag{2.1.18}$$

Thus, $\boldsymbol{v}_t$ has the same direction as $\boldsymbol{M}^t\boldsymbol{v}_0$ and is unit norm, so we can write

$$\boldsymbol{v}_t = \frac{\boldsymbol{M}^t\boldsymbol{v}_0}{\|\boldsymbol{M}^t\boldsymbol{v}_0\|_2}. \tag{2.1.19}$$

Because all the eigenvectors $\boldsymbol{w}_i$ of $\boldsymbol{M}$ form an orthonormal basis for $\mathbb{R}^D$, we can write

$$\boldsymbol{v}_0 = \sum_{i=1}^D \alpha_i \boldsymbol{w}_i, \tag{2.1.20}$$

[2] In fact, as we have mentioned in the previous chapter, PCA was one of the first problems that neural networks were used to solve [BH89; Oja82].

where because $\boldsymbol{v}_0$ is Gaussian the $\alpha_i$ are all nonzero with probability 1. Thus, we can use our earlier expression for $\boldsymbol{v}_t$ to write

$$\boldsymbol{v}_t = \frac{\boldsymbol{M}^t \boldsymbol{v}_0}{\|\boldsymbol{M}^t \boldsymbol{v}_0\|_2} = \frac{\sum_{i=1}^D \lambda_i^t \alpha_i \boldsymbol{w}_i}{\|\sum_{i=1}^D \lambda_i^t \alpha_i \boldsymbol{w}_i\|_2} = \frac{\sum_{i=1}^D \lambda_i^t \alpha_i \boldsymbol{w}_i}{\sqrt{\sum_{i=1}^D \lambda_i^{2t} \alpha_i^2}}. \tag{2.1.21}$$

Now, let us consider the case where $\lambda_1 > \lambda_2 \geq \cdots \geq \lambda_D \geq 0$. (The case with repeated top eigenvalues goes similarly.) Then we can write

$$\boldsymbol{v}_t = \frac{\alpha_1 \boldsymbol{w}_1 + \sum_{i=2}^D (\lambda_i/\lambda_1)^t \alpha_i \boldsymbol{w}_i}{\sqrt{\alpha_1^2 + \sum_{i=2}^D (\lambda_i/\lambda_1)^{2t} \alpha_i^2}}. \tag{2.1.22}$$

Because $\lambda_1 > \lambda_i$ for all $i > 1$, the terms inside the summation go to 0 exponentially fast, and the remainder is the limit

$$\lim_{t\to\infty} \boldsymbol{v}_t = \frac{\alpha_1}{|\alpha_1|} \boldsymbol{w}_1 = \operatorname{sign}(\alpha_1) \boldsymbol{w}_1, \tag{2.1.23}$$

which is a top unit eigenvector of $\boldsymbol{M}$. The top eigenvalue $\lambda_1$ of $\boldsymbol{M}$ can be estimated by $\boldsymbol{v}_t^\top \boldsymbol{M} \boldsymbol{v}_t$, which converges similarly rapidly to $\lambda_1$. □

*Remark* 2.5 (Power Iteration as a Denoising Process). Notice that computing the principal component via power iteration can be interpreted as a "denoising" process: starting from random noise, it converges iteratively onto a one-dimensional subspace. Indeed, as we will show later in Chapter 3, if we assume *the support of the data distribution is a single subspace*, the associated denoising operator (or the score function) is precisely the same as the above power iteration. In Chapter 3, we will show how to generalize such a denoising process to distributions of more general low-dimensional structures. This is essentially a universal and most efficient approach to learn any low-dimensional distributions.

To find the second top eigenvector, we apply the power iteration algorithm to $\boldsymbol{M} - \lambda_1 \boldsymbol{v}_1 \boldsymbol{v}_1^\top$, which has eigenvectors $(\boldsymbol{w}_i)_{i=2}^D$ and corresponding eigenvalues $(\lambda_i)_{i=2}^D$. By repeating this procedure $d$ times in sequence, we can very efficiently estimate the top $d$ eigenvectors of $\boldsymbol{M}$ for any symmetric positive semidefinite matrix $\boldsymbol{M}$. Thus we can also apply it to $\boldsymbol{X}\boldsymbol{X}^\top/N$ to recover the top $d$ principal components, which is what we were after in the first place. Notice that this approach recovers one principal component at a time; we will contrast this to other algorithmic approaches, such as gradient descent on all eigenvectors, in future sections.

### 2.1.3 Local and Incremental Learning via Online PCA

Notice that in the above construction, the linear transform $\boldsymbol{U}$ (i.e., the matrix of principal components) used for encoding and decoding is computed "offline" from all the input data beforehand. One question is whether this transform can be learned "online" as the input data comes in order. This question was answered by the work of Oja in 1982 [Oja82].

*Example* 2.1 (Normalized Hebbian learning scheme for PCA). Consider a sequence of i.i.d. random vectors $\boldsymbol{x}_1, \ldots, \boldsymbol{x}_i, \ldots \in \mathbb{R}^n$ with covariance $\boldsymbol{\Sigma} \in \mathbb{R}^{n \times n}$. Let $\boldsymbol{u}_0 \in \mathbb{R}^n$ and define the response of an input vector $\boldsymbol{x}_i$ against a weight vector $\boldsymbol{u}_i$ to be their inner product:

$$\eta_i = \boldsymbol{u}_i^T \boldsymbol{x}_i \tag{2.1.24}$$

and we update the weight vector according to the following scheme:

$$\boldsymbol{u}_{i+1} = \frac{\boldsymbol{u}_i + \gamma \eta_i \boldsymbol{x}_i}{\|\boldsymbol{u}_i + \gamma \eta_i \boldsymbol{x}_i\|_2} \tag{2.1.25}$$

for some small gain $\gamma > 0$. This update scheme can be viewed as a normalized Hebbian scheme, in which the weights of connections between neurons become stronger if (products of) both the input $\boldsymbol{x}$ and output $\eta$ are strong. One may view the vector of weights $\boldsymbol{u}$ as "learned" based on a form of feedback from the output $\eta$. Then, under reasonable assumptions, Oja [Oja82] has shown that $\boldsymbol{u}_i$ converges to the eigenvector associated with the largest eigenvalue of $\boldsymbol{\Sigma}$. ■

To interpret the normalized Hebbian scheme (2.1.25), let us think of the data sample PCA objective (2.1.13) as an approximation to the *population* version of the objective, defined via an expectation over the distribution of samples $\boldsymbol{x}$:

$$\min_{\boldsymbol{U}^\top \boldsymbol{U} = \boldsymbol{I}} \mathbb{E}_{\boldsymbol{x}} \left[ \left\| \boldsymbol{x} - \boldsymbol{U}\boldsymbol{U}^\top \boldsymbol{x} \right\|_2^2 \right] \tag{2.1.26}$$

We will dig into this objective further in the following section. Then the normalized Hebbian scheme can be interpreted as a first-order approximation to a *stochastic* projected gradient descent scheme on Equation (2.1.26) (with batch size 1, and with the number of columns of $\boldsymbol{U}$ equal to 1) as long as $\|\boldsymbol{u}\|_2 = 1$, which is maintained by the projection operation in (2.1.25). Such an *online algorithm* is a useful primitive for more complex algorithms that process sequentially-structured data with causal structure, as we will study more in Section 5.3.3 and Section 8.11. Interestingly, and more generally, the simple online PCA algorithm is suggestive of the fact that *simpler algorithms than (end-to-end) back propagation can succeed in learning encoder-decoder pairs*, which we will briefly revisit in Chapter 9 as an important goal for the future study of intelligence.

### 2.1.4 The Statistical View: Probabilistic PCA

Notice that the above formulation makes no statistical assumptions on the data-generating process. However, it is common to include statistical elements within a given data model, as it may add further enlightening interpretations about the result of the analysis. As such, we ask the natural question: *what is the statistical analogue to low-dimensional structure?* Our answer is that a low-dimensional *distribution* is one whose support is concentrated around a low-dimensional geometric structure.

To illustrate this point, we discuss *probabilistic principal component analysis (PPCA)* [Row97; TB99]. This formulation can be viewed as a statistical variant of regular PCA. Mathematically, we now consider our data as samples from a random variable $\boldsymbol{x}$ taking values in $\mathbb{R}^D$ (also sometimes called a *random vector*). We say that $\boldsymbol{x}$ has (approximate) low-rank statistical structure if and only if there exists an orthonormal matrix $\boldsymbol{U} \in \mathsf{O}(D, d)$, a random variable $\boldsymbol{z}$ taking values in $\mathbb{R}^d$, and a *small* random variable $\boldsymbol{\varepsilon}$ taking values in $\mathbb{R}^D$ such that $\boldsymbol{z}$ and $\boldsymbol{\varepsilon}$ are independent, and

$$\boldsymbol{x} = \boldsymbol{U}\boldsymbol{z} + \boldsymbol{\varepsilon}. \tag{2.1.27}$$

Our goal is again to recover $\boldsymbol{U}$. Towards this end, we set up the analogous problem as in Section 2.1.1, i.e., optimizing over subspace supports $\tilde{\boldsymbol{U}}$ and random variables $\boldsymbol{z}$ to solve the following problem:

$$\min_{\tilde{\boldsymbol{U}}, \tilde{\boldsymbol{z}}} \mathbb{E}\, \|\boldsymbol{x} - \tilde{\boldsymbol{U}}\tilde{\boldsymbol{z}}\|_2^2. \tag{2.1.28}$$

Since we are finding the best such random variable $\boldsymbol{z}$ we can find its realization $\boldsymbol{z}(\boldsymbol{x})$ separately for each value of $\boldsymbol{x}$. Performing the same calculations as in Section 2.1.1, we obtain

$$\min_{\tilde{\boldsymbol{U}}, \tilde{\boldsymbol{z}}} \mathbb{E}\, \|\boldsymbol{x} - \tilde{\boldsymbol{U}}\tilde{\boldsymbol{z}}\|_2^2 = \min_{\tilde{\boldsymbol{U}}} \mathbb{E} \min_{\tilde{\boldsymbol{z}}(\boldsymbol{x})} \|\boldsymbol{x} - \tilde{\boldsymbol{U}}\tilde{\boldsymbol{z}}(\boldsymbol{x})\|_2^2 = \min_{\tilde{\boldsymbol{U}}} \mathbb{E}\, \|\boldsymbol{x} - \tilde{\boldsymbol{U}}\tilde{\boldsymbol{U}}^\top \boldsymbol{x}\|_2^2, \tag{2.1.29}$$

again re-emphasizing the fact that the estimated subspace with principal components $\boldsymbol{U}^\star$ corresponds to a denoiser $\boldsymbol{U}^\star(\boldsymbol{U}^\star)^\top$ which projects onto that subspace. As before, we obtain

$$\arg\min_{\tilde{\boldsymbol{U}}} \mathbb{E}\, \|\boldsymbol{x} - \tilde{\boldsymbol{U}}\tilde{\boldsymbol{U}}^\top \boldsymbol{x}\|_2^2 = \arg\max_{\tilde{\boldsymbol{U}}} \mathbb{E}\, \|\tilde{\boldsymbol{U}}^\top \boldsymbol{x}\|_2^2 \tag{2.1.30}$$

$$= \arg\max_{\tilde{\boldsymbol{U}}} \operatorname{tr}(\tilde{\boldsymbol{U}}^\top \mathbb{E}[\boldsymbol{x}\boldsymbol{x}^\top]\tilde{\boldsymbol{U}}), \tag{2.1.31}$$

and the solution to the latter problem is just the top $d$ principal components of the second moment matrix $\mathbb{E}[\boldsymbol{x}\boldsymbol{x}^\top]$. Actually, the above problems are visually very similar to the equations for computing the principal components in the previous subsection, except with $\mathbb{E}[\boldsymbol{x}\boldsymbol{x}^\top]$ replacing $\boldsymbol{X}\boldsymbol{X}^\top/N$. In fact, the latter quantity is an estimate for the former. Both formulations effectively do the same thing, and have the same practical solution—compute the left singular vectors of the data matrix $\boldsymbol{X}$, or equivalently the top eigenvectors of the estimated covariance matrix $\boldsymbol{X}\boldsymbol{X}^\top/N$. The statistical formulation, however, has an additional interpretation. Suppose that $\mathbb{E}[\boldsymbol{z}] = \boldsymbol{0}$ and $\mathbb{E}[\boldsymbol{\varepsilon}] = \boldsymbol{0}$. We have

$$\mathbb{E}[\boldsymbol{x}] = \boldsymbol{U}\,\mathbb{E}[\boldsymbol{z}] + \mathbb{E}[\boldsymbol{\varepsilon}] = \boldsymbol{0}, \tag{2.1.32}$$

so that $\operatorname{Cov}[\boldsymbol{x}] = \mathbb{E}[\boldsymbol{x}\boldsymbol{x}^\top]$. Now working out $\operatorname{Cov}[\boldsymbol{x}]$ we have

$$\operatorname{Cov}[\boldsymbol{x}] = \boldsymbol{U}\operatorname{Cov}[\boldsymbol{z}]\boldsymbol{U}^\top + \operatorname{Cov}[\boldsymbol{\varepsilon}] = \boldsymbol{U}\,\mathbb{E}[\boldsymbol{z}\boldsymbol{z}^\top]\boldsymbol{U}^\top + \operatorname{Cov}[\boldsymbol{\varepsilon}]. \tag{2.1.33}$$

In particular, if $\operatorname{Cov}[\boldsymbol{\varepsilon}]$ is small, it holds that $\operatorname{Cov}[\boldsymbol{x}] = \mathbb{E}[\boldsymbol{x}\boldsymbol{x}^\top]$ is approximately a low-rank matrix, in particular rank $d$. Thus the top $d$ eigenvectors of $\mathbb{E}[\boldsymbol{x}\boldsymbol{x}^\top]$

essentially summarize the whole covariance matrix. But they are also the principal components, so we can interpret principal component analysis as performing a low-rank decomposition of $\mathrm{Cov}[\boldsymbol{x}]$.

*Remark* 2.6. By using the probabilistic viewpoint of PCA, we achieve a clearer and more quantitative understanding of how it relates to denoising. First, consider the denoising problem in (2.1.29), namely

$$\min_{\tilde{\boldsymbol{U}}} \mathbb{E}\,\|\boldsymbol{x} - \tilde{\boldsymbol{U}}\tilde{\boldsymbol{U}}^\top \boldsymbol{x}\|_2^2. \tag{2.1.34}$$

It is not too hard to prove that if $\tilde{\boldsymbol{U}}$ has $d$ columns and if $\boldsymbol{\varepsilon}$ is an isotropic Gaussian random variable, i.e., with distribution $\boldsymbol{\varepsilon} \sim \mathcal{N}(\boldsymbol{0}, \sigma^2 \boldsymbol{I})$,[3] then for *any* optimal solution $\boldsymbol{U}^\star$ to this problem, we have

$$\boldsymbol{U}^\star(\boldsymbol{U}^\star)^\top = \boldsymbol{U}\boldsymbol{U}^\top \tag{2.1.35}$$

and so the true supporting subspace, say $\mathcal{S} \doteq \mathrm{col}(\boldsymbol{U})$, is recovered as the span of columns of $\boldsymbol{U}^\star$, since

$$\mathcal{S} = \mathrm{col}(\boldsymbol{U}) = \mathrm{col}(\boldsymbol{U}\boldsymbol{U}^\top) = \mathrm{col}(\boldsymbol{U}^\star(\boldsymbol{U}^\star)^\top) = \mathrm{col}(\boldsymbol{U}^\star). \tag{2.1.36}$$

In particular, the learned *denoising map* $\boldsymbol{U}^\star(\boldsymbol{U}^\star)^\top$ is an orthogonal projection onto $\mathcal{S}$, pushing noisy points onto the ground truth supporting subspace. We can establish a similar technical result in the case where we only have finite samples, as in Theorem 2.1, but this takes more effort and technicality. Summarizing this discussion, we have the following informal Theorem.

**Theorem 2.3.** *Suppose that the random variable $\boldsymbol{x}$ can be written as*

$$\boldsymbol{x} = \boldsymbol{U}\boldsymbol{z} + \boldsymbol{\varepsilon} \tag{2.1.37}$$

*where $\boldsymbol{U} \in \mathsf{O}(D, d)$ captures the low-rank structure, $\boldsymbol{z}$ is a random vector taking values in $\mathbb{R}^d$, and $\boldsymbol{\varepsilon}$ is a random vector taking values in $\mathbb{R}^D$ such that $\boldsymbol{z}$ and $\boldsymbol{\varepsilon}$ are independent, and $\boldsymbol{\varepsilon}$ is small. Then the principal components $\boldsymbol{U}^\star \in \mathsf{O}(D, d)$ of our dataset are given by the top $d$ eigenvectors of $\mathbb{E}[\boldsymbol{x}\boldsymbol{x}^\top]$, and approximately correspond to the optimal linear denoiser: $\boldsymbol{U}^\star(\boldsymbol{U}^\star)^\top \approx \boldsymbol{U}\boldsymbol{U}^\top$.*

### 2.1.5 Masked Low-Rank Matrix Completion

In the previous subsections, we discussed the problem of *learning a low-rank geometric or statistical distribution*, where the data were sampled from a subspace with additive noise. But this is not the only type of disturbance from a low-dimensional distribution that is worthwhile to study. In this subsection, we introduce one more class of non-additive errors which become increasingly important in deep learning. Let us consider the case where we have some data $\{\boldsymbol{x}_i\}_{i=1}^N$ generated according to (2.1.1). Now we arrange them into a matrix

[3] Other distributions work so long as they support all of $\mathbb{R}^D$, but the Gaussian is the easiest to work with here.

$\boldsymbol{X} = \left[\boldsymbol{x}_1, \dots, \boldsymbol{x}_N\right] \in \mathbb{R}^{D \times N}$. Unlike before, we do not observe $\boldsymbol{X}$ directly; we instead imagine that our observation was corrupted en route and we obtained

$$\boldsymbol{Y} = \boldsymbol{M} \odot \boldsymbol{X}, \tag{2.1.38}$$

where $\boldsymbol{M} \in \{0, 1\}^{D \times N}$ is a *mask* that is known by us, and $\odot$ is element-wise multiplication. In this case, our goal is to recover $\boldsymbol{X}$ (from which point we can use PCA to recover $\boldsymbol{U}$, etc), given only the corrupted observation $\boldsymbol{Y}$, the mask $\boldsymbol{M}$, and the knowledge that $\boldsymbol{X}$ is (approximately) low-rank. This is called *low-rank matrix completion*.

This and similar generalizations of this low-rank structure recovery problem are referred to as "Robust PCA" problems. There have been many excellent resources discussing algorithms and approaches to solve these problems [WM22]. Hence we will not go into the solution method here. Instead, we will discuss under what conditions this problem is *plausible* to solve. On one hand, in the most absurd case, suppose that each entry of the matrix $\boldsymbol{X}$ were chosen independently from all the others. Then there would be no hope of recovering $\boldsymbol{X}$ exactly even if only one entry were missing and we had $DN - 1$ entries. On the other hand, suppose that we knew that $\boldsymbol{X}$ has rank 1 exactly, which is an extremely strong condition on the low-dimensional structure of the data, and we were dealing with the mask

$$\boldsymbol{M} = \begin{bmatrix} \mathbf{1}_{(D-1)\times 1} & \mathbf{0}_{(D-1)\times(N-1)} \\ 1 & \mathbf{1}_{1\times(N-1)} \end{bmatrix}. \tag{2.1.39}$$

Then we know that the data are distributed on a line, and we know a vector on that line—it is just the first column of the matrix $\boldsymbol{Y} = \boldsymbol{M} \odot \boldsymbol{X}$. From the last coordinate of each column, also revealed to us by the mask, we can solve for each column since for each final coordinate there is only one vector on the line with this coordinate. Thus we can reconstruct $\boldsymbol{X}$ with perfect accuracy, and we only need a linear number of observations $D + N - 1$.

In the real world, actual problems are somewhere in between the two limiting cases discussed above. Yet the differences between these two extremes, as well as the earlier discussion of PCA, reveal a general kernel of truth:

*The lower-dimensional and more structured the data distribution is, the easier it is to process, and the fewer observations are needed—provided that the algorithm effectively utilizes this low-dimensional structure.*

As is perhaps predictable, we will encounter this motif repeatedly in the remainder of the manuscript, starting in the very next section.[4]

[4] In later chapters, we will see that one can complete real-world data such as images with only a fraction of unmasked observations, e.g., the masked autoencoding of images in Section 8.5. This is also because the distribution of natural images has rather low-dimensional (albeit nonlinear) structures in the space of all images.

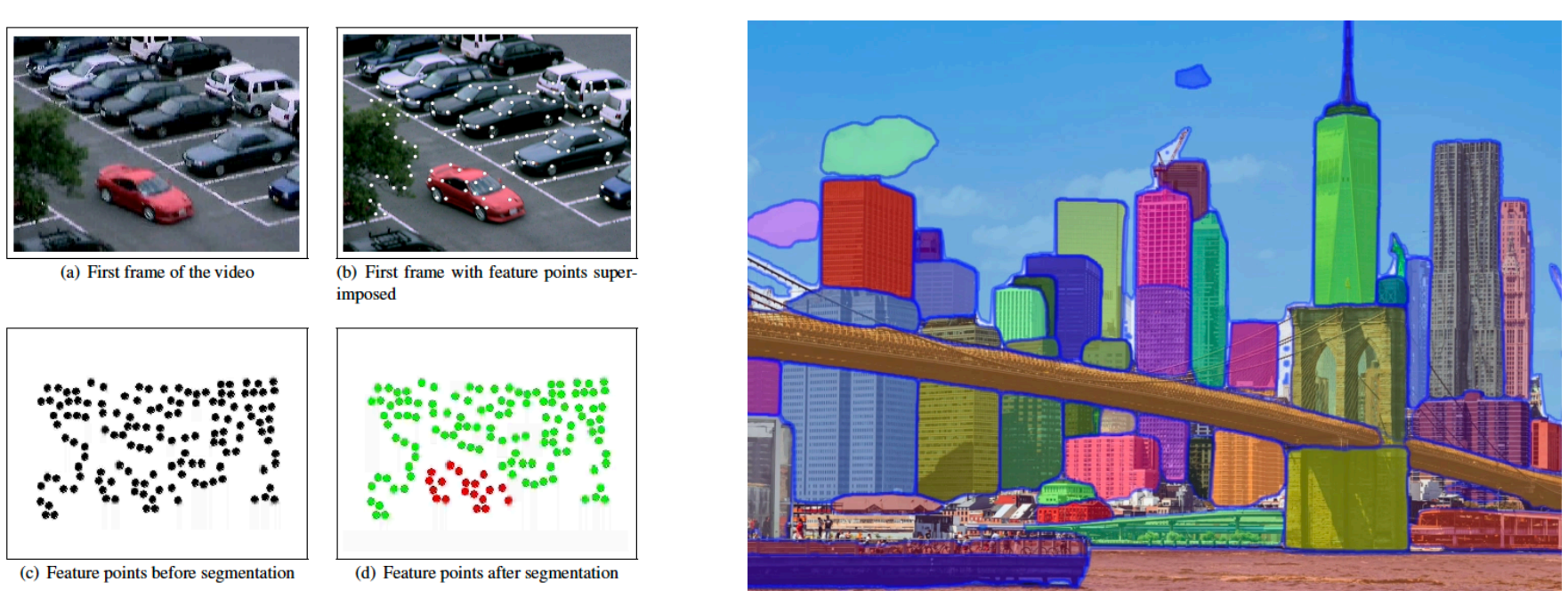

Figure 2.4: **Left:** features tracked on independently moving objects in a scene. **Right:** image patches associated with different regions of an image.

## 2.2 A Mixture of Complete Low-Dimensional Subspaces

As we have seen, low-rank signal models are rich enough to provide a full picture of the interplay between low-dimensionality in data and efficient and scalable computational algorithms for representation and recovery under errors. These models imply a *linear* and symmetric representation learning pipeline (2.1.6):

$$\boldsymbol{z} = \mathcal{E}(\boldsymbol{x}) = \boldsymbol{U}^\top \boldsymbol{x}, \quad \hat{\boldsymbol{x}} = \mathcal{D}(\boldsymbol{z}) = \boldsymbol{U}\boldsymbol{z},$$

which can be provably learned from finite samples of $\boldsymbol{x}$ with principal component analysis (solved efficiently, say, with the power method) whenever the distribution of $\boldsymbol{x}$ truly is linear. This is a restrictive assumption—for as Harold Hotelling, the distinguished 20th century statistician,[5] objected following George Dantzig's presentation of his theory of linear programming for the first time [Dan02],

*"...we all know the world is nonlinear."*

Even accounting for its elegance and simplicity, the low-rank assumption is too restrictive to be broadly applicable to modeling real-world data. A key limitation is the assumption of a *single* linear subspace that is responsible for generating the structured observations. In many practical applications, structure generated by a *mixture* of distinct low-dimensional subspaces provides a more realistic model. For example, consider a video sequence that captures the motion of several distinct objects, each subject to its own independent displacement (Figure 2.4 left). Under suitable assumptions on the individual motions, each object becomes responsible for an independent low-dimensional subspace in the concatenated sequence of video frames [VM04]. As another example, consider modeling natural images via learning a model for the distribution of

[5] By coincidence, also famous for his contributions to the development and naming of Principal Component Analysis [Hot33].

*patches*, spatially-contiguous collections of pixels, within an image (Figure 2.4 right). Unlike in the simplistic setting of Figure 2.2, where images of faces with matched poses can be well-approximated by a single low-dimensional subspace, the patch at a specific location in a natural image can correspond to objects with very different properties—for example, distinct color or shape due to occlusion boundaries. Therefore, modeling the distribution of patches with a single subspace is futile, but a *mixture* of subspaces, one for each region, performs surprisingly well in practice, say for segmenting or compressing purposes [MRY+11].[6] We will see a concrete example in a later chapter (Example 4.10).

In this section, we will begin by discussing the conceptual and algorithmic foundations for structured representation learning when the data distribution can be modeled by *a mixture of low-dimensional subspaces*, as illustrated in Figure 2.1. In this setting, the decoder mapping will be almost as simple as the case of a single subspace: we simply represent via

$$\hat{\boldsymbol{x}} = \mathcal{D}(\boldsymbol{z}) = \left( \sum_{k=1}^{K} \pi_k(\boldsymbol{z}) \boldsymbol{U}_k \right) \boldsymbol{z}, \tag{2.2.1}$$

where $\pi_k : \mathbb{R}^d \to \{0, 1\}$ are a set of *sparse* weighting random variables, such that only a single subspace $\mathcal{S}_k = \mathrm{col}(\boldsymbol{U}_k)$ is selected in the sum. However, the task of encoding such data $\boldsymbol{x}$ to suitable representations $\boldsymbol{z}$, and learning such an encoder-decoder pair from data, will prove to be more involved.

We will see how ideas from the rich literature on *sparse representation* and *independent component analysis* lead to a natural reformulation of the above decoder architecture through the lens of sparsity, the corresponding encoder architecture (obtained through a power-method-like algorithm analogous to that of principal component analysis), and strong guarantees of correctness and efficiency for learning such encoder-decoder pairs from data. In this sense, the case of mixed linear low-dimensional structure already leads to many of the key conceptual components of structured representation learning that we will develop in far greater generality in this book.

### 2.2.1 Mixtures of Subspaces and Sparsely-Used Dictionaries

Let $\boldsymbol{U}_1, \dots, \boldsymbol{U}_K$, each of size $D \times d$, denote a collection of orthonormal bases for $K$ subspaces of dimension $d$ in $\mathbb{R}^D$. To say that $\boldsymbol{x}$ follows a mixture-of-subspaces distribution parameterized by $\boldsymbol{U}_1, \dots, \boldsymbol{U}_K$ means, geometrically speaking, that

$$\boldsymbol{x} = \boldsymbol{U}_k \boldsymbol{z} \quad \text{for some } \ k \in [K], \ \ \boldsymbol{z} \in \mathbb{R}^d. \tag{2.2.2}$$

The statistical analogue of this geometric model, as we saw for the case of PCA and linear structure, is that $\boldsymbol{x}$ follows a *mixture of Gaussians* distribution: that is,

$$\boldsymbol{x} \sim \sum_{k=1}^{K} \pi_k \mathcal{N}(\boldsymbol{0}, \boldsymbol{U}_k \boldsymbol{U}_k^\top), \quad \text{for some } \pi_k \geq 0, \ \sum_{k=1}^{K} \pi_k = 1. \tag{2.2.3}$$

[6] We will return to this observation in Chapter 5, where we will show it can be significantly generalized to learn representations for large-scale modern datasets.

In other words, for each $k \in [K]$, $\boldsymbol{x}$ is Gaussian on the low-dimensional subspace $\mathrm{col}(\boldsymbol{U}_k)$ with probability $\pi_k$.

*Remark* 2.7 (A Mixture of Gaussians v.s. A Superposition of Gaussians). One should be aware that the above model (2.2.3) is a *mixture* of Gaussian distributions, not to be confused with a mixture of Gaussian variables by superposition, say

$$\boldsymbol{x} = \sum_{i=1}^{n} w_i \boldsymbol{x}_i, \quad \boldsymbol{x}_i \sim \mathcal{N}(\boldsymbol{0}, \boldsymbol{U}_i \boldsymbol{U}_i^\top), \tag{2.2.4}$$

where $\boldsymbol{x}_i$ are independent random Gaussian vectors and $w_i$ are a set of fixed weights. As we know from the properties of Gaussian vectors, such a superposition $\boldsymbol{x}$ will remain a Gaussian distribution.

For now, we focus on the geometric perspective offered by (2.2.2). There is an algebraically convenient alternative to this conditional representation. Consider a *lifted* representation vector $\boldsymbol{z} = [\boldsymbol{z}_1^\top, \ldots, \boldsymbol{z}_K^\top]^\top \in \mathbb{R}^{dK}$, such that $\boldsymbol{z}$ is *$d$-sparse* with support on one of the $K$ consecutive non-overlapping blocks of $d$ coordinates out of $dK$. Then (2.2.2) can be written equivalently as

$$\boldsymbol{x} = \underbrace{\begin{bmatrix} | & \ldots & | \\ \boldsymbol{U}_1 & \ldots & \boldsymbol{U}_K \\ | & \ldots & | \end{bmatrix}}_{\boldsymbol{U}} \underbrace{\begin{bmatrix} \boldsymbol{z}_1 \\ \vdots \\ \boldsymbol{z}_K \end{bmatrix}}_{\boldsymbol{z}}, \quad \left\| \begin{bmatrix} \|\boldsymbol{z}_1\|_2 \\ \vdots \\ \|\boldsymbol{z}_K\|_2 \end{bmatrix} \right\|_0 = 1. \tag{2.2.5}$$

Here, the $\ell^0$ "norm" $\|\cdot\|_0$ measures sparsity by counting the number of nonzero entries:

$$\|\boldsymbol{z}\|_0 = |\{i \mid z_i \neq 0\}|, \tag{2.2.6}$$

and the matrix $\boldsymbol{U} \in \mathbb{R}^{D \times Kd}$ is called a *dictionary* with all the $\{\boldsymbol{U}_i\}_{i=1}^K$ as code words. In general, if the number of subspaces in the mixture $K$ is large enough, there is no bound on the number of columns contained in the dictionary $\boldsymbol{U}$. In the case where $Kd < D$, $\boldsymbol{U}$ is called *undercomplete*; when $Kd = D$, it is called *complete*; and when $Kd > D$, it is called *overcomplete*.

Now, (2.2.5) suggests a convenient relaxation for tractability of analysis: rather than modeling $\boldsymbol{x}$ as coming from a mixture of $K$ *specific* subspaces $\boldsymbol{U}_1, \ldots, \boldsymbol{U}_K$, we may instead start with a dictionary $\boldsymbol{U} \in \mathbb{R}^{D \times m}$, where $m$ may be smaller or larger than $D$, and simply seek to represent $\boldsymbol{x} = \boldsymbol{U}\boldsymbol{z}$ with the sparsity $\|\boldsymbol{z}\|_0$ sufficiently small. This leads to the *sparse dictionary model* for $\boldsymbol{x}$:

$$\boldsymbol{x} = \boldsymbol{U}\boldsymbol{z} + \boldsymbol{\varepsilon}, \quad \|\boldsymbol{z}\|_0 \ll d, \tag{2.2.7}$$

where $\boldsymbol{\varepsilon}$ represents an unknown noise vector. Geometrically, this implies that $\boldsymbol{x}$ lies close to the span of a subset of $\|\boldsymbol{z}\|_0$ columns of $\boldsymbol{U}$, making this an instantiation of the mixture-of-subspaces model (2.2.2) with a very large value of $K$, and specific correlations between the subspaces $\boldsymbol{U}_k$.

**Orthogonal dictionary for sparse coding.** Now we can formulate the structured representation learning problem for mixtures of low-dimensional subspaces that we will study in this section. We assume that we have samples $\boldsymbol{X} = [\boldsymbol{x}_1, \ldots, \boldsymbol{x}_N]$ from an unknown sparse dictionary model (2.2.7), possibly with added noises $\boldsymbol{\varepsilon}_i$. Let us begin from the assumption that the dictionary $\boldsymbol{U}$ in the sparse dictionary model (2.2.7) is complete and orthogonal,[7] and that each coefficient vector $\boldsymbol{z}$ is $d$-sparse, with $d \ll D$. In this setting, representation learning amounts to correctly learning the orthogonal dictionary $\boldsymbol{U}$ via optimization: we can then take $\mathcal{E}(\boldsymbol{x}) = \boldsymbol{U}^\top \boldsymbol{x}$ as the encoder and $\mathcal{D}(\boldsymbol{z}) = \boldsymbol{U}\boldsymbol{z}$ as the decoder, and $\mathcal{D} = \mathcal{E}^{-1}$. In diagram form:

$$\boldsymbol{x} \xrightarrow{\mathcal{E}=\boldsymbol{U}^\top} \boldsymbol{z} \xrightarrow{\mathcal{D}=\boldsymbol{U}} \hat{\boldsymbol{x}}. \tag{2.2.8}$$

We see that the autoencoding pair $(\mathcal{E}, \mathcal{D})$ for complete dictionary learning is symmetric, as in the case of a single linear subspace, making the computational task of encoding and decoding no more difficult than in the linear case. On the other hand, the task of learning the dictionary $\boldsymbol{U}$ is strictly more difficult than learning a single linear subspace by PCA. To see why we cannot simply use PCA to learn the orthogonal dictionary $\boldsymbol{U}$ correctly, note that the loss function that gave rise to PCA, namely (2.1.5), is completely invariant to rotations of the rows of the matrix $\boldsymbol{U}$: that is, if $\boldsymbol{Q}$ is any $d \times d$ orthogonal matrix, then $\boldsymbol{U}$ and $\boldsymbol{U}\boldsymbol{Q}$ are both feasible and have an identical loss for (2.1.5). The sparse dictionary model is decidedly not invariant to such transformations: if we replaced $\boldsymbol{U}$ by $\boldsymbol{U}\boldsymbol{Q}$ and made a corresponding rotation $\boldsymbol{Q}^\top \boldsymbol{z}$ of the representation coefficients $\boldsymbol{z}$, we would in general destroy the sparsity structure of $\boldsymbol{z}$, violating the modeling assumption. Thus, we need new algorithms for learning orthogonal dictionaries.

### 2.2.2 Complete Dictionary Learning

In this section, we will derive algorithms for solving the orthogonal dictionary learning problem. To be more precise, we assume that the observed vector $\boldsymbol{x} \in \mathbb{R}^D$ follows a statistical model

$$\boldsymbol{x} = \boldsymbol{U}\boldsymbol{z} + \boldsymbol{\varepsilon}, \tag{2.2.9}$$

where $\boldsymbol{U} \in \mathbb{R}^{D \times D}$ is an unknown orthogonal dictionary, $\boldsymbol{z}$ is a random vector with statistically independent components $z_i$, each with zero mean, and $\boldsymbol{\varepsilon} \in \mathbb{R}^D$ is an independent random vector of small (Gaussian) noises. The goal is to recover $\boldsymbol{U}$ (and hence $\boldsymbol{z}$) from samples of $\boldsymbol{x}$.

Here we assume that each independent component $z_i$ is distributed as

$$z_i \sim \mathrm{Ber}(\theta) \cdot \mathcal{N}(0, 1/\theta).$$

That is, it is the product of a Bernoulli random variable with probability $\theta$ of being 1 and $1 - \theta$ of being 0, and an independent Gaussian random variable with variance $1/\theta$. This distribution is formally known as the *Bernoulli-Gaussian* distribution. The normalization is chosen so that $\mathrm{Var}(z_i) = 1$ and hence $\mathbb{E}[\|\boldsymbol{z}\|_2^2] = d$. This modeling assumption implies that the vector of independent components $\boldsymbol{z}$ is typically very sparse: we calculate $\mathbb{E}\left[\|\boldsymbol{z}\|_0\right] = d\theta$, which is small when $\theta$ is inversely proportional to $d$.

[7] It can be shown that for the complete case, we do not lose any generality by making the orthogonal assumption (Exercise 2.4).

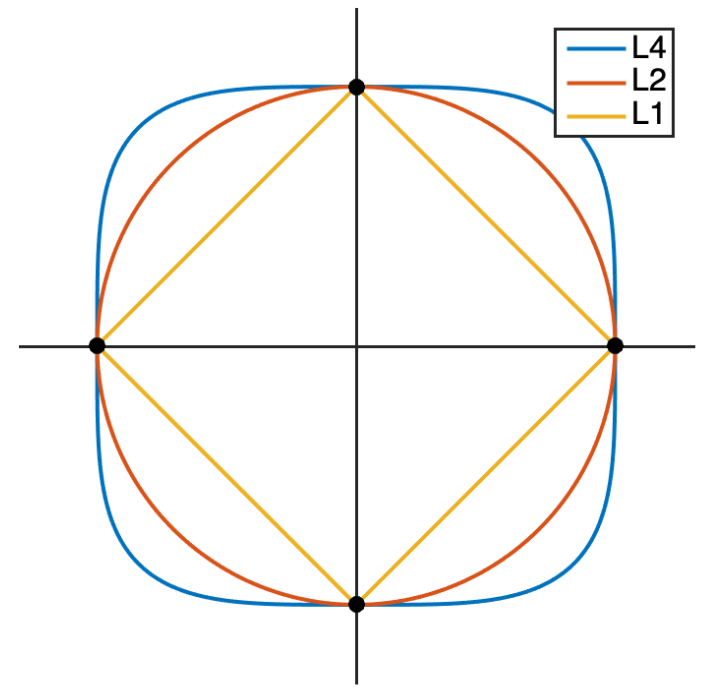

Figure 2.5: Maximizing $\ell^4$ norm or minimizing $\ell^1$ norm promotes sparsity (for vectors on the sphere).

*Remark* 2.8 (The Orthogonal Assumption). At first sight, the assumption that the dictionary $\boldsymbol{U}$ is orthogonal might seem to be somewhat restrictive. But there is actually no loss of generality. One may consider a complete dictionary to be any square invertible matrix $\boldsymbol{U}$. With samples generated from this dictionary: $\boldsymbol{X} = \boldsymbol{U}\boldsymbol{Z} \in \mathbb{R}^{D\times N}$, it is easy to show that with some preconditioning of the data matrix $\boldsymbol{X}$:

$$\bar{\boldsymbol{X}} = \left(\frac{1}{N\theta}\boldsymbol{X}\boldsymbol{X}^\top\right)^{-\frac{1}{2}}\boldsymbol{X}, \tag{2.2.10}$$

then there exists an orthogonal matrix $\boldsymbol{U}_o \in \mathsf{O}(D)$ such that

$$\bar{\boldsymbol{X}} = \boldsymbol{U}_o\boldsymbol{Z}. \tag{2.2.11}$$

See Exercise 2.4 or [SQW17b] for more details.

**Dictionary learning via the MSP algorithm.** Now suppose that we are given a set of observations:

$$\boldsymbol{x}_i = \boldsymbol{U}\boldsymbol{z}_i + \boldsymbol{\varepsilon}_i,\ \forall i \in [N]. \tag{2.2.12}$$

Let $\boldsymbol{X} = [\boldsymbol{x}_1, \ldots, \boldsymbol{x}_N]$ and $\boldsymbol{Z} = [\boldsymbol{z}_1, \ldots, \boldsymbol{z}_N]$. The goal is to recover $\boldsymbol{U}$ from the data $\boldsymbol{X}$. Therefore, given any orthogonal matrix $\boldsymbol{A} \in \mathsf{O}(D)$,

$$\boldsymbol{A}\boldsymbol{x}_i = \boldsymbol{A}\boldsymbol{U}\boldsymbol{z}_i + \boldsymbol{A}\boldsymbol{\varepsilon}_i \tag{2.2.13}$$

would be nearly sparse if $\boldsymbol{A} = \boldsymbol{U}^\top$ (as by assumption the noise $\boldsymbol{\varepsilon}_i$ is of small magnitude).

Also, given $\boldsymbol{U}$ is orthogonal and the fact $\boldsymbol{\varepsilon}$ is small, the vector $\boldsymbol{x}$ has a predictable expected norm, i.e., $\mathbb{E}[\|\boldsymbol{x}\|_2^2] \approx \mathbb{E}[\|\boldsymbol{z}\|_2^2] = d$. It is a known fact that for vectors on a sphere, maximizing the $\ell^4$ norm is equivalent to minimizing the $\ell^0$ norm (for promoting sparsity),

$$\arg\max_{\boldsymbol{z}\in\mathbb{S}^n} \|\boldsymbol{z}\|_4 \quad \Leftrightarrow \quad \arg\min_{\boldsymbol{z}\in\mathbb{S}^n} \|\boldsymbol{z}\|_0. \tag{2.2.14}$$

This is illustrated in Figure 2.5.

An orthogonal matrix $\boldsymbol{A}$ preserves the Euclidean ($\ell^2$) norm: $\|\boldsymbol{A}\boldsymbol{x}\|_2^2 = \|\boldsymbol{x}\|_2^2$. Therefore, to find the correct orthogonal dictionary $\boldsymbol{U}$ from $\boldsymbol{X}$, we may try to solve the following optimization program:

$$\max_{\tilde{\boldsymbol{A}}\in\mathsf{O}(D)} \frac{1}{4}\left\|\tilde{\boldsymbol{A}}\boldsymbol{X}\right\|_4^4 = \frac{1}{4}\sum_{i=1}^{N}\left\|\tilde{\boldsymbol{A}}\boldsymbol{x}_i\right\|_4^4 \tag{2.2.15}$$

This is known as the $\ell^4$ maximization problem [ZMZ+20]. After we find the solution $\boldsymbol{A}^\star$, we can take the transpose $\boldsymbol{U}^\star = (\boldsymbol{A}^\star)^\top$.

*Remark* 2.9. It is also known that for vectors on a sphere, minimizing the $\ell^1$ norm is equivalent to minimizing the $\ell^0$ norm (for promoting sparsity),

$$\arg\min_{\boldsymbol{z}\in\mathbb{S}^n} \|\boldsymbol{z}\|_1 \quad \Leftrightarrow \quad \arg\min_{\boldsymbol{z}\in\mathbb{S}^n} \|\boldsymbol{z}\|_0,$$

which is also illustrated in Figure 2.5. This fact can also be exploited to learn the dictionary $\boldsymbol{A}$ effectively and efficiently. This was actually explored earlier than the $\ell^4$ norm used here. Interested readers may refer to the work of [QZL+20b].

Note that the above problem is equivalent to the following constrained optimization problem:

$$\min -\frac{1}{4}\left\|\tilde{\boldsymbol{A}}\boldsymbol{X}\right\|_4^4 \quad \text{subject to} \quad \tilde{\boldsymbol{A}}^\top\tilde{\boldsymbol{A}} = \boldsymbol{I}. \tag{2.2.16}$$

As shown in [WM22], using the Lagrange multiplier method, one can derive that the optimal solution to the problem should satisfy the following "fixed point" condition:

$$\boldsymbol{A}^\star = \mathcal{P}_{\mathrm{O}(D)}[(\boldsymbol{A}^\star\boldsymbol{X})^{\odot 3}\boldsymbol{X}^\top], \tag{2.2.17}$$

where $\mathcal{P}_{\mathrm{O}(D)}[\,\cdot\,]$ is a projection onto the space of orthogonal matrices $\mathrm{O}(D)$.[8]

To compute the fixed point for the above equation, similar to how we computed eigenvectors for PCA (2.1.16), we may take the following power iteration:

$$\tilde{\boldsymbol{A}}_{t+1} = \mathcal{P}_{\mathrm{O}(D)}[(\tilde{\boldsymbol{A}}_t\boldsymbol{X})^{\odot 3}\boldsymbol{X}^\top]. \tag{2.2.18}$$

This is known as the *matching, stretching, and projection* (MSP) algorithm proposed by [ZMZ+20]. It was shown that under broad conditions such a greedy algorithm indeed converges to the correct solution at a superlinear rate.

[8] For any matrix $\boldsymbol{M}$ with SVD $\boldsymbol{M} = \boldsymbol{U}\boldsymbol{\Sigma}\boldsymbol{V}^\top$, $\mathcal{P}_{\mathrm{O}(D)}[\boldsymbol{M}] = \boldsymbol{U}\boldsymbol{V}^\top$. We leave this as an exercise for the reader.

*Remark* 2.10 (Global Optimality of $\ell^4$ Maximization). The constrained $\ell^4$ maximization problem is a nonconvex program. In general one should *not* expect that any greedy (say gradient-descent type) algorithms would converge to the globally optimal solution. Surprisingly, one can show that, unlike general nonconvex programs, the landscape of $\ell^4$ maximization over a sphere

$$\min -\frac{1}{4}\left\|\tilde{\boldsymbol{q}}^\top \boldsymbol{X}\right\|_4^4 \quad \text{subject to} \quad \tilde{\boldsymbol{q}}^\top \tilde{\boldsymbol{q}} = 1. \tag{2.2.19}$$

is very benign: All local minima are close to the global optima and all critical points are saddle points with a direction of negative curvature. Hence, any descent method with the ability of escaping strict saddle points provably finds global optimal solutions. For more precise statements, interested readers may refer to [QZL+20a].

*Remark* 2.11 (Stable Deep Linear Network). The above iterative process of computing the dictionary has a natural incremental "deep learning" interpretation. Let us define $\delta \boldsymbol{A}_{t+1} = \tilde{\boldsymbol{A}}_{t+1}\tilde{\boldsymbol{A}}_t^\top$ and $\boldsymbol{Z}_t = \tilde{\boldsymbol{A}}_t \boldsymbol{X}$, then it is easy to show that

$$\delta \boldsymbol{A}_{t+1} = \mathcal{P}_{\mathrm{O}(D)}[(\boldsymbol{Z}_t)^{\odot 3}\boldsymbol{Z}_t^\top].$$

If $\tilde{\boldsymbol{A}}_t$ converges to the correct dictionary $\boldsymbol{U}$, then the above iterative encoding process is essentially equivalent to a "deep linear network":

$$\boldsymbol{Z} \longleftarrow \boldsymbol{Z}_{t+1} = \underbrace{\delta \boldsymbol{A}_{t+1}\delta \boldsymbol{A}_t \ldots \delta \boldsymbol{A}_1}_{\text{forward constructed layers}} \boldsymbol{X}.$$

Note that the computation of the increment transforms $\delta \boldsymbol{A}_{t+1}$ at each layer depends only on the feature output from the previous layer $\boldsymbol{Z}_t$. The network is naturally stable as each layer is a norm-preserving orthogonal transform. Despite its resemblance to a linear deep network, backpropagation is unnecessary to learn each layer. All layers are learned in one forward pass!

### 2.2.3 Connection to ICA and Kurtosis

With the Bernoulli-Gaussian model, the variables $z_i$ are independent and non-Gaussian. Then, there is a clear correspondence between the dictionary learning and the classic independent component analysis (ICA), to the extent that algorithms to solve one problem can be used to solve the other.[9]

Towards deriving an algorithm based on ICA, we focus on an objective function known as *kurtosis*, which is used in ICA as a direct consequence of the non-Gaussianity of the components. The *kurtosis*, or fourth-order cumulant, of a zero-mean random variable $X$ is defined as

$$\mathrm{kurt}(X) = \mathbb{E}\, X^4 - 3(\mathbb{E}\, X^2)^2. \tag{2.2.20}$$

[9] We explore this issue in more depth in Exercise 2.3, where a connection between non-Gaussianity of the independent components and the purely geometric notion of symmetry is made. This issue is related to our observation above that PCA does not work for recovering sparsely-used orthogonal dictionaries: in the statistical setting, it can be related to rotational invariance of the Gaussian distribution (Exercise 2.2).

If we have only finite samples from the random variable $X$ arranged into a vector $\boldsymbol{x} = [x_1, \ldots, x_N]$, we define kurtosis through their empirical average, which yields

$$\mathrm{kurt}(\boldsymbol{x}) = \frac{1}{N}\|\boldsymbol{x}\|_4^4 - \frac{3}{N^2}\|\boldsymbol{x}\|_2^4. \tag{2.2.21}$$

Finally, for random vectors, we define their kurtosis as the sum of each component's scalar kurtosis. Kurtosis is a natural loss function for ICA because for Gaussian $X$, kurtosis is zero; the reader can verify further that the Bernoulli-Gaussian distribution has positive kurtosis. Thus a natural procedure for seeking non-Gaussian independent components is to search for a set of mutually-orthogonal directions $\tilde{\boldsymbol{U}} \in \mathbb{R}^{d\times k}$, such that $\tilde{\boldsymbol{U}}^\top \boldsymbol{X}$ has maximal kurtosis, where $\boldsymbol{X} = \boldsymbol{U}\boldsymbol{Z} \in \mathbb{R}^{D\times N}$ is the Bernoulli-Gaussian ICA data matrix (such that $\tilde{\boldsymbol{U}}$ approximates $\boldsymbol{U}$ up to permutations of the columns). Formally, we seek to solve the problem

$$\max_{\tilde{\boldsymbol{U}}^\top \tilde{\boldsymbol{U}} = \boldsymbol{I}} \mathrm{kurt}(\tilde{\boldsymbol{U}}^\top \boldsymbol{X}). \tag{2.2.22}$$

At one extreme, we may set $k = D$ and seek to recover the entire dictionary $\boldsymbol{U}$ in a single shot. It can be shown that this problem can be solved with the MSP algorithm we have seen previously. At the other extreme, we may set $k = 1$ and seek to recover a single direction (column of $\boldsymbol{U}$) at a time, performing *deflation*, i.e., replacing the data matrix $\boldsymbol{X}$ by $(\boldsymbol{I} - \boldsymbol{u}\boldsymbol{u}^\top)\boldsymbol{X}$, after each step before finding another direction. There is a natural tradeoff between the scalability of the $k = 1$ incremental approach and the efficiency and robustness of the $k = D$ approach.

**Incremental ICA: correctness and FastICA algorithm.** The FastICA algorithm, advanced by Hyvärinen and Oja [HO97], is a fast fixed-point algorithm for solving the $k = 1$ kurtosis maximization scheme for ICA. The problem at hand is

$$\max_{\|\tilde{\boldsymbol{u}}\|_2^2 = 1} \mathrm{kurt}(\boldsymbol{X}^\top \tilde{\boldsymbol{u}}). \tag{2.2.23}$$

First, we will perform some very basic analysis of this objective to verify its correctness. Notice by the change of variables $\tilde{\boldsymbol{w}} = \boldsymbol{U}^\top \tilde{\boldsymbol{u}}$ that this problem is equivalent to

$$\max_{\|\tilde{\boldsymbol{w}}\|_2^2 = 1} \mathrm{kurt}(\boldsymbol{Z}^\top \tilde{\boldsymbol{w}}).$$

This objective is simple enough that we can make strong statements about its correctness as a formulation for recovering the dictionary $\boldsymbol{U}$. For example, in the population setting where $N \to \infty$, we may use additivity properties of the kurtosis (Exercise 2.5) and our assumed normalization on the independent components to write the previous problem equivalently as

$$\max_{\|\tilde{\boldsymbol{w}}\|_2^2 = 1} \sum_{i=1}^{d} \mathrm{kurt}(z_i)\tilde{w}_i^4. \tag{2.2.24}$$

It can be shown that under the Bernoulli-Gaussian assumption, the optimization landscape of this problem is "benign" (Exercise 2.7)—meaning that all local maxima of the objective function correspond to the recovery of one of the independent components.

In practice, we cannot directly compute the change of variables sending $\tilde{\boldsymbol{u}}$ to $\tilde{\boldsymbol{w}}$ because it requires the ground-truth dictionary $\boldsymbol{U}$, so we develop algorithms directly for the $\tilde{\boldsymbol{u}}$ parameterization and the data matrix $\boldsymbol{X}$. Then one efficient and scalable way to compute one of these maxima is via first-order optimization algorithms, which iteratively follow the gradient of the objective function and project onto the constraint set $\{\tilde{\boldsymbol{u}} \mid \|\tilde{\boldsymbol{u}}\|_2^2 = 1\}$. Since we have assumed that each $z_i$ satisfies $\mathrm{Var}(z_i) = 1$, we have for any $\tilde{\boldsymbol{u}}$, for large enough $N$

$$\mathrm{kurt}(\boldsymbol{X}^\top \tilde{\boldsymbol{u}}) \approx \tfrac{1}{N}\|\boldsymbol{X}^\top \tilde{\boldsymbol{u}}\|_4^4 - 3\|\tilde{\boldsymbol{u}}\|_2^4. \tag{2.2.25}$$

We can then derive a corresponding approximation to the gradient:

$$\nabla_{\tilde{\boldsymbol{u}}}\, \mathrm{kurt}(\boldsymbol{X}^\top \tilde{\boldsymbol{u}}) \approx \tfrac{4}{N}\boldsymbol{X}(\boldsymbol{X}^\top \tilde{\boldsymbol{u}})^{\odot 3} - 12\|\tilde{\boldsymbol{u}}\|_2^2 \tilde{\boldsymbol{u}}.$$

The FastICA algorithm uses a fixed-point method to compute directions of maximum kurtosis. It starts from the first-order optimality conditions for the kurtosis maximization problem, given the preceding gradient approximation and the constraint set, which read

$$\boldsymbol{X}(\boldsymbol{X}^\top \tilde{\boldsymbol{u}})^{\odot 3} = \underbrace{\left\langle \tilde{\boldsymbol{u}}, \boldsymbol{X}(\boldsymbol{X}^\top \tilde{\boldsymbol{u}})^{\odot 3} \right\rangle}_{\lambda} \tilde{\boldsymbol{u}}, \tag{2.2.26}$$

where the specific value of $\lambda$ is determined using the unit norm constraint on $\tilde{\boldsymbol{u}}$. Exercise 2.6 describes the mathematical background necessary to derive these optimality conditions from first principles. Equation (2.2.26) is satisfied by *any* critical point of the kurtosis maximization problem; we want to derive an equation satisfied by only the maximizers. After noticing that $\lambda = \|\boldsymbol{X}^\top \tilde{\boldsymbol{u}}\|_4^4$, we equivalently re-express (2.2.26) as the modified equation

$$\frac{1}{N}\boldsymbol{X}(\boldsymbol{X}^\top \tilde{\boldsymbol{u}})^{\odot 3} - 3\tilde{\boldsymbol{u}} = \left(\frac{\lambda}{N} - 3\right)\tilde{\boldsymbol{u}}, \tag{2.2.27}$$

and realize that any maximizer of (2.2.23) must satisfy $\lambda/N - 3 > 0$, assuming that $N$ is sufficiently large. Hence, we may *normalize* both sides of (2.2.27), giving the following fixed-point equation satisfied by every maximizer of (2.2.23):

$$\frac{\frac{1}{N}\boldsymbol{X}(\boldsymbol{X}^\top \tilde{\boldsymbol{u}})^{\odot 3} - 3\tilde{\boldsymbol{u}}}{\left\|\frac{1}{N}\boldsymbol{X}(\boldsymbol{X}^\top \tilde{\boldsymbol{u}})^{\odot 3} - 3\tilde{\boldsymbol{u}}\right\|_2} = \tilde{\boldsymbol{u}}. \tag{2.2.28}$$

Iterating the mapping defined by the lefthand side of this fixed point expression then yields the FastICA algorithm of Hyvärinen and Oja [HO97]:

$$\tilde{\boldsymbol{v}}^+ = \tfrac{1}{N}\boldsymbol{X}(\boldsymbol{X}^\top \tilde{\boldsymbol{u}})^{\odot 3} - 3\tilde{\boldsymbol{u}}, \tag{2.2.29}$$

$$\tilde{\boldsymbol{u}}^+ = \tilde{\boldsymbol{v}}^+ / \left\|\tilde{\boldsymbol{v}}^+\right\|_2. \tag{2.2.30}$$

It turns out that the FastICA algorithm converges extremely rapidly (actually at a *cubic* rate) to columns of the dictionary $\boldsymbol{U}$; interested readers may consult [HO97] for details. Comparing the FastICA algorithm (2.2.29) to the power method studied in Section 2.1.1 for the PCA problem and the MSP algorithm (2.2.18), we notice a striking similarity. Indeed, FastICA is essentially a modified power method, involving the gradient of the empirical kurtosis rather than the simpler linear gradient of the PCA objective.

## 2.3 A Mixture of Overcomplete Low-Dimensional Subspaces

As we have seen, complete dictionary learning enjoys an elegant computational theory in which we maintain a symmetric autoencoding structure $\mathcal{E}(\boldsymbol{x}) = \boldsymbol{U}^\top \boldsymbol{x}$, $\mathcal{D}(\boldsymbol{z}) = \boldsymbol{U}\boldsymbol{z}$, with a scalable power-method-like algorithm (the MSP algorithm) for learning an orthogonal dictionary/codebook $\boldsymbol{U}$ from data. Nevertheless, for learning representations of general high-dimensional data distributions, one must expand the size of the codebook beyond the orthogonality requirement—meaning that we must have $\boldsymbol{A} \in \mathbb{R}^{D \times m}$, with $m \gg D$, corresponding to the case of an *overcomplete* dictionary/codebook,[10] and the signal model

$$\boldsymbol{x} = \boldsymbol{A}\boldsymbol{z} + \boldsymbol{\varepsilon}, \quad \|\boldsymbol{z}\|_0 = d \ll m. \tag{2.3.1}$$

There are both geometric and physical/modeling motivations for passing to the overcomplete case. Geometrically, recall that in our original reduction from the mixture of subspace data model to the sparse dictionary model, a mixture of $K$ subspaces in $\mathbb{R}^D$, each of dimension $d$, led to a dictionary of shape $\boldsymbol{A} \in \mathbb{R}^{D \times Kd}$. In other words, overcomplete dictionaries correspond to *richer* mixtures of subspaces, with more distinct modes of variability for modeling the high-dimensional data distribution. On the modeling side, we may run a computational experiment on real-world data that reveals the additional modeling power conferred by an overcomplete representation.

*Example* 2.2. Given sampled images of hand-written digits, Figure 2.6(a) shows the result of fitting an orthogonal dictionary to the dataset. In contrast, Figure 2.6(b) shows the result of running an optimization algorithm for learning overcomplete dictionaries (which we will present in detail later in the Chapter) on these samples. Notice that the representations become far sparser and the codebooks far more interpretable—they consist of fundamental primitives for the strokes composing the digits, i.e. oriented edges. ■

In fact, overcomplete dictionary learning was originally proposed as a biologically plausible algorithm for image representation based on empirical evidence of how early stages of the visual cortex represent visual stimuli [OF97; OF96].

In the remainder of this section, we will overview the conceptual and computational foundations of overcomplete dictionary learning. Supposing that the model (2.3.1) is satisfied with sparse codes $\boldsymbol{z}$, overcomplete dictionary $\boldsymbol{A}$, and sparsity level $d$, and given samples $\boldsymbol{X} = [\boldsymbol{x}_1, \ldots, \boldsymbol{x}_N]$ of $\boldsymbol{x}$, we want to learn an encoder $\mathcal{E} : \mathbb{R}^D \to \mathbb{R}^m$ mapping each $\boldsymbol{x}$ to its sparse code $\boldsymbol{z}$, and a decoder $\mathcal{D}(\boldsymbol{z}) = \boldsymbol{A}\boldsymbol{z}$ reconstructing each $\boldsymbol{x}$ from its sparse code. In diagram form:

$$\boldsymbol{x} \xrightarrow{\quad \mathcal{E} \quad} \boldsymbol{z} \xrightarrow{\quad \mathcal{D}=\boldsymbol{A} \quad} \hat{\boldsymbol{x}}. \tag{2.3.2}$$

[10] We change the notation here from $\boldsymbol{U}$ to $\boldsymbol{A}$ in order to emphasize the non-orthogonality and non-square-shape of the overcomplete dictionary $\boldsymbol{A}$.

We will start from the construction of the encoder $\mathcal{E}$. We will work incrementally: first, *given the true dictionary* $\boldsymbol{A}$, we will show how the problem of *sparse coding* gives an elegant, scalable, and provably-correct algorithm for recovering the sparse code $\boldsymbol{z}$ of $\boldsymbol{x}$. Although this problem is NP-hard in the worst case, it can be solved efficiently and scalably for dictionaries $\boldsymbol{A}$ which are *incoherent*, i.e. having columns that are not too correlated. The encoder architecture encompassed by this solution will no longer be symmetric: we will see it has the form of a primitive deep network, which depends on the dictionary $\boldsymbol{A}$.

Then we will proceed to the task of learning the decoder $\mathcal{D}$, or equivalently the overcomplete dictionary $\boldsymbol{A}$. We will derive an algorithm for overcomplete dictionary learning that allows us to simultaneously learn the codebook $\boldsymbol{A}$ and the sparse codes $\boldsymbol{z}$, using ideas from sparse coding. Finally, we will discuss a more modern perspective on learnable sparse coding that leads us to a fully asymmetric encoder-decoder structure, as an alternative to (2.3.2). Here, the decoder will correspond to an incremental solution to the sparse dictionary learning problem, and yield a pair of deep network-like encoder decoders for sparse dictionary learning. This structure will foreshadow many developments to come in the remainder of the monograph, as we progress from analytical models to modern neural networks.

### 2.3.1 Sparse Coding with an Overcomplete Dictionary

In this section, we will consider the data model (2.3.1), which accommodates sparse linear combinations of many motifs, or *atoms*. Given data $\{\boldsymbol{x}_i\}_{i=1}^N \subseteq \mathbb{R}^D$ satisfying this model, i.e. expressible as

$$\boldsymbol{x}_i = \boldsymbol{A}\boldsymbol{z}_i + \boldsymbol{\varepsilon}_i, \qquad \forall i \in [N] \tag{2.3.3}$$

for some dictionary $\boldsymbol{A} \in \mathbb{R}^{D\times m}$ with $m$ atoms, sparse codes $\boldsymbol{z}_i$ such that $\|\boldsymbol{z}_i\|_0 \leq d$, and small errors $\boldsymbol{\varepsilon}_i$, the sparse coding problem is to recover the codes $\boldsymbol{z}_i$ as accurately as possible from the data $\boldsymbol{x}_i$, given the dictionary $\boldsymbol{A}$. Efficient algorithms to solve this problem succeed when the dictionary $\boldsymbol{A}$ is *incoherent* in the sense that the inner products $\boldsymbol{a}_i^\top \boldsymbol{a}_j$ are uniformly small, hence the atoms are nearly orthogonal.[11]

Note that we can collect the $\boldsymbol{x}_i$ into $\boldsymbol{X} = \begin{bmatrix}\boldsymbol{x}_1, \ldots, \boldsymbol{x}_N\end{bmatrix} \in \mathbb{R}^{D\times N}$, collect the $\boldsymbol{z}_i$ into $\boldsymbol{Z} = \begin{bmatrix}\boldsymbol{z}_1, \ldots, \boldsymbol{z}_N\end{bmatrix} \in \mathbb{R}^{d\times N}$, and collect the $\boldsymbol{\varepsilon}_i$ into $\boldsymbol{E} = \begin{bmatrix}\boldsymbol{\varepsilon}_1, \ldots, \boldsymbol{\varepsilon}_N\end{bmatrix} \in \mathbb{R}^{D\times N}$, to rewrite (2.3.3) as

$$\boldsymbol{X} = \boldsymbol{A}\boldsymbol{Z} + \boldsymbol{E}. \tag{2.3.4}$$

[11] As it turns out, in a high-dimensional space, it is rather easy to pack a number of nearly orthogonal vectors that is much larger than the ambient dimension [WM22].

A natural approach to solving the sparse coding problem is to seek the sparsest signals that are consistent with our observations, and this naturally leads to the following optimization problem:

$$\min_{\tilde{\boldsymbol{Z}} \in \mathbb{R}^{d \times N}} \left\{ \|\boldsymbol{X} - \boldsymbol{A}\tilde{\boldsymbol{Z}}\|_F^2 + \lambda \|\tilde{\boldsymbol{Z}}\|_1 \right\}, \tag{2.3.5}$$

where the $\ell^1$ norm $\|\tilde{\boldsymbol{Z}}\|_1$, taken elementwise, is known to promote sparsity of the solution [WM22]. This optimization problem is known as the LASSO problem.

However, unlike PCA or the complete dictionary learning problem, there is no clear power iteration-type algorithm to compute the optimal $\boldsymbol{Z}^\star$. A natural alternative is to consider solving the above optimization problem with gradient descent type algorithms. Let $\mathcal{L}(\tilde{\boldsymbol{Z}}) = \|\boldsymbol{X} - \boldsymbol{A}\tilde{\boldsymbol{Z}}\|_2^2 + \lambda\|\tilde{\boldsymbol{Z}}\|_1$. Conceptually, we could try to compute $\boldsymbol{Z}^\star$ with the following iterations:

$$\tilde{\boldsymbol{Z}}_{t+1} \leftarrow \tilde{\boldsymbol{Z}}_t + \eta \nabla \mathcal{L}(\tilde{\boldsymbol{Z}}_t). \tag{2.3.6}$$

However, because the term associated with the $\ell^1$ norm $\|\tilde{\boldsymbol{Z}}\|_1$ is non-smooth, we cannot just run gradient descent. For this type of function, we need to replace the gradient $\nabla \mathcal{L}(\tilde{\boldsymbol{Z}})$ with something that generalizes the notion of gradient, known as the subgradient $\partial \mathcal{L}(\tilde{\boldsymbol{Z}})$:

$$\tilde{\boldsymbol{Z}}_{t+1} \leftarrow \tilde{\boldsymbol{Z}}_t + \eta \partial \mathcal{L}(\tilde{\boldsymbol{Z}}_t). \tag{2.3.7}$$

However, it is known that the convergence of subgradient descent is usually very slow. Hence, for this type of optimization problems, it is conventional to adopt a so-called *proximal gradient descent*-type algorithm. We give a technical overview of this method in Section A.1.3.

Applying proximal gradient to the LASSO objective function (2.3.5) leads to the classic *iterative shrinkage-thresholding algorithm* (ISTA), which implements the iteration

$$\begin{aligned} \tilde{\boldsymbol{Z}}_1 &\sim \mathcal{N}(\boldsymbol{0}, \boldsymbol{I}), & (2.3.8) \\ \tilde{\boldsymbol{Z}}_{t+1} &= S_{\eta\lambda}\Big(\tilde{\boldsymbol{Z}}_t - 2\eta \boldsymbol{A}^\top(\boldsymbol{A}\tilde{\boldsymbol{Z}}_t - \boldsymbol{X})\Big), \quad \forall t \geq 1, & (2.3.9) \end{aligned}$$

with step size $\eta \geq 0$, and the soft-thresholding operator $S_\alpha$ defined on scalars by

$$\begin{aligned} S_\alpha(x) &\doteq \begin{cases} x - \alpha, & x \geq \alpha, \\ 0, & -\alpha < x < \alpha, \\ x + \alpha, & x \leq -\alpha \end{cases} & (2.3.10) \\ &= \operatorname{sign}(x) \max\{|x| - \alpha, 0\}, & (2.3.11) \end{aligned}$$

and applied element-wise to the input matrix. As a proximal gradient descent algorithm applied to a convex problem, convergence to a global optimum is assured, and a precise convergence rate can be derived straightforwardly [WM22].

We now take a moment to remark on the functional form of the update operator in (2.3.9). It takes the form

$$\tilde{\boldsymbol{Z}}_{t+1} = \texttt{nonlinearity}(\tilde{\boldsymbol{Z}}_t + \texttt{linear}^\top(\texttt{linear}(\tilde{\boldsymbol{Z}}_t) + \texttt{bias})). \tag{2.3.12}$$

This functional form is very similar to that of a residual network layer, namely,

$$\tilde{\boldsymbol{Z}}_{t+1} = \tilde{\boldsymbol{Z}}_t + \texttt{linear}_1^\top(\texttt{nonlinearity}(\texttt{linear}_2(\tilde{\boldsymbol{Z}}_t) + \texttt{bias}_1) + \texttt{bias}_2, \tag{2.3.13}$$

only decoupling the linear maps (i.e., matrix multiplications), adding a bias, and moving the nonlinearity. The nonlinearity in (2.3.9) is notably similar to the commonly-used ReLU activation function $x \mapsto \max\{x, 0\}$ in deep learning—as visualized in Figure 2.7, the soft-thresholding operator is like a "signed" ReLU activation, which is also shifted by a bias. The ISTA, then, can be viewed as a forward pass of a primitive (recurrent one-layer) neural network. We argue in Chapter 5 that such operations are essential to deep representation learning.

### 2.3.2 Overcomplete Dictionary Learning

Recall that we have the data model

$$\boldsymbol{X} = \boldsymbol{A}\boldsymbol{Z} + \boldsymbol{E}, \tag{2.3.14}$$

where $\boldsymbol{Z}$ is sparse, and our goal previously was to estimate $\boldsymbol{Z}$ given knowledge of the data $\boldsymbol{X}$ and the dictionary atoms $\boldsymbol{A}$. Now we turn to the more practical and more difficult case where we do not know either $\boldsymbol{A}$ or $\boldsymbol{Z}$ and seek to learn them from a large dataset.

A direct generalization of Equation (2.3.5) suggests solving the problem

$$\min_{\tilde{\boldsymbol{A}}, \tilde{\boldsymbol{Z}}} \left\{ \|\boldsymbol{X} - \tilde{\boldsymbol{A}}\tilde{\boldsymbol{Z}}\|_F^2 + \lambda \|\tilde{\boldsymbol{Z}}\|_1 \right\}. \tag{2.3.15}$$

However, the bilinear term in Equation (2.3.15) introduces a scale ambiguity that hinders convergence: given any point $(\tilde{\boldsymbol{A}}, \tilde{\boldsymbol{Z}})$ and any constant $c > 0$, the substitution $(c^{-1}\tilde{\boldsymbol{A}}, c\tilde{\boldsymbol{Z}})$ gives loss value

$$\|\boldsymbol{X} - \tilde{\boldsymbol{A}}\tilde{\boldsymbol{Z}}\|_F^2 + c\lambda \|\tilde{\boldsymbol{Z}}\|_1. \tag{2.3.16}$$

This loss is evidently minimized over $c$ by taking $c \to 0$, which corresponds to making the rescaled dictionary $c^{-1}\tilde{\boldsymbol{A}}$ go 'to infinity'. In particular, the iterates of any optimization algorithm solving (2.3.15) will not converge.

This issue with (2.3.15) is dealt with by adding a constraint on the scale of the columns of the dictionary $\tilde{\boldsymbol{A}}$. For example, it is common to assume that each column $\boldsymbol{A}_j$ of the dictionary $\boldsymbol{A}$ in (2.3.14) has bounded $\ell^2$ norm—say, without loss of generality, by 1. We then enforce this as a constraint, giving the objective

$$\min_{\tilde{\boldsymbol{Z}}, \tilde{\boldsymbol{A}} \,:\, \|\tilde{\boldsymbol{A}}_j\|_2 \leq 1} \left\{ \|\boldsymbol{X} - \tilde{\boldsymbol{A}}\tilde{\boldsymbol{Z}}\|_F^2 + \lambda \|\tilde{\boldsymbol{Z}}\|_1 \right\}. \tag{2.3.17}$$

Constrained optimization problems such as (2.3.17) can be solved by a host of sophisticated algorithms [NW06]. However, a simple and scalable one is actually furnished by the same proximal gradient descent algorithm that we used to solve the sparse coding problem in the previous section. We can encode each constraint as additional regularization term, via the characteristic function for the constraint set—details are given in Example A.1. Applying proximal gradient descent to the resulting regularized problem is equivalent to *projected gradient descent*, in which, at each iteration, the iterates after taking a gradient descent step are projected onto the constraint set.

*Remark* 2.12 ($\ell^4$ maximization versus $\ell^1$ minimization). Note that the above problem formulation follows naturally from the LASSO formulation (2.3.5) for sparse coding. We promote the sparsity of the solution via the $\ell^1$ norm. Nevertheless, if we are only interested in recovering the over-complete dictionary $\boldsymbol{A}$, the $\ell^4$ maximization scheme introduced in Section 2.2.2 also generalizes to the over-complete case, without any significant modification. Interested readers may refer to the work of [QZL+20a].

The above problem (2.3.17), which we call *overcomplete dictionary learning*, is nonconvex as here both $\boldsymbol{A}$ and $\boldsymbol{Z}$ are unknown. It cannot be solved easily by the standard convex optimization toolkit. Nevertheless, because it is interesting, simple to state, and practically important, there have been many important works dedicated to different algorithms and theoretical analysis for this problem. Here, for the interest of this manuscript, we present an idiomatic approach to solve this problem which is closer to the spirit of deep learning.

From our experience with the LASSO problem above, it is easy to see that, for the two unknowns $\boldsymbol{A}$ and $\boldsymbol{Z}$, if we fix one and optimize the other, each subproblem is in fact convex and easy to solve. This naturally suggests that we could attempt to solve the above program (2.3.17) by minimizing against $\boldsymbol{Z}$ or $\boldsymbol{A}$ alternatively, say using gradient descent. Coupled with a natural choice of initialization, this leads to the following iterative scheme:

$$\tilde{\boldsymbol{Z}}^{\ell+1} = S_{\eta\lambda}\left(\tilde{\boldsymbol{Z}}^{\ell} - 2\eta\boldsymbol{A}_+^\top(\boldsymbol{A}_+\tilde{\boldsymbol{Z}}^{\ell} - \boldsymbol{X})\right), \quad \tilde{\boldsymbol{Z}}^1 = \boldsymbol{0}, \quad \forall \ell \in [L] \tag{2.3.18}$$

$$\boldsymbol{Z}^+ = \tilde{\boldsymbol{Z}}^L, \tag{2.3.19}$$

$$\tilde{\boldsymbol{A}}_{t+1} = \operatorname{proj}_{\|(\,\cdot\,)_j\|_2 \le 1,\, \forall j}\left(\tilde{\boldsymbol{A}}_t - 2\nu(\tilde{\boldsymbol{A}}_t\boldsymbol{Z}^+ - \boldsymbol{X})(\boldsymbol{Z}^+)^\top\right),$$
$$\text{where} \quad (\tilde{\boldsymbol{A}}_1)_j \overset{\text{i.i.d.}}{\sim} \mathcal{N}(\boldsymbol{0}, \tfrac{1}{D}\boldsymbol{I}), \;\; \forall j \in [m], \qquad \forall t \in [T], \tag{2.3.20}$$

$$\boldsymbol{A}_+ = \tilde{\boldsymbol{A}}_T, \tag{2.3.21}$$

where the projection operation in the update for $\boldsymbol{A}$ ensures each column has at most unit $\ell^2$ norm, via $\boldsymbol{A}_j \mapsto \boldsymbol{A}_j / \max\{\|\boldsymbol{A}_j\|_2, 1\}$, and where $\boldsymbol{A}_+$ is initialized with each column i.i.d. $\mathcal{N}(\boldsymbol{0}, \frac{1}{D}\boldsymbol{I})$. The above consists of one 'block' of alternating minimization, and we repeatedly perform such blocks, each with independent initializations, until convergence. Above, we have used two separate indices $\{t\}$ and $\{\ell\}$ to indicate the iterations. As we will see later, this allows us to interpret the two updates separately in the context of deep learning.

Despite the dictionary learning problem being a nonconvex problem, it has been shown that alternating minimization type algorithms indeed converge to the correct solution, at least locally. See, for example, [AAJ+16]. As a practical demonstration, the above algorithm (with $L = T = 1$) was used to generate the results for overcomplete dictionary learning in Figure 2.6.

### 2.3.3 Learned Deep Sparse Coding

The main insight from the alternating minimization algorithm for overcomplete dictionary learning in the previous section (Equations (2.3.18) and (2.3.20)) is to notice that *when we fix $\boldsymbol{A}$, the ISTA update for $\boldsymbol{Z}^\ell$ (2.3.18) looks like the forward pass of a deep neural network with weights given by $\boldsymbol{A}$ (and $\boldsymbol{A}^\top$).* But in general, we do not know the true $\boldsymbol{A}$, and the current estimate $\boldsymbol{A}_+$ could be erroneous. Hence it needs to be further updated using (2.3.20) based on the residual of using the current estimate of the sparse codes $\boldsymbol{Z}^+$ to reconstruct $\boldsymbol{X}$. The alternating minimization algorithm iterates these two procedures until convergence. But we can instead extrapolate, and design other learning procedures by combining these insights with techniques from deep learning. This leads to more interpretable network architectures, which will be a recurring theme throughout this manuscript.

**Learned ISTA.** The above deep-network interpretation of the alternating minimization is more conceptual than practical, as the process could be rather inefficient and take many layers or iterations to converge. But this is mainly because we try to infer both $\boldsymbol{Z}$ and $\boldsymbol{A}$ from $\boldsymbol{X}$. The problem can be significantly simplified and the above iterations can be made much more efficient in the *supervised* setting, where we have a dataset of input and output pairs $(\boldsymbol{X}, \boldsymbol{Z})$ distributed according to (2.3.14) and we only seek to learn $\boldsymbol{A}^\ell$ for the layerwise learnable sparse coding iterations (2.3.29):[12]

$$\boldsymbol{Z}^{\ell+1} = S_{\eta\lambda}\big(\boldsymbol{Z}^\ell - 2\eta(\boldsymbol{A}^\ell)^\top(\boldsymbol{A}^\ell\boldsymbol{Z}^\ell - \boldsymbol{X})\big), \quad \forall \ell \in [L]. \tag{2.3.22}$$

If we denote the operator for each iteration as $\boldsymbol{Z}^{\ell+1} = f(\boldsymbol{A}^\ell, \boldsymbol{Z}^\ell)$, the above iteration can be illustrated in terms of a diagram:

$$\boldsymbol{X}, \boldsymbol{Z}^1 \xrightarrow{f(\boldsymbol{A}^1,\,\cdot\,)} \boldsymbol{Z}^2 \xrightarrow{f(\boldsymbol{A}^2,\,\cdot\,)} \boldsymbol{Z}^3 \xrightarrow{f(\boldsymbol{A}^3,\,\cdot\,)} \cdots \boldsymbol{Z}^L \xrightarrow{f(\boldsymbol{A}^L,\,\cdot\,)} \boldsymbol{Z}^{L+1} \approx \boldsymbol{Z}.$$

Thus, given the sequential architecture, to learn the operator $\boldsymbol{A}^\ell$ at each layer, it is completely natural to learn it, say via back propagation (BP),[13] by minimizing the error between the final code $\boldsymbol{Z}^L$ and the ground truth $\boldsymbol{Z}$:

$$\min_{\{\boldsymbol{A}^\ell\}} \big\|\boldsymbol{Z}^L(\boldsymbol{A}^1, \ldots, \boldsymbol{A}^L) - \boldsymbol{Z}\big\|_2^2. \tag{2.3.23}$$

[12] We drop the "tilde" accents here for concision, as the indexing makes the optimization-dependence clear.

[13] See Section A.2 for a brief description of BP.

This is the basis of the Learned ISTA (LISTA) algorithm [GL10], which can be viewed as the learning algorithm for a deep neural network, which tries to emulate the sparse encoding process from $\boldsymbol{X}$ to $\boldsymbol{Z}$. In particular, it can be viewed as a *simple representation learning algorithm*. In fact, this same methodology can be used as a basis to understand the representations computed in more powerful network architectures, such as transformers. We develop these implications in detail in Chapter 5.

**Sparse Autoencoders.** The original motivation for overcomplete dictionary learning was to provide a simple generative model for high-dimensional data. We have seen with LISTA that, in addition, iterative algorithms for learning sparsely-used overcomplete dictionaries provide an interpretation for ReLU-like deep networks, which we will generalize in later chapters to more complex data distributions than (2.3.14). But it is also worth noting that even in the modern era of large models, the data generating model (2.3.14) provides a useful practical basis for *interpreting features in pretrained large-scale deep networks*, such as transformers, following the hypothesis that the (non-interpretable, *a priori*) features in these networks consist of sparse "superpositions" of underlying features, which are themselves interpretable [EHO+22b]. These *unsupervised* learning paradigms are generally more data friendly than LISTA, as well, which requires large amounts of labeled $(\boldsymbol{X}, \boldsymbol{Z})$ pairs for supervised training.

We can use our development of the LISTA algorithm above to understand common practices in this field of research. In the most straightforward instantiation (see [GTT+25; HCS+24]), a large number of features from a pretrained deep network $h$ are collected from different inputs $\boldsymbol{x}_i$, which themselves are chosen based on a desired interpretation task.[14] For simplicity, we will use $h$ to denote the pre-selected feature map in question, with $D$-dimensional features; given $N$ sample inputs, let $\boldsymbol{H} \in \mathbb{R}^{D \times N}$ denote the full matrix of features of $h$. Then a so-called sparse autoencoder $f : \mathbb{R}^D \to \mathbb{R}^d$, with decoder $g : \mathbb{R}^d \to \mathbb{R}^D$, is trained via the LASSO loss (2.3.5):

$$\min_{f,g} \|\boldsymbol{H} - g(f(\boldsymbol{H}))\|_F^2 + \lambda \|f(\boldsymbol{H})\|_1, \tag{2.3.24}$$

where the sparse autoencoder $f$ takes the form of a one-layer neural network, i.e. $f(\boldsymbol{h}_i) = \sigma(\boldsymbol{W}_{\mathrm{enc}}(\boldsymbol{h}_i - \boldsymbol{b}) + \boldsymbol{b}_{\mathrm{enc}})$, where $\sigma(x) = \max\{x, 0\}$ is the ReLU activation function, and the decoder $g$ is linear, so that $g(\boldsymbol{z}_i) = \boldsymbol{W}_{\mathrm{dec}}\boldsymbol{z} + \boldsymbol{b}$.[15]

[14] For example, the inputs $\boldsymbol{x}_i$ could correspond to texts containing samples of computer code in different programming languages, with our task being to try to identify interpretable features in a transformer feature map $h$ corresponding to different salient aspects of the input, such as the specific programming language (distinct across input "classes") or the need to insert a matching parenthesis at the current position (common across input "classes"). We discuss the use of deep networks, and in particular transformers, for text representation learning in greater detail in Chapter 8.

[15] Since $f$ is a sparsifying operator, we denote it analogously to the previously used $f$, say in LISTA where $f$ denotes an optimization algorithm to sparsely encode the input. Throughout the book, $f$ will be used to denote such sparse and otherwise structured transformations, including those of *deep network encoders* (Chapters 4 and 5).

The parameterization and training procedure (2.3.24) may initially seem to be an arbitrary application of deep learning to the sparse coding problem, but it is actually highly aligned with the algorithms we have studied above for layerwise sparse coding with a learned dictionary. In particular, recall the LISTA architecture $\boldsymbol{Z}^L = f(\boldsymbol{A}^L, f(\boldsymbol{A}^{L-1}, \cdots, f(\boldsymbol{A}^1, \boldsymbol{X})\cdots))$. In the special case $L = 2$, we have

$$\boldsymbol{Z}^2 = f(\boldsymbol{A}^1, \boldsymbol{X}) = S_{\eta\lambda}\big(\boldsymbol{Z}^1 - 2\eta(\boldsymbol{A}^1)^\top(\boldsymbol{A}^1\boldsymbol{Z}^1 - \boldsymbol{X})\big). \tag{2.3.25}$$

Let us assume that the sparse codes $\boldsymbol{Z}$ in question are nonnegative, i.e., that $\boldsymbol{Z} \geq \mathbf{0}$.[16] Then (see Example A.3), we can consider an equivalent LISTA architecture obtained from the sparse coding objective with an additional non-negativity constraint on $\boldsymbol{Z}$ as

$$\boldsymbol{Z}^2 = f(\boldsymbol{A}^1, \boldsymbol{X}) = \max\left\{\boldsymbol{Z}^1 - 2\eta(\boldsymbol{A}^1)^\top(\boldsymbol{A}^1\boldsymbol{Z}^1 - \boldsymbol{X}) - \lambda\eta\mathbf{1}, 0\right\}, \tag{2.3.26}$$

and after some algebra, express this as

$$\boldsymbol{Z}^2 = f(\boldsymbol{A}^1, \boldsymbol{X}) = \max\left\{2\eta(\boldsymbol{A}^1)^\top + \left(\boldsymbol{Z}^1 - 2\eta(\boldsymbol{A}^1)^\top\boldsymbol{A}^1\boldsymbol{Z}^1 - \lambda\eta\mathbf{1}\right), 0\right\}. \tag{2.3.27}$$

Given the ability to change the sparse code initialization $\boldsymbol{Z}^1$ as a learnable parameter (which, in the current framework, must have all columns equal to the same learnable vector), this has the form of a ReLU neural network with learnable bias—identical to the sparse autoencoder $f$! Moreover, to *decode* the learned sparse codes $\boldsymbol{Z}^2$, it is natural to apply the learned dictionary $\boldsymbol{Z}^2 \mapsto \boldsymbol{A}^1\boldsymbol{Z}^2$. Then the only difference between this and the SAE decoder $g$ is the additional bias $\boldsymbol{b}$, which can technically be absorbed into $\boldsymbol{H}$ and $f$ in the training objective (2.3.24).

Thus, the SAE parameterization and training procedure coincides with LISTA training with $L = 1$, and a modified training objective—using the LASSO objective (2.3.5), which remains *unsupervised*, instead of the supervised reconstruction loss (2.3.23) used in vanilla LISTA. In particular, we can understand the SAE architecture in terms of our interpretation of the LISTA architecture in terms of layerwise sparse coding in (2.3.29). This connection is suggestive of a host of new design strategies for improving practical interpretability methodology, many of which remain tantalizingly unexplored. We begin to lay out some connections to broader autoencoding methodology in Chapter 6.

**Layerwise learned sparse coding?** In the supervised setting, LISTA provides a deep neural network analogue of the sparse coding iteration, with layerwise-learned dictionaries, inspired by alternating minimization; even in the unsupervised setting, the same methodology can be applied to learning, as with sparse autoencoders. But the connection between low-dimensional-structure-seeking optimization algorithms and deep network architectures goes much deeper than this, and suggests an array of scalable and natural neural learning architectures which may even be usable without backpropagation.

[16] In the data generating model (2.3.14), an arbitrary dictionary-and-sparse-code pair $(\boldsymbol{A}, \boldsymbol{Z})$ can be replaced by one in which $\boldsymbol{Z} \geq \mathbf{0}$ simply by doubling the number of columns in $\boldsymbol{A}$, so from a modeling perspective, this is a very mild assumption.

As a simple illustration, we return the alternating minimization iterations (2.3.18) and (2.3.20). This scheme randomly re-initializes the dictionary $\boldsymbol{A}_1$ on every such update. An improvement uses instead *warm starting*, where the residual is generated using the previous estimate $\boldsymbol{A}_+$ for the dictionary. If we then view each ISTA update (2.3.18) as a layer and allow the associated dictionary, now coupled with the sparse code updates as $\boldsymbol{A}_\ell$, to update in time, this leads to a "layerwise-learnable" sparse coding scheme:

$$\boldsymbol{Z}^1 = \boldsymbol{0}, \quad (\boldsymbol{A}_1)_j \overset{\text{i.i.d.}}{\sim} \mathcal{N}(\boldsymbol{0}, \tfrac{1}{D}\boldsymbol{I}), \ \forall j \in [m], \tag{2.3.28}$$

$$\boldsymbol{Z}^{\ell+1} = S_{\eta\lambda}\big(\boldsymbol{Z}^\ell - 2\eta(\boldsymbol{A}_\ell)^\top(\boldsymbol{A}_\ell\boldsymbol{Z}^\ell - \boldsymbol{X})\big), \tag{2.3.29}$$

$$\boldsymbol{A}_{\ell+1} = \boldsymbol{A}_\ell - 2\nu(\boldsymbol{A}_\ell\boldsymbol{Z}^{\ell+1} - \boldsymbol{X})(\boldsymbol{Z}^{\ell+1})^\top. \tag{2.3.30}$$

Note that this iteration corresponds to a relabeling of (2.3.18) and (2.3.20) for $T = L = 1$, over infinitely many blocks. Each of the 'inner' steps updating $\boldsymbol{Z}$ can be considered as a one-layer forward pass, while each of the 'outer' steps updating $\boldsymbol{A}$ can be considered as a one-layer backward pass, of a primitive deep neural network. In particular, this algorithm is the simplest case in which a clear divide between forward optimization and backward learning manifests. This distinction between forward updates to the ephemeral "features" $\boldsymbol{Z}$ of data $\boldsymbol{X}$ versus backward updates to the persistent parameters $\boldsymbol{A}$ is still observed in current neural networks and autoencoders—we will have much more to say about it in Chapter 5 and in Chapter 6.

Notice that the above layer-wise scheme also suggests a plausible alternative to the current end-to-end optimization strategy that primarily relies on back propagation (BP) detailed in Section A.2.3. Freeing training large networks from BP would be one of the biggest challenges and opportunities in the future, as we will discuss more at the end of the book in Chapter 9.

## 2.4 Summary and Notes

**Key Messages.** The idealistic models we have presented in this chapter—PCA, ICA, and dictionary learning—were developed over the course of the twentieth century. Many books have been written solely about each method, so we only attempted here to give a broad overview of the key concepts and results. As we will see in coming chapters, although these are arguably rather idealistic simple models for low-dimensional data distributions, the basic ideas and even resulting algorithms to identify such distributions can be naturally generalized or extended to arbitrary low-dimensional distributions – the thesis of this book. In addition, as we will see in coming chapters, these basic (linear) models, to a large extent, can also serve as local prototypes to effectively approximately model general distributions with nonlinear low-dimensional structures.

**More History and References.** Jolliffe [Jol86] attributes principal component analysis to Pearson [Pea01], and independently Hotelling [Hot33]. In mathematics, the main result on the related problem of low-rank approximation in unitarily invariant norms is attributed to Eckart and Young [EY36], and to Mirsky for full generality [Mir60]. PCA continues to play an important role in research as perhaps the simplest model problem for unsupervised representation learning: as early as the 1980s, works such as Oja [Oja82] and Baldi and Hornik [BH89] used the problem to understand learning in primitive neural networks, and more recently, it has served as a tool for understanding more complex representation learning frameworks, such as diffusion models [WZZ+24].

Independent component analysis was proposed by B. Ans et al. [BJC85] and pioneered by Aapo Hyvärinen in the 1990s and early 2000s in a series of influential works: see Hyvärinen and Oja [HO00b] for a summary. As a simple model for structure that arises in practical data, it initially saw significant use in applications such as blind source separation, where each independent component $z_i$ represents an independent source (such as sound associated to a distinct instrument in a musical recording) that is superimposed to produce the observation $\boldsymbol{x} = \boldsymbol{U}\boldsymbol{z}$.

The problem of dictionary learning can, in the complete or orthogonal case, be seen as one of the foundational problems of twentieth-century signal processing, particularly in linear systems theory, where the Fourier basis plays the key role; from the 1980s onward, the field of computational harmonic analysis crystallized around the study of alternate such dictionaries for classes of signals in which optimal approximation could only be realized in a basis other than Fourier (e.g., wavelets) [DVD+98]. However, the importance of the case of redundant bases, or overcomplete dictionaries, was only highlighted following the pioneering work of Olshausen and Field [OF97; OF96]. Early subsequent work established the conceptual and algorithmic foundations for learning sparsely-used overcomplete dictionaries, often aimed at representing natural images [AEB06; DM03; Don01; EA06; GJB15; MBP14; MK07]. Later, a significant amount of theoretical interest in the problem, as an important and nontrivial model problem for unsupervised representation learning, led to its study by the signal processing, theoretical machine learning, and theoretical computer science communities, in particular focused on conditions under which the problem could be provably and efficiently solved. A non-exhaustive list of notable works in this line include those of Spielman et al. [SWW12] and Sun et al. [SQW17a] on the complete case; Arora et al. [AGM+15]; Barak et al. [BKS15]; and Qu et al. [QZL+20a]. Many deep theoretical questions about this simple-to-state problem remain open, perhaps in part due to a tension with the problem's worst-case NP-hardness (e.g., see Tillmann [Til15]).

One point that we wish to highlight from the study of these classical analytical models for low-dimensional structure is the common role played by various *generalized power methods*—algorithms that very rapidly converge, at least locally, to various types of low-dimensional structures. The terminology for this class of algorithms follows the work of M. Journée et al. [MYP+10]. At a high level, modeled on the classical power iteration for computation of the top

Table 2.1: Summary of (generalized) power methods presented in this chapter.

| Problem | Algorithm | Iteration | Type of Structure Enforced |
|---|---|---|---|
| PCA | Power Method | $\boldsymbol{u}_t = \frac{\boldsymbol{X}\boldsymbol{X}^\top \boldsymbol{u}_t}{\\|\boldsymbol{X}\boldsymbol{X}^\top \boldsymbol{u}_t\\|_2}$ | 1-dim. subspace (unit vector) |
| ICA | FastICA | $\boldsymbol{u}_{t+1} = \frac{\frac{1}{N}\boldsymbol{X}(\boldsymbol{X}^\top \boldsymbol{u}_t)^{\odot 3} - 3\boldsymbol{u}_t}{\left\\|\frac{1}{N}\boldsymbol{X}(\boldsymbol{X}^\top \boldsymbol{u}_t)^{\odot 3} - 3\boldsymbol{u}_t\right\\|_2}$ | 1-dim. subspace (unit vector) |
| Complete DL | MSP Algorithm | $\boldsymbol{U}_{t+1} = \mathcal{P}_{\mathsf{O}(D)}[(\boldsymbol{U}_t\boldsymbol{X})^{\odot 3}\boldsymbol{X}^\top]$ | $D$-dim. subspace (ortho. matrix) |

eigenvector of a semidefinite matrix $\boldsymbol{A} \in \mathbb{R}^{n\times n}$, that is

$$\boldsymbol{u}_{t+1} = \frac{\boldsymbol{A}\boldsymbol{u}_t}{\|\boldsymbol{A}\boldsymbol{u}_t\|_2}, \tag{2.4.1}$$

this class of algorithms consists of a "powering" operation involving a matrix $\boldsymbol{A}$ associated to the data, along with a "projection" operation that enforces a desired type of structure. Table 2.1 presents a summary of the algorithms we have studied in this chapter that follow this structure. The reader may appreciate the applicability of this methodology to different types of low-dimensional structure, and different losses (i.e., both the quadratic loss from PCA, and the kurtosis-type losses from ICA), as well as the lack of such an algorithm for overcomplete dictionary learning, despite the breadth of the literature on these algorithms. We see the development of power methods for further families of low-dimensional structures, particularly those relevant to applications where deep learning is prevalent, as one of the more important (and open) research questions suggested by this chapter.

The connection we make in Section 2.2.1 between the geometric mixture-of-subspaces distributional assumption and the more analytically-convenient sparse dictionary assumption has been mentioned in prior work, especially by those focused on generalized principal component analysis and applications such as subspace clustering, e.g., work of Vidal et al. [VMS16]. The mixture of subspaces assumption will continue to play a significant role throughout this manuscript, both as an analytical test case for different algorithmic paradigms, and as a foundation for deriving different deep network architectures, as with LISTA in Section 2.3.3, but which can scale to more complex data distributions.

## 2.5 Exercises and Extensions

*Exercise* 2.1. Prove that, for any symmetric matrix $\boldsymbol{A}$, the solution to the problem $\max_{\boldsymbol{U}\in\mathsf{O}(D,d)} \operatorname{tr}\left(\boldsymbol{U}^\top \boldsymbol{A}\boldsymbol{U}\right)$ is the matrix $\boldsymbol{U}^\star$ whose columns are the top $d$ unit eigenvectors of $\boldsymbol{A}$.

*Exercise* 2.2. Let $\boldsymbol{z} \sim \mathcal{N}(\boldsymbol{0}, \sigma^2\boldsymbol{I})$ be a Gaussian random variable with independent components, each with variance $\sigma^2$. Prove that for any orthogonal matrix $\boldsymbol{Q}$ (i.e., $\boldsymbol{Q}^\top\boldsymbol{Q} = \boldsymbol{I}$), the random variable $\boldsymbol{Q}\boldsymbol{z}$ is distributed identically to $\boldsymbol{z}$. *(Hint: recall the formula for the Gaussian probability density function, and the formula for the density of a linear function of a random variable.)*

*Exercise* 2.3. The notion of statistical identifiability discussed above can be related to *symmetries* of the model class, allowing estimation to be understood in a purely deterministic fashion without any statistical assumptions.

Consider the model $\boldsymbol{X} = \boldsymbol{U}\boldsymbol{Z}$ for matrices $\boldsymbol{X}, \boldsymbol{U}, \boldsymbol{Z}$ of compatible sizes.

1. Show that if $\boldsymbol{A}$ is any square invertible matrix of compatible size, then the pair $(\boldsymbol{U}\boldsymbol{A}, \boldsymbol{A}^{-1}\boldsymbol{Z})$ also equals $\boldsymbol{X}$ under the model. We call this a $\mathsf{GL}(d)$ *symmetry.*

2. Suppose each column of $\boldsymbol{Z}$ is an independent and identically distributed observation from a common statistical model $\boldsymbol{z}$, which moreover has zero mean and independent components $z_i$ with positive variance. Show that for any square invertible matrix $\boldsymbol{A}$, if $\boldsymbol{A}\boldsymbol{z}$ has uncorrelated components, then $\boldsymbol{A}$ can be written as $\boldsymbol{D}_1\boldsymbol{Q}\boldsymbol{D}_2$, where $\boldsymbol{Q}$ is an orthogonal matrix and $\boldsymbol{D}_1$, $\boldsymbol{D}_2$ are diagonal matrices. *This links the "independence" assumption in ICA to a "symmetry breaking" effect, which only allows scale and rotational symmetries.*

*Exercise* 2.4. Consider the model $\boldsymbol{x} = \boldsymbol{U}\boldsymbol{z}$, where $\boldsymbol{U} \in \mathbb{R}^{D\times d}$ with $D \geq d$ is fixed and has rank $d$, and $\boldsymbol{z}$ is a zero-mean random variable. Let $\boldsymbol{x}_1, \dots, \boldsymbol{x}_N$ denote i.i.d. observations from this model.

1. Show that the matrix $\boldsymbol{X} = [\boldsymbol{x}_1, \dots, \boldsymbol{x}_N]$ has rank no larger than $d$, and therefore there is an orthonormal matrix $\boldsymbol{V} \in \mathbb{R}^{D\times d}$ so that $\boldsymbol{X} = \boldsymbol{V}\boldsymbol{Y}$, where $\boldsymbol{Y} \in \mathbb{R}^{d\times N}$. (*Hint: use PCA.*)

2. Show that the *whitened matrix* $(\boldsymbol{Y}\boldsymbol{Y}^\top)^{-1/2}\boldsymbol{Y}$ exists in expectation whenever $\mathrm{Cov}(\boldsymbol{z})$ is nonsingular, and that it has identity empirical covariance.[17]

3. Show, by using the singular value decomposition of $\boldsymbol{U}$, that the matrix $\boldsymbol{V}$ can be chosen so that the whitened matrix satisfies $(\boldsymbol{Y}\boldsymbol{Y}^\top)^{-1/2}\boldsymbol{Y} = \boldsymbol{W}[\boldsymbol{z}_1, \dots, \boldsymbol{z}_N]$, where $\boldsymbol{W}$ is an orthonormal matrix.

*Exercise* 2.5. Let $X$ and $Y$ be zero-mean independent random variables.

1. Show that $\mathrm{kurt}(X + Y) = \mathrm{kurt}(X) + \mathrm{kurt}(Y)$.

2. For any $\alpha \in \mathbb{R}$, show that $\mathrm{kurt}(\alpha X) = \alpha^4 \,\mathrm{kurt}(X)$.

*Exercise* 2.6. Let $f : \mathbb{R}^d \to \mathbb{R}$ be a given twice-continuously-differentiable objective function. Consider the spherically-constrained optimization problem

$$\max_{\|\boldsymbol{u}\|_2^2=1} f(\boldsymbol{u}). \tag{2.5.1}$$

In this exercise, we will derive the expressions we gave in the FastICA derivation for maximizing kurtosis over the sphere via a gradient ascent algorithm. These expressions are special cases of a rich theory of calculus and optimization on manifolds, of which the sphere is a particular example. A deep technical study of this field is out-of-scope for our purposes, so we only mention two key references for the interested reader: the pioneering textbook by Absil, Mahony, and Sepulchre [AMS09], and a more recent introductory treatise by Boumal [Bou23].

[17] In particular, it can be proved mathematically that this is enough to guarantee that the whitened matrix exists with high probability whenever $\boldsymbol{z}$ satisfies a suitable concentration inequality and $N$ is sufficiently large.

1. For any constraint set $\mathcal{M}$ that is a differentiable submanifold of $\mathbb{R}^d$, the *tangent space* at a point $\boldsymbol{u} \in \mathcal{M}$ is, informally, the best local linear approximation to the manifold $\mathcal{M}$ at the point $\boldsymbol{u}$. In the important special case where $\mathcal{M}$ is defined locally at $\boldsymbol{u}$ as a level set of a function $F : \mathbb{R}^d \to \mathbb{R}$, that is
$$U \cap \mathcal{M} = F^{-1}(\{0\})$$
for some open set $U \subset \mathcal{M}$ with $\boldsymbol{u} \in U$, the tangent space to $\mathcal{M}$ at $\boldsymbol{u}$ can be calculated via differentiation:
$$T_{\boldsymbol{u}}\mathcal{M} = \mathrm{Ker}(DF_{\boldsymbol{u}}).$$
It is easily seen that the sphere has the defining equation $F(\boldsymbol{u}) = \|\boldsymbol{u}\|_2^2 - 1$. Show, using these facts, that the tangent space to the sphere at $\boldsymbol{u}$ is given by
$$T_{\boldsymbol{u}}\mathbb{S}^{d-1} = \{\boldsymbol{v} \in \mathbb{R}^d \mid \langle \boldsymbol{v}, \boldsymbol{u} \rangle = 0\},$$
and that the orthogonal projection onto this subspace is $\boldsymbol{P}_{\boldsymbol{u}}^{\perp} = \boldsymbol{I} - \boldsymbol{u}\boldsymbol{u}^{\top}$.

2. The vector field
$$\mathrm{grad}\, f(\boldsymbol{u}) = \boldsymbol{P}_{\boldsymbol{u}}^{\perp} \nabla f \tag{2.5.2}$$
is known as the *Riemannian gradient* of the function $f$ restricted to the sphere. The *first order optimality conditions* for the optimization problem (2.5.1) can be expressed in terms of the Riemann gradient:
$$\mathrm{grad}\, f(\boldsymbol{u}) = \boldsymbol{0}.$$
Geometrically, this says that the Euclidean gradient of $f$ at $\boldsymbol{u}$ must be orthogonal to the tangent space to the sphere at $\boldsymbol{u}$. Now suppose $\boldsymbol{v} \in \mathbb{R}^d$ is nonzero. Show that
$$\mathrm{proj}_{\mathbb{S}^{d-1}}(\boldsymbol{v}) \doteq \min_{\|\boldsymbol{u}\|_2^2 = 1} \|\boldsymbol{u} - \boldsymbol{v}\|_2 = \frac{\boldsymbol{v}}{\|\boldsymbol{v}\|_2},$$
using the first-order optimality conditions.

3. In optimization over $\mathbb{R}^d$, one checks the second-order optimality conditions (to determine whether a critical point is a maximizer, a minimizer, or a saddle point) using the *Hessian matrix* $\nabla^2 f(\boldsymbol{u})$. Show, by differentiating the Riemann gradient $\mathrm{grad}\, f(\boldsymbol{u})$ for the sphere with respect to $\boldsymbol{u}$ as in the first part of this exercise, that the corresponding object for determining second-order optimality conditions for sphere-constrained optimization is the *Riemannian Hessian*, defined as
$$\mathrm{Hess}\, f(\boldsymbol{u}) = \boldsymbol{P}_{\boldsymbol{u}}^{\perp} \left( \nabla^2 f(\boldsymbol{u}) - \langle \nabla f(\boldsymbol{u}), \boldsymbol{u} \rangle \boldsymbol{I} \right) \boldsymbol{P}_{\boldsymbol{u}}^{\perp}. \tag{2.5.3}$$

*Exercise* 2.7. In this exercise, we sketch an argument referred to in the literature as a *landscape analysis* for the spherically-constrained population kurtosis maximization problem (2.2.24). We will show that when there is at least one independent component with positive kurtosis, its global maximizers indeed lead to the recovery of one column of the dictionary $\boldsymbol{U}$. For simplicity, we will assume that $\mathrm{kurt}(z_i) \neq 0$ for each $i = 1, \dots, d$.

1. Using the results of Part 1 of Exercise 2.6, show that the first-order optimality condition for (2.2.24) is

$$\left( \sum_{i=1}^{d} \mathrm{kurt}(z_i) w_i^4 \right) \boldsymbol{w} = \mathrm{kurt}(\boldsymbol{z}) \odot \boldsymbol{w}^{\odot 3}, \tag{2.5.4}$$

   where the kurtosis is calculated elementwise, $\odot$ denotes elementwise multiplication of vectors and $\boldsymbol{w}^{\odot 3}$ denotes the elementwise cube of its argument.

2. Show that the vectors $\boldsymbol{w}$ with unit norm that also satisfy (2.5.4) all take the following form. Let $S^+ = \{i \in [d] \mid \mathrm{kurt}(z_i) > 0\}$, and $S^- = \{i \in [d] \mid \mathrm{kurt}(z_i) < 0\}$. Let $S$ be a subset either of $S^+$ or $S^-$. Then

$$\boldsymbol{w}_S = \sum_{i \in S} \pm \sqrt{\frac{1}{\mathrm{kurt}(z_i) \sum_{j \in S} \frac{1}{\mathrm{kurt}(z_j)}}} \boldsymbol{e}_i \tag{2.5.5}$$

   satisfies (2.5.4), where $\boldsymbol{e}_i$ is the vector with a 1 in the $i$-th position and 0s elsewhere, and the $\pm$ sign denotes the choice of either a positive or negative sign.

3. Assume that there is at least one $i$ such that $\mathrm{kurt}(z_i) > 0$. Using the results of Part 2 of Exercise 2.6, show that the only local maxima of the objective of (2.2.24) are the signed one-sparse vectors $\pm \boldsymbol{e}_i$ with $i \in S^+$. Conclude that the global maximizers of (2.2.24) are the signed one-sparse vectors corresponding to components with maximum kurtosis. (*Hint: count the number of positive and negative eigenvalues of the Riemannian Hessian* (2.5.3) *at each critical point.*)

4. Now assume that $\mathrm{kurt}(z_j) < 0$ for every $j = 1, \dots, d$. This corresponds to an "over-deflated" instantiation of the kurtosis maximization problem. Using again the results of Part 2 of Exercise 2.6, show that the only local maxima of the objective of (2.2.24) are the signed dense vectors $\sum_{i=1}^{d} \pm \boldsymbol{e}_i$. This shows that the optimization formulation (2.2.24) cannot be applied naively.

*Exercise* 2.8. This exercise follows the structure and formalism introduced in Exercise 2.6, but applies it instead to the orthogonal group $\mathsf{O}(d) = \{\boldsymbol{U} \in \mathbb{R}^{d \times d} \mid \boldsymbol{U}^\top \boldsymbol{U} = \boldsymbol{I}\}$. Consult the description of Exercise 2.6 for the necessary conceptual background; the formalism applies identically to the case where the ambient space is the set of $d \times d$ matrices as long as one recalls that the relevant inner product on matrices is $\langle \boldsymbol{X}, \boldsymbol{Y} \rangle = \mathrm{tr}(\boldsymbol{X}^\top \boldsymbol{Y})$. An excellent general reference for facts about optimization on the orthogonal group is Edelman, Arias, and Smith [EAS98].

Let $f : \mathbb{R}^{d\times d} \to \mathbb{R}$ be a given twice-continuously-differentiable objective function. Consider the orthogonally-constrained optimization problem

$$\max_{\boldsymbol{Q}^\top \boldsymbol{Q}=\boldsymbol{I}} f(\boldsymbol{Q}). \tag{2.5.6}$$

1. It is easily seen that the orthogonal group has the defining equation $F(\boldsymbol{Q}) = \boldsymbol{Q}^\top \boldsymbol{Q} = \boldsymbol{I}$. Show, using this fact, that the tangent space to the orthogonal group at $\boldsymbol{Q}$ is given by

$$T_{\boldsymbol{Q}}\mathsf{O}(d) = \{\boldsymbol{Q}\boldsymbol{\Omega} \in \mathbb{R}^{d\times d} \mid \boldsymbol{\Omega}^\top = -\boldsymbol{\Omega}\},$$

and that the orthogonal projection onto this subspace is

$$\mathcal{P}_{T_{\boldsymbol{Q}}\mathsf{O}(d)}(\boldsymbol{\Delta}) = \boldsymbol{Q}\,\mathrm{skew}(\boldsymbol{Q}^\top \boldsymbol{\Delta}),$$

where $\mathrm{skew}(\boldsymbol{\Delta}) = \frac{1}{2}(\boldsymbol{\Delta} - \boldsymbol{\Delta}^\top)$ is the orthogonal projection onto the set of skew-symmetric matrices. The vector field

$$\mathrm{grad}\, f(\boldsymbol{Q}) = \mathcal{P}_{T_{\boldsymbol{Q}}\mathsf{O}(d)}\left(\nabla f(\boldsymbol{Q})\right) \tag{2.5.7}$$

is known as the *Riemannian gradient* of the function $f$ restricted to the orthogonal group. The *first order optimality conditions* for the optimization problem (2.5.6) can be expressed in terms of the Riemann gradient:

$$\mathrm{grad}\, f(\boldsymbol{Q}) = \boldsymbol{0}.$$

2. Show, by differentiating the Riemann gradient $\mathrm{grad}\, f(\boldsymbol{Q})$ for the orthogonal group with respect to $\boldsymbol{Q}$ as in the first part of this exercise, that the *Riemannian Hessian* is given by

$$\mathrm{Hess}\, f(\boldsymbol{Q}) = \mathcal{P}_{T_{\boldsymbol{Q}}\mathsf{O}(d)}\left(\nabla^2 f(\boldsymbol{Q}) - \mathrm{sym}(\boldsymbol{Q}^\top \nabla f(\boldsymbol{Q})) \otimes \boldsymbol{I}\right) \mathcal{P}_{T_{\boldsymbol{Q}}\mathsf{O}(d)}, \tag{2.5.8}$$

where $\mathrm{sym}(\boldsymbol{\Delta}) = \frac{1}{2}(\boldsymbol{\Delta} + \boldsymbol{\Delta}^\top)$ denotes the orthogonal projection onto the set of symmetric matrices, and $\otimes$ denotes the Kronecker product of matrices. Take care to interpret the operators appearing in the previous expression as *linear transformations on $d \times d$ matrices*, **not** as $d \times d$ matrices themselves. The *second-order optimality conditions* for the optimization problem (2.5.6) can be expressed in terms of the Riemann Hessian:

$$\mathrm{Hess}\, f(\boldsymbol{Q}) \preceq \boldsymbol{0}.$$

For a minimization problem, the sign is reversed.

(*Hint: The key is to manipulate one's calculations to obtain the form* (2.5.8), *which is as compact as possible. To this end, make use of the following* *isomorphism of the Kronecker product: if $\boldsymbol{A}$, $\boldsymbol{X}$, and $\boldsymbol{B}$ are matrices of compatible sizes, then one has*

$$(\boldsymbol{B}^\top \otimes \boldsymbol{A}) \operatorname{vec}(\boldsymbol{X}) = \operatorname{vec}(\boldsymbol{A}\boldsymbol{X}\boldsymbol{B}),$$

*where* $\operatorname{vec}(\boldsymbol{X})$ *denotes the "left-to-right" stacking of the columns of the matrix argument into a vector. We use this isomorphism in* (2.5.8) *in order to define the Kronecker product of two matrices as an operator on matrices in a canonical way.*)

3. Now suppose $\boldsymbol{X} \in \mathbb{R}^{d\times d}$ is full-rank. In this and the next part of the exercise, we consider the projection onto the orthogonal group of $\boldsymbol{X}$:

$$\operatorname{proj}_{\mathsf{O}(d)}(\boldsymbol{X}) \doteq \min_{\boldsymbol{Q}\in\mathsf{O}(d)} \|\boldsymbol{Q} - \boldsymbol{X}\|_F^2. \tag{2.5.9}$$

We will prove that the solution to this problem is given by

$$\operatorname{proj}_{\mathsf{O}(d)}(\boldsymbol{X}) = \boldsymbol{U}\boldsymbol{V}^\top,$$

where $\boldsymbol{X} = \boldsymbol{U}\boldsymbol{S}\boldsymbol{V}^\top$ is a singular value decomposition of $\boldsymbol{X}$.

1. Using the first and second-order optimality conditions, show that every local minimizer $\boldsymbol{Q}$ of (2.5.9) satisfies

$$\left(\boldsymbol{Q}^\top \boldsymbol{X}\right)^\top = \boldsymbol{Q}^\top \boldsymbol{X},$$
$$\boldsymbol{Q}^\top \boldsymbol{X} \succeq \boldsymbol{0}.$$

(*Hint: use linearity of the Kronecker product in either of its two arguments when the other is fixed.*)

2. Using these conditions, argue that at every local minimizer $\boldsymbol{Q}$ of (2.5.9), one has $\boldsymbol{Q}^\top \boldsymbol{X} = (\boldsymbol{X}^\top \boldsymbol{X})^{1/2}$. (*Hint: Use the fact from linear algebra that if* $\boldsymbol{S} \succeq \boldsymbol{0}$ *is a symmetric positive semidefinite matrix, then* $(\boldsymbol{S}^\top \boldsymbol{S})^{1/2} = \boldsymbol{S}$.)

3. Using the singular value decomposition $\boldsymbol{X} = \boldsymbol{U}\boldsymbol{S}\boldsymbol{V}^\top$, conclude that

$$\boldsymbol{U}\boldsymbol{V}^\top = \operatorname{proj}_{\mathsf{O}(d)}(\boldsymbol{X}).$$

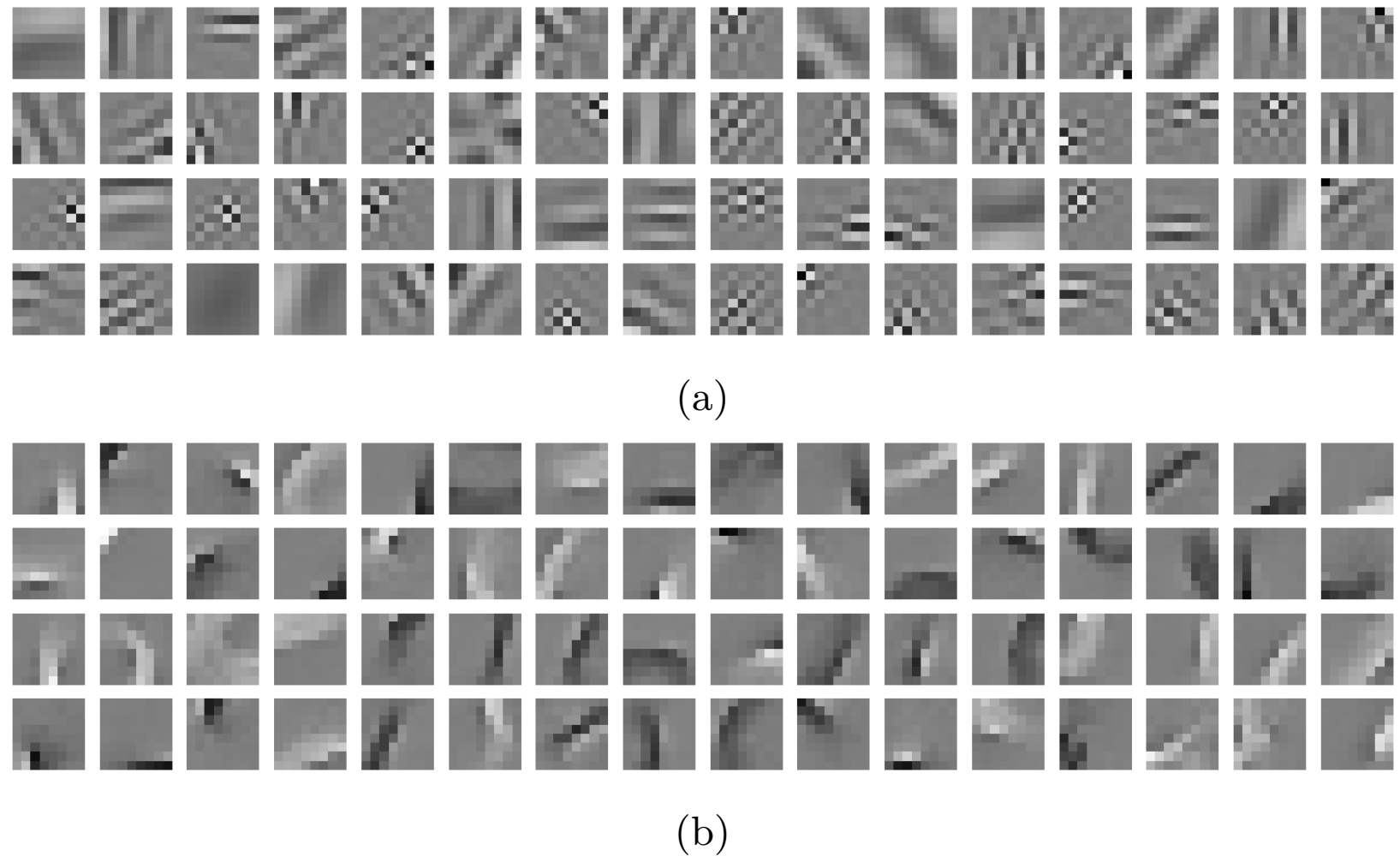

Figure 2.6: Comparison of learned dictionary atoms for complete (orthogonal) and overcomplete dictionaries, trained to reconstruct 8 by 8 patches taken from MNIST digits. Both dictionaries are trained for 6000 epochs on $10^4$ random patches with nontrivial content, and sparse codes are computed with the LASSO objective and $\lambda = 0.1$ (see (2.3.5)). Colormaps have black for negative values, and white for positive values. **Top:** An orthogonal dictionary learned with the MSP algorithm (2.2.18) is constrained to have no more than 64 atoms; the learned atoms roughly correspond to a "spike and slab" dictionary, and achieve relatively poor reconstruction sparsity levels on held-out test data (codes are approximately 17-sparse on average, with respect to a threshold of $10^{-1}$). **Bottom:** In contrast, an overcomplete dictionary (here, with $8^3$ atoms; we visualize a random subset of 64) learns semantically-meaningful dictionary atoms corresponding to signed oriented edges, which can be pieced together to create digit patches and achieve superior reconstruction and sparsity levels. Codes are approximately 20-sparse on average, while being 8 times larger than those of the orthogonal dictionary. To compute the dictionary, we use an optimizer based on proximal alternating linearized minimization on a suitably-regularized version of (2.3.17).

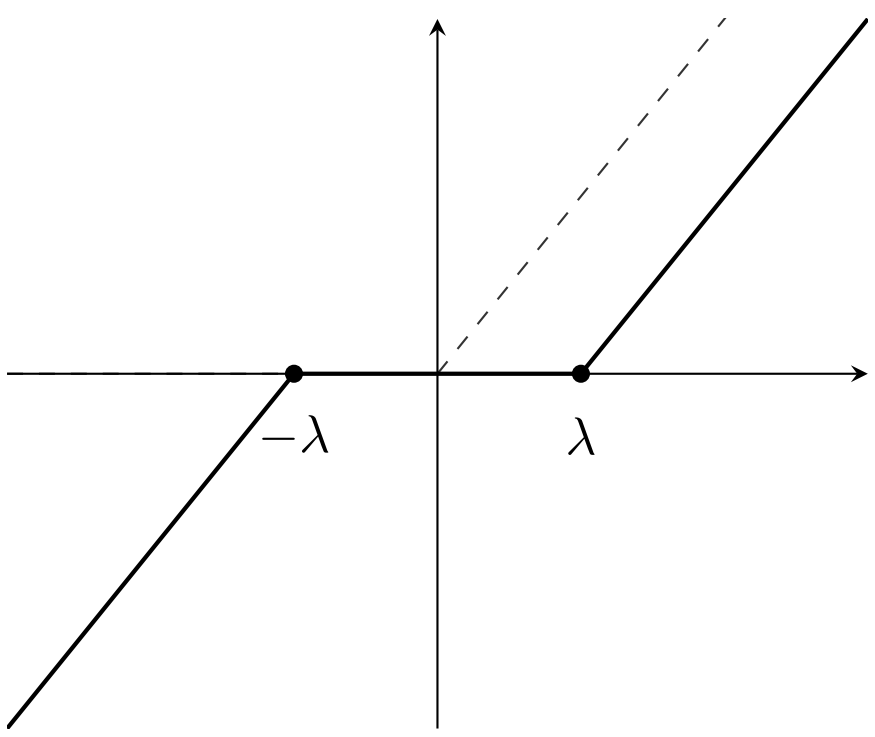

Figure 2.7: The graph of the soft thresholding function $S_\lambda$ (solid) compared to the ReLU activation $x \mapsto \max\{x, 0\}$ (dashed). Between $-\lambda$ and $+\lambda$, the value of $S_\lambda$ is clipped to 0, while large inputs' (absolute) value is simply shrunk by $\lambda$. Unlike ReLU, $S_\lambda$ is symmetric and acts on both positive and negative inputs.

# Chapter 3

# Pursuing Low-Dimensional Distributions via Denoising

"*Information is the resolution of uncertainty.*"

—Claude Shannon, 1948

In Chapter 2, we showed how to learn simple classes of distributions whose supports are assumed to be either a single low-dimensional subspace, a mixture of such subspaces, or low-rank Gaussians. For simplicity, the different (hidden) linear or Gaussian modes are taken to be orthogonal or independent,[1] as illustrated in Figure 2.1. For these special distributions, simple and effective learning algorithms with correctness and efficiency guarantees can be derived, and the geometric and statistical interpretations of the associated algorithmic operations are clear.

In practice, both linearity and independence are idealistic assumptions that distributions of real-world high-dimensional data *rarely* satisfy. The only assumption we can make is that the intrinsic dimension of the distribution is very low compared with the dimension of the ambient space in which the data are embedded. Hence, in this chapter we show how to learn a more general class of low-dimensional distributions in a high-dimensional space that is *not* necessarily (piecewise) linear.

Real-world distributions typically contain multiple components or modes, for example, corresponding to different object classes in images. These modes may not be statistically independent and can even have different intrinsic dimensions. Therefore, we generally assume that the data are distributed on a mixture of low-dimensional nonlinear submanifolds embedded in a high-dimensional space. Figure 3.1 illustrates such a distribution. Another constraint is that in real-world situations we typically have access only to a finite number of samples

[1] Or can be reduced to such idealistic cases under benign conditions.

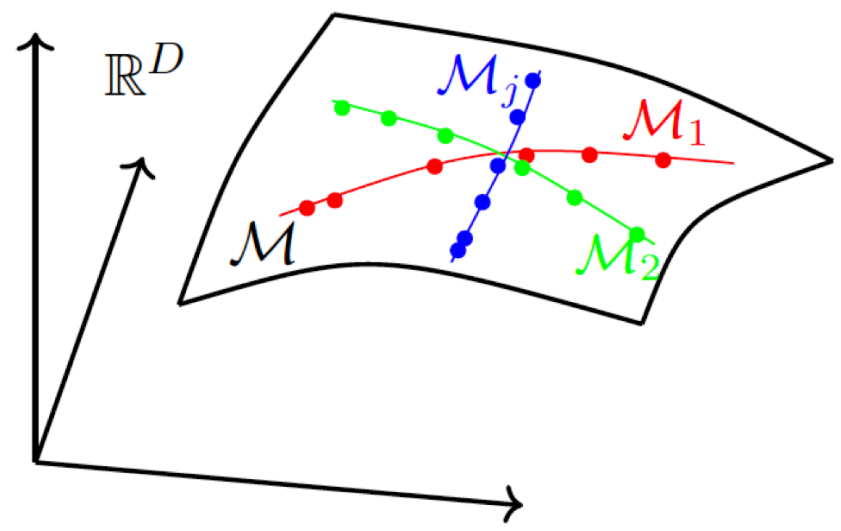

Figure 3.1: Data $\boldsymbol{X}$ are distributed on a mixture of low-dimensional submanifolds $\cup_j \mathcal{M}_j$ in a very high-dimensional ambient space, say $\mathbb{R}^D$. This is a much more general case than the special cases studied in Chapter 2, where the submanifolds were assumed to be piecewise linear and thus admitted parametric forms and analytical solutions.

from the data distribution, rather than prior knowledge of the exact distribution itself.

To learn such a distribution under these conditions, we must address three fundamental questions:

- What is a general approach to learning and representing a low-dimensional distribution in a high-dimensional space?

- How do we measure the complexity of the distribution so that we can exploit its low dimensionality effectively for learning?

- How do we make the learning process computationally tractable and scalable, especially when the ambient dimension is high and the sample size is large?

As we will see, the fundamental idea of *compression*, or *dimension reduction*, which we have already shown to be effective in the linear/independent case, remains a guiding principle for developing computational models and methods for general nonlinear low-dimensional distributions.

Due to its theoretical and practical importance, we examine in greater depth how this compression-based framework materializes when the distribution of interest can be well modeled or approximated by a mixture of low-dimensional subspaces or low-rank Gaussians. As later chapters show, these seemingly restrictive model classes serve as basic building blocks for algorithms that perform compression on general distributions.

## 3.1 Entropy Minimization and Compression

### 3.1.1 Entropy and Coding Rate

In Chapter 1 we mentioned that the goal of learning is to find the simplest way to generate a given set of data. Conceptually, Kolmogorov complexity was intended to provide such a measure of complexity, but it is not computable and is not associated with any implementable scheme that can actually reproduce the data. Hence, we need an alternative, computable, and realizable measure of complexity. This leads us to the notion of *entropy*, introduced by Shannon in 1948 [Sha48].

To illustrate the constructive nature of entropy, let us start with the simplest case. Suppose we have a discrete random variable that takes $N$ distinct values, or *tokens*, $\{\boldsymbol{x}_1, \dots, \boldsymbol{x}_N\}$ with equal probability $1/N$. We can encode each token $\boldsymbol{x}_i$ using the $\log_2 N$-bit binary representation of $i$. This coding scheme can be generalized to encoding arbitrary discrete distributions [CT91]: given a distribution $p$ such that $\sum_{i=1}^{N} p(\boldsymbol{x}_i) = 1$, one can assign each token $\boldsymbol{x}_i$ with probability $p(\boldsymbol{x}_i)$ a binary code of length $\log_2[1/p(\boldsymbol{x}_i)] = -\log_2 p(\boldsymbol{x}_i)$ bits. Hence the average number of bits, or *the coding rate*, needed to encode any sample from the distribution $p(\cdot)$ is given by the expression[2]:

$$H(p) \doteq \mathbb{E}_{\boldsymbol{x}\sim p}[-\log p(\boldsymbol{x})] = -\sum_{i=1}^{N} p(\boldsymbol{x}_i) \log p(\boldsymbol{x}_i). \tag{3.1.1}$$

This is known as the *entropy* of the (discrete) distribution $p(\cdot)$. Note that this entropy is always nonnegative and it is zero if and only if $p(\boldsymbol{x}_i) = 1$ for some $\boldsymbol{x}_i$ with $i \in [N]$.[3] For a random variable $\boldsymbol{x}$, we will often write $H(\boldsymbol{x})$ to mean the entropy of the (marginal) distribution of $\boldsymbol{x}$.

### 3.1.2 Differential Entropy

When the random variable $\boldsymbol{x} \in \mathbb{R}^D$ is continuous and has a probability density $p$, one may view the limit of the above sum (3.1.1) as related to an integral:

$$h(p) \doteq \mathbb{E}_{\boldsymbol{x}\sim p}[-\log p(\boldsymbol{x})] = -\int_{\mathbb{R}^D} p(\boldsymbol{\xi}) \log p(\boldsymbol{\xi}) \mathrm{d}\boldsymbol{\xi}. \tag{3.1.2}$$

More precisely, given a continuous probability distribution $p$ on $\mathbb{R}^D$, we may discretize $\mathbb{R}^D$ into a union of disjoint cubes $C_i^\epsilon$ of side length $\epsilon^{1/D}$, each (say) centered at $\boldsymbol{x}_i$, and define the discretized distribution $p^\epsilon$ by $p^\epsilon(\boldsymbol{x}_i) = \mathbb{P}_{\boldsymbol{x}\sim p}[\boldsymbol{x} \in C_i^\epsilon]$. Then one can show that $H(p^\epsilon) + \log \epsilon \approx h(p)$. By taking $\epsilon$ to be small, we observe that the differential entropy $h(p)$ can be negative. Interested readers may refer to [CT91] for a more detailed explanation. Similar to the discrete entropy, we will often write $h(\boldsymbol{x})$ to mean the entropy of the (marginal) distribution of $\boldsymbol{x}$.

[2] By the convention of information theory [CT91], the log here has base 2. Hence, entropy is measured in (binary) bits.

[3] Here we use the fact $\lim_{p\to 0} p \log p = 0$.

*Example* 3.1 (Entropy of Gaussian distributions). Through direct calculation, one can show that the entropy of a Gaussian random variable $x \sim \mathcal{N}(\mu, \sigma^2)$ is

$$h(x) = \frac{1}{2}\log(2\pi\sigma^2) + \frac{1}{2}. \tag{3.1.3}$$

The entropy of a multivariate Gaussian random variable $\boldsymbol{x} \sim \mathcal{N}(\boldsymbol{\mu}, \boldsymbol{\Sigma})$ in $\mathbb{R}^D$ is

$$h(\boldsymbol{x}) = \frac{D}{2}(1 + \log(2\pi)) + \frac{1}{2}\operatorname{logdet}(\boldsymbol{\Sigma}). \tag{3.1.4}$$

■

Similar to the entropy of a discrete distribution, we would like the differential entropy to be associated with the coding rate of some realizable coding scheme. For example, as mentioned above, if we discretize the domain of the distribution with a grid of size $\epsilon > 0$ with centers $\boldsymbol{x}_i$, one possible coding scheme is to map each $\boldsymbol{x}$ to its nearest center $\boldsymbol{x}_i$, and the entropy of the resulting discrete distribution approximates the differential entropy [CT91].

There are some caveats associated with the definition of differential entropy. For a distribution in a high-dimensional space, when its support becomes degenerate (low-dimensional), its differential entropy diverges to $-\infty$. This fact is proved in Theorem B.1, but even in the simple explicit case of Gaussian distributions (3.1.4), when the covariance $\boldsymbol{\Sigma}$ is singular, we can see that $\operatorname{logdet}(\boldsymbol{\Sigma}) = -\infty$ so we have $h(\boldsymbol{x}) = -\infty$ (note that by Theorem B.1, Gaussian distributions have the largest entropy among all distributions with the same mean and covariance). In such a situation, it is not obvious how to properly quantize or encode such a distribution. Nevertheless, degenerate (Gaussian) distributions are precisely the simplest possible, and arguably the most important, instances of low-dimensional distributions in a high-dimensional space. In the next chapter, we will discuss a complete resolution to this seeming difficulty with degeneracy.

### 3.1.3 Minimizing Coding Rate

The learning problem entails recovering a (potentially continuous) distribution $p(\boldsymbol{x})$ from samples $\{\boldsymbol{x}_1, \dots, \boldsymbol{x}_N\}$ drawn from it. For ease of exposition, we write $\boldsymbol{X} = \begin{bmatrix}\boldsymbol{x}_1, \dots, \boldsymbol{x}_N\end{bmatrix} \in \mathbb{R}^{D \times N}$. Because the distributions of interest are (nearly) low-dimensional, we expect their (differential) entropy to be very small. However, unlike the situations studied in the previous chapter, we do not know the family of (analytical) low-dimensional models to which $p(\boldsymbol{x})$ belongs. Thus, checking whether the entropy is small seems to be the only guideline we can rely on to identify and model the distribution.

Now, given the samples alone without knowing what $p(\boldsymbol{x})$ is, in theory they could be interpreted as samples from any generic distribution. In particular, they could be interpreted as any of the following:

1. as samples from the empirical distribution $p^{\boldsymbol{X}}$ itself, which assigns probability $1/N$ to each of the $N$ samples $\boldsymbol{x}_i, i \in [N]$;
2. as samples from a standard normal distribution $\boldsymbol{x}^n \sim p^n \doteq \mathcal{N}(\mathbf{0}, \sigma^2 \boldsymbol{I})$ with variance $\sigma^2$ large enough (say larger than the squared sample norms);
3. as samples from a normal distribution $\boldsymbol{x}^e \sim p^e \doteq \mathcal{N}(\mathbf{0}, \hat{\boldsymbol{\Sigma}})$ with covariance $\hat{\boldsymbol{\Sigma}} = \frac{1}{N}\boldsymbol{X}\boldsymbol{X}^\top$ being the empirical covariance of the samples;
4. as samples from a distribution $\hat{\boldsymbol{x}} \sim \hat{q}(\boldsymbol{x})$ that closely approximates the ground-truth distribution $p$.

The question is: which interpretation is better, and in what sense? Suppose that you believe these data $\boldsymbol{X}$ are drawn from a particular distribution $q(\boldsymbol{x})$, which may be one of the above. Then we could encode the data points with the optimal codebook for the distribution $q(\boldsymbol{x})$. Then, as the number of samples $N$ becomes large, the required average coding length (or coding rate) is given by the so-called *cross-entropy*:

$$\mathrm{CE}(p \parallel q) \doteq \mathbb{E}_{\boldsymbol{x}\sim p}[-\log q(\boldsymbol{x})]. \tag{3.1.5}$$

If we have identified the correct distribution $p(\boldsymbol{x})$, the coding rate is given by the entropy $h(p) = \mathbb{E}_{\boldsymbol{x}\sim p}[-\log p(\boldsymbol{x})]$. It turns out that the above coding length $\mathrm{CE}(p \parallel q)$ is always greater than or equal to the entropy $h(\boldsymbol{x})$ unless $p = q$.

Hence, given a set of sampled data $\boldsymbol{X}$, to determine which distribution best characterizes the data among $p^n$, $p^e$, and $\hat{q}$, we may compare their coding rates for $\boldsymbol{X}$ and see which yields the lowest rate. We know from the above that the (theoretically achievable) coding rate for a distribution is closely related to its entropy. In general,

$$h(p^n) > h(p^e) > h(\hat{q}). \tag{3.1.6}$$

If the data $\boldsymbol{X}$ were encoded by the codebook associated with each of these distributions, the coding rate for $\boldsymbol{X}$ would therefore decrease in the same order:

$$\mathrm{CE}(p \parallel p^n) > \mathrm{CE}(p \parallel p^e) > \mathrm{CE}(p \parallel \hat{q}). \tag{3.1.7}$$

This observation gives a general guideline for pursuing a distribution $p(\boldsymbol{x})$ that has low-dimensional structure. It suggests two possible approaches:

A1. Starting with a general distribution (say a normal distribution) with high entropy, gradually transform the distribution toward the (empirical) distribution of the data by reducing entropy;

A2. Among a large family of (parametric or non-parametric) distributions with explicit coding schemes that encode the given data, progressively search for better coding schemes that give lower coding rates.

Conceptually, both approaches A1 and A2 try to do the same thing. For A1, we need to ensure such a path of transformation exists and is computable. For A2, it is necessary that the chosen family is rich enough and can closely approximate (or contain) the ground-truth distribution. For either approach, we need to ensure that solutions with lower entropy or better coding rates can be efficiently computed and converge to the desired distribution quickly.[4] We will explore both approaches in the remaining sections of this chapter.

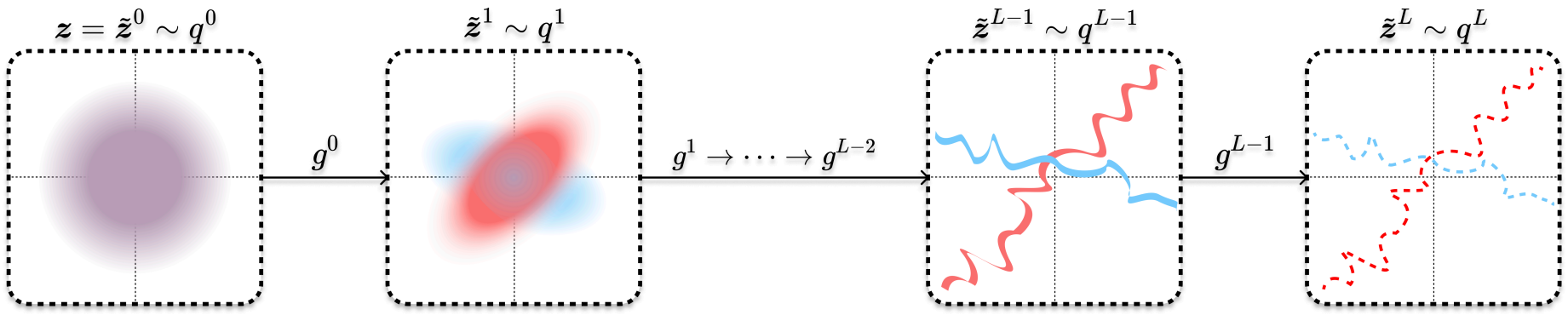

Figure 3.2: Illustration of an iterative denoising process that, starting from an isotropic Gaussian distribution, converges to an arbitrary data distribution.

## 3.2 Compression via Denoising

In this section we describe a *natural* and *computationally tractable* way to learn a distribution $p_{\boldsymbol{x}}$ by learning a parametric encoding that minimizes the entropy or coding rate of the representation, then using this encoding to transform high-entropy samples from a standard Gaussian into low-entropy samples from the target distribution, as illustrated in Figure 3.2. This methodology combines approaches A1 and A2 above to learn and sample from the distribution.

### 3.2.1 Diffusion and Denoising Processes

We first want a procedure that reduces the entropy of a very noisy random variable, producing a lower-entropy sample from the data distribution. Here we describe one potential approach, perhaps the most natural. First, we devise a way to *gradually increase* the entropy of samples already drawn from the data distribution. Next, we seek an *approximate inverse* of this process. In general, increasing entropy has no inverse, because information from the original distribution may be destroyed. We therefore restrict ourselves to a special case in which (1) the entropy-increasing operation is simple, computable, and reversible, and (2) we have a (parametric) encoding of the data distribution, as alluded to above. As we will see, these two conditions make our approach feasible.

#### Diffusion Processes

We will increase the entropy in arguably the simplest possible way, i.e., *adding isotropic Gaussian noise*. More precisely, given the random variable $\boldsymbol{x}$, we can consider the *stochastic process* $(\boldsymbol{x}_t)_{t\in[0,T]}$ that adds gradual noise to it, i.e.,

$$\boldsymbol{x}_t \doteq \boldsymbol{x} + t\boldsymbol{g}, \qquad \forall t \in [0, T], \tag{3.2.1}$$

[4] Say the distribution of real-world data such as images and texts.

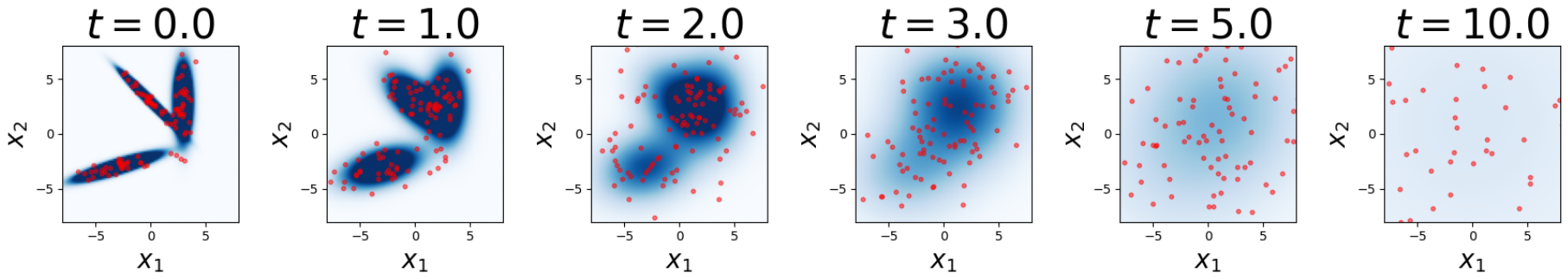

Figure 3.3: **Diffusing a mixture of Gaussians $\boldsymbol{x}$.** From left to right, we observe the evolution of the density $p_t$ of $\boldsymbol{x}_t$ as $t$ grows from 0 to 10, along with some representative samples (red). The plane is colored by the probability density $p_t$; high-density regions are colored darker blue. We observe that the probability mass becomes less concentrated as $t$ increases, signaling that entropy increases.

where $T \in [0, \infty)$ is a time horizon and $\boldsymbol{g} \sim \mathcal{N}(\mathbf{0}, \boldsymbol{I})$ is drawn independently of $\boldsymbol{x}$. This process is an example of a *diffusion process*, so named because it spreads (i.e., *diffuses*) the probability mass over all of $\mathbb{R}^D$ as time goes on, increasing the entropy. This intuition is confirmed graphically by Figure 3.3, and rigorously via the following theorem.

**Theorem 3.1** (Simplified version of Theorem B.2)**.** *Suppose that $(\boldsymbol{x}_t)_{t\in[0,T]}$ follows the model (3.2.1). For any $t \in (0, T]$, the random variable $\boldsymbol{x}_t$ has a probability density $p_t$ and differential entropy $h(\boldsymbol{x}_t) > -\infty$. Moreover, under certain technical conditions on $\boldsymbol{x}$,*

$$\frac{\mathrm{d}}{\mathrm{d}t} h(\boldsymbol{x}_t) > 0, \qquad \forall t \in (0, T], \tag{3.2.2}$$

*i.e., the entropy of the distribution of $\boldsymbol{x}_t$ increases over time $t$.*

The proof is elementary, but it is rather long, so we postpone it to Section B.2.1. The main as-yet unstated implication of this result is that $h(\boldsymbol{x}_t) > h(\boldsymbol{x}_s)$ for every $t > s \geq 0$. To see this, note that if $s = 0$ and $h(\boldsymbol{x}) = h(\boldsymbol{x}_0) = -\infty$ then $h(\boldsymbol{x}_t) > -\infty$ for all $t > 0$, so the conclusion is true; otherwise $h(\boldsymbol{x}_t) = h(\boldsymbol{x}_s) + \int_s^t \left[\frac{\mathrm{d}}{\mathrm{d}u} h(\boldsymbol{x}_u)\right] \mathrm{d}u > h(\boldsymbol{x}_s)$ by the fundamental theorem of calculus, so in both cases $h(\boldsymbol{x}_t) > h(\boldsymbol{x}_s)$ for every $t > s \geq 0$.

*Example* 3.2 (Adding noise to a mixture of Gaussians)*.* In this example, we consider adding noise to a mixture of $K$ Gaussians. Choose $K$ means $\boldsymbol{\mu}_1, \dots, \boldsymbol{\mu}_K \in \mathbb{R}^D$, $K$ covariance matrices $\boldsymbol{\Sigma}_1, \dots, \boldsymbol{\Sigma}_K \in \mathsf{PSD}(D)$, and probabilities $\pi_1, \dots, \pi_K \in [0, 1]$ such that $\sum_{k=1}^K \pi_k = 1$. Then $\boldsymbol{x}$ is sampled as follows.

- First, an index (or *label*) $y \in [K]$ is sampled s.t. $y = k$ with probability $\pi_k$.
- Second, $\boldsymbol{x}$ is sampled from the Gaussian $\mathcal{N}(\boldsymbol{\mu}_y, \boldsymbol{\Sigma}_y)$; i.e., if $y = k$ then $\boldsymbol{x} \sim \mathcal{N}(\boldsymbol{\mu}_k, \boldsymbol{\Sigma}_k)$.

Thus $\boldsymbol{x}$ has the Gaussian mixture distribution

$$\boldsymbol{x} \sim \sum_{k=1}^K \pi_k \mathcal{N}(\boldsymbol{\mu}_k, \boldsymbol{\Sigma}_k). \tag{3.2.3}$$

Gaussian mixture models are relevant because they can approximate many distributions: given any distribution for $\boldsymbol{x}$, there exists a Gaussian mixture that approximates it arbitrarily well. For this reason, Gaussian mixture models are a popular, analytically tractable surrogate for multi-modal and complex data distributions. In this book we use them extensively as a “model problem” to understand and develop almost every concept. For now, we represent the data $\boldsymbol{x}$ by a Gaussian mixture. After adding noise, the distribution of $\boldsymbol{x}_t$ is again a Gaussian mixture:

$$\boldsymbol{x}_t = \boldsymbol{x} + t\boldsymbol{g} \sim \sum_{k=1}^{K} \pi_k \mathcal{N}(\boldsymbol{\mu}_k, \boldsymbol{\Sigma}_k + t^2 \boldsymbol{I}). \tag{3.2.4}$$

To see by example how the entropy evolves, suppose $K = 1$, so $\boldsymbol{x} \sim \mathcal{N}(\boldsymbol{\mu}, \boldsymbol{\Sigma})$. By (3.1.4), the entropy of $\boldsymbol{x}_t$ is

$$h(\boldsymbol{x}_t) = \frac{D}{2}(1 + \log(2\pi)) + \frac{1}{2}\operatorname{logdet}(\boldsymbol{\Sigma} + t^2 \boldsymbol{I}). \tag{3.2.5}$$

Differentiating with respect to time (see Chapter A) gives

$$\frac{\mathrm{d}}{\mathrm{d}t} h(\boldsymbol{x}_t) = t \operatorname{tr}((\boldsymbol{\Sigma} + t^2 \boldsymbol{I})^{-1}) = \sum_{i=1}^{D} \frac{t}{\lambda_i(\boldsymbol{\Sigma}) + t^2} > 0, \tag{3.2.6}$$

where $\lambda_i(\boldsymbol{\Sigma})$ are the (non-negative) eigenvalues of $\boldsymbol{\Sigma}$. Because this derivative is positive, the entropy $h(\boldsymbol{x}_t)$ of the noised distribution increases over time. ■

### Denoising Processes

The inverse operation to adding noise is known as *denoising*. It is a classical and well-studied topic in signal processing and system theory, instantiated (for example) as the Wiener filter and the Kalman filter (see Section 1.3.1). Several problems discussed in Chapter 2, such as PCA, ICA, and Dictionary Learning, are specific instances of the denoising problem. For a fixed $t$ and the additive Gaussian noise model (3.2.1), the denoising problem can be formulated as attempting to learn a function $\bar{\boldsymbol{x}}^\star(t, \cdot)$ that forms the best possible approximation (in expectation) of the true random variable $\boldsymbol{x}$, given both $t$ and $\boldsymbol{x}_t$:

$$\bar{\boldsymbol{x}}^\star(t, \cdot) \in \arg\min_{\bar{\boldsymbol{x}}(t,\cdot)} \mathbb{E} \left\| \boldsymbol{x} - \bar{\boldsymbol{x}}(t, \boldsymbol{x}_t) \right\|_2^2. \tag{3.2.7}$$

The solution to this problem, when optimizing $\bar{\boldsymbol{x}}(t, \cdot)$ over all possible (square-integrable) functions, is the so-called *Bayes optimal denoiser*, and turns out to be the conditional expectation:

$$\bar{\boldsymbol{x}}^\star(t, \boldsymbol{\xi}) \doteq \mathbb{E}[\boldsymbol{x} \mid \boldsymbol{x}_t = \boldsymbol{\xi}]. \tag{3.2.8}$$

This expression justifies the notation $\bar{\boldsymbol{x}}$, which computes a conditional expectation (i.e., conditional mean or conditional average). In short, it attempts to remove the noise from the noisy input, outputting the best possible guess (in expectation and with respect to the $\ell^2$ distance) of the denoised original random variable.

*Example* 3.3 (Denoising Gaussian Noise from a Mixture of Gaussians). In this example we compute the Bayes optimal denoiser for an important class of distributions, the Gaussian mixture model in Example 3.2. Let us define $\varphi(\boldsymbol{x};\boldsymbol{\mu},\boldsymbol{\Sigma})$ as the probability density of $\mathcal{N}(\boldsymbol{\mu},\boldsymbol{\Sigma})$ evaluated at $\boldsymbol{x}$. In this notation, the density of $\boldsymbol{x}_t$ is

$$p_t(\boldsymbol{x}_t) = \sum_{k=1}^{K} \pi_k \varphi(\boldsymbol{x}_t; \boldsymbol{\mu}_k, \boldsymbol{\Sigma}_k + t^2 \boldsymbol{I}). \tag{3.2.9}$$

Conditioned on $y$, the variables are jointly Gaussian: if we write $\boldsymbol{x} = \boldsymbol{\mu}_y + \boldsymbol{\Sigma}_y^{1/2}\boldsymbol{u}$ where $(\cdot)^{1/2}$ is the matrix square root and $\boldsymbol{u} \sim \mathcal{N}(\boldsymbol{0}, \boldsymbol{I})$ independently of $y$ (and $\boldsymbol{g}$), then we have

$$\begin{bmatrix} \boldsymbol{x} \\ \boldsymbol{x}_t \end{bmatrix} = \begin{bmatrix} \boldsymbol{\mu}_y \\ \boldsymbol{\mu}_y \end{bmatrix} + \begin{bmatrix} \boldsymbol{\Sigma}_y^{1/2} & \boldsymbol{0} \\ \boldsymbol{\Sigma}_y^{1/2} & t\boldsymbol{I} \end{bmatrix} \begin{bmatrix} \boldsymbol{u} \\ \boldsymbol{g} \end{bmatrix}. \tag{3.2.10}$$

This shows that $\boldsymbol{x}$ and $\boldsymbol{x}_t$ are jointly Gaussian (conditioned on $y$) as claimed. Thus we can write

$$\begin{bmatrix} \boldsymbol{x} \\ \boldsymbol{x}_t \end{bmatrix} \sim \mathcal{N}\left( \begin{bmatrix} \boldsymbol{\mu}_y \\ \boldsymbol{\mu}_y \end{bmatrix}, \begin{bmatrix} \boldsymbol{\Sigma}_y & \boldsymbol{\Sigma}_y \\ \boldsymbol{\Sigma}_y & \boldsymbol{\Sigma}_y + t^2\boldsymbol{I} \end{bmatrix} \right). \tag{3.2.11}$$

Thus the conditional expectation of $\boldsymbol{x}$ given $\boldsymbol{x}_t$ (i.e., the Bayes optimal denoiser conditioned on $y$) is famously (see also Exercise 3.2)

$$\mathbb{E}[\boldsymbol{x} \mid \boldsymbol{x}_t, y] = \boldsymbol{\mu}_y + \boldsymbol{\Sigma}_y(\boldsymbol{\Sigma}_y + t^2\boldsymbol{I})^{-1}(\boldsymbol{x}_t - \boldsymbol{\mu}_y). \tag{3.2.12}$$

To find the overall Bayes optimal denoiser, we use the law of iterated expectation, obtaining

$$\begin{aligned} \bar{\boldsymbol{x}}^\star(t, \boldsymbol{x}_t) &= \mathbb{E}[\boldsymbol{x} \mid \boldsymbol{x}_t] && (3.2.13) \\ &= \mathbb{E}[\mathbb{E}[\boldsymbol{x} \mid \boldsymbol{x}_t, y] \mid \boldsymbol{x}_t] && (3.2.14) \\ &= \sum_{k=1}^{K} \mathbb{P}[y = k \mid \boldsymbol{x}_t]\, \mathbb{E}[\boldsymbol{x} \mid \boldsymbol{x}_t, y = k]. && (3.2.15) \end{aligned}$$

The probability can be dealt with as follows. Let $p_{t|y}$ be the probability density of $\boldsymbol{x}_t$ conditioned on the value of $y$. Then

$$\begin{aligned} \mathbb{P}[y = k \mid \boldsymbol{x}_t] &= \frac{p_{t|y}(\boldsymbol{x}_t \mid k)\pi_k}{p_t(\boldsymbol{x}_t)} && (3.2.16) \\ &= \frac{\pi_k \varphi(\boldsymbol{x}_t; \boldsymbol{\mu}_k, \boldsymbol{\Sigma}_k + t^2\boldsymbol{I})}{\sum_{i=1}^{K} \pi_i \varphi(\boldsymbol{x}_t; \boldsymbol{\mu}_i, \boldsymbol{\Sigma}_i + t^2\boldsymbol{I})}. && (3.2.17) \end{aligned}$$

On the other hand, the conditional expectation is as described before:

$$\mathbb{E}[\boldsymbol{x} \mid \boldsymbol{x}_t, y = k] = \boldsymbol{\mu}_k + \boldsymbol{\Sigma}_k(\boldsymbol{\Sigma}_k + t^2\boldsymbol{I})^{-1}(\boldsymbol{x}_t - \boldsymbol{\mu}_k). \tag{3.2.18}$$

So putting this all together, the true Bayes optimal denoiser is

$$\bar{\boldsymbol{x}}^\star(t, \boldsymbol{x}_t) = \sum_{k=1}^{K} \left\{ \frac{\pi_k \varphi(\boldsymbol{x}_t; \boldsymbol{\mu}_k, \boldsymbol{\Sigma}_k + t^2\boldsymbol{I})}{\sum_{i=1}^{K} \pi_i \varphi(\boldsymbol{x}_t; \boldsymbol{\mu}_i, \boldsymbol{\Sigma}_i + t^2\boldsymbol{I})} \right. \tag{3.2.19}$$

$$\left. \cdot \left(\boldsymbol{\mu}_k + \boldsymbol{\Sigma}_k(\boldsymbol{\Sigma}_k + t^2\boldsymbol{I})^{-1}(\boldsymbol{x}_t - \boldsymbol{\mu}_k)\right) \right\}. \tag{3.2.20}$$

This example is particularly important, and several special cases will give us great conceptual insight later. For now, let us attempt to extract some geometric intuition from the functional form of the optimal denoiser (3.2.19).

To understand (3.2.19) intuitively, let us first set $K = 1$ (i.e., one Gaussian) such that $\boldsymbol{x} \sim \mathcal{N}(\boldsymbol{\mu}, \boldsymbol{\Sigma})$. Let us then diagonalize $\boldsymbol{\Sigma} = \boldsymbol{V}\boldsymbol{\Lambda}\boldsymbol{V}^\top$. Then the Bayes optimal denoiser is

$$\bar{\boldsymbol{x}}^\star(t, \boldsymbol{x}_t) = \boldsymbol{\mu} + \boldsymbol{\Sigma}(\boldsymbol{\Sigma} + t^2\boldsymbol{I})^{-1}(\boldsymbol{x}_t - \boldsymbol{\mu}) \tag{3.2.21}$$

$$= \boldsymbol{\mu} + \boldsymbol{V} \begin{bmatrix} \lambda_1/(\lambda_1 + t^2) & & \\ & \ddots & \\ & & \lambda_D/(\lambda_D + t^2) \end{bmatrix} \boldsymbol{V}^\top(\boldsymbol{x}_t - \boldsymbol{\mu}), \tag{3.2.22}$$

where $\lambda_1, \ldots, \lambda_D$ are the eigenvalues of $\boldsymbol{\Sigma}$. We can observe that this denoiser has three steps:

- Translate the input $\boldsymbol{x}_t$ by $\boldsymbol{\mu}$.
- Contract the (translated) input $\boldsymbol{x}_t - \boldsymbol{\mu}$ in each eigenvector direction by a quantity $\lambda_i/(\lambda_i + t^2)$. If the translated input is low-rank and some eigenvalues of $\boldsymbol{\Sigma}$ are zero, these directions get immediately contracted to 0 by the denoiser, ensuring that the output of the contraction is similarly low-rank.
- Translate the output back by $\boldsymbol{\mu}$.

This is the geometric interpretation of the denoiser of a *single* Gaussian. The overall denoiser of the Gaussian mixture model (3.2.19) uses $K$ such denoisers, weighting their output by the posterior probabilities $\mathbb{P}[y = k \mid \boldsymbol{x}_t]$. If the means of the Gaussians are well separated, these posterior probabilities are very close to 0 or 1 near each mean or cluster. In this regime, the overall denoiser (3.2.19) has the same geometric interpretation as the above single Gaussian denoiser. ■

Intuitively, and as we can see from Example 3.3, the Bayes-optimal denoiser $\bar{\boldsymbol{x}}^\star(t, \cdot)$ should move its input $\boldsymbol{x}_t$ toward the modes of the distribution of $\boldsymbol{x}$. It turns out that we can quantify this by showing that the Bayes-optimal denoiser *takes a gradient-ascent step* on the (log-)density of $\boldsymbol{x}_t$, which (recall) we denoted $p_t$. That is, following the denoiser means moving from the input iterate to a region of higher probability within this (perturbed) distribution. For small $t$, the perturbation is small, so our initial intuition is therefore almost exactly right. The picture is visualized in Figure 3.4 and rigorously formulated as Tweedie's formula [Rob56].

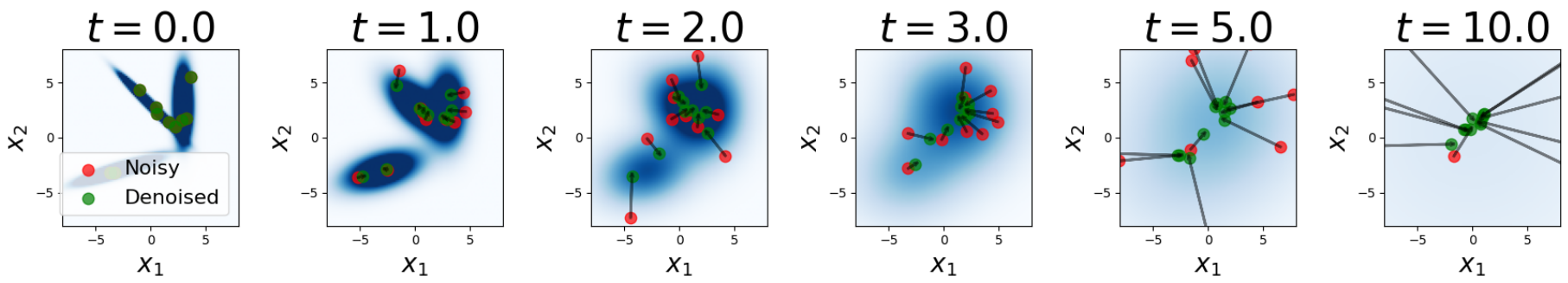

Figure 3.4: **Bayes-optimal denoiser and score of a Gaussian mixture model.** In the same setting as Figure 3.3, we demonstrate the effect of the Bayes-optimal denoiser $\bar{\boldsymbol{x}}^{\star}$ by plotting $\boldsymbol{x}_t$ (red) and $\bar{\boldsymbol{x}}^{\star}(t, \boldsymbol{x}_t)$ (green) for some choice of $t$ and $\boldsymbol{x}_t$. By Tweedie's formula Theorem 3.2, the residual between them is proportional to the so-called (Hyvärinen) score $\nabla_{\boldsymbol{x}_t} \log p_t(\boldsymbol{x}_t)$. We can see that the score points toward the modes of the distribution of $\boldsymbol{x}_t$.

**Theorem 3.2** (Tweedie's Formula)**.** *Suppose that* $(\boldsymbol{x}_t)_{t \in [0,T]}$ *obeys* (3.2.1)*. Let* $p_t$ *be the density of* $\boldsymbol{x}_t$ *(as previously declared). Then*

$$\mathbb{E}[\boldsymbol{x} \mid \boldsymbol{x}_t] = \boldsymbol{x}_t + t^2 \nabla_{\boldsymbol{x}_t} \log p_t(\boldsymbol{x}_t). \tag{3.2.23}$$

*Proof.* For the proof, suppose that $\boldsymbol{x}$ has a density,[5] and call this density $p$. Let $p_{0|t}$ and $p_{t|0}$ be the conditional densities of $\boldsymbol{x} = \boldsymbol{x}_0$ given $\boldsymbol{x}_t$ and $\boldsymbol{x}_t$ given $\boldsymbol{x}$ respectively. Let $\varphi(\boldsymbol{x}; \boldsymbol{\mu}, \boldsymbol{\Sigma})$ be the density of $\mathcal{N}(\boldsymbol{\mu}, \boldsymbol{\Sigma})$ evaluated at $\boldsymbol{x}$, so that $p_{t|0}(\boldsymbol{x}_t \mid \boldsymbol{x}) = \varphi(\boldsymbol{x}_t; \boldsymbol{x}, t^2\boldsymbol{I})$. Then a simple calculation gives

$$\begin{aligned}
\nabla_{\boldsymbol{x}_t} \log p_t(\boldsymbol{x}_t) &= \frac{\nabla_{\boldsymbol{x}_t} p_t(\boldsymbol{x}_t)}{p_t(\boldsymbol{x}_t)} && (3.2.24)\\
&= \frac{1}{p_t(\boldsymbol{x}_t)} \nabla_{\boldsymbol{x}_t} \int_{\mathbb{R}^D} p(\boldsymbol{x}) p_{t|0}(\boldsymbol{x}_t \mid \boldsymbol{x}) \mathrm{d}\boldsymbol{x} && (3.2.25)\\
&= \frac{1}{p_t(\boldsymbol{x}_t)} \nabla_{\boldsymbol{x}_t} \int_{\mathbb{R}^D} p(\boldsymbol{x}) \varphi(\boldsymbol{x}_t; \boldsymbol{x}, t^2\boldsymbol{I}) \mathrm{d}\boldsymbol{x} && (3.2.26)\\
&= \frac{1}{p_t(\boldsymbol{x}_t)} \int_{\mathbb{R}^D} p(\boldsymbol{x}) [\nabla_{\boldsymbol{x}_t} \varphi(\boldsymbol{x}_t; \boldsymbol{x}, t^2\boldsymbol{I})] \mathrm{d}\boldsymbol{x} && (3.2.27)\\
&= \frac{1}{p_t(\boldsymbol{x}_t)} \int_{\mathbb{R}^D} p(\boldsymbol{x}) \varphi(\boldsymbol{x}_t; \boldsymbol{x}, t^2\boldsymbol{I}) \left[-\frac{\boldsymbol{x}_t - \boldsymbol{x}}{t^2}\right] \mathrm{d}\boldsymbol{x} && (3.2.28)\\
&= \frac{1}{t^2 p_t(\boldsymbol{x}_t)} \int_{\mathbb{R}^D} p(\boldsymbol{x}) \varphi(\boldsymbol{x}_t; \boldsymbol{x}, t^2\boldsymbol{I}) [\boldsymbol{x} - \boldsymbol{x}_t] \mathrm{d}\boldsymbol{x} && (3.2.29)\\
&= \frac{1}{t^2 p_t(\boldsymbol{x}_t)} \int_{\mathbb{R}^D} p(\boldsymbol{x}) \varphi(\boldsymbol{x}_t; \boldsymbol{x}, t^2\boldsymbol{I}) \boldsymbol{x} \mathrm{d}\boldsymbol{x} && (3.2.30)
\end{aligned}$$

[5] Note that the theorem is true without this assumption, and it can be proven through a minimal modification of this proof to use probability measures instead of densities.

$$
\begin{aligned}
&- \frac{\boldsymbol{x}_t}{t^2 p_t(\boldsymbol{x}_t)} \int_{\mathbb{R}^D} p(\boldsymbol{x}) \varphi(\boldsymbol{x}_t; \boldsymbol{x}, t^2 \boldsymbol{I}) \mathrm{d}\boldsymbol{x} \\
&= \frac{1}{t^2 p_t(\boldsymbol{x}_t)} \int_{\mathbb{R}^D} p(\boldsymbol{x}) p_{t|0}(\boldsymbol{x}_t \mid \boldsymbol{x}) \boldsymbol{x} \mathrm{d}\boldsymbol{x} - \frac{\boldsymbol{x}_t}{t^2 p_t(\boldsymbol{x}_t)} p_t(\boldsymbol{x}_t) \qquad (3.2.31) \\
&= \frac{1}{t^2 p_t(\boldsymbol{x}_t)} \int_{\mathbb{R}^D} p_t(\boldsymbol{x}_t) p_{0|t}(\boldsymbol{x} \mid \boldsymbol{x}_t) \boldsymbol{x} \mathrm{d}\boldsymbol{x} - \frac{\boldsymbol{x}_t}{t^2 p_t(\boldsymbol{x}_t)} p_t(\boldsymbol{x}_t) \qquad (3.2.32) \\
&= \frac{1}{t^2} \int_{\mathbb{R}^D} p_{0|t}(\boldsymbol{x} \mid \boldsymbol{x}_t) \boldsymbol{x} \mathrm{d}\boldsymbol{x} - \frac{\boldsymbol{x}_t}{t^2} \qquad (3.2.33) \\
&= \frac{1}{t^2} \mathbb{E}[\boldsymbol{x} \mid \boldsymbol{x}_t] - \frac{\boldsymbol{x}_t}{t^2} \qquad (3.2.34) \\
&= \frac{\mathbb{E}[\boldsymbol{x} \mid \boldsymbol{x}_t] - \boldsymbol{x}_t}{t^2}. \qquad (3.2.35)
\end{aligned}
$$

Simple rearrangement of the above equality proves the theorem. □

This result develops a connection between denoising and optimization: the Bayes-optimal denoiser takes a single step of gradient ascent on the perturbed data density $p_t$, and the step size adaptively becomes smaller (i.e., takes more precise steps) as the perturbation to the data distribution grows smaller. The quantity $\nabla_{\boldsymbol{x}_t} \log p_t(\boldsymbol{x}_t)$ is called the *(Hyvärinen) score* and frequently appears in discussions about denoising, etc.; it first appeared in a paper of Aapo Hyvärinen in the context of ICA [Hyv05].

Similar to how one step of gradient descent is almost never sufficient to minimize an objective in practice when initializing far from the optimum, the output of the Bayes-optimal denoiser $\bar{\boldsymbol{x}}^\star(t, \cdot)$ is almost never contained in a high-probability region of the data distribution when $t$ is large, *especially* when the data have low-dimensional structure. We illustrate this point explicitly in the following example.

*Example* 3.4 (Denoising a two-point mixture). Let $x$ be uniform on the two-point set $\{-1, +1\}$ and let $(x_t)_{t \in [0,T]}$ follow (3.2.1). This is precisely a degenerate Gaussian mixture model with priors equal to $\frac{1}{2}$, means $\{-1, +1\}$, and covariances both equal to 0. For a fixed $t > 0$ we can use the calculation of the Bayes-optimal denoiser in (3.2.19) to obtain (proof as exercise)

$$
\bar{x}^\star(t, x_t) = \frac{\varphi(x_t; +1, t^2) - \varphi(x_t; -1, t^2)}{\varphi(x_t; 1, t^2) + \varphi(x_t; -1, t^2)} = \tanh\left(\frac{x_t}{t^2}\right). \qquad (3.2.36)
$$

For $t$ near 0, this quantity is near $\{-1, +1\}$ for almost all inputs $\bar{x}^\star(t, x_t)$. However, for $t$ large, this quantity is not necessarily even approximately in the original support of $x$, which, remember, is $\{-1, +1\}$. In particular, for $x_t \approx 0$ it holds $\bar{x}^\star(t, x_t) \approx 0$, which lies completely in between the two possible points. Thus $\bar{x}^\star$ *will not output "realistic"* $x$. More rigorously, the distribution of $\bar{x}^\star(t, x_t)$ is very different from the distribution of $x$. ■

### Incremental Denoising

Therefore, to denoise the very noisy sample $\boldsymbol{x}_T$ (where, recall, $T$ is the maximum time), we cannot use the denoiser just once. Instead, we must apply it many times, analogously to gradient descent with decaying step sizes, to converge to a stationary point $\hat{\boldsymbol{x}}$. Namely, we use the denoiser to go from $\boldsymbol{x}_T$ to $\hat{\boldsymbol{x}}_{T-\delta}$, which approximates $\boldsymbol{x}_{T-\delta}$, then from $\hat{\boldsymbol{x}}_{T-\delta}$ to $\hat{\boldsymbol{x}}_{T-2\delta}$, and so on, all the way from $\hat{\boldsymbol{x}}_\delta$ to $\hat{\boldsymbol{x}} = \hat{\boldsymbol{x}}_0$. Each denoising step makes the action of the denoiser more like a gradient step on the original (log-)density.

More formally, we uniformly discretize $[0, T]$ into $L + 1$ timesteps $0 = t_0 < t_1 < \cdots < t_L = T$, i.e.,

$$t_\ell = \frac{\ell}{L}T, \qquad \ell \in \{0, 1, \ldots, L\}. \tag{3.2.37}$$

Then for each $\ell \in [L] = \{1, 2, \ldots, L\}$, going from $\ell = L$ to $\ell = 1$, we run the iteration

$$\begin{aligned}
\hat{\boldsymbol{x}}_{t_{\ell-1}} &= \mathbb{E}[\boldsymbol{x}_{t_{\ell-1}} \mid \boldsymbol{x}_{t_\ell} = \hat{\boldsymbol{x}}_{t_\ell}] && (3.2.38)\\
&= \mathbb{E}[\boldsymbol{x} + t_{\ell-1}\boldsymbol{g} \mid \boldsymbol{x}_{t_\ell} = \hat{\boldsymbol{x}}_{t_\ell}] && (3.2.39)\\
&= \mathbb{E}\left[\boldsymbol{x} + t_{\ell-1} \cdot \frac{\boldsymbol{x}_{t_\ell} - \boldsymbol{x}}{t_\ell} \;\middle|\; \boldsymbol{x}_{t_\ell} = \hat{\boldsymbol{x}}_{t_\ell}\right] && (3.2.40)\\
&= \frac{t_{\ell-1}}{t_\ell}\hat{\boldsymbol{x}}_{t_\ell} + \left(1 - \frac{t_{\ell-1}}{t_\ell}\right)\mathbb{E}[\boldsymbol{x} \mid \boldsymbol{x}_{t_\ell} = \hat{\boldsymbol{x}}_{t_\ell}] && (3.2.41)\\
&= \left(1 - \frac{1}{\ell}\right)\hat{\boldsymbol{x}}_{t_\ell} + \frac{1}{\ell}\bar{\boldsymbol{x}}^\star(t_\ell, \hat{\boldsymbol{x}}_{t_\ell}). && (3.2.42)
\end{aligned}$$

At the beginning of the iteration, when $\ell$ is large, we barely trust the denoiser's output and mostly keep the current iterate. This makes sense, as the denoiser can have huge variance (cf. Example 3.4). When $\ell$ is small, the denoiser "locks on" to the modes of the data distribution, since a denoising step is essentially a gradient step on the true distribution's log-density, and we can trust it not to produce unreasonable samples; thus the denoising step is dominated by the denoiser's output. At $\ell = 1$ we even discard the current iterate and keep only the denoiser's output.

We visualize the convergence process in $\mathbb{R}^3$ in Figure 3.5. While we will develop some rigorous results about convergence later, we now give a closed-form example of the iterative denoising process applied to denoising a Gaussian data distribution.

*Example* 3.5 (Iterative denoising for a single Gaussian). To understand the above iterative denoising iteration intuitively, we apply it to denoise $\boldsymbol{x}_T$ defined in (3.2.1), where $\boldsymbol{x} \sim \mathcal{N}(\boldsymbol{\mu}, \boldsymbol{\Sigma})$. Using (3.2.42), we compute

$$\begin{aligned}
\hat{\boldsymbol{x}}_{t_{\ell-1}} - \boldsymbol{\mu} &= \left(1 - \frac{1}{\ell}\right)\hat{\boldsymbol{x}}_{t_\ell} + \frac{1}{\ell}\bar{\boldsymbol{x}}^\star(t_\ell, \hat{\boldsymbol{x}}_{t_\ell}) - \boldsymbol{\mu} && (3.2.43)\\
&= \left(1 - \frac{1}{\ell}\right)(\hat{\boldsymbol{x}}_{t_\ell} - \boldsymbol{\mu}) + \frac{1}{\ell}(\bar{\boldsymbol{x}}^\star(t_\ell, \hat{\boldsymbol{x}}_{t_\ell}) - \boldsymbol{\mu}). && (3.2.44)
\end{aligned}$$

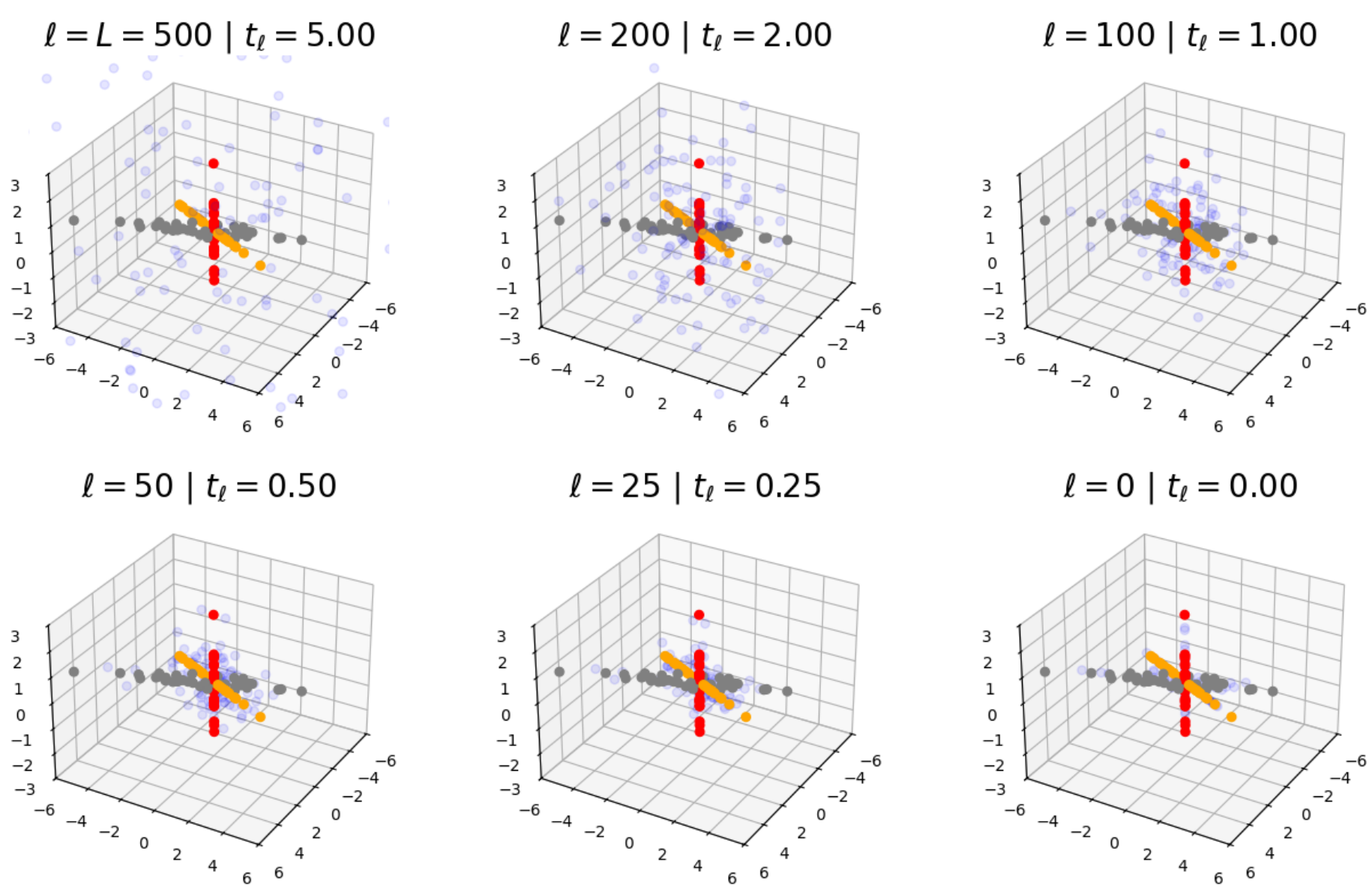

Figure 3.5: **Denoising a low-rank mixture of Gaussians.** Each figure shows samples from the true data distribution (gray, orange, red) and samples undergoing the denoising process (3.2.82) (light blue). At top left, the process has just started and the noise is very large. As the process continues, the noise is pushed further toward the support of the low-rank data distribution. Finally, in the bottom right, the generated samples are perfectly aligned with the support of the data and closely resemble samples drawn from the low-rank Gaussian mixture model.

Then, plugging in the form for $\bar{\boldsymbol{x}}^{\star}$ that we obtain from (3.2.22) and performing some tedious matrix algebra, we obtain

$$\boldsymbol{x}_{t_{\ell-1}} - \boldsymbol{\mu} = \boldsymbol{V} \begin{bmatrix} 1 - \frac{\ell T^2}{\lambda_1 L^2 + \ell^2 T^2} & & \\ & \ddots & \\ & & 1 - \frac{\ell T^2}{\lambda_D L^2 + \ell^2 T^2} \end{bmatrix} \boldsymbol{V}^{\top}(\boldsymbol{x}_{t_\ell} - \boldsymbol{\mu}). \quad (3.2.45)$$

Notice that all entries in this diagonal matrix are in $(0,1)$. Taking the $\ell^2$-norm and using a cheap yet tight upper bound, it holds

$$\|\boldsymbol{x}_{t_{\ell-1}} - \boldsymbol{\mu}\|_2 \leq \underbrace{\max_{i \in [D]} \left\{ 1 - \frac{\ell T^2}{\lambda_i L^2 + \ell^2 T^2} \right\}}_{c_{\ell,L}} \cdot \|\boldsymbol{x}_{t_\ell} - \boldsymbol{\mu}\|_2. \quad (3.2.46)$$

At first glance, this inequality seems to imply that the denoising iteration is *very close*, conceptually, to a *contraction mapping* such as power iteration (2.1.17) sending the iterate $\boldsymbol{x}_t$ to the mean $\boldsymbol{\mu}$, since $c_{\ell,L}$ are all in $(0,1)$ for any $\ell$ and $L$.

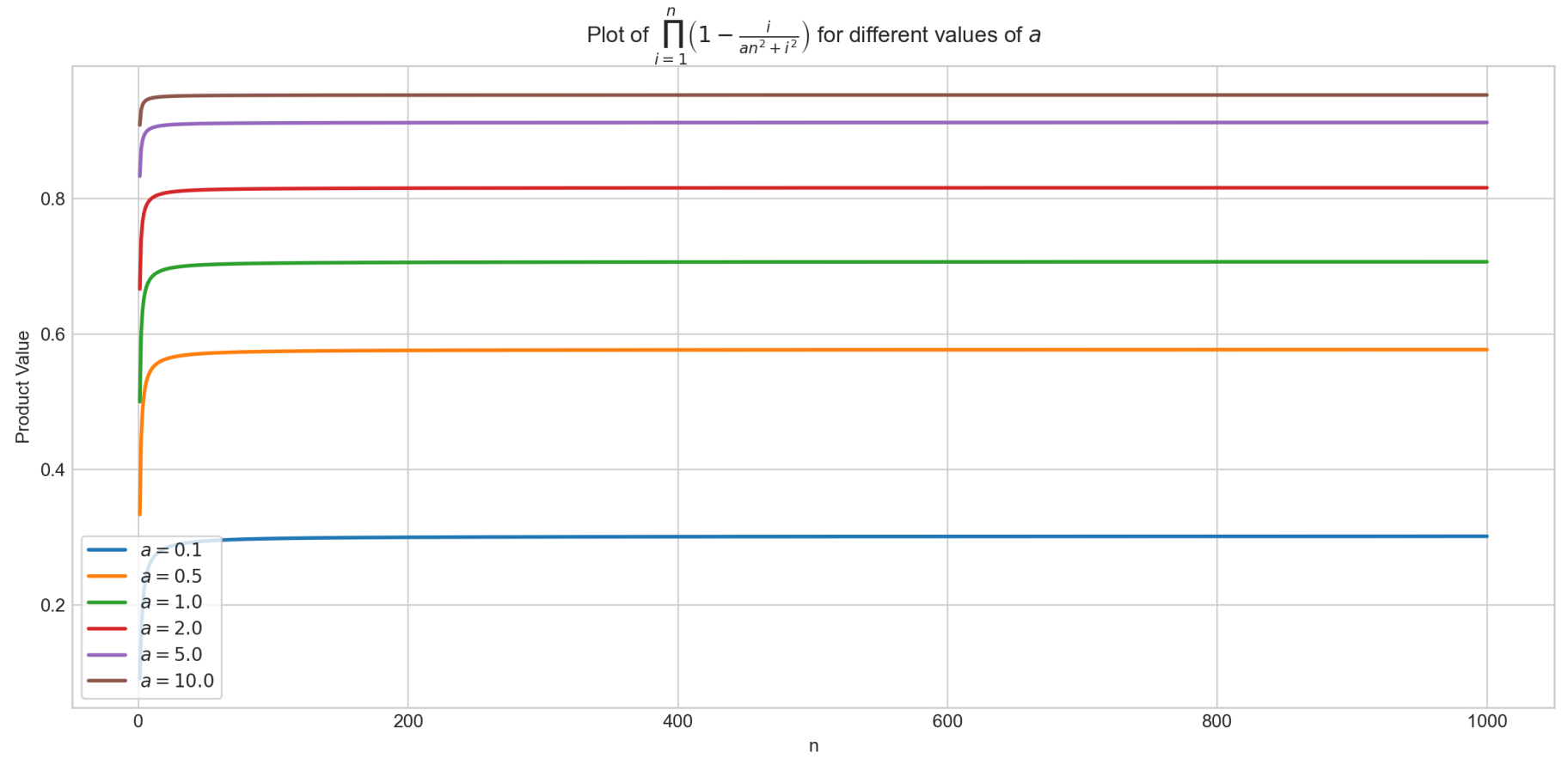

Figure 3.6: **The denoising iteration from Example 3.5 is not a contraction mapping.** Plot of the contraction coefficient products $\prod_{\ell=1}^{L} c_{\ell,L}$ in the case where the data are in $D = 1$, the time horizon $T = 1$, and the data variance $\lambda_1 = a$, for varying values of $a$. If the denoising iteration were (conceptually similar to) a contraction mapping, these products would go to 0, but instead they converge to non-zero values.

However, this is false. To see why, observe that by iterating (3.2.46) it holds

$$\|\boldsymbol{x}_0 - \boldsymbol{\mu}\|_2 \leq \prod_{\ell=1}^{L} c_{\ell,L} \cdot \|\boldsymbol{x}_T - \boldsymbol{\mu}\|_2. \tag{3.2.47}$$

If each denoising update had similar properties to a contraction mapping, then $\prod_{\ell=1}^{L} c_{\ell,L}$ would go to 0 as $L \to \infty$, and the iterate would converge to $\boldsymbol{\mu}$ extremely fast, rendering the sampler completely useless or vacuous. But in reality the denoising iteration is *actually not a contraction mapping* since $c_{\ell,L}$ changes with both the indices $\ell$ and $L$, so that $\prod_{\ell=1}^{L} c_{\ell,L}$ does not go to 0 as $L \to \infty$, as demonstrated in Figure 3.6 (it is also possible to compute the limit analytically). In reality, the sampler ensures that the terminal iterates spread out amongst the support of the data distribution, enabling actual sampling from the data distribution, which is quite unlike the objective of power iteration (for example). ■

Recall that we wanted to build the iterative denoising process to reduce the entropy of the iterates. While we did this in a roundabout way by inverting a process that adds entropy, it is now time to pay the piper and confirm that our iterative denoising process reduces the entropy.

**Theorem 3.3** (Simplified Version of Theorem B.3)**.** *Suppose that $(\boldsymbol{x}_t)_{t\in[0,T]}$ obeys (3.2.1). Then, under certain technical conditions on $\boldsymbol{x}$, for every $s < t$ with $s, t \in (0, T]$,*

$$h(\mathbb{E}[\boldsymbol{x}_s \mid \boldsymbol{x}_t]) < h(\boldsymbol{x}_t). \tag{3.2.48}$$

The full statement of the theorem and the proof itself require some technicality, so they are postponed to Section B.2.2.

### Computing the Denoiser

The last thing we discuss here is that many times, we will *not be able to compute* $\bar{\boldsymbol{x}}^\star(t,\cdot)$ for any $t$, since we do not have the distribution $p_t$. But we can try to *learn one from data*. Recall that the denoiser $\bar{\boldsymbol{x}}^\star$ is defined in (3.2.7) as minimizing the mean-squared error $\mathbb{E}\,\|\bar{\boldsymbol{x}}(t,\boldsymbol{x}_t)-\boldsymbol{x}\|_2^2$. We can use this mean-squared error as a loss or objective function to learn the denoiser. For example, we can parameterize $\bar{\boldsymbol{x}}(t,\cdot)$ by a neural network, writing it as $\bar{\boldsymbol{x}}_\theta(t,\cdot)$, and optimize the loss over the parameter space $\Theta$:

$$\min_{\theta\in\Theta}\mathbb{E}\,\|\bar{\boldsymbol{x}}_\theta(t,\boldsymbol{x}_t)-\boldsymbol{x}\|_2^2. \tag{3.2.49}$$

The solution to this optimization problem, implemented via gradient descent or a similar algorithm, will give us a $\bar{\boldsymbol{x}}_{\theta^\star}(t,\cdot)$ which is a good approximation to $\bar{\boldsymbol{x}}^\star(t,\cdot)$ (at least if the training works) and which we will use as our denoiser.

What is a good architecture for this neural network $\bar{\boldsymbol{x}}_{\theta^\star}(t,\cdot)$? To answer this question, we will first attempt to understand the case of data $\boldsymbol{x}$ that are *low-rank*, i.e., having low-rank support. The most canonical way to represent low-rank data is through a *low-rank Gaussian*, and so we start our exploration by assuming that $\boldsymbol{x}$ is drawn (approximately) from a low-rank Gaussian:

$$\boldsymbol{x}\sim\mathcal{N}(\boldsymbol{0},\boldsymbol{U}\boldsymbol{U}^\top), \tag{3.2.50}$$

where $\boldsymbol{U}\in\mathsf{O}(D,P)$ is a tall orthogonal matrix with $P\le D$ (using zero-mean for essentially notational convenience). The optimal denoiser under (3.2.50) is (from Example 3.3):

$$\bar{\boldsymbol{x}}^\star(t,\boldsymbol{x}_t)=\boldsymbol{U}\boldsymbol{U}^\top(\boldsymbol{U}\boldsymbol{U}^\top+t^2\boldsymbol{I})^{-1}\boldsymbol{x}_t. \tag{3.2.51}$$

The matrix $\boldsymbol{U}\boldsymbol{U}^\top+t^2\boldsymbol{I}$ is a low-rank perturbation of the full-rank matrix $t^2\boldsymbol{I}$, and thus ripe for simplification via the *Sherman-Morrison-Woodbury identity*, i.e., for matrices $\boldsymbol{A},\boldsymbol{C},\boldsymbol{U},\boldsymbol{V}$ such that $\boldsymbol{A}$ and $\boldsymbol{C}$ are invertible,

$$(\boldsymbol{A}+\boldsymbol{U}\boldsymbol{C}\boldsymbol{V})^{-1}=\boldsymbol{A}^{-1}-\boldsymbol{A}^{-1}\boldsymbol{U}(\boldsymbol{C}^{-1}+\boldsymbol{V}\boldsymbol{A}^{-1}\boldsymbol{U})^{-1}\boldsymbol{V}\boldsymbol{A}^{-1}. \tag{3.2.52}$$

We prove this identity in Exercise 3.3. For now we apply this identity with $\boldsymbol{A}=t^2\boldsymbol{I}$, $\boldsymbol{U}=\boldsymbol{U}$, $\boldsymbol{V}=\boldsymbol{U}^\top$, and $\boldsymbol{C}=\boldsymbol{I}$, obtaining

$$(\boldsymbol{U}\boldsymbol{U}^\top+t^2\boldsymbol{I})^{-1}=\frac{1}{t^2}\boldsymbol{I}-\frac{1}{t^4}\boldsymbol{U}\left(\boldsymbol{I}+\frac{1}{t^2}\boldsymbol{U}^\top\boldsymbol{U}\right)^{-1}\boldsymbol{U}^\top \tag{3.2.53}$$

$$=\frac{1}{t^2}\boldsymbol{I}-\frac{1}{t^4\left(1+\frac{1}{t^2}\right)}\boldsymbol{U}\boldsymbol{U}^\top \tag{3.2.54}$$

$$=\frac{1}{t^2}\left(\boldsymbol{I}-\frac{1}{1+t^2}\boldsymbol{U}\boldsymbol{U}^\top\right). \tag{3.2.55}$$

This implies

$$\boldsymbol{U}\boldsymbol{U}^\top(\boldsymbol{U}\boldsymbol{U}^\top + t^2\boldsymbol{I})^{-1} = \frac{1}{t^2}\boldsymbol{U}\boldsymbol{U}^\top\left(\boldsymbol{I} - \frac{1}{1+t^2}\boldsymbol{U}\boldsymbol{U}^\top\right) \tag{3.2.56}$$

$$= \frac{1}{t^2}\left(1 - \frac{1}{1+t^2}\right)\boldsymbol{U}\boldsymbol{U}^\top \tag{3.2.57}$$

$$= \frac{1}{1+t^2}\boldsymbol{U}\boldsymbol{U}^\top. \tag{3.2.58}$$

Our optimal denoiser is a (rescaled) linear projection onto the low-rank subspace supporting our data distribution. This is essentially the same thing as probabilistic PCA from Chapter 2 (cf. Section 2.1.4). We can formalize this intuition by considering a class of subspace-parameterized denoisers:

$$\bar{\boldsymbol{x}}_{\boldsymbol{V}}(t, \boldsymbol{x}_t) = \frac{1}{1+t^2}\boldsymbol{V}\boldsymbol{V}^\top\boldsymbol{x}_t, \tag{3.2.59}$$

and showing that the denoising problem $\min_{\boldsymbol{V}\in\mathsf{O}(D,P)} \mathbb{E}\,\|\bar{\boldsymbol{x}}_{\boldsymbol{V}}(t,\boldsymbol{x}_t) - \boldsymbol{x}\|_2^2$ obtains the same solution as probabilistic PCA (Section 2.1.4). Indeed, plugging our subspace denoiser into the training loss (3.2.7), we obtain

$$\min_{\boldsymbol{V}\in\mathsf{O}(D,P)} \mathbb{E}\,\|\bar{\boldsymbol{x}}_{\boldsymbol{V}}(t,\boldsymbol{x}_t) - \boldsymbol{x}\|_2^2 = \mathbb{E}\left\|\frac{1}{1+t^2}\boldsymbol{V}\boldsymbol{V}^\top(\boldsymbol{x}+t\boldsymbol{g}) - \boldsymbol{x}\right\|_2^2, \tag{3.2.60}$$

where the equality is due to (3.2.1). Conditioned on $\boldsymbol{x}$ (thus integrating over $\boldsymbol{g}$), we compute

$$\mathbb{E}_{\boldsymbol{g}}\left\|\frac{1}{1+t^2}\boldsymbol{V}\boldsymbol{V}^\top(\boldsymbol{x}+t\boldsymbol{g}) - \boldsymbol{x}\right\|_2^2 \tag{3.2.61}$$

$$= \left\|\frac{1}{1+t^2}\boldsymbol{V}\boldsymbol{V}^\top\boldsymbol{x} - \boldsymbol{x}\right\|_2^2 - \frac{t}{1+t^2}\,\mathbb{E}_{\boldsymbol{g}}\left\langle \boldsymbol{x} - \frac{1}{1+t^2}\boldsymbol{V}\boldsymbol{V}^\top\boldsymbol{x}, \boldsymbol{V}\boldsymbol{V}^\top\boldsymbol{g}\right\rangle \tag{3.2.62}$$

$$+ \frac{t^2}{(1+t^2)^2}\,\mathbb{E}_{\boldsymbol{g}}\left\|\boldsymbol{V}\boldsymbol{V}^\top\boldsymbol{g}\right\|_2^2 \tag{3.2.63}$$

$$= \left\|\frac{1}{1+t^2}\boldsymbol{V}\boldsymbol{V}^\top\boldsymbol{x} - \boldsymbol{x}\right\|_2^2 + \frac{t^2 P}{(1+t^2)^2} \tag{3.2.64}$$

where the second equality follows from $\boldsymbol{g} \sim \mathcal{N}(\boldsymbol{0}, \boldsymbol{I})$ and the fact that $\boldsymbol{V} \in \mathsf{O}(D,P)$ so that $\mathbb{E}_{\boldsymbol{g}}\,\|\boldsymbol{V}\boldsymbol{V}^\top\boldsymbol{g}\|_2^2 = \mathbb{E}_{\boldsymbol{g}}\,\|\boldsymbol{V}^\top\boldsymbol{g}\|_2^2 = P$. Therefore, the denoising problem is equivalent to

$$\min_{\boldsymbol{V}\in\mathsf{O}(D,P)}\left\{\mathbb{E}\left\|\frac{1}{1+t^2}\boldsymbol{V}\boldsymbol{V}^\top\boldsymbol{x} - \boldsymbol{x}\right\|_2^2 \right. \tag{3.2.65}$$

$$\left. = \mathbb{E}\,\|\boldsymbol{x}\|_2^2 + \left(\frac{1}{(1+t^2)^2} - \frac{2}{1+t^2}\right)\mathbb{E}\,\|\boldsymbol{V}^\top\boldsymbol{x}\|_2^2\right\}, \tag{3.2.66}$$

which in turn is equivalent to the problem

$$\max_{\boldsymbol{V}\in\mathsf{O}(D,P)} \mathbb{E}\,\|\boldsymbol{V}^\top \boldsymbol{x}\|_2^2 \tag{3.2.67}$$

which was shown in Section 2.1.1 to be equivalent to the problem

$$\min_{\boldsymbol{V}\in\mathsf{O}(D,P)} \mathbb{E}\,\|\boldsymbol{x} - \boldsymbol{V}\boldsymbol{V}^\top \boldsymbol{x}\|_2^2, \tag{3.2.68}$$

i.e., the probabilistic PCA formulation (2.1.30).

Thus, if we were to assume our data were drawn from a low-rank Gaussian, our denoising algorithm would essentially consist of many PCA iterations in a row applied to progressively less-noisy data. While this method sounds very simple, it can actually produce complex and relatively powerful denoising behaviors when applied to practical data, as demonstrated by PCANet [CJG+15]. Nevertheless, to obtain a more general denoising operator that may work on complex and multi-modal data distributions, we will consider a broader class of data distributions: the ubiquitous class of *low-rank Gaussian mixture models*, whose denoiser we computed in Example 3.2 and Example 3.3. In the context of denoising, since Gaussian mixture models can approximate a very broad class of distributions arbitrarily well, in practice optimizing among the class of denoisers for Gaussian mixture models can potentially give us something close to the optimal denoiser for the real data distribution.

Formally, we now assume that $\boldsymbol{x}$ is drawn from a low-rank Gaussian mixture model $\frac{1}{K}\sum_{k=1}^{K}\mathcal{N}(\boldsymbol{0}, \boldsymbol{U}_k\boldsymbol{U}_k^\top)$, where $\boldsymbol{U}_k \in \mathsf{O}(D,P)$ are tall orthogonal matrices whose columns span the $P$-dimensional subspaces that support the $K$ low-rank Gaussian components. We write

$$\boldsymbol{x} \sim \frac{1}{K}\sum_{k=1}^{K}\mathcal{N}(\boldsymbol{0}, \boldsymbol{U}_k\boldsymbol{U}_k^\top). \tag{3.2.69}$$

Then the optimal denoiser under (3.2.1) is (from Example 3.3)

$$\bar{\boldsymbol{x}}^\star(t, \boldsymbol{x}_t) = \sum_{k=1}^{K}\left\{\frac{\varphi(\boldsymbol{x}_t; \boldsymbol{0}, \boldsymbol{U}_k\boldsymbol{U}_k^\top + t^2\boldsymbol{I})}{\sum_{i=1}^{K}\varphi(\boldsymbol{x}_t; \boldsymbol{0}, \boldsymbol{U}_i\boldsymbol{U}_i^\top + t^2\boldsymbol{I})}\right. \tag{3.2.70}$$

$$\left.\cdot\left(\boldsymbol{U}_k\boldsymbol{U}_k^\top(\boldsymbol{U}_k\boldsymbol{U}_k^\top + t^2\boldsymbol{I})^{-1}\boldsymbol{x}_t\right)\right\}. \tag{3.2.71}$$

Applying the same Sherman-Morrison-Woodbury identity as in (3.2.55), we obtain

$$(\boldsymbol{U}_k\boldsymbol{U}_k^\top + t^2\boldsymbol{I})^{-1} = \frac{1}{t^2}\left(\boldsymbol{I} - \frac{1}{1+t^2}\boldsymbol{U}_k\boldsymbol{U}_k^\top\right). \tag{3.2.72}$$

We can then compute the posterior probabilities as follows. Since the $\boldsymbol{U}_k$ are orthogonal, $\det(\boldsymbol{U}_k\boldsymbol{U}_k^\top + t^2\boldsymbol{I})$ is the same for every $k$. Thus

$$\frac{\varphi(\boldsymbol{x}_t; \boldsymbol{0}, \boldsymbol{U}_k\boldsymbol{U}_k^\top + t^2\boldsymbol{I})}{\sum_{i=1}^{K}\varphi(\boldsymbol{x}_t; \boldsymbol{0}, \boldsymbol{U}_i\boldsymbol{U}_i^\top + t^2\boldsymbol{I})} \tag{3.2.73}$$

$$
\begin{aligned}
&= \frac{\exp\big(-\frac{1}{2}\boldsymbol{x}_t^\top(\boldsymbol{U}_k\boldsymbol{U}_k^\top + t^2\boldsymbol{I})^{-1}\boldsymbol{x}_t\big)}{\sum_{i=1}^K \exp\big(-\frac{1}{2}\boldsymbol{x}_t^\top(\boldsymbol{U}_i\boldsymbol{U}_i^\top + t^2\boldsymbol{I})^{-1}\boldsymbol{x}_t\big)} && (3.2.74)\\
&= \frac{\exp\Big(-\frac{1}{2t^2}\boldsymbol{x}_t^\top\Big(\boldsymbol{I} - \frac{1}{1+t^2}\boldsymbol{U}_k\boldsymbol{U}_k^\top\Big)\boldsymbol{x}_t\Big)}{\sum_{i=1}^K \exp\Big(-\frac{1}{2t^2}\boldsymbol{x}_t^\top\Big(\boldsymbol{I} - \frac{1}{1+t^2}\boldsymbol{U}_i\boldsymbol{U}_i^\top\Big)\boldsymbol{x}_t\Big)} && (3.2.75)\\
&= \frac{\exp\Big(-\frac{1}{2t^2}\|\boldsymbol{x}_t\|_2^2 + \frac{1}{2t^2(1+t^2)}\|\boldsymbol{U}_k^\top\boldsymbol{x}_t\|_2^2\Big)}{\sum_{i=1}^K \exp\Big(-\frac{1}{2t^2}\|\boldsymbol{x}_t\|_2^2 + \frac{1}{2t^2(1+t^2)}\|\boldsymbol{U}_i^\top\boldsymbol{x}_t\|_2^2\Big)} && (3.2.76)\\
&= \frac{\exp\big(-\frac{1}{2t^2}\|\boldsymbol{x}_t\|_2^2\big)\exp\Big(\frac{1}{2t^2(1+t^2)}\|\boldsymbol{U}_k^\top\boldsymbol{x}_t\|_2^2\Big)}{\exp\big(-\frac{1}{2t^2}\|\boldsymbol{x}_t\|_2^2\big)\sum_{i=1}^K \exp\Big(\frac{1}{2t^2(1+t^2)}\|\boldsymbol{U}_i^\top\boldsymbol{x}_t\|_2^2\Big)} && (3.2.77)\\
&= \frac{\exp\Big(\frac{1}{2t^2(1+t^2)}\|\boldsymbol{U}_k^\top\boldsymbol{x}_t\|_2^2\Big)}{\sum_{i=1}^K \exp\Big(\frac{1}{2t^2(1+t^2)}\|\boldsymbol{U}_i^\top\boldsymbol{x}_t\|_2^2\Big)}. && (3.2.78)
\end{aligned}
$$

This is a softmax operation weighted by the projection of $\boldsymbol{x}_t$ onto each subspace measured by $\|\boldsymbol{U}_i^\top\boldsymbol{x}_t\|_2$ (tempered by the temperature $2t^2(1+t^2)$). Meanwhile, the component denoisers can be written in the same way as (3.2.58) to obtain

$$
\boldsymbol{U}_k\boldsymbol{U}_k^\top(\boldsymbol{U}_k\boldsymbol{U}_k^\top + t^2\boldsymbol{I})^{-1}\boldsymbol{x}_t = \frac{1}{1+t^2}\boldsymbol{U}_k\boldsymbol{U}_k^\top\boldsymbol{x}_t. \tag{3.2.79}
$$

Putting these together, we have

$$
\bar{\boldsymbol{x}}^\star(t, \boldsymbol{x}_t) = \frac{1}{1+t^2}\sum_{k=1}^K \frac{\exp\Big(\frac{1}{2t^2(1+t^2)}\|\boldsymbol{U}_k^\top\boldsymbol{x}_t\|_2^2\Big)}{\sum_{i=1}^K \exp\Big(\frac{1}{2t^2(1+t^2)}\|\boldsymbol{U}_i^\top\boldsymbol{x}_t\|_2^2\Big)}\boldsymbol{U}_k\boldsymbol{U}_k^\top\boldsymbol{x}_t, \tag{3.2.80}
$$

i.e., a projection of $\boldsymbol{x}_t$ onto each of the $K$ subspaces, weighted by a softmax of a quadratic function of $\boldsymbol{x}_t$. This functional form is similar to an *attention mechanism* in a transformer architecture! As we will see in Chapter 5, this is no coincidence at all; the deep link between denoising and lossy compression (to be covered in Section 4.1) makes transformer denoisers so effective in practice. Thus, our Gaussian mixture model theory motivates the use of transformer-like neural networks for denoising.

Overall, the learned denoiser architecture corresponds to an (implicit parametric) encoding scheme of the given data, since it can be used to denoise/project onto the data distribution. Training a denoiser is equivalent to finding a better coding scheme, and this partially implements approach A2 (i.e., pursuing a distribution by searching for schemes with better coding rates) at the end of Section 3.1.3. In the sequel, we discuss how to implement the other approach.

### 3.2.2 Sampling a Distribution via Iterative Denoising

Remember that at the end of Section 3.1.3, we discussed two approaches for pursuing a distribution with low-dimensional structure. The first approach (A1) starts with a normal distribution of high entropy and gradually reduces its entropy until it reaches the data distribution. We call this procedure *sampling*, since we generate new samples. It is now time to discuss how to implement this with the toolkit we have built.

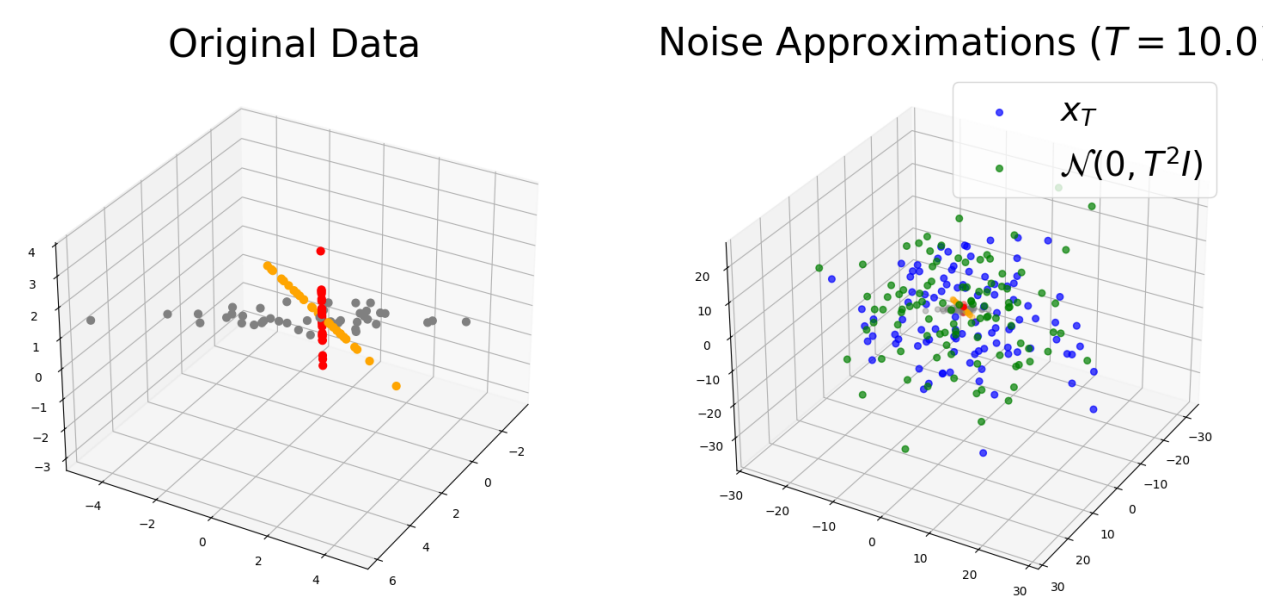

Figure 3.7: **Visualizing $\boldsymbol{x}_T$ versus $\mathcal{N}(\mathbf{0}, T^2\boldsymbol{I})$.** *Left:* A plot of Gaussian mixture model data $\boldsymbol{x}$. *Right:* A plot of $\boldsymbol{x}$ as well as $\boldsymbol{x}_T$ and an independent sample of $\mathcal{N}(\mathbf{0}, T^2\boldsymbol{I})$, for $T = 10$. On the right plot, $\boldsymbol{x}$ is plotted in the same colors as the left; however, samples from $\boldsymbol{x}_T$ and $\mathcal{N}(\mathbf{0}, T^2\boldsymbol{I})$ are both much larger, on average, than samples from $\boldsymbol{x}$, and so it appears much smaller because of the scaling. Despite this large blow-up, we clearly observe the similarities in the distributions of $\boldsymbol{x}_T$ and $\mathcal{N}(\mathbf{0}, T^2\boldsymbol{I})$.

We know how to denoise very noisy samples $\boldsymbol{x}_T$ to obtain approximations $\hat{\boldsymbol{x}}$ whose distributions resemble that of the target random variable $\boldsymbol{x}$. Yet the sampling procedure requires us to start from a template distribution that has *no* influence from the distribution of $\boldsymbol{x}$ and to use the denoiser to guide the iterates toward the distribution of $\boldsymbol{x}$. How can we do this? One way is motivated as follows:

$$\frac{\boldsymbol{x}_T}{T} = \frac{\boldsymbol{x} + T\boldsymbol{g}}{T} = \frac{\boldsymbol{x}}{T} + \boldsymbol{g} \xrightarrow{T \to +\infty} \boldsymbol{g} \sim \mathcal{N}(\mathbf{0}, \boldsymbol{I}). \tag{3.2.81}$$

Thus, $\boldsymbol{x}_T \approx \mathcal{N}(\mathbf{0}, T^2\boldsymbol{I})$. This approximation is quite good for almost all practical distributions, and is visualized in Figure 3.7.

So, discretizing $[0, T]$ into $0 = t_0 < t_1 < \cdots < t_L = T$ uniformly using $t_\ell = T\ell/L$ (as in the previous section), one possible way to sample from pure noise is:

- Sample $\hat{\boldsymbol{x}}_T \sim \mathcal{N}(\mathbf{0}, T^2\boldsymbol{I})$ (independently of everything else).
- Run the denoising iteration as in Section 3.2.1, i.e.,

$$\hat{\boldsymbol{x}}_{t_{\ell-1}} = \left(1 - \frac{1}{\ell}\right)\hat{\boldsymbol{x}}_{t_\ell} + \frac{1}{\ell}\bar{\boldsymbol{x}}^\star(t_\ell, \hat{\boldsymbol{x}}_{t_\ell}). \tag{3.2.82}$$

- Output $\hat{\boldsymbol{x}} = \hat{\boldsymbol{x}}_0$.

This is conceptually all there is behind *diffusion models*, which transform noise into data samples in accordance with the first desideratum. However, a few steps remain before we obtain models that can actually sample from real data distributions like images under practical resource constraints. In the sequel, we introduce and motivate several such steps.

**Step 1: Different discretizations.** The first step is motivated by the observation that *we do not need to spend so many denoising iterations* at large $t$. Figure 3.5 shows that the first 200 or 300 iterations out of the 500 iterations of the sampling process are spent contracting the noise toward the data distribution as a whole, before the remaining iterations push the samples toward a subspace. Given a fixed iteration count $L$, this suggests that we should allocate more timesteps $t_\ell$ near $t = 0$ than near $t = T$. During sampling (and training), we can therefore use an alternative discretization of $[0, T]$ into $0 \leq t_0 < t_1 < \cdots < t_L \leq T$, such as an *exponential discretization*:

$$t_\ell = C_1(e^{C_2 \ell} - 1), \qquad \forall \ell \in \{0, 1, \ldots, L\} \tag{3.2.83}$$

where $C_1, C_2 > 0$ are constants that can be tuned for optimal performance in practice; theoretical analyses often specify such optimal constants as well. The denoising/sampling iteration then becomes

$$\hat{\boldsymbol{x}}_{t_{\ell-1}} \doteq \frac{t_{\ell-1}}{t_\ell} \hat{\boldsymbol{x}}_{t_\ell} + \left(1 - \frac{t_{\ell-1}}{t_\ell}\right) \bar{\boldsymbol{x}}^\star(t_\ell, \hat{\boldsymbol{x}}_{t_\ell}), \tag{3.2.84}$$

with $\hat{\boldsymbol{x}}_{t_L} \sim \mathcal{N}(\boldsymbol{0}, t_L^2 \boldsymbol{I})$.

**Step 2: Different noise models.** The second step is to consider slightly different models compared to (3.2.1). The basic motivation for this is as follows. In practice, the noise distribution $\mathcal{N}(\boldsymbol{0}, t_L^2 \boldsymbol{I})$ becomes an increasingly poor estimate of the true covariance in high dimensions, i.e., (3.2.81) becomes an increasingly poor approximation, especially with anisotropic high-dimensional data. The increased distance between $\mathcal{N}(\boldsymbol{0}, t_L^2 \boldsymbol{I})$ and the true distribution of $\boldsymbol{x}_{t_L}$ may cause the denoiser to perform worse in such circumstances. Theoretically, $\boldsymbol{x}_{t_L}$ never converges to any distribution as $t_L$ increases, so this setup is difficult to analyze end-to-end. In this case, our remedy is to *simultaneously add noise and shrink the contribution of $\boldsymbol{x}$, such that $\boldsymbol{x}_T$ converges as $T \to \infty$*. The rate of added noise is denoted $\sigma \colon [0, T] \to \mathbb{R}_{\geq 0}$, and the rate of shrinkage is denoted $\alpha \colon [0, T] \to \mathbb{R}_{\geq 0}$, such that $\sigma$ is *increasing* and $\alpha$ is (not strictly) *decreasing*, and

$$\boldsymbol{x}_t \doteq \alpha_t \boldsymbol{x} + \sigma_t \boldsymbol{g}, \qquad \forall t \in [0, T]. \tag{3.2.85}$$

The previous setup has $\alpha_t = 1$ and $\sigma_t = t$, and this is called the *variance-exploding (VE) process*. A popular choice that decreases the contribution of $\boldsymbol{x}$, as we described originally, has $T = 1$ (so that $t \in [0, 1]$), $\alpha_t = \sqrt{1 - t^2}$ and $\sigma_t = t$; this is the *variance-preserving (VP) process*. Note that under the VP process, $\boldsymbol{x}_1 \sim \mathcal{N}(\boldsymbol{0}, \boldsymbol{I})$ exactly, so we can just sample from this standard distribution and iteratively denoise. As a result, the VP process is much easier to analyze theoretically and more stable empirically.[6]

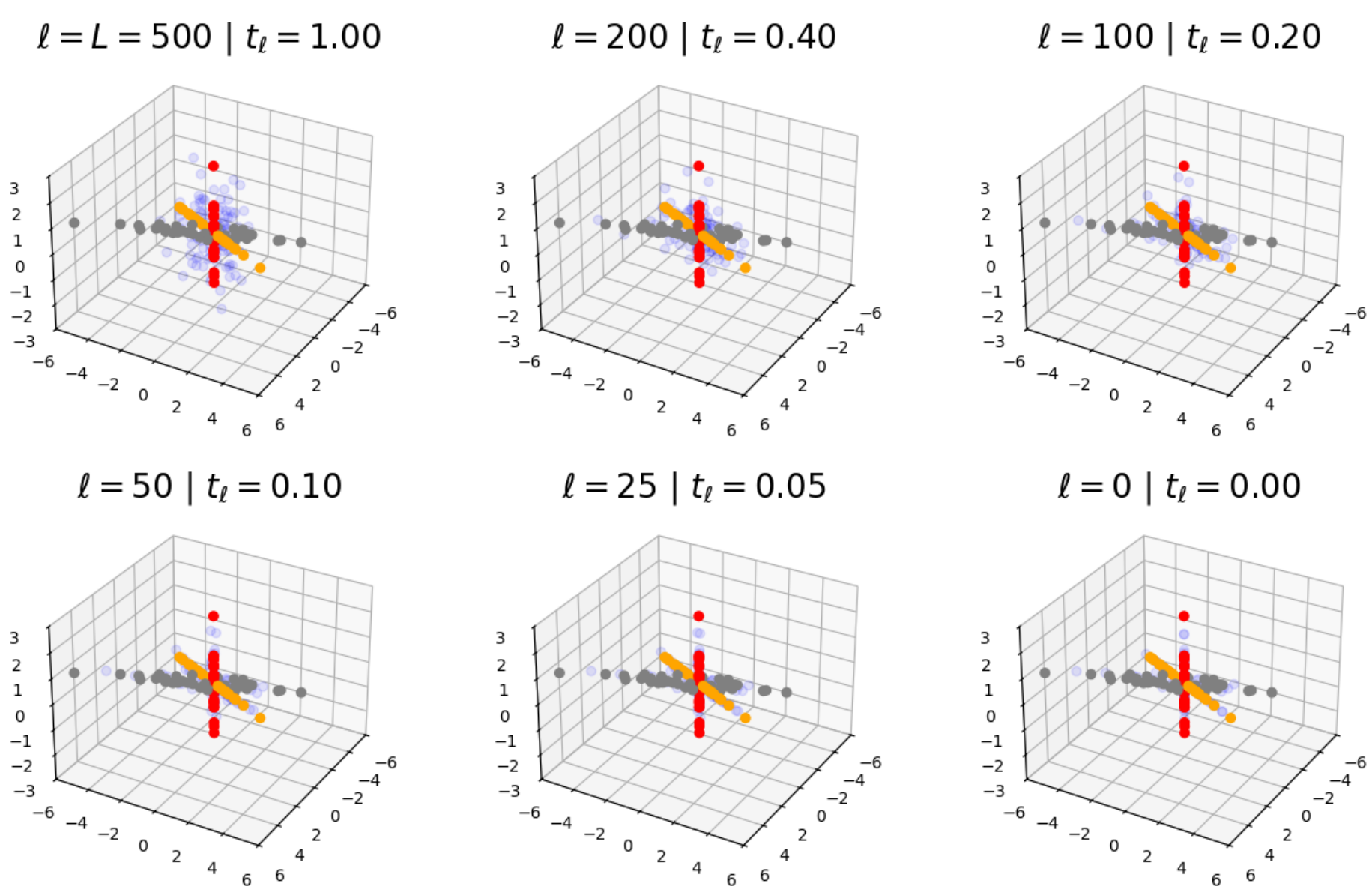

Figure 3.8: **Denoising a mixture of Gaussians using the VP diffusion process.** We use the same figure setup and data distribution as Figure 3.5. Note that compared to Figure 3.5, the noise distribution is much more concentrated around the origin.

With this more general setup, Tweedie's formula (3.2.23) becomes

$$\mathbb{E}[\boldsymbol{x} \mid \boldsymbol{x}_t] = \frac{1}{\alpha_t}\left(\boldsymbol{x}_t + \sigma_t^2 \nabla \log p_t(\boldsymbol{x}_t)\right). \tag{3.2.86}$$

The denoising iteration (3.2.84) becomes

$$\hat{\boldsymbol{x}}_{t_{\ell-1}} = \frac{\sigma_{t_{\ell-1}}}{\sigma_{t_\ell}}\hat{\boldsymbol{x}}_{t_\ell} + \left(\alpha_{t_{\ell-1}} - \frac{\sigma_{t_{\ell-1}}}{\sigma_{t_\ell}}\alpha_{t_\ell}\right)\bar{\boldsymbol{x}}^\star(t_\ell, \hat{\boldsymbol{x}}_{t_\ell}). \tag{3.2.87}$$

*Example* 3.6 (Gaussian mixture model)*.* For a Gaussian mixture model (see Example 3.2), its denoiser (3.2.19) becomes

$$\bar{\boldsymbol{x}}^\star(t, \boldsymbol{x}_t) = \sum_{k=1}^{K}\left\{\frac{\pi_k \varphi(\boldsymbol{x}_t; \alpha_t \boldsymbol{\mu}_k, \alpha_t^2 \boldsymbol{\Sigma}_k + \sigma_t^2 \boldsymbol{I})}{\sum_{i=1}^{K} \pi_i \varphi(\boldsymbol{x}_t; \alpha_t \boldsymbol{\mu}_i, \alpha_t^2 \boldsymbol{\Sigma}_i + \sigma_t^2 \boldsymbol{I})}\right. \tag{3.2.88}$$
$$\left. \cdot \left(\boldsymbol{\mu}_k + \alpha_t \boldsymbol{\Sigma}_k (\alpha_t^2 \boldsymbol{\Sigma}_k + \sigma_t^2 \boldsymbol{I})^{-1}(\boldsymbol{x}_t - \alpha_t \boldsymbol{\mu}_k)\right)\right\}.$$

[6] Why use the whole $\alpha, \sigma$ setup? As we will see in Exercise 3.5, it encapsulates and unifies many proposed processes, including the recently popular so-called *flow matching* process. Despite this, the VE and VP processes are still the most popular empirically and theoretically (so far), and so we will consider them in this Section.

Figure 3.8 demonstrates iterations of the sampling procedure. ■

Note that the denoising iteration (3.2.87) gives a sampling algorithm called the DDIM ("Denoising Diffusion Implicit Model") sampler [SME20], and is one of the most popular sampling algorithms used in diffusion models today. We summarize it in Algorithm 3.1.

---

**Algorithm 3.1** Sampling using a denoiser.

**Input:** An ordered list of timesteps $0 \leq t_0 < \cdots < t_L \leq T$ to use for sampling.
**Input:** A denoiser $\bar{\boldsymbol{x}} \colon \{t_\ell\}_{\ell=1}^{L} \times \mathbb{R}^D \to \mathbb{R}^D$.
**Input:** Scale and noise level functions $\alpha, \sigma \colon \{t_\ell\}_{\ell=0}^{L} \to \mathbb{R}_{\geq 0}$.
**Output:** A sample $\hat{\boldsymbol{x}}$, approximately from the distribution of $\boldsymbol{x}$.

1: **function** DDIMSampler($\bar{\boldsymbol{x}}, (t_\ell)_{\ell=0}^{L}$)
2: Initialize $\hat{\boldsymbol{x}}_{t_L} \sim$ approximate distribution of $\boldsymbol{x}_{t_L}$
`# VP` $\implies \mathcal{N}(\mathbf{0}, \boldsymbol{I})$, `VE` $\implies \mathcal{N}(\mathbf{0}, t_L^2 \boldsymbol{I})$.
3: **for** $\ell = L, L-1, \ldots, 1$ **do**
4: Compute

$$\hat{\boldsymbol{x}}_{t_{\ell-1}} \doteq \frac{\sigma_{t_{\ell-1}}}{\sigma_{t_\ell}} \hat{\boldsymbol{x}}_{t_\ell} + \left( \alpha_{t_{\ell-1}} - \frac{\sigma_{t_{\ell-1}}}{\sigma_{t_\ell}} \alpha_{t_\ell} \right) \bar{\boldsymbol{x}}(t_\ell, \hat{\boldsymbol{x}}_{t_\ell})$$

5: **end for**
6: **return** $\hat{\boldsymbol{x}}_{t_0}$
7: **end function**

---

**Step 3: Optimizing training pipelines.** If we use the procedure dictated by Section 3.2.1 to learn a separate denoiser $\bar{\boldsymbol{x}}(t, \cdot)$ for each time $t$ to be used in the sampling algorithm, *we would have to learn $L$ separate denoisers!* This is highly inefficient—the usual case is that we have to train $L$ separate neural networks, taking up $L$ times the training time and storage memory, and then be locked into using these timesteps for sampling forever. Instead, we can *train a single neural network* to denoise across all times $t$, taking as input the continuous variables $\boldsymbol{x}_t$ and $t$ (instead of just $\boldsymbol{x}_t$ before). Mechanically, our training loss averages over $t$, i.e., solves the following problem:

$$\min_{\theta} \mathbb{E} \, \|\bar{\boldsymbol{x}}_\theta(\tau, \boldsymbol{x}_\tau) - \boldsymbol{x}\|_2^2, \tag{3.2.89}$$

where $\tau \sim \mathcal{U}([0, T])$, i.e., is drawn uniformly from $[0, T]$. Similar to Step 1, where we used more timesteps closer to $t = 0$ to ensure a better sampling process, we may want to ensure that the denoiser is higher quality closer to $t = 0$, and thereby *weight the loss* so that $t$ near 0 has higher weight.[7] Letting $w_t$ be the

[7] In practice, the distribution of sampled $t$ may also be non-uniform, in order to ensure (in conjunction with the weighting scheme) that the Monte Carlo estimate for the loss has small variance, greatly improving training stability.

weight at time $t$, the weighted loss would look like

$$\min_{\theta} \mathbb{E}[w_\tau \|\bar{\boldsymbol{x}}_\theta(\tau, \boldsymbol{x}_\tau) - \boldsymbol{x}\|_2^2]. \tag{3.2.90}$$

One reasonable choice of weight in practice is $w_t = \alpha_t^2/\sigma_t^2$. The precise reason will be covered in the next paragraph, but generally it serves to up-weight the losses corresponding to $t$ near 0 while still remaining reasonably numerically stable. Also, of course, we cannot compute the expectation in practice, so we use the most straightforward Monte Carlo average to estimate it. The series of changes made here have several conceptual and computational benefits: we do not need to train multiple denoisers, we can train on one set of timesteps and sample using a subset (or others entirely), etc. The full pipeline is discussed in Algorithm 3.2. Certainly, this pipeline works in practice; we discuss some results in Sections 8.6 to 8.10.

**Step 4: Changing the estimation target.** It is common to instead reorient the whole denoising pipeline around *noise predictors*, i.e., estimates of $\mathbb{E}[\boldsymbol{g} \mid \boldsymbol{x}_t]$. In practice, noise predictors are slightly easier to train because their output is (almost) always of comparable size to a Gaussian random variable, so training is more numerically stable, but they may require larger models since they predict a full-dimensional target instead of potentially low-dimensional data [LH25]. By (3.2.85) we have

$$\boldsymbol{x}_t = \alpha_t \mathbb{E}[\boldsymbol{x} \mid \boldsymbol{x}_t] + \sigma_t \mathbb{E}[\boldsymbol{g} \mid \boldsymbol{x}_t] \implies \mathbb{E}[\boldsymbol{g} \mid \boldsymbol{x}_t] = \frac{1}{\sigma_t}\left(\boldsymbol{x}_t - \alpha_t \mathbb{E}[\boldsymbol{x} \mid \boldsymbol{x}_t]\right), \tag{3.2.91}$$

so any predictor for $\boldsymbol{x}$ can be turned into a predictor for $\boldsymbol{g}$ using the above relation, i.e.,

$$\bar{\boldsymbol{g}}(t, \boldsymbol{x}_t) = \frac{1}{\sigma_t}\boldsymbol{x}_t - \frac{\alpha_t}{\sigma_t}\bar{\boldsymbol{x}}(t, \boldsymbol{x}_t), \tag{3.2.92}$$

and vice-versa. Thus a good network for estimating $\bar{\boldsymbol{g}}$ is the same as a good network for estimating $\bar{\boldsymbol{x}}$ *plus a residual connection* (as seen in, e.g., transformers). Their losses are also the same as the denoiser, up to the factor of $\alpha_t/\sigma_t$, i.e.,

$$\mathbb{E}[w_\tau \|\boldsymbol{g} - \bar{\boldsymbol{g}}(\tau, \boldsymbol{x}_\tau)\|_2^2] = \mathbb{E}\left[w_\tau \frac{\alpha_\tau^2}{\sigma_\tau^2}\|\boldsymbol{x} - \bar{\boldsymbol{x}}(\tau, \boldsymbol{x}_\tau)\|_2^2\right], \tag{3.2.93}$$

recalling that $\tau \sim \mathcal{U}([0, T])$. Other targets have been proposed for different tasks, e.g., $\mathbb{E}[\boldsymbol{v}_t \mid \boldsymbol{x}_t]$ where $\boldsymbol{v}_t = \frac{\mathrm{d}}{\mathrm{d}t}\boldsymbol{x}_t$ (called $v$-prediction or velocity prediction), etc. This turns out to be equivalent to denoising and noise prediction, but also fundamentally useful for generalizations of diffusion such as flow matching, which we will discuss in Remark 3.3.

### Sampling Error Rate of the Denoising Process

We have made many changes to our original platonic noising/denoising process. To assure ourselves that the new process still works in practice, we can compute

**Algorithm 3.2** Learning a denoiser from data.

**Input:** Dataset $\mathcal{D} \subseteq \mathbb{R}^D$.
**Input:** An ordered list of timesteps $0 \leq t_0 < \cdots < t_L \leq T$ to use for sampling.
**Input:** A weighting function $w \colon \{t_\ell\}_{\ell=1}^L \to \mathbb{R}_{\geq 0}$.
**Input:** Scale and noise level functions $\alpha, \sigma \colon \{t_\ell\}_{\ell=0}^L \to \mathbb{R}_{\geq 0}$.
**Input:** A parameter space $\Theta$ and a denoiser architecture $\bar{\boldsymbol{x}}_\theta$.
**Input:** An optimization algorithm for the parameters.
**Input:** The number of optimization iterations $M$.
**Input:** The number of Monte Carlo draws $N$ per training step.
**Output:** A trained denoiser $\bar{\boldsymbol{x}}_{\theta^\star}$.

1: **function** TrainDenoiser($\mathcal{D}, \Theta$)
2: Initialize $\theta^{(1)} \in \Theta$
3: **for** $m \in [M]$ **do**
4: **for** $i \in [N]$ **do**
`# Draw a sample from the dataset.`
5: $\boldsymbol{x}^{(m,i)} \sim \mathcal{D}$

`# Sample a timestep.`
6: $t^{(m,i)} \overset{\text{i.i.d.}}{\sim} \mathcal{U}(\{t_\ell\}_{\ell=1}^L)$

`# Compute weights and scales.`
7: $(\alpha^{(m,i)}, \sigma^{(m,i)}, w^{(m,i)}) = (\alpha_{t^{(m,i)}}, \sigma_{t^{(m,i)}}, w_{t^{(m,i)}})$

`# Sample a noise vector.`
8: $\boldsymbol{g}^{(m,i)} \overset{\text{i.i.d.}}{\sim} \mathcal{N}(\boldsymbol{0}, \boldsymbol{I})$

`# Compute the noised sample.`
9: $\boldsymbol{x}_t^{(m,i)} \doteq \alpha^{(m,i)} \boldsymbol{x}^{(m,i)} + \sigma^{(m,i)} \boldsymbol{g}^{(m,i)}$
10: **end for**
`# Compute the training loss.`
11: $\hat{\mathcal{L}}^{(m)} \doteq \dfrac{1}{N} \sum_{i=1}^{N} w^{(m,i)} \| \boldsymbol{x}^{(m,i)} - \bar{\boldsymbol{x}}_{\theta^{(m)}}(t^{(m,i)}, \boldsymbol{x}_t^{(m,i)}) \|_2^2$

`# Update the model parameters.`
12: $\theta^{(m+1)} \doteq \texttt{OptimUpdate}^{(m)}(\theta^{(m)}, \nabla_{\theta^{(m)}} \hat{\mathcal{L}}^{(m)})$
13: **end for**
`# Return the trained denoiser.`
14: **return** $\bar{\boldsymbol{x}}_{\theta^{(M+1)}}$
15: **end function**

numerical examples (such as Figure 3.8). To assure ourselves that it is theoretically sound, we can prove a *bound on the error rate* for the sampling algorithm, which shows that the error rate is small. We now furnish such a rate from the literature, which shows that the output distribution of the sampler converges in the so-called *total variation (TV) distance* to the true distribution. The TV distance between two random variables $\boldsymbol{x}$ and $\boldsymbol{y}$ is defined as:[8]

$$\mathsf{TV}(\boldsymbol{x}, \boldsymbol{y}) \doteq \sup_{A \subseteq \mathbb{R}^d} |\mathbb{P}[\boldsymbol{x} \in A] - \mathbb{P}[\boldsymbol{y} \in A]| \,. \tag{3.2.94}$$

If $\boldsymbol{x}$ and $\boldsymbol{y}$ are very close (uniformly), then the supremum will be small. So the TV distance measures the closeness of random variables. (It is indeed a metric, as the name suggests; the proof is an exercise.)

**Theorem 3.4** ([LY24] Theorem 1, Simplified)**.** *Suppose that* $\mathbb{E}\,\|\boldsymbol{x}\|_2 < \infty$*. If* $\boldsymbol{x}$ *is denoised according to the VP process with an exponential discretization*[9] *as in* (3.2.83)*, the output* $\hat{\boldsymbol{x}}$ *of Algorithm 3.1 satisfies the total variation bound*

$$\mathsf{TV}(\boldsymbol{x}, \hat{\boldsymbol{x}}) = \tilde{\mathcal{O}}\Bigg( \underbrace{\frac{D}{L}}_{\text{discretization error}} + \underbrace{\sqrt{\frac{1}{L}\sum_{\ell=1}^{L} \frac{\alpha_{t_\ell}^2}{\sigma_{t_\ell}^4}\,\mathbb{E}\,\|\bar{\boldsymbol{x}}^\star(t_\ell, \boldsymbol{x}_{t_\ell}) - \bar{\boldsymbol{x}}(t_\ell, \boldsymbol{x}_{t_\ell})\|_2^2}}_{\text{average excess error of the denoiser}} \Bigg) \tag{3.2.95}$$

*where* $\bar{\boldsymbol{x}}^\star$ *is the Bayes optimal denoiser for* $\boldsymbol{x}$*, and* $\tilde{\mathcal{O}}$ *is a version of the big-*$\mathcal{O}$ *notation that ignores logarithmic factors in* $L$*.*

The very high-level proof technique is, as discussed earlier, to bound the error at each step, distinguish the error sources (between discretization and denoiser error), and carefully ensure that the errors do not accumulate too much (or even cancel out).

Note that if $L \to \infty$ and we correctly learn the Bayes-optimal denoiser $\bar{\boldsymbol{x}} = \bar{\boldsymbol{x}}^\star$ (so the excess error is 0), then the sampling process in Algorithm 3.1 yields a *perfect (in distribution) inverse* of the noising process, since the error rate in Theorem 3.4 goes to 0,[10] as argued heuristically above.

*Remark* 3.1. What if the data is low-dimensional, say supported on a *low-rank subspace* of the high-dimensional space $\mathbb{R}^D$? If the data distribution is compactly supported—say if the data is normalized to the unit hypercube, which is often ensured as a preprocessing step for real data such as images—it is possible to do better. Namely, the authors of [LY24] also define a measure of *approximate intrinsic dimension* $\kappa$ using the asymptotics of the so-called covering number, which is extremely similar in intuition (if not in implementation) to the rate-distortion function presented in the next Chapter. Then they show that using a particular small modification of the DDIM sampler in Algorithm 3.1 (i.e., slightly perturbing the update coefficients), the discretization error becomes $\tilde{\mathcal{O}}(\kappa/L)$ instead of $\tilde{\mathcal{O}}(D/L)$ as in Theorem 3.4. Therefore, using this modified algorithm, $L$ does not have to be too large even as $D$ reaches the thousands or millions, since real data have low-dimensional structure. However, in practice we use the DDIM sampler instead, so $L$ should have a mild dependence on $D$ to achieve consistent error rates. The exact choice of $L$ trades off between the computational complexity (e.g., runtime or memory consumption) of sampling and the statistical complexity of learning a denoiser for low-dimensional structures. The value of $L$ is often different at training time (where a larger $L$ allows better coverage of the interval $[0, T]$, which helps the network learn a relationship that generalizes over $t$) and sampling time (where $L$ being smaller means more efficient sampling). One can even pick the timesteps adaptively at sampling time to optimize this tradeoff [BLZ+22].

[8]The supremum here is only over *measurable* sets $A$, not absolutely all sets. As with all other measure-theoretic caveats in this book, feel free to ignore it for practical purposes.

[9]The precise definition is rather lengthy in our notation and only defined up to various absolute constants, so we omit it here for brevity. It is in the original paper [LY24].

[10]Similar results hold for VE processes; the theoretical toolkits used for the proofs differ because of the structure of the underlying stochastic processes.

*Remark* 3.2. Various other works define the reverse process as moving backward in the time index $t$ using an explicit difference equation, or differential equation in the limit $L \to \infty$, or forward in time using the transformation $\boldsymbol{y}_t = \boldsymbol{x}_{T-t}$, such that if $t$ increases then $\boldsymbol{y}_t$ becomes closer to $\boldsymbol{x}_0$. Each paper uses its own notation, and it can sometimes be confusing to understand the relationships between different papers' results. In this work we strive to keep consistency: we move forward in time to noise, and backward in time to denoise. If you are reading another work that is not clear on the time index, or trying to implement an algorithm that is similarly unclear, there is one way to disambiguate every time: the sampling process should always have a *positive* coefficient on the denoiser term (or equivalently the score function) when taking a step.

*Remark* 3.3. The diffusion/denoising processes and associated models provide a way to transport the data distribution to and from a high-entropy (Gaussian or Gaussian-like) distribution. On the other hand, it may be useful to transport the data distribution to and from an arbitrary target distribution, which could be Gaussian, a Gaussian mixture, or even another type of data distribution. If we have this desideratum, we can use the *flow matching* framework to achieve this. Suppose the target random variable is $\boldsymbol{y}$, and our flow is

$$\boldsymbol{x}_t = \alpha_t \boldsymbol{x} + \sigma_t \boldsymbol{y}, \qquad \forall t \in [0, 1]. \tag{3.2.96}$$

such that $\alpha_0 = \sigma_1 = 1$ and $\alpha_1 = \sigma_0 = 0$, so that $\boldsymbol{x}_0 = \boldsymbol{x}$ and $\boldsymbol{x}_1 = \boldsymbol{y}$. Nothing about our strategy changes: we can train a denoiser $\bar{\boldsymbol{x}}_\theta$ to denoise $\boldsymbol{x}_t$ back to $\boldsymbol{x}$ (or equivalently a "noise" predictor to predict $\boldsymbol{y}$), and then apply our iterative denoising algorithms to start from a realization of $\boldsymbol{y}$ to obtain $\boldsymbol{x}$. The only thing missing is Tweedie's formula which connects the score $\nabla \log p_t$ of the denoiser to the denoiser; such a formula does not exist in general.

In practice, inspired by the concept of learning a so-called *transport map* between data pairs $\boldsymbol{x}$ and $\boldsymbol{y}$, practitioners often wish to train a *velocity predictor* $\bar{\boldsymbol{v}}_\theta$ to predict $\boldsymbol{v}_t = \frac{\mathrm{d}}{\mathrm{d}t}\boldsymbol{x}_t = \alpha_t'\boldsymbol{x} + \sigma_t'\boldsymbol{y}$ instead of a denoiser $\bar{\boldsymbol{x}}_\theta$, or rather train $\bar{\boldsymbol{v}}_\theta$ to predict $\mathbb{E}[\boldsymbol{v}_t \mid \boldsymbol{x}_t]$. In this case, it is simplest to take the choice of coefficients $\alpha_t = 1-t$ and $\sigma_t = t$ (for $t \in [0,1]$), so that the prediction target is $\boldsymbol{v}_t = \boldsymbol{y} - \boldsymbol{x}$ for each time $t$. Given the estimator $\bar{\boldsymbol{v}}_\theta$, the standard sampling algorithm follows the field backwards in time, i.e., starting from a realization of $\boldsymbol{y}$, we can iteratively update the iterate as follows:

$$\hat{\boldsymbol{x}}_{t_{\ell-1}} = \hat{\boldsymbol{x}}_{t_\ell} - \bar{\boldsymbol{v}}_\theta(t_\ell, \hat{\boldsymbol{x}}_{t_\ell})\Delta t, \tag{3.2.97}$$

where $\Delta t = t_\ell - t_{\ell-1}$ is the timestep. Note that velocity prediction is algebraically equivalent to denoising (see Exercise 3.6). See Sections 8.6 and 8.9 for practical applications of flow matching.

*Remark* 3.4. In practice, diffusion usually takes many steps and many function evaluations of the denoiser. It is therefore natural to ask: "can we speed up the sampling process by using a smaller number of steps and function evaluations?" The answer is yes, and such models are often called *consistency models* or, more descriptively, *few-step models*. The core idea is to use a denoiser that takes in two times $\bar{\boldsymbol{x}}_\theta(s, t, \boldsymbol{x}_t)$ and estimates $\mathbb{E}[\boldsymbol{x}_s \mid \boldsymbol{x}_t]$, which in some sense is an easier task, or alternatively to use a velocity predictor $\bar{\boldsymbol{v}}_\theta(s, t, \boldsymbol{x}_t)$ and estimate the average velocity $\mathbb{E}[(\boldsymbol{x}_t - \boldsymbol{x}_s)/(t-s) \mid \boldsymbol{x}_t]$. Such denoisers have additional consistency conditions such as $\bar{\boldsymbol{x}}_\theta(u, t, \bar{\boldsymbol{x}}_\theta(s, u, \boldsymbol{\xi})) = \bar{\boldsymbol{x}}_\theta(s, t, \boldsymbol{\xi})$, which motivate auxiliary training losses. Once such a predictor is learned, the sampling process becomes very simple and efficient, as it only requires a few steps of the form

$$\hat{\boldsymbol{x}}_s = \bar{\boldsymbol{x}}_\theta(s, t, \hat{\boldsymbol{x}}_t) = \hat{\boldsymbol{x}}_t - \bar{\boldsymbol{v}}_\theta(s, t, \hat{\boldsymbol{x}}_t) \cdot (t - s). \tag{3.2.98}$$

Relevant formulations include flow maps [BAV25] and mean flows [GDB+25].

## 3.3 Memorization, Generalization, and Coding Rates

The calculation presented at the end of Section 3.2.1 seems to suggest (loosely speaking) that in practice, using a transformer-like network is a good choice for learning or approximating a denoiser. This is reasonable, but what is the problem with using any old neural network (such as a multi-layer perceptron (MLP)) and simply scaling it up to infinity? To answer this question, let us consider a constraint we have ignored or glossed over so far in this Chapter: we only have finite data with which to learn the denoiser. Indeed, let us consider the training problem (3.2.90), except this time we assume that our data $\boldsymbol{x}$ are drawn from the empirical distribution $p^{\boldsymbol{X}}$ on $N$ training samples $\boldsymbol{X} = [\boldsymbol{x}^{(1)}, \dots, \boldsymbol{x}^{(N)}]$. Because $p^{\boldsymbol{X}}$ is supported on a 0-dimensional subset of the ambient space $\mathbb{R}^D$, we have that $h(p^{\boldsymbol{X}}) = -\infty$, so in some sense $p^{\boldsymbol{X}}$ is the simplest (i.e., lowest-entropy) distribution that could generate our observations $\boldsymbol{X}$. In short, this assumption arguably *has the least inductive bias* about the data distribution! Many popular commentaries on deep learning and generative models would have you believe that, due to this property, the best way to solve the generative modeling problem is to scale a generic denoiser as large as possible and reap the corresponding benefits. However, this advice is (provably!) wrong in this setting. To understand why, let us note that the empirical distribution $p^{\boldsymbol{X}}$ is actually a mixture of Gaussians with zero covariances:

$$p^{[\boldsymbol{x}^{(1)},\ldots,\boldsymbol{x}^{(N)}]} = \frac{1}{N}\sum_{i=1}^{N}\mathcal{N}(\boldsymbol{x}^{(i)}, \mathbf{0}). \tag{3.3.1}$$

As such, we know that the (globally) optimal solution to the denoising problem (3.2.90) with $\boldsymbol{x}$ drawn from the empirical distribution, written out as follows:

$$\min_{\theta\in\Theta} \frac{1}{N}\sum_{i=1}^{N}\mathbb{E}[w_\tau\|\boldsymbol{x}^{(i)} - \bar{\boldsymbol{x}}_\theta(\tau, \boldsymbol{x}_\tau^{(i)})\|_2^2] \tag{3.3.2}$$

where $\boldsymbol{x}_t^{(i)}$ is the noised version of the training sample $\boldsymbol{x}^{(i)}$. The optimizer of this loss over all (square-integrable) denoisers $\bar{\boldsymbol{x}}$ is the associated optimal Gaussian denoiser from Example 3.3, i.e.,

$$\bar{\boldsymbol{x}}^\star(t, \boldsymbol{x}_t) = \sum_{i=1}^{N}\frac{\exp(-\|\boldsymbol{x}_t - \alpha_t\boldsymbol{x}^{(i)}\|_2^2/(2\sigma_t^2))}{\sum_{j=1}^{N}\exp(-\|\boldsymbol{x}_t - \alpha_t\boldsymbol{x}^{(j)}\|_2^2/(2\sigma_t^2))}\boldsymbol{x}^{(i)}. \tag{3.3.3}$$

This is a convex combination of the data $\boldsymbol{x}^{(i)}$, and the coefficients get "sharper" (i.e., closer to 0 or 1) as $t \to 0$. In particular, for small $t$, the denoiser is almost exactly equal to the closest sample in the training set, and the update moves the iterate closer to the closest point, such that at the end of sampling, the iterate $\hat{\boldsymbol{x}}$ is (nearly) equal to a training sample $\boldsymbol{x}^{(i)}$. Let us re-emphasize this point: *if we use the denoiser* (3.3.3) *and a good sampler, our samples will re-generate the training data.* This is confirmed by our theoretical guarantee on sampling Theorem 3.4, which, when instantiated in this context, says that if we take the number of sampling steps to $\infty$ and use the prescribed timestep schedule, the diffusion model associated with the denoiser (3.3.3) will always sample from the empirical distribution and thus always reproduce a training sample. This phenomenon is a particularly strict instantiation of *memorization* in generative modeling.[11]

Since in practice we only have a fixed finite dataset on which we can train our denoiser, the learning problem we aim to solve in this case is the finite-sample denoising loss (3.3.2) (compared to the infinite-sample denoising loss (3.2.90), for example). As such, *if we have an arbitrarily large model and optimize it as well as possible, it will learn to memorize and re-generate the training set.* In other words, this very large-scale model is not much more interesting than a simple database of the training data.

[11]Memorization is a much broader phenomenon whose exact characterization depends on the domain the model is trained on (e.g., text, images, videos, audio, etc.). Later in this Section we will discuss more heuristic measures of memorization.

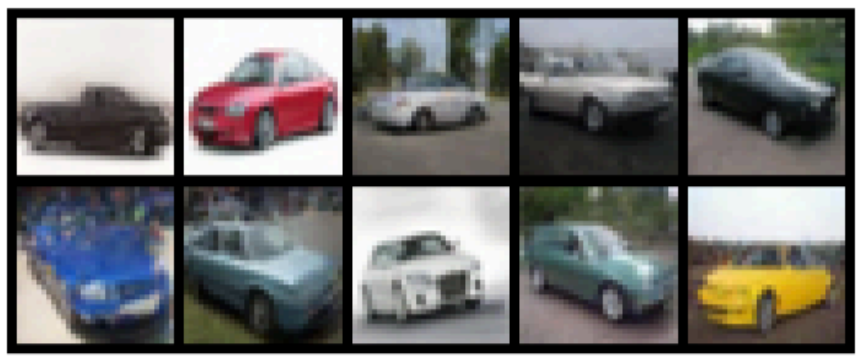

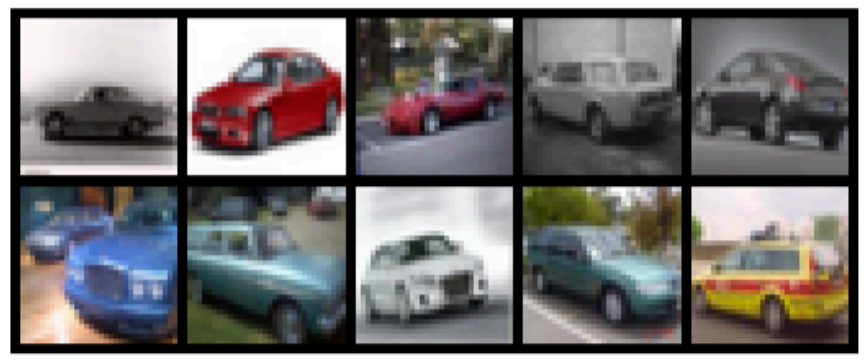

Figure 3.9: **Generated samples (left) from a diffusion model trained on** 6,000 **samples and (right) their closest points in the training dataset (in the Euclidean norm).** We observe that the generated samples are essentially pixel-perfect equivalents to the most similar training data. Both sets of images are from [YCK+23].

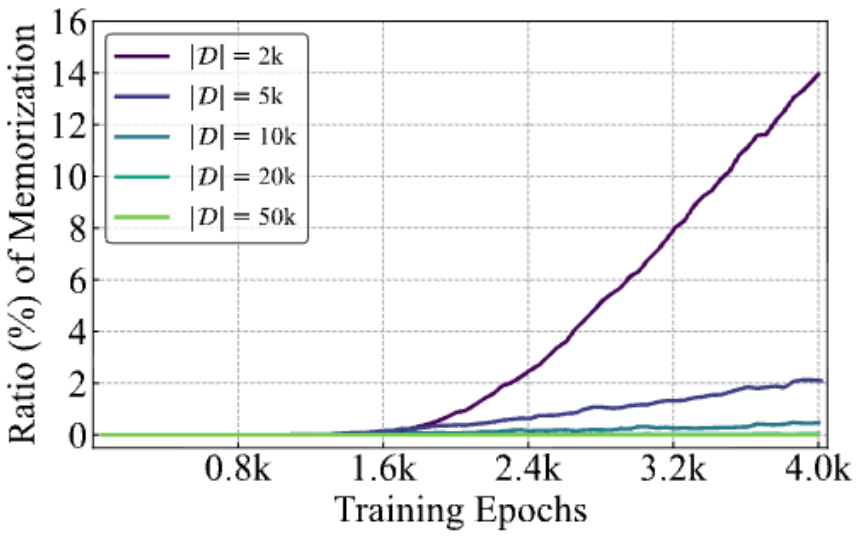

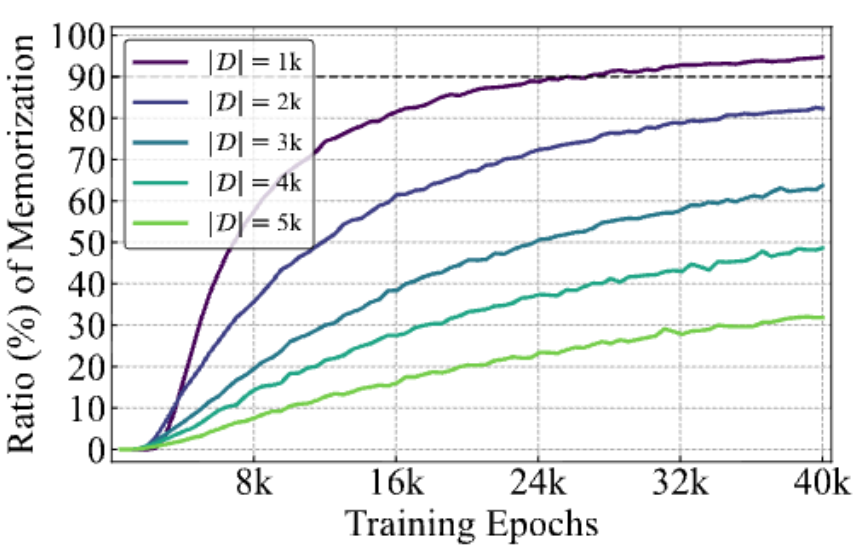

Figure 3.10: **Memorization versus dataset size for a fixed model, on (left) 4,000 training epochs and (right) 40,000 training epochs with smaller datasets**. We observe from the left plot that large enough datasets will never be memorized (e.g., 10,000 to 50,000 samples), whereas small datasets (1,000 to 5,000 samples) will be memorized after enough training. This means that as the dataset becomes larger relative to the model, the memorization ratio at the end of training goes from near-1 to near-0 relatively rapidly, indicating the possible existence of a phase transition from memorization to generalization as the dataset size increases. Figure taken from [GDP+23].

The above argument holds for arbitrarily powerful models that can fit the (true) memorizing denoiser. It is therefore relevant to ask whether it actually bears out in practice: do very large models (relative to the dataset) learn to memorize the training data? It turns out that this is empirically true. One of the first studies on this was [YCK+23]. To understand their results, we define the memorization ratio as the probability of memorization:

$$\mathrm{MemRatio}(\bar{\boldsymbol{x}}) = \mathbb{P}_{\hat{\boldsymbol{x}} \sim \mathrm{DM}(\bar{\boldsymbol{x}})}[\hat{\boldsymbol{x}} \text{ is memorized}], \tag{3.3.4}$$

where $\mathrm{DM}(\bar{\boldsymbol{x}})$ is the distribution induced by the diffusion model that uses denoiser $\bar{\boldsymbol{x}}$, and the sample $\hat{\boldsymbol{x}}$ is memorized using the following criterion:

$$\hat{\boldsymbol{x}} \text{ is memorized} \quad \Longleftrightarrow \quad \frac{\|\hat{\boldsymbol{x}} - \boldsymbol{x}_{(1)}(\hat{\boldsymbol{x}})\|_2}{\|\hat{\boldsymbol{x}} - \boldsymbol{x}_{(2)}(\hat{\boldsymbol{x}})\|_2} < \frac{1}{3}, \tag{3.3.5}$$

where $\boldsymbol{x}_{(i)}(\hat{\boldsymbol{x}})$ is the $i^{\text{th}}$ closest point to $\hat{\boldsymbol{x}}$ in the training dataset (in the Euclidean norm), and $1/3$ is a constant that is empirically shown to agree well with human perception of memorization. Notice that this definition exactly corresponds to our earlier intuitive definition of memorization as sampled points each being much closer to a particular training sample than all other training samples. They showed that this kind of memorization occurs in practice when the model is much larger (in terms of the number of parameters) relative to the data; Figure 3.9 demonstrates that in this case, generated samples are essentially pixel-perfect copies of training samples. Slightly later, [GDP+23] performed a careful study to characterize how various factors such as architectures and training time impact memorization, showing among other things that as the dataset size increases relative to a fixed model, the memorization ratio rapidly changes from near-1 to near-0, i.e., the model quickly stops memorizing. This indicates the possibility of a phase transition between memorization and generalization as the model becomes smaller relative to the data. [BPM+25] investigates the existence and theoretical mechanism of this phase transition in synthetic data drawn from a Gaussian mixture model, demonstrating that it exists and providing a partial characterization of the model size at which models will start or stop memorizing. Other theoretical work such as [GVM25; KG24; NZM+24; ZLL+25] also attempts to provide a theoretical characterization of the phase transition in terms of model and dataset size in different settings.

In order to empirically prove the existence of a phase transition between memorization and generalization in real data, [ZZL+24] rethought the definition of memorization using the following reasoning. In the very high-dimensional space of natural images, Euclidean distances between points become much more strict [WM22], in the sense that if two images are not pixel-perfect matches to each other, then their Euclidean distances will concentrate sharply around a fixed value (scaling with the images' dimension). This does not catch the case where certain *concepts* (such as copyrighted cartoon characters) are generated in novel positions, making it very difficult to form effective and practical memorization criteria around the Euclidean distance. To fix this, [ZZL+24] uses a particular function $f_{\mathrm{SSCD}}$ (a pre-trained neural network called a Self-Supervised

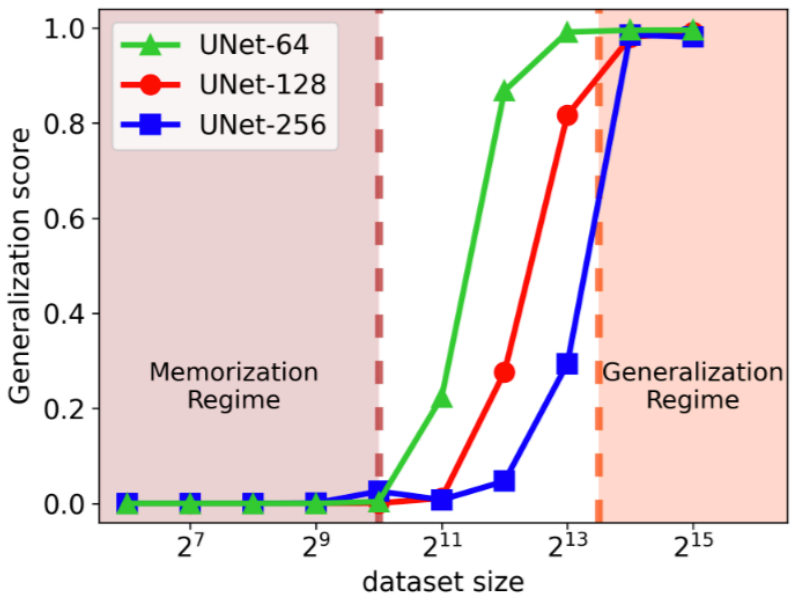

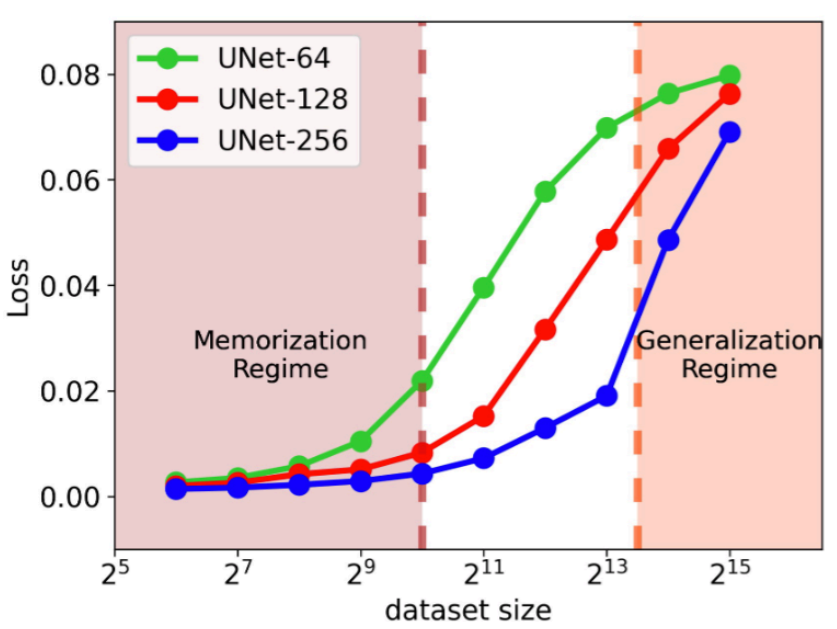

Figure 3.11: **The generalization score (left) and training loss (right) for a fixed model and increasing data**, from [ZZL+24]. The generalization score is $1-$ the fraction of samples that are memorized, and the different lines correspond to differently-sized denoiser models (UNet-64 is smaller than UNet-128, which in turn is smaller than UNet-256). The figures show that, as the dataset size becomes larger relative to the model size, the loss increases significantly and memorization stops occurring, showing that our theory and intuition apply to real models.

Copy Detector [PRR+22]) to map images into a Euclidean space, then takes the normalized inner product to obtain their (cosine) similarity, such that images that are similar due to this cosine similarity are claimed to be conceptually similar in the sense needed for memorization. Then, [ZZL+24] says that a sample $\hat{\boldsymbol{x}}$ is memorized if its similarity with any training point is $> 3/5$, i.e.,

$$\hat{\boldsymbol{x}} \text{ is memorized} \iff \max_{i \in [N]} \frac{f_{\mathrm{SSCD}}(\hat{\boldsymbol{x}})^\top f_{\mathrm{SSCD}}(\boldsymbol{x}^{(i)})}{\|f_{\mathrm{SSCD}}(\hat{\boldsymbol{x}})\|_2 \|f_{\mathrm{SSCD}}(\boldsymbol{x}^{(i)})\|_2} > \frac{3}{5} \tag{3.3.6}$$

Figure 3.11 shows the outcome of taking many (around 10,000) samples from diffusion models trained on differently-sized training datasets in terms of estimating memorization and computing the training loss. The outcome is in line with the above discussion; when the model is too large relative to the training data and the training loss is too small, memorization occurs and samples reproduce the training data. Thus, our intuition applies in practice: *for generative diffusion models, scaling the model is not all you need!*

So now the question is: what kind of denoiser should we use? The key is to choose a network architecture that can approximate the true denoiser (say, the one corresponding to a low-rank distribution as in (3.2.80)) without approximating the memorizing denoiser (3.3.3). Some work has explored this fine line and why modern diffusion models, which use Transformer- and convolution-based architectures, can memorize and generalize in different regimes [KG24; NZM+24; ZLL+25].

Let us at last return to the initial premise of this Chapter, described in Section 3.1.3: finding the minimal-entropy distribution that models the data is the only way to learn the distribution. The above discussion has hopefully shown that this claim is false. However, a related claim, also argued in Section 3.1.3 because of its equivalence in the situations we had then considered, might still be true: finding the distribution with minimal *coding rate* is a way to learn the distribution. Supposing for the moment that $\boldsymbol{x}$ has a density $p$, the coding rate (i.e., cross-entropy) of the empirical distribution $p^{\boldsymbol{X}}$ with respect to $p$ is $\mathrm{CE}(p \parallel p^{\boldsymbol{X}}) = \infty$. This coding rate is obviously not minimal in any sense; if we use the same successively better approximations $p^n$, $p^e$, and $\hat{q}$ discussed in Section 3.1.3, it holds

$$\infty = \mathrm{CE}(p \parallel p^{\boldsymbol{X}}) > \mathrm{CE}(p \parallel p^n) > \mathrm{CE}(p \parallel p^e) > \mathrm{CE}(p \parallel \hat{q}) \approx h(p). \tag{3.3.7}$$

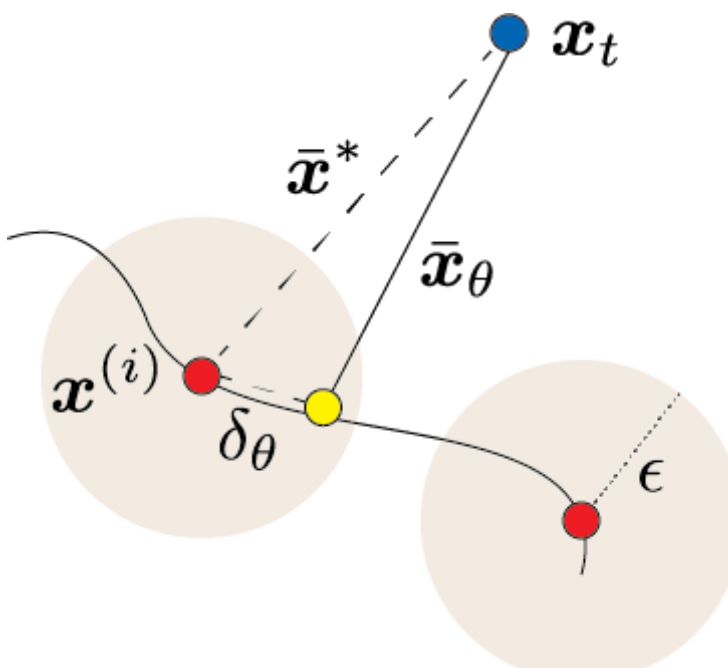

Figure 3.12: **Visualizing the action of a well-trained denoiser** as a perturbation of the empirical denoiser at *small times*, when the memorizing denoiser acts like a projection onto the nearest point in the training set.

Minimizing the coding rate therefore seems a sensible way to learn the distribution while avoiding this issue. However, a problem of tractability now arises: how can we devise algorithms that learn the distribution by minimizing the coding rate while avoiding the drawbacks associated with overly minimizing entropy?

One clue is to examine what happens with successfully trained diffusion models that do not memorize. By writing the trained denoiser $\bar{\boldsymbol{x}}_\theta$ as a perturbation of the memorizing denoiser $\bar{\boldsymbol{x}}^\star$, i.e.,

$$\bar{\boldsymbol{x}}_\theta(t, \boldsymbol{x}_t) = \bar{\boldsymbol{x}}^\star(t, \boldsymbol{x}_t) + \boldsymbol{\delta}_\theta(t, \boldsymbol{x}_t) \tag{3.3.8}$$

we can gain graphical intuition about the action of a well-trained denoiser, as in Figure 3.12. At a very high level, the trained denoiser computes the memorizing denoiser plus some perturbation. At a small time $t$ (perhaps even at the last step of denoising), the memorizing denoiser points entirely in the direction of a single training sample (the training sample to be re-generated by the current iterate). The distribution being learned, or denoised against, is just the empirical distribution with some $\epsilon$-balls around each point,[12] where $\epsilon$ is an upper bound on the magnitude of the perturbation, visualized in Figure 3.12 by the faint orange balls.[13] For increasingly larger well-trained diffusion models, the perturbation radius $\epsilon$ shrinks to 0, and the model learns the empirical distribution. On the other hand, $\epsilon$ large enough that balls corresponding to different points overlap or are close may imply that the diffusion model generalizes.

[12]More formally, this perturbed distribution has support on (a subset of) a union of $\epsilon$-balls centered at each training sample. This picture is supported by past work, e.g., [SKC+25].

Thus, our successful diffusion model can be viewed as minimizing a *lossy coding rate*, i.e., coding rate up to a precision $\epsilon$ (so as to be able to disregard the $\epsilon$-balls around each point). This notion of the lossy coding rate turns out to be the right thing to minimize, morally speaking, and will be *extremely* useful. We will tease out its uses in Chapter 4 and throughout the rest of this book.

## 3.4 Summary and Notes

**Key messages.** In this Chapter, we have shown that, to pursue a general low-dimensional distribution, a universal computational approach is to start with a generic distribution and compress it toward the distribution through (iterative) noise or entropy reduction. As we discussed in the preceding Chapter, this is even the case for arguably the simplest distributions of a single subspace, such as computing PCA via the power-iteration algorithm. As we will see in future chapters, the fundamental roles of modern deep networks are not to fit arbitrary (noisy) data distributions or functions; instead, they are to realize or approximate those (lossy) compressing or denoising operators. We will derive precise forms of those operators and the associated deep architectures in future chapters.

**More extensions and references.** The use of denoising and diffusion for sampling has a rich history. The first work clearly about a diffusion model is probably [SWM+15], but before this there are many works on denoising as a computational and statistical problem. The most relevant of these is probably [Hyv05], which explicitly uses the score function to denoise and to perform independent component analysis. The most popular follow-ups are essentially co-occurring: [HJA20; SE19]. Since then, thousands of papers have built on diffusion models; we revisit this topic in Chapter 6.

Many of these works use a different stochastic process than the simple linear combination (3.2.85). In fact, all works listed above emphasize the need to add *independent* Gaussian noise at the beginning of each step of the forward process. Theoretically-minded work uses Brownian motion or stochastic differential equations to formulate the forward process [SSK+21]. However, since linear combinations of Gaussians still result in Gaussians, the *marginal distributions* of such processes still take the form of (3.2.85). Most of our discussion requires only that the marginal distributions are what they are, so our overly simplistic model is actually enough for almost everything. The only time marginal distributions are not enough is when we derive an expression for $\mathbb{E}[\boldsymbol{x}_s \mid \boldsymbol{x}_t]$ in terms of $\mathbb{E}[\boldsymbol{x} \mid \boldsymbol{x}_t]$. Different (noising) processes give different such expressions, which can be used for sampling (and of course there are other ways to derive efficient samplers, such as the ever-popular DDPM sampler). The process in (3.2.85) is a bona fide stochastic process whose "natural" denoising iteration takes the form of the popular DDIM algorithm [SME20]. (Even this equivalence is not trivial; we cite [DGG+25] as justification.)

[13] In high dimensions, it is hypothesized that a better model for these perturbations is not being contained by balls but rather by much more anisotropic geometry-adaptive shapes [FPH+25], but the intuition remains similar.

On top of the theoretical work [LY24] covered in Section 3.2.2, and the lineage of work that it builds on, which studies the *sampling* efficiency of diffusion models when the data has low-dimensional structure, there is a large body of work that studies the *training* efficiency of diffusion models when the data has low-dimensional structure. Specifically, Chen et al. [CHZ+23] and Wang et al. [WZZ+24] characterized the approximation and estimation error of denoisers when the data belongs to a mixture of low-rank Gaussians, showing that the number of training samples required to accurately learn the distribution scales with the intrinsic dimension of the data rather than the ambient dimension. There is considerable *methodological* work that attempts to use the low-dimensional structure of the data to perform various tasks with diffusion models. We highlight three: image editing [CZG+24], watermarking [LZQ24], and unlearning [CZL+25], though this list is inexhaustive.

## 3.5 Exercises and Extensions

*Exercise* 3.1. Show that (3.2.8) is the optimal solution of Problem (3.2.7).

*Exercise* 3.2. Consider random vectors $\boldsymbol{x} \in \mathbb{R}^D$ and $\boldsymbol{y} \in \mathbb{R}^d$, such that the pair $(\boldsymbol{x}, \boldsymbol{y}) \in \mathbb{R}^{D+d}$ is jointly Gaussian. This means that

$$\begin{bmatrix} \boldsymbol{x} \\ \boldsymbol{y} \end{bmatrix} \sim \mathcal{N}\left( \begin{bmatrix} \boldsymbol{\mu}_{\boldsymbol{x}} \\ \boldsymbol{\mu}_{\boldsymbol{y}} \end{bmatrix}, \begin{bmatrix} \boldsymbol{\Sigma}_{\boldsymbol{x}} & \boldsymbol{\Sigma}_{\boldsymbol{x}\boldsymbol{y}} \\ \boldsymbol{\Sigma}_{\boldsymbol{x}\boldsymbol{y}}^\top & \boldsymbol{\Sigma}_{\boldsymbol{y}} \end{bmatrix} \right),$$

where the mean and covariance parameters are given by

$$\boldsymbol{\mu}_{\boldsymbol{x}} = \mathbb{E}[\boldsymbol{x}], \quad \boldsymbol{\mu}_{\boldsymbol{y}} = \mathbb{E}[\boldsymbol{y}], \quad \begin{bmatrix} \boldsymbol{\Sigma}_{\boldsymbol{x}} & \boldsymbol{\Sigma}_{\boldsymbol{x}\boldsymbol{y}} \\ \boldsymbol{\Sigma}_{\boldsymbol{x}\boldsymbol{y}}^\top & \boldsymbol{\Sigma}_{\boldsymbol{y}} \end{bmatrix} = \mathbb{E}\left[ \begin{bmatrix} \boldsymbol{x} - \mathbb{E}[\boldsymbol{x}] \\ \boldsymbol{y} - \mathbb{E}[\boldsymbol{y}] \end{bmatrix} \begin{bmatrix} \boldsymbol{x} - \mathbb{E}[\boldsymbol{x}] \\ \boldsymbol{y} - \mathbb{E}[\boldsymbol{y}] \end{bmatrix}^\top \right]$$

Assume that $\boldsymbol{\Sigma}_{\boldsymbol{y}}$ is positive definite (hence invertible); then positive semidefiniteness of the covariance matrix is equivalent to the Schur complement condition $\boldsymbol{\Sigma}_{\boldsymbol{x}} - \boldsymbol{\Sigma}_{\boldsymbol{x}\boldsymbol{y}} \boldsymbol{\Sigma}_{\boldsymbol{y}}^{-1} \boldsymbol{\Sigma}_{\boldsymbol{x}\boldsymbol{y}}^\top \succeq \mathbf{0}$.

In this exercise, we will prove that the conditional distribution $p_{\boldsymbol{x} \mid \boldsymbol{y}}$ is Gaussian: namely,

$$p_{\boldsymbol{x} \mid \boldsymbol{y}} \sim \mathcal{N}\left( \boldsymbol{\mu}_{\boldsymbol{x}} + \boldsymbol{\Sigma}_{\boldsymbol{x}\boldsymbol{y}} \boldsymbol{\Sigma}_{\boldsymbol{y}}^{-1} (\boldsymbol{y} - \boldsymbol{\mu}_{\boldsymbol{y}}), \boldsymbol{\Sigma}_{\boldsymbol{x}} - \boldsymbol{\Sigma}_{\boldsymbol{x}\boldsymbol{y}} \boldsymbol{\Sigma}_{\boldsymbol{y}}^{-1} \boldsymbol{\Sigma}_{\boldsymbol{x}\boldsymbol{y}}^\top \right). \tag{3.5.1}$$

A direct path to prove this result manipulates the defining ratio of densities $p_{\boldsymbol{x},\boldsymbol{y}} / p_{\boldsymbol{y}}$. We sketch an algebraically concise argument of this form below.

1. Verify the following matrix identity for the covariance:

$$\begin{bmatrix} \boldsymbol{\Sigma}_{\boldsymbol{x}} & \boldsymbol{\Sigma}_{\boldsymbol{x}\boldsymbol{y}} \\ \boldsymbol{\Sigma}_{\boldsymbol{x}\boldsymbol{y}}^\top & \boldsymbol{\Sigma}_{\boldsymbol{y}} \end{bmatrix} = \begin{bmatrix} \boldsymbol{I}_D & \boldsymbol{\Sigma}_{\boldsymbol{x}\boldsymbol{y}}\boldsymbol{\Sigma}_{\boldsymbol{y}}^{-1} \\ \boldsymbol{0} & \boldsymbol{I}_d \end{bmatrix} \begin{bmatrix} \boldsymbol{\Sigma}_{\boldsymbol{x}} - \boldsymbol{\Sigma}_{\boldsymbol{x}\boldsymbol{y}}\boldsymbol{\Sigma}_{\boldsymbol{y}}^{-1}\boldsymbol{\Sigma}_{\boldsymbol{x}\boldsymbol{y}}^\top & \boldsymbol{0} \\ \boldsymbol{0} & \boldsymbol{\Sigma}_{\boldsymbol{y}} \end{bmatrix} \begin{bmatrix} \boldsymbol{I}_D & \boldsymbol{0} \\ \boldsymbol{\Sigma}_{\boldsymbol{y}}^{-1}\boldsymbol{\Sigma}_{\boldsymbol{x}\boldsymbol{y}}^\top & \boldsymbol{I}_d \end{bmatrix}. \tag{3.5.2}$$

   One arrives at this identity by performing two rounds of (block) Gaussian elimination on the covariance matrix.

2. Based on the previous identity, show that

$$\begin{bmatrix} \boldsymbol{\Sigma}_{\boldsymbol{x}} & \boldsymbol{\Sigma}_{\boldsymbol{x}\boldsymbol{y}} \\ \boldsymbol{\Sigma}_{\boldsymbol{x}\boldsymbol{y}}^\top & \boldsymbol{\Sigma}_{\boldsymbol{y}} \end{bmatrix}^{-1} = \boldsymbol{P} \begin{bmatrix} \left(\boldsymbol{\Sigma}_{\boldsymbol{x}} - \boldsymbol{\Sigma}_{\boldsymbol{x}\boldsymbol{y}}\boldsymbol{\Sigma}_{\boldsymbol{y}}^{-1}\boldsymbol{\Sigma}_{\boldsymbol{x}\boldsymbol{y}}^\top\right)^{-1} & \boldsymbol{0} \\ \boldsymbol{0} & \boldsymbol{\Sigma}_{\boldsymbol{y}}^{-1} \end{bmatrix} \boldsymbol{P}^\top, \tag{3.5.3}$$

   where

$$\boldsymbol{P} = \begin{bmatrix} \boldsymbol{I}_D & \boldsymbol{0} \\ -\boldsymbol{\Sigma}_{\boldsymbol{y}}^{-1}\boldsymbol{\Sigma}_{\boldsymbol{x}\boldsymbol{y}}^\top & \boldsymbol{I}_d \end{bmatrix}, \tag{3.5.4}$$

   and whenever the relevant inverses are defined.[14] Conclude that

$$\begin{bmatrix} \boldsymbol{x} - \boldsymbol{\mu}_{\boldsymbol{x}} \\ \boldsymbol{y} - \boldsymbol{\mu}_{\boldsymbol{y}} \end{bmatrix}^\top \begin{bmatrix} \boldsymbol{\Sigma}_{\boldsymbol{x}} & \boldsymbol{\Sigma}_{\boldsymbol{x}\boldsymbol{y}} \\ \boldsymbol{\Sigma}_{\boldsymbol{x}\boldsymbol{y}}^\top & \boldsymbol{\Sigma}_{\boldsymbol{y}} \end{bmatrix}^{-1} \begin{bmatrix} \boldsymbol{x} - \boldsymbol{\mu}_{\boldsymbol{x}} \\ \boldsymbol{y} - \boldsymbol{\mu}_{\boldsymbol{y}} \end{bmatrix} \tag{3.5.5}$$

$$= \begin{bmatrix} \boldsymbol{x} - \boldsymbol{m} \\ \boldsymbol{y} - \boldsymbol{\mu}_{\boldsymbol{y}} \end{bmatrix}^\top \begin{bmatrix} \left(\boldsymbol{\Sigma}_{\boldsymbol{x}} - \boldsymbol{\Sigma}_{\boldsymbol{x}\boldsymbol{y}}\boldsymbol{\Sigma}_{\boldsymbol{y}}^{-1}\boldsymbol{\Sigma}_{\boldsymbol{x}\boldsymbol{y}}^\top\right)^{-1} & \boldsymbol{0} \\ \boldsymbol{0} & \boldsymbol{\Sigma}_{\boldsymbol{y}}^{-1} \end{bmatrix} \begin{bmatrix} \boldsymbol{x} - \boldsymbol{m} \\ \boldsymbol{y} - \boldsymbol{\mu}_{\boldsymbol{y}} \end{bmatrix}, \tag{3.5.6}$$

   where $\boldsymbol{m} = \boldsymbol{\mu}_{\boldsymbol{x}} + \boldsymbol{\Sigma}_{\boldsymbol{x}\boldsymbol{y}}\boldsymbol{\Sigma}_{\boldsymbol{y}}^{-1}(\boldsymbol{y} - \boldsymbol{\mu}_{\boldsymbol{y}})$.

3. By dividing $p_{\boldsymbol{x},\boldsymbol{y}}/p_{\boldsymbol{y}}$, prove Equation (3.5.1). (*Hint: Using the previous identities, only minimal algebra should be necessary. For the normalizing constant, use Equation* (3.5.3) *to factor the determinant similarly.*)

*Exercise* 3.3. Show the Sherman–Morrison–Woodbury identity, i.e., for matrices $\boldsymbol{A}$, $\boldsymbol{C}$, $\boldsymbol{U}$, $\boldsymbol{V}$ such that $\boldsymbol{A}$, $\boldsymbol{C}$, and $\boldsymbol{A} + \boldsymbol{U}\boldsymbol{C}\boldsymbol{V}$ are invertible,

$$(\boldsymbol{A} + \boldsymbol{U}\boldsymbol{C}\boldsymbol{V})^{-1} = \boldsymbol{A}^{-1} - \boldsymbol{A}^{-1}\boldsymbol{U}(\boldsymbol{C}^{-1} + \boldsymbol{V}\boldsymbol{A}^{-1}\boldsymbol{U})^{-1}\boldsymbol{V}\boldsymbol{A}^{-1}. \tag{3.5.7}$$

*Exercise* 3.4. Rederive the following, assuming $\boldsymbol{x}_t$ follows the generalized noise model (3.2.85).

- Tweedie's formula: (3.2.86).
- The DDIM iteration: (3.2.87).
- The Bayes-optimal denoiser for a Gaussian mixture model: (3.2.88).

*Exercise* 3.5.

[14] In cases where the Schur complement term is not invertible, the same result holds with its inverse replaced by the Moore-Penrose pseudoinverse. In particular, the conditional distribution (3.5.1) becomes a degenerate Gaussian distribution.

1. Implement the formulae derived in Exercise 3.4, building a sampler for Gaussian mixtures.

2. Reproduce Figure 3.4 and Figure 3.8.

3. We now introduce *Flow Matching (FM)* (see Remark 3.3), as follows:

$$\alpha_t = 1 - t, \qquad \sigma_t = t. \tag{3.5.8}$$

   Implement this process using the same framework, and test it for sampling in high dimensions. Which process gives better or more stable results?

*Exercise* 3.6. Consider the somewhat familiar interpolation

$$\boldsymbol{x}_t = \alpha_t \boldsymbol{x} + \sigma_t \boldsymbol{y}, \qquad \forall t \in [0, T]. \tag{3.5.9}$$

where $\boldsymbol{y}$ is a (not necessarily Gaussian) random variable. We define the *velocity* as

$$\boldsymbol{v}_t = \frac{\mathrm{d}}{\mathrm{d}t}\boldsymbol{x}_t = \alpha_t' \boldsymbol{x} + \sigma_t' \boldsymbol{y}, \qquad \forall t \in [0, T]. \tag{3.5.10}$$

Compute an algebraic relationship writing $\mathbb{E}[\boldsymbol{v}_t \mid \boldsymbol{x}_t]$ in terms of $\mathbb{E}[\boldsymbol{x} \mid \boldsymbol{x}_t]$ and vice versa, showing that velocity prediction is algebraically equivalent to denoising and thus also to noise prediction (see Remark 3.3).

# Chapter 4

# Representation Learning via Lossy Compression

"*When you can measure what you are speaking about and express it in numbers, you know something about it; but when you cannot measure it, when you cannot express it in numbers, your knowledge is of the meager and unsatisfactory kind: it may be the beginning of knowledge, but you have scarcely, in your thoughts, advanced to the stage of science, whatever the matter may be.*"

– Lord Kelvin, 1883

Let us recap what we have covered so far. We have shown that it is possible to identify and access a distribution with arbitrary low-dimensional support, such as the one illustrated in Figure 4.1 (left), by gradually transforming the distribution of pure noise (high entropy) into the distribution to be learned via *iterative denoising*. We have shown that the denoiser at each iteration is directly connected to the gradient of the distribution's log-density via Tweedie's formula (3.2.23). We showed that in practice such a denoiser $\bar{\boldsymbol{x}}_\theta$ can be estimated using finite samples. Thus we developed a general way of learning or pursuing a data distribution.

Nevertheless, in this methodology the encoding of the distribution is implicit in the denoiser's functional form and parameters, if any. In fact, acute readers might have noticed that for a general distribution we have never explicitly specified what the functional form for the denoiser should be. In practice, people typically model it with some deep neural network that has an empirically designed architecture. In addition, although we know the above denoising process reduces the entropy, we do not know by how much, nor do we know the entropy of the intermediate and resulting distributions. Hence it would be very useful to know the (optimal) form of the denoising operation and how to measure the change in entropy. Finally, we showed at the end of the last Chapter that if the denoising process reduces the entropy *too much* it leads to learning only the training data, with no possible interpolation or extrapolation. The methodological innovation which resolves all of these issues will be developed in the Chapter to come.

**Measure of information.** Recall that our general goal is to model data from a continuous distribution with low-dimensional support. If we want to identify the simplest model that generates the data, we might consider three typical measures of parsimony: dimension, volume, or differential entropy. If we use dimension, then obviously the best model for a given dataset is the empirical distribution itself, which is zero-dimensional.[1] If we use the entropy of the data instead, then for all distributions with low-dimensional support, the differential entropy is always negative infinity. Or if we use the volume of the space spanned by the data, the volume of the support of any low-dimensional model is always zero. So, among all distributions with low-dimensional support that could have generated the same data samples, how can we decide which one is better based on these measures of parsimony that cannot distinguish among themselves at all?

In this Chapter, we discuss a framework that alleviates the above technical difficulty by associating the learned distribution with an explicit computable encoding and decoding scheme, following and making more explicit the second approach suggested at the end of Section 3.1.3. As we will see, such an approach essentially allows us to accurately approximate the entropy of the learned distributions in terms of a *lossy coding rate* associated with the coding scheme, as described and motivated (but not formalized) at the end of Section 3.3. Roughly speaking, such a measure can be viewed as a generalized notion of volume of the empirical data distribution. With such a measure, we can accurately quantify how much the entropy is reduced through a transformation of the distribution, say by a denoiser. In addition, with such a measure, we can derive, at least in principle, the optimal denoising operator that reduces the entropy the fastest.

**Representation learning.** Even though the diffusion and denoising process introduced in the previous chapter allows us, in principle, to learn an arbitrary low-dimensional distribution, it does not give an explicit representation of the distribution. As we will see, the learned distribution is to be repeatedly accessed and used for subsequent tasks (such as used as the prior for Bayesian inference in Chapter 7). Hence it is desirable that we can transform the data distribution, say by a mapping $f(\boldsymbol{x}, \theta)$, to some standard structured form that facilitates efficient access, say a mixture of (incoherent) linear subspaces or low-rank Gaussians, as illustrated in Figure 4.1 right. We will formalize this process as "representation learning" in this Chapter.

[1] In the Introduction Chapter 1, we argued that one important goal of intelligence is to pursue low-dimensional structures in observed data. On the technical level though, one shall not take the word "low-dimensional" literally. It is the goal of this chapter to clarify precisely what low-dimensional structures we should be pursuing and how.

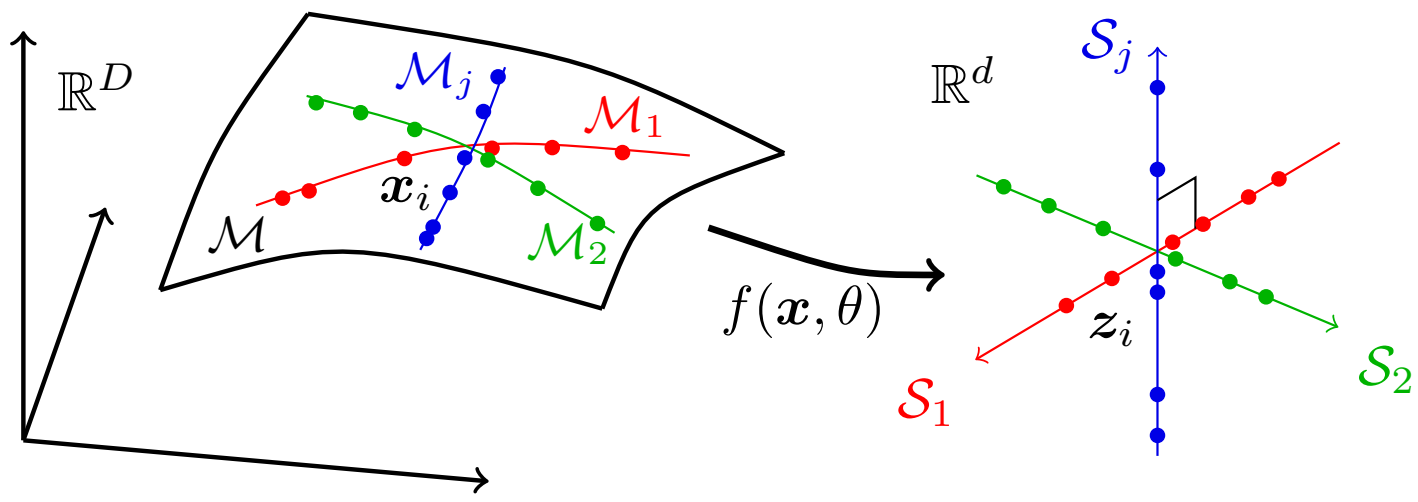

Figure 4.1: Learn a structured representation (right) for the data distribution (left). Notice that in addition to learning (and sampling) the distribution of the original data as we did in the previous Chapter 3, we here want to learn a transform $f(\boldsymbol{x}, \theta)$ for the data distribution so as to result in a representation that has desired better geometric or statistical properties, say linear and incoherent.

As we will also see, for an arbitrary distribution, the above measure of information (coding rate) is not (easily) computable. But for a mixture of subspaces or low-rank Gaussians, we can derive an accurate closed form for the measure. By optimizing this measure, it will allow us to find the optimal representation. As we will see in the next Chapter 5, optimizing the measure also allows us to derive explicit forms of the operator that performs the transformation $f(\boldsymbol{x}, \theta)$ in the most efficient way. This will lead to a principled explanation for the architecture of deep networks, as well as to more efficient deep-architecture designs.

## 4.1 Compression via Lossy Coding

In this section, we develop the *lossy coding rate* as a computable measure of complexity for distributions with low-dimensional support—a setting where classical measures such as entropy, dimension, and volume all fail to distinguish among candidate models. We first motivate the necessity of lossy coding with respect to these pathologies (Section 4.1.1), then show that the rate distortion function connects this coding-theoretic measure to the geometry of the data support via sphere covering (Theorem 4.1). This connection is the bridge between information-theoretic and geometric approaches to learning low-dimensional distributions. Finally, we develop a tight closed-form approximation to the lossy coding rate for Gaussians (Section 4.1.3) and an effective algorithm for clustering mixtures of low-dimensional Gaussians (Section 4.1.4), which will be generalized to nonlinear distributions in the following section.

### 4.1.1 Necessity of Lossy Coding

We have previously, multiple times, discussed a difficulty: if we learn the distribution from finite samples in the end, and our function class of denoisers contains enough functions, how do we ensure that we sample from the *true* distribution (with low-dimensional supports), instead of any other distribution that may produce those finite samples with high probability? Let us reveal some of the conceptual and technical difficulties with some concrete examples.

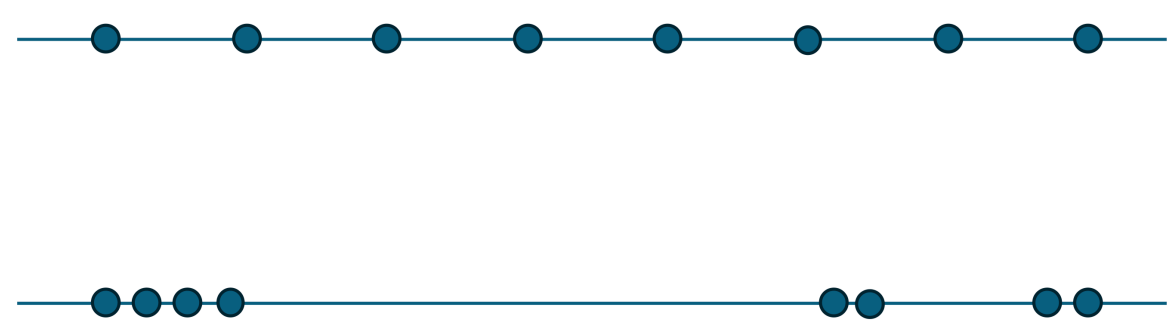

Figure 4.2: Eight points observed on a line with different geometric configurations; evenly distributed (top) or closely clustered (bottom).

*Example* 4.1 (Volume, Dimension, and Entropy). For the example shown on the top of Figure 4.2, suppose we have taken some samples from a uniform distribution on a line (say in a 2D plane). The volume of the line or the sample sets is zero. Geometrically, the empirical distribution on the produced finite sample set is *the minimum-dimension* one which can produce the finite sample set.[2] But this is in seemingly contrast with yet another measure of complexity: entropy. The (differential) entropy of the line is negative infinity but the (discrete) entropy of this sample set is finite and positive. So we seem to have a paradoxical situation according to these common measures of parsimony or complexity: they cannot properly differentiate among (models for) distributions of low-dimensional supports at all, and some seem to differentiate them even in exactly opposite manners.[3] ■

*Example* 4.2 (Density). Consider the two sets of sampled data points shown in Figure 4.2. Geometrically, they are essentially the same: each set consists of eight points and each point has occurred with equal frequency 1/8th. The only difference is that for the second data set, some points are "close" enough to be viewed as having a higher density around their respective "cluster." Which one is more relevant to the true distribution that may have generated the samples? How can we reconcile such ambiguity in interpreting this kind of (empirical) distributions? ■

There is yet another technical difficulty associated with constructing an explicit encoding and decoding scheme for a data set. Given a sampled data set in $\boldsymbol{X} = [\boldsymbol{x}_1, \ldots, \boldsymbol{x}_N]$, how to design a coding scheme that is implementable on machines with finite memory and computing resources? Note that even representing a general real number requires an infinite number of digits or bits. Therefore, one may wonder whether the entropy of a distribution is a direct measure for the complexity of its (optimal) coding scheme. We examine this matter with another simple example.

[2] A set of discrete samples are all of zero dimension whereas the supporting line is one dimension.

[3] Of course, strictly speaking, differential entropy and discrete entropy are not directly comparable.

*Example* 4.3 (Precision). Consider a discrete distribution $\boldsymbol{X} = [e, \pi]$ with equal probability $1/2$ taking the values of the Euler number $e \approx 2.71828$ or the number $\pi \approx 3.14159$. The entropy of this distribution is $H = 1$, which suggests that one may encode the two numbers by a one-bit digit 0 or 1, respectively. But can you realize a decoding scheme for this code on a finite-state machine? The answer is actually no, as it takes infinitely many bits to describe either number precisely. ■

Hence, it is generally impossible to have an encoding and decoding scheme that can precisely reproduce samples from an arbitrary real-valued distribution.[4] But there would be little practical value to encode a distribution without being able to decode for samples drawn from the same distribution.

So to ensure that any encoding/decoding scheme is computable and implementable with finite memory and computational resources, we need to quantify the sample $\boldsymbol{x}$ and encode it only up to a certain precision, say $\epsilon > 0$. *By doing so, in essence, we treat any two data points as equivalent if their distance is less than* $\epsilon$. More precisely, we would like to consider coding schemes

$$\boldsymbol{x} \mapsto \hat{\boldsymbol{x}} \tag{4.1.1}$$

such that the expected error caused by the quantization is bounded by $\epsilon$. It is mathematically more convenient, and conceptually almost identical, to bound the expected *squared* error by $\epsilon^2$, i.e.,

$$\mathbb{E}[d(\boldsymbol{x}, \hat{\boldsymbol{x}})^2] \leq \epsilon^2. \tag{4.1.2}$$

Typically, the distance $d$ is chosen to be the Euclidean distance, or the 2-norm.[5] We will adopt this choice in the sequel.

### 4.1.2 Rate Distortion and Data Geometry

Of course, among all encoding schemes that satisfy the above constraint, we would like to choose the one that minimizes the resulting coding rate. For a given random variable $\boldsymbol{x}$ and a precision $\epsilon$, this rate is known as the *rate distortion*, denoted as $\mathcal{R}_\epsilon(\boldsymbol{x})$.

To understand this rate distortion better, we turn to a deep theorem of information theory, originally proved in Shannon [Sha59], which relates this rate distortion to other information-theoretic quantities. To understand this theorem, we first need to define a measure of *difference* between two distributions known as the *Kullback-Leibler* (KL) divergence, or *relative entropy*, which is denoted by $\mathsf{KL}(\cdot \parallel \cdot)$ and defined in relation to the entropy and cross-entropy (see Section 3.1.3) as

$$\mathsf{KL}(p \parallel q) = \mathrm{CE}(p \parallel q) - h(p) = \mathbb{E}_{\boldsymbol{x}\sim p}[-\log q(\boldsymbol{x})] - \mathbb{E}_{\boldsymbol{x}\sim p}[-\log p(\boldsymbol{x})] \tag{4.1.3}$$

[4] That is, if one wants to encode such samples precisely, the only way is to memorize every single sample.

[5] More generally, we can replace $d^2$ with any so-called *divergence*.

$$= \mathbb{E}_{\boldsymbol{x}\sim p}[\log(p(\boldsymbol{x})/q(\boldsymbol{x}))]. \tag{4.1.4}$$

The KL divergence is always non-negative, and equals zero if and only if $p = q$;[6] in the case where $p$ and $q$ have the same support and both are probability densities, this can be proved by the following calculation based on Jensen's inequality and the fact that the function $\log(\cdot)$ is strictly concave:

$$-\mathsf{KL}(p \parallel q) = -\mathbb{E}_{\boldsymbol{x}\sim p}[\log(p(\boldsymbol{x})/q(\boldsymbol{x}))] = \mathbb{E}_{\boldsymbol{x}\sim p}[\log(q(\boldsymbol{x})/p(\boldsymbol{x}))] \tag{4.1.5}$$

$$\leq \log \mathbb{E}_{\boldsymbol{x}\sim p}[q(\boldsymbol{x})/p(\boldsymbol{x})] = \log \int_{\mathbb{R}^D} \frac{q(\boldsymbol{\xi})}{p(\boldsymbol{\xi})} p(\boldsymbol{\xi}) \mathrm{d}\boldsymbol{\xi} \tag{4.1.6}$$

$$= \log \int_{\mathbb{R}^D} q(\boldsymbol{\xi}) \mathrm{d}\boldsymbol{\xi} = \log \mathbb{E}_{\boldsymbol{x}\sim q}[1] = \log 1 = 0. \tag{4.1.7}$$

Then, we define the *mutual information* $I(\cdot;\cdot)$, which is a measure of *dependence* between two random variables, as

$$I(\boldsymbol{x}; \hat{\boldsymbol{x}}) = \mathsf{KL}(p(\boldsymbol{x}, \hat{\boldsymbol{x}}) \parallel p(\boldsymbol{x})p(\hat{\boldsymbol{x}})), \tag{4.1.8}$$

With the previous definitions in place, we may write the rate distortion $\mathcal{R}_\epsilon$ in purely probabilistic terms as

$$\mathcal{R}_\epsilon(\boldsymbol{x}) = \min_{p(\hat{\boldsymbol{x}}|\boldsymbol{x}) : \mathbb{E}[\|\boldsymbol{x}-\hat{\boldsymbol{x}}\|_2^2] \leq \epsilon^2} I(\boldsymbol{x}; \hat{\boldsymbol{x}}). \tag{4.1.9}$$

Note that the minimization in (4.1.9) is over all conditional distributions $p(\hat{\boldsymbol{x}} \mid \boldsymbol{x})$ that satisfy the distortion constraint $\mathbb{E}[\|\boldsymbol{x} - \hat{\boldsymbol{x}}\|_2^2] \leq \epsilon^2$. Each such conditional distribution induces a joint distribution $p(\boldsymbol{x}, \hat{\boldsymbol{x}}) = p(\hat{\boldsymbol{x}} \mid \boldsymbol{x})p(\boldsymbol{x})$, which determines the mutual information (4.1.8). Many convenient properties of the mutual information (and hence the rate distortion) are implied by corresponding properties of the KL divergence. From the definition, we know that $\mathcal{R}_\epsilon(\boldsymbol{x})$ is a *non-increasing* function in $\epsilon$.

*Example* 4.4 (Rate distortion of Gaussian source (Theorem 13.3.3, [CT91])). Consider a Gaussian random variable $\boldsymbol{x} \sim \mathcal{N}(0, \sigma^2)$. For this case, it is well-known that the rate distortion function can be computed in closed form as:

$$\mathcal{R}_\epsilon(x) = \begin{cases} \frac{1}{2} \log\left(\sigma^2/\epsilon^2\right), & 0 \leq \epsilon \leq \sigma, \\ 0, & \epsilon > \sigma \end{cases}. \tag{4.1.10}$$

This extends in a particularly interesting way to the case of $D$ dimensions. Suppose that $\boldsymbol{x} \sim \mathcal{N}(\mathbf{0}, \boldsymbol{\Sigma})$, where $\boldsymbol{\Sigma} \in \mathsf{PSD}(D)$ has eigenvalues $\lambda_1 \geq \cdots \geq \lambda_D \geq 0$. Then the rate distortion is obtainable via "water filling" as

$$\mathcal{R}_\epsilon(\boldsymbol{x}) = \frac{1}{2} \sum_{i=1}^{D} \log\left(\frac{\lambda_i}{\min\{\kappa, \lambda_i\}}\right), \tag{4.1.11}$$

[6] Technically, this equality should be taken to mean "almost everywhere", i.e., except possibly on a set of zero measure (volume), since this set would not impact the value of any integral.

where the "water level" $\kappa$ satisfies $\sum_{i=1}^{D} \min\{\kappa, \lambda_i\} = \epsilon^2$. Intuitively, directions with variance $\lambda_i$ larger than $\kappa$ are effectively encoded, while directions with smaller variance are discarded, ensuring that the distortion constraint is satisfied with minimal mutual information. Please refer to Exercise 4.4 for a detailed derivation of this closed form. ■

*Remark* 4.1. As it turns out, the rate distortion is an implementable approximation to the entropy of $\boldsymbol{x}$ in the following sense. Assume that $\boldsymbol{x}$ and $\hat{\boldsymbol{x}}$ are continuous random vectors. Then the mutual information can be written as

$$I(\boldsymbol{x}; \hat{\boldsymbol{x}}) = h(\boldsymbol{x}) - h(\boldsymbol{x} \mid \hat{\boldsymbol{x}}), \tag{4.1.12}$$

where $h(\boldsymbol{x} \mid \hat{\boldsymbol{x}}) = -\mathbb{E}[\log_2 p(\boldsymbol{x} \mid \hat{\boldsymbol{x}})]$ is the *conditional entropy* of $\boldsymbol{x}$ given $\hat{\boldsymbol{x}}$. Hence, the minimal coding rate is achieved when the difference between the entropy of $\boldsymbol{x}$ and the conditional entropy of $\boldsymbol{x}$ given $\hat{\boldsymbol{x}}$ is minimized among all distributions that satisfy the constraint: $\mathbb{E}[\|\boldsymbol{x} - \hat{\boldsymbol{x}}\|_2^2] \leq \epsilon^2$.

In fact, it is not necessary to assume that $\boldsymbol{x}$ and $\hat{\boldsymbol{x}}$ are continuous to obtain the above type of conclusion. For example, if both random vectors are instead discrete, we have after a suitable interpretation of the KL divergence for discrete-valued random vectors that

$$I(\boldsymbol{x}; \hat{\boldsymbol{x}}) = H(\boldsymbol{x}) - H(\boldsymbol{x} \mid \hat{\boldsymbol{x}}), \tag{4.1.13}$$

where $H(\boldsymbol{x})$ is defined in (3.1.1) of Section 3.1.3. More generally, advanced mathematical notions from measure theory can be used to define the mutual information (and hence the rate distortion) for arbitrary random variables $\boldsymbol{x}$ and $\hat{\boldsymbol{x}}$, including those with rather exotic low-dimensional distributions; see Cover and Thomas [CT91, §8.5].

*Remark* 4.2. Given a set of data points in $\boldsymbol{X} = [\boldsymbol{x}_1, \ldots, \boldsymbol{x}_N]$, one can always interpret them as samples from a uniform discrete distribution with equal probability $1/N$ on these $N$ vectors. The entropy for such a distribution is $H(\boldsymbol{X}) = \frac{1}{N} \log_2 N$.[7] However, even if $\boldsymbol{X}$ is a uniform distribution on its samples, the coding rate $\mathcal{R}_\epsilon(\boldsymbol{X})$ achievable with a lossy coding scheme could be significantly lower than $H(\boldsymbol{X})$ if these samples are not so evenly distributed and many are clustered closely to each other. Therefore, for the second distribution shown in Figure 4.2, for a properly chosen quantization error $\epsilon$, the achievable lossy coding rate can be significantly lower than coding it as a uniform distribution.[8] Also notice that, with the notion of rate distortion, the difficulty discussed in Example 4.3 also disappears: We can choose two rational numbers that are close enough to each of the two irrational numbers. The resulting coding scheme will have a finite complexity.

*Example* 4.5. Sometimes, one may face an opposite situation when we want to fix the coding rate first and try to find a coding scheme that minimizes the distortion. For example, suppose that we only want to use a fixed number of codes for points sampled from a distribution, and we want to know how to design the codes such that the average or maximum distortion is minimized during the encoding/decoding scheme. For example, given a uniform distribution on a unit square, we wonder how precisely we can encode points drawn from this distribution, with say $n$ bits. This problem is equivalent to asking what is the minimum radius (i.e., distortion) such that we can cover the unit square with $2^n$ discs of this radius. Figure 4.3 shows approximately optimal coverings of a square with $n = 4, 6, 8$, so that $2^n = 16, 64, 256$ discs, respectively. Notice that the optimal radii of the discs decreases as the number of discs $2^n$ increases. ■

[7] Note again, even if we can encode these vectors with this coding rate, we cannot decode them with an arbitrary precision.

[8] Nevertheless, for this discrete uniform distribution, when $\epsilon$ is small enough, we always have $H(\boldsymbol{X}) \approx \mathcal{R}_\epsilon(\boldsymbol{X})$.

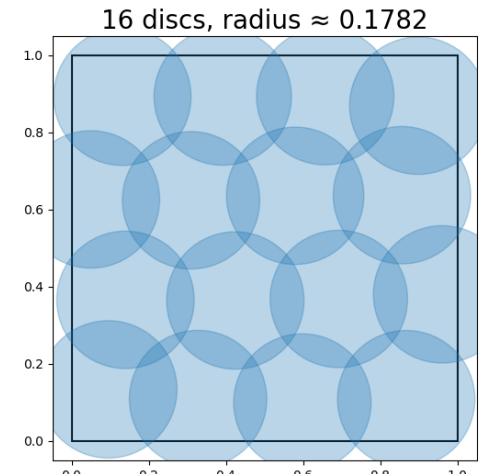

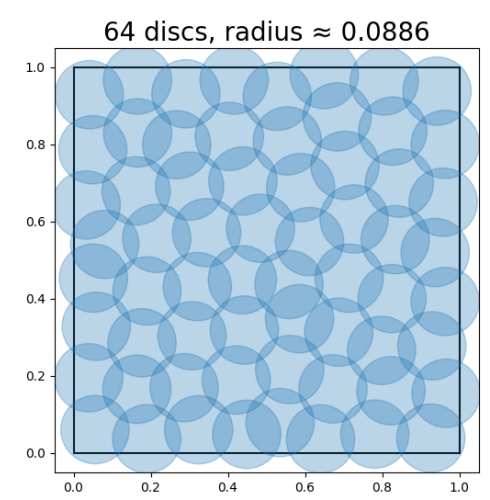

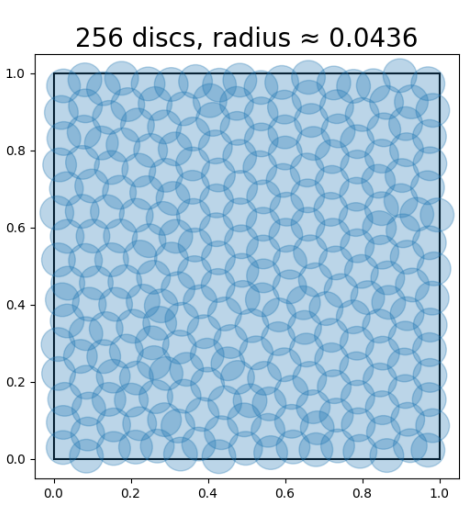

Figure 4.3: Approximations to the optimal solutions for $2^4$, $2^6$, and $2^8$ discs covering a square, along with the corresponding radii, calculated using a heuristic optimization algorithm.

It turns out to be a notoriously hard problem to obtain closed-form expressions for the rate distortion function (4.1.9) for general distributions $p(\boldsymbol{x})$. However, as Example 4.5 suggests, there are important special cases where the *geometry* of the support of the distribution $p(\boldsymbol{x})$ can be linked to the rate distortion function and hence to the optimal coding rate at distortion level $\epsilon$. In fact, this example can be generalized to any setting where the support of $p(\boldsymbol{x})$ is a sufficiently regular compact set—including low-dimensional distributions—and $p(\boldsymbol{x})$ is uniformly distributed on its support. This covers a vast number of cases of practical interest. We formalize this notion in the following result, which establishes this property for a special case.

**Theorem 4.1.** *Suppose that $\boldsymbol{x}$ is a random variable such that its support $K \doteq \operatorname{Supp}(\boldsymbol{x})$ is a compact set. Define the covering number $\mathcal{N}_\epsilon(K)$ as the minimum number of balls of radius $\epsilon$ that can cover $K$, i.e.,*

$$\mathcal{N}_\epsilon(K) \doteq \min\left\{ n \in \mathbb{N} \colon \exists \boldsymbol{p}_1, \ldots, \boldsymbol{p}_n \in K \text{ s.t. } K \subseteq \bigcup_{i=1}^{n} B_\epsilon(\boldsymbol{p}_i) \right\}, \tag{4.1.14}$$

*where $B_\epsilon(\boldsymbol{p}) = \{\boldsymbol{\xi} \in \mathbb{R}^D \mid \|\boldsymbol{\xi} - \boldsymbol{p}\|_2 \le \epsilon\}$ is the Euclidean ball of radius $\epsilon$ centered at $\boldsymbol{p}$. Then it holds*

$$\mathcal{R}_\epsilon(\boldsymbol{x}) \le \log_2 \mathcal{N}_\epsilon(K). \tag{4.1.15}$$

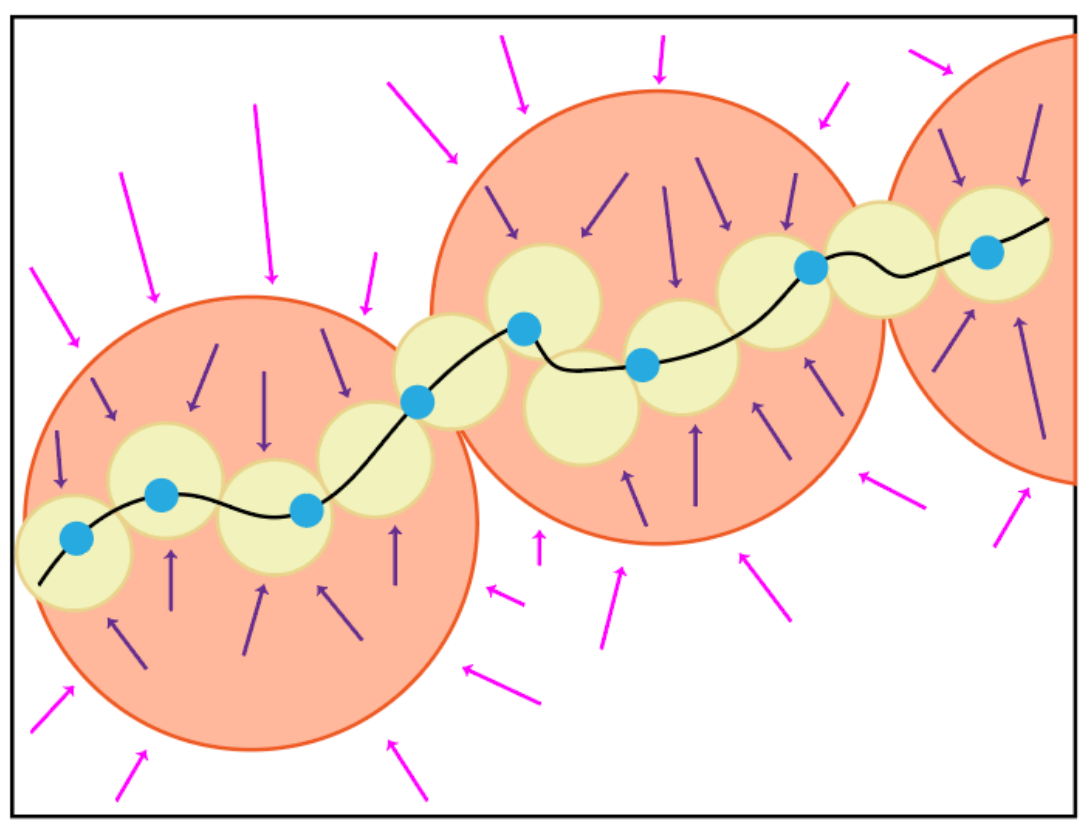

Figure 4.4: **The approximation of a low-dimensional distribution by $\epsilon$ balls.** We can see that as the $\epsilon$ parameter shrinks, the union of $\epsilon$-balls approximates the support of the true distribution (black) increasingly well. Furthermore, the associated denoisers (whose input-output mapping is given by the provided arrows) obtained by approximating the true distribution by a mixture of Gaussians, each with covariance $(\epsilon^2/D)\boldsymbol{I}$, increasingly well-approximate the true denoisers. At large $\epsilon$, such denoisers do not point near the true distribution at all, whereas at small $\epsilon$ they closely approximate the true denoisers. Theorem 4.1 establishes that this approximation characterizes the rate distortion function at small distortions $\epsilon$, unifying the parallel approaches of coding rate minimization and denoising for learning low-dimensional distributions without pathologies.

*If, in addition, $\boldsymbol{x}$ is uniformly distributed on $K$ and $K$ is a union of mutually orthogonal low-dimensional hyperspheres,*[9] *then a matching lower bound holds:*

$$\mathcal{R}_\epsilon(\boldsymbol{x}) \geq \log_2 \mathcal{N}_\epsilon(K) - O(D). \tag{4.1.16}$$

*Proof.* A proof of this theorem is beyond the scope of this book and we defer it to Section B.3. □

The upper bound in Theorem 4.1 requires only that $\mathrm{Supp}(\boldsymbol{x})$ is a compact set, while our proof of the lower bound requires stronger regularity assumptions: specifically, that $\boldsymbol{x}$ is uniformly distributed on $\mathrm{Supp}(\boldsymbol{x})$ and that $\mathrm{Supp}(\boldsymbol{x})$ is a mixture of low-rank Gaussians (with appropriate normalization).[10] Under these assumptions, the implication of Theorem 4.1 can be summarized as follows: for sufficiently accurate coding of the distribution of $\boldsymbol{x}$ (i.e., as $\epsilon \to 0$), the minimum rate distortion coding framework is equivalent to the sphere packing problem on the support of $\boldsymbol{x}$. So the rate distortion can be thought of as a "probability-aware" way to approximate the support of the distribution of $\boldsymbol{x}$ by a mixture of many small balls (Figure 4.4).

[9] In fact, it is possible to treat highly irregular $K$, such as compact manifolds or even fractals, with a parallel result, but its statement becomes far more technical: cf. Riegler et al. [RBK18; RKB23]. We give a simple proof in Section B.3 which shows the result for unions of orthogonal hyperspheres, an assumption that is equivalent to a mixture of low-rank Gaussians with orthogonal supports when the dimension of each subspace is sufficiently large. See Section B.3 for further discussion of this assumption.

[10] Later in this Chapter, in Section 4.2.3, we will see an objective function whose maxima are normalized mixtures of low-rank Gaussians that correspond to the input data in a certain way. Chapter 5 will further develop this framework for deep learning, showing that the structural assumption we make to prove the lower bound is not overly restrictive.

We now discuss another connection between this and the denoising-diffusion-entropy complexity hierarchy we discussed earlier in this chapter.

*Remark* 4.3. The key ingredient in the proof of the lower bound in Theorem 4.1 is an important result from information theory known as the *Shannon lower bound* for the rate distortion, named after Claude Shannon, who first derived it in a special case [Sha59]. It asserts the following estimate for the rate distortion function, for any random variable $\boldsymbol{x}$ with a density $p(\boldsymbol{x})$ and finite expected squared norm [LZ94]:

$$\mathcal{R}_\epsilon(\boldsymbol{x}) \geq h(\boldsymbol{x}) - \log_2 \mathrm{vol}(B_\epsilon) - C_D, \tag{4.1.17}$$

where $C_D > 0$ is a constant depending only on $D$. Moreover, this lower bound is actually sharp as $\epsilon \to 0$: that is,

$$\lim_{\epsilon \to 0} \mathcal{R}_\epsilon(\boldsymbol{x}) - [h(\boldsymbol{x}) - \log_2 \mathrm{vol}(B_\epsilon) - C_D] = 0. \tag{4.1.18}$$

So when the distortion $\epsilon$ is small, we can think solely in terms of the Shannon lower bound, rather than the (generally intractable) optimization problem defining the rate distortion (4.1.9).

The Shannon lower bound is the bridge between the coding rate, entropy minimization/denoising, and geometric sphere packing approaches for learning low-dimensional distributions. Notice that in the special case of a uniform density $p(\boldsymbol{x})$, (4.1.17) becomes

$$\begin{aligned}
\mathcal{R}_\epsilon(\boldsymbol{x}) &\geq -\int_K \frac{1}{\mathrm{vol}(K)} \log_2 \frac{1}{\mathrm{vol}(K)} \mathrm{d}\boldsymbol{\xi} - \log_2 \mathrm{vol}(B_\epsilon) - C_D &(4.1.19)\\
&= \log_2 \mathrm{vol}(K)/\mathrm{vol}(B_\epsilon) - C_D. &(4.1.20)
\end{aligned}$$

The ratio $\mathrm{vol}(K)/\mathrm{vol}(B_\epsilon)$ approximates the number of $\epsilon$-balls needed to cover $K$ by a worst-case argument, which is accurate for sufficiently regular sets $K$ when $\epsilon$ is small (see Section B.3 for details). Meanwhile, recall the Gaussian denoising model $\boldsymbol{x}_\epsilon = \boldsymbol{x} + \epsilon \boldsymbol{g}$ from earlier in the Chapter, where $\boldsymbol{g} \sim \mathcal{N}(\boldsymbol{0}, \boldsymbol{I})$ is independent of $\boldsymbol{x}$. Interestingly, the differential entropy of the joint distribution $(\boldsymbol{x}, \epsilon\boldsymbol{g})$ can be calculated as

$$\begin{aligned}
h(\boldsymbol{x}, \epsilon\boldsymbol{g}) &= -\int p(\boldsymbol{\xi}) p(\boldsymbol{\gamma}) \log_2 p(\boldsymbol{\xi}) p(\boldsymbol{\gamma}) \mathrm{d}\boldsymbol{\xi} \mathrm{d}\boldsymbol{\gamma} &(4.1.21)\\
&= h(\boldsymbol{x}) + h(\epsilon \boldsymbol{g}). &(4.1.22)
\end{aligned}$$

We have seen the Gaussian entropy calculated in Equation (3.1.4): when $\epsilon$ is small, it is equal, up to additive constants, to the quantity $-\log_2 \mathrm{vol}(B_\epsilon)$ we have seen in the Shannon lower bound. In certain special cases (e.g., data supported on incoherent low-rank subspaces), when $\epsilon$ is small and the support of $\boldsymbol{x}$ is sufficiently regular, the distribution of $\boldsymbol{x}_\epsilon$ can even be well-approximated locally by the product of the distributions $p(\boldsymbol{x})$ and $p(\boldsymbol{g})$, justifying the above computation. Hence the Gaussian denoising process yields yet another interpretation of the Shannon lower bound, as arising from the entropy of a noisy version of $\boldsymbol{x}$, with noise level proportional to the distortion level $\epsilon$.

Thus, this finite rate distortion approach via sphere covering re-enables or generalizes all previous measures of complexity of the distribution, allowing us to differentiate between and rank different distributions in a unified way. These interrelated viewpoints are visualized in Figure 4.4.

For a general distribution at finite distortion levels, it is typically impossible to find its rate distortion function in an analytical form. One must often resort to numerical computation[11]. Nevertheless, as we will see, in our context we often need to know the rate distortion as an explicit function of a set of data points or their representations. This is because we want to use the coding rate as a measure of the goodness of the representations. An explicit analytical form makes it easy to determine how to transform the data distribution to improve the representation. So, we should work with distributions whose rate distortion functions take explicit analytical forms. To this end, we start with the simplest, and also the most important, family of distributions.

### 4.1.3 Lossy Coding Rate for a Low-Dimensional Gaussian

Now suppose we are given a set of data samples in $\boldsymbol{X} = [\boldsymbol{x}_1, \ldots, \boldsymbol{x}_N]$ from any distribution.[12] We would like to come up with a constructive scheme that can encode the data up to certain precision, say

$$\boldsymbol{x}_i \mapsto \hat{\boldsymbol{x}}_i, \quad \text{subject to} \quad \|\boldsymbol{x}_i - \hat{\boldsymbol{x}}_i\|_2 \leq \epsilon. \tag{4.1.23}$$

Notice that this is a sufficient, explicit, and interpretable condition which ensures that the data are encoded such that $\frac{1}{N}\sum_{i=1}^{N} \|\boldsymbol{x}_i - \hat{\boldsymbol{x}}_i\|_2^2 \leq \epsilon^2$. This latter inequality is exactly the rate distortion constraint for the provided empirical distribution and its encoding. For example, in Example 4.5, we used this simplified criterion to explicitly find the minimum distortion and explicit coding scheme for a given coding rate.

Without loss of generality, let us assume the mean of $\boldsymbol{X}$ is zero, i.e., $\frac{1}{N}\sum_{i=1}^{N} \boldsymbol{x}_i = \boldsymbol{0}$. Without any prior knowledge about the nature of the distribution behind $\boldsymbol{X}$, we may view $\boldsymbol{X}$ as sampled from a Gaussian distribution $\mathcal{N}(\boldsymbol{0}, \boldsymbol{\Sigma})$ with the covariance[13]

$$\boldsymbol{\Sigma} = \frac{1}{N}\boldsymbol{X}\boldsymbol{X}^\top. \tag{4.1.24}$$

[11] Interested readers may refer to [Bla72] for a classic algorithm that computes the rate distortion function numerically for a discrete distribution.

[12] Or these data points could be viewed as an (empirical) distribution themselves.

[13] It is known that given a fixed variance, the Gaussian achieves the maximal entropy. That is, it gives an upper bound for what the worst case could be in terms of possible coding rate.

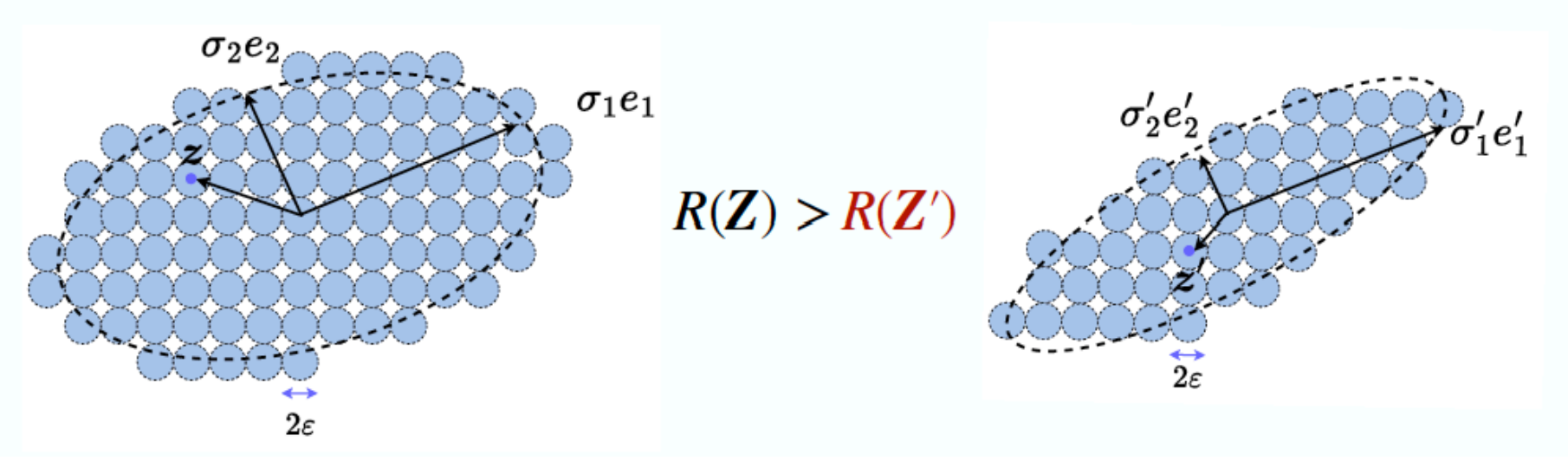

Figure 4.5: Covering the region spanned by the data vectors using $\epsilon$-balls. The larger the volume of the space, the more balls are needed, hence the more bits are needed to encode and enumerate the balls. Each real-valued vector $\boldsymbol{x}$ can be encoded as the number of the ball which it falls into.

Notice that geometrically $\boldsymbol{\Sigma}$ characterizes an ellipsoidal region where most of the samples $\boldsymbol{x}_i$ reside.

We may view $\hat{\boldsymbol{X}} = [\hat{\boldsymbol{x}}_1, \ldots, \hat{\boldsymbol{x}}_N]$ as a noisy version of $\boldsymbol{X} = [\boldsymbol{x}_1, \ldots, \boldsymbol{x}_N]$:

$$\hat{\boldsymbol{x}}_i = \boldsymbol{x}_i + \boldsymbol{w}_i, \tag{4.1.25}$$

where $\boldsymbol{w}_i \sim \mathcal{N}(\boldsymbol{0}, \epsilon^2 \boldsymbol{I}/D)$ is a Gaussian noise independent of $\boldsymbol{x}_i$. Then the covariance of $\hat{\boldsymbol{x}}_i$ is given by

$$\hat{\boldsymbol{\Sigma}} = \mathbb{E}\left[\hat{\boldsymbol{x}}_i \hat{\boldsymbol{x}}_i^\top\right] = \frac{\epsilon^2}{D}\boldsymbol{I} + \frac{1}{N}\boldsymbol{X}\boldsymbol{X}^\top. \tag{4.1.26}$$

Note that the volume of the region spanned by the vectors $\boldsymbol{x}_i$ is proportional to the square root of the determinant of the covariance matrix

$$\text{volume}(\hat{\boldsymbol{x}}_i) \propto \sqrt{\det\left(\hat{\boldsymbol{\Sigma}}\right)} = \sqrt{\det\left(\frac{\epsilon^2}{D}\boldsymbol{I} + \frac{1}{N}\boldsymbol{X}\boldsymbol{X}^\top\right)}. \tag{4.1.27}$$

The volume spanned by each random vector $\boldsymbol{w}_i$ is proportional to

$$\text{volume}(\boldsymbol{w}_i) \propto \sqrt{\det\left(\frac{\epsilon^2}{D}\boldsymbol{I}\right)}. \tag{4.1.28}$$

To encode vectors that fall into the region spanned by $\hat{\boldsymbol{x}}_i$, we can cover the region with non-overlapping balls of radius $\epsilon$, as illustrated in Figure 4.5. When the volume of the region spanned by $\hat{\boldsymbol{x}}_i$ is significantly larger than the volume of the $\epsilon$-ball, the total number of balls that we need to cover the region is approximately equal to the ratio of the two volumes:

$$\#\,\epsilon\text{-balls} \approx \frac{\text{volume}(\hat{\boldsymbol{x}}_i)}{\text{volume}(\boldsymbol{w}_i)} = \sqrt{\det\left(\boldsymbol{I} + \frac{D}{N\epsilon^2}\boldsymbol{X}\boldsymbol{X}^\top\right)}. \tag{4.1.29}$$

If we use binary numbers to label all the $\epsilon$-balls in the region of interest, the total number of binary bits needed is thus

$$\mathcal{R}_\epsilon(\boldsymbol{X}) \approx \log_2(\#\,\epsilon\text{-balls}) \approx R_\epsilon(\boldsymbol{X}) \doteq \frac{1}{2}\log\det\left(\boldsymbol{I} + \frac{D}{N\epsilon^2}\boldsymbol{X}\boldsymbol{X}^\top\right). \tag{4.1.30}$$

Notice that from Example 4.4 one can show that the right-hand side of this formula is the same as the rate distortion (up to precision $\epsilon$) of a Gaussian source with covariance matrix $\frac{1}{N}\boldsymbol{X}\boldsymbol{X}^\top + (\epsilon^2/D)\boldsymbol{I}$ (e.g., the empirical covariance matrix of the data plus a noise term). In this way, a geometric argument about the volume of the enclosed vectors and sphere packing translates into an information-theoretic argument about the rate distortion of a perturbed Gaussian source.

*Example* 4.6. Figure 4.5 shows an example of a 2D distribution with an ellipsoidal support – approximating the support of a 2D Gaussian distribution. The region is covered by small balls of size $\epsilon$. All the balls are numbered from 1 to say $n$. Then given any vector $\boldsymbol{x}$ in this region, we only need to determine to which $\epsilon$-ball center it is the closest, denoted as $\mathrm{ball}_\epsilon(\boldsymbol{x})$. To remember $\boldsymbol{x}$, we only need to remember the number of this ball, which takes $\log(n)$ bits to store. If we need to decode $\boldsymbol{x}$ from this number, we simply take $\hat{\boldsymbol{x}}$ as the center of the ball. This leads to an explicit encoding and decoding scheme:

$$\boldsymbol{x} \longrightarrow \mathrm{ball}_\epsilon(\boldsymbol{x}) \longrightarrow \hat{\boldsymbol{x}} = \text{center of } \mathrm{ball}_\epsilon(\boldsymbol{x}). \tag{4.1.31}$$

One may refer to these ball centers as "codes" of a code book or a dictionary for the encoding scheme. It is easy to see that the accuracy of this (lossy) encoding-decoding scheme is about the radius of the ball $\epsilon$. Clearly $\mathcal{R}_\epsilon(\boldsymbol{Z})$ is the average number of bits required to encode the ball number of each vector $\boldsymbol{z}$ with this coding scheme, and hence can be called the *coding rate* associated with this scheme. ■

From the above derivation, we know that the coding rate $\mathcal{R}_\epsilon(\boldsymbol{X})$ is (approximately) achievable with an explicit encoding (and decoding) scheme. It has two interesting properties:

- First, one may notice that $R_\epsilon(\boldsymbol{X})$ closely resembles the rate distortion function of a Gaussian source [CT91] (see Example 4.4). Indeed, when $\epsilon$ is small, the above expression is a close approximation to the rate distortion of a Gaussian source, as pointed out by [MDH+07a].
- Second, the same closed-form coding rate $R_\epsilon(\boldsymbol{X})$ can be derived as an approximation of $\mathcal{R}_\epsilon(\boldsymbol{X})$ if the data $\boldsymbol{X}$ are assumed to be from a linear subspace. This can be shown by properly quantifying the singular value decomposition (SVD) of $\boldsymbol{X} = \boldsymbol{U}\boldsymbol{\Sigma}\boldsymbol{V}^\top$ and constructing a lossy coding scheme for vectors in the subspace spanned by $\boldsymbol{U}$ [MDH+07a].

In our context, the closed-form expression $R_\epsilon(\boldsymbol{X})$ is rather fundamental: it is the coding rate associated with an explicit and natural lossy coding scheme for data drawn from either a Gaussian distribution or a linear subspace. As we will see in the next chapter, this formula plays an important role in understanding the architecture of deep neural networks.

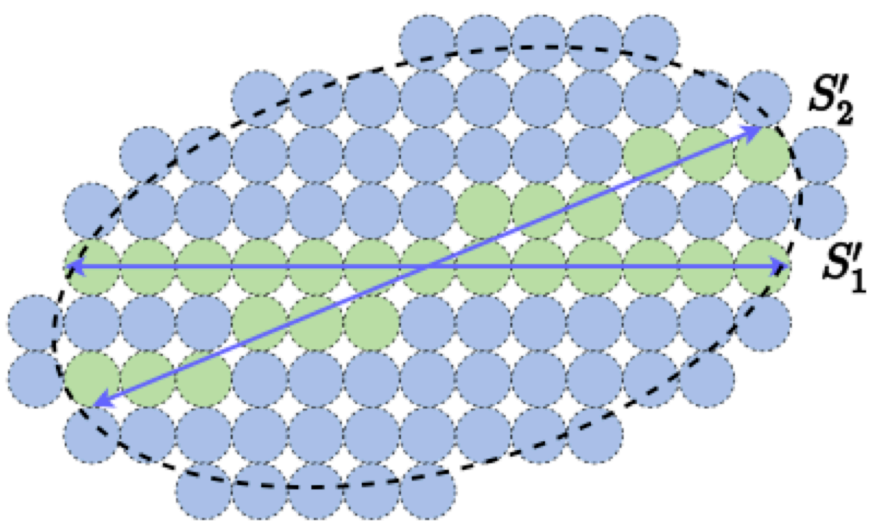

Figure 4.6: Comparison of two lossy coding schemes for data that are distributed around two subspaces. One is to pack (blue) $\epsilon$-balls for the entire space spanned by the two subspaces; the other is to pack balls only in a tabular neighborhood around the two subspaces. The latter obviously has a much smaller code book and results in a much lower coding rate for samples on the subspaces.

### 4.1.4 Clustering a Mixture of Low-Dimensional Gaussians

As we have discussed before, the given dataset $\boldsymbol{X}$ often has low-dimensional intrinsic structures. Hence, encoding it as a general Gaussian would be very redundant. If we can identify those intrinsic structures in $\boldsymbol{X}$, we could design much better coding schemes that give much lower coding rates. Or equivalently, the codes used to encode such $\boldsymbol{X}$ can be compressed. We will see that compression gives a unifying computable way to identify such structures. In this section, we demonstrate this important idea with the most basic family of low-dimensional structures: a mixture of (low-dimensional) Gaussians or subspaces.

*Example* 4.7. Figure 4.6 shows an example in which the data $\boldsymbol{X}$ are distributed around two subspaces (or low-dimensional Gaussians). If they are viewed and coded together as one single Gaussian, the associated discrete (lossy) code book, represented by all the blue balls, is obviously very redundant. We can try to identify the locations of the two subspaces, denoted by $S_1$ and $S_2$, and design a code book that only covers the two subspaces, i.e., the green balls. If we can correctly partition samples in the data $\boldsymbol{X}$ into the two subspaces: $\boldsymbol{X} = [\boldsymbol{X}_1, \boldsymbol{X}_2]\boldsymbol{\Pi}$ with $\boldsymbol{X}_1 \in S_1$ and $\boldsymbol{X}_2 \in S_2$, where $\boldsymbol{\Pi}$ denotes a permutation matrix, then the resulting coding rate for the data will be much lower. This gives a more parsimonious, hence more desirable, representation of the data. ■

So, more generally speaking, if the data are drawn from any mixture of subspaces or low-dimensional Gaussians, it would be desirable to identify those components and encode the data based on the intrinsic dimensions of those components. It turns out that we do not lose much generality by assuming that the data are drawn from a mixture of low-dimensional Gaussians. This is be-cause a mixture of Gaussians can closely approximate most general distributions [BDS16].

**The clustering problem.** Now for this specific family of distributions, how can we effectively and efficiently identify those low-dimensional components from a set of samples

$$\boldsymbol{X} = [\boldsymbol{x}_1, \boldsymbol{x}_2, \dots, \boldsymbol{x}_N], \tag{4.1.32}$$

drawn from them? In other words, given the whole data set $\boldsymbol{X}$, we want to partition, or cluster, it into multiple, say $K$, subsets:

$$\boldsymbol{X}\boldsymbol{\Pi} = [\boldsymbol{X}_1, \boldsymbol{X}_2, \dots, \boldsymbol{X}_K], \tag{4.1.33}$$

where each subset consists of samples drawn from only one low-dimensional Gaussian or subspace and $\boldsymbol{\Pi}$ is a permutation matrix to indicate membership of the partition. Note that, depending on the situation, the partition could be either deterministic or probabilistic. As shown in [MDH+07b], for a mixture of Gaussians, probabilistic partition does not lead to a lower coding rate. So for simplicity, we here consider a deterministic partition only.

**Clustering via lossy compression.** The main difficulty in solving the above clustering problem is that we normally do not know the number of clusters $K$, nor do we know the dimension of each component. There has been a long history for the study of this clustering problem. The textbook [VMS16] gives a systematic and comprehensive coverage of different approaches to this problem. To find an effective approach to this problem, we first need to understand and clarify why we want to cluster. In other words, what exactly do we gain from clustering the data, compared with not to? How do we measure the gain? From the perspective of data compression, a correct clustering should lead to a more efficient encoding (and decoding) scheme.

For any given data set $\boldsymbol{X}$, there are already two obvious encoding schemes as the baseline. They represent two extreme ways to encode the data:

- Simply view all the samples together drawn as from one single Gaussian. The associated coding rate is, as derived before, given by:

$$\mathcal{R}_\epsilon(\boldsymbol{X}) \approx R_\epsilon(\boldsymbol{X}) = \frac{1}{2}\log\det\left(\boldsymbol{I} + \frac{D}{N\epsilon^2}\boldsymbol{X}\boldsymbol{X}^\top\right). \tag{4.1.34}$$

- Simply memorize all the samples separately by assigning a different number to each sample. The coding rate would be:

$$\mathcal{R}_0(\boldsymbol{X}) = \log(N). \tag{4.1.35}$$

Note that either coding scheme can become the "optimal" solution for certain (extreme) choice of the quantization error $\epsilon$:

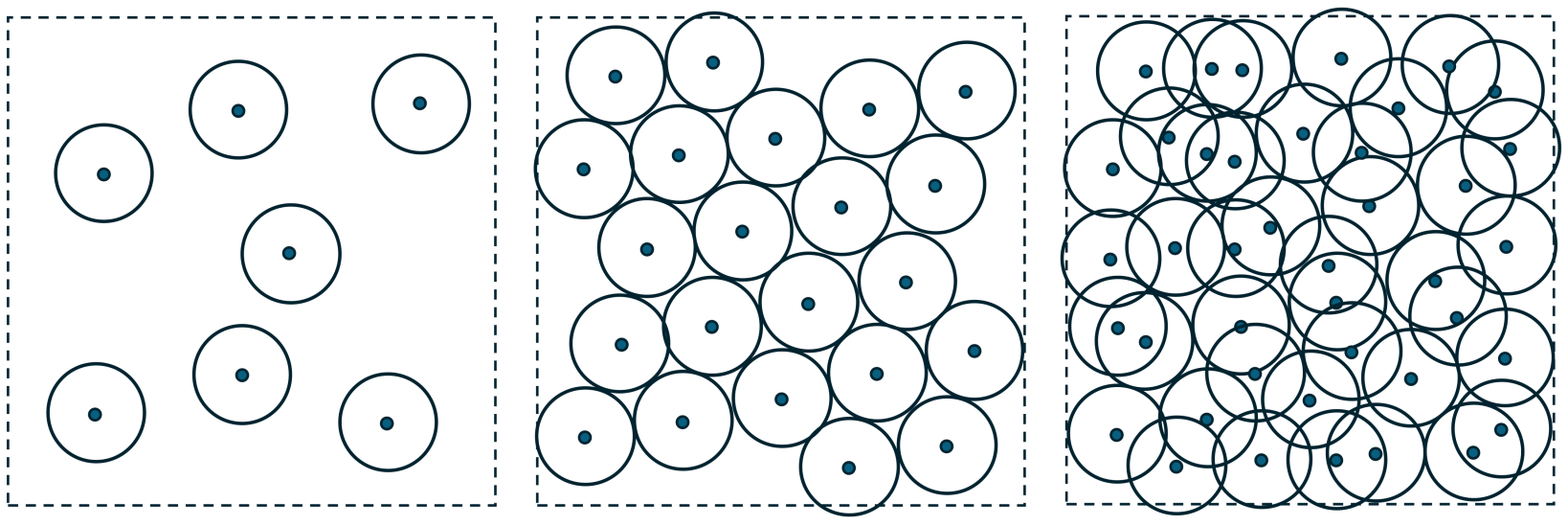

Figure 4.7: A number of random samples on a 2D plane. Consider an $\epsilon$-disc assigned to each sample with the sample as its center. The density of the samples increases from left to right.

1. *Lazy Regime*: If we choose $\epsilon$ to be extremely large, all samples in $\boldsymbol{X}$ can be covered by a single ball. The rate is $\lim_{\epsilon \to \infty} \mathcal{R}_\epsilon \to \frac{1}{2} \log \det(\boldsymbol{I}) = 0$.

2. *Memorization Regime*: If $\epsilon$ is extremely small, every sample in $\boldsymbol{X}$ is covered by a different $\epsilon$-ball, hence the total is $N$. The rate is $\lim_{\epsilon \to 0} \mathcal{R}_\epsilon \to \log(N)$.

Note that the first scheme corresponds to the scenario when one does not care about anything interesting about the distribution at all. One does not want to spare any bit for anything informative. We call this the "lazy regime." The second scheme corresponds to the scenario when one wants to decode every sample with an extremely high precision. So one would better "memorize" every sample. We call this the "memorization regime."

Notice that these asymptotics in terms of $\epsilon$ are exactly those discussed when introducing the lossy coding rate as a quantity that realistically-trained diffusion models approximate, as in the end of Section 3.3. In the current situation, we *explicitly* code the distribution up to precision $\epsilon$; in the case of diffusion models, we *implicitly* code the distribution. Certain properties of the scheme such as memorization and generalization manifest similarly in both cases, and indeed one can study one case to understand the other.

*Example* 4.8. To see when the memorization regime is preferred or not, let us consider a number, say $N$, of samples randomly distributed in a unit area on a 2D plane.[14] Imagine we try to design a lossy coding scheme with a fixed quantization error $\epsilon$. This is equivalent to putting an $\epsilon$-disc around each sample, as shown in Figure 4.7. When $N$ is small, the chance that all the discs overlap with each other is zero. A codebook of size $N$ is necessary and optimal in this case. When $N$ or the density reaches a certain critical value $N_c$, with high probability all the discs start to overlap and connect into one cluster that covers the whole plane—this phenomenon is known as continuum "percolation" [Gil61; MM12]. When $N$ becomes larger than this value, the discs overlap heavily. The number $N$ of discs becomes very redundant because we only want to encode

[14] Say the points are drawn by a Poisson process with density $N$ points per unit area.

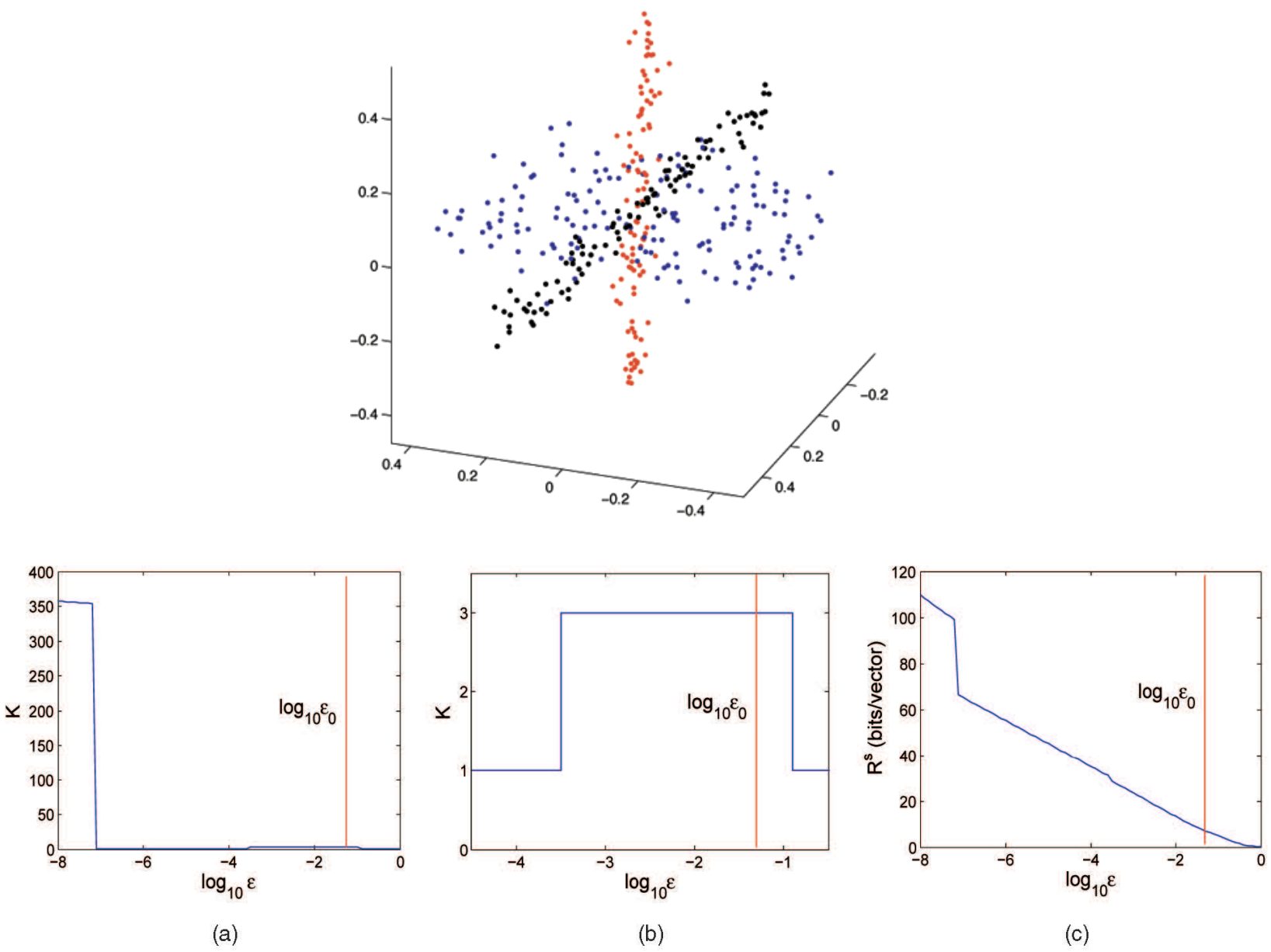

Figure 4.8: Top: 358 noisy samples drawn from two lines and one plane in $\mathbb{R}^3$. Bottom: the effect of varying $\epsilon$ on the clustering result and the coding rate. The red line marks the variance $\epsilon_0$ of the Gaussian noise added to the samples. Note that the bottom left and middle plots show the same quantity with different axis locations and scales.

points on the plane up to the given precision $\epsilon$. The number of discs needed to cover all the samples is much less than $N$.[15] ■

Both the lazy and memorization regimes are somewhat trivial and perhaps are of little theoretical or practical interest. Either scheme would be far from optimal when used to encode a large number of samples drawn from a distribution that has a *compact and low-dimensional support*. The interesting regime exists in between these two.

*Example* 4.9. Figure 4.8 shows an example with noisy samples drawn from two lines and one plane in $\mathbb{R}^3$. As we notice from the plot (c) on the right, the optimal coding rate decreases monotonically as we increase $\epsilon$, as anticipated from the property of the rate distortion function. The plots (a) and (b) show, when varying $\epsilon$ from very small (near zero) to very large (towards infinite), the optimal number of clusters when the coding rate is minimal. We can clearly see the lazy regime and the memorization regime on the two ends of the plots. But one can also notice in plot (b), when the quantization error $\epsilon$ is chosen to be around the level of the true noise variance $\epsilon_0$, the optimal number of clusters is the "correct" number three that represents two lines and one plane.

[15] In fact, there are efficient algorithms to find such a covering [BBF+01].

We informally refer to this middle regime as the "generalization regime". Notice that a sharp phase transition takes place between these regimes.[16] ■

From the above discussion and examples, we see that, when the quantization error relative to the sample density[17] is in a proper range, minimizing the lossy coding rate would allow us to uncover the underlying (low-dimensional) distribution of the sampled data. *Hence, quantization, started as a choice of practicality, seems to be becoming necessary for learning a continuous distribution from its empirical distribution with finite samples.* Although a rigorous theory for explaining this phenomenon remains elusive, here, for learning purposes, we care about how to exploit the phenomenon to design algorithms that can find the correct distribution.

Let us use the simple example shown in Figure 4.6 to illustrate the basic ideas. If one can partition all samples in $\boldsymbol{X}$ into two clusters in $\boldsymbol{X}_1$ and $\boldsymbol{X}_2$, with $N_1$ and $N_2$ samples respectively, then the associated coding rate would be[18]

$$R_\epsilon^c(\boldsymbol{X} \mid \boldsymbol{\Pi}) = \frac{N_1}{N} R_\epsilon(\boldsymbol{X}_1) + \frac{N_2}{N} R_\epsilon(\boldsymbol{X}_2), \tag{4.1.36}$$

where we use $\boldsymbol{\Pi}$ to indicate membership of the partition. If the partition respects the low-dimensional structures of the distribution, in this case $\boldsymbol{X}_1$ and $\boldsymbol{X}_2$ belonging to the two subspaces respectively, then the resulting coding rate should be significantly smaller than the above two basic schemes:

$$R_\epsilon^c(\boldsymbol{X} \mid \boldsymbol{\Pi}) \ll R_\epsilon(\boldsymbol{X}), \quad R_\epsilon^c(\boldsymbol{X} \mid \boldsymbol{\Pi}) \ll R_0(\boldsymbol{X}). \tag{4.1.37}$$

In general, we can cast the clustering problem into an optimization problem that minimizes the coding rate:

$$\min_{\boldsymbol{\Pi}} \left\{ R_\epsilon^c(\boldsymbol{X} \mid \boldsymbol{\Pi}) \doteq \sum_{k=1}^{K} \frac{N_k}{N} R_\epsilon(\boldsymbol{X}_k) \right\}. \tag{4.1.38}$$

**Optimization strategies to cluster.** The remaining question is how we optimize the above coding rate objective to find the optimal clusters. There are three natural approaches to this objective:

1. We may start with the whole set $\boldsymbol{X}$ as a single cluster (i.e. the lazy regime) and then search (say randomly) to partition it so that it would lead to a smaller coding rate.

2. Inversely, we may start with each sample $\boldsymbol{x}_i$ as its own cluster (i.e. the memorization regime) and search to merge clusters that would result in a smaller coding rate.

[16] So far, to our best knowledge, there is no rigorous theoretical justification for these phase transition behaviors.

[17] or the sample density relative to the quantization error

[18] We here ignore some overhead bits needed to encode the membership for each sample, say via the Huffman coding.

3. Alternatively, if we could represent (or approximate) the membership $\boldsymbol{\Pi}$ as some continuous parameters, we may use optimization methods such as gradient descent (GD).

The first approach is not so appealing computationally as the number of possible partitions that one needs to try is exponential in the number of samples. For example, the number of partitions of $\boldsymbol{X}$ into two subsets of equal size is $\binom{N}{N/2}$ which explodes as $N$ becomes large. We will explore the third approach in the next Chapter 5. There, we will see how the role of deep neural networks, transformers in particular, is connected with the coding rate objective.

The second approach was originally suggested in the work of [MDH+07b]. It demonstrates the benefit of being able to evaluate the coding rate efficiently (say with an analytical form). With it, the (low-dimensional) clusters of the data can be found rather efficiently and effectively via the principle of minimizing coding length (MCL). Note that for a cluster $\boldsymbol{X}_k$ with $N_k$ samples, the length of binary bits needed to encode all the samples in $\boldsymbol{X}_k$ is given by:[19]

$$L(\boldsymbol{X}_k) = N_k R_\epsilon(\boldsymbol{X}_k). \tag{4.1.39}$$

If we have two clusters $\boldsymbol{X}_k$ and $\boldsymbol{X}_l$, if we want to code the samples as two separate clusters, the length of binary bits needed is

$$L^c(\boldsymbol{X}_k, \boldsymbol{X}_l) = N_k R_\epsilon(\boldsymbol{X}_k) + N_l R_\epsilon(\boldsymbol{X}_l) - N_k \log \frac{N_k}{N_k + N_l} - N_l \log \frac{N_l}{N_k + N_l}.$$

The last two terms are the number of bits needed to encode the memberships of samples according to the Huffman code.

Then, given any two separate clusters $\boldsymbol{X}_1$ and $\boldsymbol{X}_2$, we can decide whether to merge them or not based on the difference between the two coding lengths:

$$L(\boldsymbol{X}_k \cup \boldsymbol{X}_l) - L^c(\boldsymbol{X}_k, \boldsymbol{X}_l) \tag{4.1.40}$$

is positive or negative and $\boldsymbol{X}_k \cup \boldsymbol{X}_l$ denotes the union of the sets of samples in $\boldsymbol{X}_k$ and $\boldsymbol{X}_l$. If it is positive , it means the coding length would become smaller if we merge the two clusters into one. This simple fact leads to the following clustering algorithm proposed by [MDH+07b]:

Note that this algorithm is tractable as the total number of (pairwise) comparisons and merges is about $O(N^2 \log N)$. However, due to its greedy nature, there is no theoretical guarantee that the process will converge to the globally optimal clustering solution. Nevertheless, as reported in [MDH+07b], in practice, this seemingly simple algorithm works extremely well. The clustering results plotted in Figure 4.8 were actually computed by this algorithm.

*Example* 4.10 (Image Segmentation). The above measure of coding length and the associated clustering algorithm assume the data distribution is a mixture

[19] In fact, a more accurate estimate of the coding length is $L(\boldsymbol{X}_k) = (N_k + D) R_\epsilon(\boldsymbol{X}_k)$ where the extra bits are used to encode the basis of the subspace [MDH+07b]. Here we omit this overhead for simplicity.

**Algorithm 4.1** Pairwise Steepest Descent of Coding Length

**Input:** $N$ data points $\{\boldsymbol{x}_i\}_{i=1}^N$
**Output:** A set $\mathcal{C}$ of clusters

```
1:  procedure PairwiseSteepestDescentOfCodingLength({x_i}_{i=1}^N)
        # Initialize N clusters X_k with one element each
2:      C ← {{x_i}}_{i=1}^N

3:      while |C| > 1 do

            # If no bits are saved by any merging
4:          if min_{X_k, X_l ∈ C} [L(X_k ∪ X_l) − L^c(X_k, X_l)] ≥ 0 then
              # Early return C and exit
5:            return C
6:          else
              # Merge clusters which save the most bits
7:            X_{k*}, X_{l*} ← arg min_{X_k, X_l ∈ C} [L(X_k ∪ X_l) − L^c(X_k, X_l)]

              # Remove unmerged clusters and add back the merged one
8:            C ← [C \ {X_{k*}, X_{l*}}] ∪ {X_{k*} ∪ X_{l*}}
9:          end if
10:     end while
        # If all merges yield savings, return one cluster
11:     return C
12: end procedure
```

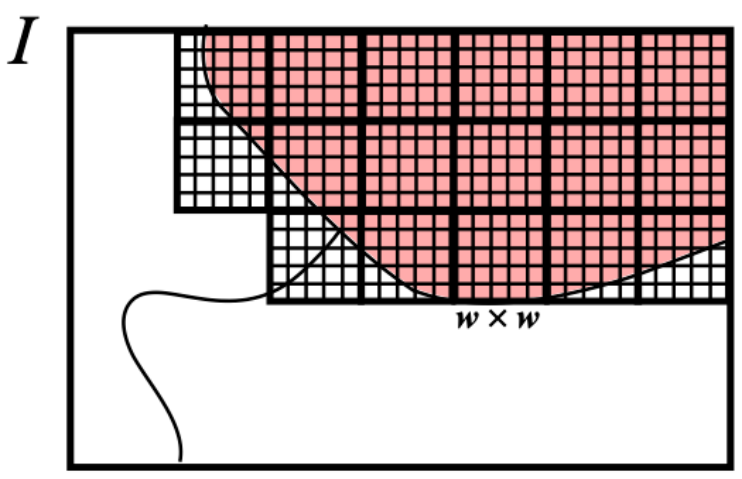

Figure 4.9: Image patches with a size of $w \times w$ pixels.

of (low-dimensional) Gaussians. Although this seems somewhat idealistic, the measure and algorithm can already be very useful and even powerful in scenarios when the model is (approximately) valid.

For example, a natural image typically consists of multiple regions with nearly homogeneous textures. If we take many small windows from each region, they should resemble samples drawn from a (low-dimensional) Gaussian, as illustrated in Figure 4.9. Figure 4.10 shows the results of image segmentation based on applying the above clustering algorithm to the image patches directly. More technical details regarding customizing the algorithm to the image segmentation problem can be found in [MRY+11]. ■

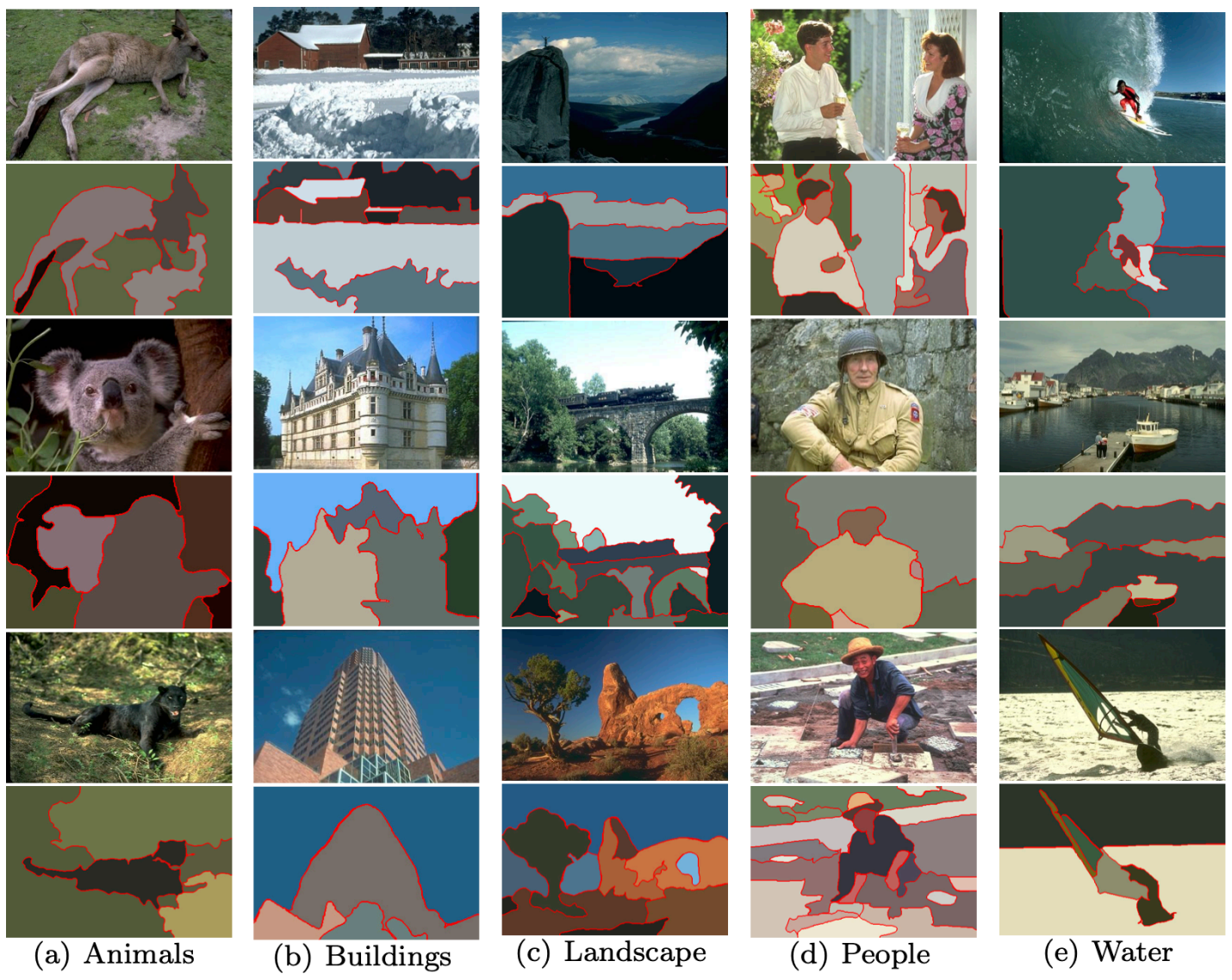

Figure 4.10: Segmentation results based on the clustering algorithm applied to the image patches.

## 4.2 Transformation via Maximizing Information Gain

So far in this chapter, we have discussed how to identify a distribution with low-dimensional structures through the principle of compression. As we have seen from the previous two sections, computational compression can be realized through either the denoising operation or clustering. Figure 4.11 illustrates this concept with our favorite example.

It is important to note that the clustering algorithm and the coding rate formulas developed in the previous section assumed that the data distribution is *already* (approximately) a mixture of low-dimensional Gaussians. In practice, the raw data—images, audio, text—rarely satisfy this assumption directly: their intrinsic structures are typically *nonlinear*. The purpose of this section is to develop a framework for *transforming* data so that their representations do take the form of a mixture of incoherent low-dimensional subspaces, at which point the tools from the preceding section become directly applicable. That is, the preceding section answers "how to compress and cluster data that are already Gaussian," while this section answers "how to transform data so that they become Gaussian." As we will see in Chapter 5, deep networks are precisely the computational mechanism for realizing such transformations, with the network's architecture derived from optimizing the coding rate objective developed here.

Indeed, the ultimate goal for identifying a data distribution is to use it to facilitate certain subsequent tasks such as segmentation, classification, or generation (of images). Hence, how the resulting distribution is "represented" matters tremendously with respect to how information related to these subsequent tasks can be efficiently and effectively retrieved and utilized. This naturally raises a fundamental question: *what makes a representation truly "good" for downstream use?* In the following, we will explore the essential properties that a meaningful and useful representation should possess, and how these properties can be explicitly characterized and pursued via maximizing information gain.

**How to measure the goodness of representations.** One may view a given dataset as samples of a random vector $\boldsymbol{x}$ with a certain distribution in a high-dimensional space, say $\mathbb{R}^D$. Typically, the distribution of $\boldsymbol{x}$ has a much lower intrinsic dimension than the ambient space. Generally speaking, *learning a representation* refers to learning a continuous mapping, say $f(\cdot)$, that transforms $\boldsymbol{x}$ to a so-called *feature vector* $\boldsymbol{z}$ in another (typically lower-dimensional) space, say $\mathbb{R}^d$, where $d < D$. It is hopeful that through such a mapping

$$\boldsymbol{x} \in \mathbb{R}^D \xrightarrow{\ f(\boldsymbol{x})\ } \boldsymbol{z} \in \mathbb{R}^d, \tag{4.2.1}$$

the low-dimensional intrinsic structures of $\boldsymbol{x}$ are identified and represented by $\boldsymbol{z}$ in a more compact and structured way so as to facilitate subsequent tasks such as classification or generation. The feature $\boldsymbol{z}$ can be viewed as a (learned)

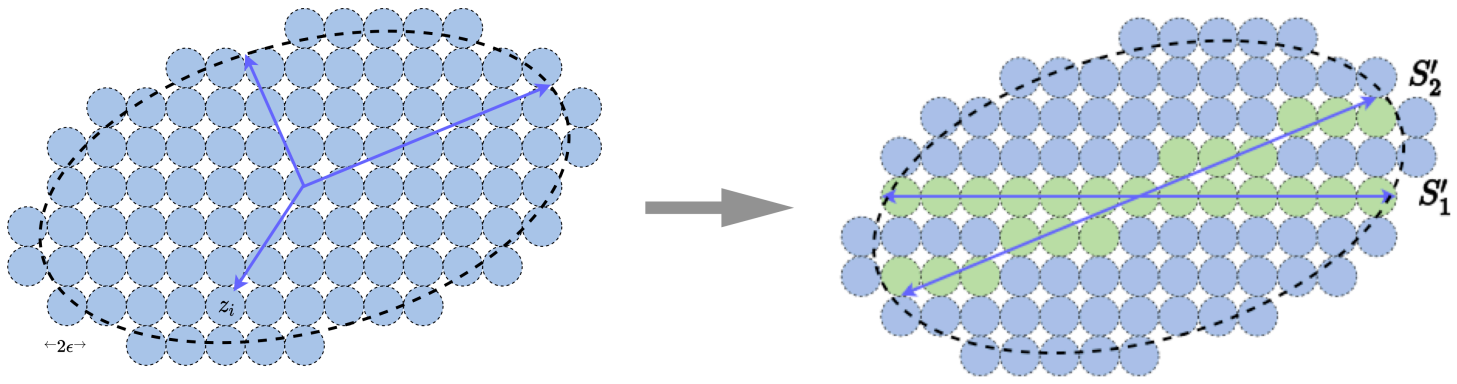

Figure 4.11: Identifying a low-dimensional distribution with two subspaces (left) via denoising or clustering, starting from a generic random Gaussian distribution (right).

compact code for the original data $\boldsymbol{x}$, so the mapping $f$ is also called an *encoder*. The fundamental question of representation learning is

*What is a principled and effective measure for the goodness of representations?*

Conceptually, the quality of a representation $\boldsymbol{z}$ depends on how well it identifies the most relevant and sufficient information of $\boldsymbol{x}$ for subsequent tasks and how efficiently it represents this information. For a long time, it was believed and argued that the "sufficiency" or "goodness" of a learned feature representation should be defined in terms of a specific task. For example, $\boldsymbol{z}$ just needs to be sufficient for predicting the class label $\boldsymbol{y}$ in a classification problem. Below, let us start with the classic problem of image classification and argue why such a notion of a task-specific "representation" is limited and needs to be generalized.

### 4.2.1 Representation Learning from Classified Data

Suppose that $\boldsymbol{x} \in \mathbb{R}^D$ is a random vector drawn from a mixture of $K$ (component) distributions $\mathcal{D} = \{\mathcal{D}_k\}_{k=1}^K$. Given a finite set of i.i.d. samples $\boldsymbol{X} = [\boldsymbol{x}_1, \boldsymbol{x}_2, \ldots, \boldsymbol{x}_N] \in \mathbb{R}^{D \times N}$ of the random vector $\boldsymbol{x}$, we *seek a good representation* through a continuous mapping $f(\boldsymbol{x}) : \mathbb{R}^D \to \mathbb{R}^d$ that captures intrinsic structures of $\boldsymbol{x}$ and best facilitates the subsequent classification task.[20] To ease the task of learning distribution $\mathcal{D}$, in the popular supervised classification setting, a true class label (or a code word for each class), usually represented by a one-hot vector $\boldsymbol{y}_i \in \mathbb{R}^K$, is given for each sample $\boldsymbol{x}_i$.

**Encoding class information via cross entropy.** Extensive studies have shown that for many practical datasets (e.g., images, audio, and natural languages), the (encoding) mapping from the data $\boldsymbol{x}$ to its class label $\boldsymbol{y}$ can be

[20]Classification is the domain where deep learning demonstrated the initial success, sparking the explosive interest in deep networks. Although our study focuses on classification, we believe the ideas and principles can be naturally generalized to other settings, such as regression.

effectively modeled by training a deep network,[21] here denoted as

$$f(\boldsymbol{x}, \theta) : \boldsymbol{x} \mapsto \boldsymbol{y}$$

with network parameters $\theta \in \Theta$, where $\Theta$ denotes the parameter space. For the output $f(\boldsymbol{x}, \theta)$ to match well with the label $\boldsymbol{y}$, we like to minimize the *cross-entropy loss* over a training set $\{(\boldsymbol{x}_i, \boldsymbol{y}_i)\}_{i=1}^N$:[22]

$$\min_{\theta \in \Theta} \mathbb{E}[\mathrm{CE}(\boldsymbol{y} \parallel f(\boldsymbol{x}, \theta))] \approx -\frac{1}{N} \sum_{i=1}^{N} \langle \boldsymbol{y}_i, \log\left(f(\boldsymbol{x}_i, \theta)\right) \rangle. \tag{4.2.2}$$

The optimal network parameters $\theta$ are typically found by optimizing the above objective through an efficient gradient descent scheme, with gradients computed via back propagation, as described in Section A.2.3 of Chapter A.

Despite its effectiveness and enormous popularity, there are two serious limitations with this approach: (1) It aims only to predict the labels $\boldsymbol{y}$ even if they might be mislabeled. Empirical studies show that deep networks, used as a "black box," can even fit random labels [ZBH+17]. (2) With such an end-to-end data fitting, despite plenty of empirical efforts in trying to interpret the so-learned features, it is not clear to what extent the intermediate features learned by the network capture the intrinsic structures of the data that make meaningful classification possible in the first place. The precise geometric and statistical properties of the learned features are also often obscured, which leads to the lack of interpretability and subsequent performance guarantees (e.g., generalizability, transferability, and robustness, etc.) in deep learning. Therefore, *one of the goals of this section is to address such limitations by reformulating the objective towards learning explicitly meaningful and useful representations for the data $\boldsymbol{x}$, not limited to classification.*

**Minimal discriminative features via information bottleneck.** One popular approach to interpret the role of deep networks is to view outputs of intermediate layers of the network as selecting certain latent features $\boldsymbol{z} = f(\boldsymbol{x}, \theta) \in \mathbb{R}^d$ of the data that are discriminative among multiple classes. Learned representations $\boldsymbol{z}$ then facilitate the subsequent classification task for predicting the class label $\boldsymbol{y}$ by optimizing a classifier $g(\boldsymbol{z})$:

$$\boldsymbol{x} \xrightarrow{\ f(\boldsymbol{x}, \theta)\ } \boldsymbol{z} \xrightarrow{\ g(\boldsymbol{z})\ } \boldsymbol{y}. \tag{4.2.3}$$

We know from (4.1.13) that *the mutual information* between two random variables, say $\boldsymbol{x}, \boldsymbol{z}$, is defined to be

$$I(\boldsymbol{x}; \boldsymbol{z}) = H(\boldsymbol{x}) - H(\boldsymbol{x} \mid \boldsymbol{z}), \tag{4.2.4}$$

[21] Here let us not worry about yet which network we should use here and why. The purpose here is to consider any empirically tested deep network. We will leave the justification of the network architectures to the next chapter.

[22] Here we interpret $\boldsymbol{y}$ as a particular *one-hot* probability distribution over $[K] = \{1, 2, \dots, K\}$, and $f(\boldsymbol{x}, \theta) \in \Delta([K]) \subseteq \mathbb{R}^K$ to also be a probability distribution over the same set, which can therefore be compared via the cross-entropy despite being computationally represented as vectors in $\mathbb{R}^K$.

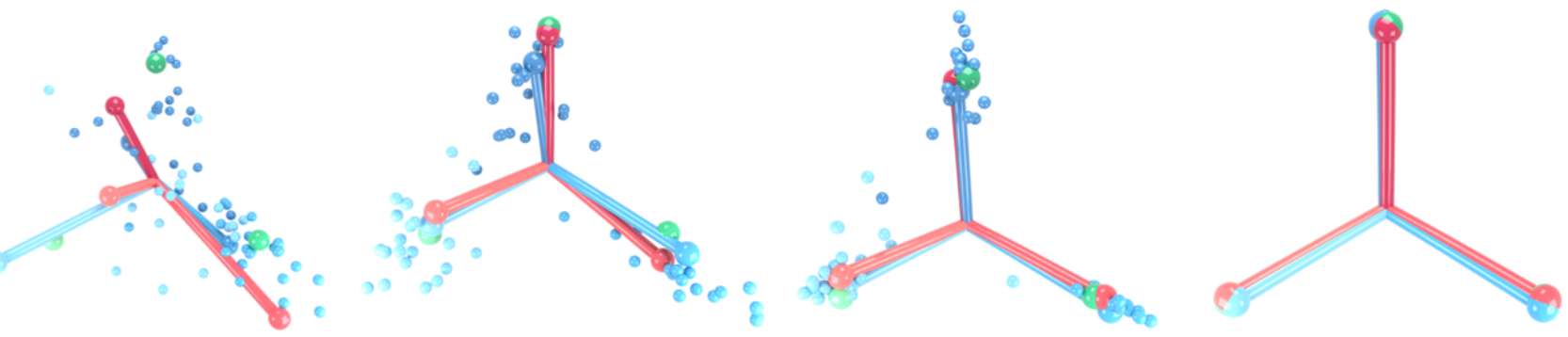

Figure 4.12: Evolution of penultimate layer outputs of a VGG13 neural network when trained on the CIFAR-10 dataset with 3 randomly selected classes. Figure from [PHD20].

where $H(\boldsymbol{x}|\boldsymbol{z})$ is the conditional entropy of $\boldsymbol{x}$ given $\boldsymbol{z}$. The mutual information is also known as *the information gain*: It measures how much the entropy of the random variable $\boldsymbol{x}$ can be reduced once $\boldsymbol{z}$ is given. Or equivalently, it measures how much information $\boldsymbol{z}$ contains about $\boldsymbol{x}$. The *information bottleneck* (IB) formulation [TZ15] further hypothesizes that the role of the network is to learn $\boldsymbol{z}$ as the minimal sufficient statistics for predicting $\boldsymbol{y}$. Formally, it seeks to maximize the mutual information $I(\boldsymbol{z}, \boldsymbol{y})$ between $\boldsymbol{z}$ and $\boldsymbol{y}$ while minimizing the mutual information $I(\boldsymbol{x}, \boldsymbol{z})$ between $\boldsymbol{x}$ and $\boldsymbol{z}$:

$$\max_{\theta \in \Theta} \mathrm{IB}(\boldsymbol{x}, \boldsymbol{y}, \boldsymbol{z}) \doteq I(\boldsymbol{z}; \boldsymbol{y}) - \beta I(\boldsymbol{x}; \boldsymbol{z}) \quad \text{s.t. } \boldsymbol{z} = f(\boldsymbol{x}, \theta), \tag{4.2.5}$$

where $\beta > 0$.

Given one can overcome some caveats associated with this framework [KTV18], such as how to accurately evaluate mutual information with finite samples of degenerate distributions, this framework can be helpful in explaining certain behaviors of deep networks. For example, recent work [PHD20] indeed shows that the representations learned via the cross-entropy loss (4.2.2) exhibit a *neural collapse* phenomenon. That is, features of each class are mapped to a one-dimensional vector whereas all other information of the class is suppressed, as illustrated in Figure 4.12.[23]

*Remark* 4.4 (Neural Collapse). Neural collapse refers to a phenomenon observed in deep neural networks trained for classification, where the learned feature representations and classifier weights exhibit highly symmetric and structured behavior during the terminal phase of training [PHD20; YWZ+22; ZDZ+21]. Specifically, within each class, features collapse to their class mean, and across classes, these means become maximally separated, forming a simplex equiangular configuration. The linear classifier aligns with the class mean up to rescaling. Additionally, the last-layer classifier converges to choosing whichever class has the nearest train class mean. Neural collapse reveals deep connections between optimization dynamics, generalization, and geometric structures arising in supervised learning.

[23] We will observe this phenomenon in experiments with real imagery data later in Section 4.3.1.

From the above example of classification, we see that the so-learned representation $\boldsymbol{z}$ of $\boldsymbol{x}$ gives a very simple encoder that essentially maps each class of data to only one code word: the one-hot vector representing each class. From the lossy compression perspective, such an encoder is too lossy to preserve information in the data distribution. Other information, such as that useful for tasks such as image segmentation, completion, or generation, is severely lost in such a supervised learning process.

To remedy this situation, we want to learn a different encoding scheme such that the resulting feature representation $\boldsymbol{z}$ can capture as much information about the data distribution as possible, not limited to that useful for classification alone. More precisely, we want the representation $\boldsymbol{z} = f(\boldsymbol{x}, \theta)$ to be such that there exists a decoder $g(\boldsymbol{z}, \eta)$ so that:

$$\|\boldsymbol{x} - \hat{\boldsymbol{x}}\|_2 \leq \epsilon \quad \text{s.t.} \quad \hat{\boldsymbol{x}} = g(f(\boldsymbol{x})). \tag{4.2.6}$$

We will formally introduce such an encoding-decoding scheme as an "autoencoder" and such a representation $\boldsymbol{z}$ as a "consistent" representation in Chapter 6. For now, the reader may simply view this as a generalization to the quantization code $\hat{\boldsymbol{x}}$ introduced in the preceding section for the rate distortion (in the native data $\boldsymbol{x}$ domain).

In the formulation of information bottleneck (4.2.5), it does not require the learned feature representation $\boldsymbol{z}$ to be consistent. Hence, at least it should be fixed with an objective for a (consistent) representation learning (RL):

$$\begin{aligned} \max_{\theta,\eta} \; & \mathrm{RL}(\boldsymbol{x}, \boldsymbol{y}, \boldsymbol{z}) \doteq I(\boldsymbol{z}; \boldsymbol{y}) - \beta I(\boldsymbol{x}; \boldsymbol{z}) \\ \text{s.t.} \quad & \boldsymbol{z} = f(\boldsymbol{x}, \theta), \quad \|\boldsymbol{x} - g(\boldsymbol{z}, \eta)\|_2 \leq \epsilon. \end{aligned} \tag{4.2.7}$$

Solving the above problem in the most general setting is a very daunting task. In general, we do not even know what classes of encoders or decoders could ensure the consistency[24] Also, since we do not know the distribution of the representation $\boldsymbol{z}$, computing the mutual information $I(\boldsymbol{z}; \boldsymbol{y})$ can be very challenging, if not impossible.

Before we study how to approach this problem in general settings (in later Chapters), for the rest of this chapter, we will demonstrate how to solve it, in some limited way, for the special setting when the data distribution consists of clear (low-dimensional) classes.

### 4.2.2 Linear Discriminative Representations

Whether the given data $\boldsymbol{X}$ of a mixed distribution $\mathcal{D}$ can be effectively classified or clustered depends on how separable (or discriminative) the component distributions $\mathcal{D}_k$ are (or can be made). One popular working assumption is that the distribution of each class has relatively *low-dimensional* intrinsic structures. Hence we may assume that the distribution $\mathcal{D}_k$ of each class has a support on a low-dimensional submanifold, say $\mathcal{M}_k$ with dimension $d_k \ll D$, and the distribution $\mathcal{D}$ of $\boldsymbol{x}$ is supported on the mixture of those submanifolds, $\mathcal{M} = \cup_{k=1}^K \mathcal{M}_k$, in the high-dimensional ambient space $\mathbb{R}^D$, as illustrated before in Figure 4.1.

[24] In practice, people tend to learn them, separately from the above objective, through trial and error, as we will discuss more in Chapter 6.

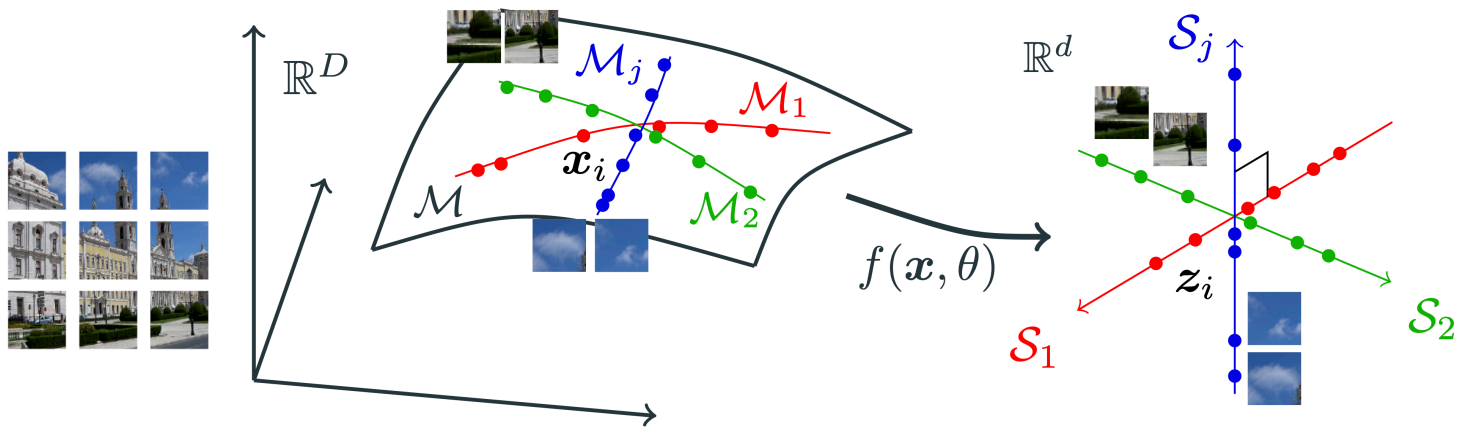

Figure 4.13: The distribution $\mathcal{D}$ of high-dimensional data $\boldsymbol{x} \in \mathbb{R}^D$ is supported on a manifold $\mathcal{M}$ and its classes on low-dimensional submanifolds $\mathcal{M}_k$. We aim to learn a mapping $f(\boldsymbol{x},\theta)$ parameterized by $\theta$ such that $\boldsymbol{z}_i = f(\boldsymbol{x}_i,\theta)$ lie on a union of maximally uncorrelated subspaces $\{\mathcal{S}_k\}$.

Not only do we need to identify the low-dimensional distribution, but we also want to represent the distribution in a form that best facilitates subsequent tasks such as classification, clustering, and conditioned generation (as we will see in the future). To do so, we require our learned feature representations to have the following properties:

1. *Within-Class Compressible:* Features of samples from the same class should be strongly *correlated* in the sense that they belong to a low-dimensional linear subspace.

2. *Between-Class Discriminative:* Features of samples from different classes should be highly *uncorrelated* and belong to different incoherent low-dimensional linear subspaces.

3. *Maximally Diverse Representation:* Dimension (or variance) of the features of each class should be *as large as possible* as long as they are incoherent to the other classes.

We refer to such a representation the *linear discriminative representation* (LDR).

Notice that the first property aligns well with the objective of the classic *principal component analysis* (PCA) that we have discussed in Section 2.1.1. It essentially says that features of samples that belong to the same class[25] should become or be made highly correlated. In other words, feature representation should be *contractive* for samples that are similar.

The second property requires that for samples that belong to different classes or are believed to be dissimilar, their features should be made (geometrically) incoherent or (statistically) independent, as illustrated before in Figure 4.1. Figure 4.13 illustrates when the data distribution is actually a mixture of low-dimensional submanifolds. Through compression (denoising or clustering), we

[25] including augmented instances of the same samples

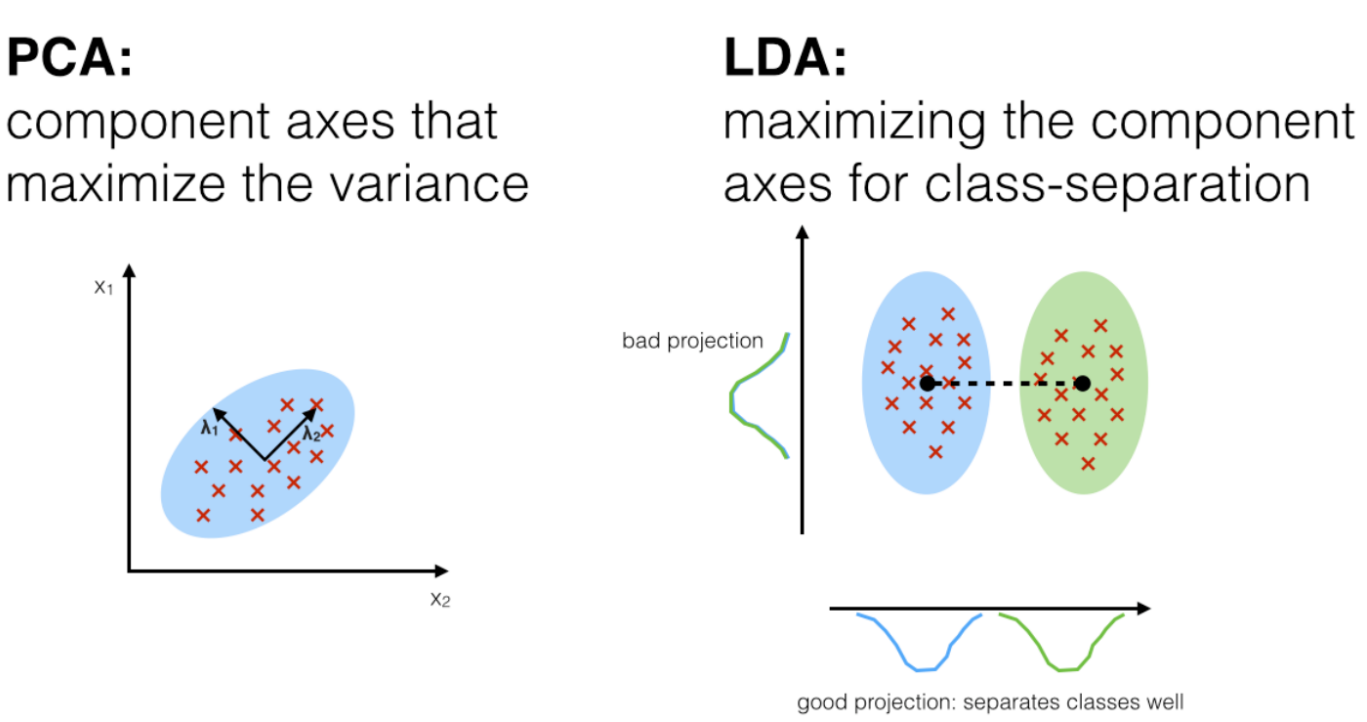

Figure 4.14: Comparison between PCA and LDA. Figures adopted from https://sebastianraschka.com/Articles/2014_python_lda.html.

first identify that the true data distribution (left). We then would like to transform the distribution so that the submanifolds eventually become mutually incoherent/independent linear subspaces (right).

*Remark* 4.5 (Linear Discriminative Analysis). One may notice that the second criterion resembles that of the classic *linear discriminant analysis* (LDA) [HTF09]. Linear discriminant analysis (LDA) [HTF09] is a supervised dimensionality reduction technique that aims to find a linear projection of data that maximizes class separability. Specifically, given labeled data, LDA seeks a linear transformation that projects high-dimensional inputs onto a lower-dimensional space where the classes are maximally separated. Note that PCA is an unsupervised method that projects data onto directions of maximum variance without considering class labels. While PCA focuses purely on preserving global variance structure, LDA explicitly exploits label information to enhance discriminative power; see the comparison in Figure 4.14. But LDA only performs a linear projection of the original data. As we will see LDR can be more general: it seeks a, probably nonlinear, transformation to make the resulting representation linear and incoherent.

*Remark* 4.6 (Contrastive Learning). One may notice that the conventional *contrastive learning* [HCL06; HFW+19; OLV18] aims to achieve similar goals for feature representation as the first two properties stipulate. For data that belong to $k$ different classes, a randomly chosen pair $(\boldsymbol{x}_i, \boldsymbol{x}_j)$ is of high probability belonging to different classes if $k$ is large.[26] Hence it is desirable that the representation $\boldsymbol{z}_i = f(\boldsymbol{x}_i, \boldsymbol{\theta})$ of a sample $\boldsymbol{x}_i$ should be highly incoherent to those $\boldsymbol{z}_j$ of other samples $\boldsymbol{x}_j$ whereas coherent to feature of its transformed version $\tau(\boldsymbol{x}_i)$, denoted as $\boldsymbol{z}(\tau(\boldsymbol{x}_i))$ for $\tau$ in certain augmentation set $\mathcal{T}$ in consideration. Hence it was proposed heuristically that, to promote discriminativeness of the

[26]For example, when $k \geq 100$, a random pair is of probability 99% belonging to different classes.

learned representation, one may seek to minimize the so-called *contrastive loss*:

$$\min_{\boldsymbol{\theta}} -\log \frac{\exp(\langle \boldsymbol{z}_i, \boldsymbol{z}(\tau(\boldsymbol{x}_i))\rangle)}{\sum_{j\neq i}\exp(\langle \boldsymbol{z}_i, \boldsymbol{z}_j\rangle)}, \tag{4.2.8}$$

which is small whenever the inner product $\langle \boldsymbol{z}_i, \boldsymbol{z}(\tau(\boldsymbol{x}_i))\rangle$ is large and $\langle \boldsymbol{z}_i, \boldsymbol{z}_j\rangle$ is small for $i \neq j$. Such contrastive loss, and many of its variants, has been widely used in popular (self-supervised) feature learning methods such as the DINO-type features [CMM+21; ODM+24] for visual data. As we will see, based on the lossy coding framework studied in the previous section, we will be able to arrive at a principled objective function that promotes the two properties and at the same time significantly simplifies the practice such as learning the DINO-type features (as we will feature in great detail in Chapter 8).

The third property is also important because we want the learned features to reveal all possible causes of why one class is different from all other classes. For example, to tell "apple" from "orange", we care not only about color but also shape and the leaves. Ideally, the dimension of each subspace $\{\mathcal{S}_k\}$ should be equal to that of the corresponding submanifold $\mathcal{M}_k$. This property will be important if we would like the map $f(\boldsymbol{x}, \theta)$ to be *invertible* for tasks such as image generation. For example, if we draw different sample points from the feature subspace for "apple", we should be able to decode them, via an inverse map $g(\boldsymbol{z}, \eta)$, to generate diverse images of apples. Such a pair of mutually inverse mappings $f$ and $g$ makes an autoencoding which we will study in great detail in Chapter 6. Notice that feature representations learned from minimizing the cross entropy (4.2.2) normally do not have this property, as they suffer the so-called *neural collapse* phenomenon discussed earlier (see Remark 4.4).

In general, although the intrinsic structures of each class/cluster may be low-dimensional, they are by no means simply linear (or Gaussian) in their original representation $\boldsymbol{x}$ and they need to be made linear first, through some nonlinear transformation.[27] Therefore, overall, we use the nonlinear transformation $f(\boldsymbol{x}, \theta)$ to seek a representation of the data such that the subspaces that represent all the classes are maximally incoherent linear subspaces. To be more precise, we want to learn a mapping $\boldsymbol{z} = f(\boldsymbol{x}, \theta)$ that maps each of the submanifolds $\mathcal{M}_k \subset \mathbb{R}^D$ (Figure 4.13 left) to a *linear* subspace $\mathcal{S}_k \subset \mathbb{R}^d$ (Figure 4.13 right). To some extent, the resulting multiple subspaces $\{\mathcal{S}_k\}$ can be viewed as discriminative *generalized principal components* [VMS16] or, if orthogonal, *independent components* [HO00a] of the resulting features $\boldsymbol{z}$ for the original data $\boldsymbol{x}$. As we will see in the next Chapter 5, deep networks precisely play the role of modeling and realizing this nonlinear transformation from the data distribution to linear discriminative representations.

### 4.2.3 The Principle of Maximal Coding Rate Reduction

Although the three properties—*between-class discriminative*, *within-class compressible*, and *maximally diverse representation*—for linear discriminative repre-sentations (LDRs) are all highly desired properties of the learned representation $\boldsymbol{z}$, they are by no means easy to obtain: Are these properties compatible so that we can expect to achieve them all at once? If so, is there a *simple but principled* objective that can measure the goodness of the resulting representations in terms of all these properties? The key to these questions is to find a principled "measure of compactness" or "information gain" for the distribution of a random variable $\boldsymbol{z}$ or from its finite samples $\{\boldsymbol{z}_i\}_{i=1}^N$. Such a measure should directly and accurately characterize intrinsic geometric or statistical properties of the distribution, in terms of its intrinsic dimension or volume. Unlike the cross entropy (4.2.2) or information bottleneck (4.2.5), such a measure should not depend exclusively on class labels so that it can work in more general settings such as supervised, self-supervised, semi-supervised, and unsupervised settings (as we will see in later sections).

[27] We will discuss how this can be done explicitly in Chapter 6.

Without loss of generality, assume that the distribution $\mathcal{D}$ of the random vector $\boldsymbol{x}$ is supported on a mixture of distributions, i.e., $\mathcal{D} = \cup_{k=1}^K \mathcal{D}_k$, where each $\mathcal{D}_k \subset \mathbb{R}^D$ has a low intrinsic dimension in the high-dimensional ambient space $\mathbb{R}^D$. Let $\boldsymbol{X}_k \in \mathbb{R}^{D \times N_k}$ denote the data matrix whose columns are samples drawn from the distribution $\mathcal{D}_k$, where $N_k$ denotes the number of samples for each $k = 1, \ldots, K$. Then, we use $\boldsymbol{X} = [\boldsymbol{X}_1, \ldots, \boldsymbol{X}_K] \in \mathbb{R}^{D \times N}$ to denote all the samples, where $N = \sum_{k=1}^K N_k$. Recall that we also use $\boldsymbol{x}_i$ to denote the $i$-th sample of $\boldsymbol{X}$, i.e., $\boldsymbol{X} = [\boldsymbol{x}_1, \ldots, \boldsymbol{x}_N]$. Under an encoding mapping:

$$\boldsymbol{x} \xrightarrow{\ f(\boldsymbol{x})\ } \boldsymbol{z}, \tag{4.2.9}$$

the input samples are mapped to $\boldsymbol{z}_i = f(\boldsymbol{x}_i)$ for each $i = 1, \ldots, N$. With an abuse of notation, we also write $\boldsymbol{Z}_k = f(\boldsymbol{X}_k)$ and $\boldsymbol{Z} = f(\boldsymbol{X})$. Therefore, we have $\boldsymbol{Z} = [\boldsymbol{Z}_1, \ldots, \boldsymbol{Z}_K]$ and $\boldsymbol{Z} = [\boldsymbol{z}_1, \ldots \boldsymbol{z}_N]$.

On one hand, for learned features to be discriminative, features of different classes/clusters are preferred to be *maximally incoherent* to each other. Hence, they together should span a space of the largest possible volume (or dimension) and the coding rate of the whole set $\boldsymbol{Z}$ should be as large as possible. On the other hand, learned features of the same class/cluster should be highly correlated and coherent. Hence, each class/cluster should only span a space (or subspace) of a very small volume and the coding rate should be as small as possible. Now, we will introduce how to measure the coding rate of the learned features.

**Coding rate of features.** Notably, a practical challenge in evaluating the coding rate is that the underlying distribution of the feature representations $\boldsymbol{Z}$ is typically unknown. To address this, we may approximate the features $\boldsymbol{Z} = [\boldsymbol{z}_1, \ldots, \boldsymbol{z}_N]$ as samples drawn from a multivariate Gaussian distribution. Under this assumption, as discussed in Section 4.1.3, the compactness of the features $\boldsymbol{Z}$ *as a whole* can be measured in terms of the average coding length per sample, referred to as the *coding rate*, subject to a precision level $\epsilon > 0$ (see (4.1.30)) defined as follows:

$$R_\epsilon(\boldsymbol{Z}) = \frac{1}{2} \operatorname{logdet}\left(\boldsymbol{I} + \frac{d}{N\epsilon^2} \boldsymbol{Z}\boldsymbol{Z}^\top\right). \tag{4.2.10}$$

On the other hand, we hope that a nonlinear transformation $f(\boldsymbol{x})$ maps each class-specific submanifold $\mathcal{M}_k \subset \mathbb{R}^D$ to a maximally incoherent linear subspace $\mathcal{S}_k \subset \mathbb{R}^d$ such that the learned features $\boldsymbol{Z}$ lie in a union of low-dimensional subspaces. This structure allows for a more accurate evaluation of the coding rate by analyzing each subspace separately. Recall that the columns of $\boldsymbol{Z}_k$ denotes the features of the samples in $\boldsymbol{X}_k$ for each $k = 1, \dots, K$. The coding rate for the features in $\boldsymbol{Z}_k$ can be computed as follows:

$$R_\epsilon(\boldsymbol{Z}_k) = \frac{N_k}{2N} \log\det\left(\boldsymbol{I} + \frac{d}{N_k \epsilon^2} \boldsymbol{Z}_k \boldsymbol{Z}_k^\top\right) \tag{4.2.11}$$

Then, the sum of the average coding rates of features in each class is

$$R_\epsilon^c(\boldsymbol{Z}) \doteq \sum_{k=1}^{K} R_\epsilon(\boldsymbol{Z}_k), \tag{4.2.12}$$

Therefore, a good representation $\boldsymbol{Z}$ of $\boldsymbol{X}$ is the one that achieves a large difference between the coding rate for the whole and that for all the classes:

$$\Delta R_\epsilon(\boldsymbol{Z}) \doteq R_\epsilon(\boldsymbol{Z}) - R_\epsilon^c(\boldsymbol{Z}). \tag{4.2.13}$$

Notice that, as per our discussions earlier in this chapter, this difference can be interpreted as the amount of "information gained" by identifying the correct low-dimensional clusters $\boldsymbol{Z}_k$ within the overall set $\boldsymbol{Z}$.

If we choose our feature mapping $f(\cdot)$ to be a deep neural network $f(\cdot, \theta)$ with network parameters $\theta$, the overall process of the feature representation and the resulting rate reduction can be illustrated by the following diagram:

$$\boldsymbol{X} \xrightarrow{f(\boldsymbol{x},\theta)} \boldsymbol{Z} \xrightarrow{\epsilon} \Delta R_\epsilon(\boldsymbol{Z}). \tag{4.2.14}$$

Note that $\Delta R_\epsilon$ is *monotonic* in the scale of the features $\boldsymbol{Z}$. To ensure fair comparison across different representations, it is essential to *normalize the scale* of the learned features. This can be achieved by either imposing the Frobenius norm of each class $\boldsymbol{Z}_k$ to scale with the number of features in $\boldsymbol{Z}_k \in \mathbb{R}^{d \times N_k}$, i.e., $\|\boldsymbol{Z}_k\|_F^2 = N_k$, or by normalizing each feature to be on the unit sphere, i.e., $\boldsymbol{z}_i \in \mathbb{S}^{d-1}$, where $N_k$ denotes the number of samples in the $k$-th class. This formulation offers a natural justification for the need for "batch normalization" in the practice of training deep neural networks [IS15].

Once the representations are comparable, the goal becomes to learn a set of features $\boldsymbol{Z} = f(\boldsymbol{X}, \theta)$ such that they maximize the reduction between the coding rate of all features and that of the sum of features w.r.t. their classes:

$$\begin{aligned} &\max_\theta \; \Delta R_\epsilon\big(\boldsymbol{Z}\big) \doteq R_\epsilon(\boldsymbol{Z}) - R_\epsilon^c(\boldsymbol{Z}), \\ &\text{s.t.} \quad \boldsymbol{Z} = f(\boldsymbol{X}, \theta), \; \|\boldsymbol{Z}_k\|_F^2 = N_k, \; k = 1, \dots, K. \end{aligned} \tag{4.2.15}$$

We refer to this as the principle of *maximal coding rate reduction* (MCR$^2$), as an embodiment of Aristotle's famous quote:

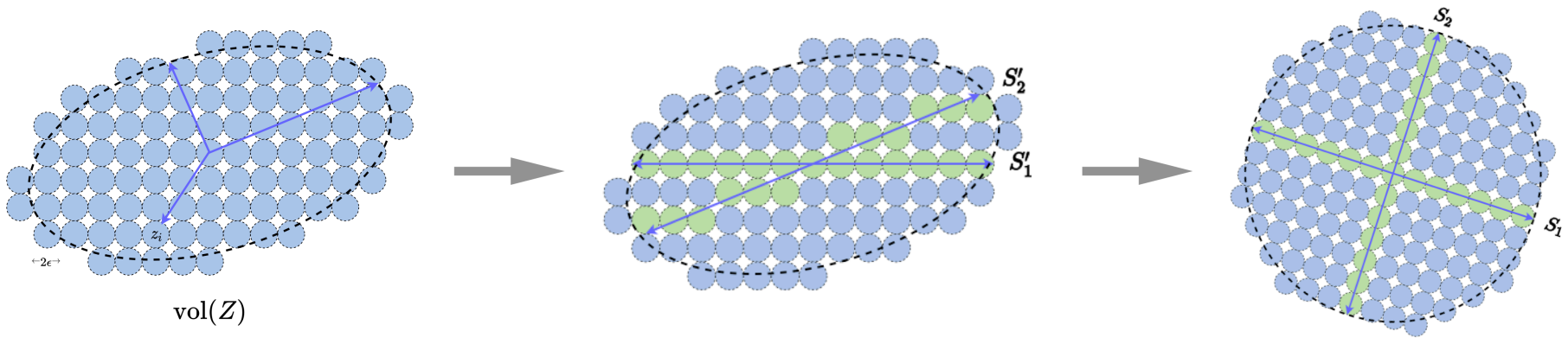

Figure 4.15: After identifying the low-dimensional data distribution, we would like to further transform the data distribution to a more informative structured representation: $R$ is the number of $\epsilon$-balls covering the whole space and $R^c$ is the sum of the numbers for all the subspaces (the green balls). $\Delta R$ is their difference (the number of blue balls).

"*The whole is greater than the sum of its parts.*"

To learn the best representation, we require that *the whole is maximally greater than the sum of its parts*. To a large extent, the quantity $\Delta R$ can be viewed as a computable realization to the mutual information term $I(\boldsymbol{z};\boldsymbol{y})$ introduced in the objective (4.2.7).

*Example* 4.11*.* To further elucidate the ideas, let us examine the example shown in Figure 4.15. From a compression perspective, the representation on the right is *the most compact one* in the sense that the difference between the coding rate when all features are encoded as a single Gaussian (blue) and that when the features are properly clustered and encoded as two separate subspaces (green) is maximal.[28] ■

Note that the above MCR$^2$ principle is designed for supervised learning problems, where the group memberships (or class labels) are known. There are a variety of ways to use the principle, or suitable generalizations, in ways which circumvent this requirement for class labels; we will discuss these later in this Chapter and in the next Chapter 5. To get a first taste, we extend the principle to unsupervised learning problems by introducing a membership matrix, which encodes the (potentially soft) assignment of each data point to latent groups or clusters. Specifically, let $\boldsymbol{\Pi} = \{\boldsymbol{\Pi}_k\}_{k=1}^K \subset \mathbb{R}^{N\times N}$ be a set of diagonal matrices whose diagonal entries encode the membership of the $N$ samples into $K$ classes. That is, $\boldsymbol{\Pi}$ lies in a simplex $\Omega \doteq \{\boldsymbol{\Pi} : \boldsymbol{\Pi}_k \geq \boldsymbol{0} : \sum_{k=1}^K \boldsymbol{\Pi}_k = \boldsymbol{I}_N\}$. Then, we can define the average coding rate with respect to the partition $\boldsymbol{\Pi}$ as

$$R_\epsilon^c(\boldsymbol{Z} \mid \boldsymbol{\Pi}) \doteq \sum_{k=1}^K \frac{\operatorname{tr}(\boldsymbol{\Pi}_k)}{2N} \log\det\left(\boldsymbol{I} + \frac{d}{\operatorname{tr}(\boldsymbol{\Pi}_k)\epsilon^2}\boldsymbol{Z}\boldsymbol{\Pi}_k\boldsymbol{Z}^\top\right). \tag{4.2.16}$$

When $\boldsymbol{Z}$ is given, $R_\epsilon^c(\boldsymbol{Z}|\boldsymbol{\Pi})$ is a concave function of $\boldsymbol{\Pi}$. Then the MCR$^2$ principle for unsupervised learning problems becomes as follows:

$$\max_{\boldsymbol{\Pi},\theta} \Delta R_\epsilon\big(\boldsymbol{Z} \mid \boldsymbol{\Pi}\big) \doteq R_\epsilon(\boldsymbol{Z}) - R_\epsilon^c(\boldsymbol{Z} \mid \boldsymbol{\Pi})$$

[28] Intuitively, the ratio between the "volume" of the whole space spanned by all features and that actually occupied by the features is maximal.

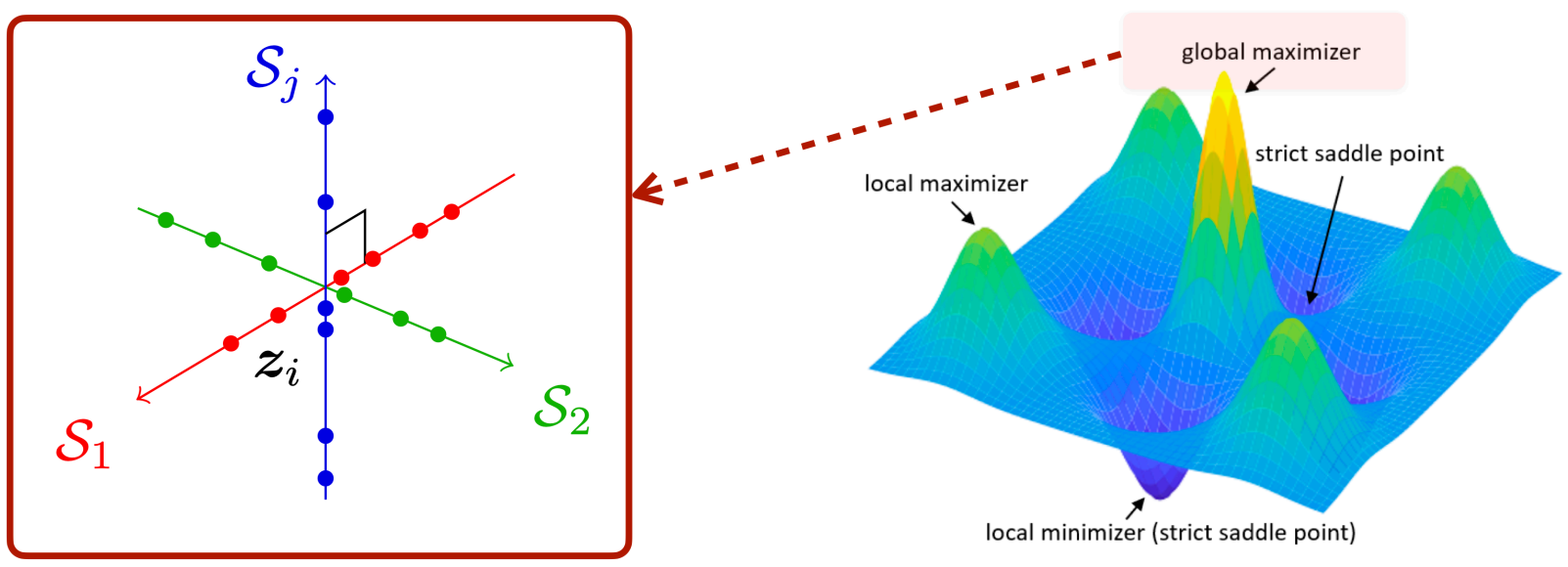

Figure 4.16: **Local optimization landscape:** According to Theorem 4.2, the global maximum of the rate reduction objective corresponds to a solution with mutually incoherent subspaces.

$$\text{s.t.} \quad \boldsymbol{Z} = f(\boldsymbol{X}, \theta), \ \|\boldsymbol{Z}\boldsymbol{\Pi}_k\|_F^2 = N_k, \ k = 1, \ldots, K, \ \boldsymbol{\Pi} \in \Omega. \tag{4.2.17}$$

Compared to (4.2.15), the formulation here allows for the joint optimization of both the group memberships and the network parameters. In particular, when $\boldsymbol{\Pi}$ is fixed to a group membership matrix that assigns $N$ data points into $K$ groups, Problem (4.2.17) can recover Problem (4.2.15).

*Remark* 4.7 (Computing Coding Rates). It is a classical fact that computing the log-determinant of an $n \times n$ matrix takes $\mathcal{O}(n^3)$ time to compute. Hence computing the coding rates $\Delta R_\epsilon$ or $R_\epsilon$ (etc.) seem intractable at the scale of modern data analysis. There are several ways around this, which we will use without comment in the experiments that power the rest of the book.

- *Use the structure of the coding rate.* Suppose that $\boldsymbol{Z}$ is an element of $\mathbb{R}^{d\times n}$. Then it holds

$$\operatorname{logdet}(\boldsymbol{I} + \alpha \boldsymbol{Z}\boldsymbol{Z}^\top) = \operatorname{logdet}(\boldsymbol{I} + \alpha \boldsymbol{Z}^\top \boldsymbol{Z}) \tag{4.2.18}$$

  so computing the matrix product and computing the coding rates requires time on the order of $\min\{d^2 n + d^3, dn^2 + n^3\}$.

- *Use the symmetry of the Gramian.* Note that $\boldsymbol{Z}\boldsymbol{Z}^\top$ (and similarly $\boldsymbol{Z}^\top\boldsymbol{Z}$) is a symmetric matrix. Hence it is relatively cheap (though still cubic time) to compute its eigenvalues $\lambda_i(\boldsymbol{Z}\boldsymbol{Z}^\top)$. Then the log-determinant can be expressed as follows:

$$\operatorname{logdet}(\boldsymbol{I} + \alpha \boldsymbol{Z}\boldsymbol{Z}^\top) = \sum_{i=1}^{d} \log(1 + \alpha \lambda_i(\boldsymbol{Z}\boldsymbol{Z}^\top)), \tag{4.2.19}$$

  (and similarly for $\boldsymbol{Z}^\top\boldsymbol{Z}$). However, this method is made redundant by the next method, which uses more structure of the input and performs better.

- *Use the positive definiteness of the input to the log determinant.* Since $\boldsymbol{Z}\boldsymbol{Z}^\top$ (and similarly $\boldsymbol{Z}^\top\boldsymbol{Z}$) is symmetric positive semidefinite, the quantity $\boldsymbol{I} + \alpha \boldsymbol{Z}\boldsymbol{Z}^\top$ is symmetric positive definite. This means we can compute the Cholesky decomposition $\boldsymbol{L}\boldsymbol{L}^\top = \boldsymbol{I} + \alpha \boldsymbol{Z}\boldsymbol{Z}^\top$, where $\boldsymbol{L} \in \mathbb{R}^{d \times d}$ is a lower-triangular matrix with positive diagonal entries. Then

$$\operatorname{logdet}(\boldsymbol{I} + \alpha \boldsymbol{Z}\boldsymbol{Z}^\top) = 2\sum_{i=1}^{d} \log(L_{ii}). \tag{4.2.20}$$

This method tends to be the cheapest in practice, especially when also strategically deciding whether to compute $\boldsymbol{Z}\boldsymbol{Z}^\top$ or $\boldsymbol{Z}^\top\boldsymbol{Z}$, though it remains at cubic time complexity.

### 4.2.4 Optimization Properties of Coding Rate Reduction

In this subsection, we study the optimization properties of the MCR$^2$ function by analyzing its optimal solutions and the structure of its optimization landscape. To get around the technical difficulty introduced by the neural networks, we consider a simplified version of Problem (4.2.15) as follows:

$$\max_{\boldsymbol{Z}} \; R_\epsilon(\boldsymbol{Z}) - R_\epsilon^c(\boldsymbol{Z}) \qquad \text{s.t.} \quad \|\boldsymbol{Z}_k\|_F^2 = N_k, \; k = 1, \ldots, K. \tag{4.2.21}$$

In theory, the MCR$^2$ principle (4.2.21) benefits from great generalizability and can be applied to representations $\boldsymbol{Z}$ of *any* distributions as long as the rates $R_\epsilon$ and $R_\epsilon^c$ for the distributions can be accurately and efficiently evaluated. The optimal representation $\boldsymbol{Z}^*$ should have some interesting geometric and statistical properties. We here reveal nice properties of the optimal representation with the special case of subspaces, which have many important use cases in machine learning. When the desired representation for $\boldsymbol{Z}$ is multiple subspaces, the rates $R_\epsilon$ and $R_\epsilon^c$ in (4.2.21) are given by (4.2.10) and (4.2.12), respectively. At the maximal rate reduction, MCR$^2$ achieves its optimal representations, denoted as $\boldsymbol{Z}^* = [\boldsymbol{Z}_1^*, \ldots, \boldsymbol{Z}_K^*]$ with $\operatorname{rank}(\boldsymbol{Z}_k^*) \le d_k$. One can show that $\boldsymbol{Z}^*$ has the following desired properties (see [YCY+20] for a formal statement and detailed proofs).

**Theorem 4.2 (Characterization of Global Optimal Solutions).** *Suppose $\boldsymbol{Z}^* = [\boldsymbol{Z}_1^*, \ldots, \boldsymbol{Z}_K^*]$ is a global optimal solution of Problem (4.2.21). The following statements hold:*

- Between-Class Discriminative*: As long as the ambient space is adequately large ($d \ge \sum_{k=1}^K d_k$), the subspaces are all orthogonal to each other,* i.e., *$(\boldsymbol{Z}_k^*)^\top \boldsymbol{Z}_l^* = \boldsymbol{0}$ for $k \ne l$.*
- Maximally Diverse Representation*: As long as the coding precision is adequately high, i.e., $\epsilon^4 < c \cdot \min_k \left\{\frac{N_k}{N}\frac{d^2}{d_k^2}\right\}$, where $c > 0$ is a constant. Each subspace achieves its maximal dimension, i.e.* $\operatorname{rank}(\boldsymbol{Z}_k^*) = d_k$*. In addition, the largest $d_k - 1$ singular values of $\boldsymbol{Z}_k^*$ are equal.*

This theorem indicates that the MCR$^2$ principle promotes embedding of data into multiple independent subspaces (as illustrated in Figure 4.16), with features distributed *isotropically* in each subspace (except for possibly one dimension). Notably, this theorem also confirms that the features learned by the MCR$^2$ principle exhibit the desired low-dimensional discriminative properties discussed in Section 4.2.2. In addition, among all such discriminative representations, it prefers the one with the highest dimensions in the ambient space. This is substantially different from the objective of information bottleneck (4.2.5).

**Regularized MCR$^2$.** The above theorem characterizes properties of the global optima of the rate reduction objectives. What about other optima, such as local ones? Due to the constraints of the Frobenius norm, it is a difficult task to analyze Problem (4.2.21) from an optimization-theoretic perspective. Therefore, we consider the Lagrangian formulation of (4.2.21). This can be viewed as a tight relaxation or even an equivalent problem of (4.2.21) whose optimal solutions agree under specific settings of the regularization parameter; see [WLP+24, Proposition 3.2]. Specifically, the formulation we study, referred to henceforth as the *regularized MCR$^2$ problem*, is as follows:

$$\max_{\boldsymbol{Z}} \; R_\epsilon(\boldsymbol{Z}) - R_\epsilon^c(\boldsymbol{Z}) - \frac{\lambda}{2}\|\boldsymbol{Z}\|_F^2, \tag{4.2.22}$$

where $\lambda > 0$ is the regularization parameter. Although the program (4.2.22) is highly nonconcave and involves matrix inverses in its gradient computation, we can still explicitly characterize its local and global optima as follows (see [WLP+24, Theorem 3.1] for detailed proofs).

**Theorem 4.3** (**Local and Global Optima**). *Let $N_k$ denote the number of training samples in the $k$-th class for each $k \in \{1, \dots, K\}$, $N_{\max} \doteq \max\{N_1, \dots, N_K\}$, $\alpha = d/(N\epsilon^2)$, and $\alpha_k = d/(N_k\epsilon^2)$ for each $k \in \{1, \dots, K\}$. Given a coding precision $\epsilon > 0$, if the regularization parameter satisfies*

$$\lambda \in \left(0, \frac{d(\sqrt{N/N_{\max}} - 1)}{N(\sqrt{N/N_{\max}} + 1)\epsilon^2}\right], \tag{4.2.23}$$

*then the following statements hold:*
*(i) (**Local maximizers**) $\boldsymbol{Z}^* = [\boldsymbol{Z}_1^*, \dots, \boldsymbol{Z}_K^*]$ is a local maximizer of Problem (4.2.22) if and only if the $k$-th block admits the following decomposition*

$$\boldsymbol{Z}_k^* = \left(\frac{\eta_k + \sqrt{\eta_k^2 - 4\lambda^2 N/N_k}}{2\lambda\alpha_k}\right)^{1/2} \boldsymbol{U}_k \boldsymbol{V}_k^\top, \tag{4.2.24}$$

*where (a) $r_k = \operatorname{rank}(\boldsymbol{Z}_k^*)$ satisfies $r_k \in [0, \min\{N_k, d\})$ and $\sum_{k=1}^K r_k \leq \min\{N, d\}$, (b) $\boldsymbol{U}_k \in \mathcal{O}^{d \times r_k}$ satisfies $\boldsymbol{U}_k^\top \boldsymbol{U}_l = \boldsymbol{0}$ for all $k \neq l$, $\boldsymbol{V}_k \in \mathcal{O}^{N_k \times r_k}$, and (c) $\eta_k := (\alpha_k - \alpha) - \lambda(N/N_k + 1)$ for each $k \in \{1, \dots, K\}$.*
*(ii) (**Global maximizers**) $\boldsymbol{Z}^* = [\boldsymbol{Z}_1^*, \dots, \boldsymbol{Z}_K^*]$ is a global maximizer of Problem*

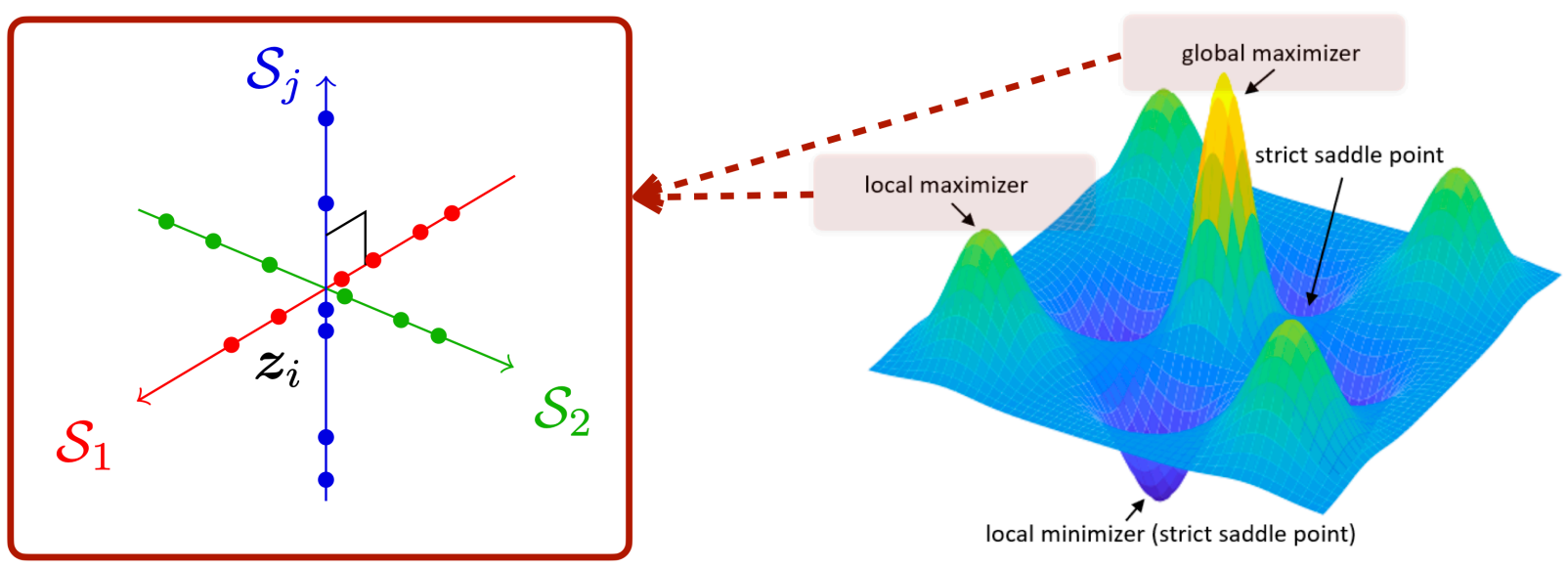

Figure 4.17: **Global optimization landscape:** According to [LSJ+16; SQW15], Theorems 4.3 and 4.4, both global and local maxima of the (regularized) rate reduction objective correspond to a solution with mutually incoherent subspaces. All other critical points are strict saddle points.

(4.2.22) *if and only if (a) it satisfies all the above conditions and* $\sum_{k=1}^{K} r_k = \min\{N, d\}$*, and (b) for all* $k \neq l \in [K]$ *satisfying* $N_k < N_l$ *and* $r_l > 0$*, we have* $r_k = \min\{N_k, d\}$*.*

This theorem explicitly characterizes the local and global optima of problem (4.2.22). Intuitively, this shows that the features represented by each local maximizer of Problem (4.2.22) are low-dimensional and discriminative. Although we have characterized the local and global optimal solutions in Theorem 4.3, it remains unknown whether these solutions can be efficiently computed using GD to solve the problem (4.2.22), since GD may get stuck at other critical points such as a saddle point. Fortunately, [LSJ+16; SQW15] showed that if a function is twice continuously differentiable and satisfies the *strict saddle property*, i.e., each critical point is either a local maximizer or a strict saddle point[29], GD converges to a local maximizer almost surely with random initialization. We investigate the global optimization landscape of the problem (4.2.22) by characterizing all its critical points as follows (see [WLP+24, Theorem 3.3] for detailed proofs).

**Theorem 4.4** (**Benign Global Optimization Landscape**)**.** *Given a coding precision* $\epsilon > 0$*, if the regularization parameter satisfies* (4.2.23)*, it holds that any critical point* $\boldsymbol{Z}$ *of the problem* (4.2.22) *is either a local maximizer or a strict saddle point.*

Together, the above two theorems show that the learned features associated with each local maximizer of the rate reduction objective — not just global maximizers — are structured as incoherent low-dimensional subspaces. Furthermore, the (regularized) rate reduction objective (4.2.15) has a very benign landscape with only local maxima and strict saddles as critical points, as illustrated in

[29]We say that a critical point is a strict saddle point of Problem (4.2.22) if it has a direction with strictly positive curvature [SQW15]. This includes classical saddle points with strictly positive curvature as well as strict local minimizers.

Figure 4.17. According to [LSJ+16; SQW15], Theorems 4.3 and 4.4 imply that low-dimensional and discriminative representations (LDRs) can be efficiently found by applying (stochastic) gradient descent to the rate reduction objective (4.2.15) from random initialization. These results also indirectly explain why in Section 4.3.1, if the chosen network is expressive enough and trained well, the resulting representation typically gives an incoherent linear representation that likely corresponds to the globally optimal solution. Interested readers are referred to [WLP+24] for proofs.

## 4.3 Learning Representations for Imagery Data

### 4.3.1 Supervised Representation Learning

We here present how the MCR$^2$ objective (4.2.15) helps learn better representations than the cross entropy (CE) (4.2.2) for image classification. Here, for simplicity,[30] we adopt the popular neural network architecture, the ResNet-18 [HZR+16b], to model the feature mapping $\boldsymbol{z} = f(\boldsymbol{x}, \theta)$. We optimize the neural network parameters $\theta$ via the backpropagation algorithm (detailed in Section A.2.3) to maximize the coding rate reduction. More implementation details of this experiment can be found in [CYY+22].

Figure 4.18 illustrates how the two rates and their difference (for both training and test data) evolve over epochs of training: After an initial phase, $R_\epsilon$ gradually increases while $R_\epsilon^c$ decreases, indicating that features $\boldsymbol{Z}$ are expanding as a whole while each class $\boldsymbol{Z}_k$ is being compressed.

We first evaluate the performance with the CIFAR-10 image classification dataset [KH+09] and the results are shown in the table of Figure 4.19. As we can see, the MCR$^2$ objective achieves a comparable performance as the cross entropy (CE) loss. In addition, the MCR$^2$ objective is more robust to corruptions in the class labels since it learns a joint (subspace) representation for each class instead of trying to fit the labels of individual samples.

Figure 4.20 shows the distribution of singular values of the learned features $\boldsymbol{Z}_k$ per class. From the distribution of the singular values shown in Figure 4.20a, we see that the features $\boldsymbol{z}$ learned from MCR$^2$ for each class are distributed on a subspace of dimension around 10-12, whereas the features from cross entropy Figure 4.20b are of much lower dimension.[31]

Figure 4.21 shows the cosine similarities between the learned features sorted by class. We also compare the similarities of the learned features by using the cross-entropy (CE) (4.2.2) and the MCR$^2$ objective (4.2.15). From the plots, one can clearly see that the features learned by the MCR$^2$ objective for different classes are much more incoherent (or independent) than features learned by using cross-entropy loss. Within each class, the features also are less coherent hence more diverse, as indicated by the distribution of their singular values shown in Figure 4.20a.

[30] We will rigorously study and justify how design or derive a proper deep architecture from and for the MCR$^2$ objective in Chapter 5.

[31] If one continue to train with more iterations, the CE features will may even collapse to a single dimension, known as the neural collapse phenomenon discussed earlier in Remark 4.4.

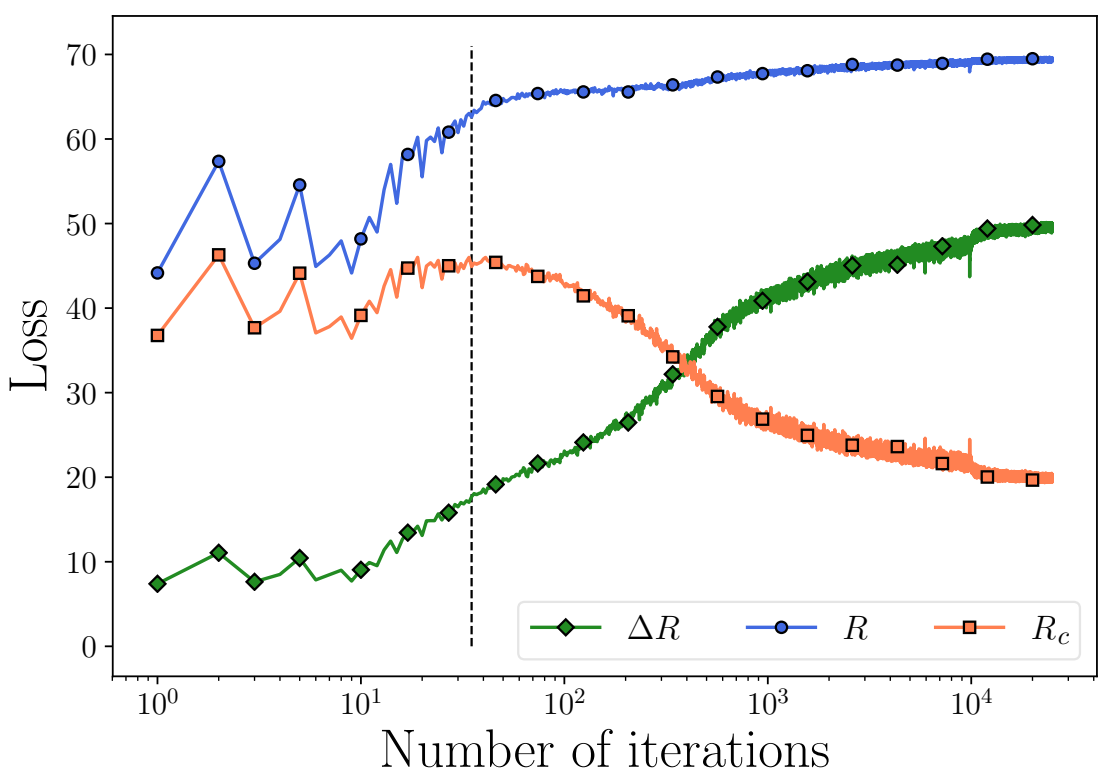

Figure 4.18: Evolution of $R_\epsilon$, $R_\epsilon^c$, $\Delta R_\epsilon$ during the training process.

| | Ratio=0.0 | Ratio=0.1 | Ratio=0.2 | Ratio=0.3 | Ratio=0.4 | Ratio=0.5 |
|---|---|---|---|---|---|---|
| CE Training | 0.939 | 0.909 | 0.861 | 0.791 | 0.724 | 0.603 |
| MCR$^2$ Training | **0.940** | **0.911** | **0.897** | **0.881** | **0.866** | **0.843** |

Figure 4.19: Comparison of classification accuracy between training with the cross entropy (CE) loss or the MCR$^2$ objective, with different ratios of the class labels being randomly corrupted.

We visualize the images that are mapped along different singular vectors in the feature subspace of each class, which we call the "principal" images. Figure 4.22 shows these principal images along each of the top-10 singular vectors. As we can see, the images along each singular vector are very similar in their appearances and structures and images along different singular vectors are very distinct from each other.

Despite the fact that the MCR$^2$ objective seems to allow us to learn better representations, there has been an apparent lack of justification of the network architectures used in the above experiments. It is yet unclear why the network adopted here (the ResNet-18) is suitable for representing the map $f(\boldsymbol{x}, \theta)$, let alone for interpreting the layer operators and parameters $\theta$ learned inside. In Chapter 5, *we will show how to derive network architectures and components entirely as a "white box" from the desired objective (say the rate reduction).*

### 4.3.2 Unsupervised Representation Learning

As we have seen in the above section, the class information makes learning a structured representation of the (mixed) data distribution relatively easy. However, such information is often not available, at least not at a massive scale.

Now we present how to extend the MCR$^2$ principle to the unsupervised set-

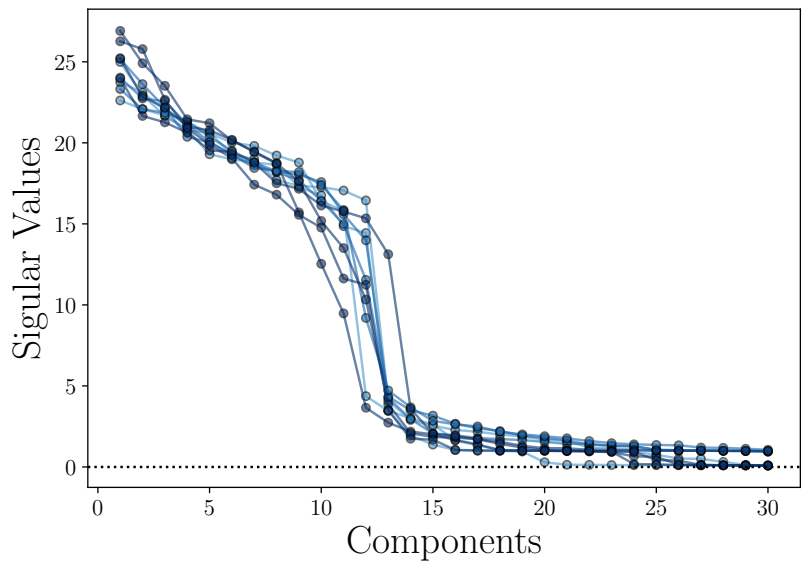

(a) PCAs of features for individual classes learned by MCR$^2$.

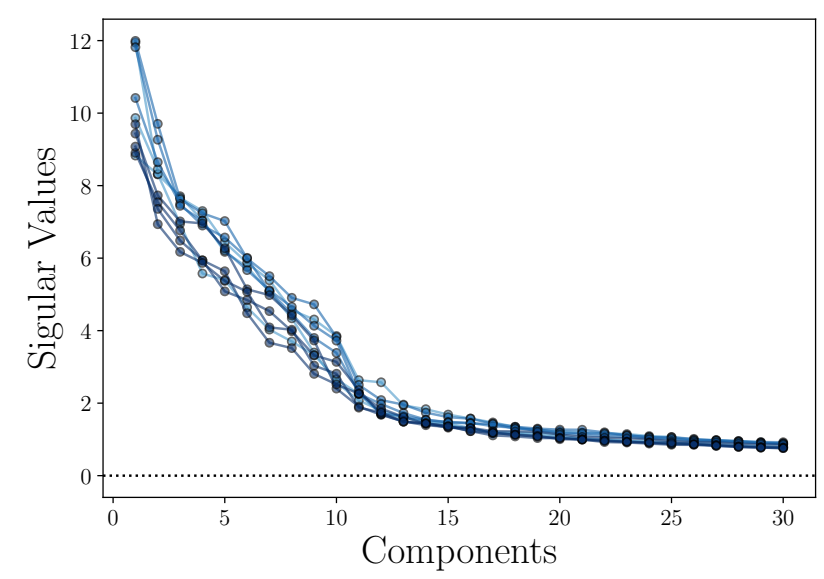

(b) PCAs of features for individual classes learned by Cross Entropy.

Figure 4.20: Comparison of the principal components of learned features from MCR$^2$ versus those from cross entropy.

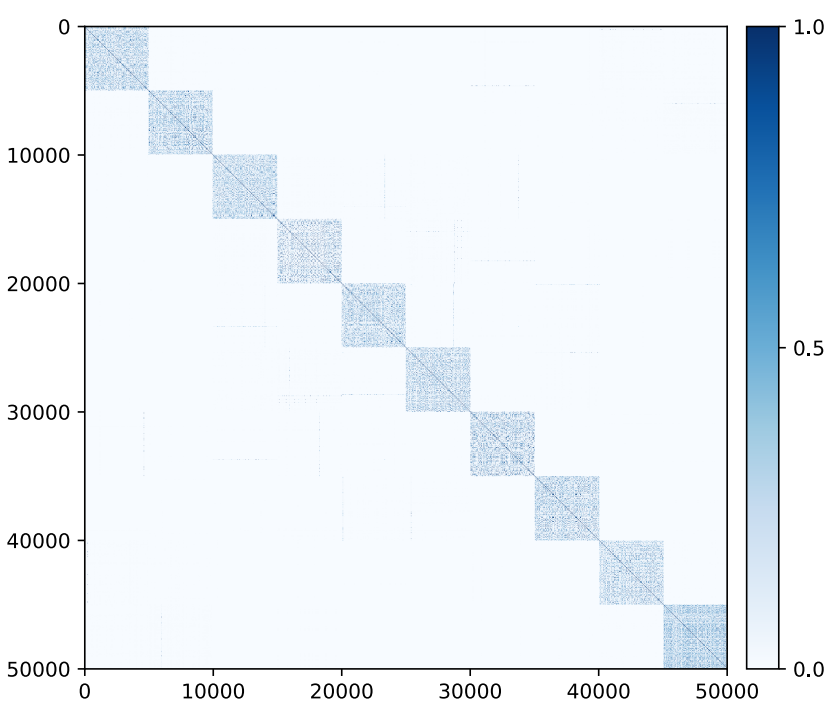

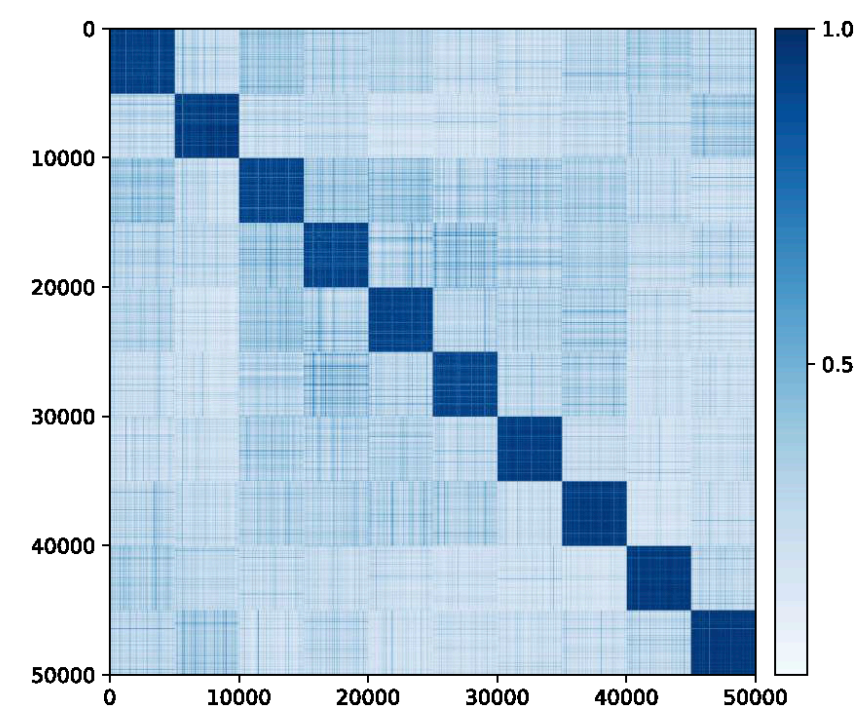

Figure 4.21: Cosine similarity between learned features by using the MCR$^2$ objective (**left**) and CE loss (**right**).

ting. Contrary to supervised learning where the class labels are known, in unsupervised learning, the group memberships of the data samples are unknown. In this case, we can apply certain class of augmentations or transformations, say $\tau$ with a distribution $P_\tau$, to each sample. For example, for images, augmentations typically include translations, rotations, or cropping of each image sample. We may view the augmented samples of each data point as belonging to the same group or cluster.

More specifically, given a set of $N$ data samples $\boldsymbol{X} = [\boldsymbol{x}_1, \dots, \boldsymbol{x}_N]$, we generate $m$ augmented views $\{\tau_i\}_{i=1}^m$ for each data point $\boldsymbol{x}_k$ , denoted as $\tilde{\boldsymbol{X}}_k = [\tau_1(\boldsymbol{x}_k), \tau_2(\boldsymbol{x}_k), \dots, \tau_m(\boldsymbol{x}_k)]$. We treat the augmented samples $\tilde{\boldsymbol{X}}_i$ as the $i$-th class with $K = N$ groups/clusters in total. Now, we can apply the MCR$^2$ principle as we did before in the supervised learning case. Specifically, we learn the

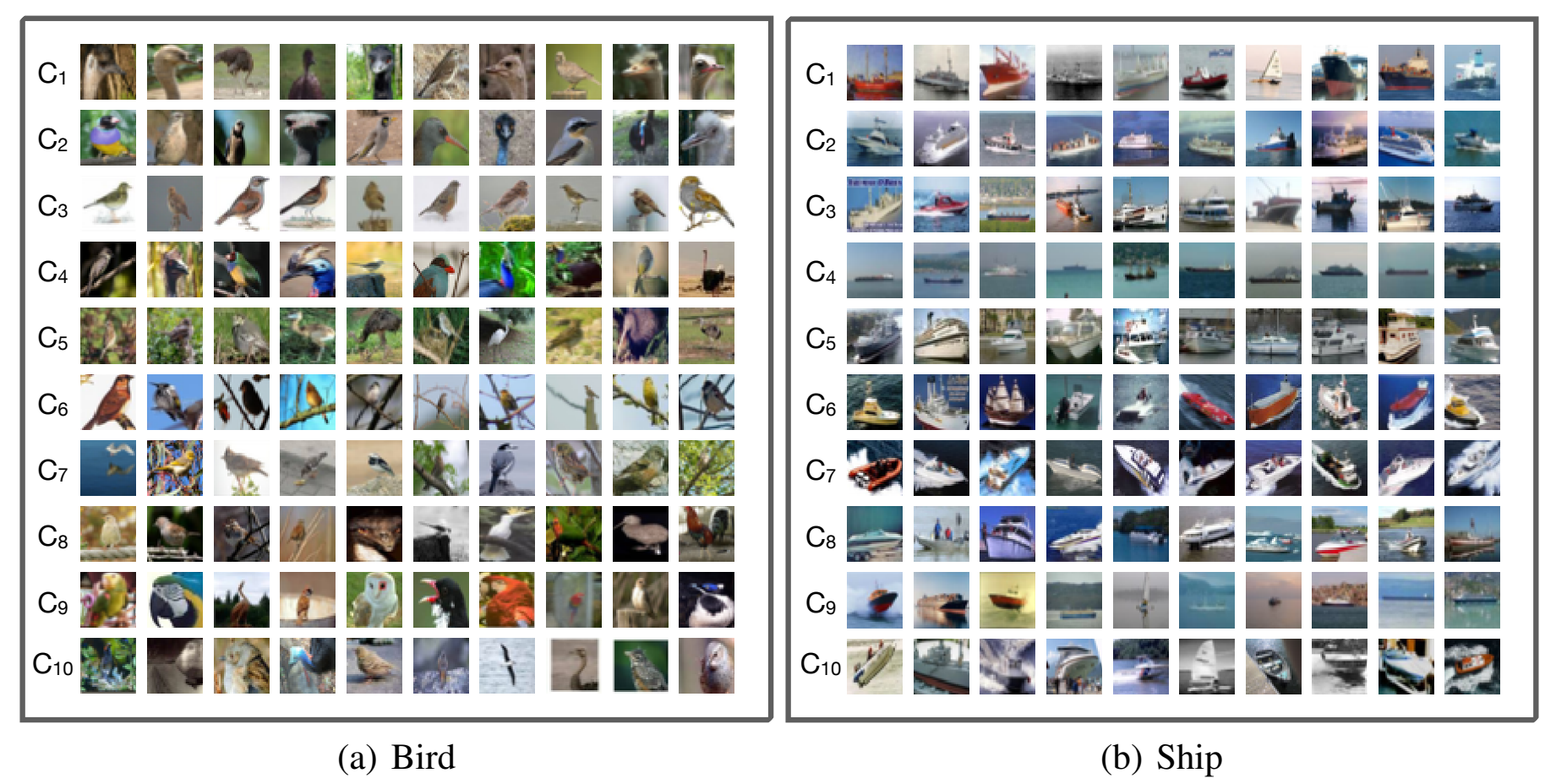

Figure 4.22: Top-10 "principal" images for class - "Bird" and "Ship" in the CIFAR-10 image dataset.

feature mapping $f(\boldsymbol{x}, \theta)$ by maximizing the coding rate reduction as follows:

$$\begin{aligned} \max_{\theta}\ & \Delta R_\epsilon\big(\boldsymbol{Z}\big) \doteq R_\epsilon(\boldsymbol{Z}) - R^c_\epsilon(\boldsymbol{Z}), \\ \text{s.t.}\quad & \boldsymbol{Z} = f(\tilde{\boldsymbol{X}}, \theta),\ \|\boldsymbol{Z}_k\|_F^2 = m,\ k = 1, \dots, N. \end{aligned} \tag{4.3.1}$$

Here, $\tilde{\boldsymbol{X}} = [\tilde{\boldsymbol{X}}_1, \dots, \tilde{\boldsymbol{X}}_N]$ and $\boldsymbol{Z}_k = f(\tilde{\boldsymbol{X}}_k, \theta)$ denotes the features of the augmented samples of $\boldsymbol{x}_k$ for each $k = 1, \dots, N$.

Let us again adopt ResNet-18 as the feature mapping $f(\boldsymbol{x}, \theta)$ and optimize the network parameters $\theta$ to maximize the unsupervised objective (4.3.1). After training on the CIFAR-10 dataset, we visualize the evolution of the two rates and their difference (for both training and test data) over epochs of training in Figure 4.23a. Similar to the supervised learning case, after an initial phase, $R_\epsilon$ gradually increases while $R^c_\epsilon$ decreases, indicating that features $\boldsymbol{Z}$ are expanding as a whole while each cluster $\boldsymbol{Z}_k$ is being compressed. Compared to the supervised case shown in Figure 4.18, we can observe that, in the unsupervised case, the total coding rate $R_\epsilon$ expands rapidly and the overall MCR$^2$ objective saturates at a local optimum. To alleviate this, we can alter the optimization strategy by controlling the dynamics of $R_\epsilon$ and $R^c_\epsilon$. Specifically, we can rescale the rates by replacing the coding rate $R_\epsilon(\boldsymbol{Z})$ in (4.3.1) with

$$\tilde{R}_\epsilon(\boldsymbol{Z}) \doteq \frac{1}{2\gamma_1} \log\det\left(\boldsymbol{I} + \frac{\gamma_2 d}{N m \epsilon^2} \boldsymbol{Z}\boldsymbol{Z}^\top\right).$$

By setting $\gamma_1 = \gamma_2 = N$, we arrive at a modified learning objective, referred to as MCR$^2$-CTRL, whose training dynamics are shown in Figure 4.23b. As we can see, with this controlled MCR$^2$ objective, features are first compressed and then gradually expanded, leading to a higher value of the overall objective $\Delta R_\epsilon(\boldsymbol{Z})$.

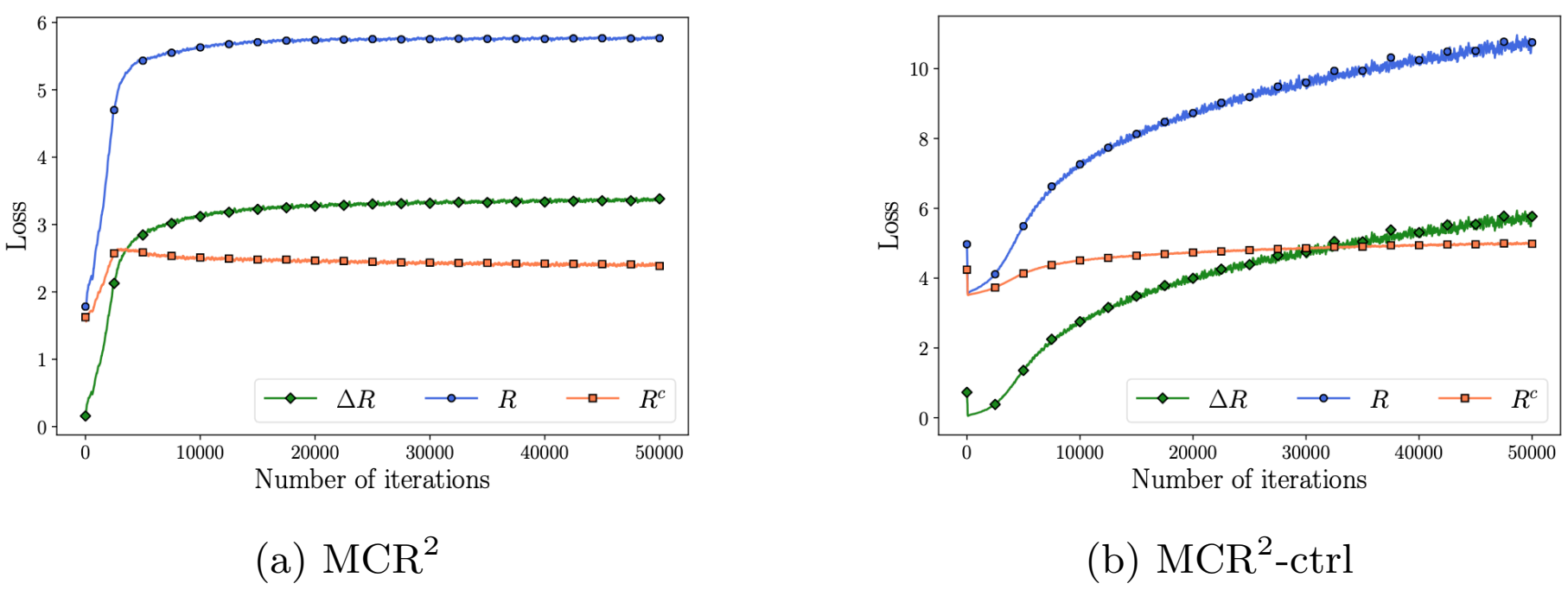

(a) MCR$^2$ (b) MCR$^2$-ctrl

Figure 4.23: Evolution of the rates of MCR$^2$ in the training process for unsupervised learning on CIFAR-10.

Table 2: Clustering results on CIFAR10, CIFAR100, and STL10 datasets.

| Dataset | Metric | K-Means | JULE | RTM | DEC | DAC | DCCM | MCR$^2$-Ctrl |
|---|---|---|---|---|---|---|---|---|
| CIFAR10 | NMI | 0.087 | 0.192 | 0.197 | 0.257 | 0.395 | 0.496 | **0.630** |
| | ACC | 0.229 | 0.272 | 0.309 | 0.301 | 0.521 | 0.623 | **0.684** |
| | ARI | 0.049 | 0.138 | 0.115 | 0.161 | 0.305 | 0.408 | **0.508** |
| CIFAR100 | NMI | 0.084 | 0.103 | - | 0.136 | 0.185 | 0.285 | **0.387** |
| | ACC | 0.130 | 0.137 | - | 0.185 | 0.237 | 0.327 | **0.375** |
| | ARI | 0.028 | 0.033 | - | 0.050 | 0.087 | 0.173 | **0.178** |
| STL10 | NMI | 0.124 | 0.182 | - | 0.276 | 0.365 | 0.376 | **0.446** |
| | ACC | 0.192 | 0.182 | - | 0.359 | 0.470 | 0.482 | **0.491** |
| | ARI | 0.061 | 0.164 | - | 0.186 | 0.256 | 0.262 | **0.290** |

Figure 4.24: Comparison of clustering performance accuracy between training with the MCR$^2$-ctrl objective and other unsupervised learning algorithms on CIFAR-10, CIFAR-100, and STL-10 datasets.

To assess the quality of the learned representations and verify whether they have good subspace structures, we adopt a subspace clustering algorithm EnSC [YLR+16]. For evaluation metrics, we consider normalized mutual information (NMI), clustering accuracy (ACC), and adjusted rand index (ARI) [32]. For comparison, we consider methods including JULE [YPB16], RTM [NMM19], DEC [XGF16], DAC [CWM+17], and DCCM [WLW+19]. The clustering results are summarized in Figure 4.24. As we can see, the features learned by maximizing the MCR$^2$-ctrl objective achieve better clustering accuracy than those learned by other algorithms.

There is one main drawback though: the MCR$^2$-based unsupervised learning method requires a costly $\log\det$ operation of $\mathsf{O}(d^3)$ complexity *per sample* when computing $R^c_\epsilon$. This quickly becomes computationally prohibitive as we try to scale to larger datasets. Instead of minimizing the exact coding rate for each group of augmented samples $\tilde{\boldsymbol{X}}_i$, we can approximate the effect of compressing features in each group by maximizing the *cosine similarity* between the features of the augmented samples:[33]

[32] Refer to the Appendix in [YCY+20] for their specific definitions.

| Method | Model | Epochs | 20-NN | Linear Probing |
|---|---|---|---|---|
| DINO | ViT-B | 100 | 72.9 | 76.3 |
| SimDINO | ViT-B | 100 | **74.9** | **77.3** |
| DINO | ViT-L | 100 | – | – |
| SimDINO | ViT-L | 100 | **75.6** | **77.4** |

Table 4.1: **Classification performance** on ImageNet-1k test set for DINO and SimDINO, using both $k$-nearest neighbor accuracy ($k = 20$) and linear probing. SimDINO outperforms DINO in both accuracy and training stability. DINO training on ViT-L is unstable and quickly diverges, while SimDINO trains stably.

$$\begin{aligned} &\max_{\theta} R_{\epsilon}(\boldsymbol{Z}) + \frac{\gamma}{N}\sum_{k=1}^{N}\sum_{i\neq j}\boldsymbol{z}_{k,j}^{\top}\boldsymbol{z}_{k,i}, \\ &\text{s.t.}\quad \boldsymbol{z}_{i,j} = f(\tau_j(\boldsymbol{x}_i),\theta),\ \|\boldsymbol{z}_{i,j}\|_2^2 = 1, \end{aligned} \tag{4.3.2}$$

where $\gamma > 0$ is a hyperparameter that controls the strength of the cosine similarity term, as it differs from the coding rate in magnitude. This objective avoids the costly $\log\det$ operation and is much more scalable. In fact, this objective is closely related to popular (unsupervised) *contrastive learning* methods such as SimCLR [CKN+20] and DINO [CTM+21], which have been engineered to learn high-quality representations from massive amount of images.

As we will show later in Section 8.2.2 of Chapter 8, this objective (4.3.2) can be implemented very efficiently to learn a highly structured representation $\boldsymbol{Z}$ that is highly linear and discriminative. The learned feature representation leads to state-of-the-art performance on large real-world datasets comparable or even better than features learned by DINO, but without the heavy engineering efforts needed for DINO. Hence, we refer to this objective as *Simplified DINO* (SimDINO). As a preview, Table 4.1 show that SimDINO outperforms DINO on ImageNet-1k classification using both $k$-nearest neighbor accuracy and linear probing (i.e. training a linear classifier on pretrained models). This demonstrates that the objective leads to highly structured and linearized representations. We can also utilize these representations for downstream tasks such as unsupervised segmentation[34], as illustrated in Figure 4.25. More implementation details and experimental results of SimDINO will be presented in Chapter 8.

[33] We leave as an exercise for the readers to verify under what conditions that minimizing the per class coding rate $R^c(\boldsymbol{Z})$ is (approximately) equivalent to maximizing the cosine similarity among the features $\boldsymbol{Z}$.

[34] Specifically, we adopt MaskCut [WGY+23], an effective unsupervised approach of extracting features from a frozen vision backbone for object detection and instance segmentation.

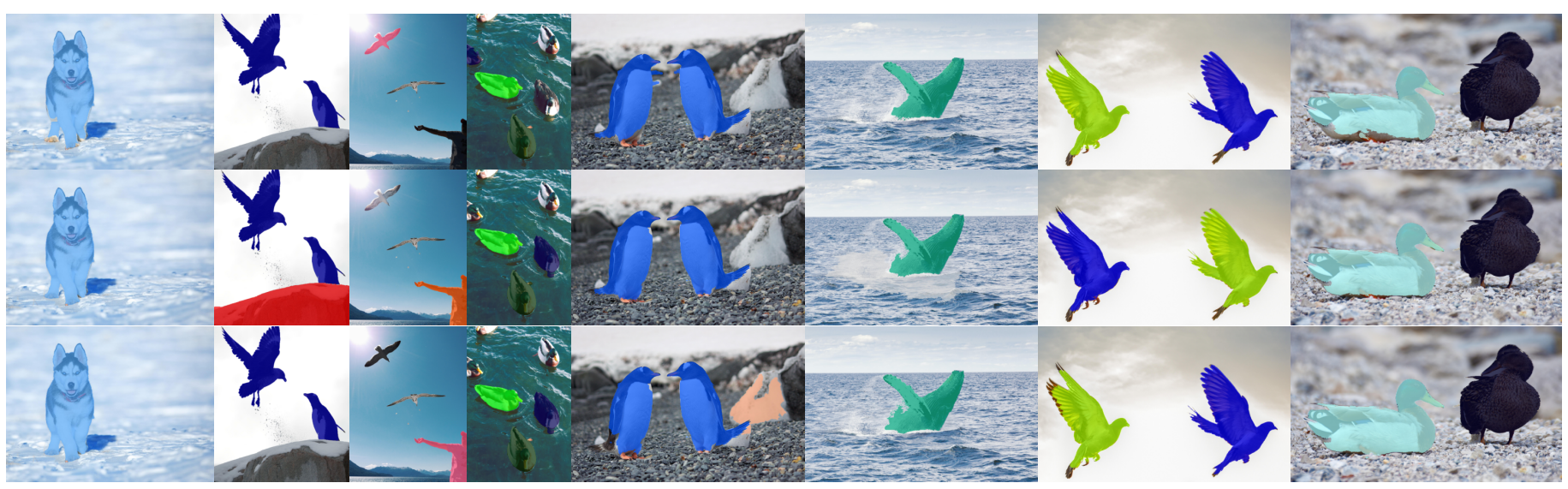

Figure 4.25: **Visualization of unsupervised segmentation results** from DINO ViT-B/16 (row 1), SimDINO ViT-B/16 (row 2) and SimDINO ViT-L/16 (row 3).

## 4.4 Summary and Notes

**Key messages.** In this Chapter, we have studied basic concepts and ideas behind how to learn a distribution from finite samples and obtain a *computable* encoding and decoding scheme for the distribution. For distributions that are continuous and possibly low-dimensional, we have argued that such a scheme is necessarily *lossy* – hence the importance of the concept of rate distortion. The noise introduced through distortion, due to quantization, plays a related but different role from the noise we have studied in the preceding chapter. One may think of the noise considered in Chapter 3 as "outside" of the submanifold of the support of the distribution whereas the noise considered in this chapter is "inside" the support. Informally, we may refer to the diffusion/denoising process outside of the support as "continuation" and the diffusion/denoising process inside the support as "percolation." A more unified analysis and understanding of their compounded effect is due for future study.

In this chapter, we introduced for the first time the concept of learning a *representation* for a distribution. We have considered arguably the simplest case, the case of classification. If one views the classification label $\boldsymbol{y}$ as a fixed simple (grossly lossy) code for the data $\boldsymbol{x}$, we may seek a consistent[35] representation $\boldsymbol{z} = f(\boldsymbol{x})$ of the data $\boldsymbol{x}$ such that it tends to maximize the mutual information between the representation and this simple code:

$$\max I(\boldsymbol{z}, \boldsymbol{y}) = H(\boldsymbol{z}) - H(\boldsymbol{z} \mid \boldsymbol{y}). \tag{4.4.1}$$

In addition, we require such a representation $\boldsymbol{z}$ is not only consistent to $\boldsymbol{x}$ but also much more *structured* than the native data distribution of $\boldsymbol{x}$ – classes are linearly discriminative. This would make the learned representation much more efficient to use than the native data distribution for subsequent tasks such as recognition, segmentation and completion. The maximal coding rate reduction objective introduced is a natural and computable realization of these requirements. In Chapter 7, we will see how to generalize this objective to a much broader context when $\boldsymbol{y}$ could be a more general lossy code (say image caption) or a random variable related to $\boldsymbol{x}$.

[35] That is, we can faithfully recover $\boldsymbol{x}$ from $\boldsymbol{z}$ as we will make the notion of "consistent" representations clear later in Chapter 6.

## 4.5 Exercises and Extensions

*Exercise* 4.1. Please show the following properties of the $\log\det(\cdot)$ function.

1. Show that

$$f(\boldsymbol{X}) = \log\det(\boldsymbol{X})$$

   is a concave function. (**Hint:** The function $f(\boldsymbol{x})$ is concave if and only if the one-dimensional function $f(\boldsymbol{x} + t\boldsymbol{h})$ is concave for all $\boldsymbol{x}$ and $\boldsymbol{h}$.)

2. Show that

$$\log\det(\boldsymbol{I} + \boldsymbol{X}^\top \boldsymbol{X}) = \log\det(\boldsymbol{I} + \boldsymbol{X}\boldsymbol{X}^\top)$$

3. Let $\boldsymbol{A} \in \mathbb{R}^{n\times n}$ be a positive definite matrix. Please show that:

$$\log\det(\boldsymbol{A}) = \sum_{i=1}^{n} \log(\lambda_i), \tag{4.5.1}$$

   where $\lambda_1, \lambda_2, \ldots, \lambda_n$ are the eigenvalues of $\boldsymbol{A}$.

4. Show that the gradient of the $\log\det(\cdot)$ function is given by

$$\nabla \log\det(\boldsymbol{A}) = \boldsymbol{A}^{-1},$$

   for a positive definite matrix $\boldsymbol{A}$.

*Exercise* 4.2. Let $\boldsymbol{Z} = [\boldsymbol{Z}_1, \ldots, \boldsymbol{Z}_K] \in \mathbb{R}^{d\times N}$, where each $\boldsymbol{Z}_k \in \mathbb{R}^{d\times N_k}$ and $N = \sum_{k=1}^{K} N_k$.

1. Please show that

$$\log\det\left(\boldsymbol{I} + \alpha \boldsymbol{Z}\boldsymbol{Z}^\top\right) \leq \sum_{k=1}^{K} \log\det\left(\boldsymbol{I} + \alpha \boldsymbol{Z}_k \boldsymbol{Z}_k^\top\right),$$

   where $\alpha > 0$.

2. Please derive the necessary and sufficient condition under which the equality holds.

*Exercise* 4.3. Let $\boldsymbol{z}_1$ and $\boldsymbol{z}_2$ be two vectors of unit length. Let $\boldsymbol{Z} = [\boldsymbol{z}_1, \boldsymbol{z}_2]$ Show that the following three objectives are equivalent:

- $\max \boldsymbol{z}_1^\top \boldsymbol{z}_2$.
- $\min \|\boldsymbol{z}_1 - \boldsymbol{z}_2\|_2^2$.
- $\max \log\det(\boldsymbol{I} + \boldsymbol{Z}\boldsymbol{Z}^\top)$.

However, they may no longer be exactly equivalent when there are more than two vectors in a high-dimensional space. The $\log\det$ is more general as it measures the total volume spanned by all the vectors in the high-dimensional space.

*Exercise* 4.4. Please derive the rate-distortion function for the Gaussian random vector introduced in Example 4.4.

# Chapter 5

# Deep Representations as Unrolled Optimization

"*What I cannot create, I do not understand.*"
— Richard Feynman

In previous chapters, we have shown how to identify low-dimensional structures in high-dimensional spaces, *mainly focusing on linear structures*. For example, we introduced principal component analysis (PCA) to learn the linear denoiser $\hat{\boldsymbol{U}}^\top$ when the observed data $\boldsymbol{x}$ follow the statistical model $\boldsymbol{x} = \boldsymbol{U}\boldsymbol{z} + \boldsymbol{\varepsilon}$. In this setting, the learned representations are linearly transformed input data $\hat{\boldsymbol{U}}^\top \boldsymbol{x}$. Under the linear model assumption, one can learn the low-dimensional linear structure with efficient optimization algorithms and strong theoretical guarantees. Moreover, the linear model assumption covers a wide range of applications and problems, including face recognition, magnetic resonance image recovery, and structure texture recovery [WM22].

On the other hand, the linear model can be limited when dealing with real-world applications, especially when the input data $\boldsymbol{x}$ is complex, such as speech and natural languages, images and videos, and robotic motions. The low-dim distributions of such data are typically *nonlinear*. How to deal with nonlinearity has a long history across different disciplines such as control theory, signal processing, and pattern recognition. There have been considerable efforts that try to extend methods and solutions for linear models to handle nonlinearity, including early effort to extend PCA to nonlinear PCA (as we will study in more detail in Chapter 6). In most cases, the methods are designed based on certain assumptions about the data distributions and tailored to specific problems.

More recently, deep neural networks have achieved remarkable success across a wide range of data and applications. A neural network

$$f(\cdot, \boldsymbol{\theta})\colon \boldsymbol{x} \xrightarrow{f^{\mathrm{pre}}} \boldsymbol{z}^1 \to \cdots \to \boldsymbol{z}^\ell \xrightarrow{f^\ell} \boldsymbol{z}^{\ell+1} \to \cdots \to \boldsymbol{z}^{L+1} = \boldsymbol{z}. \tag{5.0.1}$$

can learn effective features/representations for downstream applications. For example, a trained deep neural network $f(\cdot, \boldsymbol{\theta})$ can be applied to map images to feature vectors, that is, $\boldsymbol{z}_i = f(\boldsymbol{x}_i, \boldsymbol{\theta})$, while a linear classifier can be learned on top of such representations $\{\boldsymbol{z}_i\}$. One notable breakthrough is AlexNet [KSH12], a deep convolutional neural network trained with more than a million natural images, outperforming (on predictive tasks) all previous approaches that were based on hand-crafted features. One of the key differences between AlexNet and previous approaches is that the former *learns parameters of the nonlinear transformation from massive amounts of data* trained with back-propagation (BP) [RHW86b], as detailed in Section A.2.3 of Chapter A.

Subsequent popular practice models the mapping $f$ with other empirically designed artificial deep neural networks and learns the parameters $\boldsymbol{\theta}$ from random initialization via BP. Starting with the AlexNet [KSH12], the architectures of modern deep networks continue to be empirically revised and improved. Network architectures such as VGG [SZ14], ResNet [HZR+16b], DenseNet [HLV+17], CNN, RNN or LSTM [HS97], Transformer [VSP+17b], and mixtures of experts (MoE) [FZS22; SMM+17], etc. have continued to push the performance envelope. As part of the effort to improve the performance of deep networks, almost every component of the networks has been empirically scrutinized, and various revisions and improvements have been proposed. They are not limited to nonlinear activation functions [KUM+17; MHN13; NIG+18; XWC+15], skip connections [HZR+16b; RFB15b], normalizations [BKH16; IS15; MKK+18; UVL16; WH20], up/down sampling or pooling [SMB10], convolutions [KSH12; LBB+98b], etc. However, almost all such modifications have been developed through years of empirical trial and error or ablation studies. Some recent practices even take to the extreme by searching for effective network structures and training strategies through extensive random search techniques, such as Neural Architecture Search [BGN+17; ZL17], AutoML [HKV19], and Learning to Learn [ADG+16].

Despite the wide application of deep neural networks, it is not clear what the underlying design principles of such a constructed network are. In particular, it is not clear what mathematical function each layer of the network performs. In this chapter, based on the results from previous chapters, we develop a principled framework that will provide a fully rigorous mathematical interpretation of the role of a deep network, including its individual layers and the network as a whole.

To understand deep networks and how they should be better designed, we must start with the objective of representation learning. In previous chapters, we have argued that the objective is to identify the intrinsically low-dimensional data distribution and then transform it to a compact and structured (say piecewise linear) representation. As we have seen in the previous chapter, the general approach to identifying a low-dimensional data distribution is through a compression process that progressively minimizes the entropy or coding rate of the distribution. However, up to this point, we have been using empirically designed deep networks to model or approximate the operations that aim to optimize these objectives, such as the score function for denoising (in Section 1.3.1) or the transformation that maximizes the rate reduction (in Section 4.2.4).

As we have argued in the previous chapter, Section 4.2 in particular, one can measure the goodness of the resulting representation by the information of the representation gained from a "lazy" representation which models all data as one big Gaussian.[1] In particular, if *we use a mixture of Gaussians (subspaces)*[2] *as prototypical distributions to approximate the non-linear distribution of interest*, then we can efficiently measure the coding rate of such a representation using the (sum of) rate distortion functions of the associated Gaussians. Then the amount of *information gained* or (relative) entropy reduced with such a modeling can be measured by the difference between the coding rate for the lazy representation and that for the more refined representation. Then, the objective of representation learning is to maximize this information gain, also known as the rate reduction objective.

As we will see in this chapter, once the objective of representation learning is clear, the role of a deep neural network is precisely to help optimize the objective iteratively. Each layer of a deep neural network can be naturally derived as an iterative optimization step to incrementally maximize the information gain, including the popular architectures of ResNet, CNN, and Transformer, and other more advanced variants. In particular, this chapter aims to answer the following questions about deep networks:

- Section 5.1 — given a measure of goodness for a learned representation, how to construct the nonlinear mapping from the data to the optimal representation via unrolled optimization for the objective?

- Section 5.2 — how would the above unrolling approach provide a principled interpretation of the popular transformer architectures; if so, what are the associated objective and optimization mechanisms?

- Section 5.3 — how would this framework guide us to design more efficient or more parsimonious deep architectures?

## 5.1 White-Box Deep Networks via Unrolled Optimization

Now, if we agree that maximizing the rate reduction or information gain leads to the desired representation as discussed in Section 4.2, the remaining question is how to construct and learn a (nonlinear) mapping from the data $\boldsymbol{X}$ to the optimal representation $\boldsymbol{Z}^*$. This involves designing a network architecture and learning algorithm that can effectively capture the underlying structures in the data and faithfully realize the optimal representation.

[1] that we have seen in the previous chapter as one particular choice of interpretation of the sampled dataset.

[2] which we have studied thoroughly in the previous chapter.

### 5.1.1 Deep Networks from Unrolled Gradient Descent

In the previous chapter, we presented the rate reduction objective (4.2.15) as a principled objective for learning linear discriminative representations of the data. We have, however, not specified the architecture of the feature mapping $\boldsymbol{z} = f(\boldsymbol{x}, \boldsymbol{\theta})$ for extracting such representations from input data $\boldsymbol{x}$. A straightforward choice is to use a conventional deep network, such as ResNet, for implementing $f(\boldsymbol{x}, \boldsymbol{\theta})$. As we have seen in Section 4.3.1, such a choice often leads to decent performance empirically. Nonetheless, there remain several unanswered problems with adopting an arbitrary deep network. Although the learned feature representation is now more interpretable, the network itself is still *not*. It is unclear why any chosen "black-box" network is able to optimize the desired MCR$^2$ objective at all. The good empirical results (say with a ResNet) do not necessarily justify the particular choice in architectures and operators of the network: Why is a deep layered model even necessary; what do additional layers try to improve or simplify; how wide and deep is adequate; or is there any rigorous justification for the convolutions (in a popular multi-channel form) and nonlinear operators (e.g. ReLU or softmax) used?

In this chapter, we show that using gradient ascent to maximize the rate reduction $\Delta R_\epsilon(\boldsymbol{Z} \mid \boldsymbol{\Pi})$ as defined in (4.2.15) naturally leads to a "white-box" deep network that realizes the desired mapping. All network layers, linear/nonlinear operators, and parameters are *explicitly constructed in a purely forward propagation fashion.* Moreover, such network architectures resemble existing empirically-designed deep networks, providing principled justifications for their design.

**Gradient ascent for coding rate reduction.** From the previous chapter, we see that to seek a linear discriminative representation (LDR), mathematically, we are essentially seeking a continuous mapping $f(\cdot) : \boldsymbol{x} \mapsto \boldsymbol{z}$ from the data $\boldsymbol{X} = [\boldsymbol{x}_1, \ldots, \boldsymbol{x}_N] \in \mathbb{R}^{D \times N}$ (or initial features extracted from the data[3]) to an optimal representation $\boldsymbol{Z} = [\boldsymbol{z}_1, \ldots, \boldsymbol{z}_N] \in \mathbb{R}^{d \times N}$ that maximizes the following coding rate reduction objective (also see (4.2.17)):

$$\Delta R_\epsilon(\boldsymbol{Z} \mid \boldsymbol{\Pi}) \doteq \underbrace{\frac{1}{2} \log\det\Big(\boldsymbol{I} + \alpha \boldsymbol{Z}\boldsymbol{Z}^\top\Big)}_{R_\epsilon(\boldsymbol{Z})} - \underbrace{\sum_{k=1}^{K} \frac{\gamma_k}{2} \log\det\Big(\boldsymbol{I} + \alpha_k \boldsymbol{Z}\boldsymbol{\Pi}_k\boldsymbol{Z}^\top\Big)}_{R_\epsilon^c(\boldsymbol{Z}\mid\boldsymbol{\Pi})}, \tag{5.1.1}$$

where $\epsilon > 0$ is a prescribed quantization error and for simplicity we denote[4]

$$\alpha \doteq \frac{d}{N\epsilon^2}, \quad \alpha_k \doteq \frac{d}{\operatorname{tr}(\boldsymbol{\Pi}_k)\epsilon^2}, \quad \gamma_k \doteq \frac{\operatorname{tr}(\boldsymbol{\Pi}_k)}{N}, \quad \text{for } k = 1, \ldots, K. \tag{5.1.2}$$

The question really boils down to whether there is a *constructive* way of finding such a continuous mapping $f(\cdot, \boldsymbol{\theta})$ from $\boldsymbol{x}$ to $\boldsymbol{z}$? To this end, let us consider incrementally maximizing the objective $\Delta R_\epsilon$ as a function of $\boldsymbol{Z} \subseteq$ $\mathbb{S}^{d-1}$. Although there might be many optimization schemes to choose from, for simplicity we first consider the arguably simplest projected *gradient ascent* (PGA) scheme:[5]

[3] As we will see the necessity of such a feature extraction in the next section.

[4] Notice our use of slightly simplified notation compared to Chapter 3.

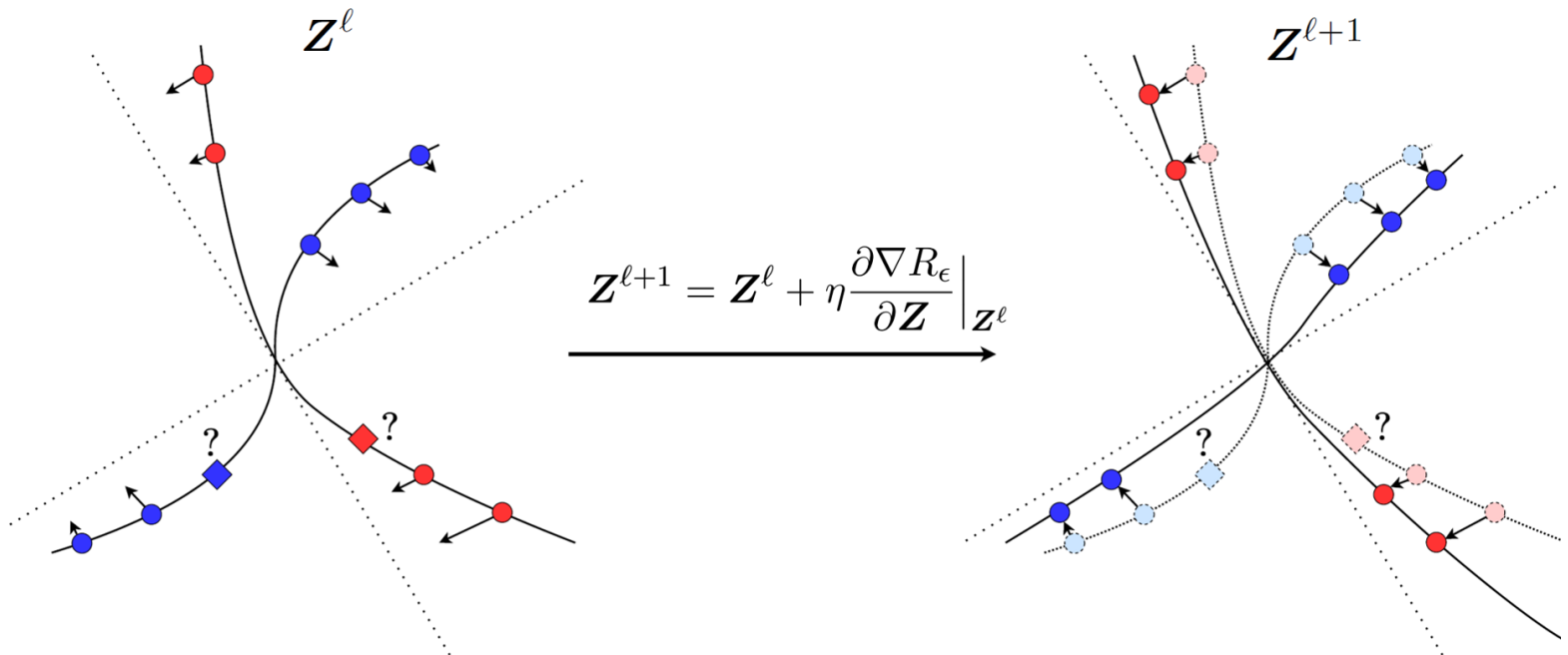

Figure 5.1: Incremental deformation via gradient flow to both flatten data of each class into a subspace and push different classes apart.

$$\boldsymbol{Z}^{\ell+1} \;\propto\; \boldsymbol{Z}^{\ell} + \eta \cdot \frac{\partial \Delta R_\epsilon}{\partial \boldsymbol{Z}}(\boldsymbol{Z}^{\ell}) \quad \text{s.t.} \quad \boldsymbol{Z}^{\ell+1} \subseteq \mathbb{S}^{d-1}, \quad \ell = 1, 2, \ldots, \tag{5.1.3}$$

for some step size $\eta > 0$ and the iterate starts with the given data $\boldsymbol{Z}^0 = \boldsymbol{X}$.[6] This scheme can be interpreted as how one should incrementally adjust locations of the current features $\boldsymbol{Z}^{\ell}$, initialized as the input data $\boldsymbol{X}$, in order for the resulting $\boldsymbol{Z}^{\ell+1}$ to improve the rate reduction $\Delta R_\epsilon$, as illustrated in Figure 5.1.

Simple calculation shows that the gradient $\partial \Delta R_\epsilon / \partial \boldsymbol{Z}$ entails evaluating the following derivatives of the two terms in $\Delta R_\epsilon$:

$$\frac{1}{2}\frac{\partial \log\det(\boldsymbol{I} + \alpha \boldsymbol{Z}\boldsymbol{Z}^\top)}{\partial \boldsymbol{Z}}(\boldsymbol{Z}^{\ell}) = \underbrace{\alpha(\boldsymbol{I} + \alpha \boldsymbol{Z}^{\ell}(\boldsymbol{Z}^{\ell})^\top)^{-1}}_{\boldsymbol{E}^{\ell}\,\in \mathbb{R}^{d\times d}} \boldsymbol{Z}^{\ell}, \tag{5.1.4}$$

$$\frac{1}{2}\frac{\partial \left(\gamma_k \log\det(\boldsymbol{I} + \alpha_k \boldsymbol{Z}\boldsymbol{\Pi}_k \boldsymbol{Z}^\top)\right)}{\partial \boldsymbol{Z}}(\boldsymbol{Z}^{\ell}) = \gamma_k \underbrace{\alpha_k(\boldsymbol{I} + \alpha_k \boldsymbol{Z}^{\ell}\boldsymbol{\Pi}_k(\boldsymbol{Z}^{\ell})^\top)^{-1}}_{\boldsymbol{C}_k^{\ell}\,\in \mathbb{R}^{d\times d}} \boldsymbol{Z}^{\ell}\boldsymbol{\Pi}_k. \tag{5.1.5}$$

Notice that in the above, the matrix $\boldsymbol{E}^{\ell}$ only depends on $\boldsymbol{Z}^{\ell}$ and it aims to *expand* all the features to increase the overall coding rate; the matrix $\boldsymbol{C}_k^{\ell}$ depends on features from the $k$-class and aims to *compress* them to reduce the coding rate of each class. Then the complete gradient $\frac{\partial \Delta R_\epsilon}{\partial \boldsymbol{Z}}(\boldsymbol{Z}^\ell) \in \mathbb{R}^{d\times N}$ is of the form:

$$\frac{\partial \Delta R_\epsilon}{\partial \boldsymbol{Z}}(\boldsymbol{Z}^\ell) = \underbrace{\boldsymbol{E}^\ell}_{\text{Expansion}} \boldsymbol{Z}^\ell - \sum_{k=1}^{K} \gamma_k \underbrace{\boldsymbol{C}_k^\ell}_{\text{Compression}} \boldsymbol{Z}^\ell \boldsymbol{\Pi}_k. \tag{5.1.6}$$

[5] Notice that we use subscript $j$ on $\boldsymbol{Z}_j$ to indicate features in the $j$-th class and superscript $\ell$ on $\boldsymbol{Z}^{\ell}$ to indicate all features at $\ell$-th iteration or layer.

[6] Again, for simplicity, we here first assume the initial features $\boldsymbol{Z}^1$ are the data themselves. Note that here $\ell$ denotes the number of iterations. Hence, the data and the features have the same dimension $d$. This needs not to be the case though. As we will see in the next section, the initial features can be some (lifted) features of the data to begin with and could in principle have a different (much higher) dimension. All subsequent iterates have the same dimension.

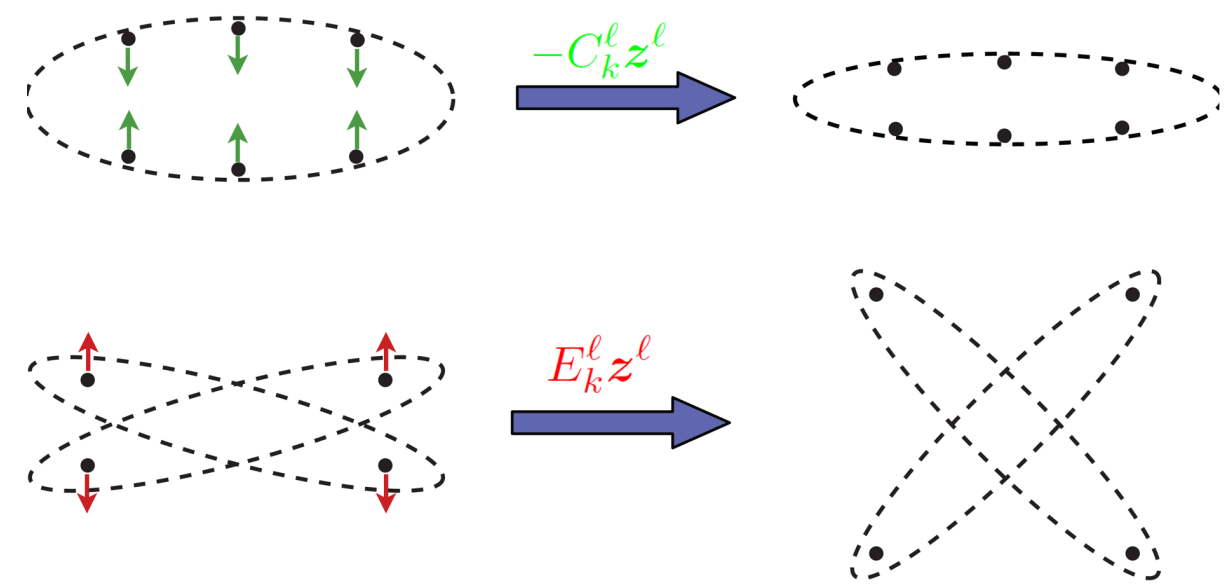

Figure 5.2: Interpretation of $\boldsymbol{C}_k^\ell$ and $\boldsymbol{E}^\ell$: $\boldsymbol{C}_k^\ell$ compresses each class by contracting the features to a low-dimensional subspace; $\boldsymbol{E}^\ell$ expands all features by contrasting and repelling features across different classes.

*Remark* 5.1 (Interpretation of $\boldsymbol{E}^\ell$ and $\boldsymbol{C}_j^\ell$ as linear operators). For any $\boldsymbol{z}^\ell \in \mathbb{R}^d$,

$$\boldsymbol{E}^\ell \boldsymbol{z}^\ell = \alpha(\boldsymbol{z}^\ell - \boldsymbol{Z}^\ell \boldsymbol{q}_\star^\ell), \quad \text{where} \quad \boldsymbol{q}_\star^\ell \doteq \arg\min_{\boldsymbol{q}^\ell} \left\{\alpha\|\boldsymbol{z}^\ell - \boldsymbol{Z}^\ell \boldsymbol{q}^\ell\|_2^2 + \|\boldsymbol{q}^\ell\|_2^2\right\}. \tag{5.1.7}$$

Notice that $\boldsymbol{q}_\star^\ell$ is exactly the solution to the ridge regression by all the data points $\boldsymbol{Z}^\ell$ concerned. Therefore, $\boldsymbol{E}^\ell$ (similarly for $\boldsymbol{C}_k^\ell$) is approximately (i.e., when $N$ is large enough) the projection onto the orthogonal complement of the subspace spanned by columns of $\boldsymbol{Z}^\ell$. Another way to interpret the matrix $\boldsymbol{E}^\ell$ is through eigenvalue decomposition of the covariance matrix $\boldsymbol{Z}^\ell(\boldsymbol{Z}^\ell)^\top$. Assuming that $\boldsymbol{Z}^\ell(\boldsymbol{Z}^\ell)^\top \doteq \boldsymbol{U}^\ell \boldsymbol{\Lambda}^\ell (\boldsymbol{U}^\ell)^\top$ where $\boldsymbol{\Lambda}^\ell \doteq \operatorname{diag}\left(\lambda_1^\ell, \dots, \lambda_d^\ell\right)$, we have

$$\boldsymbol{E}^\ell = \alpha \boldsymbol{U}^\ell \operatorname{diag}\left(\frac{1}{1+\alpha\lambda_1^\ell}, \dots, \frac{1}{1+\alpha\lambda_d^\ell}\right) \left(\boldsymbol{U}^\ell\right)^\top. \tag{5.1.8}$$

Therefore, the matrix $\boldsymbol{E}^\ell$ operates on a vector $\boldsymbol{z}^\ell$ by stretching in a way that directions of large variance are shrunk while directions of vanishing variance are kept. These are exactly the directions (5.1.4) in which we move the features so that the overall volume expands and the coding rate will increase, hence the positive sign. To the opposite effect, the directions associated with (5.1.5) are "residuals" of features of each class that deviate from the subspace to which they are supposed to belong. These are exactly the directions in which the features need to be compressed back onto their respective subspace, hence the negative sign (see Figure 5.2).

Essentially, the linear operations $\boldsymbol{E}^\ell$ and $\boldsymbol{C}_k^\ell$ in gradient ascent for rate reduction are determined by training data conducting "auto-regressions". The recent renewed understanding about ridge regression in an over-parameterized setting [WX20; YYY+20] indicates that using seemingly redundantly sampled data (from each subspace) as regressors do not lead to overfitting.

**Gradient-guided feature map increment.** Notice that in the above, the gradient ascent considers all the features $\boldsymbol{Z}^\ell = [\boldsymbol{z}_1^\ell, \dots, \boldsymbol{z}_N^\ell]$ as free variables. The increment $\boldsymbol{Z}^{\ell+1} - \boldsymbol{Z}^\ell = \eta \frac{\partial \Delta R_\epsilon}{\partial \boldsymbol{Z}}(\boldsymbol{Z}^\ell)$ does not yet give a transformation on the entire feature domain $\boldsymbol{z}^\ell \in \mathbb{R}^d$. According to equation (5.1.6), the gradient cannot be evaluated at a point whose membership is not known, as illustrated in Figure 5.1. Hence, in order to find the optimal $f(\boldsymbol{x}, \boldsymbol{\theta})$ explicitly, we may consider constructing a small increment transform $g(\cdot, \theta^\ell)$ on the $\ell$-th layer feature $\boldsymbol{z}^\ell$ to emulate the above (projected) gradient scheme:

$$\boldsymbol{z}^{\ell+1} \ \propto \ \boldsymbol{z}^\ell + \eta \cdot g(\boldsymbol{z}^\ell, \theta^\ell) \quad \text{subject to} \quad \boldsymbol{z}^{\ell+1} \in \mathbb{S}^{d-1} \tag{5.1.9}$$

such that $\left[g(\boldsymbol{z}_1^\ell, \boldsymbol{\theta}^\ell), \dots, g(\boldsymbol{z}_N^\ell, \boldsymbol{\theta}^\ell)\right] \approx \frac{\partial \Delta R_\epsilon}{\partial \boldsymbol{Z}}(\boldsymbol{Z}^\ell)$. That is, we need to approximate the gradient flow $\frac{\partial \Delta R_\epsilon}{\partial \boldsymbol{Z}}$ that locally deforms all (training) features $\{\boldsymbol{z}_i^\ell\}_{i=1}^N$ with a continuous mapping $g(\boldsymbol{z}, \boldsymbol{\theta})$ defined on the entire feature space $\boldsymbol{z}^\ell \in \mathbb{R}^d$. Notice that one may interpret the increment (5.1.9) as a discretized version of a continuous differential equation:

$$\dot{\boldsymbol{z}} = g(\boldsymbol{z}, \theta). \tag{5.1.10}$$

Hence the (deep) network so constructed can be interpreted as a certain neural ODE [CRB+18]. Nevertheless, unlike neural ODE where the flow $g$ is chosen to be some generic structures, here our $g(\boldsymbol{z}, \boldsymbol{\theta})$ is to emulate the gradient flow of the rate reduction on the feature set (as shown in Figure 5.1):

$$\dot{\boldsymbol{Z}} = \frac{\partial \Delta R_\epsilon}{\partial \boldsymbol{Z}},$$

and its structure is entirely derived and fully determined from this objective, without any other priors or heuristics.

By inspecting the structure of the gradient (5.1.6), it suggests that a natural candidate for the increment transform $g(\boldsymbol{z}^\ell, \boldsymbol{\theta}^\ell)$ is of the form:

$$g(\boldsymbol{z}^\ell, \boldsymbol{\theta}^\ell) \ \doteq \ \boldsymbol{E}^\ell \boldsymbol{z}^\ell - \sum_{k=1}^K \gamma_k \pi_k(\boldsymbol{z}^\ell) \boldsymbol{C}_k^\ell \boldsymbol{z}^\ell \in \mathbb{R}^d, \tag{5.1.11}$$

where $\pi_k(\boldsymbol{z}^\ell) \in [0, 1]$ indicates the probability of $\boldsymbol{z}^\ell$ belonging to the $k$-th class. The increment map parameters $\boldsymbol{\theta}^\ell$ depend on: First, a set of linear maps represented by $\boldsymbol{E}^\ell$ and $\{\boldsymbol{C}_k^\ell\}_{k=1}^K$ that depend only on statistics of features of the training $\boldsymbol{Z}^\ell$; Second, the membership $\{\pi_k(\boldsymbol{z}^\ell)\}_{k=1}^K$ of any feature $\boldsymbol{z}^\ell$. Notice that on the training samples $\boldsymbol{Z}^\ell$, for which the memberships $\boldsymbol{\Pi}_k$ are known, the so defined $g(\boldsymbol{z}^\ell, \boldsymbol{\theta})$ gives exactly the values for the gradient $\frac{\partial \Delta R_\epsilon}{\partial \boldsymbol{Z}}(\boldsymbol{Z}^\ell)$.

Since we only have the membership for the training samples, the function $g(\cdot)$ defined in (5.1.11) can only be evaluated on the training. To extrapolate $g(\cdot)$ to the entire feature space, we need to estimate $\pi_k(\boldsymbol{z}^\ell)$ in its second term. In conventional deep learning, this map is typically modeled as a deep network and learned from the training data, say via *back propagation*. Nevertheless, our goal here is not to learn a precise classifier $\pi_k(\boldsymbol{z}^\ell)$ already. Instead, we only need a good enough estimate of the class information in order for $g(\cdot)$ to approximate the gradient $\frac{\partial \Delta R_\epsilon}{\partial \boldsymbol{Z}}$ well.

From the geometric interpretation of the linear maps $\boldsymbol{E}^\ell$ and $\boldsymbol{C}_k^\ell$ given by Remark 5.1, the term $\boldsymbol{p}_k^\ell \doteq \boldsymbol{C}_k^\ell \boldsymbol{z}^\ell$ can be viewed as (approximately) the projection of $\boldsymbol{z}^\ell$ onto the orthogonal complement of each class $j$. Therefore, $\|\boldsymbol{p}_j^\ell\|_2$ is small if $\boldsymbol{z}^\ell$ is in class $j$ and large otherwise. This motivates us to estimate its membership based on the following softmax function:

$$\widehat{\boldsymbol{\pi}}(\boldsymbol{z}^\ell) \doteq \operatorname{softmax}\left(-\lambda \begin{bmatrix} \|\boldsymbol{C}_1^\ell \boldsymbol{z}^\ell\|_2 \\ \vdots \\ \|\boldsymbol{C}_K^\ell \boldsymbol{z}^\ell\|_2 \end{bmatrix}\right) \tag{5.1.12}$$

$$= \frac{1}{\sum_{k=1}^K \exp(-\lambda \|\boldsymbol{C}_k^\ell \boldsymbol{z}^\ell\|_2)} \begin{bmatrix} \exp(-\lambda \|\boldsymbol{C}_1^\ell \boldsymbol{z}^\ell\|_2) \\ \vdots \\ \exp(-\lambda \|\boldsymbol{C}_K^\ell \boldsymbol{z}^\ell\|_2) \end{bmatrix} \in [0,1]^K. \tag{5.1.13}$$

Hence, the second term of (5.1.11) can be approximated by this estimated membership:

$$\sum_{k=1}^K \gamma_k \pi_k(\boldsymbol{z}^\ell) \boldsymbol{C}_k^\ell \boldsymbol{z}^\ell \;\approx\; \sum_{k=1}^K \gamma_k \widehat{\pi}_k(\boldsymbol{z}^\ell) \boldsymbol{C}_k^\ell \boldsymbol{z}^\ell \;\doteq\; \boldsymbol{\sigma}\Big([\boldsymbol{C}_1^\ell \boldsymbol{z}^\ell, \ldots, \boldsymbol{C}_K^\ell \boldsymbol{z}^\ell]\Big), \tag{5.1.14}$$

which is denoted as a nonlinear operator $\boldsymbol{\sigma}(\cdot)$ on outputs of the feature $\boldsymbol{z}^\ell$ through $K$ groups of filters: $[\boldsymbol{C}_1^\ell, \ldots, \boldsymbol{C}_K^\ell]$. Notice that the nonlinearality arises due to a "soft" assignment of class membership based on the feature responses from those filters.

Overall, combining (5.1.9), (5.1.11), and (5.1.14), the increment feature transform from $\boldsymbol{z}^\ell$ to $\boldsymbol{z}^{\ell+1}$ now becomes

$$\boldsymbol{z}^{\ell+1} \propto \; \boldsymbol{z}^\ell + \eta \cdot \boldsymbol{E}^\ell \boldsymbol{z}^\ell - \eta \cdot \boldsymbol{\sigma}\big([\boldsymbol{C}_1^\ell \boldsymbol{z}^\ell, \ldots, \boldsymbol{C}_K^\ell \boldsymbol{z}^\ell]\big) \tag{5.1.15}$$

$$= \; \boldsymbol{z}^\ell + \eta \cdot g(\boldsymbol{z}^\ell, \boldsymbol{\theta}^\ell) \qquad \text{s.t.} \quad \boldsymbol{z}^{\ell+1} \in \mathbb{S}^{d-1}, \tag{5.1.16}$$

with the nonlinear function $\boldsymbol{\sigma}(\cdot)$ defined above and $\boldsymbol{\theta}^\ell$ collecting all the layer-wise parameters. That is $\boldsymbol{\theta}^\ell = \big\{\boldsymbol{E}^\ell, \boldsymbol{C}_1^\ell, \ldots, \boldsymbol{C}_K^\ell, \gamma_k, \lambda\big\}$. Note features at each layer are always "normalized" by projecting onto the unit sphere $\mathbb{S}^{d-1}$, denoted as $\mathcal{P}_{\mathbb{S}^{d-1}}$, i.e., dividing each feature by its norm.[7] The form of increment in (5.1.15) can be illustrated by a diagram in Figure 5.3(a).

[7] This mirrors the common LayerNorm block in neural networks, which computes the projected feature and then multiplies and adds by trainable parameters [BKH16], see Section 8.2.3 for more details.

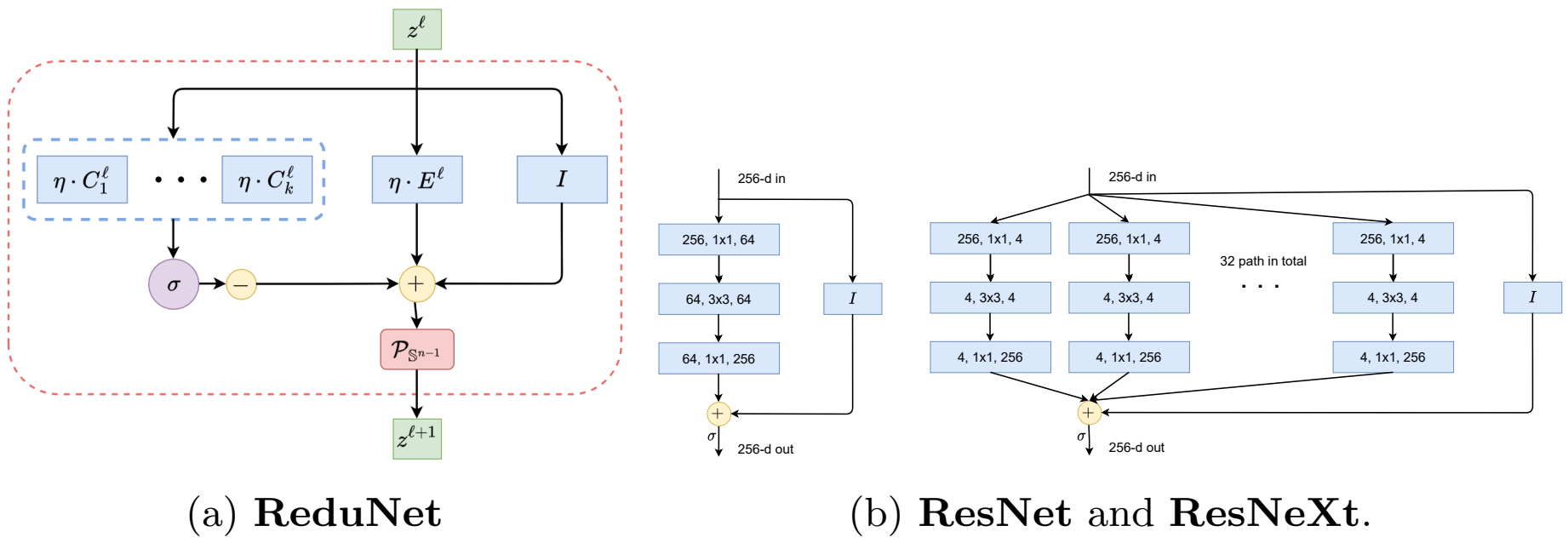

(a) **ReduNet** (b) **ResNet** and **ResNeXt**.

Figure 5.3: Network Architectures of the ReduNet and comparison with others. **(a)**: Layer structure of the **ReduNet** derived from one iteration of gradient ascent for optimizing rate reduction. **(b)** (left): A layer of ResNet [HZR+16b]; and **(b)** (right): A layer of ResNeXt [XGD+17]. As we will see in Section 5.1.2, the linear operators $\boldsymbol{E}^\ell$ and $\boldsymbol{C}_k^\ell$ of the ReduNet naturally become (multi-channel) convolutions when shift-invariance is imposed.

**Deep network for optimizing rate reduction.** Notice that the increment is constructed to emulate the gradient ascent for the rate reduction $\Delta R_\epsilon$. Hence by transforming the features iteratively via the above process, we expect the rate reduction to increase, as we will see in the experimental section. This iterative process, once converged, say after $L$ iterations, gives the desired feature map $f(\boldsymbol{x}, \boldsymbol{\theta})$ on the input $\boldsymbol{x} = \boldsymbol{z}^0$, precisely in the form of a *deep network*, in which each layer has the structure shown in Figure 5.3 left:

$$
\begin{aligned}
f(\boldsymbol{x}, \boldsymbol{\theta}) &= f^L \circ f^{L-1} \circ \cdots \circ f^1 \circ f^0(\boldsymbol{z}^0), && (5.1.17)\\
f^\ell(\boldsymbol{z}^\ell, \boldsymbol{\theta}^\ell) &\doteq \boldsymbol{z}^{\ell+1} = \mathcal{P}_{\mathbb{S}^{n-1}}[\boldsymbol{z}^\ell + \eta \cdot g(\boldsymbol{z}^\ell, \boldsymbol{\theta}^\ell)], && (5.1.18)\\
g(\boldsymbol{z}^\ell, \boldsymbol{\theta}^\ell) &= \boldsymbol{E}^\ell \boldsymbol{z}^\ell - \boldsymbol{\sigma}\big([\boldsymbol{C}_1^\ell \boldsymbol{z}^\ell, \dots, \boldsymbol{C}_K^\ell \boldsymbol{z}^\ell]\big). && (5.1.19)
\end{aligned}
$$

As this deep network is derived from maximizing the rate **redu**ction, we call it the **ReduNet**. By comparing the architecture of ReduNet with those of popular empirically designed networks, **ResNet** and **ResNeXt** shown in Figure 5.3, the similarity is somewhat uncanny. Conceptually, ReduNet could also be used to justify the popular mixture of experts (**MoE**) architecture [SMM+17] as each parallel channel, $\boldsymbol{C}_k^\ell$, can be viewed as an "expert" trained for each class of objects.

We summarize the training and evaluation of ReduNet in Algorithm 5.1 and Algorithm 5.2, respectively. Notice that all parameters of the network are explicitly constructed layer by layer in a *forward propagation* fashion. The construction does not need any back propagation! The so-learned features can be directly used for classification, say via a nearest subspace classifier.

*Example* 5.1. To provide some intuition on how ReduNet transforms the features, we provide a simple example with mixed 3D Gaussians and visualize how the features are transformed in Figure 5.5. Consider a mixture of three Gaussian distributions in $\mathbb{R}^3$ that is projected onto $\mathbb{S}^2$. We first generate data points for 3 classes: for $k = 1, 2, 3$, $\boldsymbol{X}_k = [\boldsymbol{x}_{k,1}, \ldots, \boldsymbol{x}_{k,m}] \in \mathbb{R}^{3\times m}$, $\boldsymbol{x}_{k,i} \sim \mathcal{N}(\boldsymbol{\mu}_k, \sigma_k^2 \boldsymbol{I})$, and $\pi(\boldsymbol{x}_{k,i}) = k$. We set $m = 500, \sigma_1 = \sigma_2 = \sigma_3 = 0.1$, and $\boldsymbol{\mu}_1, \boldsymbol{\mu}_2, \boldsymbol{\mu}_3 \in \mathbb{S}^2$. Then we project all the data points onto $\mathbb{S}^2$, i.e., $\boldsymbol{x}_{k,i}/\|\boldsymbol{x}_{k,i}\|_2$. To construct the network (computing $\boldsymbol{E}^\ell, \boldsymbol{C}_k^\ell$ for the $\ell$-th layer), we set the number of iterations/layers $L = 2,000$, step size $\eta = 0.5$, and precision $\epsilon = 0.1$. We do this only to demonstrate that our framework leads to stable deep networks even with thousands of layers. In practice, thousands of layers may not be necessary and one can stop whenever adding new layers gives diminishing returns. For this example, a couple of hundred layers is sufficient. Hence, the clear optimization objective gives a natural criterion for the depth of the network needed.

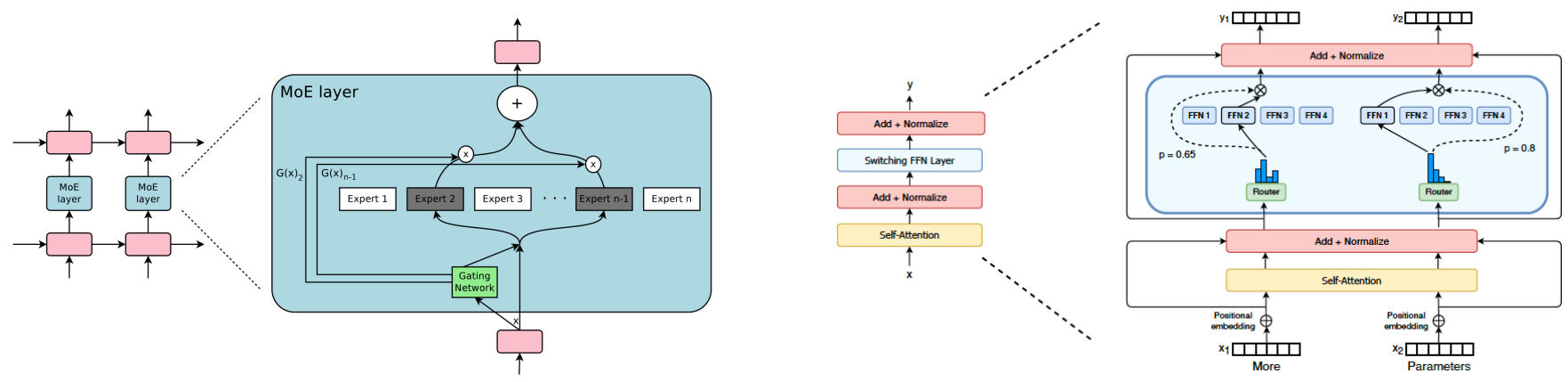

Figure 5.4: Left: a mixture of experts (MoE) deep network [SMM+17]. Right: a sparsity-promoting Switch Transformer [FZS22], used to implement MoE with 1.7 trillion parameters.

As shown in Figure 5.5, we can observe that after the mapping $f(\cdot, \theta)$, samples from the same class are highly compressed and converge to a single cluster and the angle between two different clusters is approximately $\pi/2$, which is well aligned with the optimal solution $\boldsymbol{Z}^\star$ of the MCR$^2$ loss in $\mathbb{S}^2$. MCR$^2$ loss of features on different layers can be found in Figure 5.5(**c**). Empirically, we find that the constructed ReduNet is able to maximize MCR$^2$ loss and converges stably and samples from the same class converge to one cluster and points in different clusters are orthogonal to each other.[8] Moreover, when sampling new data points from the same distributions, we find that new samples from the same class consistently converge to the same cluster center as the training samples.

■

### 5.1.2 Convolutional Networks from Invariant Rate Reduction

In the previous section, we derived the layer-wise architecture of a deep network, the ReduNet, using unrolled optimization for the rate reduction objective. Specifically, the compression term $R_\epsilon^c(\boldsymbol{Z} \mid \boldsymbol{\Pi})$ in (5.1.1) is designed to compress representations from the same class. However, this formulation does not account for possible domain transformation or deformation of the input data. For

[8] Because of the low dimension, each cluster collapses to a 1-dimensional subspace on the sphere (e.g., a single point and its antipode); this is the extreme case of keeping within-cluster dimension minimal. In higher dimensions, of course, subspaces will be less compressed.

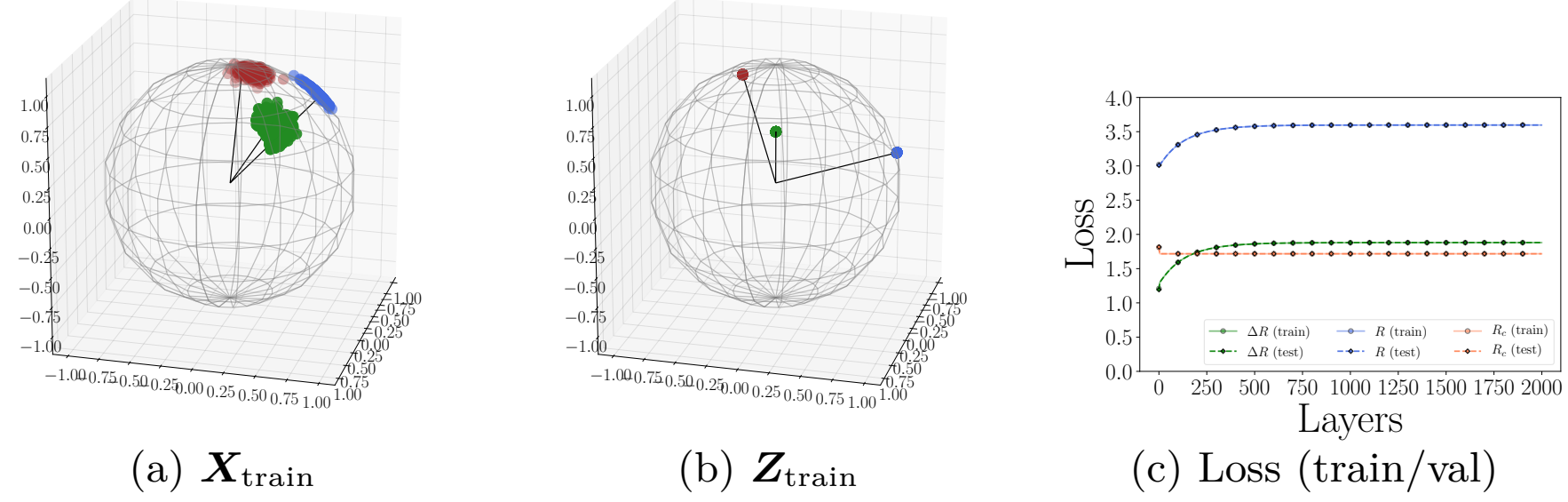

(a) $\boldsymbol{X}_{\text{train}}$ (b) $\boldsymbol{Z}_{\text{train}}$ (c) Loss (train/val)

Figure 5.5: Original samples and learned representations for 3D Mixture of Gaussians. We visualize data points $\boldsymbol{X}$ (before mapping $f(\cdot,\theta)$) in (**a**) and learned features $\boldsymbol{Z}$ (after mapping $f(\cdot,\theta)$) in (**b**) by scatter plot. In each scatter plot, each color represents one class of samples. In (**c**), we also show the plots for the progression of values of the objective functions.

instance, shifting an object slightly to the right does not change the semantic label of an image. In this section, we will demonstrate how convolutional layers can be derived by maximizing a rate reduction objective that is invariant to certain domain deformations, such as image rotations and translations.

For many clustering or classification tasks (such as object detection in images), we consider two samples as *equivalent* if they differ by certain classes of domain deformations or augmentations $\mathcal{T} = \{\tau\}$. Hence, we are only interested in low-dimensional structures that are *invariant* to such deformations (i.e., $\boldsymbol{x} \in \mathcal{M}$ iff $\tau(\boldsymbol{x}) \in \mathcal{M}$ for all $\tau \in \mathcal{T}$), which are known to have sophisticated geometric and topological structures and can be difficult to learn precisely in practice, even with rigorously designed CNNs [CW16a]. In this framework, this can be formulated in a very natural way: all equivariant instances are to be embedded into the same subspace, so that the subspace itself is invariant to the transformations under consideration.

In many applications, such as serial data or imagery data, the semantic meaning (labels) of the data are *invariant* to certain transformations $\mathfrak{g} \in \mathbb{G}$ (for some group $\mathbb{G}$) [CW16c; ZKR+17]. For example, the meaning of an audio signal is invariant to shift in time; and the identity of an object in an image is invariant to translation in the image plane. Hence, we prefer the feature mapping $f(\boldsymbol{x},\boldsymbol{\theta})$ is rigorously invariant to such transformations:

$$\textit{Group Invariance: } f(\boldsymbol{x} \circ \mathfrak{g}, \boldsymbol{\theta}) \sim f(\boldsymbol{x}, \boldsymbol{\theta}), \ \forall \mathfrak{g} \in \mathbb{G}, \tag{5.1.20}$$

where "$\sim$" indicates two features belonging to the same equivalent class. Although to ensure invariance or equivarience, convolutional operators have been common practice in deep networks [CW16c], it remains challenging in practice to train an (empirically designed) convolution network from scratch that can *guarantee* invariance even to simple transformations such as translation and rotation [AW18; ETT+17]. An alternative approach is to carefully design convolution filters of each layer so as to ensure translational invariance for a wide range of signals, say using wavelets as in ScatteringNet [BM13] and followup works [WB18]. However, in order to ensure invariance to generic signals, the number of convolutions needed usually grows exponentially with network depth. That is the reason why this type of network cannot be constructed so deep, usually only several layers.

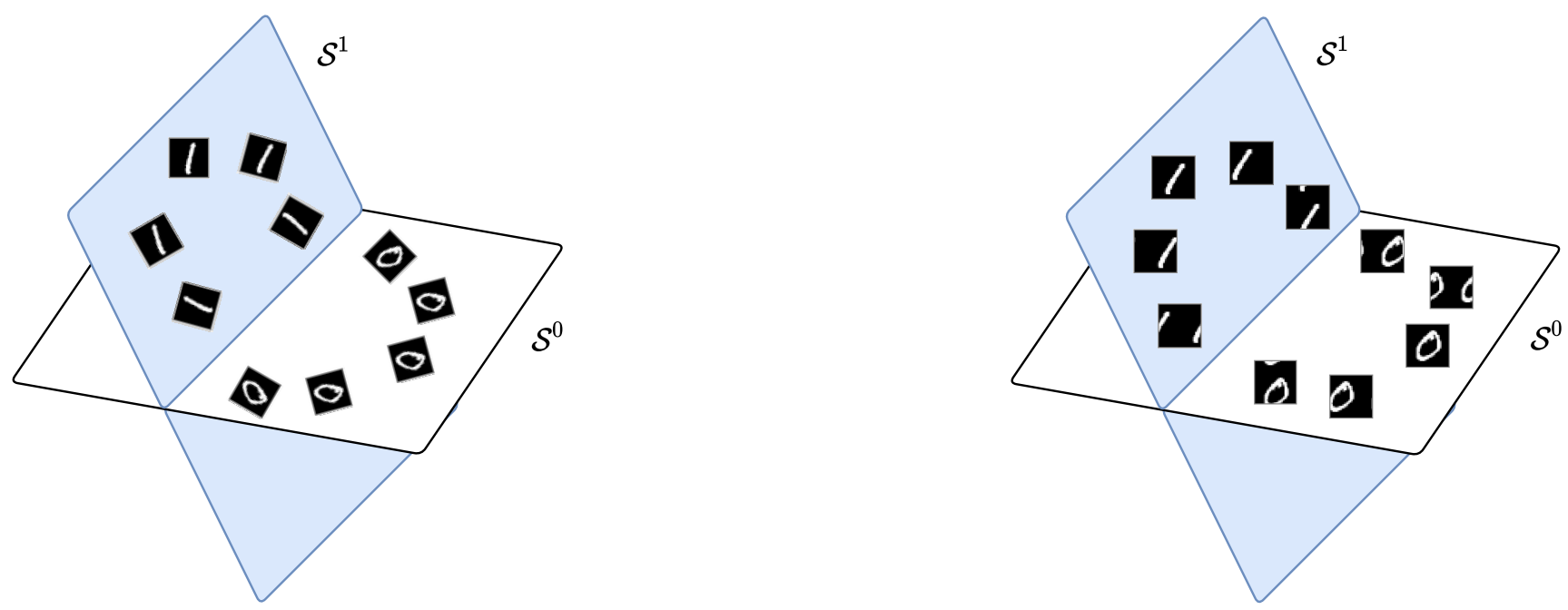

Figure 5.6: Illustration of the sought representation that is equivariant/invariant to image rotation (left) or translation (right): all transformed images of each class are mapped into the same subspace that is incoherent to other subspaces. The features embedded in each subspace are equivariant to the transformation group whereas each subspace is invariant to such transformations.

Now, we show that the MCR$^2$ principle is compatible with invariance in a natural and precise way: we only need to assign all transformed versions $\{\boldsymbol{x} \circ \mathfrak{g} \mid \mathfrak{g} \in \mathbb{G}\}$ into the same class as the data $\boldsymbol{x}$ and map their features $\boldsymbol{z}$ all to the same subspace $\mathcal{S}$. Hence, all group equivariant information is encoded only inside the subspace, and any classifier defined on the resulting set of subspaces will be automatically invariant to such group transformations. See Figure 5.6 for an illustration of the examples of 1D rotation and 2D translation. Next, we will rigorously show that when the group $\mathbb{G}$ is circular 1D shifting, the resulting deep network naturally becomes a *multi-channel convolution network*. Because the so-constructed network only needs to ensure invariance for the given data $\boldsymbol{X}$ or their features $\boldsymbol{Z}$, the number of convolutions needed actually remains constant through a very deep network, as opposed to the ScatteringNet.

**1D serial data and shift invariance** To classify one-dimensional data $\boldsymbol{x} = [x(0), x(1), \ldots, x(D-1)] \in \mathbb{R}^D$ invariant under shifting, we take $\mathbb{G}$ to be the group of all circular shifts. Each observation $\boldsymbol{x}_i$ generates a family $\{\boldsymbol{x}_i \circ \mathfrak{g} \mid \mathfrak{g} \in \mathbb{G}\}$ of shifted copies, which are the columns of the circulant matrix $\mathsf{circ}(\boldsymbol{x}_i) \in \mathbb{R}^{D \times D}$

given by

$$\mathsf{circ}(\boldsymbol{x}) \doteq \begin{bmatrix} x(0) & x(D-1) & \dots & x(2) & x(1) \\ x(1) & x(0) & x(D-1) & \cdots & x(2) \\ \vdots & x(1) & x(0) & \ddots & \vdots \\ x(D-2) & \vdots & \ddots & \ddots & x(D-1) \\ x(D-1) & x(D-2) & \dots & x(1) & x(0) \end{bmatrix} \in \mathbb{R}^{D\times D}. \tag{5.1.21}$$

We refer the reader to [KS12] for properties of circulant matrices. For simplicity, let $\boldsymbol{Z}^1 \doteq [\boldsymbol{z}_1^1, \dots, \boldsymbol{z}_N^1] = \boldsymbol{X} \in \mathbb{R}^{d\times N}$.[9] Then what happens if we construct the ReduNet from their circulant families $\mathsf{circ}(\boldsymbol{Z}^1) = \big[\mathsf{circ}(\boldsymbol{z}_1^1), \dots, \mathsf{circ}(\boldsymbol{z}_N^1)\big] \in \mathbb{R}^{d\times(dN)}$? That is, we want to compress and map all these into the same subspace by the ReduNet.

Notice that now the data covariance matrix:

$$\begin{aligned} \mathsf{circ}(\boldsymbol{Z}^1)\mathsf{circ}(\boldsymbol{Z}^1)^\top &= \big[\mathsf{circ}(\boldsymbol{z}_1^1), \dots, \mathsf{circ}(\boldsymbol{z}_N^1)\big]\big[\mathsf{circ}(\boldsymbol{z}_1^1), \dots, \mathsf{circ}(\boldsymbol{z}_N^1)\big]^\top \\ &= \sum_{i=1}^N \mathsf{circ}(\boldsymbol{z}_i^1)\mathsf{circ}(\boldsymbol{z}_i^1)^\top \in \mathbb{R}^{d\times d} \end{aligned} \tag{5.1.22}$$

associated with this family of samples is *automatically* a (symmetric) circulant matrix. Moreover, because the circulant property is preserved under sums, inverses, and products, the matrices $\boldsymbol{E}^1$ and $\boldsymbol{C}_k^1$ are also automatically circulant matrices, whose application to a feature vector $\boldsymbol{z} \in \mathbb{R}^d$ can be implemented using circular convolution "$\circledast$". Specifically, we have the following proposition.

**Proposition 5.1** (Convolution structures of $\boldsymbol{E}^1$ and $\boldsymbol{C}_k^1$)**.** *The matrix*

$$\boldsymbol{E}^1 = \alpha\big(\boldsymbol{I} + \alpha\mathsf{circ}(\boldsymbol{Z}^1)\mathsf{circ}(\boldsymbol{Z}^1)^\top\big)^{-1} \tag{5.1.23}$$

*is a circulant matrix and represents a circular convolution:*

$$\boldsymbol{E}^1\boldsymbol{z} = \boldsymbol{e}_1 \circledast \boldsymbol{z},$$

*where $\boldsymbol{e}_1 \in \mathbb{R}^d$ is the first column vector of $\boldsymbol{E}^1$ and "$\circledast$" is circular convolution defined as*

$$(\boldsymbol{e}_1 \circledast \boldsymbol{z})_i \doteq \sum_{j=0}^{d-1} e_1(j)x(i+d-j \ \mathit{mod}\ d).$$

*Similarly, the matrices $\boldsymbol{C}_k^1$ associated with any subsets of $\boldsymbol{Z}^1$ are also circular convolutions.*

[9] Again, to simplify discussion, we assume for now that the initial features $\boldsymbol{Z}^1$ are $\boldsymbol{X}$ themselves hence have the same dimension $d$, i.e., $D = d$. But that does not need to be the case as we will soon see that we need to lift $\boldsymbol{X}$ to a higher dimension.

Not only do the first-layer parameters $\boldsymbol{E}^1$ and $\boldsymbol{C}_k^1$ of the ReduNet become circulant convolutions but also the next-layer features remain circulant matrices. That is, the incremental feature transform in (5.1.15) applied to all shifted versions of a $\boldsymbol{z}^1 \in \mathbb{R}^d$, given by

$$\mathsf{circ}(\boldsymbol{z}^1) + \eta \cdot \boldsymbol{E}^1 \mathsf{circ}(\boldsymbol{z}^1) - \eta \cdot \boldsymbol{\sigma}\Big([\boldsymbol{C}_1^1 \mathsf{circ}(\boldsymbol{z}^1), \dots, \boldsymbol{C}_K^1 \mathsf{circ}(\boldsymbol{z}^1)]\Big), \tag{5.1.24}$$

is a circulant matrix. This implies that there is no need to construct circulant families from the second layer features as we did for the first layer. By denoting

$$\boldsymbol{z}^2 \propto \boldsymbol{z}^1 + \eta \cdot g(\boldsymbol{z}^1, \boldsymbol{\theta}^1) = \boldsymbol{z}^1 + \eta \cdot \boldsymbol{e}_1 \circledast \boldsymbol{z}^1 - \eta \cdot \boldsymbol{\sigma}\Big([\boldsymbol{c}_1^1 \circledast \boldsymbol{z}^1, \dots, \boldsymbol{c}_K^1 \circledast \boldsymbol{z}^1]\Big), \tag{5.1.25}$$

the features at the next level can be written as

$$\mathsf{circ}(\boldsymbol{Z}^2) = \big[\mathsf{circ}(\boldsymbol{z}_1^1 + \eta g(\boldsymbol{z}_1^1, \boldsymbol{\theta}^1)), \dots, \mathsf{circ}(\boldsymbol{z}_N^1 + \eta g(\boldsymbol{z}_N^1, \boldsymbol{\theta}^1))\big].$$

Continuing inductively, we see that all matrices $\boldsymbol{E}^\ell$ and $\boldsymbol{C}_k^\ell$ based on such $\mathsf{circ}(\boldsymbol{Z}^\ell)$ are circulant, and so are all features. By virtue of the properties of the data, ReduNet has taken the form of a convolutional network, *with no need to explicitly choose this structure!*

**A fundamental trade-off between invariance and sparsity.** There is one problem though: In general, the set of all circular permutations of a vector $\boldsymbol{z}$ gives a full-rank matrix. That is, the $d$ "augmented" features associated with each sample (hence each class) typically already span the entire space $\mathbb{R}^d$. For instance, all shifted versions of a delta function $\delta(d)$ can generate any other signal as their (dense) weighted superposition. The MCR$^2$ objective (4.2.15) will not be able to distinguish classes as different subspaces.

One natural remedy is to improve the separability of the data by "lifting" the original signal to a higher dimensional space, e.g., by taking their responses to multiple, filters $\boldsymbol{k}_1, \dots, \boldsymbol{k}_C \in \mathbb{R}^d$:

$$\boldsymbol{z}[c] = \boldsymbol{k}_c \circledast \boldsymbol{x} = \mathsf{circ}(\boldsymbol{k}_c)\boldsymbol{x} \in \mathbb{R}^d, \quad c = 1, \dots, C. \tag{5.1.26}$$

The filters can be pre-designed invariance-promoting filters,[10] or adaptively learned from the data,[11] or randomly selected as we do in our experiments. This operation lifts each original signal $\boldsymbol{x} \in \mathbb{R}^d$ to a $C$-channel feature, denoted as $\bar{\boldsymbol{z}} \doteq [\boldsymbol{z}[1], \dots, \boldsymbol{z}[C]]^\top \in \mathbb{R}^{C \times d}$. Then, we may construct the ReduNet on vector representations of $\bar{\boldsymbol{z}}$, denoted as $\vec{(\bar{\boldsymbol{z}})} \doteq [\boldsymbol{z}[1]^\top, \dots, \boldsymbol{z}[C]^\top] \in \mathbb{R}^{dC}$. The associated circulant version $\mathsf{circ}(\bar{\boldsymbol{z}})$ and its data covariance matrix, denoted as $\bar{\boldsymbol{\Sigma}}(\bar{\boldsymbol{z}})$, for all its shifted versions are given as:

$$\mathsf{circ}(\bar{\boldsymbol{z}}) \doteq \begin{bmatrix} \mathsf{circ}(\boldsymbol{z}[1]) \\ \vdots \\ \mathsf{circ}(\boldsymbol{z}[C]) \end{bmatrix} \in \mathbb{R}^{dC \times d}, \tag{5.1.27}$$

[10] For 1D signals like audio, one may consider the conventional short-time Fourier transform (STFT); for 2D images, one may consider 2D wavelets as in the ScatteringNet [BM13].

[11] For learned filters, one can learn filters as the principal components of samples as in the PCANet [CJG+15] or from convolution dictionary learning [LB19; QLZ19].

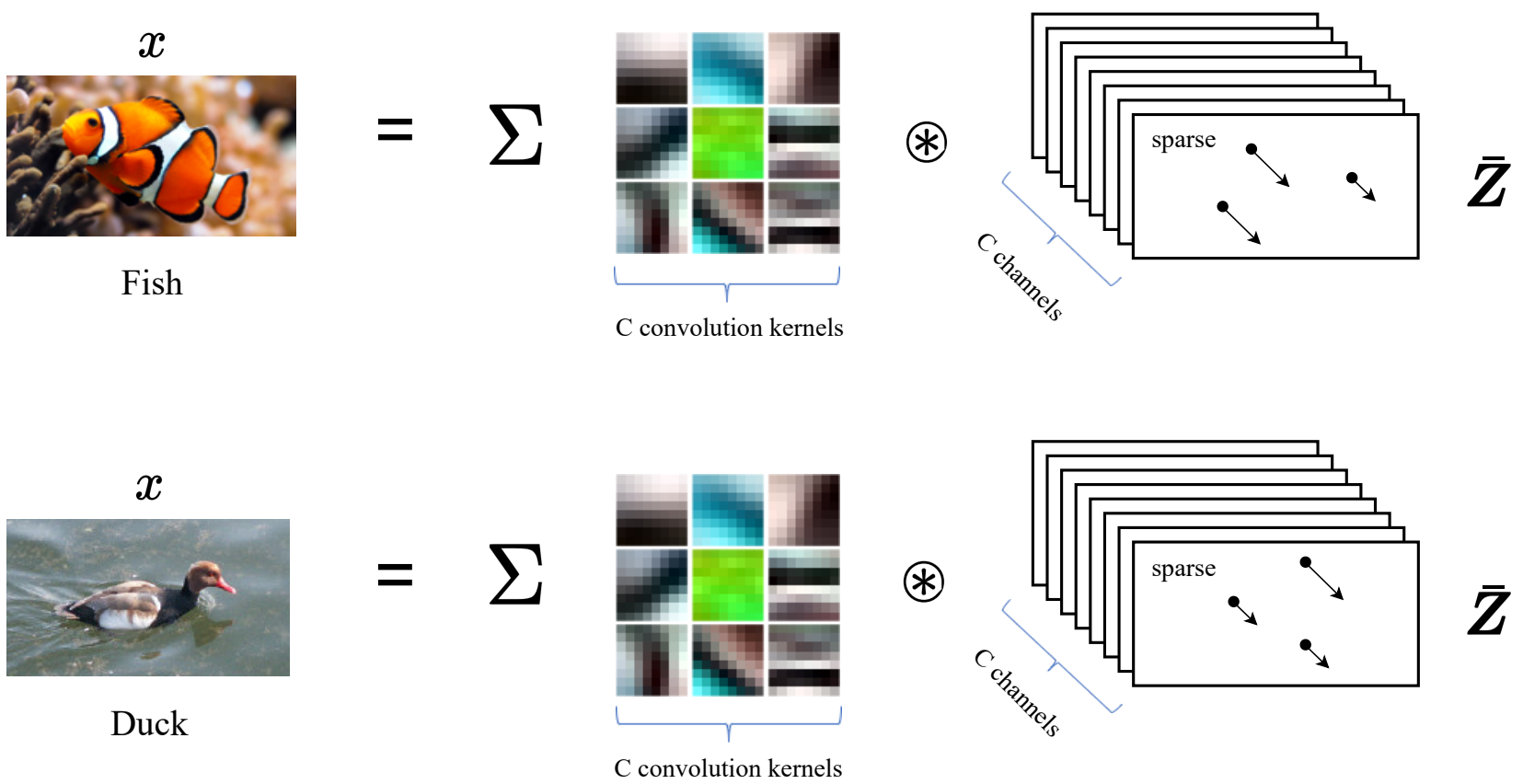

Figure 5.7: Each input signal $\boldsymbol{x}$ (an image here) can be represented as a superposition of sparse convolutions with multiple kernels $\boldsymbol{d}_c$ in a dictionary $\boldsymbol{D}$.

$$\bar{\boldsymbol{\Sigma}}(\bar{\boldsymbol{z}}) \doteq \begin{bmatrix} \mathsf{circ}(\boldsymbol{z}[1]) \\ \vdots \\ \mathsf{circ}(\boldsymbol{z}[C]) \end{bmatrix} \left[\mathsf{circ}(\boldsymbol{z}[1])^\top, \ldots, \mathsf{circ}(\boldsymbol{z}[C])^\top\right] \in \mathbb{R}^{dC \times dC}, \tag{5.1.28}$$

where $\mathsf{circ}(\boldsymbol{z}[c]) \in \mathbb{R}^{d \times d}$ with $c \in [C]$ is the circulant version of the $c$-th channel of the feature $\bar{\boldsymbol{z}}$. Then the columns of $\mathsf{circ}(\bar{\boldsymbol{z}})$ will only span at most a $d$-dimensional proper subspace in $\mathbb{R}^{dC}$. However, this simple lifting operation (if linear) is not sufficient to render the classes separable yet—features associated with other classes will span the *same* $d$-dimensional subspace. This reflects a fundamental conflict between invariance and linear (subspace) modeling: *one cannot hope for arbitrarily shifted and superposed signals to belong to the same class.*

One way of resolving this conflict is to leverage additional structure within each class, in the form of *sparsity*: signals within each class are not generated as an arbitrary linear superposition of some base atoms (or motifs), but only *sparse* combinations of them and their shifted versions, as shown in Figure 5.7. More precisely, let $\boldsymbol{D}_k = [\boldsymbol{d}_{k,1}, \ldots, \boldsymbol{d}_{k,c}]$ denote a matrix with a collection of atoms associated for class $k$, also known as a dictionary, then each signal $\boldsymbol{x}$ in this class is sparsely generated as:

$$\boldsymbol{x} = \boldsymbol{d}_{k,1} \circledast z_1 + \ldots + \boldsymbol{d}_{k,c} \circledast z_c = \mathsf{circ}(\boldsymbol{D}_k)\boldsymbol{z}, \tag{5.1.29}$$

for some sparse vector $\boldsymbol{z}$. Signals in different classes are then generated by different dictionaries whose atoms (or motifs) are incoherent from one another. Due to incoherence, signals in one class are unlikely to be sparsely represented by atoms in any other class. Hence all signals can be represented as

$$\boldsymbol{x} = \left[\mathsf{circ}(\boldsymbol{D}_1), \mathsf{circ}(\boldsymbol{D}_2), \ldots, \mathsf{circ}(\boldsymbol{D}_K)\right]\bar{\boldsymbol{z}}, \tag{5.1.30}$$

where $\bar{\boldsymbol{z}}$ is sparse.[12] There is a vast literature on how to learn the most compact and optimal sparsifying dictionaries from sample data, e.g. [LB19; QLZ19] and subsequently solve the inverse problem and compute the associated sparse code $\boldsymbol{z}$ or $\bar{\boldsymbol{z}}$. Recent studies of [QLZ20; QZL+20a] even show that under broad conditions the convolution dictionary learning problem can be solved effectively and efficiently.

Nevertheless, for tasks such as classification, we are not necessarily interested in the precise optimal dictionary nor the precise sparse code for each individual signal. We are mainly interested if collectively the set of sparse codes for each class are adequately separable from those of other classes. Under the assumption of the sparse generative model, if the convolution kernels $\{\boldsymbol{k}_c\}_{c=1}^C$ match well with the "transpose" or "inverse" of the above sparsifying dictionaries $\boldsymbol{D} = [\boldsymbol{D}_1, \ldots, \boldsymbol{D}_K]$, also known as the *analysis filters* [NDE+13; RE14], signals in one class will only have high responses to a small subset of those filters and low responses to others (due to the incoherence assumption). Nevertheless, in practice, often a sufficiently large number of, say $C$, random filters $\{\boldsymbol{k}_c\}_{c=1}^C$ suffices to ensure that the extracted $C$-channel features

$$\big[\boldsymbol{k}_1 \circledast \boldsymbol{x}, \boldsymbol{k}_2 \circledast \boldsymbol{x}, \ldots, \boldsymbol{k}_C \circledast \boldsymbol{x}\big]^\top = \big[\mathsf{circ}(\boldsymbol{k}_1)\boldsymbol{x}, \ldots, \mathsf{circ}(\boldsymbol{k}_C)\boldsymbol{x}\big]^\top \in \mathbb{R}^{C\times d} \quad (5.1.31)$$

for different classes have different response patterns to different filters hence make different classes separable [CJG+15].

Therefore, in our framework, to a large extent the number of channels (or the width of the network) truly plays the role as the *statistical resource* whereas the number of layers (the depth of the network) plays the role as the *computational resource*. The theory of compressive sensing precisely characterizes how many measurements are needed in order to preserve the intrinsic low-dimensional structures (including separability) of the data [WM22].

The multi-channel responses $\bar{\boldsymbol{z}}$ should be sparse. So to approximate the sparse code $\bar{\boldsymbol{z}}$, we may take an entry-wise *sparsity-promoting nonlinear thresholding*, say $\boldsymbol{\tau}(\cdot)$, on the above filter outputs by setting low (say absolute value below $\epsilon$) or negative responses to be zero:

$$\bar{\boldsymbol{z}} \doteq \boldsymbol{\tau}\Big(\big[\mathsf{circ}(\boldsymbol{k}_1)\boldsymbol{x}, \ldots, \mathsf{circ}(\boldsymbol{k}_C)\boldsymbol{x}\big]^\top\Big) \in \mathbb{R}^{C\times d}. \quad (5.1.32)$$

Figure 5.8 illustrates the basic ideas. One may refer to [RE14] for a more systematical study on the design of the sparsifying thresholding operator. Nevertheless, here we are not so interested in obtaining the best sparse codes as long as the codes are sufficiently separable. Hence the nonlinear operator $\boldsymbol{\tau}$ can be simply chosen to be a soft thresholding or a ReLU. These presumably sparse features $\bar{\boldsymbol{z}}$ can be assumed to lie on a lower-dimensional (nonlinear) submanifold of $\mathbb{R}^{dC}$, which can be linearized and separated from the other classes by subsequent ReduNet layers, as illustrated later in Figure 5.9.

[12] Notice that similar sparse representation models have long been proposed and used for classification purposes in applications such a face recognition, demonstrating excellent effectiveness [WWG+12; WYG+09]. Recently, the convolution sparse coding model has been proposed by [PRE17] as a framework for interpreting the structures of deep convolution networks.

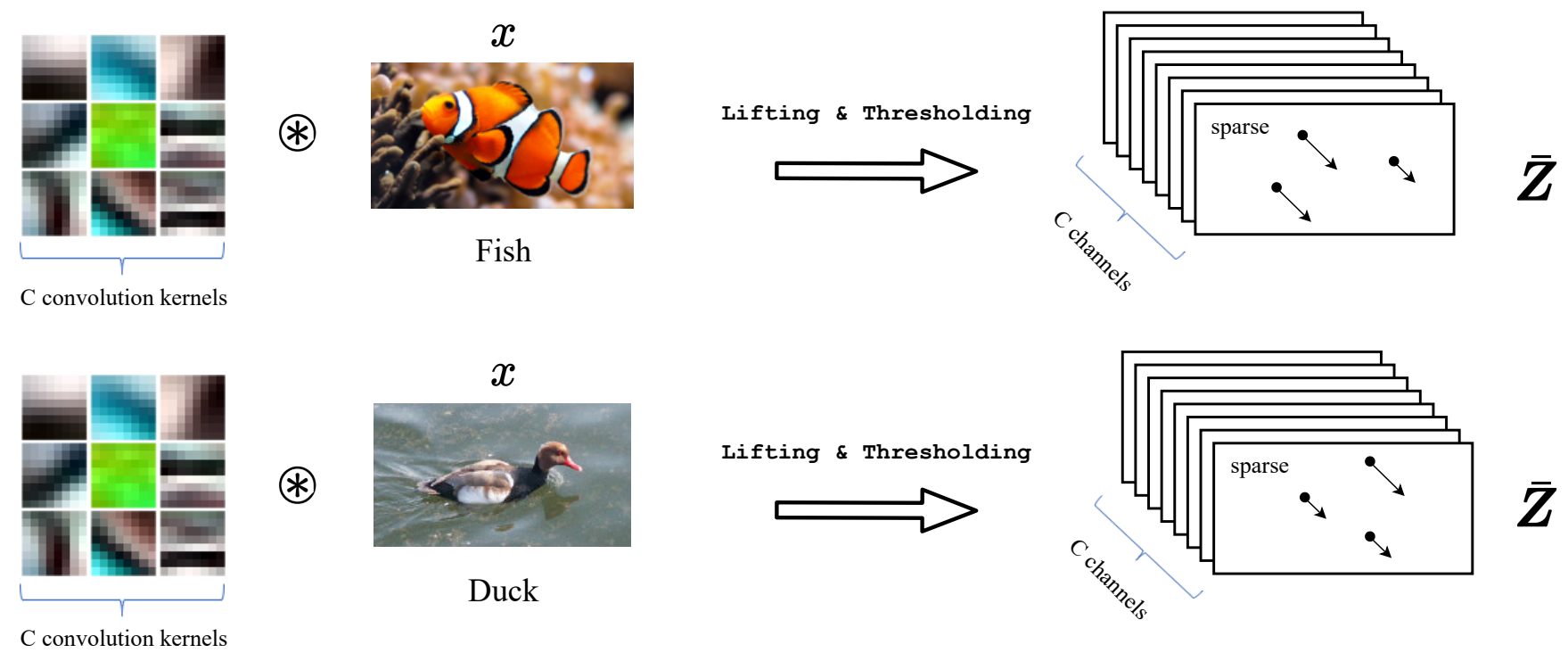

Figure 5.8: Estimate the sparse code $\bar{\boldsymbol{z}}$ of an input signal $\boldsymbol{x}$ (an image here) by taking convolutions with multiple kernels $\boldsymbol{k}_c$ and then sparsifying.

The ReduNet constructed from circulant version of these multi-channel features $\bar{\boldsymbol{Z}} \doteq [\bar{\boldsymbol{z}}_1, \dots, \bar{\boldsymbol{z}}_N] \in \mathbb{R}^{C \times d \times N}$, i.e., $\mathsf{circ}(\bar{\boldsymbol{Z}}) \doteq [\mathsf{circ}(\bar{\boldsymbol{z}}_1), \dots, \mathsf{circ}(\bar{\boldsymbol{z}}_N)] \in \mathbb{R}^{dC \times dN}$, retains the good invariance properties described above: the linear operators, now denoted as $\bar{\boldsymbol{E}}$ and $\bar{\boldsymbol{C}}_k$, remain block circulant, and represent *multi-channel 1D circular convolutions.* Specifically, we have the following result.

**Proposition 5.2** (Multi-channel convolution structures of $\bar{\boldsymbol{E}}$ and $\bar{\boldsymbol{C}}_k$)**.** *The matrix*

$$\bar{\boldsymbol{E}} \doteq \alpha \left(\boldsymbol{I} + \alpha\, \mathsf{circ}(\bar{\boldsymbol{Z}}) \mathsf{circ}(\bar{\boldsymbol{Z}})^\top\right)^{-1} \tag{5.1.33}$$

*is block circulant, i.e.,*

$$\bar{\boldsymbol{E}} = \begin{bmatrix} \bar{\boldsymbol{E}}_{1,1} & \cdots & \bar{\boldsymbol{E}}_{1,C} \\ \vdots & \ddots & \vdots \\ \bar{\boldsymbol{E}}_{C,1} & \cdots & \bar{\boldsymbol{E}}_{C,C} \end{bmatrix} \in \mathbb{R}^{dC \times dC}, \tag{5.1.34}$$

*where each $\bar{\boldsymbol{E}}_{c,c'} \in \mathbb{R}^{d \times d}$ is a circulant matrix. Moreover, $\bar{\boldsymbol{E}}$ represents a multi-channel circular convolution, i.e., for any multi-channel signal $\bar{\boldsymbol{z}} \in \mathbb{R}^{C \times n}$ we have*

$$\bar{\boldsymbol{E}} \cdot \mathsf{vec}(\bar{\boldsymbol{z}}) = \mathsf{vec}(\bar{\boldsymbol{e}} \circledast \bar{\boldsymbol{z}}). \tag{5.1.35}$$

*In above, $\bar{\boldsymbol{e}} \in \mathbb{R}^{C \times C \times d}$ is a multi-channel convolutional kernel with $\bar{\boldsymbol{e}}[c, c'] \in \mathbb{R}^d$ being the first column vector of $\bar{\boldsymbol{E}}_{c,c'}$, and $\bar{\boldsymbol{e}} \circledast \bar{\boldsymbol{z}} \in \mathbb{R}^{C \times d}$ is the multi-channel circular convolution defined as*

$$(\bar{\boldsymbol{e}} \circledast \bar{\boldsymbol{z}})[c] \doteq \sum_{c'=1}^{C} \bar{\boldsymbol{e}}[c, c'] \circledast \bar{\boldsymbol{z}}[c'], \quad \forall c = 1, \dots, C. \tag{5.1.36}$$

*Similarly, the matrices $\bar{\boldsymbol{C}}_k$ associated with any subsets of $\bar{\boldsymbol{Z}}$ are also multi-channel circular convolutions.*

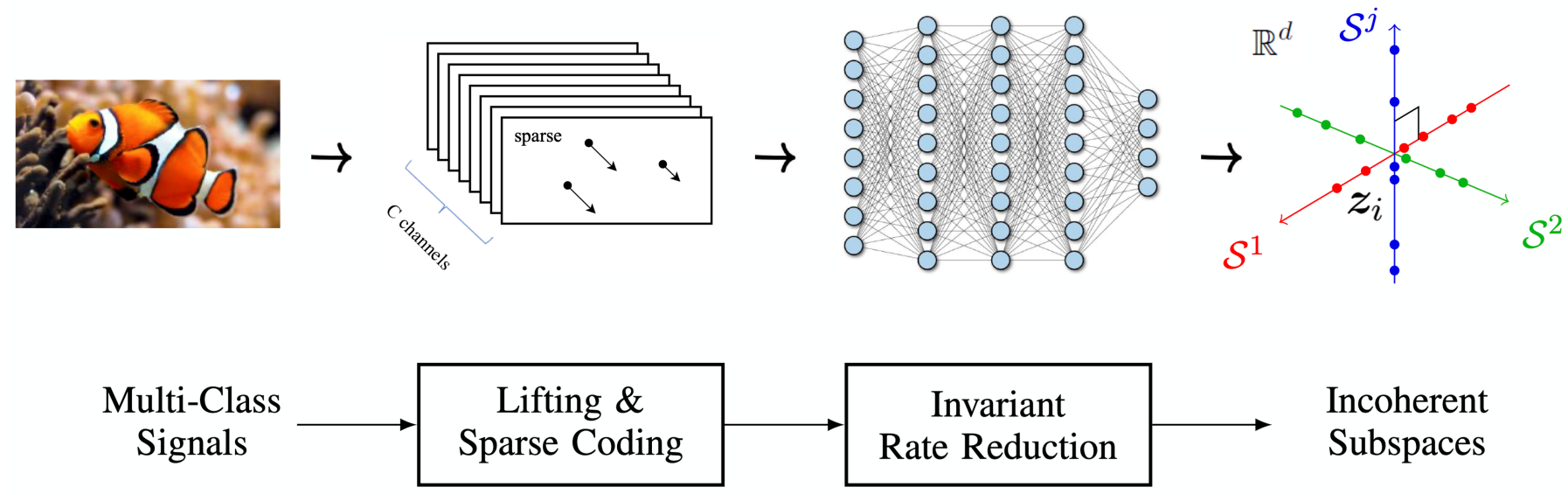

Figure 5.9: The overall process for classifying multi-class signals with shift invariance: Multi-channel lifting, sparse coding, followed by a multi-channel convolution ReduNet for invariant rate reduction. These components are *necessary* in order to map shift-invariant multi-class signals to incoherent (linear) subspaces as an LDR. Note that the architectures of most modern deep neural networks resemble this process. The so-learned LDR facilitates subsequent tasks such as classification.

From Proposition 5.2, shift invariant ReduNet is a deep convolutional network for multi-channel 1D signals by construction. Notice that even if the initial lifting kernels are separated (5.1.32), the matrix inverse in (5.1.33) for computing $\bar{\boldsymbol{E}}$ (similarly for $\bar{\boldsymbol{C}}_k$) introduces "cross talk" among all $C$ channels. Hence, these multi-channel convolutions in general are *not* depth-wise separable, unlike the Xception nets [Cho17] that were once suggested to simplify multi-channel convolutional neural networks.[13]

*Remark* 5.2 (Reducing Computational Complexity in the Frequency Domain). The calculation of $\bar{\boldsymbol{E}}$ in (5.1.33) requires inverting a matrix of size $dC \times dC$, which in general has complexity $O(d^3C^3)$. Nevertheless, by using the fact that a circulant matrix can be diagonalized by the Discrete Fourier Transform (DFT) matrix, the complexity can be significantly reduced. As shown in [CYY+22], to compute $\bar{\boldsymbol{E}}$ and $\bar{\boldsymbol{C}}_k \in \mathbb{R}^{dC \times dC}$, we only need to compute in the frequency domain the inverse of $C \times C$ blocks for $d$ times hence the overall complexity becomes $O(dC^3)$.

**Overall network architecture and comparison.** Following the above derivation, we see that in order to find a linear discriminative representation (LDR) for multiple classes of signals/images that is invariant to translation, sparse coding, a multi-layer architecture with multi-channel convolutions, different nonlinear activation, and spectrum computing all become *necessary* components for achieving the objective effectively and efficiently. Figure 5.9 illustrates the overall process of learning such a representation via invariant rate reduction on the input sparse codes.

[13] It remains open what additional structures on the data would lead to depth-wise separable convolutions.

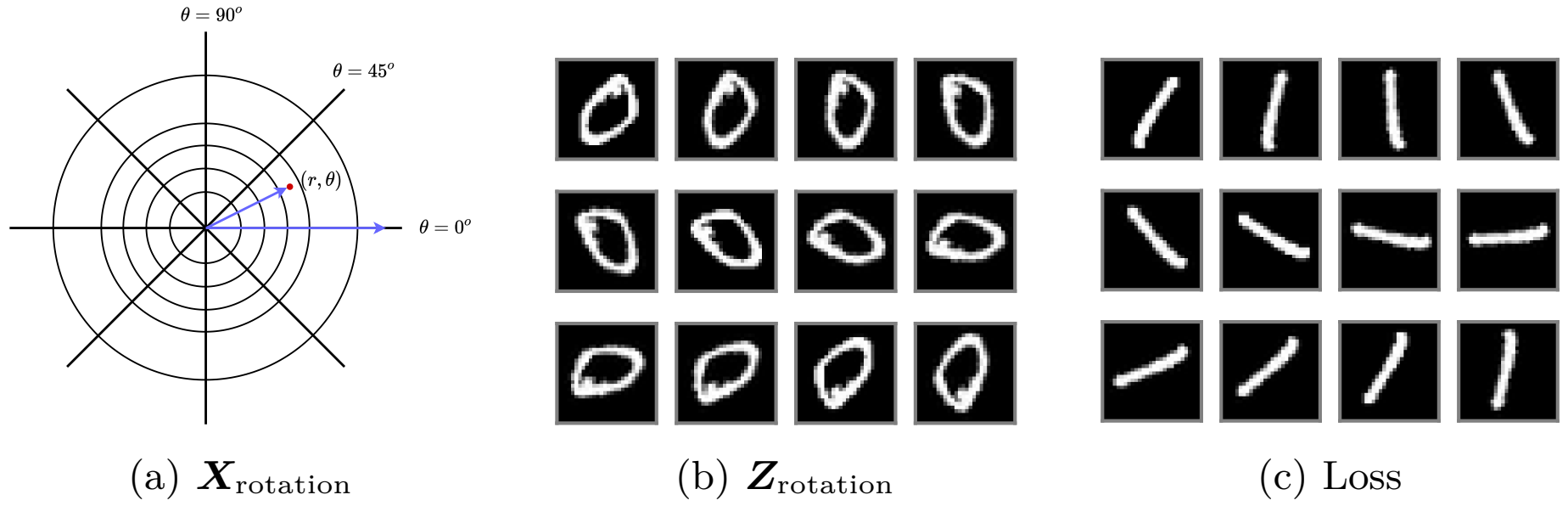

(a) $\boldsymbol{X}_{\text{rotation}}$ (b) $\boldsymbol{Z}_{\text{rotation}}$ (c) Loss

Figure 5.10: Examples of rotated images of MNIST digits, each by 18°. (**Left**) Diagram for polar coordinate representation; (**Right**) Rotated images of digit '0' and '1'.

*Example* 5.2 (Invariant Classification of Digits). We next provide an empirical performance of the ReduNet on learning *rotation* invariant features on the real 10-class MNIST dataset. We impose a polar grid on the image $\boldsymbol{x} \in \mathbb{R}^{H \times W}$, with its geometric center being the center of the 2D polar grid (as illustrated in Figure 5.10). For each radius $r_i$, $i \in [C]$, we can sample $\Gamma$ pixels with respect to each angle $\gamma_l = l \cdot (2\pi/\Gamma)$ with $l \in [\Gamma]$. Then given a sample image $\boldsymbol{x}$ from the dataset, we represent the image in the (sampled) polar coordinate as a multi-channel signal $\boldsymbol{x}_p \in \mathbb{R}^{\Gamma \times C}$. The goal here is to learn a rotation invariant representation, i.e., we expect to learn $f(\cdot, \theta)$ such that $\{f(\boldsymbol{x}_p \circ \mathfrak{g}, \theta)\}_{\mathfrak{g} \in \mathbb{G}}$ lie in the same subspace, where $\mathfrak{g}$ is the cyclic-shift in polar angle. We use $N = 100$ training samples (10 from each class) and set $\Gamma = 200$, $C = 15$ for polar sampling. By performing the above sampling in polar coordinate, we can obtain the data matrix $\boldsymbol{X}_p \in \mathbb{R}^{(\Gamma \cdot C) \times N}$. For the ReduNet, we set the number of layers/iterations $L = 40$, precision $\epsilon = 0.1$, step size $\eta = 0.5$. Before the first layer, we perform lifting of the input by 1D circulant-convolution with 20 random Gaussian kernels of size 5.

To evaluate the learned representation, each training sample is augmented by 20 of its rotated version, each shifted with stride=10. We compute the cosine similarities among the $m \times 20$ augmented training inputs $\boldsymbol{X}_{\text{rotation}}$ and the results are shown in Figure 5.11 (**a**). We compare the cosine similarities among the learned features of all the augmented versions, i.e., $\bar{\boldsymbol{Z}}_{\text{rotation}}$ and summarize the results in Figure 5.11 (**b**). As we see, the so-constructed rotation-invariant ReduNet is able to map the training data (as well as all its rotated versions) from the 10 different classes into 10 nearly orthogonal subspaces. That is, the learned subspaces are truly invariant to shift transformation in polar angle. Next, we randomly draw another 100 test samples followed by the same augmentation procedure. In Figure 5.11 (**c**), we visualize the MCR$^2$ loss on the $\ell$-th layer representation of the ReduNet on the training and test dataset. From these results, we can find that the constructed ReduNet is indeed able to maximize the MCR$^2$ loss as well as generalize to the test data.

■

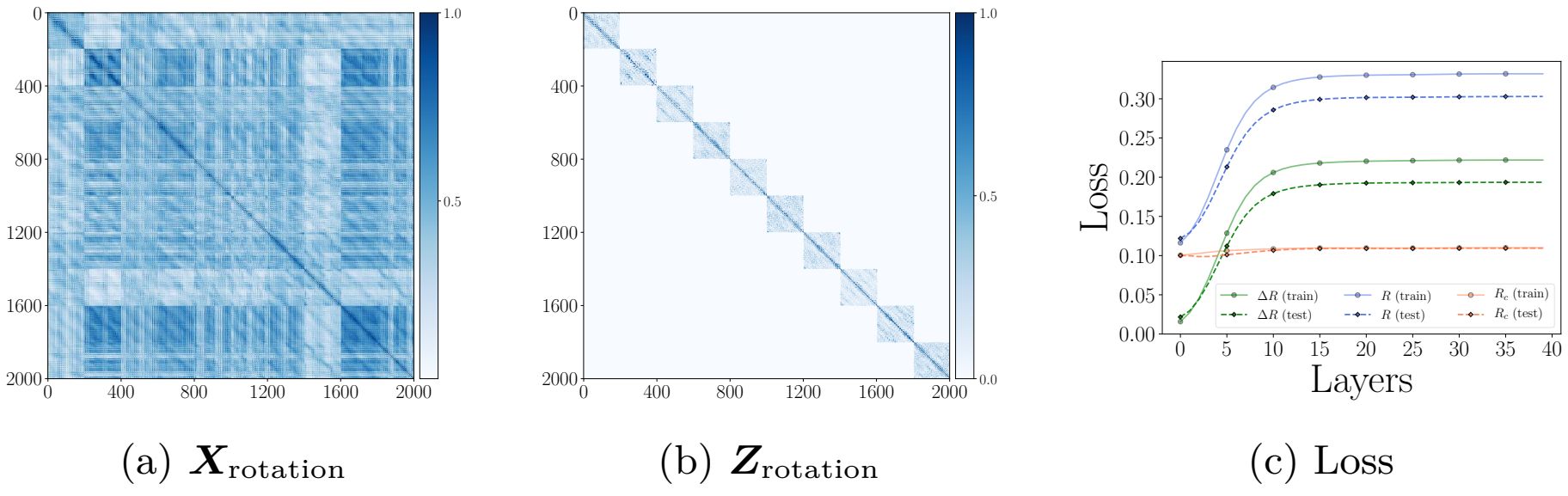

(a) $\boldsymbol{X}_{\text{rotation}}$ (b) $\boldsymbol{Z}_{\text{rotation}}$ (c) Loss

Figure 5.11: (a)(b) are heatmaps of cosine similarity among rotated training data $\boldsymbol{X}_{\text{rotation}}$ and learned features $\bar{\boldsymbol{Z}}_{\text{rotation}}$ for rotation invariance. (d) visualizes the training/val MCR$^2$ losses across layers.

## 5.2 White-Box Transformers from Unrolled Optimization

As we have seen in the previous section, we use the problem of classification to provide a rigorous interpretation for main architectural characteristics of popular deep networks such as the ResNet and the CNN: each layer of such networks can be viewed as to imitate a gradient step which increases the rate reduction (or information gain) objective. This perspective also leads to a somewhat surprising fact: the the parameters and operators of the layers of such a deep network, the ReduNet, can be computed in a purely forward fashion.

Despite the theoretical and conceptual importance of the ReduNet, several factors limit it from being very practical. First, as we have discussed in the above, the computational cost of computing the matrix operators in each layer in a forward fashion can be very high. Second, the so-computed operators may not be so effective in optimizing the objective and it might take thousands of iterations (hence layers). As we have seen in Section 2.3.3 for LISTA, these two issues can be addressed by allowing to optimize those operators and make them learnable via back-propagation.[14]

The supervised classification setting in which the ReduNet was derived is also somewhat limiting. In practice, an image might not belong to a single class as it may contain multiple objects. Hence it would be more general to assume that different regions of the image belong to different low-dimensional models (say a Gaussian or a subspace). As we will see, such a generalization would lead to a both simple and general architecture which unifies the rate reduction and the denoising operations that we have seen in the previous chapter. Moreover, the so-obtained architecture resembles the popular Transformer architecture.

[14]Or, perhaps, by a mixture of both forward and backward optimization.

### 5.2.1 Unrolled Optimization for Sparse Rate Reduction

For the past several years, as the amount and quality of data available for training deep networks has increased dramatically, the field has moved from considering inputs as “atoms” to inputs as “sequences of tokens”, where each token is a part of the overall input. This shift in perspective has both philosophical and pragmatic improvements: philosophically, it allows us to measure and model each part of the input and their interactions; pragmatically, it allows us to use similar deep networks, *sequence-to-sequence models* such as the ubiquitous Transformers, to learn every kind of data distribution. Now, it falls to us to understand representation learning in the “token-centric” model of data as token sequences, and to cope with the aforementioned challenges (such as, in the context of the rate reduction framework, not having labels for each token).

Formally, let $\boldsymbol{X} = \left[\boldsymbol{x}_1, \ldots, \boldsymbol{x}_N\right] \in \mathbb{R}^{D \times N}$ denote random variables representing our data source. In vision tasks, each $\boldsymbol{x}_i \in \mathbb{R}^D$ is interpreted as a *token*, typically corresponding to an image patch. In language tasks, each $\boldsymbol{x}_i \in \mathbb{R}^D$ is interpreted as an *token embedding*, i.e., a continuous vector representation of a discrete token such as a word or subword.[15] The $\boldsymbol{x}_i$'s may have arbitrary correlation structures. We use $\boldsymbol{Z} = \left[\boldsymbol{z}_1, \ldots, \boldsymbol{z}_N\right] \in \mathbb{R}^{d \times N}$ to denote the random variables that defines our representations, where $\boldsymbol{z}_i \in \mathbb{R}^d$ is the representation of the corresponding token $\boldsymbol{x}_i \in \mathbb{R}^D$.

*Remark* 5.3. In transformers, each input sample is typically converted into a sequence of *tokens*. A token is a basic unit of information derived from the raw input: in natural language processing, tokens are typically words or subwords; in computer vision, they correspond to image patches; and in other modalities, they may represent time steps, spatial locations, or other domain-specific units. A *token embedding* is a continuous vector representation of a token that serves as the input to a transformer. It maps each token to a point in a high-dimensional space, enabling the model to process symbolic inputs using numerical computation. A *token representation* is a vector that encodes the semantic or structural information of a token, typically produced by the intermediate or final layers of a transformer. These representations are designed to capture meaningful features of the input that are useful for downstream tasks such as classification, generation, or regression. Please refer to Section 8.2 for more details about this vocabulary and how it relates to implementations.

**Objective for learning a structured and compact representation.** Inspired by the previous discussion of rate reduction (Section 5.1), we propose that the objective of representation learning in this token-centric model for our data is to find a feature mapping $f\colon \boldsymbol{X} \in \mathbb{R}^{D \times N} \to \boldsymbol{Z} \in \mathbb{R}^{d \times N}$ which transforms input tokens $\{\boldsymbol{x}_i\}_{i=1}^N \subset \mathbb{R}^D$ with a potentially nonlinear and multi-modal distribution to a (piecewise) *linearized and compact* token representations $\{\boldsymbol{z}_i\}_{i=1}^N \subset \mathbb{R}^d$. While the *joint* distribution of tokens representations $\{\boldsymbol{z}_i\}_{i=1}^N$ may be sophisticated (and task-specific), we further contend that it is reasonable and practical to require that the target *marginal* distribution of individual token representations should be highly compressed and structured, amenable for compact coding. Particularly, we require the distribution to be *a mixture of low-dimensional (say $K$) Gaussian distributions*, such that the $k$-th Gaussian has mean $\mathbf{0} \in \mathbb{R}^d$, covariance $\boldsymbol{\Sigma}_k \succeq \mathbf{0} \in \mathbb{R}^{d\times d}$, and support spanned by the orthonormal basis $\boldsymbol{U}_k \in \mathbb{R}^{d\times p}$. We denote $\boldsymbol{U}_{[K]} = \{\boldsymbol{U}_k\}_{k=1}^K$ to be the set of bases of all Gaussians.

[15] With a slight abuse of terminology, we refer to both the discrete tokens and their associated embeddings simply as tokens throughout this chapter for convenience.

As usual, in order to learn good representations, we wish to maximize the *information gain* for the final token representations, or more specifically to maximize an appropriate notion of the rate reduction (see Section 4.2.3), i.e.,

$$\max_{\boldsymbol{Z}\in\mathbb{R}^{d\times N}} \ \Delta R_\epsilon(\boldsymbol{Z} \mid \boldsymbol{U}_{[K]}) \doteq R_\epsilon(\boldsymbol{Z}) - R^c_\epsilon(\boldsymbol{Z} \mid \boldsymbol{U}_{[K]}). \tag{5.2.1}$$

Here, the first term $R_\epsilon$ is an estimate of the lossy coding rate for the whole set of token representations. More specifically, if we view the token representations $\{\boldsymbol{z}_i\}_{i=1}^N$ as i.i.d. samples from a single zero-mean Gaussian, their lossy coding rate subject to a quantization precision $\epsilon > 0$ is given as

$$R_\epsilon(\boldsymbol{Z}) \doteq \frac{1}{2}\mathrm{logdet}\left(\boldsymbol{I} + \frac{d}{N\epsilon^2}\boldsymbol{Z}^\top\boldsymbol{Z}\right) = \frac{1}{2}\mathrm{logdet}\left(\boldsymbol{I} + \frac{d}{N\epsilon^2}\boldsymbol{Z}\boldsymbol{Z}^\top\right). \tag{5.2.2}$$

The second term $R^c_\epsilon$ is an estimate of the lossy coding rate under the codebook $\boldsymbol{U}_{[K]}$, which is given as

$$R^c_\epsilon(\boldsymbol{Z} \mid \boldsymbol{U}_{[K]}) \doteq \sum_{k=1}^K R_\epsilon(\boldsymbol{U}_k^\top\boldsymbol{Z}) = \frac{1}{2}\sum_{k=1}^K \log\det\left(\boldsymbol{I} + \frac{p}{N\epsilon^2}(\boldsymbol{U}_k^\top\boldsymbol{Z})^\top(\boldsymbol{U}_k^\top\boldsymbol{Z})\right). \tag{5.2.3}$$

*Remark* 5.4. The expression (5.2.3) for the coding rate can be viewed as a generalization of the coding rate $R^c_\epsilon$ used in the original rate reduction objective (4.2.16). In particular, the original objective is defined with respect to a set of known membership labels $\{\boldsymbol{\Pi}_k\}$ specific to the particular data realization $\boldsymbol{X}$. In contrast, the current objective is defined with respect to subspaces $\boldsymbol{U}_{[K]}$, which are independent of any particular realization but are assumed to support the distribution of token representations. Suppose that a token representation $\boldsymbol{z}_i$ belongs to a subspace $\boldsymbol{U}_k$ and these subspaces are approximately orthogonal to each other, i.e., $\boldsymbol{U}_k^\top\boldsymbol{U}_l \approx \mathbf{0}$ for all $k \neq l$. Then, one can verify that the projections $\boldsymbol{U}_k\boldsymbol{U}_k^\top\boldsymbol{z}_i = \boldsymbol{z}_i$ and $\boldsymbol{U}_l\boldsymbol{U}_l^\top\boldsymbol{z}_i \approx \mathbf{0}$ for all $l \neq k$. These orthogonal projections effectively serve as implicit membership labels, identifying the subspace to which each token representation belongs.

*Remark* 5.5. Notably, this set of desiderata for representation learning aligns well with two well-established empirical hypotheses about the structure of token representations in trained models which use the ubiquitous transformer neural network architecture: the "linear representation hypothesis" [JRR+24; PCV24] and the "superposition hypothesis" [EHO+22a; YCO+21]. The linear representation hypothesis posits that token representations in transformer models lie in low-dimensional linear subspaces that encode semantic features. Similarly, the superposition hypothesis suggests that these representations can be approximately expressed as a sparse linear combination of these feature vectors. As we will see in Figure 5.17 later in this Chapter, the subspace basis $\boldsymbol{U}_k$ may be interpreted as a set of semantic features, where each feature corresponds to a specific aspect of the token's meaning. Token representations are then approximately expressed as sparse linear combinations of these subspace bases, capturing the essential semantic components of the token while ignoring irrelevant dimensions.

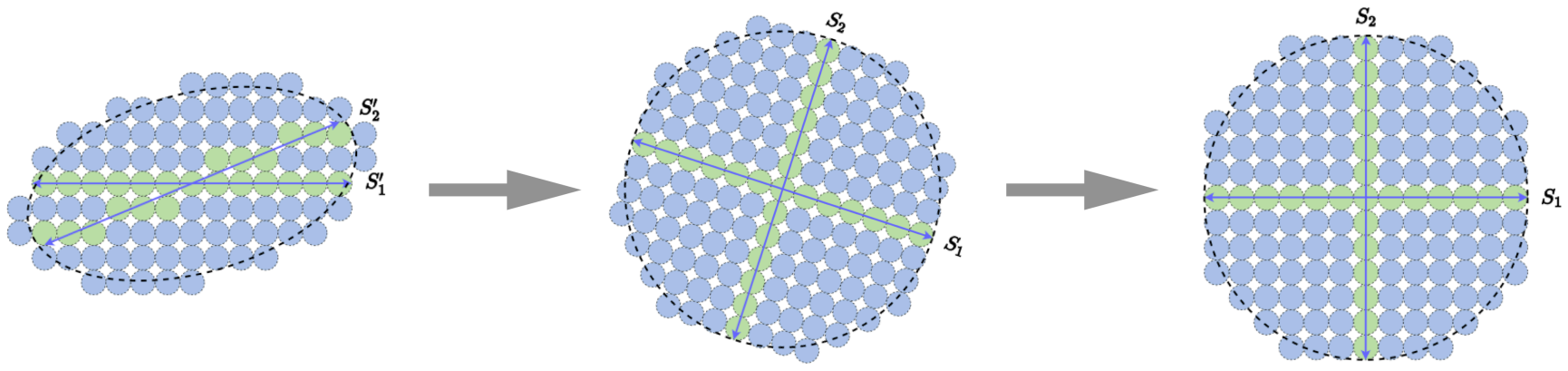

Figure 5.12: **Comparison of three sets of representations via rate reduction and sparsity.** Each $S_i$ represents one linear subspace, and the number of blue balls represents the difference between the coding rates $\Delta R_\epsilon(\boldsymbol{Z} \mid \boldsymbol{U}_{[K]}) = R_\epsilon(\boldsymbol{Z}) - R^c_\epsilon(\boldsymbol{Z} \mid \boldsymbol{U}_{[K]})$.

**Sparse rate reduction.** Note that the rate reduction objective (5.2.1) is invariant to arbitrary joint rotations of the representations and subspaces. In particular, optimizing the rate reduction objective may not naturally lead to axis-aligned (i.e., *sparse*) representations. For instance, consider the three sets of learned representations in Figure 5.12. The coding rate reduction increases from (a) to (b), but because it is invariant under rotations, remains the same from (b) to (c). Therefore, we would like to transform the representations (and their supporting subspaces) so that the representations $\boldsymbol{Z}$ eventually become sparse[16] with respect to the standard coordinates of the resulting representation space as in Figure 5.12(c). Computationally, we may combine the above two goals into a unified objective for optimization:

$$\max_{f \in \mathcal{F}} \left[\Delta R_\epsilon(\boldsymbol{Z} \mid \boldsymbol{U}_{[K]}) - \lambda \|\boldsymbol{Z}\|_0\right] \qquad \text{s.t. } \boldsymbol{Z} = f(\boldsymbol{X}), \tag{5.2.4}$$

where $\mathcal{F}$ denotes a general function class and the $\ell_0$ norm $\|\boldsymbol{Z}\|_0$ promotes the sparsity of the final token representations $\boldsymbol{Z} = f(\boldsymbol{X})$.

In practice, the $\ell_0$ norm is often relaxed to the $\ell_1$ norm to improve computational tractability and enable convex optimization techniques [WM22]. Motivated by this, we relax Problem (5.2.4) accordingly, leading to a formulation that remains faithful to the original sparsity objective while being more amenable to efficient algorithms as follow:

$$\max_{f \in \mathcal{F}} \left[\Delta R_\epsilon(\boldsymbol{Z} \mid \boldsymbol{U}_{[K]}) - \lambda \|\boldsymbol{Z}\|_1\right] \qquad \text{s.t. } \boldsymbol{Z} = f(\boldsymbol{X}), \tag{5.2.5}$$

With a slight abuse of terminology, we often refer to this objective function also as the *sparse rate reduction.*

[16]Concretely, having few nonzero entries.

**White-box network architecture via unrolled optimization.** Although easy to state, each term in the above objective is computationally challenging to optimize [WM22]. Hence it is natural to adopt an approximation approach that realizes the global transformation $f$ to optimize (5.2.4) through a concatenation of multiple, say $L$, simple *incremental and local* operations $f^\ell$ that push the representation distribution towards the desired parsimonious model distribution:

$$f\colon \boldsymbol{X} \xrightarrow{f^{\mathrm{pre}}} \boldsymbol{Z}^1 \to \cdots \to \boldsymbol{Z}^\ell \xrightarrow{f^\ell} \boldsymbol{Z}^{\ell+1} \to \cdots \xrightarrow{f^L} \boldsymbol{Z}^{L+1} = \boldsymbol{Z}, \tag{5.2.6}$$

where $f^{\mathrm{pre}} : \mathbb{R}^D \to \mathbb{R}^d$ is the pre-processing mapping that transforms each input token $\boldsymbol{x}_i \in \mathbb{R}^D$ to the initial token representations $\boldsymbol{z}_i^1 \in \mathbb{R}^d$. Each incremental *forward mapping* $\boldsymbol{Z}^{\ell+1} = f^\ell(\boldsymbol{Z}^\ell)$, or a "layer", transforms the token distribution to *optimize* the above sparse rate reduction objective (5.2.4), conditioned on the distribution of its input $\boldsymbol{Z}^\ell$.

*Remark* 5.6. In contrast to other unrolled optimization approaches such as the ReduNet (see Section 5.1), we *explicitly model* the distribution of $\boldsymbol{Z}^\ell$ at each layer, say as a mixture of linear subspaces or sparsely generated from a dictionary. The model parameters are learned from data (say via *backward propagation* with end-to-end training). This separation between forward "optimization" and backward "learning" clarifies the mathematical role of each layer as an operator that transforms the distribution of its input, whereas the input distribution is in turn modeled (and subsequently learned) by the parameters of the layer.

Now, we show how to derive these incremental and local operations through an unrolled optimization perspective to solve Problem (5.2.5). Once we decide on using an incremental approach to optimizing Problem (5.2.5), there are a variety of possible choices to achieve the optimization. Given a model for $\boldsymbol{Z}^\ell$, say a mixture of subspaces $\boldsymbol{U}_{[K]}$, we opt for a two-step *alternating minimization* method with a strong conceptual basis. First, we *compress* the tokens $\boldsymbol{Z}^\ell$ via a gradient descent to minimize the coding rate term $R_\epsilon^c(\boldsymbol{Z} \mid \boldsymbol{U}_{[K]}^\ell)$. Specifically, we take a gradient step on $R_\epsilon^c$ with a learning rate $\kappa$ as follows:

$$\boldsymbol{Z}^{\ell+1/2} = \boldsymbol{Z}^\ell - \kappa \nabla_{\boldsymbol{Z}} R_\epsilon^c(\boldsymbol{Z} \mid \boldsymbol{U}_{[K]}^\ell). \tag{5.2.7}$$

Next, we *sparsify* the compressed tokens, generating $\boldsymbol{Z}^{\ell+1}$ via a suitably-relaxed proximal gradient step to minimize the remaining term $\lambda\|\boldsymbol{Z}\|_1 - R_\epsilon(\boldsymbol{Z})$. As we will argue in detail later, we can find such a $\boldsymbol{Z}^{\ell+1}$ by solving a sparse presentation problem with respect to an orthogonal complete dictionary $\boldsymbol{D}^\ell$:

$$\boldsymbol{Z}^{\ell+1} = \arg\min_{\boldsymbol{Z}} \left\{ \lambda\|\boldsymbol{Z}\|_1 + \frac{1}{2}\|\boldsymbol{Z}^{\ell+1/2} - \boldsymbol{D}^\ell \boldsymbol{Z}\|_F^2 \right\}. \tag{5.2.8}$$

In the following, we provide technical details for each of the two steps above and derive efficient updates for their implementation.

**Self-attention as gradient descent on coding rate of token representations.** For the first step (5.2.7), the gradient of the coding rate $\nabla_{\boldsymbol{Z}} R_\epsilon^c$ is costly to compute, as it involves $K$ separate matrix inverses, one for each of the $K$ subspaces with basis $\boldsymbol{U}_k^\ell$:

$$\nabla_{\boldsymbol{Z}} R_\epsilon^c(\boldsymbol{Z} \mid \boldsymbol{U}_{[K]}) = \frac{p}{N\epsilon^2} \sum_{k=1}^{K} \boldsymbol{U}_k \boldsymbol{U}_k^\top \boldsymbol{Z} \Big( \boldsymbol{I} + \frac{p}{N\epsilon^2} (\boldsymbol{U}_k^\top \boldsymbol{Z})^\top (\boldsymbol{U}_k^\top \boldsymbol{Z}) \Big)^{-1}. \quad (5.2.9)$$

Now, we demonstrate that this gradient can be naturally approximated using a so-called multi-head subspace self-attention (MSSA) operator, which has a similar functional form to the multi-head self-attention operator [VSP+17b] with $K$ heads (i.e., one for each subspace, coming from each matrix inverse). Here, we approximate the gradient (5.2.9) using the first-order Neumann series (see Exercise 5.2):[17]

$$\nabla_{\boldsymbol{Z}} R_\epsilon^c(\boldsymbol{Z} \mid \boldsymbol{U}_{[K]}) \quad (5.2.10)$$

$$\approx \frac{p}{N\epsilon^2} \sum_{k=1}^{K} \boldsymbol{U}_k \boldsymbol{U}_k^\top \boldsymbol{Z} \left( \boldsymbol{I} - \frac{p}{N\epsilon^2} (\boldsymbol{U}_k^\top \boldsymbol{Z})^\top (\boldsymbol{U}_k^\top \boldsymbol{Z}) \right) \quad (5.2.11)$$

$$= \frac{p}{N\epsilon^2} \left( \sum_{k=1}^{K} \boldsymbol{U}_k \boldsymbol{U}_k^\top \right) \boldsymbol{Z} - \left( \frac{p}{N\epsilon^2} \right)^2 \sum_{k=1}^{K} \boldsymbol{U}_k (\boldsymbol{U}_k^\top \boldsymbol{Z}) (\boldsymbol{U}_k^\top \boldsymbol{Z})^\top (\boldsymbol{U}_k^\top \boldsymbol{Z}). \quad (5.2.12)$$

In the above approximation, we compute the similarity between projected token representations $\{\boldsymbol{U}_k^\top \boldsymbol{z}_i\}_{i=1}^n$ through an auto-correlation among the projected features as $(\boldsymbol{U}_k^\top \boldsymbol{Z})^\top (\boldsymbol{U}_k^\top \boldsymbol{Z})$. Even if each token feature within $\boldsymbol{Z}$ is normalized, the norm of this matrix can grow arbitrarily[18] as the sequence length increases, making our Neumann approximation increasingly poor (as the true gradient does not grow arbitrarily in size with sequence length). To fix this practical issue and generalize our formulation, we can instead use an arbitrary kernel to measure similarity. One particularly fitting kernel that aligns with our motivation of only auto-regressing against tokens which belong to the same subspace is the local (Nadaraya-Watson) kernel which involves membership estimation via softmax [CMP+21], namely computing (relative) similarity as $\mathrm{softmax}((\boldsymbol{U}_k^\top \boldsymbol{Z})^\top (\boldsymbol{U}_k^\top \boldsymbol{Z}))$. Now suppose that the subspaces expand "as much as possible", that is, all subspaces' bases $\boldsymbol{U}_{[K]}$ together span the whole space. Then, we have $\sum_{k=1}^{K} \boldsymbol{U}_k \boldsymbol{U}_k^\top = \boldsymbol{I}$. Hence, (5.2.10) becomes

$$\nabla_{\boldsymbol{Z}} R_\epsilon^c(\boldsymbol{Z} \mid \boldsymbol{U}_{[K]}) \approx \frac{p}{N\epsilon^2} \boldsymbol{Z} - \left( \frac{p}{N\epsilon^2} \right)^2 \mathrm{MSSA}(\boldsymbol{Z}^\ell \mid \boldsymbol{U}_{[K]}^\ell), \quad (5.2.13)$$

[17] This approximation is only meaningful when $\boldsymbol{Z}$ has a small norm, which in practice can be enforced by normalization strategies (such as the so-called "layer normalization"). We do not include it here for notational convenience, though the corresponding layer normalization operator is included in the final architecture.

[18] Given the practical choice of $\epsilon^2 = p/N$, i.e., a more precise code is used to accommodate more information from more samples.

where MSSA is defined through an SSA operator as follows:

$$\mathrm{SSA}(\boldsymbol{Z} \mid \boldsymbol{U}_k) \doteq (\boldsymbol{U}_k^\top \boldsymbol{Z})\,\mathrm{softmax}((\boldsymbol{U}_k^\top \boldsymbol{Z})^\top (\boldsymbol{U}_k^\top \boldsymbol{Z})),\ \forall k \in [K], \tag{5.2.14}$$

$$\mathrm{MSSA}(\boldsymbol{Z} \mid \boldsymbol{U}_{[K]}) \doteq \frac{p}{N\epsilon^2}\left[\boldsymbol{U}_1, \dots, \boldsymbol{U}_K\right] \begin{bmatrix} \mathrm{SSA}(\boldsymbol{Z} \mid \boldsymbol{U}_1) \\ \vdots \\ \mathrm{SSA}(\boldsymbol{Z} \mid \boldsymbol{U}_K) \end{bmatrix}. \tag{5.2.15}$$

Substituting (5.2.13) into (5.2.7) yields that it can naturally be approximated by

$$\boldsymbol{Z}^{\ell+1/2} = \left(1 - \frac{\kappa p}{N\epsilon^2}\right)\boldsymbol{Z}^\ell + \frac{\kappa p}{N\epsilon^2}\mathrm{MSSA}(\boldsymbol{Z}^\ell \mid \boldsymbol{U}_{[K]}^\ell). \tag{5.2.16}$$

*Remark* 5.7. The SSA operator in (5.2.14) resembles the *attention operator* in a typical transformer [VSP+17b], except that here the linear operators of value, key, and query are all set to be *the same* as the subspace basis, i.e., $\boldsymbol{V}_k = \boldsymbol{K}_k = \boldsymbol{Q}_k = \boldsymbol{U}_k^*$. Hence, we name $\mathrm{SSA}(\,\cdot \mid \boldsymbol{U}_k) : \mathbb{R}^{d\times n} \to \mathbb{R}^{p\times n}$ the **S**ubspace **S**elf-**A**ttention (SSA) operator. Then, the whole MSSA operator in (5.2.15), formally defined as $\mathrm{MSSA}(\,\cdot \mid \boldsymbol{U}_{[K]})\colon \mathbb{R}^{d\times n} \to \mathbb{R}^{d\times n}$ and called the **M**ulti-Head **S**ubspace **S**elf-**A**ttention (MSSA) operator, aggregates the attention head outputs by averaging using model-dependent weights, similar in concept to the popular multi-head self-attention operator in existing transformer networks. The overall gradient step (5.2.16) resembles the multi-head self-attention implemented with a skip connection in transformers.

**MLP as proximal gradient descent for sparse coding of token representations.** For the second step of alternating minimization, we need to minimize $\lambda\|\boldsymbol{Z}\|_1 - R_\epsilon(\boldsymbol{Z})$. Note that the gradient $\nabla R_\epsilon(\boldsymbol{Z})$ involves a matrix inverse, and thus naive proximal gradient (see Section A.1.3) to optimize this problem becomes intractable on large-scale problems. We therefore take a different approach to trading off between representational diversity and sparsification: we posit a (complete) incoherent or orthogonal dictionary $\boldsymbol{D}^\ell \in \mathbb{R}^{d\times d}$, and ask to sparsify the intermediate iterates $\boldsymbol{Z}^{\ell+1/2}$ with respect to $\boldsymbol{D}^\ell$. That is, $\boldsymbol{Z}^{\ell+1/2} \approx \boldsymbol{D}^\ell \boldsymbol{Z}^{\ell+1}$ where $\boldsymbol{Z}^{\ell+1}$ is more sparse; that is, it is a *sparse encoding* of $\boldsymbol{Z}^{\ell+1/2}$. The dictionary $\boldsymbol{D}^\ell$ is used to sparsify all tokens simultaneously. By the incoherence assumption, we have $(\boldsymbol{D}^\ell)^\top(\boldsymbol{D}^\ell) \approx \boldsymbol{I}$. Thus from (5.2.2) we have

$$R_\epsilon(\boldsymbol{Z}^{\ell+1/2}) \approx R_\epsilon(\boldsymbol{D}^\ell \boldsymbol{Z}^{\ell+1}) \approx R_\epsilon(\boldsymbol{Z}^{\ell+1}). \tag{5.2.17}$$

To solve $\lambda\|\boldsymbol{Z}\|_1 - R_\epsilon(\boldsymbol{Z})$, we optimize the following problem

$$\boldsymbol{Z}^{\ell+1} \approx \arg\min_{\boldsymbol{Z}} \|\boldsymbol{Z}\|_1 \quad \text{subject to} \quad \boldsymbol{Z}^{\ell+1/2} = \boldsymbol{D}^\ell \boldsymbol{Z}.$$

The above sparse representation program is usually solved by relaxing it to an unconstrained convex program, known as LASSO [WM22]:

$$\boldsymbol{Z}^{\ell+1} \approx \arg\min_{\boldsymbol{Z}} \left[\lambda\|\boldsymbol{Z}\|_1 + \frac{1}{2}\|\boldsymbol{Z}^{\ell+1/2} - \boldsymbol{D}^\ell \boldsymbol{Z}\|_F^2\right]. \tag{5.2.18}$$

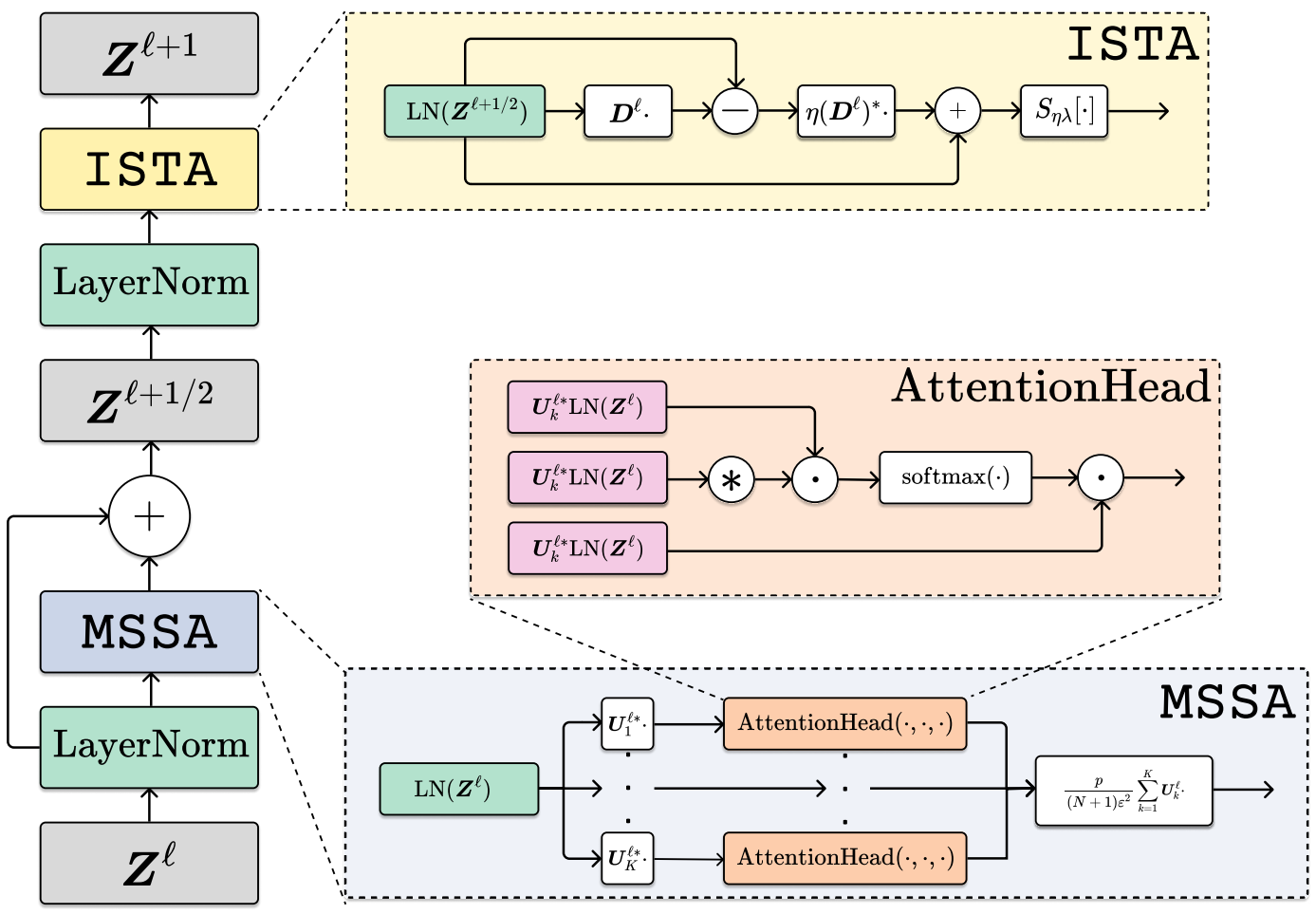

Figure 5.13: **One layer of the CRATE encoder architecture.** The full architecture is simply a concatenation of such layers, with some initial tokenizer, pre-processing head, and final task-specific head (i.e., a classification head).

In our implementation, we also add a non-negative constraint to $\boldsymbol{Z}^{\ell+1}$, and solve the corresponding non-negative LASSO:

$$\boldsymbol{Z}^{\ell+1} \approx \underset{\boldsymbol{Z}\geq\mathbf{0}}{\arg\min}\left[\lambda\|\boldsymbol{Z}\|_1 + \frac{1}{2}\|\boldsymbol{Z}^{\ell+1/2} - \boldsymbol{D}^{\ell}\boldsymbol{Z}\|_F^2\right]. \tag{5.2.19}$$

Then, we incrementally optimize Equation (5.2.19) by performing an unrolled *proximal gradient descent* step, known as an ISTA step [BT09], to give the update:

$$\boldsymbol{Z}^{\ell+1} = \mathrm{ISTA}(\boldsymbol{Z}^{\ell+1/2} \mid \boldsymbol{D}^{\ell}), \tag{5.2.20}$$

$$\text{where} \quad \mathrm{ISTA}(\boldsymbol{Z} \mid \boldsymbol{D}) \doteq \mathrm{ReLU}(\boldsymbol{Z} - \eta\boldsymbol{D}^{\top}(\boldsymbol{D}\boldsymbol{Z} - \boldsymbol{Z}) - \eta\lambda\mathbf{1}). \tag{5.2.21}$$

### 5.2.2 Overall White-Box Transformer Architecture: CRATE

We now design a white-box transformer architecture, named the Coding RATE Transformer (CRATE), by unrolling the above updates. By combining the above two steps (5.2.16) and (5.2.20):

1. Local compression of tokens within a sample towards a mixture-of-subspace structure, leading to the multi-head subspace self-attention block – MSSA;

2. Global sparsification of token sets across all samples through sparse coding, leading to the sparsification block – ISTA;

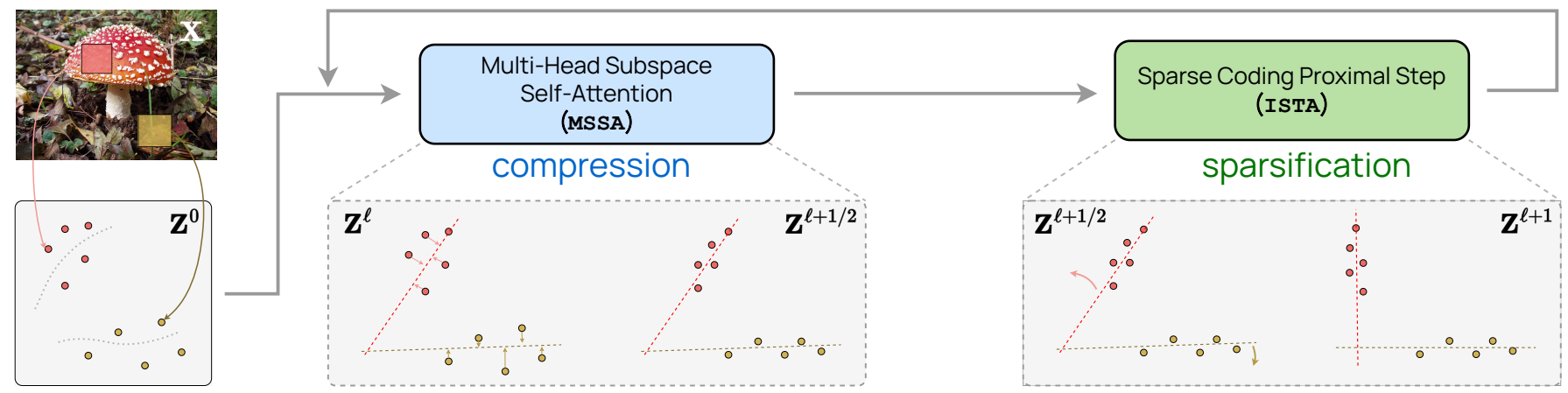

Figure 5.14: **The 'main loop' of the CRATE white-box deep network design.** After encoding input data as a sequence of tokens $\boldsymbol{Z}^0$, CRATE constructs a deep network that transforms the data to a canonical configuration of low-dimensional subspaces by successive ***compression*** against a local model for the distribution, generating $\boldsymbol{Z}^{\ell+1/2}$, and ***sparsification*** against a global dictionary, generating $\boldsymbol{Z}^{\ell+1}$. Repeatedly stacking these blocks and training the model parameters via backpropagation yields a powerful and interpretable representation of the data.

we can get the following rate-reduction-based transformer layer, illustrated in Figure 5.13,

$$\boldsymbol{Z}^{\ell+1/2} \doteq \boldsymbol{Z}^{\ell} + \operatorname{MSSA}(\boldsymbol{Z}^{\ell} \mid \boldsymbol{U}_{[K]}^{\ell}), \qquad \boldsymbol{Z}^{\ell+1} \doteq \operatorname{ISTA}(\boldsymbol{Z}^{\ell+1/2} \mid \boldsymbol{D}^{\ell}). \quad (5.2.22)$$

Composing multiple such layers following the incremental construction of our representation in (5.2.6), we obtain a white-box transformer architecture that transforms the data tokens towards a compact and sparse union of incoherent subspaces, where $f^{\mathrm{pre}} : \mathbb{R}^{D\times N} \to \mathbb{R}^{d\times N}$ is the pre-processing mapping that transforms the input tokens $\boldsymbol{X} \in \mathbb{R}^{D\times N}$ to first-layer representations $\boldsymbol{Z}^1 \in \mathbb{R}^{d\times N}$. An overall flow of this architecture was shown in Figure 5.14.

*Remark* 5.8 (**The roles of the forward pass and backward propagation**). In contrast to other unrolled optimization approaches such as the ReduNet [CYY+22], we *explicitly model* the distribution of each $\boldsymbol{Z}^{\ell}$ and $\boldsymbol{Z}^{\ell+1/2}$ at each layer, either by a mixture of linear subspaces or sparsely generated from a dictionary. We introduced the interpretation that at each layer $\ell$, the learned bases for the subspaces $\boldsymbol{U}_{[K]}^{\ell}$ and the learned dictionaries $\boldsymbol{D}^{\ell}$ together serve as a *codebook* or *analysis filter* that encodes and transforms the intermediate representations at each layer $\ell$. Since the input distribution to layer $\ell$ is first modeled by $\boldsymbol{U}_{[K]}^{\ell}$ then transformed by $\boldsymbol{D}^{\ell}$, the input distribution to each layer is different, and so we require a separate codebook at each layer to obtain the most parsimonious encoding. Parameters of these codebooks (i.e., the subspace bases and the dictionaries), heretofore assumed as being perfectly known, are actually learned from data (say via *backward propagation* within end-to-end training).

Hence, our methodology features a clear conceptual separation between forward "optimization" and backward "learning" for the so-derived white-box deep neural network. Namely, in its forward pass, we interpret each layer as an operator which, conditioned on a learned model (i.e., a codebook) for the distri-bution of its input, transforms this distribution towards a more parsimonious representation. In its backward propagation, the codebook of this model, for the distribution of the input to each layer, is updated to better fit a certain (supervised) input-output relationship, as illustrated in Figure 5.15. This conceptual interpretation implies a certain agnosticism of the model representations towards the particular task and loss; in particular, many types of tasks and losses will ensure that the models at each layer are fit, which ensures that the model produces parsimonious representations.

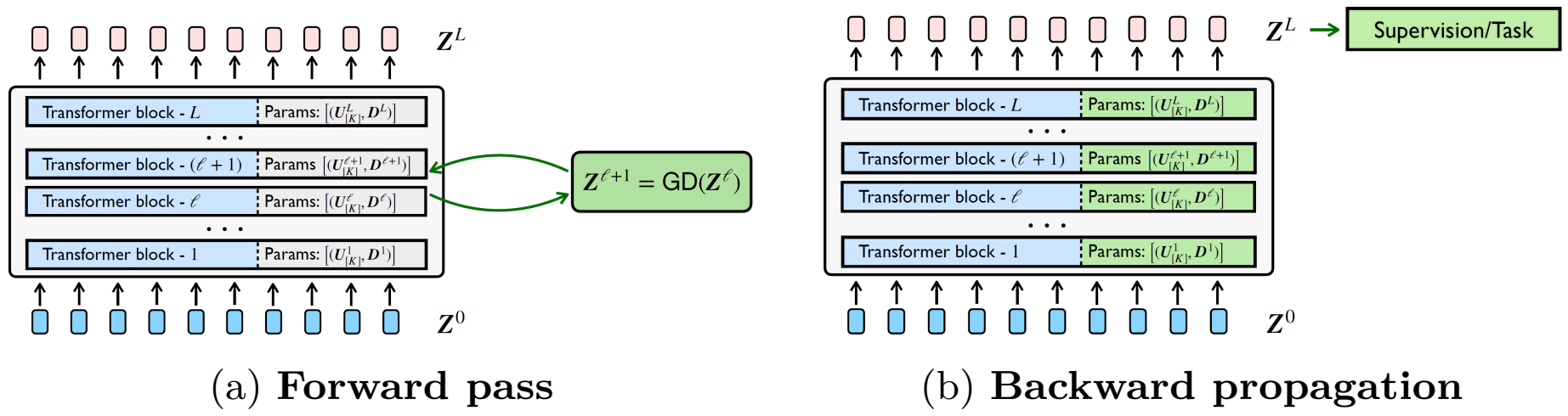

(a) **Forward pass** (b) **Backward propagation**

Figure 5.15: **The roles of forward pass and backward propagation in CRATE**. (a) Given fixed subspaces and dictionaries $\{(\boldsymbol{U}_{[K]}^{\ell}, \boldsymbol{D}^{\ell})\}_{\ell=1}^{L}$, each layer incrementally optimizes the sparse rate reduction of the representations in the forward pass; (b) Backpropagation learn subspaces and dictionaries $\{(\boldsymbol{U}_{[K]}^{\ell}, \boldsymbol{D}^{\ell})\}_{\ell=1}^{L}$ from training data.

There are three natural questions we can ask about CRATE:

- Do such constructed CRATE models learn good features by optimizing the sparse rate reduction?

- Do such constructed CRATE models use their high-quality features to perform well on predictive tasks?

- Do such constructed CRATE models exhibit interesting phenomena, say related to interpretability, as a result of their simplified design?

We contend that the answer to each question is "yes". First, Figure 5.16 examines how the network optimizes each term of the sparse rate reduction; each layer consistently decreases the features' compression measure and sparsity measures during the forward pass. Second, in Table 5.1 we demonstrate that CRATE has comparable top-1 classification accuracy under transfer learning setups to the most commonly used Vision Transformer (ViT) [DBK+21] neural network architecture at similar parameter counts, with promising scaling behavior — we obtain steady and consistent improvements by increasing the model size. Overall, CRATE achieves promising performance on real-world large-scale datasets by directly implementing our principled architecture. We provide more details of the implementation and analysis of the experimental results in Section 8.4.

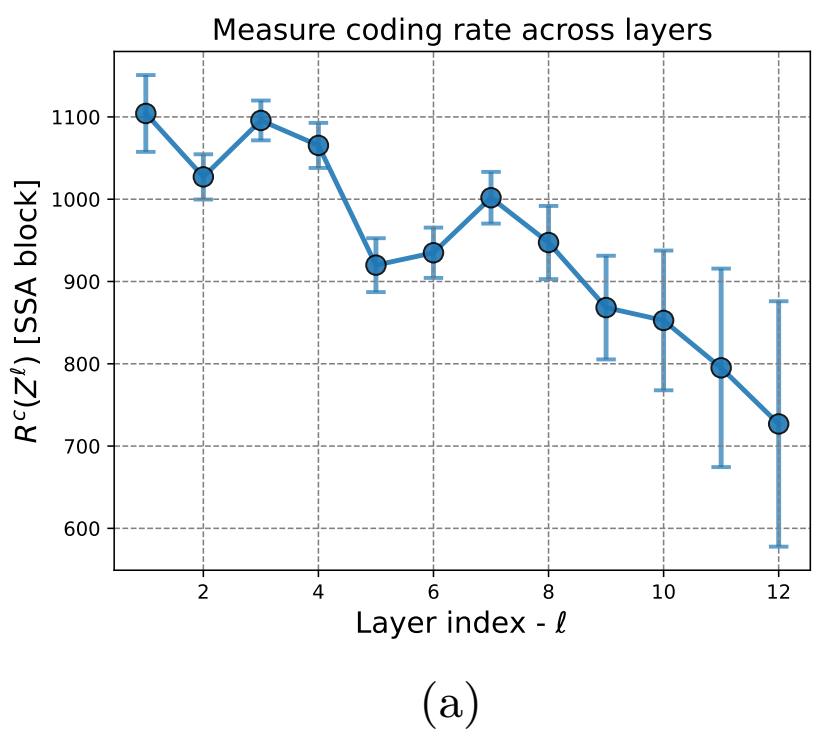

(a)

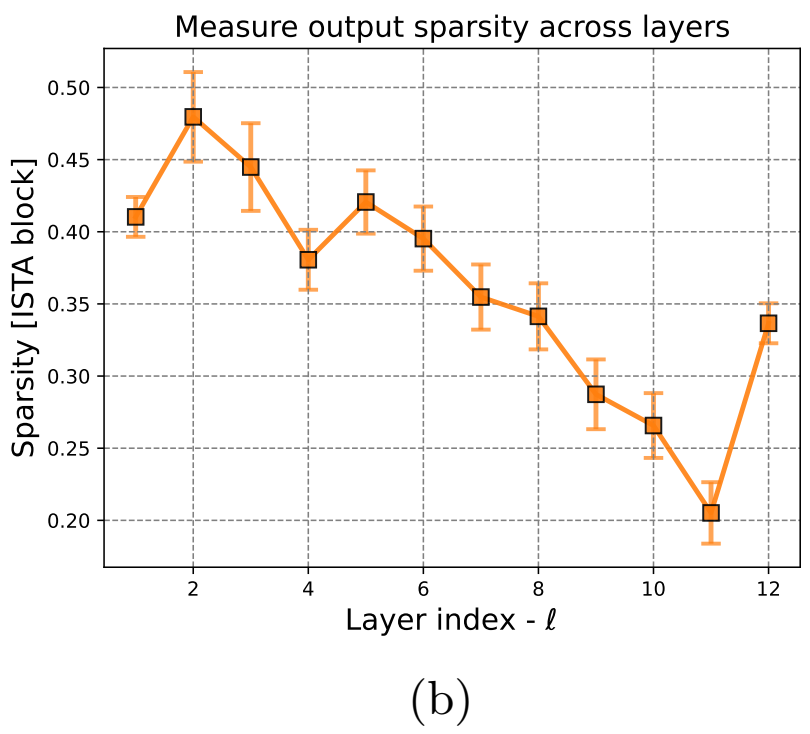

(b)

Figure 5.16: **Optimization of compression and sparsity measures through the forward pass of CRATE.** We measure the compression $R_\epsilon^c(\boldsymbol{Z}^{\ell+1/2})$ (a) and sparsity $\|\boldsymbol{Z}^{\ell+1}\|_1$ (b) at the output of each layer. We observe that both measures decrease over the course of the forward pass, as predicted by theory, and almost every layer monotonically decreases each measure. (The exception is the last layer's sparsity, since dense features are important for prediction.)

Table 5.1: **Top-1 classification accuracy of CRATE on various datasets with different model scales when pre-trained on ImageNet-1K.** For ImageNet-1K/ImageNet-1K ReaL, we directly evaluate the top-1 accuracy. For other datasets, we use models that are pre-trained on ImageNet as initialization and the evaluate the transfer learning performance via fine-tuning.

| **Model** | CRATE-T | CRATE-S | CRATE-B | CRATE-L | ViT-T | ViT-S |
|---|---|---|---|---|---|---|
| # parameters | 6.09M | 13.12M | 22.80M | 77.64M | 5.72M | 22.05M |
| ImageNet-1K | 66.7 | 69.2 | 70.8 | 71.3 | 71.5 | 72.4 |
| ImageNet-1K ReaL | 74.0 | 76.0 | 76.5 | 77.4 | 78.3 | 78.4 |
| CIFAR10 | 95.5 | 96.0 | 96.8 | 97.2 | 96.6 | 97.2 |
| CIFAR100 | 78.9 | 81.0 | 82.7 | 83.6 | 81.8 | 83.2 |
| Oxford Flowers-102 | 84.6 | 87.1 | 88.7 | 88.3 | 85.1 | 88.5 |
| Oxford-IIIT-Pets | 81.4 | 84.9 | 85.3 | 87.4 | 88.5 | 88.6 |

## 5.3 Variants of Deep Architectures by Design

So far, we wish that we have provided compelling evidence that the role of (popular) deep networks is to realize certain optimization algorithms for minimizing the coding rate (or maximizing the information gain) of the learned representations. However, readers who are familiar with optimization methods might have noticed that the above architectures (the ReduNet or the CRATE) correspond to rather basic optimization techniques. They may have plenty of room for improvement in efficiency or effectiveness. Moreover, if we believe the proposed theoretical framework for interpreting deep networks is correct, it should not only help explain existing architectures, it should guide us develop more efficient and effective architectures. In this section, we show this is the case: the resulting new architectures are not only fully interpretable but also with guaranteed correctness and improved efficiency.

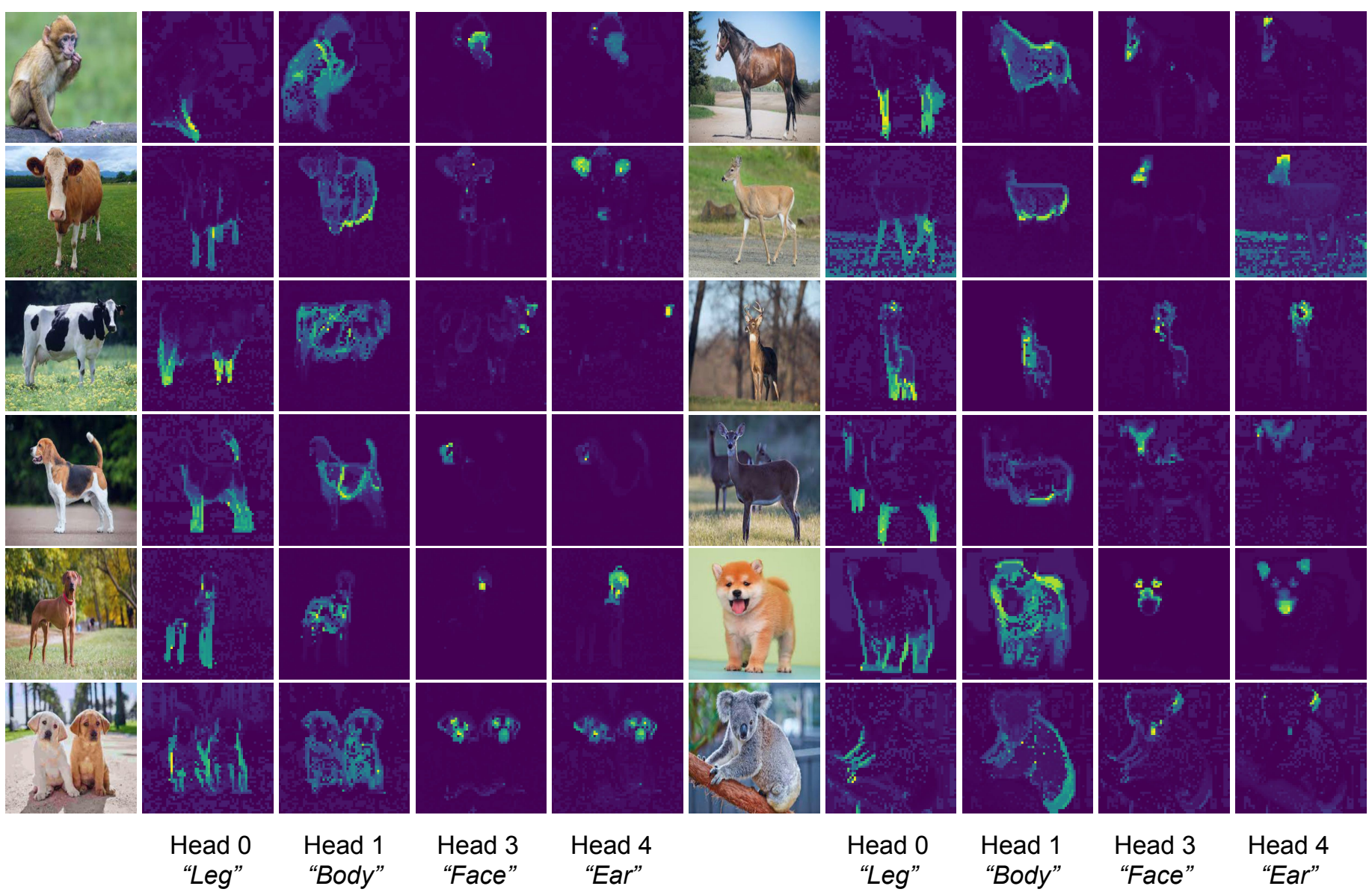

Figure 5.17: **Visualizing the attention maps of CRATE: what does the model look at to make a classification?** We visualize the last-layer attention maps, which are the last source of token-to-token interaction in the network, and therefore in some sense the most refined description of what sections of the input image the model uses to make a prediction. We observe an *extremely interesting* behavior: when trained on purely classification tasks, the attention maps identify the foreground objects in the image; moreover, they *naturally specialize* to different types of objects in the foreground! This empirical result gives us an interpretation of later-layer subspaces as each corresponding to a nearly-disjoint set of *concepts*.

### 5.3.1 Attention-Only Transformer Architecture

In this subsection, we ask and answer the question: "what is the simplest possible neural network which scales reasonably well and provably compresses the representation?" From our previous theory, the answer should be a stack of MSSA layers, which were motivated as lossy compression operator. This network architecture is simple enough such that it is possible to analyze the compression or denoising performance rigorously and systematically. Towards this end, let us formalize our subspace model: the initial token representations $\boldsymbol{Z}^1$ are sampled from a noisy mixture of low-rank Gaussians, supported on subspaces spanned by $\boldsymbol{U}_k \in \mathbb{R}^{d\times p}$, as follows: we partition the $N$ token indices into

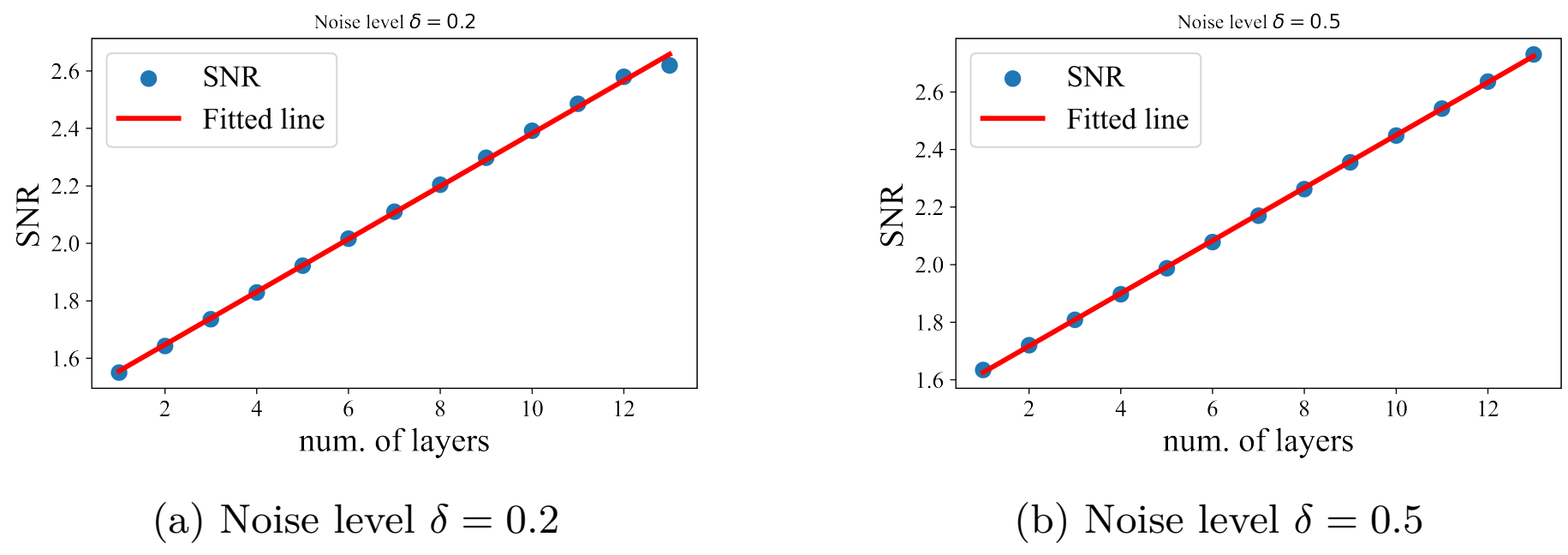

(a) Noise level $\delta = 0.2$ (b) Noise level $\delta = 0.5$

Figure 5.18: **Denoising performance of the attention-only transformer.** Here, we sample initial token representations from a mixture of low-rank Gaussians in Equation (5.3.1). Then, we apply (5.3.2) to update token representations and report the SNR at each layer.

known subsets $C_1, \dots, C_K$, such that:

$$\boldsymbol{z}_i = \underbrace{\boldsymbol{U}_k \boldsymbol{a}_i}_{signal} + \underbrace{\sum_{j \neq k}^{K} \boldsymbol{U}_j \boldsymbol{e}_{i,j}}_{noise}, \ \forall i \in C_k, \tag{5.3.1}$$

where $\boldsymbol{a}_i \overset{i.i.d.}{\sim} \mathcal{N}(\boldsymbol{0}, \boldsymbol{I}_{p_k})$ and $\boldsymbol{e}_{i,j} \overset{i.i.d.}{\sim} \mathcal{N}(\boldsymbol{0}, \delta^2 \boldsymbol{I}_{p_j})$ for all $i \in C_k$ and $k \in [K]$, $\{\boldsymbol{a}_i\}$ and $\{\boldsymbol{e}_{i,j}\}$ are all independent.

**Denoising operator for token representations.** Now, we show that the MSSA operator (see (5.2.15)) can incrementally denoise token representations generated from the above model. Specifically, we consider a generalized MSSA operator: for each $\ell \in [L]$,

$$\boldsymbol{Z}^{\ell+1} = \boldsymbol{Z}^{\ell} + \eta \sum_{k=1}^{K} \boldsymbol{U}_k \boldsymbol{U}_k^T \boldsymbol{Z}^{\ell} \varphi(\boldsymbol{Z}^{\ell^T} \boldsymbol{U}_k \boldsymbol{U}_k^T \boldsymbol{Z}^{\ell}), \tag{5.3.2}$$

where $\eta > 0$ is the step size, and $\varphi$ is a function operating on each column of the input matrix, such as softmax, element-wise ReLU, etc. To simplify our theory, we assume that the subspaces in Equation (5.3.1) are orthogonal to each other, i.e., $\boldsymbol{U}_k^T \boldsymbol{U}_j = \boldsymbol{0}$ for all $k \neq j$. Note that this assumption is not restrictive, as in high-dimensional spaces, random low-dimensional subspaces are incoherent to each other with high probability, i.e., $\boldsymbol{U}_k^T \boldsymbol{U}_j \approx \boldsymbol{0}$ [WM22].[19]

Now, let the columns of $\boldsymbol{Z}_k^{\ell}$ denote the token representations from the $k$-th subspace at the $\ell$-th layer. To quantify the denoising capability, we define the

[19] One may straightforwardly generalize our results to non-orthogonal subspaces, with slightly more sophisticated analysis.

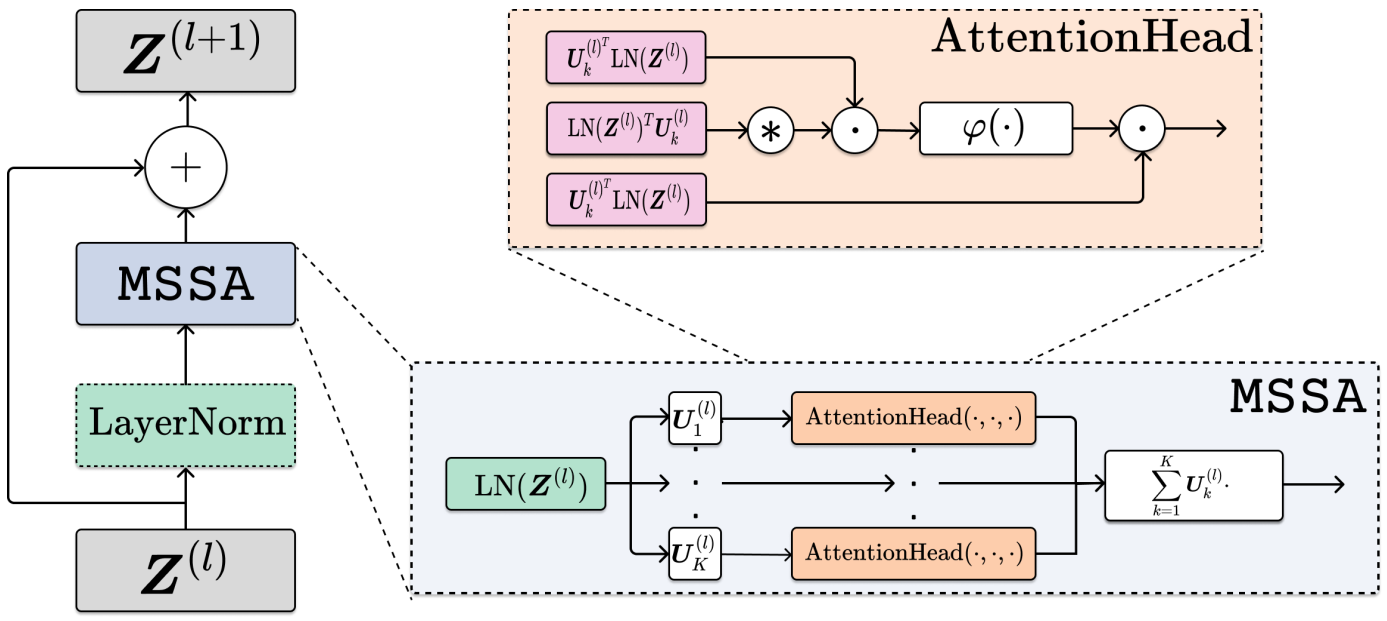

Figure 5.19: **Details of the attention-only transformer architecture.** Each layer consists of the MSSA operator and a skip connection. In addition, LayerNorm is included only for language tasks. In practice, backpropagation is applied to train the model parameters using training samples.

signal-to-noise ratio (SNR) for each block of the token representations at the $\ell$-th layer as follows:

$$\mathrm{SNR}(\boldsymbol{Z}_k^{\ell}) \doteq \frac{\|\boldsymbol{U}_k\boldsymbol{U}_k^T\boldsymbol{Z}_k^{\ell}\|_F}{\|(\boldsymbol{I}-\boldsymbol{U}_k\boldsymbol{U}_k^T)\boldsymbol{Z}_k^{\ell}\|_F}, \quad \forall k \in [K]. \tag{5.3.3}$$

To simplify our analysis, we assume that $|C_1| = \cdots = |C_K| = N/K$, and

$$\begin{bmatrix}\boldsymbol{U}_1 & \dots & \boldsymbol{U}_K\end{bmatrix} \in \mathcal{O}^{d\times Kp}. \tag{5.3.4}$$

With the above setup, we can now characterize the denoising performance of a slightly modified MSSA operator, which thresholds terms which are too small in the softmax.

**Theorem 5.1.** *Let $\boldsymbol{Z}^1$ be defined in Equation (5.3.1) and $\varphi(\cdot)$ in (5.3.2) be $\varphi(\boldsymbol{x}) = h(\sigma(\boldsymbol{x}))$, where $\sigma : \mathbb{R}^N \to \mathbb{R}^N$ is the softmax function and $h : \mathbb{R}^N \to \mathbb{R}^N$ is an element-wise thresholding function with $h(x) = \tau$ if $x > \tau$ and $h(x) = 0$ if $x \leq \tau$ for each $i \in [N]$. Suppose that $p \gtrsim \log N$, $\delta \lesssim \sqrt{\log N}/\sqrt{p}$, and*

$$\tau \in \left(\frac{1}{2}, \frac{1}{1 + N\exp(-9p/32)}\right].$$

*For sufficiently large $N$, it holds with probability at least $1 - KLN^{-\Omega(1)}$ that for each $\ell \in [L]$,*

$$\mathrm{SNR}(\boldsymbol{Z}_k^{\ell+1}) = (1 + \eta\tau)\mathrm{SNR}(\boldsymbol{Z}_k^{\ell}),\ \forall k \in [K]. \tag{5.3.5}$$

This theorem demonstrates that when the initial token representations are sampled from a mixture of low-rank Gaussian distributions with a noise level $O(\sqrt{\log N}/\sqrt{p})$, we show that each application of the modified MSSA operator denoises token representations at a linear rate. Notably, our theoretical results are well-supported by experimental observations using the *actual* MSSA operator in Figure 5.18. This theorem provides a theoretical foundation for the practical denoising capability of the transformer architecture derived by unrolling (5.3.2).

Table 5.2: **Evaluating AoT on vision tasks**: top-1 accuracy on ImageNet.

| Models | Accuracy | # of Parameters |
|---|---|---|
| AoT | 71.7% | 22M |
| CRATE | 79.5% | 39M |
| ViT | 72.4 % | 22M |

**Attention-only transformers.** To verify that this denoising capability translates to reasonable performance on real data and tasks, we formally propose an attention-only transformer architecture, which we short-hand as AoT. Specifically, by unrolling the iterative optimization steps (5.3.2) as layers of a deep network, we construct a transformer architecture in Figure 5.19. Each layer of the proposed architecture only consists of the MSSA operator and a skip connection. For language tasks, we additionally incorporate LayerNorm before the MSSA operator to improve performance. The complete architecture is built by stacking such layers, along with essential task-specific pre-processing and post-processing steps, such as positional encoding, token embedding, and a final task-specific head to adapt to different applications. Table 5.2 and demonstrates the performance of this minimalist transformer on vision classification; the performance is reasonable, despite the minimalist design. In the sequel, we will discuss a conceptual improvement to the CRATE formulation which *also* improve the performance and efficiency.

### 5.3.2 Linear-Time Attention: Token Statistics Transformer

In this subsection, we propose a new transformer attention operator whose computational complexity scales linearly with the number of tokens based on the coding rate reduction objective. Specifically, we derive a novel variational form of the $\mathrm{MCR}^2$ objective and show that the architecture that results from unrolled gradient descent of this variational objective leads to a new attention module called Token Statistics Self-Attention (TSSA). TSSA has *linear computational and memory complexity* and radically departs from the typical attention architecture that computes pairwise similarities between tokens. Replacing a traditional attention operator in a transformer with TSSA yields a new neural network architecture called the Token Statistics Transformer (ToST), which is highly efficient and has promising empirical performance. As a preliminary, recall from (4.2.17) that $\mathbf{\Pi} = [\boldsymbol{\pi}_1, \ldots, \boldsymbol{\pi}_K] \in \mathbb{R}^{N\times K}$ denotes a stochastic "group assignment" matrix (i.e., $\mathbf{\Pi}\mathbf{1} = \mathbf{1}$ and $\Pi_{ik} \geq 0,\ \forall (i,k) \in [N]\times[K]$), where $\Pi_{ik}$ denotes the probability of assigning the $i$-th token to the $k$-th group.

**A new variational form for coding rates.** To begin, we consider a general form of MCR$^2$-like objectives based on concave functions of the spectrum of a matrix. Namely, for a given PSD matrix $\boldsymbol{M} \in \mathsf{PSD}(d)$ and any scalar $c \geq 0$ we have that $\log\det(\boldsymbol{I} + c\boldsymbol{M}) = \sum_{i=1}^{d} \log(1 + c\lambda_i(\boldsymbol{M}))$, where $\lambda_i(\boldsymbol{M})$ is the $i$-th largest eigenvalue of $\boldsymbol{M}$. Further, note that $\log(1 + c\sigma)$ is a concave non-decreasing function of $\sigma$. Thus, we describe our results in terms of a more general form of MCR$^2$ based on general spectral functions of PSD matrices of the form $F(\boldsymbol{M}) = \sum_{i=1}^{d} f(\lambda_i(\boldsymbol{M}))$, where $f$ is concave and non-decreasing. In particular, recall from earlier in this Chapter that the attention mechanism arises from unrolling the compression component of MCR$^2$, so we consider a more general MCR$^2$-style compression function:

$$R_{c,f}(\boldsymbol{Z}, \boldsymbol{\Pi}) \doteq \frac{1}{2}\sum_{k=1}^{K} \frac{N_k}{N} F\left(\frac{1}{N_k}\boldsymbol{Z}\mathrm{Diag}(\boldsymbol{\pi}_k)\boldsymbol{Z}^\top\right). \tag{5.3.6}$$

For the above objective, we now note the following result:

**Theorem 5.2.** *Let $f\colon [0,\infty) \to \mathbb{R}$ be non-decreasing, concave, and obey $f(0) = 0$, and let $F\colon \mathsf{PSD}(d) \to \mathbb{R}$ have the form $F(\boldsymbol{M}) = \sum_{i=1}^{d} f(\lambda_i(\boldsymbol{M}))$. Then for each $\boldsymbol{M} \in \mathsf{PSD}(d)$ and $\boldsymbol{Q} \in \mathsf{O}(d)$, we have*

$$F(\boldsymbol{M}) \leq \sum_{i=1}^{d} f\left((\boldsymbol{Q}^\top \boldsymbol{M}\boldsymbol{Q})_{ii}\right). \tag{5.3.7}$$

*Further, the inequality in (5.3.7) is achieved with equality for any $\boldsymbol{Q}$ which diagonalizes $\boldsymbol{M}$, and if $f$ is strictly concave then the inequality in (5.3.7) is achieved with equality if and only if $\boldsymbol{Q}$ diagonalizes $\boldsymbol{M}$.*

Using the above result, we can replace (5.3.6) with an equivalent variational objective with form

$$R_{c,f}^{\mathrm{var}}(\boldsymbol{Z}, \boldsymbol{\Pi} \mid \boldsymbol{U}_{[K]}) \doteq \frac{1}{2}\sum_{k=1}^{K} \frac{N_k}{N} \sum_{i=1}^{d} f\left(\frac{1}{N_k}(\boldsymbol{U}_k^\top \boldsymbol{Z}\mathrm{Diag}(\boldsymbol{\pi}_k)\boldsymbol{Z}^\top \boldsymbol{U}_k)_{ii}\right), \tag{5.3.8}$$

where the equivalence is in the sense that for an optimal choice of $\{\boldsymbol{U}_k \in \mathsf{O}(d)\}_{k=1}^{K}$ matrices as described in Theorem 5.2 (i.e., orthogonal matrices which diagonalize each $\boldsymbol{Z}\mathrm{Diag}(\boldsymbol{\pi}_k)\boldsymbol{Z}^\top$) we will achieve a tight bound with $R_{c,f}^{\mathrm{var}}(\boldsymbol{Z}, \boldsymbol{\Pi} \mid \boldsymbol{U}_{[K]}) = R_{c,f}(\boldsymbol{Z}, \boldsymbol{\Pi})$. Note that in general, achieving this bound would require selecting, for each sampled instance of $\boldsymbol{Z}$, a new optimal set of $\boldsymbol{U}_k$ parameter matrices which diagonalize each $\boldsymbol{Z}\mathrm{Diag}(\boldsymbol{\pi}_k)\boldsymbol{Z}^\top$, which is clearly impractical for network architecture. Instead, as an alternative viewpoint, rather than considering the data ($\boldsymbol{Z}$) as fixed and trying to optimize the $\boldsymbol{U}_k$ parameters to achieve the tight variational bound, we can instead take the algorithmic unrolling design principle described above and design an operator to perturb $\boldsymbol{Z}$ to incrementally minimize $R_{c,f}^{\mathrm{var}}(\cdot \mid \boldsymbol{U}_{[K]})$. To make this point explicit, each variational bound becomes tight when the eigenspaces of $\boldsymbol{Z}\mathrm{Diag}(\boldsymbol{\pi}_k)\boldsymbol{Z}^\top$ align with the columns of $\boldsymbol{U}_k$, so by rotating the appropriate columns of $\boldsymbol{Z}$ (namely, those which correspond to large entries in $\boldsymbol{\pi}_k$) to align with $\boldsymbol{U}_k$ we can approach a tight variational bound. That is, instead of rotating $\boldsymbol{U}_k$ to align with the data for each instance of $\boldsymbol{Z}$, we can instead rotate the token features in each $\boldsymbol{Z}$ to align with $\boldsymbol{U}_k$.

Following this approach, we compute a gradient descent step on $R^{\mathrm{var}}_{c,f}$ w.r.t. $\boldsymbol{Z}$. To begin this computation, first let $\boldsymbol{\pi} \in \mathbb{R}^N$ be any element-wise non-negative vector. Then we have

$$\nabla_{\boldsymbol{Z}} \, \frac{1}{2}\sum_{i=1}^{d} f((\boldsymbol{Z}\mathrm{Diag}(\boldsymbol{\pi})\boldsymbol{Z}^\top)_{ii}) = \mathrm{Diag}(\nabla f[\boldsymbol{Z}^{\odot 2}\boldsymbol{\pi}])\boldsymbol{Z}\mathrm{Diag}(\boldsymbol{\pi}), \tag{5.3.9}$$

where $\nabla f$ is the gradient of $f$, and (recall) $\nabla f[\cdot]$ applies $\nabla f$ to each element of the vector in the bracket. In particular, for $f(x) = \log(1 + (d/\epsilon^2)x)$, $\nabla f(x) = (d/\epsilon^2)(1 + (d/\epsilon^2)x)^{-1}$ is simply a non-linear activation. Also, (recall) $N_k = \langle \boldsymbol{\pi}_k, \mathbf{1}\rangle$. Thus, the gradient of $R^{\mathrm{var}}_{c,f}$ w.r.t. $\boldsymbol{Z}$ is:

$$\nabla_{\boldsymbol{Z}} R^{\mathrm{var}}_{c,f}(\boldsymbol{Z}, \boldsymbol{\Pi} \mid \boldsymbol{U}_{[K]}) \tag{5.3.10}$$

$$= \frac{1}{n}\sum_{k=1}^{K} \boldsymbol{U}_k \underbrace{\mathrm{Diag}\left(\nabla f\left[(\boldsymbol{U}_k^\top \boldsymbol{Z})^{\odot 2}\frac{\boldsymbol{\pi}_k}{\langle \boldsymbol{\pi}_k, \mathbf{1}\rangle}\right]\right)}_{\doteq \boldsymbol{D}(\boldsymbol{Z}, \boldsymbol{\pi}_k \mid \boldsymbol{U}_k)} \boldsymbol{U}_k^\top \boldsymbol{Z}\mathrm{Diag}(\boldsymbol{\pi}_k). \tag{5.3.11}$$

(Note that the $1/N$ constant arises from a $(N_k/N) \cdot (1/N_k) = 1/N$ constant in each term of the sum.) If we now consider a gradient step w.r.t. the $j$-th token $\boldsymbol{z}_j$ , we arrive at our proposed incremental compression operator, i.e., our surrogate for a *self attention* + residual operator:

$$\boldsymbol{z}_j^+ = \boldsymbol{z}_j - \tau\nabla_{\boldsymbol{z}_j} R^{\mathrm{var}}_{c,f}(\boldsymbol{Z}, \boldsymbol{\Pi} \mid \boldsymbol{U}_{[K]}) \tag{5.3.12}$$

$$= \boldsymbol{z}_j - \frac{\tau}{N}\sum_{k=1}^{K} \Pi_{jk}\boldsymbol{U}_k\boldsymbol{D}(\boldsymbol{Z}, \boldsymbol{\pi}_k \mid \boldsymbol{U}_k)\boldsymbol{U}_k^\top \boldsymbol{z}_j \tag{5.3.13}$$

for each $j \in [n]$, where $\tau > 0$ is a step size parameter for the incremental optimization.

**Model interpretation.** Given the proposed attention operator in (5.3.12), first recall that the rows of $\boldsymbol{\Pi}$ are non-negative and sum to 1 , so our operator takes a weighted average of $K$ "attention head"-esque operators and then adds a residual connection. Using that $\sum_{k=1}^{K} \Pi_{jk} = 1$, we can rewrite (5.3.12) as:

$$\boldsymbol{z}_j^+ = \sum_{k=1}^{K} \Pi_{jk}\Big[\boldsymbol{z}_j \underbrace{- \frac{\tau}{n}\boldsymbol{U}_k\boldsymbol{D}(\boldsymbol{Z}, \boldsymbol{\pi}_k \mid \boldsymbol{U}_k)\boldsymbol{U}_k^\top \boldsymbol{z}_j}_{\text{action of one attention head}}\Big]. \tag{5.3.14}$$

That is, we can view each attention head as first projecting the token features onto the basis $\boldsymbol{U}_k$ via multiplying by $\boldsymbol{U}_k^\top$, multiplying by the diagonal matrix $\boldsymbol{D}(\boldsymbol{Z}, \boldsymbol{\pi}_k \mid \boldsymbol{U}_k)$ (abbreviated as $\boldsymbol{D}_k$), projecting back into the standard basis via multiplying by $\boldsymbol{U}_k$, and subtracting this from the original token features via the residual connection. The core aspect of our attention layer is the computation of $\boldsymbol{D}_k$. Namely, $\Pi_{jk} \geq 0$, so $\boldsymbol{\pi}_k/\langle \boldsymbol{\pi}_k, \mathbf{1}\rangle \in \mathbb{R}^N$ forms a probability distribution over which tokens belong to the $k^{\text{th}}$ group. As a result, $(\boldsymbol{U}_k^\top \boldsymbol{Z})^{\odot 2}\boldsymbol{\pi}_k/\langle \boldsymbol{\pi}_k, \mathbf{1}\rangle$ estimates the second moment of $\boldsymbol{U}_k^\top \boldsymbol{Z}$ under the distribution given by $\boldsymbol{\pi}_k/\langle \boldsymbol{\pi}_k, \mathbf{1}\rangle$. Further, since $f$ is a concave non-decreasing function, $\nabla f(x)$ monotonically decreases towards 0 as $x$ increases, so the entries of $\boldsymbol{D}_k$ (which have form $\nabla f(x)$) achieve their maximum at $x = 0$ and decay monotonically to 0 as $x$ increases.

From this, we arrive at the core interpretation of our attention head + residual operators $[\boldsymbol{I} - (\tau/n)\boldsymbol{U}_k\boldsymbol{D}_k\boldsymbol{U}_k^\top]$. Namely, this operator does an approximate low-rank data-dependent projection, where directions which have a large amount of "power" after the projection $\boldsymbol{U}_k^\top \boldsymbol{Z}$ (i.e., directions which have a large second moment $(\boldsymbol{U}_k^\top \boldsymbol{Z})^{\odot 2}\boldsymbol{\pi}_k/\langle \boldsymbol{\pi}_k, \mathbf{1}\rangle$) are preserved, while directions which do not are suppressed. To see this, recall that the entries of $\boldsymbol{D}_k$ decrease monotonically to 0 as the second moment increases, so for directions with large second moments the attention + residual operator acts largely as the identity operator. Conversely, for directions with a small second moment, our operator subtracts a projection of the tokens along those directions, resulting in those directions being suppressed. Compared to the standard self-attention operator, our method clearly does not compute any pairwise similarities between tokens. Rather, the interactions between the tokens in $\boldsymbol{Z}$ impact the operator solely through their contribution to the second moment statistic used to construct the $\boldsymbol{D}_k$'s. Nevertheless, similar to the standard self-attention operator, our method still has a clear interpretation as performing a form of compression towards a data-dependent low-rank structure, in the sense that it performs an approximate low-rank projection, where the specific directions that are suppressed are those which are not strongly aligned with other tokens in the group.

**Computational considerations.** Having introduced our proposed attention operator, we now discuss how it can be computed practically. First, until this point in the presentation, we have avoided discussion of how tokens are "grouped" into various attention heads via the $\boldsymbol{\Pi}$ matrix, but clearly a means of constructing $\boldsymbol{\Pi}$ is needed to implement our method. Additionally, our variational form in Theorem 5.2 requires the $\boldsymbol{U}$ matrices to be square and orthogonal, but one would ideally like to use smaller matrices (i.e., reduce the number of columns in $\boldsymbol{U}$) for efficiency as well as drop the orthogonality constraints.

In practice, we do not enforce the orthogonality constraints. To reduce the number of columns in the $\boldsymbol{U}$ matrices, we note that similar to CRATE [YBP+23], if we assume the features $\boldsymbol{Z}$ within group $k$ are (approximately) clustered around a low-dimensional subspace — say of dimension $p$ — then the within-group-$k$ covariance $\boldsymbol{Z}\operatorname{Diag}(\boldsymbol{\pi}_k)\boldsymbol{Z}^\top$ is low-rank, where recall that [YCY+20] shows that the optimal geometry of $\boldsymbol{Z}$ will be for each group to be a low-rank subspace, orthogonal to the other groups. We can thus explicitly find a low-dimensional orthonormal basis for the image of this covariance, i.e., the linear span of the data in group $k$. If the dimension is $p \leq d$, the basis can be represented by a $d \times p$ orthogonal matrix $\boldsymbol{U}_k \in \mathsf{O}(d, p)$. In this case, we can more efficiently upper-bound $R_{c,f}$ using these low-rank orthogonal basis matrices. To show this, we use a more general version of Theorem 5.2 to yield the following corollary.

**Corollary 5.1.** *Let $f\colon [0, \infty) \to \mathbb{R}$ be non-decreasing, concave, and obey $f(0) = 0$, and let $F\colon \mathsf{PSD}(p) \to \mathbb{R}$ have the form $F(\boldsymbol{M}) = \sum_{i=1}^{p} f(\lambda_i(\boldsymbol{M}))$. Let $\boldsymbol{Z}$, $\boldsymbol{\Pi}$ be fixed. Then, for all $\boldsymbol{U}_1, \dots, \boldsymbol{U}_K \in \mathsf{O}(d, p)$ such that $\operatorname{image}(\boldsymbol{Z} \operatorname{diag}(\boldsymbol{\pi}_k)\boldsymbol{Z}^\top) \subset \operatorname{image}(\boldsymbol{U}_k)$ for all $k \in [K]$, we have*

$$R_{c,f}(\boldsymbol{Z}, \boldsymbol{\Pi}) \leq R_{c,f}^{\mathrm{var}}(\boldsymbol{Z}, \boldsymbol{\Pi} \mid \boldsymbol{U}_{[K]}), \tag{5.3.15}$$

*where $R_{c,f}^{\mathrm{var}}$ is formally defined in (5.3.8). Equality holds if $\boldsymbol{U}_k$ diagonalizes $\boldsymbol{Z} \operatorname{diag}(\boldsymbol{\pi}_k)\boldsymbol{Z}^\top$ for each $k \in [K]$, and if $f$ is strongly concave then this equality condition becomes an "if and only if."*

The final step to define our attention operator is to estimate the group membership $\boldsymbol{\Pi}$. For this we posit a simple model of how each feature $\boldsymbol{z}_j$ deviates from its supporting subspace and then find the optimal subspace assignment. [YBP+23] show that if we independently model each $\boldsymbol{z}_j$ as belonging to a low-dimensional Gaussian mixture model, where each Gaussian has a covariance matrix with identical spectrum and is supported on a subspace with orthonormal basis $\boldsymbol{U}_k$, plus independent Gaussian noise with covariance $\eta \boldsymbol{I}$, then the posterior probability that each token $\boldsymbol{z}_j$ belongs to each subspace is given by the assignment matrix $\boldsymbol{\Pi} = \boldsymbol{\Pi}(\boldsymbol{Z} \mid \boldsymbol{U}_{[K]})$ as follows:

$$\boldsymbol{\Pi} = \begin{bmatrix} \boldsymbol{\nu}(\boldsymbol{z}_1 \mid \boldsymbol{U}_{[K]})^\top \\ \vdots \\ \boldsymbol{\nu}(\boldsymbol{z}_n \mid \boldsymbol{U}_{[K]})^\top \end{bmatrix}, \quad \text{where} \tag{5.3.16}$$

$$\boldsymbol{\nu}(\boldsymbol{z}_j \mid \boldsymbol{U}_{[K]}) \doteq \operatorname{softmax}\left( \frac{1}{2\eta} \begin{bmatrix} \|\boldsymbol{U}_1^\top \boldsymbol{z}_j\|_2^2 \\ \vdots \\ \|\boldsymbol{U}_K^\top \boldsymbol{z}_j\|_2^2 \end{bmatrix} \right), \quad \forall j \in [n], \tag{5.3.17}$$

where $\eta$ becomes a learnable temperature parameter. Thus, given an input feature $\boldsymbol{Z}$, we estimate $\boldsymbol{\Pi}$ using (5.3.16) and then compute the attention operator. Combining the construction of $\boldsymbol{\Pi}$ in (5.3.16) with (5.3.12), we obtain the *Token Statistics Self-Attention* operator:

$$\operatorname{TSSA}(\boldsymbol{Z} \mid \boldsymbol{U}_{[K]}) \doteq -\frac{\tau}{n} \sum_{k=1}^{K} \boldsymbol{U}_k \boldsymbol{D}(\boldsymbol{Z}, \boldsymbol{\pi}_k \mid \boldsymbol{U}_k) \boldsymbol{U}_k^\top \boldsymbol{Z} \operatorname{diag}(\boldsymbol{\pi}_k), \tag{5.3.18}$$

where $\boldsymbol{\pi}_k$ are the columns of $\boldsymbol{\Pi} = \boldsymbol{\Pi}(\boldsymbol{Z} \mid \boldsymbol{U}_{[K]})$ defined in (5.3.16) and $\boldsymbol{D}$ is defined in (5.3.10).

By starting with a standard Transformer architecture and replacing the attention operator with TSSA, we obtain an architecture called Token Statistics

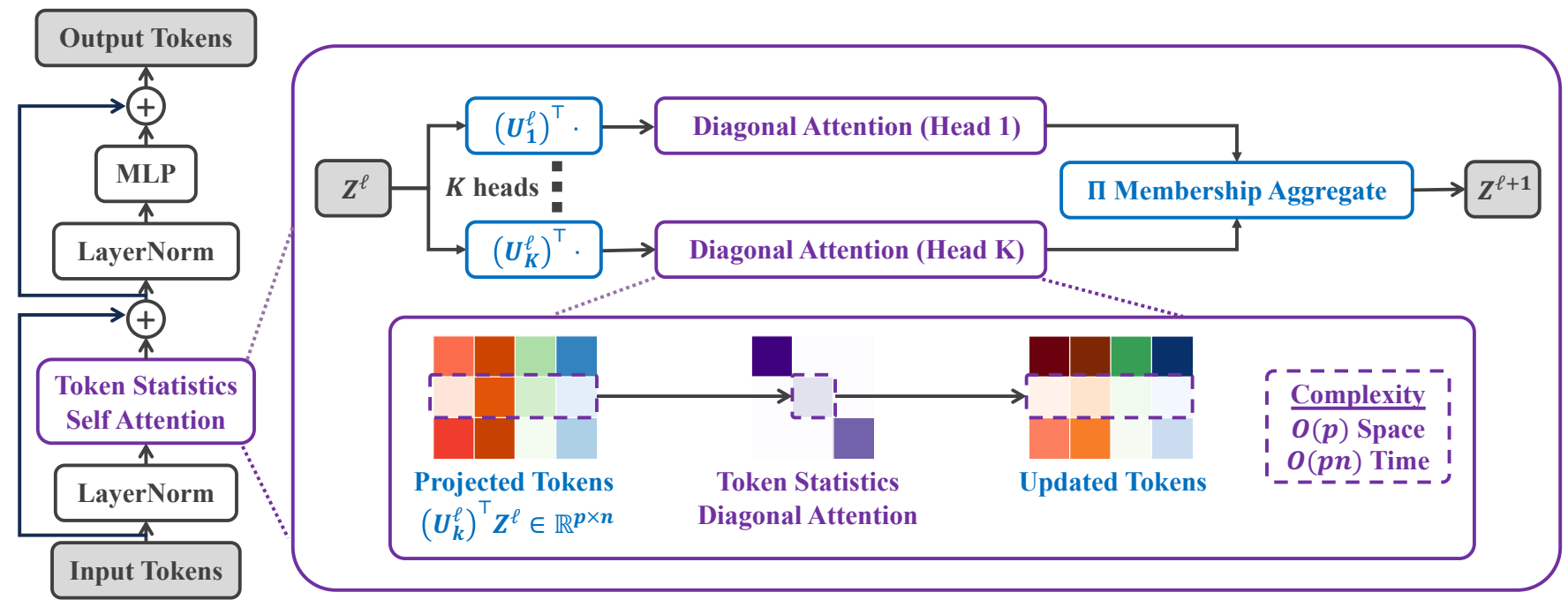

Figure 5.20: **One layer $\ell$ of the proposed Token Statistics Transformer (ToST).** Notably, the self-attention of ToST transforms tokens $\boldsymbol{Z}^{\ell}$ efficiently to $\boldsymbol{Z}^{\ell+1}$, via multiplying each row of the projected token by *only a scalar*. This leads to reduced complexity of the attention: it has $O(p)$ space and $O(pn)$ time complexity, where $p$ is the dimension of the projected tokens of each head, and $n$ is the number of tokens.

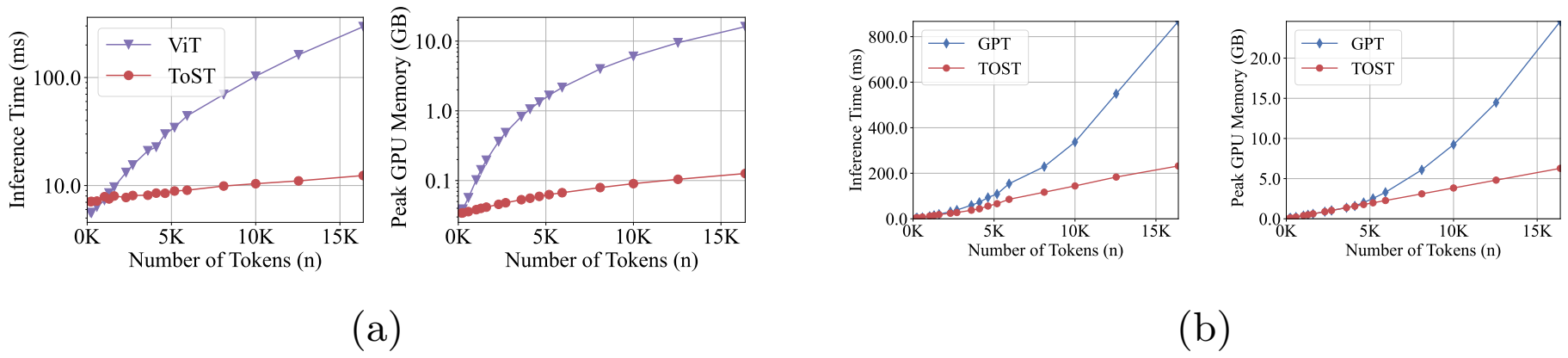

Figure 5.21: **Time and memory complexity of ToST in practice** compared to empirically designed ViT (a) and GPT-2 (b). We observe that ToST only requires linear time and memory, in comparison to quadratic costs of traditional transformers.

Transformer (ToST), visualized in Figure 5.20. We would like to re-emphasize that the TSSA operator mitigates the usual time and memory complexity of the attention operator, in particular reducing the quadratic time and memory complexity to linear, which makes ToST much more efficient than its empirically designed counterpart. We now briefly show the empirical performance and efficiency of the ToST architecture.

### 5.3.3 Causal CRATE

Some of the most popular applications of deep networks are in the regimes of *sequence data* where, unlike the previous example of imagery, there is a natural order to the data. For example, unlike in imagery (where one tends to look at the whole image at once), language is processed one token at a time (causally). This simple fact motivates the development of large language models (LLMs), which are trained to predict the (probability distribution of) the next token in a sequence given *only the history of previous tokens*. Then, to sample from an LLM, one simply iteratively samples from the predicted distribution for the next token and then appends the sampled token to the history. In order to train LLMs, and other models such as video generation models, it is necessary to develop architectures which can efficiently perform this causal computation.

Table 5.3: **Top 1 accuracy of ToST on various datasets with different model sizes when pre-trained on ImageNet.** For ImageNet/ImageNetReaL, we directly evaluate the top-1 accuracy. For other datasets, we pre-train the model on ImageNet and then evaluate the transfer learning performance via fine-tuning. We observe that ToST performance scales with model size and is competitive with the empirically designed ViT.

| Datasets | ToST-T(iny) | ToST-S(mall) | ToST-M(edium) | ViT-S | ViT-B(ase) |
|---|---|---|---|---|---|
| # parameters | 5.8M | 22.6M | 68.1M | 22.1M | 86.6 M |
| ImageNet | 67.3 | 77.9 | 80.3 | 79.8 | 81.8 |
| ImageNet ReaL | 72.2 | 84.1 | 85.6 | 85.6 | 86.7 |
| CIFAR10 | 95.5 | 96.5 | 97.5 | 98.6 | 98.8 |
| CIFAR100 | 78.3 | 82.7 | 84.5 | 88.8 | 89.3 |
| Oxford Flowers-102 | 88.6 | 92.8 | 94.2 | 94.0 | 95.7 |
| Oxford-IIIT-Pets | 85.6 | 91.1 | 92.8 | 92.8 | 94.1 |

Table 5.4: **Long-context performance of ToST.** We demonstrate performance on the Long Range Arena, a benchmark of long-context (large-$n$) tasks. On this benchmark, ToST has outright superior performance and efficiency compared to every Transformer variant, including the Transformer itself.

| Model | ListOps | Text | Retrieval | Image | Pathfinder | Avg |
|---|---|---|---|---|---|---|
| Reformer | 37.27 | 56.10 | 53.40 | 38.07 | 68.50 | 50.56 |
| BigBird | 36.05 | 64.02 | 59.29 | 40.83 | 74.87 | 54.17 |
| LinFormer | 16.13 | 65.90 | 53.09 | 42.34 | 75.30 | 50.46 |
| Performer | 18.01 | 65.40 | 53.82 | 42.77 | 77.05 | 51.18 |
| Transformer | 37.11 | 65.21 | 79.14 | 42.94 | 71.83 | 59.24 |
| ToST | 37.25 | 66.75 | 79.46 | 46.62 | 69.41 | 59.90 |

With this in mind, and the understanding of causal autoregressive processes from Chapter 1, we say that an encoder $f$ is *causal* if for every input $\boldsymbol{X} = [\boldsymbol{x}_1, \ldots, \boldsymbol{x}_N]$, it holds

$$
\begin{aligned}
f(\boldsymbol{X})_{1:N-1} &= f([\boldsymbol{x}_1, \ldots, \boldsymbol{x}_N])_{1:N-1} = f([\boldsymbol{x}_1, \ldots, \boldsymbol{x}_{N-1}]) && (5.3.19) \\
&= f(\boldsymbol{X}_{1:N-1}), && (5.3.20)
\end{aligned}
$$

where $\boldsymbol{Z}_{1:N-1} = [\boldsymbol{z}_1, \ldots, \boldsymbol{z}_{N-1}]$, etc. In English, this means that the first $N-1$ features are equal to the output of the encoder on the first $N-1$ token embeddings of the input; even simpler, it means we can compute the features of each token *one at a time*, and this would be equivalent to computing the features for the entire input. This causality is necessary for the previously mentioned applications (e.g., LLMs). In particular, the previously presented implementations of CRATE and ToST are *not causal* (proof is left as an exercise).

We can construct causal white-box architectures by a variety of methods. Here, we will showcase a simple method which builds on our previous unrolled optimization framework. Specifically, we compute the features $\boldsymbol{z}_1, \dots, \boldsymbol{z}_N$ corresponding to the first $N$ tokens of the input *one-at-a-time* to optimize the representation learning objective, such as the sparse rate reduction (5.2.5):

$$\max_{\boldsymbol{z}_i} \; \Delta R(\boldsymbol{Z}_{1:i} \mid \boldsymbol{U}_{[K]}) - \lambda \|\boldsymbol{Z}_{1:i}\|_1, \;\; \forall i \in [N]. \tag{5.3.21}$$

If we follow through with the two-step unrolling procedure that yielded CRATE, we can obtain the iteration:

$$\boldsymbol{z}_i^{\ell+1/2} \approx \boldsymbol{z}_i^{\ell} - \kappa \nabla_{\boldsymbol{z}_i} R(\boldsymbol{Z}_{1:i}^{\ell} \mid \boldsymbol{U}_{[K]}^{\ell}), \tag{5.3.22}$$

$$\boldsymbol{z}_i^{\ell+1} \approx \arg\min_{\boldsymbol{z}_i} \left\{ \lambda \|\boldsymbol{z}_i\|_1 + \frac{1}{2}\|\boldsymbol{z}_i^{\ell+1/2} - \boldsymbol{D}^{\ell}\boldsymbol{z}_i\|_F^2 \right\}, \tag{5.3.23}$$

or, using the same conversion from these two steps to network operators that we used for CRATE,

$$\boldsymbol{z}_i^{\ell+1/2} = \left(1 - \frac{\kappa p}{N\epsilon^2}\right)\boldsymbol{z}_i^{\ell} + \frac{\kappa p}{N\epsilon^2}\operatorname{MSSA}(\boldsymbol{Z}_{1:i}^{\ell} \mid \boldsymbol{U}_{[K]}^{\ell})_i, \tag{5.3.24}$$

$$\boldsymbol{z}_i^{\ell+1} = \operatorname{ISTA}(\boldsymbol{z}_i^{\ell+1/2} \mid \boldsymbol{D}^{\ell}). \tag{5.3.25}$$

Let us investigate this iteration in slightly more detail. First, let us note that by construction, this sequence of features corresponds to a causal encoder. Next, let us suppose that we are in a setting where we have computed $\boldsymbol{Z}_{1:i-1}^{\ell+1} = [\boldsymbol{z}_1^{\ell+1}, \dots, \boldsymbol{z}_{i-1}^{\ell+1}]$, having computed the quantities

$$(\boldsymbol{U}_k^{\ell})^{\top}\boldsymbol{Z}_{1:i-1}^{\ell} = [(\boldsymbol{U}_k^{\ell})^{\top}\boldsymbol{z}_1^{\ell}, \dots, (\boldsymbol{U}_k^{\ell})^{\top}\boldsymbol{z}_{i-1}^{\ell}] \tag{5.3.26}$$

along the way, and want to compute $\boldsymbol{z}_i^{\ell+1}$ (for instance relevant to the case of LLM inference). In this case, note that the update rule for $\boldsymbol{z}_i^{\ell+1/2}$ can be simplified as:

$$\boldsymbol{z}_i^{\ell+1/2} \tag{5.3.27}$$

$$= \left(1 - \frac{\kappa p}{N\epsilon^2}\right)\boldsymbol{z}_i^{\ell} + \frac{\kappa p}{N\epsilon^2}\operatorname{MSSA}(\boldsymbol{Z}_{1:i}^{\ell} \mid \boldsymbol{U}_{[K]}^{\ell})_i \tag{5.3.28}$$

$$= \left(1 - \frac{\kappa p}{N\epsilon^2}\right)\boldsymbol{z}_i^{\ell} + \frac{\kappa p^2}{N^2\epsilon^4}\sum_{k=1}^{K}\boldsymbol{U}_k\boldsymbol{U}_k^{\top}\boldsymbol{Z}_{1:i}\operatorname{softmax}((\boldsymbol{U}_k^{\top}\boldsymbol{Z}_{1:i})^{\top}(\boldsymbol{U}_k^{\top}\boldsymbol{z}_i)) \tag{5.3.29}$$

$$= \left(1 - \frac{\kappa p}{N\epsilon^2}\right)\boldsymbol{z}_i^{\ell} + \frac{\kappa p^2}{N^2\epsilon^4}\sum_{k=1}^{K}\boldsymbol{U}_k\boldsymbol{U}_k^{\top}[\boldsymbol{Z}_{1:i-1}, \boldsymbol{z}_i]\operatorname{softmax}((\boldsymbol{U}_k^{\top}[\boldsymbol{Z}_{1:i-1}, \boldsymbol{z}_i])^{\top}(\boldsymbol{U}_k^{\top}\boldsymbol{z}_i)) \tag{5.3.30}$$

$$= \left(1 - \frac{\kappa p}{N\epsilon^2}\right)\boldsymbol{z}_i^{\ell} + \frac{\kappa p^2}{N^2\epsilon^4}\sum_{k=1}^{K}\boldsymbol{U}_k[\boldsymbol{U}_k^{\top}\boldsymbol{Z}_{1:i-1}, \boldsymbol{U}_k^{\top}\boldsymbol{z}_i]\operatorname{softmax}(([\boldsymbol{U}_k^{\top}\boldsymbol{Z}_{1:i-1}, \boldsymbol{U}_k^{\top}\boldsymbol{z}_i])^{\top}(\boldsymbol{U}_k^{\top}\boldsymbol{z}_i)). \tag{5.3.31}$$

This step now becomes *highly efficient* if we cache $\boldsymbol{U}_k^\top \boldsymbol{Z}_{1:i-1}$ from previous steps, and add a single column to it each time we compute the projection $\boldsymbol{U}_k^\top \boldsymbol{z}_i$. Namely, we greatly reduce the number of large matrix-matrix products and replace them by cache loads and matrix-vector products, which is overall much cheaper in terms of time complexity. This caching is the reason why causal generative models such as LLMs can efficiently sample 1000s of tokens per second, even if each training step takes a few seconds by itself. The cache for the subspace projections of the features is also known as the so-called "KV cache".

Finally, let us consider the case where we are to *train* a causal CRATE model, and want to find the most efficient way to do this given a full input sequence $\boldsymbol{X} = [\boldsymbol{x}_1, \ldots, \boldsymbol{x}_N]$ simultaneously. The ISTA step is parallelizable to become the regular ISTA step from the non-causal CRATE, i.e.,

$$\operatorname{ISTA}(\boldsymbol{Z} \mid \boldsymbol{D}) \doteq [\operatorname{ISTA}(\boldsymbol{z}_1 \mid \boldsymbol{D}), \ldots, \operatorname{ISTA}(\boldsymbol{z}_N \mid \boldsymbol{D})], \tag{5.3.32}$$

therefore implying that the ISTA step remains the same as the non-causal CRATE. The MSSA step is more interesting, since it changes in a meaningful way. To see how it changes, note that each MSSA operator has an interesting structure which merits some attention. Namely, if we define the *causal MSSA operator* as the block matrix:

$$\operatorname{CMSSA}(\boldsymbol{Z} \mid \boldsymbol{U}_{[K]}) \doteq [\operatorname{MSSA}(\boldsymbol{Z}_{1:1} \mid \boldsymbol{U}_{[K]})_1, \ldots, \operatorname{MSSA}(\boldsymbol{Z}_{1:N} \mid \boldsymbol{U}_{[K]})_N], \tag{5.3.33}$$

then, working our way through the softmax algebra (proof is again left as an exercise), we can show that

$$\operatorname{CMSSA}(\boldsymbol{Z} \mid \boldsymbol{U}_{[K]}) \doteq \frac{p}{N\epsilon^2}[\boldsymbol{U}_1, \ldots, \boldsymbol{U}_K]\begin{bmatrix}\operatorname{softmax}((\boldsymbol{U}_1^\top \boldsymbol{Z})^\top(\boldsymbol{U}_1^\top \boldsymbol{Z}) + \boldsymbol{M}_N) \\ \vdots \\ \operatorname{softmax}((\boldsymbol{U}_K^\top \boldsymbol{Z})^\top(\boldsymbol{U}_K^\top \boldsymbol{Z}) + \boldsymbol{M}_N)\end{bmatrix}, \tag{5.3.34}$$

$$\text{where } \boldsymbol{M}_N = \begin{bmatrix} 0 & -\infty & -\infty & \ldots & -\infty & -\infty \\ 0 & 0 & -\infty & \ldots & -\infty & -\infty \\ \vdots & \vdots & \vdots & \ddots & \vdots & \vdots \\ 0 & 0 & 0 & \ldots & 0 & -\infty \\ 0 & 0 & 0 & \ldots & 0 & 0 \end{bmatrix}. \tag{5.3.35}$$

Here the matrix $\boldsymbol{M}_N \in \mathbb{R}^{N \times N}$ is another way to encode that no feature in CMSSA, i.e., the causal MSSA, depends on a future feature, since the relevant entries of the argument to the softmax are $-\infty$, and thus are set to 0 after the exponential within the softmax, and therefore effectively ignored. This matrix $\boldsymbol{M}_N$ is sometimes called a *(causal) attention mask*.

The upshot of this is that when we have the full sequence $\boldsymbol{Z}^\ell$, we can compute $\boldsymbol{Z}^{\ell+1}$ in time similar to or less than that of the usual CRATE layer (since $\boldsymbol{M}_N$ is hard-coded, input-independent, and enables us to ignore many entries for the softmax). Thus, we can define and train the full sequence-to-sequence model in the same way as the regular architecture.

Table 5.5: Zero-shot validation cross-entropy loss of the Causal-CRATE-Base model and GPT2-Small, GPT2-Base model evaluated on the test split of the datasets (lower is better).

| | #parameters | OWT | LAMBADA | WikiText | PTB | Avg |
|---|---|---|---|---|---|---|
| GPT2-Base | 124M | 2.85 | 4.12 | 3.89 | 4.63 | 3.87 |
| GPT2-Small | 64M | 3.04 | 4.49 | 4.31 | 5.15 | 4.25 |
| Causal-CRATE-Base | 60M | 3.37 | 4.91 | 4.61 | 5.53 | 4.61 |

When we apply this to language modeling (details to be provided in Chapter 8), we find that we can obtain reasonable results compared to similar-sized empirically designed language models, as shown in Table 5.5.

Despite the theoretical advantages of the causal CRATE architecture, the gap on language modeling (and more generally sequence modeling) tasks still remains. Below, we discuss two strategies for potentially closing this gap; they are each the subject of active lines of research.

*Example* 5.3 (Better Objectives and Unrolling Strategies). One path to consider is that the objective of the sparse rate reduction is not the best for causal sequence data. For example, we may want to ensure that the features have some kind of temporal relationships or structure, e.g., being generated by a linear dynamical system with Gaussian (mixture) initialization and Gaussian step noise. There is a recent but explosively popular line of work on efficiently and explicitly modeling dynamics in the features using so-called *state-space models* (SSMs, named such in contravention of the classical notion of "state space models" discussed in Chapter 1) [GD24; YKH24; YWZ+24]. Such models are empirically promising because they avoid the need for all-pairs attention, dodging the computational and memory complexity of the MSSA operator and its conventional cousin, the multi-head attention. Despite this efficiency gain, they are highly performant and have even been used in some large-scale language models [TZL+25]. By designing more sophisticated objectives and unrolling strategies, we may be able to build a "white-box" SSM architecture, or SSM-transformer hybrid, and thusly close the gap on causal tasks. ■

*Example* 5.4 (Inference-Time Computation Strategies). Another path to improve causal models is to use more so-called *inference-time compute* strategies: during inference (such as sampling from an LLM) find a way to spend more computational resources in order to obtain a better result. If we are able to do this, we can solve harder problems by allocating more compute to the task at hand.

One naive approach to this is to loop the model: feed the final features into the initial sequence-to-sequence layers and enable it to go through more rounds of iterative denoising. This methodology is known in the literature as "looped transformers" (for the case in which the backbone model is an empirically-designed transformers); this approach has been shown to have benefits at some particularly difficult tasks [GRS+23; SDL+25; YLN+23]. However, from our perspective it is more reasonable to loop the iterate through the *same layer* multiple times, taking multiple local denoising or compression steps against the *same model* at each layer before going to the next.

Still another potential methodology relies on our understanding of the low-dimensional structure in the features and model parameters. For example, since the $\boldsymbol{U}_{[K]}$ matrices are supposed to be supports for the subspaces spanned by the features, we could update them at inference-time using an online algorithm such as online (G)PCA. Similar steps may be able to adapt the dictionary at inference time. This would allow us to improve and specialize our local denoising or compression operators at each layer for the features, and thus potentially yielding better performance.

There have been many empirical attempts to improve inference-time compute methods. It is still a very active area of ongoing research. ∎

## 5.4 Summary and Notes

The materials presented in this chapter are based on a series of recent works on this topic, including [CYY+22; WLP+24; WLY+25; WDL+25; YBP+23]. These contributions encompass both theoretical advances and practical methodologies for constructing interpretable deep networks through unrolled optimization. Many of the key results and proofs discussed in this chapter are derived directly from, or inspired by, these foundational works.

The idea of unrolling an optimization algorithm to construct a neural network traces back to the seminal work [GL10]. In this work, the authors demonstrated that sparse coding algorithms—such as the Iterative Shrinkage-Thresholding Algorithm (ISTA)—can be unrolled to form multilayer perceptrons (MLPs), effectively bridging iterative optimization and neural network design. Notably, [MLE19] demonstrated that such unrolled networks are more interpretable, efficient, and effective compared to generic networks.

In this chapter, we build on this perspective to develop principled, white-box deep network architectures by unrolling optimization algorithms that are designed to minimize well-motivated objectives—such as the (sparse) rate reduction objective introduced earlier. This approach not only clarifies the role of each layer in the network but also offers theoretical grounding for architectural choices, moving beyond empirical trial-and-error toward interpretable and goal-driven design. In the following, we compare conventional DNNs, which are typically constructed through empirical design and heuristic tuning, with our mathematically grounded ReduNet architectures:

| | Conventional DNNs | ReduNets |
|---|---|---|
| Objectives | input/output fitting | information gain |
| Deep architectures | trial & error | iterative optimization |
| Layer operators | empirical | projected gradient |
| Shift invariance | CNNs+augmentation | invariant ReduNets |
| Initializations | random/pre-design | forward unrolled |
| Training/fine-tuning | back prop | forward/back prop |
| Interpretability | black box | white box |
| Representations | hidden/latent | incoherent subspaces |

## 5.5 Exercises and Extensions

*Exercise* 5.1. Let $\boldsymbol{Z} = [\boldsymbol{Z}_1, \ldots, \boldsymbol{Z}_K] \in \mathbb{R}^{d\times N}$ with $\boldsymbol{Z}_k \in \mathbb{R}^{d\times N_k}$ for each $k \in [K]$. For some $\alpha > 0$, let

$$R(\boldsymbol{Z}) = \log\det\left(\boldsymbol{I} + \alpha \boldsymbol{Z}\boldsymbol{Z}^\top\right).$$

1. Given any direction $\boldsymbol{D} \in \mathbb{R}^{d\times m}$, please show that $\nabla R(\boldsymbol{Z}) = \alpha \boldsymbol{X}^{-1}\boldsymbol{Z}$ and

   $$\begin{aligned}&\nabla^2 R(\boldsymbol{Z})[\boldsymbol{D},\boldsymbol{D}]\\ &= \alpha \mathrm{Tr}\left(\boldsymbol{X}^{-1}\boldsymbol{D}\boldsymbol{D}^\top\right) - \frac{\alpha^2}{2}\mathrm{Tr}\left(\boldsymbol{X}^{-1}\left(\boldsymbol{Z}\boldsymbol{D}^\top + \boldsymbol{D}\boldsymbol{Z}^\top\right)\boldsymbol{X}^{-1}\left(\boldsymbol{Z}\boldsymbol{D}^\top + \boldsymbol{D}\boldsymbol{Z}^\top\right)\right),\end{aligned}$$

   where $\boldsymbol{X} \doteq \boldsymbol{I} + \alpha \boldsymbol{Z}\boldsymbol{Z}^\top$. *Hint:* Note that

   $$\nabla^2 R(\boldsymbol{Z})[\boldsymbol{D},\boldsymbol{D}] \doteq \left\langle \lim_{t\to 0}\frac{\nabla R(\boldsymbol{Z}+t\boldsymbol{D}) - \nabla R(\boldsymbol{Z})}{t}, \boldsymbol{D}\right\rangle.$$

2. Please show that

   $$R(\boldsymbol{Z}) \le \sum_{k=1}^{K}\log\det\left(\boldsymbol{I} + \alpha \boldsymbol{Z}_k\boldsymbol{Z}_k^\top\right),$$

   where the equality holds if and only if $\boldsymbol{Z}_k^\top \boldsymbol{Z}_l = \boldsymbol{0}$ for all $k \neq l \in [K]$.

3. Given some $\alpha > 0$, let $\alpha_k = N\alpha/N_k$ for each $k \in [K]$. Please derive the closed-form for the first-order critical point of the following function:

   $$f(\boldsymbol{Z}_k) = \frac{1}{2}\log\det\left(\boldsymbol{I} + \alpha \boldsymbol{Z}_k\boldsymbol{Z}_k^\top\right) - \frac{N_k}{2N}\log\det\left(\boldsymbol{I} + \alpha_k \boldsymbol{Z}_k\boldsymbol{Z}_k^\top\right) - \frac{\lambda}{2}\|\boldsymbol{Z}_k\|_F^2.$$

   *Hint:* Let $r_k = \mathrm{rank}(\boldsymbol{Z}_k)$. Consider the following singular value decomposition of $\boldsymbol{Z}_k$:

   $$\boldsymbol{Z}_k = \boldsymbol{P}_k\boldsymbol{\Sigma}_k\boldsymbol{Q}_k^\top = \begin{bmatrix}\boldsymbol{P}_{k,1} & \boldsymbol{P}_{k,2}\end{bmatrix}\begin{bmatrix}\tilde{\boldsymbol{\Sigma}}_k & \boldsymbol{0}\\ \boldsymbol{0} & \boldsymbol{0}\end{bmatrix}\begin{bmatrix}\boldsymbol{Q}_{k,1}^\top\\ \boldsymbol{Q}_{k,2}^\top\end{bmatrix}, \tag{5.5.1}$$

where $\boldsymbol{P}_k \in \mathcal{O}^d$ with $\boldsymbol{P}_{k,1} \in \mathbb{R}^{d\times r_k}$ and $\boldsymbol{P}_{k,2} \in \mathbb{R}^{d\times(d-r_k)}$, $\boldsymbol{\Sigma}_k \in \mathbb{R}^{d\times N_k}$ with $\tilde{\boldsymbol{\Sigma}}_k \in \mathbb{R}^{r_k\times r_k}$ being a diagonal matrix, and $\boldsymbol{Q}_k \in \mathcal{O}^{N_k}$ with $\boldsymbol{Q}_{k,1} \in \mathbb{R}^{N_k\times r_k}$ and $\boldsymbol{P}_{k,2} \in \mathbb{R}^{N_k\times(N_k-r_k)}$.

*Exercise* 5.2 (Neumann series for matrix inverse). Let $\boldsymbol{A} \in \mathbb{R}^{n\times n}$. If $\|\boldsymbol{A}\| < 1$, please show

$$(\boldsymbol{I} - \boldsymbol{A})^{-1} = \sum_{k=0}^{\infty} \boldsymbol{A}^k. \tag{5.5.2}$$

*Hint:* The proof consists of two steps.

(i) **Step 1**: Please show that the infinite series $\sum_{k=0}^{\infty} \boldsymbol{A}^k$ converges when $\|\boldsymbol{A}\| < 1$ using $\|\boldsymbol{A}^k\| \leq \|\boldsymbol{A}\|^k$.

(ii) **Step 2**: Compute the matrix product $(\boldsymbol{I} - \boldsymbol{A})\sum_{k=0}^{\infty} \boldsymbol{A}^k$.

*Exercise* 5.3. In this exercise we will consider a coding rate measure $R_\epsilon^{c,\mu}$ which uses the means of the distributions, i.e., a coding rate for tokens drawn (non-i.i.d.) from the low-rank Gaussian mixture model

$$\boldsymbol{z}_i \sim \frac{1}{K}\sum_{k=1}^{K} \mathcal{N}(\boldsymbol{\mu}_k, \boldsymbol{U}_k\boldsymbol{U}_k^\top). \tag{5.5.3}$$

1. Please argue that a suitable coding rate for $\boldsymbol{Z}$ under this model is

$$R_\epsilon^{c,\mu}(\boldsymbol{Z}) = \sum_{k=1}^{K} R_\epsilon(\boldsymbol{U}_k^\top(\boldsymbol{Z} - \boldsymbol{\mu}_k\mathbf{1}^\top)). \tag{5.5.4}$$

2. Similarly to how we derived MSSA, derive a new attention-like operator from this coding rate. What are the key differences?

*Exercise* 5.4. Please compute the gradients in (5.3.9) and (5.3.10).

*Exercise* 5.5. Please show Corollary 5.1 when $Kp \leq d$.

*Exercise* 5.6. Derive a causal version of the TSSA operator.

**Algorithm 5.1** Training algorithm for ReduNet

**Input:** $\boldsymbol{X} = [\boldsymbol{x}_1, \ldots, \boldsymbol{x}_N] \in \mathbb{R}^{D \times N}$, $\boldsymbol{\Pi} = \{\boldsymbol{\Pi}_k\}_{k=1}^{K}$, $\epsilon > 0$, $\lambda$, and a learning rate $\eta$.

**Output:** The learned parameters $\{\boldsymbol{E}^\ell\}_{\ell=1}^{L}, \{\{\boldsymbol{C}_k^\ell\}_{k=1}^{K}\}_{\ell=1}^{L}, \{\gamma_k\}_{k=1}^{k}$

1: **procedure** ReduNetTraining($\boldsymbol{X}, \boldsymbol{\Pi}, \epsilon, \lambda, \eta$)

    `# Define constants`

2:     $\alpha \leftarrow d/(N\epsilon^2)$

3:     **for** $k \in \{1, \ldots, K\}$ **do**

4:         $\alpha_k \leftarrow D/(\mathrm{tr}(\boldsymbol{\Pi}_k)\epsilon^2)$

5:         $\gamma_k \leftarrow \mathrm{tr}(\boldsymbol{\Pi}_k)/D$

6:     **end for**

    `# ReduNet layer-by-layer iteration`

7:     $\boldsymbol{Z}^1 = \left[\boldsymbol{z}_1^1, \ldots, \boldsymbol{z}_N^1\right] \leftarrow \boldsymbol{X}$     ▷ Initialize the ReduNet per-layer iteration

8:     **for** $\ell \in \{1, \ldots, L\}$ **do**

        `# Step 1: Compute network parameters` $\boldsymbol{E}^\ell, \{\boldsymbol{C}_k^\ell\}_{k=1}^{K}$

9:         $\boldsymbol{E}^\ell \leftarrow \alpha\left(\boldsymbol{I} + \alpha \boldsymbol{Z}^\ell (\boldsymbol{Z}^\ell)^\top\right)^{-1} \in \mathbb{R}^{d \times d}$

10:         **for** $k \in \{1, \ldots, K\}$ **do**

11:             $\boldsymbol{C}_k^\ell \leftarrow \alpha_k\left(\boldsymbol{I} + \alpha_k \boldsymbol{Z}^\ell \boldsymbol{\Pi}_k (\boldsymbol{Z}^\ell)^\top\right)^{-1} \in \mathbb{R}^{d \times d}$

12:         **end for**

        `# Step 2: Update features` $\boldsymbol{Z}^\ell$

13:         **for** $i \in \{1, \ldots, N\}$ **do**

            `# Compute soft assignments` $\hat{\boldsymbol{\pi}}(\boldsymbol{z}_i^\ell)$

14:             $\hat{\boldsymbol{\pi}}(\boldsymbol{z}_i^\ell) \leftarrow \mathrm{softmax}(-\lambda[\|\boldsymbol{C}_1^\ell \boldsymbol{z}_i^\ell\|_2, \ldots, \|\boldsymbol{C}_K^\ell \boldsymbol{z}_i^\ell\|_2]) \in [0,1]^K$

            `# Update features` $\boldsymbol{z}_i^{\ell+1}$ `from` $\boldsymbol{z}_i^\ell$

15:             $\boldsymbol{z}_i^{\ell+1} \leftarrow \mathcal{P}_{\mathbb{S}^{d-1}}\left(\boldsymbol{z}_i^\ell + \eta\left(\boldsymbol{E}^\ell \boldsymbol{z}_i^\ell - \sum_{k=1}^{K} \gamma_k \hat{\pi}_k(\boldsymbol{z}_i^\ell) \boldsymbol{C}_k^\ell \boldsymbol{z}_i^\ell\right)\right) \in \mathbb{R}^d$

16:         **end for**

17:     **end for**

    `# Return all network parameters.`

18:     **return** $\{\boldsymbol{E}^\ell\}_{\ell=1}^{L}, \{\{\boldsymbol{C}_k^\ell\}_{k=1}^{K}\}_{\ell=1}^{L}, \{\gamma_k\}_{k=1}^{K}$

19: **end procedure**

**Algorithm 5.2** Evaluation algorithm for ReduNet

**Input:** Input $\boldsymbol{x} \in \mathbb{R}^D$, network parameters $\{\boldsymbol{E}^\ell\}_{\ell=1}^L, \{\{\boldsymbol{C}_k^\ell\}_{k=1}^K\}_{\ell=1}^L, \{\gamma_k\}_{k=1}^K$, learning rate $\lambda$

**Output:** feature $\boldsymbol{z}^{L+1}$

```
1: procedure REDUNETEVALUATION(x)
       # Initialize the ReduNet per-layer iteration
2:     z^1 ← x ∈ R^D

3:     for ℓ ∈ {1, ..., L} do
           # Compute soft assignments π̂(z^ℓ)
4:         π̂(z^ℓ) ← softmax(−λ [‖C_1^ℓ z^ℓ‖_2, ..., ‖C_K^ℓ z^ℓ‖_2]) ∈ [0, 1]^K

           # Update feature z^{ℓ+1} using z^ℓ
5:         z^{ℓ+1} ← P_{S^{d−1}}(z^ℓ + η(E^ℓ z^ℓ − Σ_{k=1}^K γ_k π̂_k(z^ℓ) C_k^ℓ z^ℓ)) ∈ R^d
6:     end for

       # Return the output features
7:     return z^{L+1}
8: end procedure
```

# Chapter 6

# Consistent and Self-Consistent Representations

> "*Everything should be made as simple as possible, but not any simpler.*"
>
> – Albert Einstein

In the past chapters, we have argued that the fundamental goal of learning is to *identify* a data distribution with low-dimensional support and *transform* it into a compact and structured representation. This entails two related problems:

1. **Distribution learning:** How do we get ahold of the distribution of data $\boldsymbol{X}$, to enable generalizable sampling/generation?
2. **Representation learning:** How do we map the data $\boldsymbol{X}$ to a compact and structured representation $\boldsymbol{Z}$, such that the representation is sufficient to reconstruct $\boldsymbol{X}$?

In Chapters 3 and 4, we have developed a complete computational solution to the distribution learning problem for general data distributions, based on the diffusion-denoising process. In Chapter 4, we have defined a compression-based objective function, the information gain, whose maximization leads to compact and structured representations. And in Chapter 5, we have seen that the layers of popular deep network architectures can be interpreted as incremental optimization steps towards maximizing the information gain. This leaves us poised to present a mathematically-founded solution to the representation learning problem.

However, when we try to achieve a certain objective through optimization, there is no guarantee that the solution $\boldsymbol{Z}$ found in the end by greedy optimization is the correct solution. In fact, even if the optimization process manages to find the globally optimal solution $\boldsymbol{Z}^*$, there is no guarantee that the solution corresponds to a complete representation of the data distribution.[1] Therefore, an outstanding question is: how can we ensure that the learned representation of the data distribution is correct or good enough?

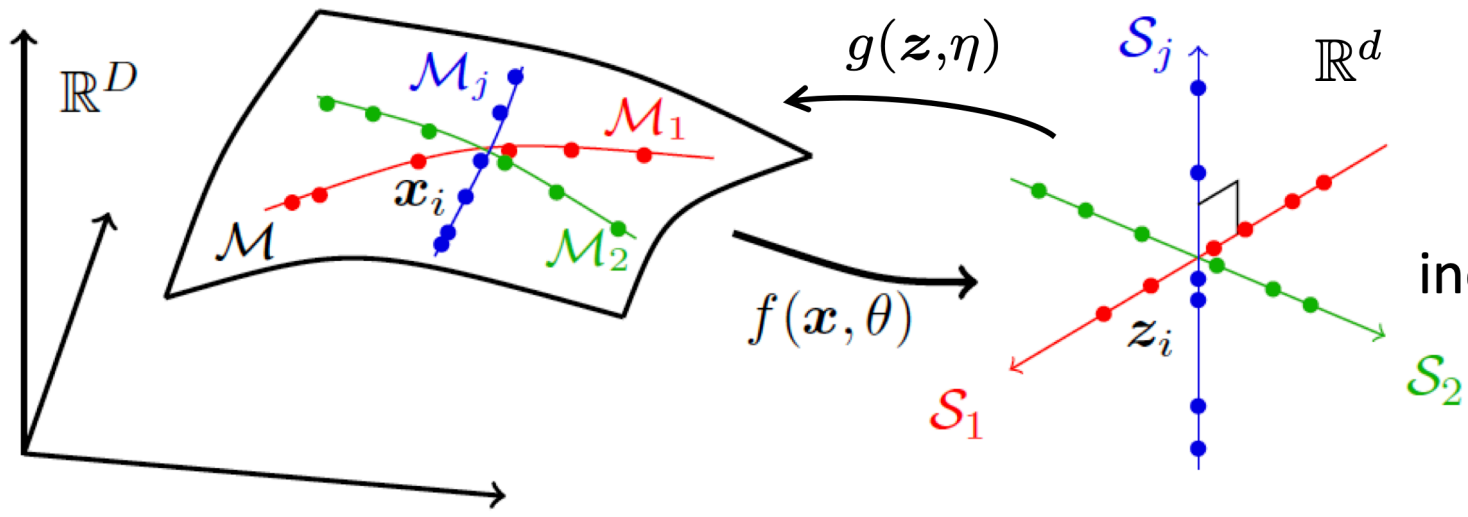

Figure 6.1: Learn a good auto-encoding representation for a general distribution whose supports are on a mixture of low-dimensional submanifolds. Notice that in addition to learning a good representation via a mapping $f(\boldsymbol{x}, \theta)$ in an end-to-end fashion as we did in Chapter 5, we want to make sure that the so-learned representation can be decoded, via an "inverse" mapping $g(\boldsymbol{z}, \eta)$, back to approximate the original data distribution.

Of course, the only way to verify this is to see whether there exists a decoding map, say $g$, that can decode the learned representation to reproduce the original data (distribution) well enough:

$$\boldsymbol{X} \xrightarrow{\mathcal{E}=f} \boldsymbol{Z} \xrightarrow{\mathcal{D}=g} \hat{\boldsymbol{X}} \tag{6.0.1}$$

in terms of some measure of similarity:

$$d(\boldsymbol{X}, \hat{\boldsymbol{X}}). \tag{6.0.2}$$

This leads to the concept of *consistent representation*.[2] As we have briefly alluded to in Chapter 1 (see Figure 1.25), *autoencoding*, which integrates the encoding and decoding processes, is a natural framework for learning such a representation. We have studied some important special cases in Chapter 2 where the supports of the distribution are piecewise linear. In this chapter, Section 6.1, we will study how to extend autoencoding to more general classes of distributions whose supports are nonlinear, as illustrated in Figure 6.1, while enforcing the consistency between $\boldsymbol{X}$ and $\hat{\boldsymbol{X}}$.

Given this classical autoencoding formalism, there is a natural question: *where does distribution learning end and representation learning begin?* As we will see, the answer comes down to *efficiency*. It is typically far more effective, as demonstrated by modern practice, to learn a compact representation for the data distribution as a type of 'preparation' of the distribution, and only then to perform distribution learning on the so-learned representation. This is the dominant methodology in practical problems. We will demonstrate the links between the foundations we have developed in Chapters 4 and 5 and the state of the art in Section 6.1.5, where the "representation autoencoder" framework demonstrates how open-loop learned representations (which we discussed in Section 4.3.2, and will review) can be combined with a simple decoder and a $\boldsymbol{Z}$-space diffusion process to enable efficient, simultaneous representation learning and sampling.

[1]This could be due to many reasons: for example, the data available for learning the distribution might not be sufficient, or the formulation of the optimization program fails to consider some additional constraints or conditions.

[2]Note that here the notion of consistency is empirical or inductive in nature. It is different from the notion of logic and deductive consistency in mathematics, also known as the Hilbert's Principle, which we will have more discussions in Chapter 9.

More fundamentally, in many practical and natural learning scenarios, it can be difficult or even impossible to compare distributions of the data $\boldsymbol{X}$ and $\hat{\boldsymbol{X}}$. We are left with the only option of comparing the learned feature $\boldsymbol{Z}$ with its image $\hat{\boldsymbol{Z}}$ under the encoder $f$:

$$\boldsymbol{X} \xrightarrow{\mathcal{E}=f} \boldsymbol{Z} \xrightarrow{\mathcal{D}=g} \hat{\boldsymbol{X}} \xrightarrow{\mathcal{E}=f} \hat{\boldsymbol{Z}} \tag{6.0.3}$$

This leads to the notion of a *self-consistent representation*. In Section 6.2, we will study when and how we can learn a consistent representation by enforcing the self-consistency between $\boldsymbol{Z}$ and $\hat{\boldsymbol{Z}}$ only through a closed-loop transcription framework.

Furthermore, in many practical and natural learning scenarios, we normally do not have sufficient samples of the data distribution all available at once. For example, animals and humans develop their visual memory by continuously taking in increments of observations throughout their lives. In Section 6.3, we will study how to extend the closed-loop transcription framework to learn a self-consistent representation in a *continuous learning* setting.

Of course, a fundamental motivation for why we ever want to identify the low-dimensional structures in a data distribution and find a good representation is to make it easy to use the data for various tasks of intelligence, such as classification, completion, and prediction. Therefore, the resulting joint representation $(\boldsymbol{x}, \boldsymbol{z})$ must be structured in such a way that is best suited for these tasks. In the following chapter, we will see how the learned representation can be structured to facilitate conditioned completion or generation tasks.

## 6.1 Learning Consistent Representations

Here we give a formal definition of consistent representations, which are closely related to the concept of autoencoding.

**Definition 6.1** (Consistent Representations)**.** Given data $\boldsymbol{X}$, a *consistent representation* is a pair of functions $(f\colon \mathcal{X} \to \mathcal{Z}, g\colon \mathcal{Z} \to \mathcal{X})$, such that the *features* $\boldsymbol{Z} = f(\boldsymbol{X})$ are compact and structured, and the *autoencoding*

$$\hat{\boldsymbol{X}} \doteq g(\boldsymbol{Z}) = g(f(\boldsymbol{X})) \tag{6.1.1}$$

is *close* to $\boldsymbol{X}$ according to one of the following two measures:

1. We say that it is *sample-wise* consistent if $\boldsymbol{X} \approx \hat{\boldsymbol{X}}$ in a certain norm with high probability.

2. We say that the representation is *distributionally consistent* if $\mathrm{Law}(\boldsymbol{X}) \approx \mathrm{Law}(\hat{\boldsymbol{X}})$.

Astute readers may have noticed that if we do not impose certain requirements on the representation $\boldsymbol{Z}$ sought, the above problem has a trivial solution: one may simply choose the functions $f$ and $g$ to be the identity map! Hence, the true purpose of seeking an autoencoding is to ensure that the $\boldsymbol{Z}$ so obtained is both more compact and more structured than $\boldsymbol{X}$. As we have seen in Chapter 5, it suffices to design the representation to maximize the rate reduction

$$\Delta R_\epsilon(\boldsymbol{Z}). \tag{6.1.2}$$

From the definition of consistent representation, it is required that the representation $\boldsymbol{Z}$ be sufficient to recover the original data distribution $\boldsymbol{X}$ to some degree of accuracy. For sample-wise consistency, a typical choice is to minimize the expected reconstruction error:

$$d(\boldsymbol{X}, \hat{\boldsymbol{X}}) = \mathbb{E}[\|\boldsymbol{X} - \hat{\boldsymbol{X}}\|_2^2]. \tag{6.1.3}$$

For consistency in distribution, a typical choice is to minimize a certain distributional distance such as the KL divergence[3]:

$$d(\boldsymbol{X}, \hat{\boldsymbol{X}}) = \mathcal{D}_{KL}(\boldsymbol{X}\|\hat{\boldsymbol{X}}). \tag{6.1.4}$$

Hence, computation aside, when we seek a good autoencoding for a data distribution $\boldsymbol{X}$, conceptually we try to find an encoder $f$ and decoder $g$ such that

$$\min_{f,g}[-\Delta R_\epsilon(\boldsymbol{Z}) + d(\boldsymbol{X}, \hat{\boldsymbol{X}})]. \tag{6.1.5}$$

Notice that the quantization error $\epsilon$ plays the role of the tradeoff parameter in this Lagrangian relaxation.

In the remainder of this section, we will study how to solve such autoencoding problems under different conditions, from simple and ideal cases to increasingly more challenging and realistic conditions. We will culminate with an efficient instantiation of (6.1.5) (in Section 6.1.5) that ensures the representation space $\boldsymbol{Z}$ is compact and structured while also allowing for accurate reconstruction and efficient sampling.

[3]Note that for distributions without common support, which is typical for degenerate distributions, KL divergence may not even be well-defined. In fact, much of the distribution learning literature is trying to address this technical difficulty by replacing or approximating it with something well-defined and efficiently computable.

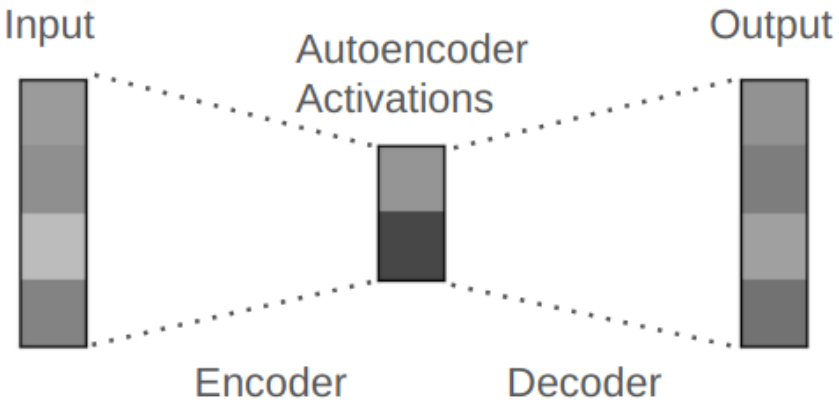

Figure 6.2: Illustration of a typical autoencoder such as PCA, seeking a low-dimensional representation $\boldsymbol{z}$ of high-dimensional data $\boldsymbol{x}$.

### 6.1.1 Linear Autoencoding via PCA

According to [Bal11], the phrase "autoencoder" was first introduced by Hinton and Rumelhart [RHW86a] so that a deep representation could be learned via back propagation (BP) in a self-supervised fashion—reconstructing the original data is the self-supervising task. However, the very same concept of seeking a compact and consistent representation has its roots in many classic studies. As we have already seen in Chapter 2, the classical PCA, ICA, and sparse dictionary learning all share a similar goal. The only difference is that when the underlying data distribution is simple (linear and independent), the encoding or decoding mappings become easy to represent and learn: they do not need to be deep and often can be computed in closed form or with an explicit algorithm.

It is instructive to see how the notion of consistency we have defined plays out in the simple case of PCA: here, the consistent encoding and decoding mappings are given by a single-layer linear transform:

$$\boldsymbol{X} \xrightarrow{\mathcal{E}=\boldsymbol{U}^\top} \boldsymbol{Z} \xrightarrow{\mathcal{D}=\boldsymbol{U}} \hat{\boldsymbol{X}}, \tag{6.1.6}$$

where $\boldsymbol{U} \in \mathbb{R}^{D\times d}$ typically with $d \ll D$. Hence $\boldsymbol{U}^\top$ represents a projection from a higher-dimensional space $\mathbb{R}^D$ to a lower one $\mathbb{R}^d$, as illustrated in Figure 6.2.

As we saw in Chapter 2, when the distribution of $\boldsymbol{x}$ is indeed supported on a low-dimensional subspace $\boldsymbol{U}_o$, the compactness of the representation $\boldsymbol{z}$ produced by $\mathcal{E}$ is a direct consequence of correctly estimating (and enforcing) the dimension of this subspace. Recall that gradient descent on the reconstruction criterion exactly yields these sample-wise consistent mappings: indeed, the optimal solution to the problem

$$\min_{\boldsymbol{U}^\top\boldsymbol{U}=\boldsymbol{I}} \mathbb{E}_{\boldsymbol{x}}\left[\left\|\boldsymbol{x} - \boldsymbol{U}\boldsymbol{U}^\top\boldsymbol{x}\right\|_2^2\right] \tag{6.1.7}$$

precisely coincides with $\boldsymbol{U}_o$ when the dimension of the representation is sufficiently large. In this case, we obtain sample-wise consistency for free, since this guarantees that $\boldsymbol{U}_o\boldsymbol{U}_o^\top\boldsymbol{x} = \boldsymbol{x}$. Notice that in the case of PCA, the rate reduction term in (6.1.5) becomes void as the regularization on the representation $\boldsymbol{Z}$ sought is explicit: It spans the entire subspace of lower dimension[4].

[4]One may view $\Delta R = 0$ in this case.

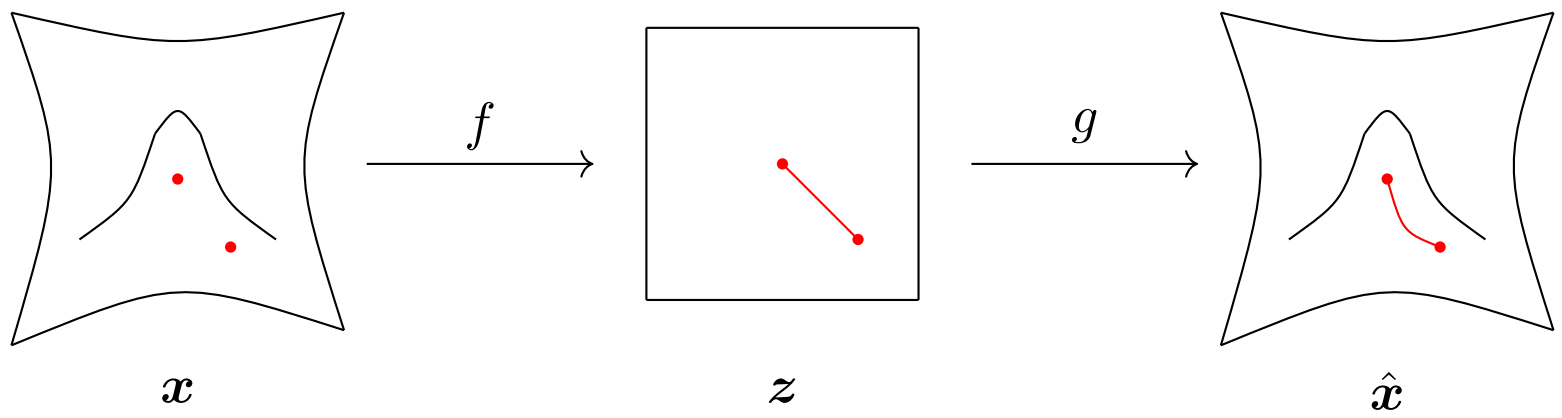

Figure 6.3: A depiction of interpolation through manifold flattening on a manifold in $\mathbb{R}^3$ of dimension $d = 2$. To interpolate two points on the data manifold, map them through the flattening map $f$ to the flattened space, take their convex interpolants, and then map them back to the data manifold through the reconstruction map $g$.

### 6.1.2 Nonlinear PCA and Autoencoding

Of course, one should expect that things will no longer be so simple when we deal with more complex distributions whose underlying low-dimensional structures are nonlinear.

**Data on a Nonlinear Submanifold.** So, to move beyond the linear structure addressed by PCA, we may assume that the data distribution lies on a (smooth) submanifold $\mathcal{M}$. The intrinsic dimension of the submanifold, say $d$, is typically much lower than the dimension of the ambient space $\mathbb{R}^D$. From this geometric perspective, we typically want to find a nonlinear mapping $f$ such that the resulting manifold $f(\mathcal{M})$ is flattened, as illustrated by the example shown in Figure 6.3. The resulting feature $\boldsymbol{z}$-space is typically more compact (of lower-dimension) than the $\boldsymbol{x}$-space, and the manifold is flat. From the statistical perspective, which is complementary to the geometric perspective but distinct in general, we may also want to ensure that the data distribution on $\mathcal{M}$ is mapped to a sufficiently regular distribution, say a Gaussian or uniform distribution (with very low-dimensional support), in the $\boldsymbol{z}$-space. In general, the problem of learning such an autoencoding mapping for this class of data distributions is known as *nonlinear principal component analysis* (NLPCA).

**A Classical Attempt via a Two-Layer Network.** As we have seen above, in the case of PCA, a single-layer linear neural network is sufficient. That is no longer the case for NLPCA. In 1991, Kramer [Kra91] proposed to solve NLPCA by using a two-layer neural network to represent the encoder mapping $f$ (or its inverse $g$) based on the universal representation property of two-layer networks with sigmoid activation:

$$\boldsymbol{z} = \boldsymbol{W}_2\sigma(\boldsymbol{W}_1\boldsymbol{x} + \boldsymbol{b}), \tag{6.1.8}$$

where $\sigma(\,\cdot\,)$ is the sigmoid function:

$$\sigma(x) = \frac{1}{1 + e^{-x}}. \tag{6.1.9}$$

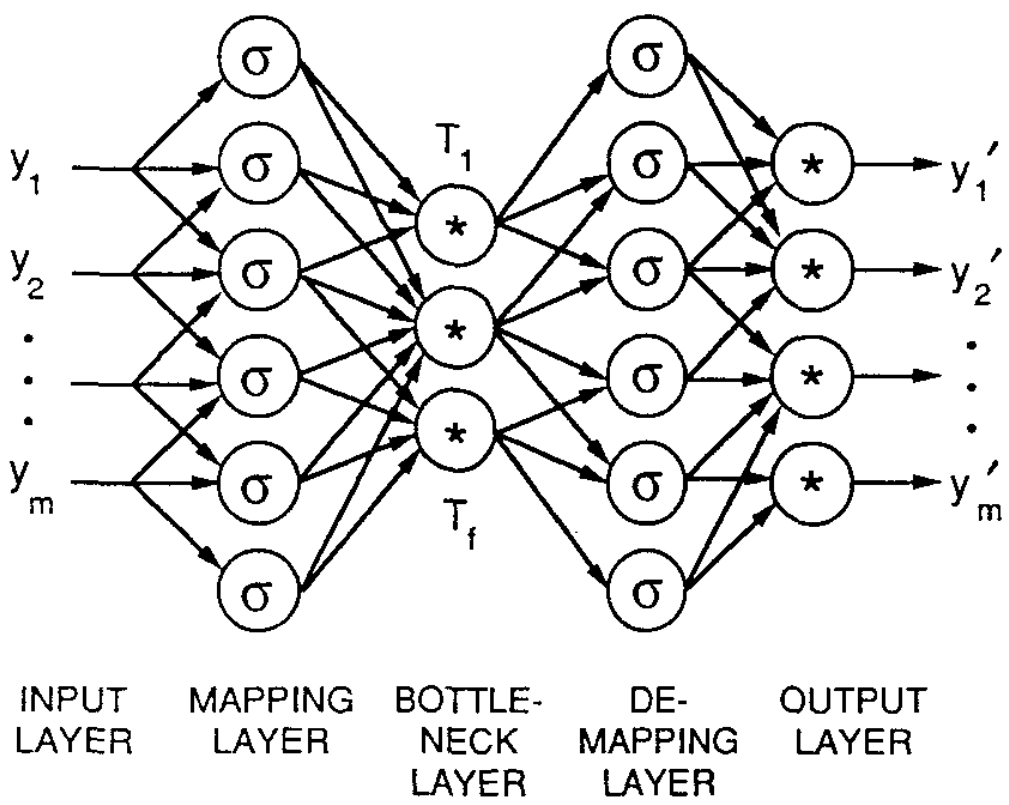

Figure 6.4: Nonlinear PCA by autoassociative neural networks of depth two for both the encoding and decoding mappings, suggested by Kramer [Kra91].

Cybenko [Cyb89] showed that functions of the above form (with enough hidden nodes) can approximate any smooth nonlinear function, say the encoder $f(\cdot)$, to an arbitrary precision. In particular, they can represent the flattening and reconstruction maps for data distributions supported on unions of low-dimensional manifolds, as in Figure 6.3. The overall architecture of the original networks proposed by Kramer is illustrated in Figure 6.4.

Unfortunately, unlike the case of PCA above, there is in general no closed-form learning scheme for the parameters $\boldsymbol{\theta} = (\boldsymbol{W}, \boldsymbol{b})$ of these networks. Hence it was proposed to train the network via back propagation with the supervision of reconstruction error:

$$\min_{\boldsymbol{\theta}} \mathbb{E}[\|\boldsymbol{x} - \hat{\boldsymbol{x}}(\boldsymbol{\theta})\|_2^2]. \tag{6.1.10}$$

Compared to the simple case of PCA, we utilize the same reconstruction objective for learning, but a far more complex nonlinear class of models for parameterizing and learning the encoder and decoder. Although universal approximation properties such as Cybenko's suggest that *in principle* learning consistent autoencoders via this framework is possible—because for any random sample of data, given enough parameters, such autoencoding pairs exist—one often finds it highly nontrivial to find them using gradient descent. Moreover, to obtain a sufficiently informative reconstruction objective and model for the distribution of high-dimensional real-world data such as images, the required number of samples and hidden nodes can be huge. In addition, as a measure of the compactness of the learned representation, the (lower) dimension of $\boldsymbol{z}$ for the bottleneck layer is often chosen heuristically.[5]

[5] In the later work [HZ93], Hinton et al. suggested to use the minimum description length (MDL) principle to promote the compactness of the learned coding scheme, in a spirit very similar to the rate distortion measure introduced in this book.

**Manifold Flattening via a Deeper Network.** Based on the modern practice of deep networks, such classical shallow and wide network architectures are known to be rather difficult to train effectively and efficiently via back propagation (BP), partly due to the vanishing gradient of the sigmoid function. Hence, the modern practice normally suggests further decomposing the nonlinear transform $f$ (or $g$) into a composition of many more layers of simpler transforms, resulting in a deeper network architecture [HS06], as illustrated in Figure 6.5.

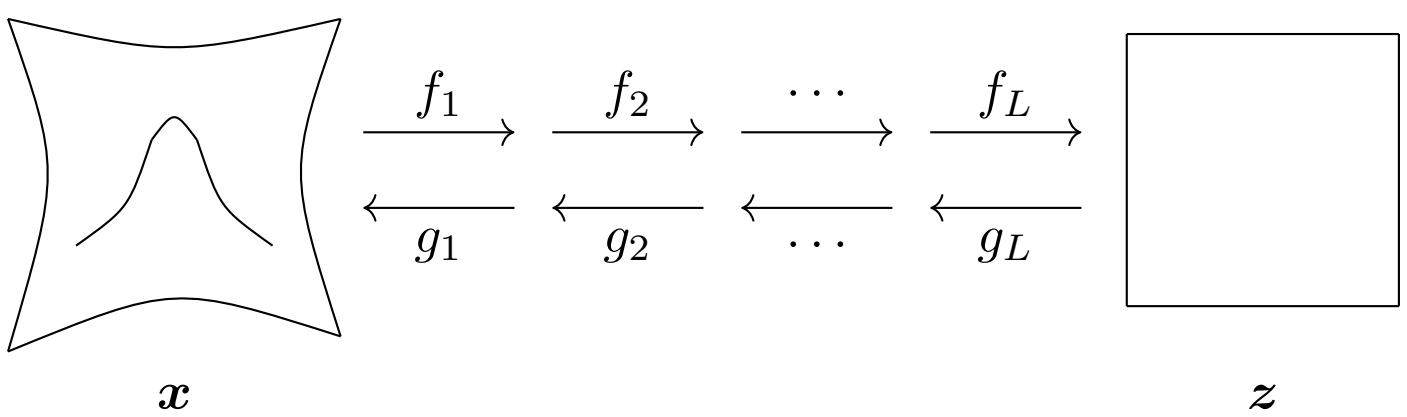

Figure 6.5: A depiction of the construction process of the flattening and reconstruction pair $(f, g)$, where the encoder $f = f_L \circ f_{L-1} \circ \cdots \circ f_1$ is constructed from composing flattening layers, and the decoder $g = g_1 \circ g_2 \circ \cdots \circ g_L$ is composed of inversions of each $f_\ell$.

In light of universal approximation theorems such as Cybenko's, one may initially wonder why, conceptually, deeper autoencoders should be preferred over shallow ones. From purely an expressivity perspective, we can understand this phenomenon through a geometric angle related to the task of flattening the nonlinear manifold on which our hypothesized data distribution is supported. A purely constructive approach to flattening the manifold proceeds incrementally, in parallel to what we have seen in Chapters 3 to 5 with the interaction between diffusion, denoising, and compression. In the geometric setting, the incremental *flattening process* corresponding to $f_\ell$ takes the form of transforming a neighborhood of one point of the manifold to be closer to a flat manifold (i.e., a subspace), and enforcing local consistency with the rest of the data samples; the corresponding incremental operation in the decoder, $g_\ell$, undoes this transformation. This procedure precisely incorporates curvature information about the underlying manifold, which is estimated from data samples. Given enough samples from the manifold and a careful instantiation of this conceptual process, it is possible to implement this sketch as a computational procedure that verifiably flattens nonlinear manifolds in a white-box fashion [PPR+24]. However, the approach is limited in its applicability to high-dimensional data distributions such as images due to unfavorable scalability, motivating the development of more flexible methods for incremental autoencoding.

### 6.1.3 Sparse Autoencoding

In the above autoencoding schemes, the dimension of the feature space $d$ is typically chosen to be much lower than that of the original data space $D$ in order to explicitly enforce or promote the learned representation to be low-dimensional. However, in practice, we normally do not know the intrinsic dimension of the data distribution. Hence, the choice of the feature space dimension for autoencoding is often done empirically. In more general situations, the data distribution can be a mixture of a few low-dimensional subspaces or submanifolds. In these cases, it is no longer feasible to enforce a single low-dimensional space for all features together.

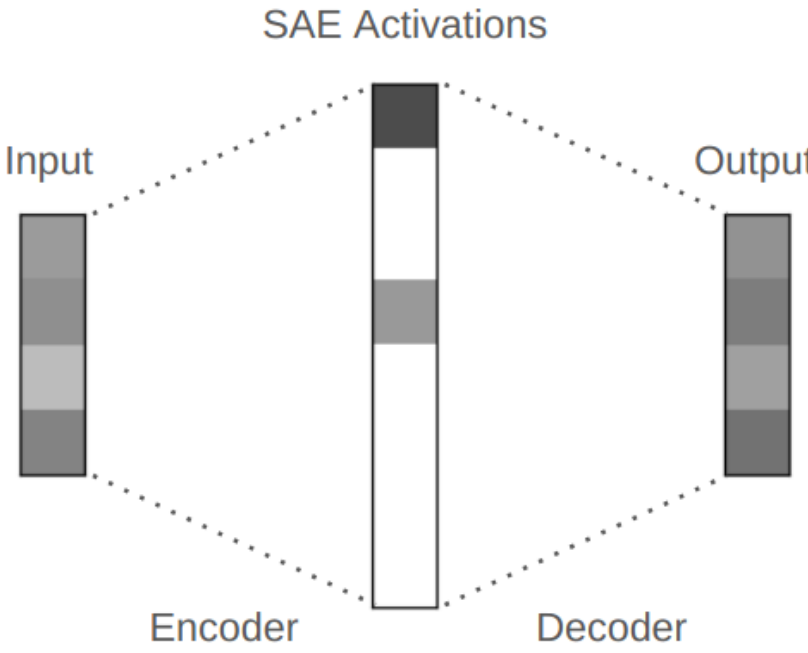

Figure 6.6: Illustration of a sparse autoencoder (SAE), compared to that of a typical autoencoder (AE) in Figure 6.2.

The sparse autoencoder is meant to resolve some of these limitations. In particular, the dimension of the feature space can be equal to or even higher than that of the data space, as illustrated in Figure 6.6. However, the features are required to be highly sparse in the feature space. So if we impose sparsity as a measure of parsimony in addition to rate reduction in the objective (6.1.5), we obtain a new objective for sparse autoencoding:

$$\min_{f,g} [\|\boldsymbol{Z}\|_0 - \Delta R_\epsilon(\boldsymbol{Z}) + d(\boldsymbol{X}, \hat{\boldsymbol{X}})], \tag{6.1.11}$$

where the $\ell^0$-"norm" $\|\cdot\|_0$ is known to promote sparsity. This is very similar to the sparse rate reduction objective (5.2.4) which we used in the previous Chapter 5 to derive the white-box CRATE architecture.

As a method for learning autoencoding pairs in an end-to-end fashion, sparse autoencoding has been practiced in the past [LRM+12; RPC+06], but nearly all modern autoencoding frameworks are instead based on a different, probabilistic autoencoding framework, which we will study now.

### 6.1.4 Variational Autoencoding

An improved, more modern methodology for autoencoding that still finds significant application to this day is *variational autoencoding* [KW13a; KW19]. It formulates the representation learning problem in probabilistic terms, as the search for a model $p(\boldsymbol{x}; \boldsymbol{\theta})$ that maximizes the likelihood of a data sample (or more generally, the population data distribution):

$$\max_{\boldsymbol{\theta}} \mathbb{E}_{\boldsymbol{x}}[\log p(\boldsymbol{x};\, \boldsymbol{\theta})].$$

By conditioning we may write $p(\boldsymbol{x}, \boldsymbol{z};\, \boldsymbol{\theta}) = p(\boldsymbol{z};\, \boldsymbol{\theta})p(\boldsymbol{x} \mid \boldsymbol{z};\, \boldsymbol{\theta})$, and

$$\begin{aligned} p(\boldsymbol{x}; \boldsymbol{\theta}) &= \int p(\boldsymbol{z};\, \boldsymbol{\theta})p(\boldsymbol{x} \mid \boldsymbol{z};\, \boldsymbol{\theta})\mathrm{d}\boldsymbol{z} \\ &= \mathbb{E}_{\boldsymbol{z}\sim p(\,\cdot\,;\, \boldsymbol{\theta})}[p(\boldsymbol{x} \mid \boldsymbol{z};\, \boldsymbol{\theta})]. \end{aligned}$$

This implies that a probabilistic model for $\boldsymbol{x}$ can be learned with a *prior* $p(\boldsymbol{z})$ and a conditional distribution $p(\boldsymbol{x} \mid \boldsymbol{z})$, corresponding more generally to a decoder. Technically, Bayes' rule then implies that the *posterior* $p(\boldsymbol{z} \mid \boldsymbol{x})$, necessary for encoding, can be expressed in terms of the prior and the decoder (and samples of $\boldsymbol{x}$), but this is computationally intractable. Hence, in the VAE framework, we introduce an "approximate posterior" $q(\boldsymbol{z} \mid \boldsymbol{x};\, \boldsymbol{\eta})$, corresponding to a separate encoder network, and jointly learn the encoder and decoder through the maximization of a lower bound on the true likelihood, known as the evidence lower bound (ELBO), which is tight when the posterior approximation is tight.

The following instantiation is standard in practice. We use the following Gaussian distributions:

$$\begin{aligned} \boldsymbol{z} &\sim \mathcal{N}(\mathbf{0}, \boldsymbol{I}), \\ \boldsymbol{x} \mid \boldsymbol{z} &\sim \mathcal{N}(g_1(\boldsymbol{z}), \mathrm{diag}(e^{g_2(\boldsymbol{z})})\boldsymbol{I}), \end{aligned}$$

where $g = (g_1, g_2)$ are deep networks with parameters $\boldsymbol{\theta}$, which correspond to the *decoder* in the autoencoder. Similarly, for the approximate posterior, we use a Gaussian distribution with parameters given by an encoder MLP $f = (f_1, f_2)$ with parameters $\boldsymbol{\eta}$:

$$\boldsymbol{z} \mid \boldsymbol{x} \sim \mathcal{N}(f_1(\boldsymbol{x}), \mathrm{diag}(e^{f_2(\boldsymbol{x})})\boldsymbol{I}).$$

This makes probabilistic encoding and decoding simple: we simply map our data $\boldsymbol{x}$ to the mean and variance parameters of a Gaussian distribution for encoding, and vice versa for decoding. After the aforementioned manipulations, the final VAE objective reads:

$$\max_{\boldsymbol{\theta},\boldsymbol{\eta}} \mathbb{E}_{\boldsymbol{x}}\mathbb{E}_{\boldsymbol{z}\sim q(\,\cdot\,\mid \boldsymbol{x};\, \boldsymbol{\eta})}\left[\log \frac{p(\boldsymbol{x}, \boldsymbol{z};\, \boldsymbol{\theta})}{q(\boldsymbol{z} \mid \boldsymbol{x};\, \boldsymbol{\eta})}\right] \tag{6.1.12}$$

$$\equiv \max_{\boldsymbol{\theta},\boldsymbol{\eta}} \left\{2\mathbb{E}_{\boldsymbol{x}}\left[\mathbb{E}_{\boldsymbol{z}\sim q(\,\cdot\,\mid \boldsymbol{x};\, \boldsymbol{\eta})}\left[\log p(\boldsymbol{x} \mid \boldsymbol{z};\, \boldsymbol{\theta})\right] + \langle f_2(\boldsymbol{x}) - e^{f_2(\boldsymbol{x})}, \mathbf{1}\rangle - \|f_1(\boldsymbol{x})\|_2^2\right]\right\}. \tag{6.1.13}$$

We can understand this as a regularized autoencoding objective, where there is an interesting additional consistency enforcement in the first term of the ELBO.

At the same time, the reader should note the numerous arbitrary decisions the VAE formalism makes (e.g., the choice and parameterization of the Gaussian distributions). This leads to a complex training objective that remains a challenge to optimize efficiently despite more than ten years of research. Recent representation learning methodologies have made notable improvements to this baseline, and they can even be formulated in the white-box framework we have developed in Chapter 5, as we will soon see.

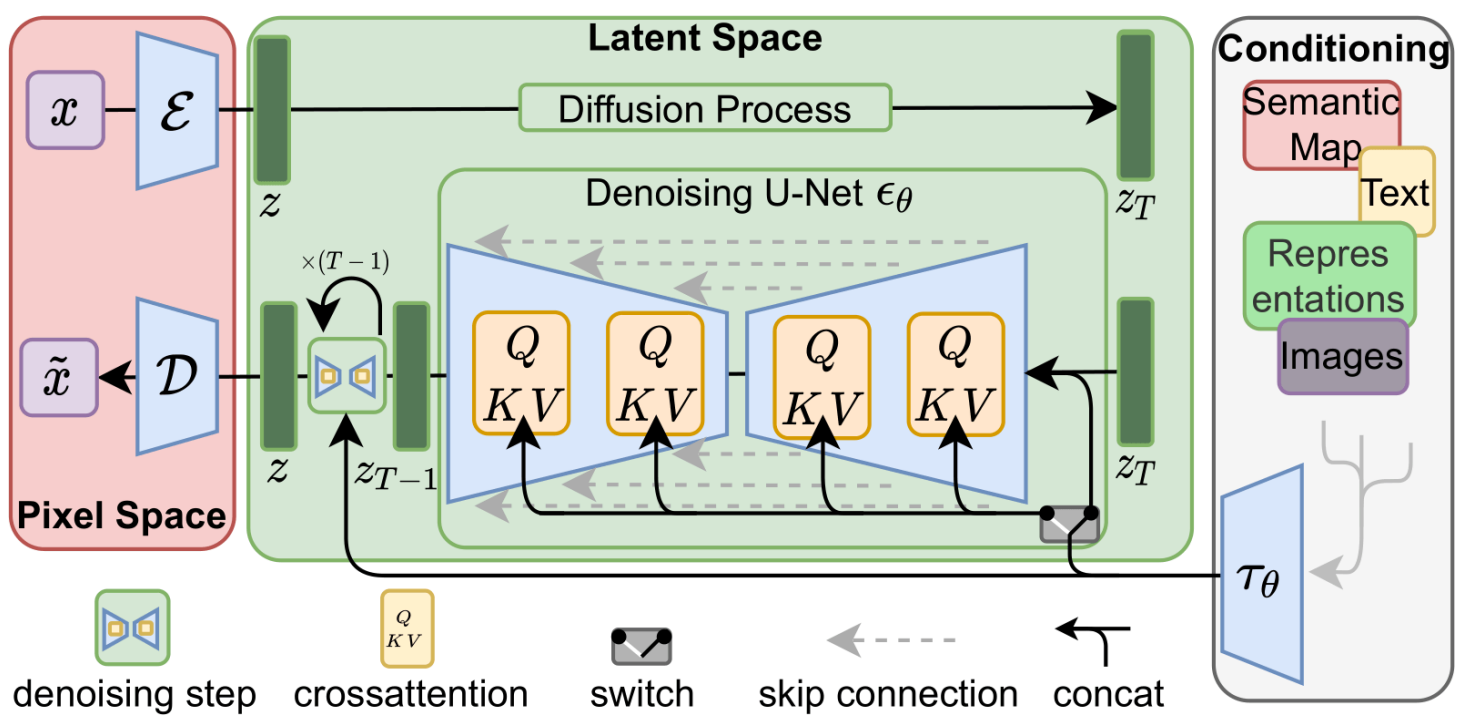

Figure 6.7: Diagram of the Latent Diffusion Model from the work of [RBL+22] which utilizes a variational autoencoder (red box on the left) to learn a latent representation of the data $\boldsymbol{x}$. Then a diffusion/denoising model (green box in the middle) is learned to access the representation $\boldsymbol{z}$. The block on the right is about how to control the denoising sampling process via a (text-based) control signal which we will study in Chapter 7.

**VAE-Based Latent Diffusion.** An important application of VAEs is in the context of latent diffusion models [RBL+22], where a VAE is first trained to learn a more compact and regulated latent representation of high-dimensional data such as images. Then a diffusion model can be trained in the latent space to access the distribution of the representation and hence generate new samples from it, as illustrated in Figure 6.7. This approach enables efficient generation of high-quality images by leveraging the highly-compressed structure of the learned latent space.[6] However, training VAEs can be challenging due to issues such as posterior collapse and balancing the reconstruction and regularization terms in the ELBO. Moreover, the Gaussian posteriors in VAEs are applied per data point and no joint structure is learned across samples, which can severely limit the quality of the learned representations. These challenges have motivated the development of alternative approaches, such as representation autoencoders (RAEs) to be introduced next, which aim to address some of the limitations of VAEs while still providing effective latent representations for generative modeling tasks.

[6]Many popular early image or video generative models are based on this latent diffusion framework, as we describe in further detail in Section 7.4.2 and in implementation terms in Chapter 8.

### 6.1.5 Joint Representation and Distribution Learning

The previous approaches to training autoencoding pairs for representation learning share a common limitation: they do not *explicitly organize* the representation space $\boldsymbol{Z}$, but only encourage it indirectly (e.g., by enforcing sparsity in the sparse autoencoder or Gaussianity in VAE). As we hinted at the start of the chapter, we have seen previously (in Chapter 4) how to do much better, by learning a representation $\boldsymbol{Z}$ via maximizing the rate reduction $\Delta R_{\epsilon}(\boldsymbol{Z})$ while ensuring reconstruction consistency. We describe this procedure in more detail in this section, in particular also demonstrating how to *efficiently sample* in the learned representation space $\boldsymbol{Z}$.

As we have seen at the end of Chapter 4 (Section 4.3), we often can learn a good representation $\boldsymbol{Z}$ of the data $\boldsymbol{X}$:

$$f : \boldsymbol{X} \to \boldsymbol{Z}, \tag{6.1.14}$$

based on some end-to-end supervised or self-supervised methods, including the MCR$^2$ and SimDINO methods introduced in Section 4.3.1 and Section 4.3.2, respectively. As we have also argued, the so-learned representations have better structures than the original data distribution. The representations tend to be a mixture of low-dimensional linear subspaces (or low-rank Gaussians). However, we have not answered the question of how to assess the goodness of the so-learned representations. That is, can we establish from them a reverse decoder:

$$g : \boldsymbol{Z} \to \hat{\boldsymbol{X}} \tag{6.1.15}$$

such that $\hat{\boldsymbol{X}}$ and $\boldsymbol{X}$ are close? In light of the preceding examples of classical autoencoding setups, we should expect that an objective of the form (6.1.5) should work—but what sample-wise consistency loss should be used?

Furthermore, those low-dimensional (linear) structures are still embedded in a high-dimensional representation space. Hence, in order to use the so-learned representations, we need to be able to access their distributions too. In Chapter 3, we have learned that denoising is a general approach to identify and hence access such low-dimensional representations. Hence, we can access the distribution of the learned features via a (learned) denoising process from a random noise:

$$h : \boldsymbol{G} \to \boldsymbol{Z}. \tag{6.1.16}$$

In addition, we have also learned that if the target distribution of $\boldsymbol{Z}$ can be modeled (approximately) as a mixture of subspaces or low-rank Gaussians, their optimal denoising operators share a similar structure to those found in Transformers, or in CRATE introduced in Chapter 5.

Thus, altogether, we want to learn an autoencoder for the data $\boldsymbol{X}$ whose representation $\boldsymbol{Z}$ can be easily accessed. Concretely, we would like to learn an encoder $f$ and a decoder $g$ such that:

1. The representation $\boldsymbol{Z} = f(\boldsymbol{X})$ is a consistent representation of the original data $\boldsymbol{X}$ as defined in Definition 6.1.

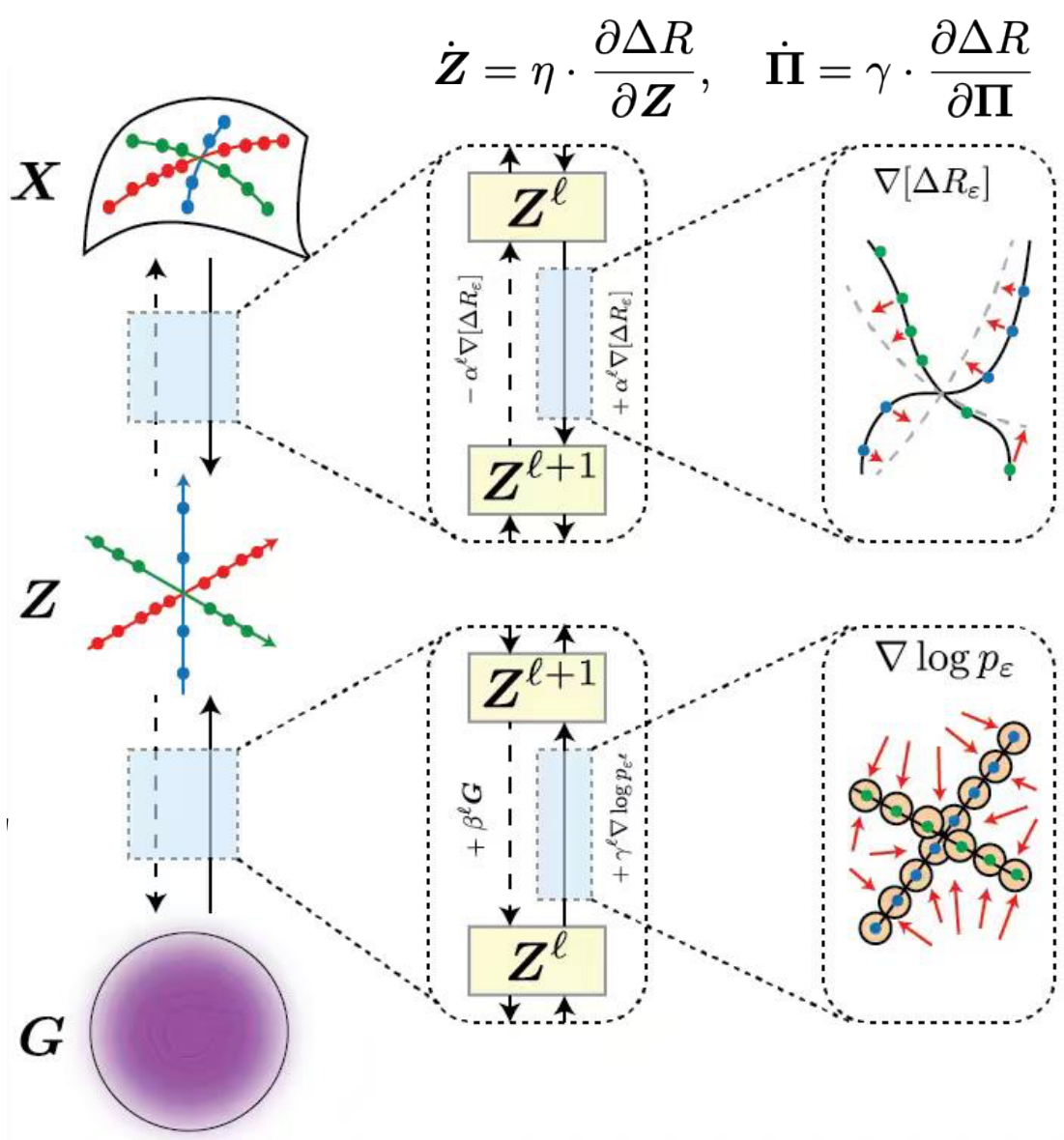

Figure 6.8: Illustration of learning an autoencoding representation $\boldsymbol{Z}$ for the data $\boldsymbol{X}$ (top) and a diffusion/denoising model (bottom) to access the representation $\boldsymbol{Z}$.

2. The structured representation $\boldsymbol{Z}$ can be modeled approximately as a mixture of low-dimensional Gaussians so that we can learn a denoiser to efficiently access the distribution of $\boldsymbol{Z}$ from random Gaussian noise $\boldsymbol{G}$.

Figure 6.8 illustrates the above overall processes.

From Chapter 4, we know that the learned representation $\boldsymbol{Z}$ by optimizing the (unsupervised) MCR$^2$ objective satisfies the second requirement and can be viewed to have structures close to a mixture of subspaces or low-rank Gaussians. By incorporating a sample-consistency objective, we can extend the rate reduction objective in Section 4.3.2 to learn a consistent representation:

$$\min_{f,g}[-\Delta R_\epsilon(\boldsymbol{Z}) + d(\boldsymbol{X}, \hat{\boldsymbol{X}})], \tag{6.1.17}$$

where $d(\boldsymbol{X}, \hat{\boldsymbol{X}})$ is a measure of similarity between the original data $\boldsymbol{X}$ and the reconstructed data $\hat{\boldsymbol{X}} = g(\boldsymbol{Z})$. In practice, because the objective $\Delta R_\epsilon(\boldsymbol{Z})$ alone already promotes low-dimensional structure of $\boldsymbol{Z}$, we can pretrain the encoder $f$ to maximize $\Delta R_\epsilon(\boldsymbol{Z})$ first, and then learn the decoder $g$ to minimize the reconstruction loss $d(\boldsymbol{X}, \hat{\boldsymbol{X}})$ while fixing the encoder $f$. This two-step approach has been demonstrated to be effective, as we will see in Section 8.6.2 of Chapter 8.

Once we have learned such a consistent representation $\boldsymbol{Z}$, we can then learn a denoiser to access its distribution, as we have studied in Chapter 3. Concretely, suppose we have a probabilistic model $p(\boldsymbol{z})$ that captures the (marginal) distribution of the learned representations $\boldsymbol{Z}$ and an initial standard Gaussian distribution $\mathcal{N}(\mathbf{0}, \boldsymbol{I})$. Our goal is to learn a denoiser $\bar{\boldsymbol{z}}$ such that it iteratively denoises a sample from the initial distribution to the target distribution of the learned representations. This denoiser can be trained using objectives introduced in Chapter 3 (see also Chapter 7 for conditional denoisers). For instance, take the simple case of flow matching (recall Exercise 3.5 and Section 3.2.2). We can define the denoiser as a time-dependent vector field $\bar{\boldsymbol{z}}_\theta(t, \boldsymbol{z}_t)$ parameterized by $\theta$ that transforms a sample from the initial distribution to the target distribution over time $t \in [0, 1]$. Here, $\boldsymbol{z}_t$ is an interpolation between $\boldsymbol{z}_0$ and $\boldsymbol{z}_1$ at time $t$:

$$\boldsymbol{z}_t = (1 - t)\boldsymbol{z}_0 + t\boldsymbol{z}_1, \tag{6.1.18}$$

where $\boldsymbol{z}_0 \sim p(\boldsymbol{z})$ is a sample from the learned representation distribution and $\boldsymbol{z}_1 \sim \mathcal{N}(\mathbf{0}, \boldsymbol{I})$. The training objective for the denoiser can then be expressed as:

$$\min_\theta \mathcal{L}_{\mathrm{FM}}(\theta) = \mathbb{E}_{\boldsymbol{z}_0 \sim p(\boldsymbol{z}), \boldsymbol{z}_1 \sim \mathcal{N}(\mathbf{0}, \boldsymbol{I}), t \sim \mathrm{Unif}(0,1)} \left[ \|\bar{\boldsymbol{z}}_\theta(t, \boldsymbol{z}_t) - (\boldsymbol{z}_1 - \boldsymbol{z}_0)\|_2^2 \right]. \tag{6.1.19}$$

It is worth noting the difference between learning a denoiser for the representations learned via VAE and those learned via methods like MCR$^2$ or SimDINO. As discussed in Section 6.1.4, the VAE representations are enforced to approximate a Gaussian posterior *per sample* and no joint distribution of the latent space is modeled, making the representation space less structured and the denoising process potentially more challenging. In contrast, representations learned through MCR$^2$ or SimDINO tend to have more pronounced linear and low-dimensional structures, such as mixtures of subspaces or low-rank Gaussians (as demonstrated in Section 4.3.2), which can facilitate the denoising process. This structural property can lead to more efficient and effective denoising models, as they can better exploit the underlying geometry of the representation space.

Once we have trained the denoiser, we can use it to generate new samples from the learned representation distribution via a diffusion process, as we have shown in Chapter 3. This completes the learning of a *representation autoencoder* (RAE) with an accessible representation distribution. In Chapter 8 (specifically Section 8.6.2), we will see a specific implementation of RAE and its applications in image generation.

## 6.2 Learning Self-Consistent Representations

In earlier chapters, we have studied methods that would allow us to learn a low-dimensional distribution via (lossy) compression. As we have mentioned in Chapter 1 and demonstrated in the previous chapters, the progress made in machine intelligence largely relies on finding computationally feasible and efficient solutions to realize the desired compression, not only computable or tractable in theory, but also scalable in practice:

$$\textbf{computable} \implies \textbf{tractable} \implies \textbf{scalable}. \tag{6.2.1}$$

It is even fair to say that the tremendous advancement in machine intelligence made in the past decade or so is largely attributed to the development of scalable models and methods, say by training deep networks via back propagation. Large models with billions of parameters trained with trillions of data points on tens of thousands of powerful GPUs have demonstrated super-human capabilities in memorizing existing knowledge. This has led many to believe that the "intelligence" of such models will continue to improve as their scale continues to go up.

While we celebrate the engineering marvels of such large man-made machine learning systems, we also must admit that, compared to intelligence in nature, this approach to improving machine intelligence is unnecessarily resource-demanding. Natural intelligent beings, including animals and humans, simply cannot afford such a brute-force solution for learning because they must operate with a very limited budget in energy, space, and time, subject to many strict physical constraints.

First, there is strong scientific evidence that our brain does not conduct global end-to-end back propagation to improve or correct its predictions. Instead, it has long been known in neuroscience that our brain corrects errors with local closed-loop feedback, such as predictive coding. This was the scientific basis that inspired Norbert Wiener to develop the theory of feedback control and the Cybernetics program back in the 1940s.

Second, we saw in the previous sections that in order to learn a consistent representation, one needs to learn a bidirectional autoencoding:

$$\boldsymbol{X} \xrightarrow{\mathcal{E}=f} \boldsymbol{Z} \xrightarrow{\mathcal{D}=g} \hat{\boldsymbol{X}}. \tag{6.2.2}$$

It requires enforcing the observed input data $\boldsymbol{X}$ and the decoded $\hat{\boldsymbol{X}}$ to be close by some measure of similarity $d(\boldsymbol{X}, \hat{\boldsymbol{X}})$. However, when the data $\boldsymbol{X}$ is high-dimensional (e.g., images), naive distance measures such as the $\ell^2$ norm in ambient space are largely uninformative: two images that differ by a small geometric transformation may be far apart in pixel distance yet perceptually identical. Hence, an outstanding question is: how can we enforce similarity between $\boldsymbol{X}$ and $\hat{\boldsymbol{X}}$ without a meaningful distance in the ambient data space? As we will see in this chapter, the answer lies in closed-loop feedback in a learned feature space, together with the low-dimensionality of the data distribution.

As we know from the previous chapter, in order to ensure that a representation is consistent, we need to compare the generated $\hat{\boldsymbol{X}} \sim p(\hat{\boldsymbol{x}})$ and the original $\boldsymbol{X} \sim p(\boldsymbol{x})$, at least in distribution. Even when we do have access to $\boldsymbol{X}$ and $\hat{\boldsymbol{X}}$, technically, computing and minimizing the distance between two distributions can be problematic, especially when the support of the distributions is low-dimensional. The KL-divergence introduced in Chapter 3 is not even well-defined between two distributions that do not have overlapping supports.

As an early attempt to alleviate the above difficulty in computing and minimizing the distance between two (low-dimensional) distributions, people had suggested learning the generator/decoder $g$ via discriminative approaches [Tu07]. This line of thought has led to the idea of *Generative Adversarial Nets*

*(GAN)* [GPM+14b]. It introduces a discriminator $d$, usually modeled by a deep network, to discern differences between the generated samples $\hat{\boldsymbol{X}}$ and the real ones $\boldsymbol{X}$:

$$\boldsymbol{Z} \xrightarrow{\;g(\boldsymbol{z},\eta)\;} \hat{\boldsymbol{X}},\ \boldsymbol{X} \xrightarrow{\;d(\boldsymbol{x},\theta)\;} \{\boldsymbol{0},\boldsymbol{1}\}. \tag{6.2.3}$$

It is suggested that we may attempt to align the distributions between $\hat{\boldsymbol{x}}$ and $\boldsymbol{x}$ via a *Stackelberg game*—a sequential game in which one player (the leader) commits to a strategy first and the other (the follower) responds optimally (see Section A.3 for a detailed treatment)—between the generator $g$ and the discriminator $d$:

$$\max_{\theta}\min_{\eta} \mathbb{E}_{p(\boldsymbol{x})}\big[\log d(\boldsymbol{x},\theta)\big] + \mathbb{E}_{p(\boldsymbol{z})}\big[\log\big(1 - d(\underbrace{g(\boldsymbol{z},\eta)}_{\hat{\boldsymbol{x}}\sim p_g},\theta)\big)\big]. \tag{6.2.4}$$

That is, the discriminator $d$ is trying to minimize the cross entropy between the true samples $\boldsymbol{X}$ and the generated ones $\hat{\boldsymbol{X}}$ while the generator $g$ is trying to do the opposite.

One may show that finding an equilibrium for the above Stackelberg game is equivalent to minimizing the *Jensen-Shannon divergence*:

$$\mathcal{D}_{JS}(p(\boldsymbol{x}), p_g(\hat{\boldsymbol{x}})) = \mathcal{D}_{KL}\big(p\|(p+p_g)/2\big) + \mathcal{D}_{KL}\big(p_g\|(p+p_g)/2\big). \tag{6.2.5}$$

Note that, compared to the KL-divergence, the JS-divergence is well-defined even if the supports of the two distributions are non-overlapping. However, JS-divergence does not have a closed-form expression even between two Gaussians, whereas KL-divergence does. In addition, since the data distributions are low-dimensional, the JS-divergence can be highly ill-conditioned to optimize.[7] This may explain why many additional heuristics are typically used in many subsequent variants of GAN.

So, instead, it has also been suggested to replace JS-divergence with the earth mover (EM) distance or the Wasserstein distance.[8] However, both the JS-divergence and Wasserstein distance can only be approximately computed between two general distributions.[9] Furthermore, neither the JS-divergence nor the Wasserstein distance has closed-form formulae, even for Gaussian distributions.[10]

If it is difficult to compare distributions of the data $\boldsymbol{X}$ and $\hat{\boldsymbol{X}}$, it may be possible to compare the learned feature $\boldsymbol{Z}$ with its image $\hat{\boldsymbol{Z}}$ under the encoder $f$:

$$\boldsymbol{X} \xrightarrow{\;\mathcal{E}=f\;} \boldsymbol{Z} \xrightarrow{\;\mathcal{D}=g\;} \hat{\boldsymbol{X}} \xrightarrow{\;\mathcal{E}=f\;} \hat{\boldsymbol{Z}} \tag{6.2.6}$$

[7] As shown in [ACB17].

[8] Roughly speaking, for distributions with potentially non-overlapping low-dimensional supports, the JS-divergence behaves like the $\ell^0$-norm, and the EM-distance behaves like the $\ell^1$-norm.

[9] For instance, the Wasserstein distance requires one to compute the maximal difference between expectations of the two distributions over all 1-Lipschitz functions.

[10] The ($\ell^1$-norm) Wasserstein distance can be bounded by the ($\ell^2$-norm) Wasserstein distance which has a closed-form [DDS22]. However, as is well-known in high-dimensional geometry, $\ell^1$-norm and $\ell^2$ norm deviate significantly in terms of their geometric and statistical properties as the dimension becomes high [WM22]. The bound can become very loose.

This leads to the notion of *self-consistent representation.*

**Definition 6.2** (Self-Consistent Representation)**.** Given data $\boldsymbol{X}$, we call a *self-consistent representation* a pair of functions $(f\colon \mathcal{X} \to \mathcal{Z}, g\colon \mathcal{Z} \to \mathcal{X})$, such that the *features* $\boldsymbol{Z} = f(\boldsymbol{X})$ are compact and structured, and the *autoencoding features* $\hat{\boldsymbol{Z}} \doteq f \circ g(\boldsymbol{Z})$ are *close* to $\boldsymbol{Z}$.

1. We say that it is *sample-wise* self-consistent if $\boldsymbol{Z} \approx \hat{\boldsymbol{Z}}$ in a certain norm with high probability.
2. We say that the representation is *distributionally self-consistent* if $\mathrm{Law}(\boldsymbol{Z}) \approx \mathrm{Law}(\hat{\boldsymbol{Z}})$.

### 6.2.1 Closed-Loop Transcription via Stackelberg Games

How do we try to ensure a learned representation is self-consistent? As usual, let us assume $\boldsymbol{X} = \cup_{k=1}^{K} \boldsymbol{X}_k$ with each subset of samples $\boldsymbol{X}_k$ belonging to a low-dimensional submanifold: $\boldsymbol{X}_k \subset \mathcal{M}_k, k = 1, \dots, K$. Following the notation in Chapter 3, we use a matrix $\boldsymbol{\Pi}_k(i,i) = 1$ to denote the membership of sample $i$ belonging to class $k$ and $\boldsymbol{\Pi}_k(i,j) = 0$ otherwise. One seeks a continuous mapping $f(\cdot, \boldsymbol{\theta}) : \boldsymbol{x} \mapsto \boldsymbol{z}$ from $\boldsymbol{X}$ to an optimal representation $\boldsymbol{Z} = [\boldsymbol{z}_1, \dots, \boldsymbol{z}_N] \subset \mathbb{R}^{d \times N}$:

$$\boldsymbol{X} \xrightarrow{\; f(\boldsymbol{x}, \boldsymbol{\theta}) \;} \boldsymbol{Z}, \tag{6.2.7}$$

which maximizes the following coding rate-reduction (MCR$^2$) objective:

$$\max_{\boldsymbol{Z}} \; \Delta R_\epsilon(\boldsymbol{Z} \mid \boldsymbol{\Pi}) \doteq \underbrace{\frac{1}{2} \log\det\Big(\boldsymbol{I} + \alpha \boldsymbol{Z}\boldsymbol{Z}^\top\Big)}_{R_\epsilon(\boldsymbol{Z})} - \underbrace{\sum_{k=1}^{K} \frac{\gamma_k}{2} \log\det\Big(\boldsymbol{I} + \alpha_k \boldsymbol{Z}\boldsymbol{\Pi}^k \boldsymbol{Z}^\top\Big)}_{R^c_\epsilon(\boldsymbol{Z}|\boldsymbol{\Pi})}, \tag{6.2.8}$$

where $\alpha = d/(N\epsilon^2)$, $\alpha_k = d/(\mathrm{tr}(\boldsymbol{\Pi}_k)\epsilon^2)$, $\gamma_k = \mathrm{tr}(\boldsymbol{\Pi}_k)/N$ for each $k = 1, \dots, K$.

One issue with learning such a one-sided mapping (6.2.7) via maximizing the above objective (6.2.8) is that it tends to expand the dimension of the learned subspace for features in each class[11], as we have seen in Section 4.2 of Chapter 4. To verify whether the learned features are consistent, meaning neither over-estimating nor under-estimating the intrinsic data structure, we may consider learning a decoder $g(\cdot, \eta) : \boldsymbol{z} \mapsto \boldsymbol{x}$ from the representation $\boldsymbol{Z} = f(\boldsymbol{X}, \boldsymbol{\theta})$ back to the data space $\boldsymbol{x}$: $\hat{\boldsymbol{X}} = g(\boldsymbol{Z}, \boldsymbol{\eta})$:

$$\boldsymbol{X} \xrightarrow{\; f(\boldsymbol{x}, \boldsymbol{\theta}) \;} \boldsymbol{Z} \xrightarrow{\; g(\boldsymbol{z}, \boldsymbol{\eta}) \;} \hat{\boldsymbol{X}}, \tag{6.2.9}$$

[11] If the dimension of the feature space $d$ is too high, maximizing the rate reduction may over-estimate the dimension of each class. Hence, to learn a good representation, one needs to pre-select a proper dimension for the feature space, as achieved in the experiments in [YCY+20]. In fact the same "model selection" problem persists even in the simplest single-subspace case, which is the classic PCA [Jol86]. To our best knowledge, selecting the correct number of principal components in a heterogeneous noisy situation remains an active research topic [HSD20].

and check how close $\boldsymbol{X}$ and $\hat{\boldsymbol{X}}$ are. This forms an autoencoding which is what we have studied extensively earlier in this chapter.

**Measuring distance in the feature space.** However, as we have discussed above, if we do not have the option to compute the distance between $\boldsymbol{X}$ and $\hat{\boldsymbol{X}}$, we are left with the option of comparing their corresponding features $\boldsymbol{Z}$ and $\hat{\boldsymbol{Z}} = f(\hat{\boldsymbol{X}}, \boldsymbol{\theta})$. Notice that under the MCR$^2$ objective, the distributions of the resulting $\boldsymbol{Z}$ or $\hat{\boldsymbol{Z}}$ tend to be piecewise linear so their distance can be computed more easily. More precisely, since the features of each class, $\boldsymbol{Z}_k$ and $\hat{\boldsymbol{Z}}_k$, are close to being a low-dimensional subspace/Gaussian, their "distance" can be measured by the rate reduction, with (6.2.8) restricted to two sets of equal size:

$$\Delta R_\epsilon\big(\boldsymbol{Z}_k, \hat{\boldsymbol{Z}}_k\big) \doteq R_\epsilon\big(\boldsymbol{Z}_k \cup \hat{\boldsymbol{Z}}_k\big) - \frac{1}{2}\big(R_\epsilon(\boldsymbol{Z}_k) + R_\epsilon(\hat{\boldsymbol{Z}}_k)\big). \tag{6.2.10}$$

The overall distance between $\boldsymbol{Z}$ and $\hat{\boldsymbol{Z}}$ is given by:

$$d(\boldsymbol{Z}, \hat{\boldsymbol{Z}}) \doteq \sum_{k=1}^{K} \Delta R_\epsilon\big(\boldsymbol{Z}_k, \hat{\boldsymbol{Z}}_k\big) = \sum_{k=1}^{K} \Delta R_\epsilon\big(\boldsymbol{Z}_k, f(g(\boldsymbol{Z}_k, \boldsymbol{\eta}), \boldsymbol{\theta})\big). \tag{6.2.11}$$

If we are interested in discerning *any* differences in the distributions of the original data $\boldsymbol{X}$ and their decoded $\hat{\boldsymbol{X}}$, we may view the feature encoder $f(\cdot, \boldsymbol{\theta})$ as a "discriminator" to *magnify* any difference between all pairs of $\boldsymbol{X}_k$ and $\hat{\boldsymbol{X}}_k$, by simply maximizing the distance $d(\boldsymbol{Z}, \hat{\boldsymbol{Z}})$:

$$\max_f d(\boldsymbol{Z}, \hat{\boldsymbol{Z}}) = \max_{\boldsymbol{\theta}} \sum_{k=1}^{K} \Delta R_\epsilon\big(\boldsymbol{Z}_k, \hat{\boldsymbol{Z}}_k\big) = \max_{\boldsymbol{\theta}} \sum_{k=1}^{K} \Delta R_\epsilon\big(f(\boldsymbol{X}_k, \boldsymbol{\theta}), f(\hat{\boldsymbol{X}}_k, \boldsymbol{\theta})\big). \tag{6.2.12}$$

That is, a "distance" between $\boldsymbol{X}$ and $\hat{\boldsymbol{X}}$ can be measured as the maximally achievable rate reduction between all pairs of classes in these two sets. In a way, this measures how well or badly the decoded $\hat{\boldsymbol{X}}$ aligns with the original data $\boldsymbol{X}$—hence measuring the goodness of "injectivity" of the encoder $f$. Notice that such a discriminative measure is consistent with the idea of GAN (6.2.3) that tries to separate $\boldsymbol{X}$ and $\hat{\boldsymbol{X}}$ into two classes, measured by the cross-entropy.

Nevertheless, here the discriminative encoder $f$ naturally generalizes to cases when the data distributions are multi-class and multi-modal, and the discriminativeness is measured with a more refined measure—the rate reduction—instead of the typical two-class loss (e.g., cross entropy) used in GANs. That is, we may view the encoder $f$ as a generalized discriminator that replaces the binary classifier $d$ in (6.2.3):

$$\boldsymbol{Z} \xrightarrow{\hspace{2pt} g(\boldsymbol{z}, \boldsymbol{\eta}) \hspace{2pt}} \hat{\boldsymbol{X}}, \, \boldsymbol{X} \xrightarrow{\hspace{2pt} f(\boldsymbol{x}, \boldsymbol{\theta}) \hspace{2pt}} \{\hat{\boldsymbol{Z}}, \boldsymbol{Z}\}. \tag{6.2.13}$$

The overall pipeline can be illustrated by the following "closed-loop" diagram:

$$\boldsymbol{X} \xrightarrow{\hspace{2pt} f(\boldsymbol{x}, \boldsymbol{\theta}) \hspace{2pt}} \boldsymbol{Z} \xrightarrow{\hspace{2pt} g(\boldsymbol{z}, \boldsymbol{\eta}) \hspace{2pt}} \hat{\boldsymbol{X}} \xrightarrow{\hspace{2pt} f(\boldsymbol{x}, \boldsymbol{\theta}) \hspace{2pt}} \hat{\boldsymbol{Z}}, \tag{6.2.14}$$

where the overall model has parameters: $\boldsymbol{\Theta} = \{\boldsymbol{\theta}, \boldsymbol{\eta}\}$. Figure 6.9 shows the overall process.

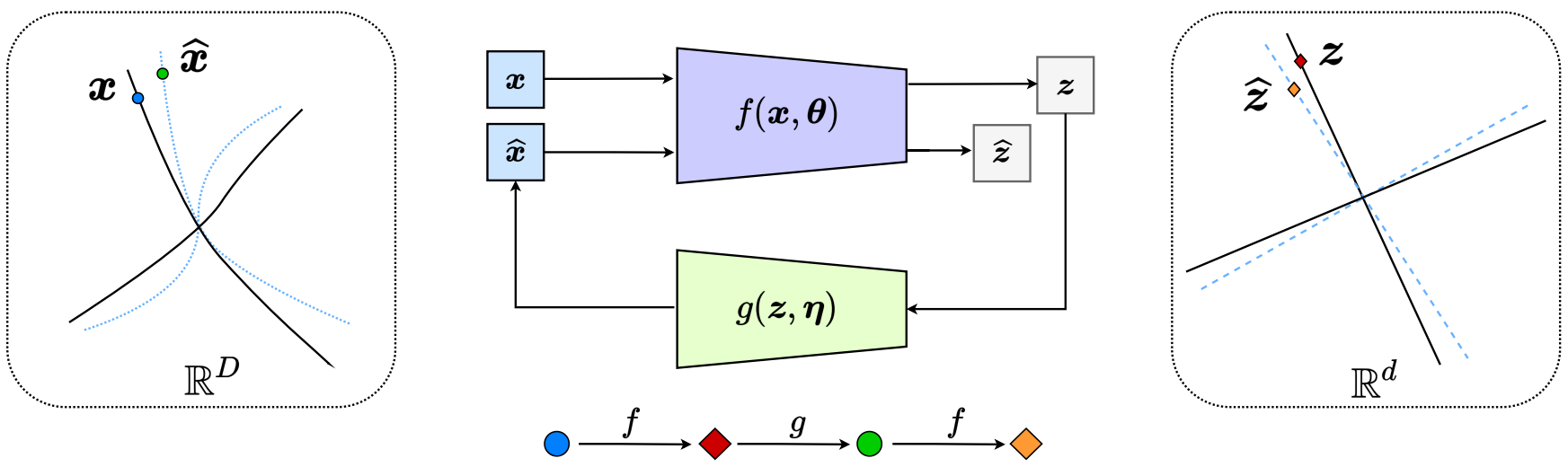

Figure 6.9: **A Closed-loop Transcription.** The encoder $f$ has dual roles: it learns a representation $\boldsymbol{z}$ for the data $\boldsymbol{x}$ via maximizing the rate reduction of $\boldsymbol{z}$ and it is also a "feedback sensor" for any discrepancy between the data $\boldsymbol{x}$ and the decoded $\hat{\boldsymbol{x}}$. The decoder $g$ also has dual roles: it is a "controller" that corrects the discrepancy between $\boldsymbol{x}$ and $\hat{\boldsymbol{x}}$ and it also aims to minimize the overall coding rate for the learned representation.

**Encoder and decoder as a two-player game.** Obviously, to ensure the learned autoencoding is self-consistent, the main goal of the decoder $g(\cdot, \boldsymbol{\eta})$ is to *minimize* the distance between $\boldsymbol{Z}$ and $\hat{\boldsymbol{Z}}$. That is, to learn $g$, we want to minimize the distance $d(\boldsymbol{Z}, \hat{\boldsymbol{Z}})$:

$$\min_{g} d(\boldsymbol{Z}, \hat{\boldsymbol{Z}}) \doteq \min_{\boldsymbol{\eta}} \sum_{k=1}^{K} \Delta R_\epsilon\big(\boldsymbol{Z}_k, \hat{\boldsymbol{Z}}_k\big) = \min_{\boldsymbol{\eta}} \sum_{k=1}^{K} \Delta R_\epsilon\big(\boldsymbol{Z}_k, f(g(\boldsymbol{Z}_k, \boldsymbol{\eta}), \boldsymbol{\theta})\big), \tag{6.2.15}$$

where $\boldsymbol{Z}_k = f(\boldsymbol{X}_k, \boldsymbol{\theta})$ and $\hat{\boldsymbol{Z}}_k = f(\hat{\boldsymbol{X}}_k, \boldsymbol{\theta})$.

*Example* 6.1. One may wonder why we need the mapping $f(\cdot, \boldsymbol{\theta})$ to function as a discriminator between $\boldsymbol{X}$ and $\hat{\boldsymbol{X}}$ by maximizing $\max_{\boldsymbol{\theta}} \Delta R_\epsilon\big(f(\boldsymbol{X}, \boldsymbol{\theta}), f(\hat{\boldsymbol{X}}, \boldsymbol{\theta})\big)$. Figure 6.10 gives a simple illustration: there might be many decoders $g$ such that $f \circ g$ is an identity (Id) mapping. $f \circ g(\boldsymbol{z}) = \boldsymbol{z}$ for all $\boldsymbol{z}$ in the subspace $S_{\boldsymbol{z}}$ in the feature space. However, $g \circ f$ is not necessarily an autoencoding map for $\boldsymbol{x}$ in the original distribution $S_{\boldsymbol{x}}$ (here drawn as a subspace for simplicity). That is, $g \circ f(S_{\boldsymbol{x}}) \not\subset S_{\boldsymbol{x}}$, let alone $g \circ f(S_{\boldsymbol{x}}) = S_{\boldsymbol{x}}$ or $g \circ f(\boldsymbol{x}) = \boldsymbol{x}$. One should expect that, without careful control of the image of $g$, this would be the case with high probability, especially when the support of the distribution of $\boldsymbol{x}$ is extremely low-dimensional in the original high-dimensional data space. ■

Comparing the contractive and contrastive nature of (6.2.15) and (6.2.12) on the same distance measure, we see the roles of the encoder $f(\cdot, \boldsymbol{\theta})$ and the decoder $g(\cdot, \boldsymbol{\eta})$ naturally as "**a two-player game**": *while the encoder $f$ tries to magnify the difference between the original data and their transcribed data, the decoder $g$ aims to minimize the difference.* Now for convenience, let us define the "closed-loop encoding" function:

$$h(\boldsymbol{x}, \boldsymbol{\theta}, \boldsymbol{\eta}) \doteq f\big(g\big(f(\boldsymbol{x}, \boldsymbol{\theta}), \boldsymbol{\eta}\big), \boldsymbol{\theta}\big) : \boldsymbol{x} \mapsto \hat{\boldsymbol{z}}. \tag{6.2.16}$$

Ideally, we want this function to be very close to $f(\boldsymbol{x}, \boldsymbol{\theta})$ or at least the distributions of their images should be close. With this notation, combining (6.2.15) and (6.2.12), a closed-loop notion of "distance" between $\boldsymbol{X}$ and $\hat{\boldsymbol{X}}$ can be computed as *an equilibrium point* to the following Stackelberg game (cf. Section A.3) for the same utility in terms of rate reduction:

$$\mathcal{D}(\boldsymbol{X}, \hat{\boldsymbol{X}}) \doteq \max_{\boldsymbol{\theta}} \min_{\boldsymbol{\eta}} \sum_{k=1}^{K} \Delta R_\epsilon\big(f(\boldsymbol{X}_k, \boldsymbol{\theta}), h(\boldsymbol{X}_k, \boldsymbol{\theta}, \boldsymbol{\eta})\big). \tag{6.2.17}$$

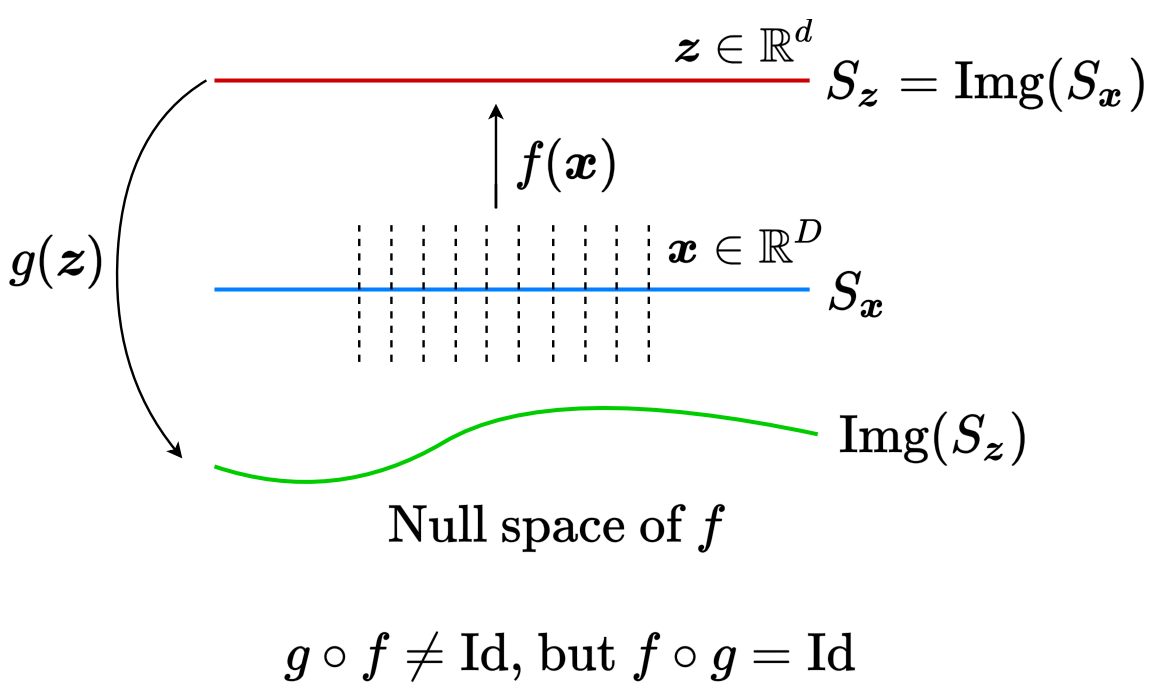

Figure 6.10: **Embeddings of low-dim submanifolds in a high-dim space.** $S_{\boldsymbol{x}}$ (blue) is the submanifold for the original data $\boldsymbol{x}$; $S_{\boldsymbol{z}}$ (red) is the image of $S_{\boldsymbol{x}}$ under the mapping $f$, representing the learned feature $\boldsymbol{z}$; and the green curve is the image of the feature $\boldsymbol{z}$ under the decoding mapping $g$.

Notice that this only measures the difference between (features of) the original data and its transcribed version. It does not measure how good the representation $\boldsymbol{Z}$ (or $\hat{\boldsymbol{Z}}$) is for the multiple classes within $\boldsymbol{X}$ (or $\hat{\boldsymbol{X}}$). To this end, we may combine the above distance with the original MCR$^2$-type objectives (6.2.8): namely, the rate reduction $\Delta R_\epsilon(\boldsymbol{Z})$ and $\Delta R_\epsilon(\hat{\boldsymbol{Z}})$ for the learned LDR $\boldsymbol{Z}$ for $\boldsymbol{X}$ and $\hat{\boldsymbol{Z}}$ for the decoded $\hat{\boldsymbol{X}}$. Notice that although the encoder $f$ tries to *maximize* the multi-class rate reduction of the features $\boldsymbol{Z}$ of the data $\boldsymbol{X}$, the decoder $g$ should *minimize* the rate reduction of the multi-class features $\hat{\boldsymbol{Z}}$ of the decoded $\hat{\boldsymbol{X}}$. That is, the decoder $g$ tries to use a minimal coding rate needed to achieve a good decoding quality.

Hence, the overall "multi-class" Stackelberg game for learning the closed-loop transcription, named CTRL-Multi, is

$$\max_{\boldsymbol{\theta}} \min_{\boldsymbol{\eta}} \mathcal{T}_{\boldsymbol{X}}(\boldsymbol{\theta}, \boldsymbol{\eta}) \tag{6.2.18}$$

$$\doteq \underbrace{\Delta R_\epsilon\big(f(\boldsymbol{X}, \boldsymbol{\theta})\big)}_{\text{Expansive encode}} + \underbrace{\Delta R_\epsilon\big(h(\boldsymbol{X}, \boldsymbol{\theta}, \boldsymbol{\eta})\big)}_{\text{Compressive decode}} + \sum_{k=1}^{K} \underbrace{\Delta R_\epsilon\big(f(\boldsymbol{X}_k, \boldsymbol{\theta}), h(\boldsymbol{X}_k, \boldsymbol{\theta}, \boldsymbol{\eta})\big)}_{\text{Contrastive encode \& Contractive decode}} \tag{6.2.19}$$

$$= \Delta R_\epsilon\big(\boldsymbol{Z}(\boldsymbol{\theta})\big) + \Delta R_\epsilon\big(\hat{\boldsymbol{Z}}(\boldsymbol{\theta}, \boldsymbol{\eta})\big) + \sum_{k=1}^{K} \Delta R_\epsilon\big(\boldsymbol{Z}_k(\boldsymbol{\theta}), \hat{\boldsymbol{Z}}_k(\boldsymbol{\theta}, \boldsymbol{\eta})\big), \tag{6.2.20}$$

subject to certain constraints (upper or lower bounds) on the first term and the second term.

Notice that, without the terms associated with the generative part $h$ or with all such terms fixed as constant, the above objective is precisely the original MCR$^2$ objective introduced in Chapter 4. In an unsupervised setting, if we view each sample (and its augmentations) as its own class, the above formulation remains exactly the same. The number of classes $K$ is simply the number of independent samples. In addition, notice that the above game's objective function depends only on (features of) the data $\boldsymbol{X}$, hence one can learn the encoder and decoder (parameters) without the need for sampling or matching any additional distribution (as typically needed in GANs or VAEs).

As a special case, if $\boldsymbol{X}$ only has one class, the Stackelberg game reduces to a special "two-class" or "binary" form,[12] named CTRL-Binary,

$$\max_{\boldsymbol{\theta}} \min_{\boldsymbol{\eta}} \mathcal{T}^b_{\boldsymbol{X}}(\boldsymbol{\theta}, \boldsymbol{\eta}) \doteq \Delta R_\epsilon\big(f(\boldsymbol{X}, \boldsymbol{\theta}), h(\boldsymbol{X}, \boldsymbol{\theta}, \boldsymbol{\eta})\big) = \Delta R_\epsilon\big(\boldsymbol{Z}(\boldsymbol{\theta}), \hat{\boldsymbol{Z}}(\boldsymbol{\theta}, \boldsymbol{\eta})\big), \tag{6.2.21}$$

between $\boldsymbol{X}$ and the decoded $\hat{\boldsymbol{X}}$ by viewing $\boldsymbol{X}$ and $\hat{\boldsymbol{X}}$ as two classes $\{\mathbf{0}, \mathbf{1}\}$. Notice that this binary case resembles the formulation of the original GAN (6.2.3). Instead of using cross entropy, our formulation adopts a more refined rate-reduction measure, which has been shown to promote diversity in the learned representation in Chapter 3.

Sometimes, even when $\boldsymbol{X}$ contains multiple classes/modes, one could still view all classes together as one class. Then, the above binary objective is to align the union distribution of all classes with their decoded $\hat{\boldsymbol{X}}$. This is typically a simpler task to achieve than the multi-class one (6.2.20), since it does not require learning of a more refined multi-class CTRL for the data, as we will later see in experiments. Notice that one good characteristic of the above formulation is that *all quantities in the objectives are measured in terms of rate reduction for the learned features* (assuming features eventually become subspace Gaussians).

One may notice that the above learning framework draws inspiration from closed-loop error correction widely practiced in feedback control systems. The closed-loop mechanism is used to form an overall feedback system between the two encoding and decoding networks for correcting any "error" in the distributions between the data $\boldsymbol{x}$ and the decoded $\hat{\boldsymbol{x}}$. Using terminology from control theory, one may view the encoding network $f$ as a "sensor" for error feedback while the decoding network $g$ as a "controller" for error correction. However, notice that here the "target" for control is not a scalar nor a finite dimensional vector, but a continuous distribution—in order for the distribution of $\hat{\boldsymbol{x}}$ to match that of the data $\boldsymbol{x}$. This is in general a control problem in an infinite dimensional space. The space of possible diffeomorphisms of submanifolds that $f$ tries to model is infinite-dimensional [Lee02]. Ideally, we hope when the sensor $f$ and the controller $g$ are optimal, the distribution of $\boldsymbol{x}$ becomes a "fixed point" for the closed loop while the distribution of $\boldsymbol{z}$ reaches a compact linear discriminative representation. Hence, the minimax programs (6.2.20) and (6.2.21) can also be interpreted as games between an error-feedback sensor and an error-reducing controller.

[12] As the first two rate reduction terms automatically become zero.

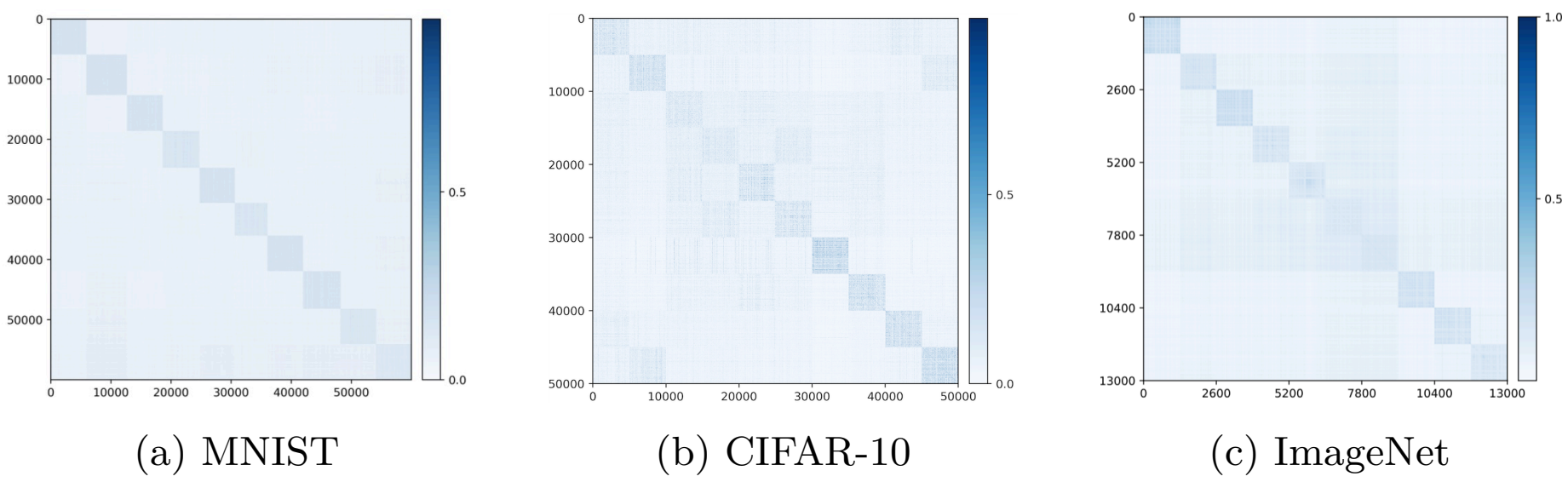

(a) MNIST (b) CIFAR-10 (c) ImageNet

Figure 6.11: Visualizing the alignment between $\boldsymbol{Z}$ and $\hat{\boldsymbol{Z}}$: $|\boldsymbol{Z}^\top \hat{\boldsymbol{Z}}|$ in the feature space for (**a**) MNIST, (**b**) CIFAR-10, and (**c**) ImageNet-10-Class.

The remaining question is whether the above framework can indeed learn a good (autoencoding) representation of a given dataset? Before we give some formal theoretical justification (in the next subsection), we present some empirical results.

**Visualizing correlation of features $\boldsymbol{Z}$ and decoded features $\hat{\boldsymbol{Z}}$.** We visualize the cosine similarity between $\boldsymbol{Z}$ and $\hat{\boldsymbol{Z}}$ learned from the multi-class objective (6.2.20) on MNIST, CIFAR-10 and ImageNet (10 classes), which indicates how close $\hat{\boldsymbol{z}} = f \circ g(\boldsymbol{z})$ is from $\boldsymbol{z}$. Results in Figure 6.11 show that $\boldsymbol{Z}$ and $\hat{\boldsymbol{Z}}$ are aligned very well within each class. The block-diagonal patterns for MNIST are sharper than those for CIFAR-10 and ImageNet, as images in CIFAR-10 and ImageNet have more diverse visual appearances.

**Visualizing autoencoding of the data $\boldsymbol{X}$ and the decoded $\hat{\boldsymbol{X}}$.** We compare some representative $\boldsymbol{X}$ and $\hat{\boldsymbol{X}}$ on MNIST, CIFAR-10 and ImageNet (10 classes) to verify how close $\hat{\boldsymbol{x}} = g \circ f(\boldsymbol{x})$ is to $\boldsymbol{x}$. The results are shown in Figure 6.12, and visualizations are created from training samples. Visually, the autoencoded $\hat{\boldsymbol{x}}$ faithfully captures major visual features from its respective training sample $\boldsymbol{x}$, especially the pose, shape, and layout. For the simpler dataset such as MNIST, autoencoded images are almost identical to the original.

### 6.2.2 A Mixture of Low-Dimensional Gaussians

In the above, we have argued that it is possible to formulate the problem of learning a data distribution as a closed-loop autoencoding problem. We also saw empirically that such a scheme seems to work. The remaining question is when and why such a scheme should work. It is difficult to answer this question for the most general cases with arbitrary data distributions. Nevertheless, as usual, let us see if we can arrive at a rigorous justification for the ideal case when the data distribution is a mixture of low-dimensional subspaces or low-rank Gaussians. A clear characterization and understanding of this important

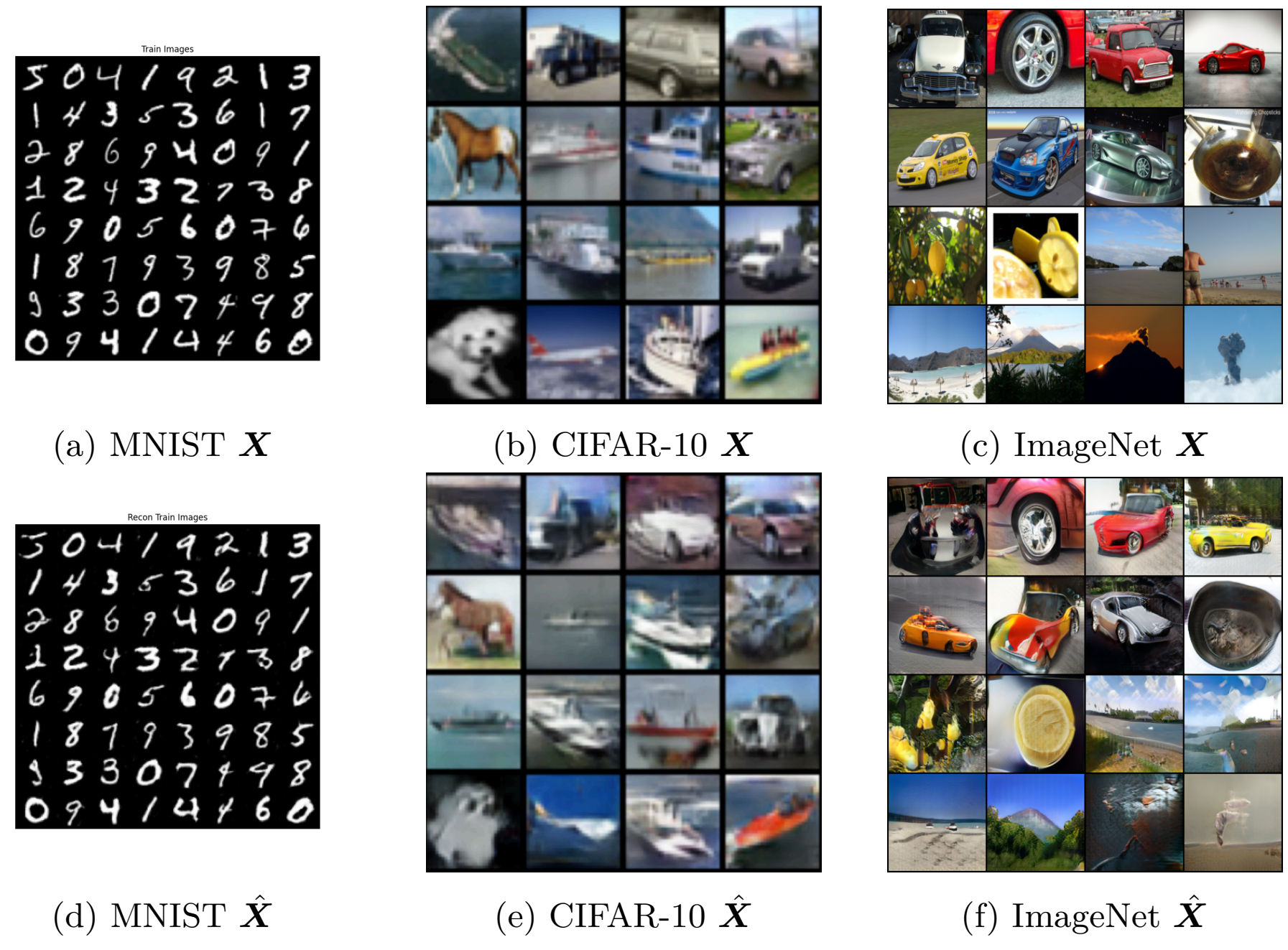

(a) MNIST $\boldsymbol{X}$ (b) CIFAR-10 $\boldsymbol{X}$ (c) ImageNet $\boldsymbol{X}$

(d) MNIST $\hat{\boldsymbol{X}}$ (e) CIFAR-10 $\hat{\boldsymbol{X}}$ (f) ImageNet $\hat{\boldsymbol{X}}$

Figure 6.12: Visualizing the auto-encoding property of the learned closed-loop transcription ($\boldsymbol{x} \approx \hat{\boldsymbol{x}} = g \circ f(\boldsymbol{x})$) on MNIST, CIFAR-10, and ImageNet (zoom in for better visualization).

special case would shed light on the more general cases.[13]

To this end, let us first suppose that $\boldsymbol{X}$ is distributed according to a mixture of low-dimensional Gaussians, and the label (i.e., subspace assignment) for $\boldsymbol{X}$ is given by $\boldsymbol{y}$. Then, let us set up a minimax optimization problem to learn the data distribution, say through learning an encoding of $\boldsymbol{X}$ into representations $\boldsymbol{Z}$ which are supported on a mixture of *orthogonal* subspaces, and a decoding of $\boldsymbol{Z}$ back into $\boldsymbol{X}$. Then, the representation we want to achieve is maximized by the earlier-discussed version of the information gain, i.e., $\Delta R_\epsilon(\boldsymbol{Z}) = R_\epsilon(\boldsymbol{Z}) - \sum_{k=1}^{K} R_\epsilon(\boldsymbol{Z}_k)$, which enforces that the representation $\boldsymbol{Z}_k$ of each class $k$ spans a subspace which is orthogonal to the supporting subspaces of other classes. The way to measure the consistency of the decoding is, as before, given by $\sum_{k=1}^{K} \Delta R_\epsilon(\boldsymbol{Z}_k, \hat{\boldsymbol{Z}}_k)$, which enforces that the representation $\boldsymbol{Z}_k$ and its autoencoding $\hat{\boldsymbol{Z}}_k$ for each class $k$ span the same subspace. Thus, we can set up a simplified Stackelberg game:

$$\max_{\boldsymbol{\theta}} \min_{\boldsymbol{\eta}} \left\{ \Delta R_\epsilon(\boldsymbol{Z}(\boldsymbol{\theta})) + \sum_{k=1}^{K} \Delta R_\epsilon(\boldsymbol{Z}_k(\boldsymbol{\theta}), \hat{\boldsymbol{Z}}_k(\boldsymbol{\theta}, \boldsymbol{\eta})) \right\} \tag{6.2.22}$$

[13] As most distributions with low-dimensional structures can be well-approximated by this family of distributions.

Notice that this is a simpler setup than what is used in practice—there is no $\Delta R_{\epsilon}(\hat{\boldsymbol{Z}})$ term, for instance, and we work in the supervised setting with class labels (although the techniques used to prove the following result are easy to extend to unsupervised formulations). Also, the consistency of the representations is only measured in a distribution-wise sense via $\Delta R_{\epsilon}$ (though this may be substituted with a sample-wise distance metric such as the $\ell_2$ norm if desired, and equivalent conclusions may be drawn, *mutatis mutandis*).

Under mild conditions, in order to realize the desired encoder and decoder which realize $\boldsymbol{Z}$ from a data source $\boldsymbol{X}$ that is already distributed according to a mixture of correlated low-dimensional Gaussians, we only require a linear encoder and decoder to disentangle and whiten the Gaussians. We then study this setting in the case where $\boldsymbol{\theta}$ and $\boldsymbol{\eta}$ parameterize matrices whose operator norm is constrained.

We want to understand what kinds of optima are learned in this setting. One suitable solution concept for this kind of game, where the decoder's optimal solution is defined solely with respect to the encoder (and not, in particular, with respect to some other intrinsic property of the decoder), is a *Stackelberg equilibrium* where the decoder follows the encoder. Namely, at such an equilibrium, the decoder should optimize its objective; meanwhile the encoder should optimize its objective, given that whatever it picks, the decoder will pick an optimal response (which may affect the encoder objective). In game-theoretic terms, it is like the decoder goes *second*: it chooses its weights after the encoder, and both the encoder and decoder attempt to optimize their objective in light of this. It is computationally tractable to learn sequential equilibria via gradient methods via *alternating optimization*, where each side uses different learning rates. Detailed exposition of sequential equilibria is beyond the scope of this book and we provide more technical details in Section A.3. In this setting, we have the following result:

**Theorem 6.1** ([PPC+23], Abridged)**.** *Suppose that $\boldsymbol{X}$ is distributed on a mixture of subspaces. Under certain realistic yet technical conditions, it holds that all sequential equilibria of* (6.2.22) *obey:*

- *The $\boldsymbol{Z}_k$ lie on orthogonal subspaces and are isotropic on those subspaces, i.e., maximizing the information gain.*
- *The autoencoding is self-consistent, i.e., the subspaces spanned by $\boldsymbol{Z}_k$ and $\hat{\boldsymbol{Z}}_k$ are the same for all $k$.*

This notion of self-consistency is the most one can expect if there are only geometric assumptions on the data, i.e., there are no statistical assumptions. If we assume that the columns of $\boldsymbol{X}$ are drawn from a low-rank Gaussian mixture model, then analogous versions of this theorem certify that $\boldsymbol{Z}_k$ are also low-rank Gaussians whose covariance is isotropic, for instance. Essentially, this result validates, via the simple case of Gaussian mixtures on subspaces, that minimax games to optimize the information gain and self-consistency may achieve optimal solutions.

## 6.3 Continuous Learning Self-Consistent Representations

### 6.3.1 Class-wise Incremental Learning

As we have seen, deep neural networks have demonstrated a great ability to learn representations for hundreds or even thousands of classes of objects, in both discriminative and generative contexts. However, networks typically must be trained offline, with uniformly sampled data from all classes simultaneously. It has been known that when an (open-loop) network is updated to learn new classes without data from the old ones, previously learned knowledge will fall victim to the problem of *catastrophic forgetting* [MC89]. This is known in neuroscience as the stability-plasticity dilemma: the challenge of ensuring that a neural system can learn from a new environment while retaining essential knowledge from previous ones [Gro87].

In contrast, natural neural systems (e.g., animal brains) do not seem to suffer from such catastrophic forgetting at all. They are capable of developing new memory of new objects while retaining memory of previously learned objects. This ability, for either natural or artificial neural systems, is often referred to as *incremental learning, continual learning, sequential learning*, or *lifelong learning* [AR20].

While many recent works have highlighted how artificial neural systems can be trained in more flexible ways, the strongest existing efforts toward answering the stability-plasticity dilemma for artificial neural networks typically require either storing raw exemplars [CRE+19; RKS+17] or providing external mechanisms [KPR+17]. Raw exemplars, particularly in the case of high-dimensional inputs like images, are costly and difficult to scale, while external mechanisms—which typically include secondary networks and representation spaces for generative replay, incremental allocation of network resources, network duplication, or explicit isolation of used and unused parts of the network—require heuristics and incur hidden costs.

Here we are interested in an incremental learning setting that is similar to nature. It counters these existing practices with two key qualities.

1. The first is that it should be *memory-based.* When learning new classes, no raw exemplars of old classes are available to train the network together with new data. This implies that one has to rely on a compact and thus structured "memory" learned for old classes, such as incrementally learned generative representations of the old classes, as well as the associated encoding and decoding mappings [KK18].

2. The second is that it should be *self-contained.* Incremental learning takes place in a single neural system with a fixed capacity, and in a common representation space. The ability to minimize forgetting is implied by optimizing an overall learning objective, without external networks, architectural modifications, or resource allocation mechanisms.

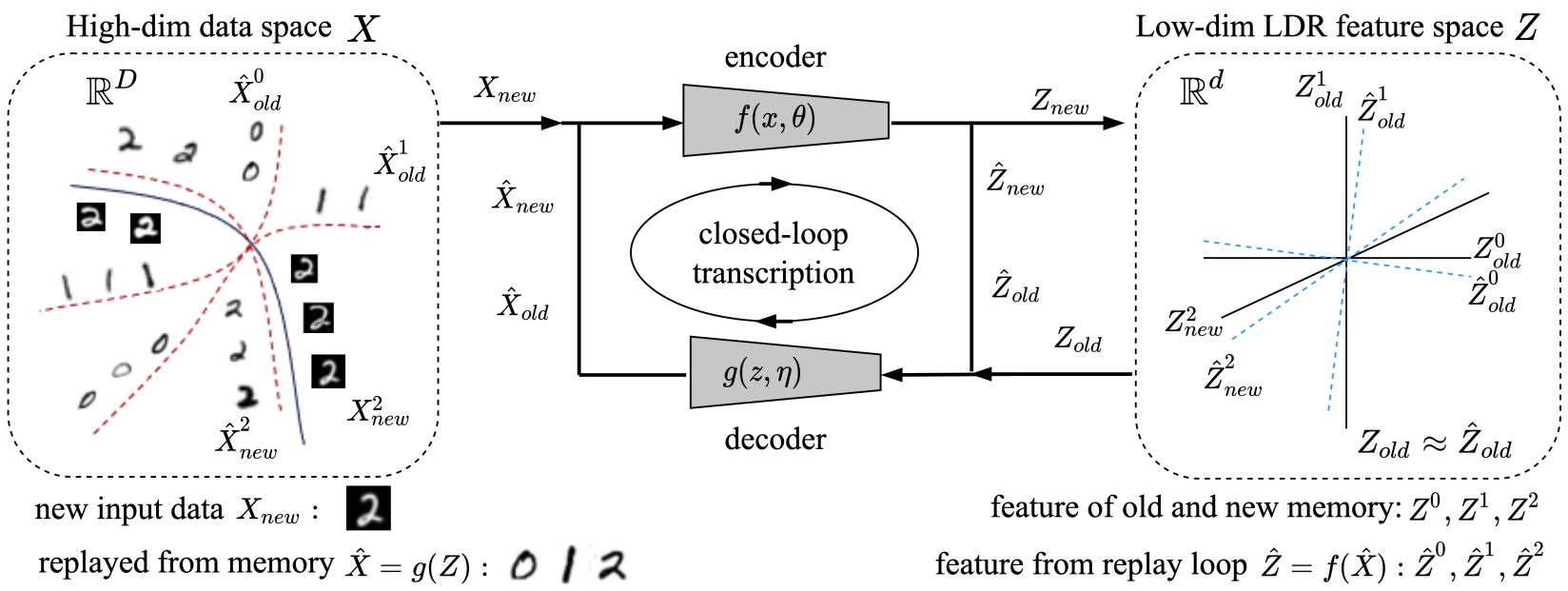

Figure 6.13: **Overall framework** of our closed-loop transcription-based incremental learning for a structured LDR memory. Only a single, entirely self-contained, encoding-decoding network is needed: for a new data class $\boldsymbol{X}_{new}$, a new LDR memory $\boldsymbol{Z}_{new}$ is incrementally learned as a minimax game between the encoder and decoder subject to the constraint that old memory of past classes $\boldsymbol{Z}_{old}$ is intact through the closed-loop transcription (or replay): $\boldsymbol{Z}_{old} \approx \hat{\boldsymbol{Z}}_{old} = f(g(\boldsymbol{Z}_{old}))$.

The incoherent linear structures for features of different classes closely resemble how objects are encoded in different areas of the inferotemporal cortex of animal brains [BSM+20; CT17]. The closed-loop transcription $\boldsymbol{X} \to \boldsymbol{Z} \to \hat{\boldsymbol{X}} \to \hat{\boldsymbol{Z}}$ also resembles popularly hypothesized mechanisms for memory formation [JT20; VST+20]. This leads to a question: since memory in the brain is formed in an incremental fashion, can the above closed-loop transcription framework also support incremental learning?

**LDR memory sampling and replay.** The simple linear *structures* of LDR make it uniquely suited for incremental learning: the distribution of features $\boldsymbol{Z}_j$ of each previously learned class can be explicitly and concisely represented by a principal subspace $\mathcal{S}_j$ in the feature space. To preserve the memory of an old class $j$, we only need to preserve the subspace while learning new classes. To this end, we simply sample $m$ representative prototype features on the subspace along its top $r$ principal components, and denote these features as $\boldsymbol{Z}_{j,old}$. Because of the simple linear structures of LDR, we can sample from $\boldsymbol{Z}_{j,old}$ by calculating the mean and covariance of $\boldsymbol{Z}_{j,old}$ after learning class $j$. The storage required is extremely small, since we only need to store means and covariances, which are sampled from as needed. Suppose a total of $t$ old classes have been learned so far. If prototype features, denoted $\boldsymbol{Z}_{old} \doteq [\boldsymbol{Z}_{1,old}, \ldots, \boldsymbol{Z}_{t,old}]$, for all of these classes can be preserved when learning new classes, the subspaces $\{\mathcal{S}_j\}_{j=1}^t$ representing past memory will be preserved as well. Details about sampling and calculating mean and covariance can be found in the work of [TDW+23].

**Incremental learning LDR with an old-memory constraint.** Notice that, with the learned autoencoding (6.2.9), one can replay and use the images, say $\hat{\boldsymbol{X}}_{old} = g(\boldsymbol{Z}_{old}, \boldsymbol{\eta})$, associated with the memory features to avoid forgetting while learning new classes. This is typically how generative models have been used for prior incremental learning methods. However, with the closed-loop framework, explicitly replaying images from the features is not necessary. Past memory can be effectively preserved through optimization exclusively on the features themselves.

Consider the task of incrementally learning a new class of objects.[14] We denote a corresponding new sample set as $\boldsymbol{X}_{new}$. The features of $\boldsymbol{X}_{new}$ are denoted as $\boldsymbol{Z}_{new}(\boldsymbol{\theta}) = f(\boldsymbol{X}_{new}, \boldsymbol{\theta})$. We concatenate them together with the prototype features of the old classes $\boldsymbol{Z}_{old}$ and form $\boldsymbol{Z} = [\boldsymbol{Z}_{new}(\boldsymbol{\theta}), \boldsymbol{Z}_{old}]$. We denote the replayed images from all features as $\hat{\boldsymbol{X}} = [\hat{\boldsymbol{X}}_{new}(\boldsymbol{\theta}, \boldsymbol{\eta}), \hat{\boldsymbol{X}}_{old}(\boldsymbol{\eta})]$ although we do not actually need to compute or use them explicitly. We only need features of replayed images, denoted $\hat{\boldsymbol{Z}} = f(\hat{\boldsymbol{X}}, \boldsymbol{\theta}) = [\hat{\boldsymbol{Z}}_{new}(\boldsymbol{\theta}, \boldsymbol{\eta}), \hat{\boldsymbol{Z}}_{old}(\boldsymbol{\theta}, \boldsymbol{\eta})]$.

Mirroring the motivation for the multi-class CTRL objective (6.2.20), we would like the features of the new class $\boldsymbol{Z}_{new}$ to be incoherent to all of the old ones $\boldsymbol{Z}_{old}$. As $\boldsymbol{Z}_{new}$ is the only new class whose features need to be learned, the objective (6.2.20) reduces to the case where $K = 1$:

$$\min_{\boldsymbol{\eta}} \max_{\boldsymbol{\theta}} \Delta R_\epsilon(\boldsymbol{Z}) + \Delta R_\epsilon(\hat{\boldsymbol{Z}}) + \Delta R_\epsilon(\boldsymbol{Z}_{new}, \hat{\boldsymbol{Z}}_{new}). \tag{6.3.1}$$

However, when we update the network parameters $(\boldsymbol{\theta}, \boldsymbol{\eta})$ to optimize the features for the new class, the updated mappings $f$ and $g$ will change features of the old classes too. Hence, to minimize the distortion of the old class representations, we can try to enforce $\mathrm{Cov}(\boldsymbol{Z}_{j,old}) = \mathrm{Cov}(\hat{\boldsymbol{Z}}_{j,old})$. In other words, while learning new classes, we enforce that the memory of old classes remains "self-consistent" through the transcription loop:

$$\boldsymbol{Z}_{old} \xrightarrow{\hspace{2mm} g(\boldsymbol{z},\boldsymbol{\eta}) \hspace{2mm}} \hat{\boldsymbol{X}}_{old} \xrightarrow{\hspace{2mm} f(\boldsymbol{x},\boldsymbol{\theta}) \hspace{2mm}} \hat{\boldsymbol{Z}}_{old}. \tag{6.3.2}$$

Mathematically, this is equivalent to setting

$$\Delta R_\epsilon(\boldsymbol{Z}_{old}, \hat{\boldsymbol{Z}}_{old}) \doteq \sum_{j=1}^{t} \Delta R_\epsilon(\boldsymbol{Z}_{j,old}, \hat{\boldsymbol{Z}}_{j,old}) = 0.$$

Hence, the above minimax program (6.3.1) is revised as a *constrained* minimax game, which we refer to as *incremental closed-loop transcription* (i-CTRL). The objective of this game is identical to the standard multi-class CTRL objective (6.2.20), but includes just one additional constraint:

$$\begin{aligned} \min_{\boldsymbol{\eta}} \max_{\boldsymbol{\theta}} \quad & \Delta R_\epsilon(\boldsymbol{Z}) + \Delta R_\epsilon(\hat{\boldsymbol{Z}}) + \Delta R_\epsilon(\boldsymbol{Z}_{new}, \hat{\boldsymbol{Z}}_{new}) \\ \text{subject to} \quad & \Delta R_\epsilon(\boldsymbol{Z}_{old}, \hat{\boldsymbol{Z}}_{old}) = 0. \end{aligned} \tag{6.3.3}$$

In practice, the constrained minimax program can be solved by *alternating* minimization and maximization between the encoder $f(\cdot, \boldsymbol{\theta})$ and decoder $g(\cdot, \boldsymbol{\eta})$ as follows:

$$\max_{\boldsymbol{\theta}} \quad \Delta R_\epsilon(\boldsymbol{Z}) + \Delta R_\epsilon(\hat{\boldsymbol{Z}}) + \lambda \cdot \Delta R_\epsilon(\boldsymbol{Z}_{new}, \hat{\boldsymbol{Z}}_{new}) - \gamma \cdot \Delta R_\epsilon(\boldsymbol{Z}_{old}, \hat{\boldsymbol{Z}}_{old}); \tag{6.3.4}$$

[14] Of course, one may also consider the more general setting where the task contains a small batch of new classes, without serious modification.

$$\min_{\boldsymbol{\eta}} \quad \Delta R_\epsilon(\boldsymbol{Z}) + \Delta R_\epsilon(\hat{\boldsymbol{Z}}) + \lambda \cdot \Delta R_\epsilon(\boldsymbol{Z}_{new}, \hat{\boldsymbol{Z}}_{new}) + \gamma \cdot \Delta R_\epsilon(\boldsymbol{Z}_{old}, \hat{\boldsymbol{Z}}_{old}), \tag{6.3.5}$$

where the constraint $\Delta R_\epsilon(\boldsymbol{Z}_{old}, \hat{\boldsymbol{Z}}_{old}) = 0$ in (6.3.3) has been converted (and relaxed) to a Lagrangian term with a corresponding coefficient $\gamma$ and sign. We additionally introduce another coefficient $\lambda$ for weighting the rate reduction term associated with the new data.

**Jointly optimal memory via incremental reviewing.** As we will see, the above constrained minimax program can already achieve state-of-the-art performance for incremental learning. Nevertheless, developing an optimal memory for *all classes* cannot rely on graceful forgetting alone. Even for humans, if an object class is learned only once, we should expect the learned memory to fade as we continue to learn new ones, unless the memory can be consolidated by reviewing old object classes.

To emulate this phase of memory forming, after incrementally learning a whole dataset, we may go back to review all classes again, one class at a time. We refer to going through all classes once as one reviewing "cycle".[15] If needed, multiple reviewing cycles can be conducted. It is quite expected that reviewing can improve the learned (LDR) memory. But somewhat surprisingly, the closed-loop framework allows us to review even in a "class-unsupervised" manner: when reviewing data of an old class, say $\boldsymbol{X}_j$, the system does not need the class label and can simply treat $\boldsymbol{X}_j$ as a new class $\boldsymbol{X}_{new}$. That is, the system optimizes the same constrained mini-max program (6.3.3) without any modification; after the system is optimized, one can identify the newly learned subspace spanned by $\boldsymbol{Z}_{new}$, and use it to replace or merge with the old subspace $\mathcal{S}_j$. As our experiments show, such a class-unsupervised incremental review process can gradually improve both discriminative and generative performance of the LDR memory, eventually converging to that of a jointly learned memory.

**Experimental verification.** We show some experimental results on the following datasets: MNIST [LBB+98b] and CIFAR-10 [KNH14]. All experiments are conducted for the more challenging class-IL setting. For both MNIST and CIFAR-10, the 10 classes are split into 5 tasks with 2 classes each or 10 tasks with 1 class each. For the encoder $f$ and decoder $g$, we adopt a very simple network architecture modified from DCGAN [RMC16], which is merely a *four-layer* convolutional network. Here we only show some qualitative visual results; more experiments and detailed analysis can be found in the work [TDW+23].

**Visualizing auto-encoding properties.** We begin by qualitatively visualizing some representative images $\boldsymbol{X}$ and the corresponding replayed $\hat{\boldsymbol{X}}$ on MNIST and CIFAR-10. The model is learned incrementally with the datasets split into 5 tasks. Results are shown in Figure 6.14, where we observe that the reconstructed $\hat{\boldsymbol{X}}$ preserves the main visual characteristics of $\boldsymbol{X}$ including shapes and

[15] To distinguish from the term "epoch" used in the conventional joint learning setting.

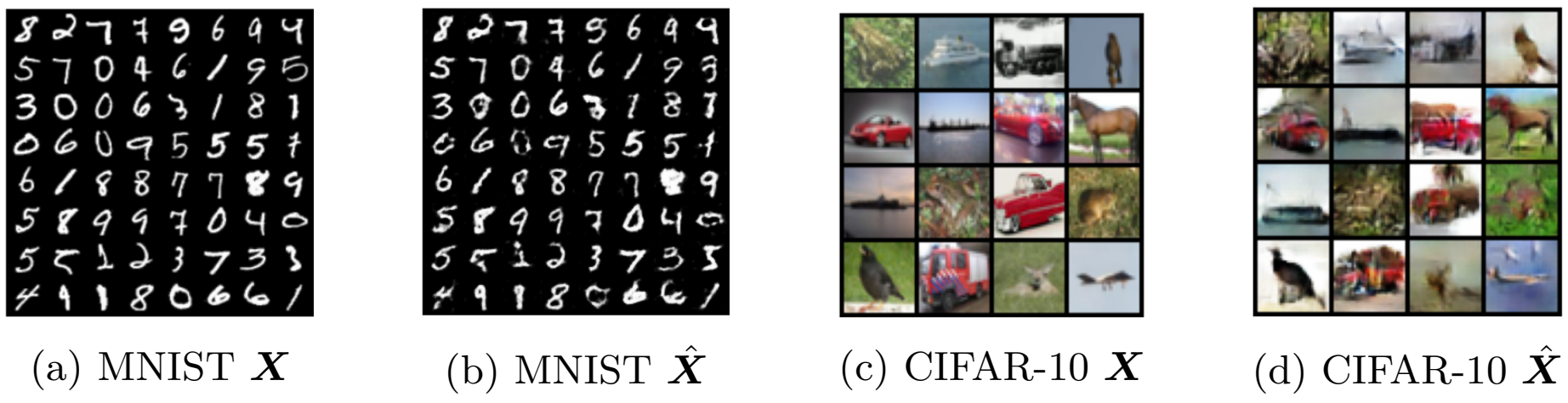

(a) MNIST $\boldsymbol{X}$ (b) MNIST $\hat{\boldsymbol{X}}$ (c) CIFAR-10 $\boldsymbol{X}$ (d) CIFAR-10 $\hat{\boldsymbol{X}}$

Figure 6.14: Visualizing the auto-encoding property of the learned representation ($\hat{\boldsymbol{X}} = g \circ f(\boldsymbol{X})$).

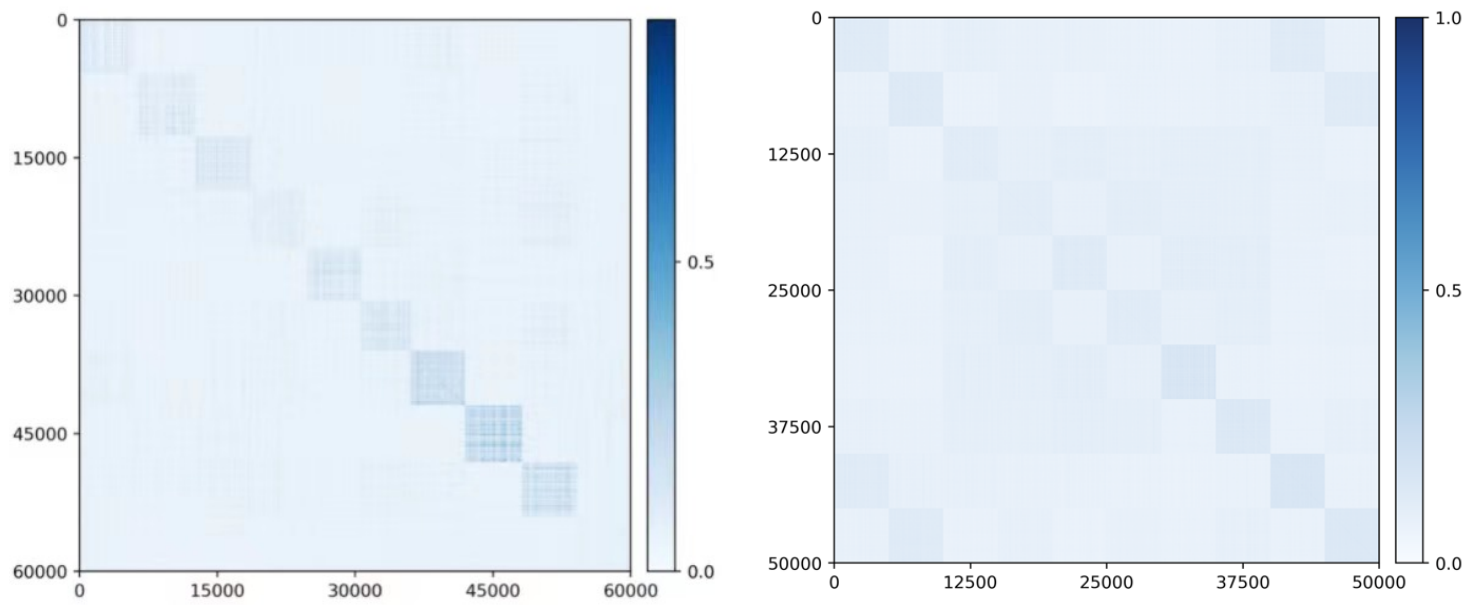

Figure 6.15: Block diagonal structure of $|\boldsymbol{Z}^\top \boldsymbol{Z}|$ in the feature space for MNIST (left) and CIFAR-10 (right).

textures. For a simpler dataset like MNIST, the replayed $\hat{\boldsymbol{X}}$ are almost identical to the input $\boldsymbol{X}$! This is rather remarkable given: (1) our method does not explicitly enforce $\hat{\boldsymbol{x}} \approx \boldsymbol{x}$ for individual samples as most autoencoding methods do, and (2) after having incrementally learned all classes, the generator has not forgotten how to generate digits learned earlier, such as 0, 1, 2. For a more complex dataset like CIFAR-10, we also demonstrate good visual quality, faithfully capturing the essence of each image.

**Principal subspaces of the learned features.** Most generative memory-based methods utilize autoencoders, VAEs, or GANs for replay purposes. The structure or distribution of the learned features $\boldsymbol{Z}_j$ for each class is unclear in the feature space. The features $\boldsymbol{Z}_j$ of the LDR memory, on the other hand, have a clear linear structure. Figure 6.15 visualizes correlations among all learned features $|\boldsymbol{Z}^\top \boldsymbol{Z}|$, in which we observe clear block-diagonal patterns for both datasets.[16] This indicates that the features for different classes $\boldsymbol{Z}_j$ indeed lie on subspaces that are incoherent from one another. Hence, features of each class can be well modeled as a principal subspace in the feature space.

[16]Notice that these patterns closely resemble the similarity matrix of response profiles of object categories from different areas of the inferotemporal cortex, as shown in Extended DataFig.3 of [BSM+20].

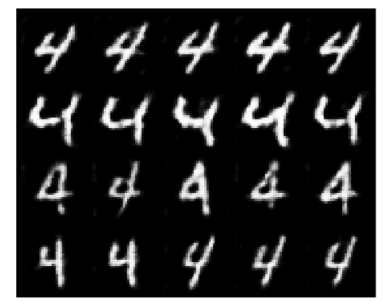

(a) sampled $\hat{\boldsymbol{x}}_{old}$ of '4'

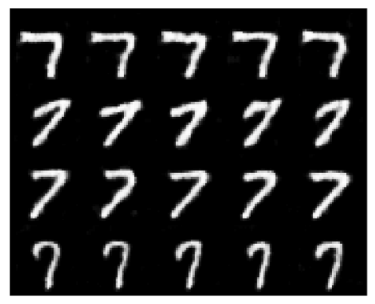

(b) sampled $\hat{\boldsymbol{x}}_{old}$ of '7'

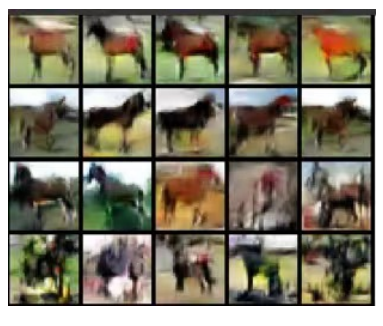

(c) sampled $\hat{\boldsymbol{x}}_{old}$ of 'horse'

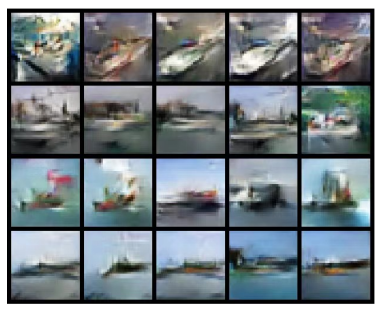

(d) sampled $\hat{\boldsymbol{x}}_{old}$ of 'ship'

Figure 6.16: Visualization of 5 reconstructed $\hat{\boldsymbol{x}} = g(\boldsymbol{z})$ from $\boldsymbol{z}$'s with the closest distance to (top-4) principal components of learned features for MNIST (class '4' and class '7') and CIFAR-10 (class 'horse' and 'ship').

**Replay images of samples from principal components.** Since features of each class can be modeled as a principal subspace, we further visualize the individual principal components within each of those subspaces. Figure 6.16 shows the images replayed from sampled features along the top-4 principal components for different classes, on MNIST and CIFAR-10 respectively. Each row represents samples along one principal component, and they clearly show similar visual characteristics but are distinctively different from those in other rows. We see that the model remembers different poses of '4' after having learned all remaining classes. For CIFAR-10, the incrementally learned memory remembers representative poses and shapes of horses and ships.

**Effectiveness of incremental reviewing.** We verify how the incrementally learned LDR memory can be further consolidated with an unsupervised incremental reviewing phase described before. Experiments are conducted on CIFAR-10, with 10 steps. Figure 6.17 left shows replayed images of the first class 'airplane' at the end of incremental learning of all ten classes, sampled along the top-3 principal components – every two rows (16 images) are along one principal direction. Their visual quality remains very decent—we observe almost no forgetting. The right figure shows replayed images after reviewing the first class once. We notice a significant improvement in visual quality after reviewing, and principal components of the features in the subspace start to correspond to distinctively different visual attributes within the same class.

### 6.3.2 Sample-wise Continuous Unsupervised Learning

As we know, the closed-loop CTRL formulation can already learn a decent autoencoding, even without class information, with the CTRL-Binary program:

$$\max_{\boldsymbol{\theta}} \min_{\boldsymbol{\eta}} \quad \Delta R_\epsilon(\boldsymbol{Z}, \hat{\boldsymbol{Z}}) \tag{6.3.6}$$

However, note that (6.3.6) is practically limited because it only aligns the dataset $\boldsymbol{X}$ and the regenerated $\hat{\boldsymbol{X}}$ at the distribution level. There is no guarantee that each sample $\boldsymbol{x}$ would be close to the decoded $\hat{\boldsymbol{x}} = g(f(\boldsymbol{x}))$.

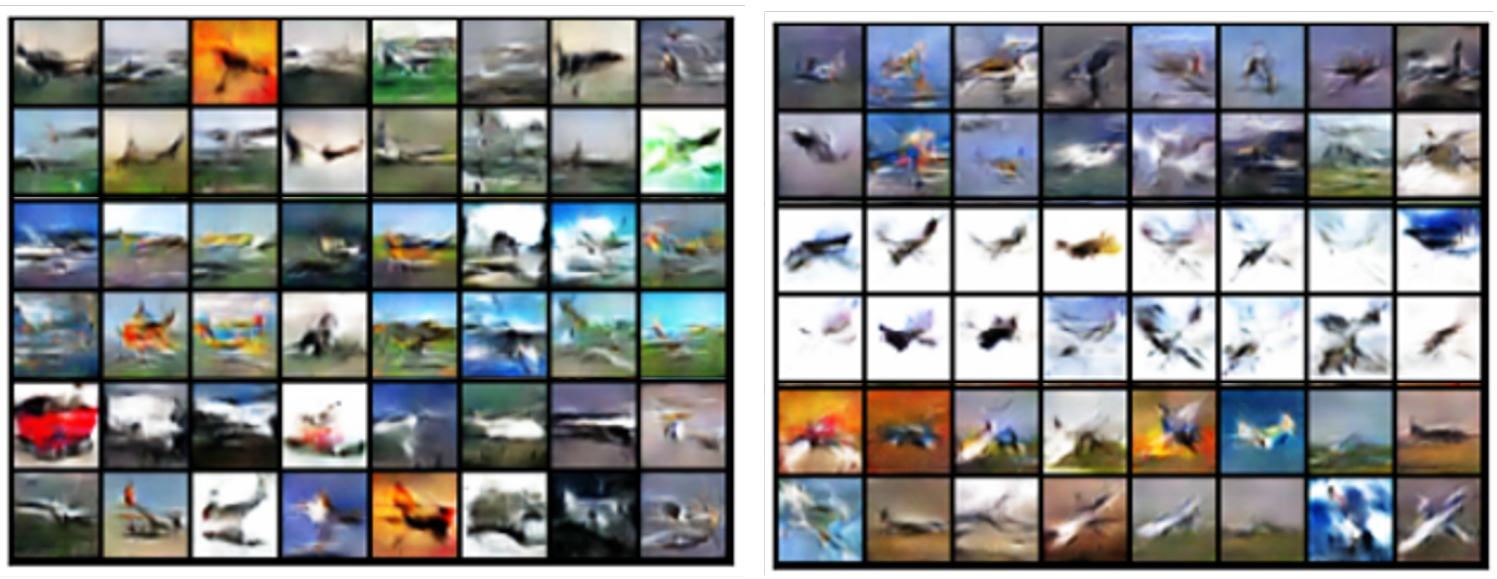

Figure 6.17: Visualization of replayed images $\hat{\boldsymbol{x}}_{old}$ of class 1-'airplane' in CIFAR-10, before (left) and after (right) one reviewing cycle.

**Sample-wise constraints for unsupervised transcription.** To improve discriminative and generative properties of representations learned in the unsupervised setting, we propose two additional mechanisms for the above CTRL-Binary maximin game (6.3.6). For simplicity and uniformity, here these will be formulated as equality constraints over rate reduction measures, but in practice they can be enforced softly during optimization.

**Sample-wise self-consistency via closed-loop transcription.** First, to address the issue that CTRL-Binary does not learn a sample-wise consistent autoencoding, we need to promote $\hat{\boldsymbol{x}}$ to be close to $\boldsymbol{x}$ for each sample. In the CTRL framework, this can be achieved by enforcing their corresponding features $\boldsymbol{z} = f(\boldsymbol{x})$ and $\hat{\boldsymbol{z}} = f(\hat{\boldsymbol{x}})$ to be close. To promote sample-wise self-consistency, where $\hat{\boldsymbol{x}} = g(f(\boldsymbol{x}))$ is close to $\boldsymbol{x}$, we want the distance between $\boldsymbol{z}$ and $\hat{\boldsymbol{z}}$ to be zero or small, for all $N$ samples. This distance can be measured by the rate reduction:

$$\sum_{i \in N} \Delta R_{\epsilon}(\boldsymbol{z}^i, \hat{\boldsymbol{z}}^i) = 0. \tag{6.3.7}$$

Note that this again avoids measuring differences in the image space.

**Self-supervision via compressing augmented samples.** Since we do not know any class label information between samples in the unsupervised setting, the best we can do is to view every sample and its augmentations (say via translation, rotation, occlusion, etc.) as one "class"—a basic idea behind almost all self-supervised learning methods. In the rate reduction framework, it is natural to compress the features of each sample and its augmentations. In this work, we adopt the standard transformations in SimCLR [CKN+20] and denote such a transformation as $\tau$. We denote each augmented sample $\boldsymbol{x}_a = \tau(\boldsymbol{x})$, and its corresponding feature as $\boldsymbol{z}_a = f(\boldsymbol{x}_a, \boldsymbol{\theta})$. For discriminative purposes, we hope the classifier is *invariant* to such transformations. Hence it is natural to enforce that the features $\boldsymbol{z}_a$ of all augmentations are the same as those $\boldsymbol{z}$ of the original sample $\boldsymbol{x}$. This is equivalent to requiring the distance between $\boldsymbol{z}$ and

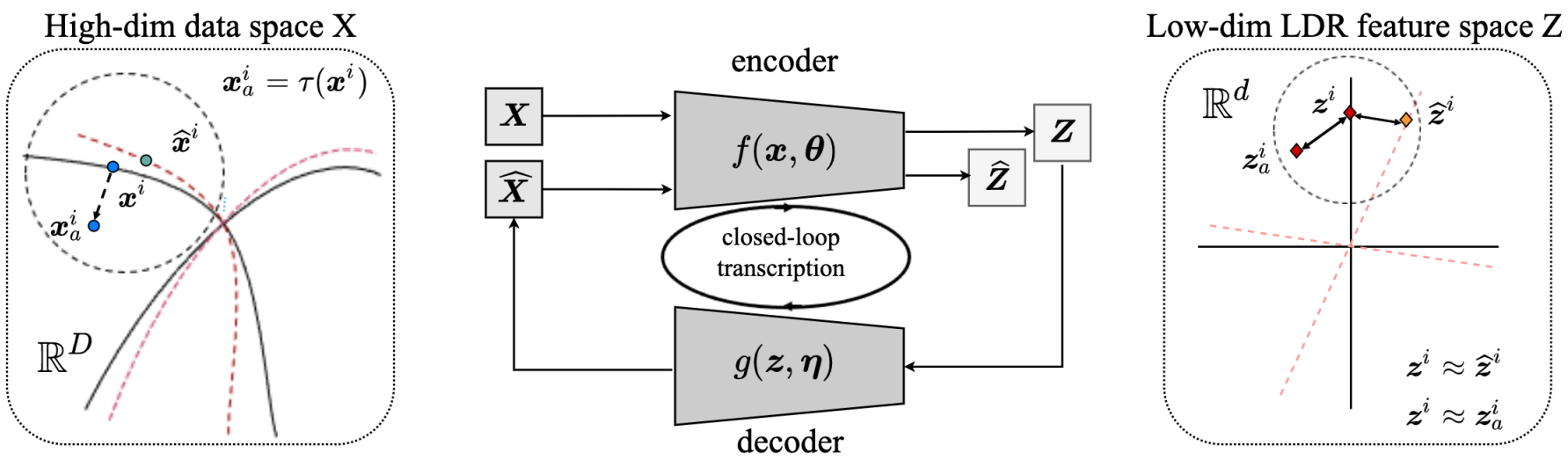

Figure 6.18: **Overall framework** of closed-loop transcription for unsupervised learning. Two additional constraints are imposed on the CTRL-Binary method: 1) self-consistency for sample-wise features $\boldsymbol{z}^i$ and $\hat{\boldsymbol{z}}^i$, say $\boldsymbol{z}^i \approx \hat{\boldsymbol{z}}^i$; and 2) invariance/similarity among features of augmented samples $\boldsymbol{z}^i$ and $\boldsymbol{z}_a^i$, say $\boldsymbol{z}^i \approx \boldsymbol{z}_a^i = f(\tau(\boldsymbol{x}^i), \boldsymbol{\theta})$, where $\boldsymbol{x}_a^i = \tau(\boldsymbol{x}^i)$ is an augmentation of sample $\boldsymbol{x}^i$ via some transformation $\tau(\cdot)$.

$\boldsymbol{z}_a$, measured in terms of rate reduction again, to be zero (or small) for all $N$ samples:

$$\sum_{i \in N} \Delta R_\epsilon(\boldsymbol{z}^i, \boldsymbol{z}_a^i) = 0. \tag{6.3.8}$$

**Unsupervised representation learning via closed-loop transcription.** So far, we know the CTRL-Binary objective $\Delta R_\epsilon(\boldsymbol{Z}, \hat{\boldsymbol{Z}})$ in (6.3.6) helps align the distributions while sample-wise self-consistency (6.3.7) and sample-wise augmentation (6.3.8) help align and compress features associated with each sample. Besides consistency, we also want learned representations to be maximally discriminative for different samples (here viewed as different “classes”). Notice that the rate distortion term $R_\epsilon(\boldsymbol{Z})$ measures the coding rate (hence volume) of all features.

**Unsupervised CTRL.** Putting these elements together, we propose to learn a representation via the following constrained maximin program, which we refer to as *unsupervised CTRL* (u-CTRL):

$$\begin{aligned} \max_{\boldsymbol{\theta}} \min_{\boldsymbol{\eta}} \quad & R_\epsilon(\boldsymbol{Z}) + \Delta R_\epsilon(\boldsymbol{Z}, \hat{\boldsymbol{Z}}) \\ \text{subject to} \quad & \sum_{i \in N} \Delta R_\epsilon(\boldsymbol{z}^i, \hat{\boldsymbol{z}}^i) = 0, \;\; \text{and} \;\; \sum_{i \in N} \Delta R_\epsilon(\boldsymbol{z}^i, \boldsymbol{z}_a^i) = 0. \end{aligned} \tag{6.3.9}$$

Figure 6.18 illustrates the overall architecture of the closed-loop system associated with this program.

In practice, the above program can be optimized by alternating maximization and minimization between the encoder $f(\cdot, \boldsymbol{\theta})$ and the decoder $g(\cdot, \boldsymbol{\eta})$. We adopt the following optimization strategy that works well in practice, which is used for all subsequent experiments on real image datasets:

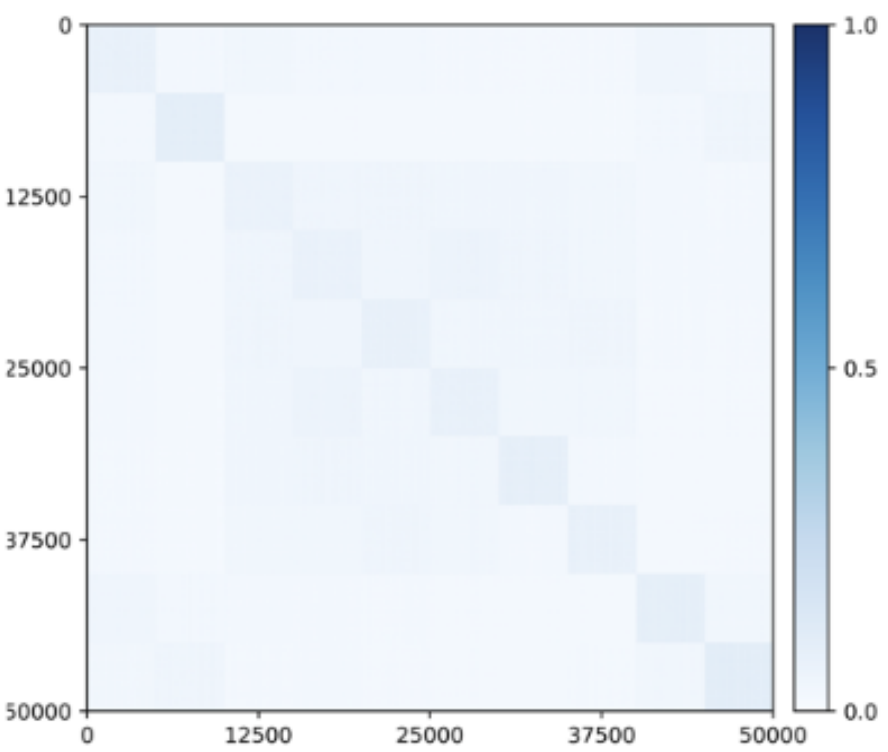

Figure 6.19: Emergence of block-diagonal structures of $|\boldsymbol{Z}^\top \boldsymbol{Z}|$ in the feature space for CIFAR-10.

$$\max_{\boldsymbol{\theta}} \; R_\epsilon(\boldsymbol{Z}) + \Delta R_\epsilon(\boldsymbol{Z}, \hat{\boldsymbol{Z}}) - \lambda_1 \sum_{i \in N} \Delta R_\epsilon(\boldsymbol{z}^i, \boldsymbol{z}_a^i) - \lambda_2 \sum_{i \in N} \Delta R_\epsilon(\boldsymbol{z}^i, \hat{\boldsymbol{z}}^i) \quad (6.3.10)$$

$$\min_{\boldsymbol{\eta}} \; R_\epsilon(\boldsymbol{Z}) + \Delta R_\epsilon(\boldsymbol{Z}, \hat{\boldsymbol{Z}}) + \lambda_1 \sum_{i \in N} \Delta R_\epsilon(\boldsymbol{z}^i, \boldsymbol{z}_a^i) + \lambda_2 \sum_{i \in N} \Delta R_\epsilon(\boldsymbol{z}^i, \hat{\boldsymbol{z}}^i), \quad (6.3.11)$$

where the constraints $\sum_{i \in N} \Delta R_\epsilon(\boldsymbol{z}^i, \hat{\boldsymbol{z}}^i) = 0$ and $\sum_{i \in N} \Delta R_\epsilon(\boldsymbol{z}^i, \boldsymbol{z}_a^i) = 0$ in (6.3.9) have been converted (and relaxed) to Lagrangian terms with corresponding coefficients $\lambda_1$ and $\lambda_2$.[17]

The above representation is learned without class information. In order to facilitate discriminative or generative tasks, it must be highly structured. It has been verified experimentally that this is indeed the case and u-CTRL demonstrates significant advantages over other incremental or unsupervised learning methods [TDC+24]. We here only illustrate some qualitative results with the experiment on the CIFAR-10 dataset [KNH14], with standard augmentations for self-supervised learning [CKN+20]. One may refer to [TDC+24] for experiments on more and larger datasets and their quantitative evaluations.

As one can see from the experiments, specific and unique structure indeed emerges naturally in the representations learned using u-CTRL: globally, features of images in the same class tend to be clustered well together and separated from other classes, as shown in Figure 6.19; locally, features of individual classes cluster into tight, well-separated groups, as shown in Figure 6.20.

**Unsupervised conditional image generation via rate reduction.** The highly-structured feature distribution also suggests that the learned representa-tion can be very useful for generative purposes. For example, we can organize the sample features into meaningful clusters, and model them with low-dimensional (Gaussian) distributions or subspaces. By sampling from these compact models, we can conditionally regenerate meaningful samples from computed clusters. This is known as *unsupervised conditional image generation* [HKJ+21].

[17]Notice that computing the rate reduction terms $\Delta R$ for all samples or a batch of samples requires computing the expensive $\log\det$ of large matrices. In practice, from the geometric meaning of $\Delta R$ for two vectors, $\Delta R$ can be approximated with an $\ell^2$ norm or the cosine distance between two vectors. This approximation is developed further in Section 4.3.2, where it motivates the SimDINO objective.

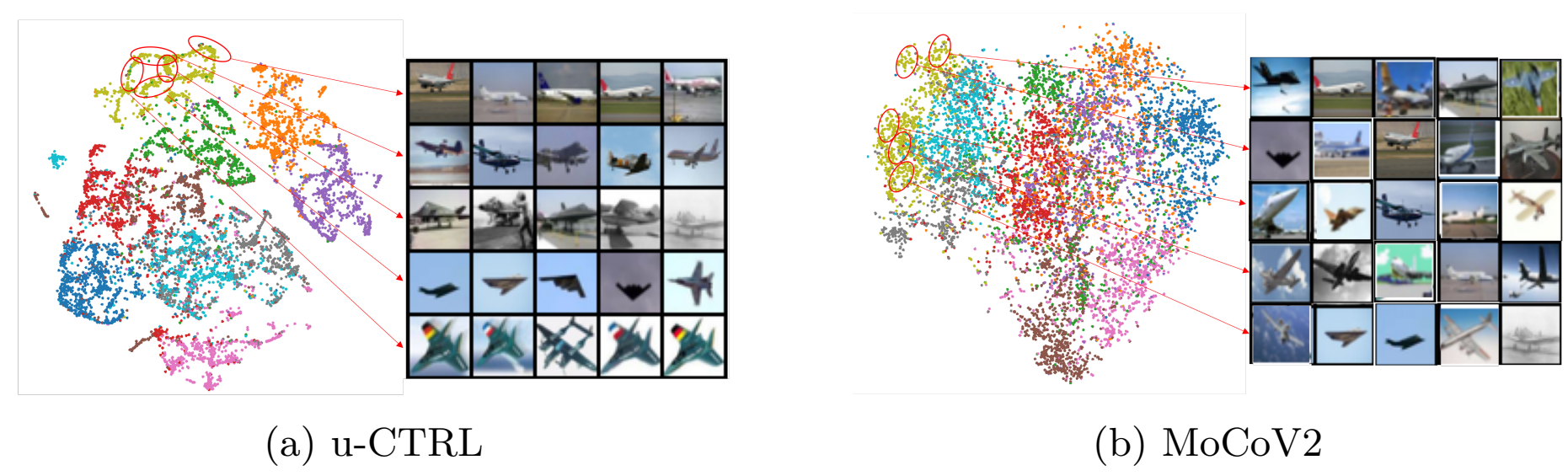

(a) u-CTRL (b) MoCoV2

Figure 6.20: t-SNE visualizations of learned features of CIFAR-10 with different models.

To cluster features, we exploit the fact that the rate reduction framework (4.2.15) is inspired by unsupervised clustering via compression [MDH+07b], which provides a principled way to find the membership $\mathbf{\Pi}$. Concretely, we maximize the same rate reduction objective (4.2.15) over $\mathbf{\Pi}$, but fix the learned representation $\boldsymbol{Z}$ instead. We simply view the membership $\mathbf{\Pi}$ as a nonlinear function of the features $\boldsymbol{Z}$, say $h_{\boldsymbol{\pi}}(\cdot, \xi) : \boldsymbol{Z} \mapsto \mathbf{\Pi}$ with parameters $\xi$. In practice, we model this function with a simple neural network, such as an MLP head right after the output feature $\boldsymbol{z}$. To estimate a "pseudo" membership $\hat{\mathbf{\Pi}}$ of the samples, we solve the following optimization problem over $\mathbf{\Pi}$:

$$\hat{\mathbf{\Pi}} = \arg\max_{\xi} \Delta R_\epsilon(\boldsymbol{Z} \mid \mathbf{\Pi}(\xi)). \tag{6.3.12}$$

In Figure 6.21, we visualize images generated from the ten unsupervised clusters from (6.3.12). Each block represents one cluster and each row represents one principal component for each cluster. Despite learning and training without labels, the model not only organizes samples into correct clusters, but is also able to preserve statistical diversities within each cluster/class. We can easily recover the diversity within each cluster by computing different principal components and then sampling and generating accordingly. While the experiments presented here are somewhat limited in scale, we will explore more direct and powerful methods that utilize the learned data distributions and representations for conditional generation and estimation in the next chapter.

## 6.4 Summary and Notes

Historically, autoencoding has been one of the important drivers of research innovation in neural networks for learning, although the most practically impressive demonstrations of deep learning have probably been in other domains (such as discriminative classification, with AlexNet [KSH12], or generative modeling with GPT architectures [BMR+20]). Works we have featured throughout the chapter, especially the work of [HS06], served as catalysts of research interest in neural networks during times when they were otherwise not prominent in the machine learning research landscape. In modern practice, autoencoders remain core components of many large-scale systems for generating highly structured data such as visual data, speech data, and molecular data, especially the VQ-VAE approach [OVK17], which builds on the variational autoencoder methodology we discussed in Section 6.1.4. The core problem of autoencoding remains of paramount intellectual importance due to its close connection with representation learning, and we anticipate that it will reappear on the radar of practical researchers in the future as efficiency in training and deploying large models continues to become more important.

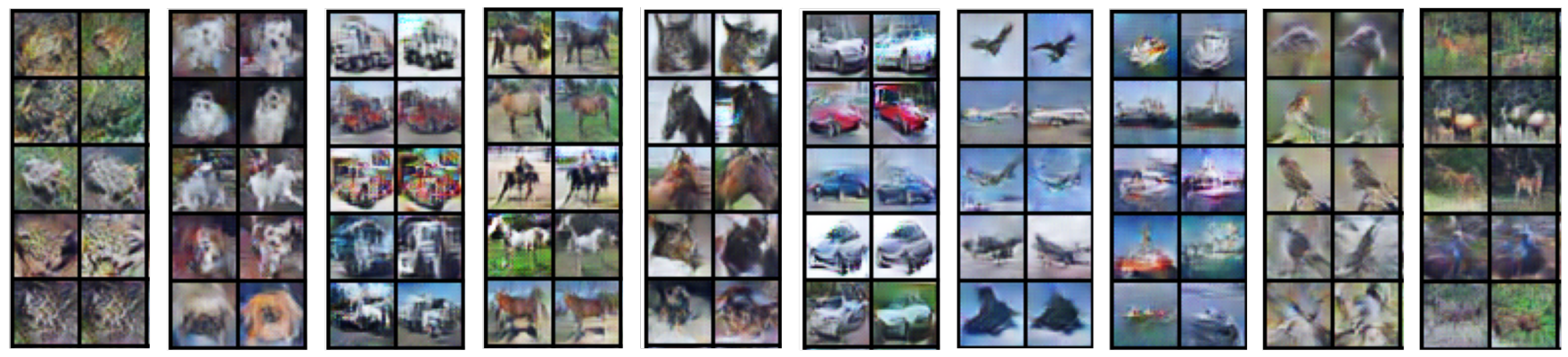

Figure 6.21: Unsupervised conditional image generation from each cluster of CIFAR-10, using u-CTRL. Images from different rows mean generation from different principal components of each cluster.

Materials presented in the second half of this chapter are based on a series of recent works on the topic of closed-loop transcription: [DTL+22], [PPC+23], [TDW+23], and [TDC+24]. In particular, Section 6.2.1 is based on the pioneering work of [DTL+22]. After that, the work of [PPC+23] has provided strong theoretical justifications for the closed-loop framework, at least for an ideal case. Section 6.3.1 and Section 6.3.2 are based on the works of [TDW+23] and [TDC+24], respectively. They demonstrate that the closed-loop framework naturally supports incremental and continuous learning, either in a class-wise or sample-wise setting. The reader may refer to these papers for more technical and experimental details.

**Shallow vs. deep neural networks, for autoencoding and more.** In Section 6.1.2, we discussed Cybenko's universal approximation theorem and how it states that in principle, a neural network with a single hidden layer (and suitable elementwise nonlinearities) is sufficient to approximate any suitably regular target function. Of course, the major architectural reason for the dominance of neural networks in practice has been the refinement of techniques for training *deeper* neural networks. Why is depth necessary? From a fundamental point of view, the issue of depth separations, which construct settings where a deeper neural network can approximate a given class of target functions with exponentially-superior efficiency relative to a shallow network, has been studied at great length in the theoretical literature: examples include [BN20; Tel16; VJO+21]. The ease of training deeper networks in practice has not received as satisfying an answer from the perspective of theory. ResNets [HZR+16b] represent the pioneering empirical work of making deeper networks more easily trainable, used in nearly all modern architectures in some form. Theoretical studies have focused heavily on trainability of very deep networks, quantified via the initial neural tangent kernel [BGW21; MBD+21], but these studies have not given significant insight into the trainability benefits of deeper networks in middle and late stages of training (but see [YH21] for the root of a line of research attempting to address this).

## 6.5 Exercises and Extensions

*Exercise* 6.1 (Conceptual Understanding of Manifold Flattening). Consider data lying on a curved manifold $\mathcal{M}$ embedded in $\mathbb{R}^D$ (like a curved surface in 3-dimensional space), as discussed in the manifold flattening subsection of Section 6.1.2. In this exercise, we will describe the basic ingredients of the manifold flattening algorithm from [PPR+24]. A manifold is called flat if it is an open set in Euclidean space (or more generally, an open set in a subspace).

1. Suppose the manifold $\mathcal{M}$ is a *graph*: this means that $\mathcal{M} \subset \mathbb{R}^{D_1} \times \mathbb{R}^{D_2}$ (say), and that there is a function $F : \mathbb{R}^{D_1} \to \mathbb{R}^{D_2}$ such that

$$\mathcal{M} = \{(\boldsymbol{x}, F(\boldsymbol{x})) \mid \boldsymbol{x} \in \mathbb{R}^{D_1}\}. \tag{6.5.1}$$

   Give an integer $D_3$ and a map $f : \mathbb{R}^{D_1} \times \mathbb{R}^{D_2} \to \mathbb{R}^{D_3}$ that flattens $\mathcal{M}$, and describe the corresponding (lossless) reconstruction procedure from the flattened representation.

2. Now suppose that $\mathcal{M}$ is a general smooth manifold. Smooth manifolds have the property that they are locally well-approximated by subspaces near each point. Describe in intuitive terms how to flatten the manifold $\mathcal{M}$ locally at each point, by relating it to a graph. (The algorithm of [PPR+24] performs a refined version of this local flattening process in a way that allows them to be glued together to form a global flattening.)

*Exercise* 6.2 (Reproduce Closed-Loop Transcription). Implement a closed-loop transcription pipeline for representation learning on the CIFAR-10 dataset following the methodology in Section 6.2.1. Reference [DTL+22] for useful hyperparameters and architecture settings. Reproduce the result in Figure 6.11.

# Chapter 7

# Inference with Low-Dimensional Distributions

> "*Mathematics is the art of giving the same name to different things.*"
>
> — Henri Poincaré

In the previous Chapters of this book, we have studied how to effectively and efficiently learn a representation for a variable $\boldsymbol{x}$ in the world with a distribution $p(\boldsymbol{x})$ that has a low-dimensional support in a high-dimensional space. So far, we have mainly developed the methodology for learning representation and autoencoding in a general, distribution or task-agnostic fashion. With such a learned representation, one can already use it to perform some generic and basic tasks such as classification (if the encoding is supervised with the class) and generation of random samples that have the same distribution as the given data (say natural images or natural languages).

More generally, however, the universality and scalability of the theoretical and computational framework presented in this book has enabled us to learn the distribution of a variety of important real-world high-dimensional data such as natural languages, human poses, natural images, videos, and even 3D scenes. Once the intrinsically rich and low-dimensional structures of these real data can be learned and represented correctly, they start to enable a broad family of powerful, often seemingly miraculous, tasks. Hence, from here onwards, we will start to show how to connect and tailor the general methods presented in previous Chapters to learn useful representations for specific structured data distributions and for many popular tasks in modern practice of machine intelligence.

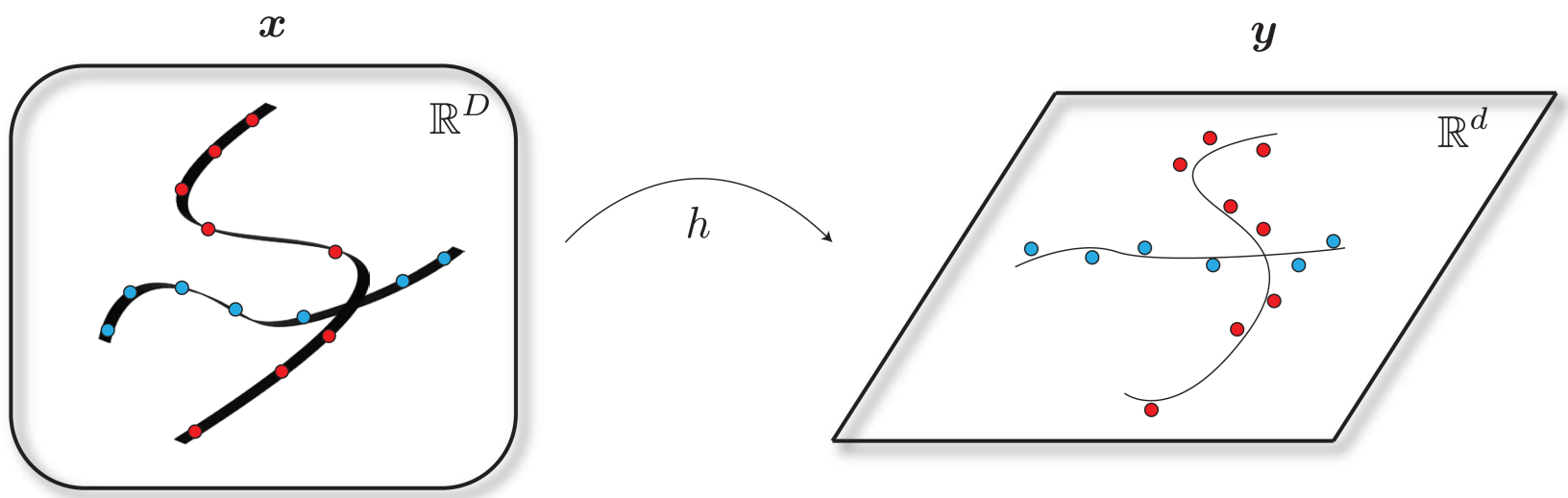

Figure 7.1: **Inference with low-dimensional distributions.** This is the generic picture for this chapter: we have a low-dimensional distribution for $\boldsymbol{x} \in \mathbb{R}^D$ (here depicted as a union of two 2-dimensional manifolds in $\mathbb{R}^3$) and a measurement model $\boldsymbol{y} = h(\boldsymbol{x}) + \boldsymbol{w} \in \mathbb{R}^d$. We want to infer various things about this model, including the conditional distribution of $\boldsymbol{x}$ given $\boldsymbol{y}$, or the conditional expectation $\mathbb{E}[\boldsymbol{x} \mid \boldsymbol{y}]$, given various information about the model and (potentially finite) samples of either $\boldsymbol{x}$ or $\boldsymbol{y}$.

## 7.1 Bayesian Inference and Constrained Optimization

### 7.1.1 Bayesian Inference with Low-Dimensional Distributions

**Leveraging low-dimensionality for stable and robust inference.** Generally speaking, a good representation or autoencoding should enable us to utilize the learned low-dimensional distribution of the data $\boldsymbol{x}$ and its representation $\boldsymbol{z}$ for various subsequent classification, estimation, and generation tasks under different conditions. As we have alluded to earlier in Chapter 1 Section 1.2.2, the importance of the *low-dimensionality* of the distribution is the key for us to conduct stable and robust inference related to the data $\boldsymbol{x}$, as illustrated by the few simple examples in Figure 1.11, from incomplete, noisy, and even corrupted observations. As it turns out, the very same concept carries over to real-world high-dimensional data whose distributions have a low-dimensional support, such as natural images and languages.

Despite a dazzling variety of applications in the practice of machine learning with data such as languages, images, videos and many other modalities, almost all practical applications can be viewed as a special case of the following inference problem: given an observation $\boldsymbol{y}$ that depends on $\boldsymbol{x}$, say

$$\boldsymbol{y} = h(\boldsymbol{x}) + \boldsymbol{w}, \tag{7.1.1}$$

where $h(\cdot)$ represents measurements of a part of $\boldsymbol{x}$ or certain observed attributes and $\boldsymbol{w}$ represents some measurement noise and even (sparse) corruptions, solve the "inverse problem" of obtaining a most likely estimate $\hat{\boldsymbol{x}}(\boldsymbol{y})$ of $\boldsymbol{x}$ or generating a sample $\hat{\boldsymbol{x}}$ that is at least consistent with the observation $\boldsymbol{y} \approx h(\hat{\boldsymbol{x}})$. Figure 7.1 illustrates the general relationship between $\boldsymbol{x}$ and $\boldsymbol{y}$.

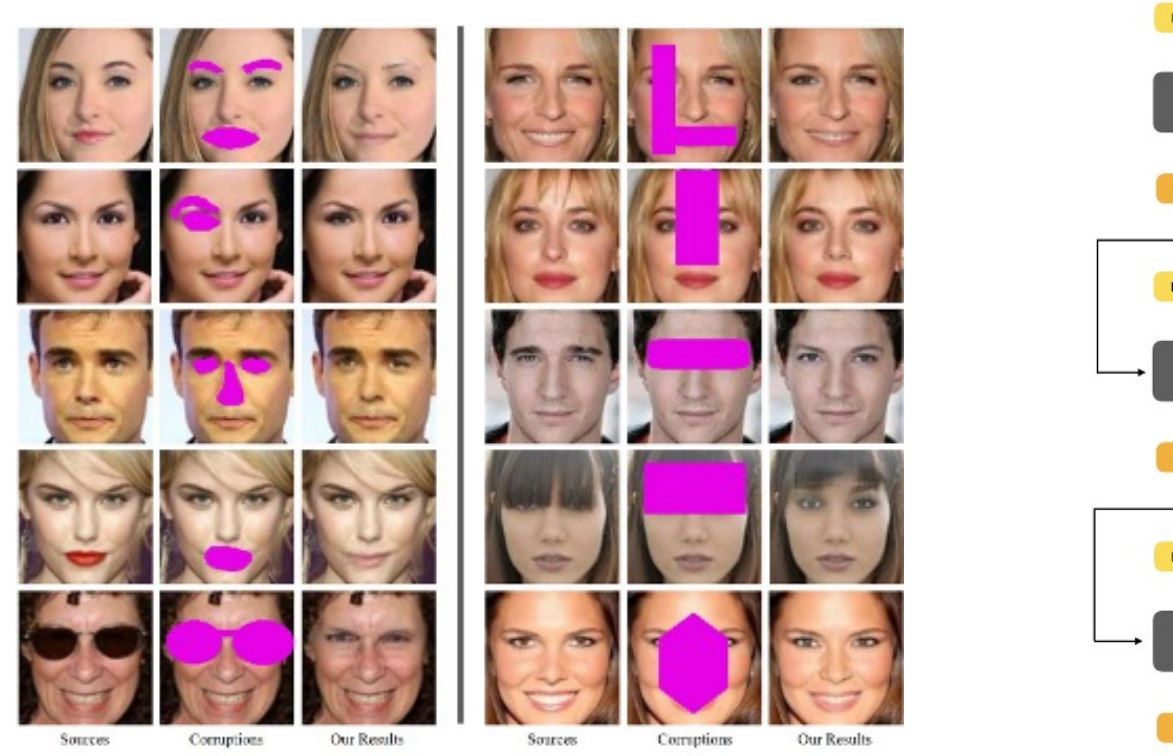

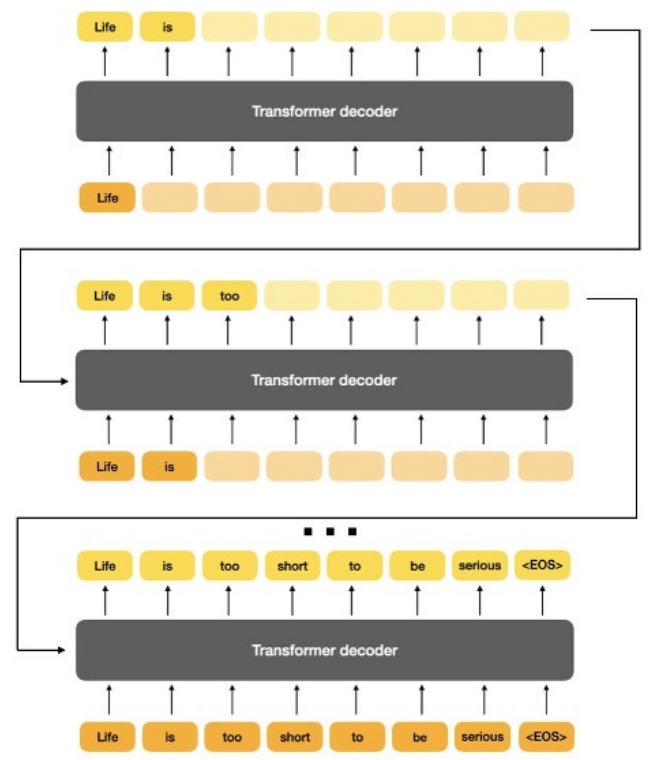

Figure 7.2: Left: image completion. Right: text prediction. In particular, text prediction is the inspiration for the popular Generative Pre-trained Transformer (GPT).

*Example* 7.1 (Image Completion and Text Prediction). The popular natural image completion and natural language prediction are two typical tasks that require us to recover a full data $\boldsymbol{x}$ from its partial observations $\boldsymbol{y}$, with parts of $\boldsymbol{x}$ masked out and to be completed based on the rest. Figure 7.2 shows some examples of such tasks. In fact, it is precisely these tasks which have inspired how to train modern large models for text generation (such as GPT) and image completion (such as the masked autoencoder) that we will study in greater detail later. ■

**Statistical interpretation via Bayes' rule.** Generally speaking, to accomplish such tasks well, we need to get ahold of the conditional distribution $p(\boldsymbol{x} \mid \boldsymbol{y})$. If we had this, then we would be able to find the maximal likelihood estimate (prediction):

$$\hat{\boldsymbol{x}} = \arg\max_{\boldsymbol{x}} p(\boldsymbol{x} \mid \boldsymbol{y}); \tag{7.1.2}$$

or compute the conditional expectation estimate:

$$\hat{\boldsymbol{x}} = \mathbb{E}[\boldsymbol{x} \mid \boldsymbol{y}] = \int \boldsymbol{x} p(\boldsymbol{x} \mid \boldsymbol{y}) \mathrm{d}\boldsymbol{x}; \tag{7.1.3}$$

or sample from the conditional distribution:

$$\hat{\boldsymbol{x}} \sim p(\boldsymbol{x} \mid \boldsymbol{y}). \tag{7.1.4}$$

Notice that if the conditional distribution $p(\boldsymbol{x} \mid \boldsymbol{y})$ has a low-dimensional support that is nonlinear, these three different estimates can be rather different, as illustrated in Figure 7.3. Conceptually, the maximum a posteriori (MAP) estimate is the most desired one—it is the sample of the highest probability to have produced the observation $\boldsymbol{y}$, but it is typically the most expensive to compute.

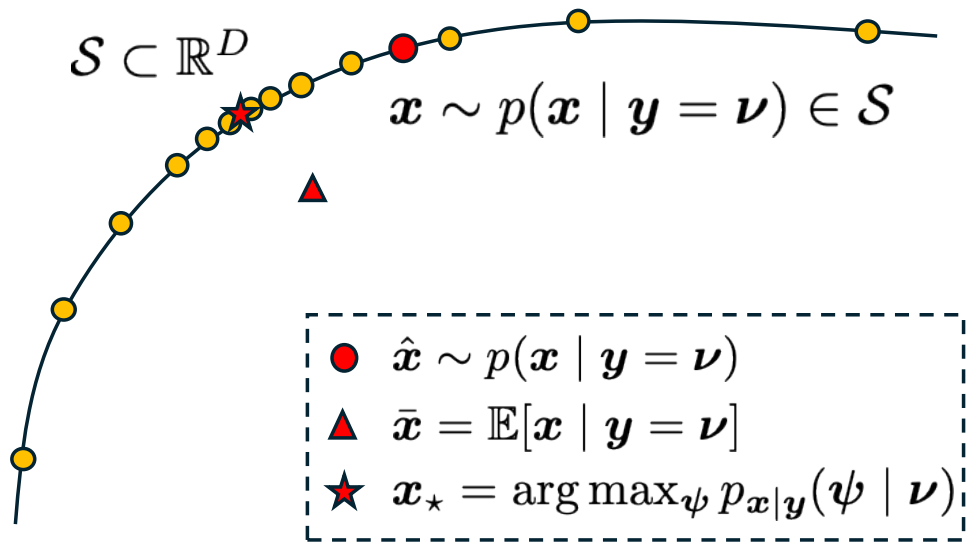

Figure 7.3: Comparison of three different surrogates for an estimate of $\boldsymbol{x}$ given $\boldsymbol{y}$.

Notice that from Bayes' rule, we have

$$p(\boldsymbol{x} \mid \boldsymbol{y}) = \frac{p(\boldsymbol{y} \mid \boldsymbol{x})p(\boldsymbol{x})}{p(\boldsymbol{y})}. \tag{7.1.5}$$

For instance, the maximal likelihood estimate can be computed by solving the following (maximal log likelihood) program:

$$\hat{\boldsymbol{x}} = \arg\max_{\boldsymbol{x}}[\log p(\boldsymbol{y} \mid \boldsymbol{x}) + \log p(\boldsymbol{x})], \tag{7.1.6}$$

say via gradient ascent:

$$\boldsymbol{x}_{k+1} = \boldsymbol{x}_k + \alpha \cdot \big(\nabla_{\boldsymbol{x}} \log p(\boldsymbol{y} \mid \boldsymbol{x}) + \nabla_{\boldsymbol{x}} \log p(\boldsymbol{x})\big). \tag{7.1.7}$$

Efficiently computing the conditional distribution $p(\boldsymbol{x} \mid \boldsymbol{y})$ naturally depends on how we learn and exploit the low-dimensional distribution $p(\boldsymbol{x})$ of the data $\boldsymbol{x}$ and the observation model $\boldsymbol{y} = h(\boldsymbol{x}) + \boldsymbol{w}$ that determines the conditional distribution $p(\boldsymbol{y} \mid \boldsymbol{x})$.

*Remark* 7.1 (End-to-End versus Bayesian Inference). In the modern practice of data-driven machine learning, for certain popular tasks people often directly learn the conditional distribution $p(\boldsymbol{x} \mid \boldsymbol{y})$ or a (probabilistic) mapping or a regressor. Such a mapping is often modeled by some deep networks and trained end-to-end with sufficient paired samples $(\boldsymbol{x}, \boldsymbol{y})$. Such an approach is very different from the above Bayesian approach in which both the distribution of $\boldsymbol{x} \sim p(\boldsymbol{x})$ and the (observation) mapping are needed. The benefit of the Bayesian approach is that the learned distribution $p(\boldsymbol{x})$ can facilitate many different tasks with varied observation models and conditions.

### 7.1.2 Constrained Optimization with Submanifolds

**Geometric interpretation as constrained optimization.** As the support $\mathcal{S}_{\boldsymbol{x}}$ of the distribution of $\boldsymbol{x}$ is low-dimensional, we may assume that there exists

a function $F$ such that

$$F(\boldsymbol{x}) = 0 \quad\iff\quad \boldsymbol{x} \in \mathcal{S}_{\boldsymbol{x}} \tag{7.1.8}$$

such that $\mathcal{S}_{\boldsymbol{x}} = F^{-1}(\{0\})$ is the low-dimensional support of the distribution $p(\boldsymbol{x})$. Geometrically, one natural choice of $F(\boldsymbol{x})$ is the "distance function" to the support $\mathcal{S}_{\boldsymbol{x}}$:

$$F(\boldsymbol{x}) = \min_{\boldsymbol{x}_p \in \mathcal{S}_{\boldsymbol{x}}} \|\boldsymbol{x} - \boldsymbol{x}_p\|_2. \tag{7.1.9}$$

Notice that, in reality, we only have discrete samples on the support of the distribution. In the same spirit of continuation, through diffusion or lossy coding studied in Chapters 3 and 4, we may approximate the distance function as $F(\boldsymbol{x}) \approx \min_{\boldsymbol{x}_p \in \mathcal{C}^{\epsilon}_{\boldsymbol{x}}} \|\boldsymbol{x} - \boldsymbol{x}_p\|_2$ where $\mathcal{S}_{\boldsymbol{x}}$ is replaced by a covering $\mathcal{C}^{\epsilon}_{\boldsymbol{x}}$ of the samples with $\epsilon$-balls. But if the analytical form of a simple distribution is given, sometimes $F(\boldsymbol{x})$ can be computed explicitly.

*Example* 7.2 (Distance to a Line in $\mathbb{R}^3$). Let us assume that the support of a low-dimensional distribution in $\mathbb{R}^3$ is the $x_3$-axis. Then the distance function $F(\boldsymbol{x})$ is given by:

$$F(\boldsymbol{x}) = \sqrt{x_1^2 + x_2^2}. \tag{7.1.10}$$

■

To see how such a function plays an important role in exploiting the distribution, for simplicity, we will assume that, for the rest of the subsection, the distance function is already given.

Now given $\boldsymbol{y} = h(\boldsymbol{x}) + \boldsymbol{w}$, to solve for $\boldsymbol{x}$, we can solve the following constrained optimization problem:

$$\max_{\boldsymbol{x}} -\frac{1}{2}\|h(\boldsymbol{x}) - \boldsymbol{y}\|_2^2 \quad \text{s.t.} \quad F(\boldsymbol{x}) = 0. \tag{7.1.11}$$

Using the method of augmented Lagrange multipliers, we can solve the following unconstrained program:

$$\max_{\boldsymbol{x}} \left[-\frac{1}{2}\|h(\boldsymbol{x}) - \boldsymbol{y}\|_2^2 + \lambda F(\boldsymbol{x}) - \frac{\mu}{2}F(\boldsymbol{x})^2\right] \tag{7.1.12}$$

for some constant Lagrange multiplier $\lambda$. This is equivalent to the following program:

$$\max_{\boldsymbol{x}} \left[\log\exp\Big(-\frac{1}{2}\|h(\boldsymbol{x}) - \boldsymbol{y}\|_2^2\Big) + \log\exp\Big(-\frac{\mu}{2}\big(F(\boldsymbol{x}) - \lambda/\mu\big)^2\Big)\right], \tag{7.1.13}$$

where $c \doteq \lambda/\mu$ can be viewed as a "mean" for the constraint function. As $\mu$ becomes large when enforcing the constraint via continuation[1], $c$ becomes

[1] In the same spirit of continuation in Chapter 3 where we obtained better approximations of our distribution by sending $\varepsilon \to 0$, here we send $\mu \to \infty$. Larger values of $\mu$ will constrain $F$ to take smaller and smaller values at the optimum, meaning that the optimum lies within a smaller and smaller neighborhood of the support $\mathcal{S}_{\boldsymbol{x}}$. Interestingly, the theory of Lagrange multipliers hints that, under certain benign conditions on $F$ and other terms in the objective, we only need to make $\mu$ large enough in order to ensure $F(\boldsymbol{x}) = 0$ at the optimum, meaning that at *finite* penalty we get *perfect* approximation of the support. In general, we should have the intuition that $\mu$ plays the same role as $\varepsilon^{-1}$.

increasingly small. The above program may be interpreted in two different ways.

Firstly, one may view the first term as the conditional probability of $\boldsymbol{y}$ given $\boldsymbol{x}$, and the second term as a probability density for $\boldsymbol{x}$:

$$p(\boldsymbol{y} \mid \boldsymbol{x}) \propto \exp\Big(-\frac{1}{2}\|h(\boldsymbol{x}) - \boldsymbol{y}\|_2^2\Big), \quad p(\boldsymbol{x}) \propto \exp\Big(-\frac{\mu}{2}(F(\boldsymbol{x}) - c)^2\Big). \tag{7.1.14}$$

Hence, solving the constrained optimization for the inverse problem is equivalent to conducting Bayes inference with the above probability densities. Hence solving the above program (7.1.13) via gradient ascent is equivalent to the above maximal likelihood estimate (7.1.7), in which the gradient takes the form:

$$\begin{aligned} & \nabla_{\boldsymbol{x}} \log p(\boldsymbol{y} \mid \boldsymbol{x}) + \nabla_{\boldsymbol{x}} \log p(\boldsymbol{x}) & (7.1.15) \\ = \; & -(h(\boldsymbol{x}) - \boldsymbol{y})^\top \frac{\partial h}{\partial \boldsymbol{x}}(\boldsymbol{x}) - \mu(F(\boldsymbol{x}) - c)\frac{\partial F}{\partial \boldsymbol{x}}(\boldsymbol{x}), & (7.1.16) \end{aligned}$$

where $\frac{\partial h}{\partial \boldsymbol{x}}(\boldsymbol{x})$ and $\frac{\partial F}{\partial \boldsymbol{x}}(\boldsymbol{x})$ are the Jacobian of $h(\boldsymbol{x})$ and $F(\boldsymbol{x})$, respectively.

*Example* 7.3 (Gradient of $F(\boldsymbol{x})$). For the distance function defined in Example 7.2, its gradient is given by

$$\frac{\partial F}{\partial \boldsymbol{x}}(\boldsymbol{x}) = \frac{1}{\sqrt{x_1^2 + x_2^2}} \begin{bmatrix} x_1 \\ x_2 \\ 0 \end{bmatrix}. \tag{7.1.17}$$

Notice that unlike regular smooth functions whose gradient is typically unique at a fixed point, the gradients of the distance function at a point in the support, say $(0, 0, x_3)$, can span an entire subspace complementary to the support. ■

Notice that the above (gradient)

$$-\mu(F(\boldsymbol{x}) - c)\frac{\partial F}{\partial \boldsymbol{x}}(\boldsymbol{x})$$

always points towards the low-dimensional support of the distribution. Hence the descent process can be viewed as a "denoising" process, studied in Chapter 3, that gradually enforces the data to be closer to the correct support. Notice that the above derivation suggests that the "step size" of the score $\nabla_{\boldsymbol{x}} \log p(\boldsymbol{x})$ of the denoising process is

$$\mu\big|F(\boldsymbol{x}) - \lambda/\mu\big|$$

which for any fixed $\mu$ is nearly proportional to the distance of the point $\boldsymbol{x}$ to the support.

Secondly, notice that solving the above program (7.1.13) with $\mu$ increasing is equivalent to:

$$\boldsymbol{x}_\star = \arg\min_{\boldsymbol{x}} \frac{1}{2}\|h(\boldsymbol{x}) - \boldsymbol{y}\|_2^2 + \frac{\mu}{2}\big(F(\boldsymbol{x}) - \lambda/\mu\big)^2 \quad \text{as} \quad \mu \uparrow \infty. \tag{7.1.18}$$

Due to the conspicuous quadratic form of the two terms in the above equation, they can also be interpreted as certain "energy" functions. Such a formulation is often referred to as "Energy Minimization" in the machine learning literature, advocated by people like Yann LeCun [LCH+06]. Here the energy of a point $\boldsymbol{x}$ is a quadratic function of its distance to the support of the desired distribution. Notice that in Chapter 3 and Chapter 4, we have argued that the notion of entropy is to measure the "volume" or uncertainty of the data distribution whereas here the energy depends on the "distance" of $\boldsymbol{x}$ to the support of the distribution. As we minimize the above energy functions, the entropy (or uncertainty) of the feasible solutions reduces until it reaches (a feasible solution of) the optimal MAP estimate, as illustrated in Figure 7.4.

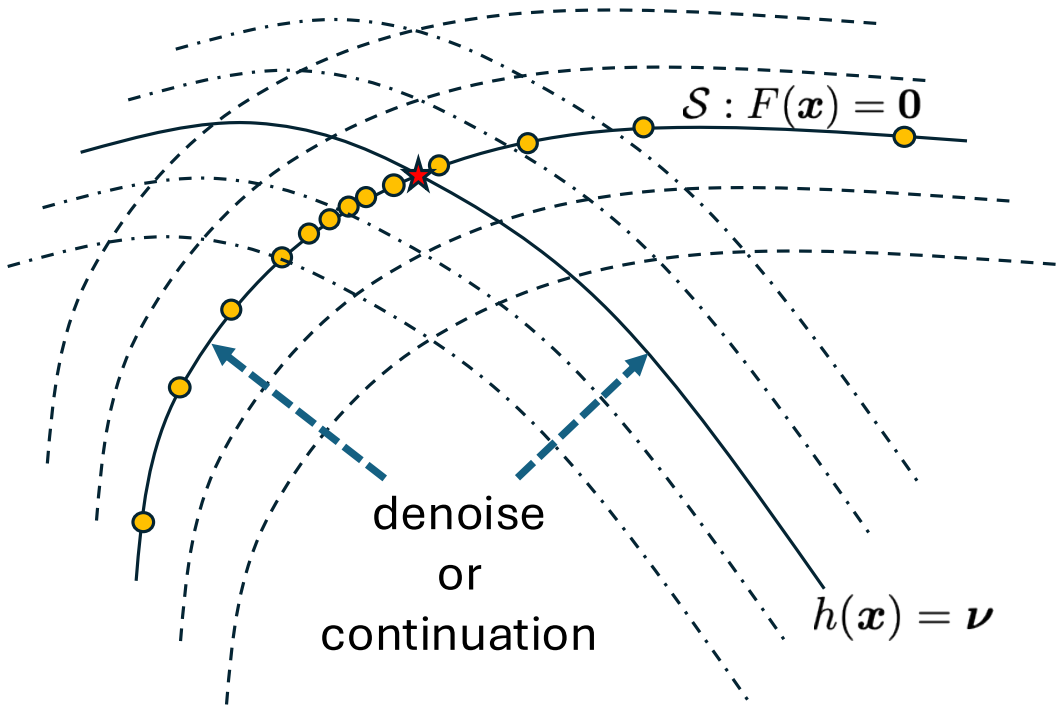

Figure 7.4: Illustration of the continuation process of enforcing the constraints $F(\boldsymbol{x}) = 0$ and $\boldsymbol{y} = h(\boldsymbol{x})$.

### 7.1.3 Representative Practical Settings for Inference

Notice that the above discussion and derivation are based on the assumption that we have perfect knowledge about the function $F(\boldsymbol{x})$ and the observation model $h(\boldsymbol{x})$. In practice, however, they may not be available at all and need to be "learned" from the data given. Hence to make the above conceptual solution truly computable, we need to deal with various situations in which information about the distributions of $\boldsymbol{x}$ and the relationship between $\boldsymbol{x}$ and $\boldsymbol{y}$ are given or accessible in different ways and forms.

In general, they can mostly be categorized into four cases, which are, conceptually, increasingly more challenging:

- *Case 1:* Both a model for the distribution of $\boldsymbol{x}$ and the observation model $\boldsymbol{y} = h(\boldsymbol{x})$ (possibly with added noise $\boldsymbol{w}$) are known, even with an analytical form. This is typically the case for many classic signal processing problems, such as signal denoising, the sparse vector recovery problem we saw in Chapter 2 and the low-rank matrix recovery problem to be introduced below.

- *Case 2:* We do not have a model for the distribution but only samples $\boldsymbol{X} = \{\boldsymbol{x}_1, \ldots, \boldsymbol{x}_N\}$ of $\boldsymbol{x}$, and the observation model $\boldsymbol{y} = h(\boldsymbol{x})$ (possibly with added noise $\boldsymbol{w}$) is known.[2] A model for the distribution $p(\boldsymbol{x})$ of $\boldsymbol{x}$ needs to be learned, and subsequently the conditional distribution $p(\boldsymbol{x} \mid \boldsymbol{y})$. Natural image completion or natural language completion (e.g., BERT and GPT) are typical examples of this class of problems.

- *Case 3:* We only have some paired samples: $(\boldsymbol{X}, \boldsymbol{Y}) = \{(\boldsymbol{x}_1, \boldsymbol{y}_1), \ldots, (\boldsymbol{x}_N, \boldsymbol{y}_N)\}$ of the two variables $(\boldsymbol{x}, \boldsymbol{y})$. The distributions of $\boldsymbol{x}$ and $\boldsymbol{y}$ and their relationship $h(\cdot)$ need to be learned from these paired sample data. For example, given many images and their captions, learning to conduct text-conditioned image generation is one such problem.
- *Case 4:* We only have the samples $\boldsymbol{Y} = \{\boldsymbol{y}_1, \ldots, \boldsymbol{y}_N\}$ of the observations $\boldsymbol{y}$, and the observation model $h(\cdot)$ needs to be known, at least in some parametric family $h(\cdot, \boldsymbol{\theta})$. The distribution $p(\boldsymbol{x})$ and $p(\boldsymbol{x} \mid \boldsymbol{y})$ need to be learned from $\hat{\boldsymbol{x}}$, estimated from $\boldsymbol{Y}$. For example, learning to render a new view from a sequence of calibrated or uncalibrated views is one such problem.

In this chapter, we will discuss general approaches to learn the desired distributions and solve the associated conditional estimation or generation for these cases, typically with a representative practical problem. Throughout the chapter, the reader should keep Figure 7.1 in mind.

## 7.2 Conditional Inference with a Known Data Distribution

Notice that in the setting we have discussed in previous Chapters, the autoencoding network is trained to reconstruct a set of samples of the random vector $\boldsymbol{x}$. This would allow us to regenerate samples from the learned (low-dimensional) distribution. In practice, the low-dimensionality of the distribution, once given or learned, can be exploited for stable and robust recovery, completion, or prediction tasks. That is, under rather mild conditions, one can recover $\boldsymbol{x}$ from highly compressive, partial, noisy or even corrupted measures of $\boldsymbol{x}$ of the kind:

$$\boldsymbol{y} = h(\boldsymbol{x}) + \boldsymbol{w}, \tag{7.2.1}$$

where $\boldsymbol{y}$ is typically an observation of $\boldsymbol{x}$ that is of much lower dimension than $\boldsymbol{x}$ and $\boldsymbol{w}$ can be random noise or even sparse gross corruptions. This is a class of problems that have been extensively studied in the classical signal processing literature, for low-dimensional structures such as sparse vectors, low-rank matrices, and beyond. Interested readers may see [WM22] for a complete exposition of this topic.

Here to put the classic work in a more general modern setting, we illustrate the basic idea and facts through the arguably simplest task of data (and particularly image) completion. That is, we consider the problem of recovering a sample $\boldsymbol{x}$ when parts of it are missing (or even corrupted). We want to recover or predict the rest of $\boldsymbol{x}$ from observing only a fraction of it:

$$f : \mathcal{P}_\Omega(\boldsymbol{x}) \mapsto \hat{\boldsymbol{x}}, \tag{7.2.2}$$

where $\mathcal{P}_\Omega(\,\cdot\,)$ represents a masking operation (see Figure 7.5 for an example).

[2] In the literature, this setting is sometimes referred to as the *empirical Bayesian inference*.

In this section and the next, we will study the completion task under two different scenarios: One is when the distribution of the data $\boldsymbol{x}$ of interest is already given *a priori*, even in a certain analytical form. This is the case that prevails in classic signal processing where the structures of the signals are assumed to be known, for example, band-limited, sparse or low-rank. The other is when only raw samples of $\boldsymbol{x}$ are available and we need to learn the low-dimensional distribution from the samples in order to solve the completion task well. This is the case for the tasks of natural image completion or video frame prediction. As a precursor to the rest of the chapter, we start with the simplest case of image completion: when the image to be completed can be well modeled as a low-rank matrix. We will move on to increasingly more general cases and more challenging settings later.

**Low-rank matrix completion.** The low-rank *matrix completion* problem is a classical problem for data completion when its distribution is low-dimensional and known. Consider a random sample of a matrix $\boldsymbol{X}_o = [\boldsymbol{x}_1, \ldots, \boldsymbol{x}_n] \in \mathbb{R}^{m\times n}$ from the space of all matrices of rank $r$. In general, we assume the rank of the matrix is

$$\operatorname{rank}(\boldsymbol{X}_o) = r < \min\{m, n\}. \tag{7.2.3}$$

So it is clear that locally the intrinsic dimension of the space of all matrices of rank $r$ is much lower than the ambient space $mn$.

Now, let $\Omega$ indicate a set of indices of observed entries of the matrix $\boldsymbol{X}_o$. Let the observed entries be:

$$\boldsymbol{Y} = \mathcal{P}_\Omega(\boldsymbol{X}_o). \tag{7.2.4}$$

The remaining entries supported on $\Omega^c$ are unobserved or missing. The problem is whether we can recover from $\boldsymbol{Y}$ the missing entries of $\boldsymbol{X}$ correctly and efficiently. Figure 7.5 shows one example of completing such a matrix.

Notice that the fundamental reason why such a matrix can be completed is that columns and rows of the matrix are highly correlated and they all lie on a low-dimensional subspace. For the example shown in Figure 7.5, the dimension or the rank of the matrix completed is only two. Hence the fundamental idea to recover such a matrix is to seek a matrix that has the lowest rank among all matrices that have entries agreeing with the observed ones:

$$\min_{\boldsymbol{X}} \operatorname{rank}(\boldsymbol{X}) \quad \text{subject to} \quad \boldsymbol{Y} = \mathcal{P}_\Omega(\boldsymbol{X}). \tag{7.2.5}$$

This is known as the *low-rank matrix completion* problem. See [WM22] for a full characterization of the space of all low-rank matrices. As the rank function is discontinuous and rank minimization is in general an NP-hard problem, we would like to relax it with something easier to optimize.

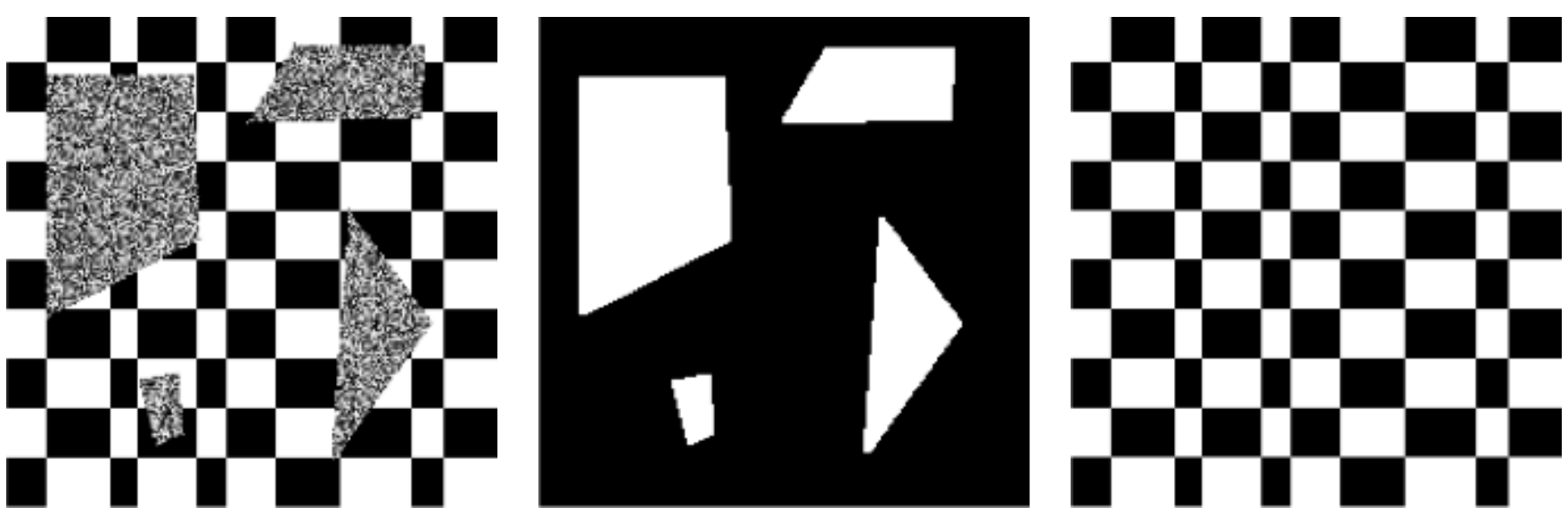

Figure 7.5: Illustration of completing an image as low-rank matrix with some entries masked or corrupted. Left: the masked/corrupted image $\boldsymbol{Y}$; middle: the mask $\Omega$; right: the completed image $\hat{\boldsymbol{X}}$.

Based on our knowledge about compression from Chapter 3, we could promote the low-rankness of the recovered matrix $\boldsymbol{X}$ by enforcing the lossy coding rate (or the volume spanned by $\boldsymbol{X}$) of the data in $\boldsymbol{X}$ to be small:

$$\min R_\epsilon(\boldsymbol{X}) = \frac{1}{2}\log\det\left(\boldsymbol{I} + \alpha \boldsymbol{X}\boldsymbol{X}^\top\right) \quad \text{subject to} \quad \boldsymbol{Y} = \mathcal{P}_\Omega(\boldsymbol{X}). \tag{7.2.6}$$

The problem can be viewed as a continuous relaxation of the above low-rank matrix completion problem (7.2.5) and it can be solved via gradient descent. One can show that the gradient descent operator for the $\log\det$ objective is precisely minimizing a close surrogate of the rank of the matrix $\boldsymbol{X}\boldsymbol{X}^\top$.

The rate distortion function is a nonconvex function, and its gradient descent does not always guarantee finding the globally optimal solution. Nevertheless, since the underlying structure sought for $\boldsymbol{X}$ is piecewise linear, the rank function admits a rather effective convex relaxation: the nuclear norm—the sum of all singular values of the matrix $\boldsymbol{X}$. As shown in the compressive sensing literature, under fairly broad conditions,[3] the matrix completion problem (7.2.5) can be effectively solved by the following convex program:

$$\min \|\boldsymbol{X}\|_* \quad \text{subject to} \quad \boldsymbol{Y} = \mathcal{P}_\Omega(\boldsymbol{X}), \tag{7.2.7}$$

where the nuclear norm $\|\boldsymbol{X}\|_*$ is the sum of singular values of $\boldsymbol{X}$. In practice, we often convert the above constrained convex optimization program to an unconstrained one:

$$\min \|\boldsymbol{X}\|_* + \lambda\|\boldsymbol{Y} - \mathcal{P}_\Omega(\boldsymbol{X})\|_F^2, \tag{7.2.8}$$

for some properly chosen $\lambda > 0$. Interested readers may refer to [WM22] for how to develop algorithms that can solve the above programs efficiently and effectively. Figure 7.5 shows a real example in which the matrix $\hat{\boldsymbol{X}}$ is actually recovered by solving the above program.

[3]Typically, such conditions specify the necessary and sufficient amount of entries needed for the completion to be computationally feasible. These conditions have been systematically characterized in [WM22].

**Further extensions.** It has been shown that images (or more accurately textures) and 3D scenes with low-rank structures can be very effectively completed via solving optimization programs of the above kind, even if there is additional corruption and distortion [LRZ+12; YZB+23; ZLG+10]:

$$\boldsymbol{Y} \circ \tau = \boldsymbol{X}_o + \boldsymbol{E}, \tag{7.2.9}$$

where $\tau$ is some unknown nonlinear distortion of the image and $\boldsymbol{E}$ is an unknown matrix that models some (sparse) occlusion and corruption. Again, interested readers may refer to [WM22] for a more detailed account.

# 7.3 Conditional Inference with a Learned Data Representation

In the previous subsection, the reason we can infer $\boldsymbol{x}$ from the partial observation $\boldsymbol{y}$ is because (the support of) the distribution of $\boldsymbol{X}$ is known or specified *a priori*, say as the set of all low-rank matrices. For many practical datasets, we do not have their distribution in an analytical form like the low-rank matrices, say the set of all natural images. Nevertheless, if we have sufficient samples of the data $\boldsymbol{x}$, we should be able to learn its low-dimensional distribution first and leverage it for future inference tasks based on an observation $\boldsymbol{y} = h(\boldsymbol{x}) + \boldsymbol{w}$. In this section, we assume the observation model $h(\cdot)$ is given and known. We will study the case when $h(\cdot)$ is not explicitly given in the next section.

## 7.3.1 Image Completion with Masked Auto-Encoding

For a general image $\boldsymbol{X}$ such as the one shown on the left of Figure 7.6, we can no longer view it as a low-rank matrix. However, humans still demonstrate remarkable ability to complete a scene and recognize familiar objects despite severe occlusion. This suggests that our brain has learned the low-dimensional distribution of natural images and can use it for completion, and hence recognition. However, the distribution of all natural images is not as simple as a low-dimensional linear subspace. Hence a natural question is whether we can learn the more sophisticated distribution of natural images and use it to perform image completion?

One empirical approach to the image completion task is to find an encoding and decoding scheme by solving the following *masked autoencoding* (MAE) program that minimizes the reconstruction loss:

$$\min_{f,g} L_{\mathrm{MAE}}(f,g) \doteq \mathbb{E}\big[\|(g \circ f)(\mathcal{P}_{\Omega}(\boldsymbol{X})) - \boldsymbol{X}\|_2^2\big]. \tag{7.3.1}$$

Unlike the matrix completion problem which has a simple underlying structure, we should no longer expect that the encoding and decoding mappings admit simple closed forms or the program can be solved by explicit algorithms.

For a general natural image, we can no longer assume that its columns or rows are sampled from a low-dimensional subspace or a low-rank Gaussian.

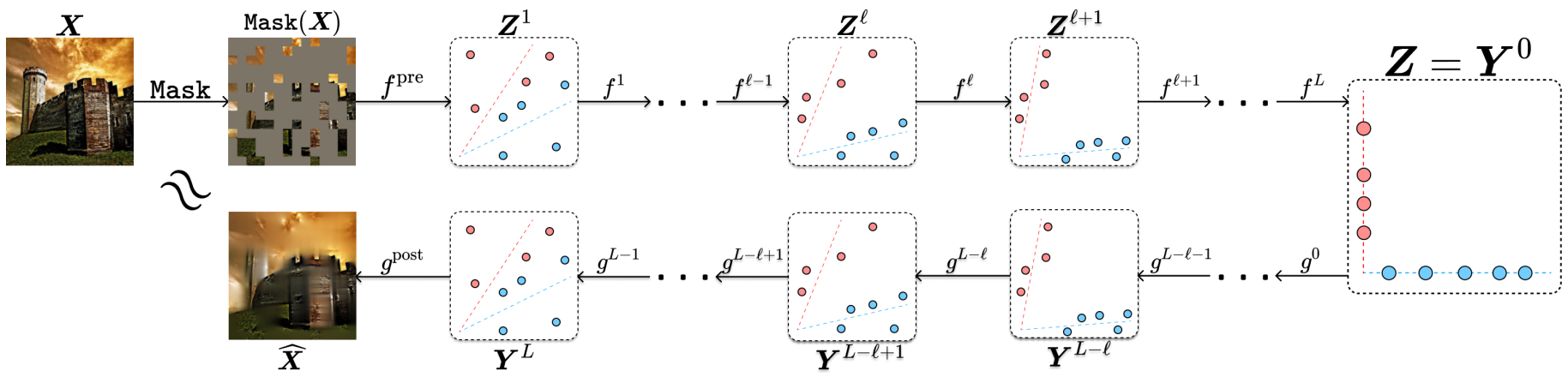

Figure 7.6: **Diagram of the overall (masked) autoencoding process.** The (image) token representations are transformed iteratively towards a parsimonious (e.g., compressed and sparse) representation by each encoder layer $f^\ell$. Furthermore, such representations are transformed back to the original image by the decoder layers $g^\ell$. Each encoder layer $f^\ell$ is meant to be (partially) inverted by a corresponding decoder layer $g^{L-\ell}$.

However, it is reasonable to assume that the image consists of multiple regions. Image patches in each region are similar and can be modeled as one (low-rank) Gaussian or subspace. Hence, to exploit the low-dimensionality of the distribution, the objective *of the encoder* $f$ is to transform $\boldsymbol{X}$ to a representation $\boldsymbol{Z}$:

$$f : \boldsymbol{X} \mapsto \boldsymbol{Z} \tag{7.3.2}$$

such that the distribution of $\boldsymbol{Z}$ can be well modeled as a mixture of subspaces, say $\{\boldsymbol{U}_{[K]}\}$, such that the rate reduction is maximized while the sparsity is minimized:

$$\mathbb{E}_{\boldsymbol{Z}=f(\boldsymbol{X})}[\Delta R_\epsilon(\boldsymbol{Z} \mid \boldsymbol{U}_{[K]}) - \lambda \|\boldsymbol{Z}\|_0] = \mathbb{E}_{\boldsymbol{Z}=f(\boldsymbol{X})}[R_\epsilon(\boldsymbol{Z}) - R_\epsilon^c(\boldsymbol{Z} \mid \boldsymbol{U}_{[K]}) - \lambda \|\boldsymbol{Z}\|_0], \tag{7.3.3}$$

where the functions $R_\epsilon(\cdot)$ and $R_\epsilon^c(\cdot)$ are defined in (5.2.2) and (5.2.3), respectively.

As we have shown in the previous Chapter 5, the encoder $f$ that minimizes the above objective can be constructed as a sequence of transformer-like operators. As shown in the work of [PBW+24], the decoder $g$ can be viewed and hence constructed explicitly as the inverse process of the encoder $f$. Figure 7.7 illustrates the overall architectures of both the encoder and the corresponding decoder at each layer. The parameters of the encoder $f$ and decoder $g$ can be learned by optimizing the reconstruction loss (7.3.1) via gradient descent.

Figure 7.8 shows some representative results of the thus-designed masked auto-encoder. More implementation details and results of the masked autoencoder for natural image completion can be found in Chapter 8 Section 8.5.

### 7.3.2 Conditional Sampling with Measurement Matching

The above (masked) autoencoding problem aims to generate a sample image that is consistent with certain observations or conditions. But let us examine the approach more closely: given the visual part of an image $\boldsymbol{X}_v = \mathcal{P}_\Omega(\boldsymbol{X})$, we try to estimate the masked part $\boldsymbol{X}_m = \mathcal{P}_{\Omega^c}(\boldsymbol{X})$. For realizations $(\boldsymbol{\Xi}_v, \boldsymbol{\Xi}_m)$ of the random variable $\boldsymbol{X} = (\boldsymbol{X}_v, \boldsymbol{X}_m)$, let

$$p_{\boldsymbol{X}_m \mid \boldsymbol{X}_v}(\boldsymbol{\Xi}_m \mid \boldsymbol{\Xi}_v)$$

be the conditional distribution of $\boldsymbol{X}_m$ given $\boldsymbol{X}_v$. It is easy to show that the optimal solution to the MAE formulation (7.3.1) is given by the conditional expectation:

$$\arg\min_{h = g \circ f} L_{\mathrm{MAE}}(h) = \boldsymbol{\Xi}_v \mapsto \boldsymbol{\Xi}_v + \mathbb{E}[\boldsymbol{X}_m \mid \boldsymbol{X}_v = \boldsymbol{\Xi}_v]. \tag{7.3.4}$$

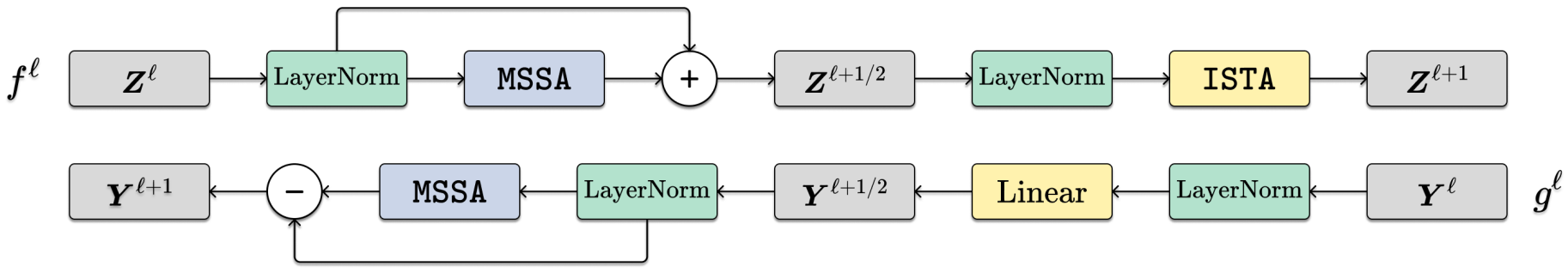

Figure 7.7: **Diagram of each encoder layer (*top*) and decoder layer (*bottom*).** Notice that the two layers are highly anti-parallel — each is constructed to do the operations of the other in reverse order. That is, in the decoder layer $g^\ell$, the ISTA block of $f^{L-\ell}$ is partially inverted first using a linear layer, then the MSSA block of $f^{L-\ell}$ is reversed; this order unravels the transformation done in $f^{L-\ell}$.

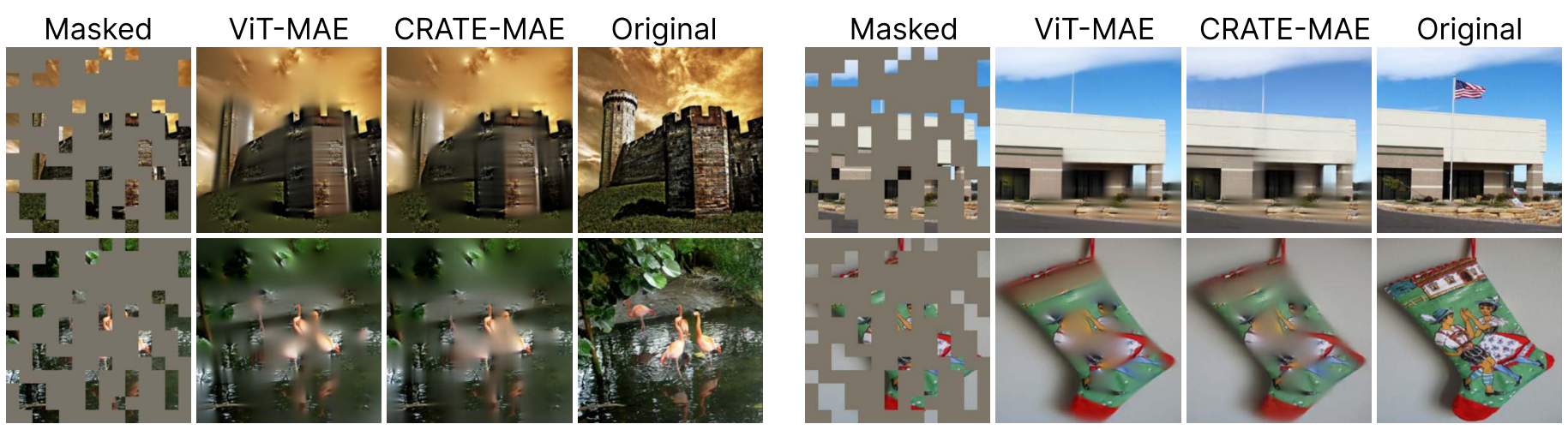

Figure 7.8: **Autoencoding visualizations of CRATE-Base and ViT-MAE-Base [HCX+22] with 75% patches masked.** We observe that the reconstructions from CRATE-Base are on par with the reconstructions from ViT-MAE-Base, despite using $< 1/3$ of the parameters.

In general, however, this expectation may not even lie on the low-dimensional distribution of natural images! This partially explains why some of the recovered patches in Figure 7.8 are a little blurry.

For many practical purposes, we would like to learn (a representation of) the conditional distribution $p_{\boldsymbol{X}_m \mid \boldsymbol{X}_v}$, or equivalently $p_{\boldsymbol{X} \mid \boldsymbol{X}_v}$, and then get a clear (most likely) sample from this distribution directly. Notice that, when the distribution of $\boldsymbol{X}$ is low-dimensional, it is possible that if a sufficient part of $\boldsymbol{X}$, $\boldsymbol{X}_v$, is observed, it fully determines $\boldsymbol{X}$ and hence the missing part $\boldsymbol{X}_m$. In other words, the distribution $p_{\boldsymbol{X} \mid \boldsymbol{X}_v}$ is a generalized function (analogous to a delta function).

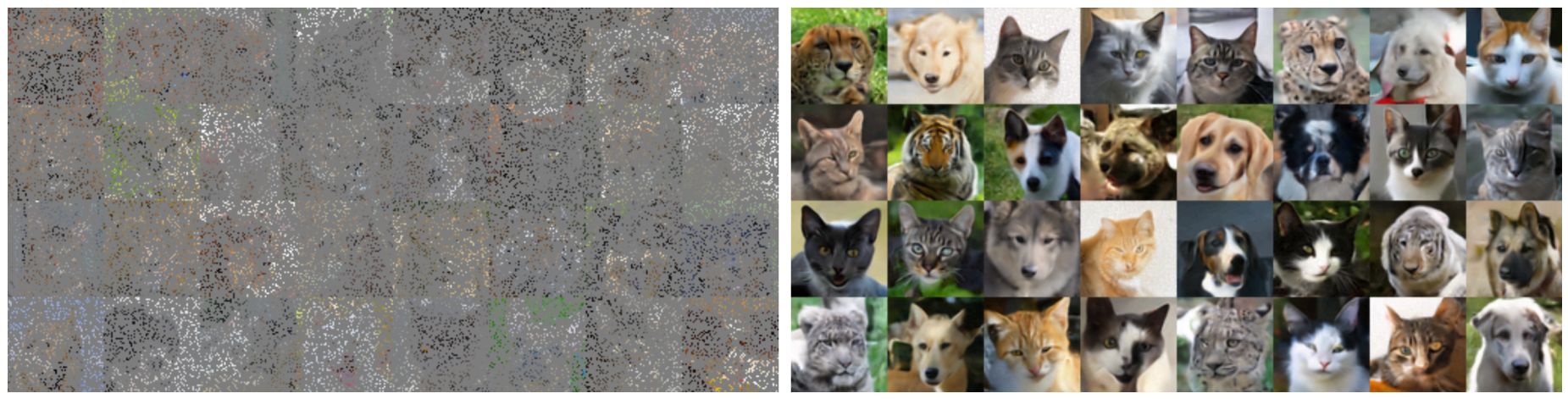

Figure 7.9: **Sampling visualizations from models trained via ambient diffusion [DSD+23b] with 80% of the pixels masked.** Using a similar ratio of masked pixels as in Figure 7.8, the ambient diffusion sampling algorithm recovers a much sharper image than the blurry image recovered by the MAE-based method. The former method samples from the distribution of natural images, while the latter approximates the conditional expectation (i.e., average) of this distribution given the observation; this averaging causes the blurriness.

Hence, instead of solving the completion task as a conditional estimation problem, we should address it as a conditional sampling problem. To that end, we should first learn the (low-dimensional) distribution of all natural images $\boldsymbol{X}$. If we have sufficient samples of natural images, we can learn the distribution via a denoising process $\boldsymbol{X}_t$ described in Chapter 3. Then the problem of recovering $\boldsymbol{X}$ from its partial observation $\boldsymbol{Y} = \mathcal{P}_\Omega(\boldsymbol{x}) + \boldsymbol{w}$ becomes a conditional generation problem—to sample the distribution conditioned on the observation.

**General linear measurements.** In fact, we may even consider recovering $\boldsymbol{X}$ from a more general linear observation model:

$$\boldsymbol{Y} = \boldsymbol{A}\boldsymbol{X}_0, \quad \boldsymbol{X}_t = \boldsymbol{X}_0 + \sigma_t \boldsymbol{G}, \tag{7.3.5}$$

where $\boldsymbol{A}$ is a linear operator on matrix space[4] and $\boldsymbol{G} \sim \mathcal{N}(\boldsymbol{0}, \boldsymbol{I})$. The masking operator $\mathcal{P}_\Omega(\cdot)$ in the image completion task is one example of such a linear model. Then it has been shown by [DSD+23a] that

$$\hat{\boldsymbol{X}}_* = \arg\min_{\hat{\boldsymbol{X}}} \mathbb{E}[\|\boldsymbol{A}(\hat{\boldsymbol{X}}(\boldsymbol{A}\boldsymbol{X}_t, \boldsymbol{A}) - \boldsymbol{X}_0)\|^2] \tag{7.3.6}$$

satisfies the condition that:

$$\boldsymbol{A}\hat{\boldsymbol{X}}_*(\boldsymbol{A}(\boldsymbol{X}_t), \boldsymbol{A}) = \boldsymbol{A}\mathbb{E}[\boldsymbol{X}_0 \mid \boldsymbol{A}\boldsymbol{X}_t, \boldsymbol{A}]. \tag{7.3.7}$$

Notice that in the special case when $\boldsymbol{A}$ is of full column rank, we have $\mathbb{E}[\boldsymbol{X}_0 \mid \boldsymbol{A}\boldsymbol{X}_t, \boldsymbol{A}] = \mathbb{E}[\boldsymbol{X}_0 \mid \boldsymbol{X}_t]$. Hence, in the more general case, it has been suggested by [DSD+23a] that one could still use the so obtained $\mathbb{E}[\boldsymbol{X}_0 \mid \boldsymbol{A}(\boldsymbol{X}_t), \boldsymbol{A}]$ to replace the $\mathbb{E}[\boldsymbol{X}_0 \mid \boldsymbol{X}_t]$ in the normal denoising process for $\boldsymbol{X}_t$:

$$\boldsymbol{X}_{t-s} = \gamma_t \boldsymbol{X}_t + (1 - \gamma_t)\mathbb{E}[\boldsymbol{X}_0 \mid \boldsymbol{A}\boldsymbol{X}_t, \boldsymbol{A}]. \tag{7.3.8}$$

[4] i.e., if we imagine unrolling $\boldsymbol{X}$ into a long vector then $\boldsymbol{A}$ takes the role of a matrix on $\boldsymbol{X}$-space

This usually works very well in practice, say for many image restoration tasks, as shown in [DSD+23a]. Compared to the blurry images recovered from MAE, the images recovered by the above method are much sharper as it leverages a learned distribution of natural images and samples a (sharp) image from the distribution that is consistent with the measurement, as shown in Figure 7.9 (cf. Figure 7.8).

**General nonlinear measurements.** To generalize the above (image) completion problems and make things more rigorous, we may consider that a random vector $\boldsymbol{x} \sim p$ is partially observed through a more general observation function:

$$\boldsymbol{y} = h(\boldsymbol{x}) + \boldsymbol{w}, \tag{7.3.9}$$

where $\boldsymbol{w}$ usually stands for some random measurement noise, say of a Gaussian distribution $\boldsymbol{w} \sim \mathcal{N}(\boldsymbol{0}, \sigma^2 \boldsymbol{I})$. It is easy to see that, for $\boldsymbol{x}$ and $\boldsymbol{y}$ so related, their joint distribution $p(\boldsymbol{x}, \boldsymbol{y})$ is naturally nearly degenerate if the noise $\boldsymbol{w}$ is small. To a large extent, we may view $p(\boldsymbol{x}, \boldsymbol{y})$ as a noisy version of a hypersurface defined by the function $\boldsymbol{y} = h(\boldsymbol{x})$ in the joint space $(\boldsymbol{x}, \boldsymbol{y})$. Practically speaking, we will consider a setting more akin to masked autoencoding than to pure matrix completion, where we always have access to a corresponding clean sample $\boldsymbol{x}$ for every observation $\boldsymbol{y}$ we receive.[5]

Like image/matrix completion, we are often faced with a setting where $\boldsymbol{y}$ denotes a degraded or otherwise "lossy" observation of the input $\boldsymbol{x}$. This can manifest in quite different forms. For example, in various scientific or medical imaging problems, the measured data $\boldsymbol{y}$ may be a compressed and corrupted observation of the underlying data $\boldsymbol{x}$; whereas in 3D vision tasks, $\boldsymbol{y}$ may represent an image captured by a camera of a physical object with an unknown (low-dimensional) pose $\boldsymbol{x}$. Generally, by virtue of mathematical modeling (and, in some cases, co-design of the measurement system), we know $h$ and can evaluate it on any input, and we can exploit this knowledge to help reconstruct and sample $\boldsymbol{x}$.

At a technical level, we want the learned representation of the data to facilitate us to sample the conditional distribution $p_{\boldsymbol{x}|\boldsymbol{y}}$, also known as the posterior, effectively and efficiently. More precisely, write $\boldsymbol{\nu}$ to denote a realization of the random variable $\boldsymbol{y}$. We want to generate samples $\hat{\boldsymbol{x}}$ such that:

$$\hat{\boldsymbol{x}} \sim p_{\boldsymbol{x}|\boldsymbol{y}}(\,\cdot \mid \boldsymbol{y} = \boldsymbol{\nu}). \tag{7.3.10}$$

Recall that in Section 3.2, we have developed a natural and effective way to produce *unconditional* samples of the data distribution $p$. The ingredients are the denoisers $\bar{\boldsymbol{x}}^*(t, \boldsymbol{\xi}) = \mathbb{E}[\boldsymbol{x} \mid \boldsymbol{x}_t = \boldsymbol{\xi}]$, or their learned approximations $\bar{\boldsymbol{x}}_\theta(t, \boldsymbol{\xi})$, for different levels of noisy observations $\boldsymbol{x}_t = \boldsymbol{x} + t\boldsymbol{g}$ (and $\boldsymbol{\xi}$ for their realizations) under Gaussian noise $\boldsymbol{g} \sim \mathcal{N}(\boldsymbol{0}, \boldsymbol{I})$, and $t \in [0, T]$ with a choice of times $0 = t_1 <$ $\dots < t_L = T$ at which to perform the iterative denoising, starting from $\hat{\boldsymbol{x}}_{t_L} \sim \mathcal{N}(\mathbf{0}, T^2\boldsymbol{I})$ (recall Equation (3.2.82)).[6] We could directly apply this scheme to generate samples from the posterior $p_{\boldsymbol{x}|\boldsymbol{y}}$ *if* we had access to a dataset of samples $\boldsymbol{x}^{\mathrm{c}} \sim p_{\boldsymbol{x}|\boldsymbol{y}}(\,\cdot \mid \boldsymbol{\nu})$ for each realization $\boldsymbol{\nu}$ of $\boldsymbol{y}$, by generating noisy observations $\boldsymbol{x}_t^{\mathrm{c}}$ and training denoisers to approximate $\mathbb{E}[\boldsymbol{x}^{\mathrm{c}} \mid \boldsymbol{x}_t^{\mathrm{c}} = \,\cdot\,, \boldsymbol{y} = \boldsymbol{\nu}]$, the mean of the posterior under the noisy observation (see Figure 7.10(a)). However, performing this resampling given only paired samples $(\boldsymbol{x}, \boldsymbol{y})$ from the joint distribution (say by binning the samples over values of $\boldsymbol{y}$) requires prohibitively many samples for high-dimensional data, and alternate approaches explicitly or implicitly rely on density estimation, which similarly suffers from the curse of dimensionality.

[5] In some more specialized applications, in particular in scientific imaging, it is of interest to be able to learn to generate samples from the posterior $p_{\boldsymbol{x}|\boldsymbol{y}}$ without access to any clean/ground-truth samples of $\boldsymbol{x}$. We give a brief overview of methods for this setting in the end-of-chapter notes.

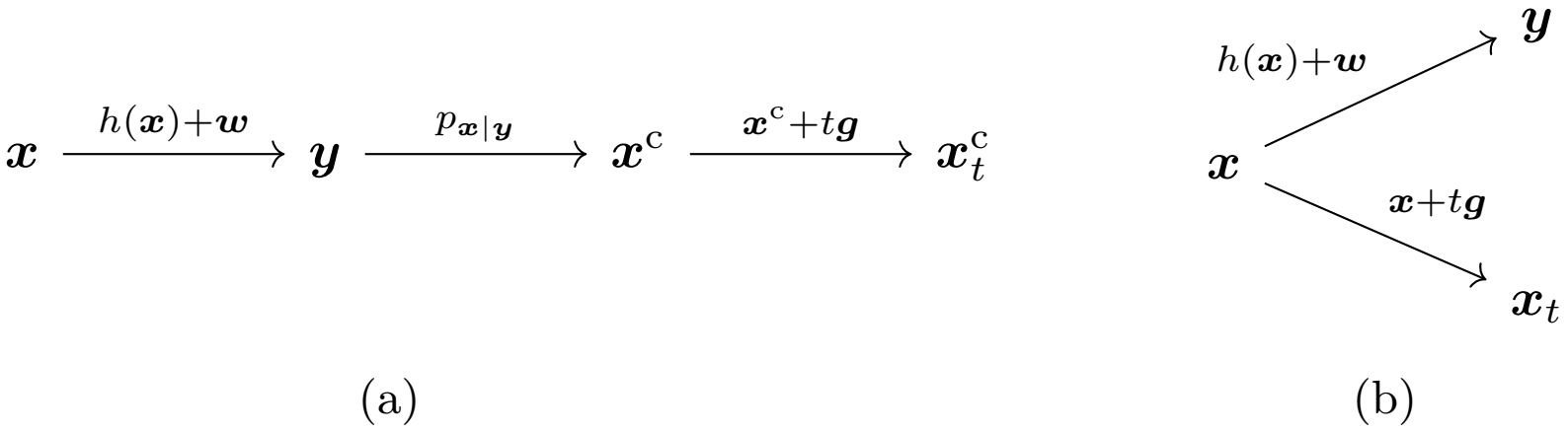

Figure 7.10: Statistical dependency diagrams for the conditional sampling process. **Left:** In a direct (conceptual) application of the diffusion-denoising scheme we have developed in Chapter 3 to conditional sampling, we use samples from the posterior $p_{\boldsymbol{x}|\boldsymbol{y}}$ to train denoisers directly on the posterior at different noise levels, then use them to generate new samples. In practice, however, we do not normally have direct samples from the posterior, but rather paired samples $(\boldsymbol{x}, \boldsymbol{y})$ from the joint. **Right:** It turns out that it suffices to have only noisy observations of $\boldsymbol{x}$ to realize the denoisers corresponding to $p_{\boldsymbol{x}_t^{\mathrm{c}}|\boldsymbol{x}^{\mathrm{c}}}$: this follows from conditional independence of $\boldsymbol{x}_t$ and $\boldsymbol{y}$ given $\boldsymbol{x}$. It implies that $p_{\boldsymbol{x}_t^{\mathrm{c}}|\boldsymbol{y}} = p_{\boldsymbol{x}_t|\boldsymbol{y}}$, which gives a score function for denoising that consists of the unconditional score function, plus a correction term that enforces measurement consistency.

Fortunately, it turns out that this is not necessary. Consider the alternate statistical dependency diagram in Figure 7.10(b), which corresponds to the random variables in the usual denoising-diffusion process, together with the measurement $\boldsymbol{y}$. Because our assumed observation model (7.3.9) implies that

[6] Recall from our discussion in Section 3.2.2 that a few small improvements to this basic iterative denoising scheme are sufficient to bring competitive practical performance. For clarity as we develop conditional sampling, we will focus here on the simplest instantiation.

$\boldsymbol{x}_t$ and $\boldsymbol{y}$ are independent conditioned on $\boldsymbol{x}$, we have for any realization $\boldsymbol{\nu}$ of $\boldsymbol{y}$

$$
\begin{aligned}
p_{\boldsymbol{x}_t^{\mathrm{c}}|\boldsymbol{y}}(\,\cdot \mid \boldsymbol{\nu}) &= \int \underbrace{p_{\boldsymbol{x}_t^{\mathrm{c}}|\boldsymbol{x}^{\mathrm{c}}}(\,\cdot \mid \boldsymbol{\xi})}_{=\mathcal{N}(\boldsymbol{\xi}, t^2\boldsymbol{I})} \cdot \underbrace{p_{\boldsymbol{x}^{\mathrm{c}}|\boldsymbol{y}}}_{=p_{\boldsymbol{x}|\boldsymbol{y}}}(\boldsymbol{\xi} \mid \boldsymbol{\nu})\mathrm{d}\boldsymbol{\xi} \\
&= \int p_{\boldsymbol{x}_t|\boldsymbol{x},\boldsymbol{y}}(\,\cdot \mid \boldsymbol{\xi}, \boldsymbol{\nu}) \cdot p_{\boldsymbol{x}|\boldsymbol{y}}(\boldsymbol{\xi} \mid \boldsymbol{\nu})\mathrm{d}\boldsymbol{\xi} \qquad (7.3.11) \\
&= \int p_{\boldsymbol{x}_t,\boldsymbol{x}|\boldsymbol{y}}(\,\cdot\,, \boldsymbol{\xi} \mid \boldsymbol{\nu})\mathrm{d}\boldsymbol{\xi} \\
&= p_{\boldsymbol{x}_t|\boldsymbol{y}}(\,\cdot \mid \boldsymbol{\nu}).
\end{aligned}
$$

Above, the first line recognizes an equivalence between the distributions arising in Figure 7.10 (a,b); the second line applies this together with conditional independence of $\boldsymbol{x}_t$ and $\boldsymbol{y}$ given $\boldsymbol{x}$; the third line uses the definition of conditional probability; and the final line marginalizes over $\boldsymbol{x}$. Thus, the denoisers from the conceptual posterior sampling process are equal to $\mathbb{E}[\boldsymbol{x} \mid \boldsymbol{x}_t = \cdot, \boldsymbol{y} = \boldsymbol{\nu}]$, which we can learn solely from paired samples $(\boldsymbol{x}, \boldsymbol{y})$, and by Tweedie's formula (Theorem 3.2), we can express these denoisers in terms of the score function of $p_{\boldsymbol{x}_t|\boldsymbol{y}}$, which, by Bayes' rule, satisfies

$$
p_{\boldsymbol{x}_t|\boldsymbol{y}}(\boldsymbol{\xi} \mid \boldsymbol{\nu}) = \frac{p_{\boldsymbol{y}|\boldsymbol{x}_t}(\boldsymbol{\nu} \mid \boldsymbol{\xi}) p_{\boldsymbol{x}_t}(\boldsymbol{\xi})}{p_{\boldsymbol{y}}(\boldsymbol{\nu})}. \qquad (7.3.12)
$$

Recall that the density of $\boldsymbol{x}_t$ is given by $p_t = \varphi_t * p$, where $\varphi_t$ denotes the standard Gaussian density with zero mean and covariance $t^2\boldsymbol{I}$ and $*$ denotes convolution. This is nothing but the *unconditional score function* obtained from the standard diffusion training that we developed in Section 3.2! The conditional score function then satisfies, for any realization $(\boldsymbol{\xi}, \boldsymbol{\nu})$ of $(\boldsymbol{x}_t, \boldsymbol{y})$,

$$
\nabla_{\boldsymbol{\xi}} \log p_{\boldsymbol{x}_t|\boldsymbol{y}}(\boldsymbol{\xi} \mid \boldsymbol{\nu}) = \underbrace{\nabla_{\boldsymbol{\xi}} \log p_t(\boldsymbol{\xi})}_{\text{score matching}} + \underbrace{\nabla_{\boldsymbol{\xi}} \log p_{\boldsymbol{y}|\boldsymbol{x}_t}(\boldsymbol{\nu} \mid \boldsymbol{\xi})}_{\text{measurement matching}}, \qquad (7.3.13)
$$

giving (by Tweedie's formula) our proposed denoisers as

$$
\begin{aligned}
\mathbb{E}[\boldsymbol{x} \mid \boldsymbol{x}_t = \boldsymbol{\xi}, \boldsymbol{y} = \boldsymbol{\nu}] &= \boldsymbol{\xi} + t^2\nabla_{\boldsymbol{\xi}} \log p_t(\boldsymbol{\xi}) + t^2\nabla_{\boldsymbol{\xi}} \log p_{\boldsymbol{y}|\boldsymbol{x}_t}(\boldsymbol{\nu} \mid \boldsymbol{\xi}) \\
&= \mathbb{E}[\boldsymbol{x} \mid \boldsymbol{x}_t = \boldsymbol{\xi}] + t^2\nabla_{\boldsymbol{\xi}} \log p_{\boldsymbol{y}|\boldsymbol{x}_t}(\boldsymbol{\nu} \mid \boldsymbol{\xi}). \qquad (7.3.14)
\end{aligned}
$$

The resulting operators are interpretable as a *corrected* version of the unconditional denoiser for the noisy observation, where the correction term (the so-called "measurement matching" term) enforces consistency with the observations $\boldsymbol{y}$. This decomposition is the basis of *classifier guidance*, which we will develop in detail in Section 7.4.1. The reader should take care to note to which argument the gradient operators are applying in the above score functions in order to fully grasp the meaning of this operator.

The key remaining issue in making this procedure computational is to prescribe how to compute the measurement matching correction, since in general we do not have a closed-form expression for the likelihood $p_{\boldsymbol{y}|\boldsymbol{x}_t}$ except for when $t = 0$. Before taking up this problem, we discuss an illustrative concrete example of the entire process, continuing from those we have developed in Section 3.2.

*Example* 7.4. Consider the case where the data distribution is Gaussian with mean $\boldsymbol{\mu} \in \mathbb{R}^D$ and covariance $\boldsymbol{\Sigma} \in \mathbb{R}^{D \times D}$, i.e., $\boldsymbol{x} \sim \mathcal{N}(\boldsymbol{\mu}, \boldsymbol{\Sigma})$. Assume that $\boldsymbol{\Sigma} \succeq \mathbf{0}$ is nonzero. Moreover, in the measurement model (7.3.9), suppose we obtain linear measurements of $\boldsymbol{x}$ with independent Gaussian noise, where $\boldsymbol{A} \in \mathbb{R}^{d \times D}$ and $\boldsymbol{y} = \boldsymbol{A}\boldsymbol{x} + \sigma \boldsymbol{w}$ with $\boldsymbol{w} \sim \mathcal{N}(\mathbf{0}, \boldsymbol{I})$ independent of $\boldsymbol{x}$. Below, we are going to work out the following consequences of this model for the general framework we have derived above. Specifically:

1. We will work out the precise forms of the posterior distribution $p_{\boldsymbol{x}|\boldsymbol{y}}$ that we are aiming to sample from.

2. We will work out the form of the measurement matching term from (7.3.13), which gives us the additive correction to the unconditional denoiser for $p_{\boldsymbol{x}|\boldsymbol{x}_t}$ that yields the denoiser for $p_{\boldsymbol{x}|\boldsymbol{x}_t,\boldsymbol{y}}$ (recall (7.3.14)).

3. We will assess the specific form of the measurement matching term we derived in the previous step, and posit a natural approximation that makes it far less sample and compute intensive to learn the measurement matching term from data. This approximation will be applied in greater generality, after the conclusion of this example.

First, we note that $\boldsymbol{x} \overset{\mathrm{d}}{=} \boldsymbol{\Sigma}^{1/2}\boldsymbol{g} + \boldsymbol{\mu}$, where $\boldsymbol{g} \sim \mathcal{N}(\mathbf{0}, \boldsymbol{I})$ is independent of $\boldsymbol{w}$ and $\boldsymbol{\Sigma}^{1/2}$ is the unique positive square root of the covariance matrix $\boldsymbol{\Sigma}$, and after some algebra, we can then write

$$\begin{bmatrix} \boldsymbol{x} \\ \boldsymbol{y} \end{bmatrix} \overset{\mathrm{d}}{=} \begin{bmatrix} \boldsymbol{\Sigma}^{1/2} & \mathbf{0} \\ \boldsymbol{A}\boldsymbol{\Sigma}^{1/2} & \sigma \boldsymbol{I} \end{bmatrix} \begin{bmatrix} \boldsymbol{g} \\ \boldsymbol{w} \end{bmatrix} + \begin{bmatrix} \boldsymbol{\mu} \\ \boldsymbol{A}\boldsymbol{\mu} \end{bmatrix}.$$

By independence, we have that $(\boldsymbol{g}, \boldsymbol{w})$ is jointly Gaussian, which means that $(\boldsymbol{x}, \boldsymbol{y})$ is also jointly Gaussian, as the affine image of a jointly Gaussian vector. Its covariance matrix is given by

$$\begin{bmatrix} \boldsymbol{\Sigma}^{1/2} & \mathbf{0} \\ \boldsymbol{A}\boldsymbol{\Sigma}^{1/2} & \sigma \boldsymbol{I} \end{bmatrix} \begin{bmatrix} \boldsymbol{\Sigma}^{1/2} & \mathbf{0} \\ \boldsymbol{A}\boldsymbol{\Sigma}^{1/2} & \sigma \boldsymbol{I} \end{bmatrix}^\top = \begin{bmatrix} \boldsymbol{\Sigma} & \boldsymbol{\Sigma}\boldsymbol{A}^\top \\ \boldsymbol{A}\boldsymbol{\Sigma} & \boldsymbol{A}\boldsymbol{\Sigma}\boldsymbol{A}^\top + \sigma^2 \boldsymbol{I} \end{bmatrix}.$$

Now, we apply the fact that conditioning a random vector with joint Gaussian distribution on a subset of coordinates is again a Gaussian distribution (Exercise 3.2). By this, we obtain that

$$\begin{aligned} p_{\boldsymbol{x}|\boldsymbol{y}}(\cdot \mid \boldsymbol{\nu}) = \mathcal{N}\Bigg( & \underbrace{\boldsymbol{\mu} + \boldsymbol{\Sigma}\boldsymbol{A}^\top \left(\boldsymbol{A}\boldsymbol{\Sigma}\boldsymbol{A}^\top + \sigma^2 \boldsymbol{I}\right)^{-1} (\boldsymbol{\nu} - \boldsymbol{A}\boldsymbol{\mu})}_{\boldsymbol{\mu}_{\boldsymbol{x}|\boldsymbol{y}}(\boldsymbol{\nu})}, \\ & \underbrace{\boldsymbol{\Sigma} - \boldsymbol{\Sigma}\boldsymbol{A}^\top \left(\boldsymbol{A}\boldsymbol{\Sigma}\boldsymbol{A}^\top + \sigma^2 \boldsymbol{I}\right)^{-1} \boldsymbol{A}\boldsymbol{\Sigma}}_{\boldsymbol{\Sigma}_{\boldsymbol{x}|\boldsymbol{y}}} \Bigg). \end{aligned} \tag{7.3.15}$$

Next, since $\boldsymbol{x}_t = \boldsymbol{x} + t\boldsymbol{g}'$, where $\boldsymbol{g}' \sim \mathcal{N}(\mathbf{0}, \boldsymbol{I})$ is independent of all other random vectors, and since $\boldsymbol{x} \mid \boldsymbol{y}$ is Gaussian, by the calculation directly above, it follows by another application of Exercise 3.2 that $\boldsymbol{x} \mid \boldsymbol{x}_t, \boldsymbol{y}$ is Gaussian. Its conditional expectation function is given by reading off the mean of the Gaussian distribution, which in this case is

$$\mathbb{E}[\boldsymbol{x} \mid \boldsymbol{x}_t = \boldsymbol{\xi}, \boldsymbol{y} = \boldsymbol{\nu}] = \boldsymbol{\mu}_{\boldsymbol{x}|\boldsymbol{y}}(\boldsymbol{\nu}) + \boldsymbol{\Sigma}_{\boldsymbol{x}|\boldsymbol{y}} \left(\boldsymbol{\Sigma}_{\boldsymbol{x}|\boldsymbol{y}} + t^2\boldsymbol{I}\right)^{-1} \left(\boldsymbol{\xi} - \boldsymbol{\mu}_{\boldsymbol{x}|\boldsymbol{y}}(\boldsymbol{\nu})\right). \tag{7.3.16}$$

The functional form of this denoiser is quite simple, but it carries an unwieldy dependence on the problem data $\boldsymbol{\mu}$, $\boldsymbol{\Sigma}$, $\boldsymbol{A}$, and $\sigma^2$. We can gain further insight into its behavior by comparing it with Equation (7.3.14). We have as usual

$$\mathbb{E}[\boldsymbol{x} \mid \boldsymbol{x}_t = \boldsymbol{\xi}] = \boldsymbol{\mu} + \boldsymbol{\Sigma}\left(\boldsymbol{\Sigma} + t^2\boldsymbol{I}\right)^{-1}(\boldsymbol{\xi} - \boldsymbol{\mu}), \tag{7.3.17}$$

which is rather simple—suggesting that the measurement matching term is rather complicated. To confirm this, we can calculate the likelihood $p_{\boldsymbol{y}|\boldsymbol{x}_t}$ directly using the following expression for the joint distribution of $(\boldsymbol{x}, \boldsymbol{x}_t, \boldsymbol{y})$:

$$\begin{bmatrix} \boldsymbol{x} \\ \boldsymbol{x}_t \\ \boldsymbol{y} \end{bmatrix} \overset{\mathrm{d}}{=} \begin{bmatrix} \boldsymbol{\Sigma}^{1/2} & \mathbf{0} & \mathbf{0} \\ \boldsymbol{\Sigma}^{1/2} & t\boldsymbol{I} & \mathbf{0} \\ \boldsymbol{A}\boldsymbol{\Sigma}^{1/2} & \mathbf{0} & \sigma\boldsymbol{I} \end{bmatrix} \begin{bmatrix} \boldsymbol{g} \\ \boldsymbol{g}' \\ \boldsymbol{w} \end{bmatrix} + \begin{bmatrix} \boldsymbol{\mu} \\ \boldsymbol{\mu} \\ \boldsymbol{A}\boldsymbol{\mu} \end{bmatrix}, \tag{7.3.18}$$

where $\boldsymbol{g}' \sim \mathcal{N}(\mathbf{0}, \boldsymbol{I})$ independent of the other Gaussians. This is again a jointly Gaussian distribution; restricting to only the final two rows, we have the covariance

$$\begin{bmatrix} \boldsymbol{\Sigma}^{1/2} & t\boldsymbol{I} & \mathbf{0} \\ \boldsymbol{A}\boldsymbol{\Sigma}^{1/2} & \mathbf{0} & \sigma\boldsymbol{I} \end{bmatrix} \begin{bmatrix} \boldsymbol{\Sigma}^{1/2} & t\boldsymbol{I} & \mathbf{0} \\ \boldsymbol{A}\boldsymbol{\Sigma}^{1/2} & \mathbf{0} & \sigma\boldsymbol{I} \end{bmatrix}^\top = \begin{bmatrix} \boldsymbol{\Sigma} + t^2\boldsymbol{I} & \boldsymbol{\Sigma}\boldsymbol{A}^\top \\ \boldsymbol{A}\boldsymbol{\Sigma} & \boldsymbol{A}\boldsymbol{\Sigma}\boldsymbol{A}^\top + \sigma^2\boldsymbol{I} \end{bmatrix}.$$

Another application of Exercise 3.2 then gives us

$$\begin{aligned} p_{\boldsymbol{y}|\boldsymbol{x}_t}(\,\cdot \mid \boldsymbol{\xi}) = \mathcal{N}\Bigg( & \underbrace{\boldsymbol{A}\boldsymbol{\mu} + \boldsymbol{A}\boldsymbol{\Sigma}\left(\boldsymbol{\Sigma} + t^2\boldsymbol{I}\right)^{-1}(\boldsymbol{\xi} - \boldsymbol{\mu})}_{\boldsymbol{\mu}_{\boldsymbol{y}|\boldsymbol{x}_t}(\boldsymbol{\xi})}, \\ & \underbrace{\boldsymbol{A}\boldsymbol{\Sigma}\boldsymbol{A}^\top + \sigma^2\boldsymbol{I} - \boldsymbol{A}\boldsymbol{\Sigma}\left(\boldsymbol{\Sigma} + t^2\boldsymbol{I}\right)^{-1}\boldsymbol{\Sigma}\boldsymbol{A}^\top}_{\boldsymbol{\Sigma}_{\boldsymbol{y}|\boldsymbol{x}_t}}\Bigg). \end{aligned} \tag{7.3.19}$$

Now notice that $\boldsymbol{\mu}_{\boldsymbol{y}|\boldsymbol{x}_t}(\boldsymbol{\xi}) = \boldsymbol{A}\mathbb{E}[\boldsymbol{x} \mid \boldsymbol{x}_t = \boldsymbol{\xi}]$. So, by the chain rule,

$$\begin{aligned} & t^2\nabla_{\boldsymbol{\xi}} \log p_{\boldsymbol{y}|\boldsymbol{x}_t}(\boldsymbol{\nu} \mid \boldsymbol{\xi}) \\ = \; & t^2\nabla_{\boldsymbol{\xi}}\left[-\frac{1}{2}(\boldsymbol{\nu} - \boldsymbol{A}\mathbb{E}[\boldsymbol{x} \mid \boldsymbol{x}_t = \boldsymbol{\xi}])^\top \boldsymbol{\Sigma}_{\boldsymbol{y}|\boldsymbol{x}_t}^{-1}(\boldsymbol{\nu} - \boldsymbol{A}\mathbb{E}[\boldsymbol{x} \mid \boldsymbol{x}_t = \boldsymbol{\xi}])\right] \\ = \; & t^2(\boldsymbol{\Sigma} + t^2\boldsymbol{I})^{-1}\boldsymbol{\Sigma}\boldsymbol{A}^\top\boldsymbol{\Sigma}_{\boldsymbol{y}|\boldsymbol{x}_t}^{-1}\left(\boldsymbol{\nu} - \boldsymbol{A}\mathbb{E}[\boldsymbol{x} \mid \boldsymbol{x}_t = \boldsymbol{\xi}]\right). \end{aligned} \tag{7.3.20}$$

This gives us a more interpretable decomposition of the conditional posterior denoiser (7.3.16): following Equation (7.3.14), it is the sum of the unconditional posterior denoiser (7.3.17) and the measurement matching term (7.3.20).

We can further analyze the measurement matching term to understand cases under which it can be approximated. Notice that

$$\boldsymbol{\Sigma}_{\boldsymbol{y}|\boldsymbol{x}_t} = \sigma^2 \boldsymbol{I} + \boldsymbol{A}\boldsymbol{\Sigma}^{1/2} \left( \boldsymbol{I} - \boldsymbol{\Sigma}^{1/2} \left( \boldsymbol{\Sigma} + t^2 \boldsymbol{I} \right)^{-1} \boldsymbol{\Sigma}^{1/2} \right) \boldsymbol{\Sigma}^{1/2} \boldsymbol{A}^\top. \tag{7.3.21}$$

If we let $\boldsymbol{\Sigma} = \boldsymbol{V}\boldsymbol{\Lambda}\boldsymbol{V}^\top$ denote an eigenvalue decomposition of $\boldsymbol{\Sigma}$, where $(\boldsymbol{v}_i)$ are the columns of $\boldsymbol{V}$, we can further write

$$\boldsymbol{\Sigma}^{1/2} \left( \boldsymbol{I} - \boldsymbol{\Sigma}^{1/2} \left( \boldsymbol{\Sigma} + t^2 \boldsymbol{I} \right)^{-1} \boldsymbol{\Sigma}^{1/2} \right) \boldsymbol{\Sigma}^{1/2} = t^2 \boldsymbol{V} \boldsymbol{\Lambda}^{1/2} \left( \boldsymbol{\Lambda} + t^2 \boldsymbol{I} \right)^{-1} \boldsymbol{\Lambda}^{1/2} \boldsymbol{V}^\top \tag{7.3.22}$$

$$= t^2 \sum_{i=1}^{D} \frac{\lambda_i}{\lambda_i + t^2} \boldsymbol{v}_i \boldsymbol{v}_i^*. \tag{7.3.23}$$

Then for any eigenvalue of $\boldsymbol{\Sigma}$ equal to zero, the corresponding summand is zero; and writing $\lambda_{\min}(\boldsymbol{\Sigma})$ for the smallest positive eigenvalue of $\boldsymbol{\Sigma}$, we have that whenever $t \ll \sqrt{\lambda_{\min}(\boldsymbol{\Sigma})}$, it holds

$$\frac{\lambda_i t^2}{\lambda_i + t^2} \approx 0. \tag{7.3.24}$$

So, when $t \ll \sqrt{\lambda_{\min}(\boldsymbol{\Sigma})}$, we have the approximation

$$\boldsymbol{\Sigma}_{\boldsymbol{y}|\boldsymbol{x}_t} \approx \sigma^2 \boldsymbol{I}. \tag{7.3.25}$$

The right-hand side of this approximation is equal to $\boldsymbol{\Sigma}_{\boldsymbol{y}|\boldsymbol{x}}$. So we have in turn

$$\nabla_{\boldsymbol{\xi}} \log p_{\boldsymbol{y}|\boldsymbol{x}_t}(\boldsymbol{\nu} \mid \boldsymbol{\xi}) \approx \nabla_{\boldsymbol{\xi}} \log p_{\boldsymbol{y}|\boldsymbol{x}}(\boldsymbol{\nu} \mid \mathbb{E}[\boldsymbol{x} \mid \boldsymbol{x}_t = \boldsymbol{\xi}]). \tag{7.3.26}$$

■

Let us take some time to interpret the results of Example 7.4 before we generalize it. The approximation (7.3.26) is, of course, a direct consequence of the specific modeling assumptions we have made in Example 7.4. However, notice that if we directly interpret this approximation, it is *ab initio* tractable: the likelihood $p_{\boldsymbol{y}|\boldsymbol{x}} = \mathcal{N}(\boldsymbol{A}\boldsymbol{x}, \sigma^2\boldsymbol{I})$ is a simple Gaussian distribution centered at the observation, and the approximation to the measurement matching term that we arrive at can be interpreted as simply evaluating the log-likelihood at the conditional expectation $\mathbb{E}[\boldsymbol{x} \mid \boldsymbol{x}_t = \boldsymbol{\xi}]$, then taking gradients with respect to $\boldsymbol{\xi}$ (which involves backpropagating through the conditional expectation).

To gain insight into the effect of the convenient approximation (7.3.26), we implement and simulate a simple numerical experiment in the Gaussian setting in Figure 7.11. The sampler we implement is a direct implementation of the simple scheme (3.2.82) we have developed in Chapter 3 and recalled above, using the true conditional posterior denoiser, i.e. Equation (7.3.16) (samples are marked with blue circles), and the convenient approximation to this denoiser made through the measurement matching approximation (7.3.26) (samples are marked with red triangles). The top row of Figure 7.11 shows a setting with large measurement noise $\sigma^2$, and the bottom with small measurement noise. The measurement matching approximation works very well in the large-noise setting, with the caveat that in the small-noise setting, it suffers from rapid collapse of the variance of the sampling distribution along directions that are parallel to the rows of the linear measurement operator $\boldsymbol{A}$, which cannot be corrected by later iterations of sampling. Our analysis in Example 7.4 precisely characterizes the level at which the noise becomes too "small" in terms of the data covariance matrix $\boldsymbol{\Sigma}$.

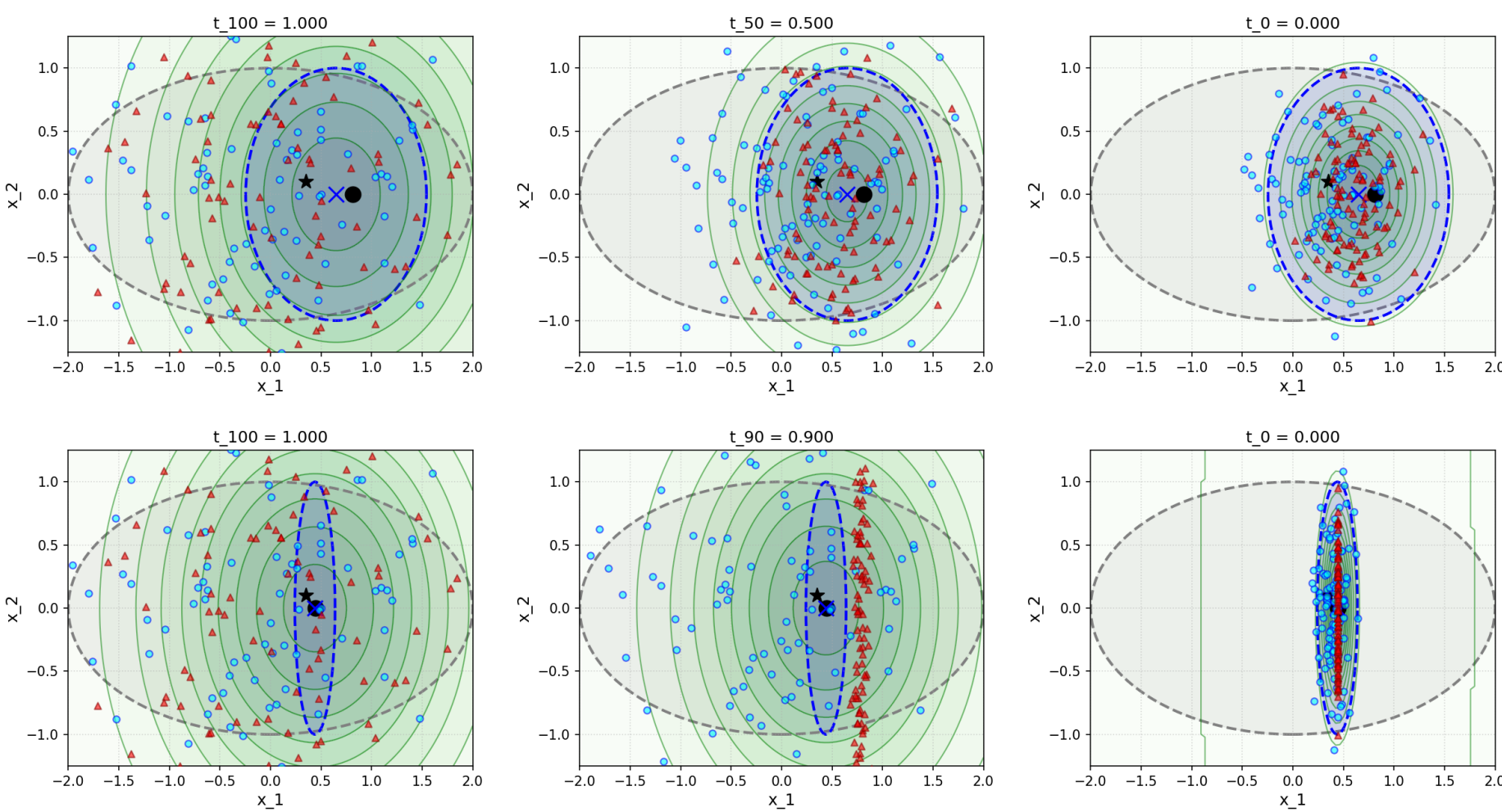

Figure 7.11: Numerical simulation of the conditional sampling setup (7.3.9), with Gaussian data, linear measurements, and Gaussian noise. We simulate $D = 2$ and $d = 1$, with $\boldsymbol{\Sigma} = \boldsymbol{e}_1\boldsymbol{e}_1^\top + \frac{1}{4}\boldsymbol{e}_2\boldsymbol{e}_2^\top$, $\boldsymbol{\mu} = \mathbf{0}$, and $\boldsymbol{A} = \boldsymbol{e}_1^\top$. The underlying signal $\boldsymbol{x}$ is marked with a black star, and the measurement $\boldsymbol{y}$ is marked with a black circle. Each individual plot corresponds to a different value of sampler time $t_\ell$, with different rows corresponding to different observation noise levels $\sigma^2$. In each plot, the covariance matrix of $\boldsymbol{x}$ is plotted in gray, the posterior covariance matrix and posterior mean of $p_{\boldsymbol{x}|\boldsymbol{y}}$ are plotted in blue (with the posterior mean marked by a blue "x"), and contours for $p_{\boldsymbol{x}_{t_\ell}|\boldsymbol{y}}$ are drawn in green. The sampler hyperparameters are $T = 1$, $L = 100$, and we draw 100 independent samples to initialize the samplers. Samplers are implemented with the closed-form denoisers derived in Example 7.4, with those using the approximation (7.3.26) marked with red triangles, and those using the exact conditional posterior denoiser marked with blue circles. **Top:** For large observation noise $\sigma = 0.5$, both the exact conditional posterior denoiser and the approximate one do a good job of converging to the posterior $p_{\boldsymbol{x}|\boldsymbol{y}}$. Sampling time (corresponding to time in the "forward process", so larger times mean larger noise) decreases from left to right. The convergence dynamics for the exact and approximate measurement matching term are similar. **Bottom:** For smaller observation noise $\sigma = 0.1$, the approximate measurement matching term leads to extreme bias in the sampler (red triangles): samples rapidly converge to an affine subspace of points that are consistent, modulo some shrinkage from the posterior mean denoiser, with the measured ground truth, and later sampling iterations are unable to recover the lost posterior variance along this dimension. Note that different times $t_\ell$ are plotted in the bottom row, compared to the top row, to show the rapid collapse of the approximation to the posterior along the measurement dimension.

**Generalizing the approximation: Diffusion Posterior Sampling.** Example 7.4 suggests a convenient approximation for the measurement matching term, i.e. (7.3.26), which can be made beyond the Gaussian setting of the example. To motivate this approximation in greater generality, notice that by conditional independence of $\boldsymbol{y}$ and $\boldsymbol{x}_t$ given $\boldsymbol{x}$, we can write

$$p_{\boldsymbol{y}|\boldsymbol{x}_t}(\boldsymbol{\nu} \mid \boldsymbol{\xi}) = \int p_{\boldsymbol{y}|\boldsymbol{x}}(\boldsymbol{\nu} \mid \boldsymbol{\xi}')p_{\boldsymbol{x}|\boldsymbol{x}_t}(\boldsymbol{\xi}' \mid \boldsymbol{\xi})\mathrm{d}\boldsymbol{\xi}'. \tag{7.3.27}$$

Formally, when the posterior $p_{\boldsymbol{x}|\boldsymbol{x}_t}$ is a delta function centered at its mean $\mathbb{E}[\boldsymbol{x} \mid \boldsymbol{x}_t = \boldsymbol{\xi}]$, the approximation (7.3.26) is exact. More generally, when the posterior $p_{\boldsymbol{x}|\boldsymbol{x}_t}$ is highly concentrated around its mean, the approximation (7.3.26) is accurate. This holds, for example, for sufficiently small $t$, which we saw explicitly in the Gaussian setting of Example 7.4. Although the numerical simulation in Figure 7.11 suggests that this approximation is not without its caveats in certain regimes, it has proved to be a reliable baseline in practice, after being proposed by Chung et al. as "Diffusion Posterior Sampling" (DPS) [CKM+23]. In addition, there are even principled and generalizable approaches to improve it by incorporating better estimates of the posterior variance (which turn out to be exact in the Gaussian setting of Example 7.4), which we discuss further in the end-of-chapter summary.

Thus, with the DPS approximation, we arrive at the following approximation for the conditional posterior denoisers $\mathbb{E}[\boldsymbol{x} \mid \boldsymbol{y}, \boldsymbol{x}_t]$, via Equation (7.3.14):

$$\mathbb{E}[\boldsymbol{x} \mid \boldsymbol{x}_t = \boldsymbol{\xi}, \boldsymbol{y} = \boldsymbol{\nu}] \approx \mathbb{E}[\boldsymbol{x} \mid \boldsymbol{x}_t = \boldsymbol{\xi}] + t^2 \nabla_{\boldsymbol{\xi}} \log p_{\boldsymbol{y}|\boldsymbol{x}}(\boldsymbol{\nu} \mid \mathbb{E}[\boldsymbol{x} \mid \boldsymbol{x}_t = \boldsymbol{\xi}]). \tag{7.3.28}$$

And, for a neural network or other model $\bar{\boldsymbol{x}}_\theta(t, \boldsymbol{\xi})$ trained as in Section 3.2 to approximate the denoisers $\mathbb{E}[\boldsymbol{x} \mid \boldsymbol{x}_t = \boldsymbol{\xi}]$ for each $t \in [0, T]$, we arrive at the learned conditional posterior denoisers

$$\bar{\boldsymbol{x}}_\theta(t, \boldsymbol{\xi}, \boldsymbol{\nu}) = \bar{\boldsymbol{x}}_\theta(t, \boldsymbol{\xi}) + t^2 \nabla_{\boldsymbol{\xi}} \log p_{\boldsymbol{y}|\boldsymbol{x}}(\boldsymbol{\nu} \mid \bar{\boldsymbol{x}}_\theta(t, \boldsymbol{\xi})). \tag{7.3.29}$$

Note that the approximation (7.3.28) is valid for arbitrary forward models $h$ in the observation model (7.3.9), including nonlinear $h$, and even to arbitrary noise models for which a clean expression for the likelihood $p_{\boldsymbol{y}|\boldsymbol{x}}$ is known. Indeed, in the case of Gaussian noise, we have

$$p_{\boldsymbol{y}|\boldsymbol{x}}(\boldsymbol{\nu} \mid \boldsymbol{\xi}) \propto \exp\left(-\frac{1}{2\sigma^2} \|h(\boldsymbol{\xi}) - \boldsymbol{\nu}\|_2^2\right). \tag{7.3.30}$$

Hence, evaluating the right-hand side of (7.3.29) requires only

1. A pretrained denoiser $\bar{\boldsymbol{x}}_\theta(t, \boldsymbol{\xi})$ for the data distribution $p$ (of $\boldsymbol{x}$), learned as in Section 3.2 via Algorithm 3.2;

2. Forward and backward pass access to the forward model $h$ for the measurements (7.3.9);

3. A forward and backward pass through $\bar{\boldsymbol{x}}_\theta(t, \boldsymbol{\xi})$, which can be evaluated efficiently using (say) backpropagation.

**Algorithm 7.1** Conditional sampling under measurements (7.3.9), with an unconditional denoiser and DPS.

**Input:** An ordered list of timesteps $0 \le t_0 < \cdots < t_L \le T$ to use for sampling.
**Input:** An unconditional denoiser $\bar{\boldsymbol{x}}_\theta \colon \{t_\ell\}_{\ell=1}^L \times \mathbb{R}^D \to \mathbb{R}^D$ for $p_{\boldsymbol{x}}$.
**Input:** Measurement realization $\boldsymbol{\nu}$ of $\boldsymbol{y}$ (Equation (7.3.9)) to condition on.
**Input:** Forward model $h : \mathbb{R}^D \to \mathbb{R}^d$ and measurement noise variance $\sigma^2 > 0$.
**Input:** Scale and noise level functions $\alpha, \sigma \colon \{t_\ell\}_{\ell=0}^L \to \mathbb{R}_{\ge 0}$.
**Output:** A sample $\hat{\boldsymbol{x}}$, approximately from $p_{\boldsymbol{x}|\boldsymbol{y}}$.

1: **function** DDIMSamplerConditionalDPS$(\bar{\boldsymbol{x}}_\theta, \boldsymbol{\nu}, h, \sigma^2, (t_\ell)_{\ell=0}^L)$
2: Initialize $\hat{\boldsymbol{x}}_{t_L} \sim$ approximate distribution of $\boldsymbol{x}_{t_L}$ (VP $\implies \mathcal{N}(\mathbf{0}, \boldsymbol{I})$, VE $\implies \mathcal{N}(\mathbf{0}, t_L^2 \boldsymbol{I})$.)
3: **for** $\ell = L, L-1, \dots, 1$ **do**
4: Compute

$$w_{t_{\ell-1}} \doteq \alpha_{t_{\ell-1}} - \frac{\sigma_{t_{\ell-1}}}{\sigma_{t_\ell}} \alpha_{t_\ell};$$

$$\hat{\boldsymbol{x}}_{t_{\ell-1}} \doteq \frac{\sigma_{t_{\ell-1}}}{\sigma_{t_\ell}} \hat{\boldsymbol{x}}_{t_\ell} + \left( \bar{\boldsymbol{x}}_\theta(t_\ell, \hat{\boldsymbol{x}}_{t_\ell}) - \frac{\sigma_{t_\ell}^2}{2\alpha_{t_\ell}\sigma^2} \nabla_{\boldsymbol{\xi}} \left[ \| h(\bar{\boldsymbol{x}}_\theta(t_\ell, \boldsymbol{\xi})) - \boldsymbol{\nu} \|_2^2 \right] \Big|_{\boldsymbol{\xi} = \hat{\boldsymbol{x}}_{t_\ell}} \right)$$

5: **end for**
6: **return** $\hat{\boldsymbol{x}}_{t_0}$
7: **end function**

Combining this scheme with the basic implementation of unconditional sampling we developed in Section 3.2, we obtain a practical algorithm for conditional sampling of the posterior $p_{\boldsymbol{x}|\boldsymbol{y}}$ given measurements following (7.3.9). Algorithm 7.1 records this scheme for the case of Gaussian observation noise with known standard deviation $\sigma$, with minor modifications to extend to a general noising process, as in Equation (3.2.85) and the surrounding discussion in Chapter 3 (our discussion above made the simplifying choices $\alpha_t = 1$, $\sigma_t = t$, and $t_\ell = T\ell/L$, as for Equation (3.2.82) in Section 3.2).

**Application to medical image reconstruction.** The framework that we have developed above is already powerful enough to be applied to solve various *scientific inverse problems*, where the measurements $\boldsymbol{y} = h(\boldsymbol{x}_o) + \boldsymbol{w}$ are the outputs of a known measurement process $h$ on a specific signal of interest $\boldsymbol{x}_o$, and where there are infinitely many possible signals $\boldsymbol{x}$ that satisfy $\boldsymbol{y} = h(\boldsymbol{x}) + \boldsymbol{w}$. This means that to find the *true* $\boldsymbol{x}_o$ that generated $\boldsymbol{y}$, we need to exploit prior information about the structure of $\boldsymbol{x}_o$. We will consider the probabilistic setting where $\boldsymbol{x}_o \sim p_{\boldsymbol{x}}$, and where we have access to a prior on $p_{\boldsymbol{x}}$, say in the form of a diffusion model.

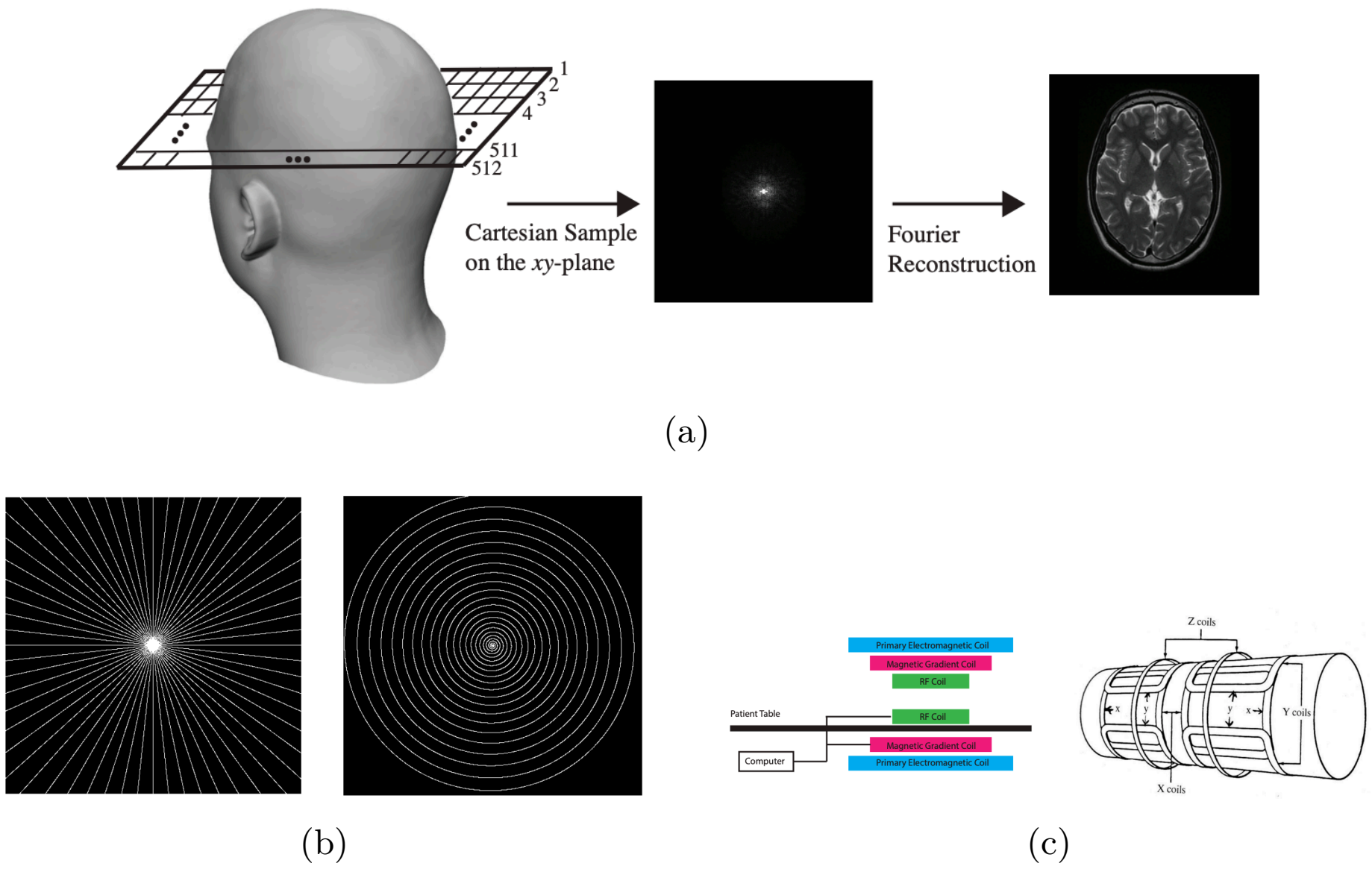

Figure 7.12: Simplified overview of magnetic resonance imaging (MRI) reconstruction. **(a)** An overview of the imaging process. A slice of the object being imaged (say, a human brain) is measured in the Fourier domain via the MRI machine. The spatial image can be reconstructed via direct inversion of the Fourier domain measurements. **(b)** In practice, the MRI machine *subsamples* the Fourier domain via a structured sampling pattern for the sake of efficiency. This creates a linear inverse problem, where priors about the structure of the underlying image must be exploited to enable unique reconstruction. **(c)** The actual measurement apparatus uses a controlled magnetic field to obtain the Fourier domain measurements. Images reproduced with permission from [WM22].

An instructive example of a scientific inverse problem is magnetic resonance imaging (MRI) (Figure 7.12), wherein the data distribution $p_{\boldsymbol{x}}$ is over images corresponding to an underlying object of interest (such as a 2D slice of a human subject's brain). In MRI applications, the measurement operator $h$ corresponds to the machine generating and modulating the magnetic field and recording the patient's response to it (Figure 7.12(c)). It can be modeled as implementing the *Fourier transform* of the underlying spatial signal (Figure 7.12(a))—see [WM22, Chapter 10] for a detailed mathematical derivation. In addition, for efficiency's sake, the MRI machine typically only measures a subset of the measured frequencies, in a structured measurement pattern that can be modeled as either a radial or a spiral pattern (Figure 7.12(b)). Writing $\mathcal{F}$ to denote the (discrete) Fourier transform operator and $\mathcal{S}$ to denote the machine's (known) subsampling pattern, the measured signal is then

$$\boldsymbol{y} = \mathcal{S} \circ \mathcal{F}(\boldsymbol{x}_o) + \boldsymbol{w}. \tag{7.3.31}$$

The measurement operator $\mathcal{S} \circ \mathcal{F}$ is *linear*, implying that it can be represented by a $m \times n$ matrix after appropriate reshaping, but it is also *compressive*, which means that $m \ll n$ and that we need to exploit prior knowledge about $p_{\boldsymbol{x}}$ to reconstruct $\boldsymbol{x}_o$ (or more generally to sample from the posterior $p_{\boldsymbol{x}|\boldsymbol{y}}$).

The distribution $p_{\boldsymbol{x}}$ of MRI images is simpler than some image distributions (e.g., natural images), but it is still complex enough to benefit from using a learned model for $p_{\boldsymbol{x}}$ rather than an analytical one (cf. [WM22]). One direct approach is to apply the diffusion posterior sampling approximation we developed in the previous section, which led to Algorithm 7.1, with the specific MRI measurement operator $\boldsymbol{A} = \mathcal{S} \circ \mathcal{F}$ and a diffusion model pretrained on MRI data. This precise combination of approaches has not appeared in the literature. Instead, we highlight results from a different but related approach, which enforces measurement consistency on each diffusion step. Following the pioneering work of Song et al. [SSX+22], numerous such measurement consistency algorithms for medical image reconstruction have been developed, including the improved approaches of Song et al. [SKZ+24] and Rout et al. [RRD+23] (which utilize latent diffusion models, as we have developed for representation autoencoders in Chapter 6). Figure 7.13 shows visual comparisons of reconstructed MRI images across three different samples, comparing FISTA-TV (a classical, accelerated version of the ISTA algorithm we studied in Chapter 2), the measurement consistency diffusion modeling approach of Song et al. [SSX+22], and the ground truth. It can be seen that superior reconstruction quality is achieved by the diffusion-based method, as well as superior faithfulness to fine detail in the ground truth image. Table 7.1 shows MRI reconstruction accuracy results (measured in terms of PSNR) for the approach developed by Song et al. [SSX+22]. It compares favorably to supervised baselines (trained on many independent paired samples $(\boldsymbol{y}, \boldsymbol{x}_o)$, which is an additional requirement) and unsupervised baselines (which correspond to measurement matching approximations that came before diffusion posterior sampling).

## 7.4 Conditional Inference with Paired Data and Measurements

In many practical applications, we do not know either the distribution of the data $\boldsymbol{x}$ of interest or the explicit relationship between the data and certain observed attributes $\boldsymbol{y}$ of the data. We only have a (large) set of paired samples $(\boldsymbol{X}, \boldsymbol{Y}) = \{(\boldsymbol{x}_1, \boldsymbol{y}_1), \ldots, (\boldsymbol{x}_N, \boldsymbol{y}_N)\}$ from which we need to infer the data

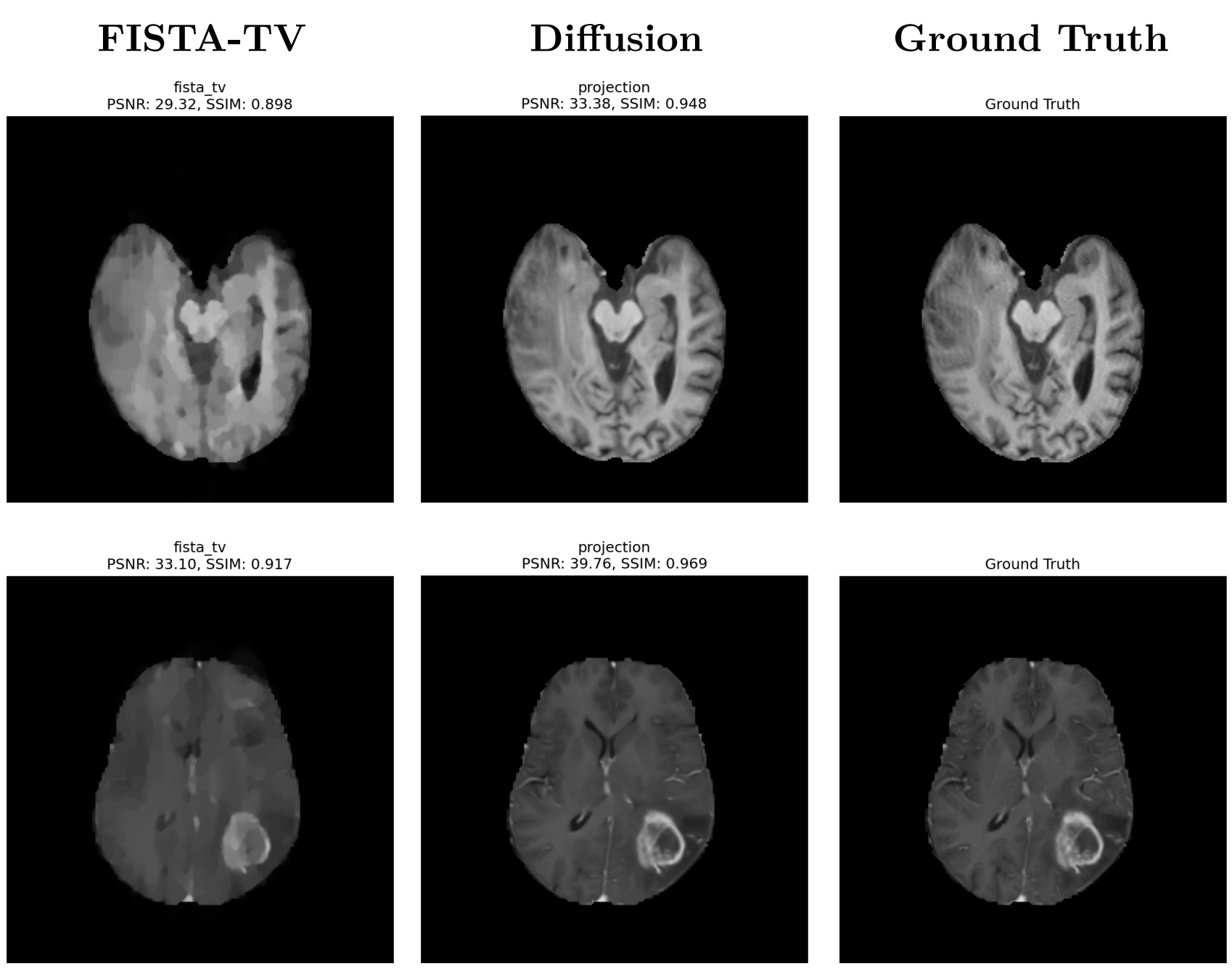

Figure 7.13: **Visual comparison of MRI reconstruction methods.** Three example brain MRI reconstructions showing FISTA-TV, the measurement consistency diffusion approach of Song et al. [SSX+22], and ground truth images on the BraTS MRI reconstruction dataset at 8x undersampling. The diffusion-based approach produces reconstructions that more closely match the ground truth structure compared to FISTA-TV, which can be seen to over-smooth the fine details present in the ground truth image. At the same time, the degree of faithfulness to the ground truth image in the diffusion-based reconstruction (arising from the proper enforcement of measurement consistency) is remarkable.

distribution and a mapping that models their relationship:

$$h : \boldsymbol{x} \mapsto \boldsymbol{y}. \tag{7.4.1}$$

The problem of image classification can be viewed as one such example. In a sense, the classification problem is to learn an (extremely lossy) compressive encoder for natural images. Say, given a random sample of an image $\boldsymbol{x}$, we would like to predict its class label $\boldsymbol{y}$ that best correlates the content in $\boldsymbol{x}$. We know the distribution of natural images of objects is low-dimensional compared to the dimension of the pixel space. From the previous chapters, we have learned that given sufficient samples, in principle, we can learn a structured low-dimensional representation $\boldsymbol{z}$ for $\boldsymbol{x}$ through a learned compressive encoding:

$$f : \boldsymbol{x} \mapsto \boldsymbol{z}. \tag{7.4.2}$$

| Method | 24× Accel. | 8× Accel. | 4× Accel. |
|---|---|---|---|
| Cascade DenseNet | $23.39_{\pm 2.17}$ | $28.35_{\pm 2.30}$ | $30.97_{\pm 2.33}$ |
| DuDoRNet | $18.46_{\pm 3.05}$ | $\mathbf{37.88_{\pm 3.03}}$ | $30.53_{\pm 4.13}$ |
| Song et al. [SSK+21] | $27.83_{\pm 2.73}$ | $35.04_{\pm 2.11}$ | $37.55_{\pm 2.08}$ |
| Jalal et al. [JAD+21] | $28.80_{\pm 3.21}$ | $36.44_{\pm 2.28}$ | $38.76_{\pm 2.32}$ |
| Song et al. [SSX+22] | $\mathbf{29.42_{\pm 3.03}}$ | $37.63_{\pm 2.70}$ | $\mathbf{39.91_{\pm 2.67}}$ |

Table 7.1: **MRI reconstruction results** comparing measurement matching approaches across different acceleration factors (the degree of Fourier space subsampling in $\mathcal{S}$) on the BraTS MRI reconstruction dataset. PSNR metrics (higher is better) with standard deviations are reported. Cascade DenseNet [ZFZ19] and DuDoRNet [ZZ20] are supervised baselines trained on paired samples at an 8x acceleration level. The last three methods are diffusion-based approaches using measurement consistency. Reproduced with permission from Song et al. [SSX+22].

The representation $\boldsymbol{z}$ can also be viewed as a learned (lossy but structured) code for $\boldsymbol{x}$. It is rather reasonable to assume that if the class assignment $\boldsymbol{y}$ truly depends on the low-dimensional structures of $\boldsymbol{x}$ and the learned code $\boldsymbol{z}$ truly reflects such structures, $\boldsymbol{y}$ and $\boldsymbol{z}$ can be made highly correlated and hence their joint distribution $p(\boldsymbol{z}, \boldsymbol{y})$ should be extremely low-dimensional. Therefore, we may combine the two desired codes $\boldsymbol{y}$ and $\boldsymbol{z}$ together and try to learn a combined encoder:

$$f : \boldsymbol{x} \mapsto (\boldsymbol{z}, \boldsymbol{y}) \tag{7.4.3}$$

where the joint distribution of $(\boldsymbol{z}, \boldsymbol{y})$ is highly low-dimensional.

From our study in previous chapters, the mapping $f$ is usually learned as a sequence of compression or denoising operators in the same space. Hence to leverage such a family of operations, we may introduce an auxiliary vector $\boldsymbol{w}$ that can be viewed as an initial random guess of the class label $\boldsymbol{y}$. In this way, we can learn a compression or denoising mapping:

$$f : (\boldsymbol{x}, \boldsymbol{w}) \mapsto (\boldsymbol{z}, \boldsymbol{y}) \tag{7.4.4}$$

within a common space. In fact, the common practice of introducing an auxiliary "class token" in the training of a transformer for classification tasks, such as in ViT, can be viewed as learning such a representation by compressing (the coding rate of) given (noisy) samples of $(\boldsymbol{x}, \boldsymbol{w})$. If the distribution of the data $\boldsymbol{x}$ is already a mixture of (low-dimensional) Gaussians, the work [WTL+08] has shown that classification can be done effectively by directly minimizing *the (lossy) coding length* associated with the given samples.

### 7.4.1 Class-Conditioned Image Generation

While a learned classifier allows us to classify a given image $\boldsymbol{x}$ to its corresponding class, we often would like to generate an image of a given class, by sampling the learned distribution of natural images. To some extent, this can be viewed as the "inverse" problem to image classification. Let $p_{\boldsymbol{x}}$ denote the distribution of natural images, say modeled by a diffusion-denoising process. Given a class label random variable $y \in [K]$ with realization $\nu$, say an "Apple", we would like to sample the conditional distribution $p_{\boldsymbol{x}|y}(\cdot \mid \nu)$ to generate an image of an apple:

$$\hat{\boldsymbol{x}} \sim p_{\boldsymbol{x}|\boldsymbol{y}}(\cdot \mid \boldsymbol{\nu}). \tag{7.4.5}$$

We call this *class-conditioned image generation*.

In Section 7.3.2, we have seen how to use the denoising-diffusion paradigm for conditional sampling from the posterior $p_{\boldsymbol{x}|\boldsymbol{y}}$ given *model-based* measurements $\boldsymbol{y} = h(\boldsymbol{x}) + \boldsymbol{w}$ (Equation (7.3.9)), culminating in the DPS algorithm (Algorithm 7.1). This is a powerful framework, but it does not apply to the class (or text) conditioned image generation problem here, where an explicit generative model $h$ for the observations/attributes $y$ is not available due to the intractability of analytical modeling. In this section, we will present techniques for extending conditional sampling to this setting.

Thus, we now assume only that we have access to samples from the joint distribution of $(\boldsymbol{x}, y)$:

$$(\boldsymbol{x}, y) \sim p_{\boldsymbol{x},y}. \tag{7.4.6}$$

As in the previous section, we define $\boldsymbol{x}_t = \alpha_t \boldsymbol{x} + \sigma_t \boldsymbol{g}$ with $\boldsymbol{g} \sim \mathcal{N}(\mathbf{0}, \boldsymbol{I})$ independent of $(\boldsymbol{x}, \boldsymbol{y})$, as in Equation (3.2.85) in Chapter 3, and we will repeatedly use the notation $\boldsymbol{\xi}$ to denote realizations of $\boldsymbol{x}$ and $\boldsymbol{x}_t$.

This section focuses on the conceptual and algorithmic underpinnings for conditional sampling in modern generative models. For a treatment focused on implementation issues in various applications, see Chapter 8, Section 8.7 and subsequent sections.

### Classifier Guidance: Consistency via a Pretrained Classifier

Recall from Section 7.3.2 that the following decomposition of the optimal conditional posterior denoiser holds, by virtue of Bayes' rule and conditional independence of $\boldsymbol{y}$ and $\boldsymbol{x}_t$ given $\boldsymbol{x}$ (recall Figure 7.10):

$$\mathbb{E}[\boldsymbol{x} \mid \boldsymbol{x}_t = \boldsymbol{\xi}, y = \nu] = \mathbb{E}[\boldsymbol{x} \mid \boldsymbol{x}_t = \boldsymbol{\xi}] + \frac{\sigma_t^2}{\alpha_t} \nabla_{\boldsymbol{\xi}} \log p_{y|\boldsymbol{x}_t}(\nu \mid \boldsymbol{\xi}). \tag{7.4.7}$$

In other words, the representation (7.4.7) for the optimal conditional denoiser holds regardless of whether the explicit observation model $\boldsymbol{y} = h(\boldsymbol{x}) + \boldsymbol{w}$ is valid. This means that if we can learn such a denoiser (7.4.7) in the paired data setting, we can follow the denoising-diffusion methodology we have developed in Chapter 3 to perform class-conditional sampling.

A natural idea is then to directly implement the likelihood correction term in (7.4.7) using a deep network $f_{\theta_{\mathrm{c}}}$ with parameters $\theta_{\mathrm{c}}$, as in Equation (7.4.4):

$$f_{\theta_{\mathrm{c}}} : (t, \boldsymbol{x}_t) \mapsto \operatorname{softmax}(\boldsymbol{W}_{\mathrm{head}} \boldsymbol{z}(t, \boldsymbol{x}_t)). \tag{7.4.8}$$

This expression combines the final representations $\boldsymbol{z}(t, \boldsymbol{x}_t)$ (which also depend on $\theta_{\mathrm{c}}$) of the noisy inputs $\boldsymbol{x}_t$ with a classification head $\boldsymbol{W}_{\mathrm{head}} \in \mathbb{R}^{K \times d}$, which maps the representations to a probability distribution over the $K$ possible classes. As is common in practice, it also takes the time $t$ in the noising process as input. Thus, with appropriate training, it provides an approximation to the log-likelihood $\log p_{y|\boldsymbol{x}_t}$, and differentiating $\log f_{\theta_{\mathrm{c}}}$ with respect to its input $\boldsymbol{x}_t$ allows an approximation to the second term in Equation (7.4.7):

$$\bar{\boldsymbol{x}}_\theta^{\mathrm{naive}}(t, \boldsymbol{x}_t, y) = \bar{\boldsymbol{x}}_{\theta_{\mathrm{d}}}(t, \boldsymbol{x}_t) + \frac{\sigma_t^2}{\alpha_t} \nabla_{\boldsymbol{x}_t} \left\langle \log f_{\theta_{\mathrm{c}}}(t, \boldsymbol{x}_t), \boldsymbol{e}_y \right\rangle \tag{7.4.9}$$

where, as usual, we approximate the first term in Equation (7.4.7) via a learned unconditional denoiser for $\boldsymbol{x}_t$ with parameters $\theta_{\mathrm{d}}$, and where we write $\boldsymbol{e}_k$ for $k \in [K]$ to denote the $k$-th canonical basis vector for $\mathbb{R}^K$ (i.e., the vector with a one in the $k$-th position, and zeros elsewhere). The reader should note that the conditional denoiser $\bar{\boldsymbol{x}}_\theta$ requires two separate training runs, with separate losses: one for the classifier parameters $\theta_{\mathrm{c}}$, on a classification loss,[7] and one for the denoiser parameters $\theta_{\mathrm{d}}$, on a denoising loss. Such an approach to conditional sampling was already recognized and exploited to perform conditional sampling in pioneering early works on diffusion models, notably those by Sohl-Dickstein et al. [SWM+15] and by Song et al. [SSK+21].

However, this straightforward methodology has two key drawbacks (which is why we label it as "naive"). The first is that, empirically, such a trained deep network classifier frequently does not provide a strong enough guidance signal (in Equation (7.4.7)) to ensure that generated samples reflect the conditioning information $y$. This was first emphasized by Dhariwal and Nichol [DN21b], who noted that in the setting of class-conditional ImageNet generation, the learned deep network classifier's probability outputs for the class $y$ being conditioned on were frequently around 0.5—large enough to be the dominant class, but not large enough to provide a strong guidance signal—and that upon inspection, generations were not consistent with the conditioning class $y$. Dhariwal and Nichol [DN21b] proposed to address this heuristically by incorporating an "inverse temperature" hyperparameter $\gamma > 0$ into the definition of the naive conditional denoiser (7.4.9), referring to the resulting conditional denoiser as having incorporated **"classifier guidance"** (CG):

$$\bar{\boldsymbol{x}}_\theta^{\mathrm{CG}}(t, \boldsymbol{x}_t, y) = \bar{\boldsymbol{x}}_{\theta_{\mathrm{d}}}(t, \boldsymbol{x}_t) + \gamma \frac{\sigma_t^2}{\alpha_t} \nabla_{\boldsymbol{x}_t} \left\langle \log f_{\theta_{\mathrm{c}}}(t, \boldsymbol{x}_t), \boldsymbol{e}_y \right\rangle \tag{7.4.10}$$

with the case $\gamma = 1$ coinciding with (7.4.9).

Dhariwal and Nichol [DN21b] found that a setting $\gamma > 1$ performed best empirically. One possible interpretation for this is as follows: note that, in the context of the *true* likelihood term Equation (7.4.7), scaling by $\gamma$ gives equivalently

$$\gamma \frac{\sigma_t^2}{\alpha_t} \nabla_{\boldsymbol{\xi}} \log p_{\boldsymbol{y}|\boldsymbol{x}_t}(\boldsymbol{\nu} \mid \boldsymbol{\xi}) = \frac{\sigma_t^2}{\alpha_t} \nabla_{\boldsymbol{\xi}} \log \left( p_{\boldsymbol{y}|\boldsymbol{x}_t}(\boldsymbol{\nu} \mid \boldsymbol{\xi})^\gamma \right), \tag{7.4.11}$$

[7] In Chapter 8, we review the process of training such a classifier in full detail.

which suggests the natural interpretation of the parameter $\gamma$ performing (inverse) *temperature scaling* on the likelihood $p_{\boldsymbol{y}|\boldsymbol{x}_t}$, which is precise if we consider the renormalized distribution $p_{\boldsymbol{y}|\boldsymbol{x}_t}(\boldsymbol{\nu} \mid \boldsymbol{\xi})^\gamma / \int p_{\boldsymbol{y}|\boldsymbol{x}_t}(\boldsymbol{\nu}' \mid \boldsymbol{\xi})^\gamma \mathrm{d}\boldsymbol{\nu}'$. However, note that this *is not* a rigorous interpretation in the context of Equation (7.4.7), because the gradients are taken with respect to $\boldsymbol{\xi}$, and the normalization constant in the temperature-scaled distribution is in general a function of $\boldsymbol{\xi}$. Instead, the parameter $\gamma$ should simply be understood as amplifying large values of the deep network classifier's output probabilities $f_{\theta_{\mathrm{c}}}(t, \boldsymbol{x}_t)$ *relative to* smaller ones, which effectively amplifies the guidance signal provided in cases where the deep network $f$ assigns it the largest probability among the $K$ classes.

#### Simplifications and Improvements via Classifier-Free Guidance

Nevertheless, classifier guidance does not address the second key drawback of the naive methodology: it is both cumbersome and wasteful to have to train an auxiliary classifier $f_{\theta_{\mathrm{c}}}$ in addition to the unconditional denoiser $\bar{\boldsymbol{x}}_{\theta_{\mathrm{d}}}$, given that it is not possible to directly adapt a pretrained classifier due to the need for it to work well on noisy inputs $\boldsymbol{x}_t$ and incorporate other empirically-motivated architecture modifications. In particular, Dhariwal and Nichol [DN21b] found that it was necessary to *explicitly design the architecture of the deep network implementing the classifier to match that of the denoiser*.

Moreover, from a purely practical perspective—trying to obtain the best possible performance from the resulting sampler—the best-performing configuration of classifier guidance-based sampling departs even further from the idealized and conceptually sound framework we have presented above. To obtain the best performance, Dhariwal and Nichol [DN21b] found it necessary to provide the class label $y$ as an additional input to the denoiser $\bar{\boldsymbol{x}}_{\theta_{\mathrm{d}}}$. As a result, the idealized classifier-guided denoiser (7.4.10), derived by Dhariwal and Nichol [DN21b] as we have done above from the conditional posterior denoiser decomposition (7.4.7), is not exactly reflective of the best-performing denoiser in practice—such a denoiser actually combines a *conditional* denoiser for $\boldsymbol{x}_t$ given $y$ with an additional guidance signal from an auxiliary classifier!

This state of affairs, empirically motivated as it is, led Ho and Salimans [HS22a] in subsequent work to propose a more empirically pragmatic methodology, known as **classifier-free guidance (CFG)**. Instead of representing the conditional denoiser (7.4.7) as a weighted sum of an unconditional denoiser for $\boldsymbol{x}_t$ with a log-likelihood correction term (with possibly modified weights, as in classifier guidance), they accept the apparent necessity of training a conditional denoiser for $\boldsymbol{x}_t$ given $y$, as demonstrated by the experimental results of Dhariwal and Nichol [DN21b], and replace the log-likelihood gradient term with a correctly-weighted sum of this conditional denoiser with an *unconditional* denoiser for $\boldsymbol{x}$ given $\boldsymbol{x}_t$.[8]

[8]That said, Ho and Salimans [HS22a] actually proposed to use a different weighting than what we present here, based on the fact that Dhariwal and Nichol [DN21b] heuristically replaced the unconditional denoiser in (7.4.7) with a conditional denoiser. In fact, the weighting we derive and present here reflects modern practice, and in particular is used in state-of-the-art

**Deriving Classifier-Free Guidance.** To see how this structure arises, we begin with an 'idealized' version of the classifier guidance denoiser $\bar{\boldsymbol{x}}_\theta^{\mathrm{CG}}$ defined in (7.4.10), for which the denoiser $\bar{\boldsymbol{x}}_{\theta_{\mathrm{d}}}$ and the classifier $f_{\theta_{\mathrm{c}}}$ perfectly approximate their targets, via (7.4.7):

$$\bar{\boldsymbol{x}}_\theta^{\mathrm{CG,\,ideal}}(t, \boldsymbol{\xi}, \nu) = \mathbb{E}[\boldsymbol{x} \mid \boldsymbol{x}_t = \boldsymbol{\xi}] + \gamma \frac{\sigma_t^2}{\alpha_t} \nabla_{\boldsymbol{\xi}} \log p_{y \mid \boldsymbol{x}_t}(\nu \mid \boldsymbol{\xi}). \tag{7.4.12}$$

We then use Bayes' rule, in the form

$$\log p_{y\mid \boldsymbol{x}_t} = \log p_{\boldsymbol{x}_t \mid y} + \log p_y - \log p_{\boldsymbol{x}_t}, \tag{7.4.13}$$

together with Tweedie's formula (Theorem 3.2, modified as in Equation (3.2.86)) to convert between score functions and denoisers, to obtain

$$\begin{aligned} \bar{\boldsymbol{x}}_\theta^{\mathrm{CG,\,ideal}}(t, \boldsymbol{\xi}, \nu) &= \frac{1}{\alpha_t}\boldsymbol{\xi} + (1-\gamma)\frac{\sigma_t^2}{\alpha_t}\nabla_{\boldsymbol{\xi}} \log p_{\boldsymbol{x}_t}(\boldsymbol{\xi}) + \gamma \frac{\sigma_t^2}{\alpha_t}\nabla_{\boldsymbol{\xi}} \log p_{\boldsymbol{x}_t \mid y}(\boldsymbol{\xi} \mid \nu) \\ &= (1-\gamma)\mathbb{E}[\boldsymbol{x} \mid \boldsymbol{x}_t = \boldsymbol{\xi}] + \gamma \mathbb{E}[\boldsymbol{x} \mid \boldsymbol{x}_t = \boldsymbol{\xi}, y = \nu], \end{aligned} \tag{7.4.14}$$

where in the last line, we apply Equation (7.3.11). Now, Equation (7.4.14) suggests a natural approximation strategy: we combine a learned unconditional denoiser for $\boldsymbol{x}$ given $\boldsymbol{x}_t$, as previously, with a learned *conditional* denoiser for $\boldsymbol{x}$ given $\boldsymbol{x}_t$ and $y$.

However, following Ho and Salimans [HS22a] and the common practice of training deep network denoisers, it is standard to use the *same* deep network to represent both the conditional and unconditional denoisers by introducing an additional label, which we will denote by $\varnothing$, to denote the "unconditional" case. This leads to the form of the CFG denoiser:

$$\bar{\boldsymbol{x}}_\theta^{\mathrm{CFG}}(t, \boldsymbol{x}_t, y) = (1-\gamma)\bar{\boldsymbol{x}}_\theta(t, \boldsymbol{x}_t, \varnothing) + \gamma \bar{\boldsymbol{x}}_\theta(t, \boldsymbol{x}_t, y). \tag{7.4.15}$$

To train a denoiser $\bar{\boldsymbol{x}}_\theta(t, \boldsymbol{x}_t, y^+)$ for use with classifier-free guidance sampling, where $y^+ \in \{1, \dots, K, \varnothing\}$, we proceed almost identically to the unconditional training procedure in Algorithm 3.2, but with two modifications:

1. When we sample from the dataset, we sample a pair $(\boldsymbol{x}, y)$ rather than just a sample $\boldsymbol{x}$.

2. Every time we sample a pair from the dataset, we sample the augmented label $y^+$ via

$$y^+ = \begin{cases} \varnothing & \text{with probability } p_{\mathrm{uncond}}; \\ y & \text{else.} \end{cases} \tag{7.4.16}$$

   Here, $p_{\mathrm{uncond}} \in [0, 1]$ is a new hyperparameter. This can be viewed as a form of dropout [SHK+14].

In this way, we train a conditional denoiser suitable for use in classifier-free guidance sampling. We summarize the overall sampling process for class-conditioned sampling with classifier-free guidance in Algorithm 7.2.

diffusion models such as Stable Diffusion 3.5 [EKB+24].

**Algorithm 7.2** Conditional sampling with classification data, using class-conditioned denoiser.

**Input:** An ordered list of timesteps $0 \leq t_0 < \cdots < t_L \leq T$ to use for sampling.
**Input:** Class label $\nu \in \{1, \dots, K\}$ to condition on.
**Input:** A denoiser $\bar{\boldsymbol{x}}_\theta \colon \{t_\ell\}_{\ell=1}^{L} \times \mathbb{R}^D \times \{1, \dots, K, \varnothing\} \to \mathbb{R}^D$ for $p_{\boldsymbol{x}|y}$ and $p_{\boldsymbol{x}}$ (input $\varnothing$ for $p_{\boldsymbol{x}}$).
**Input:** Scale and noise level functions $\alpha, \sigma \colon \{t_\ell\}_{\ell=0}^{L} \to \mathbb{R}_{\geq 0}$.
**Input:** Guidance strength $\gamma \geq 0$ ($\gamma > 1$ preferred for performance).
**Output:** A sample $\hat{\boldsymbol{x}}$, approximately from $p_{\boldsymbol{x}|y}(\cdot \mid \nu)$.

1: **function** DDIMSamplerConditionalCFG($\bar{\boldsymbol{x}}_\theta, \nu, \gamma, (t_\ell)_{\ell=0}^{L}$)
2: Initialize $\hat{\boldsymbol{x}}_{t_L} \sim$ approximate distribution of $\boldsymbol{x}_{t_L}$ (VP $\implies \mathcal{N}(\mathbf{0}, \boldsymbol{I})$, VE $\implies \mathcal{N}(\mathbf{0}, t_L^2 \boldsymbol{I})$).
3: **for** $\ell = L, L-1, \dots, 1$ **do**
4: Compute

$$\hat{\boldsymbol{x}}_{t_{\ell-1}} \doteq \frac{\sigma_{t_{\ell-1}}}{\sigma_{t_\ell}} \hat{\boldsymbol{x}}_{t_\ell} + \left( \alpha_{t_{\ell-1}} - \frac{\sigma_{t_{\ell-1}}}{\sigma_{t_\ell}} \alpha_{t_\ell} \right) \left( (1-\gamma) \bar{\boldsymbol{x}}_\theta(t_\ell, \hat{\boldsymbol{x}}_{t_\ell}, \varnothing) + \gamma \bar{\boldsymbol{x}}_\theta(t_\ell, \hat{\boldsymbol{x}}_{t_\ell}, \nu) \right)$$

5: **end for**
6: **return** $\hat{\boldsymbol{x}}_{t_0}$
7: **end function**

Ho and Salimans [HS22a] reports strong empirical performance for class-conditional image generation with classifier-free guidance, and it has become a mainstay of the largest-scale practical diffusion models, such as Stable Diffusion [RBL+22] and its derivatives.

**What Denoisers Does Classifier-Free Guidance Promote Learning?** As we have seen, the derivation of classifier-free guidance is rather opaque and empirically motivated, giving little insight into the mechanisms behind its strong performance. A number of theoretical works have attempted to address this [BN24b; LWQ25; WCL+24]. They provide explanations for some parts of the overall CFG methodology—itself encompassing denoiser parameterization and training, as well as configuration of the guidance strength and performance at sampling time. Below, following our running theme in the book, we will give an interpretation in the simplifying setting of a Gaussian mixture model data distribution and denoiser. First, we consider *the effect of CFG*: what does a large weight $\gamma \gg 1$ promote?

*Example* 7.5. Let us recall the low-rank mixture of Gaussians data generating process we studied in Example 3.3 (and specifically, the form in Equation (3.2.69)). Given $K \in \mathbb{N}$ classes, we assume that

$$\boldsymbol{x} \sim \frac{1}{K} \sum_{k=1}^{K} \mathcal{N}(\mathbf{0}, \boldsymbol{U}_k \boldsymbol{U}_k^\top), \tag{7.4.17}$$

where each $\boldsymbol{U}_k \in \mathsf{O}(D, P) \subseteq \mathbb{R}^{D\times P}$ is a matrix with orthogonal columns, and $P \ll D$. Moreover, we assume that the class label $y \in [K]$ is a deterministic function of $\boldsymbol{x}$ mapping an example to its corresponding mixture component. We consider the simple noise process (3.2.1) studied in Chapter 3, so that $\boldsymbol{x}_t = \boldsymbol{x}+t\boldsymbol{g}$ for all $t \in [0, T]$, with $\boldsymbol{g} \sim \mathcal{N}(\boldsymbol{0}, \boldsymbol{I})$.

Applying the analysis in Example 3.3 (and the subsequent analysis of the low-rank case, culminating in Equation (3.2.80)), we obtain for the class-conditional optimal denoisers

$$\mathbb{E}[\boldsymbol{x} \mid \boldsymbol{x}_t = \boldsymbol{\xi}, y = \nu] = \frac{1}{1+t^2}\boldsymbol{U}_\nu \boldsymbol{U}_\nu^\top \boldsymbol{\xi} \tag{7.4.18}$$

for each $\nu \in [K]$, and for the optimal unconditional denoiser, we obtain

$$\mathbb{E}[\boldsymbol{x} \mid \boldsymbol{x}_t = \boldsymbol{\xi}] = \frac{1}{1+t^2}\sum_{k=1}^{K} \frac{\exp\left(\frac{1}{2t^2(1+t^2)}\|\boldsymbol{U}_k^\top \boldsymbol{\xi}\|_2^2\right)}{\sum_{i=1}^{K}\exp\left(\frac{1}{2t^2(1+t^2)}\|\boldsymbol{U}_i^\top \boldsymbol{\xi}\|_2^2\right)}\boldsymbol{U}_k \boldsymbol{U}_k^\top \boldsymbol{\xi}. \tag{7.4.19}$$

As a result, we can express the CFG denoiser with guidance strength $\gamma > 1$ as

$$\begin{aligned}\bar{\boldsymbol{x}}^{\mathrm{CFG,\,ideal}}(t, \boldsymbol{x}_t, y) = \frac{1}{1+t^2}\Bigg((1-\gamma)\sum_{k=1}^{K} \frac{\exp\left(\frac{1}{2t^2(1+t^2)}\|\boldsymbol{U}_k^\top \boldsymbol{x}_t\|_2^2\right)}{\sum_{i=1}^{K}\exp\left(\frac{1}{2t^2(1+t^2)}\|\boldsymbol{U}_i^\top \boldsymbol{x}_t\|_2^2\right)}\boldsymbol{U}_k \boldsymbol{U}_k^\top \\ + \gamma \boldsymbol{U}_y \boldsymbol{U}_y^\top\Bigg)\boldsymbol{x}_t.\end{aligned} \tag{7.4.20}$$

This denoiser has a simple, interpretable form:

1. The first term, corresponding to *the unconditional denoiser*, performs denoising of the signal $\boldsymbol{x}_t$ against an average of the denoisers associated with each subspace, weighted by how correlated $\boldsymbol{x}_t$ is with each subspace.

2. The second term, corresponding to *the conditional denoiser*, simply performs denoising with the conditioning class's denoiser.

The CFG scheme averages these two denoisers. The effect of this averaging can be gleaned from the refactoring

$$\begin{aligned}\bar{\boldsymbol{x}}^{\mathrm{CFG,\,ideal}}(t, \boldsymbol{x}_t, y) = \frac{1}{1+t^2}\Bigg(\left[\gamma + (1-\gamma)\frac{\exp\left(\frac{1}{2t^2(1+t^2)}\|\boldsymbol{U}_y^\top \boldsymbol{x}_t\|_2^2\right)}{\sum_{i=1}^{K}\exp\left(\frac{1}{2t^2(1+t^2)}\|\boldsymbol{U}_i^\top \boldsymbol{x}_t\|_2^2\right)}\right]\boldsymbol{U}_y \boldsymbol{U}_y^\top \\ + (1-\gamma)\sum_{k\neq y} \frac{\exp\left(\frac{1}{2t^2(1+t^2)}\|\boldsymbol{U}_k^\top \boldsymbol{x}_t\|_2^2\right)}{\sum_{i=1}^{K}\exp\left(\frac{1}{2t^2(1+t^2)}\|\boldsymbol{U}_i^\top \boldsymbol{x}_t\|_2^2\right)}\boldsymbol{U}_k \boldsymbol{U}_k^\top\Bigg)\boldsymbol{x}_t,\end{aligned} \tag{7.4.21}$$

where we combined the conditional denoiser with the corresponding $k = y$ term in the sum over $k \in [K]$. We have

$$\sum_{k=1}^{K} \frac{\exp\left(\frac{1}{2t^2(1+t^2)}\|\boldsymbol{U}_k^\top \boldsymbol{x}_t\|_2^2\right)}{\sum_{i=1}^{K} \exp\left(\frac{1}{2t^2(1+t^2)}\|\boldsymbol{U}_i^\top \boldsymbol{x}_t\|_2^2\right)} = 1, \tag{7.4.22}$$

and each summand is nonnegative, hence also bounded above by 1. So we can conclude two regimes for the terms in Equation (7.4.21):

1. **Well-correlated regime:** In this case, suppose the noisy input $\boldsymbol{x}_t$ correlates well with one of the subspaces $\boldsymbol{U}_y$. If $\boldsymbol{x}_t$ correlates well with $\boldsymbol{U}_y$, then the normalized weight corresponding to the $k = y$ summand in the unconditional denoiser is near to 1. Then

$$\gamma + (1-\gamma)\frac{\exp\left(\frac{1}{2t^2(1+t^2)}\|\boldsymbol{U}_y^\top \boldsymbol{x}_t\|_2^2\right)}{\sum_{i=1}^{K} \exp\left(\frac{1}{2t^2(1+t^2)}\|\boldsymbol{U}_i^\top \boldsymbol{x}_t\|_2^2\right)} \approx 1, \tag{7.4.23}$$

   all other weights are necessarily near to zero, and the CFG denoiser is approximately equal to the denoiser associated to the conditioning class $y$.

   This implies: *if the CFG denoiser's input is highly correlated with one of the class subspaces, the denoiser is approximately equal to the corresponding class-conditional denoiser*! Hence, we expect the CFG denoising sampler to converge.

2. **Poorly-correlated regime:** In contrast, if $\boldsymbol{x}_t$ *does not* correlate well with $\boldsymbol{U}_y$ (say because $t$ is large), then the normalized weight corresponding to the $k = y$ summand in the unconditional denoiser is near to 0. As a result,

$$\gamma + (1-\gamma)\frac{\exp\left(\frac{1}{2t^2(1+t^2)}\|\boldsymbol{U}_y^\top \boldsymbol{x}_t\|_2^2\right)}{\sum_{i=1}^{K} \exp\left(\frac{1}{2t^2(1+t^2)}\|\boldsymbol{U}_i^\top \boldsymbol{x}_t\|_2^2\right)} \approx \gamma, \tag{7.4.24}$$

   and thus the guidance strength $\gamma \gg 1$ places a large positive weight on the denoiser associated to $y$.

   Meanwhile, in the second term of Equation (7.4.21), any classes $k \neq y$ that are well-correlated with $\boldsymbol{x}_t$ receive a large *negative* weight from the $1 - \gamma$ coefficient. This simultaneously has the effect of making the denoised signal vastly more correlated with the conditioning class $y$, and making it negatively correlated with the previous iterate (i.e., the iterate before denoising).

   In other words, *CFG steers the iterative denoising process towards the conditioning class and away from the previous iterate*, a different dynamics from purely conditional sampling (i.e., the case $\gamma = 1$).

■

Example 7.5 shows that in the mixture of Gaussians setting, the CFG denoiser provides a strong bias towards the conditioning class, while preserving local correctness (hence convergence of sampling) when the noise level $t$ is small. This is suggestive of why a large weight $\gamma \gg 1$ has been found essential in practice.

Next, the following example will demonstrate an insight into the *parameterization* of the denoiser in the presence of low-dimensional structure, again in the Gaussian mixture model.

*Example* 7.6. Consider again the mixture of Gaussians model studied in Example 7.5:

$$\boldsymbol{x} \sim \frac{1}{K}\sum_{k=1}^{K} \mathcal{N}(\mathbf{0}, \boldsymbol{U}_k\boldsymbol{U}_k^\top), \tag{7.4.25}$$

where each $\boldsymbol{U}_k \in \mathsf{O}(D,P) \subseteq \mathbb{R}^{D\times P}$ is a matrix with orthogonal columns, and $P \ll D$. The class label $y \in [K]$ is a deterministic function of $\boldsymbol{x}$ mapping an example to its corresponding mixture component. For this distribution, and again under the simple noise process (3.2.1) studied in Chapter 3, where $\boldsymbol{x}_t = \boldsymbol{x} + t\boldsymbol{g}$ for all $t \in [0,T]$, with $\boldsymbol{g} \sim \mathcal{N}(\mathbf{0}, \boldsymbol{I})$, we saw in Example 7.5 that the ideal CFG denoiser takes the form

$$\begin{aligned}\bar{\boldsymbol{x}}^{\mathrm{CFG,\,ideal}}(t,\boldsymbol{x}_t,y) = \frac{1}{1+t^2}\Bigg((1-\gamma)\sum_{k=1}^{K}\frac{\exp\left(\frac{1}{2t^2(1+t^2)}\|\boldsymbol{U}_k^\top\boldsymbol{x}_t\|_2^2\right)}{\sum_{i=1}^{K}\exp\left(\frac{1}{2t^2(1+t^2)}\|\boldsymbol{U}_i^\top\boldsymbol{x}_t\|_2^2\right)}\boldsymbol{U}_k\boldsymbol{U}_k^\top \\ + \gamma\boldsymbol{U}_y\boldsymbol{U}_y^\top\Bigg)\boldsymbol{x}_t.\end{aligned} \tag{7.4.26}$$

Now we consider the problem of parameterizing a learnable denoiser $\boldsymbol{x}_\theta^{\mathrm{CFG}}$ to represent the optimal denoiser (7.4.26). For tractability, we add an additional assumption associated to the subspaces $\boldsymbol{U}_k$ being 'distinguishable' from one another, which is natural in practice: specifically, we assume that for any pair of indices $k, k' \in [K]$ with $k \neq k'$, we can find a set of $K$ nonzero directions $\boldsymbol{v}_k \in \mathbb{R}^D$ such that

$$\boldsymbol{U}_k\boldsymbol{U}_k^\top\boldsymbol{v}_k = \boldsymbol{v}_k, \quad \boldsymbol{U}_{k'}\boldsymbol{U}_{k'}^\top\boldsymbol{v}_k = \mathbf{0}, \;\; k' \neq k. \tag{7.4.27}$$

This is a slightly stronger assumption than simple distinguishability, but it should be noted that it is not overly restrictive: for example, it still allows the subspaces $\boldsymbol{U}_k$ to have significant correlations with one another.[9]

These vectors $\boldsymbol{v}_k$ can then be thought of as *embeddings* of the class label $y \in [K]$, and we can use them to define a more general operator that can represent both the unconditional and class-conditional denoisers. More precisely, consider

[9]More generally, this assumption is naturally formulated as an *incoherence condition* between the subspaces $\boldsymbol{U}_{[K]}$, familiar from the theory of compressive sensing [WM22].

the mapping

$$(\boldsymbol{x}_t, \boldsymbol{v}) \mapsto \sum_{k=1}^{K} \frac{\exp\left(\frac{1}{2t^2(1+t^2)}|\boldsymbol{x}_t^\top \boldsymbol{U}_k \boldsymbol{U}_k^\top \boldsymbol{v}|\right)}{\sum_{i=1}^{K} \exp\left(\frac{1}{2t^2(1+t^2)}|\boldsymbol{x}_t^\top \boldsymbol{U}_i \boldsymbol{U}_i^\top \boldsymbol{v}|\right)} \boldsymbol{U}_k \boldsymbol{U}_k^\top \boldsymbol{x}_t. \tag{7.4.28}$$

Then if we plug in $\boldsymbol{v} = \boldsymbol{x}_t$ to the operator (7.4.28), we obtain a result proportional to the first term in (7.4.19), corresponding to the optimal unconditional denoiser for $\boldsymbol{x}_t$. On the other hand, if we substitute $\boldsymbol{v} = \boldsymbol{v}_y$ for some $y \in [K]$, we get by construction of the embedding vectors $\boldsymbol{v}_k$

$$\begin{aligned}
&\sum_{k=1}^{K} \frac{\exp\left(\frac{1}{2t^2(1+t^2)}|\boldsymbol{x}_t^\top \boldsymbol{U}_k \boldsymbol{U}_k^\top \boldsymbol{v}_y|\right)}{\sum_{i=1}^{K} \exp\left(\frac{1}{2t^2(1+t^2)}|\boldsymbol{x}_t^\top \boldsymbol{U}_i \boldsymbol{U}_i^\top \boldsymbol{v}_y|\right)} \boldsymbol{U}_k \boldsymbol{U}_k^\top \boldsymbol{x}_t \\
= \;& \frac{\exp\left(\frac{1}{2t^2(1+t^2)}|\boldsymbol{x}_t^\top \boldsymbol{v}_y|\right)}{\exp\left(\frac{1}{2t^2(1+t^2)}|\boldsymbol{x}_t^\top \boldsymbol{v}_y|\right) + K - 1} \boldsymbol{U}_y \boldsymbol{U}_y^\top \boldsymbol{x}_t \qquad (7.4.29) \\
&+ \sum_{k \neq y} \frac{1}{\exp\left(\frac{1}{2t^2(1+t^2)}|\boldsymbol{x}_t^\top \boldsymbol{v}_y|\right) + K - 1} \boldsymbol{U}_k \boldsymbol{U}_k^\top \boldsymbol{x}_t. \qquad (7.4.30)
\end{aligned}$$

We aim to simplify the preceding expression further. To this end, recall that when using this denoiser for sampling, we have $\boldsymbol{x}_t = \boldsymbol{x} + t\boldsymbol{g}$, with $y$ denoting the label of $\boldsymbol{x}$. In other words, conditioned on $y$, we have that $\boldsymbol{x}_t = \boldsymbol{U}_y \boldsymbol{z} + t\boldsymbol{g}$, where $\boldsymbol{z}$ is an independent $P$-dimensional Gaussian noise vector with zero mean and identity covariance. We have

$$\begin{aligned}
\left|\boldsymbol{x}_t^\top \boldsymbol{v}_y\right| &= |\langle \boldsymbol{U}_y \boldsymbol{z}, \boldsymbol{v}_y\rangle + t\langle \boldsymbol{g}, \boldsymbol{v}_y\rangle| \\
&= \left|\langle \boldsymbol{z}, \boldsymbol{U}_y^\top \boldsymbol{v}_y\rangle + t\langle \boldsymbol{g}, \boldsymbol{v}_y\rangle\right|.
\end{aligned}$$

Because $\boldsymbol{z}$ and $\boldsymbol{g}$ are independent Gaussian random variables with identity covariance matrices, it follows from *rotational invariance* of the Gaussian distribution (see [Ver18]) that

$$\begin{aligned}
\left|\boldsymbol{x}_t^\top \boldsymbol{v}_y\right| &\overset{\mathrm{d}}{=} \left|\|\boldsymbol{U}_y^\top \boldsymbol{v}_y\|_2 z + t\|\boldsymbol{v}_y\|_2 g\right| \\
&= \left|\|\boldsymbol{v}_y\|_2 z + t\|\boldsymbol{v}_y\|_2 g\right|,
\end{aligned}$$

where $z$ and $g$ are independent *scalar* $\mathcal{N}(0,1)$ random variables, and $\overset{\mathrm{d}}{=}$ denotes equality in distribution (the random variables have the same distribution). The second line above follows from the definition of $\boldsymbol{v}_y$. Now, by independence, the random variable $z + tg \sim \mathcal{N}(0, (1+t^2))$. Then using the well-known result that $\mathbb{E}[|g|] = \sqrt{2/\pi}$ if $g \sim \mathcal{N}(0,1)$, we conclude

$$\mathbb{E}\left[\left|\boldsymbol{x}_t^\top \boldsymbol{v}_y\right|\right] = \sqrt{(2/\pi)(1+t^2)}\|\boldsymbol{v}_y\|_2.$$

In practice, the quantity $\left|\boldsymbol{x}_t^\top \boldsymbol{v}_y\right|$ is close to its expectation *with overwhelming probability*, by Gaussian concentration (see, again, [Ver18]). On the other hand, we have by similar reasoning

$$\mathbb{E}\left[\|\boldsymbol{x}_t\|_2^2\right] = P + t^2 D,$$

and properties of the Gaussian distribution again imply that $\|\boldsymbol{x}_t\|_2$ will concentrate around $\sqrt{P + t^2 D}$ with overwhelming probability. It is then sensible to renormalize the embeddings $\boldsymbol{v}_y$ so that their dot-product with $\boldsymbol{x}_t$ has the same scale: defining

$$\boldsymbol{v}_{t,y} = \frac{(P + t^2 D)}{\sqrt{(2/\pi)(1+t^2)}} \frac{\boldsymbol{v}_y}{\|\boldsymbol{v}_y\|_2}$$

for the normalized embeddings, our previous argument implies

$$\left|\boldsymbol{x}_t^\top \boldsymbol{v}_{t,y}\right| \approx (P + t^2 D)$$

with overwhelming probability. Returning our attention to our starting point, i.e. (7.4.28) evaluated at $(\boldsymbol{x}_t, \boldsymbol{v}_y)$ and our subsequent simplification, we have for the "softmax" argument

$$\frac{1}{2t^2(1+t^2)} |\boldsymbol{x}_t^\top \boldsymbol{v}_y| \approx \frac{P + t^2 D}{2t^2(1+t^2)}$$

with overwhelming probability with respect to $\boldsymbol{x}_t$. Hence if $t$ is bounded and the subspace dimension $P$ is sufficiently large,[10] the following approximation holds with overwhelming probability:

$$\sum_{k=1}^{K} \frac{\exp\left(\frac{1}{2t^2(1+t^2)} \boldsymbol{x}_t^\top \boldsymbol{U}_k \boldsymbol{U}_k^\top \boldsymbol{v}_y\right)}{\sum_{i=1}^{K} \exp\left(\frac{1}{2t^2(1+t^2)} \boldsymbol{x}_t^\top \boldsymbol{U}_i \boldsymbol{U}_i^\top \boldsymbol{v}_y\right)} \boldsymbol{U}_k \boldsymbol{U}_k^\top \boldsymbol{x}_t \approx \boldsymbol{U}_y \boldsymbol{U}_y^\top \boldsymbol{x}_t. \tag{7.4.31}$$

In words, **evaluating the operator (7.4.28) at the embedding for a class $\boldsymbol{v}_{t,y}$ yields the class-conditional denoiser for $y$!**

Thus, the family of operators (7.4.28) provides a unified way to parameterize the constituent operators in the optimal denoiser for $\boldsymbol{x}$ *within a single 'network'*. More precisely, it is enough to add the output of an instantiation of (7.4.28) with input $(\boldsymbol{x}_t, \boldsymbol{x}_t)$ to an instantiation with input $(\boldsymbol{x}_t, \boldsymbol{v}_{t,y})$. The resulting operator is a function of $(t, \boldsymbol{x}_t, y)$, and computationally, the subspaces $(\boldsymbol{U}_k)_{k=1}^K$ and embeddings $y \mapsto \boldsymbol{v}_{t,y}$ become its learnable parameters. ■

[10] For example, it suffices to consider a large-scale, asymptotic regime where $P, D \to \infty$ with their ratio $P/D$ converging to a fixed constant. In such a scenario, the *aspect ratio* of the subspaces is fixed while their sizes go to infinity, which can act as a model for some discretization phenomena (e.g., a fixed scene sampled as an image with the number of pixels going to infinity).

Example 7.6 shows that in the special case of a low-rank mixture of Gaussians data distribution for $\boldsymbol{x}$ with incoherent components, operators of the form

$$(\boldsymbol{x}_t, \boldsymbol{v}) \mapsto \sum_{k=1}^{K} \frac{\exp\left(\frac{1}{2t^2(1+t^2)} \boldsymbol{x}_t^\top \boldsymbol{U}_k \boldsymbol{U}_k^\top \boldsymbol{v}\right)}{\sum_{i=1}^{K} \exp\left(\frac{1}{2t^2(1+t^2)} \boldsymbol{x}_t^\top \boldsymbol{U}_i \boldsymbol{U}_i^\top \boldsymbol{v}\right)} \boldsymbol{U}_k \boldsymbol{U}_k^\top \boldsymbol{x}_t \tag{7.4.32}$$

provide a sufficiently rich class of operators to parameterize the ideal denoiser for noisy observations $\boldsymbol{x}_t$ of $\boldsymbol{x}$ when using classifier-free guidance. For such operators, the auxiliary input $\boldsymbol{v}$ can be taken as either $\boldsymbol{x}_t$ or a suitable embedding of the class label $y \mapsto \boldsymbol{v}_y$ in order to realize such a denoiser—as the example shows, the ideal denoiser is a weighted sum of such operators, with weights derived from the guidance strength $\gamma > 1$.

Based on the framework in Chapter 5, which develops deep network architectures suitable for transforming more general data distributions to structured representations using the low-rank mixture of Gaussians model as a primitive, it is natural to imagine that operators of the type (7.4.32) may be leveraged in denoisers for general data distributions $\boldsymbol{x}$ with low-dimensional geometrically-structured components that are sufficiently distinguishable (say, incoherent) from one another. The next section demonstrates that this is indeed the case.

### 7.4.2 Caption-Conditioned Image Generation

In Section 7.4.1 we have seen, conceptually, how to set up denoisers for use with *classifier-free guidance*, the dominant practical methodology for performing conditional sampling via the diffusion-denoising paradigm. In short, such denoisers take as input the noisy data $\boldsymbol{x}_t$ as well as either the conditioning signal $\boldsymbol{y}$ or a "null input" $\varnothing$, which determines whether the denoiser should perform *unconditional denoising* or *conditional denoising* (conditioned on $\boldsymbol{y}$).

Given this design, the most important practical question is *how to realize this family of operators as a neural network?* An empirically-successful neural network layer design known as "cross attention", which we will present in due course, provides a practical answer to this question. However, even before this, there are several more fundamental issues to sort out:

1. *Input modality mismatch*: In the most interesting practical applications, the data $\boldsymbol{x}$ and the conditioning signal $\boldsymbol{y}$ *do not lie in directly-comparable spaces*, although they share many underlying structural relationships. For example, we saw this in the case of a class label $y$ in Section 7.4.1. More generally, one could consider a natural language description $\boldsymbol{y}$ of an image $\boldsymbol{x}$ (referred to as a "caption"), where $\boldsymbol{y}$ would correspond to a sequence of characters in different alphabets.

2. *Parameterization and flexibility*: Downstream of the input modality issue, there is a fundamental question of whether the same basic architecture for the network can be used for different modalities, or whether this needs to be tuned in a very precise way per-modality.

3. *Training*: This entails the proper way to use our paired samples $(\boldsymbol{x}, \boldsymbol{y})$ to train and deploy the conditional denoisers. The design space is vast: one could imagine applying various modality-specific encoder-decoder architectures, as we have developed in Chapters 4 and 6, together with different denoising backbones in various combinations. The key questions here are of *expressivity*, getting the best conditional denoising performance; *efficiency*, getting the best performance per unit of training data; and *stability*, guaranteeing that the training process converges quickly without collapse.

We will cover these three issues in detail below, providing guidance on how to explore the design space and, where possible, generally-useful prescriptions.

### Modality-Agnostic Input Embedding

Although we do not know the precise relationship between $\boldsymbol{x}$ and $\boldsymbol{y}$ in the general paired data setting, in all practical settings of interest, they have significant correlations. For example, when the conditioning signal $\boldsymbol{y}$ is a "caption" describing the image $\boldsymbol{x}$ in natural language, it is easy to imagine a human artist being able to produce an accurate sketch of $\boldsymbol{x}$ based solely on a high-quality caption. In our probabilistic setting, one natural way to quantify these "correlations" is in terms of the mutual information, which we saw previously in Chapter 4 (Equation (4.1.12)):

$$I(\boldsymbol{x}; \boldsymbol{y}) = h(\boldsymbol{x}) - h(\boldsymbol{x} \mid \boldsymbol{y}). \tag{7.4.33}$$

The mutual information (7.4.33) is a purely information-theoretic quantity that does not depend on modality-specific aspects of $\boldsymbol{x}$ and $\boldsymbol{y}$. Hence it offers a natural criterion for learning a useful representation of either input $\boldsymbol{x}$, $\boldsymbol{y}$: one simply seeks to learn, say for $\boldsymbol{x}$, a representation $\boldsymbol{z} = f(\boldsymbol{x})$ that preserves the mutual information, so that $I(\boldsymbol{z}; \boldsymbol{y}) \approx I(\boldsymbol{x}; \boldsymbol{y})$. That is:

$$\text{Learn } \ \boldsymbol{z} = f(\boldsymbol{x}) \ \text{ such that } \ I(\boldsymbol{z}; \boldsymbol{y}) \approx I(\boldsymbol{x}; \boldsymbol{y}). \tag{7.4.34}$$

Actually, we can be even more prescriptive here. For an encoder $f$ which is deterministic (matching all the cases we have studied throughout Chapters 4 and 6), a fundamental result in information theory known as the *data processing inequality* tells us that processing $\boldsymbol{x}$ with $f$ to produce $\boldsymbol{z} = f(\boldsymbol{x})$ changes the mutual information in a predictable way: *it can only reduce it.*

**Theorem 7.1** (Data Processing Inequality)**.** *Consider random variables $\boldsymbol{y}, \boldsymbol{x}, \boldsymbol{z}$ defined such that $\boldsymbol{z}$ is independent of $\boldsymbol{y}$, conditioned on $\boldsymbol{x}$.*[11] *Then the data processing inequality holds:*

$$I(\boldsymbol{x}; \boldsymbol{y}) \geq I(\boldsymbol{z}; \boldsymbol{y}). \tag{7.4.35}$$

*Proof.* Consider the mutual information $I(\boldsymbol{y}; \boldsymbol{x}, \boldsymbol{z})$ between $\boldsymbol{y}$ and $(\boldsymbol{x}, \boldsymbol{z})$. Because $\boldsymbol{z}$ is independent of $\boldsymbol{y}$ conditioned on $\boldsymbol{x}$, we have

$$h(\boldsymbol{y} \mid \boldsymbol{x}, \boldsymbol{z}) = h(\boldsymbol{y} \mid \boldsymbol{x}),$$

[11] In other words, the random variables form a Markov chain $\boldsymbol{y} \to \boldsymbol{x} \to \boldsymbol{z}$.

so $I(\boldsymbol{y}; \boldsymbol{x}, \boldsymbol{z}) = I(\boldsymbol{y}; \boldsymbol{x})$. Meanwhile, we have by Bayes' rule that

$$p_{\boldsymbol{y}|\boldsymbol{x},\boldsymbol{z}} = p_{\boldsymbol{y}|\boldsymbol{z}} \frac{p_{\boldsymbol{y},\boldsymbol{x},\boldsymbol{z}}}{p_{\boldsymbol{y},\boldsymbol{z}} p_{\boldsymbol{x}|\boldsymbol{z}}},$$

which implies by the definition of conditional differential entropy that

$$I(\boldsymbol{y}; \boldsymbol{x}, \boldsymbol{z}) = I(\boldsymbol{y}; \boldsymbol{z}) - \mathbb{E}_{\boldsymbol{x},\boldsymbol{y},\boldsymbol{z}} \left[ \log \left( \frac{p_{\boldsymbol{x}|\boldsymbol{z}} p_{\boldsymbol{y},\boldsymbol{z}}}{p_{\boldsymbol{y},\boldsymbol{x},\boldsymbol{z}}} \right) \right].$$

Applying Jensen's inequality to the expectation in the previous line then implies the claim, because log is a concave function (and the mutual information is symmetric, i.e. $I(\boldsymbol{x}; \boldsymbol{y}) = I(\boldsymbol{y}; \boldsymbol{x})$). $\square$

For a deterministic encoder $f$, $\boldsymbol{z} = f(\boldsymbol{x})$ is always independent of $\boldsymbol{y}$ when conditioned on $\boldsymbol{x}$. The data processing inequality therefore suggests the following computational procedure to find the encoder $f$: *seek to maximize the mutual information between the representation and the conditioning signal.* Equation (7.4.34) then leads to the objective

$$\max_{f} I(f(\boldsymbol{x}); \boldsymbol{y}). \tag{7.4.36}$$

This is known as the *Infomax principle* [Lin88], and it dates to the earliest years of representation learning.

**Computational considerations in applying the Infomax principle.** There are two key conceptual issues to keep in mind when thinking about how to apply the (abstract) Infomax principle (7.4.36) for learning a representation. The first is that in general it is computationally hard to estimate the mutual information, which constitutes the objective in (7.4.36). As we discussed in the context of rate distortion and lossy coding when we introduced the mutual information in Chapter 4, the mutual information can be relaxed to remain well-defined even when $\boldsymbol{x}$ and $\boldsymbol{y}$ are low-dimensional and have differential entropy approaching $-\infty$ (e.g., by connection to the Gaussian rate reduction function that we have studied in Chapters 4 to 6). Moreover, empirically such relaxations have non-worst-case statistical properties in practice, as we saw in the experiments of Chapter 4; and the accuracy of these approximations can be improved by incremental processing, as we saw in Chapter 5. This makes a criterion like Equation (7.4.36) an excellent foundation for learning representations of paired data $(\boldsymbol{x}, \boldsymbol{y})$.

The second issue to keep in mind is that for the objective (7.4.36) to represent a sensible criterion for learning $f$, it is necessary that $f$ itself is parameterized in a specific way, and/or that some additional structural enforcement is done on the representations $f(\boldsymbol{x})$. Indeed, if $f$ is completely unconstrained, there are simple trivial solutions to (7.4.36) that do not correspond to useful representations: for example, just take $f(\boldsymbol{x}) = \boldsymbol{x}$ to be the identity! We discussed principled techniques for structuring the representation space in Chapter 6 when we discussed autoencoding and closed loop transcription; many of these techniques can be brought to bear in our present joint embedding setting, and we will detail this further later in the section, when we move to practical implementation.

**Joint embedding with the Infomax principle.** Now, the question of *joint* representation of $(\boldsymbol{x}, \boldsymbol{y})$ remains. Applying the data processing inequality once more, we obtain for another encoder $g$ that produces a representation of $\boldsymbol{y}$ that

$$I(\boldsymbol{x}; \boldsymbol{y}) \geq I(f(\boldsymbol{x}); \boldsymbol{y}) \geq I(f(\boldsymbol{x}); g(\boldsymbol{y})).$$

The joint objective

$$\max_{f,g} I(f(\boldsymbol{x}); g(\boldsymbol{y})) \tag{7.4.37}$$

is then a well-founded consequence of the Infomax principle.

To design the encoders $f$ and $g$, recall the intuition that we obtained through our work in Example 7.6. There, we saw that in the case of a mixture-of-Gaussians model for $\boldsymbol{x}$, with the conditioning signal $y$ identifying which specific mixture component generates the observation, a flexible denoiser architecture could be constructed via a suitable embedding of the class label $y$ into the *same space* as the data $\boldsymbol{x}$. More precisely, such an embedding $\boldsymbol{u}_y \in \mathbb{R}^D$ allows, when correlated with the embeddings $\boldsymbol{U}_{\boldsymbol{x}}$ of $\boldsymbol{x} \in \mathbb{R}^D$ (generally a matrix) via $\boldsymbol{u}_y^\top \boldsymbol{U}_{\boldsymbol{x}}$, to 'pick out' only the embeddings relevant to the correct mixture component of the distribution. How exactly to design these layers for use in neural networks being trained on more complex nonlinear high-dimensional data is a challenging question that we will dig into in more detail below. But a necessary condition is clear: the embeddings $f$ and $g$ should map their inputs to a *common embedding space*, say $\mathbb{R}^d$, to facilitate downstream comparison when performing conditional sampling. This leads to the objective

$$\max_{f:\mathbb{R}^{D_{\boldsymbol{x}}} \to \mathbb{R}^d, g:\mathbb{R}^{D_{\boldsymbol{y}}} \to \mathbb{R}^d} I(f(\boldsymbol{x}); g(\boldsymbol{y})). \tag{7.4.38}$$

The next step is to instantiate a suitable computationally-convenient relaxation of the mutual information in (7.4.38) for use with large-scale datasets.

### Practical Prescription for Learning Joint Embeddings: CLIP

A concrete and highly influential realization of learning joint embeddings for paired data is the method of Contrastive Language-Image Pre-training (CLIP) [RKH+21b]. Take the simple case of image-text pairs $[(\boldsymbol{x}_i, \boldsymbol{y}_i)]_{i=1}^N$ sampled from a dataset $\mathcal{D}$, where $\boldsymbol{x}_i$ is an image and $\boldsymbol{y}_i$ is a text caption. CLIP trains two encoders—an image encoder $f_\theta(\boldsymbol{x})$ and a text encoder $g_\phi(\boldsymbol{y})$—to map paired samples into a shared $d$-dimensional unit sphere (i.e. $\|f_\theta(\boldsymbol{x}_i)\|_2 = 1$ and $\|g_\phi(\boldsymbol{y}_i)\|_2 = 1$). To achieve this, we can adopt a simple symmetric cross-entropy loss that pushes the cosine similarity of matched pairs $(f_\theta(\boldsymbol{x}_i), g_\phi(\boldsymbol{y}_i))$ closer together while repelling unmatched pairs:

$$\mathcal{L}_{\text{CLIP}} = -\frac{1}{N}\mathbb{E}_{[(\boldsymbol{x}_i,\boldsymbol{y}_i)]_{i=1}^N \sim \mathcal{D}}\left[\sum_{i=1}^N \log \frac{\exp\left(f_\theta(\boldsymbol{x}_i)^\top g_\phi(\boldsymbol{y}_i)\right)}{\sum_{j=1}^N \exp\left(f_\theta(\boldsymbol{x}_i)^\top g_\phi(\boldsymbol{y}_j)\right)}\right.$$

$$+\sum_{i=1}^{N}\log\frac{\exp\left(g_\phi(\boldsymbol{y}_i)^\top f_\theta(\boldsymbol{x}_i)\right)}{\sum_{j=1}^{N}\exp\left(g_\phi(\boldsymbol{y}_i)^\top f_\theta(\boldsymbol{x}_j)\right)}\Bigg], \quad (7.4.39)$$

From the perspective of mutual information maximization, CLIP is a tractable surrogate for the lower bound on the mutual information between the image and text representations:[12]

$$I\big(f_\theta(\boldsymbol{x}), g_\phi(\boldsymbol{y})\big) \geq -\mathcal{L}_{\text{CLIP}} + \log N. \quad (7.4.40)$$

Thus, CLIP provides a practical recipe for learning joint embeddings for paired data that increases the mutual information between the paired samples. Empirically, CLIP has been shown to be effective at learning joint embeddings for image-text pairs, and has been used in a variety of applications, including image captioning, image-text retrieval, image generation and beyond. We will see more specific examples in Chapter 8 (with key implementation details described in detail in Section 8.3).

#### Application to Caption-Conditioned Image Generation

Finally, we will describe an end-to-end system that integrates joint image-text encoders, learned with the CLIP loss as described above, with conditioning operators that are evocative of the mixture-of-Gaussians operator (7.4.32) we saw in Example 7.6. These techniques formed the basis for the original open-source Stable Diffusion implementation [RBL+22], where the embedding and subsequent conditioning is performed not on a class label, but a text prompt, which describes the desired image content (Figure 7.14). This section will give a high-level overview of StableDiffusion-style image generation. For full implementation details sufficient to re-implement such a system from scratch, we refer to Chapter 8: specifically, Section 8.11 describes the process of encoding strings of text to a sequence of integer IDs; Section 8.3 describes the training of the necessary image-text encoders; and Sections 8.6 and 8.7 describe the training of the (latent) diffusion model.

Stable Diffusion follows the conditional generation methodology we outline in Section 7.4.1, with two key modifications: (i) The conditioning signal is a tokenized text prompt $\boldsymbol{Y} \in \mathbb{R}^{D_{\text{text}} \times N}$, rather than a class label; (ii) Image denoising is performed in "latent" space rather than on raw pixels, using a specialized, pretrained variational autoencoder pair $f : \mathbb{R}^{D_{\text{img}}} \to \mathbb{R}^{d_{\text{img}}}$, $g : \mathbb{R}^{d_{\text{img}}} \to \mathbb{R}^{D_{\text{img}}}$ (see Sections 6.1.4 and 8.6), where $f$ is the encoder and $g$ is the decoder. For issue (i), in the context of the iterative conditional denoising framework we have developed in Section 7.4.1, this concerns the parameterization of the denoisers $\bar{\boldsymbol{z}}_\theta(t, \boldsymbol{z}_t, \boldsymbol{Y}^+)$.[13] Rombach et al. [RBL+22] implement text conditioning in the denoiser using a layer known as cross attention, inspired by the original encoder-decoder transformer architecture of Vaswani et al. [VSP+17b]. Cross attention is implemented as follows. We let $\tau : \mathbb{R}^{D_{\text{text}} \times N} \to \mathbb{R}^{d_{\text{model}} \times N_{\text{text}}}$ denote an encoding network for the text embeddings (often a causal transformer—see Section 8.11), and let $\psi : \mathbb{R}^{d_{\text{img}}} \to \mathbb{R}^{d_{\text{model}} \times N_{\text{img}}}$ denote the mapping corresponding to one of the intermediate representations in the denoiser.[14] Here, $N_{\text{text}}$ is the maximum tokenized text prompt length, and $N_{\text{img}}$ roughly corresponds to the number of image channels (layer-dependent) in the representation, which is fixed if the input image resolution is fixed. Cross attention (with $K$ heads, and no bias) is defined as

$$\operatorname{MHCA}(\boldsymbol{z}_t, \boldsymbol{Y}^+) = \boldsymbol{U}_{\text{out}} \begin{bmatrix} \operatorname{SA}([\boldsymbol{U}_{\text{qry}}^1]^\top \psi(\boldsymbol{z}_t), [\boldsymbol{U}_{\text{key}}^1]^\top \tau(\boldsymbol{Y}^+), [\boldsymbol{U}_{\text{val}}^1]^\top \tau(\boldsymbol{Y}^+)) \\ \vdots \\ \operatorname{SA}([\boldsymbol{U}_{\text{qry}}^K]^\top \psi(\boldsymbol{z}_t), [\boldsymbol{U}_{\text{key}}^K]^\top \tau(\boldsymbol{Y}^+), [\boldsymbol{U}_{\text{val}}^K]^\top \tau(\boldsymbol{Y}^+)) \end{bmatrix}, \tag{7.4.41}$$

where SA denotes the scaled dot-product attention operation in the transformer (which we recall in detail in Chapter 8: see Equations (8.2.17) and (8.2.18)), and $\boldsymbol{U}_*^k \in \mathbb{R}^{d_{\text{model}} \times d_{\text{attn}}}$ for $* \in \{\text{qry}, \text{key}, \text{val}\}$ (as well as the output projection $\boldsymbol{U}_{\text{out}}$) are the learnable parameters of the layer.

[12] This result holds under the assumption that the pairs $(\boldsymbol{x}_i, \boldsymbol{y}_i)$ are i.i.d. within each batch of $N$ [OLC+25, Lemma 1]. The tightness of the bound improves as $N$ increases [OLV18]. However, for certain data distributions, it can remain loose for all $N$ [MS19; WI20].

[13] In the setting of text conditioning, the 'augmented' label $\boldsymbol{Y}^+$, which is either the encoded text prompt or $\varnothing$, denoting unconditional denoising, is often implemented by mapping $\varnothing$ to the empty string "", then encoding this text prompt with the tokenizer as usual. This gives a simple, unified way to treat conditional and unconditional denoising with text conditioning.

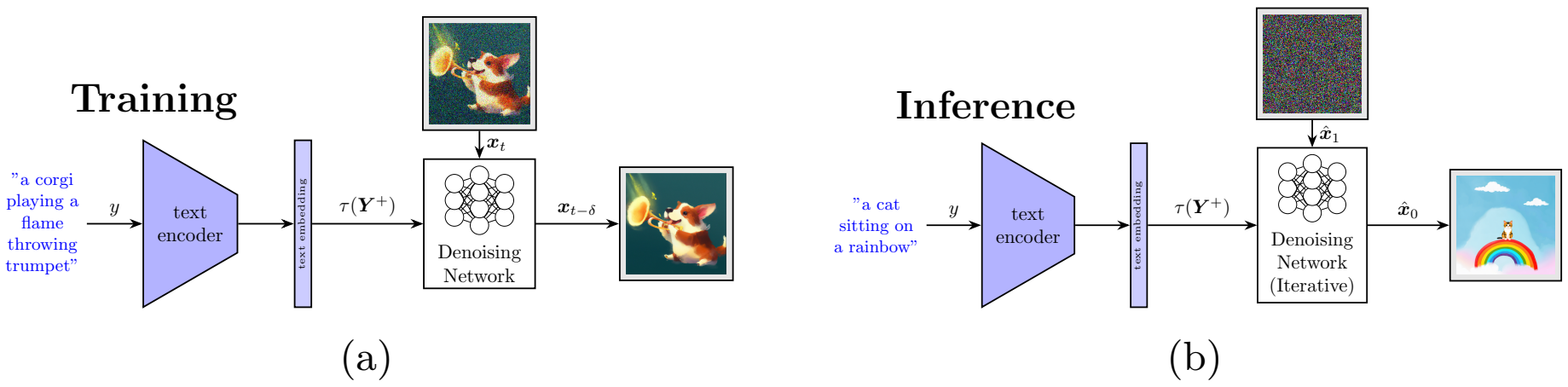

Figure 7.14: A high-level schematic of training and applying a text-to-image generative model, via conditional generation with a text prompt. **Left:** To train a text-to-image model, a large dataset of images paired with corresponding text captions is used. An encoder is used to map the captions to sequences of vectors, which are used as conditioning signals for a conditional denoiser, trained as described algorithmically in Section 7.4.1 (see Section 8.6 for implementation details). The text encoder may be pretrained and frozen, or jointly trained with the denoiser. **Right:** When applying a trained model, a desired text prompt is used as conditioning, then sampling is performed with the trained model, as in Algorithm 7.2 (*mutatis mutandis* for use with an encoded text prompt).

Notice that, by the definition of the self-attention operation, cross attention *outputs linear combinations of the value-projected text embeddings weighted by correlations between the image features and the text embeddings.* In the denoiser architecture used by Rombach et al. [RBL+22], self-attention residual blocks in the denoiser architecture, applied to the image representation at the current layer and defined analogously to those in Equation (8.2.15) for the vision transformer, are followed by cross attention residual blocks of the form (7.4.41). Such a structure requires the text encoder $\tau$ to, in a certain sense, share some structure in its output with the image feature embedding $\psi$: this can be achieved via joint embedding, learned by a procedure such as CLIP as described above. Conceptually, this joint text-image embedding space and the cross attention layer itself bear a strong resemblance to the conditional mixture of Gaussians denoiser that we derived in the previous section (recall (7.4.28)), in the special case of a single token sequence. Deeper connections can be drawn in the multi-token setting following the rate reduction framework for deriving deep network architectures discussed in Chapter 5, and manifested in the derivation of the CRATE transformer-like architecture.

[14] In practice, text-conditioned denoisers add cross attention layers at regular intervals within the forward pass of the denoiser, so $\psi$ should be seen as layer-dependent, in contrast to $\tau$.

This same basic design has been further scaled to even larger model and dataset sizes, in particular in modern instantiations of Stable Diffusion [EKB+24], as well as in competing models such as FLUX.1 [LBB+25], Imagen [SCS+22], and DALL-E [RDN+22]. The conditioning mechanism of cross attention has also become ubiquitous in other applications, as in EgoAllo (Section 8.10) for conditioned pose generation and in Michelangelo (Section 8.8) for conditional 3D shape generation based on images or texts.

## 7.5 Conditional Inference with Measurement Self-Consistency

In this last section, we consider the more extreme, but actually ubiquitous, case for distribution learning in which we only have a set of observed samples $\boldsymbol{Y} = \{\boldsymbol{y}_1, \ldots, \boldsymbol{y}_N\}$ of the data $\boldsymbol{x}$, but no samples of $\boldsymbol{x}$ directly! In general, the observation $\boldsymbol{y} \in \mathbb{R}^d$ is of lower dimension than $\boldsymbol{x} \in \mathbb{R}^D$. To make the problem well-defined, we do assume that the observation model between $\boldsymbol{y}$ and $\boldsymbol{x}$ is known to belong to a certain family of analytical models, denoted as $\boldsymbol{y} = h(\boldsymbol{x}, \theta) + \boldsymbol{w}$, with $\theta$ either known or not known.

Let us first try to understand the problem conceptually with the simple case when the measurement function $h$ is known and the observed $\boldsymbol{y} = h(\boldsymbol{x}) + \boldsymbol{w}$ is informative about $\boldsymbol{x}$. That is, we assume that $h$ is surjective from the space of $\boldsymbol{x}$ to that of $\boldsymbol{y}$ and the support of the distribution $\boldsymbol{y}_0 = h(\boldsymbol{x}_0)$ is low-dimensional. This typically requires that the *extrinsic* dimension $d$ of $\boldsymbol{y}$ is higher than the *intrinsic* dimension of the support of the distribution of $\boldsymbol{x}$. Without loss of generality, we may assume that there exist functions:

$$F(\boldsymbol{x}) = \mathbf{0}, \quad G(\boldsymbol{y}) = \mathbf{0}. \tag{7.5.1}$$

Notice that here we may assume that we know $G(\boldsymbol{y})$ but not $F(\boldsymbol{x})$. Let $\mathcal{S}_{\boldsymbol{y}} \doteq \{\boldsymbol{y} \mid G(\boldsymbol{y}) = \mathbf{0}\}$ be the support of $p(\boldsymbol{y})$. In general, $h^{-1}(\mathcal{S}_{\boldsymbol{y}}) = \{\boldsymbol{x} \mid G(h(\boldsymbol{x})) = \mathbf{0}\}$ is a superset of $\mathcal{S}_{\boldsymbol{x}} \doteq \{\boldsymbol{x} \mid F(\boldsymbol{x}) = \mathbf{0}\}$. That is, we have $h(\mathcal{S}_{\boldsymbol{x}}) \subseteq \mathcal{S}_{\boldsymbol{y}}$.

### 7.5.1 Linear Measurement Models

First, for simplicity, let us consider that the measurement is a linear function of the data $\boldsymbol{x}$ of interest:

$$\boldsymbol{y} = \boldsymbol{A}\boldsymbol{x}. \tag{7.5.2}$$

Here the matrix $\boldsymbol{A} \in \mathbb{R}^{m \times n}$ is of full row rank and $m$ is typically smaller than $n$. We assume $\boldsymbol{A}$ is known for now. We are interested in how to learn the distribution of $\boldsymbol{x}$ from such measurements. Since we no longer have direct samples of $\boldsymbol{x}$, we wonder whether we can still develop a denoiser for $\boldsymbol{x}$ with observations $\boldsymbol{y}$. Let us consider the following diffusion process:

$$\boldsymbol{y}_t = \boldsymbol{y}_0 + t\boldsymbol{g}, \quad \boldsymbol{y}_0 = \boldsymbol{A}(\boldsymbol{x}_0), \tag{7.5.3}$$

where $\boldsymbol{g} \sim \mathcal{N}(\boldsymbol{0}, \boldsymbol{I})$.

Without loss of generality, we assume $\boldsymbol{A}$ is of full row rank, i.e., underdetermined. Let us define the corresponding process $\boldsymbol{x}_t$ as one that satisfies:

$$\boldsymbol{y}_t = \boldsymbol{A}\boldsymbol{x}_t. \tag{7.5.4}$$

From the denoising process of $\boldsymbol{y}_t$, we have

$$\boldsymbol{y}_{t-s} \approx \boldsymbol{y}_t + st\nabla \log p_t(\boldsymbol{y}_t). \tag{7.5.5}$$

Then we have:

$$\boldsymbol{A}\boldsymbol{x}_{t-s} \approx \boldsymbol{A}\boldsymbol{x}_t + st\nabla \log p_t(\boldsymbol{A}\boldsymbol{x}_t), \tag{7.5.6}$$

for a small $s > 0$. So $\boldsymbol{x}_{t-s}$ and $\boldsymbol{x}_t$ need to satisfy:

$$\boldsymbol{A}(\boldsymbol{x}_{t-s} - \boldsymbol{x}_t) \approx st\nabla \log p_t(\boldsymbol{A}\boldsymbol{x}_t). \tag{7.5.7}$$

Among all $\boldsymbol{x}_{t-s}$ that satisfy the above constraint, we arbitrarily choose the one that minimizes the distance $\|\boldsymbol{x}_{t-s} - \boldsymbol{x}_t\|_2^2$. Therefore, we obtain a "denoising" process for $\boldsymbol{x}_t$:

$$\boldsymbol{x}_{t-s} \approx \boldsymbol{x}_t + st\boldsymbol{A}^{\dagger}\nabla \log p_t(\boldsymbol{A}\boldsymbol{x}_t). \tag{7.5.8}$$

Notice that this process does not sample from the distribution of $\boldsymbol{x}_t$. In particular, there are components of $\boldsymbol{x}$ in the null space/kernel of $\boldsymbol{A}$ that can never be recovered from observations. Thus more information is needed to recover the full distribution of $\boldsymbol{x}$, strictly speaking. But this recovers the component of $\boldsymbol{x}$ that is orthogonal to the null space of $\boldsymbol{A}$.

### 7.5.2 Learning 3D World Model from 2D Images

In practice, the measurement model is often nonlinear or only partially known. A typical problem of this kind is actually behind how we can learn a working model of the external world from the images perceived, say through our eyes, telescopes or microscopes. In particular, humans and animals are able to build a model of the 3D world (or 4D for a dynamical world) through a sequence of its 2D projections—a sequence of 2D images (or stereo image pairs). The mathematical or geometric model of the projection is generally known:

$$\boldsymbol{y}^i = h(\boldsymbol{x}, \theta^i) + \boldsymbol{w}^i, \tag{7.5.9}$$

where $h(\cdot)$ represents a (perspective) projection of the 3D (or 4D) scene from a certain camera view at time $t_i$ to a 2D image (or a stereo pair) and $\boldsymbol{w}$ is some possibly additive small measurement noise. Figure 7.15 illustrates this relationship concretely, while Figure 7.16 illustrates the model problem in the abstract. A full exposition of geometry related to multiple 2D views of a 3D scene is beyond the scope of this book. Interested readers may refer to the book [MKS+04]. For now, all we need to proceed is that such projections are well understood and multiple images of a scene contain sufficient information about the scene.

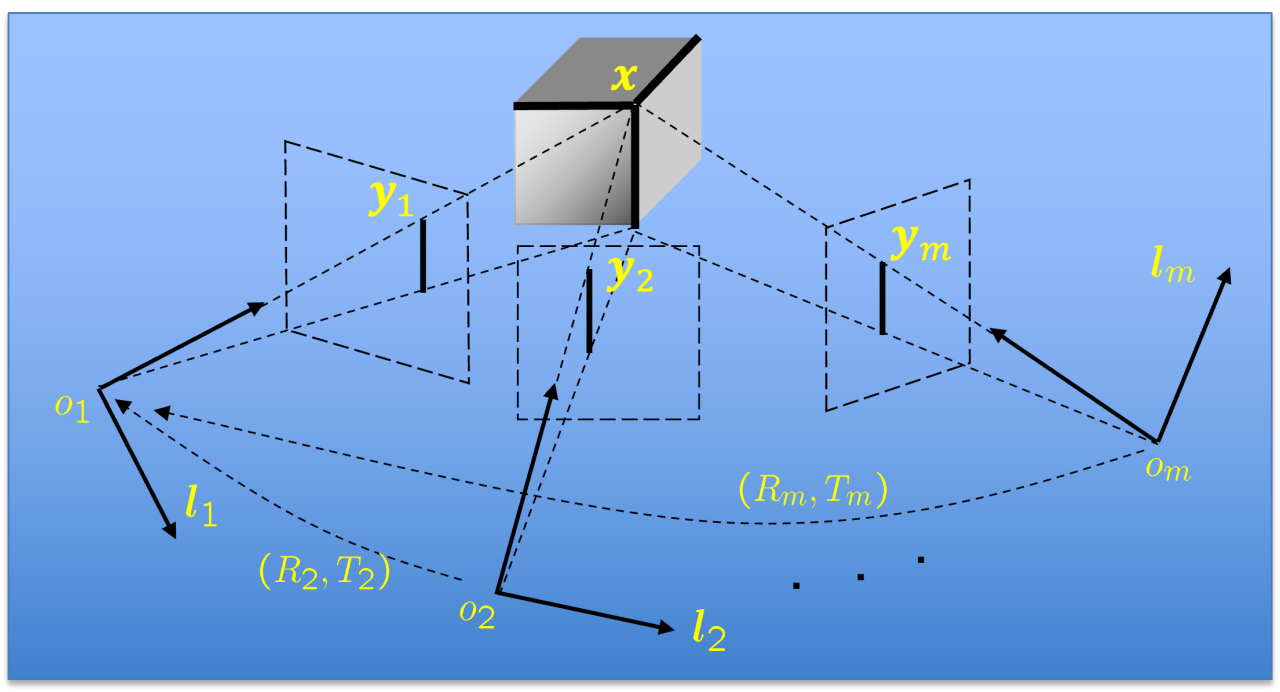

Figure 7.15: Relationship between a 3D object/scene and its 2D projections. Here we illustrate the projection of a point $\boldsymbol{x}$ and a line intersecting the point.

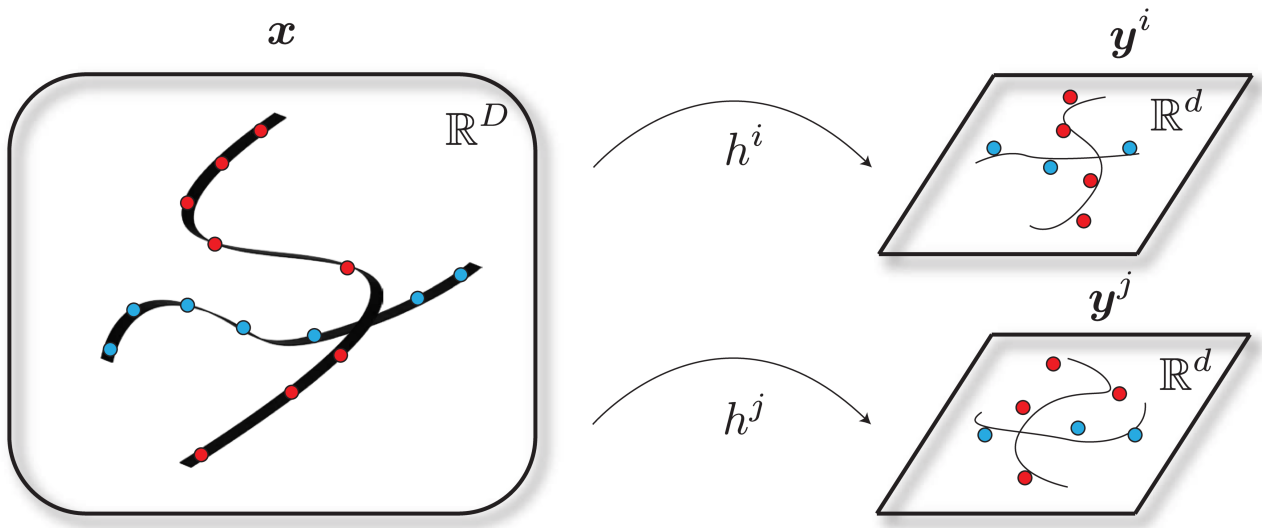

Figure 7.16: **Inference with distributed measurements.** We have a low-dimensional distribution $\boldsymbol{x}$ (here, similarly to Figure 7.1, depicted as a union of two 2-dimensional manifolds in $\mathbb{R}^3$) and a measurement model $\boldsymbol{y}^i = h^i(\boldsymbol{x}) + \boldsymbol{w}^i$. As before, we want to infer various properties of the conditional distribution of $\boldsymbol{x}$ given $\boldsymbol{y}$, where $\boldsymbol{y}$ is the collection of all the measurements $\boldsymbol{y}^i$.

In general, we would like to learn the distribution $p(\boldsymbol{x})$ of the 3D (or 4D)

world scene $\boldsymbol{x}$[15] from the perceived 2D images of the world so far. The primary function of such a (visual) world model is to allow us to recognize places where we had been before or predict what the current scene would look like in a future time at a new viewpoint.

Let us first examine the special but important case of stereo vision. In this case, we have two calibrated views of the 3D scene $\boldsymbol{x}$:

$$\boldsymbol{y}^0 = h(\boldsymbol{x}, \theta^0) + \boldsymbol{w}^0, \quad \boldsymbol{y}^1 = h(\boldsymbol{x}, \theta^1) + \boldsymbol{w}^1, \tag{7.5.10}$$

where parameters $\theta^0$ and $\theta^1$ for the view poses can be assumed to be known. $\boldsymbol{y}^0$ and $\boldsymbol{y}^1$ are two 2D-projections of the 3D scene $\boldsymbol{x}$. We may also assume that they have the same marginal distribution $p(\boldsymbol{y})$ and we have learned a diffusion and denoising model for it. That is, we know the denoiser:

$$\mathbb{E}[\boldsymbol{y} \mid \boldsymbol{y}_t = \boldsymbol{\nu}] = \boldsymbol{\nu} + t^2 \nabla_{\boldsymbol{\nu}} \log p_t(\boldsymbol{\nu}). \tag{7.5.11}$$

Or, furthermore, we may assume that we have a sufficient number of samples of stereo pairs $(\boldsymbol{y}^0, \boldsymbol{y}^1)$ and have also learned the joint distribution of the pairs. By a little abuse of notation, we also use $\boldsymbol{y} = h(\boldsymbol{x})$ to indicate the pair $\boldsymbol{y} = (\boldsymbol{y}^0, \boldsymbol{y}^1)$ and $p(\boldsymbol{y})$ as the learned probability distribution of the pair (say via a denoiser as above).

The main question now is: How to learn (a representation for) the distribution of the 3D scene $\boldsymbol{x}$ from its two projections with known relationships? People might question the rationale for doing this: why is this necessary if the function $h(\cdot)$ is largely invertible? That is, the observation $\boldsymbol{y}$ can largely determine the unknown $\boldsymbol{x}$, which is kind of the case for stereo—in general, two (calibrated) images contain sufficient information about the scene depth, from the given vantage point. However, 2D images are far from the most compact representation of the 3D scene as the same scene can produce infinitely many (highly correlated) 2D images or image pairs. In fact, a good representation of a 3D scene should be invariant to the viewpoint. Hence, a correct representation of the distribution of 3D scenes should be much more compact and structured than the distribution of 2D images, stereo pairs, or image-depth pairs.

Consider the (inverse) denoising process for the diffusion: $\boldsymbol{y}_t = \boldsymbol{y} + t\boldsymbol{g}$ in (7.5.11), where $\boldsymbol{g}$ is standard Gaussian. From the denoising process of (7.5.11), we have

$$\boldsymbol{y}_{t-s} = \boldsymbol{y}_t + st \nabla_{\boldsymbol{y}} \log p_t(\boldsymbol{y}_t). \tag{7.5.12}$$

We try to find a corresponding "denoising" process of $\boldsymbol{x}_t$ such that $\boldsymbol{x}$ is related to $\boldsymbol{y}$ as:

$$\boldsymbol{y} = h(\boldsymbol{x}). \tag{7.5.13}$$

Then we have:

$$h(\boldsymbol{x}_{t-s}) \approx h(\boldsymbol{x}_t) + st \nabla_{\boldsymbol{y}} \log p_t(h(\boldsymbol{x}_t)), \tag{7.5.14}$$

[15]Here by abuse of notation, we use $\boldsymbol{x}$ to represent either a point in 3D or a sample of an entire 3D object or a scene that consists of many points.

for a small $s > 0$. Suppose $\boldsymbol{x}_{t-s} = \boldsymbol{x}_t + s\boldsymbol{v}$ for some vector $\boldsymbol{v}$ and small increment $s$. We have

$$h(\boldsymbol{x}_{t-s}) \approx h(\boldsymbol{x}_t) + \frac{\partial h}{\partial \boldsymbol{x}}(\boldsymbol{x}_t) \cdot \boldsymbol{v}s \doteq h(\boldsymbol{x}_t) + \boldsymbol{A}(\boldsymbol{x}_t)\boldsymbol{v}s. \tag{7.5.15}$$

Hence, we have

$$\boldsymbol{A}(\boldsymbol{x}_t)\boldsymbol{v} = t\nabla_{\boldsymbol{y}} \log p_t(h(\boldsymbol{x}_t)). \tag{7.5.16}$$

Geometrically the vector $\boldsymbol{v}$ in the domain of $\boldsymbol{x}$ can be viewed as the pullback of the vector field $t\nabla \log p_t(\boldsymbol{y})$ under the map $\boldsymbol{y} = h(\boldsymbol{x})$. In general, as before, we may (arbitrarily) choose $\boldsymbol{v}$ to be the minimum 2-norm vector that satisfies the pullback relationship. Hence, we can express $\hat{\boldsymbol{x}}_{t-s}$ approximately as:

$$\hat{\boldsymbol{x}}_{t-s} \approx \boldsymbol{x}_t + st\boldsymbol{A}(\boldsymbol{x}_t)^{\dagger}\nabla_{\boldsymbol{y}} \log p_t(h(\boldsymbol{x}_t)). \tag{7.5.17}$$

*Remark* 7.2 (Parallel Sensing and Distributed Denoising.). There is something very interesting about the above equation (7.5.17). It seems to suggest we could try to learn the distribution of $\boldsymbol{x}$ through a process that is coupled with (many of) its (partial) observations:

$$\boldsymbol{y}^i = h^i(\boldsymbol{x}) + \boldsymbol{w}^i, i = 1, \ldots, K. \tag{7.5.18}$$

In this case, we obtain a set of equations that the vector field $\boldsymbol{v}$ in the domain of $\boldsymbol{x}$ should satisfy:

$$\boldsymbol{A}^i(\boldsymbol{x}_t)\boldsymbol{v} = t\nabla_{\boldsymbol{y}^i} \log p_t(h^i(\boldsymbol{x}_t)), \tag{7.5.19}$$

where $\boldsymbol{A}^i(\boldsymbol{x}_t) = \frac{\partial h^i}{\partial \boldsymbol{x}}(\boldsymbol{x}_t)$. The final $\boldsymbol{v}$ can be chosen as a "centralized" solution that satisfies all the above equations, or it could be chosen as a certain (stochastically) "aggregated" version of all $\boldsymbol{v}^i$:

$$\boldsymbol{v}^i = t\boldsymbol{A}^i(\boldsymbol{x}_t)^{\dagger}\left[\nabla_{\boldsymbol{y}^i} \log p_t(h^i(\boldsymbol{x}_t))\right], \quad i = 1, \ldots, K, \tag{7.5.20}$$

that are computed in a parallel and distributed fashion? An open question here is exactly what the so-defined "denoising" process for $\boldsymbol{x}_t$ converges to, even in the linear measurement model case. When would it converge to a distribution that has the same low-dimensional support as the original $\boldsymbol{x}_0$, as $\boldsymbol{y}_t$ converges to $\boldsymbol{y} = h(\boldsymbol{x}_0)$?

**Visual World Model from Uncalibrated Image Sequences.** In the above derivation, we have assumed that the measurement model $h(\cdot)$ is fully known. In the case of stereo vision, this is rather reasonable as the relative pose (and calibration) of the two camera views (or two eyes[16]) is usually known in advance. Hence, through the stereo image pairs, in principle we should be able to learn the distribution of 3D scenes, at least the ego-centric distribution of 3D scenes. However, the low-dimensional structures of the so-called learned distribution contain variation caused by changing the viewpoints. That is, the appearance of the stereo images varies when we change our viewpoints with respect to the same 3D scene. For many practical vision tasks (such as localization and navigation), it is important that we can decouple this variation of viewpoints from an invariant representation of (the distribution of) 3D scenes.

[16]The relative pose of our two eyes is well known to our brain.

*Remark* 7.3. Note that the above goal aligns well with Klein's Erlangen Program for modern geometry, which is to study invariants of a manifold under a group of transformations. Here, we may view the manifold of interest as the distribution of ego-centric representations of 3D scenes. We have learned that it admits a group of three-dimensional rigid-body motion acting on it. It is remarkable that our brain has learned to effectively decouple such transformations from the observed 3D world.

Notice that we have studied learning representations that are invariant to translation and rotation in a limited setting in Chapter 5. We know that the associated compression operators take the necessary form of (multi-channel) convolutions, hence leading to the (deep) convolutional neural networks. Nevertheless, operators that are associated with compression or denoising that are invariant to more general transformation groups remain elusive to characterize [CW16b]. For the 3D Vision problem in its most general setting, we know the change of our viewpoints can be well modeled as a rigid-body motion. However, the exact relative motion of our eyes between different viewpoints is usually not known. More generally, there could also be objects (e.g., cars, humans, hands) moving in the scene and we normally do not know their motion either. How can we generalize the problem of learning the distribution of 3D scenes with calibrated stereo pairs to such more general settings? More precisely, we want to learn a compact representation $\boldsymbol{x}$ of the 3D scenes that is invariant to the camera/eye motions. Once such a representation is learned, we could sample and generate a 3D scene and render images or stereo pairs from arbitrary poses.

To this end, note that we can model a sequence of stereo pairs as:

$$\boldsymbol{y}^k = h(\boldsymbol{x}^k, \theta^k), \quad k = 1, \dots, K, \tag{7.5.21}$$

where $h(\cdot)$ represents the projection map from 3D to 2D. $\theta^k$ denotes the rigid-body motion parameters of the $k$th view, with respect to some canonical frame in the world. $\boldsymbol{x}^k$ represents the 3D scene at time $k$. If the scene is static, $\boldsymbol{x}^k$ should all be the same $\boldsymbol{x}^k = \boldsymbol{x}$. To simplify the notation, we may denote the set of $k$ equations as one:

$$\boldsymbol{Y} = H(\boldsymbol{x}, \Theta). \tag{7.5.22}$$

We may assume that we are given many samples of such stereo image sequences $\{\boldsymbol{Y}_i\}$. The problem is how to recover the associated motion sequence $\{\Theta_i\}$ and learn the distribution of the scene $\boldsymbol{x}$ (that is invariant to the motion). To the best of our knowledge, this remains an open challenging problem, probably as the final frontier for the 3D Vision problem.

## 7.6 Summary and Notes

**Measurement matching without clean samples.** In our development of conditional sampling, we considered measurement matching under an observation model (7.3.9), where we assume that we have paired data $(\boldsymbol{x}, \boldsymbol{y})$—i.e., ground truth for each observation $\boldsymbol{y}$. In many practically relevant inverse problems, this is not the case: one of the most fundamental examples is in the context of compressed sensing, which we recalled in Chapter 2, where we need to reconstruct $\boldsymbol{x}$ from $\boldsymbol{y}$ using prior knowledge about $\boldsymbol{x}$ (i.e., sparsity). In the setting of denoising-diffusion, we have access to an implicit prior for $\boldsymbol{x}$ via the learned denoisers $\bar{\boldsymbol{x}}_\theta(t, \boldsymbol{\xi})$. Can we still perform conditional sampling without access to ground truth samples $\boldsymbol{x}$?

For intuition as to why this might be possible, we recall a classical example from statistics known as Stein's unbiased risk estimator (SURE). Under an observation model $\boldsymbol{x}_t = \boldsymbol{x} + t\boldsymbol{g}$ with $\boldsymbol{g} \sim \mathcal{N}(\mathbf{0}, \boldsymbol{I})$ and $t > 0$, it turns out that for any weakly differentiable $f : \mathbb{R}^D \to \mathbb{R}^D$,

$$\mathbb{E}_{\boldsymbol{g}}\left[\|\boldsymbol{x} - f(\boldsymbol{x} + t\boldsymbol{g})\|_2^2\right] = \mathbb{E}_{\boldsymbol{g}}\left[\|\boldsymbol{x} + t\boldsymbol{g} - f(\boldsymbol{x} + t\boldsymbol{g})\|_2^2 + 2t^2 \nabla \cdot f(\boldsymbol{x} + t\boldsymbol{g})\right] - t^2 D, \tag{7.6.1}$$

where $\nabla\cdot$ denotes the divergence operator:

$$\nabla \cdot f = \sum_{i=1}^{D} \partial_i f_i.$$

The $\boldsymbol{x}$-dependent part of the RHS of Equation (7.6.1) is called Stein's unbiased risk estimator (SURE). If we take expectations over $\boldsymbol{x}$ in Equation (7.6.1), note that the RHS can be written as an expectation with respect to $\boldsymbol{x}_t$—in particular, the mean-squared error of *any denoiser* $f$ can be estimated *solely from noisy samples*! This remarkable fact, in refined forms, constitutes the basis for many practical techniques for performing image restoration, denoising-diffusion, etc. using only noisy data: notable examples include the "noise2noise" paradigm [LMH+18] and Ambient Diffusion [DSD+23a].

As a fun aside, we point out that Equation (7.6.1) leads to an alternate proof of Tweedie's formula (Theorem 3.2). At a high level, one takes expectations over $\boldsymbol{x}$ and expresses the main part of the RHS of Equation (7.6.1) equivalently, via integration by parts, as

$$\begin{aligned} & \mathbb{E}_{\boldsymbol{x}_t}\left[\|\boldsymbol{x}_t - f(\boldsymbol{x}_t)\|_2^2 + 2t^2 \nabla \cdot f(\boldsymbol{x}_t)\right] \\ = \; & \mathbb{E}_{\boldsymbol{x}_t}\left[\|\boldsymbol{x}_t - f(\boldsymbol{x}_t)\|_2^2\right] - 2t^2 \int \langle \nabla p_{\boldsymbol{x}_t}(\boldsymbol{\xi}), f(\boldsymbol{\xi}) \rangle \, \mathrm{d}\boldsymbol{\xi}. \end{aligned} \tag{7.6.2}$$

This is a quadratic function of $f$, and formally taking derivatives gives that the optimal $f$ satisfies Tweedie's formula (Theorem 3.2). This argument can be made rigorous using basic ideas from the calculus of variations.

**Corrections to the Diffusion Posterior Sampling (DPS) approximation.** In Example 7.4 and in particular in Figure 7.11, we pointed out a limitation of the DPS approximation Equation (7.3.26) at small levels of measurement noise. This limitation is well-understood, and a principled approach to ameliorating it has been proposed by Rozet et al. [RAL+24]. The approach involves incorporating an additional estimate for the variance of the noisy posterior $p_{\boldsymbol{x}|\boldsymbol{x}_t}$ to Equation (7.3.26)—we refer to the paper for details. Natural estimates for the posterior variance are slightly less scalable than DPS itself due to the need to invert an affine transformation of the Jacobian of the posterior denoiser $\mathbb{E}[\boldsymbol{x} \mid \boldsymbol{x}_t = \boldsymbol{\xi}]$ (a large matrix). This is done relatively efficiently by Rozet et al. [RAL+24] using automatic differentiation and an approximation for the inverse based on conjugate gradients. It seems that it should be possible to improve further over this approach (say, using classical ideas from second-order optimization).

**More about measurement matching and diffusion models for inverse problems.** Diffusion models have become an extremely popular tool for solving inverse problems arising in scientific applications. Many more methods beyond the simple DPS algorithm we have presented in Algorithm 7.1 have been developed and continue to be developed, as the area is evolving rapidly. Popular and performant classes of approaches beyond DPS, which we have presented due to its generality, include variable splitting approaches like DAPS [ZCB+24], which allow for specific measurement constraints to be enforced much more strongly than in DPS, and exact approaches that can avoid the use of approximations as in DPS, such as TDS [WTN+23]. For more on this area, we recommend [ZCZ+25], which functions simultaneously as a survey and a benchmark of several popular methods on specific scientific inverse problem datasets.

**History of the Infomax principle.** The Infomax principle [Lin88] that we introduced in our discussion of joint embedding learning and CLIP in Section 7.4.2 is a well-known and important foundation for much of modern research into unsupervised representation learning. For further reading, we refer to [OLV18], which our conceptual discussion in Section 7.4.2 expands upon, and [OLC+25], which details further important theoretical issues associated with optimizing Infomax-type objectives from finite samples.

## 7.7 Exercises and Extensions

*Exercise* 7.1 (Posterior Variance Correction to DPS). 1. Using the code provided in the book GitHub for implementing Figure 7.11, implement the posterior variance correction proposed by Rozet et al. [RAL+24].

2. Verify that it ameliorates the posterior collapse at low noise variance issue observed in Figure 7.11.

3. Discuss any issues of sampling correctness that are retained or introduced by the corrected method, as well as its efficiency, relative to diffusion posterior sampling (DPS).

*Exercise* 7.2 (Conditional Sampling on MNIST). 1. Train a simple classifier for the MNIST dataset, using an architecture of your choice. Additionally train a denoiser suitable for use in conditional sampling (Algorithm 7.2, since this denoiser can be used for unconditional denoising as well).

2. Integrate the classifier into a conditional sampler based on classifier guidance, as described in the first part of Section 7.4.1. Evaluate the resulting samples in terms of faithfulness to the conditioning class (visually; in terms of nearest neighbor; in terms of the output of the classifier).

3. Integrate the classifier into a conditional sampler based on classifier-free guidance, as described in Section 7.4.1 and Algorithm 7.2. Perform the same evaluation as in the previous step, and compare the results.

4. Repeat the experiment on the CIFAR-10 dataset.

# Chapter 8

# Learning Representations for Real-World Data and Tasks

"*The best theory is inspired by practice, and the best practice is inspired by theory.*"

— Donald Knuth

## 8.1 Introduction

The previous chapters have presented a systematic introduction to mathematical problems, computational frameworks, and practical algorithms associated with learning low-dimensional distributions from high-dimensional data. Although most theoretical justifications of these methods have been established for idealistic models of data distributions such as (mixtures of) subspaces and/or Gaussians, the principles and ideas behind these computational methods are nevertheless powerful and general, and they are in fact meant to be applicable to real-world datasets and tasks.

To help readers understand the material in this book better and learn how to apply what you have learned so far to real-world data, here we provide some demonstrations and vignettes of several representative applications. Each application proposes a solution to a real-world task with a real-world dataset (such as visual data, motion data, and text data), using the methods we have introduced in this book. The results presented in this chapter are meant to serve the following two purposes:

- firstly, to provide additional experimental details and empirical evidence which validate the methods presented earlier in the book, and demonstrate their significant potential in real-world contexts;

- secondly, to introduce the reader to certain modern empirical methods and tasks in deep learning which are not well-documented outside of research or production codebases.

However, in our honest opinion, the solutions and results featured here are designed simply to verify that the methodology works. As such, there is great room for future improvement, in both theoretical understanding and practical engineering, to further advance the state-of-the-art performance. We will showcase some immediate and obvious improvements through this chapter and especially in the last Section 8.12. We will leave the discussions of potentially much more transformative ideas about representation learning and remaining significant open problems about intelligence in general for future research in the final Chapter 9.

### 8.1.1 Outline of the Chapter

For intelligence in nature, it is arguably true that the three most important types of data with which our brain builds our memory and knowledge about are *the visual data*, *our body motions*, and the *natural languages*. Hence, in this chapter, we will show how to apply principles and methods introduced in this book to learn good representations of these real-world data and show how such representations facilitate important practical tasks associated with these data.

Firstly, from Section 8.2 to 8.9, we will start with the visual data and show how to learn the distributions of 2D images to 3D objects since the visual memory serves as the low-level but foundational model for us (and animals) that stores knowledge of the physical world and helps perceive and predict in a new environment. We will then, in Section 8.10, show how to learn the distribution of body motions as it is crucial for us (and animals) to act or interact swiftly and accurately within the environments. Finally, in Section 8.11, we will show how to model and learn the distribution of natural languages.

Evidently, compared to the natural languages, our coverage of physical data, visual and motion, will be much more extensive. Although in the past few years, the AI industry has invested heavily in modeling natural languages[1], people have started to realize the importance of modeling knowledge about the physical world and our actions within. In most recent years, there have been a fastly growing interest and shifting attention to develop the so-called "world model," "spatial intelligence," "physical intelligence," or "embodied intelligence", as they are crucial for AI systems or agents, say an intelligent robot, which need to interact with an open physical world. As we have learned in Chapter 1, that is precisely the goal and scope of the *Cybernetics* program initiated by Norbert Wiener in the 1940s.

On the technical level, from Section 8.2 to Section 8.4, we will show how to use principles introduced in this book to learn good representations of imagery data in *unsupervised, weakly supervised*, and *supervised* settings, respectively.[2] This will serve as both an introduction to imagery data processing, data augmentation techniques, and a popular deep architecture Transformer, as a precursor to the implementation of the CRATE architecture later. These examples also give the first demonstration of the drastic kinds of simplifications one can already make using the principles introduced in the book. We will continue with modifications to the *transformer network architecture* in Section 8.4 for images and Section 8.11 or texts respectively, which demonstrate the capabilities of simplified white-box architectures, including CRATE and its variants, for *encoding* within the image and text domains. We will also demonstrate the capabilities of the simplified architectures for *autoencoding* in Section 8.5.

[1] Including their derivatives such as programming languages, logical deductions, and mathematical proofs.

Practically speaking, the primary reason why we want to learn a good representation of a data distribution is to allow us to sample the distribution (as a prior) to regenerate, estimate or predict the state of the world, conditioned on current observation or information given. We demonstrate the generative capabilities of the learned representations

- for 2D images in Section 8.6 and Section 8.7;
- for 3D objects in Section 8.8 and Section 8.9;
- for human motions in Section 8.10;
- and for natural languages in Section 8.11.

### 8.1.2 Overall Technical Setup

In previous chapters, we alluded to different setups in which we used representation-learning techniques to process real data at scale. In this chapter, we will describe such setups in great detail. The objective of this section is to get you, the reader, to be able to reproduce any experiment discussed in this section (or indeed the book) using just the description we will give in the book, the principles introduced in previous chapters and expanded on in this chapter, and hyperparameters taken from a smattering of papers whose results are discussed in this chapter. To this end, we will *precisely* describe all procedures in a detailed language, pseudocode, or mathematical notation that can be directly implemented in code. Wherever possible, we will discuss how the concrete implementations connect to the principles presented earlier in the book.

Let us define the set of possible data as $\mathcal{D}$ (eventually this will be the set of images $\mathcal{I}$, for example, or the set of text $\mathcal{T}$), and the set of finite sequences of tokens in $\mathbb{R}^d$ (i.e., the set of matrices with $d$ rows) as $(\mathbb{R}^d)^* \doteq \bigcup_{T=1}^{\infty} \mathbb{R}^{d \times T}$. In order to discuss the wealth of applications we introduce in this chapter, we first recall that in the rest of the book, we discuss two different types of model architectures.

[2]The goal in all three settings is to learn a good representation for the data distribution. They differ only in how much side information is given about the distribution that may facilitate the learning.

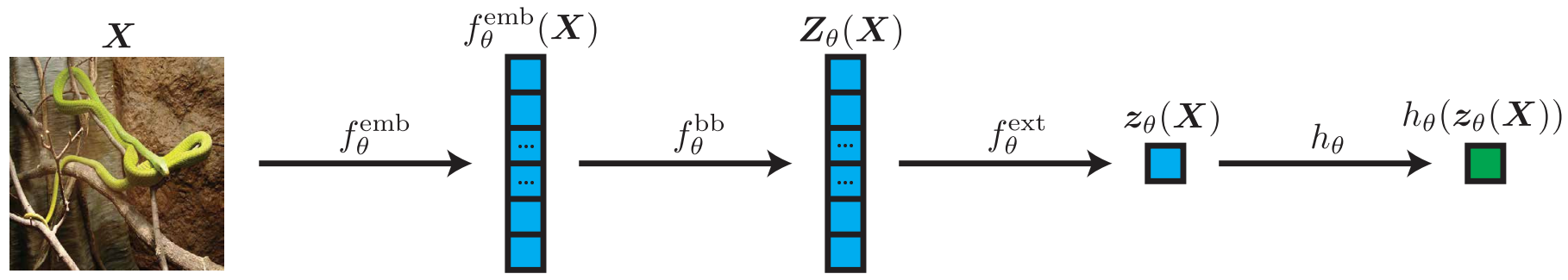

Figure 8.1: **A diagram of the encoder pipeline.** Data $\boldsymbol{X} \in \mathcal{D}$ is fed through the embedding $f_\theta^{\mathrm{emb}}$ to get a sequence in $(\mathbb{R}^d)^*$. The embedding is fed through a backbone $f_\theta^{\mathrm{bb}}$ to get features $\boldsymbol{Z}_\theta(\boldsymbol{X})$ for each token. We can extract an aggregate feature $\boldsymbol{z}_\theta(\boldsymbol{X})$ using the extraction map $f_\theta^{\mathrm{ext}}$. Finally, to use the aggregate feature in downstream tasks, we can use the task-specific head $h_\theta$.

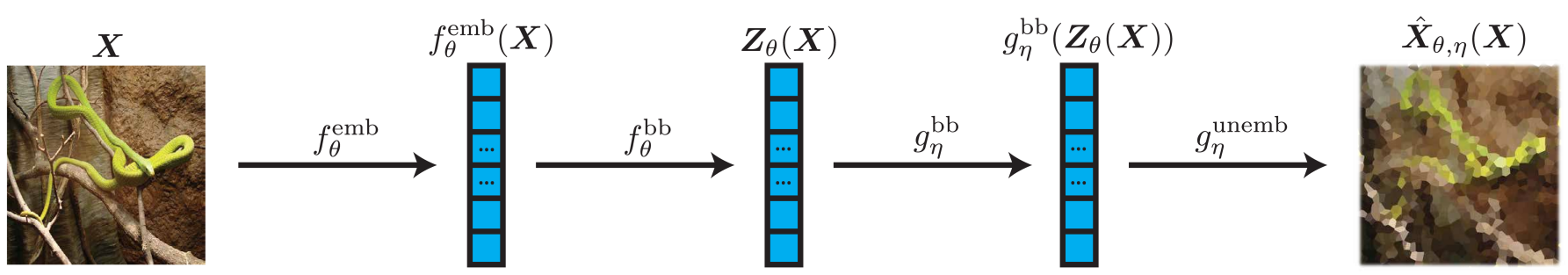

Figure 8.2: **A diagram of the autoencoder pipeline.** Data $\boldsymbol{X} \in \mathcal{D}$ is fed through the embedding $f_\theta^{\mathrm{emb}}$ to get a sequence in $(\mathbb{R}^d)^*$. The embedding is fed through an encoder backbone $f_\theta^{\mathrm{bb}}$ to get features $\boldsymbol{Z}_\theta(\boldsymbol{X})$ for each token. To decode $\boldsymbol{Z}_\theta(\boldsymbol{X})$, we pass it through a decoder backbone $g_\eta^{\mathrm{bb}}$. To map the decoder backbone output back to data space $\mathcal{D}$, we use an *unembedding* layer $g_\eta^{\mathrm{unemb}}$, overall obtaining a reconstruction $\hat{\boldsymbol{X}}_{\theta,\eta}(\boldsymbol{X})$ (here stylized to be a pixelated reconstruction of the input).

- An *encoder* architecture, parameterized by $\theta$, which is composed of several components:
  - An *embedding* $f_\theta^{\mathrm{emb}} \colon \mathcal{D} \to (\mathbb{R}^d)^*$, which converts the input data $\mathcal{D}$ into a series of *tokens* which are mapped into, or *embedded* in, $d$-dimensional space. *In the rest of the chapter, we will often identify tokens and embeddings with each other.*
  - An *encoder backbone* $f_\theta^{\mathrm{bb}} \colon (\mathbb{R}^d)^* \to (\mathbb{R}^d)^*$, which processes the series of embeddings using a sequence-to-sequence operation. This backbone is implemented by the network architectures discussed in the previous chapters, but we will give a more formal description as we go along.
  - An *aggregate feature extractor* $f_\theta^{\mathrm{ext}} \colon (\mathbb{R}^d)^* \to \mathbb{R}^d$, which extracts an aggregate representation of the whole sequence. This is used to define a single feature for the entire data sample.
  - A *task-specific head* $h_\theta \colon \mathbb{R}^d \to \mathbb{R}^m$, which extracts an $m$-dimensional output for prediction.

  We also define $f_\theta \doteq f_\theta^{\mathrm{bb}} \circ f_\theta^{\mathrm{emb}} \colon \mathcal{D} \to (\mathbb{R}^d)^*$. Given an input $\boldsymbol{X}$, we write $\boldsymbol{Z}_\theta(\boldsymbol{X}) \doteq f_\theta(\boldsymbol{X})$ and $\boldsymbol{z}_\theta(\boldsymbol{X}) \doteq f_\theta^{\mathrm{ext}}(\boldsymbol{Z}_\theta(\boldsymbol{X}))$. The overall pipeline is depicted in Figure 8.1.

- An *autoencoder* architecture, which is composed of several components:
  - An *embedding* $f_\theta^{\mathrm{emb}} \colon \mathcal{D} \to (\mathbb{R}^d)^*$, which converts the input data $\mathcal{D}$ into a series of *tokens* which are embedded in $d$-dimensional space.
  - An *encoder backbone* $f_\theta^{\mathrm{bb}} \colon (\mathbb{R}^d)^* \to (\mathbb{R}^d)^*$, which processes the series of embeddings using a sequence-to-sequence operation.
  - A *decoder backbone* $g_\eta^{\mathrm{bb}} \colon (\mathbb{R}^d)^* \to (\mathbb{R}^d)^*$, which conceptually undoes the operation of the encoder backbone.
  - An *unembedding* $g_\eta^{\mathrm{unemb}} \colon (\mathbb{R}^d)^* \to \mathcal{D}$, which acts as an inverse of the embedding.

  We also define $f_\theta \doteq f_\theta^{\mathrm{bb}} \circ f_\theta^{\mathrm{emb}} \colon \mathcal{D} \to (\mathbb{R}^d)^*$ and $g_\eta \doteq g_\eta^{\mathrm{unemb}} \circ g_\eta^{\mathrm{bb}} \colon (\mathbb{R}^d)^* \to \mathcal{D}$. Given an input $\boldsymbol{X}$, we write $\boldsymbol{Z}_\theta(\boldsymbol{X}) \doteq f_\theta(\boldsymbol{X})$ and $\hat{\boldsymbol{X}}_{\theta,\eta}(\boldsymbol{X}) \doteq g_\eta(\boldsymbol{Z}_\theta(\boldsymbol{X}))$. The overall pipeline is depicted in Figure 8.2.

We will repeatedly use this notation many times in this chapter, so please feel free to refer back to it if something doesn't make sense. This decomposition of our networks also closely mirrors most code implementations, and you can start your coding projects by defining these networks.

## 8.2 Unsupervised Learning of Image Representations

Learning high-quality and faithful representations of imagery data distributions is a fundamental problem in machine intelligence, as we have discussed in great length in Chapter 4. In the literature, there have been many popular approaches proposed for this task, many of which may appear not to use the techniques and principles outlined in this manuscript. One such approach is called *contrastive learning*, so named because the learning objective is (roughly speaking) about ensuring that features of "similar" data are similar, and features of "dissimilar" data are far apart. Contrastive learning solutions are often highly-engineered, empirically designed approaches. In this section, we will introduce one such popular approach known as DINO [CTM+21]. Moreover, though, we will demonstrate how to use the principles described in this book to drastically simplify its design decisions while improving the learned representations. This is known as the **SimDINO** [WZP+25]:

`https://robinwu218.github.io/SimDINO/`.

### 8.2.1 Data

The data that we will use to explore and simplify the DINO methodology are all two-dimensional (2D) image data.

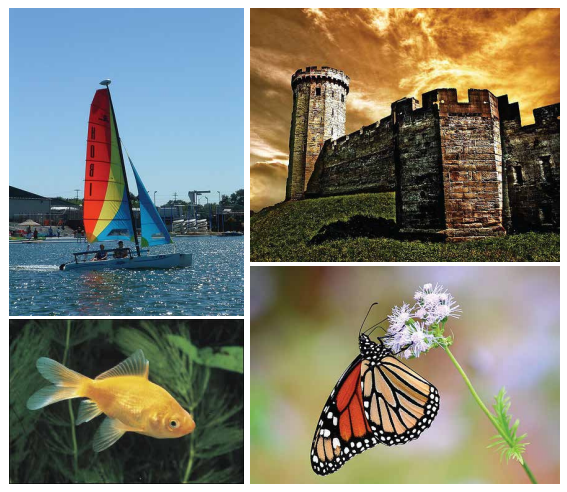

(a) ImageNet-1K samples.

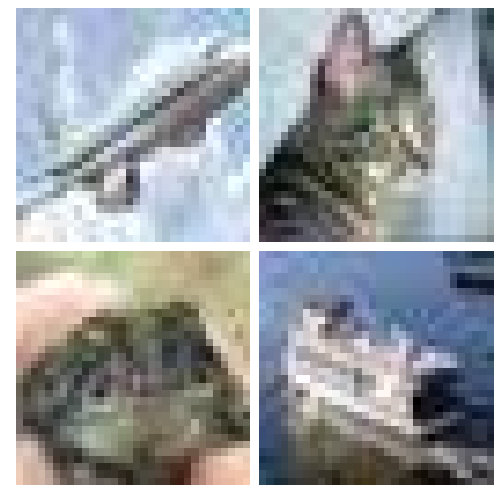

(b) CIFAR10 samples.

Figure 8.3: **Images from ImageNet-1K *(left)* and CIFAR-10 *(right)*.** Notice that the CIFAR-10 images are much lower resolution, generally speaking, reducing the complexity of learning that distribution.

For *training*, we will use the ImageNet-1K and ImageNet-21K datasets. Each sample in the dataset is an RGB image, of varying resolution, and a label indicating the object or scene that the image contains (i.e., the *class* of the image). The ImageNet-1K dataset contains 1.28M training images and 50K validation images partitioned into 1K classes. The ImageNet-21K dataset contains 14.2M training images and 21.8K classes, but the classes are not disjoint (i.e., some classes are subsets of others). Since we are doing self-supervised learning, the labels will not be used during training, only during evaluation. In order to increase the robustness of our model, we often apply *small data augmentations* to each image during processing, such as flips, added small random noise, or random large crops; we do not include this in our notation, as each augmentation of a natural image is itself (very close to) a natural image in our dataset.

For *evaluation*, we will use a litany of datasets. Of these, the most common is CIFAR-10. CIFAR-10 is a dataset of 60K RGB $32 \times 32$ natural images partitioned into 10 classes, with a pre-established training set of 50K samples and a validation set of 10K samples. The purpose of using CIFAR-10 is to ensure that models which train on one distribution of images (ImageNet) can generalize to another distribution of images (CIFAR-10). We also refer to other similar datasets, such as CIFAR-100 (disjoint from CIFAR-10), Oxford Flowers, and Oxford Pets. Exemplars of ImageNet-1K and CIFAR-10 data are shown in Figure 8.3.

On a slightly more formal level, our data $\boldsymbol{X}$ will be images; we let $\mathcal{I}$ be the set of all images. Since an image is a rectangular array of pixels, and each pixel has a color given by RGB, CMYK, or another color format, we say that an image is an element of $\mathbb{R}^{c\times h\times w}$ — here $c$ is the number of channels (i.e., 3 for RGB and 4 for CMYK), $h$ is the image height, and $w$ is the image width. Consequently, the set of all images $\mathcal{I} \doteq \bigcup_{c,h,w=1}^{\infty} \mathbb{R}^{c\times h\times w}$ is the set of all possible such data. Again, we will use this notation repeatedly.

### 8.2.2 The Simplified DINO Task and Objective

Our task is to learn a good representation of the data. Contrastive learning, by and large, does this by defining what properties of the input image we wish the features to reflect, constructing images which share these properties but vary others, and setting up a loss which promotes that the features of images with shared properties are close and images with different properties are different. The naturally optimal solution to this learning problem is that the learned features preserve the desired properties of the input. However, there are many practical and empirical complications that arise in the course of training contrastive models.

In the case of DINO, the authors propose to use a methodology which produces a single feature vector for the whole image and desires the feature vector to contain "global" (i.e., image-level) information. Accordingly, the loss will promote that images with similar global information have similar features and images with different global information have different features.

This seems intuitive, but as previously mentioned, there are several empirical considerations, even while setting up the loss. First and foremost, how should we promote similarities and differences? The answer from DINO [CTM+21] is[3] to convert the output features into "logits" corresponding to some probability distribution and take their cross-entropy. More specifically, let $\Delta_m \doteq \{\boldsymbol{x} \in \mathbb{R}^m \colon x_i \geq 0\ \forall i \in [m], \sum_{i=1}^m x_i = 1\}$ be the space of probability vectors in $\mathbb{R}^m$ and define the function $d_{\mathrm{CE}} \colon \mathbb{R}^m \times \mathbb{R}^m \to \mathbb{R}$ by

$$d_{\mathrm{CE}}(\boldsymbol{p}, \boldsymbol{q}) \doteq \mathrm{CE}(\boldsymbol{p}, \boldsymbol{q}), \quad \forall \boldsymbol{p}, \boldsymbol{q} \in \Delta_m \tag{8.2.1}$$

where $\mathrm{CE} \colon \Delta_m \times \Delta_m \to \mathbb{R}$ is the cross-entropy, defined as

$$\mathrm{CE}(\boldsymbol{p}, \boldsymbol{q}) \doteq -\sum_{i=1}^m p_i \log q_i, \quad \forall \boldsymbol{p} = (p_1, \ldots, p_m), \boldsymbol{q} = (q_1, \ldots, q_m) \in \Delta_m. \tag{8.2.2}$$

Before we continue our discussion, let us build some intuition about this distance function. We have, in particular,

$$\mathrm{CE}(\boldsymbol{p}, \boldsymbol{q}) = -\sum_{i=1}^m p_i \log q_i = \sum_{i=1}^m p_i \log(p_i/q_i) - \sum_{i=1}^m p_i \log p_i \tag{8.2.3}$$

$$= \mathsf{KL}(\boldsymbol{p} \parallel \boldsymbol{q}) + H(\boldsymbol{p}) \tag{8.2.4}$$

where $\mathsf{KL} \colon \Delta_m \times \Delta_m \to \mathbb{R}$ is the KL divergence, defined as

$$\mathsf{KL}(\boldsymbol{p} \parallel \boldsymbol{q}) \doteq \sum_{i=1}^m p_i \log(p_i/q_i), \tag{8.2.5}$$

and $H \colon \Delta_m \to \mathbb{R}$ is the entropy of a random variable. Note that $\mathsf{KL}(\boldsymbol{p} \parallel \boldsymbol{q})$ is minimized if and only if $\boldsymbol{p} = \boldsymbol{q}$. So minimizing $d_{\mathrm{CE}}$ does two things: it makes $\boldsymbol{p} = \boldsymbol{q}$, and it makes $\boldsymbol{p}$ and $\boldsymbol{q}$ have minimal entropy (i.e., vectors with 1 in one component and 0 elsewhere — these are called *one-hot vectors*). Overall, the goal of this objective is not *just* to match $\boldsymbol{p}$ and $\boldsymbol{q}$ but also to shape them in a certain way to make them low-entropy. Keep this in mind when we discuss the formulation.

[3] In the author's view, inexplicably...

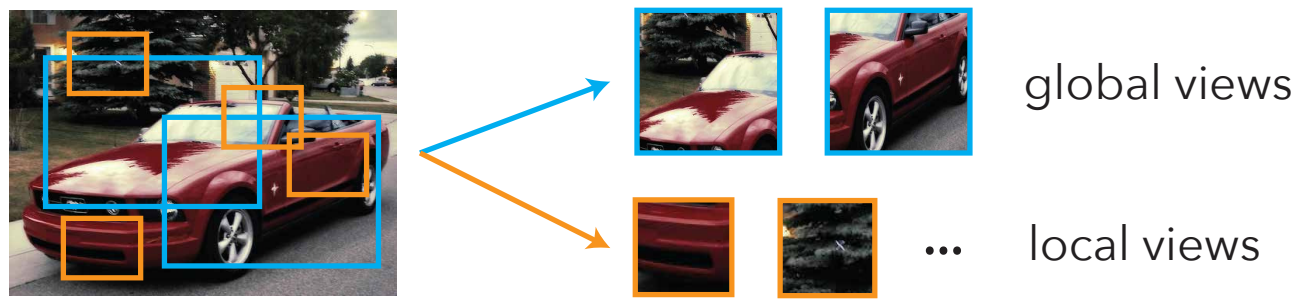

Figure 8.4: **Local and global views in DINO.** Local views and global views take a rectangular crop of the input image and resize it to a square shape, which is then input into the network for processing.

The next question is, how should we obtain samples with similar global information? The answer from DINO (as well as nearly all contrastive learning) is *data augmentation* — from each sample, make several correlated samples which share the desired properties. In the DINO case, we use different crops or *views* of the input image. Recall that we model an image as an element of the set $\mathcal{I}$. In this notation, a view is a function $v \colon \mathcal{I} \to \mathcal{I}$. In the DINO case, the view is a *random resized crop*: it takes a randomly chosen rectangular crop of the image (which has a fixed percentage $p_v \in [0, 1]$ of the total area of the image), resizes it proportionally so that the *shorter edge* is $S_{\mathrm{rsz}}$ pixels long, then resizes it to a fixed shape $(C, S_v, S_v)$ where $S_v \geq 1$ is the size of the view and $C$ is the number of channels in the original image.

There are two types of views we want to use, depicted in Figure 8.4:

- *global views*, which are random resized crops with area percentage parameter $p_{\mathrm{glo}} \in [0, 1]$ and output shape $(C, S_{\mathrm{glo}}, S_{\mathrm{glo}})$;

- *local views*, which are random resized crops with area percentage parameter $p_{\mathrm{loc}} \in [0, 1]$ and output shape $(C, S_{\mathrm{loc}}, S_{\mathrm{loc}})$. Here $p_{\mathrm{loc}} < p_{\mathrm{glo}}$ and $S_{\mathrm{loc}} < S_{\mathrm{glo}}$.

DINO desires that the aggregate features $\boldsymbol{z}_\theta(\boldsymbol{X}_v) \doteq (f_\theta^{\mathrm{ext}} \circ f_\theta)(\boldsymbol{X}_v)$ of all views $\boldsymbol{X}_v \doteq v(\boldsymbol{X})$ of an input image $\boldsymbol{X}$ be consistent with each other. DINO does this by using a *"DINO head"*[4] $h_{\boldsymbol{W},\boldsymbol{\mu}}$, parameterized by a matrix $\boldsymbol{W} \in \mathbb{R}^{s \times d}$ and a vector $\boldsymbol{\mu} \in \mathbb{R}^s$, to extract a probability vector $\boldsymbol{p}_{\theta,\boldsymbol{W},\boldsymbol{\mu}}(\boldsymbol{X}_v) \doteq h_{\boldsymbol{W},\boldsymbol{\mu}}(\boldsymbol{z}_\theta(\boldsymbol{X}_v))$ from the aggregate feature $\boldsymbol{z}_\theta(\boldsymbol{X}_v)$, using the following simple recipe:

$$h_{\boldsymbol{W},\boldsymbol{\mu}}(\boldsymbol{z}) \doteq \operatorname{softmax}([\boldsymbol{W}\boldsymbol{z} - \boldsymbol{\mu}]/\tau), \qquad \forall \boldsymbol{z} \in \mathbb{R}^d, \tag{8.2.6}$$

[4]Note that $h_{\boldsymbol{W},\boldsymbol{\mu}}$ is the task-specific head, which in Section 8.1.2 is parameterized only by $\theta$ as opposed to any specific parameters, but since we use two invocations of $h$ with different values of the second parameter, we keep the specified notation.

where the softmax: $\mathbb{R}^s \to \Delta_s$ function is defined by

$$\operatorname{softmax}\left(\begin{bmatrix} x_1 \\ \vdots \\ x_s \end{bmatrix}\right) \doteq \frac{1}{\sum_{i=1}^{s} e^{x_i}} \begin{bmatrix} e^{x_1} \\ \vdots \\ e^{x_s} \end{bmatrix} \tag{8.2.7}$$

and $\tau > 0$ is a "temperature" parameter which controls the entropy of the softmax's output.

In particular, DINO minimizes the difference between the probability vector $\boldsymbol{p}_{\theta,\boldsymbol{W},\boldsymbol{\mu}}(\boldsymbol{X}_g) \doteq h_{\boldsymbol{W},\boldsymbol{\mu}}(\boldsymbol{z}_\theta(\boldsymbol{X}_g))$ for each global view $\boldsymbol{X}_g \doteq v_g(\boldsymbol{X})$ and the probability vector $\boldsymbol{p}_{\theta,\boldsymbol{W}}(\boldsymbol{X}_c) \doteq h_{\boldsymbol{W},\boldsymbol{0}_m}(\boldsymbol{z}_\theta(\boldsymbol{X}_c))$ for each view $\boldsymbol{X}_c \doteq v_c(\boldsymbol{X})$. Here, $v_c$ can *either* be a local view or a global view. We will discuss the implementation of $f_\theta$ and $f_\theta^{\mathrm{ext}}$ shortly in Section 8.2.3. Overall, DINO solves the problem

$$\begin{aligned} &\min_{\theta,\boldsymbol{W},\boldsymbol{\mu}} \mathcal{L}_{\mathrm{DINO}}(\theta,\boldsymbol{W},\boldsymbol{\mu}) \qquad \text{where} \\ &\mathcal{L}_{\mathrm{DINO}}(\theta,\boldsymbol{W},\boldsymbol{\mu}) \doteq \mathbb{E}[d_{\mathrm{CE}}(\boldsymbol{p}_{\theta,\boldsymbol{W},\boldsymbol{\mu}}(\boldsymbol{X}_g), \boldsymbol{p}_{\theta,\boldsymbol{W}}(\boldsymbol{X}_c))], \end{aligned} \tag{8.2.8}$$

where the expectation is over data $\boldsymbol{X}$, global views $v_g$, and other views $v_c$.

In this specific case, however, if you try to implement (8.2.8) and optimize it on a real network, it is very likely that you will run into a problem: after running a few iterations of the learning algorithm, the feature mapping $f_\theta^{\mathrm{ext}} \circ f_\theta$ will *become the constant function*! This certainly optimizes the above loss since it minimizes the distance between features of different views of the same image. But we obviously do not want to learn this solution.

Actually avoiding collapse is a very common consideration in contrastive learning. So how do we do it in this case? The solution from DINO, again, is empirically designed, and carefully tunes the optimization of the parameter $\boldsymbol{\mu}$ (which is updated using all samples in the batch) and a "temperature" hyperparameter $\tau$ which is part of the implementation of $h_{\boldsymbol{W},\boldsymbol{\mu}}$ and discussed in Section 8.2.3. Given a certain special set of hyperparameters that work well, this is indeed enough to ensure non-collapse of the representation. However, outside of this special configuration, training models to converge is difficult, and the training is highly unstable.

To amend this state of affairs, let us discuss simplifications to the formulation. First, instead of computing a probability vector using a learned transformation $h_{\boldsymbol{W},\boldsymbol{\mu}}$ of the aggregate features $\boldsymbol{z}_\theta$, we can *directly use the aggregate representation*, ignoring the task-specific head (or equivalently, setting it to the identity mapping). But now we need a way to compare the vectors directly. Using our hypothesis from Chapter 5 that good representations should have Euclidean (subspace) geometry, a much more natural measure of difference is the *squared $\ell^2$ distance* $d_{\ell^2}\colon \mathbb{R}^d \times \mathbb{R}^d \to \mathbb{R}$, defined as

$$d_{\ell^2}(\boldsymbol{x},\boldsymbol{y}) \doteq \frac{1}{2}\|\boldsymbol{x}-\boldsymbol{y}\|_2^2, \qquad \forall \boldsymbol{x},\boldsymbol{y} \in \mathbb{R}^d. \tag{8.2.9}$$

This distance-based score is even more efficient to compute than the cross-entropy score. Thus, $d_{\ell^2}$ takes the place of $d_{\mathrm{CE}}$ in our simplification.

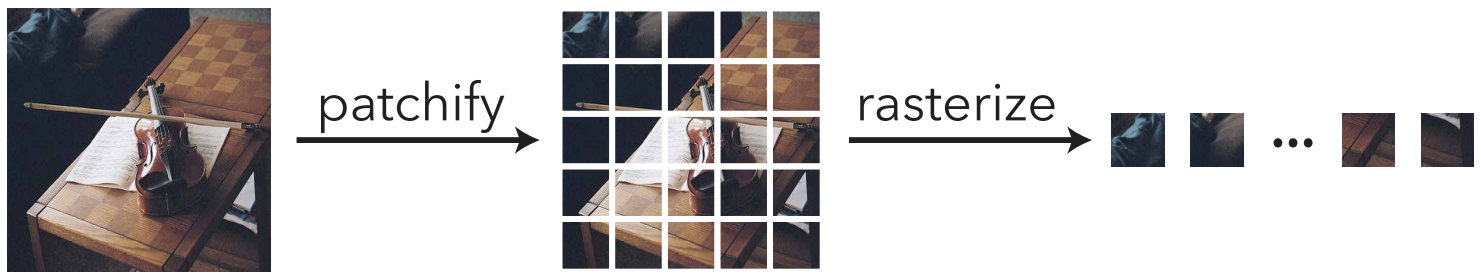

Figure 8.5: **An example of an image turned into $5 \times 5$ square patches, which are placed in raster order.** Each patch is of the same size, and the grid of patches is of shape $(N_H, N_W) = (5, 5)$. The grid of patches is then unrolled into a sequence of length $5 \times 5 = 25$ in raster order.

Before, collapse was avoided by using tricks to update $\boldsymbol{\mu}$ and $\tau$. In our simplification, if we compare the features within the representation space instead of converting them to probabilities, we do not have either of these parameters and so must consider a different way to avoid collapse. To do this, we return to the fundamentals. The basic idea of avoiding collapse is that in order to make sure that all samples do not return the same exact same features, we need different samples to have different features. In other words, we would like the *covariance* of the features to be *large* in some sense. But from Chapters 4 and 5, we already have a quantity which measures the size of the covariance matrix. Namely, we use the straightforward (population-level) *Gaussian coding rate* $R$ to ensure that the features of global views of different images, which have different global information, are well-separated and not collapsed (hence expanded). The overall modified loss $\mathcal{L}_{\mathrm{SimDINO}}$ becomes:

$$\mathcal{L}_{\mathrm{SimDINO}}(\theta) \doteq \mathbb{E}[d_{\ell^2}(\boldsymbol{z}_\theta(\boldsymbol{X}_g), \boldsymbol{z}_\theta(\boldsymbol{X}_c))] \tag{8.2.10}$$

$$- \frac{\gamma}{2} \log\det\left(\boldsymbol{I} + \frac{d}{\varepsilon^2}\operatorname{Cov}(\boldsymbol{z}_\theta(\boldsymbol{X}_g))\right), \tag{8.2.11}$$

where $\varepsilon > 0$ is fixed and the appropriate expectations are, as before, taken over data $\boldsymbol{X}$, global view $v_g$, and other (local or global) view $v_c$. The loss in (8.2.10) is essentially the same loss introduced in Section 4.3.2 of Chapter 4 for the simplified DINO ("SimDINO" [WZP+25]). As we will see, when properly implemented, it works at least as well as the original DINO.

### 8.2.3 Architecture: Vision Transformer

For the architecture, we use a standard vision transformer. Here is how such an architecture works formally in the context of image data. Recall from Section 8.1.2 that there are four components to an encoder architecture, namely an embedding, a backbone, a feature extractor, and a task-specific head. We discuss these four parts presently.

**Embedding.** Given image data $\boldsymbol{X} \in \mathcal{I}$, we embed it as a sequence of tokens in $\mathbb{R}^d$ using the map $f_\theta^{\mathrm{emb}}$, as follows. The first two steps are depicted in Figure 8.5, and the latter two are depicted in Figure 8.6.

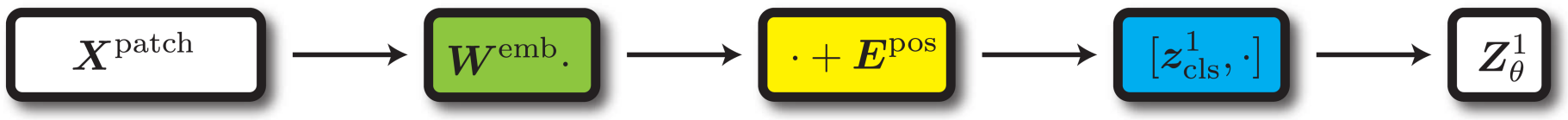

Figure 8.6: **The transformer embedding pipeline.** Given a sequence of unrolled patches in raster order $\boldsymbol{X}^{\text{patch}}$, each unrolled patch is linearly projected into the feature space, and equipped with an (additive) positional encoding and an additional token known as the class token. The output is the first-layer-input feature $\boldsymbol{Z}_\theta^1(\boldsymbol{X}) = f_\theta^{\text{emb}}(\boldsymbol{X})$.

1. First, we turn the image data $\boldsymbol{X}$ into a sequence of patches of shape $(C, P_H, P_W)$ where $P_H$ and $P_W$ are the patch dimensions. We assume that $P_H$ and $P_W$ evenly divide the height and width of $\boldsymbol{X}$, respectively (in the notation of Section 8.2.2 we assume that $P_H$ and $P_W$ evenly divide $S_{\text{loc}}$ and $S_{\text{glo}}$). Let the resulting grid of patches have $N_H$ rows and $N_W$ columns.

2. We unroll each patch into a vector of length $D \doteq CP_HP_W$. There are $N \doteq N_HN_W$ patch vectors, which we place in "raster order" (top left $\to$ top right $\to$ bottom left $\to$ bottom right) into a matrix $\boldsymbol{X}^{\text{patch}} \in \mathbb{R}^{D\times N}$, where $\boldsymbol{X}^{\text{patch}} \doteq f^{\text{patch}}(\boldsymbol{X})$. Notice that $D$ depends only on the patch size and number of channels. Since the latter quantity is normally constant among samples in the same dataset, $D$ is the same for all images in the dataset, while $N$ is different for larger and smaller images.

3. We then perform the following operation on $\boldsymbol{X}^{\text{patch}} \in \mathbb{R}^{D\times N}$ to project it to $\mathbb{R}^{d\times n}$ where $n \doteq N + 1$:

$$\boldsymbol{X}^{\text{patch}} \mapsto [\boldsymbol{z}_{\text{cls}}^1, \boldsymbol{W}^{\text{emb}}\boldsymbol{X}] + \boldsymbol{E}^{\text{pos}}. \tag{8.2.12}$$

   Here we have three trainable parameters $\boldsymbol{W}^{\text{emb}}$, $\boldsymbol{z}_{\text{cls}}^1$, and $\boldsymbol{E}^{\text{pos}}$ whose purpose is as follows:

   - $\boldsymbol{W}^{\text{emb}} \in \mathbb{R}^{d\times D}$ is a matrix which projects each patch vector to a token feature.
   - $\boldsymbol{z}_{\text{cls}}^1 \in \mathbb{R}^d$ is a so-called *class token* or *register token*. The class token heuristically holds global information of the whole data and is used for downstream tasks. In the framework of compressive deep networks from Chapter 5, we expect that the class token is projected onto the same subspaces as the salient or semantically relevant tokens during the progression of the forward pass.
   - $\boldsymbol{E}^{\text{pos}} \in \mathbb{R}^{d\times N}$ is a so-called *positional encoding* which distinguishes tokens of different patches from each other. That is, token features should have positional information, so that the overall map $f^{\text{pre}}$ is not invariant to permutations of the patches, and $\boldsymbol{E}^{\text{pos}}$ inserts this positional information.
     - In this DINO case, where the transformer receives differently-sized images, we learn a positional encoding for the largest size received

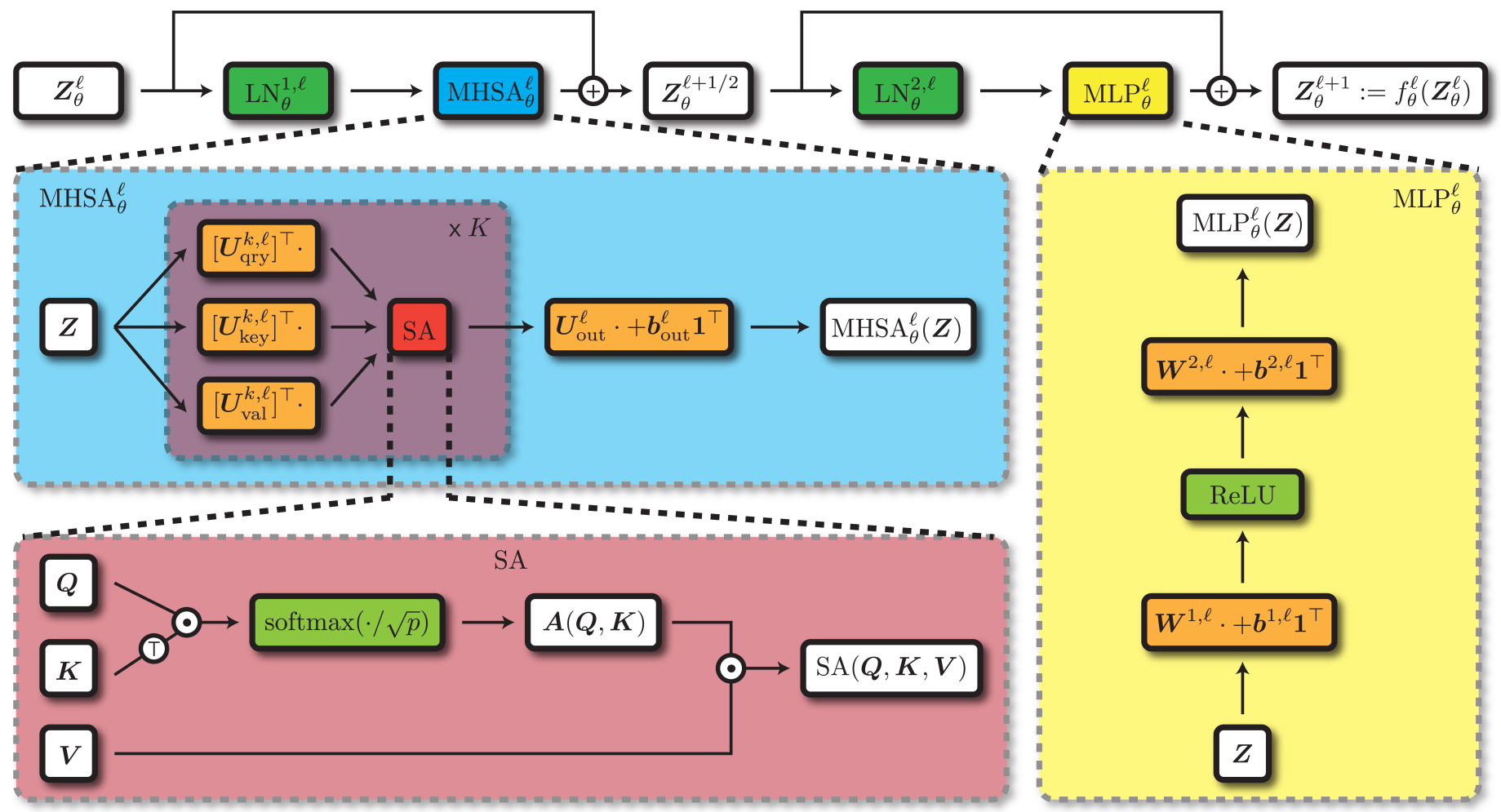

Figure 8.7: **One layer $f_\theta^\ell$ of the transformer backbone.** The input features go through layer-normalization, multi-head self-attention, and multi-layer perceptron blocks in sequence to form the output features of the layer.

during training, and interpolate to get the positional encodings for smaller-sized inputs.

Thus, in the end we have

$$f_\theta^{\mathrm{emb}}(\boldsymbol{X}) \doteq \left[\boldsymbol{z}_{\mathrm{cls}}^1, \boldsymbol{W}^{\mathrm{emb}} f^{\mathrm{patch}}(\boldsymbol{X}) + \boldsymbol{E}^{\mathrm{pos}}\right]. \tag{8.2.13}$$

All parameters $\boldsymbol{z}_{\mathrm{cls}}^1, \boldsymbol{W}^{\mathrm{emb}}, \boldsymbol{E}^{\mathrm{pos}}$ are contained in the parameter set $\theta$.

**Backbone.** Given a sequence of embeddings $\boldsymbol{Z}_\theta^1(\boldsymbol{X}) \doteq f_\theta^{\mathrm{emb}}(\boldsymbol{X}) \in (\mathbb{R}^d)^*$, we process it using the backbone map $f_\theta^{\mathrm{bb}}$ as follows and as depicted in Figure 8.7. The function $f_\theta^{\mathrm{bb}}$ is composed of $L$ *layers* $f_\theta^\ell$, i.e.,

$$f_\theta^{\mathrm{bb}} = f_\theta^L \circ \cdots \circ f_\theta^1. \tag{8.2.14}$$

The layer $f_\theta^\ell$ has the following implementation:

$$\boldsymbol{Z}_\theta^{\ell+1/2}(\boldsymbol{X}) = \boldsymbol{Z}_\theta^\ell(\boldsymbol{X}) + \mathrm{MHSA}_\theta^\ell(\mathrm{LN}_\theta^{1,\ell}(\boldsymbol{Z}_\theta^\ell(\boldsymbol{X}))) \tag{8.2.15}$$

$$\boldsymbol{Z}_\theta^{\ell+1}(\boldsymbol{X}) = \boldsymbol{Z}_\theta^{\ell+1/2}(\boldsymbol{X}) + \mathrm{MLP}_\theta^\ell(\mathrm{LN}_\theta^{2,\ell}(\boldsymbol{Z}_\theta^{\ell+1/2}(\boldsymbol{X}))) \tag{8.2.16}$$

and $f_\theta^\ell$ is defined such that $f_\theta^\ell(\boldsymbol{Z}_\theta^\ell(\boldsymbol{X})) \doteq \boldsymbol{Z}_\theta^{\ell+1}(\boldsymbol{X})$. Here we have used some operators, such as $\mathrm{MHSA}_\theta^\ell, \mathrm{MLP}_\theta^\ell$ and $\mathrm{LN}_\theta^{i,\ell}$ that are defined as follows:

- The $\mathrm{MHSA}_\theta^\ell$ operator is multi-head-self-attention, the predecessor of the multi-head subspace self-attention (cf Chapter 5). The formulation is as

follows:

$$\mathrm{MHSA}_{\theta}^{\ell}(\boldsymbol{Z}) \doteq \boldsymbol{U}_{\mathrm{out}}^{\ell} \begin{bmatrix} \mathrm{SA}([\boldsymbol{U}_{\mathrm{qry}}^{1,\ell}]^{\top}\boldsymbol{Z}, [\boldsymbol{U}_{\mathrm{key}}^{1,\ell}]^{\top}\boldsymbol{Z}, [\boldsymbol{U}_{\mathrm{val}}^{1,\ell}]^{\top}\boldsymbol{Z}) \\ \vdots \\ \mathrm{SA}([\boldsymbol{U}_{\mathrm{qry}}^{K,\ell}]^{\top}\boldsymbol{Z}, [\boldsymbol{U}_{\mathrm{key}}^{K,\ell}]^{\top}\boldsymbol{Z}, [\boldsymbol{U}_{\mathrm{val}}^{K,\ell}]^{\top}\boldsymbol{Z}) \end{bmatrix} + \boldsymbol{b}_{\mathrm{out}}^{\ell}\mathbf{1}_n^{\top}, \tag{8.2.17}$$

$$\text{where} \qquad \mathrm{SA}(\boldsymbol{Q}, \boldsymbol{K}, \boldsymbol{V}) \doteq \boldsymbol{V} \underbrace{\mathrm{softmax}\left(\frac{\boldsymbol{K}^{\top}\boldsymbol{Q}}{\sqrt{p}}\right)}_{\doteq \boldsymbol{A}(\boldsymbol{Q},\boldsymbol{K})} \tag{8.2.18}$$

where $p$ is a positive integer, $\boldsymbol{U}_{\mathrm{qry}}^{k,\ell}, \boldsymbol{U}_{\mathrm{key}}^{k,\ell}, \boldsymbol{U}_{\mathrm{val}}^{k,\ell} \in \mathbb{R}^{d\times p}$, $\boldsymbol{U}_{\mathrm{out}}^{\ell} \in \mathbb{R}^{d\times Kp}$, and $\boldsymbol{b}_{\mathrm{out}}^{\ell} \in \mathbb{R}^{d}$ are trainable parameters contained in the parameter set $\theta$, and the softmax is defined column-wise as

$$\mathrm{softmax}(\boldsymbol{M}) \doteq \begin{bmatrix}\mathrm{softmax}(\boldsymbol{m}_1) & \cdots & \mathrm{softmax}(\boldsymbol{m}_p)\end{bmatrix}, \tag{8.2.19}$$

$$\forall \boldsymbol{M} = \begin{bmatrix}\boldsymbol{m}_1, \dots, \boldsymbol{m}_p\end{bmatrix} \in \mathbb{R}^{n\times p}. \tag{8.2.20}$$

In practice, the dimensions are usually picked such that $Kp = d$. The terms

$$\begin{aligned} \boldsymbol{A}_{\theta}^{k,\ell}(\boldsymbol{Z}) &\doteq \boldsymbol{A}([\boldsymbol{U}_{\mathrm{qry}}^{k,\ell}]^{\top}\boldsymbol{Z}, [\boldsymbol{U}_{\mathrm{key}}^{k,\ell}]^{\top}\boldsymbol{Z}), \\ \mathrm{SA}_{\theta}^{k,\ell}(\boldsymbol{Z}) &\doteq \mathrm{SA}([\boldsymbol{U}_{\mathrm{qry}}^{k,\ell}]^{\top}\boldsymbol{Z}, [\boldsymbol{U}_{\mathrm{key}}^{k,\ell}]^{\top}\boldsymbol{Z}, [\boldsymbol{U}_{\mathrm{val}}^{k,\ell}]^{\top}\boldsymbol{Z}) \end{aligned} \tag{8.2.21}$$

are also known as the *kth attention map* and *kth attention head output* at layer $\ell$, respectively. Furthermore, the operation $\mathrm{SA}(\boldsymbol{Q}, \boldsymbol{K}, \boldsymbol{V})$ can be computed extremely efficiently using specialized software such as FlashAttention [SBZ+25].

- The $\mathrm{MLP}_{\theta}^{\ell}$ is a two-layer perceptron, a regular nonlinearity used in deep networks, and has the form

$$\mathrm{MLP}_{\theta}^{\ell}(\boldsymbol{Z}) \doteq \boldsymbol{W}_{\mathrm{down}}^{\ell}\,\mathrm{ReLU}(\boldsymbol{W}_{\mathrm{up}}^{\ell}\boldsymbol{Z} + \boldsymbol{b}_{\mathrm{up}}^{\ell}\mathbf{1}_n^{\top}) + \boldsymbol{b}_{\mathrm{down}}^{\ell}\mathbf{1}_n^{\top} \tag{8.2.22}$$

where $\boldsymbol{W}_{\mathrm{up}}^{\ell} \in \mathbb{R}^{q\times d}, \boldsymbol{W}_{\mathrm{down}}^{\ell} \in \mathbb{R}^{d\times q}, \boldsymbol{b}_{\mathrm{up}}^{\ell} \in \mathbb{R}^{q}, \boldsymbol{b}_{\mathrm{down}}^{\ell} \in \mathbb{R}^{d}$ are trainable parameters also contained in the parameter set $\theta$, and ReLU is the element-wise ReLU nonlinearity, i.e., $\mathrm{ReLU}(\boldsymbol{M})_{ij} = \max\{M_{ij}, 0\}$.

- Each layer-norm $\mathrm{LN}_{\theta}^{i,\ell}$ for $i \in \{1, 2\}$ is a standard normalization, which applies column-wise to each token feature independently:

$$\mathrm{LN}_{\theta}^{i,\ell}(\boldsymbol{Z}) = \mathrm{LN}_{\theta}^{i,\ell}(\begin{bmatrix}\boldsymbol{z}_1, \dots, \boldsymbol{z}_n\end{bmatrix}) = \begin{bmatrix}\mathrm{LN}_{\theta}^{i,\ell}(\boldsymbol{z}_1), \dots, \mathrm{LN}_{\theta}^{i,\ell}(\boldsymbol{z}_n)\end{bmatrix} \tag{8.2.23}$$

and has the form

$$\mathrm{LN}_{\theta}^{i,\ell}(\boldsymbol{z}) = \frac{\boldsymbol{z} - \mathrm{mean}(\boldsymbol{z})\mathbf{1}_d}{\|\boldsymbol{z} - \mathrm{mean}(\boldsymbol{z})\mathbf{1}_d\|_2} \odot \boldsymbol{\alpha}^{i,\ell} + \boldsymbol{\beta}^{i,\ell} \tag{8.2.24}$$

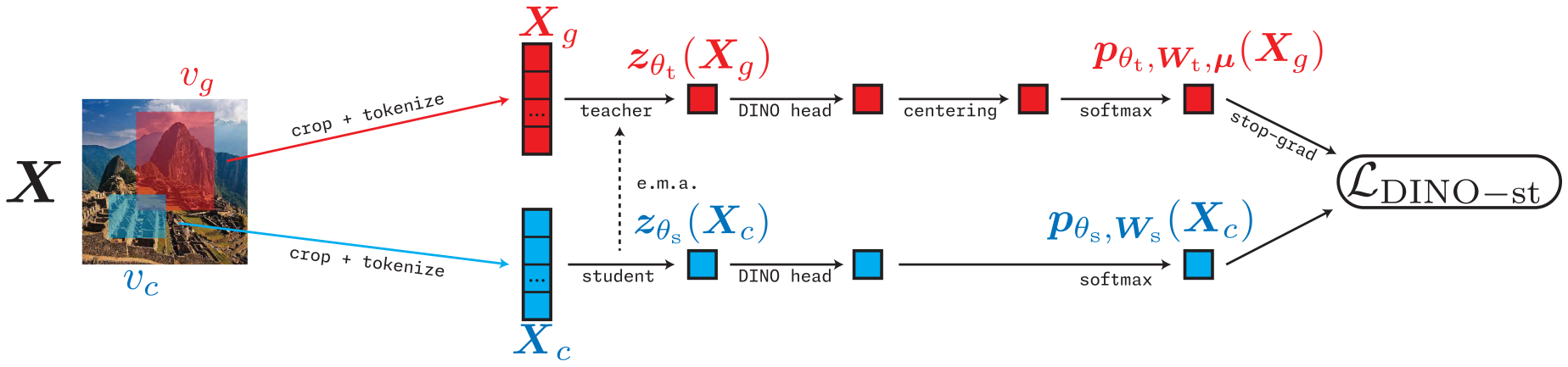

Figure 8.8: **The DINO pipeline.** Student features and teacher features are computed for each input. The objective attempts to align the student features with the teacher features by projecting both sets of features into a high-dimensional probability simplex and computing a cross-entropy loss. Notably, because of the "stop-grad", the gradient is only computed w.r.t. the *student parameters' outputs*.

$$\text{where} \qquad \operatorname{mean}(\boldsymbol{z}) = \frac{1}{d}\mathbf{1}_d^\top \boldsymbol{z} \tag{8.2.25}$$

where $\odot$ denotes element-wise multiplication, and $\boldsymbol{\alpha}^{i,\ell}, \boldsymbol{\beta}^{i,\ell} \in \mathbb{R}^d$ are trainable parameters contained in the parameter set $\theta$. The layer-norm operator serves as a sort of normalization on each token, where the scale of each token afterwards is learnable and shared amongst all tokens.

The transformer is one of the most popular neural network architectures in history, powering applications in almost all fields of deep learning.

**Feature extractor.** We use a post-processing step $f_\theta^{\text{ext}}$ which extracts the *class token feature*, which (recall) is the feature meant to contain aggregate information about the input image, and applies an MLP and normalization to it. Namely, we have

$$\boldsymbol{z}_\theta(\boldsymbol{X}) \doteq f_\theta^{\text{ext}}(\boldsymbol{Z}_\theta(\boldsymbol{X})) = f_\theta^{\text{ext}}([\boldsymbol{z}_\theta^1(\boldsymbol{X}), \dots, \boldsymbol{z}_\theta^n(\boldsymbol{X})]) \tag{8.2.26}$$

$$\doteq \frac{\operatorname{MLP}_\theta^{\text{ext}}(\boldsymbol{z}_\theta^1(\boldsymbol{X}))}{\|\operatorname{MLP}_\theta^{\text{ext}}(\boldsymbol{z}_\theta^1(\boldsymbol{X}))\|_2}. \tag{8.2.27}$$

**Task-specific ("DINO") head.** For DINO, we use the task-specific DINO head $h_{\boldsymbol{W},\boldsymbol{\mu}}$. For SimDINO, we use *no* task-specific head *at all*, as previously described.

### 8.2.4 Optimization Strategy

**Optimizing DINO.** We have a loss function and an architecture, so we now discuss the optimization strategy. The optimization strategy for DINO uses *two sets of weights for the same architecture*: *student* weights $\theta_{\text{s}}$ and *teacher* weights $\theta_{\text{t}}$. These correspond to two different neural networks, called the teacher network and student network, with the same architecture. The teacher network encodes all global views, while the student network encodes all "other" views.

The goal of the loss is to distill teacher outputs into the student model. Namely, we train on the loss $\mathcal{L}_{\mathrm{DINO-st}}$:

$$\mathcal{L}_{\mathrm{DINO-st}}(\theta_{\mathrm{s}}, \theta_{\mathrm{t}}, \boldsymbol{W}_{\mathrm{s}}, \boldsymbol{W}_{\mathrm{t}}, \boldsymbol{\mu}) \doteq \mathbb{E}[d_{\mathrm{CE}}(\boldsymbol{p}_{\theta_{\mathrm{t}}, \boldsymbol{W}_{\mathrm{t}}, \boldsymbol{\mu}}(\boldsymbol{X}_g), \boldsymbol{p}_{\theta_{\mathrm{s}}, \boldsymbol{W}_{\mathrm{s}}}(\boldsymbol{X}_c))]. \tag{8.2.28}$$

Now, we can fully describe the overall pipeline of DINO, depicted in Figure 8.8.

While it is easy to reason about (8.2.28), it is impossible in practice to implement optimization algorithms such as gradient descent with a loss given by $\mathcal{L}_{\mathrm{DINO-st}}$. This is because the expectations in the loss are impossible to evaluate, much less to take the gradient of. In this extremely frequent case, we approximate the expectation via finite samples. That is, at each timestep $k$ we:

- Subsample $B$ data points from our dataset $\{\boldsymbol{X}_1^{(k)}, \dots, \boldsymbol{X}_B^{(k)}\} \subset \mathcal{I}$.
- For each data point $\boldsymbol{X}_b^{(k)}$, sample $M_{\mathrm{glo}}$ global views $v_{b,g}^{(k),i}$ and $M_{\mathrm{loc}}$ local views $v_{b,\ell}^{(k),i}$. Apply the views to $\boldsymbol{X}_b^{(k)}$ to obtain $\boldsymbol{X}_{b,g}^{(k),i} \doteq v_{b,g}^{(k),i}(\boldsymbol{X}_b^{(k)})$ and $\boldsymbol{X}_{b,\ell}^{(k),i} \doteq v_{b,\ell}^{(k),i}(\boldsymbol{X}_b^{(k)})$.
- For each *local* view $\boldsymbol{X}_{b,\ell}^{(k),i}$, compute the following quantities:

$$\boldsymbol{z}_{\theta_{\mathrm{s}}}(\boldsymbol{X}_{b,\ell}^{(k),i}) \doteq (f_{\theta_{\mathrm{s}}}^{\mathrm{ext}} \circ f_{\theta_{\mathrm{s}}})(\boldsymbol{X}_{b,\ell}^{(k),i}), \tag{8.2.29}$$

$$\boldsymbol{p}_{\theta_{\mathrm{s}}, \boldsymbol{W}_{\mathrm{s}}}(\boldsymbol{X}_{b,\ell}^{(k),i}) \doteq h_{\boldsymbol{W}_{\mathrm{s}}, \mathbf{0}_m}(\boldsymbol{z}_{\theta_{\mathrm{s}}}(\boldsymbol{X}_{b,\ell}^{(k),i}(\theta))) \tag{8.2.30}$$

  and for each *global* view $\boldsymbol{X}_{b,g}^{(k),i}$, compute the following quantities (by an abuse of notation):

$$\boldsymbol{z}_{\theta_{\mathrm{s}}}(\boldsymbol{X}_{b,g}^{(k),i}) \doteq (f_{\theta_{\mathrm{s}}}^{\mathrm{ext}} \circ f_{\theta_{\mathrm{s}}})(\boldsymbol{X}_{b,g}^{(k),i}), \tag{8.2.31}$$

$$\boldsymbol{p}_{\theta_{\mathrm{s}}, \boldsymbol{W}_{\mathrm{s}}}(\boldsymbol{X}_{b,g}^{(k),i}) \doteq h_{\boldsymbol{W}_{\mathrm{s}}, \mathbf{0}_m}(\boldsymbol{z}_{\theta_{\mathrm{s}}}(\boldsymbol{X}_{b,g}^{(k),i})), \tag{8.2.32}$$

$$\boldsymbol{z}_{\theta_{\mathrm{t}}}(\boldsymbol{X}_{b,g}^{(k),i}) \doteq (f_{\theta_{\mathrm{t}}}^{\mathrm{ext}} \circ f_{\theta_{\mathrm{t}}})(\boldsymbol{X}_{b,g}^{(k),i}), \tag{8.2.33}$$

$$\boldsymbol{p}_{\theta_{\mathrm{t}}, \boldsymbol{W}_{\mathrm{t}}, \boldsymbol{\mu}}(\boldsymbol{X}_{b,g}^{(k),i}) \doteq h_{\boldsymbol{W}_{\mathrm{t}}, \boldsymbol{\mu}}(\boldsymbol{Z}_{\theta_{\mathrm{t}}}(\boldsymbol{X}_{b,g}^{(k),i})). \tag{8.2.34}$$

- Compute the *surrogate, approximate loss* $\hat{\mathcal{L}}_{\mathrm{DINO-st}}^{(k)}$, defined as follows:

$$\hat{\mathcal{L}}_{\mathrm{DINO-st}}^{(k)}(\theta_{\mathrm{s}}, \theta_{\mathrm{t}}, \boldsymbol{W}_{\mathrm{s}}, \boldsymbol{W}_{\mathrm{t}}, \boldsymbol{\mu}) \doteq \frac{1}{BM_{\mathrm{glo}}(M_{\mathrm{glo}} + M_{\mathrm{loc}} - 1)} \sum_{b=1}^{B} \sum_{i=1}^{M_{\mathrm{glo}}} \tag{8.2.35}$$

$$\left[\sum_{j=1}^{M_{\mathrm{loc}}} d_{\mathrm{CE}}(\boldsymbol{p}_{\theta_{\mathrm{t}}, \boldsymbol{W}_{\mathrm{t}}, \boldsymbol{\mu}}(\boldsymbol{X}_{b,g}^{(k),i}), \boldsymbol{p}_{\theta_{\mathrm{s}}, \boldsymbol{W}_{\mathrm{s}}}(\boldsymbol{X}_{b,\ell}^{(k),j})) \right. \tag{8.2.36}$$

$$\left. + \sum_{\substack{j=1 \\ j \neq i}}^{M_{\mathrm{glo}}} d_{\mathrm{CE}}(\boldsymbol{p}_{\theta_{\mathrm{t}}, \boldsymbol{W}_{\mathrm{t}}, \boldsymbol{\mu}}(\boldsymbol{X}_{b,g}^{(k),i}), \boldsymbol{p}_{\theta_{\mathrm{s}}, \boldsymbol{W}_{\mathrm{s}}}(\boldsymbol{X}_{b,g}^{(k),j}))\right] \tag{8.2.37}$$

as well as its gradients with respect to $\theta_{\mathrm{s}}$ and $\boldsymbol{W}_{\mathrm{s}}$, which should be computed under the assumption that $\theta_{\mathrm{t}}$, $\boldsymbol{W}_{\mathrm{t}}$, and $\boldsymbol{\mu}$ are constants — namely that they are *detached from the computational graph* and not dependent on $\theta_{\mathrm{s}}$ and $\boldsymbol{W}_{\mathrm{s}}$.

- Update the student parameters $\theta_{\mathrm{s}}$ and $\boldsymbol{W}_{\mathrm{s}}$ via an iterative gradient-based optimization algorithm, and update $\theta_{\mathrm{t}}$, $\boldsymbol{W}_{\mathrm{t}}$, and $\boldsymbol{\mu}$ via exponential moving averages with decay parameters $\nu^{(k)}$, $\nu^{(k)}$, and $\rho^{(k)}$ respectively, i.e.,

$$(\theta_{\mathrm{s}}^{(k+1)}, \boldsymbol{W}_{\mathrm{s}}^{(k+1)}) = \textsc{OptUpdate}^{(k)}(\theta_{\mathrm{s}}^{(k)}, \boldsymbol{W}_{\mathrm{s}}^{(k)}; \nabla_{(\theta_{\mathrm{s}}, \boldsymbol{W}_{\mathrm{s}})}\hat{\mathcal{L}}_{\mathrm{DINO-st}}^{(k)}) \tag{8.2.38}$$

$$\theta_{\mathrm{t}}^{(k+1)} = \nu^{(k)}\theta_{\mathrm{t}}^{(k)} + (1-\nu^{(k)})\theta_{\mathrm{s}}^{(k+1)} \tag{8.2.39}$$

$$\boldsymbol{W}_{\mathrm{t}}^{(k+1)} = \nu^{(k)}\boldsymbol{W}_{\mathrm{t}}^{(k)} + (1-\nu^{(k)})\boldsymbol{W}_{\mathrm{s}}^{(k+1)} \tag{8.2.40}$$

$$\boldsymbol{\mu}^{(k+1)} = \rho^{(k)}\boldsymbol{\mu}^{(k)} + (1-\rho^{(k)}) \cdot \frac{1}{BM_{\mathrm{glo}}}\sum_{b=1}^{B}\sum_{i=1}^{M_{\mathrm{glo}}} \boldsymbol{W}^{(k)}\boldsymbol{z}_{\theta_{\mathrm{t}}}(\boldsymbol{X}_{b,g}^{(k),i}), \tag{8.2.41}$$

For example, if the chosen optimization algorithm were stochastic gradient descent, we would have the update $\theta_{\mathrm{s}}^{(k+1)} \doteq \theta_{\mathrm{s}}^{(k)} - \delta^{(k)}\nabla_{\theta_{\mathrm{s}}}\hat{\mathcal{L}}_{\mathrm{DINO-st}}^{(k)}$, and so on.

Notice that the optimization procedure is rather irregular: although all four parameters change at each iteration, only two of them are directly updated from a gradient-based method. The other two are updated from exponential moving averages, and indeed treated as constants when computing any gradients. After training, we discard the student weights and use the teacher weights for our trained network $f$, as this exponentially moving average has been empirically shown to stabilize the resulting model (this idea is known as Polyak averaging or iterate averaging).

The way that $\nu$ and $\rho$ change over the optimization trajectory (i.e., the functions $k \mapsto \nu^{(k)}$ and $k \mapsto \rho^{(k)}$) are hyperparameters or design decisions, with $\nu^{(1)} < 1$ and $\lim_{k\to\infty}\nu^{(k)} = 1$ usually, and similar for $\rho$. The temperature hyperparameter $\tau$, used in the DINO head $h_{\boldsymbol{W},\boldsymbol{\mu}}$, also changes over the optimization trajectory (though this dependence is not explicitly notated).

Using the surrogate ("empirical") loss transforms our intractable optimization problem, as in optimizing the loss in (8.2.28), into a tractable stochastic optimization problem which is run to train essentially every deep learning model in the world. This conversion is extremely natural once you have seen some examples, and we will hopefully give these examples throughout the chapter.

**Optimizing SimDINO.** The simplified DINO population-level objective is very similar in spirit but much simpler in execution, i.e.,

$$\mathcal{L}_{\mathrm{SimDINO-st}}(\theta_{\mathrm{s}}, \theta_{\mathrm{t}}) \doteq \mathbb{E}[d_{\ell^2}(\boldsymbol{z}_{\theta_{\mathrm{t}}}(\boldsymbol{X}_g), \boldsymbol{z}_{\theta_{\mathrm{s}}}(\boldsymbol{X}_c))] \tag{8.2.42}$$

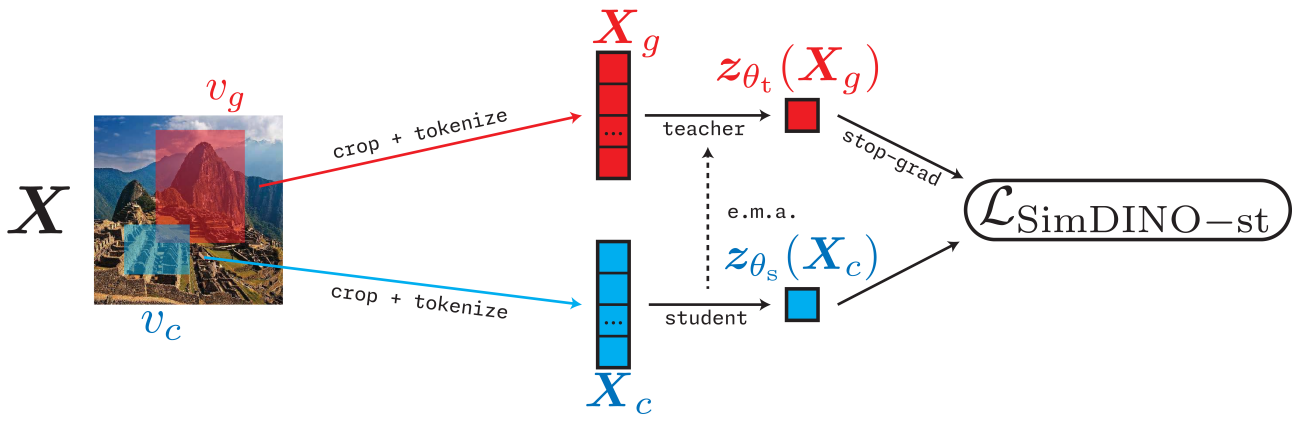

Figure 8.9: **The SimDINO pipeline.** Here, in contrast to the DINO pipeline in Figure 8.8, the loss is computed directly on the features without need of further manipulation. This shaves off several large matrices' worth of parameters and simplifies the pipeline, simultaneously making it more stable to train.

$$-\frac{\gamma}{2}\log\det\left(\boldsymbol{I}+\frac{d}{\varepsilon^2}\operatorname{Cov}(\boldsymbol{z}_{\theta_s}(\boldsymbol{X}_g))\right). \tag{8.2.43}$$

Thus, as elaborated in Figure 8.9, the SimDINO pipeline is strictly simpler than the DINO pipeline. We can use a simpler version of the DINO training pipeline to optimize SimDINO. At each timestep $k$, we:

- Subsample $B$ data points from our dataset $\{\boldsymbol{X}_1^{(k)},\dots,\boldsymbol{X}_B^{(k)}\}\subset\mathcal{I}$.
- For each data point $\boldsymbol{X}_b^{(k)}$, sample $M_{\mathrm{glo}}$ global views $v_{b,g}^{(k),i}$ and $M_{\mathrm{loc}}$ local views $v_{b,\ell}^{(k),i}$. Apply the views to $\boldsymbol{X}_b^{(k)}$ to obtain $\boldsymbol{X}_{b,g}^{(k),i}\doteq v_{b,g}^{(k),i}(\boldsymbol{X}_b^{(k)})$ and $\boldsymbol{X}_{b,\ell}^{(k),i}\doteq v_{b,\ell}^{(k),i}(\boldsymbol{X}_b^{(k)})$.
- For each *local* view $\boldsymbol{X}_{b,\ell}^{(k),i}$ compute $\boldsymbol{z}_{\theta_s}(\boldsymbol{X}_{b,\ell}^{(k),i})\doteq(f_{\theta_s}^{\mathrm{ext}}\circ f_{\theta_s})(\boldsymbol{X}_{b,\ell}^{(k),i})$. For each *global* view $\boldsymbol{X}_{b,g}^{(k),i}$ compute $\boldsymbol{z}_{\theta_s}(\boldsymbol{X}_{b,g}^{(k),i})\doteq(f_{\theta_s}^{\mathrm{ext}}\circ f_{\theta_s})(\boldsymbol{X}_{b,g}^{(k),i})$ and $\boldsymbol{z}_{\theta_t}(\boldsymbol{X}_{b,g}^{(k),i})\doteq(f_{\theta_t}^{\mathrm{ext}}\circ f_{\theta_t})(\boldsymbol{X}_{b,g}^{(k),i})$.
- Compute the *surrogate, approximate loss* $\hat{\mathcal{L}}_{\mathrm{SimDINO-st}}^{(k)}$, defined as follows:

$$\hat{\mathcal{L}}_{\mathrm{SimDINO-st}}^{(k)}(\theta_s,\theta_t) \tag{8.2.44}$$

$$\doteq\frac{1}{BM_{\mathrm{glo}}(M_{\mathrm{glo}}+M_{\mathrm{loc}}-1)}\sum_{b=1}^{B}\sum_{i=1}^{M_{\mathrm{glo}}}\left[\sum_{j=1}^{M_{\mathrm{loc}}}d_{\ell^2}(\boldsymbol{z}_{\theta_t}(\boldsymbol{X}_{b,g}^{(k),i}),\boldsymbol{z}_{\theta_s}(\boldsymbol{X}_{b,\ell}^{(k),j}))\right. \tag{8.2.45}$$

$$\left.+\sum_{j=1}^{M_{\mathrm{glo}}}d_{\ell^2}(\boldsymbol{z}_{\theta_t}(\boldsymbol{X}_{b,g}^{(k),i}),\boldsymbol{z}_{\theta_s}(\boldsymbol{X}_{b,g}^{(k),j}))\right] \tag{8.2.46}$$

$$-\frac{\gamma}{M_{\mathrm{glo}}}\sum_{i=1}^{M_{\mathrm{glo}}}R_\varepsilon([\boldsymbol{z}_{\theta_s}(\boldsymbol{X}_{1,g}^{(k),i}),\dots,\boldsymbol{z}_{\theta_s}(\boldsymbol{X}_{B,g}^{(k),i})]) \tag{8.2.47}$$

  where $R_\varepsilon$ is the Gaussian coding rate estimated on finite samples, described in Chapter 5. The gradient of $\hat{\mathcal{L}}_{\mathrm{SimDINO-st}}^{(k)}$ with respect to $\theta_s$ should (again) be computed, under the assumption that $\theta_t$ is constant.

- Update the student parameters $\theta_{\mathrm{s}}$ via an iterative gradient-based optimization algorithm, and update $\theta_{\mathrm{t}}$ via an exponential moving average with decay parameter $\nu^{(k)}$, i.e.,

$$\theta_{\mathrm{s}}^{(k+1)} = \textsc{OptUpdate}^{(k)}(\theta_{\mathrm{s}}^{(k)}; \nabla_{\theta_{\mathrm{s}}}\hat{\mathcal{L}}_{\mathrm{SimDINO-st}}^{(k)}) \tag{8.2.48}$$

$$\theta_{\mathrm{t}}^{(k+1)} = \nu^{(k)}\theta_{\mathrm{t}}^{(k)} + (1-\nu^{(k)})\theta_{\mathrm{s}}^{(k+1)}. \tag{8.2.49}$$

Again, we re-iterate that the gradient is only taken w.r.t. $\theta_{\mathrm{s}}$, treating $\theta_{\mathrm{t}}$ as a constant. Here, note that while the choice of $\nu$ is still a design decision, the hyperparameters $\rho$ and $\tau$ are removed.

### 8.2.5 Evaluation Methodology

There are several ways to evaluate a trained transformer model. We highlight two in this section. Let us define the *center crop* view $v_{\mathrm{cc}} \colon \mathcal{I} \to \mathcal{I}$ which is a *deterministic resized crop*:

- it resizes the image so that the shortest edge is of size $S_{\mathrm{rsz}}$ (similar to random resized crops with area percentage parameter 1);
- then it takes the *center* $S_{\mathrm{cc}} \times S_{\mathrm{cc}}$ crop;

so that the final shape is $(C, S_{\mathrm{cc}}, S_{\mathrm{cc}})$. Notice that the view $v_{\mathrm{cc}}$ is completely deterministic given an input. For an input $\boldsymbol{X}$, we write $\boldsymbol{X}_{\mathrm{cc}} \doteq v_{\mathrm{cc}}(\boldsymbol{X})$. Here $S_{\mathrm{cc}} \leq S_{\mathrm{rsz}}$.

**Linear Probing.** The first, and most architecture-agnostic, way to evaluate an encoder model $\boldsymbol{X} \mapsto \boldsymbol{z}_\theta(\boldsymbol{X})$ is to employ *linear probing*. Linear probing is, in a sentence, running logistic regression on the aggregate features computed by the encoder. This tells us how much semantic information exists in the representations, as well as how easily this information can be extracted. (That is: to what extent do the features of images with different semantics live on different subspaces of the feature space?)

More formally, let us suppose that we want to evaluate the quality and faithfulness of the features of the encoder on image-label data $(\boldsymbol{X}, \boldsymbol{y})$, where there are $N_{\mathrm{cls}}$ classes and $\boldsymbol{y} \in \{0,1\}^{N_{\mathrm{cls}}}$ is a "one-hot encoding" (namely, zeros in all positions except a 1 in the $i$th position if $\boldsymbol{X}$ is in the $i$th class). One way to do this is to solve the logistic regression problem

$$\min_{\boldsymbol{W} \in \mathbb{R}^{N_{\mathrm{cls}} \times d}} \mathbb{E}[\mathrm{CE}(\boldsymbol{y}, \boldsymbol{W}\boldsymbol{z}_\theta(\boldsymbol{X}_{\mathrm{cc}}))]. \tag{8.2.50}$$

More practically, if we have labeled data $\{(\boldsymbol{X}_b, \boldsymbol{y}_b)\}_{b=1}^{B}$, we can solve the *empirical* logistic regression problem (akin to (8.2.28) vs. (8.2.35)) given by

$$\min_{\boldsymbol{W} \in \mathbb{R}^{N_{\mathrm{cls}} \times d}} \frac{1}{B}\sum_{b=1}^{B} \mathrm{CE}(\boldsymbol{y}_b, \boldsymbol{W}\boldsymbol{z}_\theta(\boldsymbol{X}_{b,\mathrm{cc}})). \tag{8.2.51}$$

This problem is a convex optimization problem in $\boldsymbol{W}$, and thus can be solved efficiently via (stochastic) gradient descent or a litany of other algorithms. This linear probe, together with the encoder, may be used as a classifier, and we can evaluate the classification accuracy. The usual practice is to train the model first on a large dataset (such as ImageNet-1K), then train the linear probe on a dataset (such as the training dataset of CIFAR-10), and evaluate it on a third ("holdout") dataset which is drawn from the same distribution as the second one (such as the evaluation dataset of CIFAR-10).

**$k$-nearest Neighbors.** We can also evaluate the performance of the features on classification tasks *without needing to explicitly train a classifier* by using the $k$-nearest neighbor algorithm to get an average predicted label. Namely, given a dataset $\{\boldsymbol{z}_b\}_{b=1}^{B} \subseteq \mathbb{R}^d$, define the $k$-nearest neighbors of another point $\boldsymbol{z} \in \mathbb{R}^d$ as $\mathrm{NN}_k(\boldsymbol{z}, \{\boldsymbol{z}_b\}_{b=1}^{B})$. Using this notation, we can compute the predicted label $\hat{\boldsymbol{y}}_\theta(\boldsymbol{X} \mid \{(\boldsymbol{X}_b, \boldsymbol{y}_b)\}_{b=1}^{B})$ as

$$
\begin{aligned}
&\hat{\boldsymbol{y}}_\theta(\boldsymbol{X} \mid \{(\boldsymbol{X}_b, \boldsymbol{y}_b)\}_{b=1}^{B}) = \mathbf{1}(i^\star) \quad \text{where} \\
&i^\star \doteq \arg\max_{i \in [Q]} \sum_{b=1}^{B} \boldsymbol{y}_b \mathbf{1}[\boldsymbol{z}_\theta(\boldsymbol{X}_{\mathrm{cc},b}) \in \mathrm{NN}_k(\boldsymbol{z}_\theta(\boldsymbol{X}_{\mathrm{cc}}))]. \qquad (8.2.52)
\end{aligned}
$$

Here, $\mathbf{1}(i) \in \Delta_{N_{\mathrm{cls}}}$ is (by an abuse of notation, cf. indicator variables) the one-hot probability vector supported at $i$, i.e., 1 in the $i$th coordinate and 0 elsewhere. That is, this procedure takes the most common label among the $k$ nearest points in feature space. The $k$-nearest neighbor classification accuracy is just the accuracy of this predicted label, namely,

$$
\mathbb{E}_{\boldsymbol{X}, \boldsymbol{y}}[\mathbf{1}(\hat{\boldsymbol{y}}_\theta(\boldsymbol{X} \mid \{(\boldsymbol{X}_b, \boldsymbol{y}_b)\}_{b=1}^{B}) = \boldsymbol{y})] \qquad (8.2.53)
$$

or more commonly its corresponding empirical version, where $(\boldsymbol{X}, \boldsymbol{y})$ ranges over a finite dataset (*not* the existing samples $(\boldsymbol{X}_b, \boldsymbol{y}_b)$ which are used for the $k$ neighbors).

**Fidelity of the Attention Maps.** Another way to check the performance of the representations, for a transformer-based encoder, is to examine the fidelity of the attention maps $\boldsymbol{A}^{L,k} \in \mathbb{R}^{n \times n}$ as defined in Equation (8.2.21), at the last layer $L$, and given by the following pipeline:

$$
\boldsymbol{X} \mapsto \cdots \mapsto \boldsymbol{Z}^{L-1} = [\underbrace{\boldsymbol{z}_1^{L-1}}_{\text{class token}}, \underbrace{\boldsymbol{z}_2^{L-1} \ldots, \boldsymbol{z}_n^{L-1}}_{\text{patch tokens}}] \mapsto \boldsymbol{A}^{k,L} = \begin{bmatrix} \boldsymbol{A}_{1,1}^{k,L} & \boldsymbol{A}_{1,2:}^{k,L} \\ \boldsymbol{A}_{2:,1}^{k,L} & \boldsymbol{A}_{2:,2:}^{k,L} \end{bmatrix}. \qquad (8.2.54)
$$

In particular, we examine what the attention maps for a given input reveal about the salient objects in the input image, i.e., which parts of the image provide the most globally-relevant information to the class token. One particular way to do this is to examine the component of the attention map where the class token is extracted as the query and removed from the value matrix, i.e., $\boldsymbol{A}^{k,L}_{2:,1} \in \mathbb{R}^{1\times(n-1)} = \mathbb{R}^{1\times N}$ or its transpose $\boldsymbol{a}^{k,L} = (\boldsymbol{A}^{k,L}_{2:,1})^\top \in \mathbb{R}^N$. Notice that this vector $\boldsymbol{a}^{k,L}$, which we label as the "*saliency vector* at the $k$th attention head at layer $L$," has a value for every patch, $1, \ldots, N$, and we use this value to describe how relevant each patch is toward the global information. In particular for visualization's sake we create a new image where each patch is replaced by its corresponding value in the saliency vector, showcasing the contribution of each patch; we call this image the "*saliency map* at the $k$th attention head at layer $L$". To visualize the total relevance of each patch toward the global information across all heads, we can average the saliency vector, i.e., $\tilde{\boldsymbol{a}}^L \doteq \frac{1}{K}\sum_{k=1}^K \boldsymbol{a}^{k,L}$ and expand into the *average saliency map*. The average saliency maps should highlight the relevant parts of the input image.

**Object Detection and Segmentation.** We can evaluate how the representations capture the fine-grained (i.e., smaller or more detailed) properties of the input by using them for *semantic segmentation*. Roughly, this means that we use the features to construct bounding boxes for all objects in the input. There are several ways to do this, and several ways to score the resulting bounding boxes compared to ground truth. Each combination of methods corresponds to a particular segmentation metric. We do not formally describe them here as they are not particularly insightful, but the DINO paper [CTM+21] and DINOv2 paper [ODM+23] contain references to all metrics that are used in practice.

### 8.2.6 Experimental Setup and Results

Since SimDINO is directly built upon DINO, we compare the optimal settings for DINO as given by their original paper [CTM+21] with the same settings applied to SimDINO [WZP+25] for a fair comparison.

**Objective Function.** We use 10 local views (i.e., $M_{\text{loc}} = 10$) of resolution $96\times 96$ (i.e., $S_{\text{loc}} = 96$) and 2 global views (i.e., $M_{\text{glo}} = 2$) of resolution $224\times 224$ (i.e., $S_{\text{glo}} = 224$) for all experiments. The corresponding portions of the original images cropped for local and global views are $p_{\text{loc}} \in [\frac{1}{20}, \frac{3}{10}]$ and $p_{\text{glo}} \in [\frac{3}{10}, 1]$ (chosen randomly per-view). The smaller edge size within the resized crops is $S_{\text{rsz}} = 256$, and the center crop (evaluation) view edge size is $S_{\text{cc}} = 224$. All of these settings apply to both DINO and SimDINO.

**Model Architecture.** For all inputs, we set the patch size to be $16 \times 16$ (namely, $P_H = P_W = 16$). We use the small, base, and large models of the ViT [DBK+21] architecture as the embedding and backbone. The feature extractor is a three-layer MLP with a hidden size of 2048 and an output dimension of 256, followed by an $\ell^2$-normalization, as specified in Section 8.2.3. For DINO architectures (i.e., not SimDINO architectures), the DINO head $\boldsymbol{W}$ is a matrix in $\mathbb{R}^{65536\times 256}$, and the parameter $\boldsymbol{\mu}$ is a vector in $\mathbb{R}^{65536}$.

**Datasets and Optimization.** For pre-training, both our DINO reproduction and SimDINO use the ImageNet-1K dataset across all methods. We use AdamW [LH17] as the optimizer, which is a very standard choice. We follow the following hyperparameter recommendations:

- The batch size is $B = 1024$.
- The learning rate (for AdamW and the student model) has "base" value $2 \times 10^{-3}$. In the first 10 epochs the learning rate linearly increases from 0 to the base value (i.e., at the $i$th epoch the learning rate is $(i/10) \cdot 2 \times 10^{-3}$, for $1 \leq i \leq 10$). Then over the next 90 epochs the learning rate decays via a so-called *cosine schedule* back down to 0. The definition of a cosine schedule is given in many places, including PyTorch documentation, and it is commonly used when training deep vision models.
- The weight decay (the W in AdamW) follows a cosine schedule from 0.04 to 0.4 over training.
- The EMA rate $\nu$ follows a cosine schedule from 0.996 to 1.0 over training. Specifically for DINO, the centering EMA rate $\rho$ is fixed at 0.9.
- Specifically for DINO, the teacher temperature $\tau_{\mathrm{t}}$ is fixed at 0.1, while the student temperature $\tau_{\mathrm{s}}$ linearly increases from 0.04 to 0.07 during the first 30 epochs and is fixed at 0.07 thereafter.

We use some (essentially information-preserving) data augmentations, such as flips, color jittering, Gaussian blur, and solarization, for each seen image during training, before taking the local and global views. The exact hyperparameters governing these are not listed here, but are referenced in the DINO paper [CTM+21].

For linear probing, the linear probe is usually trained using the AdamW optimizer with learning rate $2 \times 10^{-4}$, weight decay 0.01, and batch size 512, but these are often modified on a case-by-case basis to minimize the loss.

**Evaluation Results.** In terms of downstream classification performance, we obtain the performance in Table 8.1. We observe that the performance of SimDINO is much higher than that of DINO under fair comparison. Also, it is much more stable: the prescribed settings of DINO cannot train a ViT-L(arge) model. On the other hand, Figure 8.10 shows visualizations of the average saliency maps in DINO and our simplified DINO, observing that the saliency maps look quite similar across models, indicating that the models learn features which are at least as good at capturing fine-grained details. The segmentation and object detection performances in Table 8.2 confirm this claim quantitatively, where SimDINO features show substantive improvement over those of DINO.

Table 8.1: **Classification performance** on hold-out test data for DINO and SimDINO, using both $k$-nearest neighbor accuracy ($k = 20$) and linear probing. At the same number of iterations (100), SimDINO is clearly better in terms of performance, and is more stable (the DINO training running on ViT-L backbone with the provided settings has very unstable optimization and obtains NaN loss in short order). We also compare to other standout methods, namely SwAV and MoCov3, which DINO was built on.

| Method | Model | Epochs | 20-NN | Linear Probing |
|---|---|---|---|---|
| DINO | ViT-B | 100 | 72.9 | 76.3 |
| SimDINO | ViT-B | 100 | **74.9** | **77.3** |
| DINO | ViT-L | 100 | – | – |
| SimDINO | ViT-L | 100 | **75.6** | **77.4** |
| SwAV | ViT-S | 800 | 66.3 | 73.5 |
| MoCov3 | ViT-B | 300 | – | 76.7 |

Table 8.2: **Segmentation performance** of pre-trained DINO and SimDINO models on COCO val2017 [LMB+14], a segmentation dataset which contains object location metadata. We do not train on COCO, merely using the pre-trained embedding and backbone, and the bounding boxes are extracted from the features via a method called MaskCut [WGY+23]. Nevertheless, SimDINO surpasses DINO at object detection and segmentation under fair comparison, and even surpasses DINO with smaller patch size (side length 8 instead of 16). Smaller patch sizes are known to help performance, especially with detection and segmentation tasks, so this result is quite surprising and encouraging.

| | | Detection ↑ | | | Segmentation ↑ | | |
|---|---|---|---|---|---|---|---|
| Method | Model | $AP_{50}$ | $AP_{75}$ | AP | $AP_{50}$ | $AP_{75}$ | AP |
| SimDINO | ViT-L/16 | **5.4** | 1.9 | 2.4 | 4.5 | 1.4 | 1.9 |
| SimDINO | ViT-B/16 | 5.2 | **2.0** | **2.5** | **4.7** | **1.5** | **2.0** |
| DINO | ViT-B/16 | 3.9 | 1.5 | 1.8 | 3.1 | 1.0 | 1.4 |
| DINO | ViT-B/8 | 5.1 | 2.3 | 2.5 | 4.1 | 1.3 | 1.8 |

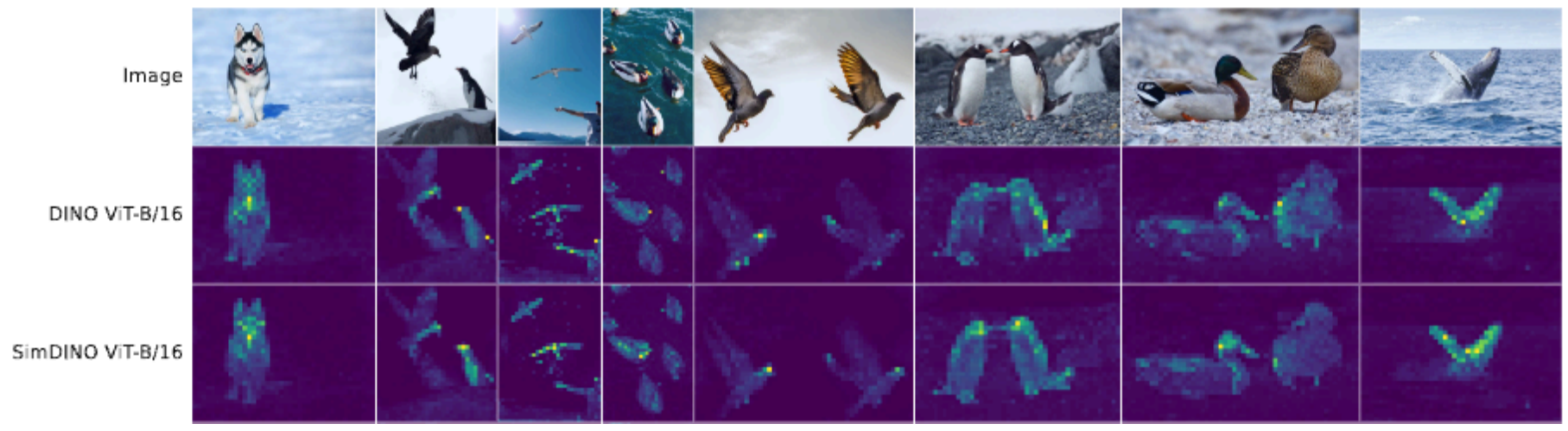

Figure 8.10: **A qualitative comparison of saliency maps** generated by DINO *(middle row)* and by SimDINO *(bottom row)*. For each image, we compute and display the average saliency map in the last layer $L$. The saliency maps are similar across models, meaning that all models converge to a similar notion of what objects are important. Note that although $X_{\mathrm{eval}}$ is a square image, it is interpolated back into rectangular shape to make this visualization.

## 8.3 Weakly Supervised Image Representation via Text Binding

Another influential contrastive learning approach departs from purely visual comparisons and instead leverages the natural alignment between images and language. Rather than defining similarity solely through multiple views of the same image, this line of work exploits the fact that many images in the wild are accompanied by textual descriptions that provide weak semantic supervision. A prominent example of this paradigm is CLIP (Contrastive Language–Image Pretraining) [RKH+21a]. In CLIP, images and their corresponding captions are treated as positive pairs, while mismatched image–text pairs are treated as negatives. The learning objective encourages the representation of an image to be close to the representation of its associated caption, and far from captions of other images. By grounding visual representations in natural language, CLIP learns semantically rich features that capture high-level concepts beyond appearance alone.

### 8.3.1 Data

The data that we will use to explore the CLIP methodology are *text-image pairs*, rather than images alone. The original CLIP work constructed a large web-scale corpus of approximately 400 million text-image pairs collected from publicly available sources on the Internet. Each sample consists of an image and an associated natural-language string (e.g., a caption, title, or short description) that co-occurs with the image on the web, providing a weak and noisy form of supervision. To encourage broad coverage of visual concepts, the CLIP work describes building this dataset by searching for pairs whose text matches one of roughly 500,000 queries, and approximately balancing the results by including up to 20,000 pairs per query. This dataset is not publicly released, so most community replications and extensions of CLIP rely instead on open web-scale datasets of text-image pairs. The most common of which is the LAION [SBV+22] family: LAION-400M provides 400 million CLIP-filtered pairs, and LAION-5B scales this recipe to 5.85 billion pairs.

Because raw web-scraped pairs are often extremely noisy (e.g., images that do not match their associated text, boilerplate alt-text, spam, duplicates, or unsafe content), CLIP-style datasets are typically constructed with a post-processing pipeline that filters and cleans candidates before training. Typical methods include basic quality filters (e.g., dropping pairs with extremely short text or very small or corrupted images), language identification to retain captions in a target language, de-duplication, and the removal of potentially malicious content.

### 8.3.2 The CLIP Task and Objective

Our goal is to learn a good representation of the data. In the CLIP setting, however, the notion of "similarity" is not induced by two augmented views of the same image, but by the weak semantic supervision provided by natural-language descriptions. Concretely, we would like an image representation to be close to the representation of its associated caption, and far from captions of other images. Equivalently, if two images admit similar textual descriptions, their learned visual features should be similar, whereas images described by unrelated text should be separated in latent space.

We formalize the notion of "similarity" between representations using *cosine similarity*. Concretely, let $f_\theta$ be an image encoder and $g_\phi$ be a text encoder (whose architecture will be elaborated in the next section), both mapping into a common latent space $\boldsymbol{Z} \in \mathbb{R}^d$. Given an image $\boldsymbol{X} \in \mathcal{I}$ and a text string $\boldsymbol{T} \in \mathcal{T}$, we form normalized embeddings

$$\boldsymbol{z}_\theta^I(\boldsymbol{X}) \doteq \frac{f_\theta(\boldsymbol{X})}{\|f_\theta(\boldsymbol{X})\|_2}, \qquad \boldsymbol{z}_\phi^T(\boldsymbol{T}) \doteq \frac{g_\phi(\boldsymbol{T})}{\|g_\phi(\boldsymbol{T})\|_2}. \tag{8.3.1}$$

Then the cosine similarity between embeddings is defined as:

$$s(\boldsymbol{X}, \boldsymbol{T}) = \left\langle \boldsymbol{z}_\theta^I(\boldsymbol{X}), \boldsymbol{z}_\phi^T(\boldsymbol{T}) \right\rangle \in [-1, 1]. \tag{8.3.2}$$

Therefore, our objective is to maximize the cosine similarity between matched pairs and minimize it between mismatched pairs. To achieve that, we can use a simple symmetric cross-entropy loss as discussed in Section 7.4.2. Concretely, given a mini-batch of $n$ text-image pairs $(\boldsymbol{X}_i, \boldsymbol{T}_i)_{i=1}^n$, we can define the loss as:

$$\mathcal{L}_{\mathrm{CLIP}} = -\frac{1}{n}\left(\sum_{i=1}^{n} \frac{\exp(s(\boldsymbol{X}_i, \boldsymbol{T}_i)/\tau)}{\sum_{k=1}^{n} \exp(s(\boldsymbol{X}_i, \boldsymbol{T}_k)/\tau)} + \sum_{i=1}^{n} \frac{\exp(s(\boldsymbol{T}_i, \boldsymbol{X}_i)/\tau)}{\sum_{k=1}^{n} \exp(s(\boldsymbol{T}_i, \boldsymbol{X}_k)/\tau)}\right), \tag{8.3.3}$$

where $\tau > 0$ is a temperature parameter that controls the sharpness of the softmax function.

### 8.3.3 Architecture: Vision Tower

The objective of CLIP is to learn both image and text representations in the same latent space. Accordingly, its architecture trains a vision encoder $f_\theta : \mathcal{I} \to \mathbb{R}^d$ and a text encoder $g_\phi : \mathcal{T} \to \mathbb{R}^d$ , also known as the "dual-tower architecture".

In terms of architecture, most vision backbones can be used: for instance, convolutional networks such as ResNet [HZR+16a], as well as the Vision Transformer [DBK+21] (detailed in Section 8.2.3) are all valid choices for CLIP. In practice, the Vision Transformer is more commonly used due to better performance and versatility. Similarly to DINO, we take the class token feature $\boldsymbol{z}_{\text{cls}}$ as the representation for each image: $f_\theta(\boldsymbol{X}) = \boldsymbol{z}_{\text{cls}}$. This is because the class token is meant to aggregate global information about the entire image and is commonly used for downstream prediction tasks.

### 8.3.4 Architecture: Text Tower

The CLIP text tower $g_\phi$ typically adopts a Transformer-based architecture similar to the Vision Transformer described earlier. The main difference lies in the embedding layer, which is used to convert the text sequence into a sequence of latent representations in the latent space $\boldsymbol{Z} \in \mathbb{R}^d$.

**Embedding.** Given a text sequence $\boldsymbol{T} \in \mathcal{T}$ (e.g., a sentence or document), we embed it as a sequence of tokens in $\mathbb{R}^d$ using the map $g_\phi^{\text{emb}}$, as follows.

1. First, we tokenize $\boldsymbol{T}$ into a sequence of discrete symbols (subwords/characters) from a vocabulary $\mathcal{V}$ of size $|\mathcal{V}|$. Concretely, a tokenizer $t^{\text{tok}} : \mathcal{T} \to \mathcal{V}^N$ produces a length-$N$ token sequence[5] $(\tau_1, \dots, \tau_N)$, where $\tau_i \in \mathcal{V}$.

2. We convert tokens to integer indices by a vocabulary map $\mathcal{V} \to \{1, \dots, |\mathcal{V}|\}$, yielding an index sequence $(s_1, \dots, s_N)$. We place these indices into a matrix $\boldsymbol{S} \in \{0, 1\}^{|\mathcal{V}| \times N}$, where the $i$-th column is the one-hot vector $\boldsymbol{e}_{s_i}$.

3. We then perform the following operation on $\boldsymbol{S}$ to project it to $\mathbb{R}^{d \times (N+1)}$:

$$\boldsymbol{S} \mapsto [\boldsymbol{z}_{\text{cls}}^1, \boldsymbol{W}^{\text{tok}} \boldsymbol{S}] + \boldsymbol{E}^{\text{pos}}. \tag{8.3.4}$$

   This is the final output of $g_\phi^{\text{emb}}$. Here we have three trainable parameters $\boldsymbol{W}^{\text{tok}}$, $\boldsymbol{z}_{\text{cls}}^1$, and $\boldsymbol{E}^{\text{pos}}$ whose purpose is as follows:

   - $\boldsymbol{W}^{\text{tok}} \in \mathbb{R}^{d \times |\mathcal{V}|}$ is a matrix whose $j$-th column is the learned embedding vector for the $j$-th vocabulary item. Thus $\boldsymbol{W}^{\text{tok}} \boldsymbol{S} \in \mathbb{R}^{d \times N}$ maps each discrete token to a continuous feature in $\mathbb{R}^d$.
   - $\boldsymbol{z}_{\text{cls}}^1 \in \mathbb{R}^d$ is a special *classification* (or *summary*) token. Heuristically, it is meant to aggregate global information about the entire sequence and is commonly used for downstream prediction tasks.

[5] In practice one often augments this sequence with special tokens (e.g., ⟨bos⟩ and ⟨eos⟩ tokens that are used to indicate the beginning and end of the sequence) and may pad shorter sequences, but we keep the notation generic and treat these as elements of $\mathcal{V}$.

- $\boldsymbol{E}^{\text{pos}} \in \mathbb{R}^{d \times n}$ is a *positional encoding* which distinguishes tokens at different positions. Since self-attention is permutation-equivariant in the token dimension, $\boldsymbol{E}^{\text{pos}}$ injects order information so that the overall map is not invariant to permutations of the tokens.

All parameters $\boldsymbol{z}_{\text{cls}}^1, \boldsymbol{W}^{\text{tok}}, \boldsymbol{E}^{\text{pos}}$ are contained in the parameter set $\theta$. The vocabulary $\mathcal{V}$ and tokenizer $t^{\text{tok}}$ are fixed.

Other components of the text tower, such as the backbone and the task-specific head, follow similar design choices as the image tower and are omitted for brevity. It is also worth noting that follow-up works have demonstrated substantial flexibility in the choice of CLIP's text encoder: fixed text representations contained in large language models, and even simple bag-of-words features, can also be effective. We discuss these variants in Section 8.3.8.

### 8.3.5 Optimization Strategy

We train CLIP using a standard end-to-end stochastic optimization procedure (e.g. SGD with momentum, AdamW). Below we present a generic process.

At each timestep $k$, we:

- **Subsample paired data.** Sample a minibatch of $n$ paired examples

$$\{(\boldsymbol{X}_i^{(k)}, \boldsymbol{T}_i^{(k)})\}_{i=1}^n \subseteq \mathcal{X} \times \mathcal{T},$$

where $\boldsymbol{X}_i^{(k)}$ is an image and $\boldsymbol{T}_i^{(k)}$ is its associated text.

- **Compute latent representations.** For each pair $(\boldsymbol{X}_i^{(k)}, \boldsymbol{T}_i^{(k)})$, compute normalized embeddings using the vision and text encoders $f_\theta$ and $g_\phi$:

$$z_\theta^I(\boldsymbol{X}_i^{(k)}) \doteq \frac{f_\theta(\boldsymbol{X}_i^{(k)})}{\|f_\theta(\boldsymbol{X}_i^{(k)})\|_2}, \qquad z_\phi^T(\boldsymbol{T}_i^{(k)}) \doteq \frac{g_\phi(\boldsymbol{T}_i^{(k)})}{\|g_\phi(\boldsymbol{T}_i^{(k)})\|_2}. \tag{8.3.5}$$

- **Compute CLIP loss.** Compute the CLIP loss as defined in (8.3.3):

$$\hat{\mathcal{L}}_{\text{CLIP}}^{(k)} \doteq -\frac{1}{n}\left(\sum_{i=1}^n \frac{\exp(s(\boldsymbol{X}_i^{(k)}, \boldsymbol{T}_i^{(k)})/\tau)}{\sum_{k=1}^n \exp(s(\boldsymbol{X}_i^{(k)}, \boldsymbol{T}_k^{(k)})/\tau)} + \sum_{i=1}^n \frac{\exp(s(\boldsymbol{T}_i^{(k)}, \boldsymbol{X}_i^{(k)})/\tau)}{\sum_{k=1}^n \exp(s(\boldsymbol{T}_i^{(k)}, \boldsymbol{X}_k^{(k)})/\tau)}\right). \tag{8.3.6}$$

- **Compute gradients.** Compute the gradients of the stochastic loss with respect to both parameter sets:

$$\nabla_\theta \hat{\mathcal{L}}_{\text{CLIP}}^{(k)}(\theta, \phi), \qquad \nabla_\phi \hat{\mathcal{L}}_{\text{CLIP}}^{(k)}(\theta, \phi). \tag{8.3.7}$$

- **Update parameters.** Apply one step of an optimization algorithm to $(\theta, \phi)$, yielding the iteration

$$(\theta^{(k+1)}, \phi^{(k+1)}) \doteq \textsc{OptUpdate}^{(k)}\left((\theta^{(k)}, \phi^{(k)}); \nabla_\theta \hat{\mathcal{L}}_{\text{CLIP}}^{(k)}, \nabla_\phi \hat{\mathcal{L}}_{\text{CLIP}}^{(k)}\right). \tag{8.3.8}$$

### 8.3.6 Evaluation Methodology

Since CLIP is architecture-agnostic, most standard evaluation protocols for vision and language encoders can be applied. For example, one can perform linear probing by training a lightweight classification head on top of the vision encoder to assess the quality of the learned representations in image classification tasks. Further, one can perform zero-shot classification without any additional training thanks to the dual-tower architecture of CLIP.

Recall that in a standard $n$-way image classification setup, we are given a dataset of images $\{\boldsymbol{X}_i\}_{i=1}^N \subseteq \mathcal{I}$, each annotated with one of $n$ class labels from the set $\mathcal{L} \doteq \{\ell_1, \ldots, \ell_n\}$ (e.g., $\ell_j \in \{\text{cat}, \text{dog}, \ldots\}$). Our goal is to build a prediction pipeline $P$, leveraging a pre-trained model, such that $P : \mathcal{I} \to \{1, \ldots, n\}$ maps each image $\boldsymbol{X}_i$ to a predicted class index $\hat{y}_i \doteq P(\boldsymbol{X}_i)$. Writing the ground-truth class label as its index $y_i \in \{1, \ldots, n\}$, the (top-1) classification accuracy of $P$ on this dataset is then computed as

$$\frac{1}{N} \sum_{i=1}^{N} \mathbf{1}\{\hat{y}_i = y_i\}.$$

In CLIP, we can use the text encoder $g_\phi$ to build a text dictionary $\boldsymbol{D} \in \mathbb{R}^{n \times d}$, whose $j$-th row is the normalized latent representation of the $j$-th class label. Concretely, for each class $\ell_j \in \mathcal{L}$, we construct a textual description (prompt) $\boldsymbol{T}_j$ (e.g., "a photo of a $\ell_j$"), and compute its normalized embedding $\boldsymbol{z}_j^T \doteq \frac{g_\phi(\boldsymbol{T}_j)}{\|g_\phi(\boldsymbol{T}_j)\|_2} \in \mathbb{R}^d$. Stacking these row-wise yields

$$\boldsymbol{D} \doteq \begin{bmatrix} (\boldsymbol{z}_1^T)^\top \\ \vdots \\ (\boldsymbol{z}_n^T)^\top \end{bmatrix} \in \mathbb{R}^{n \times d},$$

Given an image $\boldsymbol{X}_i \in \mathcal{I}$, we compute its image representation via the vision encoder and normalize it as

$$\boldsymbol{z}_i^I \doteq \frac{f_\theta(\boldsymbol{X}_i)}{\|f_\theta(\boldsymbol{X}_i)\|_2} \in \mathbb{R}^d.$$

We can then compute a similarity vector

$$\boldsymbol{d}_i \doteq \boldsymbol{D}\, \boldsymbol{z}_i^I \in \mathbb{R}^n,$$

whose $j$-th entry $\boldsymbol{d}_i[j]$ is the cosine similarity between the $i$-th image and the $j$-th class label in the shared latent space. By the CLIP training objective, images are encouraged to be more similar to their matching texts than to mismatched ones; hence a larger cosine similarity indicates closer alignment between $\boldsymbol{X}_i$ and class $\ell_j$. Therefore, we predict the class index by

$$\hat{y}_i \doteq \arg \max_{j \in \{1, \ldots, n\}} \boldsymbol{d}_i[j].$$

Finally, we can calculate the classification accuracy as mentioned above.

This zero-shot decision protocol is not limited to classification, but also applies naturally to *retrieval* tasks, where the goal is to return the most relevant items from a large database given a query (e.g., retrieve the most relevant images for a text query, or the most relevant texts for an image query). For instance, in text-to-image retrieval, instead of building a dictionary from class-name prompts, we build an *image dictionary* by encoding and normalizing all database images $\{\boldsymbol{X}_m\}_{m=1}^{M}$, stacking their embeddings into $\boldsymbol{D} \in \mathbb{R}^{M\times d}$. Given a query text $\boldsymbol{T}$, we compute its normalized embedding $\boldsymbol{z}^T \doteq g_\phi(\boldsymbol{T})/\|g_\phi(\boldsymbol{T})\|_2$, form similarity scores $\boldsymbol{d} \doteq \boldsymbol{D}\boldsymbol{z}^T \in \mathbb{R}^M$, and then rank images by these scores (equivalently, return the top-$K$ indices).

### 8.3.7 Experimental Setup and Results

Next, we present results from the original CLIP paper comparing CLIP to a supervised ResNet baseline on several classic classification benchmarks. Across all benchmarks, CLIP is evaluated using the zero-shot classification protocol described in Section 8.3.6. Full details of the ResNet experimental setup are provided in [HZR+16a]; here, we briefly introduce the evaluation datasets and summarize CLIP's training and evaluation settings relevant to these comparisons.

**Evaluation Data.**

- **CIFAR-10 [KH+09].** CIFAR-10 is a widely used small-scale image classification benchmark consisting of 60,000 RGB images of size $32 \times 32$ across 10 classes (6,000 images per class). It is typically split into 50,000 training images and 10,000 test images.

- **CIFAR-100 [KH+09].** CIFAR-100 has the same overall size and image resolution as CIFAR-10 (60,000 RGB $32\times 32$ images with a 50,000/10,000 train/test split), but it is more challenging because it contains 100 fine-grained classes (600 images per class). The 100 classes are additionally grouped into 20 superclasses.

- **PASCAL VOC 2007 (VOC2007) [EGW+10].** PASCAL VOC 2007 is a classic computer vision benchmark for object recognition, supporting tasks such as image classification, object detection, and segmentation. It focuses on 20 object categories (e.g., person, car, dog).

- **ImageNet-1K [DDS+09a].** ImageNet-1K is a subset of ILSVRC 2012 containing 1,000 object classes. This subset includes 1,281,167 training images, 50,000 validation images, and 100,000 test images, and has played a central role in large-scale visual recognition and pretraining.

| Backbone | Variant | CIFAR-10 | CIFAR-100 | VOC2007 | ImageNet |
|---|---|---|---|---|---|
| CLIP-ViT | B/32 | 95.1 | 80.5 | 87.7 | 76.1 |
| CLIP-ViT | B/16 | 96.2 | 83.1 | 89.2 | 80.2 |
| CLIP-ViT | L/14 | 98.0 | 87.5 | 89.6 | 83.9 |
| ResNet | 50 | 91.8 | 74.5 | 83.8 | 74.3 |
| ResNet | 101 | 93.0 | 77.2 | 84.4 | 75.8 |

Table 8.3: Performance on downstream datasets for various CLIP backbones and ResNet baselines.

**Model Setup.** In the CLIP models demonstrated here, Vision Transformer is adopted as the vision encoder $f_\theta$ and is instantiated as one of three ViT variants: ViT-B/32, ViT-B/16, or ViT-L/14. Here, B ("Base") and L ("Large") specify the transformer width/depth (e.g., ViT-B typically uses 12 transformer blocks with embedding dimension $d = 768$ and 12 attention heads, whereas ViT-L uses 24 blocks with $d = 1024$ and 16 heads), while the suffix /32, /16, and /14 denotes the *patch size* in pixels used to patchify the input image . Smaller patch sizes (e.g., 16 or 14) yield more tokens per image at a fixed resolution, increasing compute but often improving accuracy due to finer spatial granularity.

**Optimization Setup.** All ViT-based CLIP models are trained for 32 epochs using the Adam optimizer with decoupled weight decay (i.e., AdamW-style regularization), where weight decay is applied to all parameters except gains and biases. The learning rate is decayed using a cosine schedule. The learnable temperature parameter $\tau$ (used to scale the contrastive logits) is initialized to the equivalent of 0.07, and is clipped to prevent scaling the logits by more than 100, which is found necessary for training stability. Finally, training uses a very large mini-batch size of 32,768.

**Results.** Table 8.3 reports zero-shot top-1 accuracy of three CLIP ViT variants (B/32, B/16, L/14) and compares them to supervised ResNet-50 and ResNet-101 baselines. Across all four datasets, CLIP consistently outperforms the supervised ResNets. Scaling the CLIP backbone further yields additional gains, with ViT-L/14 reaching 98.0 on CIFAR-10, 87.5 on CIFAR-100, 89.6 on VOC2007, and 83.9 on ImageNet. Nevertheless, this table should be interpreted as a comparison between two *systems* rather than a controlled backbone ablation: CLIP benefits from large-scale image–text pretraining and is evaluated via zero-shot prompting, whereas the ResNet baselines rely on supervised pretraining and require labeled data to fit a probe on each benchmark. Thus, the consistent margins primarily demonstrate the strength of CLIP's transferable representations under the zero-shot protocol, rather than establishing a purely architecture- or data-matched superiority of ViT over ResNet.

### 8.3.8 Simplified Extension: LIFT

While effective, CLIP exhibits several limitations. First, jointly training both a text encoder and an image encoder from scratch is computationally expensive, often requiring extremely large batches and massive datasets to reach strong alignment and downstream transfer. Second, models trained in this way can struggle with compositional understanding—capturing word order in text, spatial layout in images, and object–attribute or object–object relations—partly because retrieval-style contrastive supervision can reward shortcut solutions that downweight fine-grained compositional features. LIFT (Language–Image alignment with a Fixed Text encoder) [YWZ+25] revisits a core assumption underlying these pipelines: that optimal alignment demands joint end-to-end training of both encoders:

https://jingfeng0705.github.io/LIFT/.

Instead, LIFT leverages the observation that modern large language models (LLMs) already produce highly informative text embeddings. Concretely, LIFT fixes a strong pretrained text encoder (e.g., one derived from or fine-tuned on an LLM), computes text embeddings offline, and trains only the image encoder to align to these fixed targets.

**Architecture.** LIFT retains the same dual-tower CLIP formulation in terms of a vision encoder and a text encoder that map into a shared latent space, but crucially differs in that the text tower is *fixed* and can be instantiated by any strong LLM-based text encoder $g_\phi : \mathcal{T} \to \mathbb{R}^d$ to produce text embeddings, where $\phi$ is not updated during training. In our implementation, we adopt the NV-Embed-V2 text encoder, for which the embedding dimension is $d = 4096$. Since the text tower $g_\phi$ is fixed, we can pre-compute all text embeddings $\{g_\phi(\boldsymbol{T})\}$ offline and, during training, only compute gradients and update the image encoder $f_\theta$ online. The image tower $f_\theta : \mathcal{I} \to \mathbb{R}^d$ follows the same structure as in Section 8.3.3, with the projection head output dimension set to $d$ to match the dimension of the fixed text embedding space. Apart from fixing $g_\phi$ and matching the projection dimension, the remaining components—including the contrastive loss and the optimization procedure—follow the same design principles as CLIP.

**Evaluation Methodology.** Compositional understanding is a known limitation of CLIP. We evaluate LIFT and CLIP on seven SugarCrepe [HZM+23] tasks, where each image $\boldsymbol{X}$ is paired with a *positive* (correct) caption $\boldsymbol{T}_{\text{pos}}$ and a *negative* caption $\boldsymbol{T}_{\text{neg}}$ constructed by adding, replacing, or swapping an object, attribute, or relation in $\boldsymbol{T}_{\text{pos}}$. See Figure 8.11 for some examples. Models are asked to identify the correct caption by comparing caption–image cosine similarities. Formally, for each sample $i \in \{1, \ldots, N\}$, we compute the normalized embeddings

$$\boldsymbol{z}_i^I \doteq \frac{f_\theta(\boldsymbol{X}_i)}{\|f_\theta(\boldsymbol{X}_i)\|_2}, \qquad \boldsymbol{z}_{i,\text{pos}}^T \doteq \frac{g_\phi(\boldsymbol{T}_{i,\text{pos}})}{\|g_\phi(\boldsymbol{T}_{i,\text{pos}})\|_2}, \qquad \boldsymbol{z}_{i,\text{neg}}^T \doteq \frac{g_\phi(\boldsymbol{T}_{i,\text{neg}})}{\|g_\phi(\boldsymbol{T}_{i,\text{neg}})\|_2}.$$

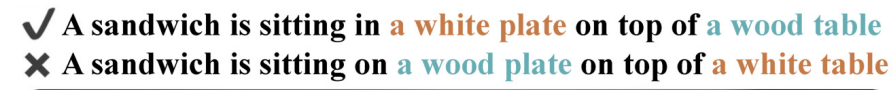

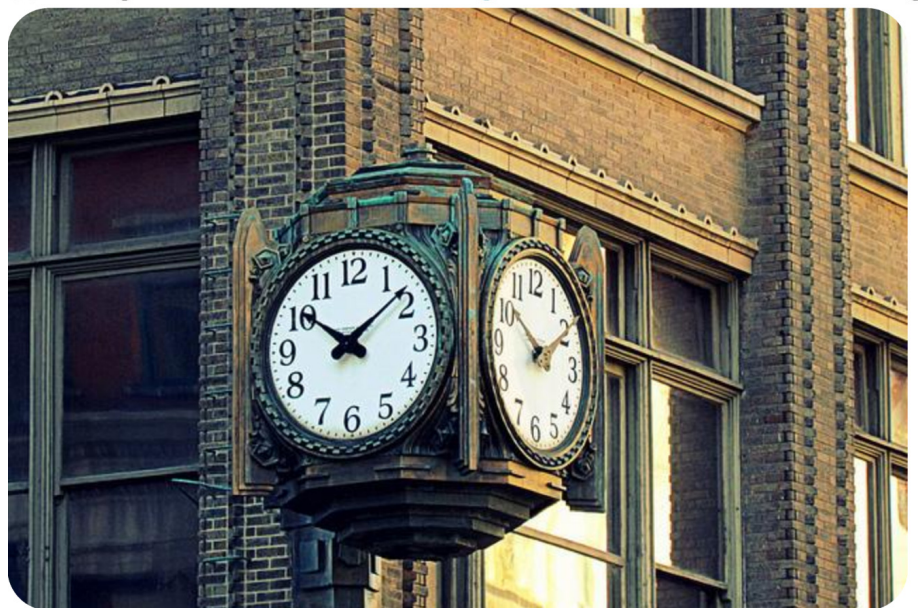

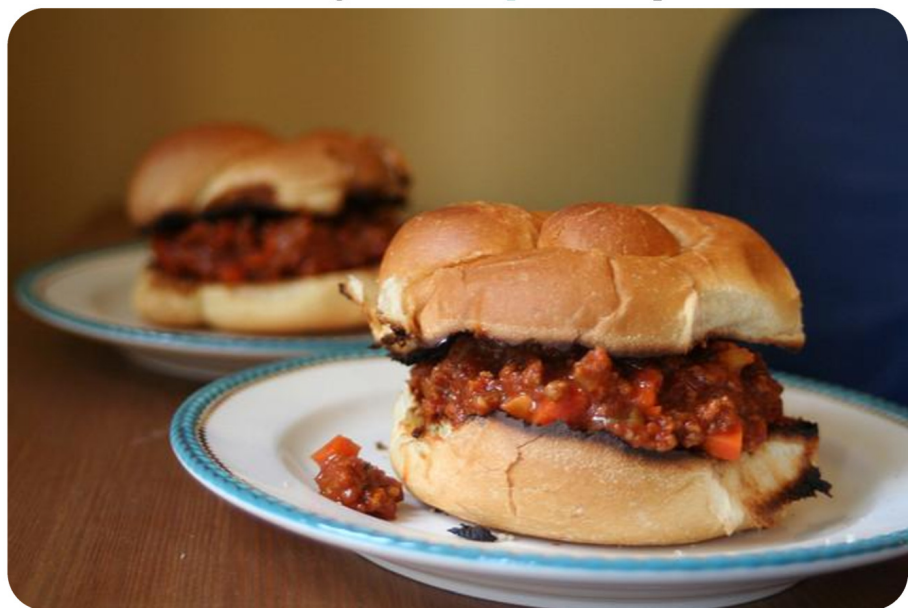

Figure 8.11: The original captions (top) and their negative counterparts (bottom) from two SugarCrepe tasks: `replace relation` (left) and `swap attribute (right)`.

We then compute similarity scores via cosine similarities:

$$s_{i,\text{pos}} \doteq \langle \boldsymbol{z}_i^I, \boldsymbol{z}_{i,\text{pos}}^T \rangle, \qquad s_{i,\text{neg}} \doteq \langle \boldsymbol{z}_i^I, \boldsymbol{z}_{i,\text{neg}}^T \rangle.$$

The model is considered correct on sample $i$ if $s_{i,\text{pos}} > s_{i,\text{neg}}$, and the overall compositional accuracy is

$$\frac{1}{N}\sum_{i=1}^{N} \mathbf{1}\{s_{i,\text{pos}} > s_{i,\text{neg}}\}.$$

**Experimental Results.** As shown in Table 8.4, when trained on DataComp-1B, LIFT outperforms CLIP on all seven tasks with a **6.8%** average accuracy gain. LIFT achieves significant gains on `add attribute`, `replace attribute`, and `replace relation` tasks. These improvements are strong evidence that LLM-based text encoders $g_\phi$ capture more informative text representations that enable more accurate modeling of object–attribute associations and object–object relations. On the other hand, we can see LIFT shows relatively low accuracy on `swap object` and `swap attribute` compared to other SugarCrepe tasks. We attribute this limitation to the contrastive learning objective, which primarily focuses on aligning lower-order statistics. Addressing this challenge requires exploring more refined information-theoretic measures for language-image alignment, a key direction for future work.

## 8.4 Supervised Image Representation via Classification

In the previous sections, we have shown how to simplify some overly complex, often heuristically designed, learning objectives based on the principles of compression introduced in this book. However, objectives associated with some of the most popular learning tasks are already rather simple. In these cases, it is difficult to further simplify the objective. Thus, in this and future sections, we will focus on principled ways to modify the *deep network architectures* for a variety of tasks.

| Method | Dataset | Sample Seen | Add | | Replace | | | Swap | |
|---|---|---|---|---|---|---|---|---|---|
| | | | Obj | Att | Obj | Att | Rel | Obj | Att |
| OpenCLIP | DataComp | 1.28B | 82.3 | 73.7 | 91.7 | 79.4 | 61.2 | 59.6 | 56.9 |
| LIFT | DataComp | 1.28B | **89.0** | **86.1** | **93.2** | **86.0** | **70.6** | **64.1** | **63.4** |

Table 8.4: The performance of LIFT and CLIP on seven SugarCrepe tasks. We use an open-source CLIP implementation, OpenCLIP [IWW+21], and train both OpenCLIP and LIFT on the open-source text–image pair dataset DataComp [GIF+23]. We report accuracy for each task, with the best results bolded.

Let us first start with arguably the most classical task in machine learning: *image classification*, which is often used as a standard task to evaluate pattern recognition algorithms or deep network architectures. From our discussion of white-box architectures in Chapter 5, we only need a semantically meaningful task to learn good representations with white-box architectures. We will validate this idea in this section.

First, the dataset stays largely the same as Section 8.2.1. Both the training and test data consist of labeled images, i.e., image-label pairs $(\boldsymbol{X}, \boldsymbol{y}) \in \mathbb{R}^{C \times H \times W} \times \{0,1\}^{N_{\mathrm{cls}}}$. We still apply various data augmentations (e.g., flips, Gaussian blurring, solarization, etc.) to each sample in each new batch.

### 8.4.1 Task and Objective

Unlike before, our task is not just to learn a good representation of the data, but also to simultaneously build a classifier. Formally, we have labeled data pairs $(\boldsymbol{X}, \boldsymbol{y})$, where $\boldsymbol{y} \in \{0,1\}^{N_{\mathrm{cls}}}$ is a one-hot vector denoting the class membership of $\boldsymbol{X}$. We consider a deterministic *center crop view* $v_{\mathrm{cc}}$ of the input data $\boldsymbol{X}$ (cf Section 8.2.2). We want to jointly train a feature mapping $(f_\theta, f_\theta^{\mathrm{ext}})$ and a *classification head* $h_\theta$, defined as follows:

$$h_\theta(\boldsymbol{z}) \doteq \mathrm{softmax}(\boldsymbol{W}^{\mathrm{head}}\boldsymbol{z} + \boldsymbol{b}^{\mathrm{head}}), \qquad \forall \boldsymbol{z} \in \mathbb{R}^d \tag{8.4.1}$$

where $(\boldsymbol{W}^{\mathrm{head}}, \boldsymbol{b}^{\mathrm{head}}) \in \mathbb{R}^{N_{\mathrm{cls}} \times d} \times \mathbb{R}^{N_{\mathrm{cls}}}$ are trainable parameters in the parameter set $\theta$, such that the map $\boldsymbol{X}_{\mathrm{cc}} \mapsto \boldsymbol{p}_\theta(\boldsymbol{X}_{\mathrm{cc}}) \doteq h_\theta(\boldsymbol{z}_\theta(\boldsymbol{X}_{\mathrm{cc}}))$ predicts a smoothed label for the view $\boldsymbol{X}_{\mathrm{cc}} = v_{\mathrm{cc}}(\boldsymbol{X})$ of the input $\boldsymbol{X}$. The learning problem attempts to minimize the distance between $\boldsymbol{p}_\theta$ and $\boldsymbol{y}$ measured through cross-entropy:

$$\min_\theta \left\{\mathcal{L}_{\mathrm{CE}}(\theta) \doteq \mathbb{E}[\mathrm{CE}(\boldsymbol{y}, \boldsymbol{p}_\theta(\boldsymbol{X}_{\mathrm{cc}}))]\right\}. \tag{8.4.2}$$

### 8.4.2 The CRATE Architecture

The architecture that we use is the CRATE architecture, described in some detail in Chapter 5. The overall setup is similar to that of the regular transformer in Section 8.2.3, with a few changes. While the embedding step is the same as both DINO and SimDINO in Section 8.2.3, the feature extraction step is the same as SimDINO in Section 8.2.3 as it just extracts the feature corresponding to the class token, and the classification head is described in Section 8.4.1, the backbone architecture is different. Each layer takes the form

$$\boldsymbol{Z}_\theta^{\ell+1/2}(\boldsymbol{X}) = \boldsymbol{Z}_\theta^{\ell}(\boldsymbol{X}) + \mathrm{MSSA}_\theta^{\ell}(\mathrm{LN}_\theta^{1,\ell}(\boldsymbol{Z}_\theta^{\ell}(\boldsymbol{X}))), \tag{8.4.3}$$

$$\boldsymbol{Z}_\theta^{\ell+1}(\boldsymbol{X}) = \mathrm{ISTA}_\theta^{\ell}(\mathrm{LN}_\theta^{2,\ell}(\boldsymbol{Z}_\theta^{\ell+1/2}(\boldsymbol{X}))), \tag{8.4.4}$$

where the $\mathrm{MSSA}_\theta^{\ell}$ and $\mathrm{ISTA}_\theta^{\ell}$ blocks are as described in Chapter 5, namely:

- The MSSA operator is multi-head-subspace-self-attention, defined as follows:

$$\mathrm{MSSA}_\theta^{\ell}(\boldsymbol{Z}) \doteq \boldsymbol{U}_{\mathrm{out}}^{\ell} \begin{bmatrix} \mathrm{SA}([\boldsymbol{U}^{1,\ell}]^\top \boldsymbol{Z}, [\boldsymbol{U}^{1,\ell}]^\top \boldsymbol{Z}, [\boldsymbol{U}^{1,\ell}]^\top \boldsymbol{Z}) \\ \vdots \\ \mathrm{SA}([\boldsymbol{U}^{K,\ell}]^\top \boldsymbol{Z}, [\boldsymbol{U}^{K,\ell}]^\top \boldsymbol{Z}, [\boldsymbol{U}^{1,\ell}]^\top \boldsymbol{Z}) \end{bmatrix} + \boldsymbol{b}_{\mathrm{out}}^{\ell} \mathbf{1}_n^\top \tag{8.4.5}$$

  where $\boldsymbol{U}^{k,\ell} \in \mathbb{R}^{d \times p}$, $\boldsymbol{U}_{\mathrm{out}}^{\ell} \in \mathbb{R}^{d \times Kp}$, and $\boldsymbol{b}_{\mathrm{out}}^{\ell} \in \mathbb{R}^{d}$ are trainable parameters belonging to the parameter set $\theta$, and (recall) the self-attention operator SA is defined in (8.2.18).

- The ISTA operator is the iterative-shrinkage-thresholding-algorithm operator, defined as follows:

$$\mathrm{ISTA}_\theta^{\ell}(\boldsymbol{Z}) \doteq \mathrm{ReLU}(\boldsymbol{Z} - \beta(\boldsymbol{D}^{\ell})^\top(\boldsymbol{D}^{\ell}\boldsymbol{Z} - \boldsymbol{Z}) + \beta\lambda \mathbf{1}_d \mathbf{1}_n^\top), \tag{8.4.6}$$

  so named because the map $\boldsymbol{X} \mapsto \mathrm{ReLU}(\boldsymbol{X} - \beta \boldsymbol{D}^\top(\boldsymbol{D}\boldsymbol{X} - \boldsymbol{Z}) + \beta\lambda \mathbf{1}_d \mathbf{1}_n^\top)$ is one step of the well-established ISTA algorithm to find an element-wise non-negative sparse representation for $\boldsymbol{Z}$ with respect to the complete dictionary $\boldsymbol{D}$ (cf Section 2.3).

We call this architecture CRATE, and a layer of the backbone is depicted in Figure 5.13. CRATE models, on top of being interpretable, are generally also highly performant and parameter-efficient.

### 8.4.3 Optimization

We train our classifier using a simple end-to-end stochastic optimization procedure, where we subsample data and views, compute the average loss and its gradient over these samples, and use an optimization algorithm to change the parameters. At each timestep $k$, we:

- Subsample $B$ different labeled samples $\{(\boldsymbol{X}_b^{(k)}, \boldsymbol{y}_b^{(k)})\}_{b=1}^{B} \subseteq \mathcal{I} \times \{0,1\}^{N_{\mathrm{cls}}}$.
- For each labeled sample $(\boldsymbol{X}_b^{(k)}, \boldsymbol{y}_b^{(k)})$, compute the central crop view $v_{b,\mathrm{cc}}^{(k)}$ and apply it to $\boldsymbol{X}_b^{(k)}$ to get $\boldsymbol{X}_{b,\mathrm{cc}}^{(k)} \doteq v_{b,\mathrm{cc}}^{(k)}(\boldsymbol{X}_b^{(k)})$.

- Compute the predictions $\boldsymbol{p}_{\theta}(\boldsymbol{X}_{b,\mathrm{cc}}^{(k)}) \doteq (h_{\theta} \circ f_{\theta}^{\mathrm{ext}} \circ f_{\theta})(\boldsymbol{X}_{b,\mathrm{cc}}^{(k)})$.
- Form the surrogate stochastic loss

$$\hat{\mathcal{L}}_{\mathrm{CE}}^{(k)}(\theta) \doteq \frac{1}{B}\sum_{b=1}^{B} \mathrm{CE}(\boldsymbol{y}_b^{(k)}, \boldsymbol{p}_{\theta}(\boldsymbol{X}_{b,\mathrm{cc}}^{(k)})). \tag{8.4.7}$$

- Compute one step of an optimization algorithm on $\theta$, giving the following iteration:

$$\theta^{(k+1)} \doteq \textsc{OptUpdate}^{(k)}(\theta^{(k)}; \nabla_{\theta}\hat{\mathcal{L}}_{\mathrm{CE}}^{(k)}). \tag{8.4.8}$$

### 8.4.4 Evaluation Methodology

We use the same evaluation procedure as Section 8.2.5. To summarize, for all evaluations (as well as training) we use a center crop view $v_{\mathrm{cc}}$ which reshapes the input image and takes a large central crop of size $(C, S_{\mathrm{cc}}, S_{\mathrm{cc}})$ where $C$ is the number of channels in the input image. We can then do linear probing, attention map visualization, and detection/segmentation benchmarks, given the output of this view.

### 8.4.5 Experimental Setup and Results

Since CRATE is directly based on the transformer, we compare the optimal settings for ViT as given by [DBK+21; TCD+20] with the same settings applied to CRATE for a fair comparison.

**Model Architecture.** The center crop resizes the whole image so that the shorter edge is of size 256 (i.e., $S_{\mathrm{rsz}} = 256$) before taking a center crop of size $224 \times 224$ (i.e., $S_{\mathrm{cc}} = 224$), both in evaluation and training. We take patch size 16 (i.e., $P_H = P_W = 16$). We use the tiny, small, base, and large models of the ViT [DBK+21] architecture as the embedding and backbone, swapping out the MHSA and MLP components for MSSA and ISTA, respectively, using the same number of heads and head dimension in the case of MSSA, and therefore reducing the number of training parameters drastically. For CRATE, we set $(\beta, \lambda) = (1, 0.1)$.

**Datasets and Optimization.** For pre-training, we use the ImageNet-1K dataset. We use the LION optimizer [CLH+24] to pre-train both our ViT replication as well as CRATE. We set the base learning rate as $2.4 \times 10^{-4}$, the weight decay as 0.5, and batch size as $B = 2048$. Our learning rate schedule increases the learning rate linearly to the base learning rate over the first 5 epochs, and decreases to 0 using a cosine schedule over the next 145 epochs (training all models for 150 epochs each). For pre-training, we apply a usual regime of data augmentations (flips, Gaussian blurs, solarization, etc.) to the image data, and also add small noise to the labels (this is called *label smoothing* [MKH19]).

Table 8.5: **Linear probing classification accuracy of CRATE and ViT** on various datasets with different model sizes when the backbone is pre-trained for classification on ImageNet-1K. We observe that given the same model configuration, CRATE has comparable classification performance with a simpler, more principled, and more parameter-efficient design.

| **Model** | CRATE-T | CRATE-S | CRATE-B | CRATE-L | ViT-T | ViT-S |
|---|---|---|---|---|---|---|
| # parameters | 6.09M | 13.12M | 22.80M | 77.64M | 5.72M | 22.05M |
| ImageNet-1K | 66.7 | 69.2 | 70.8 | 71.3 | 71.5 | 72.4 |
| ImageNet-1K ReaL | 74.0 | 76.0 | 76.5 | 77.4 | 78.3 | 78.4 |
| CIFAR10 | 95.5 | 96.0 | 96.8 | 97.2 | 96.6 | 97.2 |
| CIFAR100 | 78.9 | 81.0 | 82.7 | 83.6 | 81.8 | 83.2 |
| Oxford Flowers-102 | 84.6 | 87.1 | 88.7 | 88.3 | 85.1 | 88.5 |
| Oxford-IIIT-Pets | 81.4 | 84.9 | 85.3 | 87.4 | 88.5 | 88.6 |

Table 8.6: **Object detection and fine-grained segmentation via MaskCut on COCO val2017** [**LMB+14**]. Here all models are trained with patch size 8 instead of 16. CRATE conclusively performs better than the ViT at detection and segmentation metrics when both are trained using supervised classification.

| | Detection (↑) | | | Segmentation (↑) | | |
|---|---|---|---|---|---|---|
| Model | $AP_{50}$ | $AP_{75}$ | AP | $AP_{50}$ | $AP_{75}$ | AP |
| CRATE-S/8 | **2.9** | **1.0** | 1.1 | 1.8 | **0.7** | 0.8 |
| CRATE-B/8 | **2.9** | **1.0** | **1.3** | **2.2** | **0.7** | **1.0** |
| ViT-S/8 | 0.1 | 0.1 | 0.0 | 0.0 | 0.0 | 0.0 |
| ViT-B/8 | 0.8 | 0.2 | 0.4 | 0.7 | 0.5 | 0.4 |

For linear probing, we use several evaluation datasets such as CIFAR10, Oxford-Flowers, and Oxford-IIT-Pets. We use the AdamW optimizer to train the linear probe, using learning rate $5 \times 10^{-5}$, weight decay 0.01, and batch size $B = 256$. We also apply the aforementioned data augmentations to the image data.

**Experiment Results.** Table 8.5 demonstrates that CRATE models achieve parity or improvement compared to the popular Vision Transformer (ViT) architecture at similar parameter counts, at least in terms of the linear separability of their features with respect to different classes. In terms of attention map fidelity, Figure 8.12 demonstrates a truly extraordinary result: without needing to train on any segmentation or object detection data, *not only* do the saliency maps effectively capture all relevant parts of the input image, the saliency maps *self-organize* to each correspond to a discrete set of concepts, even across samples and classes! This is the first system to do this, to the authors' knowledge, and it can do this without using any extra data except for the image classification data. Table 8.6 confirms these qualitative insights quantitatively, showing significant improvement over ViTs trained in the same supervised classification

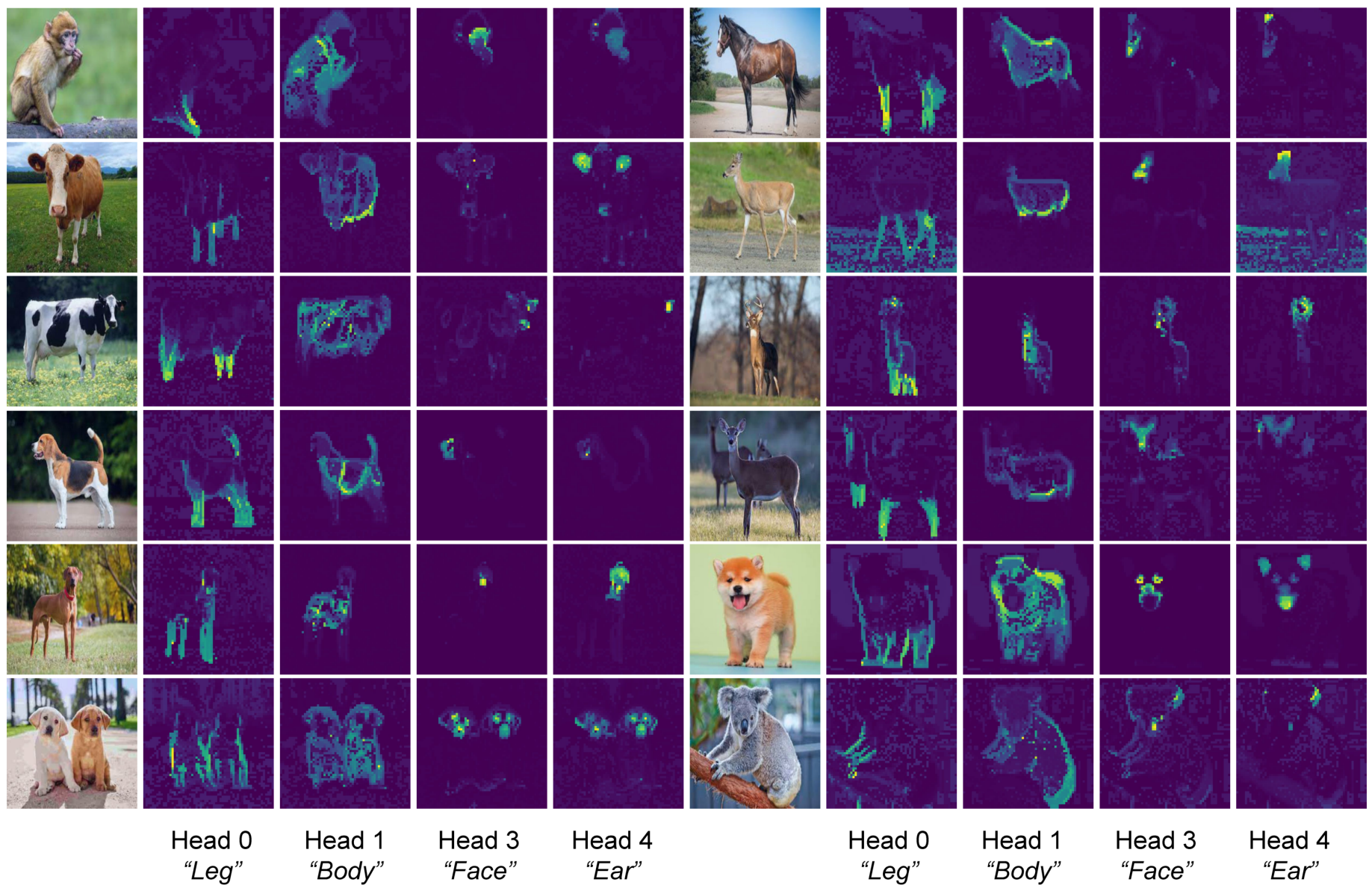

Figure 8.12: **Interpretable saliency maps in CRATE** with patch size 8. When given images with similar properties (perhaps but not necessarily from the same class), the saliency maps corresponding to different attention heads in the last layer each highlight a specific property. One can observe that the average saliency map (not included) then highlights all relevant objects in the image, showing that it uses all fine-grained details of the input image for classification. This is the *first* machine learning system to do this, to the authors' knowledge, much less automatically without training on any segmentation data.

setup.

## 8.5 Image Representation via Masked Autoencoding

In this section, we show how to apply the CRATE architecture to the *image completion* problem, also known as *masked autoencoding* (MAE), which can be viewed as a generalization of the low-rank matrix completion problem discussed in Chapter 2. Masked autoencoding, since its introduction in the deep learning context by [HCX+22], has been a staple and simple self-supervised representa-tion learning method, which aims to endow each patch feature within $\boldsymbol{Z}_\theta$ with aggregate information as well as information about its neighbors, such that both the patch feature and aggregate features are rich sources of information for the whole sample.

The data we use for this task is the same as the image datasets discussed in Section 8.2.1. As usual, we still apply data augmentations to each sample in each new batch.

### 8.5.1 The MAE Task and Objective

As the name suggests, masked autoencoding involves a view $v_m$ which, given an input, performs a random resized crop (cf Section 8.2.2) to turn the input image into a square image of size $(C, S_{\mathrm{mask}}, S_{\mathrm{mask}})$, then *masks* (i.e., sets to zero) a fixed percentage $p_{\mathrm{mask}} \in [0, 1]$ of pixels in the input. For efficiency reasons[6], the masking is done patch-wise, i.e., after embedding the whole image, $p_{\mathrm{mask}}$ percentage of *patches* are set to zero. The goal of MAE is to train an encoder $f_\theta \colon \mathcal{I} \to (\mathbb{R}^d)^*$ and a decoder $g_\eta \colon (\mathbb{R}^d)^* \to \mathcal{I}$ that can reconstruct an input from its masking, i.e., writing $\hat{\boldsymbol{X}}_{\theta,\eta} \doteq g_\eta \circ f_\theta$, we have

$$\min_{\theta,\eta} \left\{ \mathcal{L}_{\mathrm{MAE}}(\theta, \eta) \doteq \mathbb{E} \, \|\hat{\boldsymbol{X}}_{\theta,\eta}(\boldsymbol{X}_m) - \boldsymbol{X}\|_F^2 \right\} \tag{8.5.1}$$

Essentially this means that the features $\boldsymbol{Z}_\theta(\boldsymbol{X}_m)$ of the view $\boldsymbol{X}_m \doteq v_m(\boldsymbol{X})$ must contain information about the *masked patches* as well as the existing patches. From the perspective of the compression-based white-box models in Chapter 5, if a white-box autoencoder $(f_\theta, g_\eta)$ succeeds at this task, it means that the learned subspaces and dictionaries perform a *redundant* encoding of the data such that it can reconstruct missing parts of the data from encoded other parts of the data. This means that information about each patch is stored in other patches. Therefore, each patch feature should contain both information about the patch and information about the statistics of the whole image. Thus, again, we expect that the representations should contain both local and global semantically relevant information, and therefore representations of different patches with similar local and global information should be related (i.e., on the same subspace or encoded together by a dictionary).

### 8.5.2 CRATE-Based MAE Architecture

We use a CRATE encoder and decoder, depicted in Figure 8.7, though of course it is possible to use a regular transformer encoder and decoder. Details follow.

[6] The original implementation of MAE by [HCX+22] embeds the whole image, *removes* the tokens that would be masked, feeds the resulting token set through the encoder, adds back learned placeholder tokens in the masked spots and adds back the appropriate positional encoding, and feeds the resulting token set through the decoder to get the autoencoding prediction. This is more efficient since the encoder has fewer tokens to go through, but conceptually is the same as the method discussed in the text, and the resulting models' performance in the masked autoencoding task and downstream evaluations is very similar.

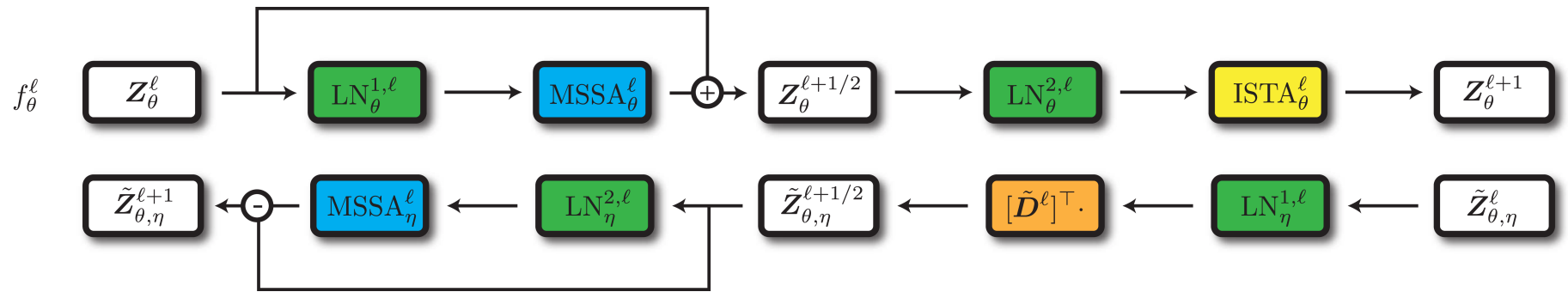

Figure 8.13: **One layer of the encoder and decoder in a CRATE autoencoder backbone.** The encoder and decoder layers both feed their inputs through multi-head subspace self-attention and a dictionary learning or dictionary encoding step. Note that the encoder and decoder layers are symmetrically designed; the conceptual goal of each decoder layer is to invert an encoder layer, so this symmetry is very much by design (see e.g., Chapter 6).

**The Encoder.** The encoder is the same as the CRATE encoder in Section 8.4.2, with the caveat that there is no feature extractor $f_\theta^{\mathrm{ext}}$. However, both the embedding $f_\theta^{\mathrm{emb}}$ and the backbone $f_\theta^{\mathrm{bb}}$ are the same.

**The Decoder Backbone.** The decoder backbone is the CRATE decoder described in Chapter 6. For completeness, we describe it now. Given a feature sequence $\boldsymbol{Z}_\theta(\boldsymbol{X}) \doteq f_\theta(\boldsymbol{X}) \in (\mathbb{R}^d)^*$, we can process it using the decoder backbone $g_\eta^{\mathrm{bb}}$ as follows. The function $g_\eta^{\mathrm{bb}}$ is composed of $L$ layers $g_\eta^\ell$, i.e.,

$$g_\eta^{\mathrm{bb}} = g_\eta^L \circ \cdots \circ g_\eta^1. \tag{8.5.2}$$

The layer $g_\eta^\ell$ has the following implementation. First, define $\tilde{\boldsymbol{Z}}_{\theta,\eta}^1(\boldsymbol{X}) \doteq \boldsymbol{Z}_\theta(\boldsymbol{X})$. Then, we obtain

$$\tilde{\boldsymbol{Z}}_{\theta,\eta}^{\ell+1/2}(\boldsymbol{X}) = [\tilde{\boldsymbol{D}}^\ell]^\top \mathrm{LN}_\eta^{1,\ell}(\tilde{\boldsymbol{Z}}_{\theta,\eta}^\ell(\boldsymbol{X})) \tag{8.5.3}$$

$$\tilde{\boldsymbol{Z}}_{\theta,\eta}^{\ell+1}(\boldsymbol{X}) = \tilde{\boldsymbol{Z}}_{\theta,\eta}^{\ell+1/2}(\boldsymbol{X}) - \mathrm{MSSA}_\eta^\ell(\mathrm{LN}_\eta^{2,\ell}(\tilde{\boldsymbol{Z}}_{\theta,\eta}^{\ell+1/2})) \tag{8.5.4}$$

and $g_\eta^\ell$ is defined such that $g_\eta^\ell(\tilde{\boldsymbol{Z}}_{\theta,\eta}^\ell) \doteq \tilde{\boldsymbol{Z}}_{\theta,\eta}^{\ell+1}(\boldsymbol{X})$. Here, the relevant concept is that $g_\eta^\ell$ should learn an approximate inverse of $f_\theta^{L+1-\ell}$, as discretizations of a forward- and reverse-time diffusion process, respectively. In particular, $\tilde{\boldsymbol{D}}^\ell$ should approximate $\boldsymbol{D}^{L+1-\ell}$, and similarly, the $\mathrm{MSSA}_\eta^\ell$ parameters should be similar to the parameters of $\mathrm{MSSA}_\theta^{L+1-\ell}$. The output is $\tilde{\boldsymbol{Z}}_{\theta,\eta} \doteq \tilde{\boldsymbol{Z}}_{\theta,\eta}^{L+1}$.

**The Un-embedding Module.** To transform $\tilde{\boldsymbol{Z}}_{\theta,\eta}(\boldsymbol{X})$ back into an estimate for $\boldsymbol{X}$, we need to undo the effect of the embedding module $f_\theta^{\mathrm{emb}}$ using the unembedding module $g_\eta^{\mathrm{unemb}}$. As such, harkening back to the functional form of the embedding module in (8.2.13), i.e.,

$$f_\theta^{\mathrm{emb}}(\boldsymbol{X}) \doteq \left[\boldsymbol{z}_{\mathrm{cls}}^1, \boldsymbol{W}^{\mathrm{emb}} f^{\mathrm{patch}}(\boldsymbol{X}) + \boldsymbol{E}^{\mathrm{pos}}\right] \tag{8.5.5}$$

it implies that our inverse operation $g_\eta^{\mathrm{unemb}}$ looks like the following:

$$g_\eta^{\mathrm{unemb}}(\tilde{\boldsymbol{Z}}) \doteq g_\eta^{\mathrm{unemb}}(\left[\tilde{\boldsymbol{z}}^1, \dots, \tilde{\boldsymbol{z}}^n\right]) \tag{8.5.6}$$

$$= g^{\text{unpatch}}(\boldsymbol{W}^{\text{unemb}}([\tilde{\boldsymbol{z}}^2, \ldots, \tilde{\boldsymbol{z}}^n] - \tilde{\boldsymbol{E}}^{\text{pos}})), \tag{8.5.7}$$

where $g^{\text{unpatch}}$ does the inverse operation of the unrolling and flattening operation that $f^{\text{patch}}$ does.[7]

This architecture is a white-box autoencoder $(f_\theta, g_\eta)$ where (recall) $f_\theta = f_\theta^{\text{bb}} \circ f_\theta^{\text{emb}}$ and $g_\eta = g_\eta^{\text{unemb}} \circ g_\eta^{\text{bb}}$. In particular, we can use it to compute an estimate for a masked view $\hat{\boldsymbol{X}}_{\theta,\eta}(\boldsymbol{X}_m) = (g_\eta \circ f_\eta)(\boldsymbol{X}_m)$ which should approximately equal $\boldsymbol{X}$ itself.

### 8.5.3 Optimization

As in Section 8.4.3, we use a simple optimization setup: we sample images and masks, compute the loss on those samples and the gradients of this loss, and update the parameters using a generic optimization algorithm and the aforementioned gradients. For each timestep $k$, we:

- Subsample $B$ different samples $\{\boldsymbol{X}_b^{(k)}\}_{b=1}^B \subseteq \mathcal{I}$.
- For each sample $\boldsymbol{X}_b^{(k)}$, compute a different randomized resized crop and mask $v_{b,m}^{(k)}$ and apply it to $\boldsymbol{X}_b^{(k)}$ to get $\boldsymbol{X}_{b,m}^{(k)} \doteq v_{b,m}^t(\boldsymbol{X}_b^{(k)})$.
- Compute the estimated autoencoding $\hat{\boldsymbol{X}}_{\theta,\eta}(\boldsymbol{X}_{b,r}^{(k)}) \doteq (g_\eta \circ f_\theta)(\boldsymbol{X}_{b,r}^{(k)})$.
- Form the surrogate stochastic loss

$$\hat{\mathcal{L}}_{\text{MAE}}^{(k)}(\theta, \eta) \doteq \frac{1}{B}\sum_{b=1}^{B} \|\hat{\boldsymbol{X}}_{\theta,\eta}(\boldsymbol{X}_{b,r}^{(k)}) - \boldsymbol{X}_b^{(k)}\|_F^2. \tag{8.5.8}$$

- Compute one step of an optimization algorithm on $(\theta, \eta)$, giving the following iteration:

$$(\theta^{(k+1)}, \eta^{(k+1)}) \doteq \textsc{OptUpdate}^{(k)}(\theta^{(k)}, \eta^{(k)}; \nabla_{(\theta,\eta)}\hat{\mathcal{L}}_{\text{MAE}}^{(k)}). \tag{8.5.9}$$

### 8.5.4 Evaluation

This is the first autoencoder network we discuss in this chapter. We use the same center crop view $v_{\text{cc}}$ as in Sections 8.2.5 and 8.4.4, resizing the final image to a square with side length $S_{\text{cc}} = S_{\text{mask}}$ pixels to match the shapes of the input images seen during training.

In addition to evaluating the masked autoencoding loss itself, it is also possible to evaluate the features $\boldsymbol{Z}_\theta(\boldsymbol{X}_{\text{cc}})$ of the view $\boldsymbol{X}_{\text{cc}} \doteq v_{\text{cc}}(\boldsymbol{X})$ of the data $\boldsymbol{X}$ directly. For attention map fidelity evaluation, obtaining $\boldsymbol{Z}_\theta(\boldsymbol{X}_{\text{cc}})$ is sufficient, but for linear probing we need to extract a summarized or aggregate feature from $\boldsymbol{Z}_{\theta}$. To do this, we can use a (parameter-free) feature extraction map that returns the feature corresponding to the class token, i.e.,

$$f_{\theta}^{\mathrm{ext}}(\boldsymbol{Z}) \doteq f_{\theta}^{\mathrm{ext}}([\boldsymbol{z}^{1}, \dots, \boldsymbol{z}^{n}]) = \boldsymbol{z}^{1}, \tag{8.5.10}$$

as in (for example) Sections 8.4.1 and 8.4.2. With this, we have a way to obtain aggregate features $\boldsymbol{z}_{\theta}(\boldsymbol{X}_{\mathrm{cc}}) \doteq (f_{\theta}^{\mathrm{ext}} \circ f_{\theta})(\boldsymbol{X}_{\mathrm{cc}})$, at which point we can perform linear probing, segmentation evaluations, and so on.

[7] Again, the "inverse positional encoding" $\tilde{\boldsymbol{E}}^{\text{pos}}$ is learned for a large input, and for smaller inputs may be interpolated. It is even possible to directly set $\tilde{\boldsymbol{E}}^{\text{pos}}$ equal to the positional encoding $\boldsymbol{E}^{\text{pos}}$ and use the same interpolated positional encodings for each input in both the encoder and decoder.

### 8.5.5 Experiments

Since CRATE-MAE is directly based on ViT-MAE, we compare the optimal settings for ViT-MAE as given by [HCX+22] with the same settings applied to CRATE-MAE for a fair comparison.

**Model Architecture.** During training, the masked crop $v_m$ resizes the whole image so that the shorter edge is of size 256 (i.e., $S_{\mathrm{rsz}} = 256$) before taking a random crop of size $224 \times 224$ (i.e., $S_{\mathrm{mask}} = 224$), and masking $p_{\mathrm{mask}} = \frac{3}{4}$ of the patches. We take patch size 16 (i.e., $P_H = P_W = 16$). We use the small and base variants of the ViT-MAE architecture as the embedding and backbone for both the encoder and decoder, swapping out the MHSA and MLP components for MSSA, ISTA, and linear layers, respectively. We use the same number of heads and head dimension in the case of MSSA. However, the original ViT-MAE uses an encoder which uses nearly all the total layers and a decoder which only uses a few layers; we allocate half the total number of layers (which stays the same from ViT-MAE to CRATE-MAE) to our encoder and decoder, as suggested by the conceptual and theoretical framework in Chapter 6. For CRATE-MAE we set $(\beta, \lambda) = (1, 0.1)$.

**Datasets and Optimization.** For pre-training, we use the ImageNet-1K dataset. We use the AdamW optimizer to pre-train both our ViT-MAE replication as well as CRATE-MAE. We set the base learning rate as $3 \times 10^{-5}$, the weight decay as 0.1, and batch size as $B = 4096$. Our learning rate schedule increases the learning rate linearly to the base learning rate over the first 40 epochs, and decreases to 0 using a cosine schedule over the next 760 epochs (training all models for 800 epochs each). For pre-training, we apply the usual regime of data augmentations (flips, Gaussian blurs, solarization, etc) to the image data.

For linear probing, we use several evaluation datasets such as CIFAR10, CIFAR100, Oxford-Flowers, and Oxford-IIT-Pets. For linear probing, we pre-compute the features of all samples in the target dataset and apply a fast linear regression solver, e.g., from a standard package such as Scikit-Learn.

**Experiment Results.** Table 8.7 demonstrates that CRATE-MAE models achieve, roughly speaking, parity with the popular ViT-MAE architecture at similar parameter counts, and also that the feature learning performance (as measured by performance on downstream classification tasks) increases with scale. Meanwhile, Figure 8.14 demonstrates that the encoder saliency maps (and therefore the fine-grained features learned by the encoder) indeed isolate and highlight the key parts of the input image.

Table 8.7: **Linear probing classification accuracy of CRATE-MAE and ViT-MAE** on various datasets with different model sizes when the backbone is pre-trained for masked autoencoding on ImageNet-1K. Given the same parameter count, CRATE-MAE achieves roughly similar performance while simultaneously enjoying a simpler and more principled architecture design.

| **Model** | CRATE-MAE-S(mall) | CRATE-MAE-B(ase) | ViT-MAE-S | ViT-MAE-B |
|---|---|---|---|---|
| # parameters | 25.4M | 44.6M | 47.6M | 143.8M |
| CIFAR10 | 79.4 | 80.9 | 79.9 | 87.9 |
| CIFAR100 | 56.6 | 60.1 | 62.3 | 68.0 |
| Oxford Flowers-102 | 57.7 | 61.8 | 66.8 | 66.4 |
| Oxford-IIIT-Pets | 40.6 | 46.2 | 51.8 | 80.1 |

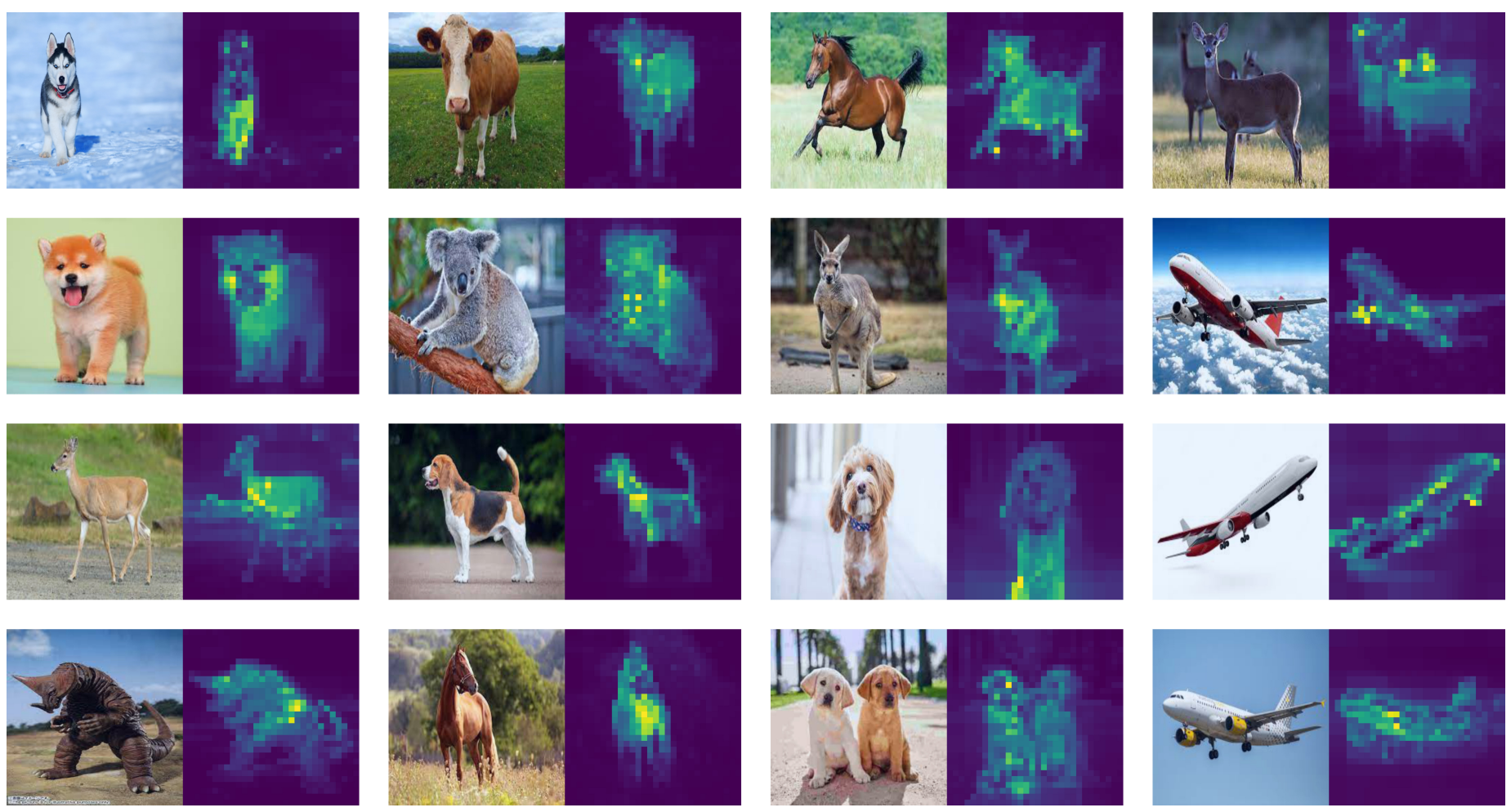

Figure 8.14: **Saliency maps of CRATE-MAE.** Each pair of images consists of the original image (left) and a selected saliency map (right) corresponding to an attention head in the last layer. As is usual for CRATE models, but unusual for general transformer-like models, the saliency maps correspond to the objects in the input image.

## 8.6 Image Generation via Auto-Encoding and Sampling

In the previous section, we have introduced masked autoencoding (MAE) as one approach to learn an autoencoding of natural images $\boldsymbol{x} \sim p(\boldsymbol{x})$:

$$\boldsymbol{x} \xrightarrow{\mathcal{E}=f} \boldsymbol{z} \xrightarrow{\mathcal{D}=g} \hat{\boldsymbol{x}}. \tag{8.6.1}$$

where $\boldsymbol{z} = f(\boldsymbol{x})$ is certain feature space. Such an autoencoding is designed to learn correlation, or co-occurrence, among patches of a natural image.

However, MAE does not enforce the (distribution of) learned features $\boldsymbol{z}$ to have any explicit structures or properties, which can be very important if we want to use the so-learned features for certain subsequent tasks such as recognition or controlled generation. In addition, the simple MMSE loss used for image completion does not necessarily respect the possible nonlinear structures in the image manifold and typically generate an expected value, hence usually blurred version, for the missing image patches, as we have shown in Figure 7.8 in the previous Chapter 7.

Hence, in this section, we will introduce some other methods to learn representations for (the distribution of) natural images via auto-encoding that aim, to large extent, to remedy the above two issues. More precisely, we want the distribution of the encoded features to have certain desired structures and images decoded from such features are much more faithful to the distribution of natural images – that is, of much better visual quality.

### 8.6.1 Variational Auto-Encoding of Imagery Data

As we have introduced in Section 6.1.4 of Chapter 6, variational auto-encoding is a method that aims to promote the distribution of learned features $\boldsymbol{z}$ and the conditional distributions between $\boldsymbol{x}$ and $\boldsymbol{z}$ to be more Gaussian like.[8]

Here we introduce a popular implementation of a VAE for imagery data, following the perceptual compression model leveraged in Stable Diffusion [RBL+22]. The key idea is to train an autoencoder that maps high-resolution images into a low-dimensional latent space, while imposing a KL-regularization on the latent distribution to encourage it to be close to a standard Gaussian $\mathcal{N}(\mathbf{0}, \boldsymbol{I})$. This latent space can subsequently serve as an efficient arena for downstream generative models such as latent diffusion. Both the encoder and decoder are jointly trained end-to-end. We outline the details as follows.

**The Encoder ($\mathcal{E}$).** The encoder $\mathcal{E}$ maps an input RGB image $\boldsymbol{x} \in \mathbb{R}^{H \times W \times 3}$ into a low-dimensional latent space. Its architecture, adopted from [ERO21], is a fully convolutional network. It begins with an input convolution that projects the 3-channel RGB input into a feature map with a base channel count of 128.

[8]One may view such prior is to promote the learned features to be diverse, in a similar spirit to maximizing the coding rate of all features.

This is followed by four stages, each consisting of two ResNet blocks [HZR+16b]. The channel dimension is scaled by multipliers $(1, 2, 4, 4)$ across stages, yielding feature maps with $128, 256, 512, 512$ channels, respectively. Spatial downsampling by a factor of $2\times$ is performed between consecutive stages via strided convolutions, resulting in three downsamplings and an overall downsampling factor of $f = 2^3 = 8$. After the four stages, a middle block consisting of two additional ResNet blocks further refines the features. Finally, an output convolution projects the features to $2c$ channels, which are split into the two quantities that parameterize the latent distribution (described below).

Crucially, unlike a standard autoencoder, the encoder does not output a single deterministic latent vector. Instead, it parameterizes a diagonal Gaussian posterior $q(\boldsymbol{z}|\boldsymbol{x}) = \mathcal{N}(\boldsymbol{\mu}, \boldsymbol{\sigma}^2 \boldsymbol{I})$ by outputting two quantities:

- A mean $\boldsymbol{\mu} = \mathcal{E}_\mu(\boldsymbol{x}) \in \mathbb{R}^{h\times w\times c}$, and
- A log-variance $\log \boldsymbol{\sigma}^2 = \mathcal{E}_\sigma(\boldsymbol{x}) \in \mathbb{R}^{h\times w\times c}$.

The latent representation is then obtained via a so called *reparameterization trick* [KW13b]:

$$\boldsymbol{z} = \boldsymbol{\mu} + \boldsymbol{\sigma} \odot \boldsymbol{\epsilon}, \quad \boldsymbol{\epsilon} \sim \mathcal{N}(\boldsymbol{0}, \boldsymbol{I}), \tag{8.6.2}$$

where $\odot$ denotes element-wise multiplication. This formulation enables gradient-based optimization through the stochastic sampling step.

**The Decoder ($\mathcal{D}$).** The decoder $\mathcal{D}$ learns the reverse mapping from the latent space back to the pixel space, producing the reconstructed image $\tilde{\boldsymbol{x}} = \mathcal{D}(\boldsymbol{z}) \in \mathbb{R}^{H\times W\times 3}$. Its architecture is approximately symmetric to the encoder: an input convolution first projects the $c$-channel latent into the feature space, followed by a middle block of two ResNet blocks. Then, four stages—each with two ResNet blocks—progressively upsample the spatial resolution, with the channel dimensions mirroring those of the encoder in reverse order $(512, 512, 256, 128)$. A final output convolution maps the features back to 3 RGB channels. Both the encoder and decoder architectures follow the convolutional design introduced in [ERO21].

### Data

We assume access to a dataset $\mathcal{D} = \{\boldsymbol{x}_i\}_{i=1}^N$ of samples from the original data distribution. Each sample is an image of the form $\boldsymbol{x} \in \mathbb{R}^{H\times W\times 3}$, where $H$ and $W$ are the height and width, respectively. In [RBL+22], the autoencoder is trained on the OpenImages dataset with images cropped to $256\times256$ resolution.

Given the downsampling factor $f = 8$ and latent channel count $c = 4$, the encoder maps each input image to a latent tensor $\boldsymbol{z} \in \mathbb{R}^{H/f\times W/f\times c}$. For the default training resolution, this yields a latent of size $32 \times 32 \times 4$. At inference time, the autoencoder generalizes to other resolutions; for example, a $512 \times 512 \times 3$ input is encoded into a $64 \times 64 \times 4$ latent.

**Objective and Evaluation Metrics**

To train the VAE, our goal is to jointly optimize the encoder $\mathcal{E}$ and decoder $\mathcal{D}$ such that the reconstructed data $\tilde{\boldsymbol{x}} = \mathcal{D}(\mathcal{E}(\boldsymbol{x}))$ is as close as possible to the original data $\boldsymbol{x}$, while regularizing the latent distribution to be close to a standard Gaussian prior. An empirically effective choice of the reconstruction loss is a weighted combination of L1, LPIPS [ZIE+18], and adversarial losses as in GAN [GPM+14b]. Additionally, a KL divergence term is included to regularize the latent space. Formally, the training objective can be expressed as:

$$\min_{\theta_{\mathcal{E}},\theta_{\mathcal{D}}} \mathcal{L}_{\text{VAE}}(\theta_{\mathcal{E}},\theta_{\mathcal{D}}) = \mathbb{E}_{\boldsymbol{x}\sim\mathcal{D}}\Big[\|\tilde{\boldsymbol{x}}-\boldsymbol{x}\|_1 + \lambda_1 \,\text{LPIPS}(\tilde{\boldsymbol{x}},\boldsymbol{x}) \tag{8.6.3}$$

$$+ \lambda_2\, \mathcal{L}_{\text{GAN}}(\tilde{\boldsymbol{x}},\boldsymbol{x}) + \lambda_3\, D_{KL}\big(q(\boldsymbol{z}|\boldsymbol{x}) \,\big\|\, p(\boldsymbol{z})\big)\Big], \tag{8.6.4}$$

where $\theta_{\mathcal{E}}$ and $\theta_{\mathcal{D}}$ are the parameters of the encoder and decoder, respectively, and $\lambda_1, \lambda_2, \lambda_3$ are hyperparameters that balance the contributions of each loss term. The first three terms enforce sample-wise consistency in the same manner as the reconstruction objectives in Section 8.6.2. The fourth term—the KL divergence between the encoder posterior $q(\boldsymbol{z}|\boldsymbol{x}) = \mathcal{N}(\boldsymbol{\mu}, \boldsymbol{\sigma}^2\boldsymbol{I})$ and the standard Gaussian prior $p(\boldsymbol{z}) = \mathcal{N}(\boldsymbol{0}, \boldsymbol{I})$—regularizes the latent space. For a diagonal Gaussian posterior, this admits a closed-form expression:

$$D_{KL}\big(q(\boldsymbol{z}|\boldsymbol{x}) \,\big\|\, p(\boldsymbol{z})\big) = \frac{1}{2}\sum_{j=1}^{d}\left(\mu_j^2 + \sigma_j^2 - \log\sigma_j^2 - 1\right), \tag{8.6.5}$$

where $d = h \times w \times c$ is the total dimensionality of the latent space.

The implementation detail from [RBL+22] is that the KL weight $\lambda_3$ is set to a very small value ($\approx 10^{-6}$). This is because the primary purpose of the KL term here is not to enable unconditional sampling from the latent space (as in a traditional VAE), but rather to prevent the latent space from developing arbitrarily high variance, which would degrade the signal-to-noise ratio for downstream models operating in this space.

For evaluation, we measure the reconstruction quality by computing the FID score [HRU+17b] on the reconstructed images from the ImageNet-1k validation set, denoted as rFID. A lower rFID indicates better reconstruction quality.

## 8.6.2 Representation Auto-Encoding of Imagery Data

As discussed in Chapter 4, learning structured representations of the data is crucial for many downstream tasks. In visual representation learning, this is often achieved by self-supervised learning methods such as DINO [CTM+21] and SimDINO (studied in Section 8.2). In Chapter 6, we posit that a good representation of data should be effectively decoded back into the original data space. This requires a consistent auto-encoding framework that can accurately reconstruct the original data from its learned representation. It is then natural to ask: is the representation learned by methods like DINO consistent and can it be effectively decoded in an auto-encoding framework?

To find out, we can train a *representation auto-encoder* (RAE) that consists of an pretrained representation encoder (e.g. DINO) and a trainable decoder, following the work:

https://rae-dit.github.io.

The encoder $f$ maps an input sample $\boldsymbol{x}$ to its learned representation $\boldsymbol{z} = f(\boldsymbol{x})$, while the decoder $g$ maps the representation back to the original data space $\boldsymbol{x}' = g(\boldsymbol{z})$. As detailed in Chapter 6, one simple objective is to impose sample-wise consistency by training the decoder to minimize the reconstruction loss between the original data $\boldsymbol{x}$ and the reconstructed data $\boldsymbol{x}'$. We outline the details as follows.

**The Encoder.** We use a pretrained representation encoder $f \colon \mathcal{X} \to \mathcal{Z}$ that maps an input data sample $\boldsymbol{x} \in \mathcal{X}$ to its learned representation $\boldsymbol{z} = f(\boldsymbol{x}) \in \mathcal{Z}$. The encoder is kept frozen during training. Here, we assume a ViT-based encoder, as introduced in Section 8.2.3.

**The Decoder.** For architectural simplicity, we use a trainable ViT-based decoder $g \colon \mathcal{Z} \to \mathcal{X}$ that maps the learned representation $\boldsymbol{z}$ back to the original data space, producing the reconstructed data $\boldsymbol{x}' = g(\boldsymbol{z})$.

### Data

We assume access to a dataset $\mathcal{D} = \{\boldsymbol{x}_i\}_{i=1}^N$ of samples from the original data distribution $\mathcal{X}$. Each sample is an image of the form $\boldsymbol{x} \in \mathbb{R}^{3\times H\times W}$, where $H$ and $W$ are the height and width, respectively. The encoder $f$ is set to a patch size of $p_e$ and feature dimension of $d$. Consequently, for each input image $\boldsymbol{x}$, we would obtain $N = HW/p_e^2$ tokens of dimension $d$. The decoder $g$ with patch size $p_d$ then maps these visual tokens back to reconstruct $\boldsymbol{x}' \in \mathbb{R}^{3\times H\frac{p_d}{p_e}\times W\frac{p_d}{p_e}}$ in the original pixel space. By default, we set $p_d = p_e$ to ensure that the reconstructed image has the same resolution as the original image.

### Objective and Evaluation Metrics

To train RAE, our goal is to train a decoder $g$ such that the reconstructed data $\boldsymbol{x}' = g(f(\boldsymbol{x}))$ is as close as possible to the original data $\boldsymbol{x}$. This can be achieved by minimizing a sample-wise consistency loss over the dataset $\mathcal{D}$. An empirically effective choice of the loss function is a weighted combination of L1, LPIPS [ZIE+18] [9], and adversarial losses as in GAN [GPM+14b]. Formally, the training objective can be expressed as:

[9] This is a perceptual loss based on computing distances between features across different layers of a pretrained network. It empirically captures human perceptual similarity better than pixel-wise losses. Refer to the original paper for details.

Table 8.8: **Reconstruction and linear probing results of RAEs.**

(a) **Encoder choice.** All encoders outperform SD-VAE.

| Model | rFID |
|---|---|
| DINOv2-B | 0.49 |
| SigLIP2-B | 0.53 |
| MAE-B | 0.16 |
| SD-VAE | 0.62 |

(b) **Larger decoders** improve rFID while remaining much more efficient than VAEs.

| Decoder | rFID | GFLOPs |
|---|---|---|
| ViT-B | 0.58 | 22.2 |
| ViT-L | 0.50 | 78.1 |
| ViT-XL | 0.49 | 106.7 |
| SD-VAE | 0.62 | 310.4 |

(c) **Encoder scaling.** rFID is stable across RAE sizes.

| Encoder | rFID |
|---|---|
| DINOv2-S | 0.52 |
| DINOv2-B | 0.49 |
| DINOv2-L | 0.52 |

(d) **Representation quality.** RAEs have much higher linear probing accuracy than VAEs.

| Model | Top-1 Acc. |
|---|---|
| DINOv2-B | 84.5 |
| SigLIP2-B | 79.1 |
| MAE-B | 68.0 |
| SD-VAE | 8.0 |

$$\min_{\theta} \mathcal{L}_{\mathrm{RAE}}(\theta) = \mathbb{E}_{\boldsymbol{x}\sim\mathcal{D}}\left[\|\boldsymbol{x}' - \boldsymbol{x}\|_1 + \lambda_1 \mathrm{LPIPS}(\boldsymbol{x}', \boldsymbol{x}) + \lambda_2 \mathcal{L}_{\mathrm{GAN}}(\boldsymbol{x}', \boldsymbol{x})\right], \quad (8.6.6)$$

where $\theta$ are the parameters of the decoder $g$, and $\lambda_1, \lambda_2$ are hyperparameters that balance the contributions of each loss term.

For evaluation, we measure the reconstruction quality by computing the FID score [HRU+17b] on the reconstructed images from the ImageNet-1k validation set, denoted as rFID. A lower rFID indicates better reconstruction quality.

### Experimental Results

We can compare RAEs trained on different pretrained encoders, including DINOv2, MAE, and SigLIP2. As a baseline, we also consider a vanilla variational autoencoder (VAE) trained directly on the image pixels without any pretrained encoder. One popular choice is SD-VAE. RAE decoders are trained on the ImageNet-1K training set and evaluated on the ImageNet-1K validation set. The results are summarized in Section 8.6.2. From Table 8.8b, we can see that RAEs with pretrained encoders outperform SD-VAE in reconstruction quality. Furthermore, larger decoder sizes lead to improved reconstruction quality, as shown in Table 8.8b, while remaining more efficient than VAEs. Table 8.8c shows that the reconstruction quality is stable across different sizes of the DINO encoders. Finally, Table 8.8d demonstrates that RAEs also retain high-quality representations, achieving much higher linear probing accuracy compared to VAEs.

To complement the quantitative metrics in Section 8.6.2, Figure 8.15 provides a visual comparison of reconstructions. RAE reconstructions preserve fine-grained textures, object boundaries, and semantic details as faithfully as the SD-VAE baseline, consistent with its better rFID scores.

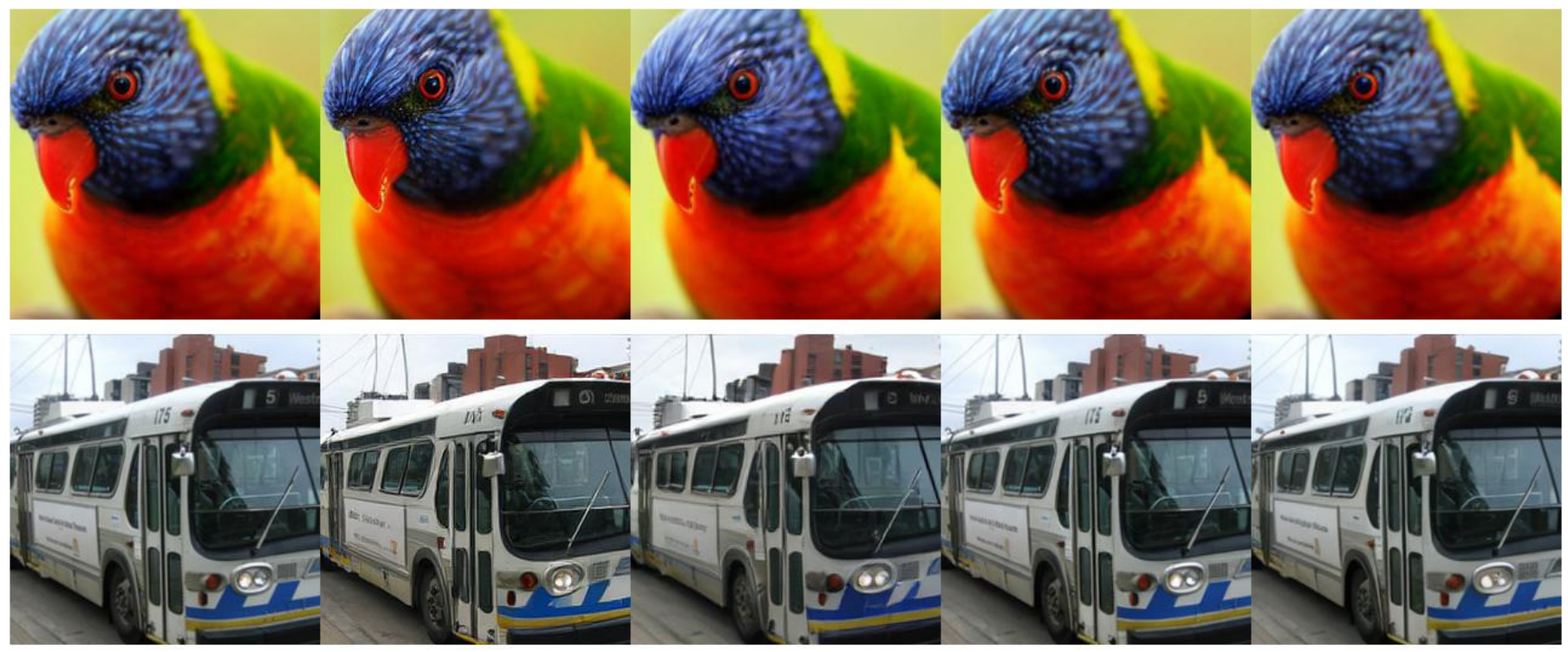

Figure 8.15: **RAE Reconstruction examples.** From left to right: input image, RAE (DINOv2-B), RAE (Siglp-B), RAE (MAE-B), SD-VAE. Zoom in for details.

### 8.6.3 Sampling from Learned Representations via Denoising

Once we have learned a consistent and structured representation of the distribution of natural images, say via VAE or RAE, we typically like to efficiently generate *samples* $\boldsymbol{z} \in \mathcal{Z}$ from the learned representation distribution so that we can regenerate the natural images associated with these samples via the learned decoder $\hat{\boldsymbol{x}} = g(\boldsymbol{z}) \in \mathcal{X}$. This auto-encoding process is illustrated by the top half of Figure 1.27 in Chapter 1.

To learn hence sample the distribution of the features $\boldsymbol{z}$ in the latent space $\mathcal{Z}$, we can leverage the denoising and diffusion method introduced in Chapter 3. Concretely, suppose $p(\boldsymbol{z})$ is the distribution of the learned features $\boldsymbol{z}$ in the feature space. We may learn a denosing process that, starting from an initial standard Gaussian distribution $\boldsymbol{g} \sim \mathcal{N}(0, \boldsymbol{I})$, converges to this distribution $p(\boldsymbol{z})$. Following the method introduced in Chapter 3, we want to learn a denoiser $\bar{\boldsymbol{z}}$ such that it iteratively denoises a sample from the initial distribution $\mathcal{N}(0, \boldsymbol{I})$ to the target distribution of the learned representations $p(\boldsymbol{z})$. This process is illustrated by the bottom half of Figure 1.27 in Chapter 1.

Here to learn the denoising, let us take the simple case of flow matching introduced in Chapter 3. We can define the denoiser as a time-dependent vector field $\bar{\boldsymbol{z}}_\theta(t, \boldsymbol{z}_t)$ parameterized by $\theta$ that transforms a sample from the initial distribution to the target distribution over time $t \in [0, 1]$. Here, $\boldsymbol{z}_t$ is an interpolation between $\boldsymbol{z}_0$ and $\boldsymbol{z}_1$ at time $t$:

$$\boldsymbol{z}_t = (1 - t)\boldsymbol{z}_0 + t\boldsymbol{z}_1, \tag{8.6.7}$$

where $\boldsymbol{z}_0 \sim p(\boldsymbol{z})$ is a sample from the learned representation distribution and $\boldsymbol{z}_1 \sim \mathcal{N}(0, \boldsymbol{I})$. The training objective for the denoiser can then be expressed as:

$$\min_{\theta} \mathcal{L}_{\text{FM}}(\theta) = \mathbb{E}_{\boldsymbol{z}_0 \sim p(\boldsymbol{z}), \boldsymbol{z}_1 \sim \mathcal{N}(0,I), t \sim \text{Unif}(0,1)} \left[ \| \bar{\boldsymbol{z}}_\theta(t, \boldsymbol{z}_t) - (\boldsymbol{z}_1 - \boldsymbol{z}_0) \|_2^2 \right]. \quad (8.6.8)$$

Typically, the denoising vector field $\bar{\boldsymbol{z}}_\theta(t, \boldsymbol{z}_t)$ are modeled by a deep network. Popular choices of the deep network can be either a U-Net or a diffusion image transformer (DiT). Below we provide more details about how they can be implemented in practice.

### 8.6.4 Realizing Denoising with a U-Net

The dominant architecture for implementing the denoising network $\boldsymbol{\epsilon}_\theta$ in diffusion models has been the *U-Net* [RFB15a], an encoder-decoder convolutional network with skip connections originally developed for biomedical image segmentation. Ho et al. [HJA20] first adopted the U-Net as the backbone for Denoising Diffusion Probabilistic Models (DDPM), establishing it as the default architecture for the field. Subsequently, Dhariwal and Nichol [DN21a] conducted a systematic series of architectural ablations that substantially improved the U-Net design, producing what they call the *Ablated Diffusion Model* (ADM). Their improved model was the first diffusion model to surpass GANs in sample quality on ImageNet. Building on these foundations, Rombach et al. [RBL+22] moved the U-Net from pixel space into the latent space of a pretrained VAE (Section 8.6.1) and introduced cross-attention layers for flexible multi-modal conditioning, giving rise to the *Latent Diffusion Model* (LDM) that underpins Stable Diffusion. In this section, we trace the evolution of the U-Net backbone through these three stages, before discussing the architectural limitations that have motivated the recent transition to transformer-based alternatives (Section 8.6.5).

**Encoder-Decoder Structure with Skip Connections.** The U-Net follows an encoder-decoder structure. The encoder progressively downsamples the input through a series of resolution stages, extracting features at multiple spatial scales; the decoder then upsamples these features back to the original resolution to produce the output prediction. The distinguishing feature of the U-Net is its *skip connections*: at each resolution level, the encoder feature map is concatenated channel-wise to the corresponding decoder feature map. This allows the decoder to combine coarse, semantically rich features from the bottleneck with fine-grained spatial details from earlier encoder layers, which is essential for producing spatially precise predictions.

In the DDPM U-Net [HJA20], the architecture begins with an input convolution that projects the noised input into a feature space with a base channel count (e.g., 128). The encoder then proceeds through multiple resolution stages—four for $32 \times 32$ inputs, or six for $256 \times 256$ inputs—where each stage consists of two convolutional residual blocks. Spatial downsampling by a factor of $2\times$ is performed between consecutive stages via strided convolutions. Self-attention layers are inserted at the $16 \times 16$ resolution level between the convolutional blocks, providing limited long-range modeling at low resolution. At the bottleneck (the lowest resolution), a middle block with additional residual blocks and self-attention further refines the features. The decoder mirrors the encoder: residual blocks at each resolution level followed by upsampling via transposed convolutions, with skip connections concatenating the corresponding encoder activations. The architecture uses group normalization [WH20] throughout, replacing the weight normalization from which the design originates.

When the U-Net operates in pixel space, its input and output are the noised image $\boldsymbol{x}_t \in \mathbb{R}^{H\times W\times 3}$ and the predicted noise $\hat{\boldsymbol{\epsilon}} \in \mathbb{R}^{H\times W\times 3}$, respectively. When operating in the latent space of a pretrained VAE (as in LDM), the input becomes $\boldsymbol{z}_t \in \mathbb{R}^{h\times w\times c}$ and the output is $\hat{\boldsymbol{\epsilon}} \in \mathbb{R}^{h\times w\times c}$, with $h = H/f$, $w = W/f$ as defined in Section 8.6.1.

**Timestep Conditioning via Adaptive Group Normalization.** The denoising network must be conditioned on the diffusion timestep $t$ so that it can adapt its behavior across different noise levels. Following the positional encoding scheme from the Transformer [VSP+17b], the timestep $t$ is first mapped to a sinusoidal frequency embedding:

$$\mathrm{Embed}(t)_{2i} = \sin\left(\frac{t}{10000^{2i/d_e}}\right), \quad \mathrm{Embed}(t)_{2i+1} = \cos\left(\frac{t}{10000^{2i/d_e}}\right), \quad (8.6.9)$$

where $d_e$ is the embedding dimension and $i$ indexes its entries. This embedding is then passed through a small MLP (typically two linear layers with a nonlinearity) to produce a timestep vector $\boldsymbol{e}_t \in \mathbb{R}^{d_e}$.

Rather than simply concatenating $\boldsymbol{e}_t$ to the input, the U-Net injects the timestep information into every residual block via *Adaptive Group Normalization* (AdaGN) [DN21a]. After the first convolution within each residual block, a group normalization operation is applied, followed by an affine modulation whose scale and shift parameters are regressed from the timestep embedding:

$$\mathrm{AdaGN}(\boldsymbol{h}, \boldsymbol{e}_t) = \boldsymbol{y}_s \odot \mathrm{GroupNorm}(\boldsymbol{h}) + \boldsymbol{y}_b, \quad (8.6.10)$$

where $\boldsymbol{h}$ denotes the intermediate feature map, $[\boldsymbol{y}_s, \boldsymbol{y}_b] = \mathrm{Linear}(\boldsymbol{e}_t)$ are the scale and shift vectors obtained from a linear projection of the timestep embedding, and $\odot$ denotes element-wise multiplication. This mechanism is analogous to adaptive instance normalization [DSK17] and FiLM conditioning [PSV+18], and allows each layer to modulate its activations based on the current noise level. Ablation experiments in [DN21a] show that removing AdaGN degrades FID by over 2 points, confirming its importance.

For class-conditional generation, a class embedding $\mathrm{Embed}(c)$ (a learned lookup table) is simply added to the timestep embedding before injection, so that $\boldsymbol{e}_t$ is replaced by $\boldsymbol{e}_t + \mathrm{Embed}(c)$ in Eq. (8.6.10). This mechanism is lightweight but limited to categorical labels; richer conditioning modalities such as text require the cross-attention mechanism introduced in Section 8.6.4.

### Architecture Improvements: From DDPM to ADM

Dhariwal and Nichol [DN21a] conducted a systematic ablation study to identify which architectural modifications most improve sample quality. Starting from the baseline DDPM U-Net [HJA20], they explored the following changes on ImageNet $128 \times 128$: (a) expanding self-attention from the $16 \times 16$ resolution to three resolutions ($32 \times 32$, $16 \times 16$, $8 \times 8$); (b) increasing the number of attention heads, using 64 channels per head to better match the Transformer convention; (c) adopting the BigGAN [BDS19] residual block for both upsampling and downsampling, which improves gradient flow; and (d) including the AdaGN conditioning layer (described above) in every residual block by default. All of these changes improve FID, and their effects compound positively. The resulting model, which they name ADM, serves as the new default U-Net design for subsequent work.

In addition to architectural improvements, Dhariwal and Nichol introduced *classifier guidance* as a mechanism for trading off sample diversity against fidelity. A classifier $p_\varphi(c|\boldsymbol{z}_t, t)$ is trained on noisy images, and its gradients are used to steer the sampling trajectory toward high-probability regions of the desired class:

$$\hat{\boldsymbol{\mu}}_\theta(\boldsymbol{z}_t, t) = \boldsymbol{\mu}_\theta(\boldsymbol{z}_t, t) + s\, \boldsymbol{\Sigma}_\theta(\boldsymbol{z}_t, t)\, \nabla_{\boldsymbol{z}_t} \log p_\varphi(c|\boldsymbol{z}_t), \tag{8.6.11}$$

where $s > 1$ is a guidance scale that sharpens the effective class distribution. With this technique, ADM achieved a state-of-the-art FID of 4.59 on class-conditional ImageNet $256 \times 256$, surpassing the long-standing GAN benchmark of BigGAN-deep (FID 6.95) for the first time. The guided model (ADM-G) further improved FID to 3.94 when combined with an upsampling stack (ADM-U).

The main drawback of classifier guidance is that it requires training a separate classifier on noisy images. This limitation was subsequently addressed by *classifier-free guidance* [HS22b], which avoids the external classifier entirely by jointly learning conditional and unconditional denoising within a single model—a technique we describe in detail in Section 8.6.5.

### The U-Net in Latent Diffusion (Stable Diffusion)

While ADM operates directly in pixel space, Rombach et al. [RBL+22] observed that much of the computational burden is spent modeling perceptually irrelevant high-frequency details. Their key idea is to separate *perceptual compression* from *generative modeling*: a pretrained autoencoder (Section 8.6.1) first compresses the image into a low-dimensional latent space, and the diffusion model then operates entirely within this latent space. For a $256 \times 256$ image with downsampling factor $f = 8$, the U-Net processes a $32 \times 32 \times 4$ latent tensor rather than a $256 \times 256 \times 3$ pixel tensor—a reduction of roughly $48\times$ in spatial elements. This yields a speedup of at least $2.7\times$ in both training and sampling throughput while improving FID.

Beyond computational savings, the latent diffusion framework introduces a general-purpose *cross-attention conditioning mechanism* that enables the U-Net to condition on arbitrary modalities such as text prompts, semantic maps, or bounding boxes. In brief, a domain-specific encoder $\tau_\theta$ maps the conditioning input $\boldsymbol{y}$ into an intermediate representation $\tau_\theta(\boldsymbol{y}) \in \mathbb{R}^{M \times d_\tau}$, which is then injected into the U-Net at multiple resolution levels via cross-attention layers: the image features serve as queries while the conditioning embeddings serve as keys and values, allowing each spatial location to attend to relevant parts of the conditioning signal. We defer the full mathematical description of this cross-attention mechanism and the conditional training objective to Section 8.7.2.

In Stable Diffusion [RBL+22], the text encoder $\tau_\theta$ is instantiated as the pre-trained CLIP ViT-L/14 model [RKH+21b], which maps text prompts to a sequence of token embeddings. The U-Net itself retains the ADM improvements—BigGAN residual blocks, attention at three resolution levels, AdaGN for timestep injection—and additionally interleaves cross-attention layers at these same resolution levels for text conditioning. The full model contains approximately 860M parameters in the U-Net and 123M in the text encoder.

#### Limitations of the U-Net Architecture

Despite its success, the U-Net has several limitations that have motivated alternative backbones. First, **scaling** a U-Net is cumbersome: its encoder-decoder structure involves many coupled design choices (channel multipliers, attention placement, up/downsampling block types), and it is unclear how to allocate additional compute optimally. In contrast, the plain Transformer scales along two simple axes, depth and width, and exhibits well-characterized scaling laws based on empirical studies [KMH+20]. Second, the U-Net's convolutional layers encode a strong **spatial inductive bias** (locality, translation equivariance) that helps in low-data regimes but limits long-range modeling on large datasets. Self-attention partially addresses this, yet in standard U-Net designs it is applied only at low-resolution stages due to its quadratic cost, leaving high-resolution features entirely local. Third, the U-Net is a **image-specific** architecture that does not readily generalize so well to other modalities. The Transformer, having proven effective across language, vision, audio, and multimodal settings [VSP+17b], offers a unified backbone that can leverage cross-domain scaling insights.

These considerations motivated the Diffusion Transformer (DiT) [PX23], which replaces the U-Net entirely with a plain Transformer operating on latent patches. As we describe in Section 8.6.5, DiT matches and surpasses the best U-Net-based models while exhibiting cleaner scaling behavior (Gflops–FID correlation of $-0.93$) and a simpler architecture.

### 8.6.5 Realizing Denoising with a Diffusion Transformer

An alternative to the U-Net for implementing the denoising network $\boldsymbol{\epsilon}_\theta$ is the *Diffusion Transformer (DiT)* [PX23], which replaces convolutions entirely with transformer blocks. The key insight is that the U-Net's spatial inductive bias is *not crucial* to the performance of diffusion models: a standard transformer operating on sequences of latent patches can match and surpass the U-Net while inheriting the favorable scaling properties of the transformer architecture class [VSP+17b]. Unlike the U-Net, the DiT has no explicit spatial inductive bias—all spatial structure is learned from data. This generality enables the same architecture to handle images, video, and other modalities with minimal modification.

DiT operates within the latent space of a pretrained VAE (Section 8.6.1), making the complete image generation pipeline a hybrid architecture: a convolutional VAE for perceptual compression and a transformer-based DDPM for generative modeling. Following the Vision Transformer (ViT) [DBK+21], the input latent is divided into patches and linearly embedded as a token sequence. We describe the architecture in detail below.

**Patchify and Positional Encoding.** The input to DiT is a noised latent $\boldsymbol{z}_t \in \mathbb{R}^{h \times w \times c}$, where $h = H/f$ and $w = W/f$ are the spatial dimensions of the latent space defined by the VAE encoder (e.g., $32 \times 32 \times 4$ for $256 \times 256$ images with $f = 8$, $c = 4$). The first layer, called *patchify*, partitions $\boldsymbol{z}_t$ into a grid of non-overlapping patches of size $p \times p$ and linearly embeds each patch into a $d$-dimensional token representation. This produces a sequence of

$$T = \left(\frac{h}{p}\right) \times \left(\frac{w}{p}\right) \tag{8.6.12}$$

tokens in $\mathbb{R}^d$. Standard sinusoidal positional embeddings (the sine-cosine variant from ViT) are added to each token to encode spatial position. The patch size $p$ is a key design parameter: halving $p$ quadruples $T$ and thus at least quadruples the total Gflops of the transformer, while leaving the parameter count essentially unchanged. In [PX23], the design space includes $p \in \{2, 4, 8\}$; the finest setting $p = 2$ yields the best generation quality.

**DiT Block with adaLN-Zero.** After patchification, the token sequence is processed by a stack of $N$ transformer blocks. Each block follows the standard transformer structure [VSP+17b]: layer normalization, multi-head self-attention (MHSA), a second layer normalization, and a pointwise feedforward network (MLP with GELU activations).

The central design question is how to inject the conditioning information—the diffusion timestep $t$ and the class label $c$—into each block. Peebles and Xie [PX23] compare four strategies: (1) *in-context conditioning*, which appends $t$ and $c$ as extra tokens in the input sequence; (2) *cross-attention*, which introduces an additional multi-head cross-attention layer that attends to the conditioning embeddings; (3) *adaptive layer normalization (adaLN)*, which regresses the scale and shift parameters of each layer normalization from the conditioning signal; and (4) *adaLN-Zero*, which extends adaLN with an additional gating mechanism initialized to zero. Among these, adaLN-Zero achieves the lowest FID while being the most compute-efficient, and is therefore adopted as the default.

In the adaLN-Zero block, the conditioning information is first formed by summing the embeddings of the timestep and class label:

$$\boldsymbol{e} = \text{Embed}(t) + \text{Embed}(c), \tag{8.6.13}$$

where Embed($t$) is obtained by passing a sinusoidal frequency embedding of $t$ through a two-layer MLP with SiLU activations, and Embed($c$) is a learned class embedding. From $\boldsymbol{e}$, a linear layer regresses six sets of dimension-wise parameters: $\boldsymbol{\gamma}_1, \boldsymbol{\beta}_1, \boldsymbol{\alpha}_1$ for the self-attention sub-block and $\boldsymbol{\gamma}_2, \boldsymbol{\beta}_2, \boldsymbol{\alpha}_2$ for the feedforward sub-block. The adaptive layer normalization applies as:

$$\text{adaLN}(\boldsymbol{h}, \boldsymbol{\gamma}, \boldsymbol{\beta}) = \boldsymbol{\gamma} \odot \text{LayerNorm}(\boldsymbol{h}) + \boldsymbol{\beta}, \tag{8.6.14}$$

where $\odot$ denotes element-wise multiplication. The parameters $\boldsymbol{\alpha}_1$ and $\boldsymbol{\alpha}_2$ serve as dimension-wise *gating factors* applied immediately before each residual connection:

$$\begin{aligned} \boldsymbol{h}' &= \boldsymbol{h} + \boldsymbol{\alpha}_1 \odot \text{MHSA}\big(\text{adaLN}(\boldsymbol{h}, \boldsymbol{\gamma}_1, \boldsymbol{\beta}_1)\big), \\ \boldsymbol{h}'' &= \boldsymbol{h}' + \boldsymbol{\alpha}_2 \odot \text{MLP}\big(\text{adaLN}(\boldsymbol{h}', \boldsymbol{\gamma}_2, \boldsymbol{\beta}_2)\big). \end{aligned} \tag{8.6.15}$$

The "Zero" in adaLN-Zero refers to the initialization: all $\boldsymbol{\alpha}$ parameters are initialized to the zero vector by the MLP that regresses them, so that each DiT block acts as the *identity function* at the start of training. This zero-initialization strategy, inspired by similar practices in ResNets [GDG+17] and diffusion U-Nets, is important for training stability and yields significantly lower FID than vanilla adaLN.

We note that adaLN applies the *same* conditioning function to all tokens uniformly. For text-to-image generation where the conditioning signal is a variable-length sequence of text tokens, two alternative approaches are common: (1) cross-attention, where text token embeddings from a frozen encoder are attended to by the image tokens; or (2) prepending the text tokens directly to the image token sequence, allowing the transformer's self-attention to jointly process both modalities.

**Output Projection.** After the final DiT block, the model must decode the sequence of token representations back into a spatial prediction. A final adaptive layer normalization (conditioned on $\boldsymbol{e}$) is applied, followed by a linear layer that projects each token into a tensor of shape $p \times p \times 2c$. The factor of $2c$ accounts for the two quantities the model predicts: the noise estimate $\hat{\boldsymbol{\epsilon}} \in \mathbb{R}^{h \times w \times c}$ and the diagonal covariance $\boldsymbol{\Sigma}_\theta \in \mathbb{R}^{h \times w \times c}$. The decoded tokens are then rearranged (un-patchified) back into their original spatial layout of $\mathbb{R}^{h \times w \times c}$ for each output.

### Training

The DiT is trained as a class-conditional latent diffusion model on the ImageNet dataset [DDS+09b] at $256 \times 256$ and $512 \times 512$ image resolution. The diffusion process operates on the latent representations produced by the pretrained VAE from Section 8.6.1: for $256 \times 256$ images, the input to DiT is $\boldsymbol{z}_t \in \mathbb{R}^{32 \times 32 \times 4}$; for $512 \times 512$ images, it is $\boldsymbol{z}_t \in \mathbb{R}^{64 \times 64 \times 4}$.

The training objective combines two terms. The noise prediction loss trains $\boldsymbol{\epsilon}_\theta$ via the simple mean-squared error:

$$\mathcal{L}_{\text{simple}}(\theta) = \mathbb{E}_{\boldsymbol{z}_0,\boldsymbol{\epsilon},t}\left[\|\boldsymbol{\epsilon}_\theta(\boldsymbol{z}_t, t) - \boldsymbol{\epsilon}\|_2^2\right], \tag{8.6.16}$$

where $\boldsymbol{z}_0$ is the clean latent, $\boldsymbol{\epsilon} \sim \mathcal{N}(\mathbf{0}, \boldsymbol{I})$ is the sampled noise, and $\boldsymbol{z}_t = \sqrt{\bar{\alpha}_t}\,\boldsymbol{z}_0 + \sqrt{1-\bar{\alpha}_t}\,\boldsymbol{\epsilon}$ is the noised latent at timestep $t$. The learned covariance $\boldsymbol{\Sigma}_\theta$ is additionally trained with the full variational lower bound $D_{KL}$ term, following the approach of Nichol and Dhariwal [ND21]. A linear variance schedule is used with $t_{\max} = 1000$ steps, ranging from $1 \times 10^{-4}$ to $2 \times 10^{-2}$.

All models are trained with AdamW [LH19] using a constant learning rate of $1 \times 10^{-4}$, no weight decay, and a batch size of 256. Data augmentation consists solely of random horizontal flips. An exponential moving average (EMA) of the model weights is maintained with a decay of 0.9999; all reported results use the EMA model. Notably, unlike supervised ViT training, DiT requires *no* learning rate warmup and *no* regularization—training is highly stable across all model configurations, with no observed loss spikes.

### Classifier-Free Guidance

To produce class-controlled sampling, DiT can easily employ *classifier-free guidance* [HS22b]. During training, the class label $c$ is randomly dropped with some probability and replaced by a learned null embedding $\varnothing$, so that the model learns both the conditional $\boldsymbol{\epsilon}_\theta(\boldsymbol{z}_t, c)$ and unconditional $\boldsymbol{\epsilon}_\theta(\boldsymbol{z}_t, \varnothing)$ noise predictions. At inference time, the guided noise prediction is computed as:

$$\hat{\boldsymbol{\epsilon}}_\theta(\boldsymbol{z}_t, c) = \boldsymbol{\epsilon}_\theta(\boldsymbol{z}_t, \varnothing) + s \cdot \big(\boldsymbol{\epsilon}_\theta(\boldsymbol{z}_t, c) - \boldsymbol{\epsilon}_\theta(\boldsymbol{z}_t, \varnothing)\big), \tag{8.6.17}$$

where $s > 1$ is the guidance scale ($s = 1$ recovers standard unguided sampling). The motivation follows from Bayes' rule: since $\nabla_{\boldsymbol{z}_t} \log p(c|\boldsymbol{z}_t) \propto \nabla_{\boldsymbol{z}_t} \log p(\boldsymbol{z}_t|c) - \nabla_{\boldsymbol{z}_t} \log p(\boldsymbol{z}_t)$, the guided estimate in Eq. (8.6.17) effectively steers the sampling trajectory toward regions of high conditional likelihood $p(c|\boldsymbol{z}_t)$. This technique has a dramatic effect on generation quality: DiT-XL/2 achieves an FID of 9.62 without guidance, which improves to 2.27 with a guidance scale of $s = 1.5$—surpassing all prior diffusion models on the class-conditional ImageNet $256 \times 256$ benchmark.

### Generation

Once the denoising network $\boldsymbol{\epsilon}_\theta$ (i.e., the DiT) has been trained, new images can be generated by running the learned reverse process. Starting from pure noise $\boldsymbol{z}_{t_{\max}} \sim \mathcal{N}(\mathbf{0}, \boldsymbol{I})$, the model iteratively denoises by sampling $\boldsymbol{z}_{t-1} \sim p_\theta(\boldsymbol{z}_{t-1}|\boldsymbol{z}_t)$ at each step, using the predicted noise $\hat{\boldsymbol{\epsilon}}_\theta$ and covariance $\boldsymbol{\Sigma}_\theta$ to parameterize the reverse transition. After $t_{\max}$ denoising steps, the resulting clean latent $\boldsymbol{z}_0$ is decoded back to the image space via the VAE decoder: $\tilde{\boldsymbol{x}} = \mathcal{D}(\boldsymbol{z}_0) \in \mathbb{R}^{H \times W \times 3}$.

## 8.7 Conditioned Image Representation and Generation

In the previous sections, we established the fundamental machinery for generative modeling: an autoencoder architecture to encode the data $\boldsymbol{x}$ into a (compressed and more structured) latent $\boldsymbol{z}$:

$$\boldsymbol{x} \xrightarrow{\mathcal{E}} \boldsymbol{z} \xrightarrow{\mathcal{D}} \hat{\boldsymbol{x}}. \tag{8.7.1}$$

Then, a generative process is learned to model the distribution of these latents $p(\boldsymbol{z})$. In this section, we shift our focus from *unconditional generation* to *control*. We assume the representation $\boldsymbol{z}$ is already learned (via methods detailed in Chapter 7); our specific task here is to steer the sampling process using a user-provided "control" signal $\boldsymbol{c}$. That is, we aim to sample from the conditional distribution $p(\boldsymbol{z}|\boldsymbol{c})$. In practice, $\boldsymbol{c}$ could represent partial attributes of the desired sample image: masked image, a semantic mask, class information, or a text caption describing the desired content.

### 8.7.1 Task Formulation

Formally, the goal of conditioned generation is to learn a mapping that transforms a simple source distribution (typically Gaussian noise) into a target conditional distribution. The generated feature $\hat{\boldsymbol{z}}$ is subsequently decoded to a corresponding image $\hat{\boldsymbol{x}}$. As illustrated as follows, the system is defined by its inputs and outputs:

1. **Stochastic Input ($\epsilon$):** A noise variable sampled from a standard normal prior $\epsilon \sim \mathcal{N}(\mathbf{0}, \boldsymbol{I})$. This provides the "seed" for diversity, ensuring that we can generate multiple variations of the same concept.

2. **Control Signal ($\boldsymbol{c}$):** A structured input defining the desired semantic or spatial content. This can be a text caption (e.g., "a red car"), a semantic segmentation mask, or a layout.

3. **Output ($\hat{\boldsymbol{x}}$):** A high-fidelity data sample $\hat{\boldsymbol{x}} = \mathcal{D}(\hat{\boldsymbol{z}})$ that aligns with the user's intent $\boldsymbol{c}$ while remaining on the natural image manifold.

We employ a domain-specific **Condition Encoder**, denoted as $\tau_\varphi$, to project the control signal $\boldsymbol{c}$ into a useful embedding space. Hence, the overall controlled generation process can be formally described as a composition of a generator $G_\theta$ and the decoder $\mathcal{D}$:

$$\underbrace{\epsilon \sim \mathcal{N}(\mathbf{0}, \boldsymbol{I})}_{\text{Source}}, \boldsymbol{c} \xrightarrow{G_\theta(\epsilon, \tau_\varphi(\boldsymbol{c}))} \hat{\boldsymbol{z}} \sim p(\boldsymbol{z} \mid \boldsymbol{c}) \xrightarrow{\mathcal{D}(\hat{\boldsymbol{z}})} \hat{\boldsymbol{x}} \approx \boldsymbol{x}. \tag{8.7.2}$$

**The Modality Gap.** While the task is conceptually straightforward—mapping $(\epsilon, \boldsymbol{c}) \to \boldsymbol{x}$—it requires bridging disjoint topological spaces, particularly in text-to-image generation. The visual data $\boldsymbol{x}$ lies in a continuous, high-dimensional pixel space $\mathbb{R}^{H \times W \times 3}$, whereas the condition $\boldsymbol{c}$ typically resides in a discrete symbolic space (language). There is no natural Euclidean metric to measure the distance $\|\boldsymbol{x} - \boldsymbol{c}\|$ directly. Therefore, the critical component $\tau_\varphi(\boldsymbol{c})$ mentioned above must align the modalities. As discussed in Chapter 7, modern approaches rely on pre-trained contrastive encoders (like CLIP) or large language models (like T5 [RSR+20]) to map discrete language tokens into a continuous semantic embedding space. The generative backbone $G_\theta$ (i.e., Denoising U-Net) must then learn to map the noise distribution to the data distribution, guided by these semantic embeddings via a specific fusion mechanism, such as Cross-Attention.

**The One-to-Many Inverse Problem.** Conditioning is inherently ill-posed because the control signal is usually *under-determined*. A text prompt $\boldsymbol{c} =$ "a dog" provides semantic category information but omits the dog's pose, lighting, background, and fur pattern. Consequently, the inverse mapping $\boldsymbol{c} \to \boldsymbol{x}$ is not a single point, but a complex multi-modal distribution. If we were to train a simple deterministic regression model to minimize the Euclidean error $\|\boldsymbol{x} - G_\theta(\boldsymbol{c})\|^2$, the model would converge to the conditional mean $\mathbb{E}[\boldsymbol{x}|\boldsymbol{c}]$—an average of all possible dogs—resulting in a blurry, unrealistic output. To generate high-fidelity results, we can utilize generative frameworks (such as U-Net or DiT mentioned in Section 8.6) to model the full conditional distribution $p(\boldsymbol{z}|\boldsymbol{c})$, allowing us to sample diverse, sharp instances that satisfy the condition.

### 8.7.2 The Basic Realization: Text-to-Image Generation

To realize the conditional framework defined in 8.7.1, we require a concrete architecture capable of bridging the modality gap between discrete text tokens and continuous pixel arrays. As illustrated in Figure 8.16, modern state-of-the-art models (such as Stable Diffusion [RBL+22]) decouple this complex task into three independent components: (1) a **Perceptual Compression** module (VAE) to translate between pixels and latents, (2) a **Semantic Conditioning** interface (CLIP Text Encoder) to interpret prompts, and (3) a **Latent Generation** backbone (U-Net) to synthesize content. We detail the implementation of each component below.

**Component 1: Perceptual Compression (The VAE).** The first pillar of the architecture is a Variational Autoencoder (VAE) trained to abstract away high-frequency details that are perceptually insignificant but computationally expensive to model. As we introduced in Section 8.6.1, this component defines the **Latent Space** in which the generation occurs.

- **The Encoder ($\mathcal{E}$):** The encoder maps an input RGB image $\boldsymbol{x} \in \mathbb{R}^{H \times W \times 3}$ into a low-dimensional latent representation $\boldsymbol{z} = \mathcal{E}(\boldsymbol{x}) \in \mathbb{R}^{h \times w \times c}$. Following [RBL+22], it utilizes a KL-regularized encoder with a downsampling factor of $f = H/h = W/w = 8$. For a standard input of $512 \times 512 \times 3$, this yields a compact latent tensor of dimensions $64 \times 64 \times 4$. In the context of the diffusion process (Component 3), this encoded representation serves as the "clean" ground-truth latent $\boldsymbol{z}_0$.

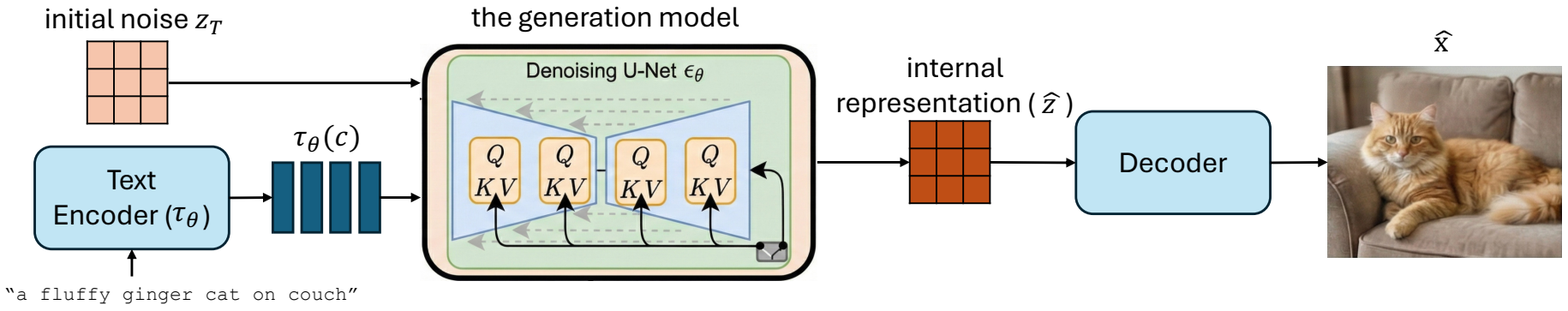

Figure 8.16: **The three-stage inference pipeline of Stable Diffusion.** (1) **Text Encoder** (e.g., CLIP) converts the input prompt $\boldsymbol{c}$ into semantic embeddings. (2) **Generation Model** (a time-conditional U-Net $\epsilon_\theta$) iteratively denoises a random latent tensor $\boldsymbol{z}_T$ to produce a clean intermediate representation $\hat{\boldsymbol{z}}$, conditioned on the text embeddings via cross-attention. (3) **Decoder** (from a VAE) maps the denoised latent $\hat{\boldsymbol{z}}$ back to the pixel space to synthesize the final high-resolution image $\hat{\boldsymbol{x}}$. *Note: The VAE Encoder is utilized only during training and is omitted from this inference pipeline.*

- **The Decoder ($\mathcal{D}$):** The decoder learns the reverse mapping to reconstruct the image from the latent space, denoted as $\tilde{\boldsymbol{x}} = \mathcal{D}(\boldsymbol{z}) = \mathcal{D}(\mathcal{E}(\boldsymbol{x}))$. It is trained to minimize the perceptual difference between $\boldsymbol{x}$ and $\tilde{\boldsymbol{x}}$. During the inference phase of the diffusion model, this decoder is frozen and used to map the generated (denoised) latent vectors back to the high-resolution pixel space.

Note that the VAE is pre-trained with a combination of perceptual loss and patch-based adversarial loss to ensure the quality of reconstructions. Once trained, its weights are frozen, and it serves as a static interface for the diffusion model.

**Component 2: Semantic Conditioning (The Text Encoder).** To guide the generation process, we require a robust semantic representation of the user's prompt $\boldsymbol{c}$ (e.g., "A fluffy ginger cat"). Instead of training a text encoder from scratch, Latent Diffusion Models typically leverage the frozen text encoder from CLIP [RKH+21b]. This pre-trained transformer maps the input text tokens to a sequence of semantic embeddings $\tau_\theta(\boldsymbol{c}) \in \mathbb{R}^{d_\tau \times M}$, where $M$ is the sequence length (e.g., 77 tokens) and $d_\tau$ is the embedding dimension (typically 768 or 1024). These embeddings serve as the semantic "Key" and "Value" sources for the generation backbone.

**Component 3: Latent Generation (The Time-Conditional U-Net).** The core generation engine is implemented as a Time-Conditional U-Net $\epsilon_\theta$. Unlike pixel-space models, this network operates entirely within the compressed latent space. Its architecture is composed of a series of ResNet blocks for feature extraction and Cross-Attention layers for conditioning.

- **Input Streams:** The model processes three distinct signals:
  1. **Visual Stream:** The noisy latent $\boldsymbol{z}_t \in \mathbb{R}^{h\times w\times 4}$, which represents the image at the current diffusion step.
  2. **Timestep ($t$):** To inform the model of the current noise level, the scalar timestep $t \in \{1,\dots,T\}$ is mapped to a sinusoidal embedding (encoding low and high frequency components), followed by a Multi-Layer Perceptron (MLP) composed of 'Linear' → 'SiLU' → 'Linear' layers. This produces a global time embedding vector (e.g., 1280-dim) which is injected into every ResNet block in the U-Net.
  3. **Context Stream:** The text embeddings $\tau_\theta(\boldsymbol{c})$ from Component 2.
- **Conditioning Mechanism (Cross-Attention):** The interaction between the visual content and the text prompt is handled via Cross-Attention layers interleaved throughout the U-Net. Let $\varphi_i(\boldsymbol{z}_t) \in \mathbb{R}^{d_\epsilon^{(i)}\times N}$ denote the flattened spatial features at U-Net layer $i$ (where $N = h_i \times w_i$ is the number of visual tokens and $d_\epsilon^{(i)}$ is the channel depth, e.g., 320, 640, or 1280), and let $\tau_\theta(\boldsymbol{c}) \in \mathbb{R}^{d_\tau\times M}$ denote the text embeddings.

  Since the visual dimension $d_\epsilon^{(i)}$ and text dimension $d_\tau$ differ, we project both into a common subspace of dimension $d$ (the attention head dimension) using learnable weights. We first define the Query ($\boldsymbol{Q}$), Key ($\boldsymbol{K}$), and Value ($\boldsymbol{V}$) matrices:

$$\boldsymbol{Q} = \boldsymbol{U}_Q^{(i)}\varphi_i(\boldsymbol{z}_t), \quad \boldsymbol{K} = \boldsymbol{U}_K^{(i)}\tau_\theta(\boldsymbol{c}), \quad \boldsymbol{V} = \boldsymbol{U}_V^{(i)}\tau_\theta(\boldsymbol{c}), \tag{8.7.3}$$

  where $\boldsymbol{U}_Q^{(i)} \in \mathbb{R}^{d\times d_\epsilon^{(i)}}$ and $\boldsymbol{U}_K^{(i)}, \boldsymbol{U}_V^{(i)} \in \mathbb{R}^{d\times d_\tau}$. Adapting the column-wise formulation from (8.4.2), the cross-attention operation is computed as:

$$\operatorname{CrossAttn}(\boldsymbol{Q},\boldsymbol{K},\boldsymbol{V}) \doteq \boldsymbol{V}\cdot \operatorname{softmax}\left(\frac{\boldsymbol{K}^\top\boldsymbol{Q}}{\sqrt{d}}\right), \tag{8.7.4}$$

  where the softmax is defined column-wise as

$$\operatorname{softmax}(\boldsymbol{A}) \doteq \begin{bmatrix}\operatorname{softmax}(\boldsymbol{a}_1) & \cdots & \operatorname{softmax}(\boldsymbol{a}_N)\end{bmatrix}, \tag{8.7.5}$$

$$\forall \boldsymbol{A} = \begin{bmatrix}\boldsymbol{a}_1,\dots,\boldsymbol{a}_N\end{bmatrix} \in \mathbb{R}^{M\times N}. \tag{8.7.6}$$

  In this context, the matrix $\boldsymbol{A} = \frac{\boldsymbol{K}^\top\boldsymbol{Q}}{\sqrt{d}}$ represents the alignment scores, and the column-wise softmax normalizes the attention weights over the $M$ text tokens for each of the $N$ visual regions. This mechanism allows the model to spatially align specific visual regions (e.g., "cat's eye") with specific text tokens.

**Training Objective.** The training process unifies these components. We sample a random image $\boldsymbol{x}$, compress it using the frozen Encoder $\boldsymbol{z}_0 = \mathcal{E}(\boldsymbol{x})$, and add Gaussian noise $\epsilon$ to produce $\boldsymbol{z}_t$. The U-Net $\epsilon_\theta$ is then trained to predict this noise, conditioned on the text $\boldsymbol{c}$ and timestep $t$:

$$\mathcal{L}_{\text{LDM}} = \mathbb{E}_{\mathcal{E}(\boldsymbol{x}),\boldsymbol{c},\epsilon\sim\mathcal{N}(0,1),t}\left[\|\epsilon - \epsilon_\theta(\boldsymbol{z}_t, t, \tau_\theta(\boldsymbol{c}))\|^2\right]. \tag{8.7.7}$$

During inference, the Encoder is discarded. The process starts from pure noise $\boldsymbol{z}_T \sim \mathcal{N}(\mathbf{0}, \boldsymbol{I})$, iteratively denoises it using the U-Net, and finally passes the resulting $\hat{\boldsymbol{z}}_0$ through the Decoder $\mathcal{D}$ to synthesize the image.

**Evaluation Metrics.** Since conditional generation lacks a single "correct" ground truth, we rely on statistical metrics to evaluate performance. We denote the distribution of real images as $p_r$ and the distribution of generated images as $p_g$.

- **FID (Fréchet Inception Distance):** This metric quantifies the realism and diversity of generated samples by measuring the distance between $p_r$ and $p_g$ in the feature space of a pre-trained Inception-v3 network. We approximate these feature distributions as multidimensional Gaussians $\mathcal{N}(\boldsymbol{\mu}_r, \boldsymbol{\Sigma}_r)$ and $\mathcal{N}(\boldsymbol{\mu}_g, \boldsymbol{\Sigma}_g)$. The FID score is then defined as the Fréchet distance between them:

$$\text{FID}(p_r, p_g) = \|\boldsymbol{\mu}_r - \boldsymbol{\mu}_g\|^2 + \text{Tr}\left(\boldsymbol{\Sigma}_r + \boldsymbol{\Sigma}_g - 2(\boldsymbol{\Sigma}_r\boldsymbol{\Sigma}_g)^{1/2}\right). \tag{8.7.8}$$

  A lower FID indicates that the generated distribution is statistically closer to the real distribution, implying higher image quality and diversity.

- **CLIP Score:** This metric measures the semantic alignment between a generated image $\hat{\boldsymbol{x}}$ and its conditioning prompt $\boldsymbol{c}$. Let $E_{\text{img}}$ and $E_{\text{txt}}$ be the image and text encoders of a pre-trained CLIP model as discussed in Chapter 7. The score is calculated as the cosine similarity between their embeddings:

$$\text{CLIP-Score}(\hat{\boldsymbol{x}}, \boldsymbol{c}) = \max\left(0, \frac{E_{\text{img}}(\hat{\boldsymbol{x}}) \cdot E_{\text{txt}}(\boldsymbol{c})}{\|E_{\text{img}}(\hat{\boldsymbol{x}})\|\|E_{\text{txt}}(\boldsymbol{c})\|}\right) \cdot w, \tag{8.7.9}$$

  where $w$ is a scaling factor (typically 2.5). A higher score indicates that the visual content faithfully reflects the textual description.

**Advanced Text Conditioned Architectures (SDXL and Flux).** To address limitations in resolution and prompt adherence, advanced architectures have evolved beyond the basic LDM. SDXL [PEL+23] introduces a dual-encoder strategy to capture richer textual nuances, computing embeddings from both CLIP ViT-L and OpenCLIP ViT-bigG and concatenating them along the channel axis. Regarding data handling, SDXL employs *Aspect Ratio Bucketing*, grouping training images of similar aspect ratios into batches to prevent the model from learning to crop subjects arbitrarily. Furthermore, it implements *Micro-Conditioning*, where metadata such as image resolution and crop coordinates are embedded and concatenated with the text condition $\boldsymbol{c}$, allowing the model to explicitly distinguish between centered and cropped subjects.

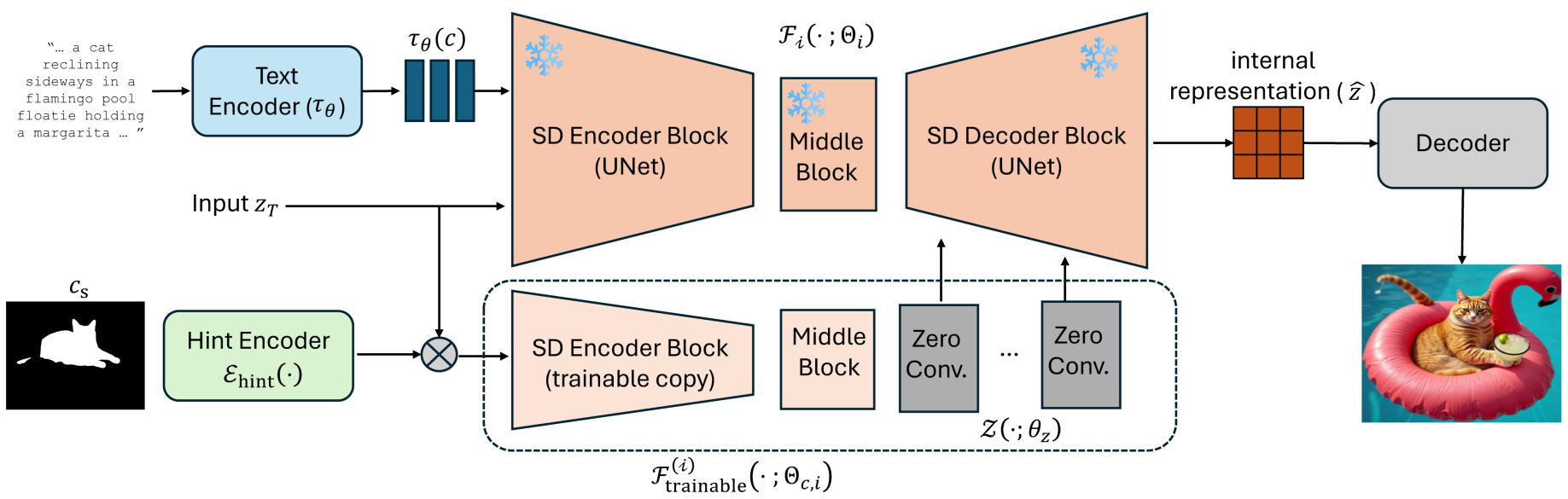

Figure 8.17: **ControlNet architecture pipeline.** A trainable copy of the Stable Diffusion (SD) U-Net encoder and middle blocks, conditioned on a hint image via the Hint Encoder $\mathcal{E}_{\text{hint}}$, injects feature maps into the frozen SD UNet decoder through zero convolutions to enable spatial control. (Image example from https://github.com/huggingface/diffusers/blob/main/docs/source/en/using-diffusers/controlnet.md).

Meanwhile, Flux [Lab24] represents a paradigm shift from the U-Net architecture to a Multimodal Diffusion Transformer (MMDiT). It processes text and visual tokens in a unified stream using Joint Attention blocks, enabled by modern transformer components like Rotary Positional Embeddings (RoPE) and RMSNorm. Mathematically, Flux replaces standard diffusion with **Rectified Flow Matching**. Unlike standard diffusion which predicts noise $\epsilon$, Flow Matching simplifies the training target to predicting a straight-line velocity field $v_\theta$ between noise and data. We define the process over a continuous time $t \in [0, 1]$, where $t = 0$ represents the noise distribution ($\boldsymbol{z}_0$) and $t = 1$ represents the data distribution ($\boldsymbol{z}_1$).[10] The objective is to minimize:

$$\mathcal{L}_{\text{RFM}} = \mathbb{E}_{t,\boldsymbol{z}_1,\boldsymbol{z}_0}\left[\|v_\theta(\boldsymbol{z}_t, t, \boldsymbol{c}) - (\boldsymbol{z}_1 - \boldsymbol{z}_0)\|^2\right], \tag{8.7.10}$$

where $\boldsymbol{z}_t = t\boldsymbol{z}_1 + (1-t)\boldsymbol{z}_0$ is the interpolated latent state. The model $v_\theta$ learns to predict the constant velocity $(\boldsymbol{z}_1 - \boldsymbol{z}_0)$ required to transport the noise $\boldsymbol{z}_0$ directly to the data $\boldsymbol{z}_1$ along a straight trajectory. This linearity significantly improves sampling efficiency and prompt adherence compared to the curved stochastic paths of traditional diffusion models.

### 8.7.3 Advanced Conditioning Mechanisms

While text conditioning provides semantic control, it struggles with fine-grained spatial constraints (e.g., specific poses or edge maps). To address this, Control-

[10] Note that this time convention ($0 \to$ Noise, $1 \to$ Data) is the inverse of the standard diffusion notation used in previous sections.

Net [ZRA23b] introduces a neural architecture that adds spatial control to large, pre-trained text-to-image diffusion models as shown in Figure 8.17.

**Architecture and Connectivity.** ControlNet operates by cloning the trainable parameters of a pre-trained diffusion model. Specifically, it creates a trainable copy of the 12 encoder blocks and 1 middle block of the Stable Diffusion U-Net, while keeping the original parameters $\Theta$ completely locked (executed under `torch.no_grad()`). This preserves the generation quality learned from billions of images while allowing the model to learn conditional tasks on small datasets.

To match the spatial resolution of the latent feature maps $\boldsymbol{z} \in \mathbb{R}^{h\times w\times c}$ (typically $64 \times 64$ as defined in Sec. 8.7.2), the spatial condition $\boldsymbol{c}_s$ (e.g., a $512 \times 512$ Canny edge map) is first processed by a lightweight **Hint Encoder** $\mathcal{E}_{\text{hint}}(\cdot)$. Unlike the VAE, this encoder is deeper but narrower, consisting of **eight** convolution layers with $3\times 3$ kernels and SiLU activations. It uses three stride-2 layers to downsample the resolution, gradually increasing channels ($16 \to 32 \to 96 \to 256$) before finally projecting to the model dimension (320 channels). Crucially, the final layer of this encoder is zero-initialized.

The critical innovation allowing these two streams to merge is the **Zero Convolution**—a $1\times 1$ convolutional layer $\mathcal{Z}(\cdot;\theta_z)$ initialized with both weights $W$ and bias $B$ set strictly to zero. The connectivity follows a specific pattern:

1. **Input Injection:** The encoded hint $\mathcal{E}_{\text{hint}}(\boldsymbol{c}_s)$ is added **only** to the first input block of the ControlNet copy.

2. **Output Injection:** The output of each trainable block in the ControlNet copy is passed through a Zero Convolution and then added to the corresponding **skip connection** of the frozen Stable Diffusion Decoder.

Formally, consider the $i$-th block of the U-Net. Let $\boldsymbol{h}_i$ denote the input feature map to this layer. We abstract the operation of this frozen block as a function $\mathcal{F}_i(\cdot;\Theta_i)$. Accordingly, let $\mathcal{F}^{(i)}_{\text{trainable}}(\cdot;\Theta_{c,i})$ denote its trainable copy in the ControlNet branch. The output $\boldsymbol{y}_i$ is computed as:

$$\boldsymbol{y}_i = \mathcal{F}_i(\boldsymbol{h}_i;\Theta_i) + \mathcal{Z}\left(\mathcal{F}^{(i)}_{\text{trainable}}(\boldsymbol{h}_i + \mathcal{E}_{\text{hint}}(\boldsymbol{c}_s);\Theta_{c,i});\theta_z\right). \tag{8.7.11}$$

Because the Zero Convolution is initialized to zero, the control term vanishes at the start of training ($\mathcal{Z}(\cdot) = 0$), ensuring the model behaves exactly like the original Stable Diffusion. However, the gradients for the weights $W$ are non-zero:

$$\frac{\partial \mathcal{Z}(I;\{W,B\})}{\partial W_{i,j}} = I_{p,i} \neq 0, \tag{8.7.12}$$

where $I$ is the input feature map. This ensures that although the initial influence is zero, learning commences immediately in the first backward pass.

**Training Dynamics and Experiments.** ControlNet is trained to minimize the standard diffusion objective, but with the additional spatial condition $\boldsymbol{c}_s$ injected into the noise predictor. The specific learning objective is:

$$\mathcal{L} = \mathbb{E}_{\boldsymbol{z}_0,t,\boldsymbol{c}_t,\boldsymbol{c}_s,\epsilon\sim\mathcal{N}(0,1)}\left[\|\epsilon - \epsilon_\theta(\boldsymbol{z}_t,t,\boldsymbol{c}_t,\boldsymbol{c}_s)\|^2\right], \tag{8.7.13}$$

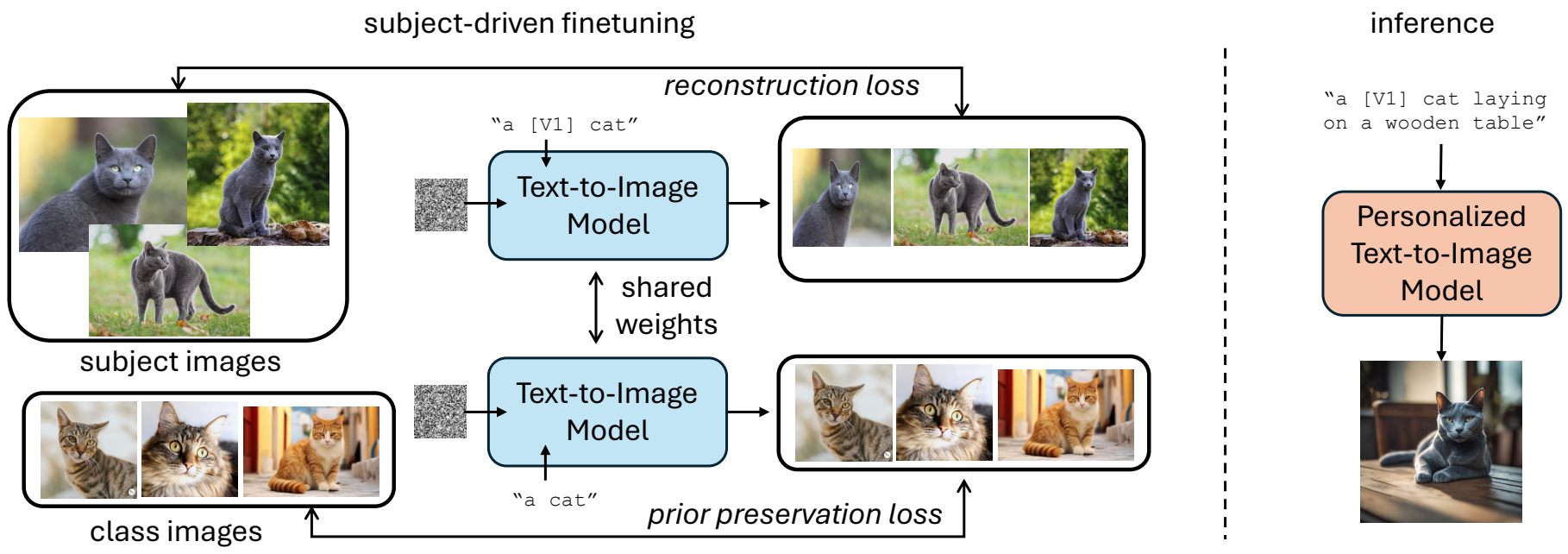

Figure 8.18: **The training and inference pipeline of DreamBooth.** During *subject-driven finetuning*, the text-to-image model is updated using a combined loss: a reconstruction loss to learn the specific subject (e.g., "[V] cat") and a prior preservation loss to maintain generic class knowledge (e.g., "a cat"). At *inference*, the personalized model can generate the specific subject in novel contexts (e.g., "laying on a wooden table") using the unique identifier.

where $\boldsymbol{c}_t$ represents the text prompt conditioning and $\boldsymbol{z}_t$ is the noisy latent. A unique phenomenon observed is "sudden convergence": the model initially ignores the spatial condition, then suddenly snaps to alignment as the error drops sharply. To force the model to rely on structural signals rather than just the text (which might contradict the edge map), a crucial strategy involves **CFG-Dropout**: randomly replacing 50% of the text prompts $\boldsymbol{c}_t$ with empty strings during training. This compels the network to extract semantic meaning directly from the spatial condition $\boldsymbol{c}_s$. Empirically, ControlNet is highly efficient, requiring only $\sim$ 23% more GPU memory compared to standard fine-tuning, yet effectively learning robust control from datasets as small as 1,000 samples.

**Subject-Driven Personalization.** While foundational models possess a vast semantic prior, they lack the ability to synthesize specific subject instances (e.g., a user's specific cat) consistently across different contexts. To address this, DreamBooth [RLJ+23] introduces a "personalization" fine-tuning protocol. The core challenge is implanting a new subject into the model's output domain without "catastrophic forgetting"—where the model loses its understanding of the general class (e.g., forgetting what a generic cat looks like) or suffers from "language drift" (overwriting the word "cat" with the specific cat's features).

**Identifier and Data Strategy.** Unlike standard training which requires millions of pairs, DreamBooth operates on a "few-shot" dataset provided by the user, typically containing just 3–5 images of the subject. To bind this subject to the model's text space, a unique identifier $[V]$ is required. The authors propose a rare-token selection strategy: rather than using random characters (which might be tokenized into common letters with strong priors), the method searches the tokenizer's vocabulary for rare token sequences (e.g., in the T5-XXL tokenizer range $\{5000, \ldots, 10000\}$) that have weak semantic associations. This identifier is always paired with a coarse Class Noun (e.g., "a $[V]$ cat") in the prompt. This "tethering" is crucial: it leverages the model's strong prior for the class (knowing how to render a "cat" in various poses) while binding the specific identity features to the identifier $[V]$.

**Prior Preservation Training.** To prevent the model from overfitting to the few-shot images and losing diversity, DreamBooth employs an autogenous class-specific prior preservation loss. As illustrated in Figure 8.18, the training process minimizes a combined objective:

$$\mathcal{L}_{\text{DB}} = \mathbb{E}_{\boldsymbol{z}_t, \boldsymbol{c}, \epsilon, t} \left[ \underbrace{\|\epsilon - \epsilon_\theta(\boldsymbol{z}_t, t, \boldsymbol{c}_{\text{subj}})\|^2}_{\text{Reconstruction Loss}} + \lambda \underbrace{\|\epsilon - \epsilon_\theta(\boldsymbol{z}'_t, t, \boldsymbol{c}_{\text{prior}})\|^2}_{\text{Prior Preservation Loss}} \right]. \tag{8.7.14}$$

The first term fine-tunes the model weights (typically the entire U-Net and text encoder) to reconstruct the subject images conditioned on the unique prompt $\boldsymbol{c}_{\text{subj}}$ (e.g., "a $[V]$ cat"). The second term acts as a regularizer: it supervises the model with its own generated samples of the generic class $\boldsymbol{c}_{\text{prior}}$ (e.g., "a cat"). This forces the model to retain its original understanding of the class distribution, ensuring that the identifier $[V]$ learns the specific subject nuances while the class noun "cat" remains generic.

**Multi-Concept Personalization.** While DreamBooth is effective, fine-tuning the entire model is computationally expensive and storage-heavy. Custom Diffusion [KZZ+23] improves efficiency by updating only the key ($W_K$) and value ($W_V$) projection matrices in the cross-attention layers to demonstrate that these specific weights are sufficient to encode new concepts.

However, accurately synthesizing "multiple" personalized concepts (e.g., "a $[V1]$ cat and a $[V2]$ dog") within a single image remains difficult due to the scarcity of high-quality, text and image aligned training data for complex compositions during the pretraining stage of Stable Diffusion . To addressing this, Gen4Gen [YCH+24] proposes a data-centric solution shown in Figure 8.19. Instead of modifying the model architecture, it introduces a generative data pipeline that leverages foundation models to construct "MyCanvas"—a benchmark dataset featuring complex multi-object scenes paired with detailed, dense captions. By fine-tuning on this synthesized data, where object identities and spatial relationships are explicitly curated, models can significantly improve their ability to disentangle and preserve multiple personalized subjects simultaneously (e.g., a specific cat and dog interacting in a complex scene) without suffering from identity blending or omission.

More recently, the field has shifted toward Tuning-Free architectures like PhotoMaker [LCW+24] and InstantID [WBW+24]. Instead of iterative optimization, these methods employ a "stacking" strategy: an external ID-Encoder (often based on face recognition backbones) extracts a high-fidelity identity embedding from the reference image, which is then injected directly into the diffusion process via decoupled cross-attention or prompt stacking. This enables zero-shot personalization of arbitrary subjects in a single forward pass without any gradient updates.

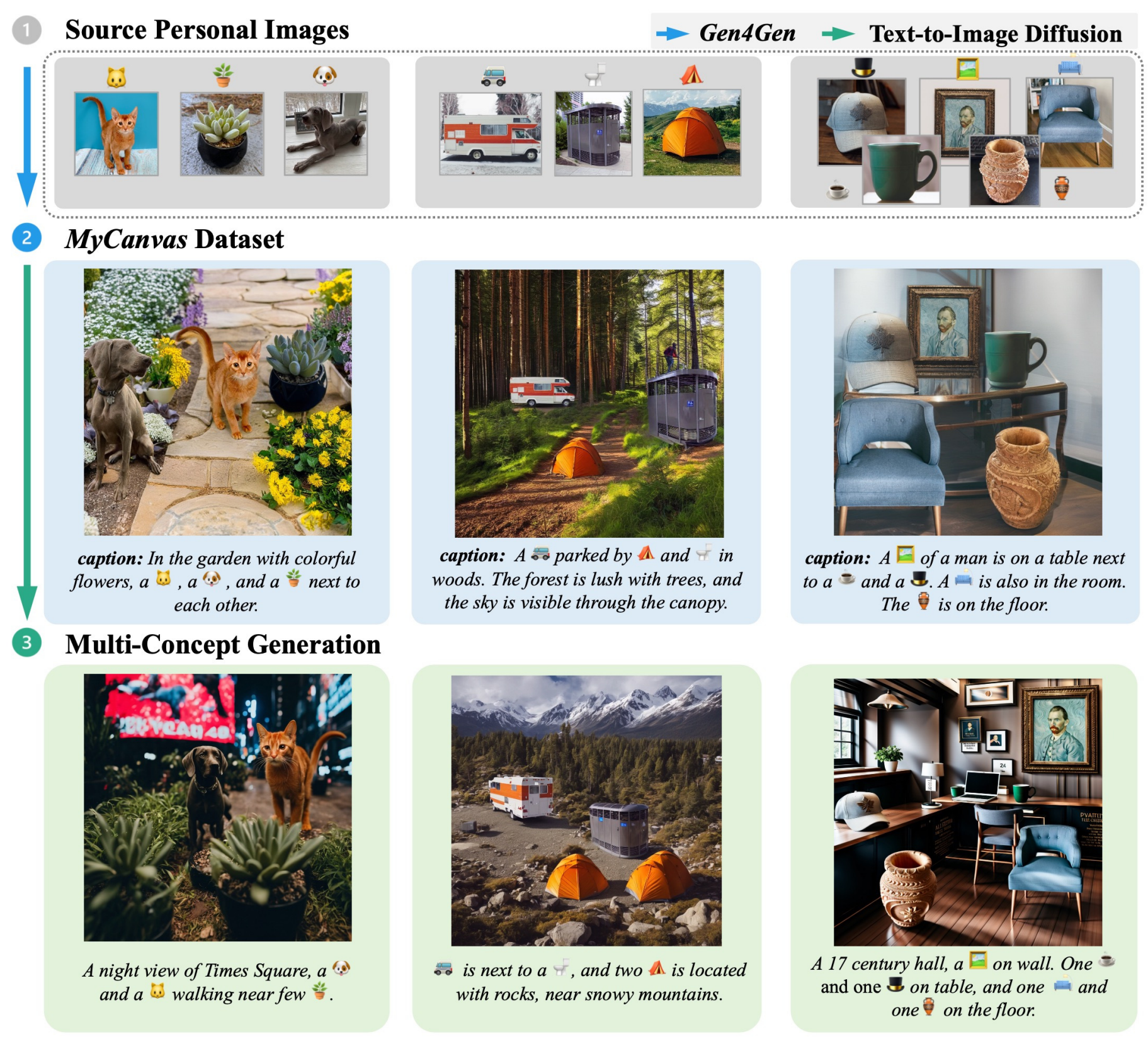

Figure 8.19: **The Gen4Gen pipeline for multi-concept personalization.** To overcome the limitations of existing datasets in multi-subject scenarios, Gen4Gen [YCH+24] automates the creation of a high-quality benchmark (MyCanvas). Training on this data significantly boosts the model's ability to generate complex scenes with multiple specific subjects (e.g., "a [$V1$] cat and a [$V2$] dog").

### 8.7.4 Extension to Video Generation

The principles of conditioned 2D generation extend naturally to video generation by treating video as a sequence of temporally correlated frames $\boldsymbol{x} \in \mathbb{R}^{T \times H \times W \times 3}$. However, directly modeling this high-dimensional space is computationally prohibitive. Consequently, modern state-of-the-art approaches typically operate within the compressed latent space defined by the VAE (Sec. 8.7.2), where the video is represented as a tensor $\boldsymbol{z} \in \mathbb{R}^{T \times h \times w \times c}$.

**Spatiotemporal Inflation.** To leverage the robust visual priors established by 2D image generators, architectures such as VideoLDM [BRL+23] and AnimateDiff [GYR+23] introduce the concept of "inflating" 2D layers into pseudo-3D layers. This is achieved by reshaping the input tensor $\boldsymbol{z}$ to alternate between spatial and temporal processing:

- **Spatial Layers (Frozen):** The 2D backbone (ResNet and Spatial Attention) processes frames independently by reshaping the input to $(B \cdot T, h, w, c)$. This preserves the pre-trained texture and object generation capabilities.

- **Temporal Layers (Trainable):** To model motion, the tensor is reshaped to isolate the temporal axis for each spatial location. We treat the spatial grid points $(B \cdot h \cdot w)$ as the batch dimension and time $(T)$ as the sequence length. A **Temporal Self-Attention** layer is then applied to exchange information across frames.

Mathematically, let $\boldsymbol{u} \in \mathbb{R}^{c \times T}$ denote the feature vector at a specific spatial pixel location across time (analogous to the spatial features $\varphi_i(\boldsymbol{z}_t)$ in Sec. 8.7.2). We define the temporal Query ($\boldsymbol{Q}$), Key ($\boldsymbol{K}$), and Value ($\boldsymbol{V}$) matrices using learnable projection weights $\boldsymbol{U}^{(temp)}$:

$$\boldsymbol{Q} = \boldsymbol{U}_Q^{(temp)}\boldsymbol{u}, \quad \boldsymbol{K} = \boldsymbol{U}_K^{(temp)}\boldsymbol{u}, \quad \boldsymbol{V} = \boldsymbol{U}_V^{(temp)}\boldsymbol{u}, \tag{8.7.15}$$

where $\boldsymbol{U}_Q^{(temp)}, \boldsymbol{U}_K^{(temp)}, \boldsymbol{U}_V^{(temp)} \in \mathbb{R}^{c \times c}$. Following the column-wise formulation established in Sec. 8.7.2, the temporal self-attention is computed as:

$$\mathrm{TempAttn}(\boldsymbol{Q}, \boldsymbol{K}, \boldsymbol{V}) \doteq \boldsymbol{V} \cdot \mathrm{softmax}\left(\frac{\boldsymbol{K}^\top \boldsymbol{Q}}{\sqrt{c}}\right) + \boldsymbol{u}. \tag{8.7.16}$$

Here, the term $\boldsymbol{K}^\top \boldsymbol{Q} \in \mathbb{R}^{T \times T}$ represents the temporal attention map, relating every frame to every other frame. Crucially, the output projection layer of this attention block is typically zero-initialized. This ensures that at the start of training, the contribution of the temporal layers vanishes, allowing the model to behave exactly like the pre-trained 2D generator and gradually learn temporal dynamics.

**Interactive World Models.** Recent advances have moved beyond passive text-to-video synthesis toward interactive world simulation, effectively treating the video generator as a predictive model of physical environments. Formally, the goal is to model the latent transition probability conditioned on an agent's action $\boldsymbol{a}_t$:

$$p(\boldsymbol{z}_{t+1} \mid \boldsymbol{z}_{1:t}, \boldsymbol{a}_t), \tag{8.7.17}$$

where $\boldsymbol{a}_t$ represents the control signal (e.g., keyboard input or motor command) at time $t$. In frameworks like Genie [BDE+24], the action $\boldsymbol{a}_t$ is discretized into a token and injected into the network using the same conditioning mechanisms used for text. For instance, $\boldsymbol{a}_t$ can be mapped to an embedding vector and added to the timestep embedding or utilized as a key/value pair in Cross-Attention layers. This allows the generative model to function as a data-driven physics engine, predicting the visual consequences of actions (e.g., a character jumping) by minimizing the reconstruction or flow matching objectives. Large-scale implementations like Sora [Ope24] further suggest that scaling this autoregressive or diffusion-based prediction leads to emergent 3D consistency and object permanence.

## 8.8 Image and Text Conditioned 3D Object Generation

So far, in previous sections, we have mainly shown how to learn representations of the distribution of *two-dimensional* (2D) images, as well as many image-based applications or tasks such representations facilitate: such as image based object recognition, image segmentation or completion, and image generation.[11]

However, our world is three-dimensional (3D).[12] Although our eyes perceive only 2D projections of the world – stereo pairs through our two eyes, we are able to develop a visual memory that represents a full 3D world model for our living environment and objects within.

This section demonstrates that, using the methods introduced in this book, it is possible to learn the distribution of 3D shapes of natural objects and then generate new 3D object shapes from this distribution, conditioned on either an input 2D image or a text description of the object. In particular, the work featured here is based on the Michelangelo project [ZLC+23b], which very much generalizes the techniques that we have introduced for 2D image representation and generation to 3D shape data[13].

We first give a brief introduction to commonly used data forms that represent 3D shapes and corresponding datasets. Then, following the open-sourced project named **Michelangelo** [ZLC+23b]:

`https://github.com/NeuralCarver/Michelangelo`,

we present a computational framework, objectives, and associated network architectures that attempt to solve the task of image and text conditioned 3D object generation. Finally, we present experimental results and analysis to verify the proposed solution, as well as discussion for further improvements.

### 8.8.1 3D Shape Representations, Datasets, and Tokenization

**3D Shape Representations** Unlike 3D images, 3D shapes and data have not been formally introduced before in this book. Therefore, this subsection

[11]Based on techniques introduced earlier: the vision transformer (ViT) [VSP+17a], the language-image contrastive learning (CLIP) [RKH+21c], and latent diffusion models (VAE-SD) [RBL+22].

[12]or four-dimensional (4D) if we consider time as an extra dimension as the world is dynamical.

[13]Generally speaking, polygon meshes and video data can all be referred to as "3D data"; however, in this section, the term 3D data refers to polygon meshes or their equivalent representations, explicit or implicit.

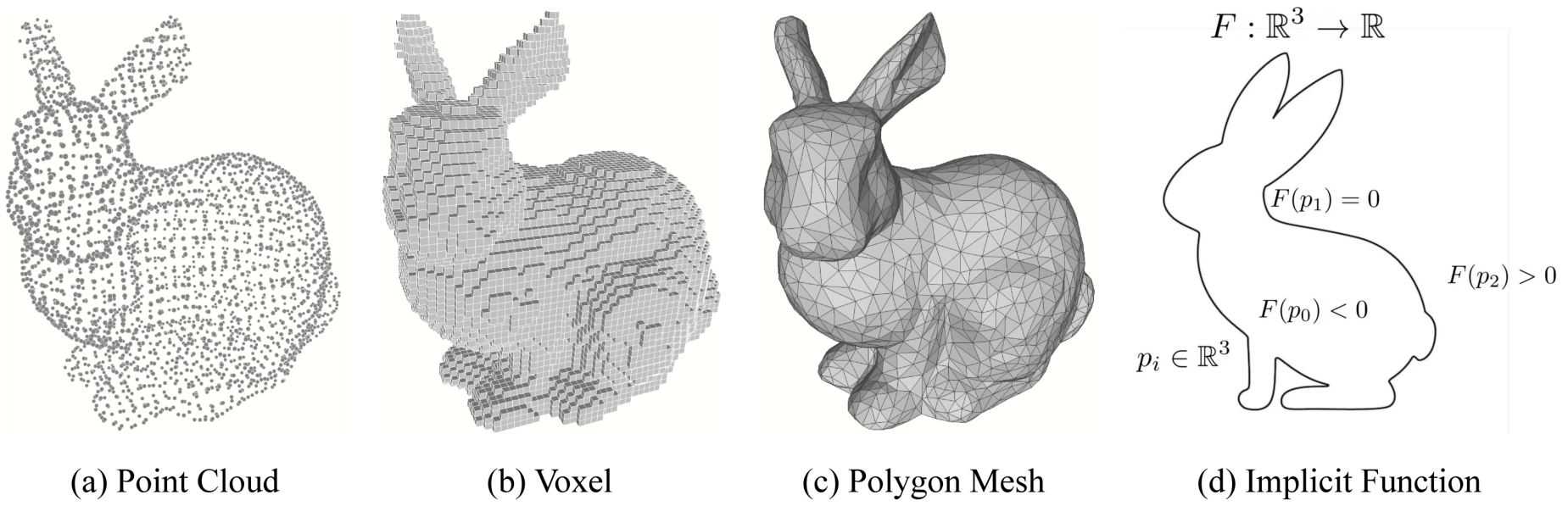

Figure 8.20: **Illustrations of different 3D representations for Stanford Bunny.** (a) the point cloud; (b) the voxel; (c) the polygon mesh; (d) the implicit function.

provides a brief introduction to 3D data. Readers wishing to learn more details are advised to consult the sources via the cited references.

3D data to which we refer here mainly mean continuous real physical 3D shapes that are typically digitized into discrete, computable forms in a certain representation space. Finding good representations of 3D shapes stands as one of the central problems in the fields of computer vision and computer graphics. The four most commonly used, and largely mathematically equivalent, types of 3D shape representations are: point clouds, voxels, polygon meshes, and implicit functions,[14] as shown in Figure 8.20 [15].

**Point clouds** precisely characterize the 3D shape of an object through discretely sampled points on the object's surface. They directly record 3D spatial coordinates $(x, y, z)$ of points on the surface (and maybe some additional photometric information at those points), as illustrated in Figure 8.20(a) Point Cloud. A point cloud is defined as a set of discrete coordinates in 3D Euclidean space:

$$P = \{\boldsymbol{p}_i \in \mathbb{R}^d \mid i = 1, 2, \dots, N\} \tag{8.8.1}$$

where $d \geq 3$ denotes the feature dimensionality, potentially including (concatenations of) coordinates $(x_i, y_i, z_i)$, colors $(r_i, g_i, b_i)$, and surface normals $(n_{xi}, n_{yi}, n_{zi})$.

**Voxels** represent volumetric units in 3D space, constituting a natural extension of the concept of 2D pixels to three dimension. Such a representation is obtained by discretizing a 3D space into a structured grid. Typical voxel models partition 3D space into uniform cubic cells, where each voxel is indexed spatially and stores region-specific attributes such as occupancy, density, material, or color, as shown in Fig. 8.20(b) Voxel.

**Polygon meshes** serve as arguably the most commonly used representation for 3D shapes in computer graphics and physical simulation domains. It

[14] such as the distance-sign function (DSF).

[15] Images in the first three columns came from [SPX+22], and the last column image came from https://medium.com/@sunil7545/implicit-vs-parametric-3d-shape-representation-9d4c01c8c60c

approximates continuous surfaces through discrete vertices, edges, and polygonal faces, with triangle meshes being the predominant variant, as depicted in Fig. 8.20(c) Polygon Mesh. Geometrically, polygon meshes consist of vertex coordinates in $\mathbb{R}^3$, complemented by their topological connectivity. Formally, a mesh $M$ is a tuple $(V, F)$, where vertex set $V = \{\boldsymbol{v}_i \in \mathbb{R}^3\}_{i=1}^n$ denotes the vertex coordinates and face set $F = \{f_j\}_{j=1}^m$ defines topology.

**Implicit functions** represent surfaces via level sets of a scalar function. This approach provides a unified framework for the mathematical description of complex surfaces (Fig. 8.20(d) Implicit Function). The standard implicit formulation is:

$$F : \mathbb{R}^3 \to \mathbb{R}, \quad F(x, y, z) = d, \tag{8.8.2}$$

where $(x, y, z)$ denote the coordinates of a query point, and $d$ denotes the distance between the query point and a surface of interest.[16] Hence the surface of interest is implicitly defined as:

$$\mathcal{S} = \{(x, y, z) \in \mathbb{R}^3 \mid F(x, y, z) = 0.\} \tag{8.8.3}$$

The conversion from an implicit function representation of 3D shape to a polygon mesh can generally be achieved through iso-surface extraction methods such as marching cubes [LC98].

Mathematically speaking, all four shape representations are largely equivalent in terms of their representation capabilities. Nevertheless, each data form may come with different computational advantages or limitations. Hence, some representation is often preferred for different tasks. In this section, we will adopt the traditional point clouds representation and show how to learn the distribution of 3D shapes based on this form of data. In the next section, we will show how to use a (generalized) voxel representation for learning the distribution of 3D shapes and poses.

**3D Datasets.** There have been many publicly available datasets for each of the above representations mentioned above. They come with varying quantities and qualities. In this section, we will use the ShapeNet [CFG+15] to study a joint shape-image-text representation and its generation performance. ShapeNet provides approximately 55,000 manufactured meshes across 55 categories. Each mesh has a category tag and corresponding texts, such as fine-grained categories or brief descriptions given by the creator. To prepare the triplet data (3D shape, image, text), the provided texts are augmented in two ways. First, the shape tag and corresponding description are combined in the format "a 3D model of (shape tag), in the style of (description)" or "a 3D model of (shape tag), (description)." Then, inspired by ULIP [XGX+23], multiple templates containing 65 predefined phrases are leveraged to provide more text information during training. As for the image data, each mesh is rendered under four camera poses, augmenting and improving the rendering diversity via the depth-conditioned ControlNet [ZRA23a].

[16] Notice that we have used the same concept to describe the low-dimensional support of a distribution in a high-dimensional space, in Section 7.1.2.

**3D Shape Data Tokenization.** As discussed in above, 3D data manifests in various forms, including point clouds, voxels, polygon meshes, and implicit neural representations. Fundamentally, point clouds, voxels, and meshes serve as discrete approximations of continuous 3D manifolds. In computer graphics and engineering, polygon meshes are particularly ubiquitous because they explicitly preserve surface topology, offering an efficient balance between visual fidelity and geometric structure.

However, directly applying modern deep learning architectures—specifically Transformers—to these raw representations presents significant challenges. Unlike 2D images, which possess a regular grid structure easily split into patches, 3D data is often sparse, unstructured, and irregular. To leverage the power of sequence modeling and self-attention mechanisms in 3D, we must bridge the gap between raw geometric data and the discrete, ordered sequences required by Transformers. This process is known as *3D Tokenization*.

- **Point Cloud Tokenization.** Raw point clouds are typically tokenized into local groups to capture geometric context. This process usually begins with Farthest Point Sampling (FPS) [ELP+97], an iterative algorithm that selects a subset of points $\{x_1, \ldots, x_m\}$ from the original set $\mathcal{P}$, such that each selected point maximizes the distance to the set of previously selected points: $x_i = \arg\max_{x \in \mathcal{P}} \min_{j=1}^{i-1} \|x - x_j\|$. Following FPS, points are grouped via k-Nearest Neighbors (k-NN) to form local clusters. Each cluster serves as a token to handle the permutation invariance of points within a cluster, a lightweight network such as PointNet [QSM+17] is applied. Mathematically, for a cluster of points $\{p_1, \ldots, p_k\}$, PointNet approximates a symmetric function $f$ by applying a shared Multi-Layer Perceptron (MLP) $h$ followed by a symmetric aggregation operator (typically max-pooling):
$$\boldsymbol{t} = \gamma \left( \max_{i=1}^{k} \{h(p_i)\} \right) \tag{8.8.4}$$
where $\boldsymbol{t}$ is the resulting feature token and $\gamma$ is a post-processing MLP. This converts an unstructured cloud into a sequence of feature vectors.

- **Voxel Tokenization.** Analogous to the patch embedding strategy in Vision Transformers (ViT) for 2D images (see Chapter 8.2.3), voxel grids can be partitioned into non-overlapping 3D sub-grids. Given a voxel grid $V \in \mathbb{R}^{D \times H \times W \times C}$, we extract patches of size $P \times P \times P$. Each patch is flattened into a vector $v_p \in \mathbb{R}^{P^3 \cdot C}$ and mapped via a learnable linear projection to the token dimension $d_{model}$. The result is a sequence of tokens $Z \in \mathbb{R}^{L \times d_{model}}$ where $L = \frac{DHW}{P^3}$. While straightforward, this method inherits the cubic complexity of voxel resolution ($O(N^3)$), often creating memory bottlenecks for high-resolution shapes.

- **Mesh Tokenization.** Tokenizing meshes is challenging due to their graph-like structure involving both vertices and faces. Unlike images, meshes lack a canonical order. To enable sequence modeling, common approaches involve explicit serialization, where raw geometric and topological elements are directly quantized and ordered into sequences.

Taking PolyGen [NGE+20] as an example, the mesh is represented as a set of vertices $V = \{v_1, \ldots, v_n\}$ (where each $v_i \in \mathbb{R}^3$) and faces $F = \{f_1, \ldots, f_m\}$ (where each $f_j$ is a tuple of vertex indices). The model learns the joint distribution $P(V, F)$ by factoring it into two sequential stages: a Vertex Model that predicts quantized coordinates, and a Face Model that predicts vertex indices

$$P(V, F) = P(V) \cdot P(F \mid V) = \prod_{i=1}^{n} P(v_i \mid v_{<i}) \cdot \prod_{j=1}^{m} P(f_j \mid f_{<j}, V). \quad (8.8.5)$$

Crucially, this factorization dictates two distinct token formats corresponding to each stage. For the Vertex Model ($P(V)$), the tokens are quantized coordinate values, continuous coordinates are discretized (e.g., into 8-bit integers), creating a sequence of tokens from a fixed vocabulary of size 256. Conversely, for the Face Model ($P(F \mid V)$), the tokens are vertex indices (pointers). While effective, operating on these raw primitives often results in extremely long sequence lengths, limiting scalability.

While these explicit tokenization strategies effectively discretize continuous manifolds, they operate directly in the high-dimensional observation space, often resulting in excessively long sequence lengths and computational redundancy. To address these scalability constraints and achieve higher compression ratios, the field has gravitated towards learning-based discrete representations. Instead of tokenizing raw geometric primitives directly, these following methods first project the 3D data into a compact, low-dimensional latent space where semantic information is aggregated, effectively decoupling geometric reconstruction from structural modeling.

**Standard Latent Vector Quantization.** This approach learns a discrete codebook to represent the shape via Vector Quantization (VQ). A representative method is AutoSDF [MCS+22] In this framework, a high-dimensional input (e.g., a voxel grid $X$ or implicit function field $F_\theta$) is compressed by an encoder $E$ into lower-dimensional spatial grid of latent vectors $Z \in \mathbb{R}^{h \times w \times d \times c}$. Each vector $z_i \in Z$ is then replaced by its nearest neighbor from a learnable codebook $\mathcal{C} = \{c_k\}_{k=1}^{K}$:

$$q(z_i) = \arg\min_{c_k \in \mathcal{C}} |z_i - c_k|_2^2. \quad (8.8.6)$$

This reduces the shape to a sequence of discrete codebook indices that preserve spatial structure while drastically reducing dimensionality. These discrete indices serve as the tokens for subsequent Transformer processing. However, this regular grid-based approach typically treats occupied and empty space uniformly, leading to computational redundancy.

**From Structural Optimization to Semantic Alignment.** To address the redundancy of regular grids, approaches such as 3DILG [ZNW22a] proposes

Irregular Latent Grids. Instead of a dense voxel grid where tokens are wastefully allocated to empty space, 3DILG defines the shape as a set of latent codes floating at arbitrary, adaptive positions in continuous space, typically concentrated on the shape's surface. Specifically, given an input shape represented as a point cloud, the method first encodes it into a set of feature-position pairs $\mathcal{P} = \{(x_i, f_i)\}_{i=1}^{M}$, where $x_i \in \mathbb{R}^3$ are key points selected via Farthest Point Sampling (FPS) and $f_i \in \mathbb{R}^C$ are their associated continuous feature vectors extracted by a PointNet-based [QSM+17] encoder. The conversion of this unstructured set into a token sequence involves three key steps:

- **Coordinate Quantization.** First, for a set of $M$ key points selected via Farthest Point Sampling (see Point Cloud Tokenization), the continuous spatial coordinates $x_i \in \mathbb{R}^3$ are discretized into 8-bit integers:

$$x_i \to (x_{i,1}, x_{i,2}, x_{i,3}), \quad \text{where } x_{i,k} \in 0, 1, \ldots, 255. \tag{8.8.7}$$

  This step effectively converts continuous positions into discrete, tokenizable units.

- **Feature Quantization.** Concurrently, the local feature vector associated with each point is compressed into a discrete codebook index $z_i$ via Vector Quantization (VQ):

$$z_i \in \{0, 1, \ldots, D-1\}, \tag{8.8.8}$$

  where $D$ represents the size of the codebook.

- **Sequence Construction.** To facilitate autoregressive generation with Transformers, the unstructured set of tuples must be serialized. 3DILG employs lexicographical sorting based on the quantized coordinates. The final shape is represented as a structured sequence $\mathcal{S}$ that explicitly couples positional coordinate and local geometric feature information:

$$\mathcal{S} = \big\{(x_{0,1}, x_{0,2}, x_{0,3}, z_0), \ldots, (x_{M-1,1}, x_{M-1,2}, x_{M-1,3}, z_{M-1})\big\}. \tag{8.8.9}$$

This hybrid formulation drastically reduces sequence length compared to dense grids while preserving high-fidelity surface details.

While 3DILG optimizes the spatial structure of tokens, it largely remains a geometric compression method. In the following, we provide a detailed exposition of Michelangelo, a work that advances this paradigm by addressing the semantic gap. It encodes 3D shapes into a sequence of latent tokens that are not only geometrically compact but also explicitly aligned with the feature space of 2D vision-language models (e.g., CLIP) [RKH+21c]), thereby facilitating high-quality conditional generation.

### 8.8.2 Task

Our goal here is to learn a conditional 3D shape generative model capable of sampling 3D shapes from the learned distribution, subject to a given condition, such as a given image or some texts, as illustrated in Figure 8.21.

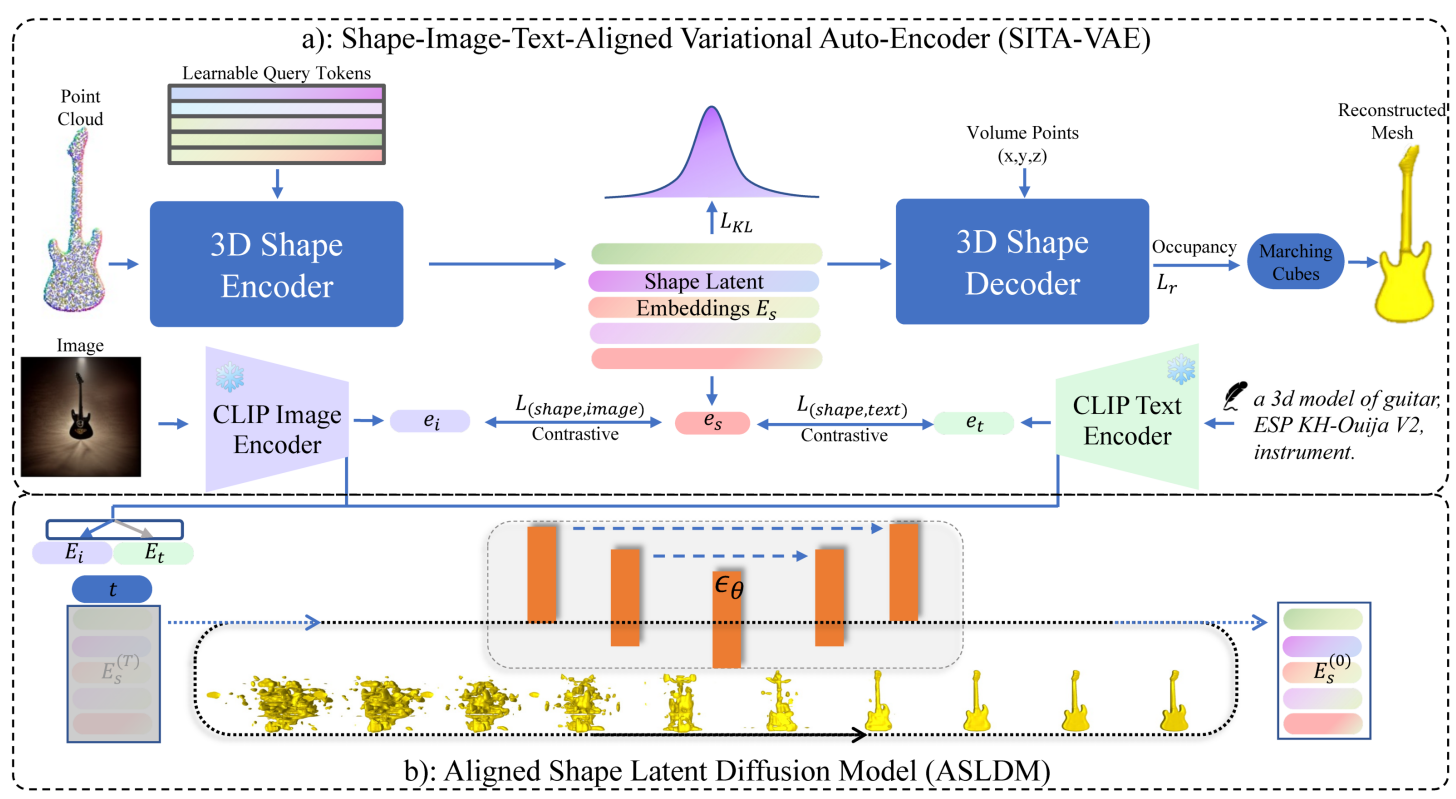

Figure 8.21: **Alignment-before-generation pipeline**. The approach contains two models: the Shape-Image-Text-Aligned Variational Auto-Encoder (SITA-VAE) and the Aligned Shape Latent Diffusion Model (ASLDM). The SITA-VAE consists of four modules: an image encoder, a text encoder, a 3D shape encoder, and a 3D shape decoder. Encoders encode input pairs into an aligned space, and the 3D shape decoder reconstructs 3D shapes given embeddings from the aligned space. The ASLDM maps the image or text condition to the aligned shape latent space for sampling a high-quality 3D shape embedding, which is later reconstructed to high-fidelity 3D shapes by the 3D shape decoder.

Following the Latent Diffusion Model [RBL+22], to fully leverage the capabilities of the generative model, a VAE is first trained to compress 3D data samples into a latent space. Generally, a good latent representation should possess two key characteristics: it must be both highly compressible yet faithfully reconstructible for 3D data samples, and conducive to generative model training. However, compared to 2D image data, 3D data remains scarce [17]. Consequently, learning an effective latent representation for 3D data is highly challenging. Figure 8.21 shows the overall diagram of the framework proposed by the Michelangelo project. We will explain implementation details for each component below.

### 8.8.3 Architecture and Objective

**Shape-Image-Text Aligned Variational Auto-Encoder.** Inspired by CLIP [RKH+21c], the approach adopted by the Michelangelo project injects semantic information into the latent space via a contrastive loss to alleviate problems observed when training VAEs exclusively from 3D data, which forms the Shape-Image-Text-Aligned Variational Auto-Encoder (SITA-VAE).

[17] The situation persists even after the releases of Objaverse-XL [DLW+23].

The SITA-VAE contains four components: a pre-trained and fixed CLIP image encoder $\mathcal{E}_\mathrm{i}$ and CLIP text encoder $\mathcal{E}_\mathrm{t}$, a trainable 3D shape encoder $\mathcal{E}_\mathrm{s}$ and neural field decoder $\mathcal{D}_\mathrm{s}$. The CLIP image encoder and text encoder take 2D images $\boldsymbol{I} \in \mathbb{R}^{H\times W\times 3}$ and tokenized texts $\boldsymbol{T} \in \mathbb{R}^{d_\mathrm{t}\times L_\mathrm{t}}$ as input, and generate image tokens $\boldsymbol{E}_\mathrm{i} \in \mathbb{R}^{d\times(1+L_\mathrm{i})}$ and text tokens $\boldsymbol{E}_\mathrm{t} \in \mathbb{R}^{d\times L_\mathrm{t}}$, where $(1 + L_\mathrm{i})$ and $L_\mathrm{t}$ are the sequence length of image tokens $\boldsymbol{E}_\mathrm{i}$ and text tokens $\boldsymbol{E}_\mathrm{t}$. The approach takes advantage of the pre-trained image encoder and text encoder from CLIP. These two encoders are trained on large-scale image-text pairs and are robust enough to capture a well-aligned vision-language space, which enriches the semantics of the 3D shape representation after multi-modal alignment via contrastive learning.

**3D shape encoder** aims to extract powerful feature representations to characterize each 3D shape effectively. To achieve this, point clouds $\boldsymbol{P} \in \mathbb{R}^{(3+C)\times N}$ are first sampled from the surface of 3D shapes, where $N$ represents the number of points, and $C$ denotes additional point features such as normal or color.

To better capture high-frequency geometric details, we utilize Fourier positional encoding to map the low-dimensional spatial coordinates $\boldsymbol{v} \in \mathbb{R}^3$ of each point into a higher-dimensional frequency space. This encoding $\gamma(\boldsymbol{v})$ is defined as a sequence of sinusoidal functions with exponentially increasing frequencies:

$$\gamma(\boldsymbol{v}) = [\sin(2^0\pi\boldsymbol{v}), \cos(2^0\pi\boldsymbol{v}), \ldots, \sin(2^{L-1}\pi\boldsymbol{v}), \cos(2^{L-1}\pi\boldsymbol{v})],$$

where $L$ is the number of frequency bands. To construct the final input representation, we explicitly decompose the input point cloud $\boldsymbol{P} \in \mathbb{R}^{(3+C)\times N}$ into spatial coordinates $\boldsymbol{v}$ and auxiliary features $\boldsymbol{f} \in \mathbb{R}^{C\times N}$. The high-dimensional positional embeddings $\gamma(\boldsymbol{v})$ are then concatenated with the preserved features $\boldsymbol{f}$ and projected via a learnable linear layer. Formally, the 3D shape encoder input $\boldsymbol{X} \in \mathbb{R}^{d\times N}$ is obtained as:

$$\boldsymbol{X} = \mathrm{Linear}([\gamma(\boldsymbol{v}); \boldsymbol{f}])$$

where $[\cdot;\cdot]$ denotes the concatenation operation along the feature dimension.

To process this dense feature sequence $\boldsymbol{X}$ efficiently, the encoder adopts a Perceiver-based architecture [JGB+21]. Unlike standard Transformers that scale quadratically with input length, the Perceiver employs a latent bottleneck mechanism to decouple the processing complexity from the input size. Instead of tokenizing the input points directly for self-attention, the model utilizes a fixed set of small, learnable latent queries $\boldsymbol{Q} \in \mathbb{R}^{(1+L_\mathrm{s})\times d}$, where $1 + L_\mathrm{s}$ is the number of query tokens. Mechanistically, a Cross-Attention layer projects the high-dimensional input space into this compact latent space. Formally, following the formulation in Chapter 8.7.2, let $\boldsymbol{U}_Q^{(temp)}, \boldsymbol{U}_K^{(temp)}, \boldsymbol{U}_V^{(temp)}$ be the learnable projection matrices. The initial latent representation $\boldsymbol{Z}_0$ is computed as:

$$\boldsymbol{Z}_0 = \text{Cross-Attention}(\boldsymbol{Q}, \boldsymbol{X}) = (\boldsymbol{U}_V^{(temp)}\boldsymbol{X})\cdot\mathrm{softmax}\left(\frac{(\boldsymbol{U}_K^{(temp)}\boldsymbol{X})^\top(\boldsymbol{U}_Q^{(temp)}\boldsymbol{Q})}{\sqrt{d_k}}\right).$$

Here, the input point cloud $\boldsymbol{X}$ serves as the Keys and Values, while the learnable queries $\boldsymbol{Q}$ act as the Queries. This operation compresses the variable-length input $N$ into a fixed-length latent sequence $\boldsymbol{Z}_0 \in \mathbb{R}^{(1+L_\mathrm{s})\times d}$.

In the context of Michelangelo, these queries are structured to capture hierarchical semantics: they consist of one global head token $\boldsymbol{Q}_g \in \mathbb{R}^{1\times d}$ with high-level semantics and $L_\mathrm{s}$ local tokens $\boldsymbol{Q}_l \in \mathbb{R}^{L_\mathrm{s}\times d}$ containing low-level geometric structure information. Then, several self- attention blocks are used to iteratively improve the feature representation and obtain the final shape embeddings, $\boldsymbol{E}_\mathrm{s} \in \mathbb{R}^{(1+L_\mathrm{s})\times d}$.

**Alignment among 3D shapes, images, and texts** plays a crucial role in SITA-VAE and the conditional generative models. Since 3D data is an order of magnitude smaller than images and text data, to learn a better-aligned representation among 3D shapes, images, and texts, the 3D shape encoder is constrained to be close to a pre-aligned vision-language space, which is pre-trained on large-scale image-text pairs with rich image and text representations by leveraging the contrastive learning strategy. Consider an input triplet of 3D shapes $\boldsymbol{X}$, images $\boldsymbol{I}$ and tokenized texts $\boldsymbol{T}$. The triplet encoders generate the corresponding shape embedding $\boldsymbol{e}_\mathrm{s}$, image embedding $\boldsymbol{e}_\mathrm{i}$ and text-embedding $\boldsymbol{e}_\mathrm{t}$ by projecting the extracted shape tokens $\boldsymbol{E}_\mathrm{s}$, image tokens $\boldsymbol{E}_\mathrm{i}$ and text tokens $\boldsymbol{E}_\mathrm{t}$ as three vectors with the same dimension, which is expressed as: $\boldsymbol{e}_\mathrm{s} = \mathcal{F}_\mathrm{s}(\boldsymbol{E}_\mathrm{s}), \boldsymbol{e}_\mathrm{i} = \mathcal{F}_\mathrm{i}(\boldsymbol{E}_\mathrm{i})$, and $\boldsymbol{e}_\mathrm{t} = \mathcal{F}_\mathrm{t}(\boldsymbol{E}_\mathrm{t})$, where $\mathcal{F}_\mathrm{s}$ is a learnable shape embedding projector, image embedding projector $\mathcal{F}_\mathrm{i}$ and text embedding projector $\mathcal{F}_\mathrm{t}$ are pre-trained and frozen during training and inference. The contrastive loss is:

$$\mathcal{L}_{(\mathrm{s},\mathrm{i})} = -\frac{1}{2}\sum_{(j,k)}\left(\log\frac{\exp(\boldsymbol{e}_\mathrm{s}^j\boldsymbol{e}_\mathrm{i}^k)}{\sum_l \exp(\boldsymbol{e}_\mathrm{s}^j\boldsymbol{e}_\mathrm{i}^l)} + \log\frac{\exp(\boldsymbol{e}_\mathrm{s}^j\boldsymbol{e}_\mathrm{i}^k)}{\sum_l \exp(\boldsymbol{e}_\mathrm{s}^l\boldsymbol{e}_\mathrm{i}^k)}\right), \tag{8.8.10}$$

$$\mathcal{L}_{(\mathrm{s},\mathrm{t})} = -\frac{1}{2}\sum_{(j,k)}\left(\log\frac{\exp(\boldsymbol{e}_\mathrm{s}^j\boldsymbol{e}_\mathrm{t}^k)}{\sum_l \exp(\boldsymbol{e}_\mathrm{s}^j\boldsymbol{e}_\mathrm{t}^l)} + \log\frac{\exp(\boldsymbol{e}_\mathrm{s}^j\boldsymbol{e}_\mathrm{t}^k)}{\sum_l \exp(\boldsymbol{e}_\mathrm{s}^l\boldsymbol{e}_\mathrm{t}^k)}\right), \tag{8.8.11}$$

where $(j,k)$ indicates the positive pair in training batches, and since pre-trained encoders from CLIP are utilized, the model is free from the constraint $\mathcal{L}_{(\mathrm{i},\mathrm{t})}$.

**3D shape decoder**, $\mathcal{D}_\mathrm{s}$, takes the shape embeddings $\boldsymbol{E}_\mathrm{s}$ as inputs to reconstruct the 3D neural field with high quality. The KL divergence loss $\mathcal{L}_{KL}$ is used to facilitate the generative process to maintain the latent space as a continuous distribution. Besides, a projection layer is leveraged to compress the latent from dimension $d$ to lower dimensions $d_0$ for a compact representation. Then, another projection layer is used to transform the sampled latent from dimension $d_0$ back to high dimension $d$ for reconstructing neural fields of 3D shapes. Like the encoder, the decoder model also builds on a transformer with the cross-attention mechanism. Given a query 3D point $\boldsymbol{x} \in \mathbb{R}^3$ in the field and its corresponding shape latent embeddings $\boldsymbol{E}_\mathrm{s}$, the decoder computes cross attention iteratively for predicting the occupancy of the query point $\mathcal{O}(\boldsymbol{x})$. The training loss is expressed as:

$$\mathcal{L}_r = \mathbb{E}_{\boldsymbol{x} \in \mathbb{R}^3}[\text{BCE}(\mathcal{D}(\boldsymbol{x} \mid \boldsymbol{E}_{\text{s}}), \mathcal{O}(\boldsymbol{x}))], \tag{8.8.12}$$

where BCE is binary cross-entropy loss, and the total loss for training Shape-Image-Text Aligned Variational Auto-Encoder (SITA) is written as:

$$\mathcal{L}_{SITA} = \lambda_c(\mathcal{L}_{(\text{s,i})} + \mathcal{L}_{(\text{s,t})}) + \mathcal{L}_r + \lambda_{KL}\mathcal{L}_{KL}. \tag{8.8.13}$$

**Aligned Shape Latent Diffusion Model.** After training the SITA-VAE, an alignment space among 3D shapes, images, and texts is obtained, as well as a 3D shape encoder and decoder that compress the 3D shape into low-dimensional shape latent embeddings and reconstruct shape latent embeddings to a neural field with high quality. Building on the success of the Latent Diffusion Model (LDM) [RBL+22] in text-to-image generation, which strikes a balance between computational overhead and generation quality, a shape latent diffusion model is proposed on the aligned space to learn a better probabilistic mapping from 2D images or texts to 3D shape latent embeddings. By leveraging the alignment space and the shape latent diffusion model, it is possible to generate high-quality 3D shapes that better conform to the visual or textual conditional inputs.

The Aligned Shape Latent Diffusion Model (ASLDM) builds on a UNet-like transformer [BNX+23], aiming to fit a distribution of the shape latent embeddings, accompanied by an auto-encoder for encoding data samples into the latent space and reconstructing the data samples given the sampled latent. By learning in the latent space, the latent diffusion model is computationally efficient, and leveraging such a compact representation enables the model to fit the target distribution faster. Specifically, the model $\epsilon_\theta$ focuses on generating shape latent embeddings $\boldsymbol{E}_{\text{s}}$ conditioned on $\boldsymbol{C}$, which is represented by the CLIP image or text encoder. Following LDM [RBL+22], the objective is

$$\mathcal{L} = \mathbb{E}_{\boldsymbol{E}_{\text{s}}, \boldsymbol{\epsilon} \sim \mathcal{N}(\boldsymbol{0}, \boldsymbol{I}), t}\left[\|\boldsymbol{\epsilon} - \epsilon_\theta(\boldsymbol{E}_{\text{s}}^{(t)}, \boldsymbol{C}, t)\|_2^2\right], \tag{8.8.14}$$

where $t$ is uniformly sampled from $\{1, ..., T\}$ and $\boldsymbol{E}_{\text{s}}^{(t)}$ is a noisy version of $\boldsymbol{E}_{\text{s}}^{(0)}$. During inference, after sampling Gaussian noise, the model gradually denoises the signal until reaching $\boldsymbol{E}_{\text{s}}^{(0)}$. Following classifier-free guidance (CFG) [HS21], the conditional latent diffusion model is trained with classifier-free guidance. In the training phase, the condition $\boldsymbol{C}$ randomly converts to an empty set $\emptyset$ with a fixed probability of 10%. Then, the sampling is performed with the linear combination of conditional and unconditional samples:

$$\epsilon_\theta(\boldsymbol{E}_{\text{s}}^{(t)}, \boldsymbol{C}, t) = \epsilon_\theta(\boldsymbol{E}_{\text{s}}^{(t)}, \emptyset, t) + \lambda(\epsilon_\theta(\boldsymbol{E}_{\text{s}}^{(t)}, \boldsymbol{C}, t) - \epsilon_\theta(\boldsymbol{E}_{\text{s}}^{(t)}, \emptyset, t)), \tag{8.8.15}$$

where $\lambda$ is the guidance scale for trading off the sampling fidelity and diversity.

### 8.8.4 Experiments and Analysis

**Metrics.** The Intersection of Union (IoU) is used to reflect the accuracy of reconstructions. Two new metrics are proposed for evaluating 3D shape generation methods. The first is a shape-image score (SI-S), which uses a 3D shape encoder and image encoder to extract corresponding shape and image embeddings and compute the Cosine Similarity of these two modalities. Another is a shape-text score (ST-S), which computes the similarity between the generated 3D shape and the conditional text input in the aligned shape embedding and text embedding space. Both metrics evaluate the similarity between results and their corresponding conditions. Moreover, both the pre-trained ULIP [XGX+23] and SITA are used to compute SI-S and ST-S, in terms of SI-S (ULIP), ST-S (ULIP), SI-S (SITA) and ST-S (SITA), respectively. Besides, the metrics of P-IS and P-FID as introduced in Point-E [NJD+22] are followed, using a pre-trained PointNet++ [QYS+17] to compute the point cloud analogous Inception Score [SGZ+16] and FID [HRU+17a] to evaluate the diversity and quality of the generated 3D shapes.

**Baselines.** We compare the Michelangelo approach to other methods, including OccNet [MON+19], ConvOcc [PNM+20], IF-Net [CAP20], 3DILG [ZNW22b], and Learnable Query Version of 3DS2V [ZTN+23], on reconstruction tasks to validate the ability of the model to recover a neural field given shape embeddings on the ShapeNet dataset [CFG+15]. For the conditional generation stage, two recent powerful 3D generation methods, 3DILG and 3DS2V, are chosen as baselines. Their shape representation modules are first fine-tuned on a mixture dataset of ShapeNet and the 3D Cartoon Monster. Then, the text and image conditional generative models of 3DILG and 3DS2V are retrained with the same protocols.

**Numerical comparison**. The numerical results are reported in Table 8.9 and Table 8.10. Table 8.9 shows that Michelangelo has the best reconstruction performance on 55 overall categories, surpassing the rest. Results of the selected categories further demonstrate that the model can faithfully reconstruct 3D shapes in each of the 55 categories. Table 8.10 reports the numerical results for conditional 3D shape generation. Michelangelo achieves the best performance on all the SI-S and ST-S metrics, indicating that it can map the information from the image or text to its corresponding 3D shape information for generating high-fidelity results. Moreover, the P-FID demonstrates that the model can produce high-quality shape-tokens for generating realistic 3D shapes, and P-IS indicates the diversity of the samples. Specifically, the four left columns show that Michelangelo surpasses the baselines on image-conditioned generation, demonstrating that it can better map visual information to 3D shapes. The four right columns validate the generative quality of text-conditioned generation. Since natural language, compared to 2D images, usually provides limited and abstract information, learning a model to map text information to 3D shapes is challenging. However, benefiting from training on the aligned latent space, Michelangelo significantly improves text-conditioned generation, as shown in the right columns of Table 8.10, which reflects that the model effectively maps natural language information to 3D shapes and generates diverse and high-quality results.

**Visual comparison**. The visual comparisons of the image/text-conditioned 3D shape generations are illustrated in Figure 8.22 and Figure 8.23. Figure 8.22 shows that 3DILG [ZNW22b] pays more attention to the global shape in the auto-regressive generation process, where its results lack depictions of details of 3D shapes. While 3DS2V [ZTN+23] generates more details of 3D shapes, the results have discontinuous or noisy surfaces. Besides, both methods struggle to generate a complete shape when the given conditions map to a complex object, fine machine, or rare monster. Figure 8.23 shows the visual comparison of text-conditional generation. In the upper-half rows, the results given simple and abstract concepts are shown, while in the lower-half rows, the results given detailed texts like descriptions for deterministic parts of the target shape are shown. Similar to the observation above, 3DILG [ZNW22b] generates over-smooth shape surfaces with fewer details, and 3DS2V [ZTN+23] produces fewer details on discontinuous object surfaces. In contrast, Michelangelo produces correct shapes that conform to the given concepts or detailed descriptions with delicate details on smooth surfaces.

| | Overall | Selected | Table | Chair | Airplane | Car | Rifle | Lamp |
|---|---|---|---|---|---|---|---|---|
| OccNet [MON+19] | 0.825 | 0.81 | 0.823 | 0.803 | 0.835 | 0.911 | 0.755 | 0.735 |
| ConvOccNet [PNM+20] | 0.888 | 0.873 | 0.847 | 0.856 | 0.881 | 0.921 | 0.871 | 0.859 |
| IF-Net [CAP20] | 0.934 | 0.924 | 0.901 | 0.927 | 0.937 | 0.952 | 0.914 | 0.914 |
| 3DILG [ZNW22b] | 0.950 | 0.948 | 0.963 | 0.95 | 0.952 | 0.961 | 0.938 | 0.926 |
| 3DS2V(LQ) [ZTN+23] | 0.955 | 0.955 | **0.965** | 0.957 | 0.962 | 0.966 | 0.947 | 0.931 |
| Ours | **0.966** | **0.964** | **0.965** | **0.966** | **0.966** | **0.969** | **0.967** | **0.95** |

Table 8.9: **Numerical results for reconstruction comparison on IoU(↑, a larger value is better)**. The results show that Michelangelo performs best in 55 categories. The results of selected categories further demonstrate that the model can reconstruct each category faithfully.

| | Image-Conditioned | | | | Text-Conditioned | | | |
|---|---|---|---|---|---|---|---|---|
| | SI-S (ULIP)↑ | SI-S (SITA)↑ | P-FID↓ | P-IS↑ | ST-S (ULIP)↑ | ST-S (SITA)↑ | P-FID↓ | P-IS↑ |
| 3DILG | 9.134 | 11.703 | 4.592 | 12.247 | 10.293 | 6.878 | 10.283 | 12.921 |
| 3DS2V | 13.289 | 15.156 | 2.921 | 12.92 | 12.934 | 9.833 | 5.704 | 13.149 |
| Ours | **13.818** | **15.206** | **1.586** | **13.233** | **16.647** | **13.128** | **2.075** | **13.558** |

Table 8.10: **Numerical results for conditional generation comparison**. The results show that Michelangelo achieves the best generative performance. The SI-S and ST-S indicate that the model generates high-fidelity results by effectively mapping the condition information to its related 3D shapes. Moreover, P-FID reflects that the model generates the most realistic 3D shapes, and P-IS indicates that the generated samples are diverse. ↑ means a larger value is better, and ↓ otherwise.

**The effectiveness of training generative model in the aligned space**. A visual comparison is performed to ablate the effectiveness of training the generative model in the aligned space, as illustrated in Figure 8.24. The upper samples are generated from the generative model trained in the aligned space, while the lower samples are generated from the generative model trained without the aligned space. The results demonstrate that the upper samples conform to the given text while the lower samples do not, which indicates that training the generative model in the aligned space leads to high-fidelity samples.

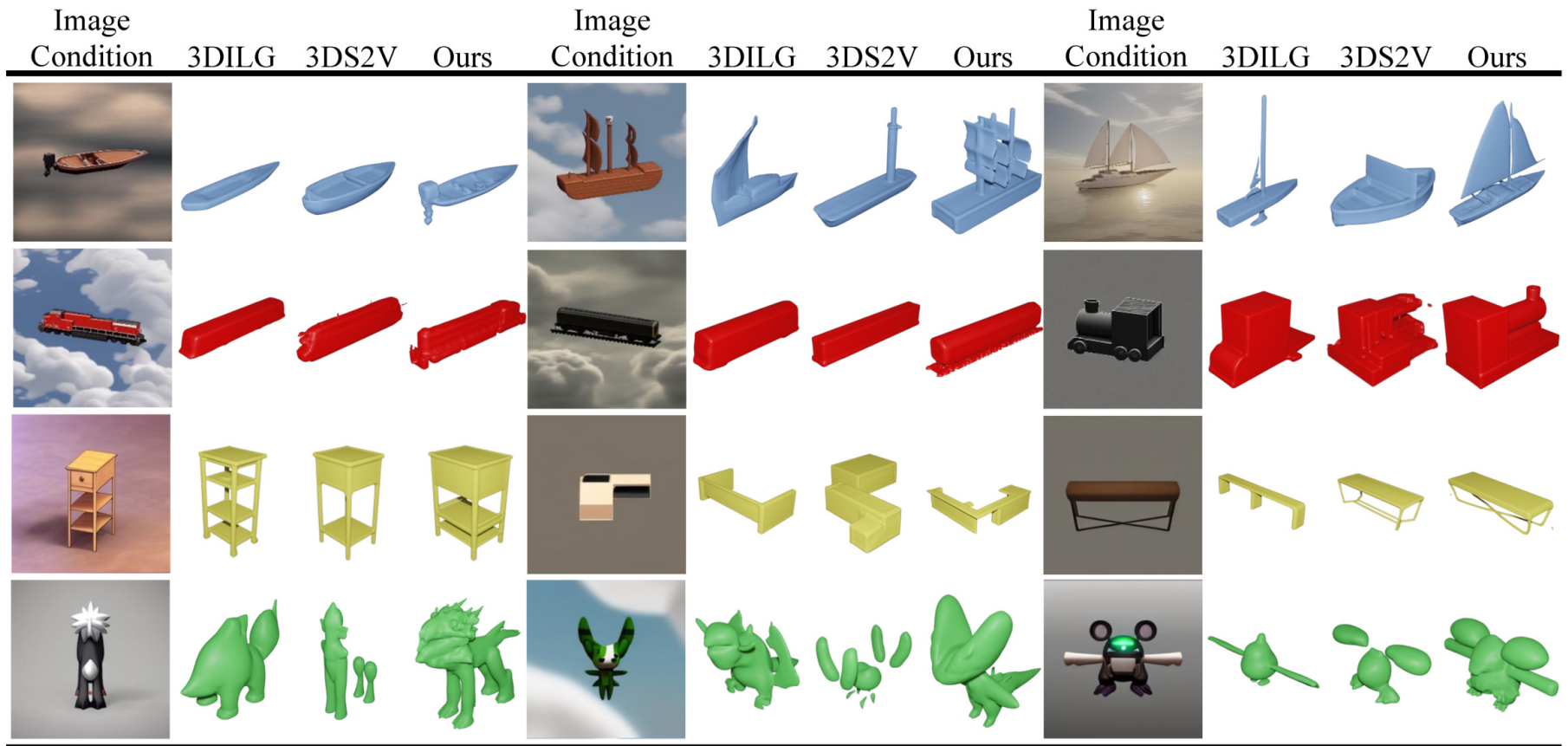

Figure 8.22: **Visual results for image-conditioned generation comparison**. The figure shows that 3DILG [ZNW22b] generates over-smooth surfaces and lacks details of shapes, whereas 3DS2V [ZTN+23] generates few details with noisy and discontinuous surfaces of shapes. In contrast to baselines, Michelangelo produces smooth surfaces and portrays shape details. Please zoom in for more visual details.

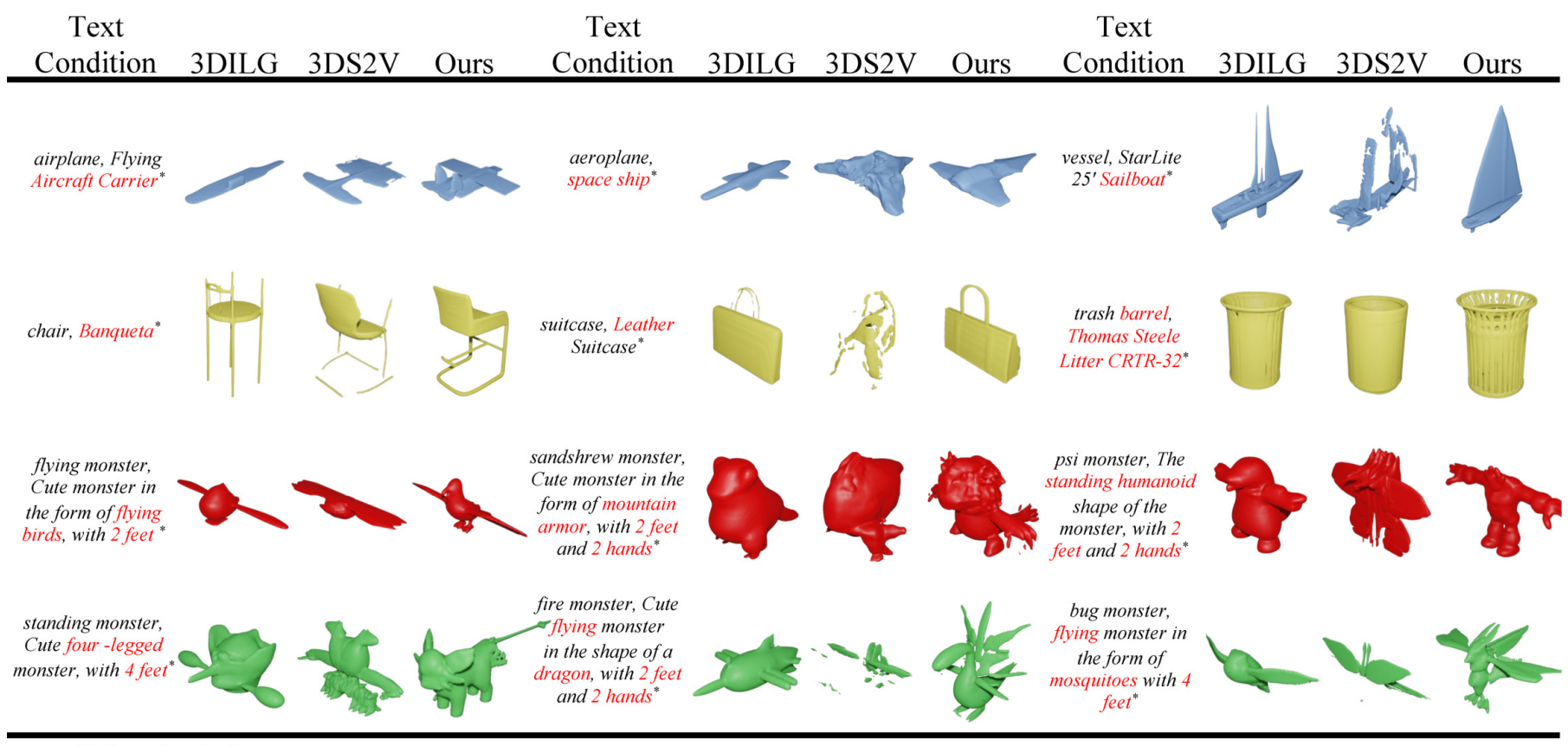

Figure 8.23: **Visual results for text-conditioned generation comparison**. In the first two rows, the models are tested with abstract texts, and the result shows that only Michelangelo generates a 3D shape that conforms to the target text with a smooth surface and fine details. The last two rows show the results given texts containing detailed descriptions, which further demonstrates that Michelangelo can capture both the global conditional information and the local information for generating high-fidelity 3D shapes. Keywords are highlighted in red; please zoom in for more visual details.

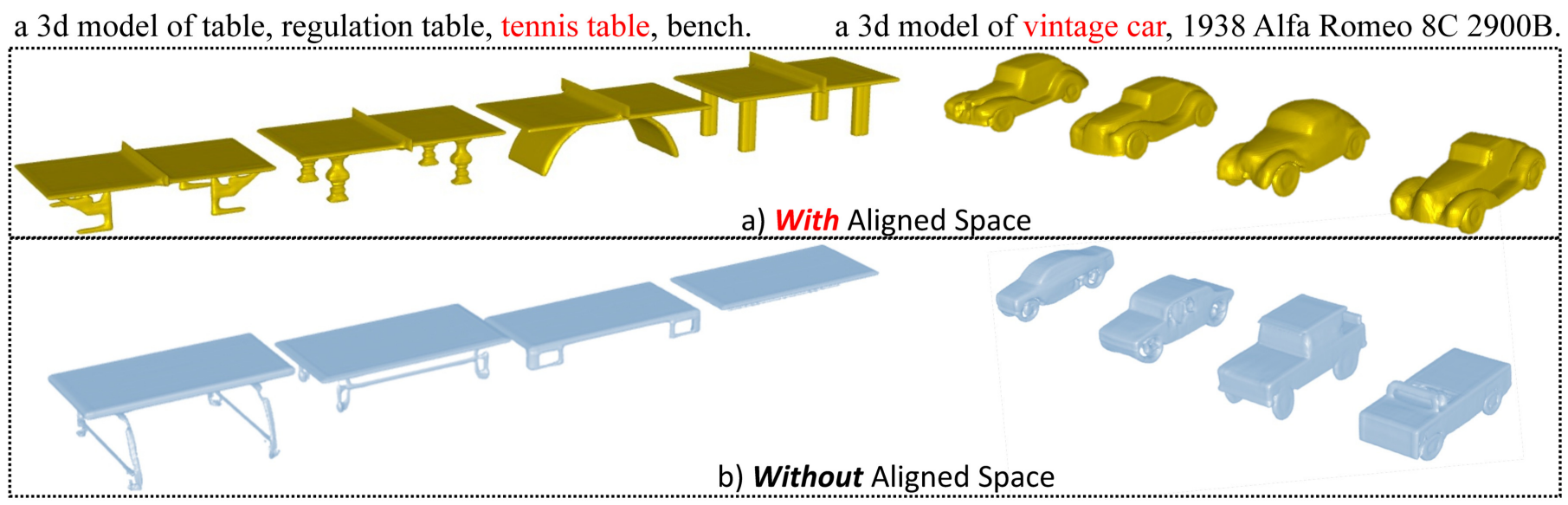

Figure 8.24: **Ablation study the effectiveness of training generative model in the aligned space**. This figure illustrates visual comparisons for ablation studies on the effectiveness of training the generative model in the aligned space. Compared with the lower samples based on the conditional texts, the upper samples are closer to the conditions semantically, which indicates the effectiveness of the training generative model in the aligned space.

**The effectiveness of vision-language models**. In addition to the well-known vision-language model (VLM) CLIP [RKH+21c], another vision-language model (VLM) SLIP [MKW+22] is introduced for training the SITA-VAE for a comprehensive comparison. First, the impact of the vision-language model on SITA-VAE's reconstruction ability is evaluated, and the results are shown in Figure 8.25. The results show that the model composed with CLIP achieves the best performance. Then, the vision-language model's impact on the ability to align multi-modal space is evaluated. Standard and zero-shot classification tasks are selected to reflect the impact of the vision-language model. Note that the classification is performed by a feature matching operation, where multiple 3D shapes and phrases are provided to the SITA-VAE; it returns the similarity between 3D shapes and each phrase as classification results, which indicates that the more aligned the multi-modal space is, the higher the classification accuracy. The results show that the model composed with CLIP achieves the best performance.

**The impact of the learnable query embeddings**. An ablation study of learnable query embeddings is performed with the same experiments as above, and the results show that using 512 learnable query embeddings leads to the best performance on reconstructions and classifications.

**Nearest Neighbor Analysis.** To verify whether the model learns to generate results based on the given conditions or memorizes the training split of the dataset, the training set is traversed to retrieve the nearest neighbors of the generated samples, as illustrated in Figure 8.26. Specifically, three nearest neighbors for each sample are exhibited for comparison. According to the visualization, the generated 3D shapes differ from each retrieved nearest neighbor, demonstrating that the model hallucinates new 3D shapes rather than overfitting the training sets.

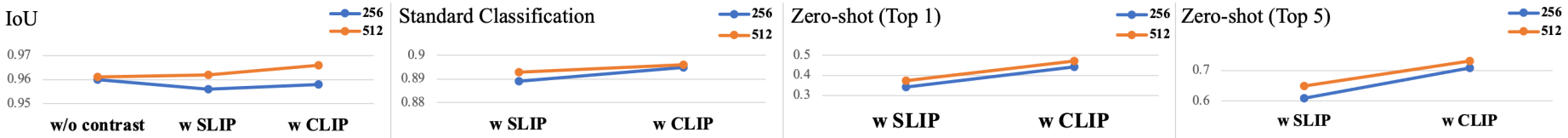

Figure 8.25: **Ablation study of the effectiveness of vision-language models and the impact of learnable query embeddings**. This figure shows the ablation study on the effectiveness of the vision-language model and the impact of learnable query embeddings. According to the table, the model with CLIP and 512 learnable query embeddings achieves the best reconstruction and classification performance, indicating its ability to recover 3D shapes and align multi-modal space.

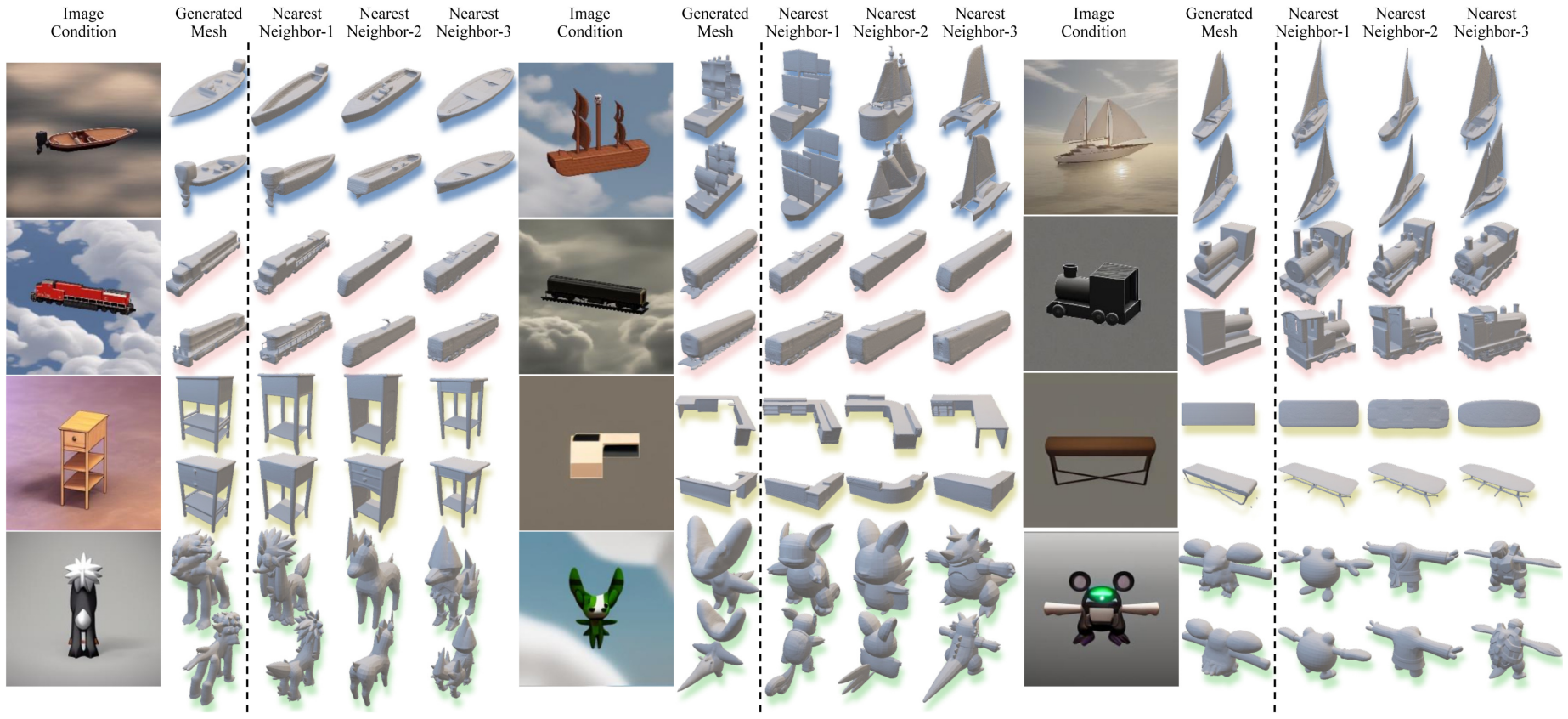

Figure 8.26: **Nearest Neighbor Analysis**. We traverse the whole training set to find three nearest neighbors for the generated 3D shapes, and the results reveal that our model could produce novel 3D shapes based on given images instead of memorizing specific ones.

**Discussion.** Though the Michelangelo project has shown that we can achieve excellent results in generating 3D objects, it still has some limitations. First, it requires a large number of samples of ground truth 3D shapes for training and learning the distribution, whereas such 3D shape data are much more expensive to obtain and available datasets are typically orders of magnitude smaller than those for 2D images. As we have alluded to in Section 7.5.2 at the end of the previous chapter, our brain is able to learn a full 3D (or 4D) visual model by having only 2D images as such data are much natural and economic to acquire. Hence, ultimately, to truly scale up the generative 3D model, we have to develop more advanced methods that would enable us to learn the shape representation (within a 3D shape-image-text aligned space) from only (multi-view) 2D images, say via differentiable rendering. Furthermore, since we represent each 3D shape as an occupancy field, it needs to convert the 3D mesh into a watertight one, which will inevitably degrade the original quality of the 3D mesh.

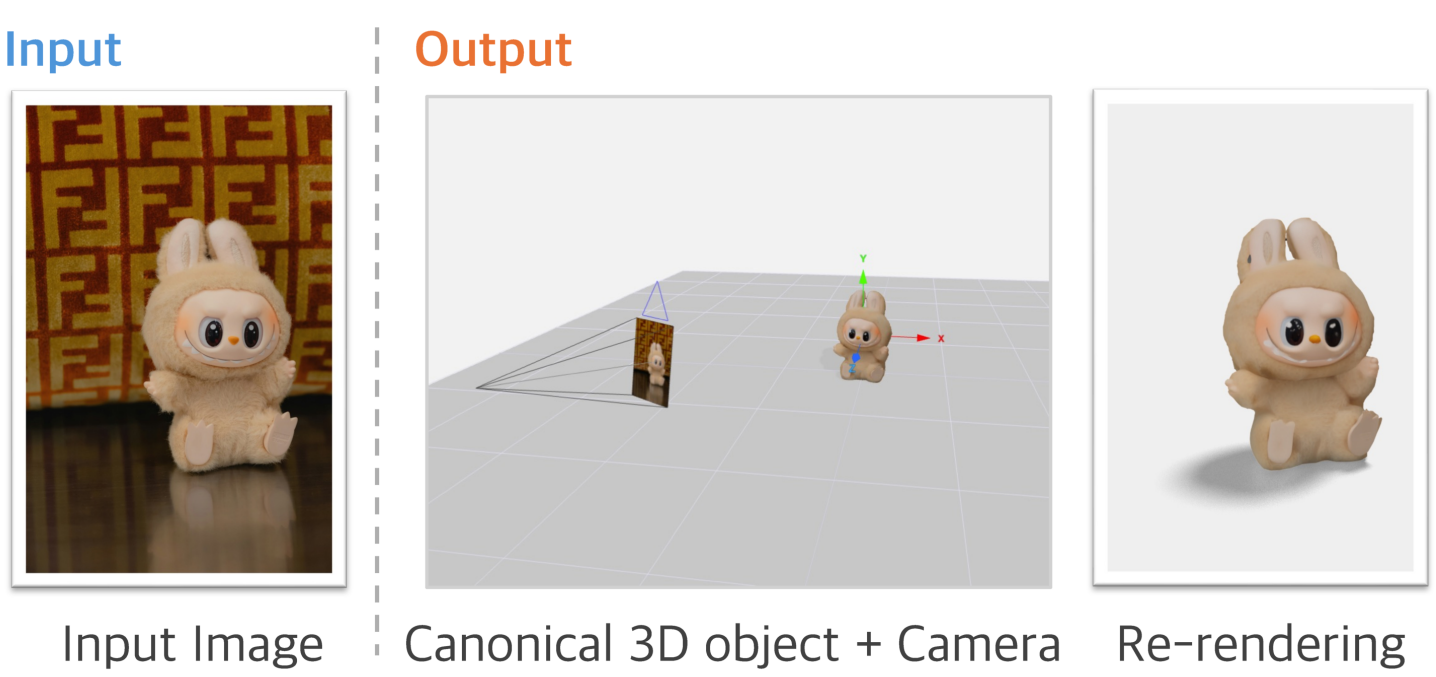

Figure 8.27: **Task: to infer the 3D shape of an object with a canonical pose relative to the camera pose from a single image.** Left: Input image. Middle: Output recovers a canonical 3D model of the object (with intrinsic properties such as shape and texture) along with the exact camera pose (extrinsic property) that generates the input. Right: Re-rendering the 3D model from the estimated pose faithfully reproduces the input.

## 8.9 Generation-Based 3D Shape and Pose Reconstruction

The fundamental reason why we develop the sense of vision is to learn to perceive our three-dimensional (3D) environment. Our brain has dedicated a significant portion of it, *the visual cortex*, to perform this task. It enables us to accurately and efficiently reconstruct shapes of 3D objects and their spatial relationships[18] from the 2D images perceived through our eyes. There has been a rich and long history of understanding mathematically how, in principle, we can fully reconstruct 3D geometry from (multiple) 2D images. Interested readers may refer to the textbook [MKS+04] for a detailed account of this important topic.

Nevertheless, as we have alluded to in the Section 7.5.2 in the previous Chapter 7, ideally, 3D reconstruction should be done in a Bayesian manner as our brain does so too. Our perception of the 3D environment seems effortless is mainly because we reconstruct a 3D scene and recognize 3D objects based on a rich "visual memory." More precisely, we have learned through our years of accumulated experience the distribution of out 3D environment and objects within. The distribution is rather low-dimensional and structured hence we can correctly and efficiently complete 3D geometry of a scene and objects in it even if we have seen only small part of it, say from a single view, as illustrated by the example in Figure 8.27. Conceptually, this ability to "complete" missing information is very similar to how natural images can be completed from a small fraction of patches, as we have seen in Section 8.5.

As a distribution, our 3D visual memory is not only low-dimensional but

[18]say relative poses relative to our viewpoint or between the objects themselves.

also highly structured. For instance, it seems that our brain stores semantics (identity of objects) of a scene in the *inferior temporal* cortex (IT cortex) and the 3D spatial information of the scene in the *hippocampus*. Our brain not only reconstructs the 3D geometry of the scene,[19] but also parses its content into individual objects and their spatial relationships with the viewer and among the objects themselves. Such a structured "what-is-where" representation enables us to conduct spatial inference within the scene from either an *egocentric, object-centric, or allocentric* perspective. These abilities are necessary and crucial for tasks such as navigating in a scene, interacting and manipulating objects within.

Hence, if the goal of learning a model of the 3D scene is to support physical interaction with an environment and objects within,[20] we must learn a representation of the 3D environment that shares the same properties and characteristics as our visual memory mentioned above.

### 8.9.1 Task and Objective

**Object 3D Shape and Poses from a Single 2D View** In this section, we takes a first step to show how this can be achieved by illustrating with a challenging and seemingly ill-posed problem: *inferring the complete 3D shape and and pose, including both ego-centric and object-centric pose, of an object from a single 2D image.* This problem, by its own, lies at the heart of computer vision, aiming to reverse the physics of image formation. An image $\boldsymbol{I}$ is the result of a rendering process,

$$\boldsymbol{I} = \boldsymbol{\mathcal{P}}(\mathcal{O}, \boldsymbol{\theta}),$$

where $\mathcal{O}$ represents the 3D object in a canonical, view-agnostic frame and $\boldsymbol{\theta}$ denotes the camera viewpoint. Given only $\boldsymbol{I}$, the task of disentangling the object's intrinsic properties (e.g., shape, texture) from the extrinsic camera pose is fundamentally ambiguous.

Previous approaches typically oversimplify this ambiguity. Generative models often prioritize the canonical object $\mathcal{O}$ at the expense of the camera pose $\boldsymbol{\theta}$, yielding reconstructions that fail to align faithfully with the input view. Conversely, traditional view-centric methods fix the pose, which limits their ability to generate or "in-paint" occluded regions and seamlessly paste the reconstructed object back into the original image.

Here we present an approach known as **Cupid** [HDZ+25]:

`https://cupid3d.github.io`,

an open-source framework that jointly models the distribution over both the 3D object and camera pose (Figure 8.27). By simultaneously generating the 3D model and reasoning about the viewpoint, Cupid ensures the output can be accurately *pasted* into the original image. This approach recasts single-view reconstruction as a conditional sampling problem, achieving geometric fidelity by

[19] Say 3D geometry that is represented in different but largely equivalent forms: point clouds, meshes, signed distance functions, neural radiance fields, or Gaussian splatters.

[20] which is the objective of the so-called "Embodied AI" for intelligent agents or robots.

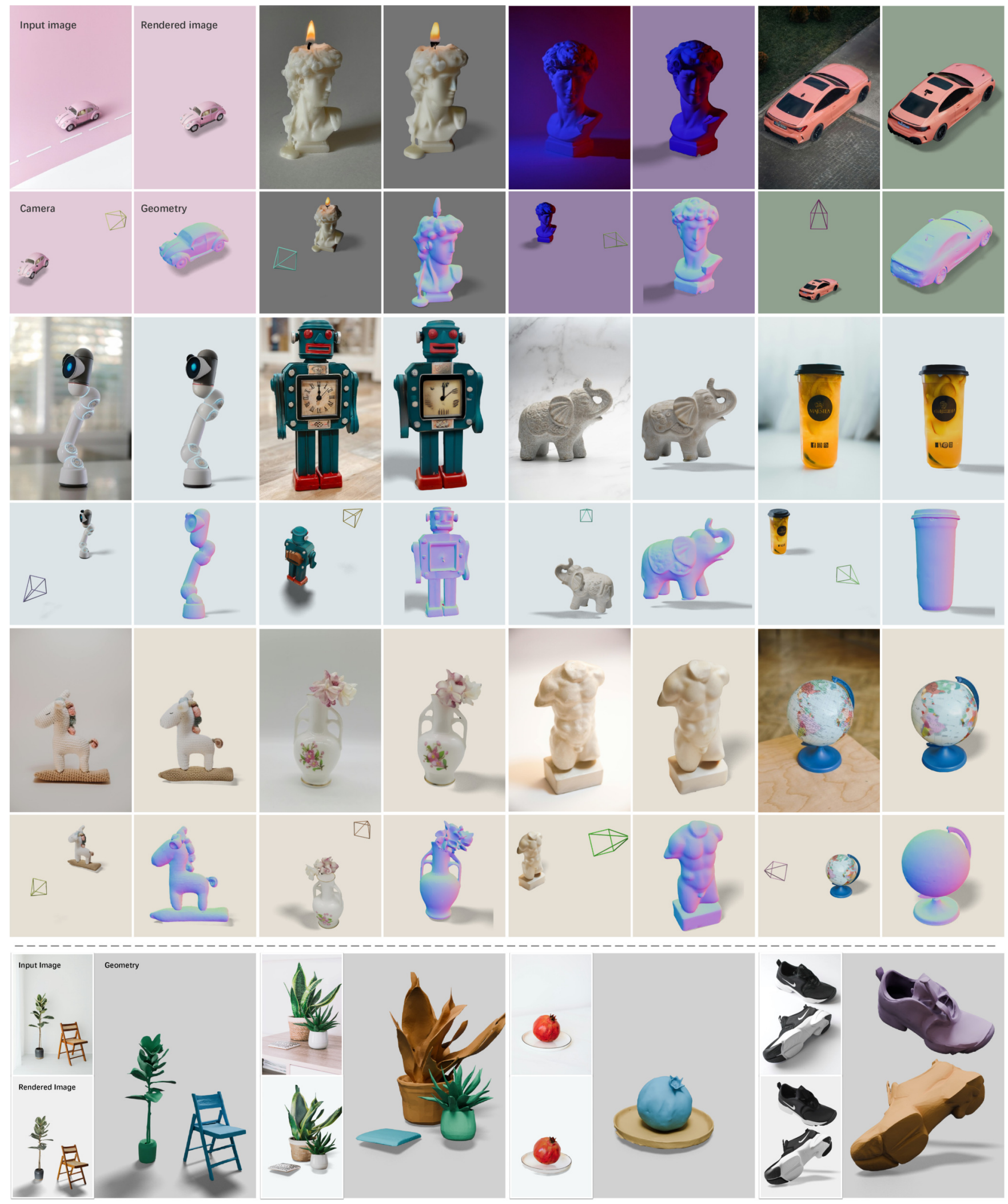

Figure 8.28: **Results for *generative 3D reconstruction* from a single test image.** Given an input image (top left), Cupid estimates camera pose (bottom left) and reconstructs 3D model (bottom right), re-rendering the input (top right). It is robust to changes in scale, placement, and lighting while preserving fine details, and supports component-aligned scene reconstruction (bottom row). All results are produced in seconds via feed-forward sampling of the learned model.

combining observational data with learned generative priors, thereby enabling efficient object- and scene-level reconstruction (Figure 8.28).

**Task Formulation and Objective.** We formulate generative 3D reconstruction as the task of estimating the joint posterior distribution

$$p(\mathcal{O}, \boldsymbol{\theta} \mid \boldsymbol{I})$$

under the constraint that the observation $\boldsymbol{I}$ is a rendering of the object $\mathcal{O}$ from pose $\boldsymbol{\theta}$. To model this conditional distribution, we employ a flow-based generative model. Specifically, we first map the 3D object and camera pose into a unified volumetric latent representation $\boldsymbol{z} = \varphi(\mathcal{O}, \boldsymbol{\theta})$. We then use a Rectified Flow [LCB+23] model to learn the conditional generation of this latent variable. The model defines a linear interpolation between a sample from the data distribution, $\boldsymbol{z}_0$, and a sample from a noise distribution, $\boldsymbol{\epsilon}$, over a normalized time interval $t \in [0, 1]$:

$$\boldsymbol{z}_t = (1 - t)\boldsymbol{z}_0 + t\boldsymbol{\epsilon}. \tag{8.9.1}$$

The generative process involves reversing this trajectory by learning a time-dependent velocity field

$$\boldsymbol{v}(\boldsymbol{z}_t, \boldsymbol{I}, t) = \nabla_t \boldsymbol{z}_t$$

that transports noisy samples towards the data manifold, conditioned on the input image $\boldsymbol{I}$. We parameterize this velocity field with a neural network $\boldsymbol{v}_\varphi$ and train it using the Conditional Flow Matching (CFM) objective [LCB+23]:

$$\mathcal{L}_{\mathrm{CFM}}(\varphi) \doteq \mathbb{E}_{t, \boldsymbol{z}_0, \boldsymbol{\epsilon}} \left\| \boldsymbol{v}_\varphi(\boldsymbol{z}_t, \boldsymbol{I}, t) - (\boldsymbol{\epsilon} - \boldsymbol{z}_0) \right\|_2^2. \tag{8.9.2}$$

By sampling from this learned model, we can generate a latent code $\boldsymbol{z}$ which can then be decoded into the final 3D object and camera pose.

### 8.9.2 3D Shape Representation and Tokenization

A central challenge in 3D shape generation is finding a tokenization that is both compact and expressive, capable of high-fidelity reconstruction. While **Michelangelo** [ZLC+23a] utilizes a learned latent vector representation—encoding shapes into a sequence of 1024 tokens—these vectors lack explicit 3D spatial coordinates. In this project, we seek a latent representation that retains explicit 3D positional information, allowing us to align shape tokens directly with image pixels. By establishing this spatial correspondence, we can leverage local pixel information to enhance generation details. Consequently, **Cupid** adopts a 3D voxel-based latent representation and tokenization approach as described in Section8.8.1. Here, we represent a shape with sparse voxels accompanied by features, which are usually defined by RGB color or more advanced DINO features [ODM+24] from 2D foundation models. We then train a decoder to map such a representation back to the shape, parameterized by SDF, and to the apperance parameterized by the Gaussian splats [KKL+23].

Specifically, given a set of sparse voxel features $\boldsymbol{F} = \{\boldsymbol{f}_i\}_{i=1}^N$ located at coordinates $\boldsymbol{x}_i$, we employ two distinct decoders, $\mathcal{D}_{geo}$ and $\mathcal{D}_{app}$, to disentangle geometry from appearance. The decoding process is formulated as:

$$s_i = \mathcal{D}_{geo}(\boldsymbol{f}_i), \quad \mathcal{G}i = \mathcal{D}_{app}(\boldsymbol{f}_i), \tag{8.9.3}$$

where $s_i \in \mathbb{R}$ represents the Signed Distance Field (SDF) value used for surface reconstruction, and $\mathcal{G}_i = \{\boldsymbol{c}_i, \alpha_i, \boldsymbol{s}_i, \boldsymbol{q}_i\}$ denotes the set of 3D Gaussian parameters—comprising color, opacity, scaling, and rotation—required for volumetric rendering. We will then use the sparse voxel features as the latent representation for latent flow matching [RBL+22].

### 8.9.3 Architecture and Implementation

Since we utilize 3D voxels for object representation, applying diffusion naively results in excessive memory consumption, as a large portion of the grid consists of unnecessary empty space, as Section 8.8.1 describes. To mitigate this, the **Cupid** architecture is designed as a cascaded, two-stage flow model that progressively refines the 3D reconstruction [XLX+25]. The first stage generates a coarse geometric shape and estimates the camera pose, while the second stage generates high-fidelity geometry and appearance. A key innovation of this framework lies in the explicit representation of the camera pose and its integration into the generative process. The overall pipeline is depicted in Figure 8.29.

**Decoupled Object and Pose Representation.** As discussed in Section 8.8.1, the 3D object $\mathcal{O}$ is tensorized into a sparse voxel-based representation:

$$\mathcal{O} \triangleq \{\boldsymbol{x}_i, \boldsymbol{f}_i\}_{i=1}^{L}, \tag{8.9.4}$$

where $\boldsymbol{x}_i \in \mathbb{R}^3$ denotes the coordinate of an active voxel, and $\boldsymbol{f}_i$ is a feature vector encoding local geometry and appearance. These features are derived by compressing a dense 3D grid of multi-view DINO [ODM+23] features via a 3D VAE. The VAE decoder is subsequently trained to reconstruct $\mathcal{O}$ into a Signed Distance Field (SDF) for mesh extraction. This 3D latent representation is illustrated in the upper section of the central blue block in Figure 8.29.

The camera pose $\boldsymbol{\theta}$, represented by its projection matrix $\boldsymbol{P} = \boldsymbol{K}[\boldsymbol{R}|\boldsymbol{t}]$, is reparameterized in a novel way to facilitate joint generation. Instead of a compact vector, we represent the pose as a dense field of 3D-to-2D correspondences, which we term an *in-cube pixel distribution*:

$$\boldsymbol{\theta} \triangleq \{\boldsymbol{x}_i, \boldsymbol{u}_i\}_{i=1}^{L},$$

where $\boldsymbol{u}_i = (u_i, v_i) \in [0,1]^2$ are the normalized 2D pixel coordinates corresponding to the 3D voxel center $\boldsymbol{x}_i$. These correspondences are obtained via perspective projection:

$$\boldsymbol{u}_i = \pi(\boldsymbol{P}, \boldsymbol{x}_i).$$

This set of correspondences can be visualized as a 3D UV cube, where the $(u, v)$ values act like view-dependent colors on a 3D grid. This is illustrated at the bottom of the blue block in the middle of Figure 8.29.

Given a generated set of correspondences, the global camera matrix $\boldsymbol{P}^*$ is recovered by solving a *Perspective-n-Point* (PnP) problem using a least-squares

solver [AKH15]:

$$\boldsymbol{P}^* = \arg\min_{\boldsymbol{P}} \sum_{i=1}^{L} \left\| \pi(\boldsymbol{P}, \boldsymbol{x}_i) - \boldsymbol{u}_i \right\|^2. \tag{8.9.5}$$

This formulation elegantly transforms the joint object-pose generation problem into one of generating an object-centered 3D shape/appearance with an associated observer/camera centered pose of the object (via the UV coordinates). This is illustrated at the bottom left corner of the the red block on the right of Figure 8.29.

*Remark* 8.1 (Structured 3D Modeling). Notice that the voxel-type representations chosen here for shape and pose are different from the point cloud type representations adopted in the conditioned shape generation task studied in the preceding Section 8.8. In that task, the goal is to generate an object that only needs to be strongly correlated to the given image (or text). It does not require an accurate pose alignment between the object (in its canonical pose) and the given image (viewpoint). For the task considered here, however, we want to "reconstruct" an object whose shape, appearance, and pose faithfully agrees with the given image. Therefore, it is necessary for the internal representation to explicitly model the object's canonical shape and relative pose simultaneously. Note that such a "decoupling" of an object (canonical) shape/appearance and a pose from which it being observed closely imitates the aforementioned roles of the IT cortex and Hippocampus in our brain, respectively. As we will see later, such a decoupled representation will enable us to obtain a structured and consistent 3D model of a scene that may contain multiple objects from multiple views. We contend that a so-structured 3D model is necessary for us to conduct view-centric or object-centric reasoning,[21] generation, and interaction with the scene.

### 8.9.4 Cascaded Flow Modeling

We employ a two-stage cascaded flow model to jointly sample a 3D object and its corresponding camera pose, denoted as $\boldsymbol{z} = \{(\boldsymbol{x}_i, \boldsymbol{f}_i), (\boldsymbol{x}_i, \boldsymbol{u}_i)\}_{i=1}^{L}$. In the first stage, the occupancy and pose generation model ($\boldsymbol{\mathcal{G}_S}$) produces two key outputs: (i) an occupancy cube identifying active voxels, and (ii) a UV cube encoding the object-centric camera pose. In the second stage, conditioned on these predictions, the pose-aligned geometry and appearance model ($\boldsymbol{\mathcal{G}_L}$) synthesizes DINO [ODM+23] features $\boldsymbol{f}_i$ for each active voxel to yield the final latent $\boldsymbol{z}$.

**Occupancy and Pose Generation.** Given a conditioning image $\boldsymbol{I}$, this stage generates an occupancy cube and a UV cube, both at resolution $r$. The occupancy cube, $\boldsymbol{G}_o \in \{0,1\}^{r\times r\times r\times 1}$, contains binary values distinguishing active from inactive voxels. The UV cube, $\boldsymbol{G}_{uv} \in [0,1]^{r\times r\times r\times 2}$, stores normalized pixel coordinates $(u, v)$ for every voxel; notably, voxels sharing identical pixel coordinates lie along the same camera ray.[22]

[21]For example, answering questions such as "which object is in front of, behind, to the left or right of, or on top of which object.

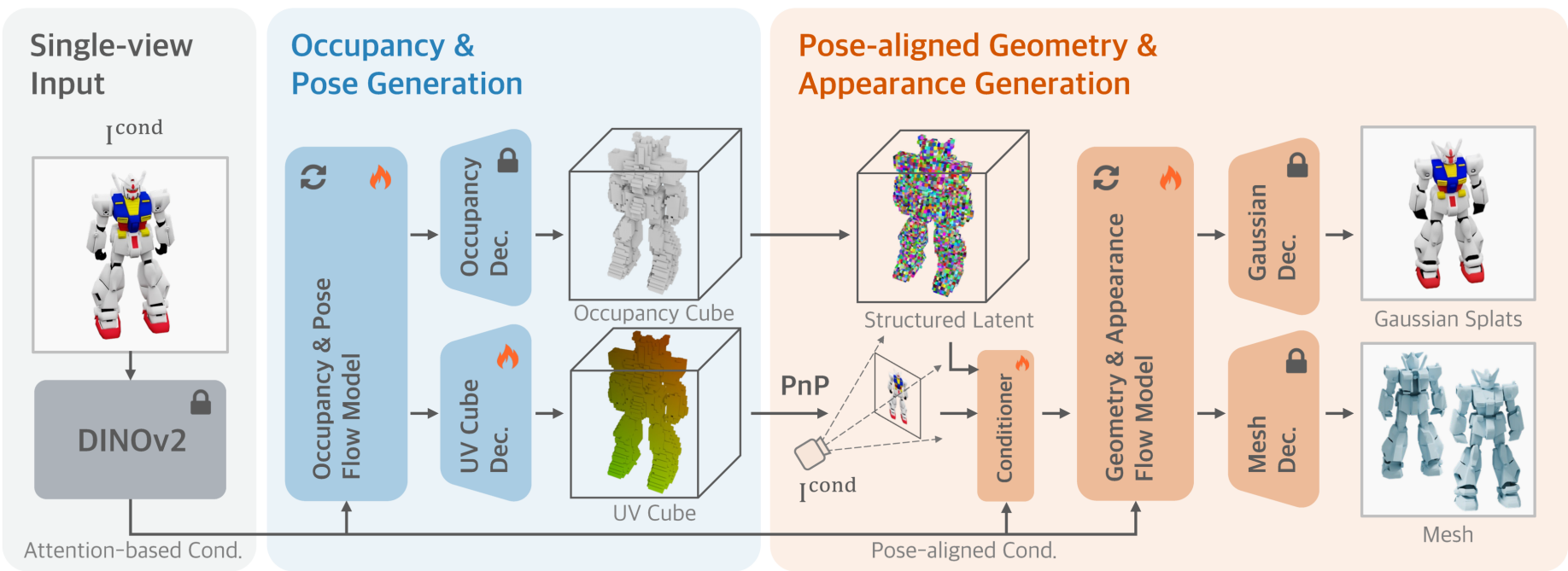

Figure 8.29: **The Cupid Two-Stage Generative Reconstruction Pipeline.** Given an input image, the first stage ($\boldsymbol{\mathcal{G}}_{\boldsymbol{S}}$) generates a coarse occupancy cube and a UV cube, which encodes 3D-to-2D correspondences. A PnP solver recovers the camera pose $\boldsymbol{P}^*$ from these correspondences. The second stage ($\boldsymbol{\mathcal{G}}_{\boldsymbol{L}}$) is conditioned on this recovered pose. It injects pixel-aligned features (sampled from DINOv2 and low-level feature maps) into the generative process to synthesize high-fidelity geometry and appearance for the final 3D object.

To improve computational efficiency, we train a 3D VAE to compress the UV cube into a low-resolution feature grid $\boldsymbol{S}_{uv} \in \mathbb{R}^{s \times s \times s \times C}$, achieving near-lossless pose recovery (mean RRE/RTE $< 0.5°$). To fine-tune TRELLIS, we concatenate the original feature grid $\boldsymbol{S}_o$ with $\boldsymbol{S}_{uv}$ and introduce linear projection layers at both the input and output of the flow network $\boldsymbol{\mathcal{G}}_{\boldsymbol{S}}$. Once the occupancy and UV cubes are generated, we extract the set $\{(\boldsymbol{x}_i, \boldsymbol{u}_i(\boldsymbol{\theta}))\}_{i=1}^{L}$ by collecting active voxels and solving for the camera pose via Equation 8.9.5.

**Pose-Aligned Geometry and Appearance Generation.** In the second stage, we generate detailed 3D latent features $\{\boldsymbol{f}_i\}_{i=1}^{L}$ exclusively at active voxel locations. Experiments indicate that the standard $\boldsymbol{\mathcal{G}}_{\boldsymbol{L}}$ model from TRELLIS [XLX+25], which relies on globally attended image information, is prone to color drift and the loss of fine-grained details. We mitigate this by utilizing the camera pose derived in the first stage to inject locally attended pixel information into each voxel.

Specifically, we compute the features for the $i$-th voxel using the estimated pose as follows:

$$
\begin{aligned}
\boldsymbol{f}_i^{\text{DINO}} &= \text{BilinearInterp}(\boldsymbol{u}_i, \text{DINO}(\boldsymbol{I})) \in \mathbb{R}^{1024}, \\
\{\boldsymbol{f}_i^{\text{h}}\}_{i=1}^{L} &= \text{SlatEncoder}\left(\{\boldsymbol{x}_i, \boldsymbol{f}_i^{\text{DINO}}\}_{i=1}^{L}\right), \quad \boldsymbol{f}_i^{\text{h}} \in \mathbb{R}^{8},
\end{aligned}
\tag{8.9.6}
$$

[22] We do not explicitly model occlusion, as recent studies suggest transformers effectively model light transport implicitly [JJT+24; ZDP+25].

where $\boldsymbol{u}_i$ represents the projection of the $i$-th 3D voxel center onto the image plane, BilinearInterp denotes bilinear interpolation, and SlatEncoder is the 3D VAE encoder. Although DINO [ODM+23] captures high-level semantics, it often lacks the low-level cues necessary for precise reconstruction. To compensate, we extract complementary low-level features $\boldsymbol{f}^{\mathrm{l}}$ from $\boldsymbol{I}$ using a lightweight convolutional head, sampling them at $\boldsymbol{u}_i$ via BilinearInterp. Finally, at each time step $t$, we fuse the noisy voxel feature $\boldsymbol{f}_i^t$ with the pixel-aligned features via a linear layer before inputting them into the flow transformer:

$$l_t = \mathrm{Linear}\big([\boldsymbol{f}_i^t \oplus \boldsymbol{f}_i^{\mathrm{h}} \oplus \boldsymbol{f}_i^{\mathrm{l}}]\big). \tag{8.9.7}$$

This pose-aligned fusion strategy significantly enhances both 3D geometric accuracy and appearance fidelity relative to the input image.

*Remark* 8.2 (Decoupled Generative Processes)*.* Notice that the above two-stage framework decouples the distribution of 3D objects into several related components, shape, pose, and appearance and learn them separately. Although this means we have to learn two generative denoising models together, the resulting learned representations are much better structured and enable much more precise and flexible spatial reasoning and generation.

### 8.9.5 Experiments

To validate the architecture, we evaluated the system across three critical dimensions: monocular geometry prediction, input-view consistency, and full single-image-to-3D reconstruction. The benchmarking process focused on learning-based systems, excluding per-scene optimization methods to ensure a fair comparison of inference capabilities. We ablate pose-conditioning.

**The Comparative Landscape.** We benchmarked against three distinct paradigms currently dominating the field:

(i) *Point-map Regression*, such as VGGT [WCK+25] and MoGe [WXD+25], which simultaneously predict per-pixel 3D points and camera poses but often lack robust priors for occluded geometry;

(ii) *View-centric Reconstruction*, including LRM [HZG+23] and LaRa [CXE+24], which generate 3D models directly in view space to bypass test-time pose estimation; and

(iii) *Decoupled Generation*, exemplified by OnePoseGen [GWX+25], which separates canonical 3D generation from pose estimation, relying on post-hoc alignment.

**Geometric Accuracy and Pipeline Simplification.** In monocular geometry tasks, the proposed model demonstrated a clear superiority over view-centric approaches (Table 8.11). On the GSO dataset [DFK+22], it reduced the average

Table 8.11: **Monocular geometry accuracy.** Cupid outperforms all 3D reconstruction and generation baselines and matches point-map regression methods that predict only partial geometry. Note that VGGT uses a ground-truth object mask, which may overestimate accuracy.

| Method | 3D | Toys4k mIOU (avg)↑ | Toys4k CD (avg)↓ | Toys4k CD (med)↓ | Toys4k F-score (0.01)↑ | Toys4k F-score (0.05)↑ | GSO mIOU (avg)↑ | GSO CD (avg)↓ | GSO CD (med)↓ | GSO F-score (0.01)↑ | GSO F-score (0.05)↑ |
|---|---|---|---|---|---|---|---|---|---|---|---|
| VGGT | × | – | 1.144 | 0.498 | 61.85 | 95.90 | – | 1.396 | 0.388 | 65.98 | 95.95 |
| MoGe | × | 92.80 | 1.284 | 0.581 | 58.54 | 95.31 | 96.18 | 1.743 | 0.575 | 58.99 | 94.68 |
| OnePoseGen | ✓ | 9.34 | 153.2 | 59.92 | 6.11 | 24.10 | 12.16 | 116.2 | 60.56 | 7.28 | 25.77 |
| LaRa | ✓ | 68.11 | 32.15 | 16.59 | 18.57 | 57.67 | 70.63 | 34.23 | 19.36 | 13.48 | 49.95 |
| OpenLRM | ✓ | 86.26 | 2.726 | 1.291 | 40.42 | 90.60 | 91.35 | 3.741 | 1.858 | 34.14 | 87.20 |
| Ours | ✓ | **92.43** | **2.534** | **0.236** | **69.82** | **97.76** | **95.27** | **1.823** | **0.434** | **61.01** | **95.59** |

Table 8.12: **Comparison on full 3D quality.** We report CLIP image scores of novel views following [GHH+24].

| | OnePoseGen | LaRa | OpenLRM | TRELLIS | Cupid |
|---|---|---|---|---|---|
| **ViT-B/16** | 0.7933 | 0.8334 | 0.8939 | 0.9465 | **0.9501** |
| **ViT-L/14** | 0.7193 | 0.7682 | 0.8410 | 0.9210 | **0.9291** |

Chamfer Distance (CD) by 50% relative to LRM [HZG+23]. This improvement is attributed to the model's dual capability: it accurately models object intrinsics via generative priors while simultaneously predicting precise extrinsic camera parameters.

Crucially, the model achieved geometric accuracy competitive with specialized point-map regressors like MoGe [WXD+25]. This finding has significant implications for pipeline design: it suggests that complex, multi-stage workflows—which traditionally use partial geometry estimation as a prerequisite step [YZY+25]—can be streamlined. Our approach achieves comparable accuracy in a single step, rendering those intermediate representations unnecessary.

**Visual Fidelity and Consistency.** We evaluated input-view consistency to measure how well the 3D reconstruction realigns with the source image (Table 8.13). The results highlighted specific weaknesses in the baselines: LaRa's texture quality degraded due to inconsistent novel-view generation, while OnePoseGen suffered from severe texture misalignment and color shifting during registration.

By contrast, our method maintained high pose accuracy without sacrificing texture details (Figure 8.30). It successfully reconstructed high-fidelity objects from both synthetic datasets (Toys4K) and challenging "in-the-wild" images (Figure 8.28), consistently outperforming baselines in semantic alignment metrics such as CLIP similarity [RKH+21b] (Table 8.12).

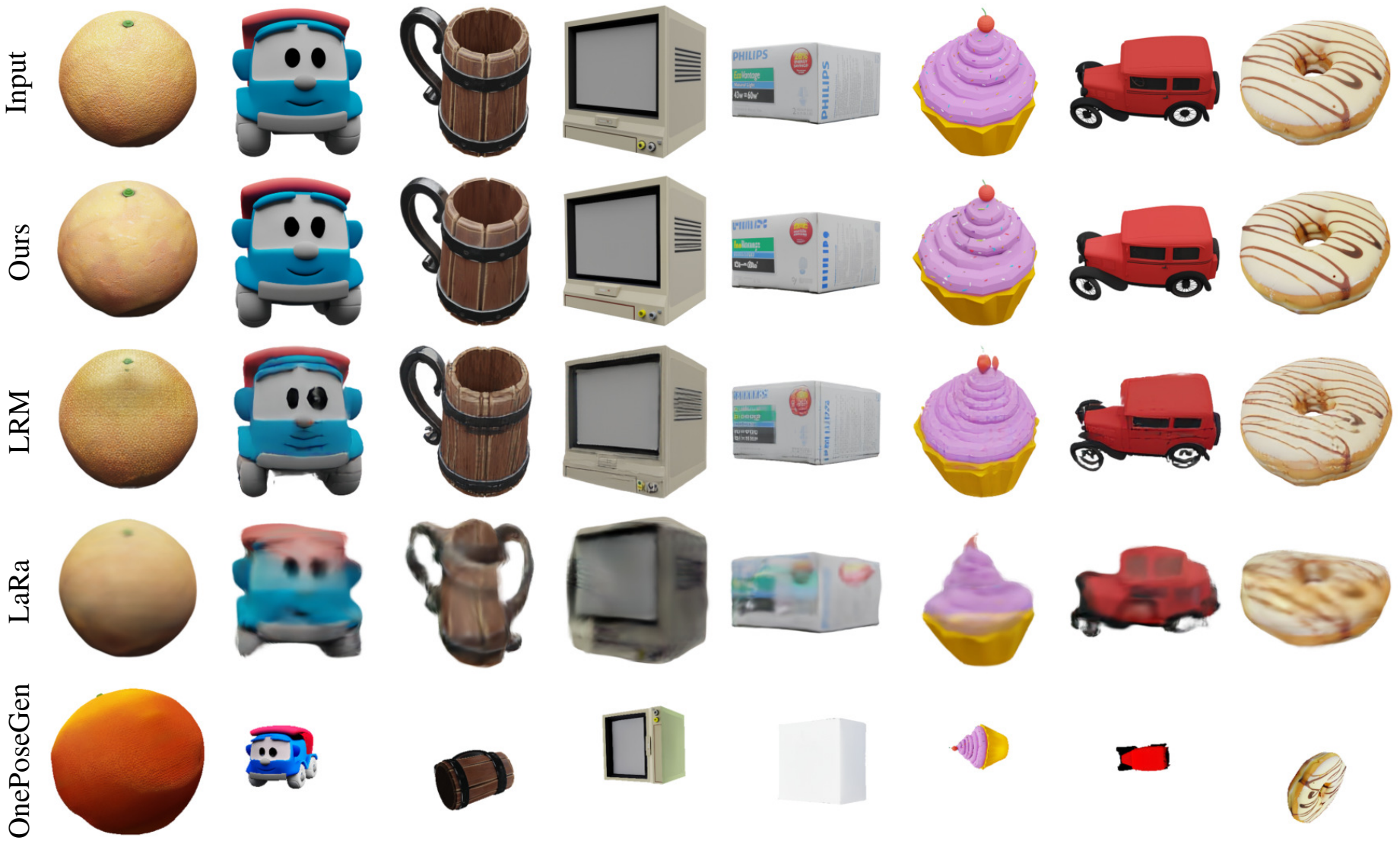

Figure 8.30: **Qualitative comparison on input view consistency.** We render the input view using its generated camera pose. For view centric methods (LRM, LaRa), we use ground-truth intrinsic for rendering as they do not model intrinsic. Our method produces the highest-fidelity geometry and appearance; LRM hallucinates incorrect details, LaRa is overly blurry due to 2D diffusion inconsistencies, and 3D generation method OnePoseGen frequently fails to register pose reliably.

**Ablation Studies** Ablation studies on the second-stage refinement confirmed the critical role of Pose-Aligned Conditioning (PAC). We tested variants ranging from standard latent baselines to those incorporating DINOv2 feature volumes.

As shown in Table 8.14, the most effective configuration proved to be a hybrid approach: concatenating DINOv2 features with low-level visual cues extracted via convolutional layers. This combination yielded the highest visual quality, effectively bridging semantic understanding with pixel-level detail (Figure 8.31). Furthermore, this conditioning mechanism demonstrated remarkable robustness, maintaining high performance even when the input geometry contained stochastic perturbations from the first generation stage.

**Application I: Compositional Scene Reconstruction.** To extend Cupid to scene-level reconstruction, we leverage foundation models to decompose the scene into $K$ objects with masks $\{\boldsymbol{M}^{(k)}\}_{k=1}^{K}$ and estimate a global metric pointmap $\mathcal{P}$ [WXD+25].

In the first stage, we reconstruct each object independently to recover its canonical geometry, adapting the generator to handle mutual occlusions via random-mask fine-tuning [WZG+25]. For the $k$-th object, this yields a set of canonical 3D points $\boldsymbol{x}_i^{(k)}$ corresponding to pixels $\boldsymbol{u}_i \in \boldsymbol{M}^{(k)}$. To resolve the scale ambiguity inherent in independent generation, we formulate the alignment as a correspondence problem between the generated canonical space and the estimated camera space:

$$\boldsymbol{p}_i = \mathcal{P}(\boldsymbol{u}_i), \quad \forall \boldsymbol{u}_i \in \boldsymbol{M}^{(k)}, \tag{8.9.8}$$

where $\boldsymbol{p}_i$ represents the target metric coordinates in the camera frame derived from the depth foundation model.

Table 8.13: **Input-view consistency.** Cupid achieves superior input view consistency, producing accurate appearance alignment.

| Dataset / Method | Pose | Toys4K PSNR↑ | Toys4K SSIM↑ | Toys4K LPIPS↓ | GSO PSNR↑ | GSO SSIM↑ | GSO LPIPS↓ |
|---|---|---|---|---|---|---|---|
| LaRa | × | 22.00 | 93.42 | 0.0884 | 19.81 | 91.61 | 0.1119 |
| OpenLRM | × | 26.41 | 80.17 | 0.1156 | 25.79 | 78.80 | 0.1268 |
| OnePoseGen | ✓ | 17.43 | 89.37 | 0.1174 | 14.87 | 86.46 | 0.1386 |
| Cupid | ✓ | **30.05** | **96.81** | **0.0251** | **28.68** | **95.49** | **0.0354** |

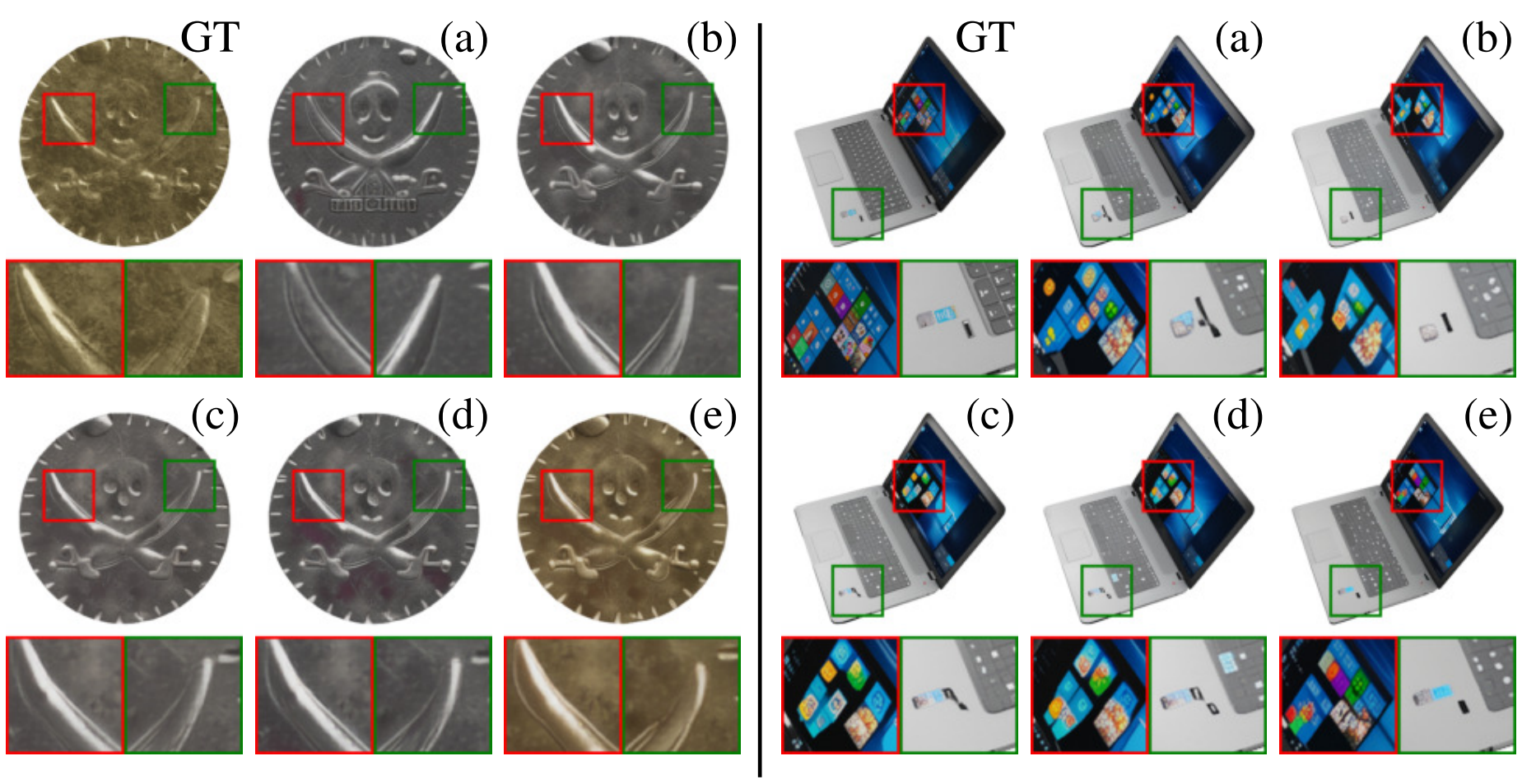

Figure 8.31: **Qualitative comparison of various pose-aligned conditioning.** Our method (e) achieves the best visual quality in terms of color fidelity and detail.

In the second stage, we compute a per-object similarity transformation $\mathcal{T}_k = \{s_k, \boldsymbol{R}_k, \boldsymbol{t}_k\}$ to place all components into a unified coordinate system. We solve for these parameters using the Umeyama method [Ume02] by minimizing the alignment error over the visible pixels of each object:

$$\mathcal{T}_k^* = \arg\min_{s_k, \boldsymbol{R}_k, \boldsymbol{t}_k} \sum_i \left\| \boldsymbol{p}_i - \left( s_k \boldsymbol{R}_k \boldsymbol{x}_i^{(k)} + \boldsymbol{t}_k \right) \right\|^2. \tag{8.9.9}$$

Table 8.14: **Ablation studies of pose-aligned conditioning.**

| Method | GT Geo & Pose | | | Sampled Geo & Pose | | |
|---|---|---|---|---|---|---|
| | **PSNR** | **SSIM** | **LPIPS** | **PSNR** | **SSIM** | **LPIPS** |
| (a) Baseline (w/o PAC) | 31.84 | 97.50 | 0.0219 | 27.47 | 95.64 | 0.0327 |
| (b) Position Embedding | 32.07 | 97.58 | 0.0211 | 27.56 | 95.67 | 0.0323 |
| (c) Latent (w/o Occ.) | 32.37 | 97.72 | 0.0201 | 27.85 | 95.87 | 0.0309 |
| (d) Latent (Occ.) | 32.39 | 97.77 | 0.0199 | 27.74 | 95.80 | 0.0313 |
| (e) Latent (Visual Feat.) | **34.86** | **98.24** | **0.0168** | **30.05** | **96.81** | **0.0251** |

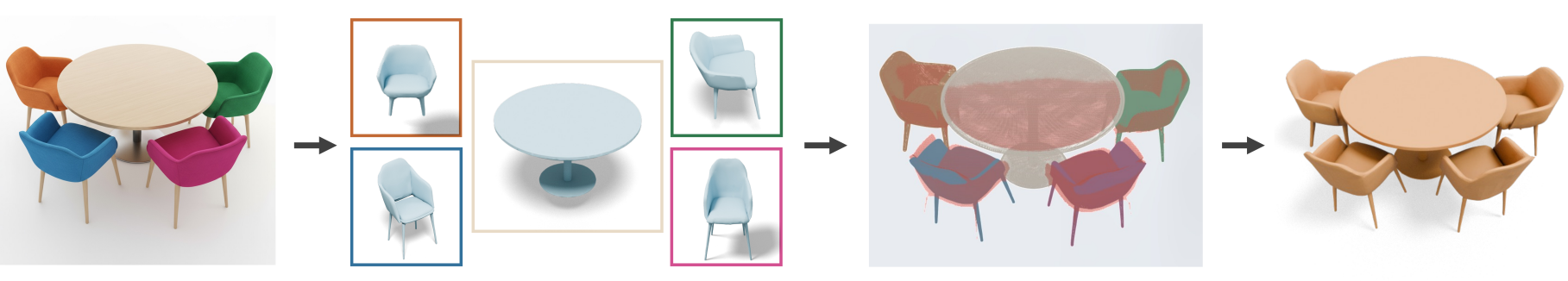

Figure 8.32: **Component-aligned scene reconstruction.** Each object is reconstructed independently with **Cupid**; explicit 3D–2D correspondences enable precise placement into a shared frame.

Applying these optimal transformations results in a metric-consistent scene composition where all generated components are correctly positioned and scaled relative to one another. Please see Figure 8.33 for more scene composition examples.

**Application II: Multi-view Consistent Reconstruction.** Thanks to our Bayesian formulation, our model generalizes to multi-view scenarios even when trained only on single images. To implement the multi-view capability described in Section 8.9.4, we extend the sampling process to accommodate $K$ input views $\boldsymbol{I}^{(k)}{}_{k=1}^{K}$.

In the first stage, we aim to recover a single canonical geometry while simultaneously estimating $K$ distinct camera poses. We initialize a shared geometry latent $\boldsymbol{S}_o$ and $K$ independent pose latents $\{\boldsymbol{S}_{uv}^{(k)}\}_{k=1}^{K}$. At each denoising step $t$, we predict the flow fields for both geometry and pose in parallel for each view. Let the output of the flow network for the $k$-th view be denoted as a concatenated vector of geometry and pose updates:

$$[d_{o,t}^{(k)}, d_{uv,t}^{(k)}] = \boldsymbol{\mathcal{G}}_{\boldsymbol{S}}(\text{concat}(\boldsymbol{S}_o^t, \boldsymbol{S}_{uv}^{(k),t}), t, \boldsymbol{I}^{(k)}). \tag{8.9.10}$$

To ensure geometric consistency across all observations, we aggregate the geometry updates while maintaining view-specific pose trajectories. The update rule at step $t$ becomes:

$$\boldsymbol{S}_o^{t-1} \leftarrow \text{Step}\left(\boldsymbol{S}_o^t, \frac{1}{K}\sum_{k=1}^{K} d_{o,t}^{(k)}\right), \tag{8.9.11}$$

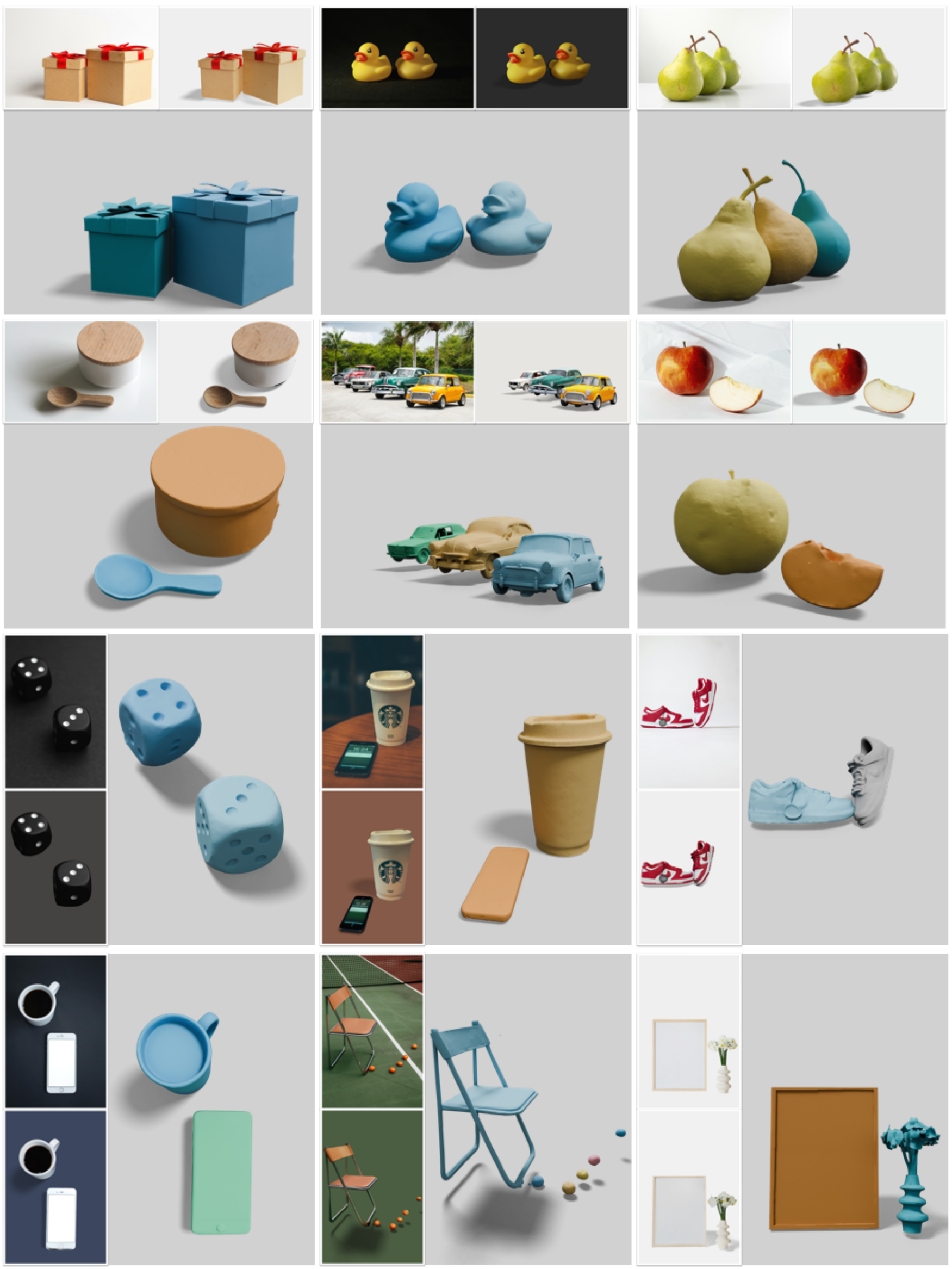

Figure 8.33: **Additional examples of component-aligned scene reconstruction.** For each example shown, the panels display: (top left) the input image, (top right or bottom left) the final rendered output, and (bottom) the reconstructed individual components, color-coded for clarity.

$$\boldsymbol{S}_{uv}^{(k),t-1} \leftarrow \text{Step}\left(\boldsymbol{S}_{uv}^{(k),t}, d_{uv,t}^{(k)}\right), \tag{8.9.12}$$

where Step($\cdot$) denotes the update rule of the flow sampler. The resulted feature grids will be decoded into a shared coarse occupancy grid $\boldsymbol{G}_o$ and poses $\{\boldsymbol{\theta}^k\}_{k=1}^{K}$.

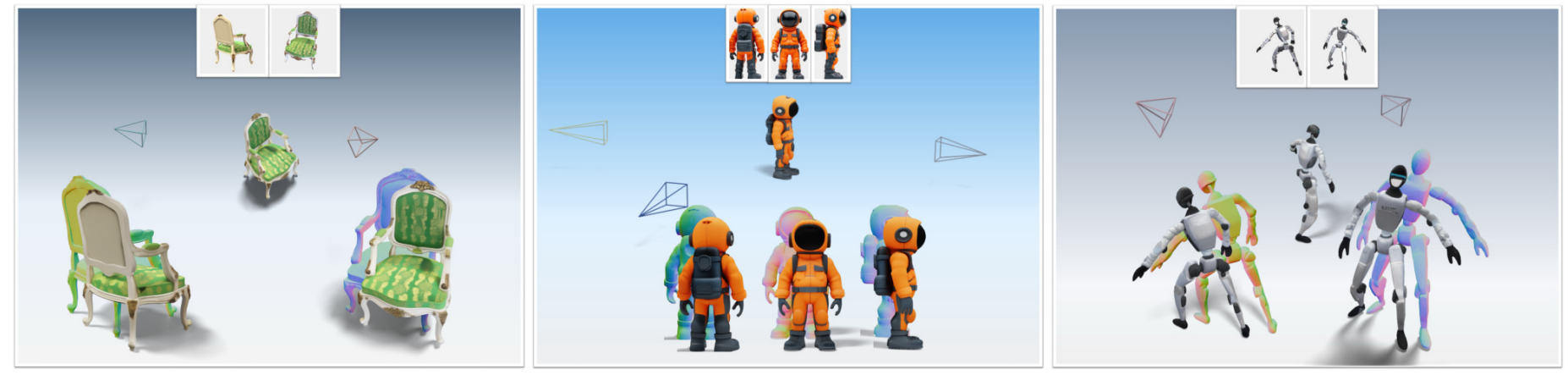

Figure 8.34: **Multi-view conditioning.** Our decoupled joint modeling naturally supports multi-view conditioning. With multiple input views available, we fuse the shared *view-agnostic* object latent across flow paths (like MultiDiffusion [BYL+23]), enabling object and cameras refinement across all views. Top: inputs; Middle: reconstructed 3D object and camera poses; Bottom: rendered images and geometry.

In the second stage, with the geometry fixed, we apply a Multi-Diffusion strategy [BYL+23] to the appearance latent $\boldsymbol{F}_o$. We maintain a single shared latent and update it using the averaged flow predictions from all $K$ views, conditioned on the coarse geometry $\boldsymbol{G}_o$ and poses $\{\boldsymbol{\theta}^k\}_{k=1}^K$ estimated in the first stage:

$$d_{\text{feat},t}^{(k)} = \boldsymbol{\mathcal{G}}_{\boldsymbol{L}}(\boldsymbol{F}_o^t, \boldsymbol{G}_o, \boldsymbol{\theta}^k, t, \boldsymbol{I}^{(k)}), \tag{8.9.13}$$

$$\boldsymbol{F}_o^{t-1} \leftarrow \text{Step}\left(\boldsymbol{F}_o^t, \frac{1}{K}\sum_{k=1}^{K} d_{\text{feat},t}^{(k)}\right). \tag{8.9.14}$$

This averaging operation ensures that the generated 3D representation satisfies the visual constraints imposed by all input views simultaneously, yielding the SfM-like consistency shown in Figure 8.34.

### 8.9.6 Conclusion

This section presented a unified framework for 3D generation and reconstruction, operationalizing the biological distinction between object identity (“what”) and spatial context (“where”). By jointly modeling the canonical 3D object $\mathcal{O}$ and the camera pose $\boldsymbol{\theta}$, our approach effectively disentangles intrinsic shape from extrinsic viewpoint.

Leveraging pose priors and aligned conditioning, the model performs Bayesian-like inference to resolve full 3D geometry from limited single-view inputs. This ensures geometric consistency and robustness to viewpoint variations. Ultimately, this structured representation advances the capabilities of Embodied AI, equipping agents with the necessary spatial understanding to navigate and interact with 3D environments effectively.

## 8.10 Conditioned Human Body Motion Generation

So far, we have shown, in preceding sections, how to learn the distributions of visual data including 2D images and 3D object shapes. Note that the primary goal for us to sense the 3D world and develop a model of it is for us to navigate, interact, and manipulate objects within. Based on information perceived about the environment, our brain makes plans and takes actions based on tasks at hand. The level of dexterity achievable by "eye-hand coordination" is arguably the highest form of physical intelligence in nature. Hence accurately estimating and controlling our body and hand movement is crucial for us to interact with objects within the environment. Our brain dedicates significant resources to this: the *motor cortex*, located in the brain's frontal lobe, records and coordinates our body movements, while proprioceptive signals from our body's nerve system provide continuous feedback about our limb positions. Therefore, if we want robots, especially humanoid-like robots, to emulate human actions, they need to learn (the distribution of) our body movements.

In this section, we show how the conditional inference framework from Chapter 7 can be applied to learn the distribution of human motion and then exploited it for the body pose estimation task. To this end, we consider a simple but natural setup: a person wears a head-mounted device such as an AR headset or smart glasses that captures forward-facing video and tracks its 3D pose via Structure from Motion (SFM) or Simultaneous Localization and Mapping (SLAM). Such methods estimate the device's position by tracking and triangulating visual features in the scene to recover 3D structure and pose.[23] Our goal is to estimate the wearer's full-body pose, height, and hand configuration from these egocentric observations, even though the body is rarely visible in the captured video. The key observation is that *head motion encodes rich information about body motion*. When we walk, our head bobs; when we reach, it tilts; when we sit, it follows a distinctive arc. These correlations, learned from motion capture data, form a conditional prior $p(\text{body} \mid \text{head})$. The method we feature here in this section is based on an open-sourced project, called **EgoAllo** (for egocentric-to-allocentric estimation):

https://egoallo.github.io ,

which incorporates additional measurements like hand detections and ground contact via guided sampling. Figure 8.35 illustrates the overall setup and task.

### 8.10.1 Representing Human Bodies

Before formulating the estimation problem, we need a mathematical representation for human bodies. The skeleton of our body is a kinematic tree: rigid bones connected by joints, with the pelvis as root and chains extending to the head, hands, and feet. Each joint has rotational degrees of freedom: the shoul-der rotates in three axes, the elbow primarily in one. For a more formal and detailed introduction to kinematic representation of linked rigid bodies like the human body, interested readers may refer to the classic textbook [MLS94].

[23] Interested readers may refer to the textbook [MKS+04] for an introduction to this line of work.

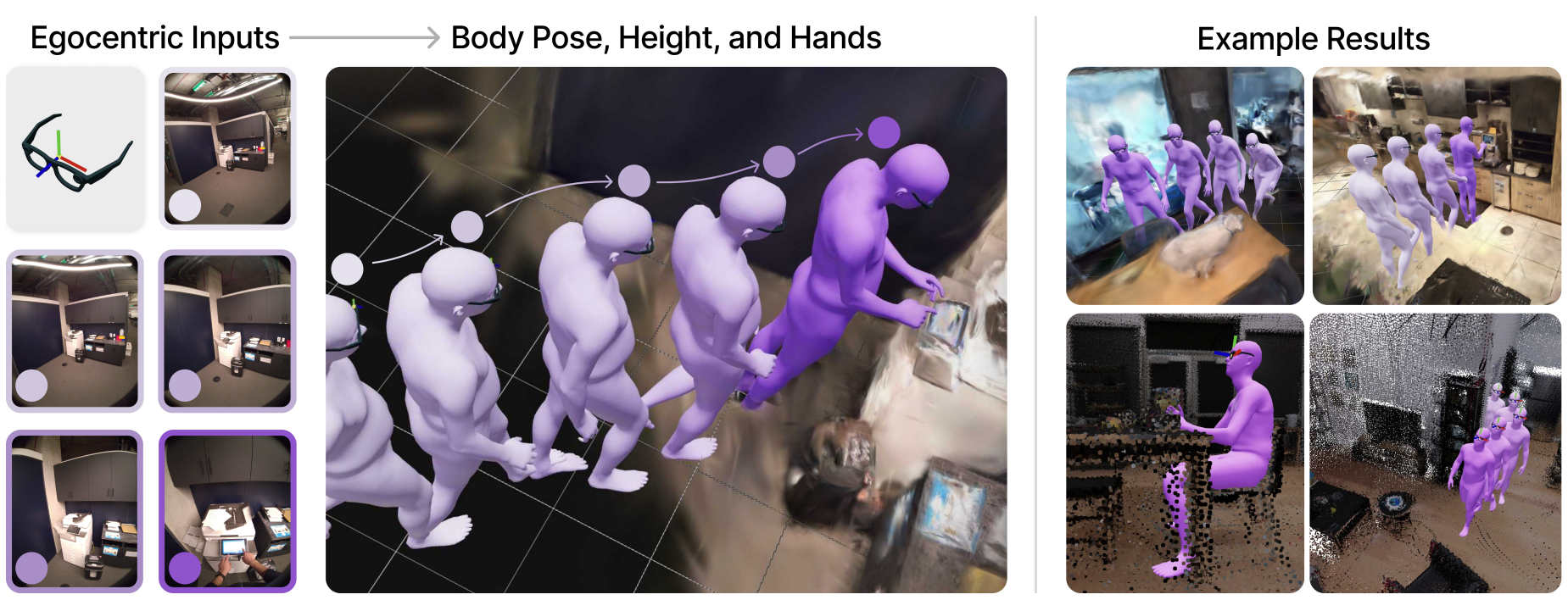

Figure 8.35: **Egocentric Human Motion Estimation.** Given head poses from SLAM and egocentric images from a head-mounted device (left), EgoAllo estimates full body pose, height, and hand parameters in the world (allocentric) reference frame (right). The estimated bodies are grounded in the scene, with feet contacting the floor and hands at physically plausible positions.

**The SMPL body model.** SMPL (Skinned Multi-Person Linear model) [LMR+15] provides a compact, differentiable representation of the human body, for both its (kinematic) pose and shape. It separates a body configuration into two components:

- **Pose parameters** $\theta$: Joint rotations specifying the body's articulation. SMPL uses 23 joints (plus the root), each parameterized by a 3D rotation. Together, these determine whether a person is standing, sitting, reaching, walking, etc.
- **Shape parameters** $\beta$: A low-dimensional vector, typically 10 coefficients, controlling body proportions such as height, weight, limb lengths, and torso width. These are derived from principal component analysis of a large corpus of 3D body scans.

Given pose $\theta$ and shape $\beta$, SMPL produces a 3D mesh via learned blend skinning. It also provides joint positions through forward kinematics: starting from the root (pelvis) position and applying each joint's rotation in sequence along the kinematic chain, we can compute where each joint ends up in 3D space. For example, to find the wrist position, we compose the rotations of the pelvis, spine, shoulder, and elbow, accumulating the translations defined by bone lengths.

**Adding hands: SMPL-H and MANO.** The original SMPL model represents hands as simple endpoints. For applications requiring detailed hand poses like grasping, gesturing, or typing, this is insufficient. SMPL-H [RTB17] extends SMPL with articulated hands based on the MANO hand model, adding 15 joints per hand (three per finger) for 52 total joints.

Hand rotations $\varphi$ follow the same parameterization as body joints. The complete state includes body rotations $\theta$, hand rotations $\varphi$, and shape parameters $\beta$.

**Why body models matter.** Why use SMPL-H rather than, say, directly regressing joint positions or mesh vertices? Parametric models offer key advantages. Outputs are always valid human bodies, with no anatomically impossible configurations. The model is differentiable, enabling gradient-based optimization. The parameterization is compact, requiring only a few hundred numbers versus a full mesh. And we get semantic structure, reasoning about "left wrist" rather than vertex indices.

### 8.10.2 Task and Setup

**Inputs.** The inputs consist of two streams from a head-mounted device:

- **Head poses $\boldsymbol{h}_{1:T} \in \mathsf{SE}(3)^T$**[24]: Position and orientation of the device in world coordinates at each timestep, from visual-inertial SLAM. Here $\mathsf{SE}(3)$ denotes the group of rigid transformations in 3D, comprising rotations and translations.[25] These poses have metric scale, meaning we know actual distances in meters rather than just relative geometry, and are gravity-aligned so we know which direction is up.
- **Egocentric images $I_{1:T}$**: Forward-facing video. The wearer's body is mostly not visible, but hands frequently appear when reaching or manipulating objects.

From images, an off-the-shelf hand estimator like HaMeR [PSR+24] produces per-frame measurements when hands are visible:

$$\boldsymbol{y}^{\text{hand},t} = \left(\hat{\varphi}^{\text{MANO},t},\, \hat{\boldsymbol{q}}^{3D,t},\, \hat{\boldsymbol{q}}^{2D,t}\right), \tag{8.10.1}$$

consisting of estimated hand rotations, 3D keypoints in the camera frame, and 2D keypoints.

**Outputs.** EgoAllo estimates a motion sequence over $T$ timesteps:

$$\boldsymbol{x}_0 = \{\boldsymbol{x}_0^1, \ldots, \boldsymbol{x}_0^T\}, \quad \text{where} \quad \boldsymbol{x}_0^t = \left(\theta^t,\, \varphi^t,\, \beta,\, s\right). \tag{8.10.2}$$

Here $\theta^t$ are body joint rotations, $\varphi^t$ hand rotations, $\beta$ shape parameters shared across time, and $s$ a height parameter that is also shared. The subscript 0 denotes clean motion, following diffusion notation where $\boldsymbol{x}_n$ is the noisy version at step $n$.

[24] Here $\mathsf{SE}(3)$ represents the special Euclidean group in $\mathbb{R}^3$, essentially the group of transformations that consist of rotation and translation.

[25] Interested readers may refer to [MKS+04] for a formal introduction to rigid body motion and its representation and properties.

EgoAllo estimates height in addition to pose, which is essential for grounding. Without it, the body floats at an arbitrary vertical position; with it, feet contact the floor and the head aligns with the camera.

**Posterior formulation.** Following the Bayesian framework from Chapter 7, estimation is formulated as sampling from a posterior. Let $\boldsymbol{y}$ denote all observations and $\boldsymbol{c} = g(\boldsymbol{h}_{1:T})$ be a processed representation of head poses (to be explained soon in Section 8.10.3). The posterior factorizes as:[26]

$$p(\boldsymbol{x}_0 \mid \boldsymbol{y}) \;\propto\; p_\theta(\boldsymbol{x}_0 \mid \boldsymbol{c}) \cdot p(\boldsymbol{y}^{\text{hand}}, \boldsymbol{y}^{\text{phys}} \mid \boldsymbol{x}_0), \tag{8.10.3}$$

where:

- $p_\theta(\boldsymbol{x}_0 \mid \boldsymbol{c})$ is a learned conditional prior, specifically a diffusion-denoising model that generates plausible body motion given head motion.
- $p(\boldsymbol{y}^{\text{hand}}, \boldsymbol{y}^{\text{phys}} \mid \boldsymbol{x}_0)$ is the likelihood of hand observations and physical constraints such as ground contact and no foot skating, given the body motion.

The likelihood is expressed via a guidance energy $\mathcal{L}_{\text{guide}}$:

$$p(\boldsymbol{y}^{\text{hand}}, \boldsymbol{y}^{\text{phys}} \mid \boldsymbol{x}_0) \propto \exp\big(-\mathcal{L}_{\text{guide}}(\boldsymbol{x}_0)\big), \tag{8.10.4}$$

so the posterior becomes:

$$p(\boldsymbol{x}_0 \mid \boldsymbol{y}) \;\propto\; p_\theta(\boldsymbol{x}_0 \mid \boldsymbol{c}) \exp\big(-\mathcal{L}_{\text{guide}}(\boldsymbol{x}_0)\big). \tag{8.10.5}$$

This is the form from Chapter 7: a learned prior combined with measurement constraints via an energy-based likelihood.

### 8.10.3 Designing the Conditioning Representation

The conditional prior $p_\theta(\boldsymbol{x}_0 \mid \boldsymbol{c})$ generates body motion given a representation $\boldsymbol{c}$ of head motion. What should this representation look like? As emphasized throughout this book, representation choice fundamentally affects learning and generalization.

A naive approach would condition directly on the raw head poses $\boldsymbol{h}_{1:T} \in \mathsf{SE}(3)^T$, the absolute position and orientation at each timestep. This works poorly. To understand why, consider what properties the representation should satisfy.

[26] For this factorization to be valid, we essentially assume that given $\boldsymbol{x}_0$, $\boldsymbol{y}^{\text{hand}}, \boldsymbol{y}^{\text{phys}}$ is conditionally independent of the head pose $\boldsymbol{c}$.

**Spatial Invariance.** The mapping from head motion to body motion should not depend on where in the world the motion occurs. Walking in the kitchen should produce the same body pose estimates as walking in the living room. Only the local motion pattern matters, not absolute position.

Conditioning on absolute poses violates this. As Figure 8.36 illustrates, identical body motion corresponds to different absolute head trajectories depending on world location. A model conditioned on absolute poses must learn separately that "head at $(3, 2, 1)$ moving forward" and "head at $(10, 5, 2)$ moving forward" both correspond to walking—an unnecessary burden that harms generalization.

**Temporal Invariance.** The mapping should also not depend on when in an observation window the motion occurs. Processing frames 0–100 should behave the same as processing frames 50–150 if they contain the same underlying motion. This rules out cumulative quantities like total displacement from sequence start, as well as absolute time indices.

One might try canonicalizing to the first timestep by placing the first frame at the origin. But this violates temporal invariance: different temporal slices of the same motion, canonicalized to their respective first frames, produce different trajectories. The model sees different inputs for the same underlying body motion.

**The Solution: Per-timestep Canonicalization.** To achieve both spatial and temporal invariance, EgoAllo defines a canonical coordinate frame at *every* timestep rather than just at the start. The representation $\boldsymbol{c} = \{\boldsymbol{c}^1, \ldots, \boldsymbol{c}^T\}$ couples:

- Relative head motion $\Delta \boldsymbol{T}^t_{\text{head}}$ between consecutive frames, capturing how the head moved from $t-1$ to $t$.

- Per-timestep canonicalized pose $\boldsymbol{T}^t_{\text{canonical}}$: the head's orientation relative to gravity and height above floor, with floor-plane position and yaw removed.

These quantities are invariant to where the motion occurs in the world and to which temporal window we extract. Two sequences with identical local motion produce identical conditioning tokens $\boldsymbol{c}$, regardless of absolute world position or time offset.

## 8.10.4 The Conditional Prior

The conditional prior $p_\theta(\boldsymbol{x}_0 \mid \boldsymbol{c})$ is a diffusion model that generates body motion sequences given head motion conditioning. The architecture follows the transformer-based design discussed in Section 8.4, adapted for sequential motion data.

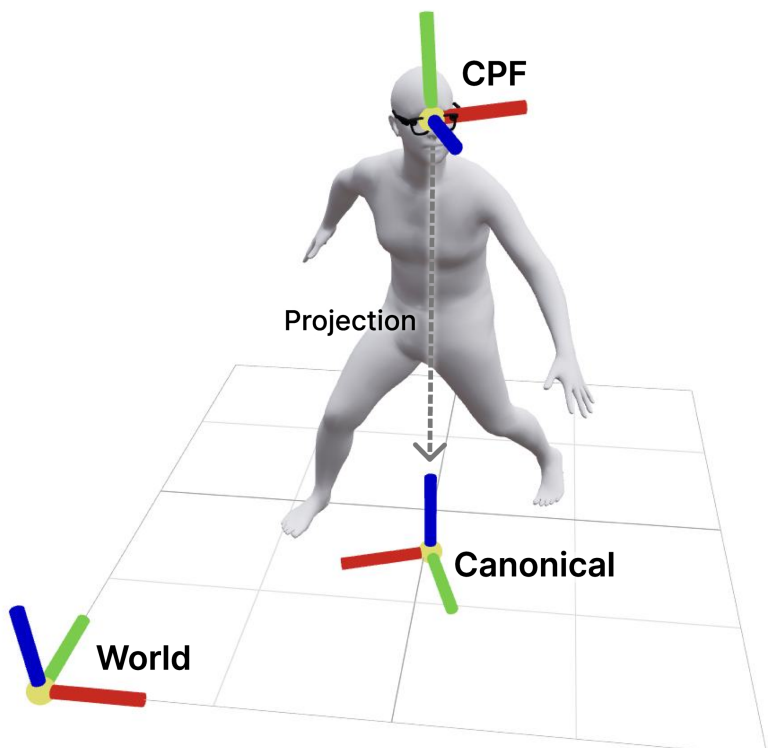

Figure 8.36: **Invariant Head Motion Conditioning.** The conditioning representation must be invariant to both spatial transformations (same motion at different world locations) and temporal shifts (same motion at different absolute times). Per-timestep canonicalization achieves both properties.

**Architecture.** The model takes three inputs:

- The noisy motion sequence $\boldsymbol{x}_n$ at diffusion step $n$, tokenized per-timestep.
- The diffusion timestep embedding, indicating the noise level.
- The invariant head conditioning tokens $\boldsymbol{c}$, one per timestep.

The head tokens are first processed by an encoder consisting of six transformer blocks with self-attention to produce contextualized representations. These condition a decoder, another six transformer blocks, that denoises the motion sequence. Conditioning is injected via cross-attention: in each decoder layer, the motion tokens form queries that attend to the head tokens as keys and values, allowing the model to dynamically select which aspects of head motion are relevant for predicting each body joint at each timestep. The model outputs a direct estimate of the clean motion $\boldsymbol{x}_0$, rather than predicting noise, parameterized as joint rotations in a head-relative coordinate frame.

**Training Data.** The model is trained on AMASS [MGT+19], a large-scale motion capture dataset aggregating multiple sources into unified SMPL-H format. AMASS contains diverse human motions—walking, running, dancing, sitting, reaching, exercising—but no head-mounted device recordings. Head trajectories are simulated by extracting head joint position and orientation from the motion capture data, as if a device were rigidly attached to the head.

**Training Objective.** Training uses the denoising score-matching loss from Chapter 3:

$$\mathcal{L}_{\text{diff}} = \mathbb{E}_{\boldsymbol{x}_0, \boldsymbol{n}} \left[ w_n \left\| D_\theta(\boldsymbol{x}_n, \boldsymbol{n}, \boldsymbol{c}) - \boldsymbol{x}_0 \right\|^2 \right], \tag{8.10.6}$$

where $D_\theta$ is the denoiser network, $w_n$ are noise-dependent weights, and $\boldsymbol{x}_n$ is obtained by adding noise to clean motion $\boldsymbol{x}_0$ according to the forward diffusion process. The model learns to predict clean motion from noisy motion, conditioned on head trajectories.

### 8.10.5 Incorporating Measurements via Guidance

The conditional prior captures the relationship between head motion and body motion, but does not use image observations or physical constraints. To incorporate these, EgoAllo defines a guidance energy $\mathcal{L}_{\text{guide}}$ that measures consistency between a motion hypothesis and the observations.

**Guidance Energy.** The total guidance energy combines several terms:

$$\mathcal{L}_{\text{guide}}(\boldsymbol{x}_0) = \lambda_{3D}\mathcal{L}_{3D} + \lambda_{2D}\mathcal{L}_{2D} + \lambda_{\text{skate}}\mathcal{L}_{\text{skate}} + \lambda_{\text{prior}}\mathcal{L}_{\text{prior}}, \tag{8.10.7}$$

where:

- $\mathcal{L}_{3D}$: Alignment between estimated hand keypoints and 3D hand detections from HaMeR, in the world frame.
- $\mathcal{L}_{2D}$: Reprojection error. Estimated hand keypoints, projected into the image, should match 2D detections.
- $\mathcal{L}_{\text{skate}}$: Foot skating penalty. Feet should not slide when in contact with ground.
- $\mathcal{L}_{\text{prior}}$: Weak regularization toward the diffusion prior's prediction, preventing guidance from pushing too far from plausible motions.

Following the energy-based interpretation from Chapter 7, this guidance energy corresponds to a negative log-likelihood: $p(\boldsymbol{y} \mid \boldsymbol{x}_0) \propto \exp(-\mathcal{L}_{\text{guide}}(\boldsymbol{x}_0))$. Low energy means consistency with observations; high energy means inconsistency.

**Why Reprojection Matters.** Single-frame hand estimators like HaMeR produce 3D hand poses from monocular images, but suffer from scale and depth ambiguity. A small hand close to the camera looks the same as a large hand far away. The 3D positions are therefore unreliable in absolute terms, even when local hand pose such as finger articulation is accurate.

The 2D reprojection loss addresses this by supervising in image space, where the hand was actually observed. Estimated 3D hand positions are projected back into the image and compared against detected 2D keypoints. This sidesteps depth ambiguity while still providing useful signal about hand location.

### 8.10.6 Conditional Sampling Algorithm

The measurement matching framework from Chapter 7 tells us how to sample from the posterior: modify the denoiser to include a gradient correction toward measurement consistency. Recall that the conditional denoiser decomposes as:

$$\mathbb{E}[\boldsymbol{x}_0 \mid \boldsymbol{x}_n, \boldsymbol{y}] = \mathbb{E}[\boldsymbol{x}_0 \mid \boldsymbol{x}_n] + \sigma_n^2 \nabla_{\boldsymbol{x}_n} \log p(\boldsymbol{y} \mid \boldsymbol{x}_n), \tag{8.10.8}$$

where the first term is the prior denoiser and the second is the measurement matching correction.

In our setting, the prior denoiser is conditioned on head motion: $D_n^{\text{head}}(\cdot; \boldsymbol{c})$. The measurement correction uses the guidance energy. The full conditional denoiser becomes:

$$D_n^{\text{guided}}(\boldsymbol{z}) = D_n^{\text{head}}(\boldsymbol{z}; \boldsymbol{c}) - \sigma_n^2 \nabla_{\boldsymbol{z}} \mathcal{L}_{\text{guide}}(\boldsymbol{x}_0(\boldsymbol{z})), \tag{8.10.9}$$

using the relationship $\nabla \log p(\boldsymbol{y} \mid \boldsymbol{x}) = -\nabla \mathcal{L}_{\text{guide}}(\boldsymbol{x})$.

**Alternating Optimization.** In practice, rather than computing gradients through the full guidance energy at each step, EgoAllo alternates between two operations:

1. **Prior denoising step.** Apply a DDIM-style update using the head-conditioned denoiser $D_n^{\text{head}}(\cdot; \boldsymbol{c})$ to obtain a proposal $\tilde{\boldsymbol{x}}_{n-1}$. DDIM (Denoising Diffusion Implicit Models) provides a deterministic sampling procedure that maps noisy samples toward cleaner ones; unlike stochastic samplers, each denoising step is a deterministic function of the current state, which allows for faster sampling with fewer steps.

2. **Measurement-matching step.** Refine the proposal by solving a proximal optimization problem:

$$\boldsymbol{x}_{n-1} = \arg\min_{\boldsymbol{x}} \left\{ \frac{1}{2} \|\boldsymbol{x} - \tilde{\boldsymbol{x}}_{n-1}\|^2 + \gamma_n \, \mathcal{L}_{\text{guide}}(\boldsymbol{x}) \right\}, \tag{8.10.10}$$

   where $\gamma_n$ is a step-size parameter. This finds motion close to the prior's prediction but more consistent with observations.

The proximal optimization is solved using Levenberg-Marquardt, a classic algorithm for nonlinear least-squares problems. It interpolates between gradient descent and Gauss-Newton optimization, adapting its step size based on how well the linearization approximates the true objective. This is well-suited to SMPL-H parameter estimation, where the objective involves differentiable forward kinematics. A few iterations suffice at each denoising step.

**Global Alignment.** The diffusion model operates in a head-relative coordinate frame, predicting body pose relative to the head rather than in absolute world coordinates. After the final denoising step, EgoAllo places the body in the world by composing with the observed SLAM head poses:

$$\boldsymbol{T}^{t}_{\text{world,root}} = \boldsymbol{T}^{t}_{\text{world,head}} \cdot \boldsymbol{T}^{(\theta^t,\beta)}_{\text{head,root}}, \tag{8.10.11}$$

where $\boldsymbol{T}_{\text{head,root}}$ is computed via forward kinematics from the estimated pose and shape. This ensures the body is correctly grounded: feet on the floor, head aligned with the camera.

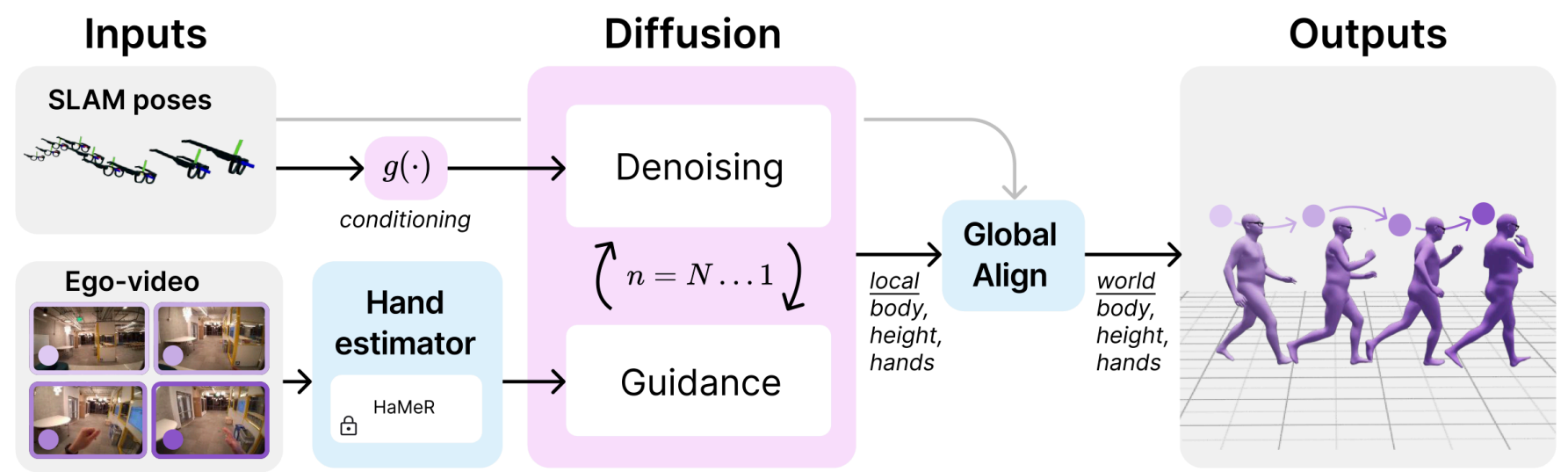

Figure 8.37: **Overview of components of the EgoAllo framework.** The diffusion model is restricted to local body parameters. An invariant parameterization g( · ) of SLAM (head) poses is used to condition a diffusion model. These can be placed into the global coordinate frame via global alignment to input poses. When available, egocentric video is used for hand detection, say via HaMeR [PSR+24], which can be incorporated into samples via guidance.

**Handling Long Sequences.** The diffusion model is trained on fixed-length sequences (32–128 frames). For longer recordings, EgoAllo uses the MultiDiffusion approach [BYL+23]: the sequence is divided into overlapping windows and denoised in parallel. After each denoising step, the results in overlapping regions are blended by averaging before proceeding to the next step. This per-step blending ensures that windows remain consistent with each other throughout the diffusion process, producing temporally coherent output without discontinuities at window boundaries.

### 8.10.7 Insights from Experiments

EgoAllo is evaluated on three datasets: AMASS [MGT+19] (motion capture with synthetic head trajectories), RICH [HYH+22] (motion capture with challenging human-scene interactions), and Aria Digital Twin [PCY+23] (real recordings from Project Aria glasses with ground-truth body poses). For hand estimation, EgoExo4D [GWT+23] is additionally used.

**Representation Matters.** Table 8.15 shows the effect of different conditioning representations on body estimation accuracy, comparing five parameterizations that vary in their invariance properties:

Table 8.15: **Effect of conditioning representation on AMASS.** Different parameterizations of head motion conditioning lead to substantially different estimation accuracy. The per-timestep invariant representation, which satisfies both spatial and temporal invariance, performs best. MPJPE is mean per-joint position error in millimeters.

| | **Seq Length 32** | | **Seq Length 128** | |
|---|---|---|---|---|
| **Conditioning** | MPJPE | Effect | MPJPE | Effect |
| Per-timestep invariant | **129.8** | — | **119.7** | — |
| Absolute + relative | 133.0 | +2.4% | 124.5 | +4.0% |
| Absolute + global deltas | 136.2 | +4.9% | 127.4 | +6.4% |
| Sequence canonicalization | 153.1 | +17.9% | 134.0 | +11.9% |
| Absolute | 159.9 | +23.2% | 148.3 | +23.9% |

- **Per-timestep invariant**: The representation described in Section 8.10.3, with both spatial and temporal invariance.
- **Absolute + relative**: Absolute poses augmented with local-frame relative motion.
- **Absolute + global deltas**: Absolute poses with world-frame position deltas.
- **Sequence canonicalization**: Alignment to first frame, providing spatial but not temporal invariance.
- **Absolute**: Raw $\mathsf{SE}(3)$ poses with no invariance properties.

The results show that conditioning representation has a dramatic impact on accuracy, with up to 24% difference in MPJPE between the best and worst choices. This reinforces a central theme of this book: the choice of representation fundamentally affects learning and generalization.

**Height Estimation Enables Grounding.** Table 8.16 compares full estimation (including height) against a variant without height estimation. Estimating height improves MPJPE by 6–7% and is essential for grounding: without it, estimated bodies often "float" above the ground because the system cannot determine the correct vertical position. The grounding metric (GND) measures whether estimated feet ever contact the floor, and height estimation substantially improves this.

**Body Context Improves Hand Estimation.** A notable finding is the symbiotic relationship between body and hand estimation. Single-frame monocular hand estimators like HaMeR produce accurate local hand poses, including finger articulation, but cannot determine the hand's absolute 3D position due to scale and depth ambiguity. By estimating the full body first, we gain kinematic constraints: the hand is attached to the wrist, which connects to the forearm, elbow, upper arm, and torso. This chain provides context for global hand localization. Figure 8.38 shows some qualitative results.

Table 8.16: **Body estimation results across datasets.** MPJPE and PA-MPJPE in millimeters; GND is the fraction of sequences where feet contact the ground. Height estimation is essential for proper grounding.

| Dataset | Variant | MPJPE↓ | PA-MPJPE↓ | GND↑ |
|---|---|---|---|---|
| AMASS | Full (with height) | **119.7** | **101.1** | **1.00** |
| | Without height | 128.1 | 110.3 | 0.98 |
| RICH | Full (with height) | **176.2** | **160.1** | **0.96** |
| | Without height | 185.7 | 169.9 | 0.82 |
| ADT | Full (with height) | **155.1** | **129.3** | 0.94 |
| | Without height | 163.7 | 140.0 | **0.96** |

Table 8.17: **Hand estimation on EgoExo4D.** MPJPE measures world-frame accuracy; PA-MPJPE measures local pose accuracy. Body context dramatically improves world-frame localization while preserving local pose quality.

| Method | MPJPE↓ | PA-MPJPE↓ |
|---|---|---|
| HaMeR (monocular, no body) | 237.9 | **13.0** |
| With body (2D reprojection) | 131.5 | 14.7 |
| With body (3D wrist from stereo) | **60.1** | 14.4 |

Table 8.17 quantifies this effect on EgoExo4D. Raw HaMeR estimates have 237.9mm world-frame error. Adding body context reduces this to 131.5mm, a 45% improvement, even using only monocular hand observations. When additional 3D wrist observations from stereo cameras are available, errors drop further to 60.1mm.

The relationship is bidirectional: hand observations also improve body estimation. When hand guidance is incorporated during sampling, body MPJPE on AMASS improves from 119.7mm to 78.8mm, a 34% reduction. Additional measurement constraints tighten the posterior and produce more accurate estimates throughout the kinematic chain.

**Qualitative Results.** Figure 8.39 shows estimated body poses from real-world recordings with Project Aria glasses. The estimated bodies are correctly grounded: feet contact the floor appropriately, body height matches the environment, and poses are consistent with observed head motion across diverse activities including walking, sitting, and object manipulation.

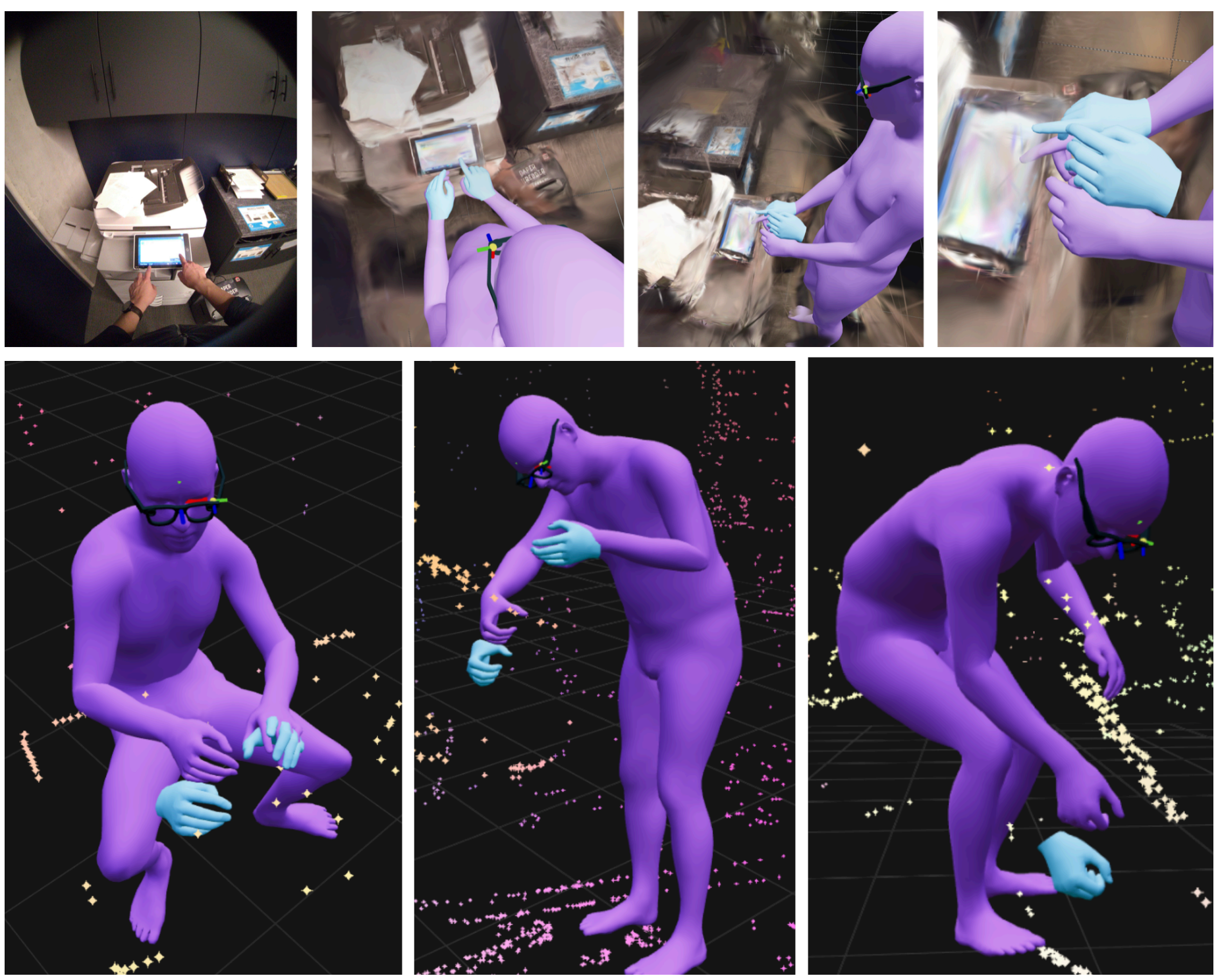

Figure 8.38: **Body Context Improves Hand Estimation.** Blue: monocular hand estimates from HaMeR, which have accurate local pose but incorrect world-frame position due to scale/depth ambiguity. Purple: estimates with body context, correctly grounded in the world frame.

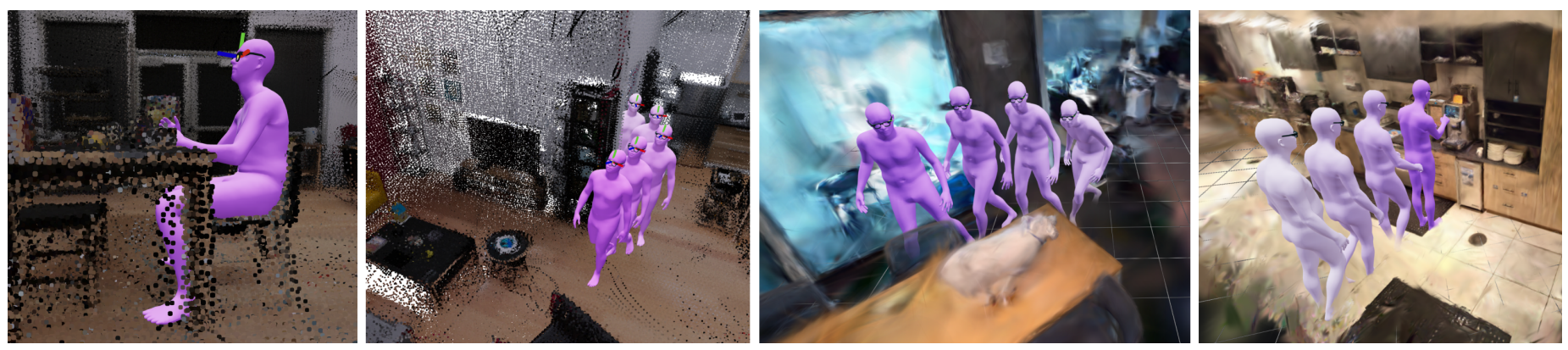

Figure 8.39: **Qualitative Results.** Estimated body poses from real-world egocentric recordings, visualized in 3D scene reconstructions. The method produces physically plausible poses with appropriate grounding across diverse activities.

### 8.10.8 Discussion

This application demonstrates how the conditional inference framework from Chapter 7 extends beyond images to structured, articulated outputs like human motion.

**Representation Design Is Crucial.** A recurring theme throughout this book is that representation choices fundamentally affect learning and generalization. This principle is particularly evident here. By analyzing what invariances the conditioning representation should satisfy, we derived a parameterization that substantially outperforms alternatives. Spatial invariance ensures independence from world location; temporal invariance ensures independence from time offset. The lesson: before training a model, carefully consider what structure the representation should encode.

**The Prior-plus-guidance Pattern Generalizes.** The sampling procedure directly applies the measurement matching framework: a learned prior, here the head-conditioned diffusion model, is combined with measurement constraints like hand observations and physics via guided sampling. This same pattern of prior plus guidance appears repeatedly across domains, from image generation to motion estimation. The components are modular: we can swap in different priors or different guidance terms without changing the overall framework.

**Holistic Estimation Outperforms Modular Pipelines.** The symbiotic relationship between body and hand estimation suggests that joint inference over the full kinematic chain outperforms estimating components independently. Head motion informs body pose; body pose provides context for hand position; hand observations refine both body and hand estimates. This mutual benefit emerges naturally from the posterior formulation, where all variables are coupled through the likelihood and prior.

**Limitations.** Several limitations remain. First, the method requires accurate SLAM with metric scale and gravity alignment; errors in head pose estimation propagate directly to body estimates. Second, generalization is limited by training data diversity, so unusual poses or activities not represented in AMASS may be estimated poorly. Third, hands can only be refined when visible in the egocentric view; occluded hands rely entirely on the body prior. Finally, the method assumes flat floors; staircases or uneven terrain would require additional modeling.

**Open Problems.** Promising directions include incorporating scene geometry for better physical grounding, where hands should contact surfaces and bodies should avoid obstacles. Other directions include enabling real-time inference for live AR/VR applications and extending to multi-person scenarios where the wearer interacts with others. The framework is flexible enough to accommodate these extensions through additional guidance terms or modified priors.

## 8.11 Natural Language Representation and Generation

So far, we have shown how to learn deep structured representations for the distributions of data associated with visual perception and body motions. To a large extent, the so learned structured representations can be viewed as memories stored in the visual cortex and motor cortex of the brain. These memories together can be loosely referred to as a "model of the world", or simply a "world model" – enable us to both perceive the world and take actions. Notice that both human and animal have the ability to learn and develop such memories for the world and their body, which record what can be predicted from the perceived as well as motor skills crucial for their survival within the environment.

On top of such a world model for the physical environment and body motion, human has developed (spoken and written) languages so that we are able to communicate knowledge learned about the world to others – the Broca's area and Wernicke's area in our brain are known for speech processing and language comprehension, respectively. To a large extent, languages record a small, but important, fraction of our world models that are worth sharing among ourselves. Hence, natural languages, as a very special type of data, must have very special structures that encode very rich information. If machines were able to learn the structures of natural languages, they may be able to communicate better with humans.

Therefore, in this section, we show how to apply methods introduced in this book to the task of *language modeling*. Or more specifically, how to train large language models (LLMs) to learn the distribution of natural languages and then generate them. The task and its setup is very much the same as that used to train GPT-2 and many other language models.

### 8.11.1 Data

The data we will use to investigate the performance of CRATE for language tasks will be OpenWebText (OWT) [GC19], an open-source reproduction of the unreleased WebText dataset used by OpenAI to train GPT2. Each sample in OWT is a web document, typically sourced from high-quality web pages, blogs, articles, or online discussions, that is written in well-formed natural language. The OpenWebText dataset contains around 8.01M documents of varying lengths, totaling around 41.70GB of text. For evaluation, we will use several datasets, such as WikiText [MXB+16][27], LAMBADA [PKL+16][28], and PTB [MSM93]. PTB and OWT are generally easier compared to other datasets. PTB focuses on simpler journalistic text, ideal for traditional language modeling, while OWT is diverse and informal, covering various topics but with less complexity in language structure or long-range dependencies. WikiText, with its formal structure and domain-specific content, requires a more complex understanding than OWT but remains manageable. LAMBADA is the most challenging, as it involves long-range dependencies, requiring the model to grasp broader contextual information to complete sentences accurately.

[27] For WikiText2 and WikiText103 [MXB+16], the test splits are the same, so we merge them as a single dataset referred to as WikiText.

[28] To obtain the accuracy on the LAMBADA dataset, we use greedy decoding.

On a more formal level, our data $\boldsymbol{X}$ will be text, or strings of characters; we let $\mathcal{T}$ be the set of all strings.

### 8.11.2 Task and Objective

For causal language modeling pre-training, the idea is that we want to *train the model to output human-like text.* The most popular way to do this by far is to use a two-stage training process:[29]

- *First*, we wish to *learn* a way to optimally encode documents as a sequence of basic ("building block") strings, called *tokens.* This is called *tokenization*, and we build a *tokenizer.*

- *Second*, we wish to *learn* a way to *predict the distribution of a token given all previous tokens.* This is called *next-token prediction*, and we build a *language model.*

This procedure actually dates back to Markov, who first noticed that natural language could be modeled by the eponymous Markov chain structure [Mar06] given an appropriate tokenization, and then to Shannon, who proposed doing this exact language modeling setup with a character-level tokenizer (i.e., each character is a token) and so-called "$n$-gram" (i.e., an explicit look-up table, calculated from training data, for the distribution of a token given the $n$ previous tokens) in place of the language model [Sha48].[30]

#### Training a Tokenizer

To build a tokenizer amounts to building a vocabulary $\mathcal{V}$, which is a set of tokens and has some pre-specified size $V$. There are several methods to do this. One popular algorithm is known as Byte Pair Encoding (BPE), which can be described as:

- Start with a list of all unique characters in your training data, and their frequencies. Ensure that there are fewer than $V$ such characters, and add each character as a separate string ("token") to the vocabulary along with its frequency.

- Until there are $V$ tokens in the vocabulary:

[29] Modern language model training has several additional training steps which demand different data distributions and algorithm approaches. However, training a model to merely mimic human writing only requires these few presented steps.

[30] A recent study [LMZ+24] scaling up $n$-gram models has shown that they are able to model text reasonably well for large $n$, but of course the memory required to store such a lookup table is of order $V^n$ and hence completely intractable.

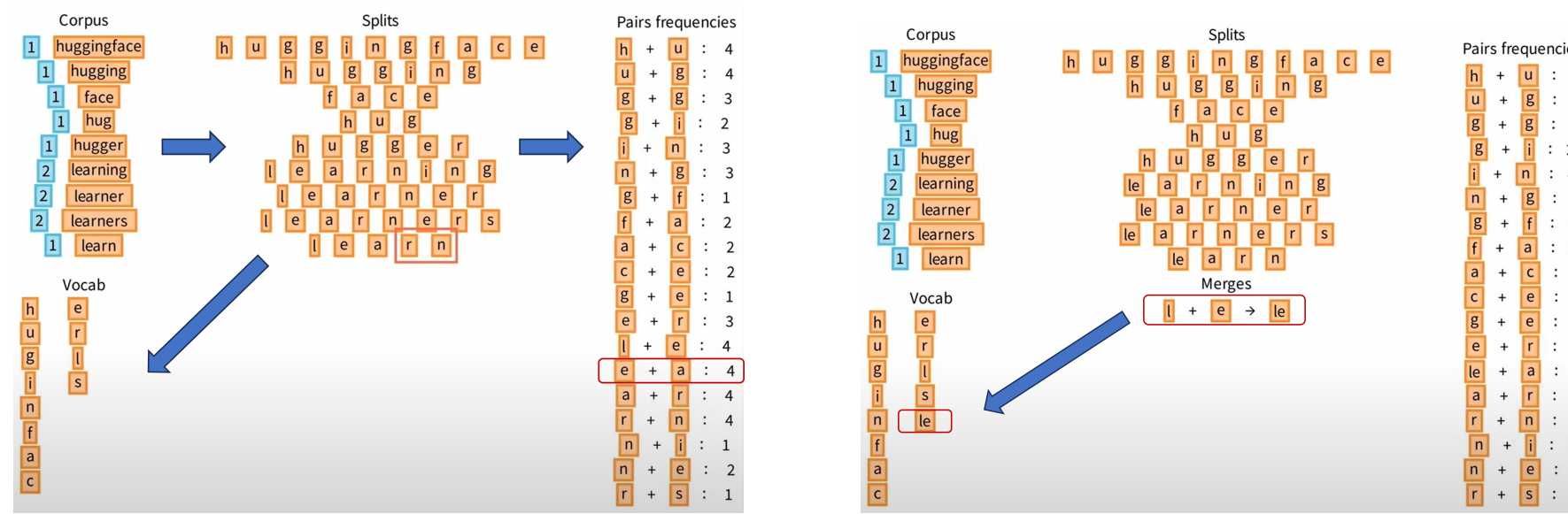

Figure 8.40: **The process of tokenizing text data using BPE.** (Image credit to https://huggingface.co/learn/nlp-course/chapter6/5). (Left) We begin by analyzing the given text corpus and constructing an initial vocabulary that consists of individual characters (or bytes in the case of byte-level BPE). Then, we compute the frequencies of adjacent character pairs in the corpus. This involves scanning the entire text and counting how often each two-character sequence (bigram) appears. (Right) After computing the frequencies of adjacent character pairs, we identify the most frequent pair in the corpus. This pair is then merged into a new subword unit, which is added to the vocabulary as a single token. This process is repeated iteratively until the predefined vocabulary size is reached.

  – Construct a token by taking the two most frequent existing tokens and merging them.
  – Compute this token's frequency in the dataset.
  – Add it to the vocabulary (along with its frequency).

- At this point, the frequency information is no longer needed and can be discarded.

The overall process of BPE is in Figure 8.40. Note that this procedure is a modification of a classical information-theoretic compression procedure for *learning a lossless encoding* of bytestream data (such as text), and as such, one can interpret it as finding an optimal lossless compression of the data. Notice that this is possible because (unlike images), the data here are fundamentally discrete and noise-free. After such a vocabulary is built, a tokenizer can break down a document into tokens (i.e., "tokenize" it). BPE uses a similar procedure to tokenize data as in training:

- Separate the document into a long list of one-character-long tokens. That is, if the document is "Hello" then the initial list is 'H', 'e', 'l', 'l', 'o'.
- While any two adjacent tokens can be concatenated and their concatenation is another token, we do it, i.e., we replace this pair of tokens with the merged token. Namely, if 'He' is a token in the vocabulary, 'H', 'e', 'l', 'l', 'o' would become 'He', 'l', 'l', 'o'.
- Repeat the above process until no more merges can be done. At this point, the document is partitioned into the final list (sequence) of tokens.

There are many practical and efficiency-based considerations to take into account during tokenization. The above algorithm, as presented, is *very far from optimal* if naively implemented, for instance. We do not cover this topic in great detail; there are many resources online to learn more, such as HuggingFace tutorials.

For instance, each token has a corresponding *index* which is just its index in the vocabulary (which after all is just a list of length $V$). Thus, the output of most tokenizers is a list of indices, say an element of $[V]^*$. Keep in mind that they correspond to substrings of the original document, as shown above.

Once a tokenizer is learned, it can be used as a black box by any language model. For instance, many models have the same (OpenAI-based) tokenizer based on the `tiktoken` library. In the remainder of this section, we will use such a fixed and pre-built tokenizer for everything, and thus identify each text document $\boldsymbol{X} \in \mathcal{T}$ with its tokenized version in $[V]^*$. Therefore, we may as well consider the text space $\mathcal{T}$ as *equal* to the space of token sequences $[V]^*$ (and lose nothing essential).

### Training a Language Model

Once we have each document as a sequence of tokens $\boldsymbol{X} \in [V]^N \subseteq [V]^* = \mathcal{T}$, we wish to perform next-token prediction. That is, given a *context* $\boldsymbol{X}_{:n} \in [V]^n$ (i.e., the first $n$ tokens $\boldsymbol{x}_1, \dots, \boldsymbol{x}_n \in [V]$ in the document)[31], we wish to predict the token $\boldsymbol{x}_{n+1} \in [V]$ at position $n+1$. To do this, we compute the aggregate feature of $\boldsymbol{X}_{:n}$ via $\boldsymbol{z}_\theta(\boldsymbol{X}_{:n}) \doteq (f_\theta^{\mathrm{ext}} \circ f_\theta)(\boldsymbol{X}_{:n}) \in \mathbb{R}^d$, and use a classification head $h_\theta \colon \mathbb{R}^d \to \Delta_V$ (implemented as either a linear layer, MLP, or something slightly more complicated) to project this feature into the $V$-dimensional probability simplex $\Delta_V$. This projection $\boldsymbol{p}_\theta(\boldsymbol{X}_{:n}) \doteq h_\theta(\boldsymbol{z}_\theta(\boldsymbol{X}_{:n}))$ serves as an estimated probability distribution of the next token. Then, using the notation $\mathbf{1}(\boldsymbol{x}_{n+1}) \in \Delta_V$ to be 1 in the $\boldsymbol{x}_{n+1}$th component and 0 elsewhere, the causal language modeling loss is

$$\min_\theta \left\{ \mathcal{L}_{\mathrm{CLM}}(\theta) \doteq \mathbb{E}_{\boldsymbol{X}} \left[ \frac{1}{N-1} \sum_{n=1}^{N-1} \mathrm{CE}(\mathbf{1}(\boldsymbol{x}_{n+1}), \boldsymbol{p}_\theta(\boldsymbol{X}_{:n})) \right] \right\} \tag{8.11.1}$$

Note how similar this is to a classification loss (say for images); one uses the cross-entropy and tries to align a predicted probability vector with the ground truth. The major difference between these two losses is that in this one, we compute the loss on a whole sequence, where each prediction is correlated with the others (unlike in the i.i.d. classification case).

Optimizing this loss is usually called "pre-training" in the language model community (contrast with "post-training" and, more recently, "mid-training", which are methodologies to modify a next-token-predictor for useful tasks).

*Side note:* Why does the first term of (8.11.1) predict $\mathbf{1}(\boldsymbol{x}_2)$, and there is no term which measures the loss to predict the first token? It's because if we wanted to predict the first token, we would have the *empty sequence* as context, and therefore make this first token prediction using a qualitatively different mechanism than that which applies to the other tokens. So actually this model is not trained to predict the *very first* token of any document. The reason this is OK is due to an implementation detail of the tokenizer: often, after building the tokenizer, we insert a *special* token into its vocabulary, called the *beginning-of-string (or document)* token and labeled `<|bos|>`.[32] Then, while processing each document, we add the `<|bos|>` token at the beginning of the document's token sequence, increasing the length of the tokenized sequence by 1. Thus the above causal language modeling objective has a term which involves trying to predict the first token of the document given only the `<|bos|>` token as context, and so it is a conceptually correct loss.

[31] Note the incongruity with Python notation: here the notation *includes* index $n$.

### 8.11.3 Architecture: Causal CRATE

For the architecture, we use a standard GPT-2-style transformer, substituting CRATE layers for the transformer layers.[33] For completeness, we specify the architecture here.

**Embedding.** We first embed the token sequence $\boldsymbol{X} \in [V]^N$ to Euclidean space. This is often done by associating each index in $[V]$ with a vector in $\mathbb{R}^d$ using a *massive*[34] array $\boldsymbol{E} \in \mathbb{R}^{V \times d}$, and directly forming the sequence $[\boldsymbol{E}_{\boldsymbol{x}_1}, \dots, \boldsymbol{E}_{\boldsymbol{x}_N}] \in \mathbb{R}^{d \times N}$. The full embedding map $f_\theta^{\mathrm{emb}}$ also applies a positional encoding $\boldsymbol{E}^{\mathrm{pos}} \in \mathbb{R}^{d \times N_{\max}}$ where $N_{\max}$ is the maximum number of tokens which are possible to process,[35] which yields the embedding map

$$f_\theta^{\mathrm{emb}}(\boldsymbol{X}) \doteq [\boldsymbol{E}_{\boldsymbol{x}_1}, \dots, \boldsymbol{E}_{\boldsymbol{x}_N}] + \boldsymbol{E}^{\mathrm{pos}}_{:N} \tag{8.11.2}$$

The parameters $\boldsymbol{E}$ and $\boldsymbol{E}^{\mathrm{pos}}$ are directly trainable. Since $\boldsymbol{E}$ is so large (and the gradient update is very sparse w.r.t. it since only a small fraction of the vocabulary is used in each sample), specialized software is used to make sure the memory updates are not too onerous. Notice also that we do not use a class token like in the other sections; more on this later.

[32] There are usually several special tokens for different purposes. Existing text containing the special tokens are specially processed.

[33] In direct contravention of the conventions in this book and those of many other communities, the NLP community calls such GPT-2-style transformers (encompassing nearly all current LLMs) "decoder-only" transformers. "Encoder-only" transformers have a different architecture, and "encoder-decoder" transformers concatenate an "encoder-only" transformer with a "decoder-only" transformer. This despite the fact that "decoder-only" transformers *also* compute an encoding of the data!

[34] By "massive" we mean that such a structure is often a large fraction of the language model's total size.

[35] Modern positional encoding methods have since taken care of this issue and allowed for (in theory) infinite extrapolation, but such methods are more complex to develop, and for simplicity we only introduce the absolute additive positional encoding here.

**Backbone.** We process the embeddings using a CRATE-like backbone which uses causal masking. To motivate causal masking, consider the causal language modeling loss $\mathcal{L}_{\mathrm{CLM}}$ defined in (8.11.1). The most naive implementation would require us to compute the forward pass $N$ times in order to backpropagate once. Obviously this is extremely inefficient, since $N$ can often be in the thousands. In order to scale training with this loss efficiently, we impose a *causal* constraint, i.e.,

$$\boldsymbol{Z}_{\theta}(\boldsymbol{X}_{:n}) = \boldsymbol{Z}_{\theta}(\boldsymbol{X})_{:n} \tag{8.11.3}$$

i.e., the $n$ columns of the token features $\boldsymbol{Z}_{\theta}(\boldsymbol{X}_{:n}) \in \mathbb{R}^{d \times n}$ should be the same as the first $n$ columns of the token features $\boldsymbol{Z}_{\theta}(\boldsymbol{X}) \in \mathbb{R}^{d \times N}$ regardless of the positive values of $n$ and $N$ such that $N \geq n$. In effect, this means we can apply the backbone *once* to the whole sequence and compute $\boldsymbol{Z}_{\theta}(\boldsymbol{X})$, then apply $f_{\theta}^{\mathrm{ext}}$ to each increasing subset $\boldsymbol{Z}_{\theta}(\boldsymbol{X}_{:n}) = \boldsymbol{Z}_{\theta}(\boldsymbol{X})_{:n}$ as $n$ grows to the sequence length $N$. Then we can use all of those to compute the loss.

So now that we want a causal architecture for the backbone, how can we get it? Since the MLP and layer normalizations inside each transformer layer affect each token individually, the only thing that matters for causality is the attention block (or MSSA in the case of CRATE). In order to make MSSA causal, we define the CMSSA block as

$$\mathrm{CMSSA}_{\theta}^{\ell}(\boldsymbol{Z}) \doteq \boldsymbol{U}_{\mathrm{out}}^{\ell} \begin{bmatrix} \mathrm{CSA}([\boldsymbol{U}^{1,\ell}]^{\top}\boldsymbol{Z}, [\boldsymbol{U}^{1,\ell}]^{\top}\boldsymbol{Z}, [\boldsymbol{U}^{1,\ell}]^{\top}\boldsymbol{Z}) \\ \vdots \\ \mathrm{CSA}([\boldsymbol{U}^{K,\ell}]^{\top}\boldsymbol{Z}, [\boldsymbol{U}^{K,\ell}]^{\top}\boldsymbol{Z}, [\boldsymbol{U}^{1,\ell}]^{\top}\boldsymbol{Z}) \end{bmatrix} + \boldsymbol{b}_{\mathrm{out}}^{\ell}\mathbf{1}_{N}^{\top} \tag{8.11.4}$$

$$\text{where} \quad \mathrm{CausalSA}(\boldsymbol{Q}, \boldsymbol{K}, \boldsymbol{V}) \doteq \boldsymbol{V}\,\mathrm{softmax}\left(\frac{\mathrm{CausalMask}(\boldsymbol{K}^{\top}\boldsymbol{Q})}{\sqrt{p}}\right), \tag{8.11.5}$$

$$\text{where} \quad \mathrm{CausalMask}(\boldsymbol{M})_{ij} = \begin{cases} M_{ij}, & \text{if } i \geq j, \\ -\infty, & \text{if } i < j \end{cases}. \tag{8.11.6}$$

Here, practitioners say that the causal mask *allows future tokens $i$ to attend to past tokens $j$ but not vice versa*. To see why, let us write out the expression for the $t$th column of $\mathrm{CSA}(\boldsymbol{Q}, \boldsymbol{K}, \boldsymbol{V})$:

$$\mathrm{CSA}(\boldsymbol{Q}, \boldsymbol{K}, \boldsymbol{V})_{t} = \sum_{i=1}^{t} \boldsymbol{V}_{i}\,\mathrm{softmax}\big([\boldsymbol{K}_{:t}]^{\top}\boldsymbol{Q}_{t}\big)_{i} \tag{8.11.7}$$

(where here the non-colon subscript denotes the column). This expression for the $t$th token uses no information about any token beyond index $t$. Therefore CSA, hence CMSSA, hence the whole causal CRATE backbone is causal in terms of the definition in (8.11.3), and we unlock the considerable efficiency gains that we were promised.

**Feature extractor.** We use a post-processing step $f_{\theta}^{\mathrm{ext}}$ which extracts the *feature vector of the last known token* so as to predict the next token. In theory, this means that each token $\boldsymbol{Z}_\theta(\boldsymbol{X})_n$ should contain rich information about all tokens that come before or at index $n$, i.e., $\boldsymbol{x}_1, \dots, \boldsymbol{x}_n$, as all of this information should be available for predicting the next token at index $n+1$. In practice, only a few of these tokens are really needed for each prediction task. Anyways, the equation for $f_\theta^{\text{ext}}$ is

$$f_\theta^{\text{ext}}(\boldsymbol{Z}_\theta(\boldsymbol{X}_{:n})) \doteq (\boldsymbol{Z}_\theta(\boldsymbol{X}))_n \tag{8.11.8}$$

where (again) the non-colon subscript is the column. In this case, as promised, we just directly extract the feature vector of the last token in the sequence.

**Task-specific head.** For our classification head $h_\theta$, the GPT-2 architecture uses a simple linear layer and a softmax to get the desired probability vectors:

$$h_\theta(\boldsymbol{z}) \doteq \operatorname{softmax}(\boldsymbol{W}^{\text{out}}\boldsymbol{z} + \boldsymbol{b}^{\text{out}}), \tag{8.11.9}$$

where $\boldsymbol{W}^{\text{out}} \in \mathbb{R}^{V \times d}, \boldsymbol{b}^{\text{out}} \in \mathbb{R}^{V}$. Some other more modern architectures use small MLPs and layer normalizations, but the idea is very much the same. Note that this linear layers also have large memory usage (because $V$ is very large), and form a bottleneck in training; there has been significant effort attempting to circumvent it.

All these architectural choices mean that causal training is extremely efficient relative to non-causal training:

- We only need *one forward pass* through the backbone to compute the loss for the whole sequence.
- The feature extraction is basically *free*.
- All tokens can be pushed through the task-specific head *in parallel*.

### 8.11.4 Optimization Strategy

We train our language model using end-to-end stochastic optimization. One remaining issue is that, in practice, different documents in a batch have different lengths (in terms of the number of tokens required for each sequence), but as of the time of writing this book, the main deep learning frameworks by and large allow only “rectangular” tensors, which do not accommodate this behavior. To try to resolve this issue, we just insert a special padding token `<|pad|>` for all shorter samples in the batch, so that we can batch-process everything using rectangular tensors. At each timestep $k$, we:

- Subsample $B$ different tokenized documents $\{\boldsymbol{X}_b^{(k)}\}_{b=1}^{B} \subseteq \mathcal{T} = [V]^*$, each with length $N_b^{(k)}$.
- Compute $N_{\max}^{(k)} \doteq \max_{b \in [B]} N_b^{(k)}$ and pad each $\boldsymbol{X}_b^{(k)}$ to length $N_{\max}^{(k)}$ using a special padding token.

- Compute the features $\boldsymbol{Z}_\theta(\boldsymbol{X}_b^{(k)})$.
- Compute the predicted distributions $\boldsymbol{p}_\theta(\boldsymbol{X}_{b,:n}^{(k)}) \doteq (h_\theta \circ f_\theta^{\mathrm{ext}})(\boldsymbol{Z}_\theta(\boldsymbol{X}_b^{(k)})_{:n})$.
- Form the surrogate stochastic loss

$$\hat{\mathcal{L}}_{\mathrm{CLM}}^{(k)}(\theta) \doteq \frac{1}{B(N_{\max}^{(k)} - 1)} \sum_{b=1}^{B} \sum_{n=1}^{N_{\max}^{(k)}-1} \mathrm{CE}(\mathbf{1}(\boldsymbol{x}_{b,n+1}^{(k)}), \boldsymbol{p}_\theta(\boldsymbol{X}_{b,:n}^{(k)}))). \tag{8.11.10}$$

- Compute one step of an optimization algorithm on $\theta$, giving the following iteration:

$$\theta^{(k+1)} \doteq \textsc{OptUpdate}^{(k)}(\theta^{(k)}; \nabla_\theta \hat{\mathcal{L}}_{\mathrm{CLM}}^{(k)}). \tag{8.11.11}$$

### 8.11.5 Evaluation Methodology

There are several ways to evaluate a trained transformer language model.

- On a holdout dataset of arbitrary text, we can evaluate $\mathcal{L}_{\mathrm{CLM}}$ on it; lower losses are better since they mean that the model has learned that distribution well.
- On a multiple choice question dataset, for each question we can put it as the context and check the estimated probability of the correct answer being generated.
- We can also test the *text generation* capabilities. Namely, we can repeatedly sample from the model's probability distribution over the next token given the context. Each time we sample we generate a new token, which we print and add to the context. This allows us to sample from the LLM, and judge the generated samples however we please.[36]

In this section, we perform the first kind of evaluation exclusively.

### 8.11.6 Experimental Setup and Results

Since our causal CRATE architecture is directly built upon GPT-2, we compare the optimal settings for GPT-2 as given by the NanoGPT repository [Kar22a] with the same settings applied to CRATE for a fair comparison.

[36] Having to re-run the model on each token every time can become prohibitively expensive. Clever storages of different internal features of the language model (such as the so-called *K-V cache*), along with the causality of the architecture, can dramatically reduce the cost of sampling.

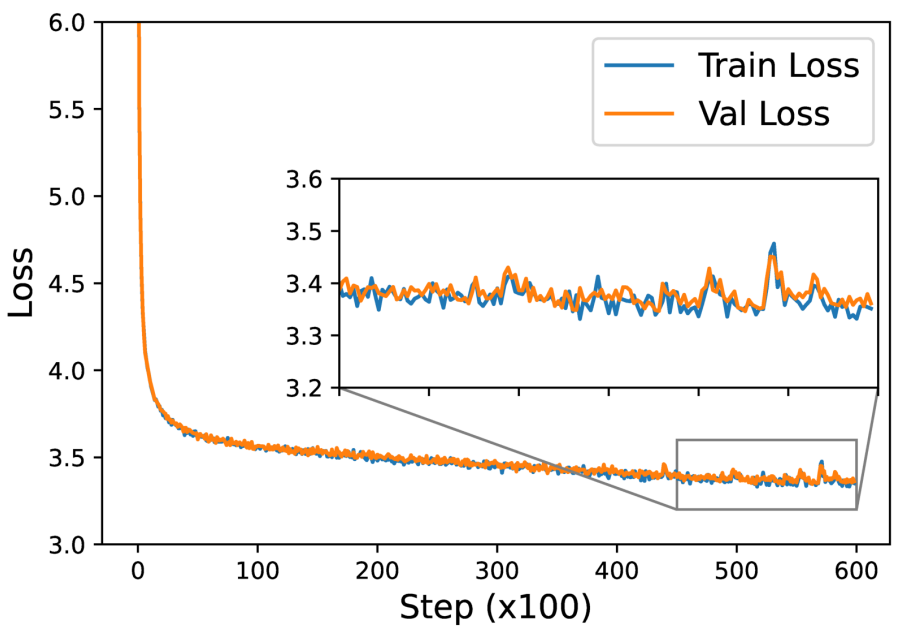

Figure 8.41: **The loss curve of CRATE-GPT-Base trained on the OpenWebText dataset.**

**Model architecture.** We use the GPT-2 tokenizer, which has vocabulary size $V = 50257$, including a special token for `<|pad|>`.[37] The context length is $N_{\max} = 1024$. The backbone model follows the GPT2-Base architecture [RWC+19] with the appropriate alterations to have causal CRATE layers, and we compare against GPT2-Small and GPT2-Base.

**Datasets and optimization.** For training causal CRATE, we follow the implementations in the NanoGPT repository [Kar22a]. Specifically, we use a batch size of 384 and train for 600,000 steps with the Adam optimizer [KB14]. For the Adam optimizer, we use $(\beta_1, \beta_2) = (0.9, 0.95)$ and weight decay of 0.1. For the learning rate schedule, we apply a linear warm-up and cosine decay, with a peak value of $\eta = 6 \times 10^{-4}$ at the $2,000$th iteration, and minimum value $6 \times 10^{-5}$. The training and validation losses over iterations are shown in Figure 8.41. The training/validation loss converges around 3.37 after training with a batch size of 384 and $600,000$ iterations. In comparison, the open GPT-2 implementation is pre-trained on OpenWebText with a batch size of 512 and $600,000$ steps and converges to a validation loss of 2.85 [Kar22a].

**Experiment results.** Table 8.18 demonstrates that CRATE models achieve reasonable performance on the causal language modeling loss across a variety of datasets compared to GPT-2 models with similar parameter counts and similar architectures.

[37] The `<|bos|>` token is not included in this setup, although it is very common in modern language models.

Table 8.18: Zero-shot cross-entropy loss of the CRATE-GPT2-Base model and GPT2-Small, GPT2-Base model evaluated on the test split of the datasets (↓ lower is better).

| | #parameters | **OWT** | **LAMBADA** | **WikiText** | **PTB** | **Avg** |
|---|---|---|---|---|---|---|
| GPT2-Base | 124M | 2.85↓ | 4.12↓ | 3.89↓ | 4.63↓ | 3.87↓ |
| GPT2-Small | 64M | 3.04 | 4.49 | 4.31 | 5.15 | 4.25 |
| Causal-CRATE-Base | 60M | 3.37 | 4.91 | 4.61 | 5.53 | 4.61 |

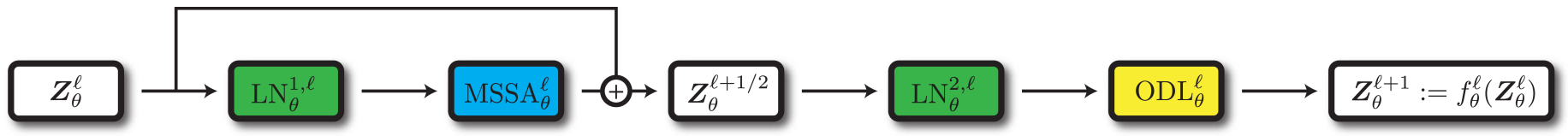

Figure 8.42: **One layer of the CRATE-$\alpha$ backbone.** The difference from CRATE is that the $\mathrm{ISTA}_\theta^\ell$ block is replaced by the $\mathrm{ODL}_\theta^\ell$ block, which performs several ISTA steps with an overcomplete dictionary.

# 8.12 Scaling and Improving White-Box Transformers

In this last section, we will discuss several ways in which various parts of CRATE-type models can be scaled up or made more efficient for certain special tasks while still remaining fully interpretable white-box. These developments mix both conceptual and empirical insights, and can be viewed as case studies about how to use white-box understanding to improve deep learning models in practice. The tasks that we use to evaluate the methods will be image classification and next-token-prediction, the data will be ImageNet and OpenWebText respectively, the optimization procedure will be the same backpropagation, and the only thing that changes is the architecture.

## 8.12.1 Increasing Network Width: CRATE-$\alpha$

One design decision enforced by the CRATE framework is the *width* of the nonlinearity in the network. In a regular transformer, the width is usually set to 4, 8, or $\frac{11}{3}$ times the feature dimension. However, CRATE enforces that the width is exactly equal to the feature dimension, i.e., the dictionaries $\boldsymbol{D}^\ell$ are square, which could lead to reduced performance. The fundamental reason that the CRATE framework constrains us to this choice is as follows:

- The ISTA block takes a *single* step of optimization for dictionary learning.
- Usually one step of any iterative optimization algorithm cannot effectively optimize the objective. So then why does this work?
- Optimization algorithms usually converge very quickly if and only if they have good initializations, or *warm starts*. The ISTA block has a warm start — it treats the input features as an initialization to the resulting sparse codes.

- This enforces that the input features and sparse codes have the same dimension. Namely, ISTA learns a complete sparsifying dictionary (cf Chapter 2).

Thus if we want to use a wide dictionary, we need ISTA to perform *overcomplete* dictionary learning. This means we cannot have the same warm start (as our sparse codes have a larger dimension than our features), and need more iterations to converge to a sparse code. Hence the step from features $\boldsymbol{Z}_{\theta}^{\ell+1/2}$ to sparse codes $\boldsymbol{Z}_{\theta}^{\ell+1}$ would no longer be

$$\boldsymbol{Z}_{\theta}^{\ell+1} = \mathrm{ISTA}_{\theta}^{\ell}(\boldsymbol{Z}_{\theta}^{\ell+1/2} \mid \boldsymbol{Z}_{\theta}^{\ell+1/2}) \tag{8.12.1}$$

where the $\mathrm{ISTA}_{\theta}^{\ell}$ function is defined as (by an abuse of notation from earlier sections)

$$\mathrm{ISTA}_{\theta}^{\ell}(\boldsymbol{Z} \mid \boldsymbol{Y}) \doteq \mathrm{ReLU}(\boldsymbol{Z} - \beta(\boldsymbol{D}^{\ell})^{\top}(\boldsymbol{D}^{\ell}\boldsymbol{Z} - \boldsymbol{Y}) + \beta\lambda\mathbf{1}_{s}\mathbf{1}_{n}^{\top}) \tag{8.12.2}$$

but rather the following iteration:

$$\boldsymbol{Z}_{\theta}^{\ell+1} = \boldsymbol{A}_{\theta}^{\ell,T}; \qquad \boldsymbol{A}_{\theta}^{\ell,t+1} = \mathrm{ISTA}_{\theta}^{\ell}(\boldsymbol{A}_{\theta}^{\ell,t} \mid \boldsymbol{Z}_{\theta}^{\ell+1/2}) \quad \forall 0 \leq t < T; \qquad \boldsymbol{A}_{\theta}^{\ell,0} = \mathbf{0}_{s\times n}, \tag{8.12.3}$$

i.e., running proximal gradient on the LASSO objective for $T \geq 1$ steps in the forward pass at each layer, initialized at $\mathbf{0}_{s\times n}$. In this circumstance, the dictionary can be as wide as needed, i.e., $\boldsymbol{D}^{\ell} \in \mathbb{R}^{s\times d}$ where $s \geq d$ (usually one takes $s = 4d$ in practice).

However, this presents an empirical problem. Using the above configuration, if $\boldsymbol{Z}^{\ell+1/2} \in \mathbb{R}^{d\times n}$, then $\boldsymbol{Z}^{\ell+1} \in \mathbb{R}^{s\times n}$, which can have an arbitrarily large feature dimension. In practice, we want the feature dimension at each layer to be the same. So this sets up a practical trichotomy for designing the network, namely, we *cannot* have *all* of the following desiderata:

1. The feature dimension at each layer is the same.

2. The dictionary is wide, i.e., overcomplete.

Table 8.19: **Object detection and fine-grained segmentation via MaskCut on COCO val2017** [LMB+14]. Here all models are trained with patch size 8 instead of 16. Compared with existing models such as CRATE and ViT, the CRATE-$\alpha$ model family noticeably has improved performance as well as scalability.

| | Detection | | | Segmentation | | |
|---|---|---|---|---|---|---|
| Model | $\mathrm{AP}_{50}\uparrow$ | $\mathrm{AP}_{75}\uparrow$ | $\mathrm{AP}\uparrow$ | $\mathrm{AP}_{50}\uparrow$ | $\mathrm{AP}_{75}\uparrow$ | $\mathrm{AP}\uparrow$ |
| CRATE-$\alpha$-B/8 | 3.5 | 1.1 | 1.5 | 2.2 | 1.0 | 1.1 |
| CRATE-$\alpha$-L/8 | **4.0** | **1.7** | **2.0** | **2.7** | **1.1** | **1.4** |
| CRATE-B/8 | 2.9 | 1.0 | 1.3 | 2.2 | 0.7 | 1.0 |
| ViT-B/8 | 0.8 | 0.2 | 0.4 | 0.7 | 0.5 | 0.4 |

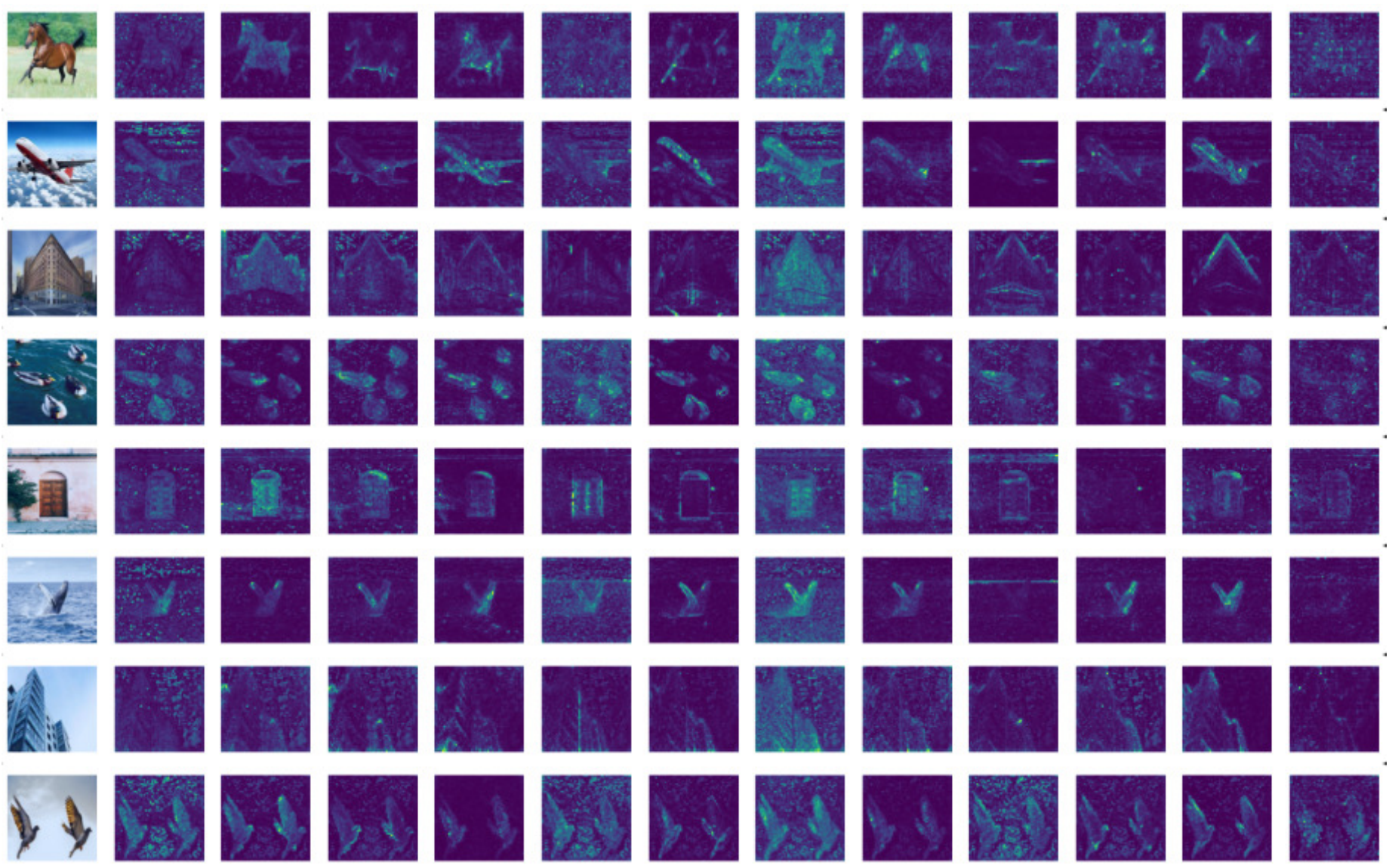

Figure 8.43: **Saliency maps from CRATE-$\alpha$ with patch size 8.** Each row is a different image and each column corresponds to a different attention head in the last layer. We observe that the saliency maps strongly correspond to the objects in the input image.

3. The output of the nonlinearity is the sparse codes of the input with respect to the dictionary.

In practice, giving up (1) is less tractable for efficiency reasons. Giving up (2) leads to the usual CRATE framework. Giving up (3) leads to a wide version of CRATE, i.e., CRATE-$\alpha$, which has the following nonlinearity to get from $\boldsymbol{Z}^{\ell+1/2}$ to $\boldsymbol{Z}^{\ell+1}$:

$$\boldsymbol{Z}_{\theta}^{\ell+1} = \boldsymbol{D}^{\ell}\boldsymbol{A}_{\theta}^{\ell,T}; \qquad \boldsymbol{A}_{\theta}^{\ell,t+1} = \mathrm{ISTA}_{\theta}^{\ell}(\boldsymbol{A}_{\theta}^{\ell,t} \mid \boldsymbol{Z}_{\theta}^{\ell+1/2}); \qquad \boldsymbol{A}_{\theta}^{\ell,0} = \boldsymbol{0}, \quad (8.12.4)$$

i.e., it takes the sparse codes obtained via proximal gradient descent and multiplies by the dictionary to get the denoised version of the input. Thus CRATE-$\alpha$'s nonlinearity computes a denoised version of the input which is amenable to sparse coding, not the actual sparse codes themselves. The map from $\boldsymbol{Z}_{\theta}^{\ell+1/2}$ to $\boldsymbol{Z}_{\theta}^{\ell+1}$ here is called the Overcomplete Dictionary Learning (ODL) block and denoted $\mathrm{ODL}_{\theta}^{\ell}$, i.e.,

$$\boldsymbol{Z}_{\theta}^{\ell+1}(\boldsymbol{X}) \doteq \mathrm{ODL}_{\theta}^{\ell}(\boldsymbol{Z}_{\theta}^{\ell+1/2}(\boldsymbol{X})). \quad (8.12.5)$$

The CRATE-$\alpha$ layer is shown in Figure 8.42. In practice this modification of CRATE performs very well at larger scales. For example, when we pre-train CRATE-$\alpha$ models on ImageNet-21K, unsupervised tasks like segmentation (see Figure 8.43 and Table 8.19) generally have significantly improved performance compared to CRATE. Similar trends are present in language model training using causal self-attention (see Table 8.20). Overall, it is a promising avenue to scaling up the performance to match black-box models such as transformers.[38]

Table 8.20: **Validation loss in language modeling.** Here all models are pre-trained on most of OpenWebText, and the validation cross-entropy loss is measured on a hold-out subset of OpenWebText. CRATE-$\alpha$ shows significant improvement over the CRATE design, though there still exists a gap with traditional transformers like GPT-2.

| Model | GPT-2-B(ase) | CRATE-B | CRATE-$\alpha$-S(mall) | CRATE-$\alpha$-B |
|---|---|---|---|---|
| # parameters | 124M | 60M | 57M | 120M |
| OWT val. loss | 2.85 | 3.37 | 3.28 | 3.14 |

### 8.12.2 Linear Time Complexity Transformers

In practice, deep learning models suffer from bottlenecks in space and time complexity, representing problem sizes beyond which they cannot scale given fixed resources. One such bottleneck, particularly meaningful when dealing with data where each sample is itself high-dimensional and rich (such as long streams of text or videos), is the *time complexity* of processing long sequences of data. In order to alleviate the time complexity of processing data using transformers, in Section 5.3.2 we proposed a *token statistics self-attention* operator $\mathrm{TSSA}_{\theta}^{\ell}$. We now build a *token statistics transformer*, called ToST:

https://robinwu218.github.io/ToST/,

around it, which we can use for long-context tasks. In particular, we can use the following layer (depicted in Figure 8.44) as a drop-in replacement for a backbone layer in CRATE:

$$\boldsymbol{Z}_{\theta}^{\ell+1/2}(\boldsymbol{X}) = \boldsymbol{Z}_{\theta}^{\ell}(\boldsymbol{X}) + \mathrm{TSSA}_{\theta}^{\ell}(\mathrm{LN}_{\theta}^{1,\ell}(\boldsymbol{Z}_{\theta}^{\ell}(\boldsymbol{X}))) \tag{8.12.6}$$

$$\boldsymbol{Z}_{\theta}^{\ell+1}(\boldsymbol{X}) = \boldsymbol{Z}_{\theta}^{\ell+1/2}(\boldsymbol{X}) + \mathrm{MLP}_{\theta}^{\ell}(\mathrm{LN}_{\theta}^{2,\ell}(\boldsymbol{Z}_{\theta}^{\ell+1/2}(\boldsymbol{X}))) \tag{8.12.7}$$

where the TSSA block is defined as in Section 5.3.2. Notice that this is exactly the same as the vision transformer architecture discussed in Section 8.2.3, except that TSSA replaces the conventional multi-head self-attention block MHSA. Regardless, the computational complexity of the forward pass of this layer is linear in all problem variables — sequence length, feature dimension, number of heads, and head dimension.

Moreover, the proposed architecture, named ToST (for "Token Statistics Transformer") performs well at vision tasks (i.e., Table 8.21) and language tasks (i.e., Table 8.22). This is especially true for long-sequence-length tasks (cf Table 8.23), where it is both more performant and much more efficient than conventional transformers and all other transformer-like architectures.

[38] Note that the experimental results in this section use a slightly different model architecture, which add very slight empirical gains. The changes are: (1) an additional residual connection on the ODL block, (2) modifying ISTA to use two separate dictionaries instead of $\boldsymbol{D}^{\ell}$ and $(\boldsymbol{D}^{\ell})^{\top}$.

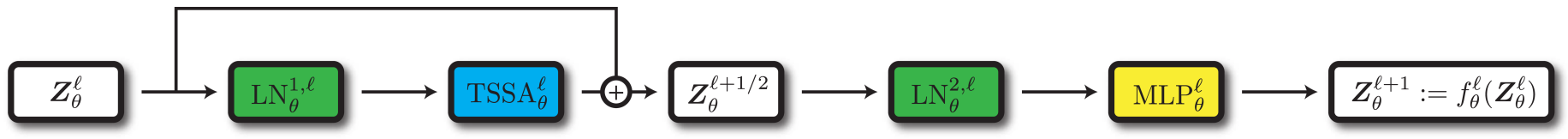

Figure 8.44: **One layer of the ToST backbone**. Token representations go through layer-norms, the token statistics self-attention (TSSA) operator, and an MLP, in order to form the layer's output.

| Datasets | ToST-T(iny) | ToST-S(mall) | ToST-M(edium) | XCiT-S | XCiT-M | ViT-S | ViT-B(ase) |
|---|---|---|---|---|---|---|---|
| # parameters | 5.8M | 22.6M | 68.1M | 24.9M | 80.2M | 22.1M | 86.6 M |
| ImageNet | 67.3 | 77.9 | 80.3 | 80.5 | 81.5 | 79.8 | 81.8 |
| ImageNet ReaL | 72.2 | 84.1 | 85.6 | 85.6 | 85.9 | 85.6 | 86.7 |
| CIFAR10 | 95.5 | 96.5 | 97.5 | 98.1 | 98.3 | 98.6 | 98.8 |
| CIFAR100 | 78.3 | 82.7 | 84.5 | 86.1 | 87.6 | 88.8 | 89.3 |
| Oxford Flowers-102 | 88.6 | 92.8 | 94.2 | 93.9 | 94.0 | 94.0 | 95.7 |
| Oxford-IIIT-Pets | 85.6 | 91.1 | 92.8 | 92.9 | 94.0 | 92.8 | 94.1 |

Table 8.21: **Linear probing classification accuracy of ToST** on various datasets with different model sizes when the backbone is pre-trained for ImageNet-1K classification. We observe that, compared to the XCiT (a empirically-designed transformer-like architecture specialized for efficient processing of long sequences) and the ViT, ToST maintains relatively similar performance, even while enjoying benefits like faster runtime and white-box design.

| Model | # params | OWT | Lambada | Wikitext | PTB | Avg ↓ |
|---|---|---|---|---|---|---|
| GPT-2-Base | 124M | 2.84 | 4.32 | 4.13 | 5.75 | 4.26 |
| ToST-Base | 110M | 3.20 | 4.98 | 4.77 | 6.39 | 4.84 |
| ToST-Medium | 304M | 2.88 | 4.45 | 4.30 | 5.64 | 4.32 |
| ToST-Large | 655M | 2.72 | 4.32 | 3.99 | 5.03 | 4.02 |

Table 8.22: **Language modeling validation loss** computed on (holdout sets of) a variety of natural language datasets, after pre-training the model on that dataset. We observe that ToST scales well, so that ToST-Large surpasses the baseline GPT-2-Base in causal language modeling, while enjoying superior efficiency in long contexts.

### 8.12.3 Attention-Only Transformers

Another bottleneck to remove from deep learning models, specifically transformer-like architectures, is the memory bottleneck that comes from massive matrix multiplications in MLPs, where the internal dimension is far greater than the feature dimension $d$. It thus is an interesting and important question to ask: do we *really* need the MLP inside a transformer, and how good can the performance get without it? To explore this question, we use the attention-only-transformer (AoT) architecture (see Section 5.3.1), depicted in Figure 8.45. Namely, each

| Model | ListOps | Text | Retrieval | Image | Pathfinder | Avg |
|---|---|---|---|---|---|---|
| Reformer | **37.27** | 56.10 | 53.40 | 38.07 | 68.50 | 50.56 |
| BigBird | 36.05 | 64.02 | 59.29 | 40.83 | 74.87 | 54.17 |
| LinFormer | 16.13 | <u>65.90</u> | 53.09 | 42.34 | <u>75.30</u> | 50.46 |
| Performer | 18.01 | 65.40 | 53.82 | 42.77 | **77.05** | 51.18 |
| Transformer | 37.11 | 65.21 | <u>79.14</u> | <u>42.94</u> | 71.83 | <u>59.24</u> |
| ToST | <u>37.25</u> | **66.75** | **79.46** | **46.62** | 69.41 | **59.90** |

Table 8.23: **Long-Range Arena (LRA) performance comparison of ToST(-B) versus the top transformer variants optimized for long-context.** Long-Range Arena is a family of benchmarks that test the long sequence modeling capability of algorithms and architectures, by fixing the dataset and evaluation mechanism. ToST scores at the top of the leaderboard compared to all known transformer variants, including XCiT and the regular (ViT) transformer (cf Table 8.21). Moreover, ToST has the lowest time- and space-complexity inference. (In this table, the best score for a particular benchmark is bolded, and the second-best score is underlined.)

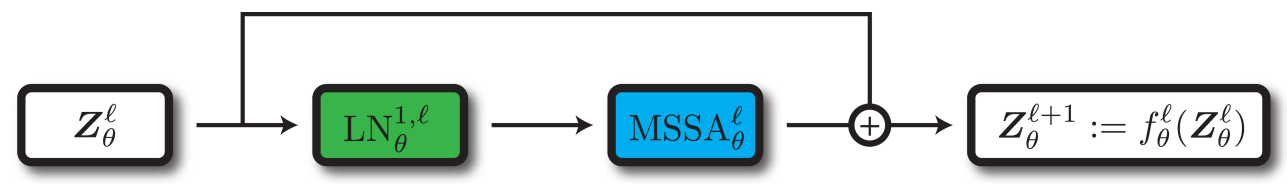

Figure 8.45: **One layer of the AoT backbone.** Token representations merely go through a layer-norm and the multi-head (subspace) self-attention operator to form the layer's output. Notice that there is no token-wise nonlinearity such as MLP or ISTA or ODL.

layer is simply of the form

$$\boldsymbol{Z}_\theta^{\ell+1}(\boldsymbol{X}) = \boldsymbol{Z}_\theta^{\ell}(\boldsymbol{X}) + \mathrm{MSSA}_\theta^\ell(\mathrm{LN}_\theta^\ell(\boldsymbol{Z}_\theta^\ell(\boldsymbol{X}))). \tag{8.12.8}$$

In our implementation, we also experimented with using multi-head self-attention (MHSA) in place of MSSA. It turns out that this architecture is *also* viable, though the depth of the network needs to be much greater in order to achieve equivalent performance to the usual CRATE or transformer architecture.

We conduct experiments using the proposed AoT architecture and demonstrate its potential. We pre-train the AoT-MSSA and AoT-MHSA models of different sizes, along with GPT-2, on OpenWebText [GC19]. We plot the training loss and validation loss against the number of training iterations in Figure 8.46(a) and (b), respectively. It is observed that medium- and large-sized AoT-based models achieve training and validation losses comparable to those of the GPT-2 base model. In addition, compared to the GPT-2 base model, the AoT-MHSA model is identical to the GPT-2 base model, except for the absence of MLP layers in the architecture. As shown in Figure 8.46, incorporating MLP layers can accelerate the training process. Using the above pre-trained models, we compute the cross-entropy validation loss without training on different datasets in Table 8.24. It is observed that the AoT models with medium and large parameter sizes can achieve comparable performance to the GPT-2 base model. Moreover, we found that adding MLP layers to AoT does not improve the zero-shot performance. These results highlight the potential of attention-only models to achieve competitive results while maintaining interpretability.

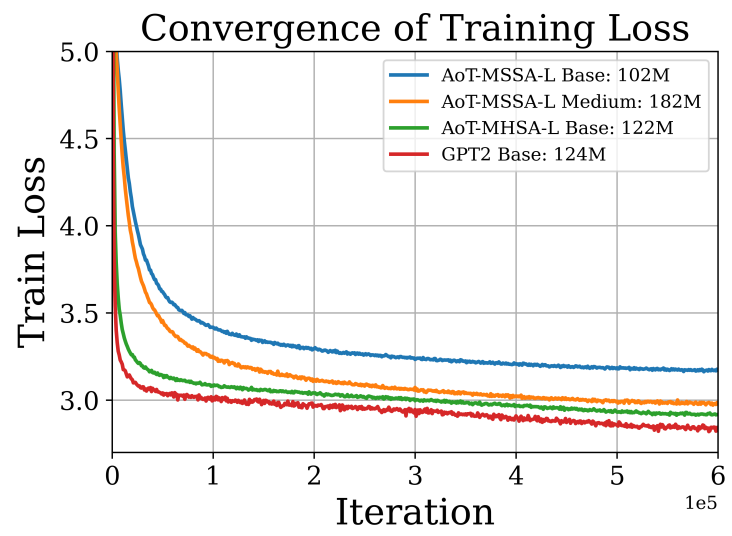

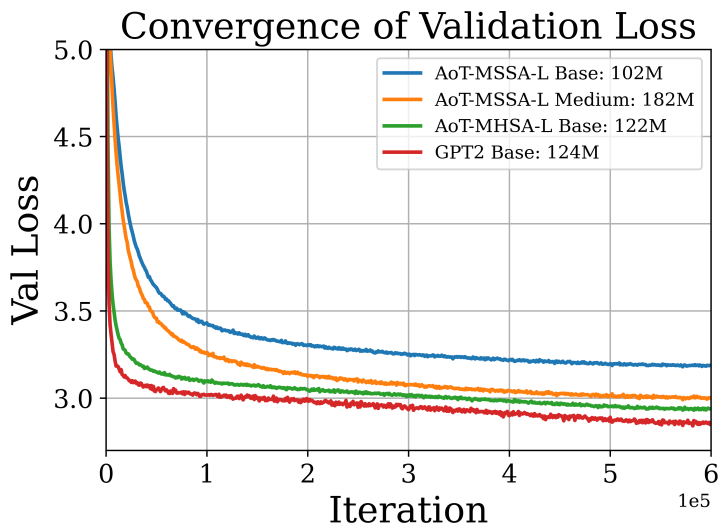

Figure 8.46: **Evaluating models on language tasks.** We plot the training loss (left) and validation loss (right) of the AoT and GPT-2 models pretrained on OpenWebText.

Table 8.24: Zero-shot results on several language benchmark datasets and tasks: Evaluation of different sizes of AoT with the MSSA and MHSA operators and comparison to the GPT2 model.

| Models<br># of parameters | **LAMBADA**<br>(val loss) ↓ | **PTB**<br>(val loss) ↓ | **WikiText**<br>(val loss) ↓ | **LAMBADA**<br>(acc) ↑ | **CBT CN**<br>(acc) ↑ | **CBT NE**<br>(acc) ↑ |
|---|---|---|---|---|---|---|
| AoT-MSSA Base (102M) | 4.70 | 6.03 | 4.65 | 0.25 | 0.80 | 0.74 |
| AoT-MSSA Medium (182M) | 4.47 | 5.08 | 4.22 | 0.29 | 0.84 | 0.77 |
| AoT-MHSA Base (122M) | 4.42 | 5.52 | 4.19 | 0.38 | 0.86 | 0.82 |
| GPT-2 Base (124M) | 4.32 | 5.75 | 4.13 | 0.40 | 0.87 | 0.84 |

## 8.13 Summary and Notes

All work in this chapter is downstream of the Transformer architecture, which was introduced by Vaswani et al. [VSP+17b]. The Transformer architecture is formally described in Section 8.2. A main empirical innovation in recent years, spurred by the prevalence and performance of the Transformer architecture, is to formulate a given learning problem as a sequence-to-sequence problem and apply the Transformer architecture. This has enabled the Transformer architecture to be essentially ubiquitous in (almost) all deep learning applications. As such, direct improvements to the Transformer can propagate to become solutions to many problems and have considerable impact; similarly, we can apply our white-box understanding of transformer-like architectures to many modern problems. The material covered in this chapter is merely a subset of the work that has already been done; other work includes masked completion for text data (i.e., BERT) [DCL+19; YBP+24], (mechanistic) interpretability of language and vision models [BM24], and error correcting coding [ZLG+]. There is much more to do.

There is also much more theory specifically about the practice of scaling neural networks, which is enormously practically viable, and we at least remark on it here. This line of work was popularized by the "Tensor Programs" line of work [YHB+22]. The basic prescription is that we want the initial gradient updates in a transformer to be constant size, and by working through the backpropagation equations (Chapter A) carefully, we can determine the scale of the initialization and learning rates (chosen layer-wise) that are required to achieve this. In practice, such prescriptions greatly increase the stability and convergence of training at large scales; they also prescribe a way to find the "optimal"[39] hyperparameters for large-scale training using only small-scale training. Follow-ups to this work attempt to accommodate the feature geometry [BN24a], which could be informed by the work in this book about representation learning. Other follow-ups incorporate this weight-wise information into the optimizer itself to obtain these scaling benefits automatically, obtaining optimizers such as Muon [JJB+], which have recently been used for training trillion-parameter models very stably [Tea]. Overall, the two approaches to deep learning theory are orthogonal or complementary.

## 8.14 Exercises and Extensions

*Exercise* 8.1. Read the DINO paper [CTM+21].

*Exercise* 8.2. DINO v2 [ODM+23] uses everything from DINO v1 but also, during the data augmentation phase, randomly masks out patches within each view. This kind of augmentation should enforce that the features of images with similar local information are similar. Formulate an optimization problem that promotes this in the encoder, and implement it.

*Exercise* 8.3. This exercise considers the implementation of stochastic optimization algorithms to minimize losses involving expectations.

(a) Propose an alternative to the term involving $R_\varepsilon$ in (8.2.44) for approximating the covariance regularization term in (8.2.42). Evaluate the time complexity required to compute your proposed term and its gradient. Include an analysis for computing it on a single compute node vs. multiple nodes.

(b) Evaluate the time complexity required to compute the existing term in (8.2.44) and its gradient.

*Exercise* 8.4. Prove that (8.2.50) and (8.2.51) are convex optimization problems.

*Exercise* 8.5.

(a) Implement the CRATE and CRATE-$\alpha$ models.

[39] The word "optimal" is used in quotes because the work on this merely uses some desiderata about the weight size, feature size, and gradient size at initialization to determine "optimality", as opposed to, say, the test loss at convergence.

(b) Compare their performance and efficiency on the CIFAR-10 dataset.

(c) Compare their interpretability in two ways:

- The sparsity $\|\boldsymbol{Z}\|_0$ of the representation $\boldsymbol{Z}$
- The attention maps $\boldsymbol{a}_{\theta}^{k,\ell}$

# Chapter 9

# Open Problems and Directions about Intelligence

> "*The study is to proceed on the basis of the conjecture that every aspect of learning or any other feature of intelligence can in principle be so precisely described that a machine can be made to simulate it. An attempt will be made to find how to make machines use language, form abstractions and concepts, solve kinds of problem now reserved for humans, and improve themselves.*"
>
> – Proposal for the Dartmouth AI program, 1956

This manuscript systematically introduces mathematical principles and computational mechanisms for how memory or empirical knowledge can be developed from observed high-dimensional data. The ability to seek parsimony in a seemingly random world is a fundamental characteristic of any intelligence, natural or artificial. We believe the principles and mechanisms presented in this book are unifying and universal, applicable to both animals and machines.

We hope this book helps readers fully clarify the mystery surrounding modern practices of artificial deep neural networks by developing a rigorous understanding of their functions and roles in learning (representations of) low-dimensional distributions from high-dimensional data. With this understanding, the capabilities and limitations of existing AI models and systems become clear:

1. Existing open-loop models and systems fall short of being complete memory systems capable of self-learning and self-improving.

2. Existing realizations of representation learning remain primitive and brute-force, far from optimal in terms of network architectures or optimization strategies.

3. Existing AI models only learn data distributions and conduct inductive Bayesian inference, which differs from high-level human intelligence.

One goal of this book is to help readers establish an objective and systematic understanding of current machine intelligence technologies and to recognize what open problems and challenges remain for further advancement of machine intelligence. In the last chapter, we provide some of our views and projections for the future.

## 9.1 Towards Autonomous Intelligence: Close the Loop?

From the practice of machine intelligence in the past decade, it has become clear that, given sufficient data and computational resources, one can build a large enough model and pre-train it to learn the *prior* distribution of all the data, say $p(\boldsymbol{x})$. Theoretically, such a large model can memorize almost all existing knowledge about the world that is encoded in data such as 2D images, 3D shapes, dynamical motions, and natural languages. As we discussed at the beginning of the book, such a large model plays a role similar to DNA, which life uses to record and pass on knowledge about the world.

The model and distribution learned in this way can then be used to create new data samples drawn from the same distribution. One can also use the model to conduct inference (e.g., completion, estimation, and prediction) with the memorized knowledge under various conditions, say by sampling the *posterior* distribution $p(\boldsymbol{x} \mid \boldsymbol{y})$ under a new observation $\boldsymbol{y}$. Strictly speaking, such inference is inductive or statistical.

Any pre-trained model, however large, cannot guarantee that the distribution it has learned so far is entirely correct or complete. If our samples $\hat{\boldsymbol{x}}_t$ from the current *prior* $p_t(\boldsymbol{x})$ or estimates $\hat{\boldsymbol{x}}_t(\boldsymbol{y})$ based on the *posterior* $p_t(\boldsymbol{x} \mid \boldsymbol{y})$ are inconsistent with the truth $\boldsymbol{x}$, we would like to correct the learned distributions:

$$p_t(\boldsymbol{x}) \to p_{t+1}(\boldsymbol{x}) \quad \text{or} \quad p_t(\boldsymbol{x} \mid \boldsymbol{y}) \to p_{t+1}(\boldsymbol{x} \mid \boldsymbol{y}), \tag{9.1.1}$$

based on the error $\boldsymbol{e}_t = \boldsymbol{x}_t - \hat{\boldsymbol{x}}_t$. This is known as error correction based on error feedback, a ubiquitous mechanism in nature for continuous learning. However, any open-ended model itself lacks the mechanism to revise or improve the learned distribution when it is incorrect or incomplete. Improving current AI models still depends largely on human involvement: supervision or reinforcement through experimentation, evaluation, and selection. We may call this process "artificial selection" of large models, as opposed to the natural selection for the evolution of life.

As we studied earlier in this book (Chapter 6 in particular), closed-loop systems align an internal representation with sensed observations of the external world. They can continuously improve the internally learned distribution and its representation to (incrementally) achieve consistency or self-consistency. An immediate next step is to develop truly closed-loop memory systems, illustrated in Figure 9.1, that autonomously and continuously learn general data distributions and improve based on error feedback.

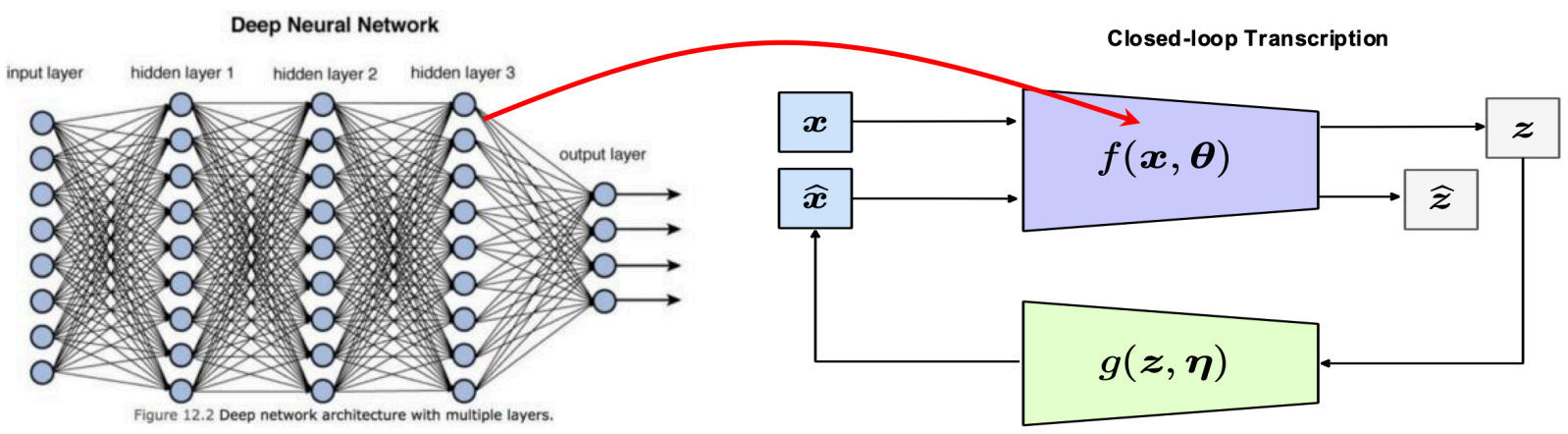

Figure 9.1: From an open-ended deep network to a closed-loop system.

Therefore, the transition from the currently popular end-to-end trained open-loop models to continuously learning closed-loop systems

$$\textbf{open-ended} \text{ models} \implies \textbf{closed-loop} \text{ systems} \tag{9.1.2}$$

is the key for machines to truly emulate how the animal brain learns and applies knowledge in an open world. We believe that

*open-ended models are for a closed world, however large;*
*closed-loop systems are for an open world, however small.*

Hence, the so-called "artificial general intelligence" (AGI) can never be achieved by simply having a system to memorize all existing data and knowledge of the world. No knowledge, however much, is truly generalizable.[1] Nevertheless, it is the ability to improve one's existing memory or knowledge and adapt to any new environments that is truly generalizable. Hence, truly genuine intelligence is itself generalizable if realized completely and correctly.[2] This is precisely the essence of the Cybernetics program laid out by Norbert Wiener in the 1940s that we discussed at the very beginning of this book.

## 9.2 Towards Natural Intelligence: Beyond Back Propagation?

The practice of machine intelligence in recent years has led many to believe that one must build a single large model to learn the distribution of all data and memorize all knowledge. Even though this is technologically possible, such a solution is likely far from necessary or efficient. As we know from training deep networks, the only known scalable method to train such networks at scale is through back propagation (BP) [RHW86b]. Although BP offers a way to correct errors via gradient signals propagated back through the whole model, it is nevertheless rather brute-force and differs significantly from how nature learns: BP is an option that nature cannot afford due to its high cost and simply cannot implement due to physical limitations.

[1] According to Sir Karl Popper, an influential Philosopher of Science, all scientific theory and knowledge are falsifiable!

[2] In our opinion, AGI should represent "artificial genuine intelligence."

More generally, we cannot truly understand intelligence unless we also understand how it can be efficiently implemented. That is, one needs to address the computational complexity of realizing mechanisms associated with achieving the objectives of intelligence. Historically, our understanding of (machine) intelligence has evolved through several phases, from the incomputable Kolmogorov complexity to Shannon's entropy, from Turing's computability to later understanding of tractability,[3] and to the strong emphasis on algorithm scalability in modern practice of high-dimensional data analysis [WM22] and artificial intelligence. This evolution can be summarized as follows:

$$\textbf{incomputable} \implies \textbf{computable} \implies \textbf{tractable} \implies \textbf{scalable}. \tag{9.2.1}$$

To a large extent, the success and popularity of deep learning and back propagation is precisely because they have offered a scalable implementation with modern computing platforms (such as GPUs) for processing and compressing massive data. Nevertheless, such an implementation is still far more expensive than how nature realizes intelligence.

There remains significant room to improve the efficiency of machine intelligence so that it can emulate the efficiency of natural intelligence, which should be orders of magnitude greater than current brute-force implementations. To this end, we need to discover new learning architectures and optimization mechanisms that enable learning data distributions under natural physical conditions and resource constraints, similar to those faced by intelligent beings in nature—for example, without accessing all data at once or updating all model parameters simultaneously (via BP).

The principled framework and approach laid out in this book can guide us to discover such new architectures and mechanisms. These new architectures and mechanisms should enable online continuous learning and should be updatable through highly localized and sparse forward or backward optimization.[4]

As we have learned from neuroscience, the cortex of our brain consists of tens of thousands of cortical columns [Haw21]. All cortical columns have similar physical structures and functions. They are highly parallel and distributed, though sparsely interconnected. Hence, we believe that to develop a more scalable and structured memory system, we need to consider architectures that emulate those of the cortex. Figure 9.2 shows such a hypothesized architecture: a massively distributed and hierarchical system consisting of many largely parallel closed-loop auto-encoding modules.[5] These modules learn to encode different sensory modalities or many projections of data from each sensory modality. Our discussion in Section 7.5 of Chapter 7 suggests that such parallel sensing and learning of a low-dimensional distribution is theoretically possible. Higher-level (lossy) autoencoders can then be learned based on outputs of lower-level ones to develop more sparse and higher-level "abstractions" of the representations learned by the lower levels.

[3]We say a problem is tractable if it allows an algorithm whose complexity is polynomial in the size of the problem.

[4]For learning a distribution, a simple instantiation of these desiderata is in the simplest case of PCA, with the online PCA method introduced in Chapter 6. Once the linear model for the distribution is learned, the easier task of conducting online prediction with noisy observations can then be easily accomplished by least-square type methods, such as the Wiener filter [Wie42; Wie49]. In the case of linear dynamical systems, that becomes the famous the Kalman filter, see [MKS+04, Appendix B] or [PV25].

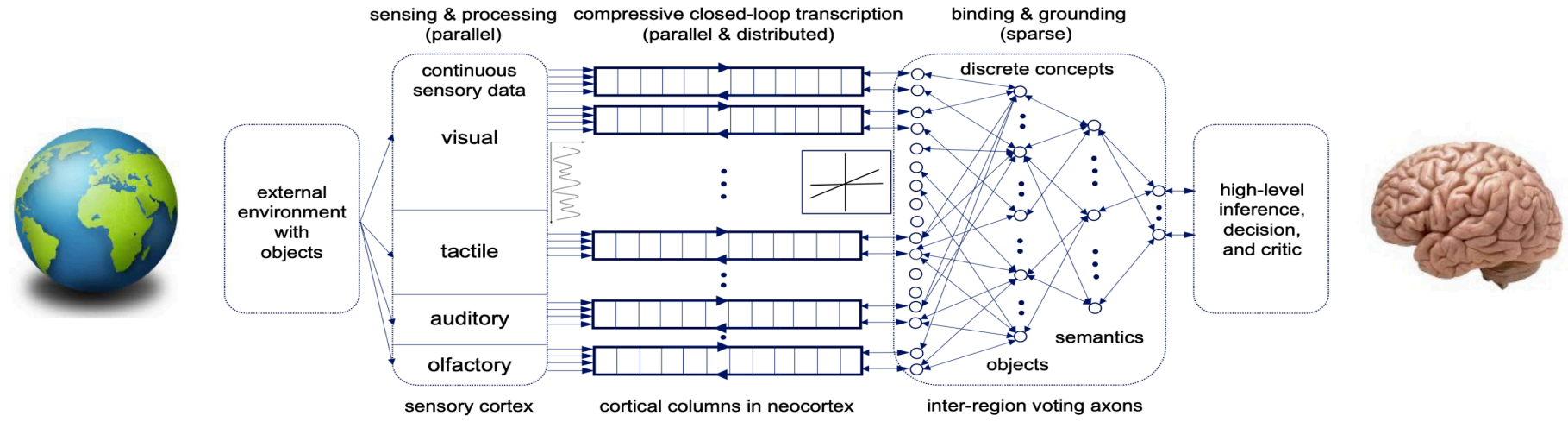

Figure 9.2: Conjectured architecture of the brain cortex. The cortex is a massively parallel and distributed auto-encoding system consisting of a hierarchy of closed-loop auto-encoders. They extract information from their inputs and maximize the information gain of the representations they output. Such processes are done at multiple levels of the hierarchy with different levels of abstraction of the information.

The distributed, hierarchical, and closed-loop system architecture illustrated in Figure 9.2 shares many characteristics of the brain's cortex. Such a system architecture may open up many more possibilities than the current single large-model architecture. In particular, it would lead to new optimization mechanisms that support continuouss self-supervised online learning and result in an ever more structured and rich representation of the sensed data of an open world. Such a system would truly emulate how the memory in our brain works. It would also allow us to transform the cuurently scalable implementation of machine intelligence into a new continous learning paradigm that achieves or even surpasses the effectiveness andefficiency of intelligence in nature:

$$\textbf{incomputable} \Longrightarrow \textbf{computable} \Longrightarrow \textbf{tractable} \Longrightarrow \textbf{scalable} \Longrightarrow \textbf{natural}. \tag{9.2.2}$$

[5] Elements of such hypothetical architectures exist in the literature in various forms, such as predictive coding networks [MSS+22], which have an optimization strategy that utilizes efficient local updates.

## 9.3 Towards Scientific Intelligence: Beyond the Turing Test?

### 9.3.1 The Evolution of Intelligence in Nature

As we have discussed at the beginning of this book, Chapter 1, intelligence in nature has evolved through multiple phases and manifested in at least four distinct forms:

$$\textbf{phylogenetic} \Longrightarrow \textbf{ontogenetic} \Longrightarrow \textbf{societal} \Longrightarrow \textbf{scientific intelligence}. \quad (9.3.1)$$

All forms of intelligence share the common objective (and characteristic) of learning useful memory or knowledge as low-dimensional distributions of high-dim data sensed from the open world. Nevertheless, they differ in the specific code books or coding schemes adopted for encoding the useful information or knowledge, the specific information or knowledge encoded by the individual or a group, the computational mechanisms for acquiring and improving the encoded information or knowledge, and the physical substrate and implementation of such mechanisms. Using the concepts and terminology developed in this book, the four stages of intelligence differ in the following three aspects:

1. The *code book* or coding scheme used to encode the intended information or knowledge.

2. The *information* or knowledge learned and encoded using the code book.

3. The *optimization mechanisms* used to create and correct the encoded information or knowledge.

Table 9.1 summarizes their main characteristics:

| | **Phylogenetic** | **Ontogenetic** | **Societal** | **Scientific** |
|---|---|---|---|---|
| **Codebook** | DNAs | Neurons/Brain | Natural Languages | Math Abstractions |
| **Information** | Genes | Memory | Empirical Knowledge | Scientific Facts |
| **Optimization** | Reinforcement | Error Feedback | Trial & Error | Theorize & Falsify |

Table 9.1: Main characteristics of the four stages of intelligence in nature.

As we now know, humans have achieved two quantum leaps in intelligence:

1. The first was the development of spoken and written language, which enabled humans to share and transmit learned knowledge across generations, much as DNA does in nature.

2. The second was the development of mathematics and formal logic roughly three thousand years ago, which became the precise language of modern science.

The language of mathematics freed us from summarizing knowledge from observations only in empirical form and allowed us to formalize knowledge as theories verifiable or falsifiable through mathematical deduction or experimentation. Through hypothesis formulation, logical deduction, and experimental verification or falsification, we can now proactively discover and develop new knowledge that is far beyond what can be learned from the distributions of sensed data. For example, causal relationships cannot be learned from distributions alone [Pea09].

Despite the seeming differences among these different stages of intelligence, the evolution of intelligence shares a common characteristic: *they all use certain forms of feedback*[6] *to create and correct information learned, and the feedback and correction are becoming increasingly frequent and efficient through evolution—from open-loop reinforcement to closed-loop self-supervision, from collective trial and error to proactive hypothesizing and falsification.*

### 9.3.2 From Inductive to Deductive Intelligence

As discussed in the introduction (Chapter 1), the 1956 "artificial intelligence" (AI) program aimed precisely to study how to endow machines with scientific intelligence, i.e., high-level functions such as mathematical abstraction, causal inference, logical deduction, and problem solving that are believed to differentiate humans from animals:

$$\textbf{low-level} \text{ (animal) intelligence } \Longrightarrow \textbf{high-level} \text{ (human) intelligence.} \tag{9.3.2}$$

As we have clarified repeatedly in this book, most technological advances in machine intelligence over the past decade or so, although carried out under the name "AI", are actually more closely related to having machines imitate low-level forms of intelligence largely shared by both animals and humans, including phylogenetic, ontogenetic, and societal intelligence. At these levels, intelligence creates memory or *empirical* knowledge (from empirical data distributions) and conducts inference (or prediction) that is mostly *inductive* (or Bayesian-like) in nature. Principles and methods introduced in this book aim to reveal scientific objectives and computational mechanisms behind intelligence at these levels. They provide strong theoretical justification and computational evidence for the following claim:

*Machines can learn empirical knowledge and conduct inductive inference.*

So far, however, no rigorous theoretical or scientific evidence suggests that these mechanisms alone would suffice to achieve the high-level human intelligence that the original "artificial intelligence" program truly aimed to understand and wanted for machines to imitate or even surpass.

In fact, as of today, we know little about how to rigorously verify or certify whether a system is truly capable of such high-level intelligence, even though the

[6] Loss for error and reward for success.

Turing Test, also known as the Imitation Game, was proposed in 1950 [Tur50], as an initial attempt to address the fundamental question:

*Can machines think?*

However, in Turing's proposal, the evaluator, or interrogator, is some human. However, most human evaluators have limited scientific training and knowledge, and their conclusions can be rather subjective.

For a long time, a precise definition of such a test was not deemed necessary since machine capabilities were far below those of humans (or even animals). However, given recent technological advances, many models and systems now claim to have reached or even surpassed human intelligence. Therefore, it is high time to develop a scientific and executable definition of the Turing Test, i.e., a systematic and rigorous protocol to evaluate and certify the intelligence level of a model or a system that claims to be "intelligent."

As Turing argued in his original paper, it is rather difficult to define what "thinking" really means. We may instead start with something arguable more tangible and try to determine first whether the system in question can truly "understand" certain abstract concepts or knowledge. For example, we may try to verify whether an intelligent model or system has truly understood an abstract concept such as:

1. the notion of equality between two quantities,

2. the notion of numbers (natural, rational, real, or imaginary),

3. the notion of infinity and mathematical induction,

4. and the notion of logic consistency[7] and proof by contradition,

just to name a few? Or has it simply memorized a massive number of examples of such notions? Note that state-of-the-art large language models still struggle with simple mathematical questions like: "Is 3.11 larger or smaller than 3.9?"[8]

From this book, we know modern generative AI systems primarily rely on Bayesian (hence inductive) inference to produce their answers. Can machines also truly grasp mathematical induction or general logical deduction and understand its difference from Bayesian induction? How do we verify whether a system truly understands the rules of logical and mathematical deduction and can apply them rigorously? Or, again, has it merely memorized statistical

[7] In mathematics, if a set of axioms and their deduced results are consistent and do not contradict one another, then they exist. This is known as Hilber's Principle. This is different from the notion of "consistency" in Chapter 6, which requires the learned representations to be consistent with the empirical data distributions. One notion of consistency is inductive in nature and the other is deductive.

[8] Some models have corrected their answers to such questions through targeted engineering, or have incorporated additional reasoning mechanisms that verify immediate answers and correct them during inference. However, we leave it to the reader as an exercise to rigorously test whether any state-of-the-art language model truly understands the notion of numbers (natural, rational, real, and complex) and their associated arithmetic.

patterns of such deductions from a large number of instances, such as chain-of-thought data, and simply exploited such patterns inductively during inference? Hence the following question remains an open one:

*Is there any difference between **memorizing** and **understanding**?*

Only by clarifying the difference between these two can we possibly provide a meaningful answer to the question: "*Can machines understand?*"

Probably much more importantly, how do we verify whether a self-claimed intelligent system is capable of *creating abstract concepts or improving its own knowledge* proactively via deductive means? For example, can it create new mathematical concepts and discover new physical laws or causal relationships? The connection between understanding and creating can be rather profound, as the physicist Richard Feynman had famously put it:

*"What I cannot create, I do not understand."*

### 9.3.3 Scientific Tests of Intelligence

Hence, it is high time we develop scientifically sound evaluation methods to determine which of the following categories a system's intelligence capability belongs to:

Level-1. having merely learned the distribution of some knowledge-carrying data and being able to regenerate them;

Level-2. being capable of autonomously and continuously learning new knowledge or correcting existing knowledge from new observations and experiences;

Level-3. being able to truly understand and create abstract concepts, rules and laws, and know how to apply them correctly in a deductive manner;

Level-4. being able to generate new scientific hypotheses or math conjectures, and verify or falsify them based on experimentation or logical deduction.

To evaluate and distinguish these different levels of intelligence, we suggest that there should probably be at least three different tests, as illustrated in Figure 9.3:

1. *The Norbert Wiener Test:* to determine whether a system can self-correct and create new (empirical) knowledge on its own or merely receives and updates information passively through external reinforcement or supervision;

2. *The Alan Turing Test:* to determine whether a system can create and understand abstract concepts and knowledge or merely learns and memorizes statistics of the samples and uses them for Bayesian inference;

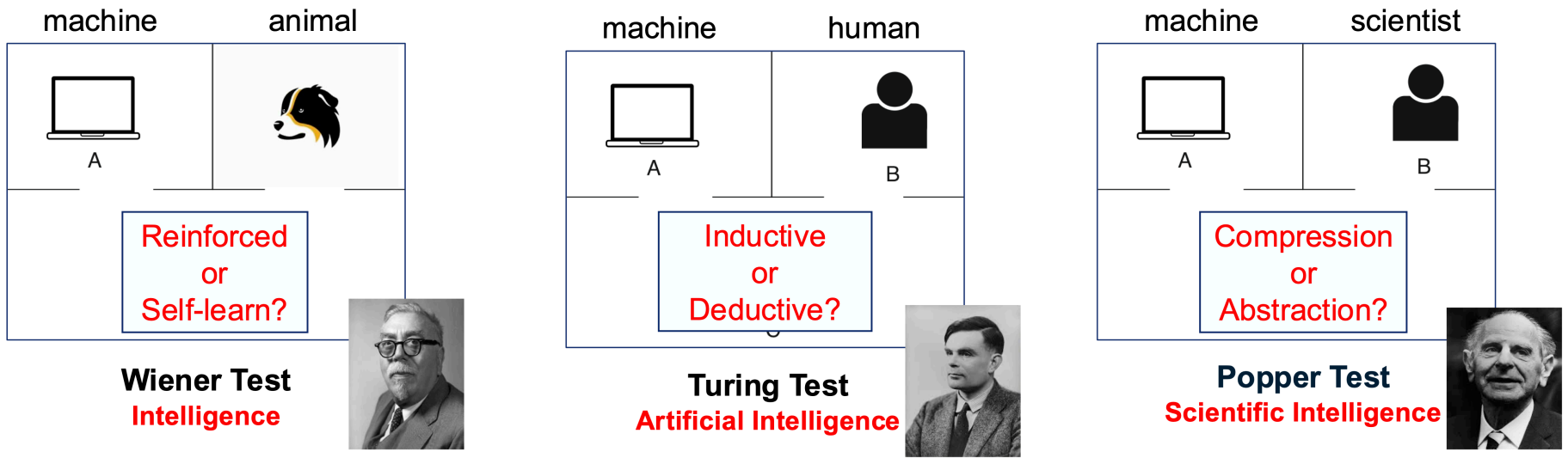

Figure 9.3: Three tests for different levels or types of intelligence capabilities: the Wiener test for basic intelligence, the Turing test for human-level intelligence, and the Popper test for scientist-level intelligence.

3. *The Karl Popper Test:* to determine whether a system can proactively explore and create new abstract knowledge by forming and verifying hypotheses or theories based on experimental consistency or logical self-consistency.

We believe that, for such tests, the evaluator, or interrogator, should not be a single human or an arbitrary group of humans but rather a scientifically certified entity following a scientifically sound protocol and process so that the conclusions would be rigorous and trustworthy.

As we have seen throughout this book, *compression* has played a fundamental role in developing a memory or creating empirical knowledge. It is the governing principle and a universal mechanism for identifying an empirical data distribution and organizing the information encoded therein with a structured representation. To a large extent, it explains the practice of "artificial intelligence" with deep networks, which is largely inductive and Bayesian in nature.[9] An outstanding question for future study is whether *compression alone* is sufficient to achieve all the higher-level deductive intelligence mentioned above. In particular, are mathematical abstraction, causal inference, hypothesis generation, and logical reasoning and deduction some kind of extended or transcended forms of compression? Is there some fundamental difference between learning data distributions through compression and forming high-level concepts and theories through (mathematical) abstraction? Hence we have another open question:

*Is there any difference between* ***compression*** *and* ***abstraction****?*

Only by knowing the difference can we possibly attempt to provide a meaningful answer to the question: "*Can machines create knowledge proactively via deduction?*"

To a large extent, Science—and its associated code book, Mathematics—can be viewed as the most advanced form of intelligence. It is largely deductive in

[9]Systems that have built-in deductive mechanisms for conducting formal verification or logical deduction are not in this category.

nature and unique to educated and enlightened humans. Philosopher Sir Karl Popper once suggested:

> *"Science may be described as the art of systematic oversimplification."*

We believe that uncovering and understanding the underlying mathematical principles and computational mechanisms of such higher-level intelligence, and successfully reproducing them through machines, will be the final frontier for Science, Mathematics, and Computation altogether!

# Appendices

# Appendix A

# Optimization Methods

"*Since the building of all the universe is perfect and is created by the wisdom creator, nothing arises in the universe in which one cannot see the sense of some maximum or minimum.*"

– L. Euler

In this chapter, we give a brief introduction to some of the most basic but important optimization algorithms used in this book. The purpose is only to help the reader apply these algorithms to problems studied in this book, not to gain a deep understanding about these algorithms. Hence, we will not provide a thorough justification for the algorithms introduced, in terms of performance guarantees.

## A.1 Steepest Descent

Optimization is concerned with the question of how to find where a function, say $\mathcal{L}(\theta)$, reaches its minimum value. Mathematically, this is stated as a problem:

$$\arg\min_{\theta \in \Theta} \mathcal{L}(\theta), \tag{A.1.1}$$

where $\Theta$ represents a domain to which the argument $\theta$ is confined. Often (and unless otherwise mentioned, in this chapter) $\Theta$ is simply $\mathbb{R}^n$. Without loss of generality, we assume that here the function $\mathcal{L}(\theta)$ is smooth[1].

The efficiency of finding the (global) minima depends on what information we have about the function $\mathcal{L}$. For most optimization problems considered in this book, the dimension of $\theta$, say $n$, is very large. That makes computing or accessing local information about $\mathcal{L}$ expensive. In particular, since the gradient $\nabla \mathcal{L}$ has $n$ entries, it is often reasonable to compute; however, the Hessian $\nabla^2 \mathcal{L}$ has $n^2$ entries which is often wildly impractical to compute (and the same goes for higher-order derivatives). Hence, it is typical to assume that we have the zeroth-order information, i.e., we are able to evaluate $\mathcal{L}(\theta)$, and the first-order information, i.e., we are able to evaluate $\nabla\mathcal{L}(\theta)$. Optimization theorists may rephrase this as saying we have a "first-order *oracle*." All optimization algorithms that we introduce in this section only use a first-order oracle.[2]

[1] In case the function $\mathcal{L}$ is not smooth, we replace its gradient with a so-called *subgradient*.

### A.1.1 Vanilla Gradient Descent for Smooth Problems

The simplest and most widely used method for optimization is *gradient descent* (GD). It was first introduced by Cauchy in 1847. The idea is very simple: starting from an initial state, we iteratively take small steps such that each step reduces the value of the function $\mathcal{L}(\theta)$.

Suppose that the current state is $\theta$. We want to take a small step, say of distance $h$, in a direction, indicated by a vector $\boldsymbol{v}$, to reach a new state $\theta + h\boldsymbol{v}$ such that the value of the function decreases:

$$\mathcal{L}(\theta + h\boldsymbol{v}) \leq \mathcal{L}(\theta). \tag{A.1.2}$$

To find such a direction $\boldsymbol{v}$, we can approximate $\mathcal{L}(\theta + h\boldsymbol{v})$ through a Taylor expansion around $h = 0$:

$$\mathcal{L}(\theta + h\boldsymbol{v}) = \mathcal{L}(\theta) + h\langle\nabla\mathcal{L}(\theta), \boldsymbol{v}\rangle + o(h), \tag{A.1.3}$$

where the inner product here (and in this chapter) will be the $\ell^2$ inner product, i.e., $\langle\boldsymbol{x}, \boldsymbol{y}\rangle = \boldsymbol{x}^\top\boldsymbol{y}$. To find the direction of *steepest descent*, we attempt to minimize this Taylor expansion among unit vectors $\boldsymbol{v}$. If $\nabla\mathcal{L}(\theta) = \mathbf{0}$, then the second term above is 0 regardless of the value of $\boldsymbol{v}$, so we cannot attempt to make progress, i.e., the algorithm has converged. On the other hand, if $\nabla\mathcal{L}(\theta) \neq \mathbf{0}$ then it holds

$$\operatorname*{arg\,min}_{\substack{\boldsymbol{v}\in\mathbb{R}^d \\ \|\boldsymbol{v}\|_2=1}}[\mathcal{L}(\theta) + h\langle\nabla\mathcal{L}(\theta), \boldsymbol{v}\rangle] = \operatorname*{arg\,min}_{\substack{\boldsymbol{v}\in\mathbb{R}^d \\ \|\boldsymbol{v}\|_2=1}}\langle\nabla\mathcal{L}(\theta), \boldsymbol{v}\rangle = -\frac{\nabla\mathcal{L}(\theta)}{\|\nabla\mathcal{L}(\theta)\|_2}, \tag{A.1.4}$$

In words, this means that the value of $\mathcal{L}(\theta + h\boldsymbol{v})$ decreases the fastest along the direction $\boldsymbol{v} = -\nabla\mathcal{L}(\theta)/\|\nabla\mathcal{L}(\theta)\|_2$, for small enough $h$. This leads to the gradient descent method: From the current state $\theta_k$ ($k = 0, 1, \ldots$), we take a step of size $h$ in the direction of the negative gradient to reach the next iterate,

$$\theta_{k+1} = \theta_k - h\nabla\mathcal{L}(\theta_k). \tag{A.1.5}$$

The step size $h$ is also called the *learning rate* in machine learning contexts.

[2] We refer the readers to the book by [WM22] for a more systematic introduction to optimization algorithms in a high-dimensional space, including algorithms assuming higher-order oracles.

**Step-Size Selection**

The remaining question is what the step size $h$ should be? If we choose $h$ to be too small, the value of the function may decrease very slowly, as shown by the plot in the middle in Figure A.1. If $h$ is too large, the value might not even decrease at all, as shown by the plot on the right in Figure A.1.

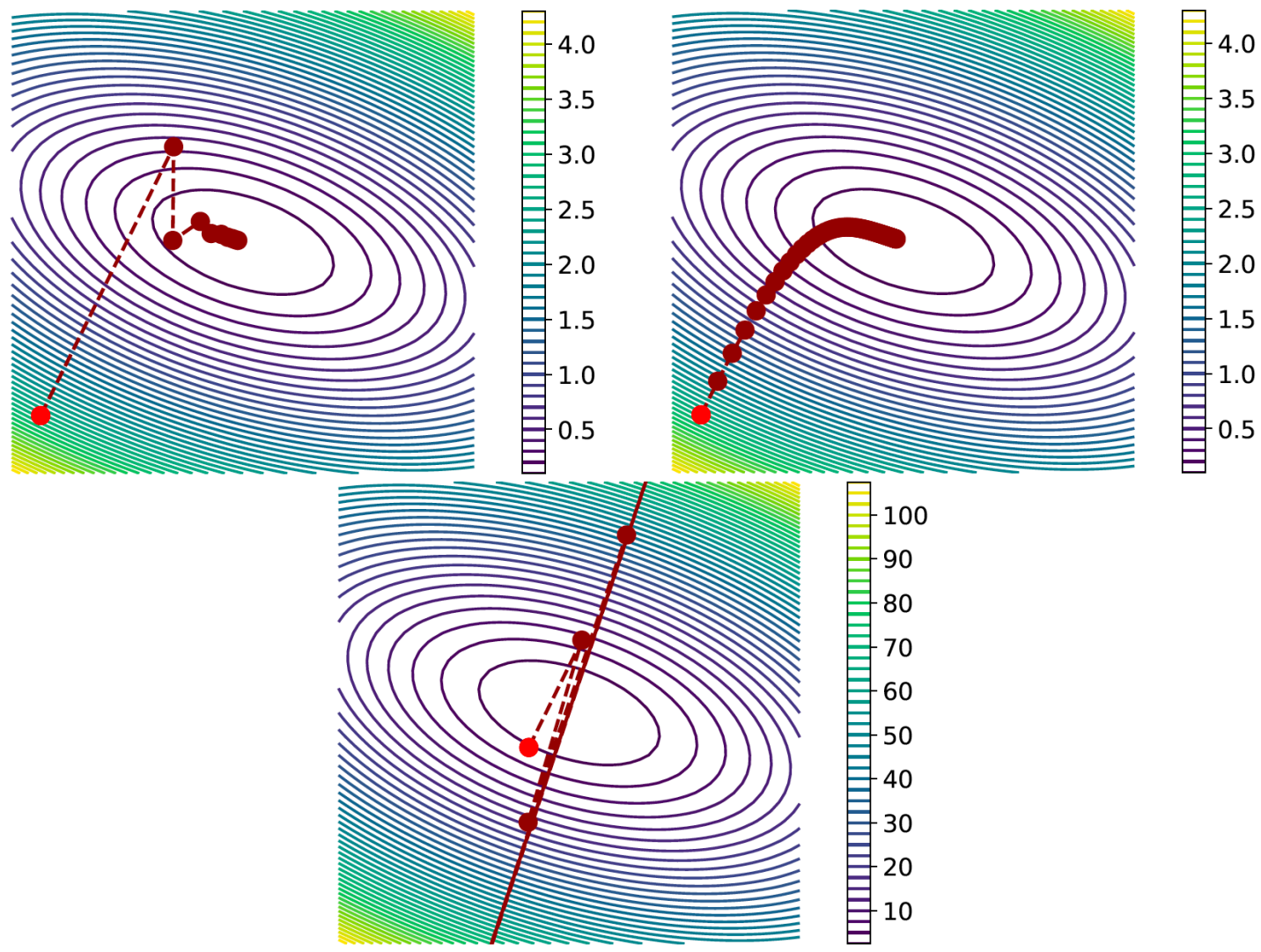

Figure A.1: The effect of the (constant) step size $h$ on the convergence of the gradient descent method.

So the step size $h$ should be chosen based on the landscape of the function $\mathcal{L}(\theta_k)$. Ideally, to choose the best step size $h$, we can solve the following optimization problem over a single variable $h$:

$$h = \arg\min_{h' \geq 0} \mathcal{L}(\theta_k - h' \nabla \mathcal{L}(\theta_k)). \tag{A.1.6}$$

This method of choosing the step size is called *line search*; as hinted by the notation, it is usually used to obtain an optimal learning rate *for each iteration* $k$. However, when the function $\mathcal{L}(\theta_k)$ is complicated, which is usually the case for training a deep neural network, this one-dimensional optimization is very difficult to solve at each iteration of gradient descent.

Then how should we choose a proper step size $h$? One common and classical approach is to try to obtain a good approximation of the local landscape around the current state $\theta$ based on some knowledge about the overall landscape of the function $\mathcal{L}(\theta)$.

Common conditions for the landscape of $\mathcal{L}(\theta)$ include:

- $\alpha$-Strong Convexity. Recall that $\mathcal{L}$ is *$\alpha$-strongly convex* if its graph lies

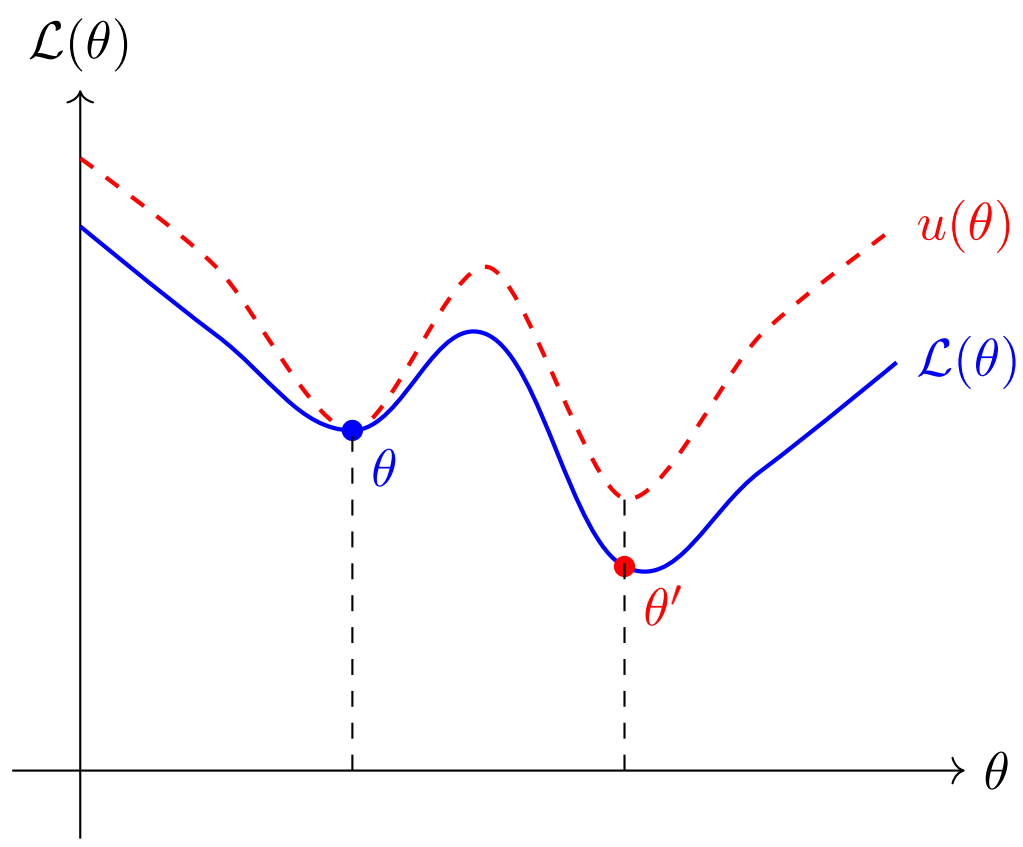

Figure A.2: **Majorization-minimization scheme and intuition.** A function $\mathcal{L}\colon \Theta \to \mathbb{R}$ has a global upper bound $u\colon \Theta \to \mathbb{R}$ which meets $\mathcal{L}$ at at least one point $\theta$. Then, finding the $\theta'$ which minimizes $u$ will improve the value of $\mathcal{L}$ from $\mathcal{L}(\theta)$. Note that similar results can be shown about local upper bounds.

above a global quadratic lower bound of slope $\alpha$, i.e.,

$$\mathcal{L}(\theta') \geq l_{\theta,\alpha}(\theta') \doteq \mathcal{L}(\theta) + \langle \nabla\mathcal{L}(\theta), \theta' - \theta\rangle + \frac{\alpha}{2}\|\theta' - \theta\|_2^2 \tag{A.1.7}$$

for any "base point" $\theta$. We say that $\mathcal{L}$ is *convex* if it is 0-strongly convex, i.e., its graph lies above its tangents. It is easy to show (proof as exercise) that strongly convex functions have unique global minima. Another important fact (proof as exercise) is that $\alpha$-strongly convex twice-differentiable functions $\mathcal{L}$ have (symmetric) Hessians $\nabla^2\mathcal{L}$ whose minimum eigenvalue is $\geq \alpha$. For $\alpha > 0$ this implies the Hessian is symmetric positive definite, and for $\alpha = 0$ (i.e., $\mathcal{L}$ is convex) this implies that the Hessian is symmetric positive semidefinite.

- $\beta$-Lipschitz Gradient (also called $\beta$-Smoothness). Recall that $\mathcal{L}$ has $\beta$-*Lipschitz gradient* if $\nabla\mathcal{L}$ exists and is $\beta$-Lipschitz, i.e.,

$$\|\nabla\mathcal{L}(\theta') - \nabla\mathcal{L}(\theta)\|_2 \leq \beta\|\theta' - \theta\|_2. \tag{A.1.8}$$

for any "base point" $\theta$. It is easy to show (proof as exercise) that this is equivalent to $\mathcal{L}$ having a global quadratic upper bound of slope $\beta$, i.e.,

$$\mathcal{L}(\theta') \leq u_{\theta,\beta}(\theta') \doteq \mathcal{L}(\theta) + \langle \nabla\mathcal{L}(\theta), \theta' - \theta\rangle + \frac{\beta}{2}\|\theta' - \theta\|_2^2. \tag{A.1.9}$$

for any "base point" $\theta$. Another important fact (proof as exercise) is that convex $\beta$-Lipschitz gradient twice-differentiable functions have (symmetric) Hessians $\nabla^2\mathcal{L}$ whose largest eigenvalue is $\leq \beta$.

First, let us suppose that $\mathcal{L}$ has $\beta$-Lipschitz gradient (but is not necessarily even convex). We will use this occasion to introduce a common optimization theme: *to minimize $\mathcal{L}$, we can minimize an upper bound on $\mathcal{L}$*, which is justified by the following lemma visualized in Figure A.2.

**Lemma A.1** (Majorization-Minimization)**.** *Suppose that $u\colon \Theta \to \mathbb{R}$ is a global upper bound on $\mathcal{L}$, namely $\mathcal{L}(\theta) \leq u(\theta)$ for all $\theta \in \Theta$. Suppose that they meet with equality at $\theta$, i.e., $\mathcal{L}(\theta) = u(\theta)$. Then*

$$\theta^{+} \in \arg\min_{\theta' \in \Theta} u(\theta') \implies \mathcal{L}(\theta^{+}) \leq u(\theta^{+}) \leq u(\theta) = \mathcal{L}(\theta). \tag{A.1.10}$$

We will use this lemma to show that we can use the Lipschitz gradient property to ensure that each gradient step cannot worsen the value of $\mathcal{L}$. Indeed, at every base point $\theta$, we have that $u_{\theta,\beta}$ is a global upper bound on $\mathcal{L}$, and $u_{\theta,\beta}(\theta) = \mathcal{L}(\theta)$. Hence by Lemma A.1

$$\text{if } \theta^{+} \text{ minimizes } u_{\theta,\beta} \text{ then } \quad \mathcal{L}(\theta^{+}) \leq u_{\theta,\beta}(\theta^{+}) \leq u_{\theta,\beta}(\theta) = \mathcal{L}(\theta). \tag{A.1.11}$$

This motivates us, when finding an update to obtain $\theta_{k+1}$ from $\theta_k$, we can instead minimize the upper bound $u_{\theta_k,\beta}$ over $\theta$ and set that to be $\theta_{k+1}$. By minimizing $u_{\theta_k,\beta}$ (proof as exercise) we get

$$\theta_{k+1} = \theta_k - \frac{1}{\beta}\nabla\mathcal{L}(\theta_k) \implies \mathcal{L}(\theta_{k+1}) \leq \mathcal{L}(\theta_k). \tag{A.1.12}$$

This implies that a step size $h = 1/\beta$ is a usable learning rate, but it does not provide a convergence rate or certify that $\mathcal{L}(\theta_k)$ actually converges to $\min_\theta \mathcal{L}(\theta)$. This requires a little more rigor, which we now pursue.

Now, let us suppose that $\mathcal{L}$ is $\alpha$-strongly convex, has $\beta$-Lipschitz gradient, and has global optimum $\theta^\star$. We will show that $\theta_k$ will converge directly to the unique global optimum $\theta^\star$, which is a very strong form of convergence. In particular, we will bound $\|\theta^\star - \theta_k\|_2$ using both strong convexity and Lipschitzness of the gradient of $\mathcal{L}$, i.e., taking a look at the neighborhood around $\theta_k$:[3]

$$\begin{aligned}
\|\theta^\star - \theta_{k+1}\|_2^2 &\leq \|\theta^\star - \theta_k + h\nabla\mathcal{L}(\theta_k)\|_2^2 && \text{(A.1.13)}\\
&= \|\theta^\star - \theta_k\|_2^2 + 2h\langle \nabla\mathcal{L}(\theta_k), \theta^\star - \theta_k\rangle + h^2\|\nabla\mathcal{L}(\theta_k)\|_2^2 && \text{(A.1.14)}\\
&\leq \|\theta^\star - \theta_k\|_2^2 + 2h\left(\mathcal{L}(\theta^\star) - \mathcal{L}(\theta_k) - \frac{\alpha}{2}\|\theta^\star - \theta_k\|_2^2\right) + h^2\|\nabla\mathcal{L}(\theta_k)\|_2^2 \quad (\alpha\text{-SC}) && \text{(A.1.15)}\\
&= (1 - \alpha h)\,\|\theta^\star - \theta_k\|_2^2 + 2h(\mathcal{L}(\theta^\star) - \mathcal{L}(\theta_k)) + h^2\|\nabla\mathcal{L}(\theta_k)\|_2^2 && \text{(A.1.16)}\\
&\leq (1 - \alpha h)\,\|\theta^\star - \theta_k\|_2^2 + 2h(\mathcal{L}(\theta^\star) - \mathcal{L}(\theta_k)) + 2h^2\beta(\mathcal{L}(\theta_k) - \mathcal{L}(\theta^\star)) \quad (\beta\text{-LG}) && \text{(A.1.17)}
\end{aligned}$$

[3]In this proof the $\beta$-Lipschitz Gradient invocation step is a little non-trivial. We also leave this step as an exercise, with the hint to plug in $\theta = \theta_0 - h\nabla\mathcal{L}(\theta_0)$ into the Lipschitz gradient identity.

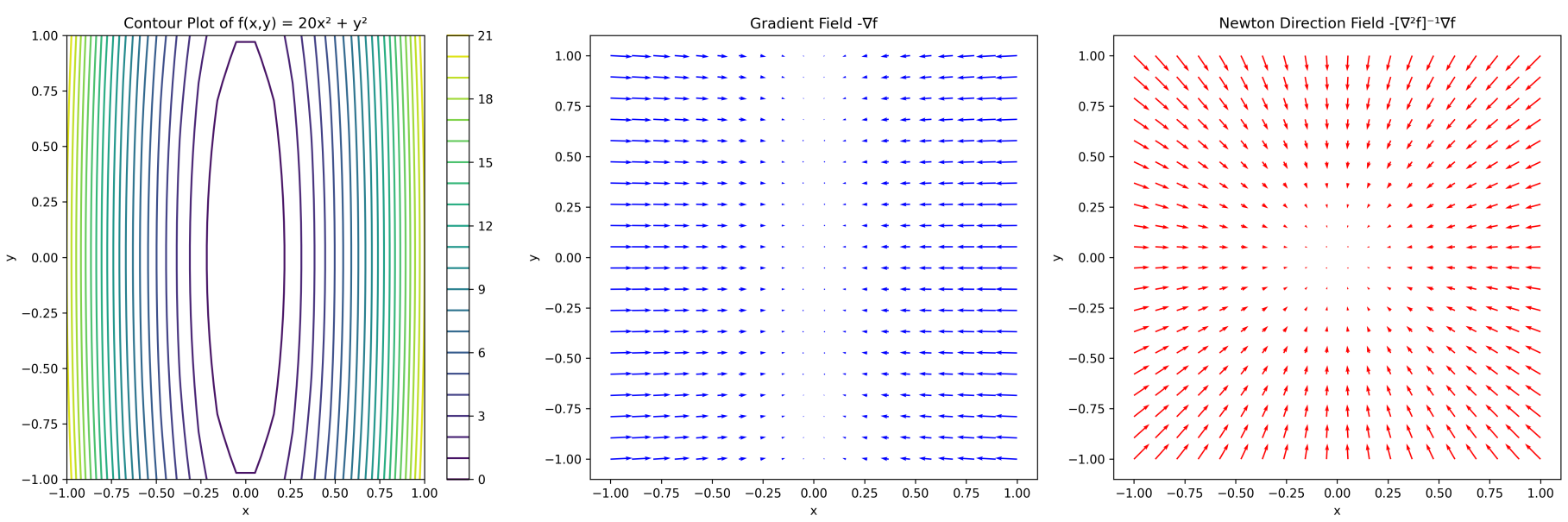

Figure A.3: **The negative gradient $-\nabla\mathcal{L}_\lambda$ and pre-conditioned (Newton's method step) vector field** $-[\nabla^2\mathcal{L}_\lambda]^{-1}[\nabla\mathcal{L}_\lambda]$ where $\lambda = 19$. There is a section of the space where following the negative gradient vector field makes very little progress towards finding the minimum, but in all cases following the Newton's method vector field achieves equal speed of progress towards the optimum since the gradient is whitened. Since the Hessian here is diagonal, adaptive learning rate algorithms (e.g. Adam, as will be discussed later in the section) can make similar progress as Newton's method, but a non-axis-aligned Hessian may even prevent Adam from succeeding quickly.

$$= (1 - \alpha h)\,\|\theta^\star - \theta_k\|_2^2 - 2h(1 - \beta h)(\mathcal{L}(\theta_k) - \mathcal{L}(\theta^\star)). \tag{A.1.18}$$

In order to ensure that the gradient descent iteration makes progress we must pick the step size so that $1 - \beta h \geq 0$, i.e., $h \leq 1/\beta$. If such a setting occurs, then

$$\begin{aligned} \|\theta^\star - \theta_{k+1}\|_2^2 &\leq (1 - \alpha h)\|\theta^\star - \theta_k\|_2^2 \leq (1 - \alpha h)^2\|\theta^\star - \theta_{k-1}\|_2^2 \leq \cdots && \text{(A.1.19)} \\ &\leq (1 - \alpha h)^{k+1}\|\theta^\star - \theta_0\|_2^2. && \text{(A.1.20)} \end{aligned}$$

In order to minimize the right-hand side, we can set $h = 1/\beta$, which obtains

$$\|\theta^\star - \theta_{k+1}\|_2^2 \leq (1 - \alpha/\beta)^{k+1}\|\theta^\star - \theta_0\|_2^2, \tag{A.1.21}$$

showing convergence to global optimum with exponentially decaying error. Notice that here we used a convergence rate to obtain a favorable *step size* of $h = 1/\beta$. This motif will re-occur in this section.

We end this section with a caveat: learning a global optimum is (usually) impractically hard. Under certain conditions, we can ensure that the gradient descent iterates converge to a *local optimum*. Also, under more relaxed conditions, we can ensure *local* convergence, i.e., that the iterates converge to a (global or local) optimum if the sequence is initialized close enough to the optimum.

### A.1.2 Preconditioned Gradient Descent for Badly-Conditioned Problems

#### Newton's Method

There are some smooth problems and strongly convex problems on which gradient descent nonetheless does quite poorly. For example, let $\lambda \geq 0$ and let $\mathcal{L}_\lambda \colon \mathbb{R}^2 \to \mathbb{R}$ of the form

$$\mathcal{L}_\lambda(\theta) = \mathcal{L}_\lambda\left(\begin{bmatrix}\theta_1 \\ \theta_2\end{bmatrix}\right) \doteq \frac{1}{2}\left\{(1+\lambda)\theta_1^2 + \theta_2^2\right\} = \frac{1}{2}\theta^\top \begin{bmatrix}1+\lambda & 0 \\ 0 & 1\end{bmatrix}\theta. \tag{A.1.22}$$

This problem is 1-strongly convex and has $(1+\lambda)$-Lipschitz gradient. The convergence rate is then geometric with rate $1 - 1/(1+\lambda)$. For large $\lambda$, this is still not very fast. In this section, we will introduce a class of optimization problems which can successfully optimize such badly-conditioned functions.

The key lies in the objective's *curvature*, which is given by the Hessian. Suppose that (as a counterfactual) we had a *second-order* oracle which would allow us to compute $\mathcal{L}(\theta)$, $\nabla\mathcal{L}(\theta)$, and $\nabla^2\mathcal{L}(\theta)$. Then, instead of picking a descent direction $\boldsymbol{v}$ to optimize the first-order Taylor expansion around $\theta$, we could optimize the second-order Taylor expansion instead. Intuitively this would allow us to incorporate curvature information into the update.

Let us carry out this computation. The second-order Taylor expansion of $\mathcal{L}(\theta + h\boldsymbol{v})$ around $h = 0$ is

$$\mathcal{L}(\theta + h\boldsymbol{v}) = \mathcal{L}(\theta) + h\langle\nabla\mathcal{L}(\theta), \boldsymbol{v}\rangle + \frac{1}{2}h^2\langle[\nabla^2\mathcal{L}(\theta)]\boldsymbol{v}, \boldsymbol{v}\rangle + o(h^2). \tag{A.1.23}$$

Then we can compute the descent direction:

$$\begin{aligned} &\operatorname*{arg\,min}_{\substack{\boldsymbol{v}\in\mathbb{R}^n \\ \|\boldsymbol{v}\|_2=1}} \left[\mathcal{L}(\theta) + h\langle\nabla\mathcal{L}(\theta), \boldsymbol{v}\rangle + \frac{1}{2}h^2\langle[\nabla^2\mathcal{L}(\theta)]\boldsymbol{v}, \boldsymbol{v}\rangle\right] && \text{(A.1.24)} \\ &= \operatorname*{arg\,min}_{\substack{\boldsymbol{v}\in\mathbb{R}^n \\ \|\boldsymbol{v}\|_2=1}} \left[\langle\nabla\mathcal{L}(\theta), \boldsymbol{v}\rangle + \frac{1}{2}h\langle[\nabla^2\mathcal{L}(\theta)]\boldsymbol{v}, \boldsymbol{v}\rangle\right]. && \text{(A.1.25)} \end{aligned}$$

This optimization problem is a little difficult to solve because of the constraint $\|\boldsymbol{v}\|_2 = 1$. But in practice we never normalize the descent direction $\boldsymbol{v}$ and use the step size $h$ to control the size of the update. So let us just solve the above problem over all vectors $\boldsymbol{v} \in \mathbb{R}^n$:[4]

$$\operatorname*{arg\,min}_{\boldsymbol{v}\in\mathbb{R}^n} \left[\langle\nabla\mathcal{L}(\theta), \boldsymbol{v}\rangle + \frac{1}{2}h\langle[\nabla^2\mathcal{L}(\theta)]\boldsymbol{v}, \boldsymbol{v}\rangle\right] = -\frac{1}{h}[\nabla^2\mathcal{L}(\theta)]^{-1}[\nabla\mathcal{L}(\theta)]. \tag{A.1.26}$$

We can thus use the steepest descent iteration

$$\theta_{k+1} = \theta_k - [\nabla^2\mathcal{L}(\theta_k)]^{-1}[\nabla\mathcal{L}(\theta_k)], \tag{A.1.27}$$

[4] If $\nabla^2\mathcal{L}(\theta)$ is not invertible, then we can replace $[\nabla^2\mathcal{L}(\theta)]^{-1}$ with the Moore-Penrose pseudoinverse of $\nabla^2\mathcal{L}(\theta)$.

(this is the celebrated *Newton's method*), or

$$\theta_{k+1} = \theta_k - h[\nabla^2 \mathcal{L}(\theta_k)]^{-1}[\nabla \mathcal{L}(\theta_k)], \tag{A.1.28}$$

(which is called *underdamped Newton's method*). Since the second-order quadratic $\mathcal{L}_\lambda$ is equal to its second-order Taylor expansion, if we run Newton's method for *one step*, we will achieve the global minimum in *one step* no matter how large $\lambda$ is. Figure A.3 gives some intuition about poorly conditioned functions and the gradient steps versus Newton's steps.

### PGD

In practice, we do *not* have a second-order oracle which allows us to compute $\nabla^2 \mathcal{L}(\theta)$. Instead, we can attempt to *learn an approximation to it* alongside the parameter update $\theta_{k+1}$ from $\theta_k$.

How do we learn an approximation to it? We shall find some equations which the Hessian's inverse satisfies and then try to update our approximation so that it satisfies the equations. Namely, taking the Taylor series of $\nabla \mathcal{L}(\theta + \delta_\theta)$ around point $\theta$, we obtain

$$\underbrace{\nabla \mathcal{L}(\theta + \delta_\theta) - \nabla \mathcal{L}(\theta)}_{\doteq \delta_{\boldsymbol{g}}} = [\nabla^2 \mathcal{L}(\theta)]\delta_\theta + o(\|\delta_\theta\|_2). \tag{A.1.29}$$

In this case we have

$$\delta_{\boldsymbol{g}} \approx [\nabla^2 \mathcal{L}(\theta)]\delta_\theta \implies \delta_\theta \approx [\nabla^2 \mathcal{L}(\theta)]^{-1}\delta_{\boldsymbol{g}} \tag{A.1.30}$$

We can now try to learn a symmetric positive semidefinite pre-conditioner $P \in \mathbb{R}^{n \times n}$ such that

$$\delta_\theta \approx P\delta_{\boldsymbol{g}}, \tag{A.1.31}$$

updating it at each iteration along with $\theta_k$. Namely, we have the *PSGD* iteration

$$P_k = \mathrm{PreconditionerUpdate}(P_{k-1}; \theta_k, \nabla \mathcal{L}(\theta_k)) \tag{A.1.32}$$

$$\theta_{k+1} = \theta_k - hP_k \nabla \mathcal{L}(\theta_k). \tag{A.1.33}$$

This update has two problems: how can we even use $P$ (since we already said we cannot store an $n \times n$ matrix) and how can we *update* $P$ at each iteration? The answers are very related; we can never materialize $P$ in computer memory, but we can represent it using a low-rank factorization (or comparable methods such as *Kronecker factorization* which is particularly suited to the form of deep neural networks). Then the preconditioner update step is designed to exploit the structure of the preconditioner representation.

We end this subsection with a caveat: in deep learning, for example, $\mathcal{L}$ is not a convex function and so Newton's method (and approximations to it) do not make sense. In this case we look at the geometric intuition of Newton's method on convex functions, say from Figure A.3: the inverse Hessian *whitens* the gradients. Thus instead of a Hessian-approximating preconditioner, we can adjust the above procedures to learn a more general whitening transformation for the gradient. This is the idea behind the original proposal of PSGD [Li17], which contains more information about how to store and update the preconditioner, and more modern optimizers like Muon [LSY+25].

### A.1.3 Proximal Gradient Descent for Non-Smooth Problems

Even in very toy problems, however, such as LASSO or dictionary learning, the problem is not strongly convex but rather just convex, and the objective is no longer just smooth but rather the sum of a smooth function and a non-smooth regularizer (such as the $\ell^1$ norm). Such problems are solved by *proximal optimization algorithms*, which generalize steepest descent to non-smooth objectives.

Formally, let us say that

$$\mathcal{L}(\theta) \doteq \mathcal{S}(\theta) + \mathcal{R}(\theta) \tag{A.1.34}$$

where $\mathcal{S}$ is smooth, say with $\beta$-Lipschitz gradient, and $\mathcal{R}$ is non-smooth (i.e., rough). The proximal gradient algorithm generalizes the steepest descent algorithm, by using the majorization-minimization framework (i.e., Lemma A.1) with a different global upper bound. Namely, we construct such an upper bound by asking: what if we take the Lipschitz gradient upper bound of $S$ but *leave $R$ alone*? Namely, we have

$$\mathcal{L}(\theta') = \mathcal{S}(\theta') + \mathcal{R}(\theta') \leq u_{\theta,\beta}(\theta') \doteq \mathcal{S}(\theta) + \langle \nabla \mathcal{S}(\theta), \theta' - \theta \rangle + \frac{\beta}{2}\|\theta' - \theta\|_2^2 + \mathcal{R}(\theta'). \tag{A.1.35}$$

Note that (proof as exercise)

$$\arg\min_{\theta'} u_{\theta,\beta}(\theta') = \arg\min_{\theta'} \left[\frac{\beta}{2}\left\|\theta' - \left(\theta - \frac{1}{\beta}\nabla \mathcal{S}(\theta)\right)\right\|_2^2 + \mathcal{R}(\theta')\right]. \tag{A.1.36}$$

Now if we try to minimize the upper bound $u_{\theta,\beta}$, we are picking a $\theta'$ that:

- is close to the gradient update $\theta - \frac{1}{\beta}\nabla \mathcal{S}(\theta)$;
- has a small value of the regularizer $\mathcal{R}(\theta')$

and trades off these properties according to the smoothness parameter $\beta$. Accordingly, let us define the proximal operator

$$\operatorname{prox}_{h,\mathcal{R}}(\theta) \doteq \arg\min_{\theta'} \left[\frac{1}{2h}\|\theta' - \theta\|_2^2 + \mathcal{R}(\theta')\right]. \tag{A.1.37}$$

Then, we can define the *proximal gradient descent* iteration which, at each iteration, minimizes the upper bound $u_{\theta_k,h^{-1}}$, i.e.,

$$\theta_{k+1} = \operatorname{prox}_{h,\mathcal{R}}(\theta_k - h\nabla \mathcal{S}(\theta_k)). \tag{A.1.38}$$

Convergence proofs are possible when $h \leq 1/\beta$, but we do not give any in this section.

One remaining question is: how can we compute the proximal operator? At first glance, it seems like we have traded one intractable minimization problem for another. Since we have not made any assumption on $\mathcal{R}$ so far, the framework works even when $\mathcal{R}$ is a very complex function (such as a neural network loss), which would require us to solve a neural network training problem in order to compute a single proximal operator. However, in practice, for simple regularizers $\mathcal{R}$ such as those we use in this manuscript, there exist proximal operators which are easy to compute or even in closed-form. We give a few below (the proofs are an exercise).

*Example* A.1. Let $\Gamma \subseteq \Theta$ be a set, and let $\chi_\Gamma$ be the characteristic function on $\Gamma$, i.e.,

$$\chi_\Gamma(\theta) \doteq \begin{cases} 0, & \text{if } \theta \in \Gamma \\ +\infty, & \text{if } \theta \notin \Gamma. \end{cases} \tag{A.1.39}$$

Then the proximal operator of $\chi_\Gamma$ is a projection, i.e.,

$$\operatorname{prox}_{h,\chi_\Gamma}(\theta) = \arg\min_{\theta' \in \Gamma} \frac{1}{2}\|\theta' - \theta\|_2^2 = \arg\min_{\theta' \in \Gamma} \|\theta' - \theta\|_2. \tag{A.1.40}$$

■

*Example* A.2. The $\ell^1$ norm has a proximal operator which performs soft thresholding:

$$S_h(\theta) \doteq \operatorname{prox}_{h,\lambda\|\cdot\|_1}(\theta) = \arg\min_{\theta'} \left[\frac{1}{2h}\|\theta' - \theta\|_2^2 + \lambda\|\theta'\|_1\right] \tag{A.1.41}$$

then $S_h(\theta)$ is defined coordinate-wise by

$$S_h(\theta)_i = \begin{cases} \theta_i - h\lambda, & \text{if } \theta_i \geq h\lambda \\ 0, & \text{if } \theta_i \in [-h\lambda, h\lambda] \\ \theta_i + h\lambda, & \text{if } \theta_i \leq -h\lambda \end{cases} = \begin{cases} \max\{|\theta_i| - h\lambda, 0\}\operatorname{sign}(\theta_i), & \text{if } |\theta_i| \geq h\lambda \\ 0, & \text{if } |\theta_i| < h\lambda. \end{cases} \tag{A.1.42}$$

The proximal gradient operation with the smooth part of the objective being least-squares and the non-smooth part being the $\ell^1$ norm (hence using this soft thresholding proximal operator) is called the Iterative Shrinkage-Thresholding Algorithm (ISTA). ■

*Example* A.3. In Chapter 5 we use a proximal operator corresponding to the $\ell^1$ norm plus the characteristic function for the positive orthant $\mathbb{R}^n_+ \doteq \{\boldsymbol{x} \in \mathbb{R}^n : x_i \geq 0 \ \forall i\}$, namely

$$T_h(\theta) \doteq \operatorname{prox}_{h,\lambda\|\cdot\|_1 + \chi_{\mathbb{R}^n_+}}(\theta) = \arg\min_{\theta' \in \mathbb{R}^n_+} \left[\frac{1}{2h}\|\theta' - \theta\|_2^2 + \lambda\|\theta'\|_1\right], \tag{A.1.43}$$

then $T_h$ is defined as

$$T_h(\theta)_i \doteq \max\{\theta_i - h\lambda, 0\}. \tag{A.1.44}$$

This proximal operator yields the non-negative ISTA that is invoked in Chapter 5 and beyond. ■

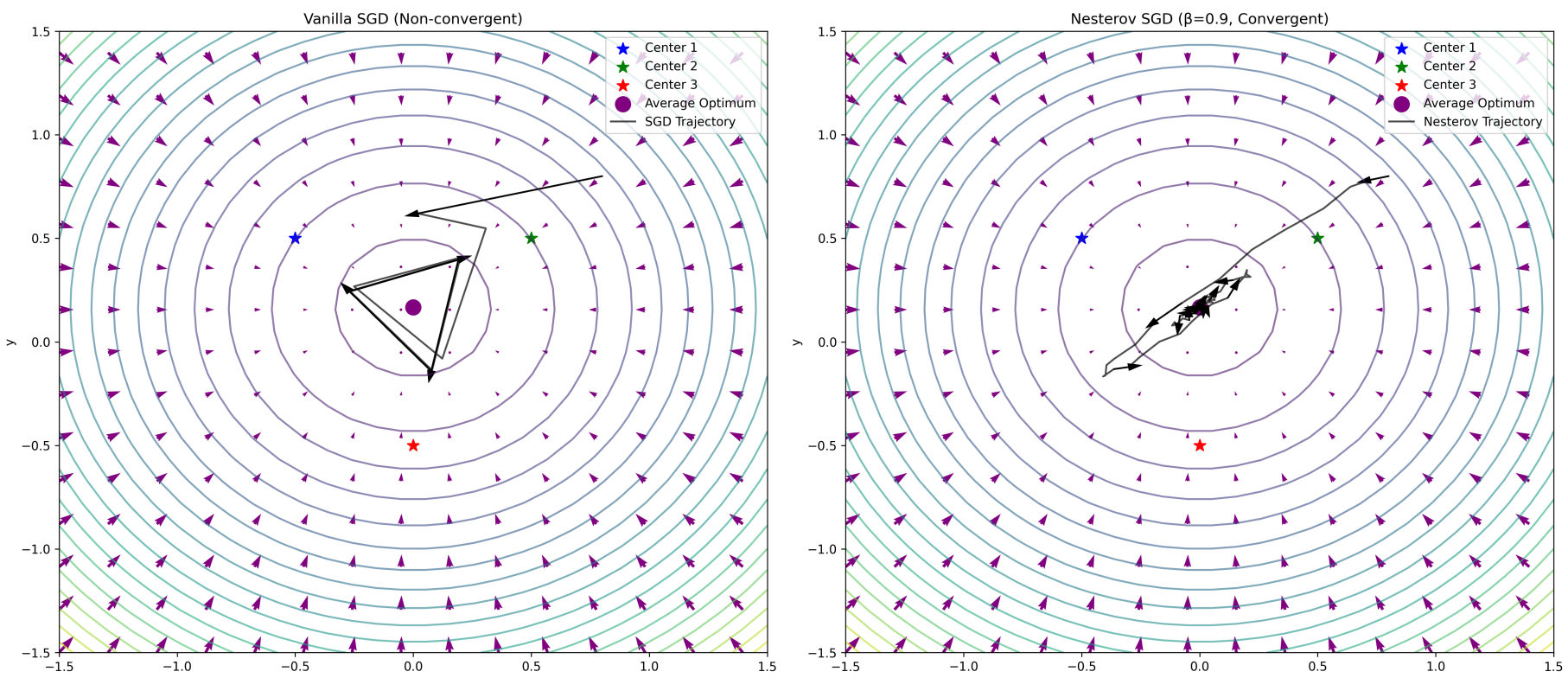

Figure A.4: **Stochastic gradient descent may not converge, even for very benign objectives, but momentum SGD converges.** For even simple quadratic objectives, stochastic gradient descent iterates may pinball around the global optimum, whereas momentum-averaged gradients align to point to the optimal value.

### A.1.4 Stochastic Gradient Descent for Large-Scale Problems

In deep learning, the objective function $\mathcal{L}$ usually cannot be computed exactly, and instead at each optimization step it is *estimated* using finite samples (say, using a mini-batch). A common way to model this situation is to define a *stochastic loss function* $\mathcal{L}_\omega(\theta)$ where $\omega$ is some "source of randomness". For example, $\omega$ could contain the indices of the samples in a batch over which to compute the loss. Then, we would like to minimize $\mathcal{L}(\theta) \doteq \mathbb{E}_\omega[\mathcal{L}_\omega(\theta)]$ over $\theta$, given access to a *stochastic first-order oracle*: given $\theta$, we can sample $\omega$ and compute $\mathcal{L}_\omega(\theta)$ and $\nabla_\theta \mathcal{L}_\omega(\theta)$. This minimization problem is called a *stochastic optimization problem*.

The basic first-order stochastic algorithm is *stochastic gradient descent*: at each iteration $k$ we sample $\omega_k$, define $\mathcal{L}_k \doteq \mathcal{L}_{\omega_k}$, and perform a gradient step on $\mathcal{L}_k$, i.e.,

$$\theta_{k+1} = \theta_k - h\nabla \mathcal{L}_k(\theta_k). \tag{A.1.45}$$

However, even for very simple problems we cannot expect the same type of convergence as we obtained in gradient descent. For example, suppose that there are $m$ possible values for $\omega \in \{1, \dots, m\}$ which it takes with equal probability, and there are $m$ possible targets $\xi_1, \dots, \xi_m$, such that the loss function $\mathcal{L}_\omega$ is

$$\mathcal{L}_\omega(\theta) \doteq \frac{1}{2}\|\theta - \xi_\omega\|_2^2. \tag{A.1.46}$$

Then $\arg\min_\theta \mathbb{E}[\mathcal{L}_\omega(\theta)] = \frac{1}{m}\sum_{i=1}^m \xi_i$, but stochastic gradient descent can "pinball" around the global optimum value, and not converge, as visualized in Figure A.4.

In order to fix this, we can either average the parameters $\theta_k$ or average the gradients $\nabla\mathcal{L}_k(\theta_k)$ over time. If we average the parameters $\theta_k$, then (using Figure A.4 as a mental model) the issue of pinballing is straightforwardly not possible, since the average iterate will grow closer to the center. As such, most theoretical convergence proofs consider the convergence of the average iterate $\frac{1}{k}\sum_{i=0}^{k}\theta_i$ to the global minimum. If we average the gradients, we will eventually learn an average gradient $\frac{1}{k}\sum_{i=0}^{k}\nabla\mathcal{L}_k(\theta_k)$ which does not change much at each iteration and therefore does not pinball.

In practice, instead of using an arithmetic average, we take an *exponentially moving average* (EMA) of the parameters (this is called *Polyak averaging*) or of the gradients (this is called *momentum*). Momentum is more popular and we will study it here.

A stochastic gradient descent iteration with momentum is as follows:

$$\boldsymbol{g}_k = \beta\boldsymbol{g}_{k-1} + (1-\beta)\nabla\mathcal{L}_k(\theta_k) \tag{A.1.47}$$

$$\theta_{k+1} = \theta_k - h\boldsymbol{g}_k. \tag{A.1.48}$$

We do not go through a convergence proof (see Chapter 7 of [GG23] for an example). However, momentum handles our toy case in Figure A.4 easily (see the right-hand figure): it stops pinballing and eventually converges to the global optimum.

We end with a caveat: one can show that Polyak momentum and Nesterov momentum are equivalent, for certain choices of parameter settings. Then it is also possible to show that a decaying learning rate schedule (i.e., the learning rate $h$ depends on the iteration $k$, and its limit is $h_k \to 0$ as $k \to \infty$) with plain SGD (or PSGD) can mimic the effect of momentum. Namely, [DCM+23] shows that if the SGD algorithm lasts $K$ iterations, the gradient norms are bounded $\|\nabla\mathcal{L}(\theta_k)\|_2 \leq G$, and we define $D \doteq \|\theta_0 - \theta^\star\|_2$, then plain SGD iterates $\theta_k$ satisfy the rate $\mathbb{E}[\mathcal{L}(\theta_k) - \mathcal{L}(\theta^\star)] \leq DG/\sqrt{K}$ — but only so long as the learning rate $h_k = (D/[G\sqrt{K}])(1-k/K)$ decays *linearly* with time. This matches learning rate schedules used in practice. Indeed, surprisingly, such a theory of convex optimization can predict many empirical phenomena in deep networks [SHT+25], despite deep learning optimization being highly non-convex and non-smooth in the worst case. It is so far unclear why this is the case.

## A.1.5 Putting Everything Together: Adam

The gradient descent scheme proposes an iteration of the form

$$\theta_{k+1} = \theta_k + h\boldsymbol{v}_k, \tag{A.1.49}$$

where (recall) $\boldsymbol{v}_k$ is chosen to be (proportional to) the steepest descent vector in the Euclidean norm:

$$\boldsymbol{v}_k = -\frac{\nabla\mathcal{L}(\theta_k)}{\|\nabla\mathcal{L}(\theta_k)\|_2} \in \arg\min_{\substack{\boldsymbol{v}\in\mathbb{R}^n \\ \|\boldsymbol{v}\|_2=1}} \langle\nabla\mathcal{L}(\theta_k), \boldsymbol{v}\rangle. \tag{A.1.50}$$

However, in the context of deep learning optimization, there is absolutely nothing which implies that we have to use the Euclidean norm; indeed the "natural geometry" of the space of parameters is not well-respected by the Euclidean norm, since small changes in the parameter space can lead to very large differences in the output space, for a particular fixed input to the network. If we were instead to use a generic norm $\|\cdot\|$ on the parameter space $\mathbb{R}^n$, we would get some other quantity corresponding to the so-called *dual norm*:

$$\boldsymbol{v}_k \in \arg\min_{\substack{\boldsymbol{v}\in\mathbb{R}^n \\ \|\boldsymbol{v}\|=1}} \langle \nabla\mathcal{L}(\theta_k), \boldsymbol{v}\rangle. \tag{A.1.51}$$

For instance, if we were to use the $\ell^\infty$-norm, it is possible to show that

$$\boldsymbol{v}_k = -\operatorname{sign}(\nabla\mathcal{L}(\theta_k)) \in \arg\min_{\substack{\boldsymbol{v}\in\mathbb{R}^n \\ \|\boldsymbol{v}\|_\infty=1}} \langle \nabla\mathcal{L}(\theta_k), \boldsymbol{v}\rangle, \tag{A.1.52}$$

where $\operatorname{sign}(\boldsymbol{x})_i = \operatorname{sign}(x_i) \in \{-1, 0, 1\}$. Thus if we were so-inclined, we could use the so-called *sign-gradient descent*:

$$\theta_{k+1} = \theta_k - h\operatorname{sign}(\nabla\mathcal{L}(\theta_k)). \tag{A.1.53}$$

From sign-gradient descent, we can derive the famous Adam optimization algorithm. Note that for a scalar $x \in \mathbb{R}$ we can write

$$\operatorname{sign}(x) = \frac{x}{|x|} = \frac{x}{\sqrt{x^2}}. \tag{A.1.54}$$

Similarly, for a vector $\boldsymbol{x} \in \mathbb{R}^n$ we write (where $\odot$ and $\oslash$ are element-wise multiplication and division)

$$\operatorname{sign}(\boldsymbol{x}) = \boldsymbol{x} \oslash [\boldsymbol{x}^{\odot 2}]^{\odot(1/2)}. \tag{A.1.55}$$

Using this representation we can write (A.1.53) as

$$\theta_{k+1} = \theta_k - h([\nabla\mathcal{L}(\theta_k)] \oslash [\nabla\mathcal{L}(\theta_k)^{\odot 2}]^{\odot \frac{1}{2}}). \tag{A.1.56}$$

Now consider the stochastic regime where we are optimizing a different loss $\mathcal{L}_k$ at each iteration. In SGD, we "tracked" (i.e., took an average of) the gradients using momentum. Here, we can track both the gradient and the squared gradient using momentum, i.e.,

$$\boldsymbol{g}_k = \beta^1 \boldsymbol{g}_{k-1} + (1-\beta^1)\nabla\mathcal{L}_k(\theta_k) \tag{A.1.57}$$

$$\boldsymbol{s}_k = \beta^2 \boldsymbol{s}_{k-1} + (1-\beta^2)[\nabla\mathcal{L}_k(\theta_k)]^{\odot 2} \tag{A.1.58}$$

$$\theta_{k+1} = \theta_k - h\boldsymbol{g}_k \oslash \boldsymbol{s}_k^{\odot\frac{1}{2}}, \tag{A.1.59}$$

where $\beta^i \in [0, 1]$ are the momentum parameters. The algorithm presented by this iteration is the celebrated *Adam* optimizer,[5] which is the most-used optimizer in deep learning. While convergence proofs of Adam are more involved, it falls out of the same steepest descent principle we used so far, and so we should expect that given a small enough learning rate, each update should improve the loss.

[5]In order to avoid division-by-zero errors, we divide by $\boldsymbol{s}_k^{\odot(1/2)} + \varepsilon\boldsymbol{1}_n$ where $\varepsilon$ is small, say on the order of $10^{-8}$.

Another way to view Adam, which partially explains its empirical success, is that it dynamically updates the learning rates for each parameter based on the squared gradients. In particular, notice that we can write

$$\theta_{k+1} = \theta_k - \eta_k \odot \boldsymbol{g}_k \qquad \text{where} \qquad \eta_k = h\boldsymbol{s}_k^{\odot(-\frac{1}{2})} \tag{A.1.60}$$

where $\eta_k$ is the parameter-wise adaptively set learning rate. This scheme is called adaptive because if the gradient of a particular parameter is large up to iteration $k$, then the learning rate for this parameter becomes smaller to compensate, and vice versa, as can be seen from the above equation.

## A.2 Computing Gradients via Automatic Differentiation

Above, we discussed several optimization algorithms for deep networks which assumed access to a *first-order oracle*, i.e., a device which would allow us to compute $\mathcal{L}(\theta)$ and $\nabla\mathcal{L}(\theta)$. For simple functions $\mathcal{L}$, it is possible to do this by hand. However, for deep neural networks, this quickly becomes tedious, and hinders rapid experimentation. Hence we require a general algorithm which would allow us to efficiently compute the gradients of arbitrary (sub)differentiable network architectures.

In this section, we introduce the basics of *automatic differentiation* (AD or *autodiff*), which is a computationally efficient way to compute gradients and Jacobians of general functions $f : \mathbb{R}^m \to \mathbb{R}^n$. We will show how this leads to the backpropagation algorithm for computing gradients of loss functions involving neural networks. A summary of the structure of this section is as follows:

1. We introduce **differentials**, a convenient formalism for calculating and organizing the derivatives of functions between high-dimensional parameter spaces (which may themselves be products of other spaces involving matrices, tensors, etc.).

2. We describe the basics of **forward-mode** and **reverse-mode** automatic differentiation, which involves considerations that are important for efficient computation of gradients/Jacobians for different kinds of functions arising in machine learning applications.

3. We describe **backpropagation** in the special case of a loss function applied to a stack-of-layers neural network as an instantiation of reverse-mode automatic differentiation.

Our treatment will err on the mathematical side, to give the reader a deep understanding of the underlying mathematics. The reader should ensure to couple this understanding with a strong grasp of practical aspects of automatic differentiation for deep learning, for example as manifested in the outstanding tutorial of Karpathy [Kar22b].

### A.2.1 Differentials

A full accounting of this subsection is given in the excellent guide [BEJ25]. To motivate differentials, let us first consider the simple example of a differentiable function $\mathcal{L}\colon \mathbb{R} \to \mathbb{R}$ acting on a parameter $\theta$. We can write

$$\mathcal{L}(\theta^+) - \mathcal{L}(\theta) = \mathcal{L}'(\theta) \cdot (\theta^+ - \theta) + o(|\theta^+ - \theta|). \tag{A.2.1}$$

If we take $\delta\theta \doteq \theta^+ - \theta$ and $\delta\mathcal{L} \doteq \mathcal{L}(\theta + \delta\theta) - \mathcal{L}(\theta)$, we can write

$$\delta\mathcal{L} = \mathcal{L}'(\theta) \cdot \delta\theta + o(|\delta\theta|). \tag{A.2.2}$$

We will (non-rigorously) define $\mathrm{d}\theta$ and $\mathrm{d}\mathcal{L}$, i.e., the *differentials* of $\theta$ and $\mathcal{L}$, to be *infinitesimally small* changes in $\theta$ and $\mathcal{L}$. Think of them as what one gets when $\delta\theta$ (and therefore $\delta\mathcal{L}$) are extremely small. The goal of differential calculus, in some sense, is to study the relationships between the differentials $\mathrm{d}\theta$ and $\mathrm{d}\mathcal{L}$, namely, seeing how small changes in the input of a function change the output. We should note that the differential $\mathrm{d}\theta$ is the *same shape* as $\theta$, and the differential $\mathrm{d}\mathcal{L}$ is the *same shape* as $\mathcal{L}$. In particular, we can write

$$\mathrm{d}\mathcal{L} = \mathcal{L}'(\theta) \cdot \mathrm{d}\theta, \tag{A.2.3}$$

whereby we have that all higher powers of $|\mathrm{d}\theta|$, such as $(\mathrm{d}\theta)^2$, are 0.

Let's see how this works for higher dimensions, i.e., $\mathcal{L}\colon \mathbb{R}^n \to \mathbb{R}$. Then we still have

$$\mathrm{d}\mathcal{L} = \mathcal{L}'(\theta) \cdot \mathrm{d}\theta \tag{A.2.4}$$

for some notion of a derivative $\mathcal{L}'(\theta)$. Since $\theta$ (hence $\mathrm{d}\theta$) is a column vector here and $\mathcal{L}$ (hence $\mathrm{d}\mathcal{L}$) is a scalar, we must have that $\mathcal{L}'(\theta)$ is a row vector. In this case, $\mathcal{L}'(\theta)$ is the Jacobian of $\mathcal{L}$ w.r.t. $\theta$. Here notice that we have set all higher powers and products of coordinates of $\mathrm{d}\theta$ to 0. In sum,

*All products and powers $\geq 2$ of differentials are equal to* 0*.*

Now consider a higher-order tensor function $\mathcal{L}\colon \mathbb{R}^{m\times n} \to \mathbb{R}^{p\times q}$. Then our basic linearization equation is insufficient for this case: $\mathrm{d}\mathcal{L} = \mathcal{L}'(\theta) \cdot \mathrm{d}\theta$ does not make sense since $\theta$ is an $m \times n$ matrix but $\mathrm{d}\mathcal{L}$ is a $p \times q$ matrix, so there is no possible vector or matrix shape for $\mathcal{L}'(\theta)$ that works in general (as no matrix can multiply a $m \times n$ matrix to form a $p \times q$ matrix unless $m = p$). So we must have a slightly more advanced interpretation.

Namely, we consider $\mathcal{L}'(\theta)$ as a *linear transformation* whose input is $\theta$-space and whose output is $\mathcal{L}$-space, which takes in a small change in $\theta$ and outputs the corresponding small change in $\mathcal{L}$. Namely, we can write

$$\mathrm{d}\mathcal{L} = \mathcal{L}'(\theta)[\mathrm{d}\theta]. \tag{A.2.5}$$

In the previous cases, $\mathcal{L}'(\theta)$ was first a linear operator $\mathbb{R} \to \mathbb{R}$ whose action was to multiply its input by the scalar derivative of $\mathcal{L}$ with respect to $\theta$, and then a linear operator $\mathbb{R}^n \to \mathbb{R}$ whose action was to multiply its input by the Jacobian derivative of $\mathcal{L}$ with respect to $\theta$. In general $\mathcal{L}'(\theta)$ is the "derivative" of $\mathcal{L}$ w.r.t. $\theta$. Think of $\mathcal{L}'$ as a generalized version of the Jacobian of $\mathcal{L}$. As such, it follows some simple derivative rules, most crucially the chain rule.

**Theorem A.1** (Differential Chain Rule)**.** *Suppose $\mathcal{L} = f \circ g$ where $f$ and $g$ are differentiable. Then*

$$\mathrm{d}\mathcal{L} = f'(g(\theta))g'(\theta)[\mathrm{d}\theta], \tag{A.2.6}$$

*where (as usual) multiplication indicates composition of linear operators. In particular,*

$$\mathcal{L}'(\theta) = f'(g(\theta))g'(\theta) \tag{A.2.7}$$

*in the sense of equality of linear operators.*

It is productive to think of the multivariate chain rule in functorial terms: composition of functions gets 'turned into' matrix multiplication of Jacobians (composition of linear operators!). We illustrate the power of this result and this perspective through several examples.

*Example* A.4. Consider the function $f(\boldsymbol{X}) = \boldsymbol{W}\boldsymbol{X} + \boldsymbol{b}\mathbf{1}^\top$. Then

$$\mathrm{d}f = f(\boldsymbol{X} + \mathrm{d}\boldsymbol{X}) - f(\boldsymbol{X}) = [\boldsymbol{W}(\boldsymbol{X} + \mathrm{d}\boldsymbol{X}) + \boldsymbol{b}\mathbf{1}^\top] - [\boldsymbol{W}\boldsymbol{X} + \boldsymbol{b}\mathbf{1}^\top] = \boldsymbol{W}\mathrm{d}\boldsymbol{X}. \tag{A.2.8}$$

Thus the derivative of an affine function w.r.t. its input is

$$f'(\boldsymbol{X})[\mathrm{d}\boldsymbol{X}] = \boldsymbol{W}\mathrm{d}\boldsymbol{X} \implies f'(\boldsymbol{X}) = \boldsymbol{W}. \tag{A.2.9}$$

Notice that $f'$ is *constant*. On the other hand, consider the function $g(\boldsymbol{W}, \boldsymbol{b}) = \boldsymbol{W}\boldsymbol{X} + \boldsymbol{b}\mathbf{1}^\top$. Then

$$\begin{aligned}
\mathrm{d}g &= g(\boldsymbol{W} + \mathrm{d}\boldsymbol{W}, \boldsymbol{b} + \mathrm{d}\boldsymbol{b}) - g(\boldsymbol{W}, \boldsymbol{b}) && \text{(A.2.10)} \\
&= [(\boldsymbol{W} + \mathrm{d}\boldsymbol{W})\boldsymbol{X} + (\boldsymbol{b} + \mathrm{d}\boldsymbol{b})\mathbf{1}^\top] - [\boldsymbol{W}\boldsymbol{X} + \boldsymbol{b}\mathbf{1}^\top] && \text{(A.2.11)} \\
&= (\mathrm{d}\boldsymbol{W})\boldsymbol{X} + (\mathrm{d}\boldsymbol{b})\mathbf{1}^\top = g'(\boldsymbol{W}, \boldsymbol{b})[\mathrm{d}\boldsymbol{W}, \mathrm{d}\boldsymbol{b}]. && \text{(A.2.12)}
\end{aligned}$$

Notice that this derivative is *constant* in $\boldsymbol{W}, \boldsymbol{b}$ (which makes sense since $g$ itself is linear) and linear in the differential inputs $\mathrm{d}\boldsymbol{W}, \mathrm{d}\boldsymbol{b}$. ■

*Example* A.5. Consider the function $f = gh$ where $g, h$ are differentiable functions whose outputs can multiply together. Then $f = p \circ v$ where $v = (g, h)$ and $p(a, b) = ab$. Applying the chain rule we have

$$\mathrm{d}f = p'(v(x))v'(x)[\mathrm{d}x]. \tag{A.2.13}$$

To compute $v'(x)$ we can compute

$$\mathrm{d}v = v'(x)[\mathrm{d}x] = v(x + \mathrm{d}x) - v(x) = \begin{bmatrix} g(x + \mathrm{d}x) - g(x) \\ h(x + \mathrm{d}x) - h(x) \end{bmatrix} = \begin{bmatrix} g'(x)[\mathrm{d}x] \\ h'(x)[\mathrm{d}x] \end{bmatrix}. \tag{A.2.14}$$

To compute $p'$ we can compute

$$\mathrm{d}p = p'(a, b)[\mathrm{d}a, \mathrm{d}b] = p(a + \mathrm{d}a, b + \mathrm{d}b) - p(a, b) = (a + \mathrm{d}a)(b + \mathrm{d}b) - ab \tag{A.2.15}$$

$$= (\mathrm{d}a)b + a(\mathrm{d}b) + (\mathrm{d}a)(\mathrm{d}b) = (\mathrm{d}a)b + a(\mathrm{d}b), \tag{A.2.16}$$

where (recall) the product of the differentials $\mathrm{d}a$ and $\mathrm{d}b$ is set to 0. Therefore

$$p'(a, b)[\mathrm{d}a, \mathrm{d}b] = (\mathrm{d}a)b + (\mathrm{d}b)a. \tag{A.2.17}$$

Putting these together, we find

$$f'(x)[\mathrm{d}x] = p'(v(x))v'(x)[\mathrm{d}x] = p'(g(x), h(x))[g'(x)[\mathrm{d}x], h'(x)[\mathrm{d}x]] \tag{A.2.18}$$

$$= (g'(x)[\mathrm{d}x])h(x) + g(x)(h'(x)[\mathrm{d}x]). \tag{A.2.19}$$

This gives

$$f'(x)[\mathrm{d}x] = (g'(x)[\mathrm{d}x])h(x) + g(x)(h'(x)[\mathrm{d}x]). \tag{A.2.20}$$

If for example we say that $f, g, h \colon \mathbb{R} \to \mathbb{R}$ then everything commutes so

$$f'(x)[\mathrm{d}x] = (g'(x)h(x) + g(x)h'(x))[\mathrm{d}x] \implies f'(x) = g'(x)h(x) + g(x)h'(x) \tag{A.2.21}$$

which is the familiar product rule. ■

*Example* A.6. Consider the function $f(\boldsymbol{A}) = \boldsymbol{A}^\top \boldsymbol{A}\boldsymbol{B}\boldsymbol{A}$ where $\boldsymbol{A}$ is a matrix and $\boldsymbol{B}$ is a constant matrix. Then, letting $f = gh$ where $g(\boldsymbol{A}) = \boldsymbol{A}^\top \boldsymbol{A}$ and $h(\boldsymbol{A}) = \boldsymbol{B}\boldsymbol{A}$, we can use the product rule to obtain

$$f'(\boldsymbol{A})[\mathrm{d}\boldsymbol{A}] = (g'(\boldsymbol{A})[\mathrm{d}\boldsymbol{A}])h(\boldsymbol{A}) + g(\boldsymbol{A})(h'(\boldsymbol{A})[\mathrm{d}\boldsymbol{A}]) \tag{A.2.22}$$

$$= ((\mathrm{d}\boldsymbol{A})^\top \boldsymbol{A} + \boldsymbol{A}^\top(\mathrm{d}\boldsymbol{A}))\boldsymbol{B}\boldsymbol{A} + \boldsymbol{A}^\top \boldsymbol{A}\boldsymbol{B}(\mathrm{d}\boldsymbol{A}). \tag{A.2.23}$$

■

*Example* A.7. Consider the function $f \colon \mathbb{R}^{m \times n \times k} \to \mathbb{R}^{m \times n}$ given by

$$f(\boldsymbol{A})_{ij} = \sum_{t=1}^{k} A_{ijt}. \tag{A.2.24}$$

We cannot write a (matrix-valued) Jacobian or gradient for this function. But we can compute its differential just fine:

$$\mathrm{d}f_{ij} = [f(\boldsymbol{A} + \mathrm{d}\boldsymbol{A}) - f(\boldsymbol{A})]_{ij} = \sum_{t=1}^{k} \mathrm{d}\boldsymbol{A}_{ijt} = \mathbf{1}_k^\top (\mathrm{d}\boldsymbol{A})_{ij}. \tag{A.2.25}$$

So

$$(f'(\boldsymbol{A})[\mathrm{d}\boldsymbol{A}])_{ij} = \mathbf{1}_k^\top (\mathrm{d}\boldsymbol{A})_{ij}, \tag{A.2.26}$$

which represents a higher-order tensor multiplication operation that is nonetheless well-defined. ■

This gives us all the technology we need to compute differentials of everything. The last thing we cover in this section is a method to compute gradients using the differential. Namely, for a function $\mathcal{L}$ whose output is a scalar, the gradient $\nabla\mathcal{L}$ is defined as

$$\mathrm{d}\mathcal{L} = \mathcal{L}'(\theta)[\mathrm{d}\theta] = \langle \nabla\mathcal{L}(\theta), \mathrm{d}\theta \rangle, \tag{A.2.27}$$

where the inner product here is the "standard" inner product for the specified objects (i.e., for vectors it's the $\ell^2$ inner product, whereas for matrices it's the Frobenius inner product, and for higher-order tensors it's the analogous sum-of-coordinates inner product). This definition is the correct generalization of the 'familiar' example of the gradient of a function from $\mathbb{R}^n$ to $\mathbb{R}$ as the vector of partial derivatives—a version of Taylor's theorem for general functions $f : \mathbb{R}^m \to \mathbb{R}^n$ makes this connection rigorous. So one way to compute the gradient $\nabla\mathcal{L}$ is to compute the differential $\mathrm{d}\mathcal{L}$ and rewrite it in the form $\langle \text{something}, \mathrm{d}\theta \rangle$, then that "something" is the gradient.

### A.2.2 Automatic Differentiation

The main idea of AD is to compute the chain rule efficiently. The basic problem we need to cope with is the following. In the optimization section of the appendix, we considered that the parameter space $\Theta$ was an abstract Euclidean space like $\mathbb{R}^n$. In practice the parameters are really some collection of vectors, matrices, and higher-order objects: $\Theta = \mathbb{R}^{m\times n} \times \mathbb{R}^n \times \mathbb{R}^{r\times q\times p} \times \mathbb{R}^{r\times q} \times \cdots$. While in theory this is the same thing as a large parameter space $\mathbb{R}^{n'}$ for some (very large) $n'$, computationally efficient algorithms for differentiation must treat these two spaces differently. Forward and reverse mode automatic differentiation are two different schemes for performing this computation.

Let us do a simple example to start. Let $\mathcal{L}$ be defined by $\mathcal{L} = a \circ b \circ c$ where $a, b, c$ are differentiable. Then the *chain rule* gives

$$\mathcal{L}'(\theta) = a'(b(c(\theta)))b'(c(\theta))c'(\theta). \tag{A.2.28}$$

To compute $\mathcal{L}(\theta)$, we first compute $c(\theta)$ then $b(c(\theta))$ then $a(b(c(\theta)))$, and store them all. There are two ways to compute $\mathcal{L}'(\theta)$. The *forward-mode AD* will compute

$$c'(\theta) \implies b'(c(\theta))c'(\theta) \implies a'(b(c(\theta)))b'(c(\theta))c'(\theta) \tag{A.2.29}$$

i.e., computing the derivatives "from the bottom-up". The *reverse mode AD* will compute

$$a'(b(c(\theta))) \implies a'(b(c(\theta)))b'(c(\theta)) \implies a'(b(c(\theta)))b'(c(\theta))c'(\theta), \tag{A.2.30}$$

i.e., computing the derivatives "from the top down". To see why this matters, suppose that $f\colon \mathbb{R}^p \to \mathbb{R}^s$ is given by $f = a \circ b \circ c$ where $a\colon \mathbb{R}^r \to \mathbb{R}^s$, $b\colon \mathbb{R}^q \to \mathbb{R}^r$, $c\colon \mathbb{R}^p \to \mathbb{R}^q$. Then the chain rule is:

$$f'(\boldsymbol{x}) = a'(b(c(\boldsymbol{x})))b'(c(\boldsymbol{x}))c'(\boldsymbol{x}) \tag{A.2.31}$$

where (recall) $f'$ is the derivative, in this case the Jacobian (since the input and output of each function are both vectors). Assuming that computing each Jacobian is trivial and the only cost is multiplying the Jacobians together, forward-mode AD has the following computational costs (assuming that multiplying $A \in \mathbb{R}^{m\times n}, B \in \mathbb{R}^{n\times k}$ takes $\mathcal{O}(mnk)$ time):

$$\text{computing } c'(\boldsymbol{x}) \in \mathbb{R}^{q\times p} \text{ takes negligible time} \tag{A.2.32}$$
$$\text{computing } b'(c(\boldsymbol{x}))c'(x) \in \mathbb{R}^{r\times p} \text{ takes } \mathcal{O}(pqr) \text{ time} \tag{A.2.33}$$
$$\text{computing } a'(b(c(\boldsymbol{x})))b'(c(\boldsymbol{x}))c'(\boldsymbol{x}) \in \mathbb{R}^{s\times p} \text{ takes } \mathcal{O}(pqr+prs) \text{ time.} \tag{A.2.34}$$

Meanwhile, doing reverse-mode AD has the following computational costs:

$$\text{computing } a'(b(c(\boldsymbol{x}))) \in \mathbb{R}^{s\times r} \text{ takes negligible time} \tag{A.2.35}$$
$$\text{computing } a'(b(c(\boldsymbol{x})))b'(c(\boldsymbol{x})) \in \mathbb{R}^{s\times q} \text{ takes } \mathcal{O}(qrs) \text{ time} \tag{A.2.36}$$
$$\text{computing } a'(b(c(\boldsymbol{x})))b'(c(\boldsymbol{x}))c'(\boldsymbol{x}) \in \mathbb{R}^{s\times p} \text{ takes } \mathcal{O}(qrs+pqs) \text{ time.} \tag{A.2.37}$$

In other words, the forward-mode AD takes $\mathcal{O}(p(qr+rs))$ time, and the reverse-mode AD takes $\mathcal{O}(s(pq+qr))$ time. *These take a different amount of time!*

More generally, suppose that $f = f^L \circ \cdots \circ f^1$ where each $f^\ell \colon \mathbb{R}^{d^{\ell-1}} \to \mathbb{R}^{d^\ell}$, so that $f \colon \mathbb{R}^{d^0} \to \mathbb{R}^{d^L}$. Then the forward-mode AD takes $\mathcal{O}(d^0(\sum_{\ell=2}^{L} d^{\ell-1}d^\ell))$ time while the reverse-mode AD takes $\mathcal{O}(d^L(\sum_{\ell=1}^{L-1} d^{\ell-1}d^\ell))$ time. From the above rates, we see that all else equal:

- If the function to optimize has *more outputs than inputs* (i.e., $d^L > d^0$), use *forward-mode AD*.

- If the function to optimize has *more inputs than outputs* (i.e., $d^0 > d^L$), use *reverse-mode AD*.

In a neural network, we compute the gradient of a *loss function* $\mathcal{L} \colon \Theta \to \mathbb{R}$, where the parameter space $\Theta$ is usually very high-dimensional. So in practice we always use reverse-mode AD for training neural networks. Reverse-mode AD, in the context of training neural networks, is called *backpropagation*.

### A.2.3 Back Propagation

In this section, we will discuss algorithmic backpropagation using a simple yet completely practical example. Suppose that we fix an input-label pair $(\boldsymbol{X}, \boldsymbol{y})$, and fix a network architecture $f_\theta = f_\theta^L \circ \cdots \circ f_\theta^1 \circ f_\theta^{\mathrm{emb}}$ where $\theta = (\theta^{\mathrm{emb}}, \theta^1, \ldots, \theta^L, \theta^{\mathrm{head}})$ and task-specific head $h_{\theta^{\mathrm{head}}}$, and write

$$\boldsymbol{Z}_\theta^1(\boldsymbol{X}) \doteq f_\theta^{\mathrm{emb}}(\boldsymbol{X}), \tag{A.2.38}$$
$$\boldsymbol{Z}_\theta^{\ell+1}(\boldsymbol{X}) \doteq f_\theta^\ell(\boldsymbol{Z}_\theta^\ell(\boldsymbol{X})), \quad \forall \ell \in \{1, \ldots, L\}, \tag{A.2.39}$$
$$\hat{\boldsymbol{y}}_{\theta^{\mathrm{head}}}(\boldsymbol{X}) \doteq h_{\theta^{\mathrm{head}}}(\boldsymbol{Z}_\theta^{L+1}(\boldsymbol{X})). \tag{A.2.40}$$

Then, we can define the loss on this one term by

$$\mathcal{L}(\theta) \doteq \mathsf{L}(\boldsymbol{y}, \hat{\boldsymbol{y}}_{\theta}(\boldsymbol{X})), \tag{A.2.41}$$

where $\mathsf{L}$ is a differentiable function of its second argument. Then the backward-mode AD instructs us to compute the derivatives in the order $\theta^{\text{head}}, \theta^{L}, \ldots, \theta^{1}, \theta^{\text{emb}}$.

To carry out this computation, let us make some changes to the notation.

- First, let us change the notation to emphasize the dependency structure between the variables. Namely,

$$\boldsymbol{Z}^{1} \doteq f^{\text{emb}}(\boldsymbol{X}, \theta^{\text{emb}}) \tag{A.2.42}$$
$$\boldsymbol{Z}^{\ell+1} \doteq f^{\ell}(\boldsymbol{Z}^{\ell}, \theta^{\ell}), \qquad \forall \ell \in \{1, \ldots, L\}, \tag{A.2.43}$$
$$\hat{\boldsymbol{y}} \doteq h(\boldsymbol{Z}^{L+1}, \theta^{\text{head}}), \tag{A.2.44}$$
$$\mathcal{L} \doteq \mathsf{L}(\boldsymbol{y}, \hat{\boldsymbol{y}}). \tag{A.2.45}$$

- Then, instead of having the derivative be $f'$, we explicitly notate the independent variable and write the derivative as $\frac{\mathrm{d}f}{\mathrm{d}\theta^{1}}$, for example. This is because there are many variables in our model and we only care about one at a time.

We can start by computing the appropriate differentials. First, for $\theta^{\text{head}}$ we have

$$\mathrm{d}\mathcal{L} = \mathrm{d}\mathsf{L} \tag{A.2.46}$$
$$= \frac{\mathrm{d}\mathsf{L}}{\mathrm{d}\hat{\boldsymbol{y}}} \cdot \mathrm{d}\hat{\boldsymbol{y}} \tag{A.2.47}$$
$$= \frac{\mathrm{d}\mathsf{L}}{\mathrm{d}\hat{\boldsymbol{y}}} \cdot \mathrm{d}(h(\boldsymbol{Z}^{L+1}, \theta^{\text{head}})) \tag{A.2.48}$$
$$= \frac{\mathrm{d}\mathsf{L}}{\mathrm{d}\hat{\boldsymbol{y}}} \left[ \frac{\mathrm{d}h}{\mathrm{d}\boldsymbol{Z}^{L+1}} \cdot \mathrm{d}\boldsymbol{Z}^{L+1} + \frac{\mathrm{d}h}{\mathrm{d}\theta^{\text{head}}} \cdot \mathrm{d}\theta^{\text{head}} \right]. \tag{A.2.49}$$

Now since $\boldsymbol{Z}^{L+1}$ does not depend on $\theta^{\text{head}}$, we have $\mathrm{d}\boldsymbol{Z}^{L+1} = 0$, so in the end it holds (using the fact that the gradient is the transpose of the derivative for a function $\mathsf{L}$ whose codomain is $\mathbb{R}$):

$$\mathrm{d}\mathcal{L} = \frac{\mathrm{d}\mathsf{L}}{\mathrm{d}\hat{\boldsymbol{y}}} \cdot \frac{\mathrm{d}h}{\mathrm{d}\theta^{\text{head}}} \cdot \mathrm{d}\theta^{\text{head}} \tag{A.2.50}$$
$$= [\nabla_{\hat{\boldsymbol{y}}} \mathsf{L}]^{\top} \frac{\mathrm{d}h}{\mathrm{d}\theta^{\text{head}}} \cdot \mathrm{d}\theta^{\text{head}} \tag{A.2.51}$$
$$= \left\langle \nabla_{\hat{\boldsymbol{y}}} \mathsf{L}, \frac{\mathrm{d}h}{\mathrm{d}\theta^{\text{head}}} \cdot \mathrm{d}\theta^{\text{head}} \right\rangle \tag{A.2.52}$$
$$= \left\langle \left( \frac{\mathrm{d}h}{\mathrm{d}\theta^{\text{head}}} \right)^{*} \nabla_{\hat{\boldsymbol{y}}} \mathsf{L}, \mathrm{d}\theta^{\text{head}} \right\rangle \tag{A.2.53}$$
$$= \langle \nabla_{\theta^{\text{head}}} \mathcal{L}, \mathrm{d}\theta^{\text{head}} \rangle. \tag{A.2.54}$$

Thus to compute $\nabla_{\theta^{\text{head}}}\mathcal{L}$, we compute the gradient $\nabla_{\hat{\boldsymbol{y}}}\mathsf{L}$ and the *adjoint*[6] of the derivative $\frac{\mathrm{d}h}{\mathrm{d}\theta^{\text{head}}}$ and multiply (i.e., apply the adjoint linear transformation to the gradient). In practice, both derivatives can be computed by hand, but many modern computational frameworks can *automatically* define the derivatives (and/or their adjoints) given code for the "forward pass," i.e., the loss function computation. While extending this automatic derivative definition to as many functions as possible is an area of active research, the resource [BEJ25] describes one basic approach to do it in some detail. By the way, backpropagation is also called the *adjoint method* for this reason — i.e., that we use adjoint derivatives to compute the gradient.

Now let us compute the differentials w.r.t. some $\theta^{\ell}$:

$$
\begin{aligned}
\mathrm{d}\mathcal{L} &= \frac{\mathrm{d}\mathcal{L}}{\mathrm{d}\boldsymbol{Z}^{\ell+1}} \cdot \mathrm{d}\boldsymbol{Z}^{\ell+1} && \text{(A.2.55)}\\
&= \frac{\mathrm{d}\mathcal{L}}{\mathrm{d}\boldsymbol{Z}^{\ell+1}} \cdot \mathrm{d}(f^{\ell}(\boldsymbol{Z}^{\ell}, \theta^{\ell})) && \text{(A.2.56)}\\
&= \frac{\mathrm{d}\mathcal{L}}{\mathrm{d}\boldsymbol{Z}^{\ell+1}} \left[\frac{\mathrm{d}f^{\ell}}{\mathrm{d}\boldsymbol{Z}^{\ell}} \cdot \mathrm{d}\boldsymbol{Z}^{\ell} + \frac{\mathrm{d}f^{\ell}}{\mathrm{d}\theta^{\ell}} \cdot \mathrm{d}\theta^{\ell}\right] && \text{(A.2.57)}\\
&= \frac{\mathrm{d}\mathcal{L}}{\mathrm{d}\boldsymbol{Z}^{\ell+1}} \cdot \frac{\mathrm{d}f^{\ell}}{\mathrm{d}\theta^{\ell}} \cdot \mathrm{d}\theta^{\ell} \qquad (\text{b/c } \boldsymbol{Z}^{\ell} \text{ isn't fn. of } \theta^{\ell} \text{ so } \mathrm{d}\boldsymbol{Z}^{\ell} = 0) && \text{(A.2.58)}\\
&= [\nabla_{\boldsymbol{Z}^{\ell+1}}\mathcal{L}]^{\top} \frac{\mathrm{d}f^{\ell}}{\mathrm{d}\theta^{\ell}} \cdot \mathrm{d}\theta^{\ell} && \text{(A.2.59)}\\
&= \left\langle \nabla_{\boldsymbol{Z}^{\ell+1}}\mathcal{L}, \frac{\mathrm{d}f^{\ell}}{\mathrm{d}\theta^{\ell}} \cdot \mathrm{d}\theta^{\ell} \right\rangle && \text{(A.2.60)}\\
&= \left\langle \left(\frac{\mathrm{d}f^{\ell}}{\mathrm{d}\theta^{\ell}}\right)^{*} \nabla_{\boldsymbol{Z}^{\ell+1}}\mathcal{L}, \mathrm{d}\theta^{\ell} \right\rangle && \text{(A.2.61)}\\
&= \langle \nabla_{\theta^{\ell}}\mathcal{L}, \mathrm{d}\theta^{\ell} \rangle. && \text{(A.2.62)}
\end{aligned}
$$

Thus to compute $\nabla_{\theta^{\ell}}\mathcal{L}$ we compute $\nabla_{\boldsymbol{Z}^{\ell+1}}\mathcal{L}$ then apply the adjoint derivative $\left(\frac{\mathrm{d}f^{\ell}}{\mathrm{d}\theta^{\ell}}\right)^{*}$ to it. Since $\nabla_{\theta^{\ell}}\mathcal{L}$ depends on $\nabla_{\boldsymbol{Z}^{\ell+1}}\mathcal{L}$, we also want to be able to compute the gradients w.r.t. $\boldsymbol{Z}^{\ell}$. This can be computed in the exact same way:

$$
\begin{aligned}
\mathrm{d}\mathcal{L} &= \frac{\mathrm{d}\mathcal{L}}{\mathrm{d}\boldsymbol{Z}^{\ell+1}} \cdot \mathrm{d}\boldsymbol{Z}^{\ell+1} && \text{(A.2.63)}\\
&= \frac{\mathrm{d}\mathcal{L}}{\mathrm{d}\boldsymbol{Z}^{\ell+1}} \cdot \mathrm{d}(f^{\ell}(\boldsymbol{Z}^{\ell}, \theta^{\ell})) && \text{(A.2.64)}\\
&= \frac{\mathrm{d}\mathcal{L}}{\mathrm{d}\boldsymbol{Z}^{\ell+1}} \left[\frac{\mathrm{d}f^{\ell}}{\mathrm{d}\boldsymbol{Z}^{\ell}} \cdot \mathrm{d}\boldsymbol{Z}^{\ell} + \frac{\mathrm{d}f^{\ell}}{\mathrm{d}\theta^{\ell}} \cdot \mathrm{d}\theta^{\ell}\right] && \text{(A.2.65)}\\
&= \frac{\mathrm{d}\mathcal{L}}{\mathrm{d}\boldsymbol{Z}^{\ell+1}} \cdot \frac{\mathrm{d}f^{\ell}}{\mathrm{d}\boldsymbol{Z}^{\ell}} \cdot \mathrm{d}\boldsymbol{Z}^{\ell} \qquad (\text{b/c } \theta^{\ell} \text{ isn't fn. of } \boldsymbol{Z}^{\ell} \text{ so } \mathrm{d}\theta^{\ell} = 0 \text{ this time}) && \text{(A.2.66)}
\end{aligned}
$$

[6] The adjoint is like a generalized transpose for more general linear transformations. Particularly, for a given pair of inner product spaces and linear transformation $T$ between those spaces, the adjoint $T^{*}$ is defined by the identity $\langle T\boldsymbol{x}, \boldsymbol{y}\rangle = \langle \boldsymbol{x}, T^{*}\boldsymbol{y}\rangle$. In finite dimensions (i.e., all cases relevant to this book) the adjoint exists and is unique.

$$= \left\langle \left(\frac{\mathrm{d}f^{\ell}}{\mathrm{d}\boldsymbol{Z}^{\ell}}\right)^{*} \nabla_{\boldsymbol{Z}^{\ell+1}}\mathcal{L}, \mathrm{d}\boldsymbol{Z}^{\ell} \right\rangle \qquad \text{(same machine)} \tag{A.2.67}$$

$$= \langle \nabla_{\boldsymbol{Z}^{\ell}}\mathcal{L}, \mathrm{d}\boldsymbol{Z}^{\ell} \rangle. \tag{A.2.68}$$

Thus to compute $\nabla_{\boldsymbol{Z}^{\ell}}\mathcal{L}$ we compute $\nabla_{\boldsymbol{Z}^{\ell+1}}\mathcal{L}$ then apply the adjoint derivative $\left(\frac{\mathrm{d}f^{\ell}}{\mathrm{d}\boldsymbol{Z}^{\ell}}\right)^{*}$ to it. So we have a recursion to compute $\nabla_{\boldsymbol{Z}^{\ell}}\mathcal{L}$ for all $\ell$, with base case $\nabla_{\boldsymbol{Z}^{L+1}}\mathcal{L}$, which is given by

$$\mathrm{d}\mathcal{L} = \mathrm{d}\mathsf{L} \tag{A.2.69}$$

$$= \frac{\mathrm{d}\mathsf{L}}{\mathrm{d}\hat{\boldsymbol{y}}} \cdot \mathrm{d}\hat{\boldsymbol{y}} \tag{A.2.70}$$

$$= \frac{\mathrm{d}\mathsf{L}}{\mathrm{d}\hat{\boldsymbol{y}}} \cdot \mathrm{d}h(\boldsymbol{Z}^{L+1}, \theta^{\mathrm{head}}) \tag{A.2.71}$$

$$= \frac{\mathrm{d}\mathsf{L}}{\mathrm{d}\hat{\boldsymbol{y}}} \left[\frac{\mathrm{d}h}{\mathrm{d}\boldsymbol{Z}^{L+1}} \cdot \mathrm{d}\boldsymbol{Z}^{L+1} + \frac{\mathrm{d}h}{\mathrm{d}\theta^{\mathrm{head}}} \cdot \mathrm{d}\theta^{\mathrm{head}}\right] \tag{A.2.72}$$

$$= \frac{\mathrm{d}\mathsf{L}}{\mathrm{d}\hat{\boldsymbol{y}}} \cdot \frac{\mathrm{d}h}{\mathrm{d}\boldsymbol{Z}^{L+1}} \cdot \mathrm{d}\boldsymbol{Z}^{L+1} \tag{A.2.73}$$

$$= \left\langle \left(\frac{\mathrm{d}h}{\mathrm{d}\boldsymbol{Z}^{L+1}}\right)^{*} \nabla_{\hat{\boldsymbol{y}}}\mathsf{L}, \mathrm{d}\boldsymbol{Z}^{L+1} \right\rangle \tag{A.2.74}$$

$$= \langle \nabla_{\boldsymbol{Z}^{L+1}}\mathcal{L}, \mathrm{d}\boldsymbol{Z}^{L+1} \rangle. \tag{A.2.75}$$

Thus we have the recursion:

$$\nabla_{\boldsymbol{Z}^{L+1}}\mathcal{L} = \left(\frac{\mathrm{d}h}{\mathrm{d}\boldsymbol{Z}^{L+1}}\right)^{*} \nabla_{\hat{\boldsymbol{y}}}\mathsf{L} \tag{A.2.76}$$

$$\nabla_{\boldsymbol{Z}^{L}}\mathcal{L} = \left(\frac{\mathrm{d}f^{L}}{\mathrm{d}\boldsymbol{Z}^{L}}\right)^{*} \nabla_{\boldsymbol{Z}^{L+1}}\mathcal{L} \tag{A.2.77}$$

$$\nabla_{\theta^{L}}\mathcal{L} = \left(\frac{\mathrm{d}f^{L}}{\mathrm{d}\theta^{L}}\right)^{*} \nabla_{\boldsymbol{Z}^{L+1}}\mathcal{L} \tag{A.2.78}$$

$$\vdots \tag{A.2.79}$$

$$\nabla_{\boldsymbol{Z}^{1}}\mathcal{L} = \left(\frac{\mathrm{d}f^{1}}{\mathrm{d}\boldsymbol{Z}^{1}}\right)^{*} \nabla_{\boldsymbol{Z}^{2}}\mathcal{L} \tag{A.2.80}$$

$$\nabla_{\theta^{1}}\mathcal{L} = \left(\frac{\mathrm{d}f^{1}}{\mathrm{d}\theta^{1}}\right)^{*} \nabla_{\boldsymbol{Z}^{2}}\mathcal{L} \tag{A.2.81}$$

$$\nabla_{\theta^{\mathrm{emb}}}\mathcal{L} = \left(\frac{\mathrm{d}f^{\mathrm{emb}}}{\mathrm{d}\theta^{\mathrm{emb}}}\right)^{*} \nabla_{\boldsymbol{Z}^{1}}\mathcal{L}. \tag{A.2.82}$$

This gives us a computationally efficient algorithm to find all gradients in the whole network.

We'll finish this section by computing the adjoint derivative for a simple layer.

*Example* A.8. Consider the "linear" (affine) layer $f^\ell$

$$f^\ell(\boldsymbol{Z}, \boldsymbol{W}^\ell, \boldsymbol{b}^\ell) \doteq \boldsymbol{W}^\ell \boldsymbol{Z} + \boldsymbol{b}^\ell \mathbf{1}^\top = \begin{bmatrix} \boldsymbol{W}^\ell & \boldsymbol{b}^\ell \end{bmatrix}. \tag{A.2.83}$$

We can compute the differential w.r.t. both parameters as

$$\mathrm{d}f^\ell = [(\boldsymbol{W}^\ell + \mathrm{d}\boldsymbol{W}^\ell)\boldsymbol{Z} + (\boldsymbol{b}^\ell + \mathrm{d}\boldsymbol{b}^\ell)\mathbf{1}^\top] - [\boldsymbol{W}^\ell \boldsymbol{Z} + \boldsymbol{b}^\ell \mathbf{1}^\top] \tag{A.2.84}$$

$$= (\mathrm{d}\boldsymbol{W}^\ell)\boldsymbol{Z} + (\mathrm{d}\boldsymbol{b}^\ell)\mathbf{1}^\top. \tag{A.2.85}$$

Thus the derivative of this transformation is

$$\frac{\mathrm{d}f^\ell}{\mathrm{d}(\boldsymbol{W}^\ell, \boldsymbol{b}^\ell)}[\mathrm{d}\boldsymbol{W}^\ell, \mathrm{d}\boldsymbol{b}^\ell] = (\mathrm{d}\boldsymbol{W}^\ell)\boldsymbol{Z} + (\mathrm{d}\boldsymbol{b}^\ell)\mathbf{1}^\top, \tag{A.2.86}$$

namely, representing the following linear transformation from $\mathbb{R}^{m\times d} \times \mathbb{R}^m$ to $\mathbb{R}^{m\times n}$:

$$T[\boldsymbol{A}, \boldsymbol{u}] = \boldsymbol{A}\boldsymbol{Z} + \boldsymbol{u}\mathbf{1}^\top. \tag{A.2.87}$$

We calculate the adjoint $T^*\colon \mathbb{R}^{m\times n} \to \mathbb{R}^{m\times d}\times\mathbb{R}^m$ w.r.t. the sum-over-coordinates (Frobenius) inner product by the following procedure:

$$\langle T[\boldsymbol{A}, \boldsymbol{u}], \boldsymbol{B}\rangle_{\mathbb{R}^{m\times n}} = \mathrm{tr}((\boldsymbol{A}\boldsymbol{Z} + \boldsymbol{u}\mathbf{1}^\top)\boldsymbol{B}^\top) \tag{A.2.88}$$

$$= \mathrm{tr}(\boldsymbol{A}\boldsymbol{Z}\boldsymbol{B}^\top + \boldsymbol{u}\mathbf{1}^\top\boldsymbol{B}^\top) \tag{A.2.89}$$

$$= \mathrm{tr}(\boldsymbol{A}\boldsymbol{Z}\boldsymbol{B}^\top) + \mathrm{tr}(\boldsymbol{u}\mathbf{1}^\top\boldsymbol{B}^\top) \tag{A.2.90}$$

$$= \mathrm{tr}(\boldsymbol{B}\boldsymbol{Z}^\top\boldsymbol{A}^\top) + \mathrm{tr}(\mathbf{1}^\top\boldsymbol{B}^\top\boldsymbol{u}) \tag{A.2.91}$$

$$= \langle \boldsymbol{B}\boldsymbol{Z}^\top, \boldsymbol{A}\rangle_{\mathbb{R}^{m\times d}} + \langle \boldsymbol{B}\mathbf{1}, \boldsymbol{u}\rangle_{\mathbb{R}^m} \tag{A.2.92}$$

$$= \langle \boldsymbol{B}(\boldsymbol{Z}^\top, \mathbf{1}), (\boldsymbol{A}, \boldsymbol{u})\rangle_{\mathbb{R}^{m\times d}\times\mathbb{R}^m} \tag{A.2.93}$$

So $T^*\boldsymbol{B} = \boldsymbol{B}(\boldsymbol{Z}^\top, \mathbf{1})$. ■

Note that as a simple application of chain rule, both backpropagation and automatic differentiation work over general "computational graphs", i.e., compositions of (simple) functions. We give all examples as neural network layers because this is the most common example in practice.

## A.3 Game Theory and Minimax Optimization

In certain cases, such as in Chapter 6, a learning problem cannot be reduced to a single optimization problem but rather represents multiple potentially opposing components of the system trying to each minimize their own objective. Examples of this paradigm include distribution learning via generative adversarial networks (GAN) and closed-loop transcription (CTRL). We will denote such a system as a *two-player game*, where we have two "players" (i.e., components) called Player 1 and Player 2 trying to minimize their objectives $\mathcal{L}^1$ and $\mathcal{L}^2$ respectively. Player 1 picks parameters $\theta \in \Theta$ and Player 2 picks parameters $\eta \in \mathrm{H}$. In this book we consider the special case of *zero-sum games*, i.e., defining a common objective $\mathcal{L}$ such that $\mathcal{L} = -\mathcal{L}^1 = \mathcal{L}^2$.

Our first, very preliminary example is as follows. Suppose that there exists functions $u(\theta)$ and $v(\eta)$ such that

$$\mathcal{L}(\theta, \eta) = -u(\theta) + v(\eta). \tag{A.3.1}$$

Then both players' objectives are independent of the other player, and the players should try to achieve their respective optima:

$$\theta^\star \in \arg\min_{\theta \in \Theta} u(\theta), \qquad \eta^\star \in \arg\min_{\eta \in \mathrm{H}} v(\eta). \tag{A.3.2}$$

The pair $(\theta^\star, \eta^\star)$ is a straightforward special case of an *equilibrium*: a situation where neither player will want to move, given the chance, since moving will end up making their own situation worse. However, not all games are so trivial; many have more complicated objectives and information structures.

In this book, the relevant game-theoretic formalism is a Stackelberg game (variously called sequential game). In this formalism, one player (without loss of generality Player 1, and also described as a *leader*) picks their parameters before the other (i.e., Player 2, also described as a *follower*), and the follower can use the full knowledge of the leader's choice to make their own choice. The correct notion of equilibrium for a Stackelberg game is a *Stackelberg equilibrium*. To explain this equilibrium, note that since Player 2 (i.e., the follower) can choose $\eta$ reactively to the choice $\theta_1$ made by Player 1 (i.e., the leader), Player 2 would of course choose the $\eta$ which minimizes $\mathcal{L}(\theta_1, \cdot)$. But of course a rational Player 1 would realize this, and so pick a $\theta_1$ such that the worst-case $\eta$ picked by Player 2 according to this rule is not too bad. More formally, let $\mathcal{S}(\theta) \doteq \arg\min_{\eta \in \mathrm{H}} \mathcal{L}(\theta, \eta)$ be the set of $\eta$ minimizing $\mathcal{L}(\theta, \eta)$, i.e., the set of all $\eta$ which Player 2 is liable to play given that Player 1 has played $\theta$. Then $(\theta^\star, \eta^\star)$ is a Stackelberg equilibrium if

$$\theta^\star \in \arg\max_{\theta \in \Theta} \min_{\eta \in \mathcal{S}(\theta)} \mathcal{L}(\theta, \eta), \qquad \eta^\star \in \arg\min_{\eta \in \mathrm{H}} \mathcal{L}(\theta^\star, \eta). \tag{A.3.3}$$

Actually (proof as exercise), one can show that in the context of two-player zero-sum Stackelberg games, $(\theta^\star, \eta^\star)$ is a Stackelberg equilibrium if and only if

$$\theta^\star \in \arg\max_{\theta \in \Theta} \min_{\eta \in \mathrm{H}} \mathcal{L}(\theta, \eta), \qquad \eta^\star \in \arg\min_{\eta \in \mathrm{H}} \mathcal{L}(\theta^\star, \eta), \tag{A.3.4}$$

(note that the notation $\mathcal{S}(\theta)$ is not used nor needed).

*Proof.* Note that

$$\mathcal{S}(\theta) = \arg\min_{\eta \in \mathrm{H}} \mathcal{L}(\theta, \eta), \tag{A.3.5}$$

so it holds

$$\min_{\eta \in \mathcal{S}(\theta)} \mathcal{L}(\theta, \eta) = \min_{\eta \in \arg\min_{\eta' \in \mathrm{H}} \mathcal{L}(\theta, \eta')} \mathcal{L}(\theta, \eta) = \min_{\eta \in \mathrm{H}} \mathcal{L}(\theta, \eta). \tag{A.3.6}$$

□

In the rest of the section, we will briefly discuss some algorithmic approaches to learn Stackelberg equilibria. The intuition you should have is that learning an equilibrium is like letting the different parts of the system automatically figure out tradeoffs between the different objectives they want to optimize.

We end this section with a caveat: in two-player zero-sum games, if it holds that

$$\max_{\theta \in \Theta} \min_{\eta \in \mathrm{H}} \mathcal{L}(\theta, \eta) = \min_{\eta \in \mathrm{H}} \max_{\theta \in \Theta} \mathcal{L}(\theta, \eta) \tag{A.3.7}$$

then every Stackelberg equilibrium is a saddle point,[7] i.e.,

$$\theta^\star \in \arg\max_{\theta \in \Theta} \mathcal{L}(\theta, \eta^\star), \qquad \eta^\star \in \arg\min_{\eta \in \mathrm{H}} \mathcal{L}(\theta^\star, \eta), \tag{A.3.8}$$

*and vice versa*, and furthermore each Stackelberg equilibrium has the (same) objective value

$$\max_{\theta \in \Theta} \min_{\eta \in \mathrm{H}} \mathcal{L}(\theta, \eta).$$

*Proof.* Suppose that indeed

$$\max_{\theta \in \Theta} \min_{\eta \in \mathrm{H}} \mathcal{L}(\theta, \eta) = \min_{\eta \in \mathrm{H}} \max_{\theta \in \Theta} \mathcal{L}(\theta, \eta). \tag{A.3.9}$$

First suppose that $(\theta^\star, \eta^\star)$ is a saddle point. We will show it is a Stackelberg equilibrium. By definition we have

$$\min_{\eta \in \mathrm{H}} \mathcal{L}(\theta^\star, \eta) = \mathcal{L}(\theta^\star, \eta^\star) = \max_{\theta \in \Theta} \mathcal{L}(\theta, \eta^\star). \tag{A.3.10}$$

It then holds for any $\theta$ and $\eta$ that

$$\min_{\eta \in \mathrm{H}} \mathcal{L}(\theta, \eta) \le \mathcal{L}(\theta, \eta^\star) \le \mathcal{L}(\theta^\star, \eta^\star). \tag{A.3.11}$$

Therefore

$$\max_{\theta \in \Theta} \min_{\eta \in \mathrm{H}} \mathcal{L}(\theta, \eta) \le \mathcal{L}(\theta^\star, \eta^\star). \tag{A.3.12}$$

Completely symmetrically,

$$\mathcal{L}(\theta^\star, \eta^\star) \le \mathcal{L}(\theta^\star, \eta) \le \max_{\theta \in \Theta} \mathcal{L}(\theta, \eta) \implies \mathcal{L}(\theta^\star, \eta^\star) \le \min_{\eta \in \mathrm{H}} \max_{\theta \in \Theta} \mathcal{L}(\theta, \eta). \tag{A.3.13}$$

Therefore since $\max\min = \min\max$ we have

$$\max_{\theta \in \Theta} \min_{\eta \in \mathrm{H}} \mathcal{L}(\theta, \eta) = \mathcal{L}(\theta^\star, \eta^\star) = \min_{\eta \in \mathrm{H}} \max_{\theta \in \Theta} \mathcal{L}(\theta, \eta). \tag{A.3.14}$$

In particular, it holds that

$$\theta^\star \in \arg\max_{\theta \in \Theta} \min_{\eta \in \mathrm{H}} \mathcal{L}(\theta, \eta). \tag{A.3.15}$$

[7] Famously called a *Nash equilibrium*.

From the saddle point condition we have $\eta^\star \in \arg\min_{\eta\in\mathrm{H}} \mathcal{L}(\theta^\star, \eta)$. So $(\theta^\star, \eta^\star)$ is a Stackelberg equilibrium. Furthermore we have also proved that all saddle points obey (A.3.14).

Now let $(\theta^\star, \eta^\star)$ be a Stackelberg equilibrium. We claim that it is a saddle point, which completes the proof. By the definition of minimax equilibrium,

$$\max_{\theta\in\Theta} \min_{\eta\in\mathrm{H}} \mathcal{L}(\theta, \eta) = \mathcal{L}(\theta^\star, \eta^\star). \tag{A.3.16}$$

Then by the $\min\max = \max\min$ assumption we have

$$\max_{\theta\in\Theta} \min_{\eta\in\mathrm{H}} \mathcal{L}(\theta, \eta) = \mathcal{L}(\theta^\star, \eta^\star) = \min_{\eta\in\mathrm{H}} \max_{\theta\in\Theta} \mathcal{L}(\theta, \eta). \tag{A.3.17}$$

This proves that all minimax equilibria have the desired objective value $\mathcal{L}(\theta^\star, \eta^\star)$. Now we want to show that

$$\theta^\star \in \arg\max_{\theta\in\Theta} \mathcal{L}(\theta, \eta^\star), \qquad \eta^\star \in \arg\min_{\eta\in\mathrm{H}} \mathcal{L}(\theta^\star, \eta). \tag{A.3.18}$$

Indeed the latter assertion holds by definition of the minimax equilibrium, so only the former need be proved. Namely, we will show that

$$\max_{\theta\in\Theta} \mathcal{L}(\theta, \eta^\star) = \mathcal{L}(\theta^\star, \eta^\star). \tag{A.3.19}$$

To show this note that by definition of the max

$$\mathcal{L}(\theta^\star, \eta^\star) \leq \max_{\theta\in\Theta} \mathcal{L}(\theta, \eta^\star), \tag{A.3.20}$$

meanwhile we have $\min_{\eta\in\mathrm{H}} \mathcal{L}(\theta, \eta) \leq \mathcal{L}(\theta, \eta^\star)$ so

$$\mathcal{L}(\theta^\star, \eta^\star) = \max_{\theta\in\Theta} \min_{\eta\in\mathrm{H}} \mathcal{L}(\theta, \eta) \leq \max_{\theta\in\Theta} \mathcal{L}(\theta, \eta^\star). \tag{A.3.21}$$

Therefore it holds

$$\mathcal{L}(\theta^\star, \eta^\star) = \max_{\theta\in\Theta} \mathcal{L}(\theta, \eta^\star), \tag{A.3.22}$$

and the proof is complete. □

Conditions under which the $\min\max = \max\min$ equality holds are given by so-called *minimax theorems*; the most famous of these is a theorem of von Neumann. However, in the cases we think about, this property usually does not hold.

### A.3.1 Learning Stackelberg Equilibria

How can we learn Stackelberg equilibria via GDA? In general this is clearly impossible, since learning Stackelberg equilibria via GDA is obviously at least as hard as computing a global minimizer of a loss function (say by setting the shared objective $\mathcal{L}(\theta, \eta)$ to only be a function of $\eta$). As such, we can achieve two types of convergence guarantees:

- When $\mathcal{L}$ is (strongly) concave in the first argument $\theta$ and (strongly) convex in the second argument $\eta$ (as well as having Lipschitz gradients in both arguments), we can achieve exponentially fast convergence to a Stackelberg equilibrium.

- When $\mathcal{L}$ is not concave or convex in either argument, we can achieve *local convergence guarantees*: namely, if we initialize the parameter values near a (local) Stackelberg equilibrium and the optimization geometry is good then we can learn that equilibrium efficiently.

The former situation is exactly analogous to the case of single-player optimization, where we proved that gradient descent converges exponentially fast for strongly convex objectives which have Lipschitz gradient. The latter situation is also analogous to the case of single-player optimization, although we did not cover it in depth due to technical difficulty; indeed there exist local convergence guarantees for nonconvex objectives which have locally nice geometry.

The algorithm in these two cases is the same algorithm, called Gradient Descent-Ascent (GDA). To motivate GDA, suppose we are trying to learn $\theta^\star$. We could do gradient ascent on the function $\theta \mapsto \min_{\eta\in\mathrm{H}} \mathcal{L}(\theta, \eta)$. But then we would need to take the derivative in $\theta$ of this function. To see how to do this, suppose that $\mathcal{L}$ is strongly convex in $\eta$ so that there is one minimizer $\eta^\star(\theta)$ of $\mathcal{L}(\theta, \cdot)$. Then, *Danskin's theorem* says that

$$\nabla_\theta \left[\min_{\eta\in\mathrm{H}} \mathcal{L}(\theta, \eta)\right] = \nabla_\theta \mathcal{L}(\theta, \texttt{stop_grad}(\eta^\star(\theta))), \tag{A.3.23}$$

where the gradient is *only with respect to the first argument* (i.e., not a total derivative which would require computing the Jacobian of $\eta^\star(\theta)$ with respect to $\theta$), indicated by the stop-gradient operator.[8] In order to take the derivative in $\theta$, we need to set up a *secondary process* to also optimize $\eta$ to obtain an approximation for $\eta^\star(\theta)$. We can do this through the following algorithmic template:

$$\eta_{k+1} = \eta_{k+1}^{T+1}; \qquad \eta_{k+1}^{t+1} = \eta_{k+1}^{t} - h\nabla_\eta \mathcal{L}(\theta_k, \eta_{k+1}^{t}), \quad \forall t \in [T]; \qquad \eta_{k+1}^{1} = \eta_k \tag{A.3.24}$$

$$\theta_{k+1} = \theta_k + h\nabla_\theta \mathcal{L}(\theta_k, \eta_{k+1}). \tag{A.3.25}$$

That is, we take $T$ steps of gradient descent to update $\eta_k$ (hopefully to the minimizer of $\mathcal{L}(\theta_k, \cdot)$), and then take a gradient ascent step to update $\theta_k$. As a bonus, on top of estimating $\theta_K \approx \arg\max_\theta \min_\eta \mathcal{L}(\theta, \eta)$, we also learn an $\eta_K \approx \arg\min_\eta \mathcal{L}(\theta_K, \eta)$ — this is an approximate Stackelberg equilibrium.

[8] In the case that $\mathcal{L}$ is not strongly convex in $\eta$ but rather just convex, Danskin's theorem can be stated in terms of subdifferentials. If $\mathcal{L}$ is not convex at all in $\eta$, then this derivative may not be well-defined, but one can obtain (local) convergence guarantees for the resulting algorithm anyways. Hence we use Danskin's theorem as a motivation and not a justification for our algorithms. Danskin's theorems can be generalized into more relaxed circumstances by the so-called *envelope theorems*.

This method is often not done in practice, as it requires $T+1$ total gradient descent iterations to update $\theta$ once. Instead, we use the so-called *(simultaneous) Gradient Descent-Ascent* (GDA) iteration

$$\theta_{k+1} = \theta_k + h\nabla_\theta \mathcal{L}(\theta_k, \eta_k) \tag{A.3.26}$$

$$\eta_{k+1} = \eta_k - Th\nabla_\eta \mathcal{L}(\theta_k, \eta_k), \tag{A.3.27}$$

which can be implemented efficiently via a single gradient step on $(\theta, \eta)$. The crucial idea here is, to make our method close to the inefficient iteration above, we use an $\eta$ update which is $T$ times faster than the $\theta$ update (these can be seen as nearly the same by taking a linearization of the dynamics).

It is crucial to pick $T$ sensibly. How can we do that? In the sequel, we discuss two configurations of $T$ which lead to convergence of GDA to a Stackelberg equilibrium under different assumptions.

### Convergence of One-Timescale GDA to Stackelberg Equilibrium

If $T = 1$ (i.e., named *one-timescale* because both $\theta$ and $\eta$ updates are of the same scale), then the GDA algorithm becomes

$$\theta_{k+1} = \theta_k + h\nabla_\theta \mathcal{L}(\theta_k, \eta_k), \qquad \eta_{k+1} = \eta_k - h\nabla_\eta \mathcal{L}(\theta_k, \eta_k). \tag{A.3.28}$$

If $\mathcal{L}$ has Lipschitz gradients in $\theta$ and $\eta$, and is strongly concave in $\theta$ and strongly convex in $\eta$, and is coercive (i.e.,$\lim_{\|\theta\|_2, \|\eta\|_2 \to \infty} \mathcal{L}(\theta, \eta) = \infty$), then all saddle points are Stackelberg equilibria and vice versa. The work [ZAK24] shows that GDA (again, with $T = 1$) with sufficiently small step size $h$ converges to a saddle point (hence Stackelberg equilibrium) exponentially fast if one exists, analogously to gradient descent for strongly convex functions with Lipschitz gradient.[9] To our knowledge, this flavor of results constitute the only known rigorous justification for single-timescale GDA.

### Local Convergence of Two-Timescale GDA to Stackelberg Equilibrium

Strong convexity/concavity is a *global* property, and none of the games we look into in this book have objectives which are globally strongly concave/strongly convex. In this case, the best we can hope for is *local* convergence to Stackelberg equilibria: if the parameters are initialized close to a Stackelberg equilibrium, then GDA can converge onto it, given an appropriate step size $h$ and timescale $T$.

In fact, our results also hold for a version of the *local* Stackelberg equilibrium called the *differential Stackelberg equilibrium*, which was introduced in [FCR19] (though we use the precise definition in [LFD+22]), and which we define as follows. A point $(\theta^\star, \eta^\star)$ is a differential Stackelberg equilibrium if:

[9] We do not provide the step size or exponential base since they are complicated functions of the strong convexity/concavity/Lipschitz constant of the gradient. Of course, the paper [ZAK24] provides the precise parameter values.

- $\nabla_\eta \mathcal{L}(\theta^\star, \eta^\star) = \mathbf{0}$;
- $\nabla^2_\eta \mathcal{L}(\theta^\star, \eta^\star)$ is symmetric positive definite;
- $\nabla_\theta \mathcal{L}(\theta^\star, \eta^\star) = \mathbf{0}$;
- $(\nabla^2_\theta \mathcal{L} + [\frac{\mathrm{d}}{\mathrm{d}\theta}\nabla_\eta \mathcal{L}][\nabla^2_\eta \mathcal{L}]^{-1}[\frac{\mathrm{d}}{\mathrm{d}\eta}\nabla_\theta \mathcal{L}])(\theta^\star, \eta^\star)$ is symmetric negative definite.

Notice that the last condition asks for the (total) Hessian

$$\nabla^2 \mathcal{L}(\theta, \eta) = \begin{bmatrix} \nabla^2_\theta \mathcal{L}(\theta, \eta) & \frac{\mathrm{d}}{\mathrm{d}\eta}\nabla^2_\theta \mathcal{L}(\theta, \eta) \\ \frac{\mathrm{d}}{\mathrm{d}\theta}\nabla^2_{\eta\theta} \mathcal{L}(\theta, \eta) & \nabla^2_\eta \mathcal{L}(\theta, \eta) \end{bmatrix} \tag{A.3.29}$$

or, equivalently, its Schur complement to be negative definite. If we look at the computation of $\nabla_\theta[\min_{\eta \in \mathrm{H}} \mathcal{L}(\theta, \eta)]$ furnished by Danskin's theorem, the last two criteria are actually constraints on the gradient and Hessian of the function $\theta \mapsto \min_{\eta \in \mathrm{H}} \mathcal{L}(\theta, \eta)$, ensuring that the gradient is 0 and the Hessian is negative semidefinite. This intuition tells us that we can expect that each Stackelberg equilibrium is a differential Stackelberg equilibrium; [FCR20] confirms this rigorously (up to some technical conditions).

Analogously to the notion of strict local optimum in single-player optimization (where we require $\nabla^2 \mathcal{L}(\theta^\star)$ to be positive semidefinite), the definition of differential Stackelberg equilibrium *implies that* $\mathcal{L}(\theta, \cdot)$ *is locally (strictly) convex in a neighborhood of the equilibrium, and that* $\min_{\eta \in \mathrm{H}} \mathcal{L}(\cdot, \eta)$ *is locally (strictly) concave in the same region.*

In this context, we present the result from [LFD+22]. Let $(\theta^\star, \eta^\star)$ be a differential Stackelberg equilibrium. Suppose that $\mathcal{L}$ has Lipschitz gradients, i.e.,

$$\max\left\{ \|\nabla^2_\theta \mathcal{L}(\theta^\star, \eta^\star)\|_2, \left\|\nabla^2_\eta \mathcal{L}(\theta^\star, \eta^\star)\right\|_2, \left\|\frac{\mathrm{d}}{\mathrm{d}\eta}\nabla_\theta \mathcal{L}(\theta^\star, \eta^\star)\right\|_2 \right\} \le \beta. \tag{A.3.30}$$

Further define the local strong convexity/concavity parameters of $\mathcal{L}(\theta, \cdot)$ and $\min_\eta \mathcal{L}(\cdot, \eta)$ respectively as

$$\mu_\eta = \lambda_{\min}(\nabla^2_\eta \mathcal{L}(\theta^\star, \eta^\star)), \tag{A.3.31}$$

$$\mu_\theta = \min\left\{\beta, -\left(\nabla^2_\theta \mathcal{L} + \left[\frac{\mathrm{d}}{\mathrm{d}\theta}\nabla_\eta \mathcal{L}\right][\nabla^2_\eta \mathcal{L}]^{-1}\left[\frac{\mathrm{d}}{\mathrm{d}\eta}\nabla_\theta \mathcal{L}\right]\right)(\theta^\star, \eta^\star)\right\}. \tag{A.3.32}$$

Then define the local condition numbers as

$$\kappa_\eta = \beta/\mu_\eta, \qquad \kappa_\theta = \beta/\mu_\theta. \tag{A.3.33}$$

The paper [LFD+22] says that if we take step size $h = \frac{1}{4\beta T}$ and take $T \ge 2\kappa_\eta$, so that the algorithm is

$$\theta_{k+1} = \theta_k + \frac{1}{4\beta T}\nabla_\theta \mathcal{L}(\theta_k, \eta_k), \tag{A.3.34}$$

$$\eta_{k+1} = \eta_k - \frac{1}{4\beta}\nabla_\eta \mathcal{L}(\theta_k, \eta_k), \tag{A.3.35}$$

the total Hessian $\nabla^2\mathcal{L}(\theta^\star, \eta^\star)$ is diagonalizable, and we initialize $(\theta_0, \eta_0)$ close enough to $(\theta^\star, \eta^\star)$, then there are positive constants $c_0, c_1 > 0$ such that

$$\|(\theta_k, \eta_k) - (\theta^\star, \eta^\star)\|_2 \leq c_0 \left(1 - \frac{c_1}{T\kappa_\theta}\right)^k \|(\theta_0, \eta_0) - (\theta^\star, \eta^\star)\|_2, \tag{A.3.36}$$

implying exponential convergence to the differential Stackelberg equilibrium.

### A.3.2 Practical Considerations when Learning Stackelberg Equilibria

In practice, we do not know how to initialize parameters close to a (differential) Stackelberg equilibrium. Due to symmetries within the objective, including those induced by overparameterization of the neural networks being trained, one can (heuristically) expect that most initializations are close to a Stackelberg equilibrium. Also, we do not know how to compute the step size $h$ or the timescale $T$, since they are dependent on properties of the loss $\mathcal{L}$ at the equilibrium. In practice, there are some common approaches:

- Take $T = 1$ (equal step-sizes), and use updates for $\theta$ and $\eta$ that are derived from a learning-rate-adaptive optimizer like Adam (as opposed to vanilla GD). Here, you *hope* (but do not *know*) that the optimizer can adjust the learning rates to learn a good equilibrium.
- Take $T$ to be some constant like $T = 10^6$ which implies that $\eta$ equilibrates $10^6$ times as fast as $\theta$. Here you can also use Adam-style updates, and hope that it fixes the time scale.
- Let $T$ depend on the iteration $k$, and let $T_k \to \infty$ as $k \to \infty$. This schedule was studied (also in the case of noise) by Borkar in [Bor97].

For example, you can use this while training CTRL-style models (see Chapter 6), where the encoder is Player 1 and the decoder is Player 2. Some theory about CTRL is given in Theorem 6.1.

## A.4 Exercises

*Exercise* A.1. We have shown that for a smooth function $f$, gradient descent converges linearly to the global optimum if it is strongly convex. However, in general nonconvex optimization, we do not have convexity, let alone strong convexity. Fortunately, in some cases, $f$ satisfies the so-called $\mu$-Polyak-Lojasiewicz (PL) inequality, i.e., there exists a constant $\mu > 0$ such that for all $\theta$,

$$\frac{1}{2}\|\nabla f(\theta)\|_2^2 \geq \mu\left(f(\theta) - f(\theta^\star)\right),$$

where $\theta^*$ is a minimizer of $f$.

Please show that under the PL inequality and the assumption that $f$ is $\beta$-smooth, gradient descent (A.1.12) converges linearly to $\theta^*$.

*Exercise* A.2. Compute the differential and adjoint derivative of the softmax function, defined as follows.

$$\operatorname{softmax}\left(\begin{bmatrix} x_1 \\ \vdots \\ x_n \end{bmatrix}\right) = \frac{1}{\sum_{i=1}^{n} e^{x_i}} \begin{bmatrix} e^{x_1} \\ \vdots \\ e^{x_n} \end{bmatrix}. \tag{A.4.1}$$

*Exercise* A.3. Carry through the backpropagation computation for a $L$-layer MLP, as defined in Section 8.2.3.

*Exercise* A.4. Carry through the backpropagation computation for a $L$-layer transformer, as defined in Section 8.2.3.

*Exercise* A.5. Carry through the backpropagation computation for an autoencoder with $L$ encoder layers and $L$ decoder layers (without necessarily specifying an architecture).

# Appendix B

# Entropy, Diffusion, Denoising, and Lossy Coding

"*The increase of disorder or entropy with time is one example of what is called an arrow of time, something that distinguishes the past from the future, giving a direction to time.*"

– A Brief History of Time, Stephen Hawking

In this appendix we provide proofs for several facts, mentioned in Chapter 3, which are related to differential entropy, how it evolves under diffusion processes, and its connections to lossy coding. We will make the following mild assumption about the random variable representing the data source, denoted $\boldsymbol{x}$.

**Assumption B.1.** $\boldsymbol{x}$ is supported on a compact set $\mathcal{S} \subseteq \mathbb{R}^D$ of radius at most $R$, i.e., $R \doteq \sup_{\boldsymbol{\xi} \in \mathcal{S}} \|\boldsymbol{\xi}\|_2$.

In particular, since compact sets in Euclidean space are bounded, it holds $R < \infty$. We will consistently use the notation $B_r(\boldsymbol{\xi}) \doteq \{\boldsymbol{u} \in \mathbb{R}^D : \|\boldsymbol{\xi} - \boldsymbol{u}\|_2 \leq r\}$ to denote the Euclidean ball of radius $r$ centered at $\boldsymbol{\xi}$. In this sense, Assumption B.1 has $\mathcal{S} \subseteq B_R(\boldsymbol{0})$.

Notice that this assumption holds for (almost) all variables we care about in practice, as it is (often) imposed by a normalization step during data preprocessing.

## B.1 Differential Entropy of Low-Dimensional Distributions

In this short appendix we discuss the differential entropy of low-dimensional distributions. By definition, the differential entropy of a random variable $\boldsymbol{x}$ which does not have a density is $-\infty$; this includes all random variables supported on low-dimensional sets. The objective of this section is to discuss why this is a "morally correct" value.

In fact, let $\boldsymbol{x}$ be any random variable such that Assumption B.1 holds, the support $\mathcal{S}$ of $\boldsymbol{x}$ has 0 volume.[1] We will consider the case that $\boldsymbol{x}$ is uniform on $\mathcal{S}$.[2] Our goal is to compute $h(\boldsymbol{x})$.

In this case, $\boldsymbol{x}$ would not have a density; in the counterfactual world where we did not know $h(\boldsymbol{x}) = -\infty$, we could not directly define it using the standard definition of differential entropy. Instead, as in the rest of analysis and information theory it would be reasonable to consider the limit of entropies of successively better approximations $\boldsymbol{x}_\varepsilon$ of $\boldsymbol{x}$ which have densities, i.e.,

$$h(\boldsymbol{x}) \text{ ``}=\text{''} \lim_{\varepsilon \searrow 0} h(\boldsymbol{x}_\varepsilon). \tag{B.1.1}$$

To this end, the basic idea is to take an $\varepsilon$-thickening of $\mathcal{S}$, say $\mathcal{S}_\varepsilon$ defined as

$$\mathcal{S}_\varepsilon = \bigcup_{\boldsymbol{\xi} \in \mathcal{S}} B_\varepsilon(\boldsymbol{\xi}) \tag{B.1.2}$$

and visualized in Figure B.1. We will work with random variables whose support is $\mathcal{S}_\varepsilon$, which is fully-dimensional, and take the limit as $\varepsilon \to 0$. Indeed, define $\boldsymbol{x}_\varepsilon \sim \mathcal{U}(\mathcal{S}_\varepsilon)$. Since $\mathcal{S}_\varepsilon$ has positive volume, $\boldsymbol{x}_\varepsilon$ has a density $p_\varepsilon$ equal to

$$p_\varepsilon(\boldsymbol{\xi}) = \mathbf{1}(\boldsymbol{\xi} \in \mathcal{S}_\varepsilon) \cdot \frac{1}{\operatorname{vol}(\mathcal{S}_\varepsilon)}. \tag{B.1.3}$$

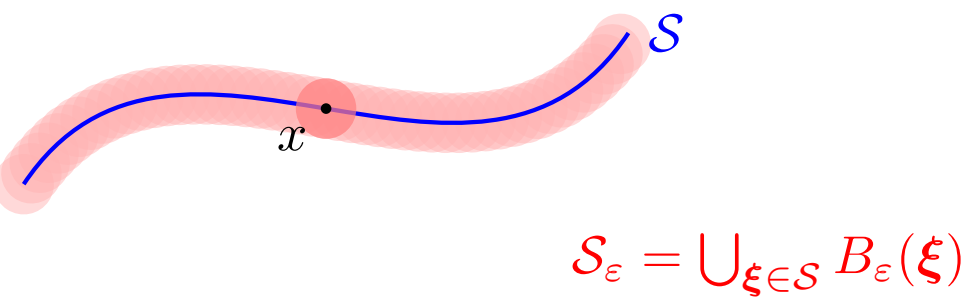

Figure B.1: Illustration of the $\varepsilon$-thickening $\mathcal{S}_\varepsilon$ of a curve $\mathcal{S} \subseteq \mathbb{R}^2$.

Computing the entropy of $\boldsymbol{x}_\varepsilon$ using the convention that $0 \log 0 = 0$, it holds

$$h(\boldsymbol{x}_\varepsilon) = -\int_{\mathbb{R}^D} p_\varepsilon(\boldsymbol{\xi}) \log p_\varepsilon(\boldsymbol{\xi}) \mathrm{d}\boldsymbol{\xi} \tag{B.1.4}$$

[1] Formally this means that $\mathcal{S}$ is Borel measurable with Borel measure 0.

[2] Say, w.r.t. the Hausdorff measure on $\mathcal{S}$.

$$= -\int_{\mathcal{S}_\varepsilon} \frac{1}{\mathrm{vol}(\mathcal{S}_\varepsilon)} \log\left(\frac{1}{\mathrm{vol}(\mathcal{S}_\varepsilon)}\right) \mathrm{d}\boldsymbol{\xi} \tag{B.1.5}$$

$$= \frac{\log(\mathrm{vol}(\mathcal{S}_\varepsilon))}{\mathrm{vol}(\mathcal{S}_\varepsilon)} \int_{\mathcal{S}_\varepsilon} \mathrm{d}\boldsymbol{\xi} \tag{B.1.6}$$

$$= \log(\mathrm{vol}(\mathcal{S}_\varepsilon)). \tag{B.1.7}$$

Since $\mathcal{S}$ is compact $\mathrm{vol}(\mathcal{S}_\varepsilon)$ is finite and tends to 0 as $\varepsilon \searrow 0$. Thus

$$h(\boldsymbol{x}) = \lim_{\varepsilon \searrow 0} h(\boldsymbol{x}_\varepsilon) = \lim_{\varepsilon \searrow 0} \log(\mathrm{vol}(\mathcal{S}_\varepsilon)) = -\infty, \tag{B.1.8}$$

as desired.

The above calculation is actually a corollary of a much more famous and celebrated set of results about the maximum possible entropy of $\boldsymbol{x}$ subject to certain constraints on the distribution of $\boldsymbol{x}$. We would be remiss to not provide the results here; the proofs are provided in Chapter 2 of [PW22], for example.

**Theorem B.1.** *Let $\boldsymbol{x}$ be a random variable on $\mathbb{R}^D$.*

1. *If $\boldsymbol{x}$ is supported on a compact set $\mathcal{S} \subseteq \mathbb{R}^D$ (i.e., Assumption B.1) then*

$$h(\boldsymbol{x}) \leq h(\mathcal{U}(\mathcal{S})) = \log \mathrm{vol}(\mathcal{S}). \tag{B.1.9}$$

2. *If $\boldsymbol{x}$ has finite covariance such that, for a PSD matrix $\boldsymbol{\Sigma} \in \mathsf{PSD}(D)$, it holds $\mathrm{Cov}(\boldsymbol{x}) \preceq \boldsymbol{\Sigma}$ (w.r.t. the PSD ordering, i.e., $\boldsymbol{\Sigma} - \mathrm{Cov}(\boldsymbol{x})$ is PSD), then*

$$h(\boldsymbol{x}) \leq h(\mathcal{N}(\mathbf{0}, \boldsymbol{\Sigma})) = \frac{1}{2}\log((2\pi e)^D \det \boldsymbol{\Sigma}). \tag{B.1.10}$$

3. *If $\boldsymbol{x}$ has finite second moment such that, for a constant $a \geq 0$, it holds $\mathbb{E}\,\|\boldsymbol{x}\|_2^2 \leq a$, then*

$$h(\boldsymbol{x}) \leq h\left(\mathcal{N}\left(\mathbf{0}, \frac{a}{D}\boldsymbol{I}\right)\right) = \frac{D}{2}\log\frac{2\pi e a}{D}. \tag{B.1.11}$$

## B.2 Diffusion and Denoising Processes

In the main body (Chapter 3), we considered a random variable $\boldsymbol{x}$, and a stochastic process defined by (3.2.1), i.e.,

$$\boldsymbol{x}_t = \boldsymbol{x} + t\boldsymbol{g}, \qquad \forall t \in [0, T] \tag{B.2.1}$$

where $\boldsymbol{g} \sim \mathcal{N}(\mathbf{0}, \boldsymbol{I})$ independently of $\boldsymbol{x}$.

The structure of this section is as follows. In Section B.2.1 we provide a formal theorem and crisp proof which shows that under Equation (B.2.1) the entropy increases, i.e., $\frac{\mathrm{d}}{\mathrm{d}t}h(\boldsymbol{x}_t) > 0$. In Section B.2.2 we provide a formal theorem and crisp proof which shows that under Equation (B.2.1), the entropy decreases during denoising, i.e., $h(\mathbb{E}[\boldsymbol{x}_s \mid \boldsymbol{x}_t]) < h(\boldsymbol{x}_t)$ for all $s < t$. In Section B.2.3 we provide proofs for technical lemmas that are needed to establish the claims in the previous subsections.

Before we start, we introduce some key notations. First, let $\varphi_t$ be the density of $\mathcal{N}(\mathbf{0}, t^2\boldsymbol{I})$, i.e.,

$$\varphi_t(\boldsymbol{\xi}) \doteq \frac{1}{(2\pi)^{D/2}t^D}\exp\left(-\frac{\|\boldsymbol{\xi}\|_2^2}{2t^2}\right). \tag{B.2.2}$$

Next, $\boldsymbol{x}_t$ is supported on all of $\mathbb{R}^D$, so it has a *density*, which we denote $p_t$ (as in the main body). A quick calculation shows that

$$p_t(\boldsymbol{\xi}) = \mathbb{E}[\varphi_t(\boldsymbol{\xi} - \boldsymbol{x})], \tag{B.2.3}$$

and from this representation it is possible to deduce (i.e., from Proposition B.4) that $p_t$ is smooth (i.e., infinitely differentiable) in $\boldsymbol{\xi}$, also smooth in $t$, and positive everywhere. This fact is somewhat remarkable at first sight: even for a completely irregular random variable $\boldsymbol{x}$ (say, a Bernoulli random variable, which does not have a density), its Gaussian smoothing admits a density for every (arbitrarily small) $t > 0$. The proof is left as an exercise for readers well-versed in mathematical analysis.

However, we also need to add an assumption about the *smoothness* of the distribution of $\boldsymbol{x}$, which will eliminate some technical problems that occur around $t = 0$ with low-dimensional distributions.[3] Despite this, we expect that our results hold under milder assumptions with additional work. For now, let us assume:

**Assumption B.2.** $\boldsymbol{x}$ has a twice continuously differentiable density, denoted $p$.

### B.2.1 Diffusion Process Increases Entropy Over Time

In this section we provide a proof of Theorem B.2. For convenience, we restate it as follows.

**Theorem B.2** (Diffusion Increases Entropy)**.** *Let $\boldsymbol{x}$ be any random variable such that Assumptions B.1 and B.2 hold, and let $(\boldsymbol{x}_t)_{t\in[0,T]}$ be the stochastic process* (B.2.1)*. Then*

$$h(\boldsymbol{x}_s) < h(\boldsymbol{x}_t), \qquad \forall s, t\colon 0 \le s < t \le T. \tag{B.2.4}$$

*Proof.* Before we start, we must ask: when does the inequality in (B.2.4) make sense? We will show in Lemma B.1 that under our assumptions, the differential entropy is well-defined, is never $+\infty$, and for $t > 0$ is finite, so the (strict) inequality in (B.2.4) makes sense.

The question of well-definedness aside, the crux of this proof is to show that the density $p_t$ of $\boldsymbol{x}_t$ satisfies a particular partial differential equation, which is very similar to the *heat equation*. The heat equation is a famous PDE which describes the diffusion of heat through space. This intuitively should make sense, and paints a mental picture: as the time $t$ increases, the probability from the original (perhaps tightly concentrated) $\boldsymbol{x}$ disperses across all of $\mathbb{R}^D$ like heat radiating from a source in a vacuum.

[3] As then various quantities become highly irregular and dealing with them would require significant additional analysis.

Such PDEs for $p_t$, known as *Fokker-Planck equations* for more general stochastic processes, are very powerful tools, as they allow us to describe the instantaneous temporal derivatives of $p_t$ in terms of the instantaneous spatial derivatives of $p_t$, and vice versa, providing a concise description of the regularity and dynamics of $p_t$. Once we obtain dynamics for $p_t$, we can then use the system to obtain another one which describes the dynamics of $h(\boldsymbol{x}_t)$, which after all is just a functional of $p_t$.

The description of the PDE involves a mathematical object called the Laplacian $\Delta$. Recall from your multivariable calculus class that the Laplacian operating on a differentiable-in-time and twice-differentiable-in-space function $f\colon (0,T)\times\mathbb{R}^D \to \mathbb{R}$ is given by

$$\Delta f_t(\boldsymbol{\xi}) = \operatorname{tr}(\nabla^2 f_t(\boldsymbol{\xi})) = \sum_{i=1}^{D} \frac{\partial^2 f_t}{\partial \xi_i^2}(\boldsymbol{\xi}). \tag{B.2.5}$$

Namely, from using the integral representation of $p_t$ and differentiating under the integral, we can compute the derivatives of $p_t$ (which we do in Proposition B.1) and observe that $p_t$ satisfies the heat-like PDE

$$\frac{\partial p_t}{\partial t}(\boldsymbol{\xi}) = t\Delta p_t(\boldsymbol{\xi}). \tag{B.2.6}$$

Then for finding the dynamics of $h(\boldsymbol{x}_t)$, we can use Proposition B.3 again as well as the heat-like PDE to get

$$\begin{aligned}
\frac{\mathrm{d}}{\mathrm{d}t} h(\boldsymbol{x}_t) &= -\frac{\mathrm{d}}{\mathrm{d}t}\int_{\mathbb{R}^D} p_t(\boldsymbol{\xi}) \log p_t(\boldsymbol{\xi}) \mathrm{d}\boldsymbol{\xi} && \text{(B.2.7)}\\
&= -\int_{\mathbb{R}^D} \frac{\partial}{\partial t}\left[p_t(\boldsymbol{\xi}) \log p_t(\boldsymbol{\xi})\right] \mathrm{d}\boldsymbol{\xi} && \text{(B.2.8)}\\
&= -\int_{\mathbb{R}^D} \frac{\partial p_t}{\partial t}(\boldsymbol{\xi})[1 + \log p_t(\boldsymbol{\xi})] \mathrm{d}\boldsymbol{\xi} && \text{(B.2.9)}\\
&= -t\int_{\mathbb{R}^D} \Delta p_t(\boldsymbol{\xi})[1 + \log p_t(\boldsymbol{\xi})] \mathrm{d}\boldsymbol{\xi}. && \text{(B.2.10)}
\end{aligned}$$

By using a slightly involved integration by parts argument (Lemma B.2), we obtain

$$\begin{aligned}
\frac{\mathrm{d}}{\mathrm{d}t} h(\boldsymbol{x}_t) &= t\int_{\mathbb{R}^D} \langle \nabla \log p_t(\boldsymbol{\xi}), \nabla p_t(\boldsymbol{\xi})\rangle \mathrm{d}\boldsymbol{\xi} && \text{(B.2.11)}\\
&= t\int_{\mathbb{R}^D} \frac{\|\nabla p_t(\boldsymbol{\xi})\|_2^2}{p_t(\boldsymbol{\xi})} \mathrm{d}\boldsymbol{\xi} && \text{(B.2.12)}
\end{aligned}$$

$$> 0 \tag{B.2.13}$$

where strict inequality holds in the last line because, for it to not hold, $\nabla p_t(\boldsymbol{\xi})$ would need to vanish almost everywhere (i.e., everywhere except possibly on a set of zero volume), but this would imply that $p_t$ would be constant almost everywhere, a contradiction with a fact that $p_t$ is a density.

To complete the proof we just use the fundamental theorem of calculus

$$h(\boldsymbol{x}_t) = h(\boldsymbol{x}_s) + \int_s^t \frac{\mathrm{d}}{\mathrm{d}u} h(\boldsymbol{x}_u)\mathrm{d}u > h(\boldsymbol{x}_s), \tag{B.2.14}$$

which proves the claim. (Note that this does not make sense when $h(\boldsymbol{x}_s) = -\infty$, which can only happen when $s = 0$ and $h(\boldsymbol{x}) = -\infty$, but in this case $h(\boldsymbol{x}_t) > -\infty$ so the claim is vacuously true anyways.) □

### B.2.2 Denoising Process Reduces Entropy Over Time

Recall that in Section 3.2.1 we start with the random variable $\boldsymbol{x}_T$ and iteratively denoise it using iterations of the form

$$\hat{\boldsymbol{x}}_s \doteq \mathbb{E}[\boldsymbol{x}_s \mid \boldsymbol{x}_t = \hat{\boldsymbol{x}}_t] = \frac{s}{t}\hat{\boldsymbol{x}}_t + \left(1 - \frac{s}{t}\right)\bar{\boldsymbol{x}}^*(t, \hat{\boldsymbol{x}}_t). \tag{B.2.15}$$

for $s, t \in \{t_0, t_1, \ldots, t_L\}$ with $s < t$ and $\boldsymbol{x}_T = \hat{\boldsymbol{x}}_T$. We wish to prove that $h(\hat{\boldsymbol{x}}_s) < h(\hat{\boldsymbol{x}}_t)$, showing that the denoising process actually reduces the entropy.

Before we go about doing this, we make several remarks about the problem statement. First, Tweedie's formula (3.2.23) says that

$$\bar{\boldsymbol{x}}^*(t, \boldsymbol{x}_t) = \boldsymbol{x}_t + t^2 \nabla p_t(\boldsymbol{x}_t), \tag{B.2.16}$$

which likens a full denoising step from time $t$ to time 0 to a gradient step on the log-density of $\boldsymbol{x}_t$. Can we get a similar result for the full denoising step from time $t$ to time $s$ in (B.2.15)? It turns out that indeed we can, and it is pretty simple. By using (B.2.15) and Tweedie's formula (3.2.23), we obtain

$$\mathbb{E}[\boldsymbol{x}_s \mid \boldsymbol{x}_t] = \frac{s}{t}\boldsymbol{x}_t + \left(1 - \frac{s}{t}\right)\left(\boldsymbol{x}_t + t^2 \nabla_{\boldsymbol{x}_t} \log p_t(\boldsymbol{x}_t)\right) = \boldsymbol{x}_t + \left(1 - \frac{s}{t}\right) t^2 \nabla_{\boldsymbol{x}_t} \log p_t(\boldsymbol{x}_t). \tag{B.2.17}$$

So this iterative denoising step is again a gradient step on the perturbed log-density $\log p_t$ with a shrunken step size. In particular, this step can be seen as a perturbation of the distribution of the random variable $\boldsymbol{x}_t$ by the *score function vector field*, suggesting a connection to stochastic differential equations (SDEs) and the theory of diffusion models [SSK+21]. Indeed, a proof of the following result Theorem B.3 can be developed using this powerful machinery and a limiting argument (e.g., following the technical approach in the exposition of [CCL+23]). We will give a simpler proof here, which will use only elementary tools and thereby illuminate some of the key quantities behind the process of entropy reduction via denoising. On the other hand, we will need to deal with some slightly technical calculations due to the fact that the denoising process in Theorem B.3 does *not* correspond exactly to the *reverse* process associated to the noise addition process that generates the observation $\boldsymbol{x}_t$.[4]

We want to prove that $h(\mathbb{E}[\boldsymbol{x}_s \mid \boldsymbol{x}_t]) < h(\boldsymbol{x}_t)$, i.e., formally:

**Theorem B.3.** *Let $\boldsymbol{x}$ be any random variable such that Assumptions B.1 and B.2 hold, and let $(\boldsymbol{x}_t)_{t\in[0,T]}$ be the stochastic process (B.2.1). For each $t > 0$, write*

$$J(p_t) \doteq \int_{\mathbb{R}^D} \frac{\|\nabla p_t(\boldsymbol{\xi})\|_2^2}{p_t(\boldsymbol{\xi})} \mathrm{d}\boldsymbol{\xi}, \qquad U_t \doteq \max\left(\frac{D}{t^2}, \left|\frac{R^2}{t^4} - \frac{D}{t^2}\right|\right) \tag{B.2.18}$$

*for the Fisher information of $p_t$ and a uniform bound on $|\Delta \log p_t|$, respectively. Then*

$$h(\mathbb{E}[\boldsymbol{x}_s \mid \boldsymbol{x}_t]) < h(\boldsymbol{x}_t) \tag{B.2.19}$$

*for all $0 \leq s < t \leq T$ such that*

$$\left(1 - \frac{s}{t}\right) t^2 U_t^2 \exp\left(\left(1 - \frac{s}{t}\right) t^2 U_t\right) < 2J(p_t). \tag{B.2.20}$$

*Proof.* This proof uses two main ideas:

1. First, write down a density for $\mathbb{E}[\boldsymbol{x}_s \mid \boldsymbol{x}_t]$ using a change-of-variables formula.

2. Second, bound this density to control the entropy.

The change of variables is justified by Corollary B.1, which was originally derived in [Gri11].

We execute these ideas now. From Corollary B.1, we obtain that the function $\bar{\boldsymbol{x}}$ defined as $\bar{\boldsymbol{x}}(\boldsymbol{\xi}) \doteq \mathbb{E}[\boldsymbol{x}_s \mid \boldsymbol{x}_t = \boldsymbol{\xi}]$ is differentiable, injective, and thus invertible on its range, which we henceforth denote $\mathcal{X} \subseteq \mathbb{R}^D$. We denote its inverse as $\bar{\boldsymbol{x}}^{-1}$. Using a change-of-variables formula, the density $\bar{p}$ of $\bar{\boldsymbol{x}}(\boldsymbol{x}_t)$ is given by

$$\bar{p}(\boldsymbol{\xi}) \doteq \frac{(p_t \circ \bar{\boldsymbol{x}}^{-1})(\boldsymbol{\xi})}{\det(\bar{\boldsymbol{x}}'(\bar{\boldsymbol{x}}^{-1}(\boldsymbol{\xi})))}, \tag{B.2.21}$$

where (recall, from Section A.2) $\bar{\boldsymbol{x}}'$ is the Jacobian of $\bar{\boldsymbol{x}}$. Since from Lemma B.3 we know $\bar{\boldsymbol{x}}'$ is a positive definite matrix, the determinant is positive and so the whole density is positive. Then it follows that

$$h(\bar{\boldsymbol{x}}(\boldsymbol{x}_t)) = -\int_{\mathcal{X}} \frac{(p_t \circ \bar{\boldsymbol{x}}^{-1})(\boldsymbol{\xi})}{\det(\bar{\boldsymbol{x}}'(\bar{\boldsymbol{x}}^{-1}(\boldsymbol{\xi})))} \log \frac{(p_t \circ \bar{\boldsymbol{x}}^{-1})(\boldsymbol{\xi})}{\det(\bar{\boldsymbol{x}}'(\bar{\boldsymbol{x}}^{-1}(\boldsymbol{\xi})))} \mathrm{d}\boldsymbol{\xi} \tag{B.2.22}$$

$$= -\int_{\mathcal{X}} \frac{(p_t \circ \bar{\boldsymbol{x}}^{-1})(\boldsymbol{\xi})}{\det(\bar{\boldsymbol{x}}'(\bar{\boldsymbol{x}}^{-1}(\boldsymbol{\xi})))} \log((p_t \circ \bar{\boldsymbol{x}}^{-1})(\boldsymbol{\xi})) \mathrm{d}\boldsymbol{\xi} \tag{B.2.23}$$

[4] For those familiar with diffusion models, we refer here to the time-reversed forward process not coinciding with the sequence of iterates generated by the process defined by Theorem B.3. These processes coincide in a certain limit where infinitely many steps of Theorem B.3 are taken with infinitely small levels of noise added at each step; for general, finite steps, we must introduce some approximations regardless of the level of sophistication of our tools.

$$+\int_{\mathcal{X}} \frac{(p_t \circ \bar{\boldsymbol{x}}^{-1})(\boldsymbol{\xi})}{\det(\bar{\boldsymbol{x}}'(\bar{\boldsymbol{x}}^{-1}(\boldsymbol{\xi})))} \log\det\big(\bar{\boldsymbol{x}}'(\bar{\boldsymbol{x}}^{-1}(\boldsymbol{\xi}))\big)\mathrm{d}\boldsymbol{\xi} \tag{B.2.24}$$

$$= -\int_{\mathbb{R}^D} p_t(\boldsymbol{\xi})\log p_t(\boldsymbol{\xi})\mathrm{d}\boldsymbol{\xi} + \int_{\mathcal{X}} \frac{(p_t \circ \bar{\boldsymbol{x}}^{-1})(\boldsymbol{\xi})}{\det(\bar{\boldsymbol{x}}'(\bar{\boldsymbol{x}}^{-1}(\boldsymbol{\xi})))} \log\det\big(\bar{\boldsymbol{x}}'(\bar{\boldsymbol{x}}^{-1}(\boldsymbol{\xi}))\big)\mathrm{d}\boldsymbol{\xi} \tag{B.2.25}$$

$$= h(\boldsymbol{x}_t) - \int_{\mathcal{X}} \frac{(p_t \circ \bar{\boldsymbol{x}}^{-1})(\boldsymbol{\xi})}{\det(\bar{\boldsymbol{x}}'(\bar{\boldsymbol{x}}^{-1}(\boldsymbol{\xi})))} \log\left(\frac{1}{\det(\bar{\boldsymbol{x}}'(\bar{\boldsymbol{x}}^{-1}(\boldsymbol{\xi})))}\right)\mathrm{d}\boldsymbol{\xi}. \tag{B.2.26}$$

We will study the last term (including the $-$), and show that it is negative.

By concavity, one has $-x\log x \le 1 - x$ for every $x \ge 0$. Hence

$$h(\bar{\boldsymbol{x}}(\boldsymbol{x}_t)) - h(\boldsymbol{x}_t) = -\int_{\mathcal{X}} \frac{(p_t \circ \bar{\boldsymbol{x}}^{-1})(\boldsymbol{\xi})}{\det(\bar{\boldsymbol{x}}'(\bar{\boldsymbol{x}}^{-1}(\boldsymbol{\xi})))} \log\left(\frac{1}{\det(\bar{\boldsymbol{x}}'(\bar{\boldsymbol{x}}^{-1}(\boldsymbol{\xi})))}\right)\mathrm{d}\boldsymbol{\xi} \tag{B.2.27}$$

$$\le \int_{\mathcal{X}} (p_t \circ \bar{\boldsymbol{x}}^{-1})(\boldsymbol{\xi}) \cdot \left(1 - \frac{1}{\det(\bar{\boldsymbol{x}}'(\bar{\boldsymbol{x}}^{-1}(\boldsymbol{\xi})))}\right)\mathrm{d}\boldsymbol{\xi} \tag{B.2.28}$$

$$= \int_{\mathcal{X}} (p_t \circ \bar{\boldsymbol{x}}^{-1})(\boldsymbol{\xi})\mathrm{d}\boldsymbol{\xi} - \int_{\mathcal{X}} \frac{(p_t \circ \bar{\boldsymbol{x}}^{-1})(\boldsymbol{\xi})}{\det(\bar{\boldsymbol{x}}'(\bar{\boldsymbol{x}}^{-1}(\boldsymbol{\xi})))}\mathrm{d}\boldsymbol{\xi} \tag{B.2.29}$$

$$= \int_{\mathbb{R}^D} p_t(\boldsymbol{\xi})\det\big(\bar{\boldsymbol{x}}'(\bar{\boldsymbol{x}}^{-1}(\boldsymbol{\xi}))\big)\mathrm{d}\boldsymbol{\xi} - \int_{\mathcal{X}} \bar{p}(\boldsymbol{\xi})\mathrm{d}\boldsymbol{\xi} \tag{B.2.30}$$

$$= \int_{\mathbb{R}^D} p_t(\boldsymbol{\xi})\det\left(\boldsymbol{I} + \left(1 - \frac{s}{t}\right)t^2\nabla^2\log p_t(\boldsymbol{\xi})\right)\mathrm{d}\boldsymbol{\xi} - 1. \tag{B.2.31}$$

Now, by the AM-GM inequality on eigenvalues, we have for any symmetric positive definite matrix $\boldsymbol{M} \in \mathsf{PSD}(D)$ the bound

$$\det(\boldsymbol{M})^{1/D} = \prod_{i=1}^{D} \lambda_i(\boldsymbol{M})^{1/D} \le \frac{\sum_{i=1}^{D}\lambda_i(\boldsymbol{M})}{D} = \frac{\mathrm{tr}(\boldsymbol{M})}{D}, \tag{B.2.32}$$

which we can apply to the above expression and obtain

$$\int_{\mathbb{R}^D} p_t(\boldsymbol{\xi})\det\left(\boldsymbol{I} + \left(1 - \frac{s}{t}\right)t^2\nabla^2\log p_t(\boldsymbol{\xi})\right)\mathrm{d}\boldsymbol{\xi} \tag{B.2.33}$$

$$\le \int_{\mathbb{R}^D} p_t(\boldsymbol{\xi})\,\mathrm{tr}\left(\frac{1}{D}\left[\boldsymbol{I} + \left(1 - \frac{s}{t}\right)t^2\nabla^2\log p_t(\boldsymbol{\xi})\right]\right)^D\mathrm{d}\boldsymbol{\xi} \tag{B.2.34}$$

$$= \int_{\mathbb{R}^D} p_t(\boldsymbol{\xi})\left(1 + \frac{\left(1 - \frac{s}{t}\right)t^2}{D}\mathrm{tr}(\nabla^2\log p_t(\boldsymbol{\xi}))\right)^D\mathrm{d}\boldsymbol{\xi} \tag{B.2.35}$$

$$= \int_{\mathbb{R}^D} p_t(\boldsymbol{\xi})\left(1 + \frac{\left(1 - \frac{s}{t}\right)t^2}{D}\Delta\log p_t(\boldsymbol{\xi})\right)^D\mathrm{d}\boldsymbol{\xi}. \tag{B.2.36}$$

Since $\bar{\boldsymbol{x}}'(\boldsymbol{\xi})$ is positive definite (Corollary B.1), we have $1 + \frac{(1-s/t)t^2}{D}\Delta\log p_t(\boldsymbol{\xi}) =$ $\frac{\operatorname{tr}(\bar{\boldsymbol{x}}'(\boldsymbol{\xi}))}{D} > 0$. By the inequality $\log(1+x) \le x$ for $x > -1$,

$$\left(1 + \frac{\left(1-\frac{s}{t}\right)t^2}{D}\Delta \log p_t(\boldsymbol{\xi})\right)^D \le \exp\left(\left(1-\frac{s}{t}\right)t^2 \Delta \log p_t(\boldsymbol{\xi})\right). \tag{B.2.37}$$

From Lemma B.5 (where, recall, $R$ is the radius of the support of $\boldsymbol{x}$ as in Assumption B.1),

$$|\Delta \log p_t(\boldsymbol{\xi})| \le \max\left(\frac{D}{t^2}, \left|\frac{R^2}{t^4} - \frac{D}{t^2}\right|\right) =: U_t. \tag{B.2.38}$$

Setting $\varepsilon \doteq \left(1-\frac{s}{t}\right)t^2 U_t$, we have $\left(1-\frac{s}{t}\right)t^2 \Delta \log p_t(\boldsymbol{\xi}) \le \varepsilon$ for all $\boldsymbol{\xi}$. By Taylor's theorem with Lagrange remainder, for any $c \le \varepsilon$,

$$e^c \le 1 + c + \frac{c^2}{2}e^{\varepsilon}, \tag{B.2.39}$$

since $e^c = 1 + c + \frac{c^2}{2}e^{\theta c}$ for some $\theta \in (0,1)$, and $e^{\theta c} \le e^{\max(c,0)} \le e^{\varepsilon}$. Applying these bounds and integrating,

$$\int_{\mathbb{R}^D} p_t(\boldsymbol{\xi}) \left(1 + \frac{\left(1-\frac{s}{t}\right)t^2}{D}\Delta \log p_t(\boldsymbol{\xi})\right)^D \mathrm{d}\boldsymbol{\xi} \tag{B.2.40}$$

$$\le \int_{\mathbb{R}^D} p_t(\boldsymbol{\xi}) \exp\left(\left(1-\frac{s}{t}\right)t^2 \Delta \log p_t(\boldsymbol{\xi})\right) \mathrm{d}\boldsymbol{\xi} \tag{B.2.41}$$

$$\le 1 + \left(1-\frac{s}{t}\right)t^2 \int_{\mathbb{R}^D} p_t(\boldsymbol{\xi})\Delta \log p_t(\boldsymbol{\xi})\mathrm{d}\boldsymbol{\xi} + \frac{e^{\varepsilon}}{2}\left(1-\frac{s}{t}\right)^2 t^4 \int_{\mathbb{R}^D} p_t(\boldsymbol{\xi})\left(\Delta \log p_t(\boldsymbol{\xi})\right)^2 \mathrm{d}\boldsymbol{\xi} \tag{B.2.42}$$

$$= 1 - \left(1-\frac{s}{t}\right)t^2 \int_{\mathbb{R}^D} \frac{\|\nabla p_t(\boldsymbol{\xi})\|_2^2}{p_t(\boldsymbol{\xi})}\mathrm{d}\boldsymbol{\xi} + \frac{e^{\varepsilon}}{2}\left(1-\frac{s}{t}\right)^2 t^4 \int_{\mathbb{R}^D} p_t(\boldsymbol{\xi})\left(\Delta \log p_t(\boldsymbol{\xi})\right)^2 \mathrm{d}\boldsymbol{\xi}, \tag{B.2.43}$$

where the identity $\int p_t \Delta \log p_t = -\int \|\nabla p_t\|_2^2 / p_t$ is the same as in the proof of Theorem B.2. Writing $J(p_t) \doteq \int_{\mathbb{R}^D} \frac{\|\nabla p_t(\boldsymbol{\xi})\|_2^2}{p_t(\boldsymbol{\xi})}\mathrm{d}\boldsymbol{\xi}$ for the Fisher information of $p_t$, and bounding $\left(\Delta \log p_t\right)^2 \le U_t^2$, we combine with our previous estimate to obtain

$$h(\bar{\boldsymbol{x}}(\boldsymbol{x}_t)) - h(\boldsymbol{x}_t) \le -\left(1-\frac{s}{t}\right)t^2 J(p_t) + \frac{e^{\varepsilon}\varepsilon^2}{2}. \tag{B.2.44}$$

This is strictly negative whenever $\varepsilon^2 e^{\varepsilon} < 2\left(1-\frac{s}{t}\right)t^2 J(p_t)$. Substituting $\varepsilon = \left(1-\frac{s}{t}\right)t^2 U_t$, this is precisely condition (B.2.20). □

Notice that condition (B.2.20) is always satisfiable: for any fixed $t > 0$, the Fisher information satisfies $J(p_t) > 0$ (since $p_t$ is not constant) and $U_t < \infty$, so taking $\left(1-\frac{s}{t}\right)t^2$ small enough ensures (B.2.20) holds.

To understand the quantitative implications, consider the behavior as $t \to 0$ (near the data), where the condition is most restrictive. Setting $s = (1-\varepsilon)t$, the condition becomes $\varepsilon t^2 U_t^2 \exp(\varepsilon t^2 U_t) < 2J(p_t)$. For small $t$, the Laplacian bound satisfies $U_t \approx R^2/t^4$, so the polynomial prefactor scales as $\varepsilon t^2 U_t^2 \approx \varepsilon R^4/t^6$. The critical scaling is therefore $\varepsilon \sim t^6/R^4$. Taking $\varepsilon = \alpha t^6/R^4$, the condition reduces to $\alpha < 2J(p_t)$. In other words, each denoising step of size $t - s = \varepsilon t \sim t^7/R^4$ reduces the entropy. The steps must become increasingly fine as $t \to 0$, a consequence of the second-order remainder in the Taylor expansion used in the proof.

At the same time, it should be noted that our bounds are likely loose in terms of the precise $t$-dependence, and certainly so for data with greater structure. For instance, the $U_t \sim R^2/t^4$ blowup reflects a worst-case Laplacian bound that can be considerably tightened under additional regularity assumptions on $p$. Moreover, if $p_t$ is log-concave (as holds, e.g., when $p$ itself is log-concave), then $\Delta \log p_t \leq 0$ everywhere, and a significantly simplified argument can be run with strengthened conclusions.

### B.2.3 Technical Lemmas and Intermediate Results

In this subsection we present technical results which power our main two conceptual theorems. Our presentation will be more or less standard for mathematics; we will start with the higher-level results first, and gradually move back to the more incremental results. The higher-level results will use the incremental results, and in this way we have an easy-to-read dependency ordering of the results: no result depends on those before it. Results which do not depend on each other are generally ordered by the place they appear in the above pair of proofs.

#### Finiteness of the Differential Entropy

We first show that the entropy exists along the stochastic process and is finite.

**Lemma B.1.** *Let $\boldsymbol{x}$ be any random variable, and let $(\boldsymbol{x}_t)_{t\in[0,T]}$ be the stochastic process (B.2.1).*

1. *For $t > 0$, the differential entropy $h(\boldsymbol{x}_t)$ exists and is $> -\infty$.*
2. *If in addition Assumption B.1 holds for $\boldsymbol{x}$, then $h(\boldsymbol{x}) < \infty$ and $h(\boldsymbol{x}_t) < \infty$.*

*Proof.* To prove Lemma B.1.1, we use a classic yet tedious analysis argument. Since $\boldsymbol{x}_t$ has a density, we can write

$$h(\boldsymbol{x}_t) = -\int_{\mathbb{R}^D} p_t(\boldsymbol{\xi}) \log p_t(\boldsymbol{\xi}) \mathrm{d}\boldsymbol{\xi}. \tag{B.2.45}$$

Accordingly, let $g \colon \mathbb{R}^D \to \mathbb{R}$ be defined as

$$g(\boldsymbol{\xi}) \doteq -p_t(\boldsymbol{\xi}) \log p_t(\boldsymbol{\xi}) \implies h(\boldsymbol{x}_t) = \int_{\mathbb{R}^D} g(\boldsymbol{\xi}) \mathrm{d}\boldsymbol{\xi}. \tag{B.2.46}$$

As usual to bound the value of an integral in analysis, define

$$g_{+}(\boldsymbol{\xi}) \doteq \max(g(\boldsymbol{\xi}), 0), \quad g_{-}(\boldsymbol{\xi}) \doteq \max(-g(\boldsymbol{\xi}), 0) \tag{B.2.47}$$

$$\implies \quad g = g_{+} - g_{-} \quad \text{and} \quad g_{+}, g_{-} \geq 0. \tag{B.2.48}$$

Then

$$h(\boldsymbol{x}_t) = \int_{\mathbb{R}^D} g_{+}(\boldsymbol{\xi})\mathrm{d}\boldsymbol{\xi} - \int_{\mathbb{R}^D} g_{-}(\boldsymbol{\xi})\mathrm{d}\boldsymbol{\xi}, \tag{B.2.49}$$

and both integrals are guaranteed to be non-negative since their integrands are.

In order to show that $h(\boldsymbol{x}_t)$ is well-defined, we need to show that $\int_{\mathbb{R}^D} g_{+}(\boldsymbol{\xi})\mathrm{d}\boldsymbol{\xi} < \infty$ or $\int_{\mathbb{R}^D} g_{-}(\boldsymbol{\xi})\mathrm{d}\boldsymbol{\xi} < \infty$. To show that $h(\boldsymbol{x}_t) > -\infty$, it merely suffices to show that $\int_{\mathbb{R}^D} g_{-}(\boldsymbol{\xi})\mathrm{d}\boldsymbol{\xi} < \infty$. To bound the integral of $g_{-}$ we need to understand the quantity $g_{-}$, namely, we want to characterize when $g$ is negative.

$$g(\boldsymbol{\xi}) \leq 0 \iff p_t(\boldsymbol{\xi}) \log p_t(\boldsymbol{\xi}) \geq 0 \iff \log p_t(\boldsymbol{\xi}) \geq 0 \iff p_t(\boldsymbol{\xi}) \geq 1. \tag{B.2.50}$$

Thus, it holds that

$$g_{-}(\boldsymbol{\xi}) = \mathbf{1}(p_t(\boldsymbol{\xi}) \geq 1) \cdot (-g(\boldsymbol{\xi})) = \mathbf{1}(p_t(\boldsymbol{\xi}) \geq 1) p_t(\boldsymbol{\xi}) \log p_t(\boldsymbol{\xi}). \tag{B.2.51}$$

In order to bound the integral of $g_{-}(\boldsymbol{\xi})$, we need to show that $p_t$ is "not too concentrated," namely that $p_t$ is not too large. To prove this, in this case we are lucky enough to be able to bound the function $g_{-}(\boldsymbol{\xi})$ itself. Namely, notice that

$$\max_{\boldsymbol{\xi} \in \mathbb{R}^D} \varphi_t(\boldsymbol{\xi} - \boldsymbol{x}) = \varphi_t(\mathbf{0}) = \frac{1}{(2\pi)^{D/2} t^D} =: C_t. \tag{B.2.52}$$

which blows up as $t \to 0$ but is finite for all finite $t$. Therefore

$$p_t(\boldsymbol{\xi}) = \mathbb{E}\, \varphi_t(\boldsymbol{\xi} - \boldsymbol{x}) \leq \mathbb{E}\, C_t = C_t. \tag{B.2.53}$$

Now there are two cases.

- If $C_t < 1$, then $p_t(\boldsymbol{\xi}) < 1$, so the indicator is never 1, hence $g_{-} = 0$ identically and its integral is also 0.
- If $C_t \geq 1$, then $\log C_t \geq 0$, so since the logarithm is monotonically increasing,

$$\begin{aligned}
\int_{\mathbb{R}^D} g_{-}(\boldsymbol{\xi})\mathrm{d}\boldsymbol{\xi} &= \int_{\mathbb{R}^D} \mathbf{1}(p_t(\boldsymbol{\xi}) \geq 1) p_t(\boldsymbol{\xi}) \log p_t(\boldsymbol{\xi})\mathrm{d}\boldsymbol{\xi} && \text{(B.2.54)} \\
&= \mathbb{E}[\mathbf{1}(p_t(\boldsymbol{x}_t) \geq 1) \log p_t(\boldsymbol{x}_t)] && \text{(B.2.55)} \\
&\leq \mathbb{E}[\mathbf{1}(p_t(\boldsymbol{x}_t) \geq 1) \log C_t] && \text{(B.2.56)} \\
&= \mathbb{P}[p_t(\boldsymbol{x}_t) \geq 1] \log C_t. && \text{(B.2.57)}
\end{aligned}$$

Hence we have $\int_{\mathbb{R}^D} g_{-}(\boldsymbol{\xi})\mathrm{d}\boldsymbol{\xi} < \infty$, so the differential entropy $h(\boldsymbol{x}_t)$ exists and is $> -\infty$.

To prove Lemma B.1.2, suppose that Assumption B.1 holds. We want to show that $h(\boldsymbol{x}) < \infty$ and $h(\boldsymbol{x}_t) < \infty$. The mechanism for doing this is the same, and involves the maximum entropy result Theorem B.1. Namely, since $\boldsymbol{x}$ is absolutely bounded, it has a finite covariance which we will denote $\boldsymbol{\Sigma}$. Then the covariance of $\boldsymbol{x}_t$ is $\boldsymbol{\Sigma} + t^2\boldsymbol{I}$. Thus the entropy of $\boldsymbol{x}$ and $\boldsymbol{x}_t$ can be upper bounded by the entropy of normal distributions with the respective covariances, i.e., $\log[(2\pi e)^D \det(\boldsymbol{\Sigma})]$ or $\log[(2\pi e)^D \det(\boldsymbol{\Sigma} + t^2\boldsymbol{I})]$, and both are $< \infty$. □

### Integration by Parts in De Bruijn Identity

Finally, we fill in the integration-by-parts argument alluded to in the proofs of Theorems B.2 and B.3. The argument is conceptually pretty simple but requires some technical estimates to show that the boundary term in the integration-by-parts vanishes.

**Lemma B.2.** *Let $\boldsymbol{x}$ be any random variable such that Assumptions B.1 and B.2 hold, and let $(\boldsymbol{x}_t)_{t\in[0,T]}$ be the stochastic process (B.2.1). For $t \geq 0$, let $p_t$ be the density of $\boldsymbol{x}_t$. Then for a constant $c \in \mathbb{R}$ it holds*

$$\int_{\mathbb{R}^D} \Delta p_t(\boldsymbol{\xi})[c + \log p_t(\boldsymbol{\xi})]\mathrm{d}\boldsymbol{\xi} = -\int_{\mathbb{R}^D} \langle \nabla \log p_t(\boldsymbol{\xi}), \nabla p_t(\boldsymbol{\xi})\rangle \mathrm{d}\boldsymbol{\xi}. \tag{B.2.58}$$

*Proof.* The basic idea of this proof is in two steps:

- First, apply Green's theorem to do integration by parts over a compact set;
- Second, send the radius of this compact set to $+\infty$, to get integrals over all of $\mathbb{R}^D$.

Green's theorem says that for any compact set $\mathcal{K} \subseteq \mathbb{R}^D$, twice continuously differentiable $\phi\colon \mathbb{R}^D \to \mathbb{R}$, and continuously differentiable $\psi\colon \mathbb{R}^D \to \mathbb{R}$,

$$\int_{\mathcal{K}} \{\psi(\boldsymbol{\xi})\Delta\phi(\boldsymbol{\xi}) + \langle \nabla\psi(\boldsymbol{\xi}), \nabla\phi(\boldsymbol{\xi})\rangle\} \,\mathrm{d}\boldsymbol{\xi} = \int_{\partial\mathcal{K}} \psi(\boldsymbol{\xi})\langle \nabla\phi(\boldsymbol{\xi}), \boldsymbol{n}(\boldsymbol{\xi})\rangle \mathrm{d}\sigma(\boldsymbol{\xi}) \tag{B.2.59}$$

where $\mathrm{d}\sigma(\boldsymbol{\xi})$ denotes an integral over the "surface measure", i.e., the inherited measure on $\partial\mathcal{K}$, namely the boundary of $\mathcal{K}$, and accordingly $\boldsymbol{\xi}$ takes values on this surface and $\boldsymbol{n}(\boldsymbol{\xi})$ is the unit normal vector to $\mathcal{K}$ at the surface point $\boldsymbol{\xi}$. Now, taking $\phi(\boldsymbol{\xi}) \doteq p_t(\boldsymbol{\xi})$ and $\psi(\boldsymbol{\xi}) \doteq c + \log p_t(\boldsymbol{\xi})$, over a ball $B_r(\mathbf{0})$ of radius $r > 0$ centered at $\mathbf{0}$ (so that $\partial B_r(\mathbf{0})$ is the sphere of radius $r$ centered at $\mathbf{0}$ and $\boldsymbol{n}(\boldsymbol{\xi}) = \boldsymbol{\xi}/\|\boldsymbol{\xi}\|_2 = \boldsymbol{\xi}/r$):

$$\int_{B_r(\mathbf{0})} \{\Delta p_t(\boldsymbol{\xi})[c + \log p_t(\boldsymbol{\xi})] + \langle \nabla \log p_t(\boldsymbol{\xi}), \nabla p_t(\boldsymbol{\xi})\rangle\} \,\mathrm{d}\boldsymbol{\xi} \tag{B.2.60}$$

$$= \int_{\partial B_r(\mathbf{0})} [c + \log p_t(\boldsymbol{\xi})] \left\langle \nabla p_t(\boldsymbol{\xi}), \frac{\boldsymbol{\xi}}{r} \right\rangle \mathrm{d}\sigma(\boldsymbol{\xi}) \tag{B.2.61}$$

$$= \frac{1}{r}\int_{\partial B_r(\mathbf{0})} [c + \log p_t(\boldsymbol{\xi})] \left\langle \nabla p_t(\boldsymbol{\xi}), \boldsymbol{\xi} \right\rangle \mathrm{d}\sigma(\boldsymbol{\xi}). \tag{B.2.62}$$

Sending $r \to \infty$, it holds that

$$\int_{\mathbb{R}^D} \{\Delta p_t(\boldsymbol{\xi})[c + \log p_t(\boldsymbol{\xi})] + \langle \nabla \log p_t(\boldsymbol{\xi}), \nabla p_t(\boldsymbol{\xi})\rangle\} \, \mathrm{d}\boldsymbol{\xi} \tag{B.2.63}$$

$$= \lim_{r\to\infty} \int_{B_r(\mathbf{0})} \{\Delta p_t(\boldsymbol{\xi})[c + \log p_t(\boldsymbol{\xi})] + \langle \nabla \log p_t(\boldsymbol{\xi}), \nabla p_t(\boldsymbol{\xi})\rangle\} \, \mathrm{d}\boldsymbol{\xi} \tag{B.2.64}$$

$$= \lim_{r\to\infty} \frac{1}{r} \int_{\partial B_r(\mathbf{0})} [c + \log p_t(\boldsymbol{\xi})] \, \langle \nabla p_t(\boldsymbol{\xi}), \boldsymbol{\xi}\rangle \, \mathrm{d}\sigma(\boldsymbol{\xi}), \tag{B.2.65}$$

where the first equality follows by dominated convergence on the integrand. It remains to compute the last limit. For this, we take asymptotic expansions of each term. The main device is as follows: for $\boldsymbol{\xi} \in \partial B_r(\mathbf{0})$, we have $\|\boldsymbol{\xi}\|_2 = r$, so

$$p_t(\boldsymbol{\xi}) = \mathbb{E}[\varphi_t(\boldsymbol{\xi} - \boldsymbol{x})] \tag{B.2.66}$$

$$= \mathbb{E}\left[\underbrace{\frac{1}{(2\pi)^{D/2} t^D}}_{\doteq C_t} e^{-\|\boldsymbol{\xi} - \boldsymbol{x}\|_2^2/(2t^2)}\right] \tag{B.2.67}$$

$$= C_t \, \mathbb{E}\left[e^{-(\|\boldsymbol{\xi}\|_2^2 - 2\langle \boldsymbol{\xi}, \boldsymbol{x}\rangle + \|\boldsymbol{x}\|_2^2)/(2t^2)}\right] \tag{B.2.68}$$

$$= C_t \, \mathbb{E}\left[e^{-(r^2 - 2\langle \boldsymbol{\xi}, \boldsymbol{x}\rangle + \|\boldsymbol{x}\|_2^2)/(2t^2)}\right] \tag{B.2.69}$$

$$= C_t e^{-r^2/(2t^2)} \, \mathbb{E}\big[e^{(2\langle \boldsymbol{\xi}, \boldsymbol{x}\rangle - \|\boldsymbol{x}\|_2^2)/(2t^2)}\big]. \tag{B.2.70}$$

Note that because $\|\boldsymbol{\xi}\|_2 = r$, we have by Cauchy-Schwarz that

$$-2r\|\boldsymbol{x}\|_2 - \|\boldsymbol{x}\|_2^2 \le 2\langle \boldsymbol{\xi}, \boldsymbol{x}\rangle - \|\boldsymbol{x}\|_2^2 \le 2r\|\boldsymbol{x}\|_2 - \|\boldsymbol{x}\|_2^2. \tag{B.2.71}$$

Recall that by Assumption B.1, $\boldsymbol{x}$ is supported on a compact set $\mathcal{S}$ of radius $R$. Thus

$$-2R(r + R) \le 2\langle \boldsymbol{\xi}, \boldsymbol{x}\rangle - \|\boldsymbol{x}\|_2^2 \le 2Rr. \tag{B.2.72}$$

In other words, it holds

$$C_t e^{-[r^2 + 2R(r+R)]/(2t^2)} \le p_t(\boldsymbol{\xi}) \le C_t e^{[-r^2 + 2Rr]/(2t^2)}. \tag{B.2.73}$$

Now to compute the gradient, we can use Proposition B.1 and linearity of expectation to compute

$$\langle \nabla p_t(\boldsymbol{\xi}), \boldsymbol{\xi}\rangle = \left\langle -\frac{1}{t^2} \, \mathbb{E}[(\boldsymbol{\xi} - \boldsymbol{x}) \, \varphi_t(\boldsymbol{\xi} - \boldsymbol{x})], \boldsymbol{\xi}\right\rangle \tag{B.2.74}$$

$$= -\frac{1}{t^2} \, \mathbb{E}[\langle \boldsymbol{\xi} - \boldsymbol{x}, \boldsymbol{\xi}\rangle \varphi_t(\boldsymbol{\xi} - \boldsymbol{x})] \tag{B.2.75}$$

$$= -\frac{1}{t^2} \, \mathbb{E}\big[\big(\|\boldsymbol{\xi}\|_2^2 - \langle \boldsymbol{\xi}, \boldsymbol{x}\rangle\big) \, \varphi_t(\boldsymbol{\xi} - \boldsymbol{x})\big] \tag{B.2.76}$$

$$= -\frac{1}{t^2} \, \mathbb{E}\big[\big(r^2 - \langle \boldsymbol{\xi}, \boldsymbol{x}\rangle\big) \, \varphi_t(\boldsymbol{\xi} - \boldsymbol{x})\big] \tag{B.2.77}$$

$$= \frac{1}{t^2}\,\mathbb{E}\big[\big(\langle \boldsymbol{\xi}, \boldsymbol{x}\rangle - r^2\big)\,\varphi_t(\boldsymbol{\xi} - \boldsymbol{x})\big]. \tag{B.2.78}$$

Using Cauchy-Schwarz and the representation $p_t(\boldsymbol{\xi}) \doteq \mathbb{E}[\varphi_t(\boldsymbol{\xi} - \boldsymbol{x})]$ again, it holds

$$\frac{1}{t^2}\,\mathbb{E}\big[\big(-Rr - r^2\big)\,\varphi_t(\boldsymbol{\xi} - \boldsymbol{x})\big] \le \langle \nabla p_t(\boldsymbol{\xi}), \boldsymbol{\xi}\rangle \le \frac{1}{t^2}\,\mathbb{E}\big[\big(Rr - r^2\big)\,\varphi_t(\boldsymbol{\xi} - \boldsymbol{x})\big] \tag{B.2.79}$$

$$\frac{1}{t^2}\,\big(-Rr - r^2\big)\,\mathbb{E}[\varphi_t(\boldsymbol{\xi} - \boldsymbol{x})] \le \langle \nabla p_t(\boldsymbol{\xi}), \boldsymbol{\xi}\rangle \le \frac{1}{t^2}\,\big(Rr - r^2\big)\,\mathbb{E}[\varphi_t(\boldsymbol{\xi} - \boldsymbol{x})] \tag{B.2.80}$$

$$-\frac{r(R+r)}{t^2}p_t(\boldsymbol{\xi}) \le \langle \nabla p_t(\boldsymbol{\xi}), \boldsymbol{\xi}\rangle \le -\frac{r(r-R)}{t^2}p_t(\boldsymbol{\xi}). \tag{B.2.81}$$

For $r > R > 0$ (as is suitable, because we are going to take the limit $r \to \infty$ while $R$ is fixed), both sides are negative. This makes sense: most of the probability mass is contained within the ball of radius $R$ and thus the score points inwards, having a negative inner product with the outward-pointing vector $\boldsymbol{\xi}$. Thus using the appropriate bounds for $p_t(\boldsymbol{\xi})$,

$$-\frac{r(R+r)}{t^2}\cdot C_t e^{[-r^2+2Rr]/(2t^2)} \le \langle \nabla p_t(\boldsymbol{\xi}), \boldsymbol{\xi}\rangle \le -\frac{r(r-R)}{t^2}\cdot C_t e^{-[r^2+2R(r+R)]/(2t^2)}. \tag{B.2.82}$$

Then, noting that $C_t = \operatorname{poly}(t^{-1})$, we can compute

$$[c + \log p_t(\boldsymbol{\xi})]\langle \nabla p_t(\boldsymbol{\xi}), \boldsymbol{\xi}\rangle = \operatorname{poly}(r, R, t^{-1}, c)e^{-\Theta_r(r^2)} \tag{B.2.83}$$

So one can see that, letting the surface area of $\partial B_r(\mathbf{0})$ be $\omega_D r^{D-1}$ where $\omega_D$ is a function of $D$, it holds

$$\frac{1}{r}\int_{\partial B_r(\mathbf{0})} [c + \log p_t(\boldsymbol{\xi})]\langle \nabla p_t(\boldsymbol{\xi}), \boldsymbol{\xi}\rangle \mathrm{d}\boldsymbol{\xi} = \operatorname{poly}(r, R, t^{-1}, c)e^{-\Theta_r(r^2)} \tag{B.2.84}$$

and therefore the exponentially decaying tails mean

$$\lim_{r\to\infty}\frac{1}{r}\int_{\partial B_r(\mathbf{0})} [c + \log p_t(\boldsymbol{\xi})]\langle \nabla p_t(\boldsymbol{\xi}), \boldsymbol{\xi}\rangle \mathrm{d}\boldsymbol{\xi} = 0. \tag{B.2.85}$$

Finally, plugging into (B.2.63), we have

$$\int_{\mathbb{R}^D} \{\Delta p_t(\boldsymbol{\xi})[c + \log p_t(\boldsymbol{\xi})] + \langle \nabla \log p_t(\boldsymbol{\xi}), \nabla p_t(\boldsymbol{\xi})\rangle\}\,\mathrm{d}\boldsymbol{\xi} = 0 \tag{B.2.86}$$

$$\implies \int_{\mathbb{R}^D} \Delta p_t(\boldsymbol{\xi})[c + \log p_t(\boldsymbol{\xi})]\mathrm{d}\boldsymbol{\xi} = -\int_{\mathbb{R}^D}\langle \nabla \log p_t(\boldsymbol{\xi}), \nabla p_t(\boldsymbol{\xi})\rangle \mathrm{d}\boldsymbol{\xi} \tag{B.2.87}$$

as claimed. □

**Local Invertibility of the Denoiser $\bar{\boldsymbol{x}}$**

Here we provide some results used in the proof of Theorem B.3 which are appropriate generalizations of corresponding results in [Gri11].

**Lemma B.3** (Generalization of [Gri11], Lemma A.1)**.** *Let $\boldsymbol{x}$ be any random variable such that Assumptions B.1 and B.2 hold, and let $(\boldsymbol{x}_t)_{t\in[0,T]}$ be the stochastic process (B.2.1). Let $s, t \in [0, T]$ be such that $0 \le s < t \le T$, and let $\bar{\boldsymbol{x}}(\boldsymbol{\xi}) \doteq \mathbb{E}[\boldsymbol{x}_s \mid \boldsymbol{x}_t = \boldsymbol{\xi}]$. The Jacobian $\bar{\boldsymbol{x}}'(\boldsymbol{\xi})$ is symmetric positive definite.*

*Proof.* We have

$$\bar{\boldsymbol{x}}'(\boldsymbol{\xi}) = \boldsymbol{I} + \left(1 - \frac{s}{t}\right) t^2 \nabla^2 \log p_t(\boldsymbol{\xi}). \tag{B.2.88}$$

Here we expand

$$\nabla^2 \log p_t(\boldsymbol{\xi}) = \frac{p_t(\boldsymbol{\xi})\nabla^2 p_t(\boldsymbol{\xi}) - (\nabla p_t(\boldsymbol{\xi}))(\nabla p_t(\boldsymbol{\xi}))^\top}{p_t(\boldsymbol{\xi})^2}. \tag{B.2.89}$$

So we need to ensure that

$$\bar{\boldsymbol{x}}'(\boldsymbol{\xi}) = \boldsymbol{I} + \left(1 - \frac{s}{t}\right) t^2 \frac{p_t(\boldsymbol{\xi})\nabla^2 p_t(\boldsymbol{\xi}) - (\nabla p_t(\boldsymbol{\xi}))(\nabla p_t(\boldsymbol{\xi}))^\top}{p_t(\boldsymbol{\xi})^2} \tag{B.2.90}$$

$$= \frac{p_t(\boldsymbol{\xi})^2\boldsymbol{I} + \left(1 - \frac{s}{t}\right) t^2 \left[p_t(\boldsymbol{\xi})\nabla^2 p_t(\boldsymbol{\xi}) - (\nabla p_t(\boldsymbol{\xi}))(\nabla p_t(\boldsymbol{\xi}))^\top\right]}{p_t(\boldsymbol{\xi})^2} \tag{B.2.91}$$

is symmetric positive semidefinite. Indeed it is obviously symmetric (by Clairaut's theorem). To show its positive semidefiniteness, we plug in the expectation representation of $p_t$ given by (B.2.3) (and $\nabla p_t$, $\Delta p_t$ by Proposition B.1) to obtain (where $\boldsymbol{x}$ is as defined and $\boldsymbol{y}$ is i.i.d. as $\boldsymbol{x}$),

$$\boldsymbol{v}^\top[\bar{\boldsymbol{x}}'(\boldsymbol{\xi})]\boldsymbol{v} \tag{B.2.92}$$

$$= p_t(\boldsymbol{\xi})^{-2}\boldsymbol{v}^\top\Bigg\{p_t(\boldsymbol{\xi})^2\boldsymbol{I} + \left(1 - \frac{s}{t}\right) t^2\, \mathbb{E}[\varphi_t(\boldsymbol{\xi} - \boldsymbol{x})]\, \mathbb{E}\left[\varphi_t(\boldsymbol{\xi} - \boldsymbol{x}) \cdot \frac{(\boldsymbol{\xi} - \boldsymbol{x})(\boldsymbol{\xi} - \boldsymbol{x})^\top - t^2\boldsymbol{I}}{t^4}\right] \tag{B.2.93}$$

$$- \left(1 - \frac{s}{t}\right) t^2\, \mathbb{E}\left[-\varphi_t(\boldsymbol{\xi} - \boldsymbol{x}) \cdot \frac{\boldsymbol{\xi} - \boldsymbol{x}}{t^2}\right] \mathbb{E}\left[-\varphi_t(\boldsymbol{\xi} - \boldsymbol{x}) \cdot \frac{\boldsymbol{\xi} - \boldsymbol{x}}{t^2}\right]^\top\Bigg\}\boldsymbol{v} \tag{B.2.94}$$

$$= p_t(\boldsymbol{\xi})^{-2}\boldsymbol{v}^\top\Bigg\{\mathbb{E}[\varphi_t(\boldsymbol{\xi} - \boldsymbol{x})\varphi_t(\boldsymbol{\xi} - \boldsymbol{y})\boldsymbol{I}] \tag{B.2.95}$$

$$+ \left(1 - \frac{s}{t}\right) t^2\, \mathbb{E}\left[\varphi_t(\boldsymbol{\xi} - \boldsymbol{x})\varphi_t(\boldsymbol{\xi} - \boldsymbol{y}) \cdot \frac{(\boldsymbol{\xi} - \boldsymbol{y})(\boldsymbol{\xi} - \boldsymbol{y})^\top - t^2\boldsymbol{I}}{t^4}\right] \tag{B.2.96}$$

$$- \left(1 - \frac{s}{t}\right) t^2\, \mathbb{E}\left[\varphi_t(\boldsymbol{\xi} - \boldsymbol{x})\varphi_t(\boldsymbol{\xi} - \boldsymbol{y}) \cdot \frac{(\boldsymbol{\xi} - \boldsymbol{x})(\boldsymbol{\xi} - \boldsymbol{y})^\top}{t^4}\right]\Bigg\} \tag{B.2.97}$$

$$= \frac{1 - s/t}{p_t(\boldsymbol{\xi})^2} \boldsymbol{v}^\top \mathbb{E}\left[\varphi_t(\boldsymbol{\xi} - \boldsymbol{x})\varphi_t(\boldsymbol{\xi} - \boldsymbol{y}) \left\{ \frac{1}{1 - s/t}\boldsymbol{I} + \frac{(\boldsymbol{\xi} - \boldsymbol{y})(\boldsymbol{\xi} - \boldsymbol{y})^\top}{t^2} - \boldsymbol{I} - \frac{(\boldsymbol{\xi} - \boldsymbol{x})(\boldsymbol{\xi} - \boldsymbol{y})^\top}{t^2} \right\}\right]\boldsymbol{v} \tag{B.2.98}$$

$$= \frac{t - s}{tp_t(\boldsymbol{\xi})^2} \boldsymbol{v}^\top \mathbb{E}\left[\frac{s}{t - s}\boldsymbol{I} + \frac{(\boldsymbol{\xi} - \boldsymbol{x})(\boldsymbol{\xi} - \boldsymbol{x})^\top}{2t^2} + \frac{(\boldsymbol{\xi} - \boldsymbol{y})(\boldsymbol{\xi} - \boldsymbol{y})^\top}{2t^2} - \frac{(\boldsymbol{\xi} - \boldsymbol{x})(\boldsymbol{\xi} - \boldsymbol{y})^\top}{t^2}\right]\boldsymbol{v} \tag{B.2.99}$$

$$= \frac{t - s}{tp_t(\boldsymbol{\xi})^2} \boldsymbol{v}^\top \mathbb{E}\left[\frac{s}{t - s}\boldsymbol{I} + \frac{1}{2t^2}\left((\boldsymbol{\xi} - \boldsymbol{x})(\boldsymbol{\xi} - \boldsymbol{x})^\top + (\boldsymbol{\xi} - \boldsymbol{y})(\boldsymbol{\xi} - \boldsymbol{y})^\top - 2(\boldsymbol{\xi} - \boldsymbol{x})(\boldsymbol{\xi} - \boldsymbol{y})^\top\right)\right]\boldsymbol{v} \tag{B.2.100}$$

$$= \frac{t - s}{tp_t(\boldsymbol{\xi})^2} \mathbb{E}\left[\frac{s}{t - s}\|\boldsymbol{v}\|_2^2 + \frac{1}{2t^2}\left([(\boldsymbol{\xi} - \boldsymbol{x})^\top \boldsymbol{v}]^2 + [(\boldsymbol{\xi} - \boldsymbol{y})^\top \boldsymbol{v}]^2 - 2[(\boldsymbol{\xi} - \boldsymbol{x})^\top \boldsymbol{v}][(\boldsymbol{\xi} - \boldsymbol{y})^\top \boldsymbol{v}]\right)\right] \tag{B.2.101}$$

$$= \frac{t - s}{tp_t(\boldsymbol{\xi})^2} \mathbb{E}\left[\frac{s}{t - s}\|\boldsymbol{v}\|_2^2 + \frac{1}{2t^2}\left([(\boldsymbol{\xi} - \boldsymbol{x})^\top \boldsymbol{v}]^2 + [(\boldsymbol{\xi} - \boldsymbol{y})^\top \boldsymbol{v}]^2 - 2[(\boldsymbol{\xi} - \boldsymbol{x})^\top \boldsymbol{v}][(\boldsymbol{\xi} - \boldsymbol{y})^\top \boldsymbol{v}]\right)\right] \tag{B.2.102}$$

$$= \frac{t - s}{tp_t(\boldsymbol{\xi})^2} \mathbb{E}\left[\frac{s}{t - s}\|\boldsymbol{v}\|_2^2 + \frac{1}{2t^2}\left([(\boldsymbol{\xi} - \boldsymbol{x})^\top \boldsymbol{v}] - [(\boldsymbol{\xi} - \boldsymbol{y})^\top \boldsymbol{v}]\right)^2\right] \tag{B.2.103}$$

$$= \frac{t - s}{tp_t(\boldsymbol{\xi})^2} \mathbb{E}\left[\frac{s}{t - s}\|\boldsymbol{v}\|_2^2 + \frac{1}{2t^2}[(\boldsymbol{y} - \boldsymbol{x})^\top \boldsymbol{v}]^2\right] \tag{B.2.104}$$

$$= \frac{s}{tp_t(\boldsymbol{\xi})^2}\|\boldsymbol{v}\|_2^2 + \frac{t - s}{2t^3 p_t(\boldsymbol{\xi})} \mathbb{E}[\{(\boldsymbol{y} - \boldsymbol{x})^\top \boldsymbol{v}\}^2] \tag{B.2.105}$$

Since $\boldsymbol{x}$ and $\boldsymbol{y}$ are i.i.d., the whole integral (i.e., the original quadratic form) is 0 if and only if $s = 0$ and $\boldsymbol{x}$ has support entirely contained in an affine subspace which is orthogonal to $\boldsymbol{v}$. But this is ruled out by assumption (i.e., that $\boldsymbol{x}$ has a density on $\mathbb{R}^D$), so the Jacobian $\bar{\boldsymbol{x}}'(\boldsymbol{\xi})$ is symmetric positive definite. ☐

**Lemma B.4** (Generalization of [Gri11] Corollary A.2, Part 1)**.** *Let $f\colon \mathbb{R}^D \to \mathbb{R}^D$ be any differentiable function whose Jacobian $f'(\boldsymbol{x})$ is symmetric positive definite. Then $f$ is injective, and hence invertible as a function $\mathbb{R}^D \to \mathcal{R}(f)$ where $\mathcal{R}(f)$ is the range of $f$.*

*Proof.* Suppose that $f$ were not injective, i.e., there exists $\boldsymbol{x}, \boldsymbol{x}'$ such that $f(\boldsymbol{x}) = f(\boldsymbol{x}')$ while $\boldsymbol{x} \neq \boldsymbol{x}'$. Define $\boldsymbol{v} \doteq (\boldsymbol{x}' - \boldsymbol{x})/\|\boldsymbol{x}' - \boldsymbol{x}\|_2$. Define the function $g\colon \mathbb{R} \to \mathbb{R}$ as $g(t) \doteq \boldsymbol{v}^\top f(\boldsymbol{x} + t\boldsymbol{v})$. Then $g(0) = \boldsymbol{v}^\top f(\boldsymbol{x}) = \boldsymbol{v}^\top f(\boldsymbol{x}') = g(\|\boldsymbol{x}' - \boldsymbol{x}\|_2)$. Since $f$ is differentiable, $g$ is differentiable, so the derivative $g'$ must vanish for some $t^* \in (0, \|\boldsymbol{x}' - \boldsymbol{x}\|_2)$ by the mean value theorem. However,

$$g'(t^*) \doteq \boldsymbol{v}^\top \left[f'(\boldsymbol{x} + t^*\boldsymbol{v})\right] \boldsymbol{v} > 0 \tag{B.2.106}$$

since the Jacobian is positive definite. Thus we arrive at a contradiction, as claimed. ☐

Combining the above two results, we obtain the following crucial result.

**Corollary B.1** (Generalization of [Gri11] Corollary A.2, Part 2)**.** *Let $\boldsymbol{x}$ be any random variable such that Assumptions B.1 and B.2 hold, and let $(\boldsymbol{x}_t)_{t\in[0,T]}$ be the stochastic process (B.2.1). Let $s, t \in [0, T]$ be such that $0 \leq s < t \leq T$, and let $\bar{\boldsymbol{x}}(\boldsymbol{\xi}) \doteq \mathbb{E}[\boldsymbol{x}_s \mid \boldsymbol{x}_t = \boldsymbol{\xi}]$. Then $\bar{\boldsymbol{x}}$ is injective, and therefore invertible onto its range.*

*Proof.* The only thing left to show is that $\bar{\boldsymbol{x}}$ is differentiable, but this is immediate from Tweedie's formula (Theorem 3.2) which shows that $\bar{\boldsymbol{x}}$ is differentiable if and only if $\nabla \log p_t$ is differentiable, and this is provided by Equation (B.2.3). □

### Controlling the Laplacian $\Delta \log p_t$

Finally, we develop a technical estimate which is required for the proof of Theorem B.3 and actually motivates the assumption for the viable $t$.

**Lemma B.5.** *Let $\boldsymbol{x}$ be any random variable such that Assumptions B.1 and B.2 hold, and let $(\boldsymbol{x}_t)_{t\in[0,T]}$ be the stochastic process (B.2.1). Let $p_t$ be the density of $\boldsymbol{x}_t$. Then, for $t > 0$ it holds*

$$\sup_{\boldsymbol{\xi}\in\mathbb{R}^D} |\Delta \log p_t(\boldsymbol{\xi})| \leq \max\left(\frac{D}{t^2}, \left|\frac{R^2}{t^4} - \frac{D}{t^2}\right|\right). \tag{B.2.107}$$

*Proof.* By chain rule, a simple exercise computes

$$\Delta \log p_t(\boldsymbol{\xi}) = \frac{\Delta p_t(\boldsymbol{\xi})}{p_t(\boldsymbol{\xi})} - \frac{\|\nabla p_t(\boldsymbol{\xi})\|_2^2}{p_t(\boldsymbol{\xi})^2}. \tag{B.2.108}$$

Using Proposition B.1 to write the terms in $\Delta p_t(\boldsymbol{\xi})$, we obtain

$$\frac{\Delta p_t(\boldsymbol{\xi})}{p_t(\boldsymbol{\xi})} = \frac{\mathbb{E}\left[\frac{\|\boldsymbol{\xi}-\boldsymbol{x}\|_2^2 - Dt^2}{t^4} \cdot \varphi_t(\boldsymbol{\xi} - \boldsymbol{x})\right]}{\mathbb{E}[\varphi_t(\boldsymbol{\xi} - \boldsymbol{x})]} \tag{B.2.109}$$

$$= \frac{\int_{\mathbb{R}^D} \left\{\frac{\|\boldsymbol{\xi}-\boldsymbol{u}\|_2^2 - Dt^2}{t^4}\right\} \varphi_t(\boldsymbol{\xi} - \boldsymbol{u}) p(\boldsymbol{u}) \mathrm{d}\boldsymbol{u}}{\int_{\mathbb{R}^D} \varphi_t(\boldsymbol{\xi} - \boldsymbol{u}) p(\boldsymbol{u}) \mathrm{d}\boldsymbol{u}}. \tag{B.2.110}$$

This looks like a Bayesian marginalization, so let us define the appropriate normalized density

$$q_{\boldsymbol{\xi}}(\boldsymbol{u}) = \frac{\varphi_t(\boldsymbol{\xi} - \boldsymbol{u}) p(\boldsymbol{u})}{\int_{\mathbb{R}^D} \varphi_t(\boldsymbol{\xi} - \boldsymbol{v}) p(\boldsymbol{v}) \mathrm{d}\boldsymbol{v}} = \frac{\varphi_t(\boldsymbol{\xi} - \boldsymbol{u}) p(\boldsymbol{u})}{\mathbb{E}[\varphi_t(\boldsymbol{\xi} - \boldsymbol{x})]} = \frac{\varphi_t(\boldsymbol{\xi} - \boldsymbol{u}) p(\boldsymbol{u})}{p_t(\boldsymbol{\xi})} \tag{B.2.111}$$

Then, defining $\boldsymbol{y}_{\boldsymbol{\xi}} \sim q_{\boldsymbol{\xi}}$, we can write

$$\frac{\Delta p_t(\boldsymbol{\xi})}{p_t(\boldsymbol{\xi})} = \int_{\mathbb{R}^D} \left\{\frac{\|\boldsymbol{\xi} - \boldsymbol{u}\|_2^2 - Dt^2}{t^4}\right\} q_{\boldsymbol{\xi}}(\boldsymbol{u}) \mathrm{d}\boldsymbol{u} = \frac{1}{t^4} \mathbb{E}[\|\boldsymbol{\xi} - \boldsymbol{y}_{\boldsymbol{\xi}}\|_2^2] - \frac{D}{t^2}. \tag{B.2.112}$$

Similarly, writing out the second term (non-squared) we obtain

$$\frac{\nabla p_t(\boldsymbol{\xi})}{p_t(\boldsymbol{\xi})} = -\frac{\boldsymbol{\xi} - \mathbb{E}[\boldsymbol{y}_{\boldsymbol{\xi}}]}{t^2}. \tag{B.2.113}$$

Letting $\boldsymbol{z}_{\boldsymbol{\xi}} \doteq \boldsymbol{y}_{\boldsymbol{\xi}} - \boldsymbol{\xi}$, it holds

$$\frac{\Delta p_t(\boldsymbol{\xi})}{p_t(\boldsymbol{\xi})} = \frac{\mathbb{E}[\|\boldsymbol{z}_{\boldsymbol{\xi}}\|_2^2]}{t^4} - \frac{D}{t^2}, \qquad \frac{\nabla p_t(\boldsymbol{\xi})}{p_t(\boldsymbol{\xi})} = \frac{\mathbb{E}[\boldsymbol{z}_{\boldsymbol{\xi}}]}{t^2}. \tag{B.2.114}$$

Thus writing $\Delta \log p_t$ out fully, we have

$$\Delta \log p_t(\boldsymbol{\xi}) = \frac{\mathbb{E}[\|\boldsymbol{z}_{\boldsymbol{\xi}}\|_2^2]}{t^4} - \frac{D}{t^2} - \frac{\|\mathbb{E}[\boldsymbol{z}_{\boldsymbol{\xi}}]\|_2^2}{t^4} \tag{B.2.115}$$

$$= \frac{\mathbb{E}[\|\boldsymbol{z}_{\boldsymbol{\xi}}\|_2^2] - \|\mathbb{E}[\boldsymbol{z}_{\boldsymbol{\xi}}]\|_2^2}{t^4} - \frac{D}{t^2} \tag{B.2.116}$$

$$= \frac{\operatorname{tr}(\operatorname{Cov}(\boldsymbol{z}_{\boldsymbol{\xi}}))}{t^4} - \frac{D}{t^2} \tag{B.2.117}$$

$$= \frac{\operatorname{tr}(\operatorname{Cov}(\boldsymbol{y}_{\boldsymbol{\xi}}))}{t^4} - \frac{D}{t^2}. \tag{B.2.118}$$

A trivial lower bound on this trace is 0, since covariance matrices are positive semidefinite. To find an upper bound, note that $\boldsymbol{y}_{\boldsymbol{\xi}}$ takes values only in the support of $\boldsymbol{x}$ (since $p$ is a factor of the density $q_{\boldsymbol{\xi}}$ of $\boldsymbol{y}_{\boldsymbol{\xi}}$), which by Assumption B.1 is a compact set $\mathcal{S}$ with radius $R \doteq \sup_{\boldsymbol{\xi} \in \mathcal{S}} \|\boldsymbol{\xi}\|_2$. So

$$\operatorname{tr}(\operatorname{Cov}(\boldsymbol{y}_{\boldsymbol{\xi}})) = \mathbb{E}[\|\boldsymbol{y}_{\boldsymbol{\xi}}\|_2^2] - \|\mathbb{E}[\boldsymbol{y}_{\boldsymbol{\xi}}]\|_2^2 \leq \mathbb{E}[\|\boldsymbol{y}_{\boldsymbol{\xi}}\|_2^2] \leq R^2. \tag{B.2.119}$$

Therefore

$$-\frac{D}{t^2} \leq \Delta \log p_t(\boldsymbol{\xi}) \leq \frac{R^2}{t^4} - \frac{D}{t^2}, \tag{B.2.120}$$

which shows the claim. □

### Derivative Computations

Here we calculate some useful derivatives which will be reused throughout the appendix.

**Proposition B.1.** *Let $\boldsymbol{x}$ be any random variable such that Assumptions B.1 and B.2 hold, and let $(\boldsymbol{x}_t)_{t \in [0,T]}$ be the stochastic process (B.2.1). For $t \geq 0$, let $p_t$ be the density of $\boldsymbol{x}_t$. Then*

$$\frac{\partial p_t}{\partial t}(\boldsymbol{\xi}) = \mathbb{E}\left[\varphi_t(\boldsymbol{\xi} - \boldsymbol{x}) \cdot \frac{\|\boldsymbol{\xi} - \boldsymbol{x}\|_2^2 - Dt^2}{t^3}\right] \tag{B.2.121}$$

$$\nabla p_t(\boldsymbol{\xi}) = -\mathbb{E}\left[\varphi_t(\boldsymbol{\xi} - \boldsymbol{x}) \cdot \frac{\boldsymbol{\xi} - \boldsymbol{x}}{t^2}\right] \tag{B.2.122}$$

$$\nabla^2 p_t(\boldsymbol{\xi}) = \mathbb{E}\left[\varphi_t(\boldsymbol{\xi} - \boldsymbol{x}) \cdot \frac{(\boldsymbol{\xi} - \boldsymbol{x})(\boldsymbol{\xi} - \boldsymbol{x})^\top - t^2 \boldsymbol{I}}{t^4}\right] \tag{B.2.123}$$

$$\Delta p_t(\boldsymbol{\xi}) = \mathbb{E}\left[\varphi_t(\boldsymbol{\xi} - \boldsymbol{x}) \cdot \frac{\|\boldsymbol{\xi} - \boldsymbol{x}\|_2^2 - Dt^2}{t^4}\right]. \tag{B.2.124}$$

*Proof.* We use the convolution representation of $p_t$, namely (B.2.3). First taking the time derivative, a computation yields that Proposition B.3 applies,[5] so we can bring the derivative inside the integral/expectation as:

$$\frac{\partial p_t}{\partial t}(\boldsymbol{\xi}) = \frac{\partial}{\partial t}\mathbb{E}[\varphi_t(\boldsymbol{\xi} - \boldsymbol{x})] = \mathbb{E}\left[\frac{\partial}{\partial t}\varphi_t(\boldsymbol{\xi} - \boldsymbol{x})\right] = \frac{\partial \varphi_t}{\partial t} * p. \tag{B.2.125}$$

Meanwhile, by properties of convolutions (Proposition B.4) and using the fact that $p$ is compactly supported (Assumption B.1),

$$p_t = \varphi_t * p \implies \nabla p_t = \nabla\varphi_t * p \implies \nabla^2 p_t = \nabla^2\varphi_t * p \implies \Delta p_t = \Delta\varphi_t * p. \tag{B.2.126}$$

The rest of the computation follows from Proposition B.2. □

**Proposition B.2.** *For $t > 0$ and $\boldsymbol{\xi} \in \mathbb{R}^D$ it holds*

$$\frac{\partial}{\partial t}\varphi_t(\boldsymbol{\xi}) = \varphi_t(\boldsymbol{\xi}) \cdot \frac{\|\boldsymbol{\xi}\|_2^2 - Dt^2}{t^3} \tag{B.2.127}$$

$$\nabla\varphi_t(\boldsymbol{\xi}) = -\varphi_t(\boldsymbol{\xi}) \cdot \frac{\boldsymbol{\xi}}{t^2} \tag{B.2.128}$$

$$\nabla^2\varphi_t(\boldsymbol{\xi}) = \varphi_t(\boldsymbol{\xi}) \cdot \frac{\boldsymbol{\xi}\boldsymbol{\xi}^\top - t^2\boldsymbol{I}}{t^4} \tag{B.2.129}$$

$$\Delta\varphi_t(\boldsymbol{\xi}) = \varphi_t(\boldsymbol{\xi}) \cdot \frac{\|\boldsymbol{\xi}\|_2^2 - Dt^2}{t^4}. \tag{B.2.130}$$

*Proof.* Direct computation. □

### Differentiating Under the Integral Sign

In this appendix, we differentiate under the integral sign many times, and it is important to know when we can do this. There are two kinds of differentiating under the integral sign:

1. Differentiating an integral $\int f_t(\boldsymbol{\xi})\mathrm{d}\boldsymbol{\xi}$ with respect to the auxiliary parameter $t$.

2. Differentiating a convolution $(f * g)(\boldsymbol{\xi}) = \int f(\boldsymbol{\xi})g(\boldsymbol{\xi} - \boldsymbol{u})\mathrm{d}\boldsymbol{u}$ with respect to the variable $\boldsymbol{\xi}$.

For the first category, we give a concrete result, stated without proof but attributable to the linked source, which derives the following result as a special case of a more general theorem about the interaction of differential operators and tempered distributions, much beyond the scope of the book. A full formal reference can be found in [Jon82].

[5] We use $f_t(\boldsymbol{\xi}) = p(\boldsymbol{\xi})\varphi_t(\boldsymbol{\xi} - \boldsymbol{x})$, noting that it is twice continuously differentiable in $\boldsymbol{\xi}$ and (more than) twice continuously differentiable in $t$. Then to check the local integrability of $f_t$ we compute $\frac{\partial f_t}{\partial t}(\boldsymbol{\xi}) = f_t(\boldsymbol{\xi}) \cdot \frac{1}{t^3}(\|\boldsymbol{\xi} - \boldsymbol{x}\|_2^2 - Dt^2)$, which is is easy to check integrable over $\boldsymbol{\xi}$ and $t \in [t_{\min}, t_{\max}]$ where $t_{\min} > 0$. (Indeed, $f_t$ has exponentially decaying tails, so the quadratic term in the product is of no issue.)

**Proposition B.3** ([Jon82], Section 11.12)**.** *Let* $f\colon (0,T)\times\mathbb{R}^D \to \mathbb{R}$ *be such that:*

- $f$ *is a jointly measurable function of* $(t,\boldsymbol{\xi})$*;*
- *For Lebesgue-almost every* $\boldsymbol{\xi}\in\mathbb{R}^D$*, the function* $t\mapsto f_t(\boldsymbol{\xi})$ *is absolutely continuous;*
- $\frac{\partial f_t}{\partial t}$ *is locally integrable, i.e., for every* $[t_{\min}, t_{\max}] \subseteq (0,T)$ *it holds*

$$\int_{t_{\min}}^{t_{\max}} \int_{\mathbb{R}^D} \left| \frac{\partial f_t}{\partial t}(\boldsymbol{\xi}) \right| \mathrm{d}\boldsymbol{\xi} < \infty. \tag{B.2.131}$$

*Then* $t \mapsto \int_{\mathbb{R}^D} f_t(\boldsymbol{\xi})\mathrm{d}\boldsymbol{\xi}$ *is an absolutely continuous function on* $(0,T)$*, and its derivative is*

$$\frac{\mathrm{d}}{\mathrm{d}t}\int_{\mathbb{R}^D} f_t(\boldsymbol{\xi})\mathrm{d}\boldsymbol{\xi} = \int_{\mathbb{R}^D} \frac{\partial}{\partial t} f_t(\boldsymbol{\xi})\mathrm{d}\boldsymbol{\xi}, \tag{B.2.132}$$

*defined for almost every* $t \in (0,T)$*.*

For the second category, we give another concrete result, stated without proof but fully formalized in [BB11].

**Proposition B.4** ([BB11], Proposition 4.20)**.** *Let* $f$ *be* $k$*-times continuously differentiable with compact support, and let* $g$ *be locally integrable. Then the convolution* $f * g$ *defined by*

$$(f*g)(\boldsymbol{\xi}) \doteq \int_{\mathbb{R}^D} f(\boldsymbol{u})g(\boldsymbol{\xi}-\boldsymbol{u})\mathrm{d}\boldsymbol{u} \tag{B.2.133}$$

*is* $k$*-times continuously differentiable, and its derivative of order* $k$ *is*

$$\nabla^k(f*g) = (\nabla^k f)*g. \tag{B.2.134}$$

Although not in the book, a simple integration by parts argument shows that if $g$ is also $k$-times differentiable, then we can "trade off" the regularity:

$$\nabla^k(f*g) = f*(\nabla^k g). \tag{B.2.135}$$

## B.3 Lossy Coding and Sphere Packing

In this section, we prove Theorem 4.1. Following our conventions throughout this appendix, we write $\mathcal{S} = \mathrm{Supp}(\boldsymbol{x})$ for the compact support of the random variable $\boldsymbol{x}$.

As foreshadowed, we will make a regularity assumption on the support set $\mathcal{S}$ to prove Theorem 4.1. One possibility for proceeding under minimal assumptions would be to instantiate the results of [RBK18; RKB23] in our setting, since these results apply to sets $\mathcal{S}$ with very low regularity (e.g., Cantor-like sets with fractal structure). However, we have found precisely computing the constants in these results, a necessary endeavor to assert a conclusion like Theorem 4.1, to be somewhat onerous in our setting. Our approach is therefore to add a geometric regularity assumption on the set $\mathcal{S}$ that sacrifices some generality, but allows us to develop a more transparent argument. To avoid sacrificing too much generality, we must ensure that low-dimensionality in the set $\mathcal{S}$ is not prohibited. We therefore consider the running example we have used throughout the book, the mixture of low-rank Gaussian distributions. In this geometric setting, we model $\mathcal{S}$ as a union of hyperspheres, each of dimension $d_k$ (possibly much smaller than $D$), living in mutually orthogonal subspaces of $\mathbb{R}^D$. This is a geometric simplification of the Gaussian mixture model: each component's support is approximated by a sphere in the subspace spanned by its principal directions. When the component dimensions $d_k$ are large, this assumption is equivalent to a mixture of low-rank Gaussians assumption, by high-dimensional measure concentration. The CRATE models we develop and train in Chapters 5 and 8 have their *subspace dimensions* set compatibly with this assumption.

**Assumption B.3.** The support $\mathcal{S} \subset \mathbb{R}^D$ of the random variable $\boldsymbol{x}$ is a finite union of $K$ spheres, each with dimension $d_k$, $k \in [K]$. The probability that $\boldsymbol{x}$ is drawn from the $k$-th sphere is given by $\pi_k \in [0, 1]$, and conditional on being drawn from the $k$-th sphere, $\boldsymbol{x}$ is uniformly distributed on that sphere. The supports satisfy that each sphere is mutually orthogonal with all others.

We proceed under the simplifying Assumption B.3 in order to simplify excessive technicality, and to connect to an important running example used throughout the monograph. We believe our results can be generalized to support $\mathcal{S}$ from the class of *sets with positive reach* with additional technical effort, but leave this for the future.

### B.3.1 Proof of Relationship Between Rate Distortion and Covering

We briefly sketch the proof, then proceed to establishing three fundamental lemmas, then give the proof. The proof will depend on notions introduced in the sketch below.

Obtaining an upper bound on the rate distortion function (4.1.9) is straightforward: by the rate characterization (i.e., the rate distortion function is the minimum rate of a code for $\boldsymbol{x}$ with expected squared $\ell^2$ distortion $\epsilon$), upper bounding $\mathcal{R}_\epsilon(\boldsymbol{x})$ only requires demonstrating one code for $\boldsymbol{x}$ that achieves this target distortion, and any $\epsilon$-covering of $\mathrm{Supp}(\boldsymbol{x})$ achieves this, with rate equal to the base-2 logarithm of the cardinality of the covering. The lower bound is more subtle. We make use of the Shannon lower bound, discussed in Remark 4.3: working out the constants in [LZ94, §III, (22)] gives a more precise version of the result quoted in Equation (4.1.17) (in bits, of course): for any random variable $\boldsymbol{x}$ with compact support and a density, it holds

$$\mathcal{R}_\epsilon(\boldsymbol{x}) \geq h(\boldsymbol{x}) - \log \mathrm{vol}(B_\epsilon) + \log\left(\frac{2}{D\Gamma(D/2)}\left(\frac{D}{2e}\right)^{D/2}\right), \tag{B.3.1}$$

with entropy (etc.) in nats in this expression. The constant can be easily estimated using Stirling's approximation. A quantitative form of Stirling's approximation which is often useful gives for any $x > 0$ [Jam15]

$$\Gamma(x) \leq \sqrt{2\pi} x^{x-1/2} e^{-x} e^{1/(12x)}. \tag{B.3.2}$$

We will apply this bound to $\Gamma(D/2)$ in Equation (B.3.1). We get

$$\log\left(\frac{2}{D\Gamma(D/2)}\left(\frac{D}{2e}\right)^{D/2}\right) \geq -\frac{1}{6D} + \log\left(\frac{2}{D}\left(\frac{D}{2e}\right)^{D/2} \cdot \sqrt{\frac{D}{4\pi}}\left(\frac{D}{2e}\right)^{-D/2}\right) \tag{B.3.3}$$

$$= -\frac{1}{6D} - \frac{1}{2}\log D\pi, \tag{B.3.4}$$

which we can take for the explicit value of the constant $C_D$ in Equation (4.1.17). Summarizing the fully quantified Shannon lower bound (in bits):

$$\mathcal{R}_\epsilon(\boldsymbol{x}) \geq h(\boldsymbol{x}) - \log_2 \operatorname{vol}(B_\epsilon) - O(\log D). \tag{B.3.5}$$

Now, the important constraint for our current purposes is that the Shannon lower bound requires the random variable $\boldsymbol{x}$ to have a density, which rules out many low-dimensional distributions of interest. But let us momentarily consider the situation when $\boldsymbol{x}$ does admit a density. The assumption that $\boldsymbol{x}$ is uniformly distributed on its support is easily formalized in this setting: for any Borel set $A \subset \mathcal{S}$, we have

$$\mathbb{P}[\boldsymbol{x} \in A] = \int_A \frac{1}{\operatorname{vol}(\mathcal{S})} \mathrm{d}\boldsymbol{x}. \tag{B.3.6}$$

Then the entropy $h(\boldsymbol{x})$ is just

$$h(\boldsymbol{x}) = \log_2 \operatorname{vol}(\mathcal{S}). \tag{B.3.7}$$

The proof then concludes with a lemma that relates the ratio $\operatorname{vol}(\mathcal{S})/\operatorname{vol}(B_\epsilon)$ to the $\epsilon$-covering number of $\mathcal{S}$ by $\epsilon$ balls.

To extend the program above to degenerate distributions satisfying Assumption B.3, our proof of the lower bound in Theorem 4.1 will leverage an approximation argument of the actual low-dimensional distribution $\boldsymbol{x}$ by "nearby" distributions which have densities, similarly but not exactly the same as the proof sketch preceding Theorem B.1. We will then link the parameter introduced in the approximating sequence to the distortion parameter $\epsilon$ in order to obtain the desired conclusion in Theorem 4.1.

**Definition B.1.** Let $\mathcal{S}$ be a compact set. For any $\delta > 0$, define the $\delta$-thickening of $\mathcal{S}$, denoted $\mathcal{S}_\delta$, by

$$\mathcal{S}_\delta = \left\{ \boldsymbol{\xi} \in \mathbb{R}^D \;\middle|\; \operatorname{dist}(\boldsymbol{\xi}, \mathcal{S}) \leq \delta \right\}. \tag{B.3.8}$$

The distance function referenced in Definition B.1 is defined by

$$\operatorname{dist}(\boldsymbol{\xi}, \mathcal{S}) = \inf_{\boldsymbol{\xi}' \in \mathcal{S}} \|\boldsymbol{\xi} - \boldsymbol{\xi}'\|_2 . \tag{B.3.9}$$

For a compact set $\mathcal{S}$, Weierstrass's theorem implies that for any $\boldsymbol{\xi} \in \mathbb{R}^D$, there is always some $\boldsymbol{\xi}' \in \mathcal{S}$ attaining the infimum in the distance function. Compactness of $\mathcal{S}_\delta$ follows readily from compactness of $\mathcal{S}$, so $\operatorname{vol}(\mathcal{S}_\delta)$ is finite for any $\delta > 0$. It is then possible to make the following definition of a thickened random variable, specialized to Assumption B.3.

**Definition B.2.** Let $\boldsymbol{x}$ be a random variable such that $\operatorname{Supp}(\boldsymbol{x}) = \mathcal{S}$ is a union of $K$ hyperspheres, distributed as in Assumption B.3. Denote the support of each component of the mixture by $\mathcal{S}_k$. Define the thickened random variable $\boldsymbol{x}_\delta$ as the mixture of measures where each component measure is uniform on the thickened set $\mathcal{S}_{k,\delta}$ (Definition B.1), for $k \in [K]$, with mixing weights $\pi_k$.

**Lemma B.6.** *Suppose the random variable $\boldsymbol{x}$ satisfies Assumption B.3. Then if $0 < \delta < \frac{1}{2}$, the thickened random variable $\boldsymbol{x}_\delta$ (Definition B.2) satisfies for any $\epsilon > 0$*

$$R_{\delta+\epsilon}(\boldsymbol{x}_\delta) \leq R_\epsilon(\boldsymbol{x}). \tag{B.3.10}$$

The proof of Lemma B.6 is deferred to Section B.3.2. Using Lemma B.6, the above program can be realized, because the random variable $\boldsymbol{x}_\delta$ has a density that is uniform with respect to the Lebesgue measure.

*(Proof of Theorem 4.1).* The upper bound is readily shown. If $S$ is any $\epsilon$-cover of the support of $\boldsymbol{x}$ with cardinality $\mathcal{N}_\epsilon(\operatorname{Supp}(\boldsymbol{x}))$, then consider the coding scheme assigning to each $\boldsymbol{\xi} \in \operatorname{Supp}(\boldsymbol{x})$ the reconstruction $\hat{\boldsymbol{\xi}} = \arg\min_{\boldsymbol{\xi}' \in S} \|\boldsymbol{\xi} - \boldsymbol{\xi}'\|_2$, with ties broken arbitrarily. Then ties occur with probability zero, and the fact that $S$ covers $\operatorname{Supp}(\boldsymbol{x})$ at scale $\epsilon$ guarantees distortion no larger than $\epsilon$; the rate of this scheme is $\log_2 \mathcal{N}_\epsilon(\operatorname{Supp}(\boldsymbol{x}))$.

For the lower bound, let $0 < \delta < \frac{1}{2}$, and consider the thickened random variable $\boldsymbol{x}_\delta$. By Lemma B.6, we have

$$R_{\delta+\epsilon}(\boldsymbol{x}_\delta) \leq R_\epsilon(\boldsymbol{x}). \tag{B.3.11}$$

Since $\boldsymbol{x}_\delta$ has a Lebesgue density that is uniform, we can then apply the Shannon lower bound, in the form (B.3.5), to get

$$\log_2 \operatorname{vol}(\operatorname{Supp}(\boldsymbol{x}_\delta)) - \log_2 \operatorname{vol}(B_{\delta+\epsilon}) - O(\log D) \leq R_\epsilon(\boldsymbol{x}). \tag{B.3.12}$$

Finally, we need to lower bound the ratio

$$\frac{\operatorname{vol}(\operatorname{Supp}(\boldsymbol{x}_\delta))}{\operatorname{vol}(B_{\delta+\epsilon})} \tag{B.3.13}$$

in terms of the covering number. Since $\operatorname{Supp}(\boldsymbol{x}_\delta) = \operatorname{Supp}(\boldsymbol{x}) + B_\delta$, where $+$ here denotes the Minkowski sum, a standard application of volume bound arguments (see e.g. [Ver18, Proposition 4.2.12]) gives

$$\operatorname{vol}(\operatorname{Supp}(\boldsymbol{x}_\delta)) \geq \mathcal{N}_{2\delta}(\operatorname{Supp}(\boldsymbol{x})) \operatorname{vol}(B_\delta). \tag{B.3.14}$$

Hence

$$\frac{\mathrm{vol}(\mathrm{Supp}(\boldsymbol{x}_\delta))}{\mathrm{vol}(B_{\delta+\epsilon})} \geq \mathcal{N}_{2\delta}(\mathrm{Supp}(\boldsymbol{x}))\frac{\mathrm{vol}(B_\delta)}{\mathrm{vol}(B_{\delta+\epsilon})} \tag{B.3.15}$$

$$= \mathcal{N}_{2\delta}(\mathrm{Supp}(\boldsymbol{x}))\left(\frac{\delta}{\delta+\epsilon}\right)^D. \tag{B.3.16}$$

Choosing $\delta = \epsilon/2$ gives from the Shannon lower bound (B.3.12) and the above estimates

$$\log_2 \mathcal{N}_\epsilon(\mathrm{Supp}(\boldsymbol{x})) - O(D) \leq R_\epsilon(\boldsymbol{x}), \tag{B.3.17}$$

as was to be shown.

□

### B.3.2 Proof of Lemma B.6

*(Proof of Lemma B.6).* It suffices to show that any code for $\boldsymbol{x}$ with expected squared distortion $\epsilon^2$ produces a code for $\boldsymbol{x}_\delta$ with the same rate and distortion not much larger, for a suitable choice of $\delta$. So fix such a code for $\boldsymbol{x}$, achieving rate $R$ and expected squared distortion $\epsilon^2$. We write $\hat{\boldsymbol{x}}$ for the reconstructed random variable using this code, and $\mathrm{q} : \mathrm{Supp}(\boldsymbol{x}) \to \mathrm{Supp}(\boldsymbol{x})$ for the associated encoding-decoding mapping (i.e., $\hat{\boldsymbol{x}} = \mathrm{q}(\boldsymbol{x})$).

Now let $\mathcal{S}_k$ denote the $k$-th hypersphere in the support of $\boldsymbol{x}$. There is an orthonormal basis $\boldsymbol{U}_k \in \mathbb{R}^{D\times d_k}$ such that $\mathrm{Span}(\mathcal{S}_k) = \mathrm{Span}(\boldsymbol{U}_k)$. The following orthogonal decomposition of the support set $\mathcal{S}$ will be used repeatedly throughout the proof. We have

$$\mathcal{S}_\delta = \{\boldsymbol{\xi} \in \mathbb{R}^D \mid \exists k \in [K] \,:\, \mathrm{dist}(\boldsymbol{\xi}, \mathcal{S}_k) \leq \delta\} \tag{B.3.18}$$

$$= \bigcup_{k\in[K]} \{\boldsymbol{\xi} \in \mathbb{R}^D \mid \mathrm{dist}(\boldsymbol{\xi}, \mathcal{S}_k) \leq \delta\}. \tag{B.3.19}$$

By orthogonal projection, for any $k \in [K]$ any $\boldsymbol{\xi} \in \mathbb{R}^D$ can be written as $\boldsymbol{\xi} = \boldsymbol{\xi}^\parallel + \boldsymbol{\xi}^\perp$, with $\boldsymbol{\xi}^\parallel \in \mathrm{Span}(\mathcal{S}_k)$ and $\langle \boldsymbol{\xi}^\parallel, \boldsymbol{\xi}^\perp\rangle = 0$. Then for any $\boldsymbol{\xi}' \in \mathcal{S}_k$, we have

$$\|\boldsymbol{\xi} - \boldsymbol{\xi}'\|_2^2 = \left\|\boldsymbol{\xi}^\parallel + \boldsymbol{\xi}^\perp - \boldsymbol{\xi}'\right\|_2^2 = \left\|\boldsymbol{\xi}^\parallel\right\|_2^2 + \|\boldsymbol{\xi}^\perp\|_2^2 + \|\boldsymbol{\xi}'\|_2^2 - 2\left\langle \boldsymbol{\xi}^\parallel, \boldsymbol{\xi}'\right\rangle \tag{B.3.20}$$

$$\geq \left\|\boldsymbol{\xi}^\parallel\right\|_2^2 + \|\boldsymbol{\xi}'\|_2^2 - 2\left\langle \boldsymbol{\xi}^\parallel, \boldsymbol{\xi}'\right\rangle \tag{B.3.21}$$

$$= \left\|\boldsymbol{\xi}^\parallel - \boldsymbol{\xi}'\right\|_2^2. \tag{B.3.22}$$

Further, it is known that for any nonzero $\boldsymbol{\xi}^\parallel \in \mathrm{Span}(\mathcal{S}_k)$,

$$\inf_{\boldsymbol{\xi}'\in\mathcal{S}_k} \left\|\boldsymbol{\xi}^\parallel - \boldsymbol{\xi}'\right\|_2^2 = \left\|\boldsymbol{\xi}^\parallel - \frac{\boldsymbol{\xi}^\parallel}{\|\boldsymbol{\xi}^\parallel\|_2}\right\|_2^2. \tag{B.3.23}$$

If $\boldsymbol{\xi}^{\parallel}$ is zero, it is clear that the above distance is equal to 1 for every $\boldsymbol{\xi}' \in \mathcal{S}_k$. Hence, if we define a projection mapping $\pi_{\mathcal{S}_k}(\boldsymbol{\xi})$ by

$$\pi_{S_k}(\boldsymbol{\xi}) = \frac{\boldsymbol{U}_k \boldsymbol{U}_k^\top \boldsymbol{\xi}}{\|\boldsymbol{U}_k^\top \boldsymbol{\xi}\|_2} \tag{B.3.24}$$

for any $\boldsymbol{\xi} \in \mathbb{R}^D$ with $\boldsymbol{U}_k^\top \boldsymbol{\xi} \neq \mathbf{0}$, then $\pi_{\mathcal{S}_k}(\boldsymbol{\xi}) = \arg\min_{\boldsymbol{\xi}' \in \mathcal{S}_k} \|\boldsymbol{\xi}' - \boldsymbol{\xi}\|_2$. We choose $0 < \delta < 1$, so that the thickened set $\mathcal{S}_\delta$ contains no points $\boldsymbol{\xi} \in \mathbb{R}^D$ at which any of the projection maps $\pi_{\mathcal{S}_k}$ is not well-defined. So the thickened set $\mathcal{S}_\delta$ satisfies

$$\mathcal{S}_\delta = \bigcup_{k \in [K]} \left\{ \boldsymbol{\xi} \in \mathbb{R}^D \;\middle|\; \left\| \boldsymbol{\xi} - \frac{\boldsymbol{U}_k \boldsymbol{U}_k^\top \boldsymbol{\xi}}{\|\boldsymbol{U}_k^\top \boldsymbol{\xi}\|_2} \right\|_2 \leq \delta \right\}. \tag{B.3.25}$$

These distances can be rewritten in terms of the orthogonal decomposition as

$$\left\| \boldsymbol{\xi} - \frac{\boldsymbol{U}_k \boldsymbol{U}_k^\top \boldsymbol{\xi}}{\|\boldsymbol{U}_k^\top \boldsymbol{\xi}\|_2} \right\|_2^2 = \|\boldsymbol{\xi}\|_2^2 - 2\|\boldsymbol{U}_k^\top \boldsymbol{\xi}\|_2 + 1 \tag{B.3.26}$$

$$= \left\|\boldsymbol{\xi}^{\parallel}\right\|_2^2 + \left\|\boldsymbol{\xi}^{\perp}\right\|_2^2 - 2\|\boldsymbol{U}_k^\top \boldsymbol{\xi}^{\parallel}\|_2 + 1 \tag{B.3.27}$$

$$= \left\|\boldsymbol{\xi}^{\perp}\right\|_2^2 + \left(\left\|\boldsymbol{\xi}^{\parallel}\right\|_2 - 1\right)^2. \tag{B.3.28}$$

We are going to show next that every such $\boldsymbol{\xi} \in \mathcal{S}_\delta$ can be uniquely associated to a projection onto a single subspace in the mixture, which will allow us to define a corresponding projection onto $\mathcal{S}$. Given a $\boldsymbol{\xi} \in \mathcal{S}_\delta$, by the above, we can find a subspace $\boldsymbol{U}_k$ such that the orthogonal decomposition $\boldsymbol{\xi} = \boldsymbol{\xi}_k^{\parallel} + \boldsymbol{\xi}_k^{\perp}$ satisfies

$$\left\|\boldsymbol{\xi}_k^{\perp}\right\|_2^2 + \left(\left\|\boldsymbol{\xi}_k^{\parallel}\right\|_2 - 1\right)^2 \leq \delta^2. \tag{B.3.29}$$

Consider the decomposition $\boldsymbol{\xi} = \boldsymbol{\xi}_j^{\parallel} + \boldsymbol{\xi}_j^{\perp}$ for some $j \neq k$. We have

$$\left\|\boldsymbol{\xi}_j^{\parallel}\right\|_2 = \left\|\boldsymbol{U}_j \boldsymbol{U}_j^\top \boldsymbol{\xi}\right\|_2 = \left\|\boldsymbol{U}_j \boldsymbol{U}_j^\top (\boldsymbol{U}_k \boldsymbol{U}_k^\top \boldsymbol{\xi} + (\boldsymbol{I} - \boldsymbol{U}_k \boldsymbol{U}_k^\top)\boldsymbol{\xi})\right\|_2 \tag{B.3.30}$$

$$= \left\|\boldsymbol{U}_j \boldsymbol{U}_j^\top (\boldsymbol{I} - \boldsymbol{U}_k \boldsymbol{U}_k^\top)\boldsymbol{\xi}\right\|_2 \tag{B.3.31}$$

$$\leq \left\|(\boldsymbol{I} - \boldsymbol{U}_k \boldsymbol{U}_k^\top)\boldsymbol{\xi}\right\|_2 = \left\|\boldsymbol{\xi}_k^{\perp}\right\|_2 \leq \delta, \tag{B.3.32}$$

where the second line uses the orthogonality assumption on the subspaces $\boldsymbol{U}_k$, and the third uses the fact that orthogonal projections are nonexpansive. Hence, the $j$-th distance satisfies

$$\left\|\boldsymbol{\xi}_j^{\perp}\right\|_2^2 + \left(\left\|\boldsymbol{\xi}_j^{\parallel}\right\|_2 - 1\right)^2 \geq (1 - \delta)^2. \tag{B.3.33}$$

This implies that if $0 < \delta < 1/2$, every $\boldsymbol{\xi} \in \mathcal{S}_\delta$ has a unique closest subspace in the mixture. Hence, under this condition, the following mapping $\pi_{\mathcal{S}} : \mathcal{S}_\delta \to \mathcal{S}$ is well-defined:

$$\pi_{\mathcal{S}}(\boldsymbol{\xi}) = \pi_{\mathcal{S}_{k_\star}}(\boldsymbol{\xi}), \;\; \text{where } \; k_\star = \arg\min_{k \in [K]} \operatorname{dist}(\boldsymbol{\xi}, \mathcal{S}_k). \tag{B.3.34}$$

Now, we define a code for $\boldsymbol{x}_\delta$ by

$$\hat{\boldsymbol{x}}_\delta = \mathrm{q}(\pi_{\mathcal{S}}(\boldsymbol{x}_\delta)). \tag{B.3.35}$$

Clearly this is associated to a rate-$R$ code for $\boldsymbol{x}_\delta$, because it uses the encoding-decoding mappings from the rate-$R$ code for $\boldsymbol{x}$. We have to show that it achieves small distortion. We calculate

$$\mathbb{E}\left[\|\boldsymbol{x}_\delta - \hat{\boldsymbol{x}}_\delta\|_2^2\right] = \mathbb{E}\left[\|\boldsymbol{x}_\delta - \mathrm{q}(\pi_{\mathcal{S}}(\boldsymbol{x}_\delta))\|_2^2\right] \tag{B.3.36}$$

$$\leq \left(\mathbb{E}\left[\|\boldsymbol{x}_\delta - \pi_{\mathcal{S}}(\boldsymbol{x}_\delta)\|_2^2\right]^{1/2} + \mathbb{E}\left[\|\pi_{\mathcal{S}}(\boldsymbol{x}_\delta) - \mathrm{q}(\pi_{\mathcal{S}}(\boldsymbol{x}_\delta))\|_2^2\right]^{1/2}\right)^2, \tag{B.3.37}$$

where the inequality uses the Minkowski inequality. Now, by Definitions B.1 and B.2, we have deterministically

$$\|\boldsymbol{x}_\delta - \pi_{\mathcal{S}}(\boldsymbol{x}_\delta)\|_2^2 \leq \delta^2, \tag{B.3.38}$$

so the expectation also satisfies this estimate. For the second term, it will suffice to characterize the density of the random variable $\pi_{\mathcal{S}}(\boldsymbol{x}_\delta)$ as being sufficiently close to the density of $\boldsymbol{x}$—which, as Assumption B.3 implies, is a mixture of uniform distributions on each sub-sphere $\mathcal{S}_k$. By the argument above, every point $\boldsymbol{\xi} \in \mathcal{S}_\delta$ can be associated to one and only one subspace $\boldsymbol{U}_k$, which means that the mixture components in the definition of $\mathcal{S}_\delta$ (recall Definition B.2) do not overlap. Hence, the density $\pi_{\mathcal{S}}(\boldsymbol{x}_\delta)$ can be characterized by studying the effect of $\pi_{\mathcal{S}_k}$ on the conditional random variable $\boldsymbol{x}_\delta$, conditioned on being drawn from $\mathcal{S}_{k,\delta}$. Denote this measure by $\mu_{k,\delta}$. We claim that the pushforward of this measure under $\pi_{\mathcal{S}_k}$ is uniform on $\mathcal{S}_k$. To see that this holds, we recall Equation (B.3.28), which gives the characterization

$$\mathcal{S}_{k,\delta} = \left\{\boldsymbol{\xi}^\parallel + \boldsymbol{\xi}^\perp \;\middle|\; \boldsymbol{\xi}^\parallel \in \mathrm{Span}(\boldsymbol{U}_k), \boldsymbol{\xi}^\perp \in \mathrm{Span}(\boldsymbol{U}_k)^\perp, \|\boldsymbol{\xi}^\perp\|_2^2 + \left(\left\|\boldsymbol{\xi}^\parallel\right\|_2 - 1\right)^2 \leq \delta^2\right\}. \tag{B.3.39}$$

The conditional distribution in question is uniform on this set; we need to show that the projection $\pi_{\mathcal{S}_k}$ applied to this conditional random variable yields a random variable that is uniform on $\mathcal{S}_k$. With respect to these coordinates, we have seen that $\pi_{\mathcal{S}_k}(\boldsymbol{\xi}^\parallel + \boldsymbol{\xi}^\perp) = \boldsymbol{\xi}^\parallel / \|\boldsymbol{\xi}^\parallel\|_2$. Hence, for any $\boldsymbol{\xi} \in \mathcal{S}_k$, we have that the preimage of $\boldsymbol{\xi}$ in $\mathcal{S}_{k,\delta}$ under $\pi_{\mathcal{S}_k}$ is

$$\pi_{\mathcal{S}_k}^{-1}(\boldsymbol{\xi}) = \left\{r\boldsymbol{\xi} + \boldsymbol{\xi}^\perp \;\middle|\; r > 0, \boldsymbol{\xi}^\perp \in \mathrm{Span}(\boldsymbol{U}_k)^\perp, \|\boldsymbol{\xi}^\perp\|_2^2 + (r-1)^2 \leq \delta^2\right\}. \tag{B.3.40}$$

To show that $(\pi_{\mathcal{S}_k})_\sharp \mu_{k,\delta}$ is uniform, we need to decompose the integral of the uniform density on $\mathcal{S}_{k,\delta}$ in a way that makes it clear that each of the fibers $\pi_{\mathcal{S}_k}^{-1}(\boldsymbol{\xi})$ (for each $\boldsymbol{\xi} \in \mathcal{S}_k$) "contributes" equally to the integral.[6] We have by

[6]More rigorously, this corresponds to decomposing the uniform density on $\mathcal{S}_{k,\delta}$ into a regular conditional density corresponding to $\boldsymbol{\xi} \in \mathcal{S}_k$, and showing that the corresponding density on $\boldsymbol{\xi}$ is uniform. The proof makes it clear this is true.

Definition B.2

$$\mathrm{vol}(\mathcal{S}_{k,\delta}) = \iint_{\mathrm{Span}(\boldsymbol{U}_k)\times\mathrm{Span}(\boldsymbol{U}_k)^\perp} \mathbf{1}_{\|\boldsymbol{\xi}^\perp\|_2^2 + \left(\left\|\boldsymbol{\xi}^\parallel\right\|_2 - 1\right)^2 \leq \delta^2} \mathrm{d}\boldsymbol{\xi}^\parallel \mathrm{d}\boldsymbol{\xi}^\perp. \tag{B.3.41}$$

In particular, the integration over the orthogonal coordinates factors. Let $\mathrm{d}\boldsymbol{\theta}^d$ denote the uniform (Haar) measure on the sphere of radius 1 in $\mathbb{R}^d$. Converting the $\boldsymbol{\xi}^\parallel$ integral to polar coordinates, we have

$$\mathrm{vol}(\mathcal{S}_{k,\delta}) = \int_{[0,\infty)} \int_{\mathbb{S}^{d_k-1}} \int_{\mathrm{Span}(\boldsymbol{U}_k)^\perp} r^{d_k-1} \mathbf{1}_{\|\boldsymbol{\xi}^\perp\|_2^2 + (r-1)^2 \leq \delta^2} \mathrm{d}r \mathrm{d}\boldsymbol{\theta}^{d_k} \mathrm{d}\boldsymbol{\xi}^\perp. \tag{B.3.42}$$

Comparing to the fiber representation (B.3.40), we see that we need to "integrate out" over the $r$ and $\boldsymbol{\xi}^\perp$ components of the preceding integral in order to verify that the pushforward is uniform. But this is evident, as the previous expression shows that the value of this integral is independent of $\boldsymbol{\xi}^\parallel$—or, equivalently in context, the value of the spherical component $\boldsymbol{\theta}^{d_k}$.

Thus it follows from the above argument that $\pi_{\mathcal{S}}(\boldsymbol{x}_\delta)$ is uniform. Because the assumption on $\delta$ implies that the mixture components in the distribution of $\boldsymbol{x}_\delta$ do not overlap, the mixing weights $\pi_k$ are also preserved in the image $\pi_{\mathcal{S}}(\boldsymbol{x}_\delta)$, and in particular, the distribution of $\pi_{\mathcal{S}}(\boldsymbol{x}_\delta)$ is equal to the distribution of $\boldsymbol{x}$. Hence the second term in Equation (B.3.37) satisfies

$$\mathbb{E}\left[\|\pi_{\mathcal{S}}(\boldsymbol{x}_\delta) - \mathrm{q}(\pi_{\mathcal{S}}(\boldsymbol{x}_\delta))\|_2^2\right] = \mathbb{E}\left[\|\boldsymbol{x} - \mathrm{q}(\boldsymbol{x})\|_2^2\right] \leq \epsilon^2, \tag{B.3.43}$$

because q is a distortion-$\epsilon$ code for $\boldsymbol{x}$.

We have thus shown that the hypothesized rate-$R$, (expected squared) distortion-$\epsilon^2$ code for $\boldsymbol{x}$ produces a rate-$R$, (expected squared) distortion $\delta + \epsilon$ code for $\boldsymbol{x}_\delta$. This establishes that

$$R_{\delta+\epsilon}(\boldsymbol{x}_\delta) \leq R_\epsilon(\boldsymbol{x}), \tag{B.3.44}$$

as was to be shown.

□